\documentclass[preprint,12pt,authoryear]{elsarticle}

\usepackage[table, dvipsnames]{xcolor}

\usepackage{amssymb}

\usepackage[hyphens]{url}
\usepackage[colorlinks,linkcolor=blue]{hyperref}
\usepackage{booktabs}
\usepackage{lipsum,afterpage,refcount}

\usepackage[capitalize]{cleveref}

\usepackage{subfigure}
\usepackage{comment} 
\usepackage{makecell}
\usepackage{pdflscape}
\usepackage{float}

\usepackage{highlight}
\usepackage{multirow}
\usepackage{slashbox}
\usepackage{tabularx}

\usepackage[symbol]{footmisc}

\newcommand{\setfootnotemark}{%
	\refstepcounter{footnote}%
	\footnotemark[\value{footnote}]}

\usepackage[normalem]{ulem} 

\usepackage{multicol}
\usepackage{highlight}
\usepackage{bbding}

\makeatletter
\renewcommand\paragraph{\@startsection{paragraph}{4}{\z@}%
	{-3.25ex\@plus -1ex \@minus -.2ex}%
	{1.5ex \@plus .2ex}%
	{\normalfont\normalsize\bfseries}}
\makeatother

\begin{document}

\begin{frontmatter}



\title{An evaluation of Deep Learning based stereo dense matching dataset shift from aerial images and a large scale stereo dataset \footnote[1]{The original paper is published on https://www.sciencedirect.com/science/article/pii/S1569843224000694. This is the long version.}}


\author{Teng Wu, Bruno Vallet, Marc Pierrot-Deseilligny, Ewelina Rupnik}

\address{Univ Gustave Eiffel, ENSG, IGN, LASTIG, F-94160 Saint-Mandé, France}

\begin{abstract}
Dense matching is crucial for 3D scene reconstruction since it enables the recovery of scene 3D geometry from image acquisition. Deep Learning (DL)-based methods have shown effectiveness in the special case of epipolar stereo disparity estimation in the computer vision community.
DL-based methods depend heavily on the quality and quantity of training datasets. However, generating ground-truth disparity maps for real scenes remains a challenging task in the photogrammetry community.
To address this challenge, we propose a method for generating ground-truth disparity maps directly from Light Detection and Ranging (LiDAR) and images to produce a large and diverse dataset for six aerial datasets across four different areas and two areas with different resolution images.
We also introduce a LiDAR-to-image co-registration refinement to the framework that takes special precautions regarding occlusions and refrains from disparity interpolation to avoid precision loss. 
Evaluating 11 dense matching methods across datasets with diverse scene types, image resolutions, and geometric configurations, which are deeply investigated in dataset shift, GANet performs best with identical training and testing data, and PSMNet shows robustness across different datasets, and we proposed the best strategy for training with a limit dataset.
We will also provide the dataset and training models; more information can be found at \url{https://github.com/whuwuteng/Aerial_Stereo_Dataset}.
\end{abstract}




\end{frontmatter}

\section{Introduction}

Dense image matching is a traditional topic in 3D reconstruction, and can be performed in stereo-pair (with only two views) \citep{scharstein2002taxonomy} or multi-view stereo (MVS) \citep{jensen2014large}.
In this paper, we focus on stereo pair dense matching in the specific case of epipolar stereo pairs (i.e., image correspondences have the same $y$ coordinate) because most of the recent deep learning (DL) approaches are limited to this simple configuration.
 
While many traditional methods show good performance \citep{scharstein2002taxonomy} and are widely used in the geospatial context,
the recent successes of DL-based dense matching methods in the computer vision community \citep{laga2020survey} raise the question of their applicability in the geospatial context.
Considering the large differences between the general image datasets of computer vision and the specific nature of geospatial photogrammetry, the training dataset has an important impact on the learning thus on the performance of DL-based methods. This paper investigates the relevance of the training dataset on the 3D accuracy and the transferability of the learned models in the realm of geospatial applications. Given the difficulty of finding epipolar stereo datasets with ground truth disparity maps in aerial photogrammetry, producing such datasets with good quality, diversity and quantity is also an important work reported hereafter.

This paper is an extension of the work presented in  \citep{wu2021new} and the contributions can be summarized as follows:
\begin{enumerate}
	\item We release a large aerial dataset consisting of epipolar stereo pairs along with corresponding sparse ground truth disparity maps computed from a LiDAR acquisitions
	\item We propose an efficient and accurate image-to-LiDAR registration method to improve their geometric alignment; through the produced quality dataset we ensure fair evaluation.
	\item We study and discuss the influence of the base to height ratio ($B/H$) on DL-based dense matching methods.
	\item We use this dataset to explore the dataset shift between different areas, resolutions and geometric configurations for DL-based methods.
	\item We evaluate 11 stereo matching methods, and include 7 additional recent DL-based methods.
\end{enumerate}

The paper is structured as follows: we start by a review of the \textit{state-of-the-art} stereo dense matching methods and benchmark datasets (Sections~\ref{SOTA} and \ref{sec:bench}).
Then, we present our method to generate sparse but accurate ground truth disparity maps from LiDAR point clouds and our six proposed datasets (Section~\ref{sec:aerialdatagen}). An in-depth evaluation of 11 methods is then proposed, that includes a study on dataset shift for learning based methods (Section~\ref{sec:eval}), the last part is the conclusion.

\section{State of the art for stereo dense matching}
\label{SOTA}

Depending on the strategies of proposed approaches, stereo dense matching methods can be divided into two categories: traditional and learning based methods. There has been many surveys in the past twenty years on traditional methods \citep{scharstein2002taxonomy} and recent surveys now focus on DL-based methods \citep{laga2020survey, hamid2020stereo}. In this paper, we review and evaluate both categories.


\subsection{Traditional methods}
Traditional dense matching methods \citep{hirschmuller2005accurate} usually can be decomposed into four steps: hand-crafted features computation, feature matching across images (cost volume computation), cost aggregation (optimization) and disparity refinement.
For the cost aggregation step, traditional methods can be divided into local and global methods.

\subsubsection{Local methods}
Local methods mainly take into consideration the local features (hand-crafted feature) \citep{hirschmuller2007evaluation} and use a local region in cost aggregation \citep{tombari2008classification} or a cost volume filter \citep{hosni2012fast}. Support-weights can be used to improve the window based match \citep{yoon2006adaptive}. Segmentation can also be used to select the local support region \citep{tombari2007segmentation}. Cross Based Cost Aggregation(CBCA) uses a local support by a shape-adaptive region to make the cost aggregation \citep{zhang2009cross}. The limitation is that it is difficult to decide the local window size, but local methods are very fast.

\subsubsection{Global methods}
Global methods mainly add an optimization in the cost aggregation step, based on a Markov random field (MRF) model, dynamic programming \citep{van2002hierarchical}, belief propagation \citep{sun2003stereo}, tree-reweighted message passing \citep{kolmogorov2005convergent} or Graphcut optimization \citep{boykov2004experimental}.  Semi global matching(SGM) is a popular pyramid based method combining dynamic programming optimization on several directions and using mutual information \citep{hirschmuller2005accurate}. A GPU implementation of SGM \citep{hernandez2016embedded} is also evaluated in this paper, as well as a Graphcut based method using plane constraints \citep{taniai2017continuous}.
Graphcut based methods are slower than SGM which is often considered to offer the best balance between efficiency and accuracy \citep{bethmann2015semi}.


\subsection{Learning based method}

Recently, machine learning (ML) techniques have been applied to the stereo dense matching problem and have demonstrated significant improvement over traditional methods on some benchmark datasets \citep{batsos2018cbmv}. With the development of DL, the performance improvement is even more significant \citep{laga2020survey}.
Learning based methods can be divided into traditional ML and DL-based methods. For ML methods, only a small amount of training data is sufficient, while DL-based methods need a much larger amount of training data.

\subsubsection{Traditional machine learning}

Traditional ML methods, for instance support vector machines (SVM) and random forests(RF) have been widely applied to stereo dense matching.
An approach consists in learning the matching behaviors of local methods and combine them probabilistically \citep{kong2006stereo}. Learning the hyper-parameters of the stereo algorithm has also been proposed \citep{pal2012learning}. RF can be applied to select the ground control points \citep{spyropoulos2014learning}. We also analyse a method which combines man-crafted features with RF to produce similarity, and then uses SGM optimization \citep{batsos2018cbmv}. Finally in \citep{li2008learning} SVM are used to learn a linear discriminant function. 

\subsubsection{Deep learning}

DL has proven successful in image recognition \citep{girshick2015fast}, semantic segmentation \citep{long2015fully}, depth prediction \citep{laina2016deeper} etc. With the proliferation of DL-based methods, both hybrid methods and end-to-end methods have been applied to stereo dense matching \citep{laga2020survey, poggi2021synergies}. While hybrid methods use DL to solve one step in the traditional pipeline, end-to-end methods aim at replacing the full pipeline, i.e. to produce a network that takes an epipolar image pair as input and returns a disparity for each pixel of the left image or both images. 
Some frameworks also aim at multi-task learning, which combines other tasks such as semantic segmentation with stereo dense matching \citep{yang2018segstereo}, which is out of the scope of this paper.

\paragraph{Hybrid methods} 
The initial researches aimed at applying DL to improve a single step in the traditional pipeline \citep{poggi2021synergies}. 
For the matching cost generation step, 2D convolutional neural networks (CNN) proved effective in feature extraction \citep{vzbontar2016stereo}. In order to make CNN efficient, feature matching can be treated as a multi-class classification problem \citep{luo2016efficient}. After feature extraction, traditional optimization is used to obtain the final result.
In the optimization step, SGM-Nets uses a CNN network to provide learned penalties for SGM step \citep{seki2017sgm}.
In the refinement step, variational networks can be used to refine the disparity \citep{knobelreiter2021learned}.

\paragraph{End to end methods}
There are two main types of architectures for end to end methods: 2D convolution and 3D convolution based methods. 2D architectures use an encoder-decoder structure with a 2D CNN while 3D architectures use 2D CNN for the feature concatenation and then construct a 3D cost volume processed by a 3D CNN to produce the final disparity.

For 2D architectures, DispNet was the first end-to-end method \citep{mayer2016large}. Inspired by FlowNet \citep{dosovitskiy2015flownet}, DispNetS uses a 2D CNN encoder-decoder structure, while DispNetC investigates correlation and a 2D CNN is adopted for cost aggregation. DispNet was later combined to a residual learning as a two stage training architecture \citep{pang2017cascade}. Residual learning was also used in \citep{liang2018learning} to refine the disparity. Finally,  \citep{tonioni2019real} uses a multi-scale approach to refine the disparity with 2D dilated convolutions networks.

For 3D architectures, the pioneer work GC-Net uses a 3D CNN based network as cost aggregation in the cost volume\citep{kendall2017end}.
Pyramid Stereo Matching network uses spatial pyramid pooling and 3D CNN \citep{chang2018pyramid}.
High resolution stereo network structures use upscale in the 2D CNN network, so the 3D cost volume can be downscaled \citep{yang2019hierarchical}.
For high-resolution image matching, to handle the high memory consumption, prune the 3D cost volume with a differential patch match method is proposed \citep{duggal2019deeppruner}.
From the penalty definition of SGM \citep{hirschmuller2005accurate}, a Semi-Global Guided Aggregation(SGA) layer and a Local Guided Aggregation(LGA) layer are integrated in a 3D CNN named GANet \citep{zhang2019ga}.
More recently, reinforce learning such as neural architecture search is applied to the stereo dense matching problem \citep{cheng2020hierarchical}.

\subsubsection{Dataset shift}

Most of the works listed above are from computer vision or robotics community. To evaluate the performance of the models, the training and evaluation data is usually of the same type. The models are usually trained on general image datasets such as KITTI or Middlebury which are very different from aerial photogrammetry. 
The aerial images in these kinds of data are very diverse in the type of scene, area or season. Because it is not always possible to obtain ground truth data adapted to train for any specific collection, \textit{domain adaptation} or \textit{dataset shift} ability is an important issue. \textit{Dataset shift} refers to the problem where a test image pictures a scene or an object never seen in (or significantly different from) the training data \citep{quinonero2009dataset}. 

Dataset shift raises three main issues: (1) differences in color and brightness, that may come from different sensors and lighting conditions; (2)  differences in the range of cost values, range of disparity and cost volume distribution, that come from different $B/H$ and scene types because of different scene types or viewing conditions; (3) differences in image resolution, scale.

To address dataset shift, based on an inaccurate disparity map produced by traditional methods which are domain agnostic, then obtain the confidence map of the disparity map, then combined the inaccurate disparity map with a confidence loss to train the final model from a DL stereo \citep{tonioni2019unsupervised}.
Considering that DL-based features depend on the learning dataset, \citep{cai2020matching} provides hand-made feature as input, and then uses a 3D CNN only for the cost aggregation.
Based on GANet\citep{zhang2019ga}, considering that batch normalization depends on the training data, a domain norm has been proposed for the normalization step \citep{zhang2020domain}.
\cite{song2021adastereo} use the target data without ground truth and add a color transfer pre-processing, a cost normalization layer and a self-supervised occlusion-aware reconstruction to improve the domain adaption.

\section{Benchmark dataset review}\label{sec:bench}

To evaluate stereo dense matching methods, many benchmark datasets have been released, especially in computer vision community. There are two main categories of stereo dense matching datasets: synthetic and real-world scenes. 

\subsection{Synthetic datasets}

Some authors propose to generate virtual datasets based on advanced computer graphics \citep{yang2019hierarchical, mayer2016large, tonioni2019learning}. CARLA is a synthetic dataset widely used in mobile mapping \citep{dosovitskiy2017carla}.
In a synthetic dataset, the occlusion is easy to obtain \citep{Butler2012}.
Due to the large volume of synthetic datasets, SYNTHIA-Seqs\citep{ros2016synthia} or Scene Flow \citep{mayer2016large} are usually used for pre-training \citep{zhang2019online}.
Virtual KITTI 2 also provides stereo disparity and is widely used in autonomous driving \citep{cabon2020virtual}.
IRS is an indoor dataset that also provides the surface normal \citep{wang2019irs}.
\Cref{Table:syndata} lists the most widely used synthetic dataset for stereo dense matching.

\begin{table}[!ht]
	\caption{Synthetic datasets for disparity estimation.}
	\label{Table:syndata}
	\centering
	\resizebox{\textwidth}{!}{
	\begin{tabular}{c|cccc}
		\toprule
		\multirow{2}*{\bfseries Data} & \multirow{2}*{\bfseries Year} & \multirow{2}*{\bfseries Scene}  & {\bfseries Stereo} & \multirow{2}*{\bfseries Citation}\\
		~ & ~ & ~ &  {\bfseries number} & ~ \\
		\hline
		MPI-Sintel & 2012 & outdoor & 1150 & \citep{Butler2012} \\
		Scene Flow & 2016 & indoor + outdoor & 39824 & \citep{mayer2016large} \\
		SYNTHIA-Seqs & 2016 & outdoor(driving) & 10000  & \citep{ros2016synthia} \\
		Virtual KITTI 2 & 2020 & outdoor(driving) & 17000  & \citep{cabon2020virtual} \\
		IRS & 2021 & indoor & 100025 & \citep{wang2019irs} \\
		\bottomrule
	\end{tabular}
	}
\end{table}

\subsection{Real-world datasets}

For real-world datasets, image pairs are acquired in an indoor or outdoor environments and a separate acquisition (in general with LiDAR) is performed on the same scene to produce a ground truth depth for each image pixel.
Middlebury is a popular indoor dataset \citep{scharstein2014high}. The KITTI dataset focuses on automobile driving, uses a stereo camera and LiDAR to obtain the ground truth. It contains two versions, KITTI 2012 \citep{geiger2012we} and KITTI 2015 \citep{mayer2016large}.
The ETH3D benchmark is another example of high-resolution stereo pairs \citep{schops2017multi} contains both indoor and outdoor scenes.
The strong ongoing research activity on autonomous driving has also resulted in several dedicated datasets \citep{yang2019drivingstereo, huang2019apolloscape}.
\cite{meister2012outdoor} propose an outdoor dataset in a challenging environment, but does not provide a ground truth depth.
\Cref{Table:data} lists some benchmarks for stereo dense matching on real scenes proposed by the Robotics and computer vision communities.

\begin{table}[!ht]
    \centering
	\caption{Benchmark datasets for disparity estimation. Stereo number is the total number of stereo pairs used for both training and testing.}
	\label{Table:data}
	\resizebox{\textwidth}{!}{
	\begin{tabular}{c|ccccc}
		\toprule
		\multirow{2}*{\bfseries Data} & \multirow{2}*{\bfseries Year} & \multirow{2}*{\bfseries Scene}  & {\bfseries Stereo} & {\bfseries Disparity} & \multirow{2}*{\bfseries Citation}\\
		~ & ~ & ~ &  {\bfseries number} & {\bfseries density} & ~ \\
		\hline
		Middlebury \textit{V2} & 2003-2006 & indoor & 32 & dense & \citep{scharstein2003high} \\
		Middlebury \textit{V3} & 2014 & indoor & 33 & dense & \citep{scharstein2014high} \\
		ETH3D & 2017 & indoor + outdoor & 47 & dense & \citep{schops2017multi} \\
		InStereo2K & 2020 & indoor & 2050 & dense & \citep{bao2020instereo2k} \\
		\hline
		KITII\textit{2012} & 2012 & outdoor(driving) & 389 & sparse & \citep{geiger2012we} \\
		KITTI\textit{2015} & 2015 & outdoor(driving) & 1600 & sparse & \citep{mayer2016large} \\
		Cityscapes & 2016 & outdoor(driving) & 24998 & dense & \citep{cordts2016cityscapes} \\
		Drivingstereo & 2019 & outdoor(driving) & 182188 & sparse & \citep{yang2019drivingstereo} \\
		ApolloScape & 2019 & outdoor(driving) & 5165 & sparse & \citep{huang2019apolloscape} \\
		\hline
		SatStereo  & 2019\setfootnotemark\label{first} & satellite & 72 & dense & \citep{patil2019new} \\
		DFC2019 & 2019 & satellite  & 8634  & dense & \citep{bosch2019semantic} \\
		\bottomrule
	\end{tabular}
	}
	\afterpage{\footnotetext[\getrefnumber{first}]{The IARPA dataset was published in 2016.}}
\end{table}

\subsection{Photogrammetry datasets}

In aerial photogrammetry, benchmark datasets focus mainly on MVS dense matching \citep{cavegn2014benchmarking}. For satellite imagery, evaluation is usually performed on the Digital Surface Model (DSM) or height map \citep{rupnik20183d, bosch2019semantic}. The pipeline contains other steps, for example, point cloud fusion and DSM generation \citep{cournet2020ground}, and the data is to evaluate the whole pipeline, not the sole stereo dense matching approach. 
IARPA dataset \citep{bosch2016multiple} is a MVS 3D reconstruction challenge for satellite images and provides the aerial corresponding LiDAR.
A recent satellite image stereo benchmark from LiDAR DSM \citep{patil2019new} uses the IARPA dataset . With more and more DL-based methods applied in the geospatial context, there are more satellite training data \citep{bosch2019semantic}.

\section{Aerial Benchmark data generation}\label{sec:aerialdatagen}

While ML, especially DL-based dense matching methods outperform traditional ones, they usually need a lot of training data (including the ground truth) to achieve good performance. Moreover, the fair evaluation requires a test set separate from training data. Generating such training data from real scenes is challenging. Moreover, the quality of the training data directly impacts the performance of the trained model, so great care must be taken when producing it. 
To avoid possible errors introduced by grid interpolation of point cloud, we propose a method to generate sparse disparity ground truth from images and LiDAR point cloud which can be registered if they are not well aligned, then we produce epipolar pairs for training DL architectures, and finally, we evaluate both traditional ML-based and DL-based dense epipolar matching approaches. 


In this section, we describe the production of these epipolar images and corresponding ground truth disparity maps from the original aerial images and LiDAR. Our workflow is mainly based on the open-source MicMac library \citep{pierrot2014micmac}. As far as we know, such training data are rare in the photogrammetry community. Thus, we openly release our dataset (including epipolar pairs and corresponding ground truth disparity maps).

\subsection{Ground truth sparse disparity generation}

Considering the configuration of the stereo dense matching methods investigated in the experiments, we generate our ground truth as a sparse disparity in epipolar geometry.
Epipolar image generation is described in Section \ref{sec:epipolar}. In real scenes, several factors influence the quality of a LiDAR-based disparity ground truth :
\begin{itemize}
    \item Image orientation accuracy (inner consistence) which mainly depends on the quality of the bundle adjustment and should be controlled in order to guarantee sub-pixel accuracy.
    \item Co-registration accuracy between image and LiDAR (external consistence): this is addressed in Section{\ref{sec:ImageLiDARReg}}
    \item Occlusions between the image and the LiDAR: Even if an image and a LiDAR point cloud are acquired from the same platform, the scene is acquired from multiple different viewpoints. Consequently, the whole LiDAR point cloud is probably not seen in the image to which it projects. To address those scenarios we developped a specific occlusion handling scheme described in Section \ref{sec:occlusion}. 
    \item Changes between the image and the LiDAR: if the images and LiDAR are not acquired simultaneously, some disparity evaluations were poor not because of the evaluated methods but due to changes (constructions, vegetation growth,...) in the scene between the image and LiDAR acquisitions. In DL-based semantic segmentation, such inconsistencies in the ground truth are called \textit{noise label} and  several research works such as self-ensemble label filtering \citep{nguyen2019self} address this issue. Using a similar idea, we propose a two-step based strategy to enhance the ground truth which is described in the section \ref{sec:change}.
\end{itemize}

\subsubsection{Image and LiDAR registration}
\label{sec:ImageLiDARReg}

In practice, it is not easy to find well-registered images and LiDAR due to the different purpose of data collection, in particular if they are not collected simultaneously. If the image and LiDAR is not well-registered, we propose to refine the registration in order to improve their consistency. 
Registration between aerial optical images and LiDAR is another difficult topic \citep{zhou2016fast}. 
In our case, given that the initial orientation is usually quite accurate (close to the real orientation), a simple iterative refinement based on ICP \citep{huang2018registration} is proposed. 
We start by computing the 3D tie points corresponding to the feature points matched during the bundle adjustment. As the initial orientation is close enough to the real position, these tie points lie close to LiDAR points.
As the noise in LiDAR is small, the closest LiDAR point is regarded as a ground control point (GCP).
Using these GPCs, the orientation can be estimated again, and the whole process can be repeated until convergence, iteratively improving the orientation.
It is difficult to know when to the terminate the loop, we evaluated this approach by measuring the disparity error between an SGM result and the LiDAR to indicate the accuracy of the orientation of images, the analysis is refer to section \ref{sec:ori}. 

\subsubsection{Epipolar image generation}
\label{sec:epipolar}

The learning based dense matching algorithms exploited assume an epipolar rectification of the images where the counterpart of a pixel in one image lies in the same line of the other image. Moreover some approaches are limited in the maximum size of the input image. The first step of our pipeline is to generate epipolar image pairs, and only the image pairs with sufficient footprint overlap are considered. Coarse image footprints are obtained using an approximate height of the scene and the Computational Geometry Algorithms Library(CGAL) \citep{cgal:f-i-20b} is used to compute their intersections. In our experiment, only images pairs with an intersection area above half the image footprint are considered for epipolar rectification, which was done using a recent module of the MicMac library \citep{deseilligny2020epipolar}. The corresponding orientation parameter files are generated in order to allow the direct projection of 3D point in the rectified images. 

\subsubsection{Occlusion aware point cloud projection}
\label{sec:occlusion}

The main question of occlusion aware projection is to ensure that a LiDAR point is only projected in an (epipolar) image if the corresponding 3D point is actually seen in this image. As images can only capture the apparent surface, we only keep the first LiDAR echo. Also, we filter out outliers based on the point cloud density using a grid \citep{cho2004pseudo}, if there is no neighborhood points for one points in radius 3m, this point is noise point, and removed.

The ground truth disparity maps used for both training and evaluation are computed from the LiDAR point cloud.
A perspective projection model is used to project the 3D points to the image focal plane.
The disparity $d$ is defined in \Cref{equation:disparity}.
\begin{equation}
\label{equation:disparity}
d = x_l - x_r
\end{equation}
where $x_l$ and $x_r$ are the projection coordinate on the $x$ axis in the image plane after projecting the 3D point cloud to the left and right images. By construction, they project at the same $y$ value.

\begin{figure}[t]
	\centering
	\subfigure[Occlusion in stereo]{
		\label{Figure.occlusion:a}
		\centering
		\includegraphics[width=0.5\linewidth]{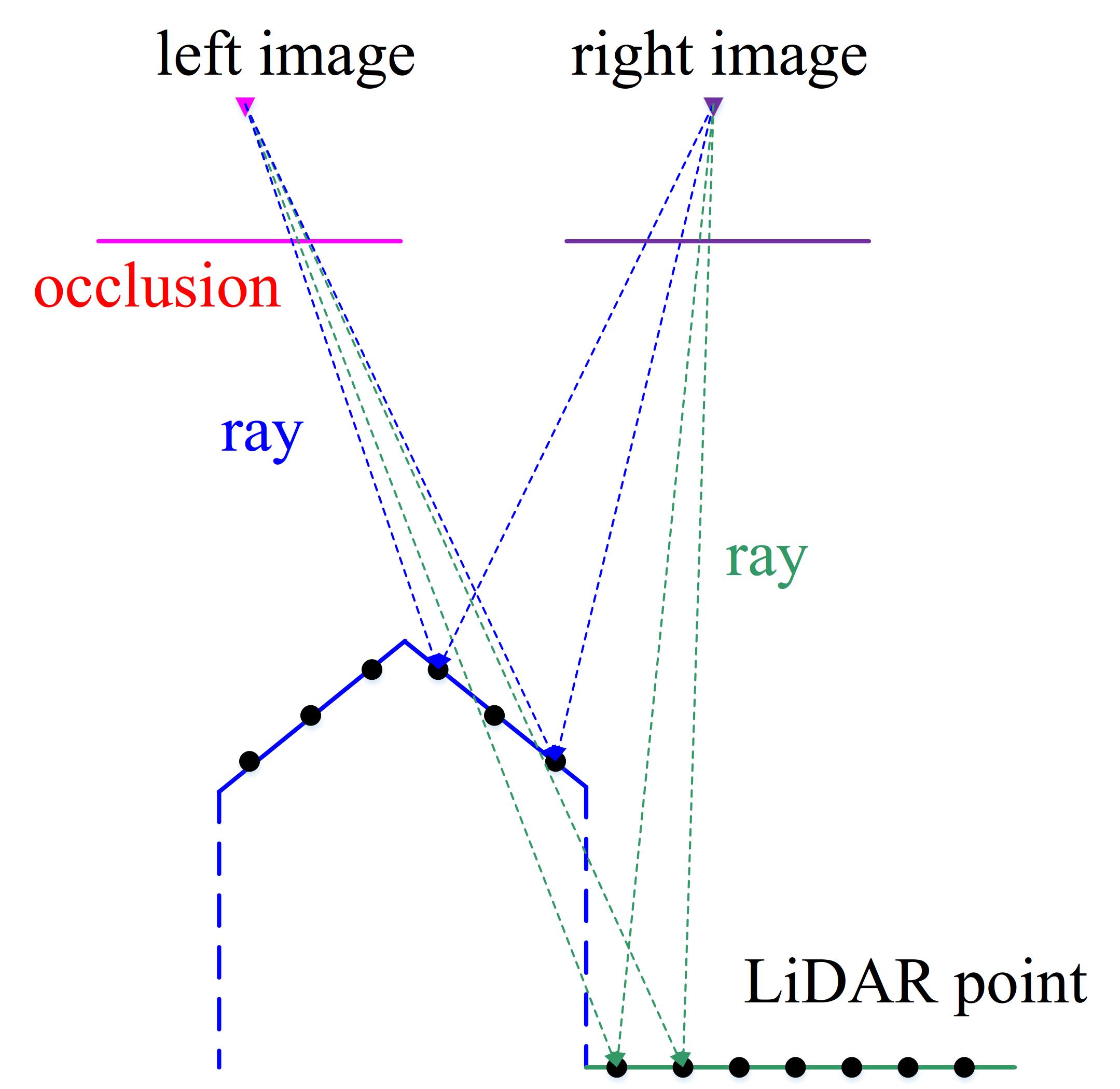}
	}
	\subfigure[Points on image plane]{
		\label{Figure.occlusion:b}
		\centering
		\includegraphics[width=0.4\linewidth]{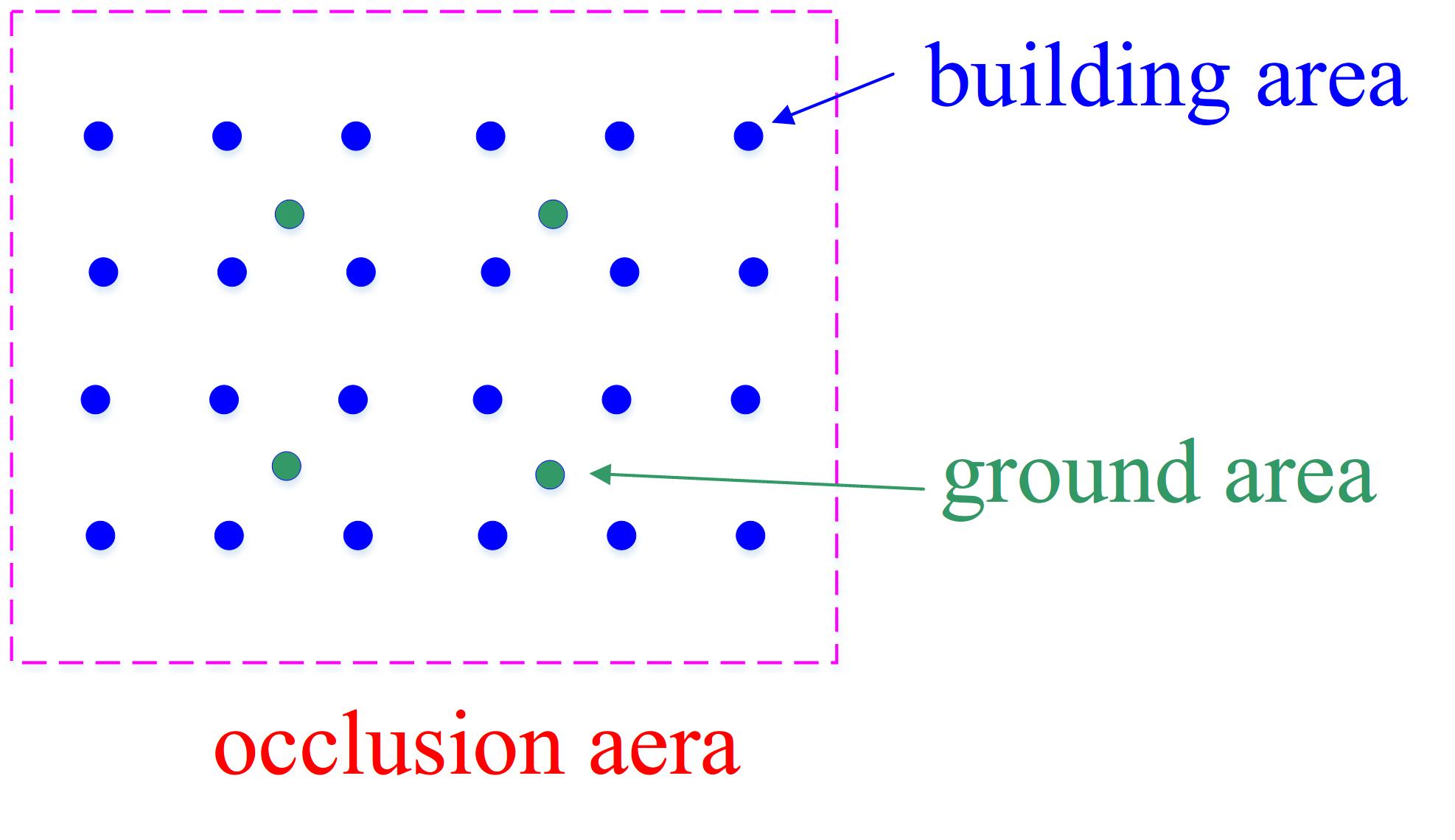}
	}
	\caption{Occlusion in projection.}
	\label{Figure.occlusion}
\end{figure}

Because the point cloud is sparse, it is difficult to predict the visibility of each point in the image, which is a well known issue \citep{bevilacqua2017joint}.
An example of occlusion (in the left image) is shown in \Cref{Figure.occlusion}. Points from the ground area should be removed as in reality, they are occluded by the roof.

In our previous work \citep{wu2021new}, an image space filter based on the disparity was used. The idea is that if a point is surrounded by closer points in image space, it is probably occluded. But we found that disparity discontinuities (on vertical surfaces for instance) were not well handled by this filter as the disparity varies rapidly there.
To solve this issue, we propose a ray tracing based occlusion detection before applying the filter. The idea is to reconstruct a continuous 3D surface from the points, then use ray tracing to detect the occlusions as intersections of the sight ray (from a 3D point to an image optical center) with this surface. However 3D surface reconstruction from point clouds is a complex and time consuming problem. Considering the globally 2.5D nature of aerial LiDAR, we propose a simple 2D Delaunay triangulation \citep{shewchuk2008two} in the horizontal plane to generate a simple surface mesh from the point cloud.
As illustrated in Equation \ref{equation:occlusion},  $z$ relates to the disparity $d$ directly. Because the height ($z$ value) is not considered in 2D Delaunay triangulation, this can cause errors in areas where the $z$ value can be multiple (overlaying roof and ground for instance). To handle this issue we filter out triangle for which the difference in the $z$ coordinate is too large before ray tracing. In our experiment, the maximum $z$ difference for a triangle is 2 meters.
As we generate ground truth disparity maps for the right image, we trace 2 rays from each LiDAR point to the optical center of the right and left images. We use the AABB Tree implemented in CGAL \citep{cgal:f-i-20b} to accelerate the intersection of these rays with the valid triangles of the mesh. If both rays do not intersect the mesh, the point is projected in both images and the disparity is computed as the difference between the column indices in the right and left images. Because the pair is epipolar, the projections have the same line index.
Note that we could have kept points seen only in the right image. In practice, adding this points decreases the quality of the result as presented in section \ref{sec:occ_analysis}. 

In epipolar geometry, the disparity at a pixel of the right image is related to the depth of the 3D point seen by that pixel \citep{jain1995machine}:
\begin{equation}
\label{equation:occlusion}
z = \frac{b \cdot f}{d} 
\end{equation}
where $z$ is the depth, $b$ is base line length, $f$ is the focal length, $d$ is the disparity. The disparity is negatively correlated to the depth.
For aerial images, the disparity is related to the height of objects, such that ground points have larger disparities.

Finally, after ray tracing, we apply an image space disparity filter to handle areas with large $z$ variance as proposed in \citep{biasutti2019visibility}: if the difference between the disparity and its median in an image space neighborhood is larger than a threshold, as shown in Equation \ref{equation:occlusion_threshold}, $d_{max}$ and $d_{min}$ is the maximum and minimum disparity in the window,  then the point is removed, which filters occluded points quite effectively.
\begin{equation}
\label{equation:occlusion_threshold}
\alpha = \frac{d - d_{min}}{d_{max} - d_{min}} 
\end{equation}

\subsubsection{Removing areas with scene changes}
\label{sec:change}

Because the wrong training data will deteriorates the performance, our aim is not to detect changes precisely but to remove the epipolar pairs in which significant changes occurred between image and LiDAR acquisitions from our dataset. If both are acquired simultaneouly, this step can be skipped.

First, we use SGM(CUDA) \citep{hernandez2016embedded} to select high accuracy (with high evaluation score) stereo pairs. Then, we fine-tune PSMNet \citep{chang2018pyramid} on these pairs, then use the fine-tuned model on the whole dataset. Finally the  stereo pairs with large errors are removed.

In this paper, we evaluate the quality of dense matching using cumulative histograms (proportion of pixels where the error between the estimated disparity and the ground truth disparity map is smaller than $N$) which are the exact opposite of the $N$-pixel error (proportion of pixels where the error is larger than $N$).
While 1-pixel error is a good indicator for the precision of the disparity (we count only pixels where the disparity is perfectly estimated), larger $N$s indicate the robustness of the method (ability to give a correct, even if inaccurate, disparity in every situation). 
In order to cope with possible scene changes, we decided to discard all stereo pairs for which the 1-pixel error of SGM(CUDA) is over 60\%.
SGM(CUDA) was chosen as it is our baseline, it is the fastest method and the quality of its results is quite independent from the scene.
A 1-pixel error over the 60\% threshold indicates that the matching problem is too hard (vegetation, water or large untextured areas), or that there are significant changes in the scene between the image and LiDAR acquisitions.
As we do not want to filter out "hard" cases (which would bias our evaluation), we train PSMNet on the "easy" dataset (with SGM 1-pixel error below 60\%) and finally filter out the stereo pairs for which the PSMNet 1-pixel error is over 40\% which we consider is a good indicator of change.
More accurate change filtering can be obtained by iterating this second step (learning PSMNet on the new "easy/unchanged" dataset to have a better model to filter out changes again).
This strategy can also be used to handle the asynchronous image and LiDAR acquisition.

\subsection{Datasets presentation}
\begin{figure}[tp]
	\centering
	\subfigure[ISPRS Vaihingen]{
		\label{Figure.data:a}
		\centering
		\includegraphics[width=0.4\linewidth]{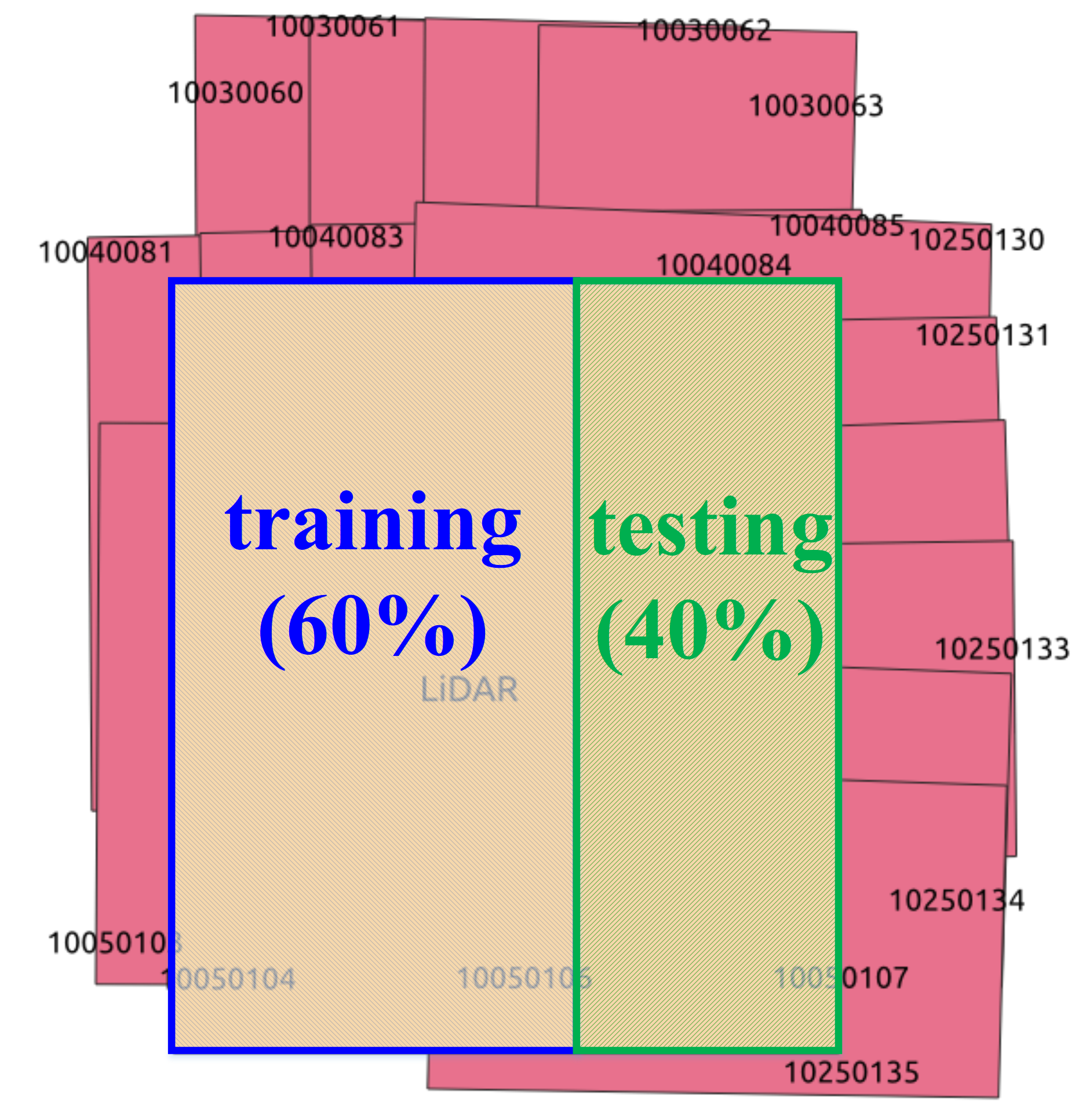}
	}
	\subfigure[EuroSDR Vahingen]{
		\label{Figure.data:b}
		\centering
		\includegraphics[width=0.4\linewidth]{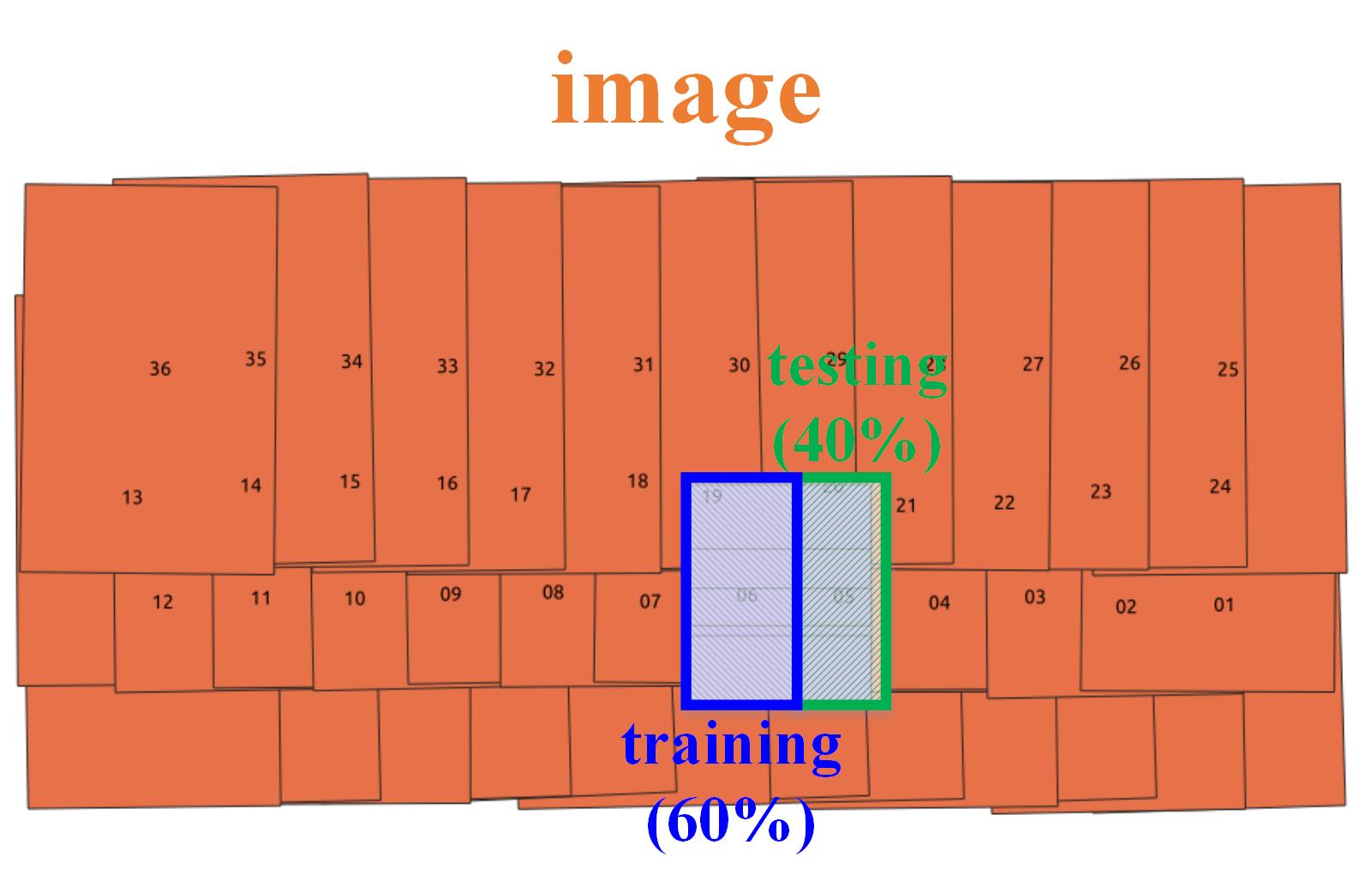}
	}
	
	\subfigure[Toulouse UMBRA]{
		\label{Figure.data:c}
		\centering
		\includegraphics[width=0.4\linewidth]{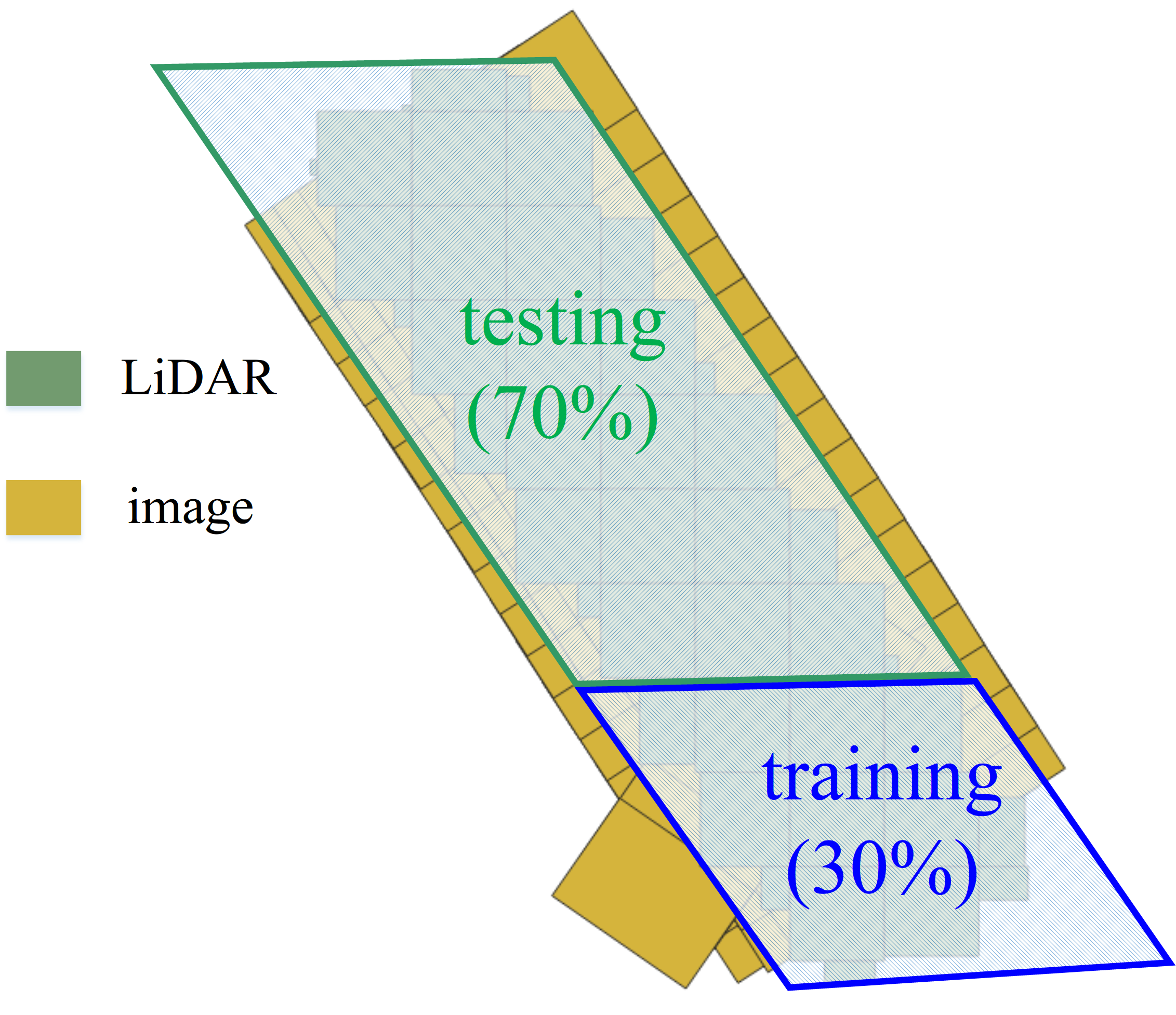}
	}
	\subfigure[Toulouse Metropole]{
		\label{Figure.data:d}
		\centering
		\includegraphics[width=0.4\linewidth]{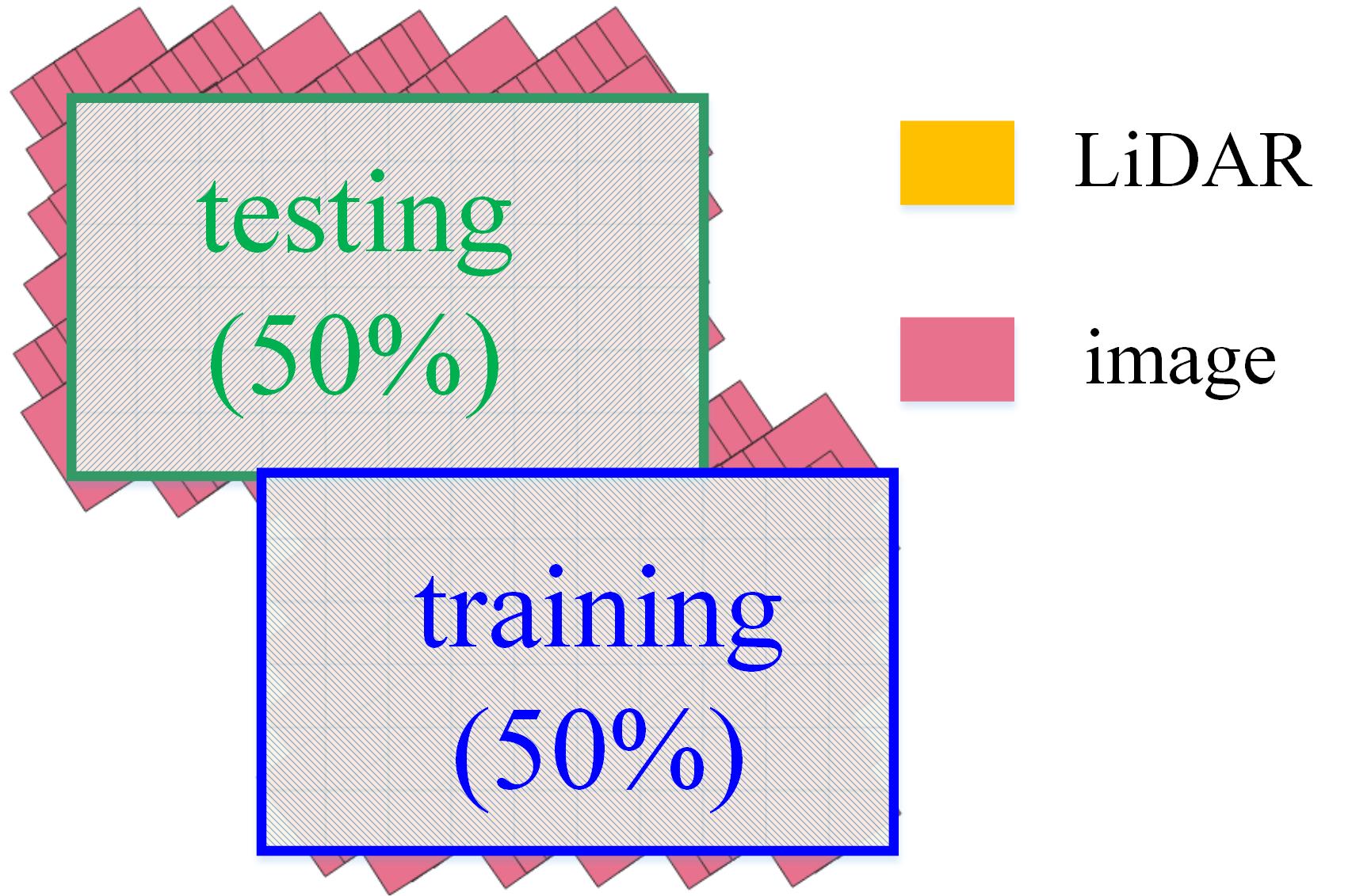}
	}

	\subfigure[Enschede]{
		\label{Figure.data:e}
		\centering
		\includegraphics[width=0.4\linewidth]{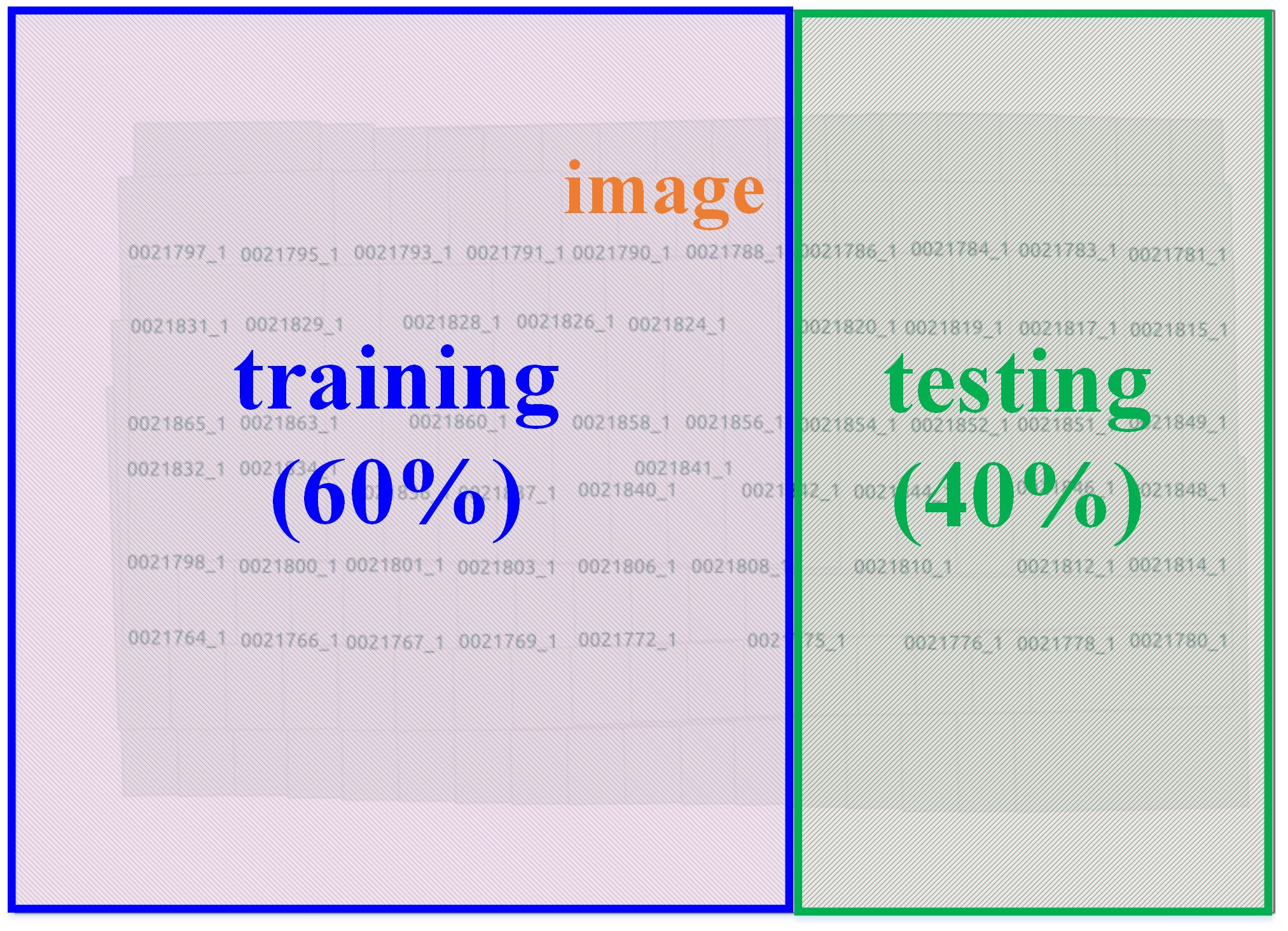}
	}
	\subfigure[DublinCity]{
		\label{Figure.data:f}
		\centering
		\includegraphics[width=0.4\linewidth]{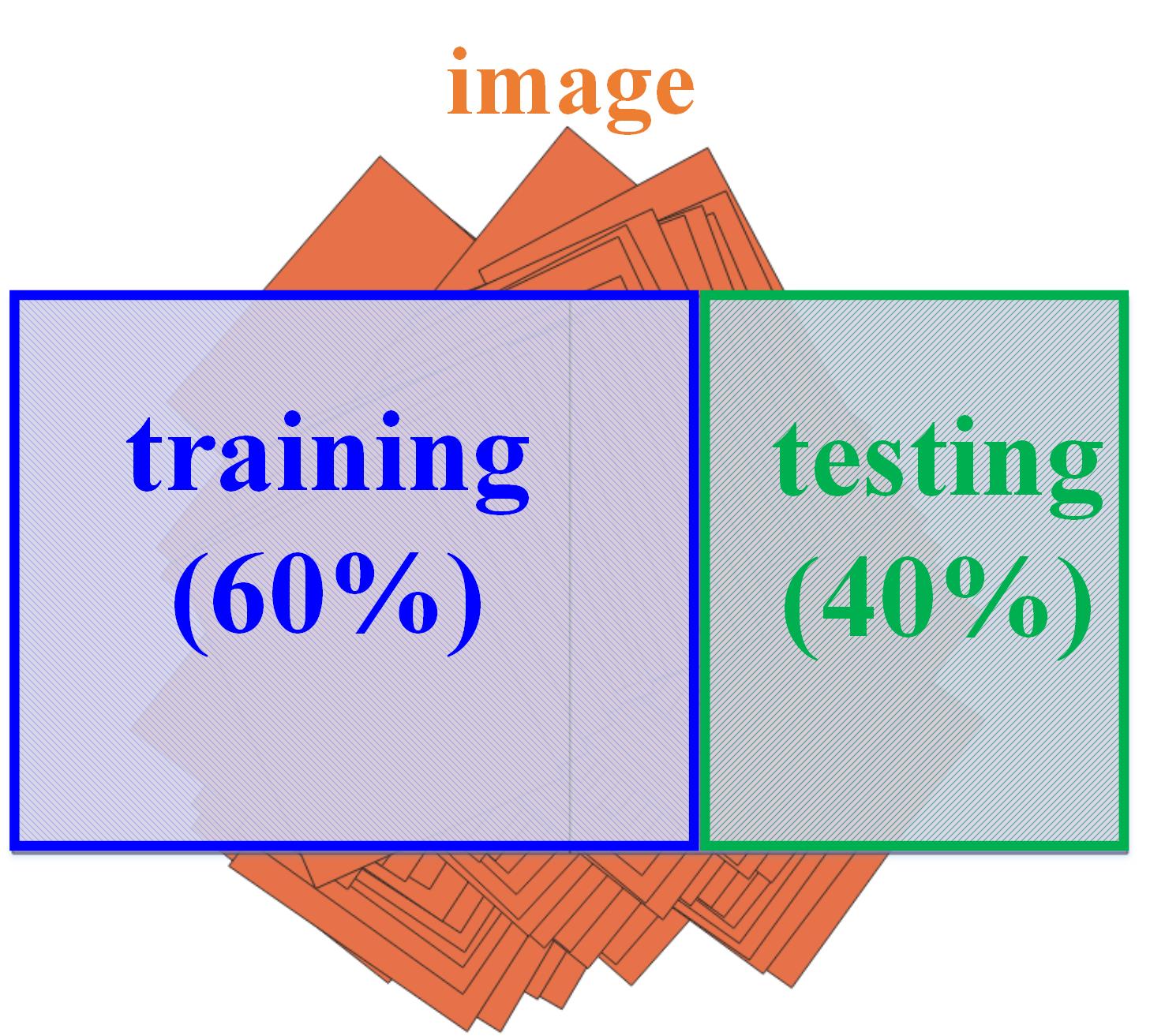}
	}
	\caption{All the dataset introduction.}
	\label{Figure.data}
\end{figure}

Along with this paper, we publish 6 datasets, as listed in \Cref{Table:aerial} produced with the pipeline described above applied to images and LiDAR from different countries. For certain areas, there are several datasets with different resolutions and acquired at different times as summarized in \cref{Figure.data}.

\begin{table}[!ht]
	\caption{processed dataset for experiment.}
	\label{Table:aerial}
	\resizebox{\textwidth}{!}{
	\begin{tabular}{l|ccccc}
		\toprule
		{\bfseries dataset} & {\bfseries color} & {\bfseries GSD(cm)} & {\bfseries LiDAR (pt/$m^2$)}  & {\bfseries train+test} \\
		\hline
		ISPRS-Vaihingen   & {IR-R-G} & {8} & {6.7} & {640 + 516}  \\
		\hline
		EuroSDR-Vaihingen & {RGB} & {20} & {6.7} & {421 + 353} \\
		\hline
		Toulouse-UMBRA & RGB & 12.5 & 2-4 & 7409 + 21458 \\
		\hline
		Toulouse-Metropole & RGB & 5 & 8 & 9813 + 14241 \\
		\hline
		Enschede & RGB & 10 & 10 & 871 + 819 \\
		\hline
		DublinCity & RGB & 3.4 & 250-348 & 35128+6799  \\
		\bottomrule
	\end{tabular}
	}
\end{table}

\subsubsection{ISPRS Vaihingen dataset}
The Vaihingen dataset is from the ISPRS 3D reconstruction benchmark which provides a good registration of images and LiDAR point cloud. The dataset is composed of 20 images with a depth of 11bit \citep{rottensteiner2012isprs}. As this dataset has an accurate geolocation, and image and LiDAR were collected nearly at same time, we did not apply the image/LiDAR registration and change filtering presented above, such that  all the epipolar pairs are used and split between train and test sets.

\subsubsection{EuroSDR Vaihingen dataset}
This data is an aerial image matching benchmark dataset \citep{haala2013landscape}, the resolution is different from the ISPRS Vaihingen dataset, but the LiDAR is same. Because the resolution is low, and  the experiment area is not large, all the possible pairs which have overlap are used. The image data was collected nearly the same time as ISPRS-Vaihingen \citep{cramer2010dgpf}. As the dataset size is smaller than \textit{ISPRS Vaihingen}, all the epipolar pairs are used and split into training and testing set.

\subsubsection{Toulouse UMBRA dataset}
The dataset is published in \citep{adeline2013description}, even though there is a NIR band in the origin dataset, only RGB bands are used. The image orientation is not accurate so we applied our image/LiDAR registration.
The image was collected in 2012 and LiDAR was collected in 2013, so we also applied our change filtering. As the dataset is very large, the training (1200 pairs) and testing (870 pairs) subsets are randomly selected.

\subsubsection{Toulouse Metropole dataset}
The dataset was provided to the AI4GEO project (see Section \ref{sec:ack}) by the metropole of Touloude. The image and LiDAR were collected at nearly same time so we did not apply change filtering. The dataset has a higher resolution than the Toulouse UMBRA dataset, and covers a larger area which contains the Toulouse UMBRA area. Given its large size, the traning (1200 pairs) and testing (1116 pairs) subsets are randomly selected.

\subsubsection{Enschede dataset}
This dataset is used and presented in \citep{zhang2018patch}. The image was collected in 2011 and the LiDAR is from AHN2 which was collected in 2012 in the Enschede area \citep{ahn2021} so we applied our change filtering. The original dataset has an oblique setting with 5 cameras. Considering that the other datasets do not contain oblique images, we selected only the nadir images to generate the epipolar pairs. The base to height ratio is large in this dataset. 
The orientation is generated using Micmac with ground control points and our image/LiDAR registration is applied.
All the epipolar pairs are used and split into training and testing subsets.

\subsubsection{DublinCity dataset}
The dataset is published in \citep{laefer2015, zolanvari2019dublincity}, the image and LiDAR were collected nearly same time so we did not filter changes.  The original dataset contains very high density LiDAR data and both  vertical and oblique images. Once again, only the vertical images are used. Because the dataset large size, we randomly sampled 1200 pairs for training and 257 pairs for testing. 
For DublinCity dataset, the orientation files are provided, but the orientation is not accurate, so the image/LiDAR registration refinement is applied.

\section{Evaluation and discussion}\label{sec:eval}

In our experiments, we evaluated traditional (providing a baseline), machine learning (ML) and deep learning (DL) methods. For DL, we have tried to cover the most significant works from the litterature but we had to discard:
\begin{enumerate}
    \item DispNet (first end-to-end DL-based stereo dense matching method), as it requires a dense disparity ground truth and ours is sparse.
    \item GC-Net (first dense matching method using a 3D CNN to process the cost volume) as it has no official code release.
\end{enumerate}
As listed in \Cref{Table:method}, 11 methods were selected in the experiments, categorized with traditional method, machine learning based method, DL-based method and end-to-end method.



\begin{table}[t]
	\centering
	\caption{A synthesis of the explored methods.}
	\label{Table:method}
	\resizebox{\textwidth}{!}{
            \tiny
            \begin{tabularx}{\textwidth}{>{\hsize=0.7\hsize}X|>{\hsize=1.0\hsize}X|>{\hsize=0.9\hsize}X|>{\hsize=1.4\hsize}X|>{\hsize=1.0\hsize}X}
			\noalign{\hrule height 1.2pt}
			{\bfseries method} &  {\bfseries feature extraction} &  {\bfseries cost aggregation}  &  {\bfseries brief introduction} &  {\bfseries citation} \\
			\noalign{\hrule height 1.2pt}
			MICMAC & NCC based feature &  dynamic programming & variant of SGM implementation &  \citep{mpd:06:sgm} \\
			\hline
			SGM (GPU) & census based feature & dynamic programming & SGM implementation based on GPU  & \citep{hernandez2016embedded} \\
			\hline
			GraphCuts & intensity and gradient consistency  & local expansion & add plane surface assumption in GraphCuts framework & \citep{taniai2017continuous} \\
			\noalign{\hrule height 1pt}
			CBMV (SGM) &  man-crafted features with  RF fusion & dynamic programming &  Combine man-crafted features with RF to produce similarity, use SGM optimization  & \citep{batsos2018cbmv} \\
			\hline
			CBMV (GraphCuts) &  man-crafted features with RF fusion & local expansion &  Combine man-crafted features with RF to produce similarity, use GraphCuts optimization  & \citep{batsos2018cbmv} \\
			\noalign{\hrule height 1pt}
			MC-CNN & window-based 2D CNN feature & dynamic programming & crop the image into patchs, 2D CNN to extract feature similarity, then use SGM optimization & \citep{zbontar2016stereo} \\
			\hline
			DeepFeature & window-based 2D CNN feature &  dynamic programming & more efficient 2D CNN to extract feature similarity, then use SGM optimization & \citep{luo2016efficient} \\
			\noalign{\hrule height 1pt}
			PSM net & 2D CNN encoder-decoder feature & 3D CNN encoder-decoder & a typical DL-based dense matching pipeline  & \citep{chang2018pyramid}  \\
			\hline
			HRS net & 2D CNN encoder-decoder feature & 3D CNN encoder-decoder &  in the feature extraction step, unpooling to obtain high resolution feature  &  \citep{yang2019hierarchical} \\
			\hline
			DeepPruner & 2D CNN encoder-decoder feature & 3D CNN encoder-decoder &  prune the search range using CNN based patch match & \citep{duggal2019deeppruner} \\
			\hline
			GANet & 2D CNN encoder-decoder feature & 3D CNN encoder-decoder & guided aggregation network in 3D CNN &  \citep{zhang2019ga} \\
			\hline
			LEAStereo & 2D CNN encoder-decoder feature & 3D CNN encoder-decoder & Use Neural Architecture Search to search the network architecture & \citep{cheng2020hierarchical} \\
			\noalign{\hrule height 1.2pt}
		\end{tabularx}
	}
\end{table}

\subsection{Experiment details}

In order to evaluate the performance without bias, all the experimented methods use the official code release. 
\textit{MICMAC} is implemented in MicMac written in C++ \citep{micmac2020}. \textit{SGM (GPU)} is based on C++ and CUDA \citep{sgm2020}. 
\textit{GraphCuts} is written in C++ \citep{LocalExpStereo2020}. \textit{CBMV} is written in C++, CUDA (SGM implementation) and Python \citep{CBMV2020}. CBMV has two pipelines in optimization step, dynamic programming optimization (\textit{CBMV (SGM)}) and Graphcut (\textit{CBMV (GraphCuts)}) which has the same implementation with \textit{GraphCuts}. 
\textit{MC-CNN} is based on C++, CUDA (SGM implementation) and Lua torch \citep{mccnn2020}. \textit{DeepFeature} \citep{cvpr16stereo} is similiar to \textit{MC-CNN}.
\textit{PSM net} \citep{PSMNet2020}, \textit{HRS net} \citep{HRSnet2020}, \textit{DeepPruner} \citep{DeepPruner2020}, and \textit{LEAStereo} \citep{LEAStereo2021} are based on Pytorch. \textit{GANet} is based on CUDA and Pytorch \citep{GANet2020}. 

For DL-based methods, the experimental setup and especially the hyperparameters are important.
In our experiments, we compare DL-based methods pretrained on the KITTI dataset and fine-tuned on our new dataset.
We found that batch size has a strong impact on the memory requirements and on the accuracy. 

The training data size influences the performance of the training model: a larger, more diverse training dataset usually gives a better result.
For each dataset, the number of available stereo pairs is different. In order to make the comparison more impartial, we set the maximum number of training stereo pairs used per dataset to 1200 (it can be less if the dataset has less than 1200 pairs).

The training strategy differs between methods: hybrid methods are pixel based, while end-to-end based methods are stereo pair based, and some methods crop the training images. The training time is also varies significantly between different methods. Finally, we select the result with the best loss for each method.
In the experiment, we use the cluster of French National Centre for Space Studies(CNES), the GPU card is NVIDIA Tesla V100, the memory size is 32GB. In order to make the statistics of the computation time of training, all the experiments are training on Toulouse Metropole dataset, the detail is in the \Cref{Table:time}. The training strategy of \textit{MC-CNN} and \textit{DeepFeature} is different from end-to-end method, because learning similarity is pixel pair based, and all the valid pixels which has LiDAR points are processed in the training, but in the training of end-to-end methods randomly crop the input image, the cropped size uses the default value, this means not all the data is used in the training in the end-to-end method.
Considering the total image number is same, so the train time is training time of each epoch.

\begin{table}[!ht]
	\centering
	\caption{Training configuration and time for each method.}
	\label{Table:time}
	\begin{tabular}{c|ccccc}
		\toprule
		name & {\bfseries epoch} &  {\bfseries croped} & {\bfseries batch size} & {\bfseries train time/s} & {\bfseries pairs} \\
		\midrule
		MC-CNN & 10 & $\times$ & 128 & 17591 & 1200 \\
		DeepFeature & 20 & $\times$ & 128 & 4121 & 1200 \\
		\midrule
		PSMNet & 500 & 256x512 & 12 & 216 & 1200 \\
		HRS net & 700 & 512x768 & 28 & 108 & 1200 \\
		DeepPruner & 900 & 256x512 & 16 & 108 & 1200 \\
		GANet  & 500 & 256x256 & 4 & 1260 & 1200  \\
		LEAStereo & 500 & 288x576 & 4 & 451 & 1200 \\
		\bottomrule
	\end{tabular}
\end{table}

\subsection{Orientation accuracy}
\label{sec:ori}
\begin{figure}[tp]
	\centering
	\includegraphics[width=0.7\linewidth]{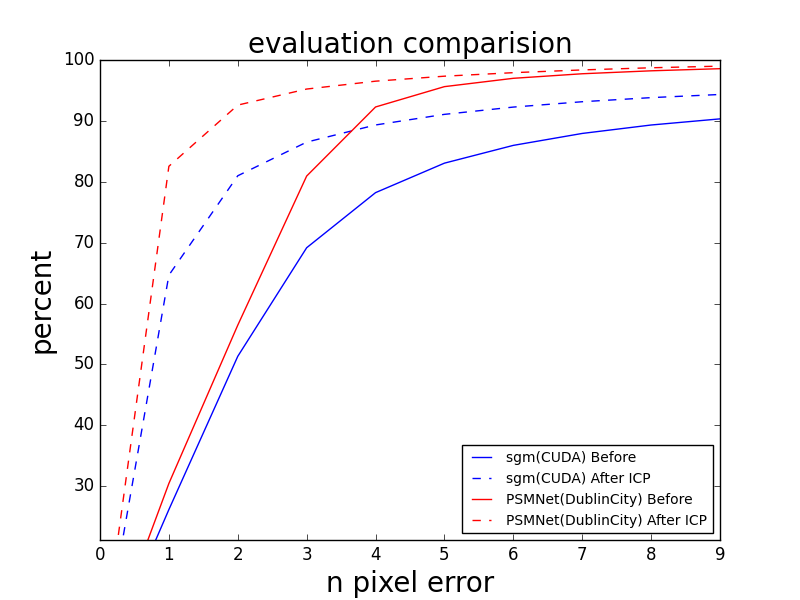}
	\caption{Cumulated histogram of the disparity error between the SGM and PSMNet results and the LiDAR.}
	\label{Figure.icp}
\end{figure}

In our experiments, we found that the accuracy of the image orientations influences the evaluation of the stereo methods. According to our experience, the 1-pixel error is a good indicator for orientation accuracy which is why we used it for image/LiDAR registration. The results on the DublinCity dataset \Cref{Figure.icp} show that our registration allows a significant decrease in this error.

\subsection{Occlusion analysis}
\label{sec:occ_analysis}
Occlusion can be handled in two different ways in the ground truth generation: a ground truth disparity map is produced at a pixel location of the left image if a 3D LiDAR point projecting on that pixel:
\begin{enumerate}
    \item can be seen in both views
    \item can be seen at least in the left view
\end{enumerate} 
We led an experiment to evaluate which strategy is the best.
We divided the pixels into two categories: (seen) pixels which can be seen in two views, and (occluded) pixel which can only be seen in the left view.   
Because the occluded pixels form only a small part of the whole (seen + occluded) set, we propose two train settings: training with the whole (seen + occluded) dataset and training with only the seen pixels.
For the evaluation, we propose three settings: evaluating on \textbf{all} (seen + occluded) pixels, on \textbf{seen} pixels only and on \textbf{occ}(luded) pixels only.
Note that the DublinCity dataset has a much denser LiDAR than other datasets and many tall buildings so there is a larger proportion of occluded pixels. As shown in \cref{Table:occlu}, adding occluded pixels to the training set slightly improves the evaluation on occluded pixel, but there is no significant difference on the seen pixels, because the occluded pixels have a small percentage in the whole image, so for the evaluation of all pixels, using occluded pixels decreases the accuracy on seen pixels. So in the other experiments, only the non occluded pixels are analysed.

\begin{table}[!ht]
	\centering
	\caption{Influence of training with and without occ(luded) pixels for \textit{PSM net}.}
	\label{Table:occlu}
	\begin{tabular}{c|c|ccccc}
		\toprule
		training & testing & {\bfseries \textless 1-pixel} & {\bfseries \textless 2-pixel} & {\bfseries \textless 3-pixel} & {\bfseries \textless 5-pixel} & {\bfseries \textless 9-pixel} \\
		\midrule
		all & all & 84.62 & 92.83 & 95.31 & 97.37 &  98.99 \\
		seen & all & 84.69 & 92.84 & 95.31 & 97.37 & 99.00 \\
		\midrule
		all & seen & 84.62 & 92.83 & 95.31 & 97.37 & 98.99 \\
		seen & seen & 84.69 & 92.84 & 95.31 & 97.37 & 99.00 \\
		\midrule
		all & occ & 25.13 & 37.38 & 43.41 & 50.28 & 59.48 \\
		seen & occ & 24.45 & 36.71 & 43.12 & 49.97 & 59.38 \\
		\bottomrule
	\end{tabular}
\end{table}

\subsection{ML performance on a single dataset}
\begin{figure}[tp]
	\centering
	\subfigure[Result on ISPRS Vaihingen]{
		\label{Figure.evalute:a}
		\centering
		\includegraphics[width=0.45\linewidth]{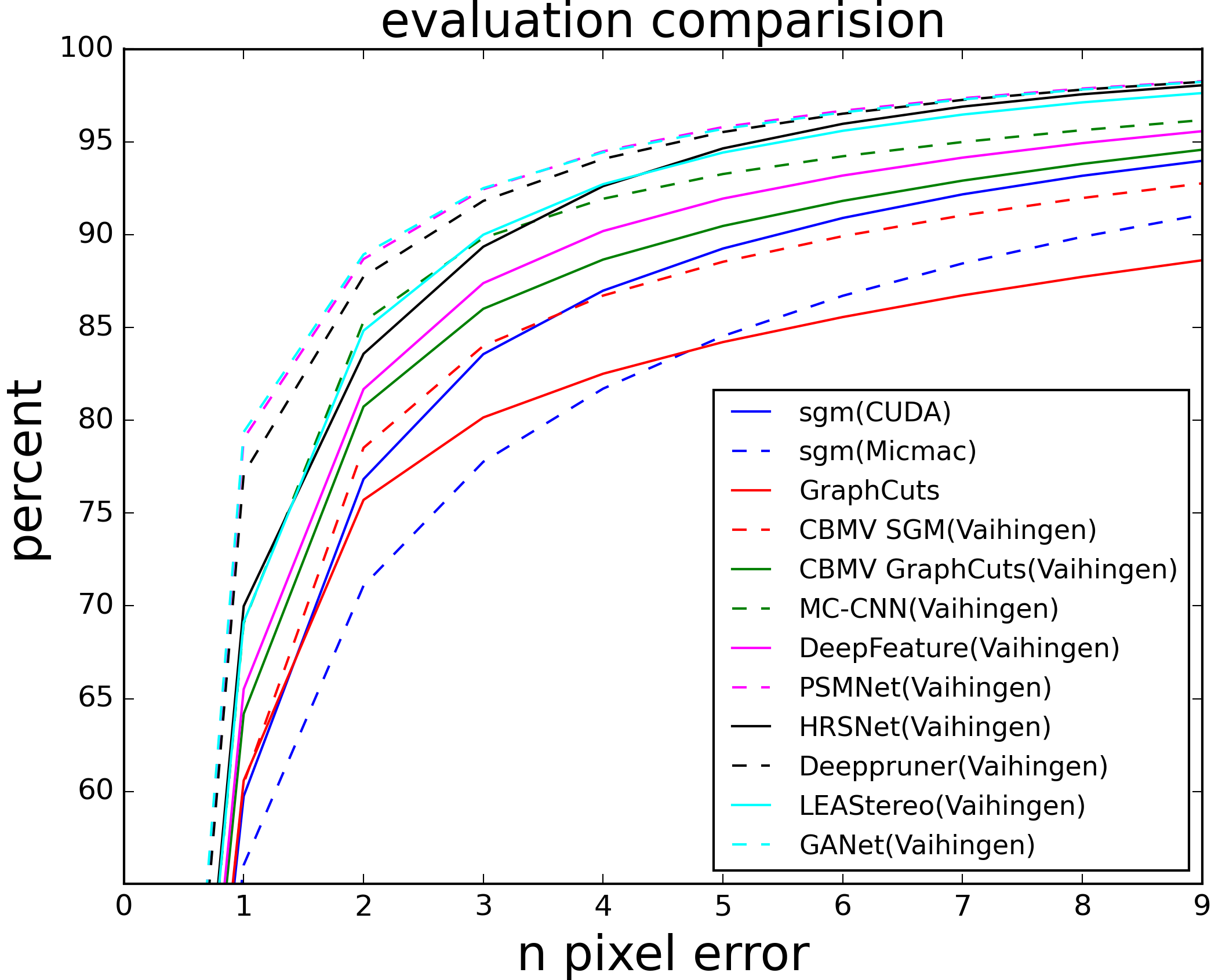}
	}
	\subfigure[Result on EuroSDR Vaihingen]{
		\label{Figure.evalute:b}
		\centering
		\includegraphics[width=0.45\linewidth]{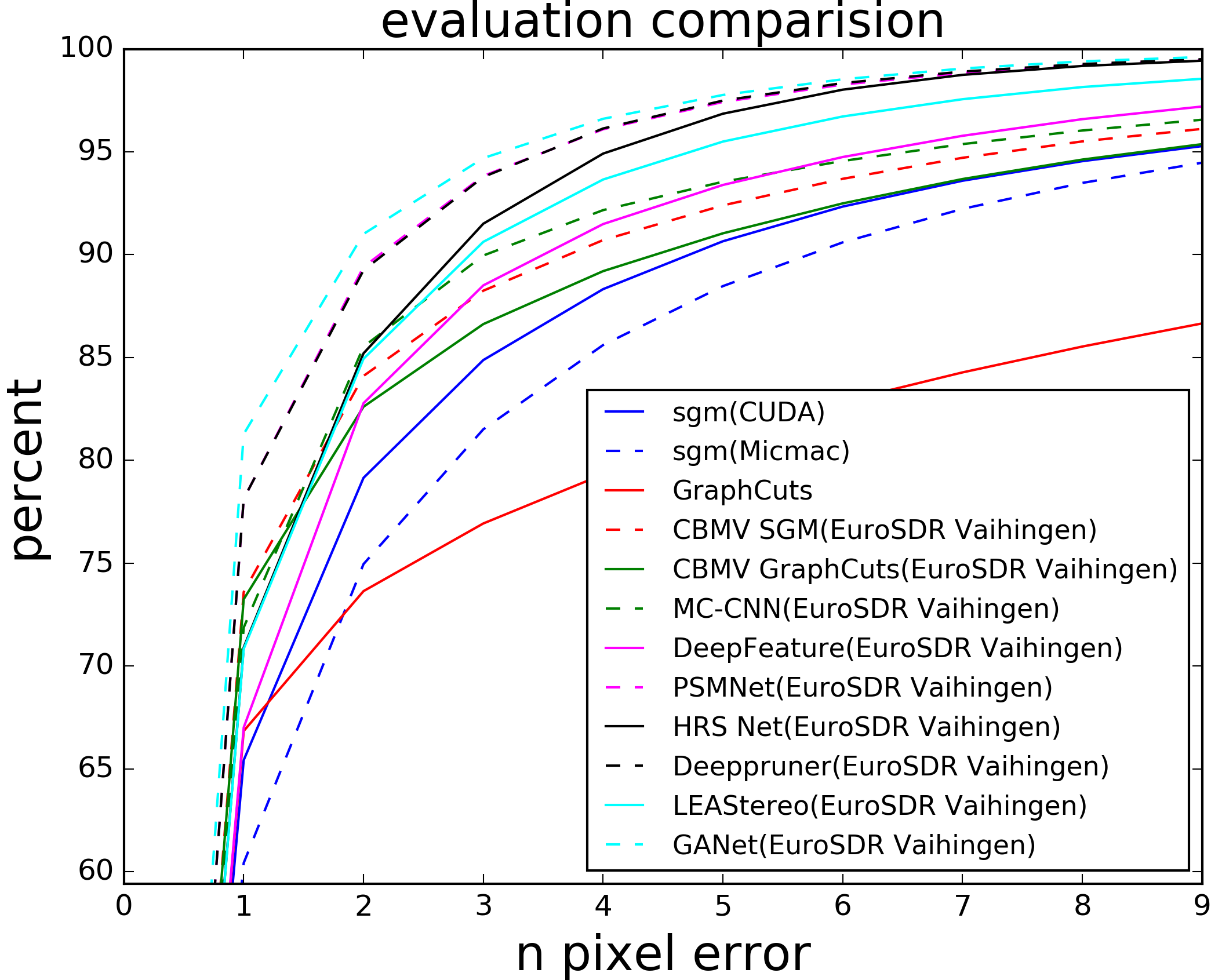}
	}
		
	\subfigure[Result on Toulouse UMBRA]{
		\label{Figure.evalute:c}
		\centering
		\includegraphics[width=0.45\linewidth]{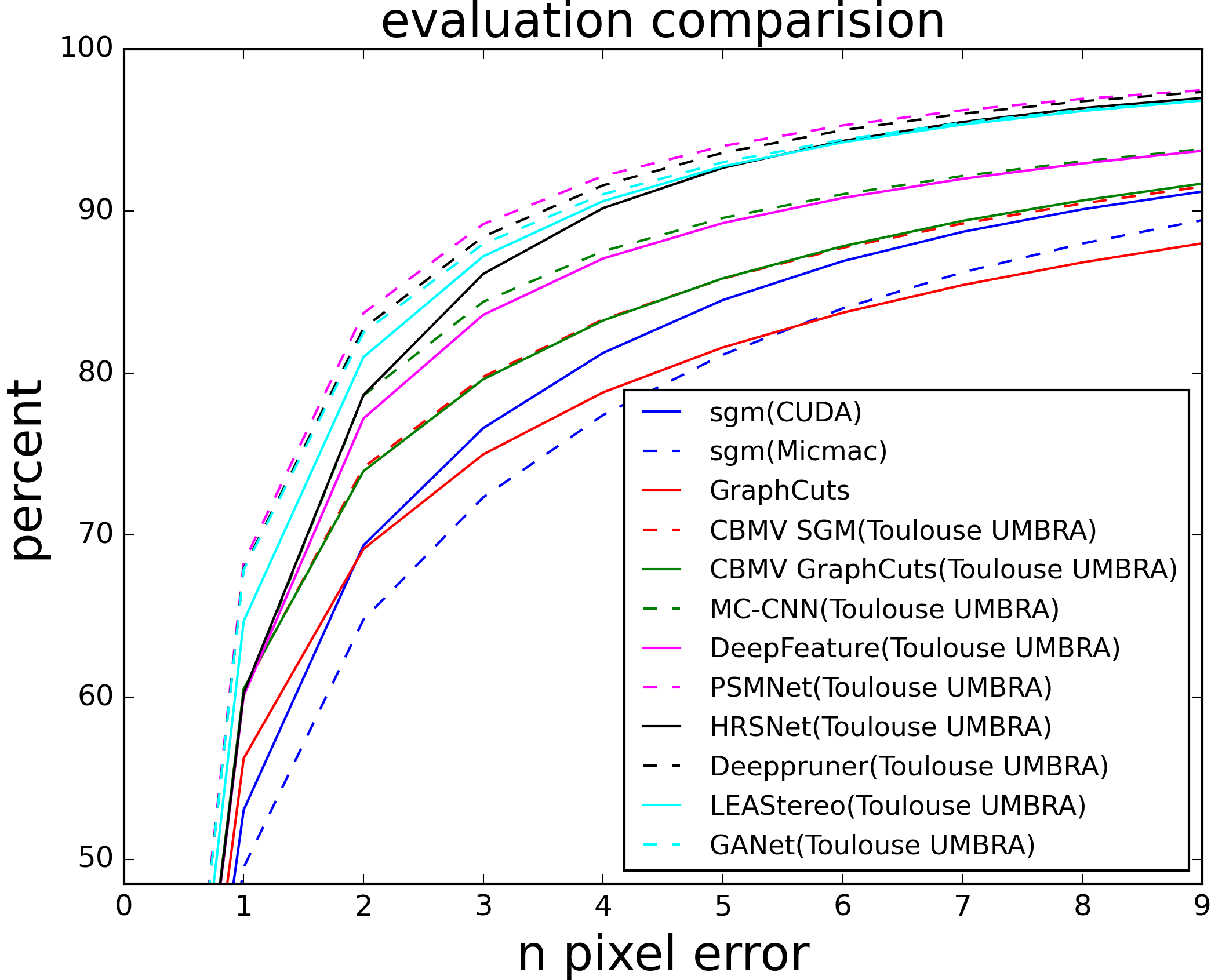}
	}
	\subfigure[Result on Toulouse Metropolet]{
		\label{Figure.evalute:d}
		\centering
		\includegraphics[width=0.45\linewidth]{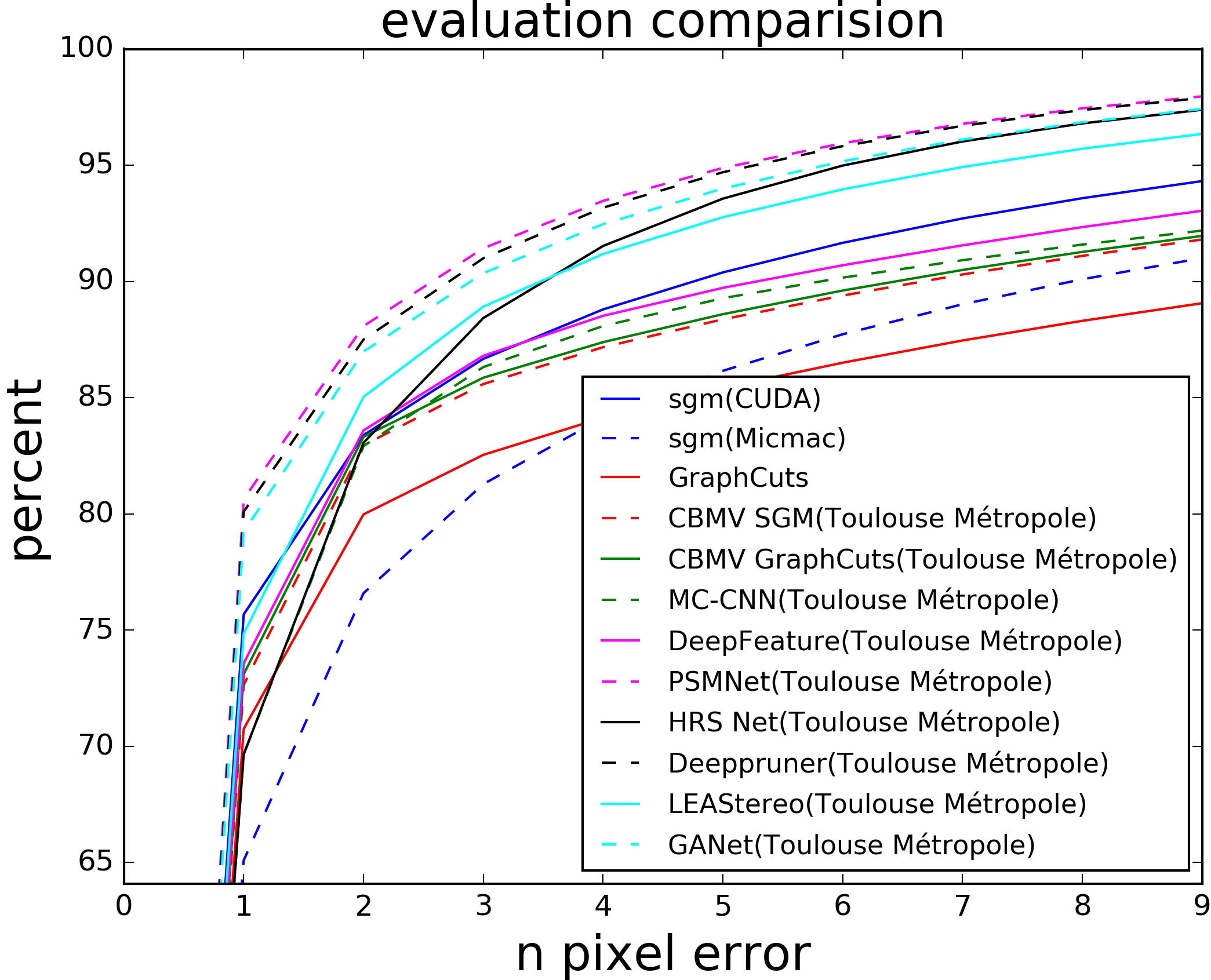}
	}
	
	\subfigure[Result on Enschede]{
		\label{Figure.evalute:e}
		\centering
		\includegraphics[width=0.45\linewidth]{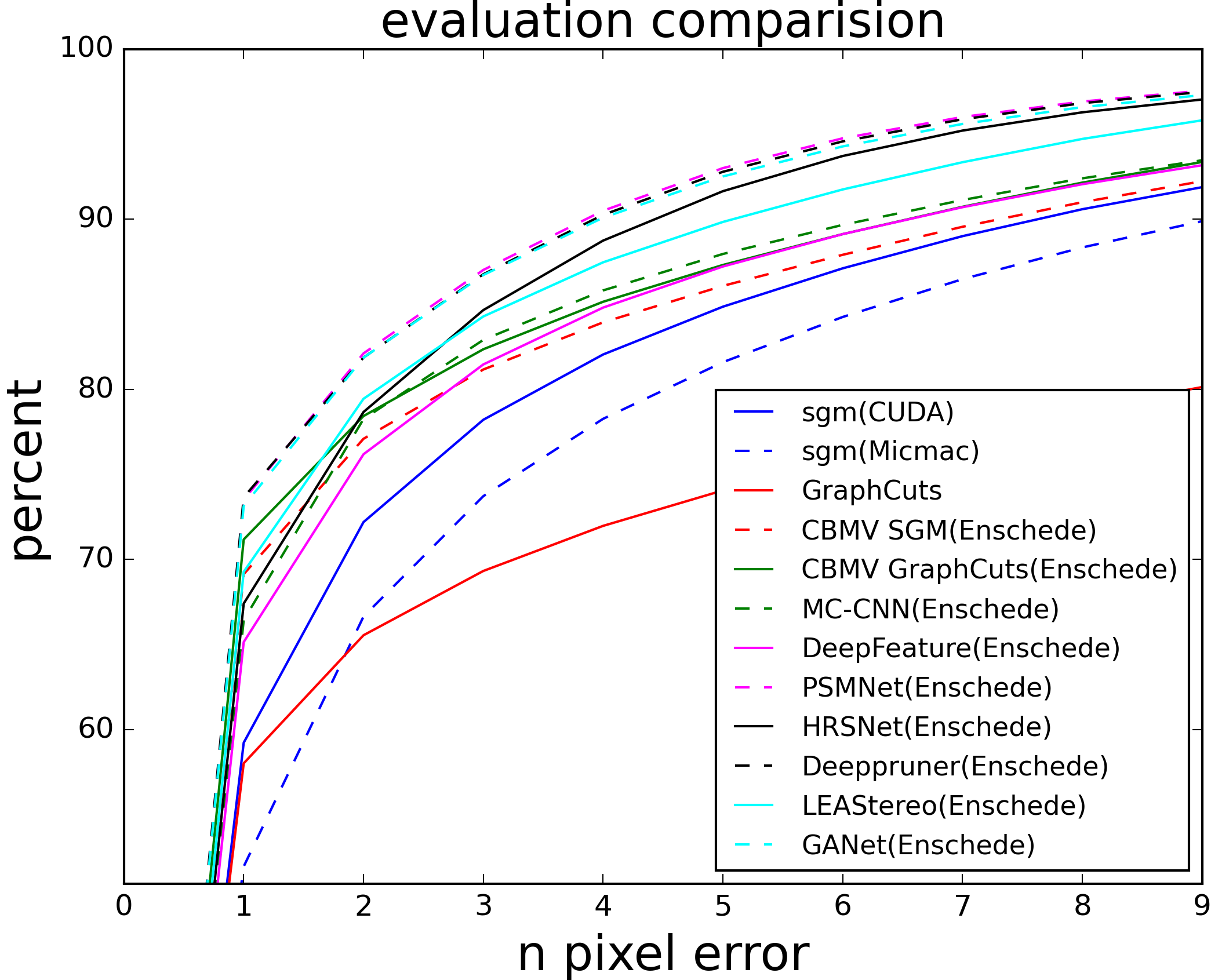}
	}
	\subfigure[Result on DublinCity]{
		\label{Figure.evalute:f}
		\centering
		\includegraphics[width=0.45\linewidth]{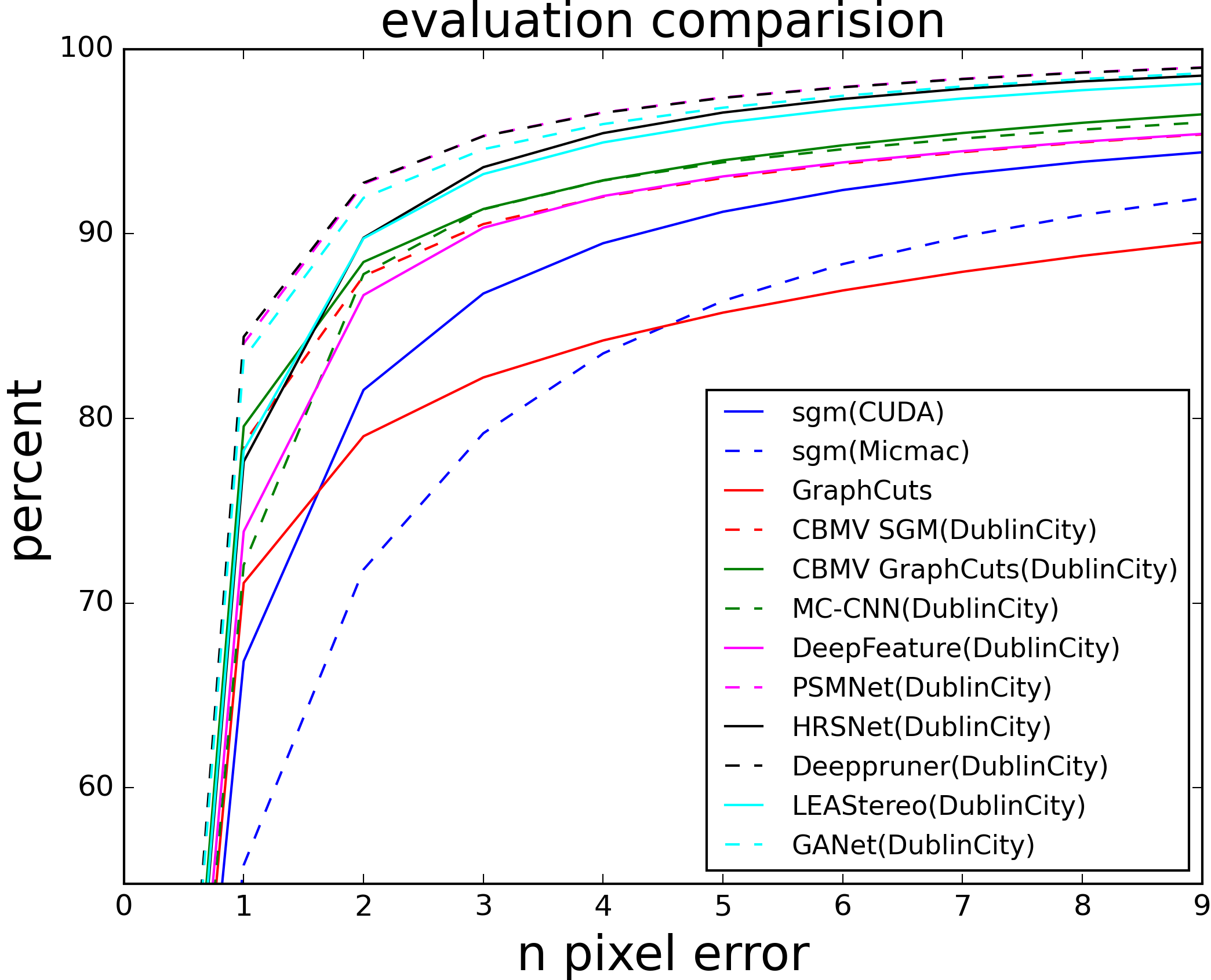}
	}
	\caption{All the method test on our dataset.}
	\label{Figure.evalute}
\end{figure}

To make the comparison fair, all learning methods are trained and tested on the same dataset.
The CBMV model is completely learned from the dataset as it only requires a small amount of training data.
DL-based methods are finetuned from the KITTI trained model provided officially.
The results for the six aerial dataset are detailed in \Cref{Figure.evalute} displaying the $N$-pixel error. For traditional method, due to the plane constraint, Graphcuts is better at 1-pixel error (more accurate), but worse than SGM method at large pixel error (less robust). For CBMV, the results are similar between SGM and Graphcuts optimization.
DL-based methods finetuned on the aerial dataset improve significantly the quality of the dense matching compared to the baseline SGM method.
Among DL-based methods, end to end methods perform better than hybrid methods. 
Generally, in the plane area, CBMV(Graphcuts) mehtod is better than traditional methods, which means that replacing handcrafted features by learned features can improve traditional pipelines.
For DL-based methods, fine-tuning on (different blocks from) the same data area significantly improves the performance. 
The rank of DL-based methods can vary between areas. We believe that this is due to variations in the proportion of large base to height ratio ($B/H$) stereo pairs as they are harder to match than small $B/H$ pairs, and some methods might be more or less sensitive to this $B/H$ induced difficulty. This is why we investigate this sensitivity to the $B/H$ in the next section.

\subsection{Influence of the base to height ratio}
\begin{figure}[tp]
	\subfigure[ISPRS Vaihingen]{
		\label{Figure.bhratio:a}
		\centering
		\includegraphics[width=0.3\textwidth]{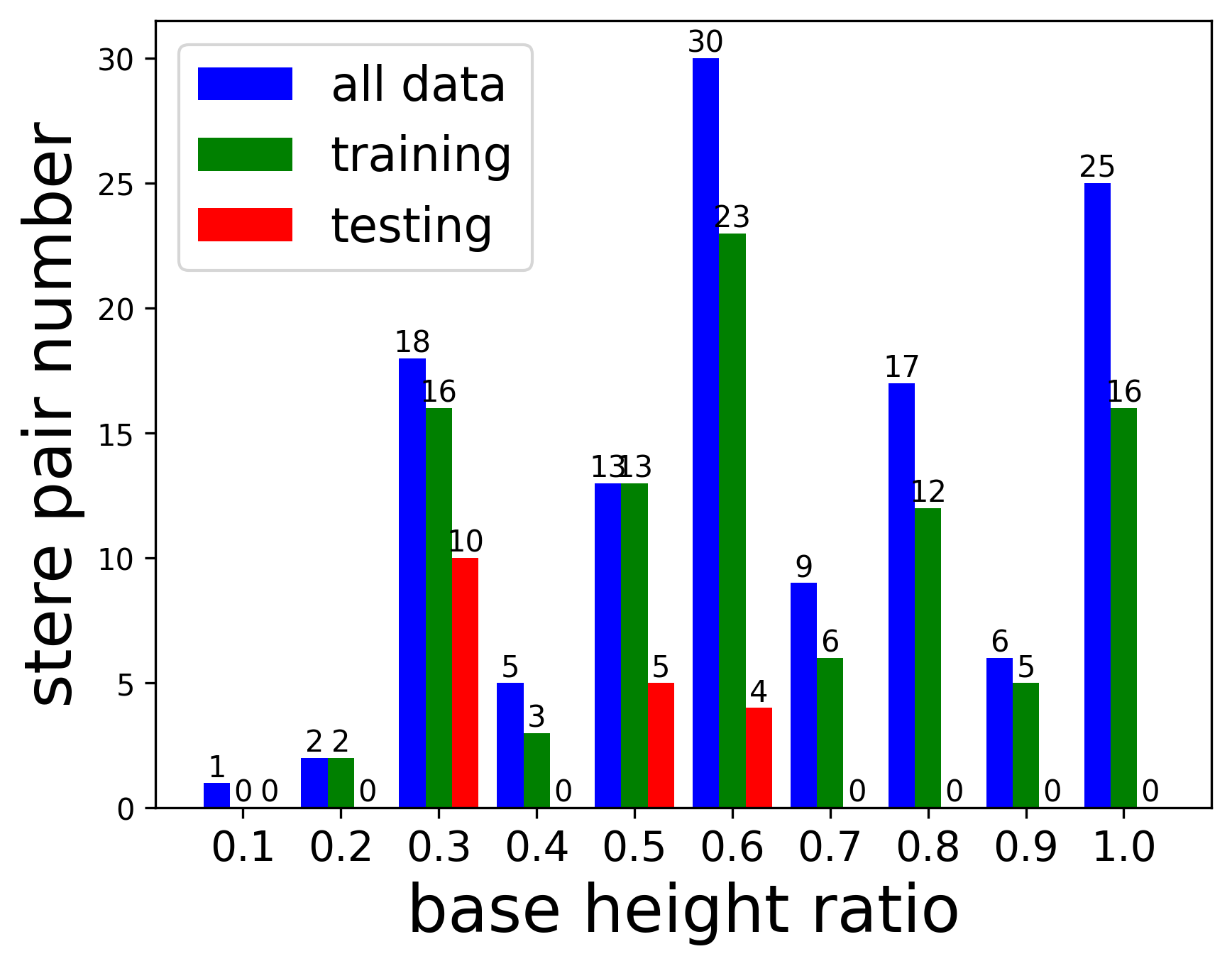}
	}\hfill%
	\subfigure[EuroSDR Vaihingen]{
		\label{Figure.bhratio:b}
		\centering
		\includegraphics[width=0.3\textwidth]{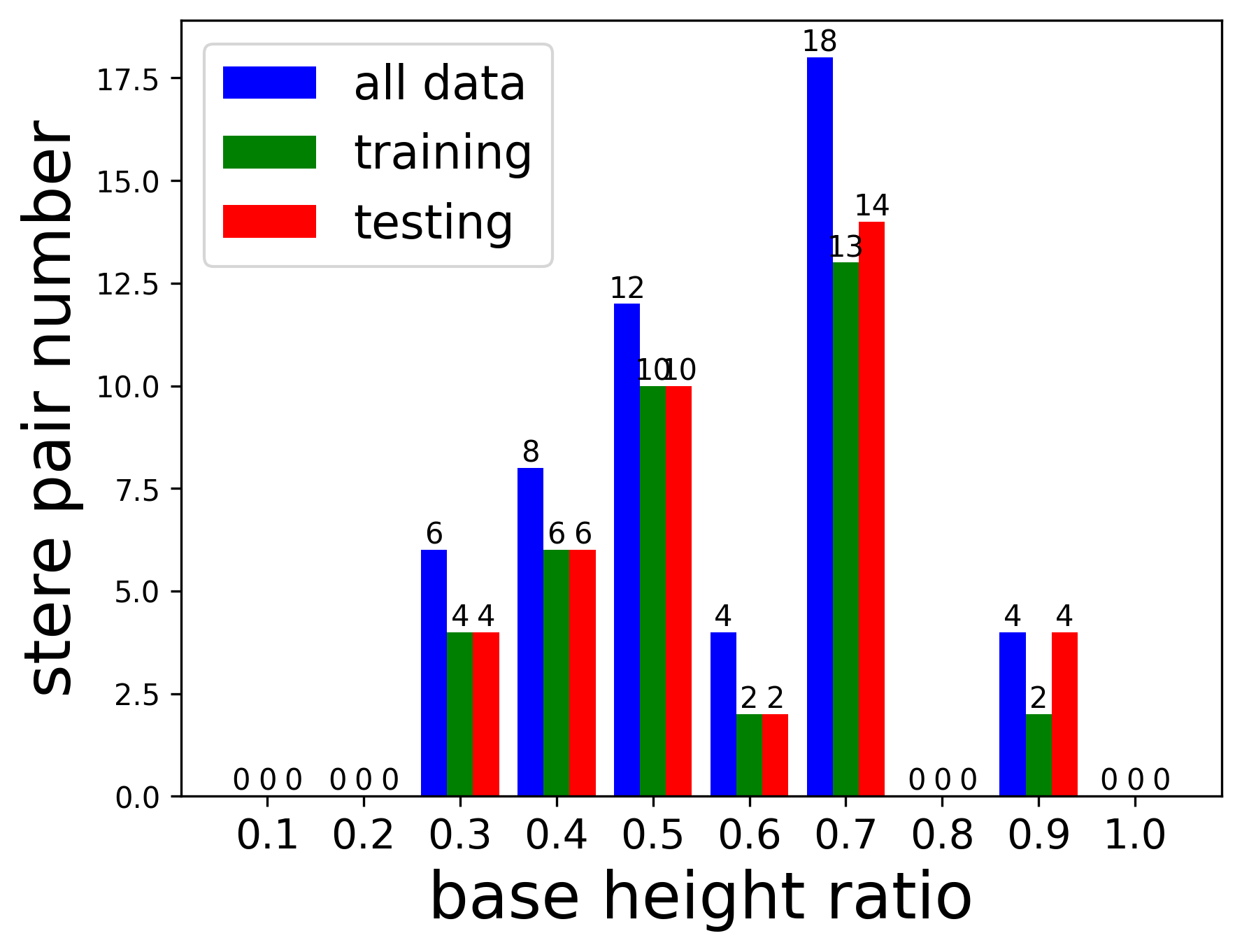}
	}\hfill%
	\subfigure[Toulouse UMBRA]{
		\label{Figure.bhratio:c}
		\centering
		\includegraphics[width=0.3\textwidth]{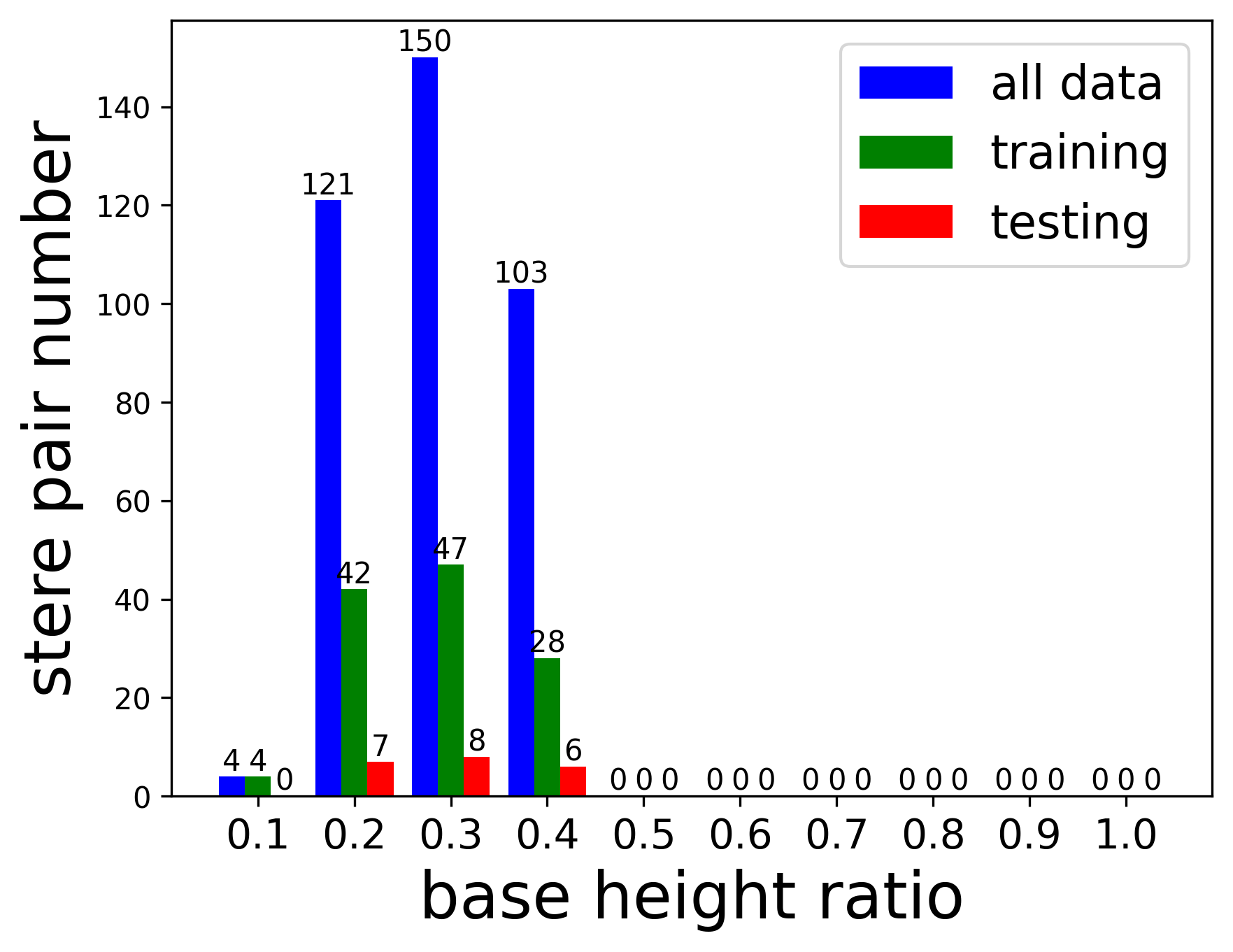}
	}
	
	\subfigure[Toulouse Metropole]{
		\label{Figure.bhratio:d}
		\centering
		\includegraphics[width=0.3\textwidth]{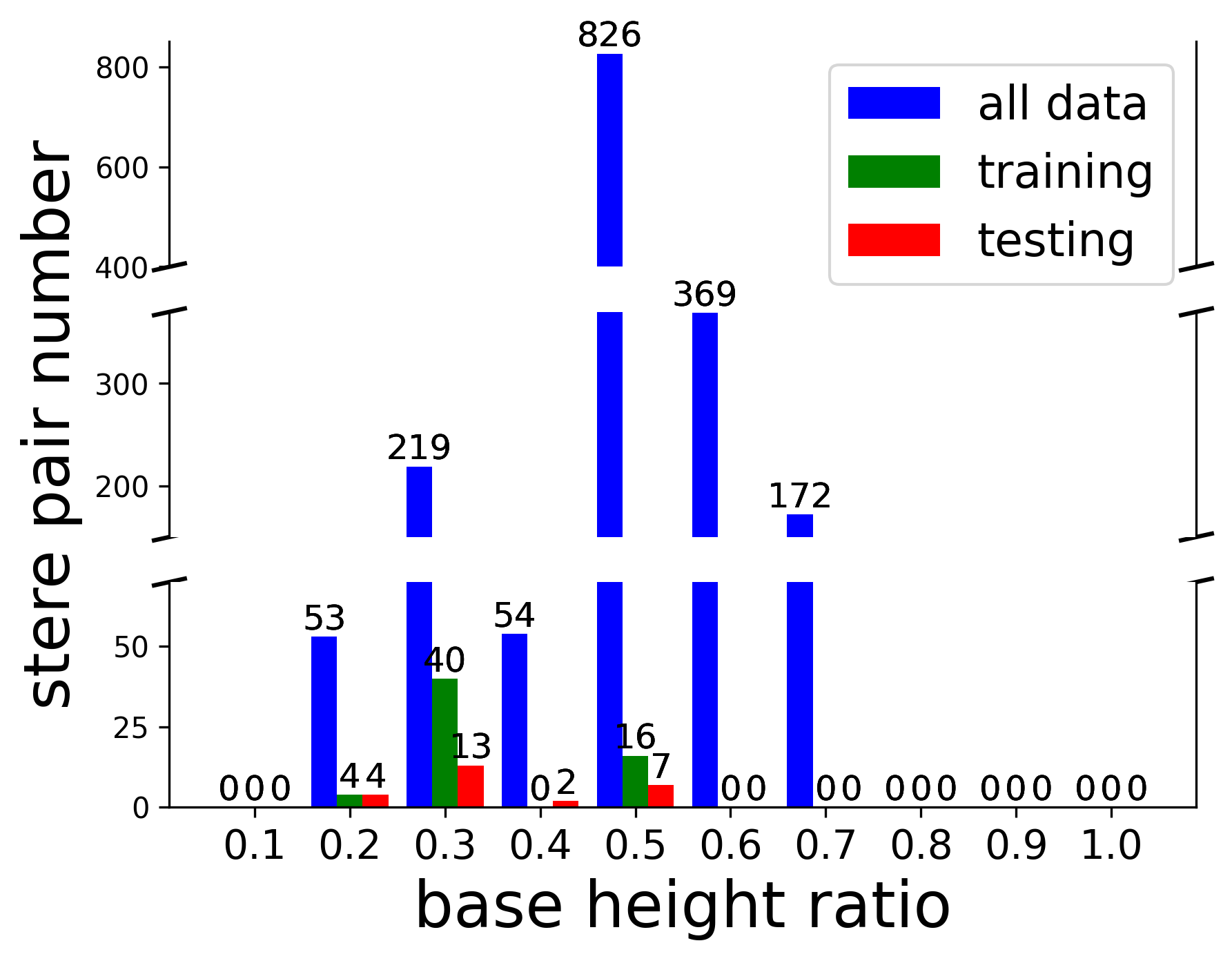}
	}\hfill%
	\subfigure[Enschede]{
		\label{Figure.bhratio:e}
		\centering
		\includegraphics[width=0.3\textwidth]{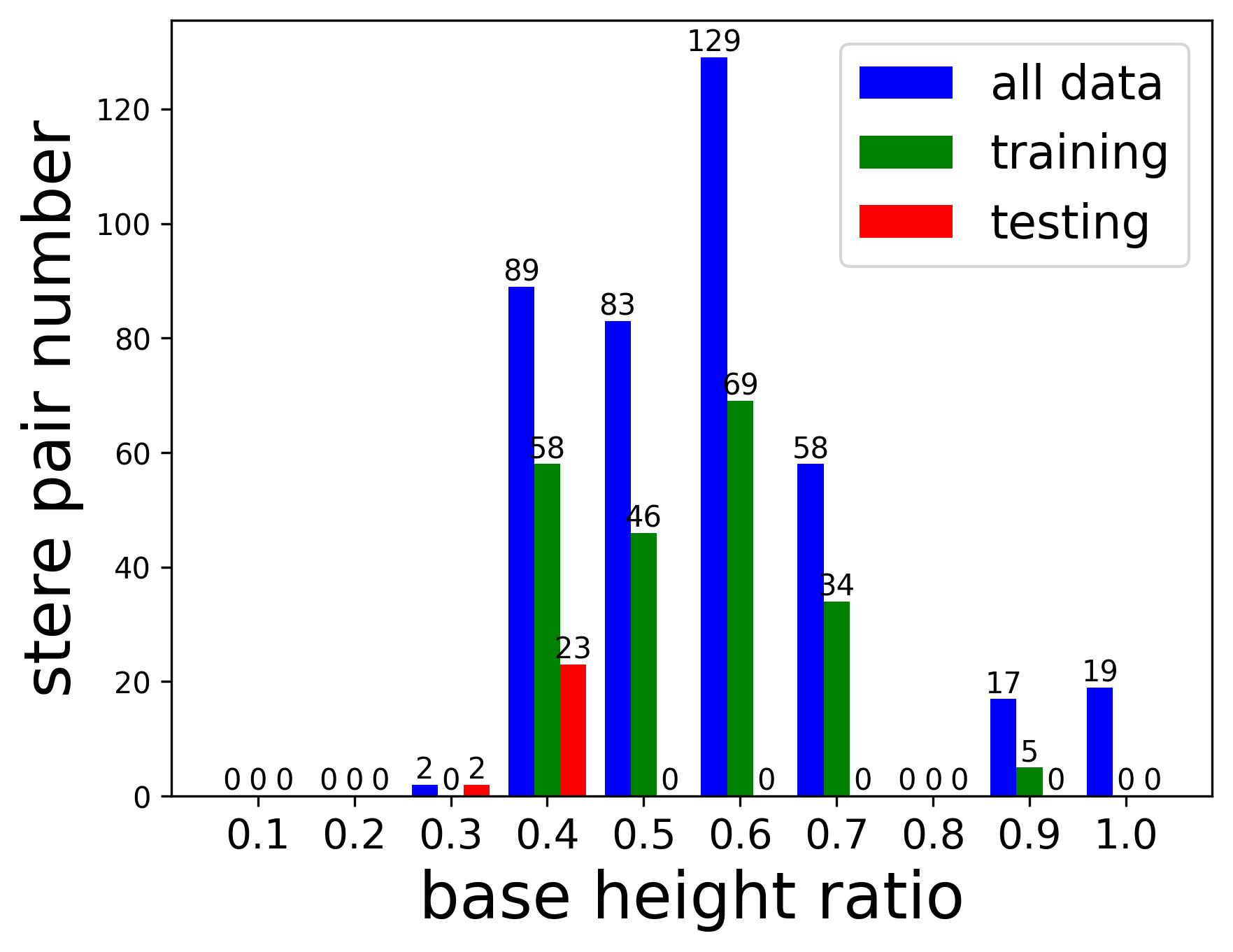}
	}\hfill%
	\subfigure[DublinCity]{
		\label{Figure.bhratio:f}
		\centering
		\includegraphics[width=0.3\textwidth]{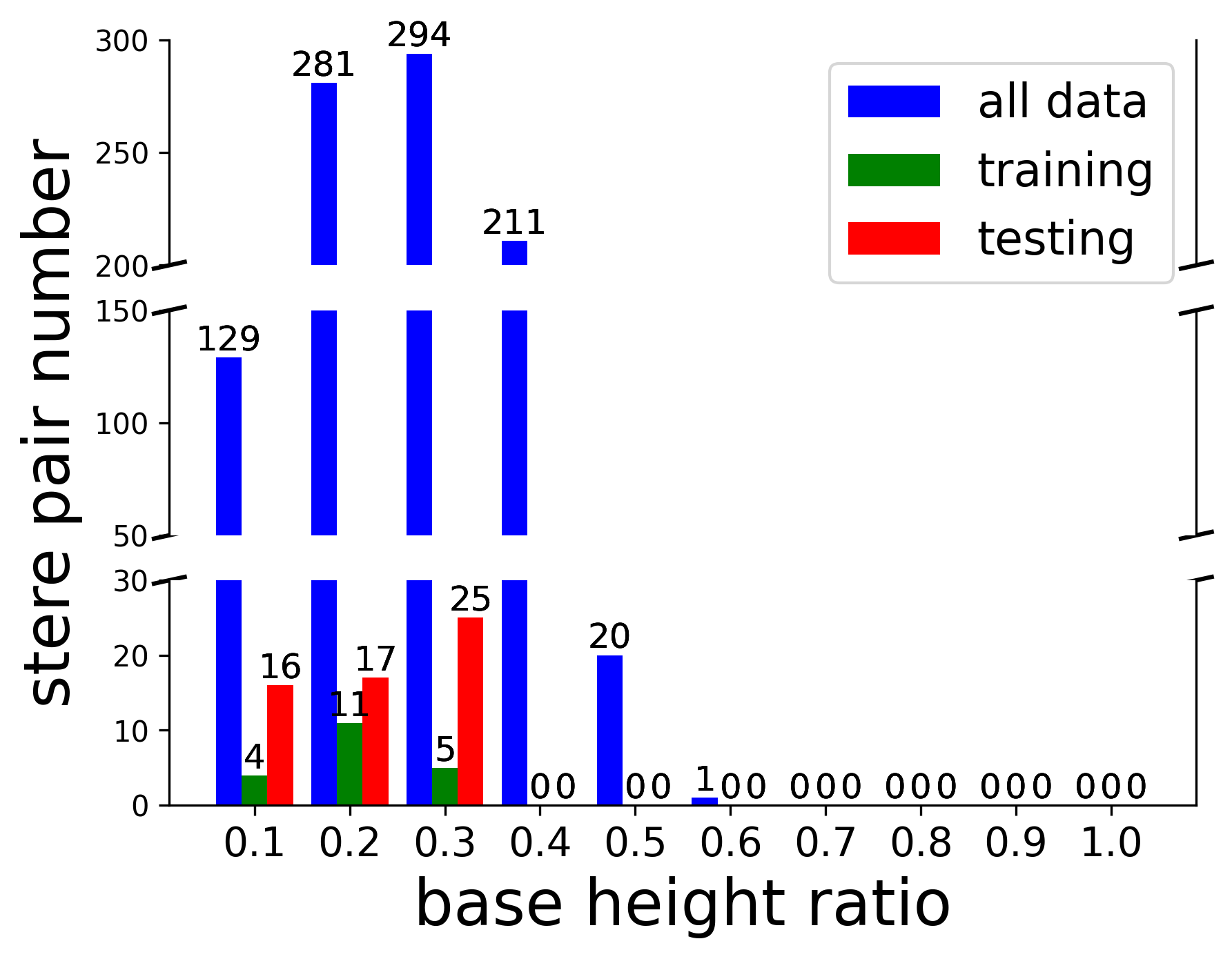}
	}
        \centering
	\caption{The base to height ratio distribution of the dataset.}
	\label{Figure.bhratio}
\end{figure}

Comparing to KITTI dataset of which the $B/H$ is fixed, in our experiments, we found that the $B/H$ of the epipolar stereo pairs has an important influence on the evaluation result. Images along-strip are usually used for dense matching because they have a large overlap and small $B/H$. Images across-strip have a larger $B/H$ which can increase the intersection accuracy, but leads to more errors because of a larger perspective distortion and more occlusions \citep{tola2008fast}.
For different stereo pair and dataset, the $B/H$ is different. The $B/H$ influences the disparity range and distribution. As shown in \Cref{Figure.bhratio}, $B/H$ is calculated from the orientation information. EuroSDR Vaihingen and Enschede have a large $B/H$ compared with the others, as shown in \Cref{Figure.bhratio:b} and \Cref{Figure.bhratio:e}.

To evaluate the influence of the $B/H$, we decided to use the \textit{Toulouse Metropole} dataset which has a high quality and is large enough to split by $B/H$.
The original dataset consisting in 20 aerial images decomposed in 1349 large epipolar pairs is split into 26775 training cropped pairs and 39711 testing cropped pairs.
The training set is further split into small ($B/H<0.4$, 129 pairs), middle ($B/H=$ 0.4 \~{} 0.6, 511 pairs) and large ($B/H>0.6$, 65 pairs) subsets.
We evaluated different settings for the training as detailed in \Cref{Table:trainingset}.
For the \textit{selected} setting, we have trained a specialized network for each category (small, middle and large $B/H$) and select the specialized model corresponding to the actual $B/H$ of the pair to process.
Because there are 200 pair are used as a validation set, so there are 600 pair in the \textit{full} set.

\begin{table}[t]
    \centering
	\caption{Training set for the $B/H$ experiment.}
	\label{Table:trainingset}
	\begin{tabular}{c|ccc|c}
		\noalign{\hrule height 1.2pt}
		{\bfseries abbreviative name} & {\bfseries small $B/H$} & {\bfseries middle $B/H$} & {\bfseries large $B/H$} & {\bfseries total} \\
		\noalign{\hrule height 1.2pt}
	    small   &  1200 & 0 & 0 & 1200 \\
		\hline
		middle  &  0 & 1200 & 0 & 1200 \\
		\hline
		large   &  0 & 0 & 1200 & 1200 \\
		\hline
		fusion  &  600 & 0 & 600 & 1200 \\
		\hline
		ave   &  400 & 400 & 400 & 1200 \\
        \hline
		random & 249 & 857 & 94 & 1200 \\
		\noalign{\hrule height 1pt}
		all & 1200 & 1200 & 1200  & 3600 \\
		\noalign{\hrule height 1pt}
		full & 6776 & 18069 & 1330 & 26175 \\
		\noalign{\hrule height 1.2pt}
	\end{tabular}
\end{table}

\begin{figure}[tp]
	\centering
	\subfigure[Small $B/H$]{
		\label{Figure.bhexperimnt:a}
		\centering
		\includegraphics[width=0.45\linewidth]{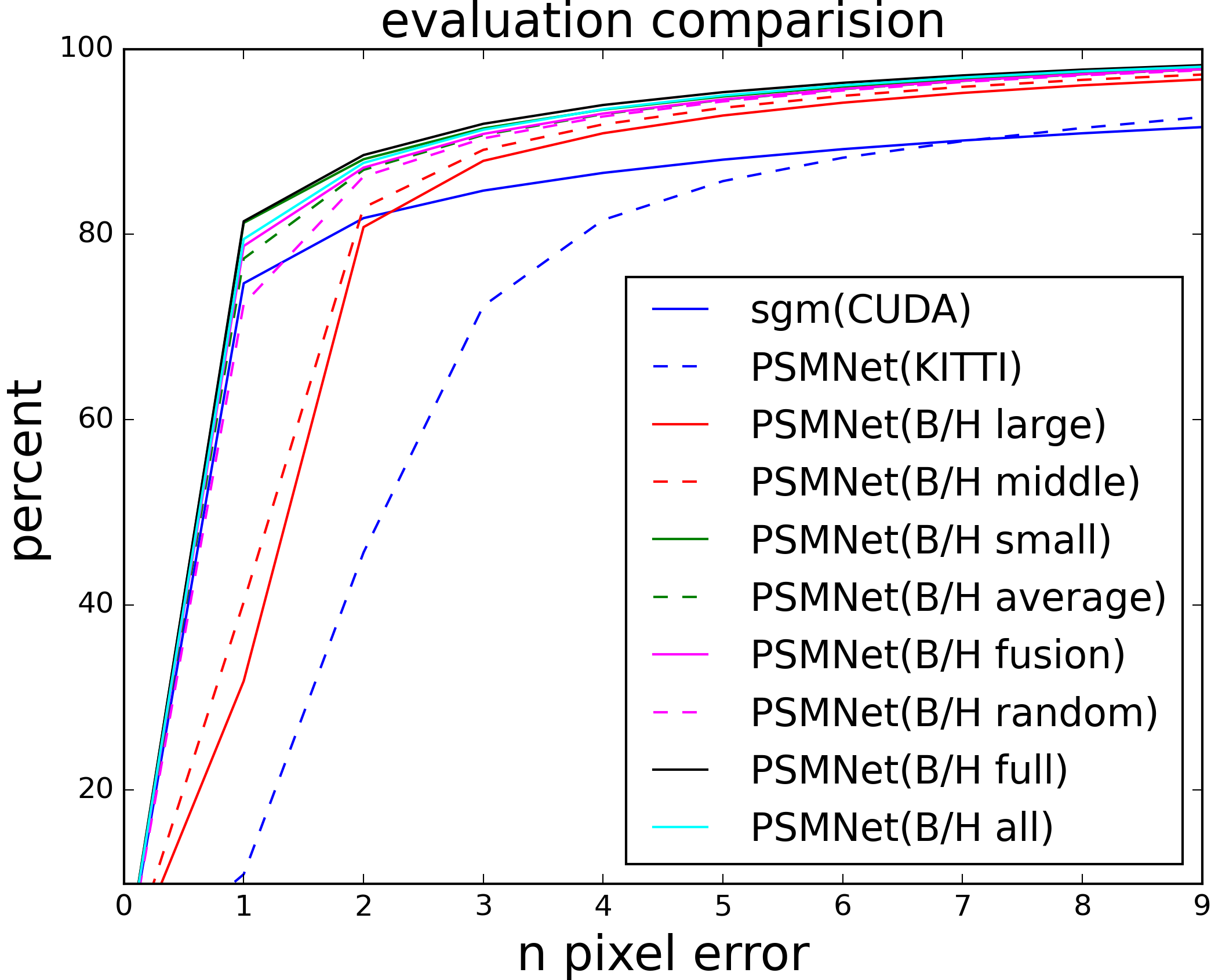}
	}
	\subfigure[Middle $B/H$]{
		\label{Figure.bhexperimnt:b}
		\centering
		\includegraphics[width=0.45\linewidth]{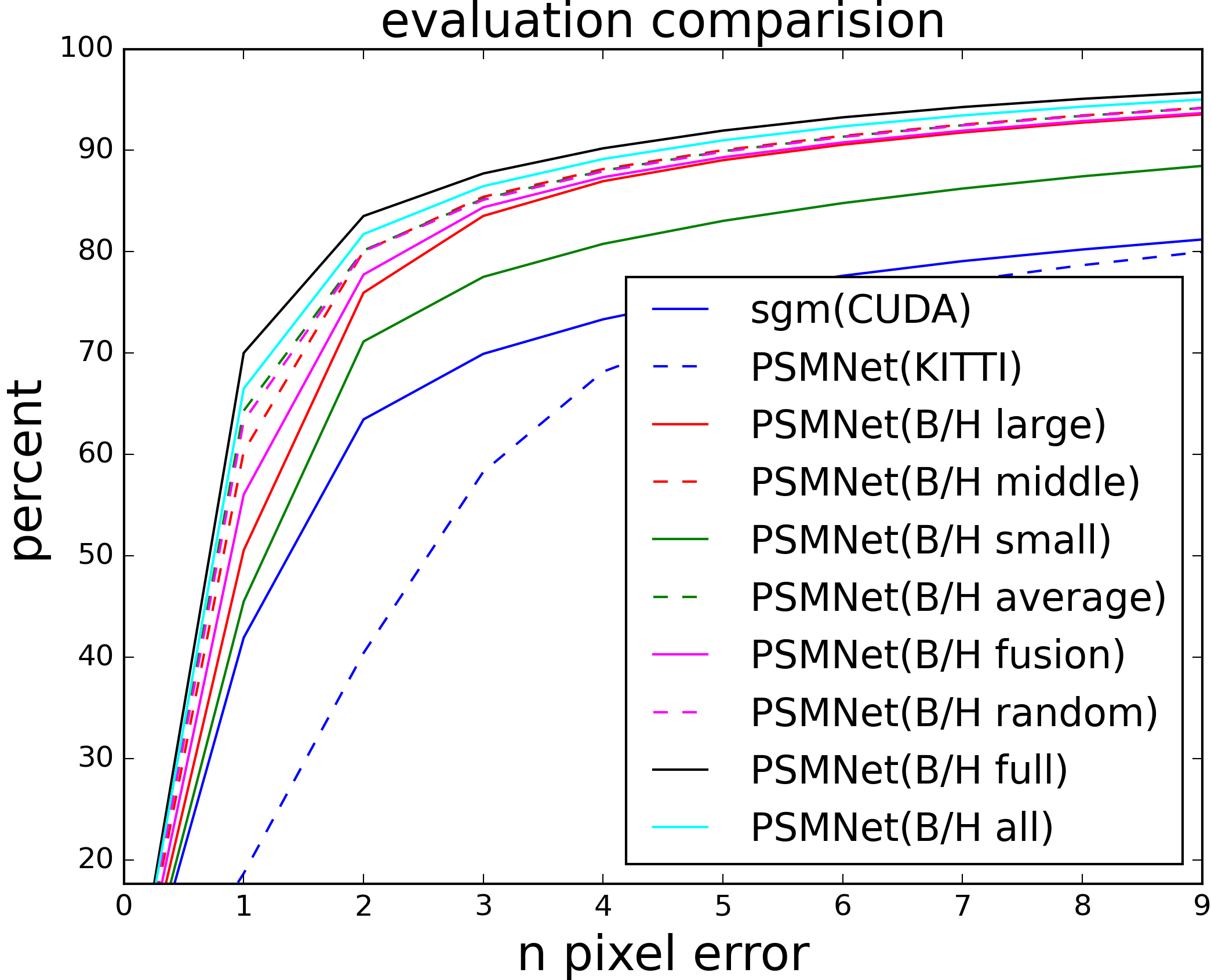}
	}
	
	\subfigure[Large $B/H$]{
		\label{Figure.bhexperimnt:c}
		\centering
		\includegraphics[width=0.45\linewidth]{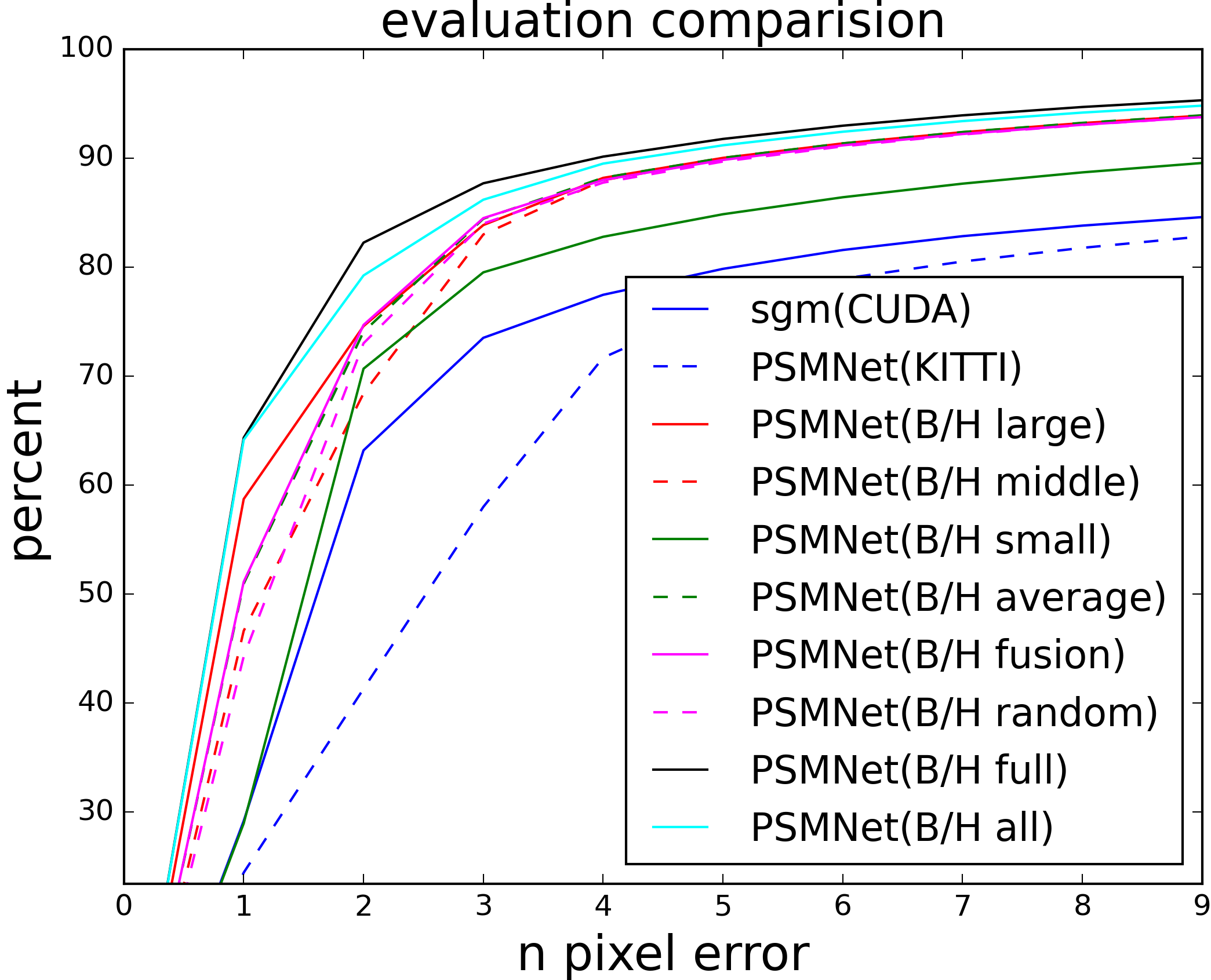}
	}
	\subfigure[All data]{
		\label{Figure.bhexperimnt:d}
		\centering
		\includegraphics[width=0.45\linewidth]{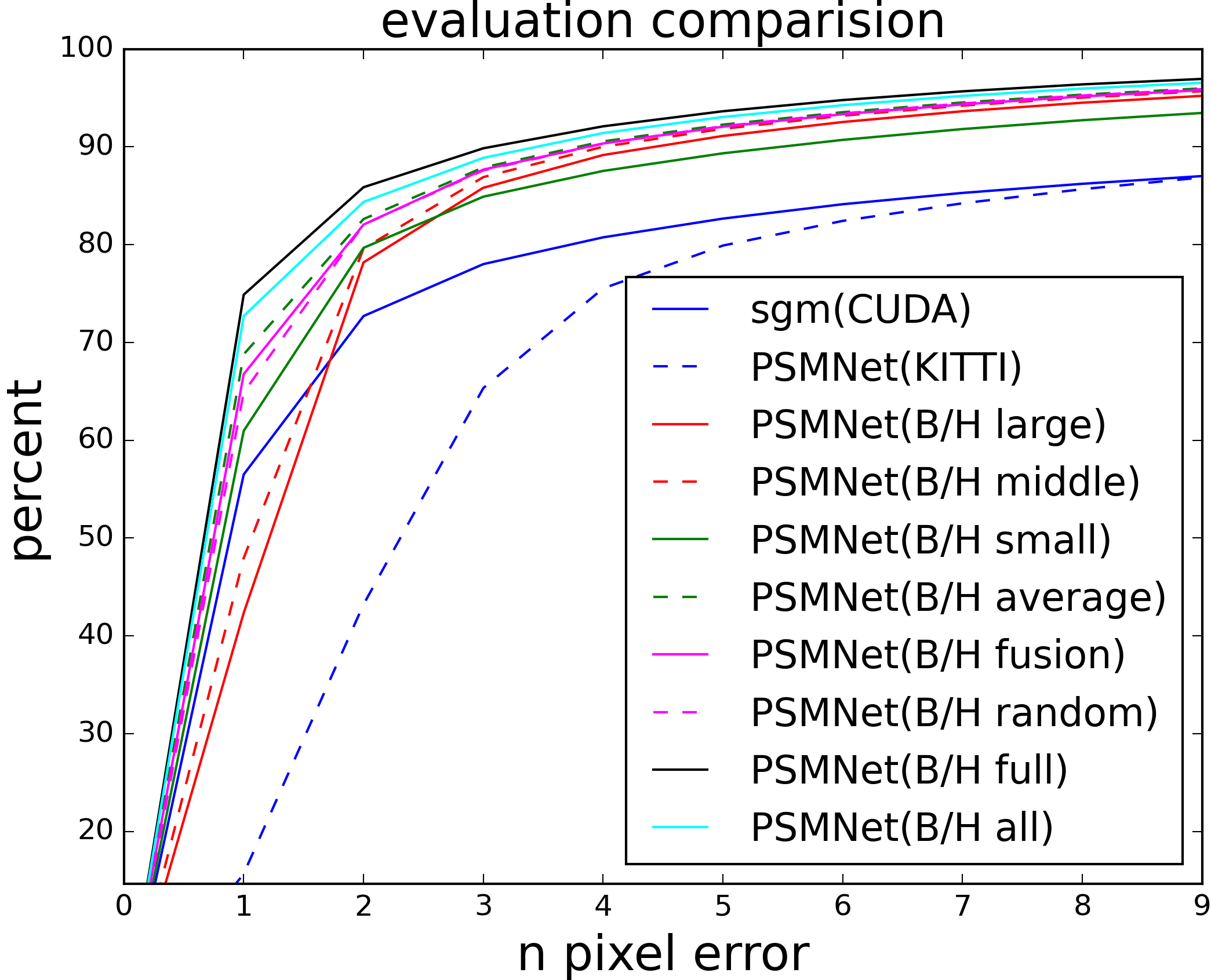}
	}
	\caption{The base to height ratio experiment on Toulouse Metropole.}
	\label{Figure.bhexperimnt}
\end{figure}

We see in \Cref{Figure.bhexperimnt} that the proportion of each $B/H$ category in the evaluation dataset influences the evaluation of the method and that using more training pairs improves the performance.
For the small $B/H$ dataset, training with small $B/H$ only gives an equivalent result as training with the whole dataset, as shown in \Cref{Figure.bhexperimnt:a}, but training with large $B/H$ only gives even worse results than SGM(CUDA).
But for middle and large $B/H$, using more training data always improve performance, as shown in \Cref{Figure.bhexperimnt:b} and \Cref{Figure.bhexperimnt:c}.
In \Cref{Figure.bhexperimnt:c}, we see that training with large $B/H$ improves the performance.
However, because using more data to train means more computation time, we can look for the best strategy for an equivalent number of training pairs.
The experiments show that the optimal training strategy is then to balance the 3 $B/H$ classes.

\subsection{Dataset shift}
\begin{figure}[tp]
	\centering
	\subfigure[ISPRS Vaihingen]{
		\label{Figure.image_hist:a}
		\centering
		\includegraphics[width=0.25\linewidth]{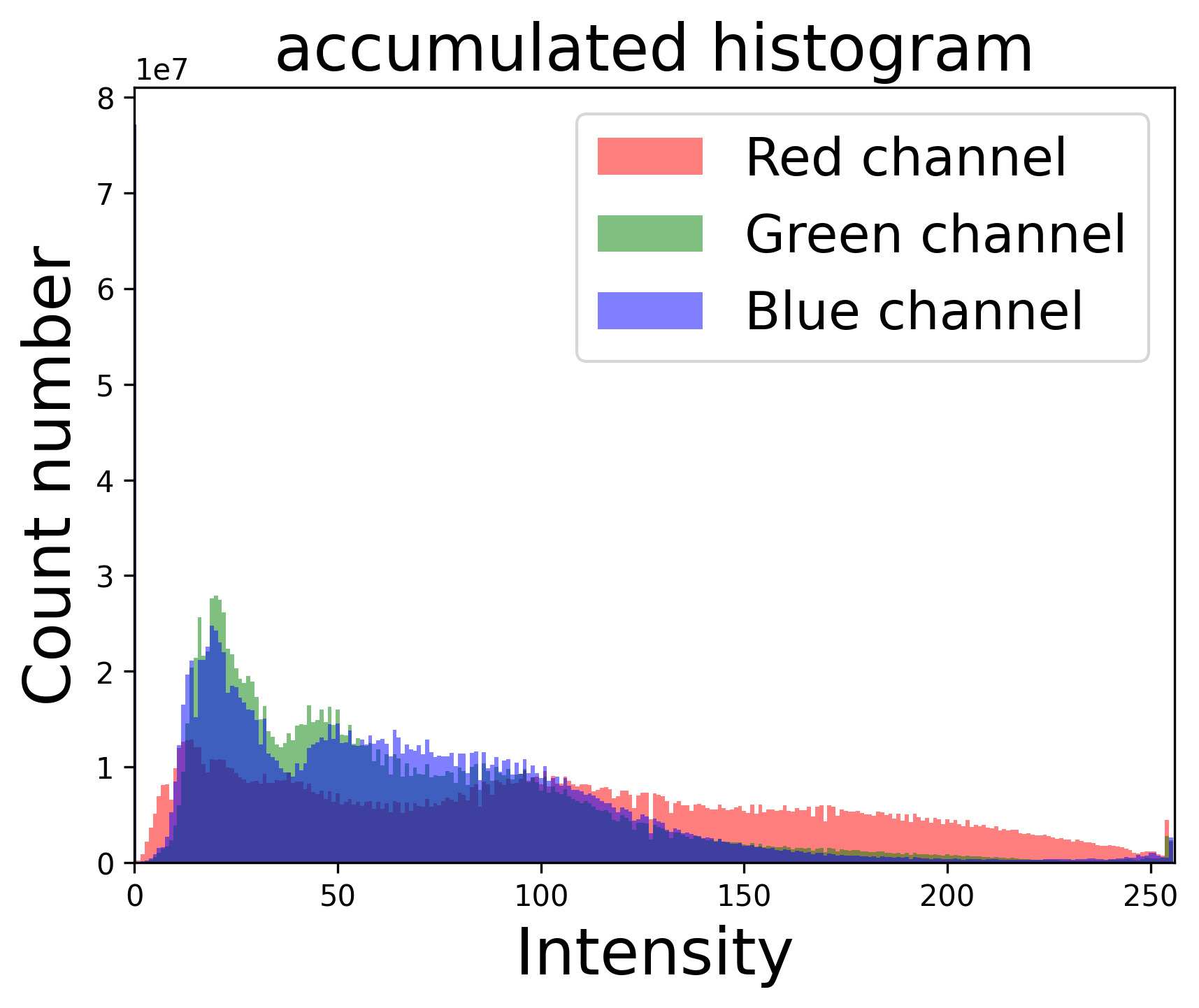}
	}
	\subfigure[EuroSDR Vaihingen]{
		\label{Figure.image_hist:b}
		\centering
		\includegraphics[width=0.25\linewidth]{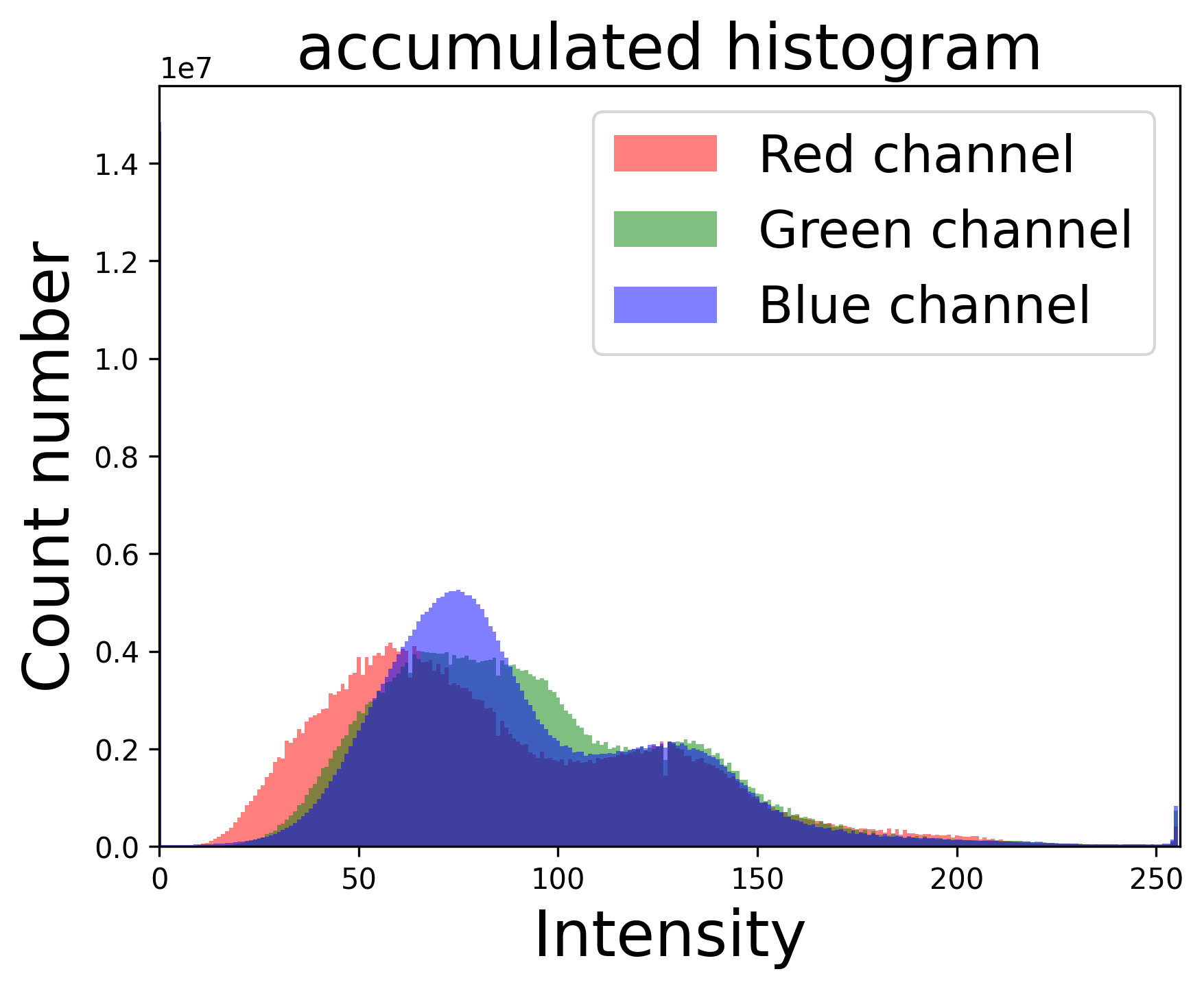}
	}
	\subfigure[Toulouse UMBRA]{
		\label{Figure.image_hist:c}
		\centering
		\includegraphics[width=0.25\linewidth]{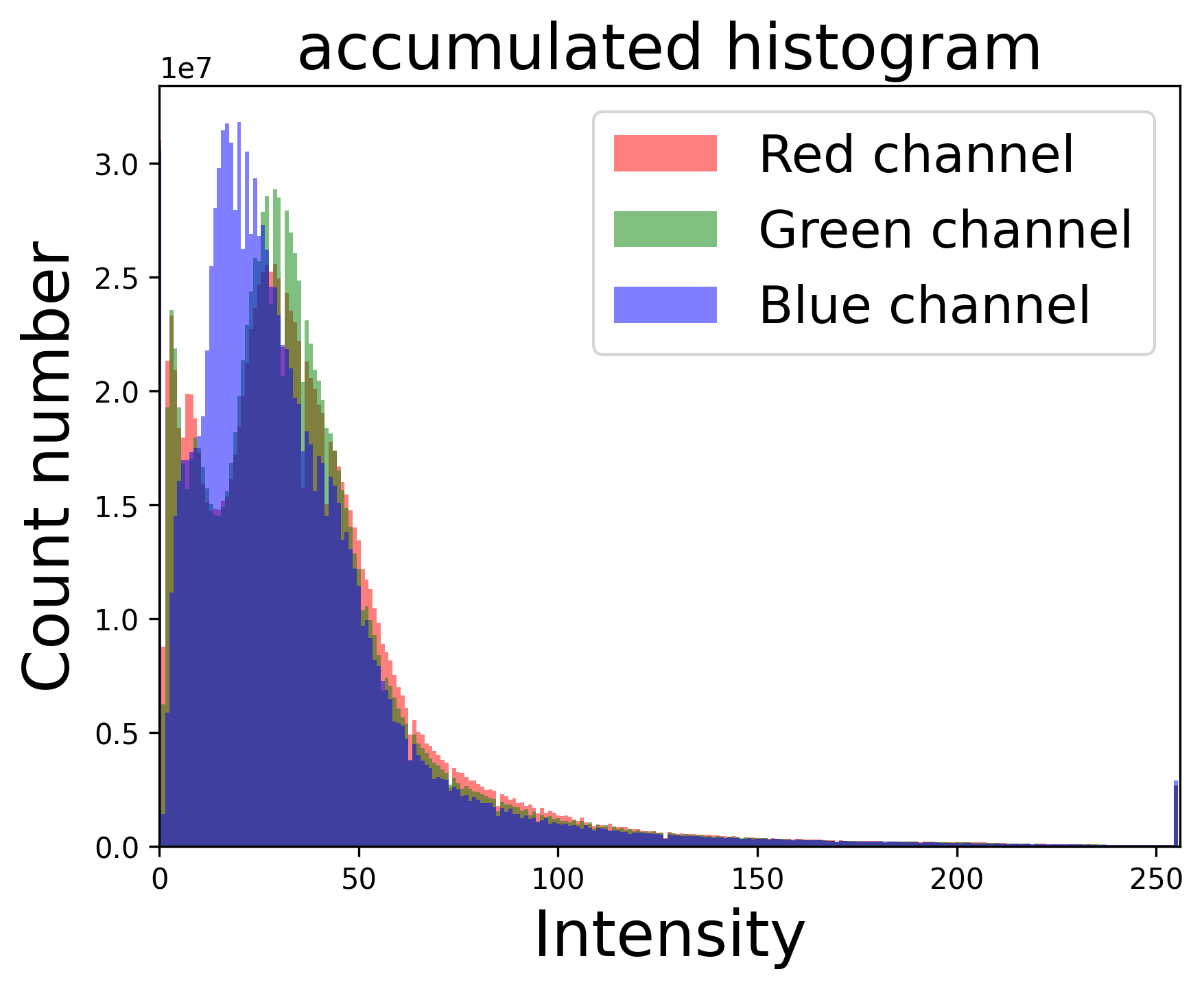}
	}
	
	\subfigure[Toulouse Metropole]{
		\label{Figure.image_hist:d}
		\centering
		\includegraphics[width=0.25\linewidth]{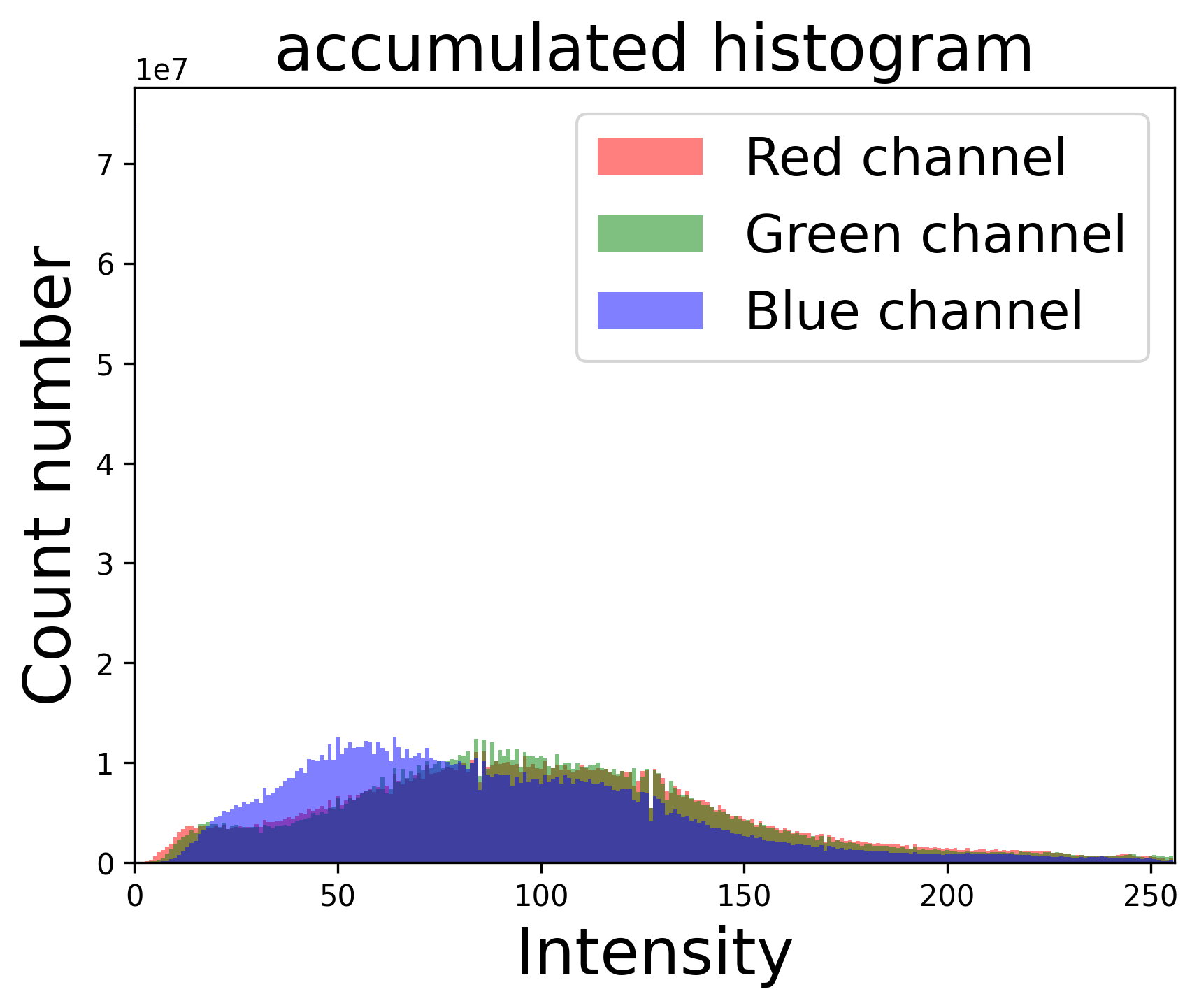}
	}
	\subfigure[Enschede]{
		\label{Figure.image_hist:e}
		\centering
		\includegraphics[width=0.25\linewidth]{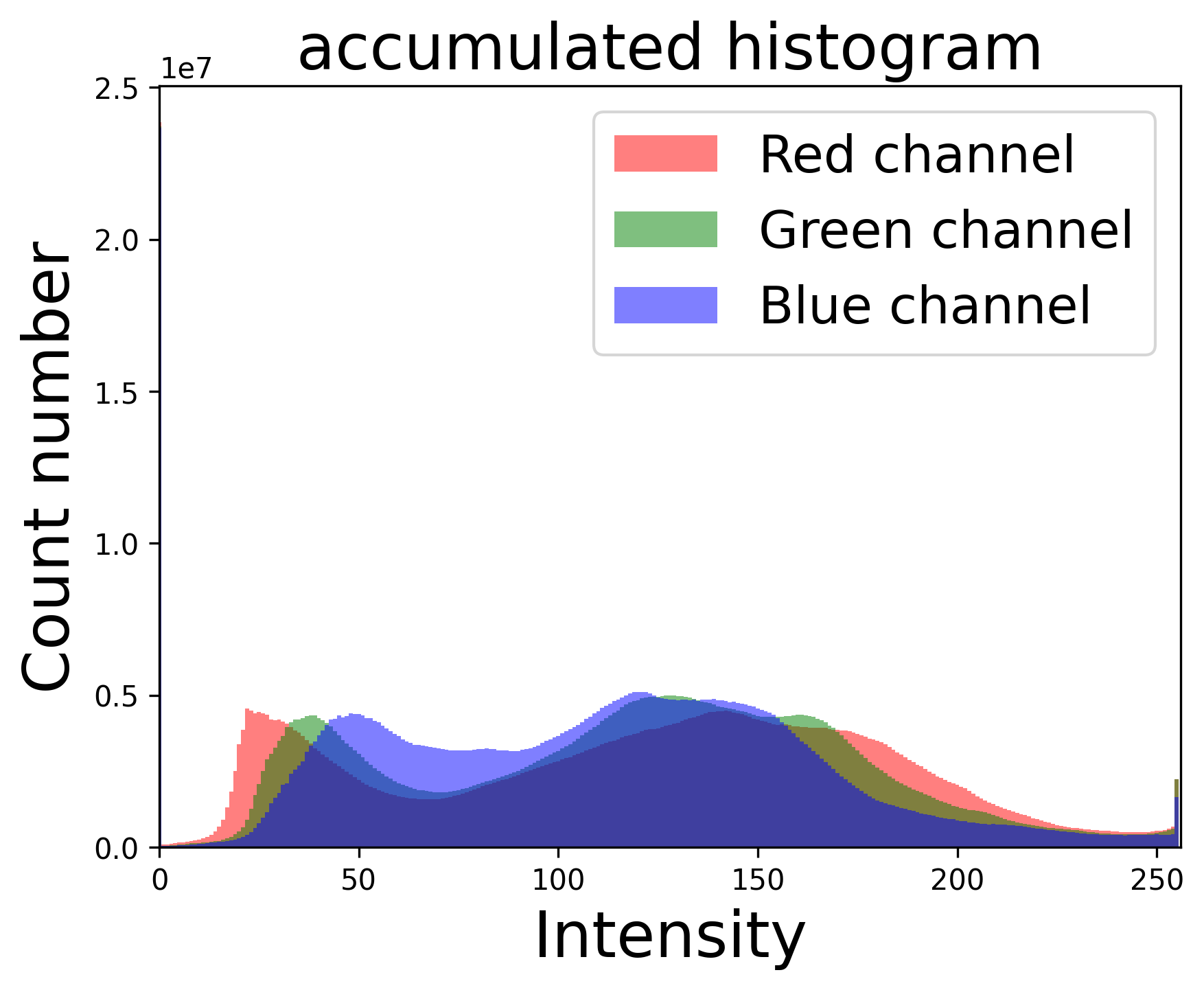}
	}
	\subfigure[DublinCity]{
		\label{Figure.image_hist:f}
		\centering
		\includegraphics[width=0.25\linewidth]{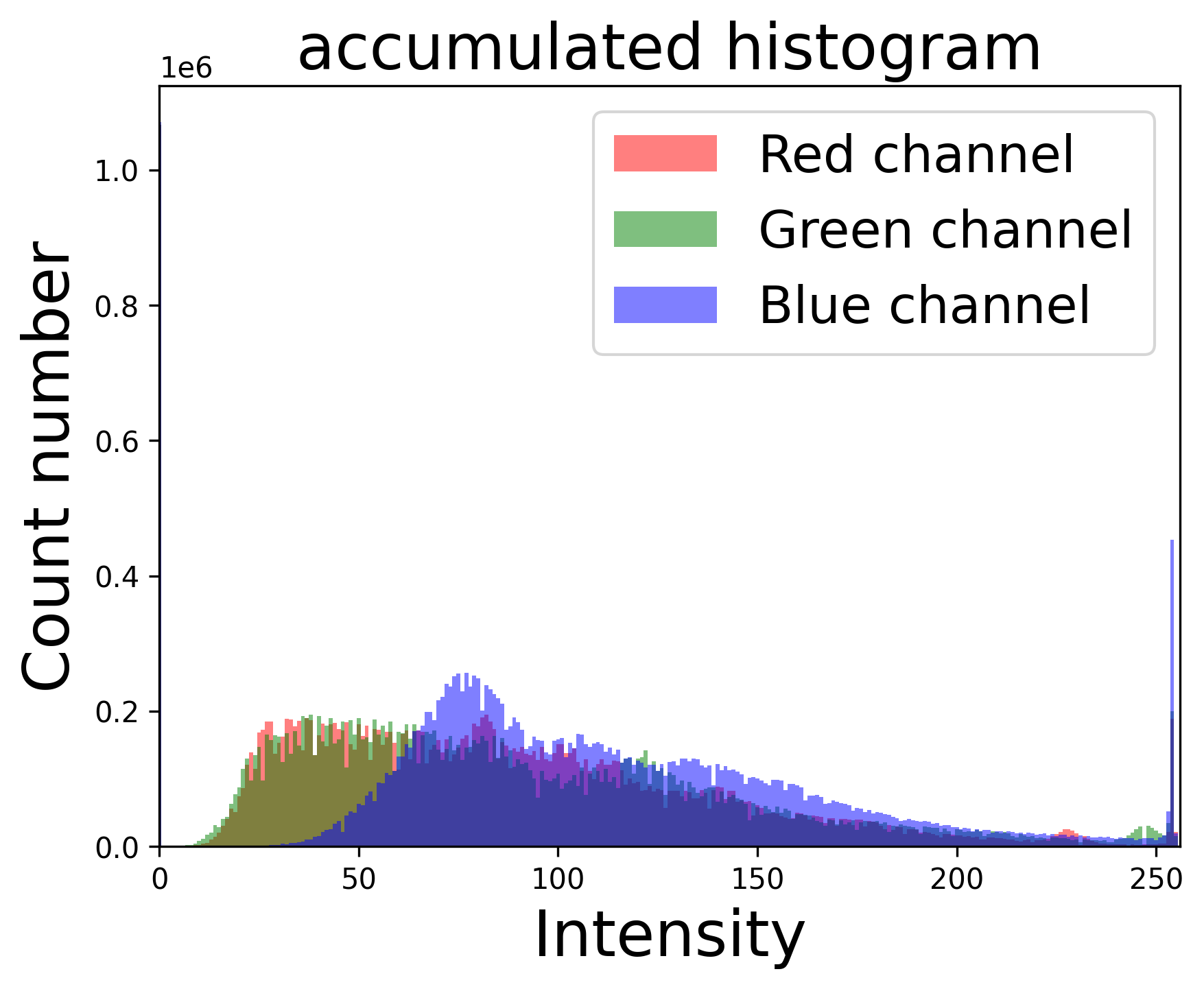}
	}
	\caption{The image histogram distribution of the dataset.}
	\label{Figure.image_hist}
\end{figure}

In photogrammetric applications, it is hard and costly to obtain training data for every scene, so we need to train on a limited diversity of areas and apply the learned model to other areas, which is commonly referred to as transferbility or\textbf{ dataset shift} \citep{quinonero2009dataset}.
We propose some experiments to study this dataset shift, considering that our 6 aerial datasets span 4 different geographic areas.
From our experience, several factors influence the performance of DL-based dense matching methods : (1) different scene type; (2) different image color and brightness, as shown in \cref{Figure.image_hist}, it varies from dataset; (3) different resolution of image; (4) different disparity range; (5) different distribution in cost volume. 
These factors influence the method transfer from one dataset to another.
Considering these factors, the transfer learning strategies and experiments are conducted in three aspects :
\begin{itemize}
	\item Dataset shift between different types of scenes. While training datasets are abundent in the computer vision community, training data is quite rare in photogrammetry. Thus analysing the shift between computer vision and photogrammetry datasets is interesting. To investigate this topic, we compare training on our proposed datasets with training on the KITTI dataset. This is made easy considering that all tested deep learing methods provide a model trained on KITTI.  
	\item Dataset shift on scenes of the same type (remote sensing images) but between different geographical areas.
	\item Dataset shift on the same geographical area but with different image resolutions. Among our datasets, two areas have data with different resolutions: ISPRS-Vaihingen has 8 $cm$ GSD while EuroSDR-Vaihingen has 20 $cm$ GSD, and Toulouse UMBRA has 12.5 $cm$ GSD while Toulouse Metropole has 5 $cm$ GSD.
\end{itemize}

To discuss the performance of remote sensing images dataset shift, training on a single area is more distinct, on the hand, the influence of more than one training areas is also an interesting point, so we split the experiment into two parts : (1) single data training, mainly discuss the above these factors in dataset shift, refer to Section \ref{sec:single_shift}; and (2) multi-data training, mainly discuss dataset shift with more than one training data, refer to Section \ref{sec:multi_shift}.

\subsubsection{Single data training in dataset shift}
\label{sec:single_shift}
\begin{figure}[tp]
	\centering
	\subfigure[Result on ISPRS Vaihingen]{
		\label{Figure.transferlearning:a}
		\centering
		\includegraphics[width=0.45\linewidth]{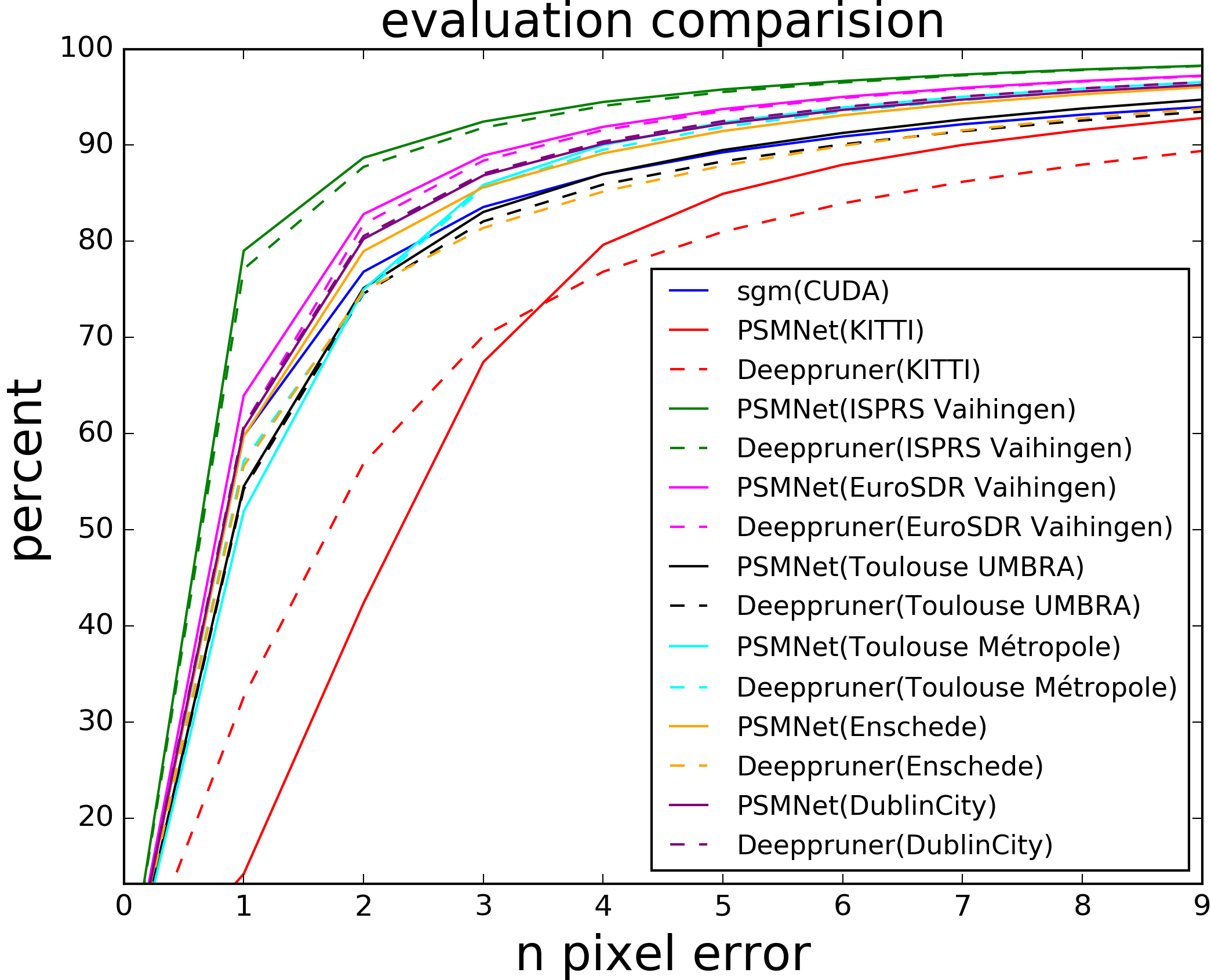}
	}
	\subfigure[Result on EuroSDR Vaihingen]{
		\label{Figure.transferlearning:b}
		\centering
		\includegraphics[width=0.45\linewidth]{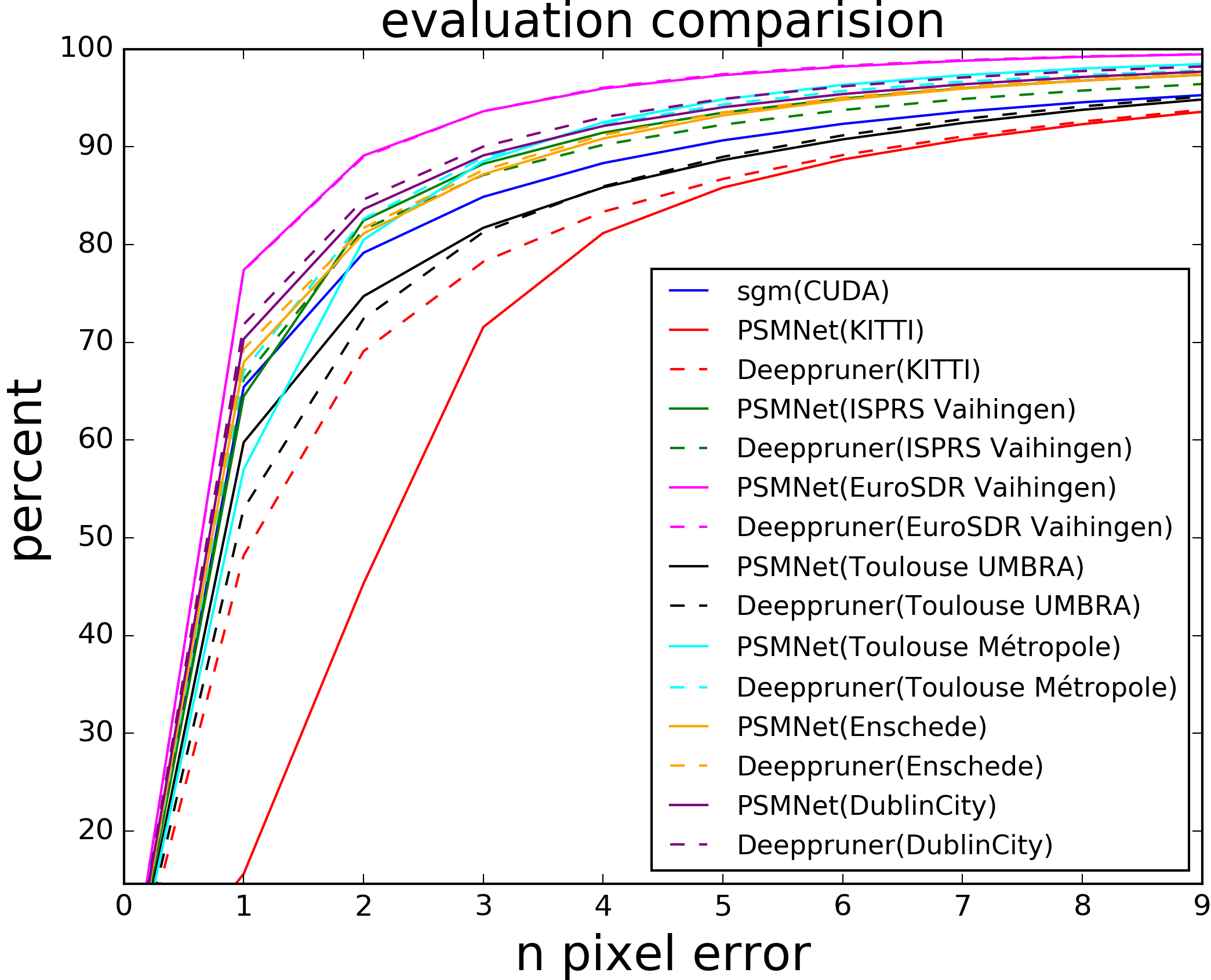}
	}
	
	\subfigure[Result on Toulouse UMBRA]{
		\label{Figure.transferlearning:d}
		\centering
		\includegraphics[width=0.45\linewidth]{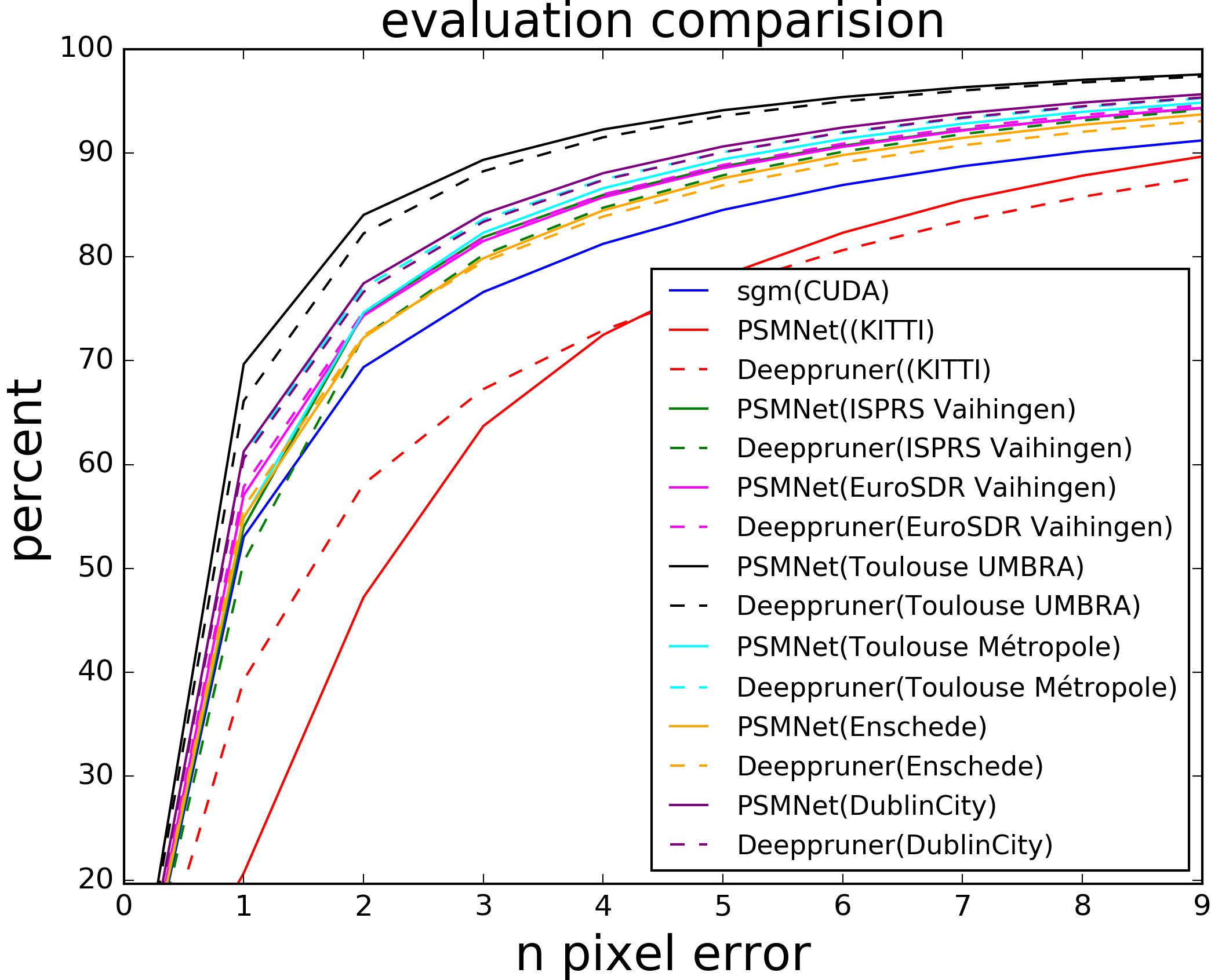}
	}
	\subfigure[Result on Toulouse Metropole]{
		\label{Figure.transferlearning:c}
		\centering
		\includegraphics[width=0.45\linewidth]{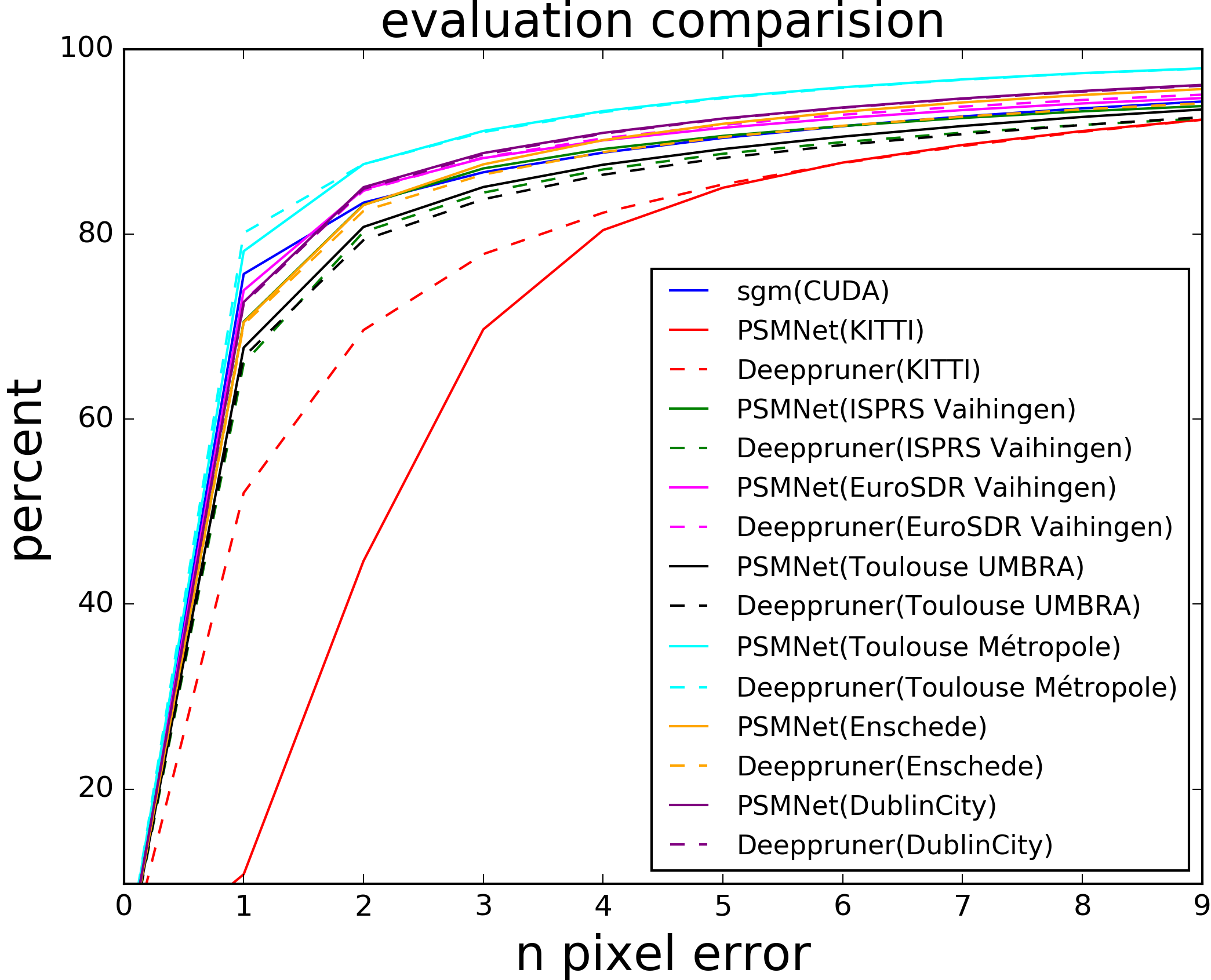}
	}

	\subfigure[Result on Enschede]{
		\label{Figure.transferlearning:e}
		\centering
		\includegraphics[width=0.45\linewidth]{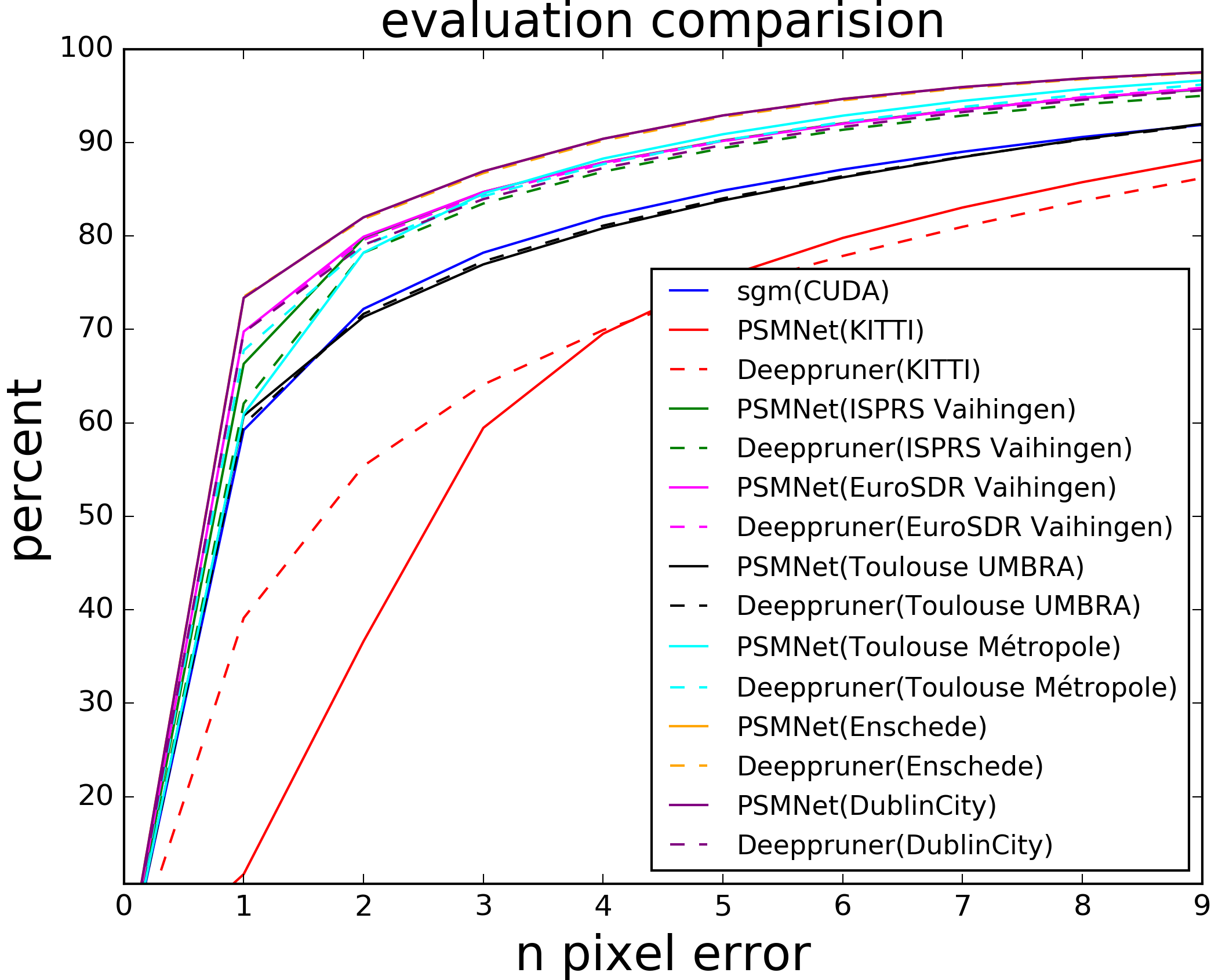}
	}
	\subfigure[Result on DublinCity]{
		\label{Figure.transferlearning:f}
		\centering
		\includegraphics[width=0.45\linewidth]{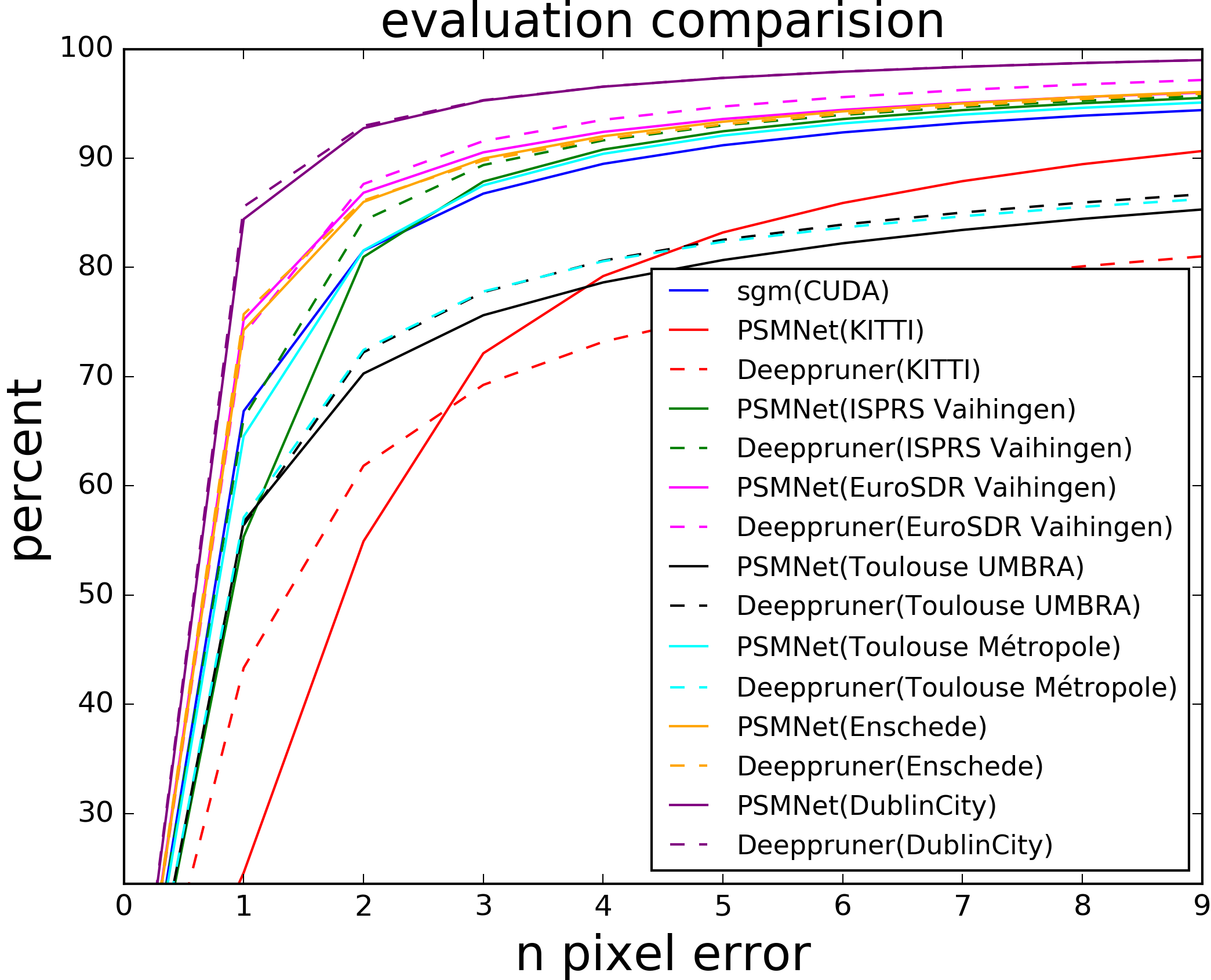}
	}
	
	\caption{Dataset shift on a single dataset.}
	\label{Figure.transferlearning}
\end{figure}

For the single data training dataset shift experiment we limited the number of lines in \cref{Figure.transferlearning} by only evaluating PSMnet and DeepPruner together. SGM(CUDA) is chosen as the baseline and we used the pre-trained KITTI model to analyse the dataset shift between different types of scenes. The name after each method indicates the training data.

\cref{Figure.transferlearning} shows that the KITTI models perform globally worse than the SGM(CUDA) baseline. DL-based methods perform better than the baseline but this improvement is around twice less when the model is trained on a different area. 
PSMNet has a better transferability than DeepPruner, which shows that transferability depends on the network structure.
 In our experiments, the model trained on Toulouse UMBRA performs badly. This is probably due to the fact that it has quite different brightness from others as shown in \cref{Figure.image_hist:c}. This shows that brightness  negatively influences transferability.
 
In \Cref{Figure.transferlearning:a}, the pink line shows that the model trained on the EuroSDR Vaihingen is better than the others, which shows that quality degrades more when training on a different area than when training at a different resolution.
EuroSDR Vaihingen is an exception: \cref{Figure.transferlearning:b} shows that PSMNet trained on DublinCity is slightly better than trained on ISPRS Vaihingen. All models except the one trained on Toulouse UMBRA have a better performance than SGM(CUDA).
In \Cref{Figure.transferlearning:c}, all the trained models are better than SGM(CUDA), because the contrast is low, which seems to have more impact on traditional methods. Training on Toulouse Metropole and testing on Toulouse UMBRA is the just a little worse than training from DublinCity which is our highest resolution dataset, which shows the importance of the resolution of the dataset used for training.
In \Cref{Figure.transferlearning:d}, the base to height ratio is small and SGM(CUDA) has a good performance while the model trained on Toulouse UMBRA performs badly.
For Enschede and DublinCity, there is only one dataset in the area. Because the $B/H$ is large in Enschede (cf \cref{Figure.bhratio}), the model trained on EuroSDR Vaihingen has a better performance, as show in \cref{Figure.transferlearning:e} (pink line and pink dot line).
Even though DublinCity has a very high resolution, the PSMNet model trained on EuroSDR Vaihingen has the best performance as shown in \cref{Figure.transferlearning:f}.

\subsubsection{Multi-data training in dataset shift}
\label{sec:multi_shift}
Based on the conclusions of the single datatset transferability experiment, we believe that training a model on multiple geographic areas can lead to a more generic model that performs well on unseen areas (areas not in the training set) which is a very important issue for the practical application to photogrammetry, as we cannot expect to have ground truth learning data everywhere.
As the performance of a network is obviously very dependant on the dataset used to train it, we propose an experiment with two settings : \textbf{\textit{Aerial all}} train the model on all available datasets; \textbf{\textit{Aerial real}} train the model on all datasets, except the ones that are on the same geographical area than the test area. We believe that \textbf{\textit{Aerial all}} is the best model that we can train based on our datasets, but we cannot evaluate how well it transfers to unseen areas. So \textbf{\textit{Aerial real}} is a more realistic setting where the test area is unseen in training. Considering that there are 4 different areas, the experiment is testing on one dataset (from one area) and training on datasets from the other 3 areas.
We evaluated dataset shift for all learning based methods but only present here the results for PSMNet, the results of the other method are provided in the appendix.

\begin{figure}[tp]
	\centering
	\subfigure[Result on ISPRS Vaihingen]{
		\label{Figure.datafusion:a}
		\centering
		\includegraphics[width=0.45\linewidth]{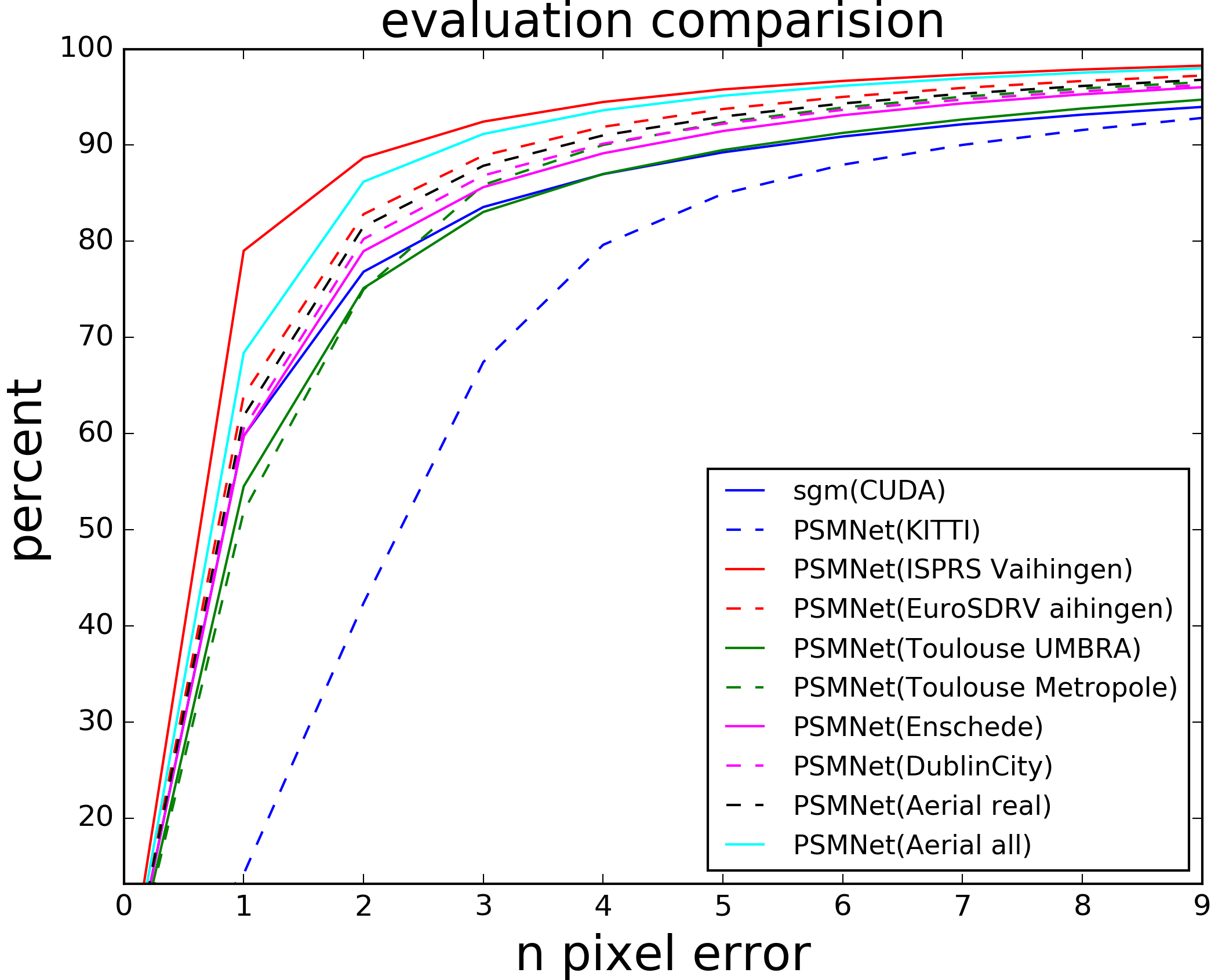}
	}
	\subfigure[Result on EuroSDR Vaihingen]{
		\label{Figure.datafusion:b}
		\centering
		\includegraphics[width=0.45\linewidth]{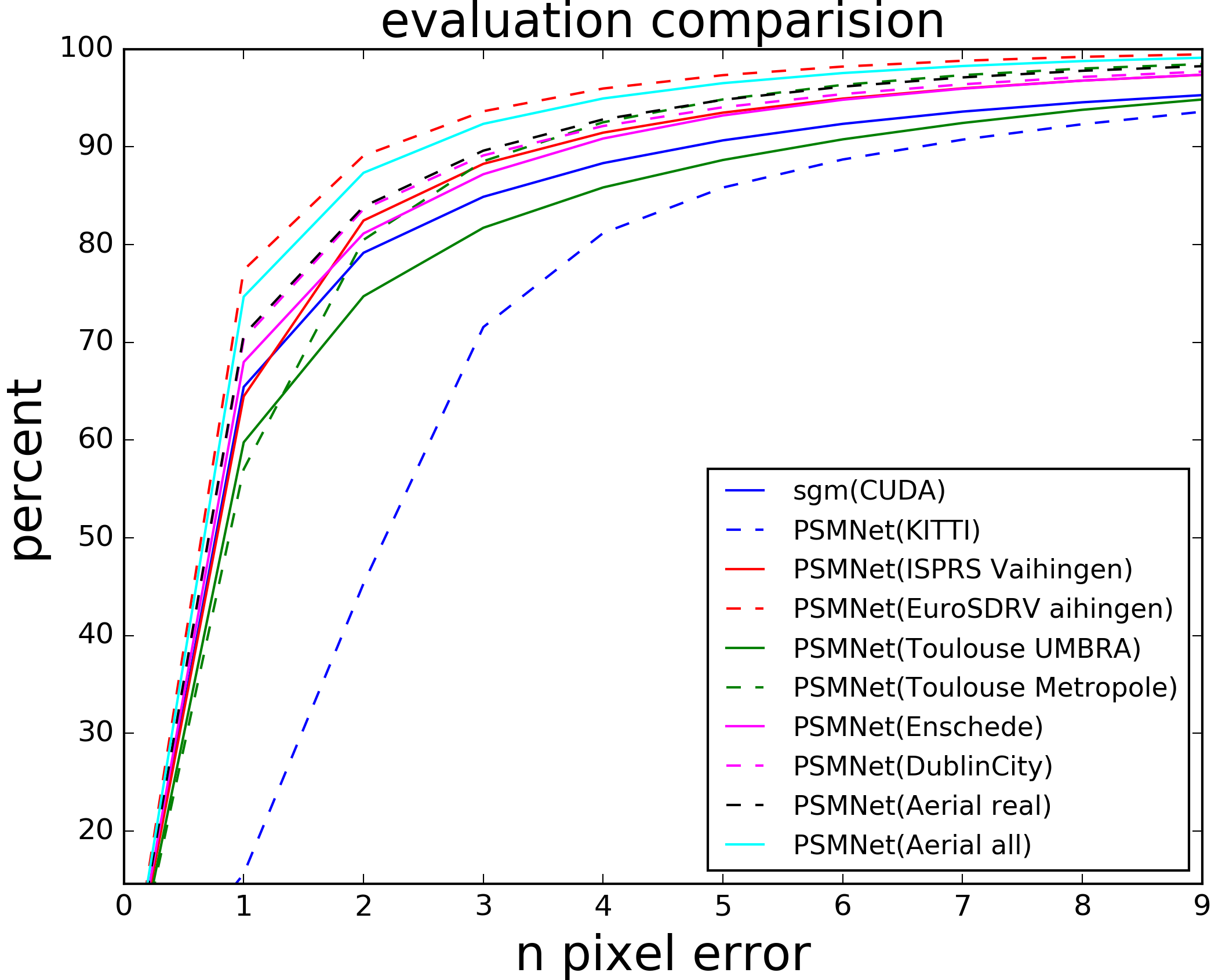}
	}
	
	\subfigure[Result on Toulouse UMBRA]{
		\label{Figure.datafusion:c}
		\centering
		\includegraphics[width=0.45\linewidth]{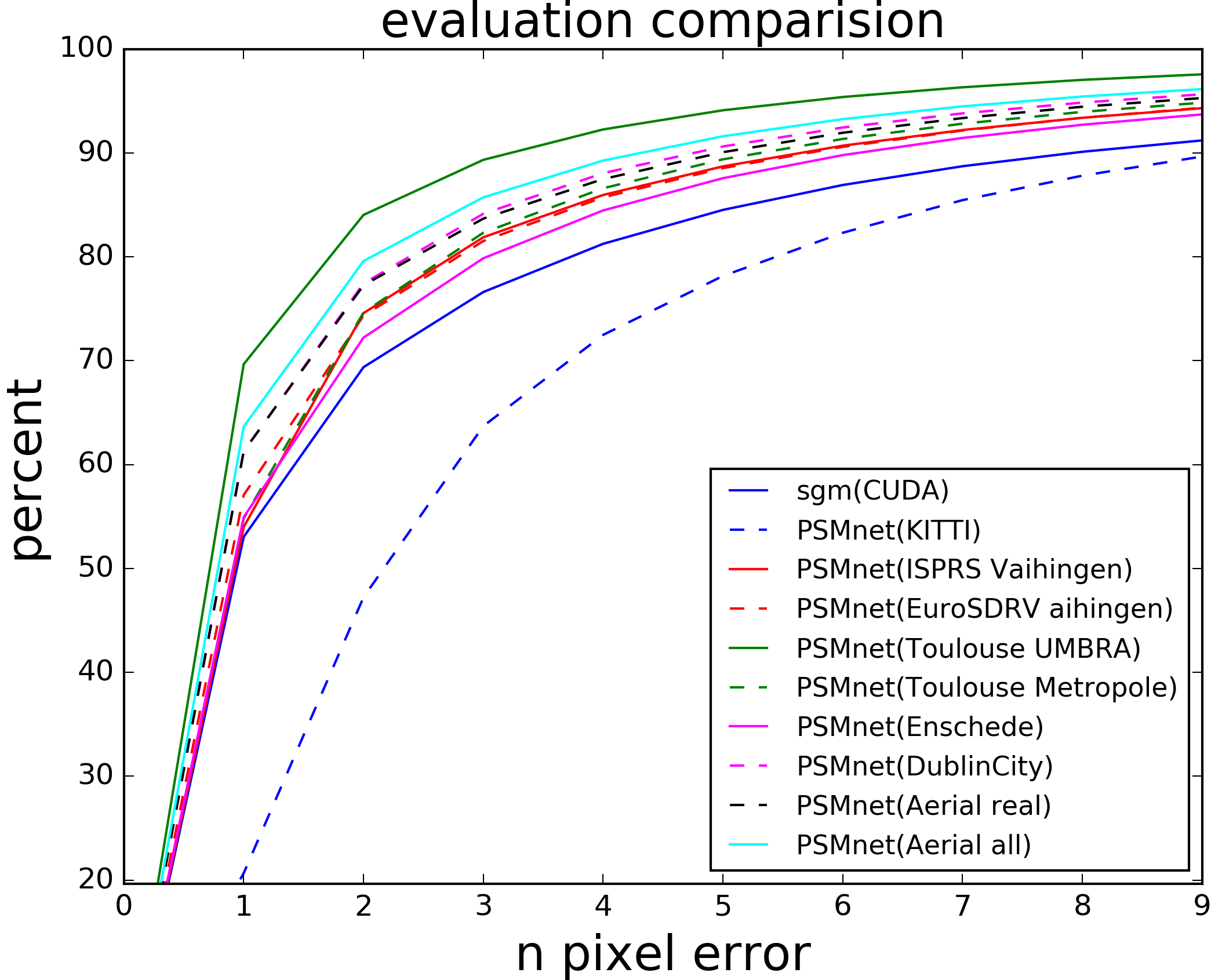}
	}
	\subfigure[Result on Toulouse Metropole]{
		\label{Figure.datafusion:d}
		\centering
		\includegraphics[width=0.45\linewidth]{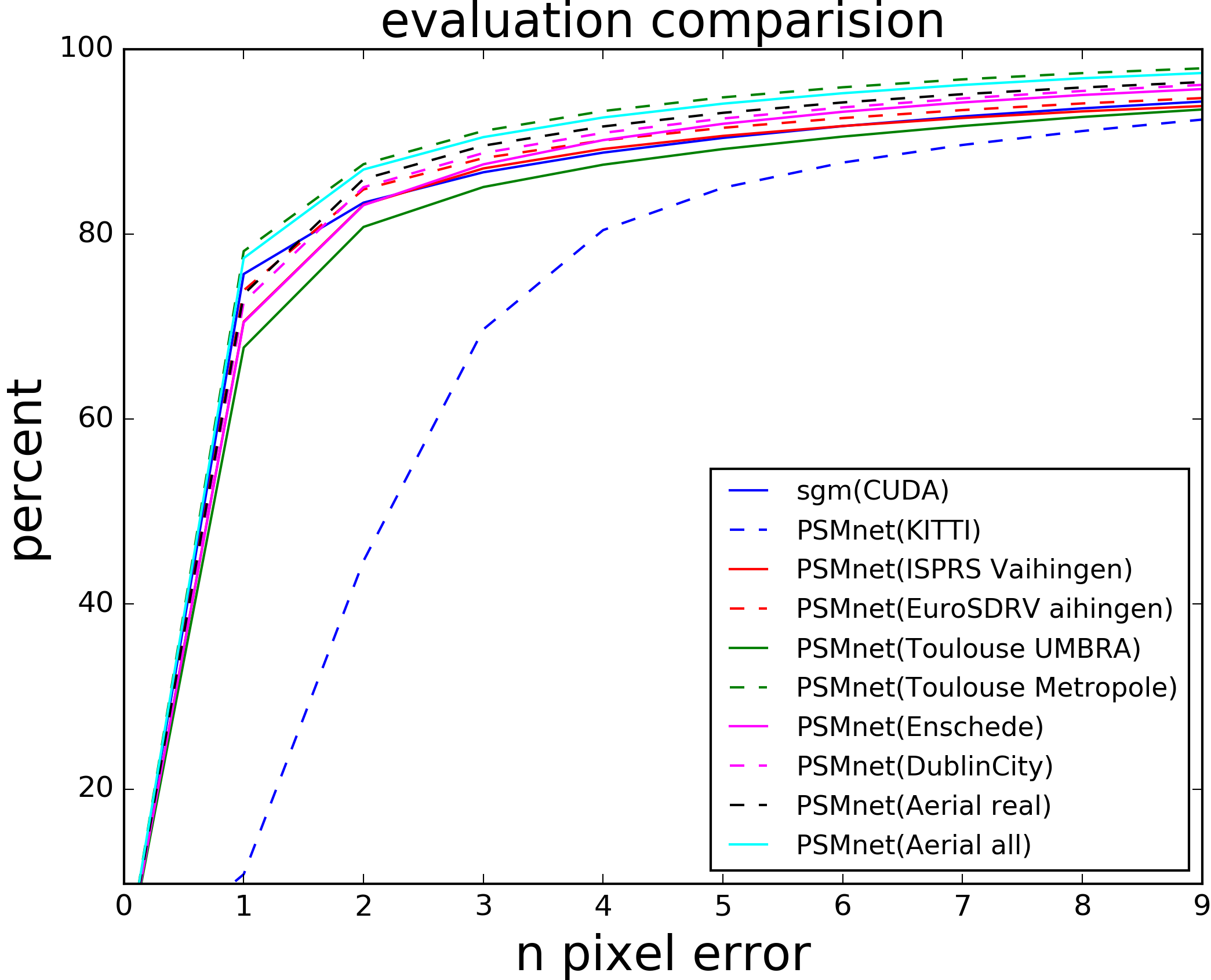}
	}
	
	\subfigure[Result on Enschede]{
		\label{Figure.datafusion:e}
		\centering
		\includegraphics[width=0.45\linewidth]{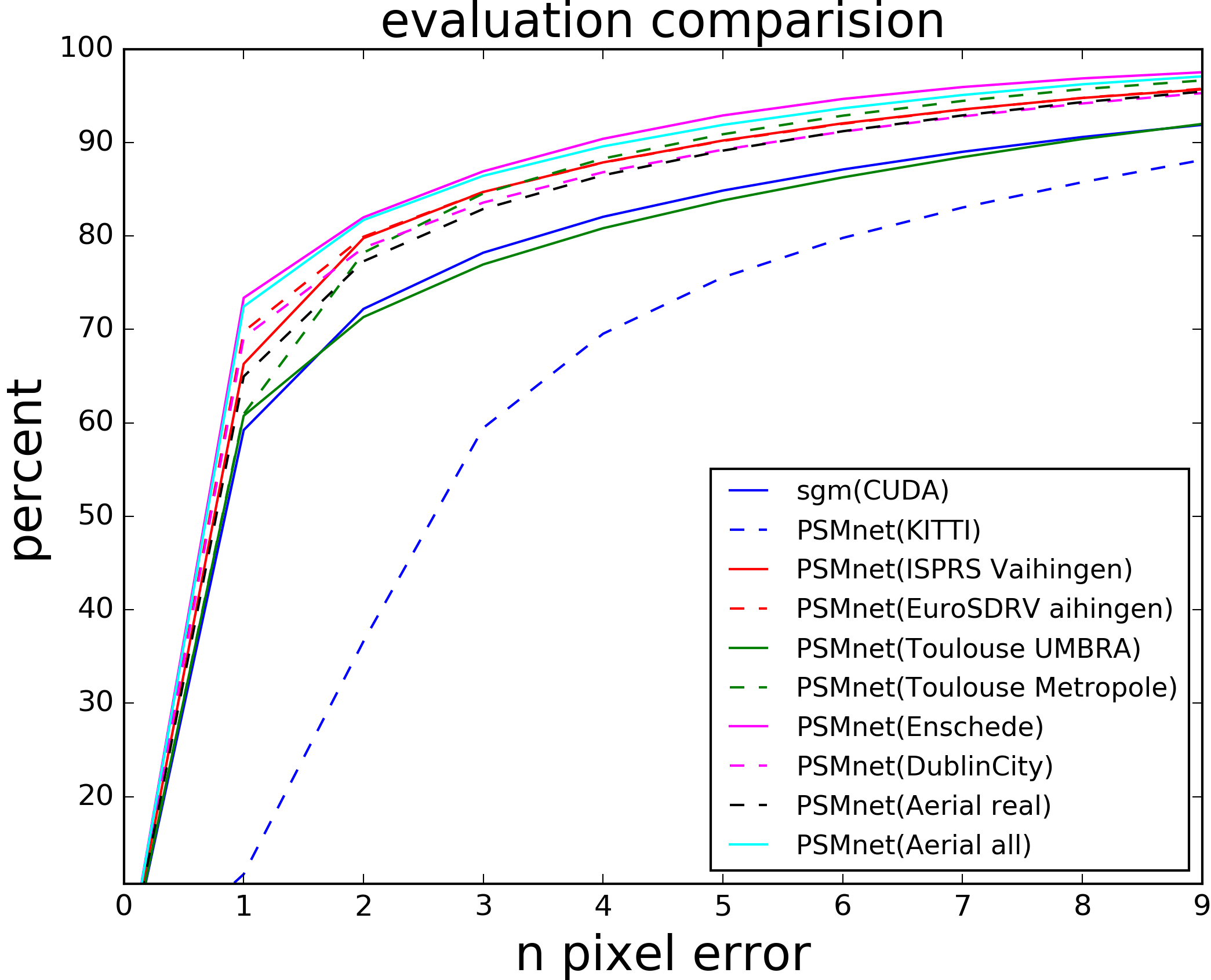}
	}
	\subfigure[Result on DublinCity]{
		\label{Figure.datafusion:f}
		\centering
		\includegraphics[width=0.45\linewidth]{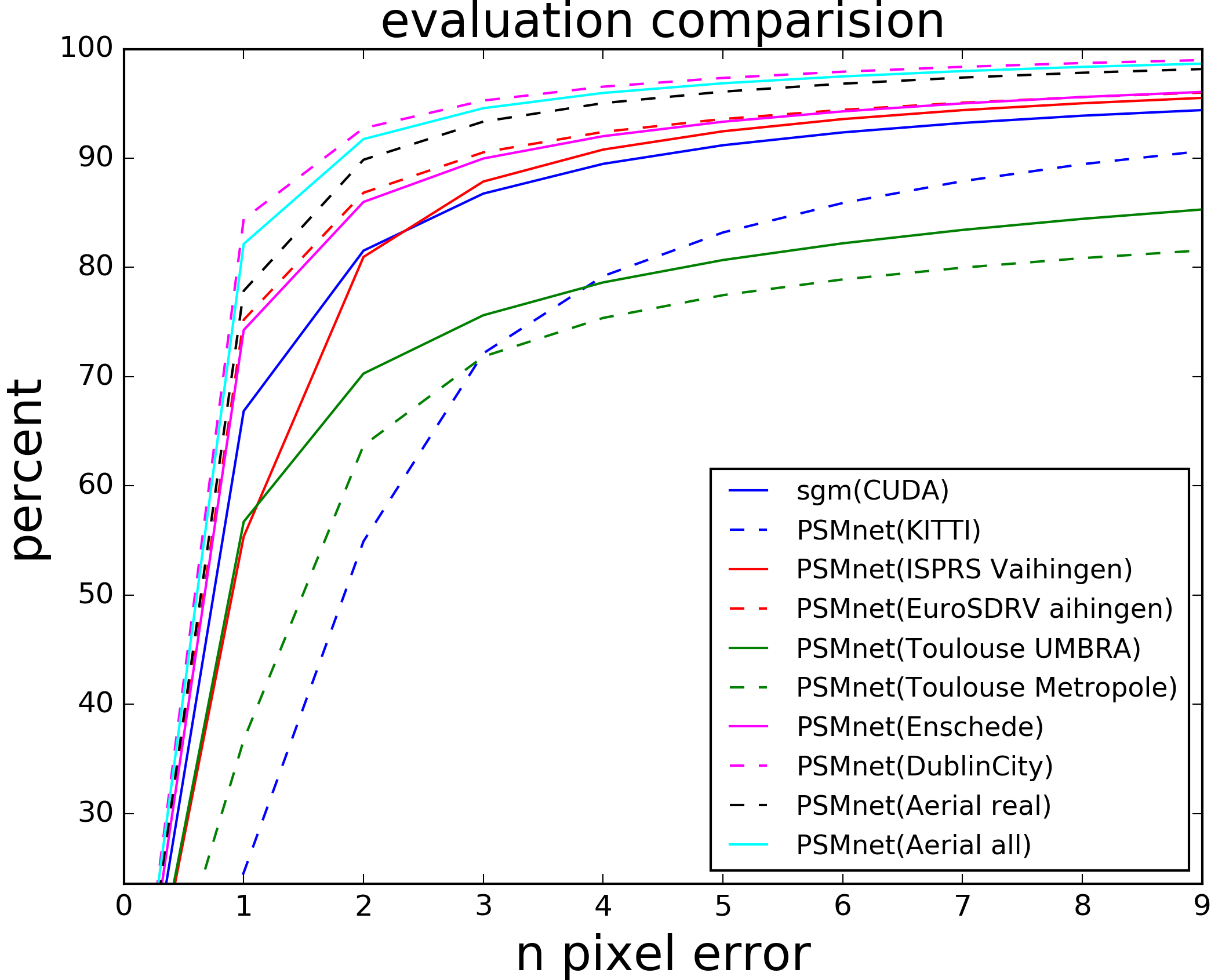}
	}
			
	\caption{Training with multi-dataset.}
	\label{Figure.datafusion}
\end{figure}
As shown in \Cref{Figure.datafusion}, training on a single dataset and testing on the same dataset is better than the model from \textbf{\textit{Aerial all}}. The model \textbf{\textit{Aerial all}} is always the second best. Models trained with datasets from different geographical areas \textbf{\textit{Aerial real}} transfer better to unseen areas than when the learning is on a single (different)area.
In \Cref{Figure.datafusion:a}, training from EuroSDR Vaihigen is better than the model from \textbf{\textit{Aerial real}}. 
In \Cref{Figure.datafusion:b,Figure.datafusion:d,Figure.datafusion:f}, the model training from \textbf{\textit{Aerial real}} is better than traing from a single dataset.
In \Cref{Figure.datafusion:c}, for the Toulouse UMBRA dataset, training modle from \textbf{\textit{Aerial real}} is near to Dublin and is better than the other training from a single dataset.
In \Cref{Figure.datafusion:e}, training from the model training from \textbf{\textit{Aerial real}} is only better than training from Toulouse UMBRA and Toulouse Metropole.


\subsubsection{Quantitative evaluation}
In order to make a quantitative evaluation of transferability, similar to the "relative transfer gain" in \citep{mensink2021factors}, we propose to estimate a "\textbf{relative shift gain}":
\begin{equation}
\label{equation:quantity}
R_{gain} = (\frac{p}{p_{base}} - 1 ) * 100
\end{equation}
In equation \Cref{equation:quantity}, $p$ is the exact opposite of the $N$-pixel error , once again, we use SGM(CUDA) as the baseline $p_{base}$ to analyse the "relative shift gain" for all the other methods.

While we displayed cumulative histograms in \Cref{Figure.datafusion}, we need single metrics here so we restrict to the 3-pixel error.
To analysis the base to height ratio, we add a new configuration Toulouse Metropole($B/H$), for the training, there are 400 pairs from small ($B/H<0.4$), 400 pairs from middle ($B/H=$ 0.4 \~{} 0.6) and 400 pairs from large ($B/H>0.6$), for the testing, there are also 20 stereo pairs from small $B/H$, 20 stereo pairs from large $B/H$, and 20 stereo pairs from large $B/H$. 
For a more detailed analysis, 1-pixel and average errors are included in the appendix.
In \Cref{Table:quantity_analysis}, in order to show the dataset shift, the results are highlighted by dataset. The color-coding in the tables ranges from red to green, the entire column indicates the transferability of a dataset, while the entire row indicates the transferability of a method.
From the experiment, we can make a summary, for the hybrid methods:
\begin{itemize}
 	\item Less sensitive to transfer.
 	\item Less improvement after fine-tuned compared to end-to-end methods.
\end{itemize}

For the end-to-end methods, the transferability depends on the methods:
\begin{itemize}
    \item PSMNet: the difference between the best case and the worst case is large.
    \item GANet and LEAStereo: the difference is smaller, indicating that they are less sensitive than PSMNet.
\end{itemize}

Concerning datasets:
\begin{itemize}
    \item Toulouse UMBRA: all the relative shift gains are large, which means that training on any dataset can improve the performance.
    \item Toulouse Metropole: training does not bring much improvement. But when the testing data contain large base to height ratio pairs, training can improve the performance.
\end{itemize}
In the Toulouse UMBRA and Toulouse Metropole columns, relative shift gains are often negative, which means that these dataset are probably too different from the others to hope for a good transfer.
On the other hand, for ISPRS Vaihingen and EuroSDR Vaihingen, relative shift gains are often positive. Even the input bands are different, the means the bands dosen't influence much in the DL-based methods.
Finally, we see that DublinCity performs best among the 6 dataset. As it is by far the dataset with the best resolution, a simple conclusion is that a better resolution makes a dataset better for training.

\begin{table}[!ht]
	\centering
	\caption{Relative shift gain on 3-pixel error based on SGM(CUDA)}
	\label{Table:quantity_analysis}
	\resizebox{\textwidth}{!}{
			\begin{tabular}{c|c|c|cc|ccc|c|c|cc}
			\noalign{\hrule height 2pt}
			\multirow{2}{*}{\backslashbox{Test}{Train}} & \multirow{2}{*}{Method} & \multirow{2}{*}{KITTI} &  ISPRS  & EuroSDR  & Touluose & Touluose  & Toulouse  & \multirow{2}{*}{Enschede} & Dublin & Aerial & Aerial \\
			~ & ~ & ~  &  Vaihingen &  Vaihingen &  UMBRA &  Metropole &  Metropole(B/H) & ~ & City & (real) & (all) \\
			\noalign{\hrule height 2pt}
			\multirow{7}{*}{ISPRS} & MC-CNN & \gradient{2.45} & \gradient{7.50} & \gradient{2.23} & \gradient{2.82} & \gradient{0.21} & \gradient{4.64} & \gradient{-0.75} & \gradient{3.41} & \gradient{1.80} & \gradient{4.23} \\
			\cline{2-10}
			~ & EfficientDeep & \gradient{-0.84} & \gradient{4.58} & \gradient{0.92} & \gradient{0.47} & \gradient{-0.66} & \gradient{-0.02} & \gradient{-4.41} & \gradient{0.36} & \gradient{0.41} & \gradient{1.23} \\
			\cline{2-10}
			~ & PSMNet & \gradient{-19.30} & \gradient{10.63} & \gradient{6.41} & \gradient{-0.61} & \gradient{2.75} & \gradient{4.46} & \gradient{2.47} & \gradient{3.93} & \gradient{5.14} & \gradient{9.10} \\
			\cline{2-10}
			~ & Deeppruner & \gradient{-16.06} & \gradient{9.88} & \gradient{5.77} & \gradient{-1.79} & \gradient{2.39} & \gradient{3.11} & \gradient{-2.60} & \gradient{4.17} & \gradient{5.42} & \gradient{8.69} \\
			\cline{2-10}
			~ & HRS Net & \gradient{-2.07} & \gradient{6.93} & \gradient{6.93} & \gradient{-4.87} & \gradient{-0.16} & \gradient{1.30} & \gradient{0.41} & \gradient{-1.23} & \gradient{1.72} & \gradient{5.23} \\
			\cline{2-10}
			~ & GANet & \gradient{-4.93} & \gradient{10.70} & \gradient{6.95} & \gradient{-1.44} & \gradient{2.31} & \gradient{3.26} & \gradient{1.82} & \gradient{1.36} & \gradient{3.05} & \gradient{8.20} \\
			\cline{2-10}
			~ & LEAStereo & \gradient{-7.67} & \gradient{7.70} & \gradient{5.58} & \gradient{0.55} & \gradient{4.45} & \gradient{4.79} & \gradient{6.03} & \gradient{4.46} & \gradient{5.08} & \gradient{5.96} \\
			\cline{2-10}
			\noalign{\hrule height 2pt}
			\multirow{7}{*}{EuroSDR} & MC-CNN & \gradient{3.26} & \gradient{5.99} & \gradient{7.87} & \gradient{3.72} & \gradient{4.93} & \gradient{5.39} & \gradient{4.91} & \gradient{4.07} & \gradient{5.82} & \gradient{6.60} \\
			\cline{2-10}
			~ & EfficientDeep & \gradient{-1.24} & \gradient{0.92} & \gradient{4.28} & \gradient{-1.83} & \gradient{0.62} & \gradient{0.51} & \gradient{3.78} & \gradient{0.45} & \gradient{0.32} & \gradient{0.32} \\
			\cline{2-10}
			~ & PSMNet & \gradient{-15.70} & \gradient{3.97} & \gradient{10.31} & \gradient{-3.73} & \gradient{4.29} & \gradient{5.82} & \gradient{2.72} & \gradient{4.99} & \gradient{5.56} & \gradient{8.79} \\
			\cline{2-10}
			~ & Deeppruner & \gradient{-7.82} & \gradient{2.63} & \gradient{10.32} & \gradient{-4.25} & \gradient{4.72} & \gradient{4.99} & \gradient{3.22} & \gradient{6.06} & \gradient{4.47} & \gradient{9.10} \\
			\cline{2-10}
			~ & HRS Net & \gradient{-4.51} & \gradient{-3.50} & \gradient{7.14} & \gradient{-6.12} & \gradient{1.05} & \gradient{1.61} & \gradient{0.58} & \gradient{0.43} & \gradient{4.34} & \gradient{4.23} \\
			\cline{2-10}
			~ & GANet & \gradient{-5.26} & \gradient{5.31} & \gradient{11.57} & \gradient{-2.43} & \gradient{5.42} & \gradient{6.88} & \gradient{3.97} & \gradient{5.15} & \gradient{6.24} & \gradient{7.92} \\
			\cline{2-10}
			~ & LEAStereo & \gradient{-11.78} & \gradient{3.20} & \gradient{6.77} & \gradient{-1.79} & \gradient{3.21} & \gradient{4.39} & \gradient{5.38} & \gradient{4.07} & \gradient{4.53} & \gradient{5.21} \\
			\cline{2-10}
			\noalign{\hrule height 2pt}
			\multirow{7}{*}{UMBRA} & MC-CNN & \gradient{5.62} & \gradient{7.19} & \gradient{4.66} & \gradient{10.17} & \gradient{5.60} & \gradient{6.87} & \gradient{3.21} & \gradient{5.92} & \gradient{6.30} & \gradient{8.80} \\
			\cline{2-10}
			~ & EfficientDeep & \gradient{5.35} & \gradient{6.98} & \gradient{1.78} & \gradient{9.13} & \gradient{5.55} & \gradient{5.83} & \gradient{1.10} & \gradient{4.10} & \gradient{5.22} & \gradient{7.39} \\
			\cline{2-10}
			~ & PSMNet & \gradient{-16.84} & \gradient{6.88} & \gradient{6.40} & \gradient{16.61} & \gradient{7.45} & \gradient{7.48} & \gradient{4.24} & \gradient{9.83} & \gradient{9.23} & \gradient{11.90} \\
			\cline{2-10}
			~ & Deeppruner & \gradient{-12.20} & \gradient{4.67} & \gradient{6.78} & \gradient{15.18} & \gradient{9.11} & \gradient{5.88} & \gradient{3.81} & \gradient{8.82} & \gradient{8.98} & \gradient{11.91} \\
			\cline{2-10}
			~ & HRS Net & \gradient{-0.20} & \gradient{2.06} & \gradient{0.77} & \gradient{12.33} & \gradient{2.06} & \gradient{3.03} & \gradient{3.27} & \gradient{4.29} & \gradient{7.01} & \gradient{10.42} \\
			\cline{2-10}
			~ & GANet & \gradient{5.35} & \gradient{8.46} & \gradient{6.42} & \gradient{14.84} & \gradient{8.33} & \gradient{6.18} & \gradient{5.25} & \gradient{9.25} & \gradient{9.53} & \gradient{11.24} \\
			\cline{2-10}
			~ & LEAStereo & \gradient{-11.05} & \gradient{7.56} & \gradient{5.83} & \gradient{13.84} & \gradient{6.24} & \gradient{6.12} & \gradient{9.73} & \gradient{9.43} & \gradient{8.47} & \gradient{9.55} \\
			\cline{2-10}
			\noalign{\hrule height 2pt}
			\multirow{7}{*}{Metropole} & MC-CNN & \gradient{-2.44} & \gradient{-0.41} & \gradient{0.64} & \gradient{-0.51} & \gradient{1.32} & \gradient{1.73} & \gradient{-0.04} & \gradient{-1.39} & \gradient{0.92} & \gradient{1.94} \\
			\cline{2-10}
			~ & EfficientDeep & \gradient{-1.23} & \gradient{-2.58} & \gradient{-0.13} & \gradient{-1.34} & \gradient{0.15} & \gradient{-0.02} & \gradient{-1.34} & \gradient{-0.63} & \gradient{-0.63} & \gradient{-0.35} \\
			\cline{2-10}
			~ & PSMNet & \gradient{-19.59} & \gradient{0.48} & \gradient{1.77} & \gradient{-1.84} & \gradient{5.16} & \gradient{4.91} & \gradient{0.99} & \gradient{1.23} & \gradient{3.32} & \gradient{3.32} \\
			\cline{2-10}
			~ & Deeppruner & \gradient{-10.21} & \gradient{-2.54} & \gradient{1.85} & \gradient{-3.36} & \gradient{5.02} & \gradient{4.86} & \gradient{-0.28} & \gradient{2.23} & \gradient{2.28} & \gradient{3.39} \\
			\cline{2-10}
			~ & HRS Net & \gradient{-4.44} & \gradient{-2.06} & \gradient{-0.47} & \gradient{-5.38} & \gradient{1.66} & \gradient{2.31} & \gradient{-0.64} & \gradient{-0.83} & \gradient{0.52} & \gradient{1.02} \\
			\cline{2-10}
			~ & GANet & \gradient{-1.23} & \gradient{1.54} & \gradient{2.81} & \gradient{-0.83} & \gradient{4.25} & \gradient{4.70} & \gradient{1.01} & \gradient{2.05} & \gradient{2.35} & \gradient{3.65} \\
			\cline{2-10}
			~ & LEAStereo & \gradient{-11.44} & \gradient{1.01} & \gradient{1.63} & \gradient{-0.42} & \gradient{2.59} & \gradient{2.31} & \gradient{2.17} & \gradient{1.76} & \gradient{1.92} & \gradient{2.23} \\
			\cline{2-10}
			\noalign{\hrule height 2pt}
			\multirow{7}{*}{Metropole(B/H)} & MC-CNN & \gradient{1.21} & \gradient{2.97} & \gradient{3.45} & \gradient{2.45} & \gradient{6.50} & \gradient{7.79} & \gradient{3.21} & \gradient{1.31} & \gradient{5.26} & \gradient{10.81} \\
			\cline{2-10}
			~ & EfficientDeep & \gradient{1.94} & \gradient{0.42} & \gradient{3.64} & \gradient{1.81} & \gradient{4.76} & \gradient{7.77} & \gradient{8.04} & \gradient{3.98} & \gradient{4.73} & \gradient{4.51} \\
			\cline{2-10}
			~ & PSMNet & \gradient{-17.62} & \gradient{5.46} & \gradient{6.85} & \gradient{1.44} & \gradient{12.20} & \gradient{12.67} & \gradient{7.09} & \gradient{7.59} & \gradient{9.64} & \gradient{10.95} \\
			\cline{2-10}
			~ & Deeppruner & \gradient{-9.74} & \gradient{2.53} & \gradient{7.18} & \gradient{0.21} & \gradient{11.40} & \gradient{12.03} & \gradient{5.95} & \gradient{6.95} & \gradient{8.59} & \gradient{9.69} \\
			\cline{2-10}
			~ & HRS Net & \gradient{-1.22} & \gradient{0.18} & \gradient{4.85} & \gradient{-1.68} & \gradient{5.70} & \gradient{8.15} & \gradient{3.99} & \gradient{2.70} & \gradient{5.85} & \gradient{6.66} \\
			\cline{2-10}
			~ & GANet & \gradient{-0.94} & \gradient{4.96} & \gradient{7.77} & \gradient{3.25} & \gradient{11.40} & \gradient{11.80} & \gradient{6.11} & \gradient{6.26} & \gradient{8.73} & \gradient{9.55} \\
			\cline{2-10}
			~ & LEAStereo & \gradient{-12.47} & \gradient{3.98} & \gradient{6.03} & \gradient{2.22} & \gradient{5.42} & \gradient{6.22} & \gradient{6.58} & \gradient{5.92} & \gradient{6.72} & \gradient{6.03} \\
			\cline{2-10}
			\noalign{\hrule height 2pt}
			\multirow{7}{*}{Enschede} & MC-CNN & \gradient{2.03} & \gradient{2.02} & \gradient{5.20} & \gradient{-0.06} & \gradient{-0.12} & \gradient{1.76} & \gradient{6.00} & \gradient{4.00} & \gradient{5.04} & \gradient{5.62} \\
			\cline{2-10}
			~ & EfficientDeep & \gradient{2.52} & \gradient{2.61} & \gradient{4.11} & \gradient{0.97} & \gradient{2.89} & \gradient{3.07} & \gradient{4.16} & \gradient{3.05} & \gradient{3.18} & \gradient{3.56} \\
			\cline{2-10}
			~ & PSMNet & \gradient{-23.97} & \gradient{8.30} & \gradient{8.31} & \gradient{-1.60} & \gradient{8.10} & \gradient{8.21} & \gradient{11.13} & \gradient{6.85} & \gradient{5.98} & \gradient{10.51} \\
			\cline{2-10}
			~ & Deeppruner & \gradient{-18.05} & \gradient{6.72} & \gradient{7.98} & \gradient{-1.19} & \gradient{7.71} & \gradient{7.07} & \gradient{10.88} & \gradient{7.33} & \gradient{10.49} & \gradient{9.87} \\
			\cline{2-10}
			~ & HRS Net & \gradient{1.53} & \gradient{4.22} & \gradient{5.67} & \gradient{-3.06} & \gradient{6.14} & \gradient{5.98} & \gradient{8.32} & \gradient{3.48} & \gradient{5.46} & \gradient{7.08} \\
			\cline{2-10}
			~ & GANet & \gradient{-8.26} & \gradient{7.68} & \gradient{9.26} & \gradient{-0.71} & \gradient{8.27} & \gradient{7.81} & \gradient{10.88} & \gradient{6.20} & \gradient{7.26} & \gradient{8.87} \\
			\cline{2-10}
			~ & LEAStereo & \gradient{-15.35} & \gradient{8.26} & \gradient{7.94} & \gradient{2.03} & \gradient{6.82} & \gradient{6.98} & \gradient{7.76} & \gradient{6.72} & \gradient{7.76} & \gradient{8.08} \\
			\cline{2-10}
			\noalign{\hrule height 2pt}
			\multirow{7}{*}{DublinCity} & MC-CNN & \gradient{1.60} & \gradient{-1.18} & \gradient{2.96} & \gradient{-1.12} & \gradient{0.55} & \gradient{2.02} & \gradient{3.16} & \gradient{5.25} & \gradient{2.04} & \gradient{4.90} \\
			\cline{2-10}
			~ & EfficientDeep & \gradient{0.51} & \gradient{-4.33} & \gradient{2.76} & \gradient{-3.72} & \gradient{1.47} & \gradient{1.53} & \gradient{1.63} & \gradient{4.10} & \gradient{2.40} & \gradient{3.79} \\
			\cline{2-10}
			~ & PSMNet & \gradient{-16.85} & \gradient{1.27} & \gradient{4.35} & \gradient{-12.84} & \gradient{-17.20} & \gradient{-0.48} & \gradient{3.70} & \gradient{9.81} & \gradient{7.60} & \gradient{9.01} \\
			\cline{2-10}
			~ & Deeppruner & \gradient{-20.18} & \gradient{3.01} & \gradient{5.54} & \gradient{-10.40} & \gradient{-10.33} & \gradient{-10.14} & \gradient{3.49} & \gradient{9.86} & \gradient{6.13} & \gradient{9.13} \\
			\cline{2-10}
			~ & HRS Net & \gradient{1.79} & \gradient{0.66} & \gradient{3.86} & \gradient{-5.73} & \gradient{-5.45} & \gradient{-0.32} & \gradient{4.29} & \gradient{7.88} & \gradient{5.22} & \gradient{7.28} \\
			\cline{2-10}
			~ & GANet & \gradient{-11.00} & \gradient{-2.90} & \gradient{4.73} & \gradient{-1.15} & \gradient{0.32} & \gradient{-6.74} & \gradient{6.39} & \gradient{9.00} & \gradient{6.85} & \gradient{8.47} \\
			\cline{2-10}
			~ & LEAStereo & \gradient{-13.36} & \gradient{3.84} & \gradient{5.71} & \gradient{2.24} & \gradient{3.11} & \gradient{4.26} & \gradient{7.12} & \gradient{7.45} & \gradient{5.17} & \gradient{7.16} \\
			\cline{2-10}
			\noalign{\hrule height 2pt}
		\end{tabular}
	}
\end{table}

\section{Visual assessment}

\begin{figure}[tp]
	\begin{minipage}[t]{0.19\textwidth}
		\includegraphics[width=0.098\linewidth]{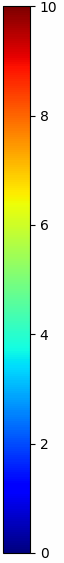}
		\includegraphics[width=0.85\linewidth]{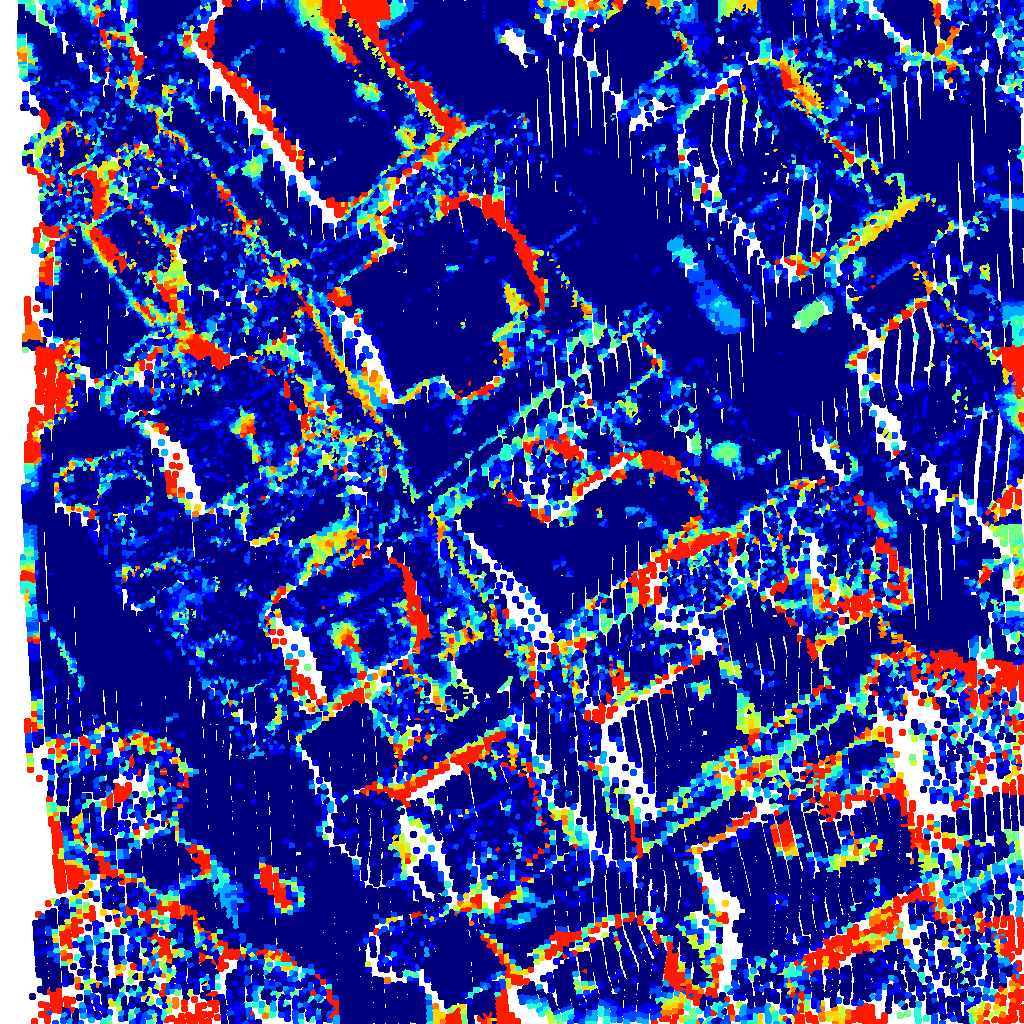}
		\includegraphics[width=\linewidth]{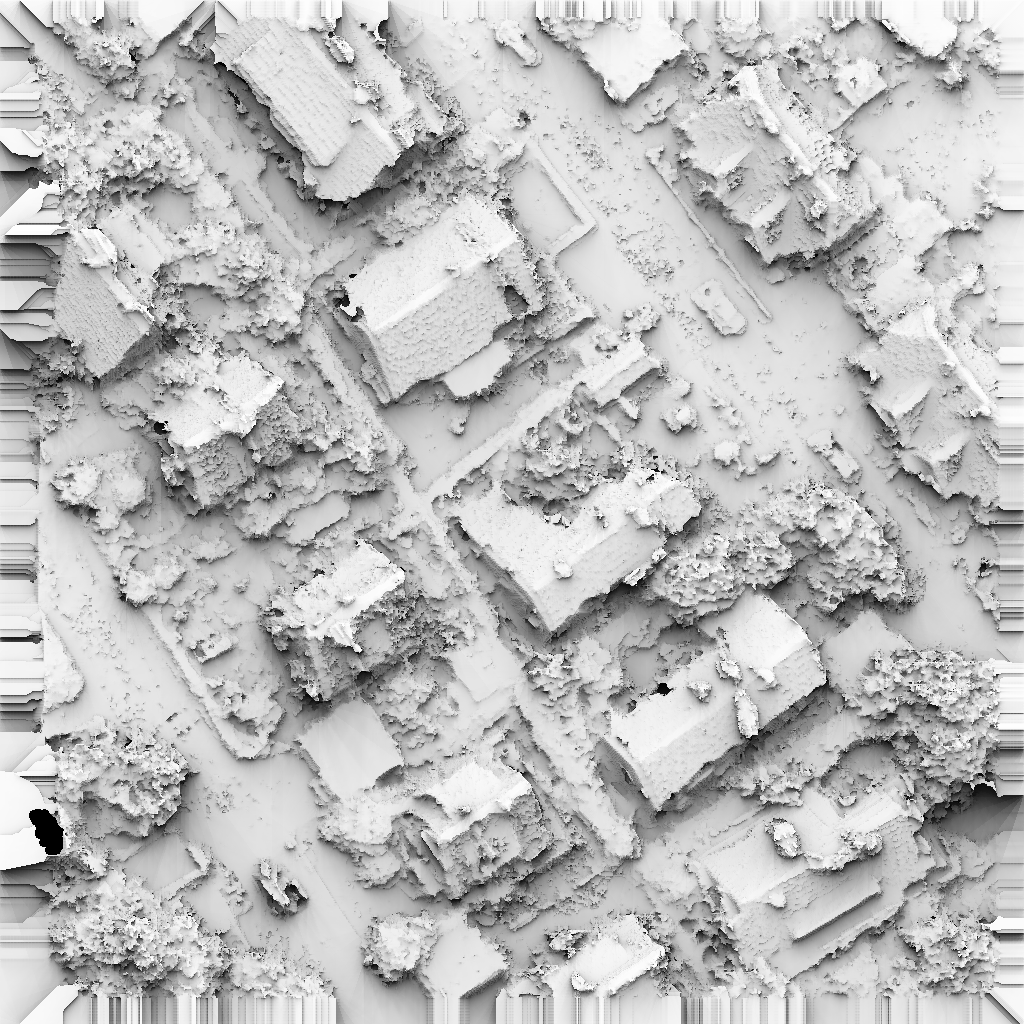}
		\centering{\tiny MICMAC}
	\end{minipage}
	\begin{minipage}[t]{0.19\textwidth}	
		\includegraphics[width=0.098\linewidth]{figures/color_map.png}
		\includegraphics[width=0.85\linewidth]{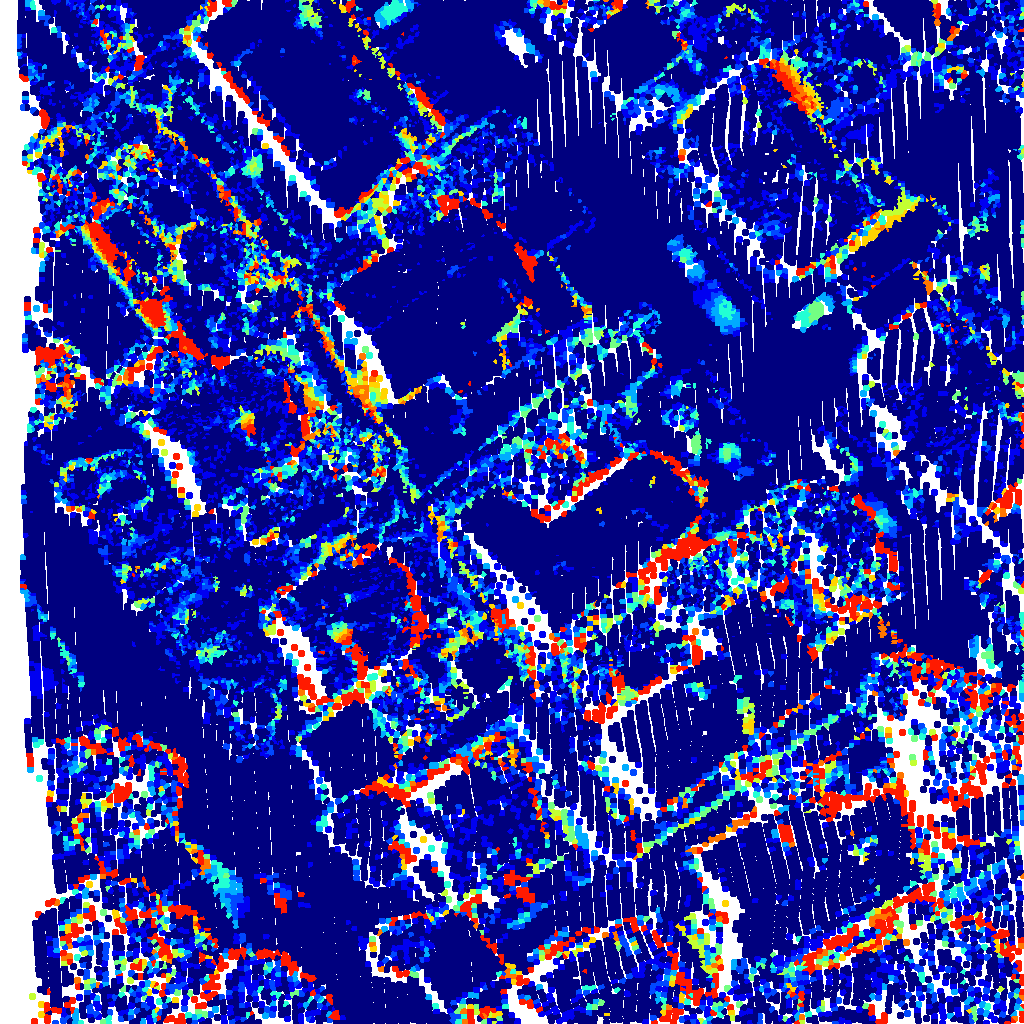}
		\includegraphics[width=\linewidth]{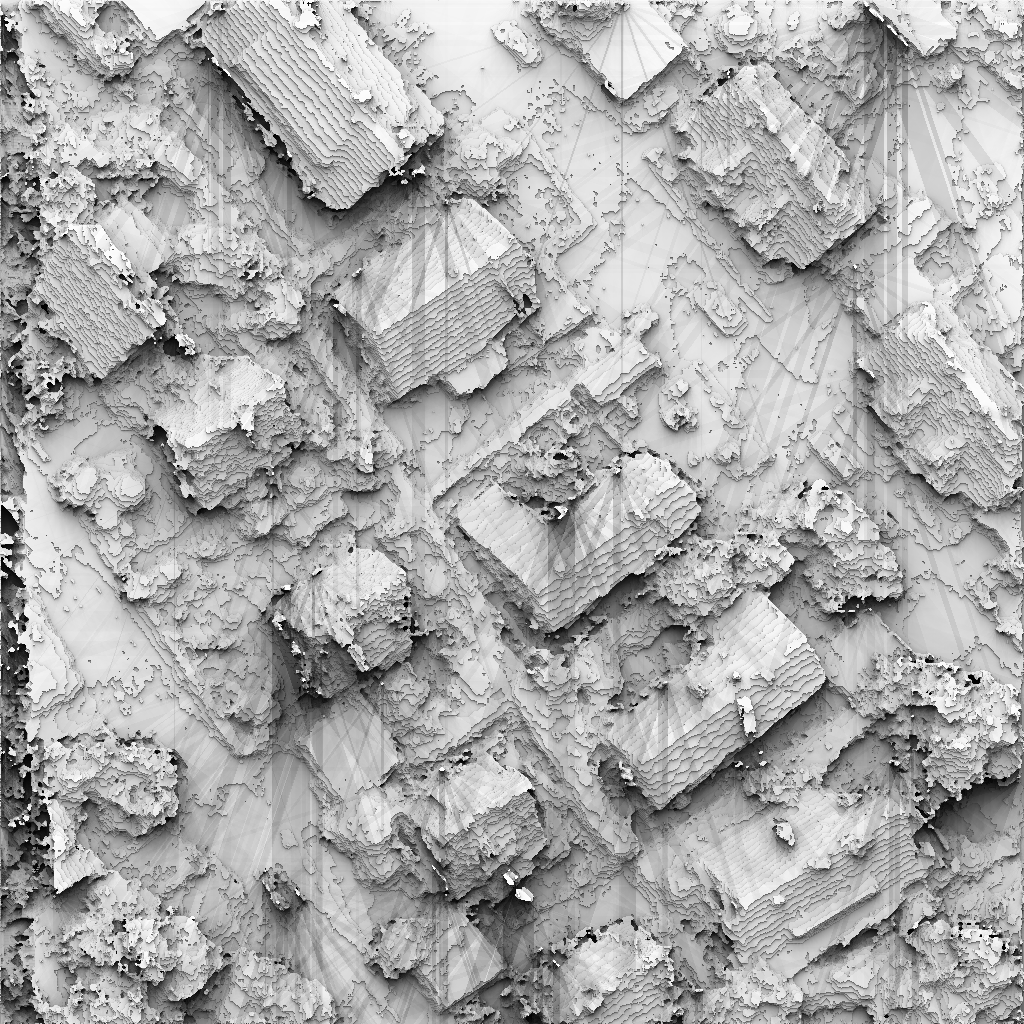}
		\centering{\tiny SGM(CUDA)}
	\end{minipage}
	\begin{minipage}[t]{0.19\textwidth}	
		\includegraphics[width=0.098\linewidth]{figures/color_map.png}
		\includegraphics[width=0.85\linewidth]{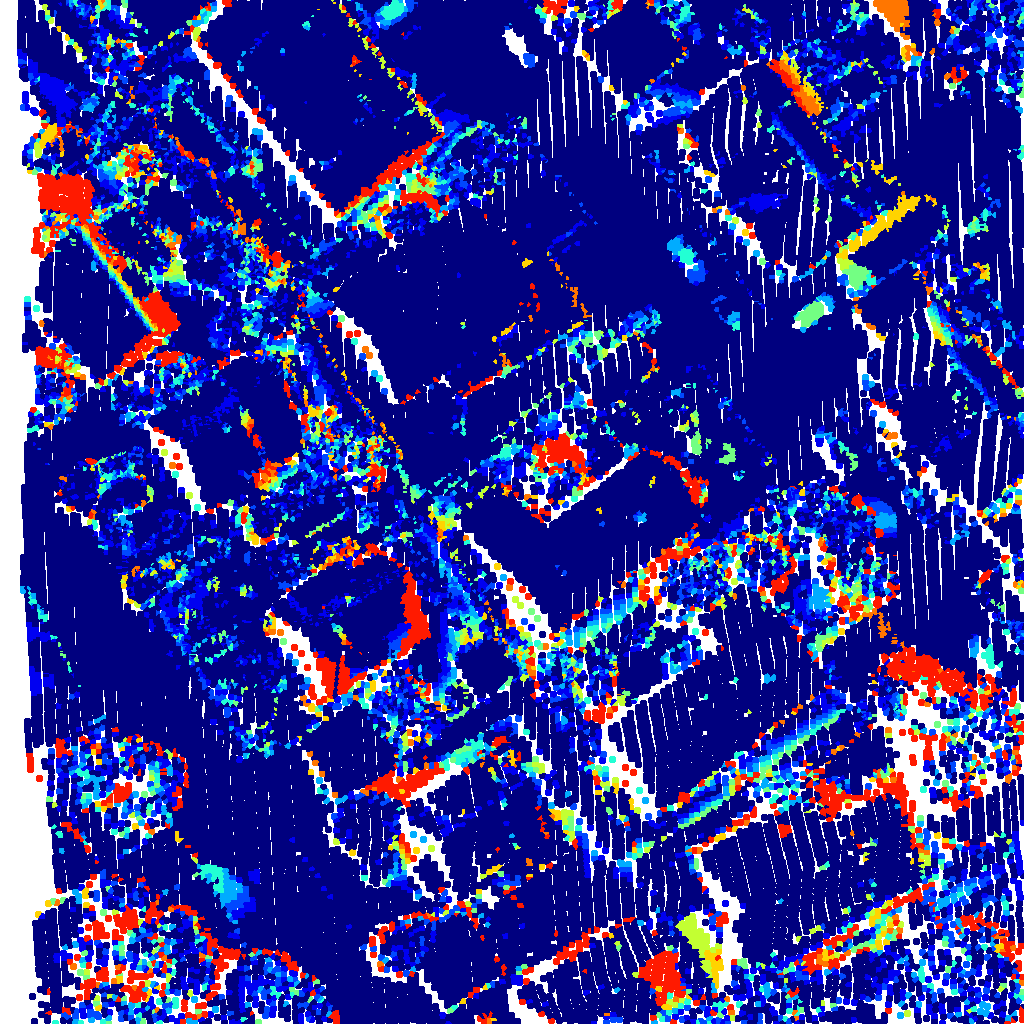}
		\includegraphics[width=\linewidth]{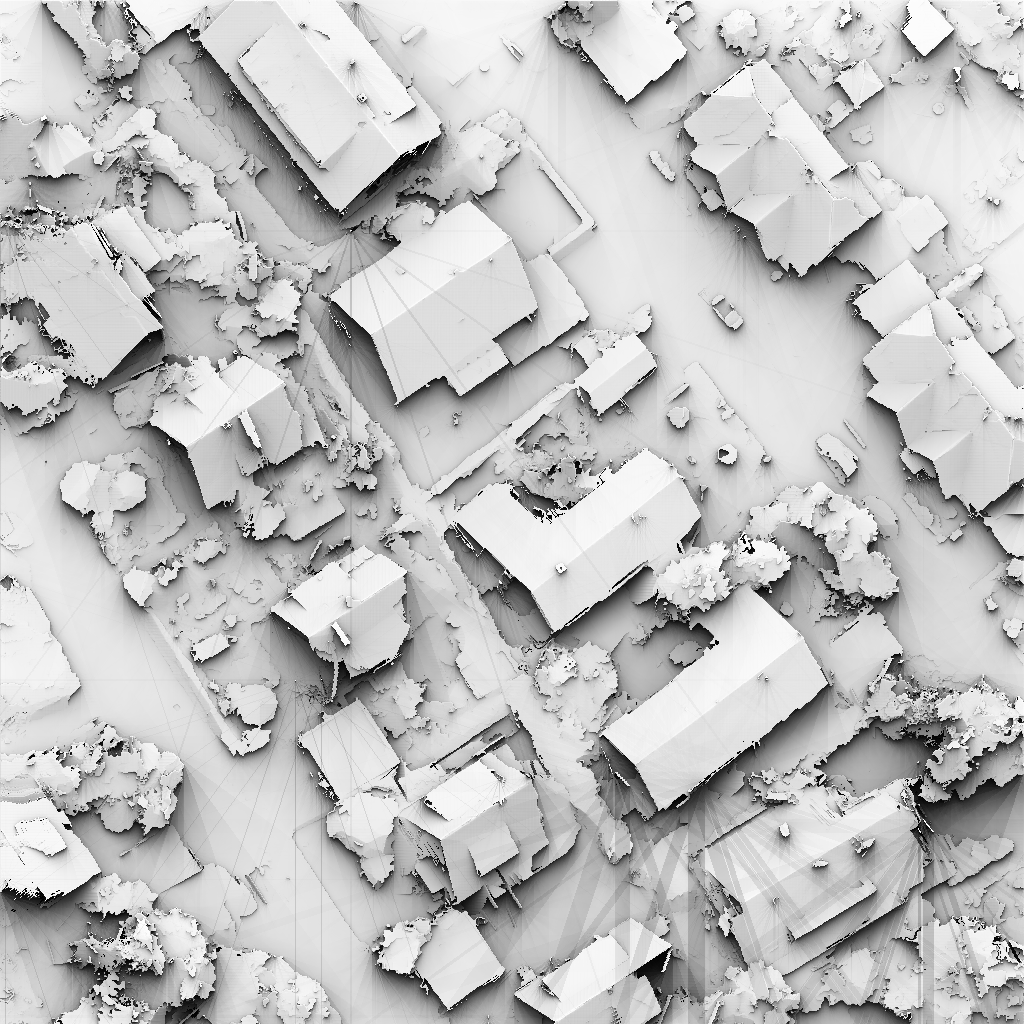}
		\centering{\tiny GraphCuts}
	\end{minipage}
	\begin{minipage}[t]{0.19\textwidth}	
		\includegraphics[width=0.098\linewidth]{figures/color_map.png}
		\includegraphics[width=0.85\linewidth]{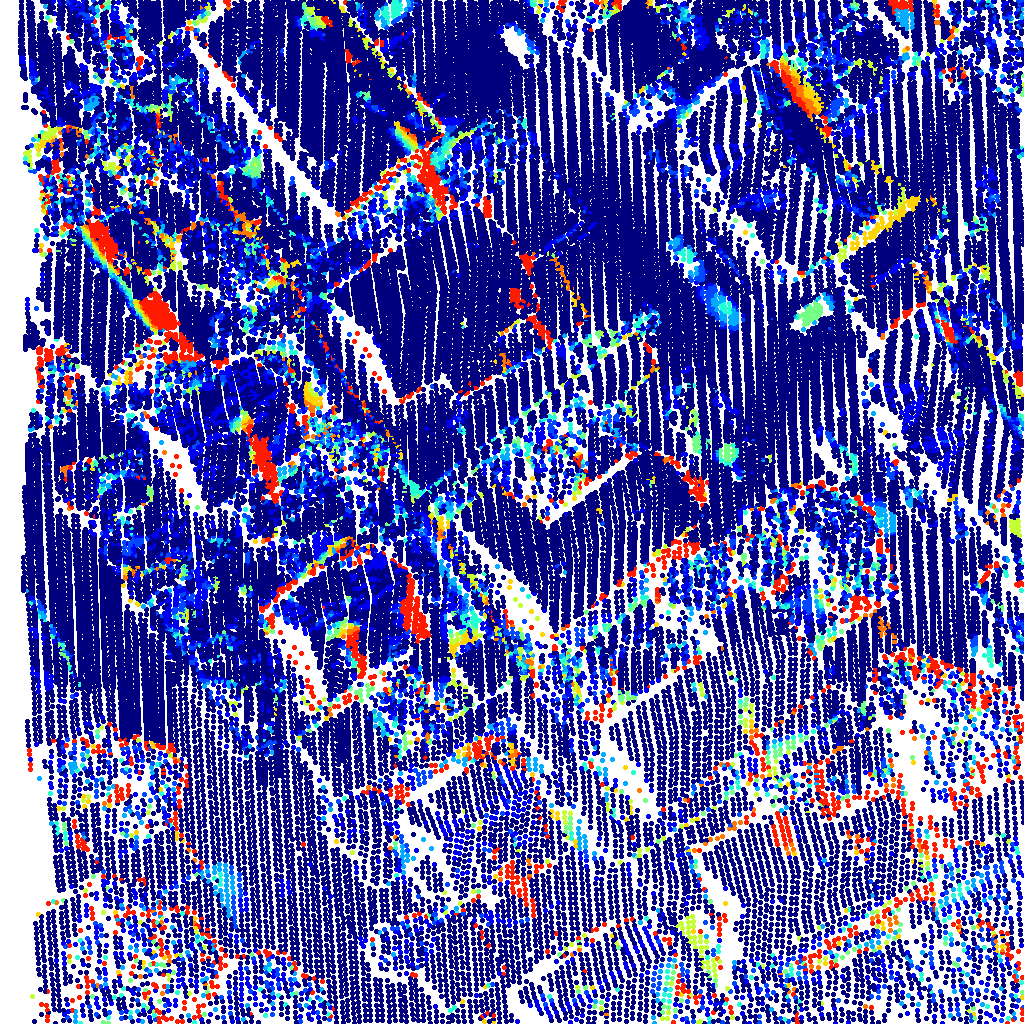}
		\includegraphics[width=\linewidth]{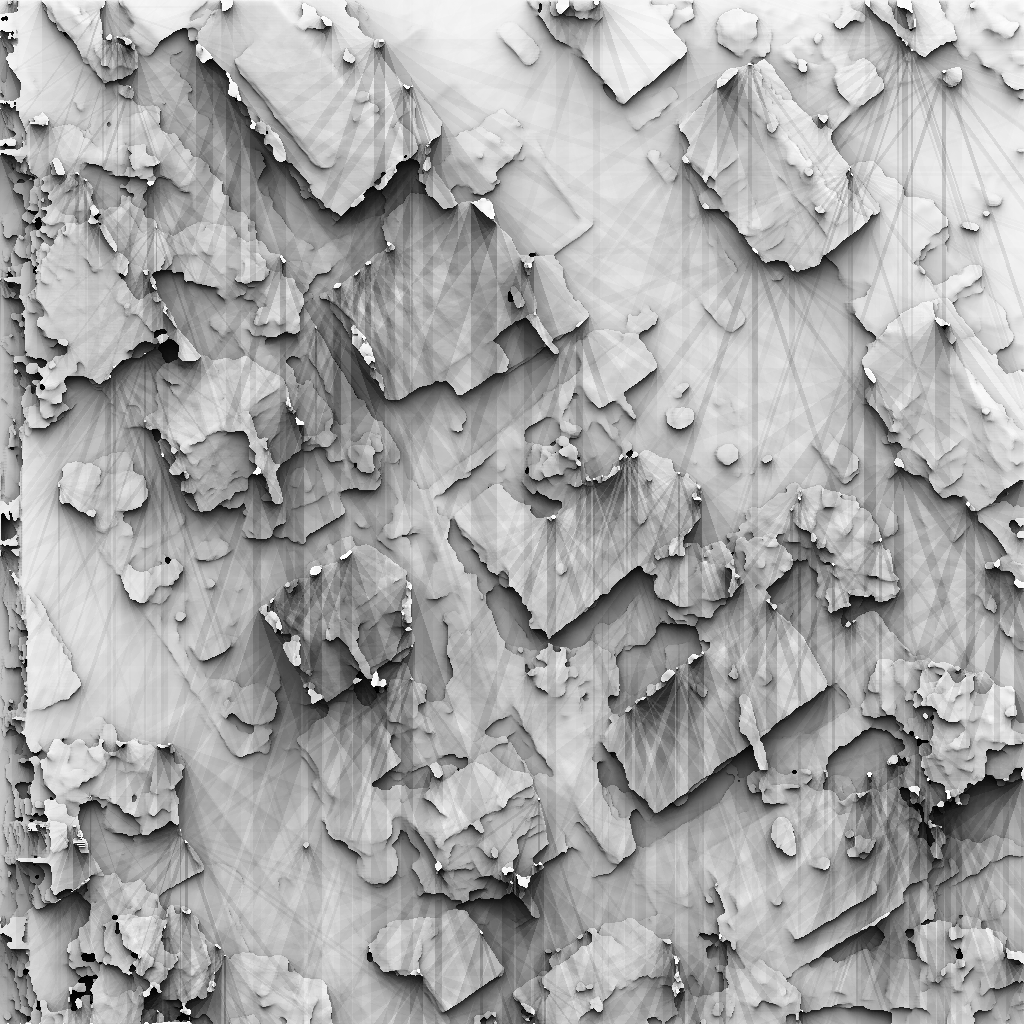}
		\centering{\tiny CBMV(SGM)}
	\end{minipage}
	\begin{minipage}[t]{0.19\textwidth}	
		\includegraphics[width=0.098\linewidth]{figures/color_map.png}
		\includegraphics[width=0.85\linewidth]{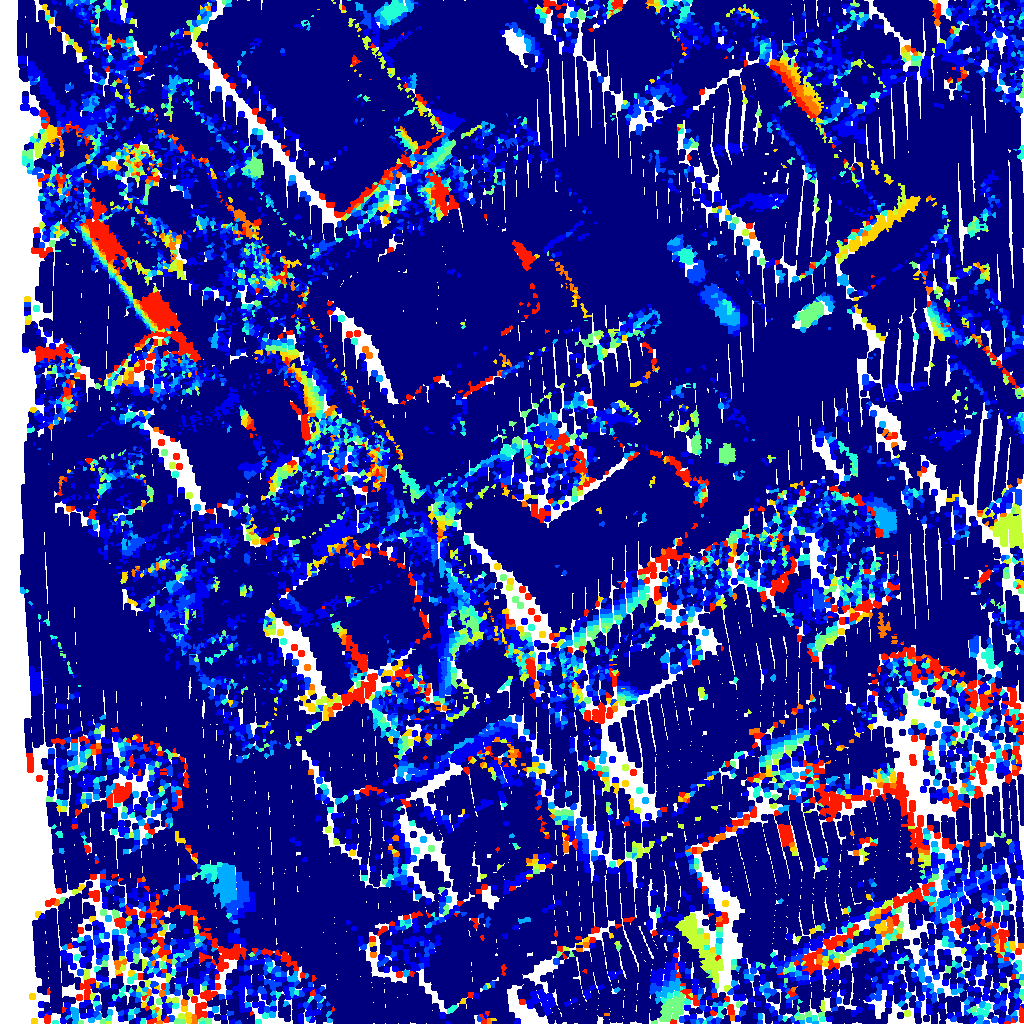}
		\includegraphics[width=\linewidth]{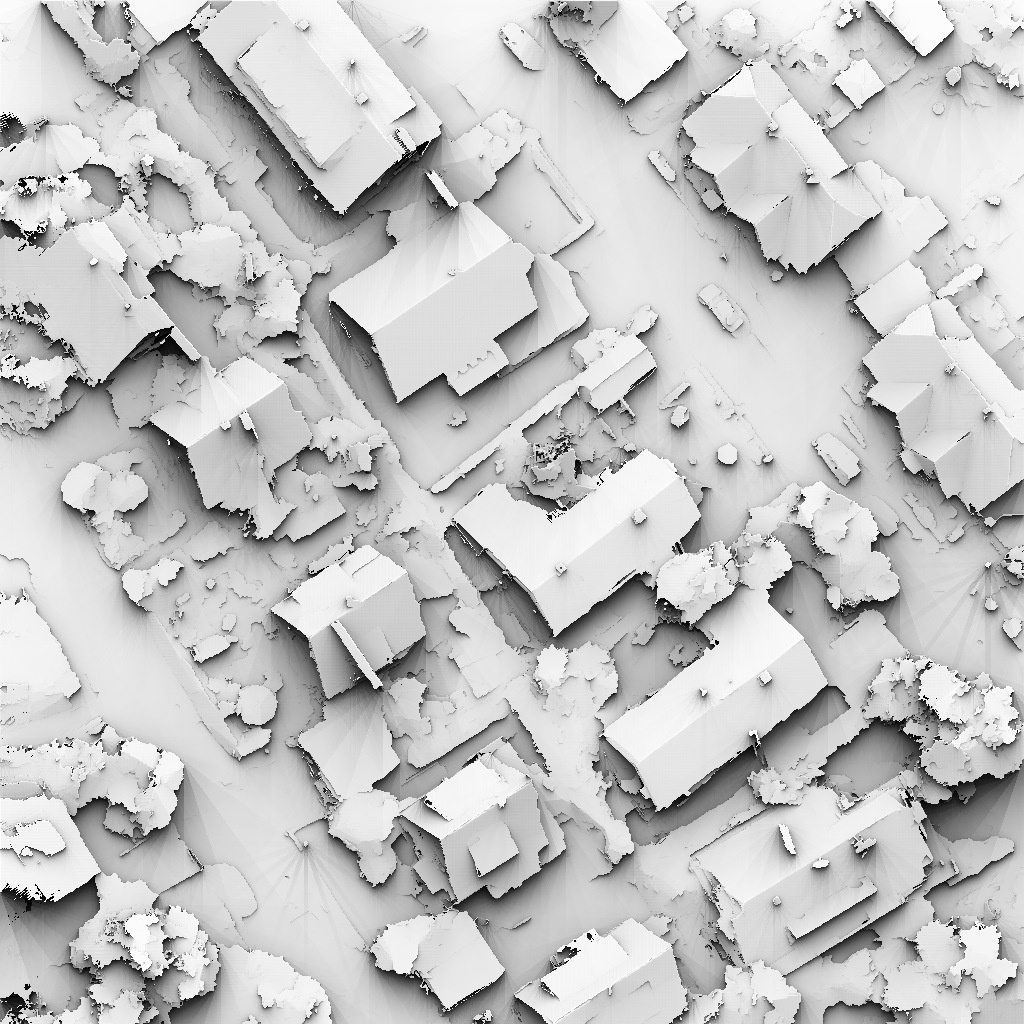}
		\centering{\tiny CBMV(GraphCuts)}
	\end{minipage}
	\begin{minipage}[t]{0.19\textwidth}	
		\includegraphics[width=0.098\linewidth]{figures/color_map.png}
		\includegraphics[width=0.85\linewidth]{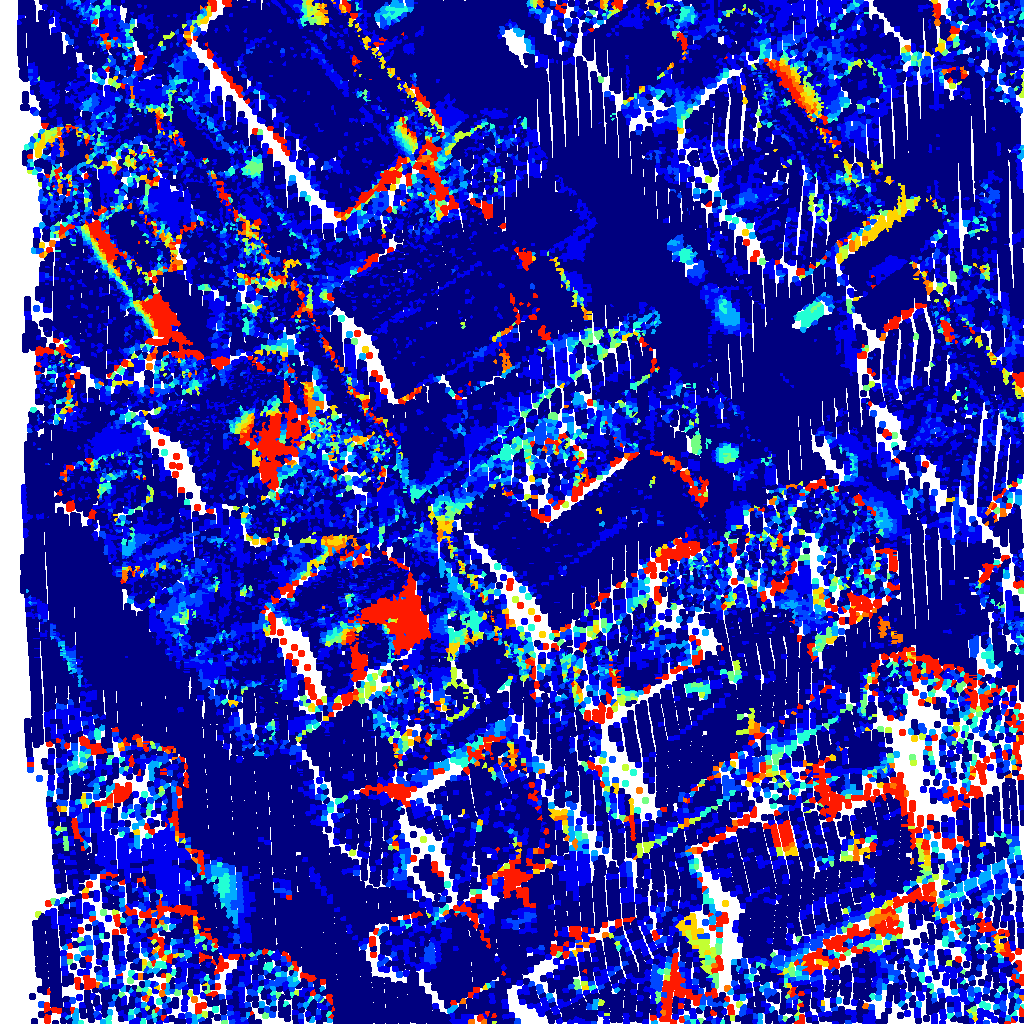}
		\includegraphics[width=\linewidth]{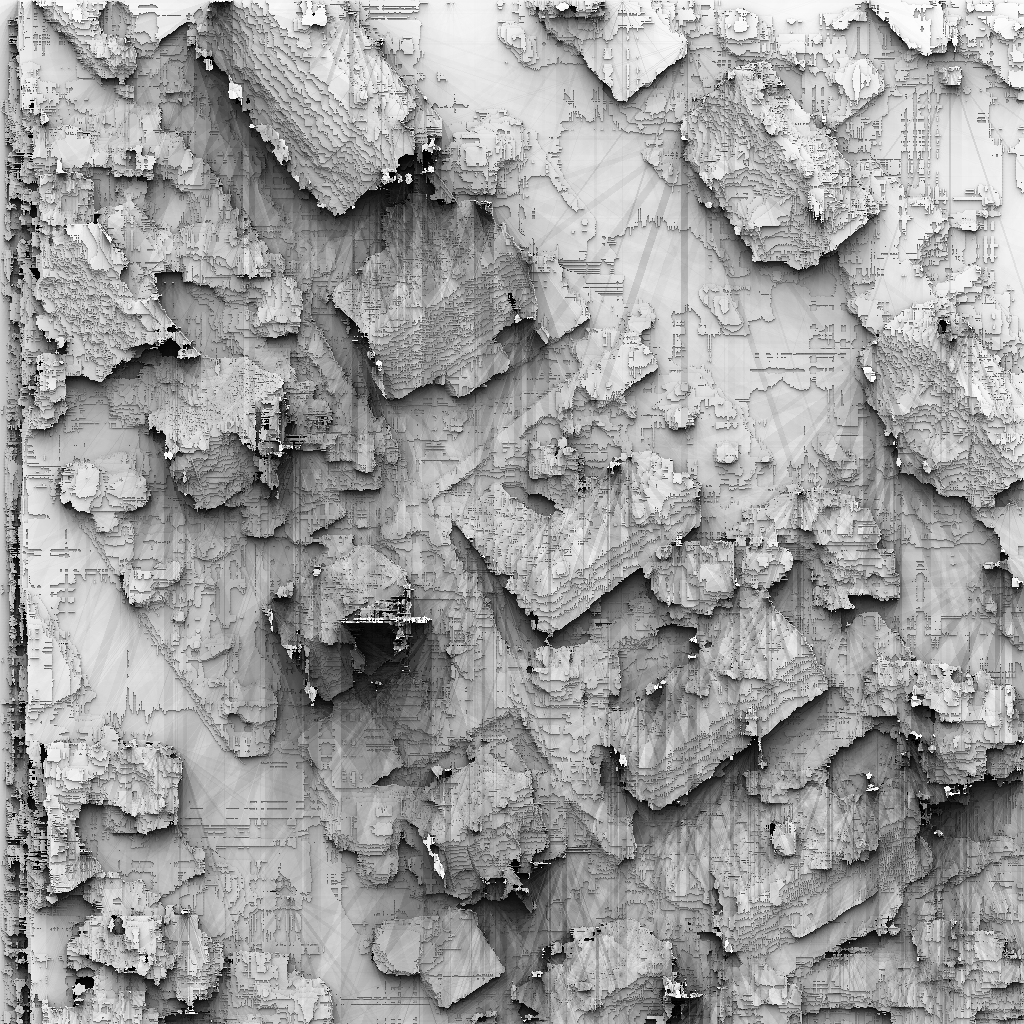}
		\centering{\tiny MC-CNN(KITTI)}
	\end{minipage}
	\begin{minipage}[t]{0.19\textwidth}	
		\includegraphics[width=0.098\linewidth]{figures/color_map.png}
		\includegraphics[width=0.85\linewidth]{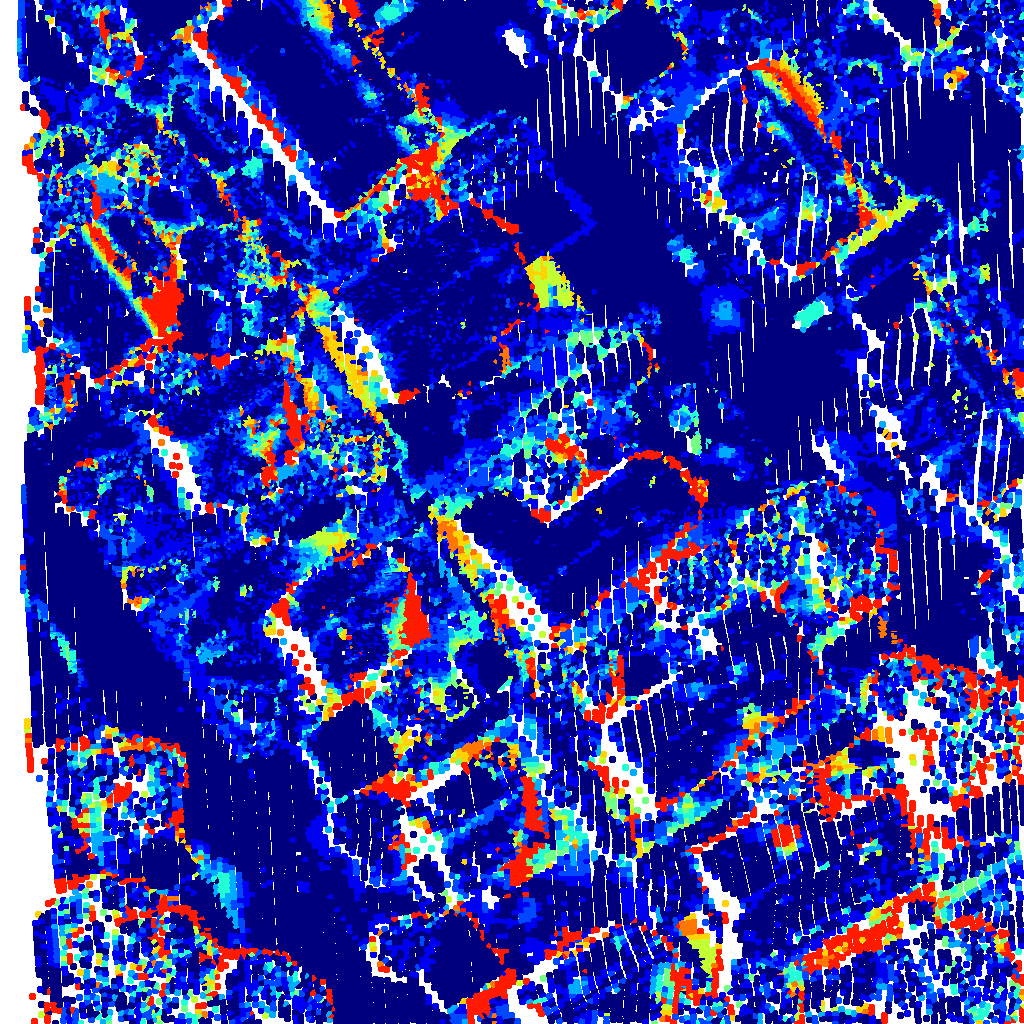}
		\includegraphics[width=\linewidth]{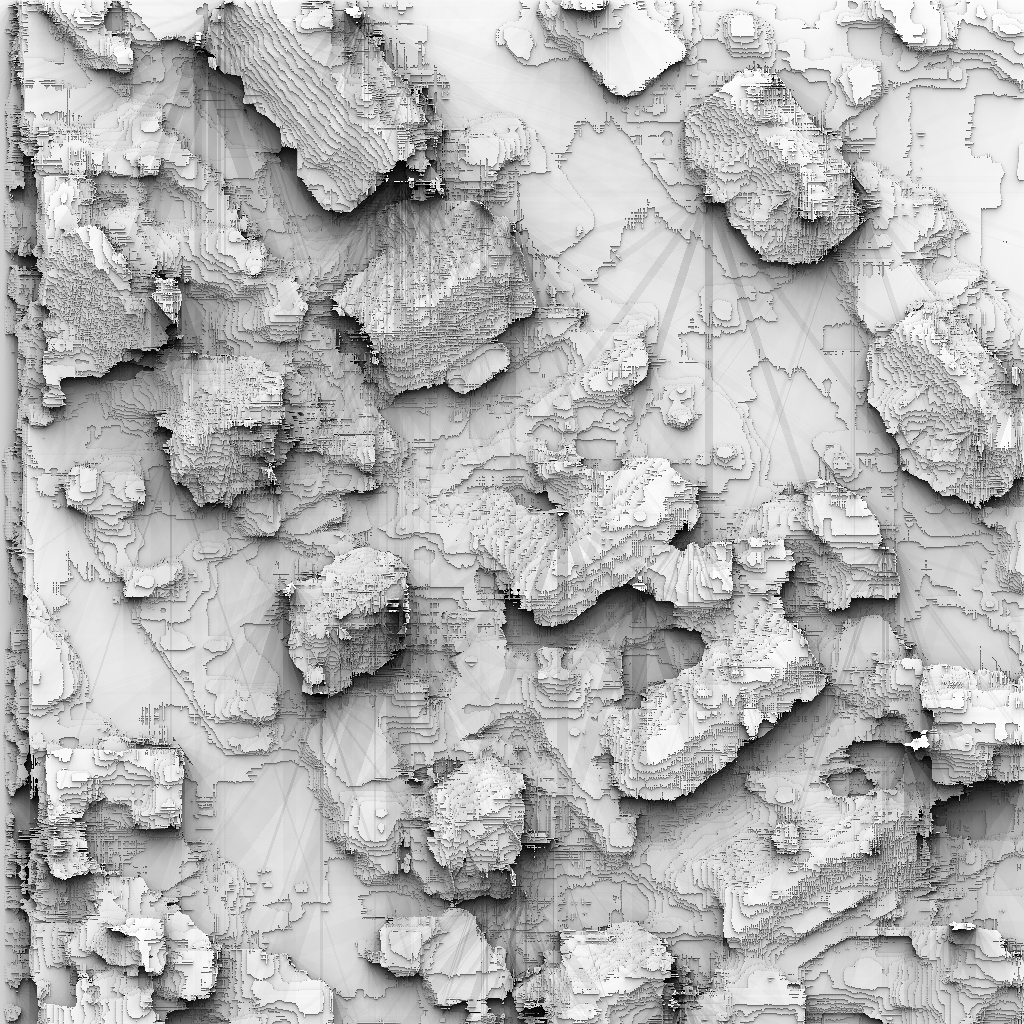}
		\centering{\tiny DeepFeature(KITTI)}
	\end{minipage}
	\begin{minipage}[t]{0.19\textwidth}	
		\includegraphics[width=0.098\linewidth]{figures/color_map.png}
		\includegraphics[width=0.85\linewidth]{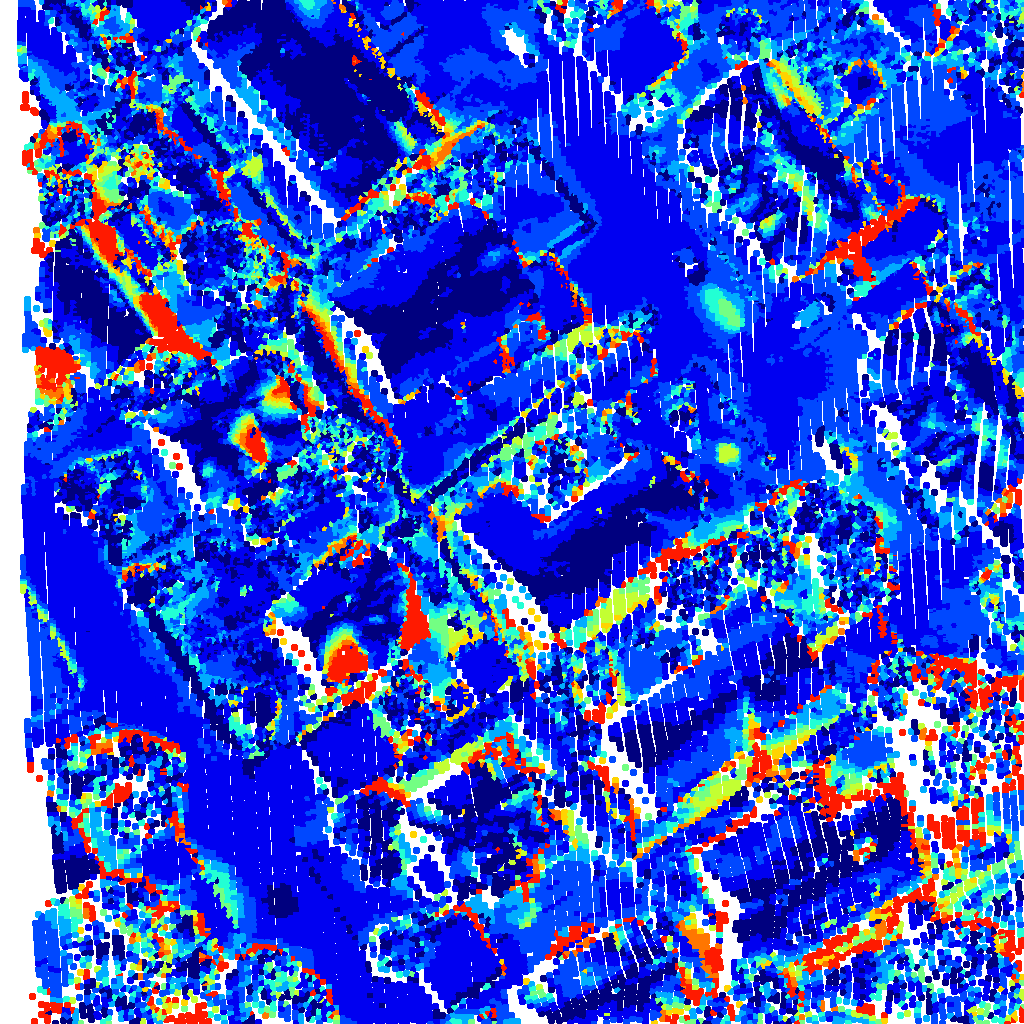}
		\includegraphics[width=\linewidth]{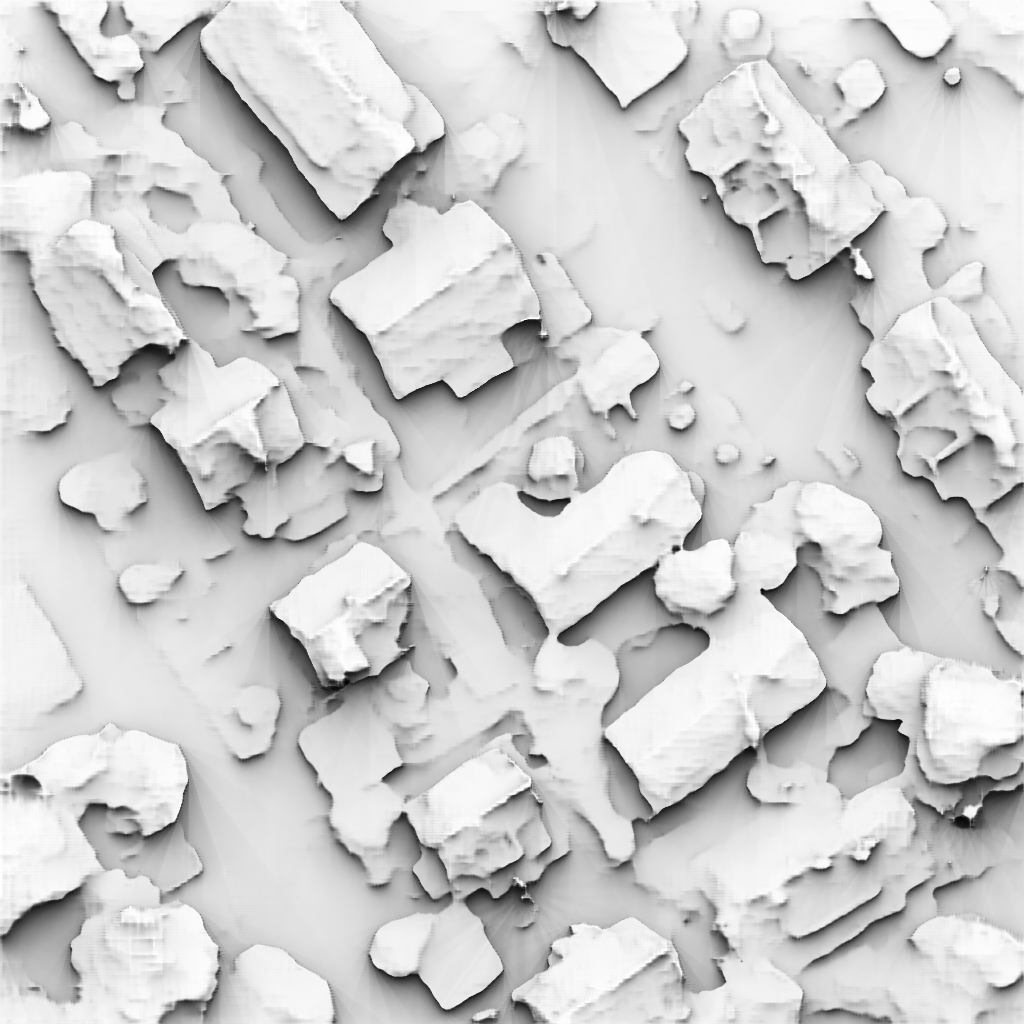}
		\centering{\tiny PSM net(KITTI)}
	\end{minipage}
	\begin{minipage}[t]{0.19\textwidth}	
		\includegraphics[width=0.098\linewidth]{figures/color_map.png}
		\includegraphics[width=0.85\linewidth]{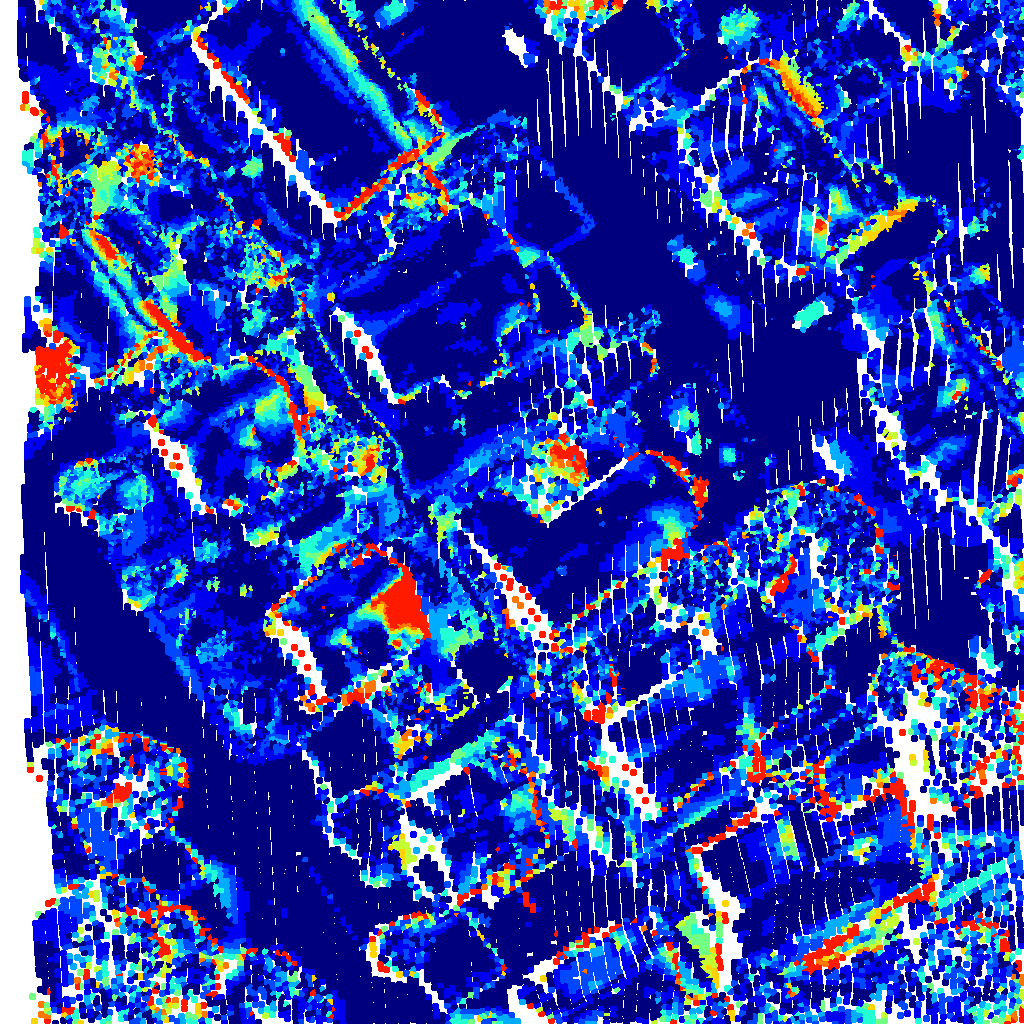}
		\includegraphics[width=\linewidth]{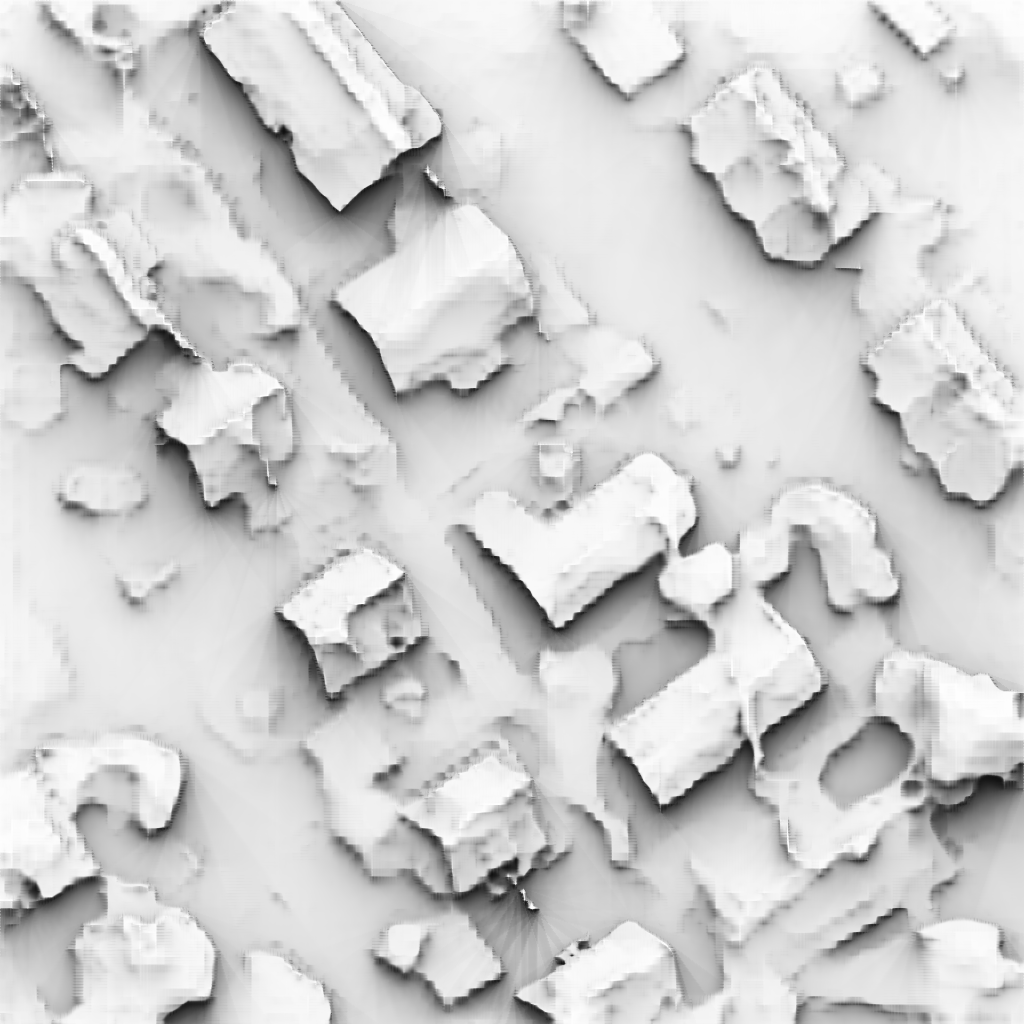}
		\centering{\tiny HRS net(KITTI)}
	\end{minipage}
	\begin{minipage}[t]{0.19\textwidth}	
		\includegraphics[width=0.098\linewidth]{figures/color_map.png}
		\includegraphics[width=0.85\linewidth]{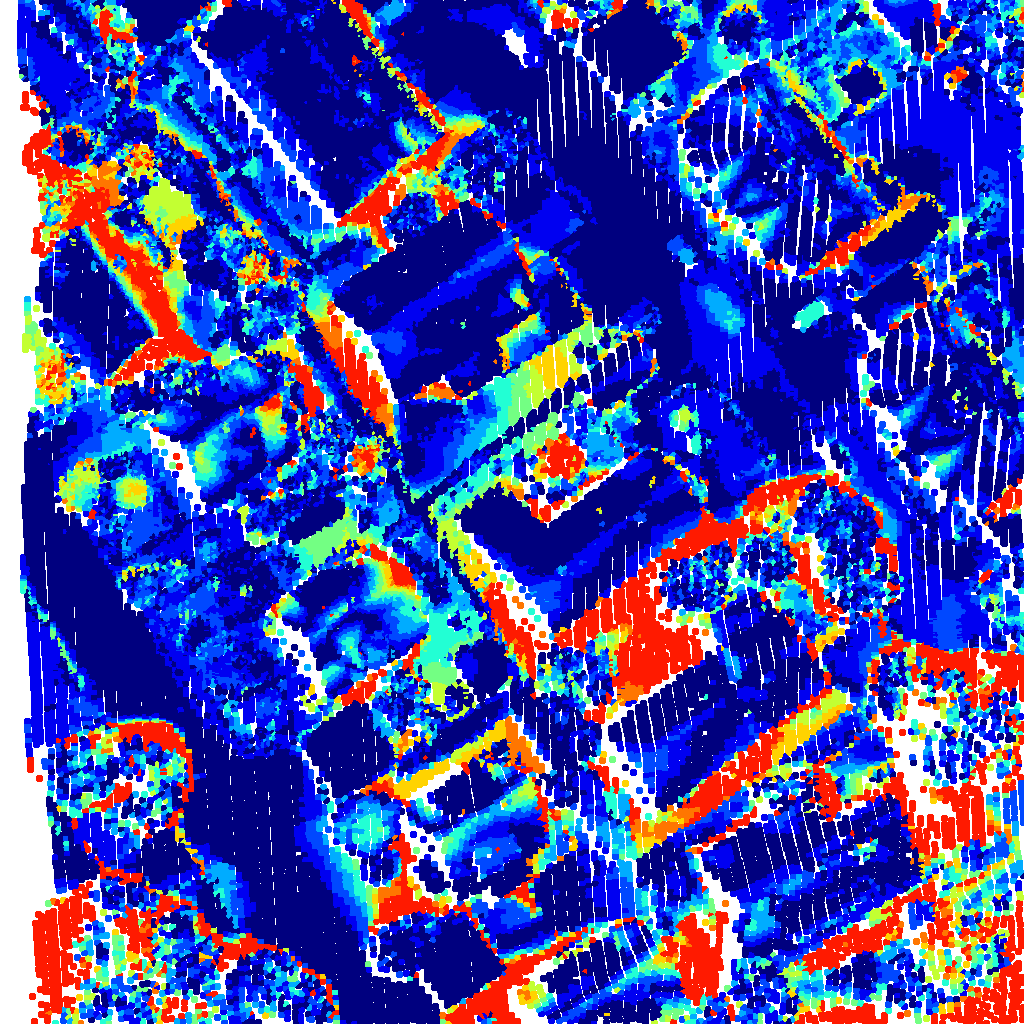}
		\includegraphics[width=\linewidth]{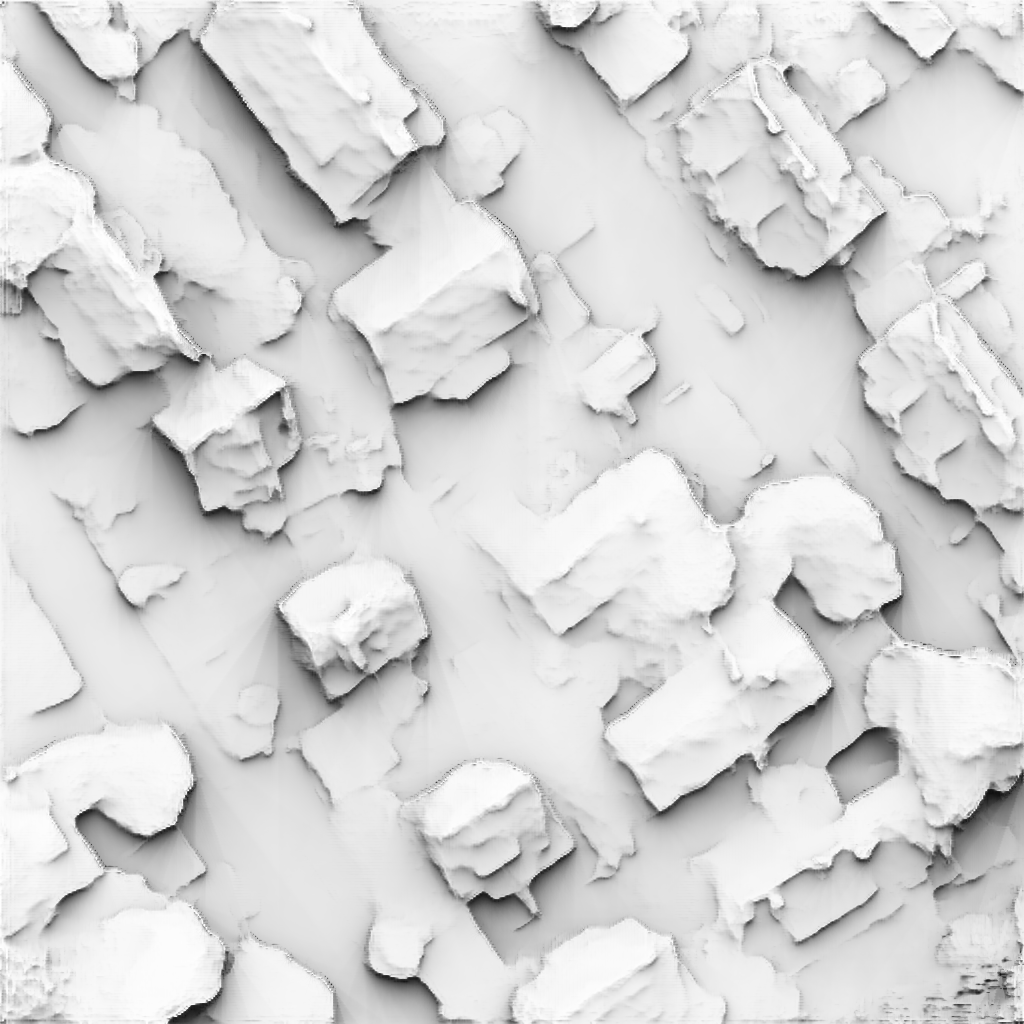}
		\centering{\tiny DeepPruner(KITTI)}
	\end{minipage}
	\begin{minipage}[t]{0.19\textwidth}
		\includegraphics[width=0.098\linewidth]{figures/color_map.png}
		\includegraphics[width=0.85\linewidth]{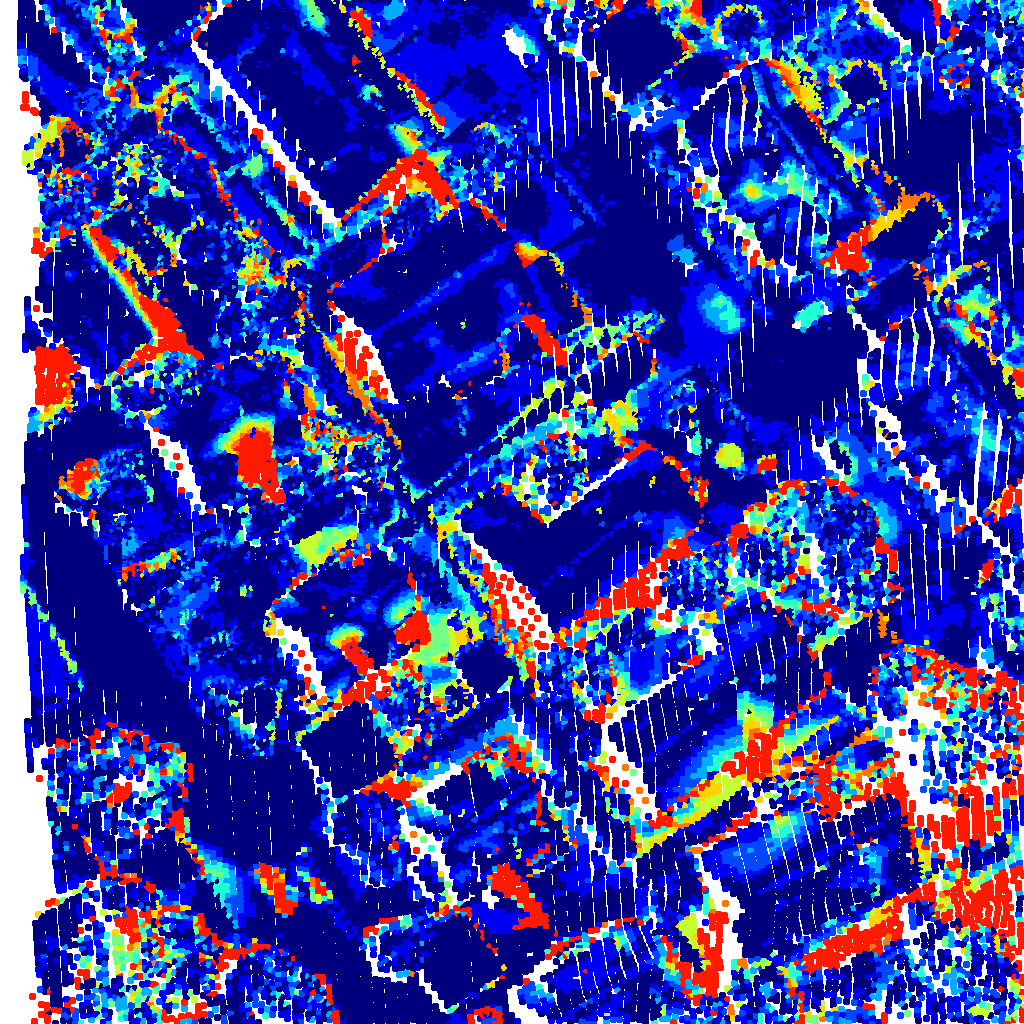}
		\includegraphics[width=\linewidth]{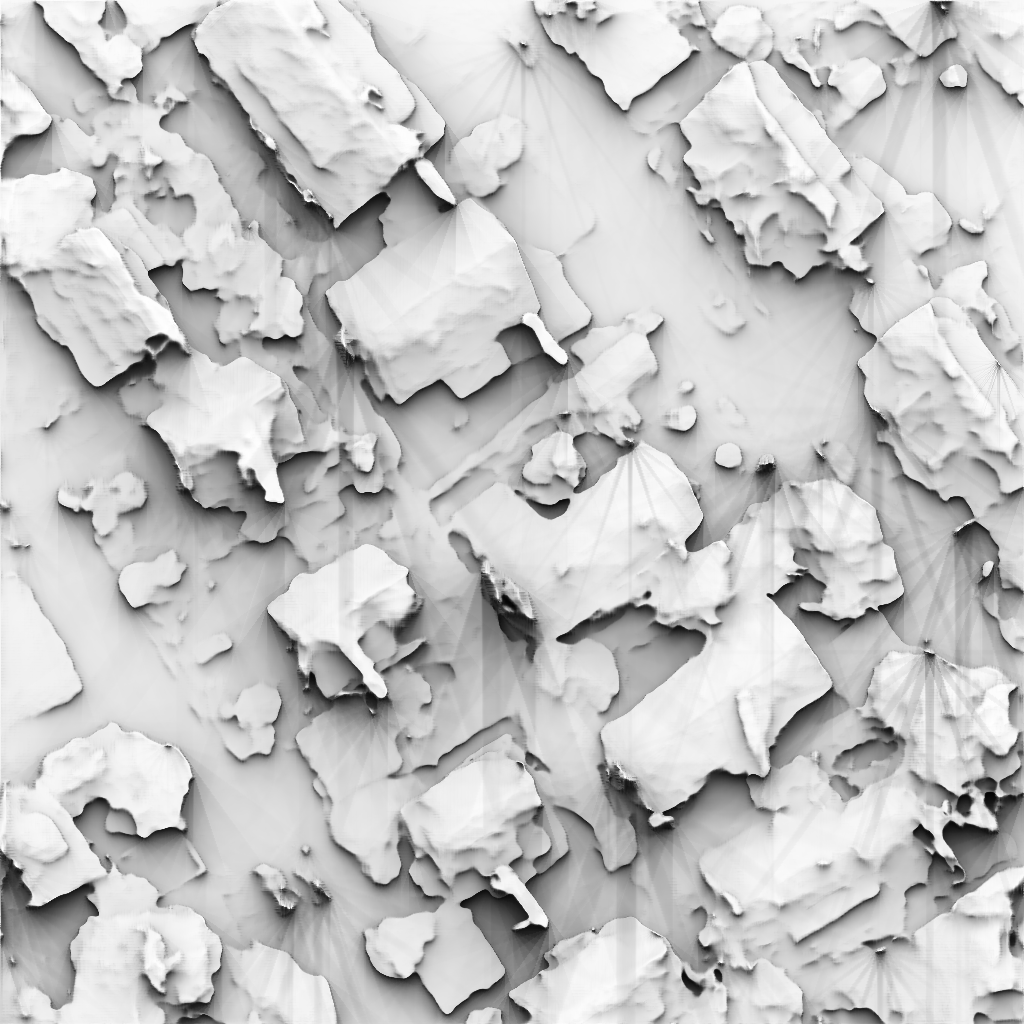}
		\centering{\tiny GANet(KITTI)}
	\end{minipage}
	\begin{minipage}[t]{0.19\textwidth}	
		\includegraphics[width=0.098\linewidth]{figures/color_map.png}
		\includegraphics[width=0.85\linewidth]{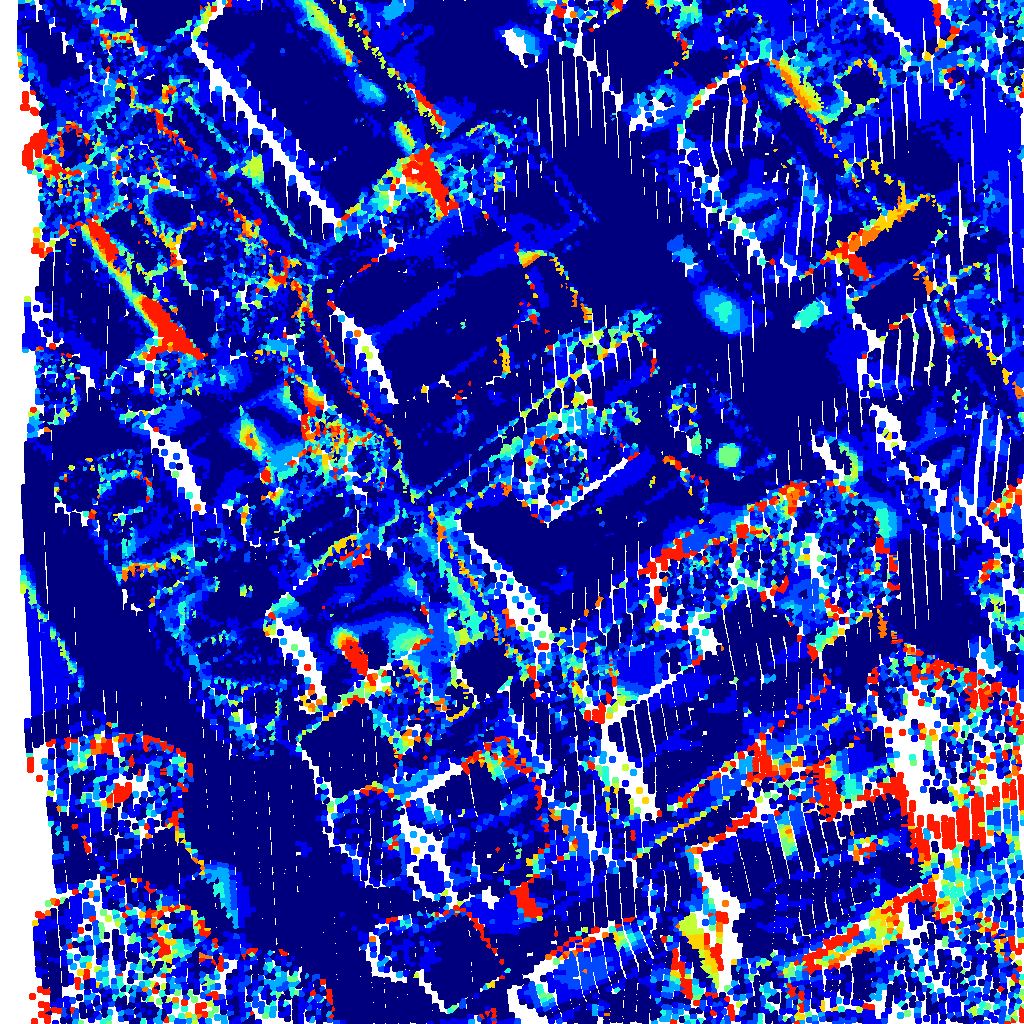}
		\includegraphics[width=\linewidth]{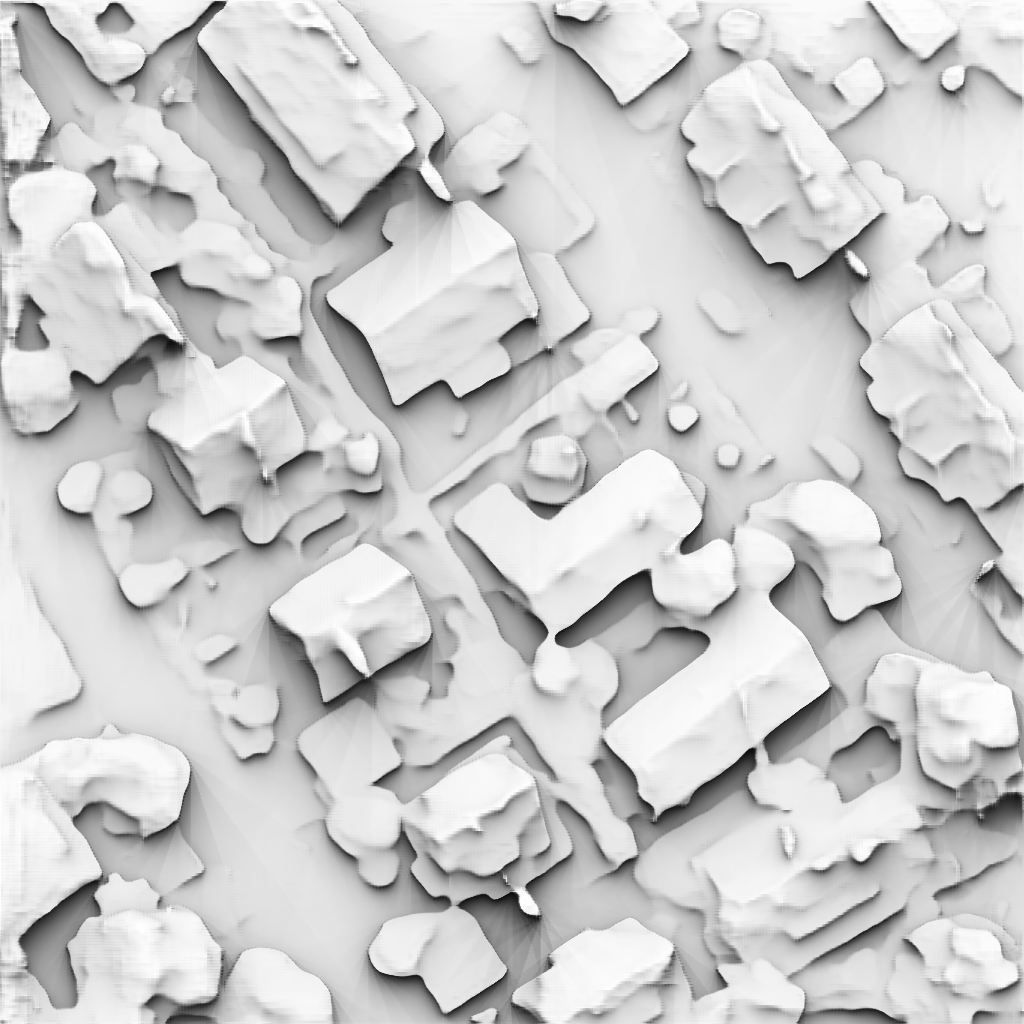}
		\centering{\tiny LEAStereo(KITTI)}
	\end{minipage}
	\begin{minipage}[t]{0.19\textwidth}	
		\includegraphics[width=0.098\linewidth]{figures/color_map.png}
		\includegraphics[width=0.85\linewidth]{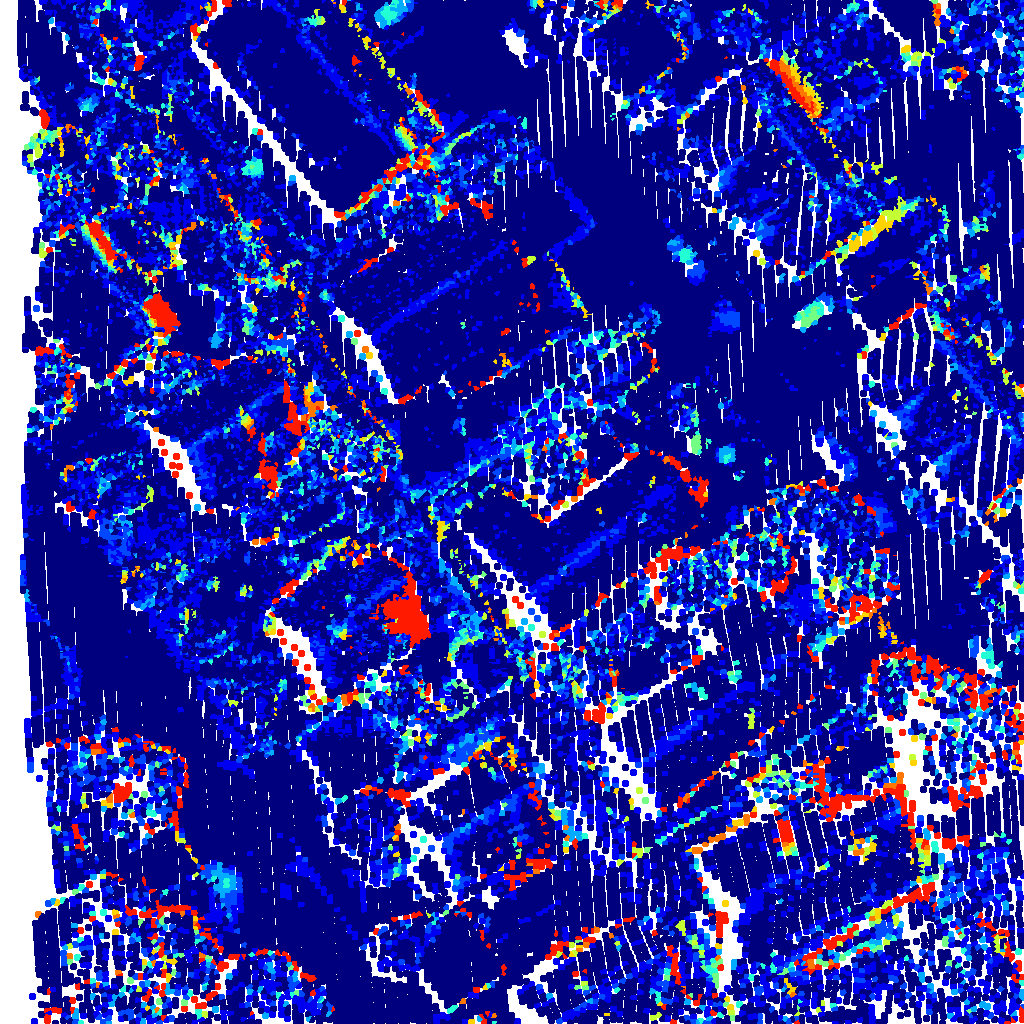}
		\includegraphics[width=\linewidth]{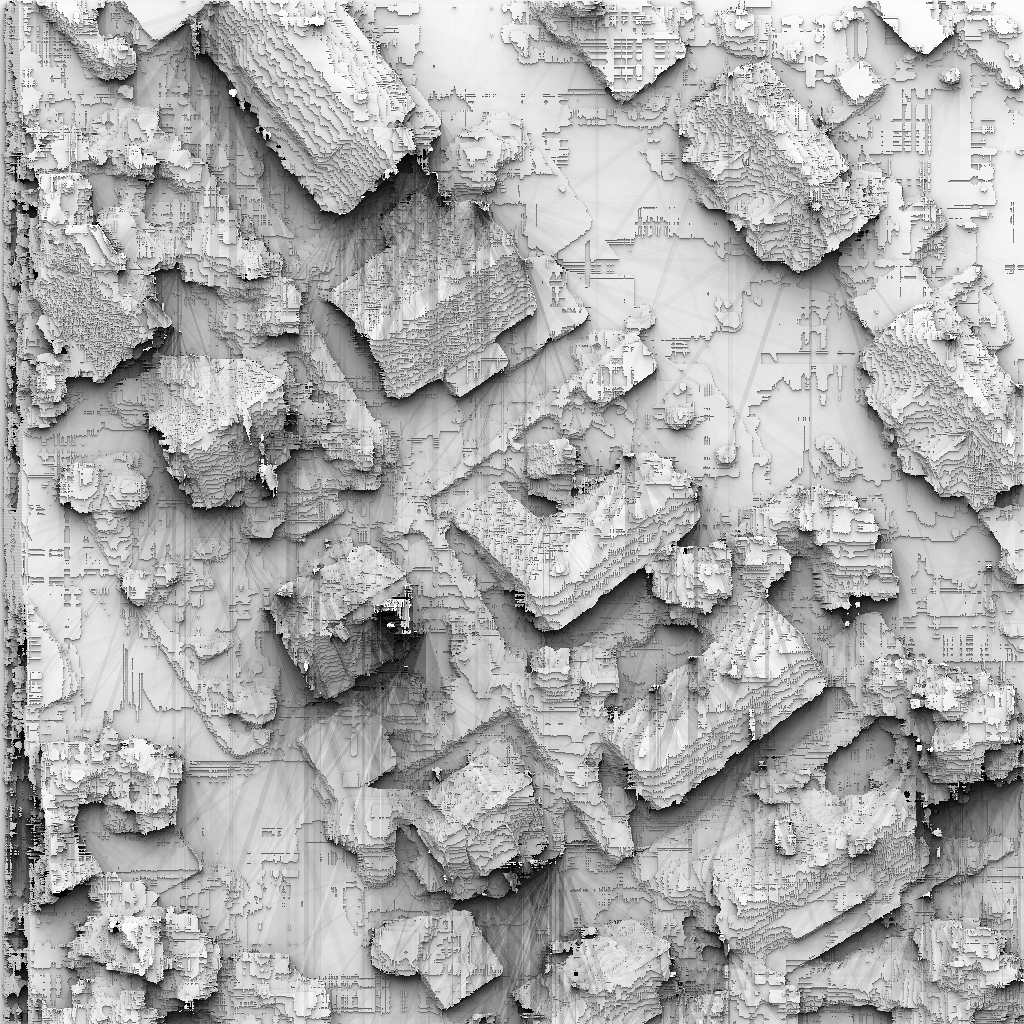}
		\centering{\tiny MC-CNN}
	\end{minipage}
	\begin{minipage}[t]{0.19\textwidth}	
		\includegraphics[width=0.098\linewidth]{figures/color_map.png}
		\includegraphics[width=0.85\linewidth]{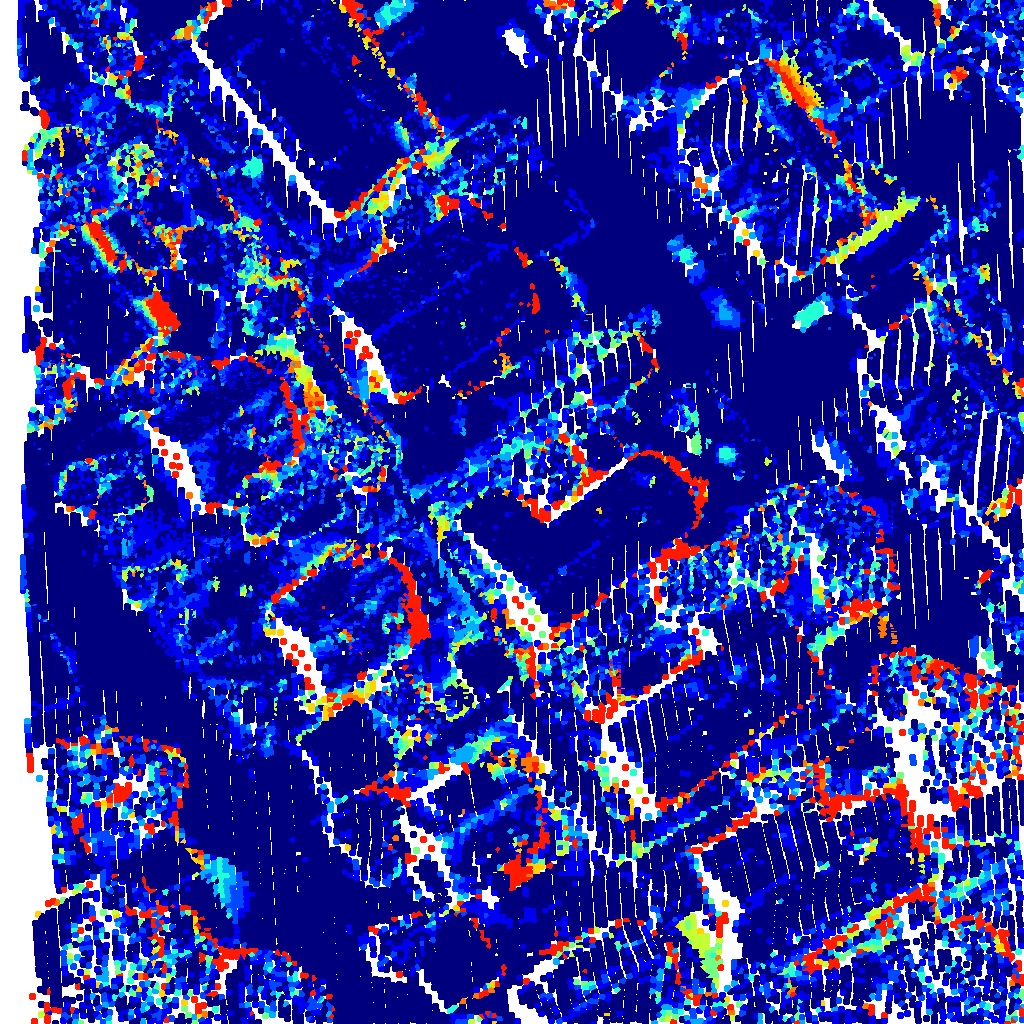}
		\includegraphics[width=\linewidth]{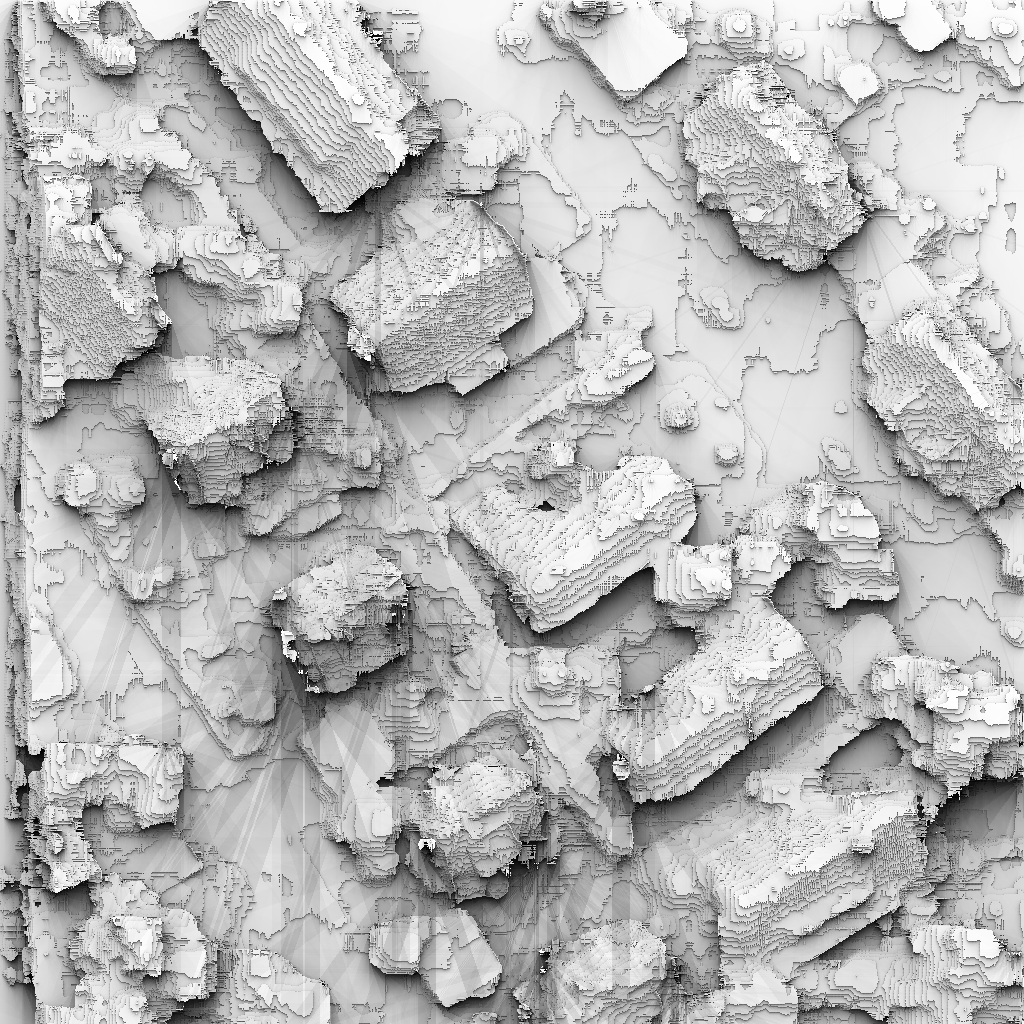}
		\centering{\tiny DeepFeature}
	\end{minipage}
	\begin{minipage}[t]{0.19\textwidth}	
		\includegraphics[width=0.098\linewidth]{figures/color_map.png}
		\includegraphics[width=0.85\linewidth]{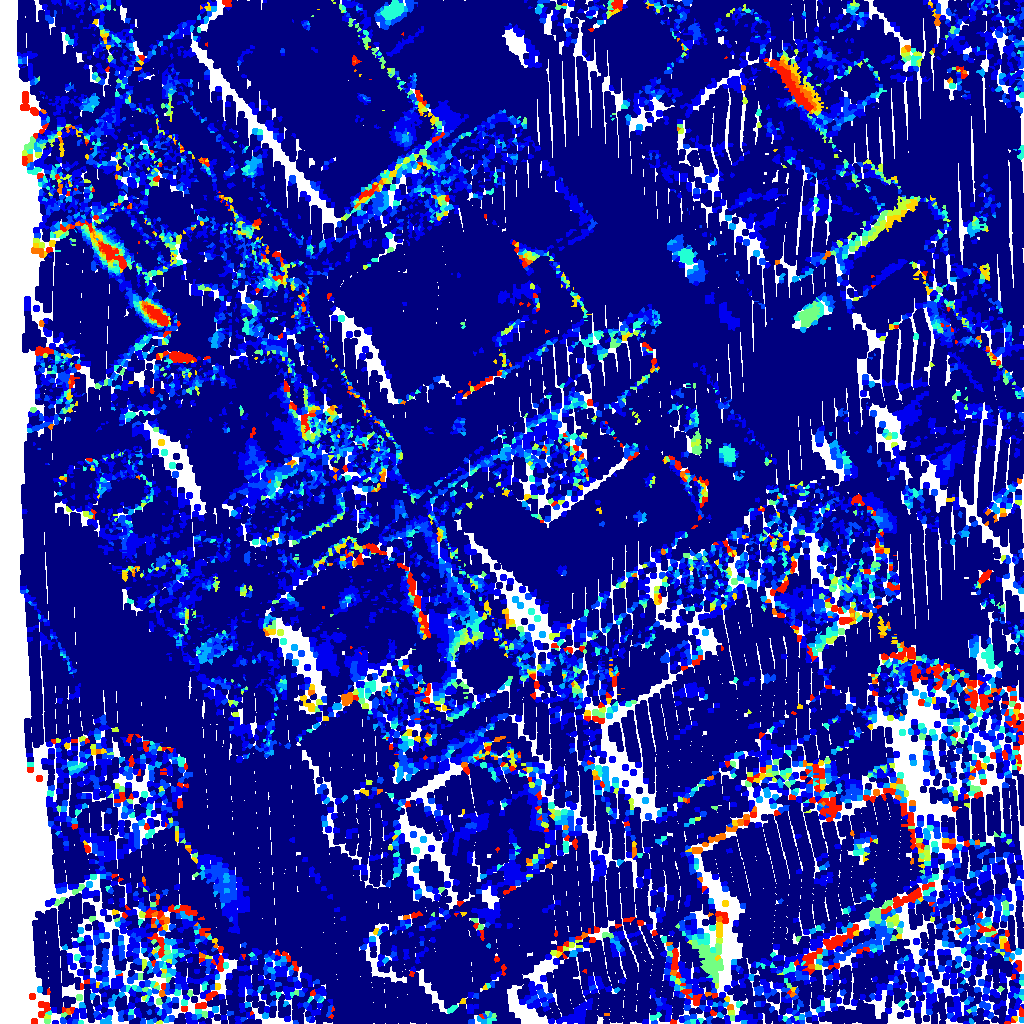}
		\includegraphics[width=\linewidth]{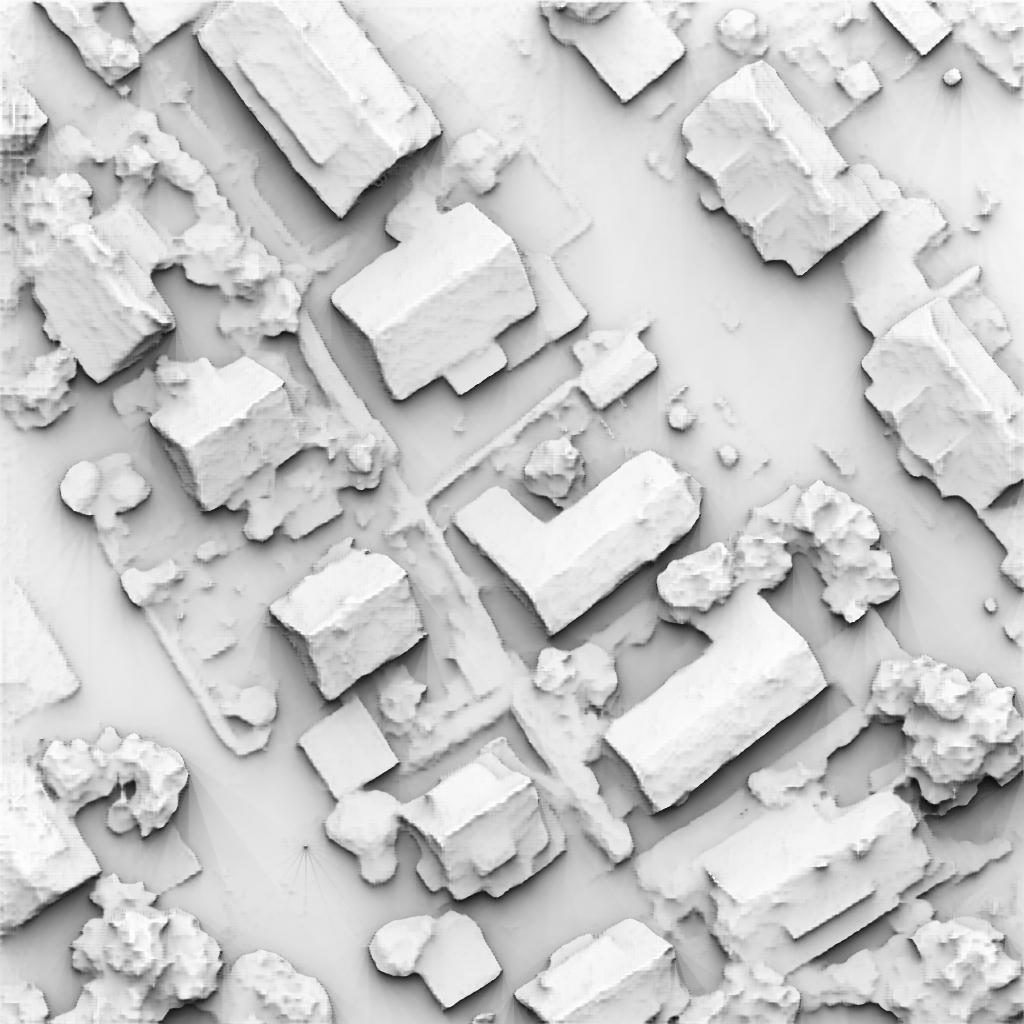}
		\centering{\tiny PSM net}
	\end{minipage}
	\begin{minipage}[t]{0.19\textwidth}	
		\includegraphics[width=0.098\linewidth]{figures/color_map.png}
		\includegraphics[width=0.85\linewidth]{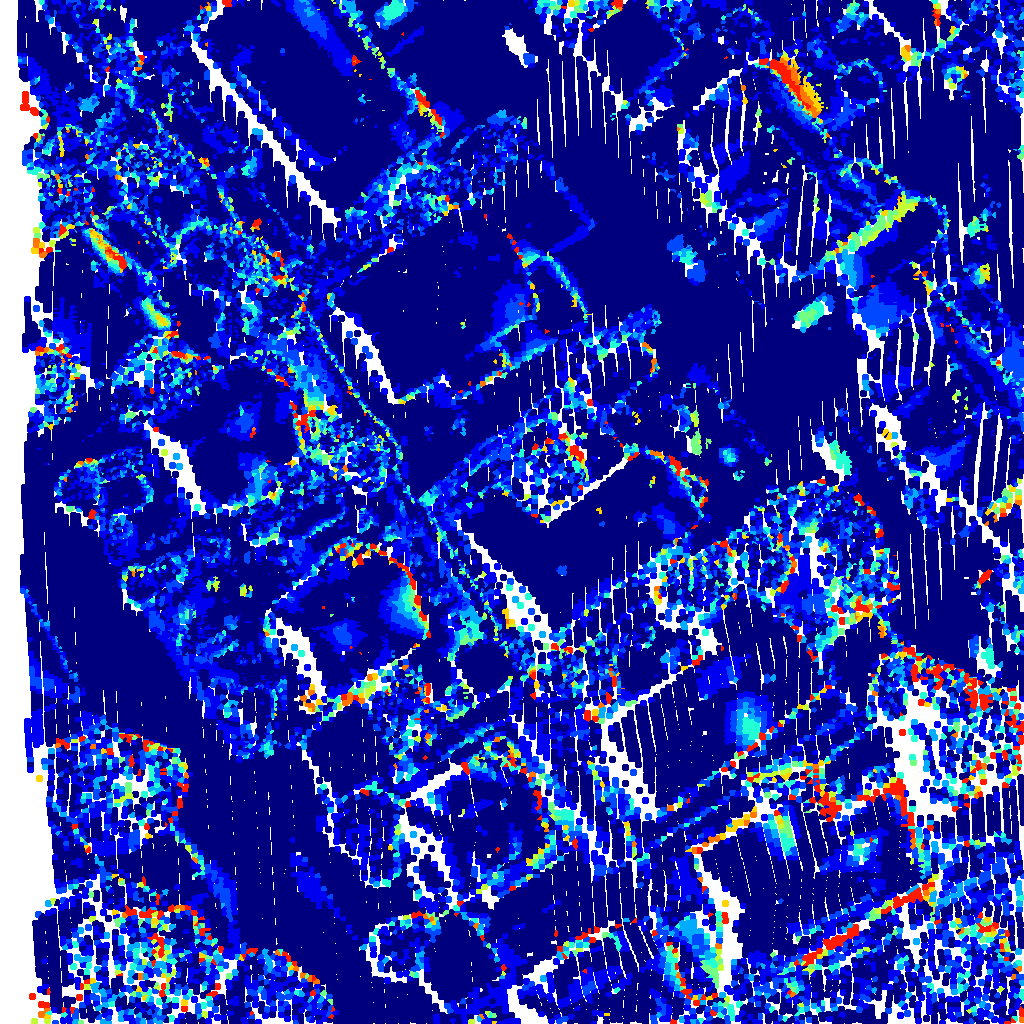}
		\includegraphics[width=\linewidth]{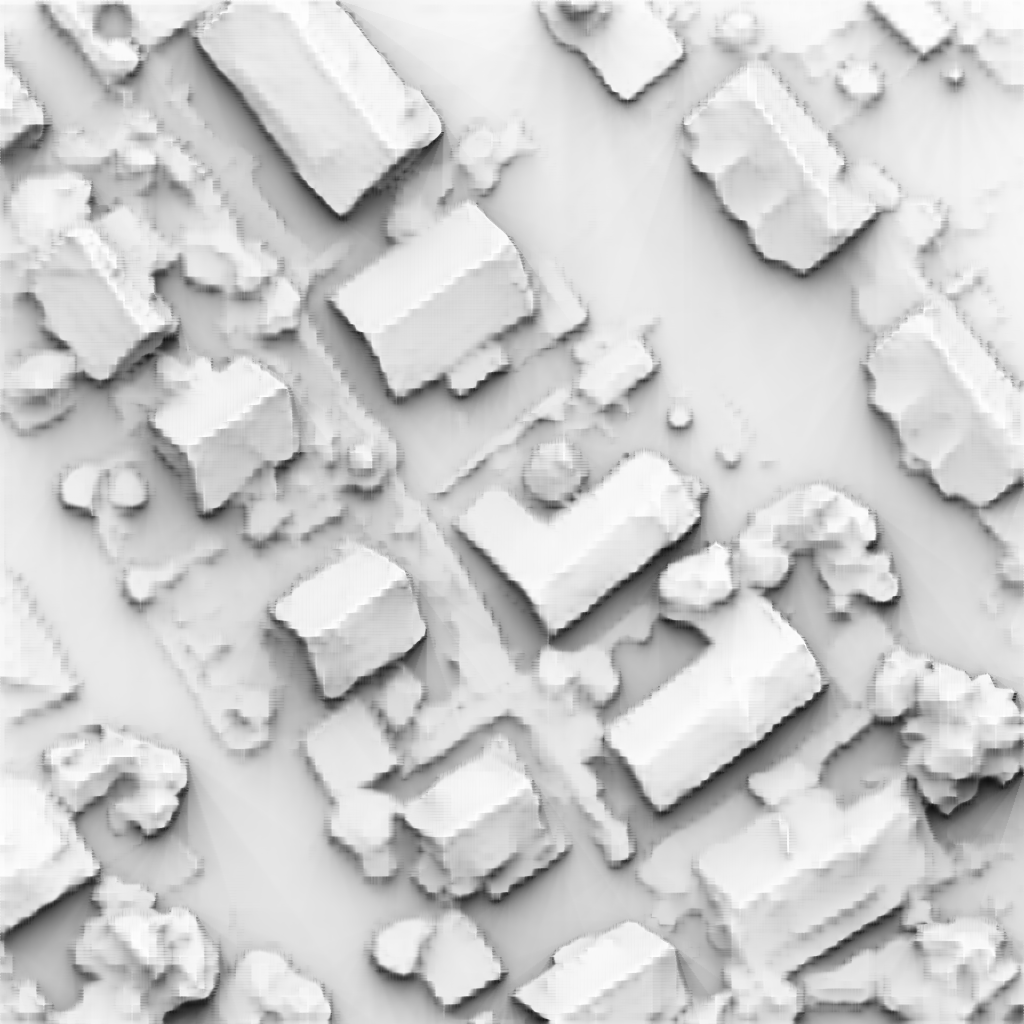}
		\centering{\tiny HRS net}
	\end{minipage}
	\begin{minipage}[t]{0.19\textwidth}	
		\includegraphics[width=0.098\linewidth]{figures/color_map.png}
		\includegraphics[width=0.85\linewidth]{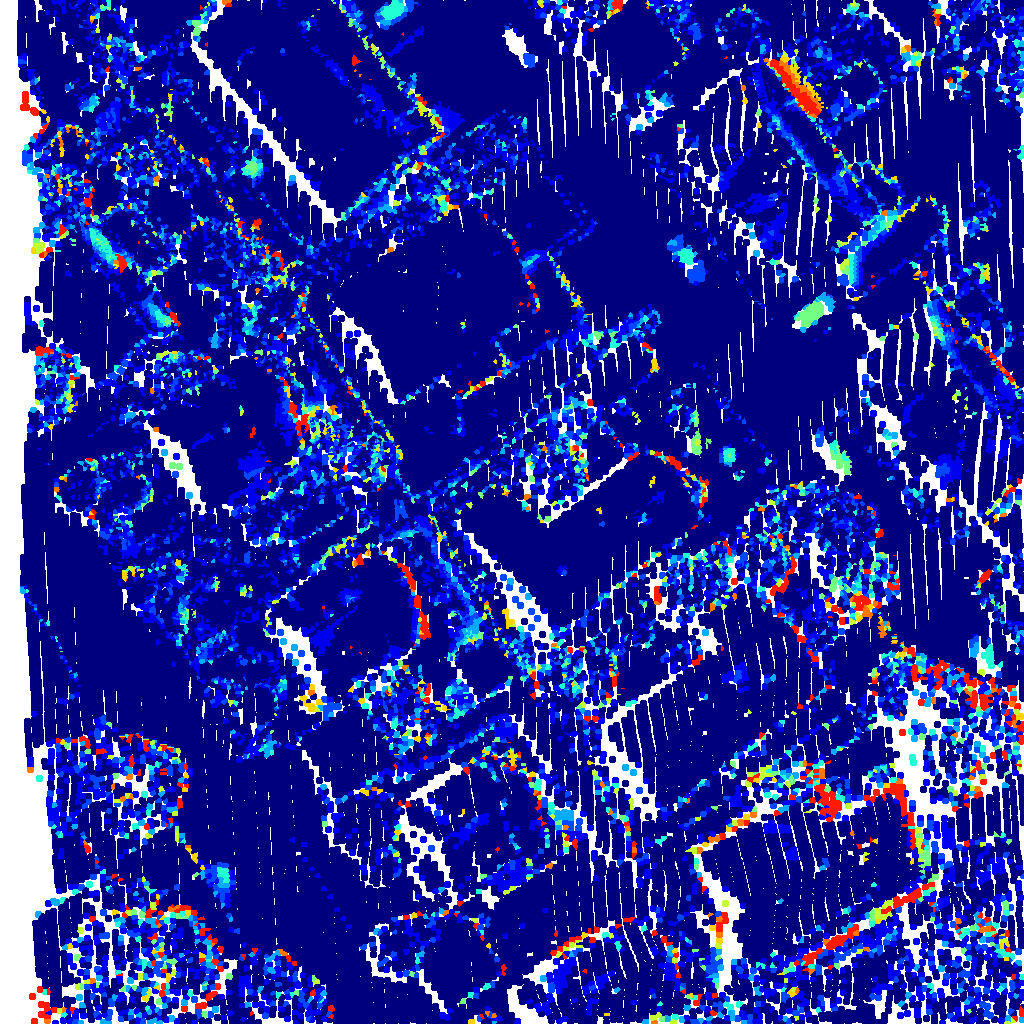}
		\includegraphics[width=\linewidth]{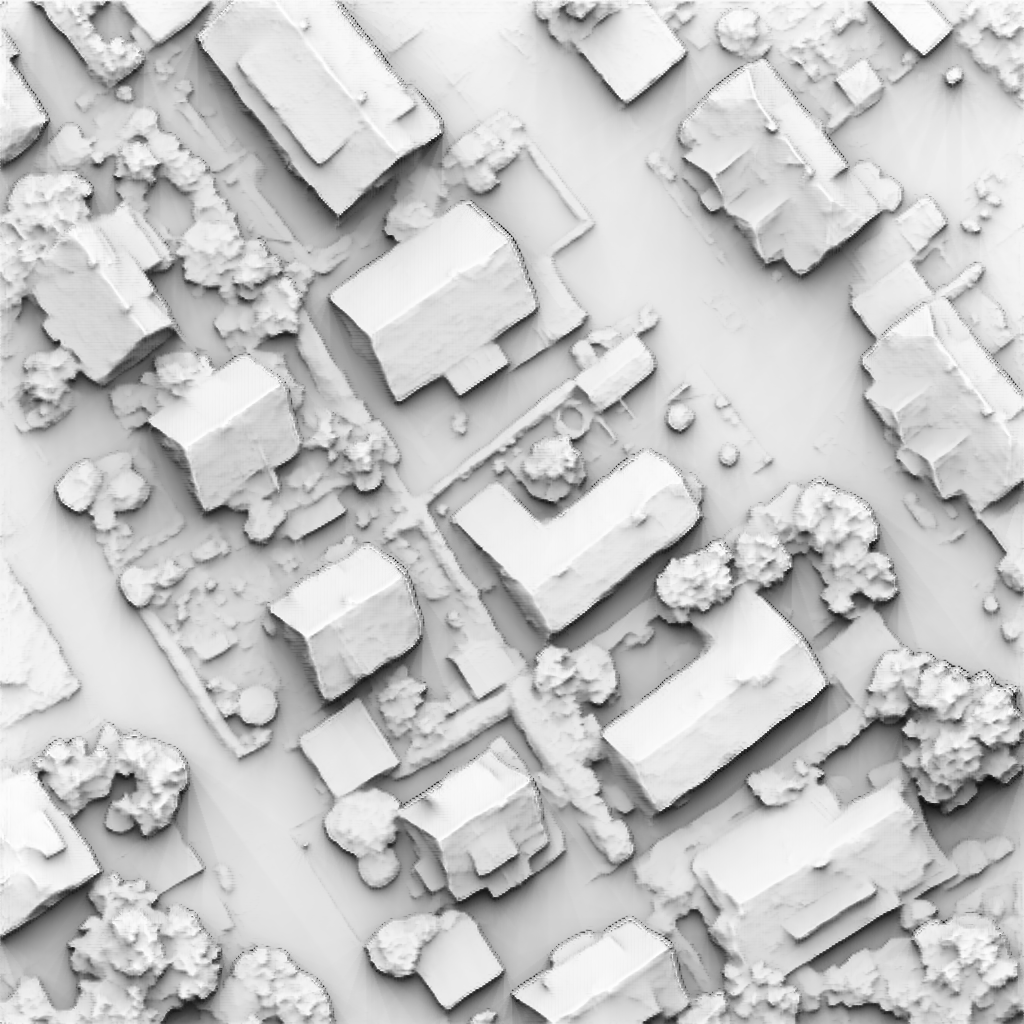}
		\centering{\tiny DeepPruner}
	\end{minipage}
	\begin{minipage}[t]{0.19\textwidth}
		\includegraphics[width=0.098\linewidth]{figures/color_map.png}
		\includegraphics[width=0.85\linewidth]{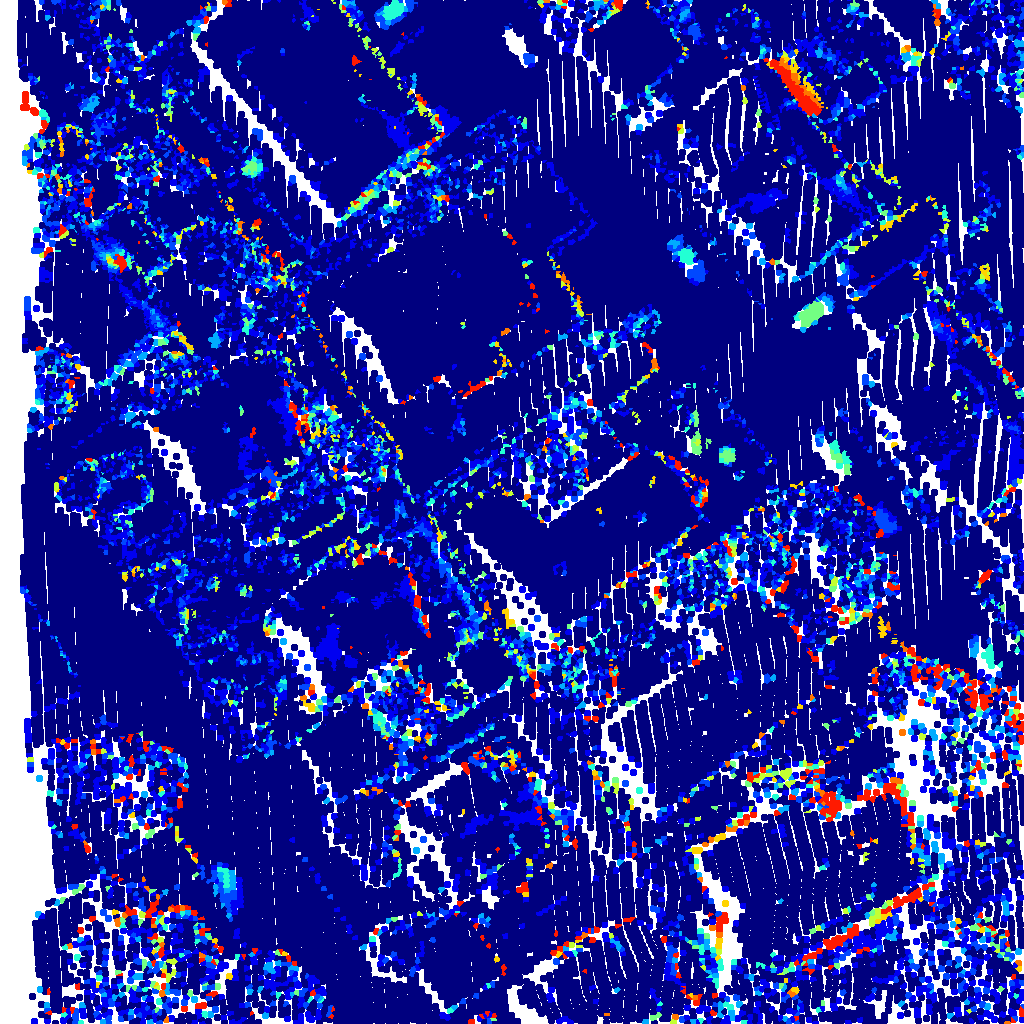}
		\includegraphics[width=\linewidth]{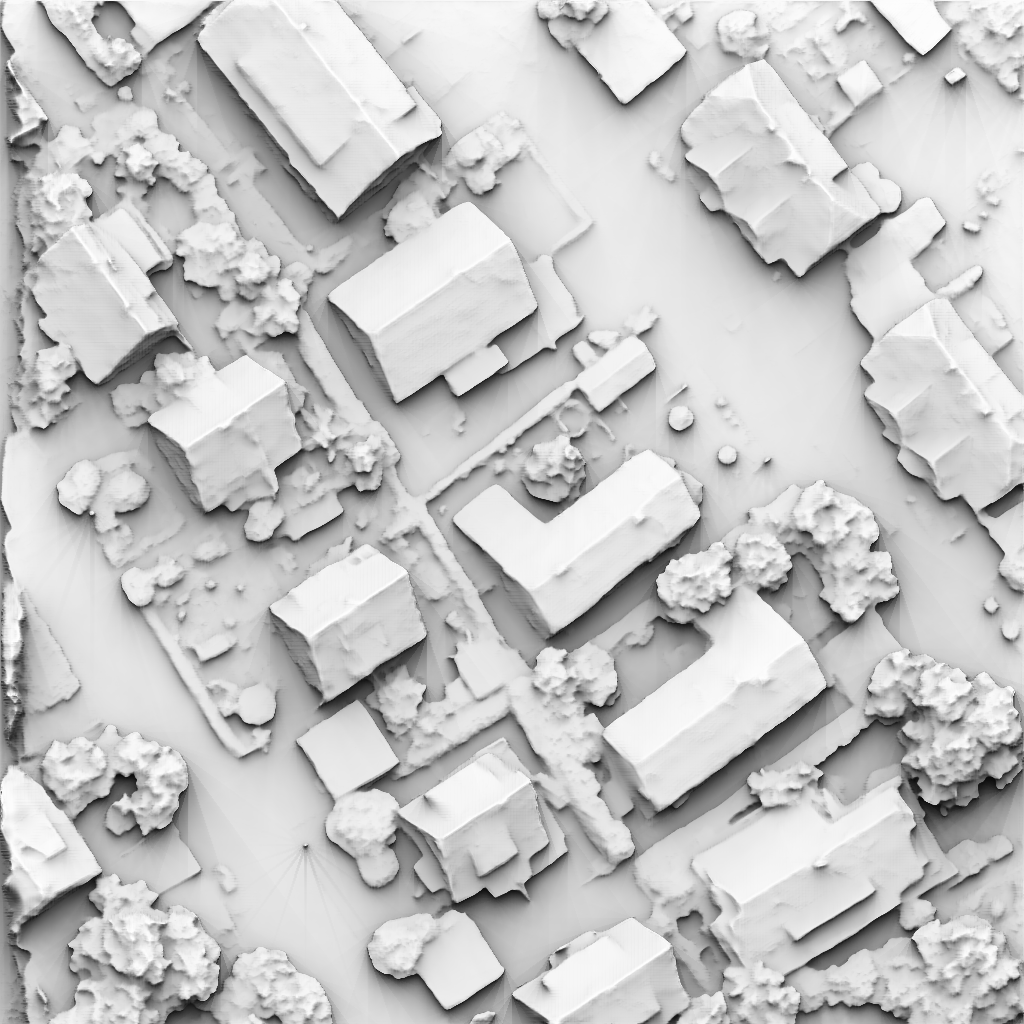}
		\centering{\tiny GANet}
	\end{minipage}
	\begin{minipage}[t]{0.19\textwidth}	
		\includegraphics[width=0.098\linewidth]{figures/color_map.png}
		\includegraphics[width=0.85\linewidth]{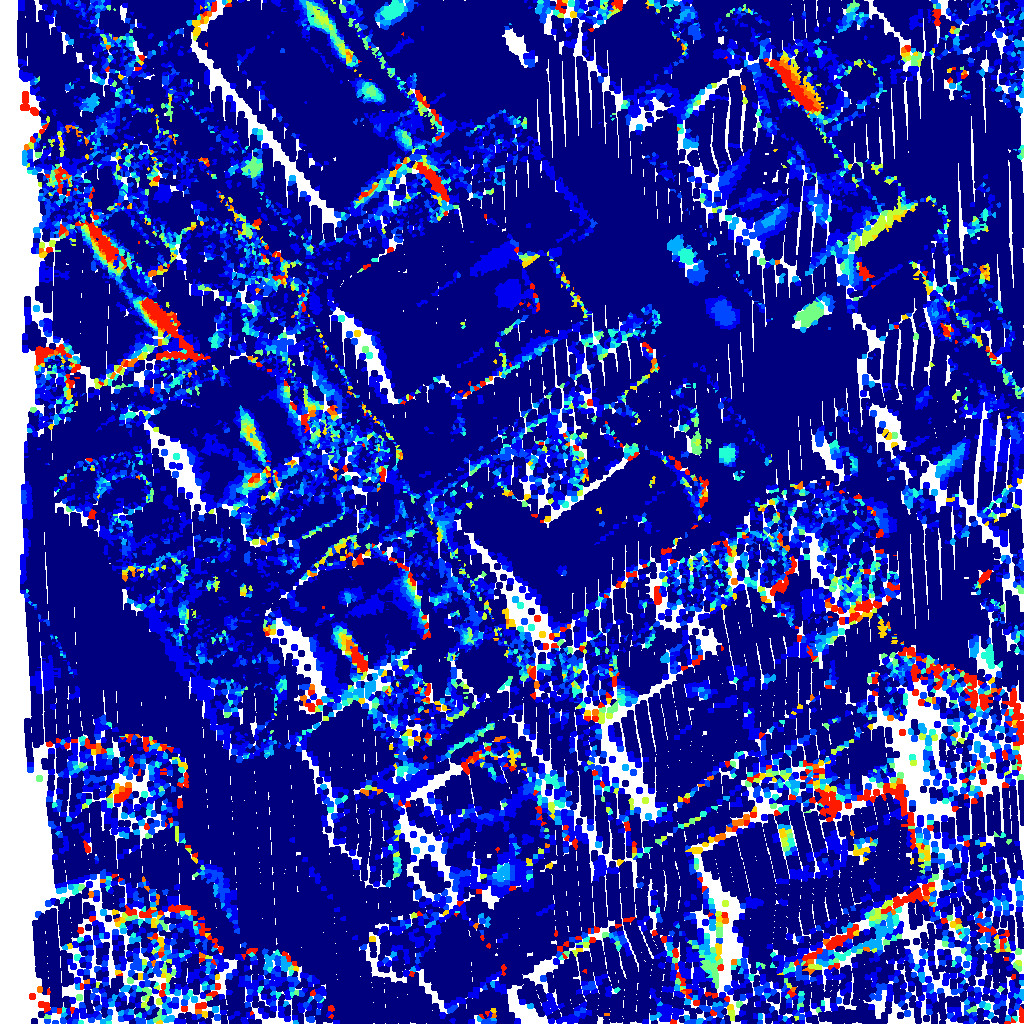}
		\includegraphics[width=\linewidth]{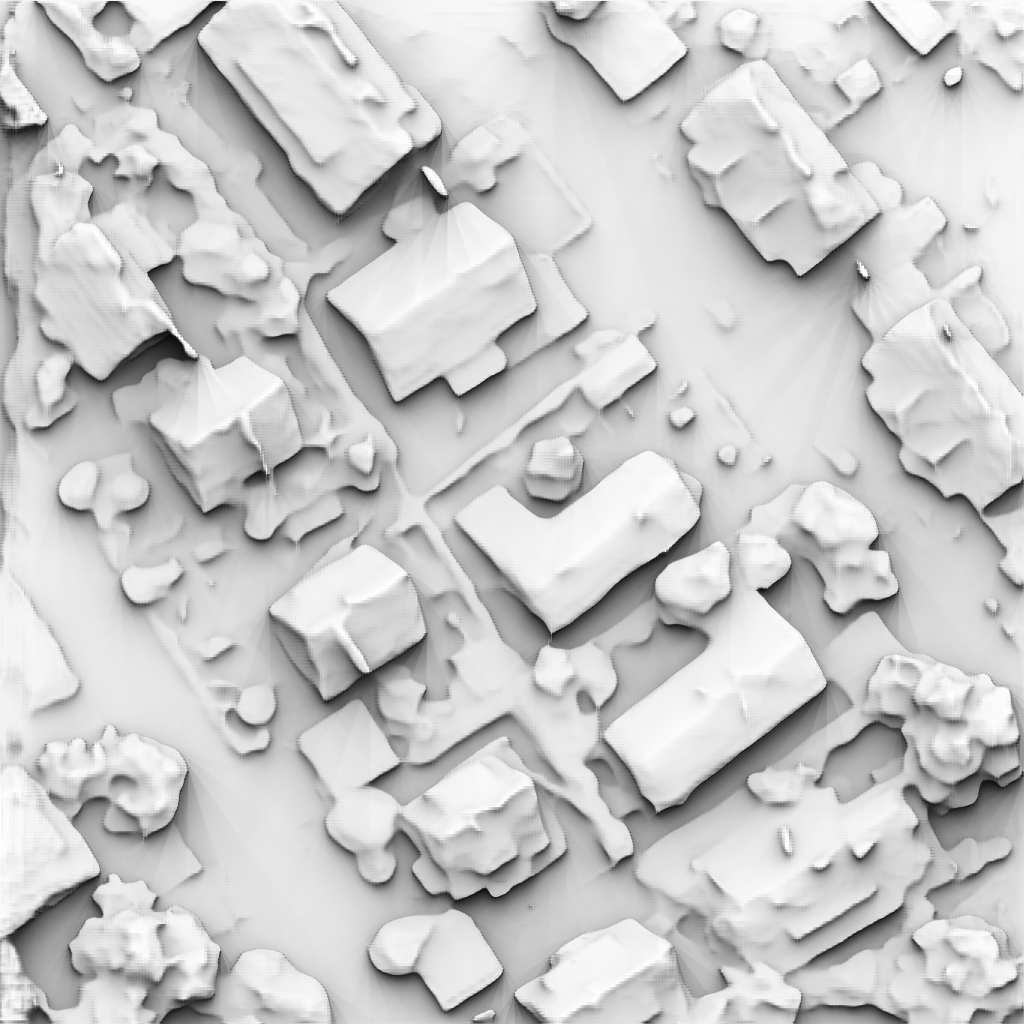}
		\centering{\tiny LEAStereo}
	\end{minipage}
	\caption{Error map and disparity visualization on building area for ISPRS Vaihingen dataset.}
	\label{Figure.vaihingenbulding}
\end{figure}

\begin{figure}[tp]
	\begin{minipage}[t]{0.19\textwidth}
		\includegraphics[width=0.098\linewidth]{figures/color_map.png}
		\includegraphics[width=0.85\linewidth]{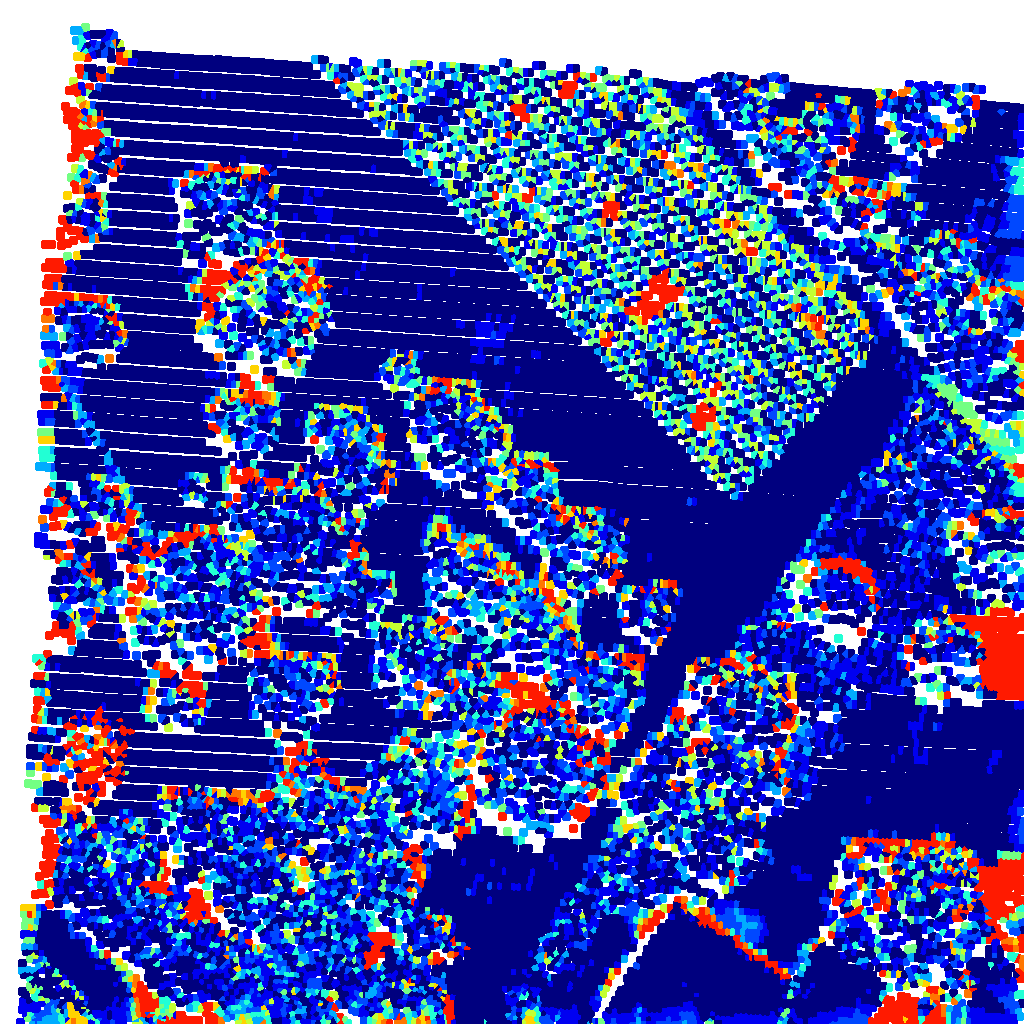}
		\includegraphics[width=\linewidth]{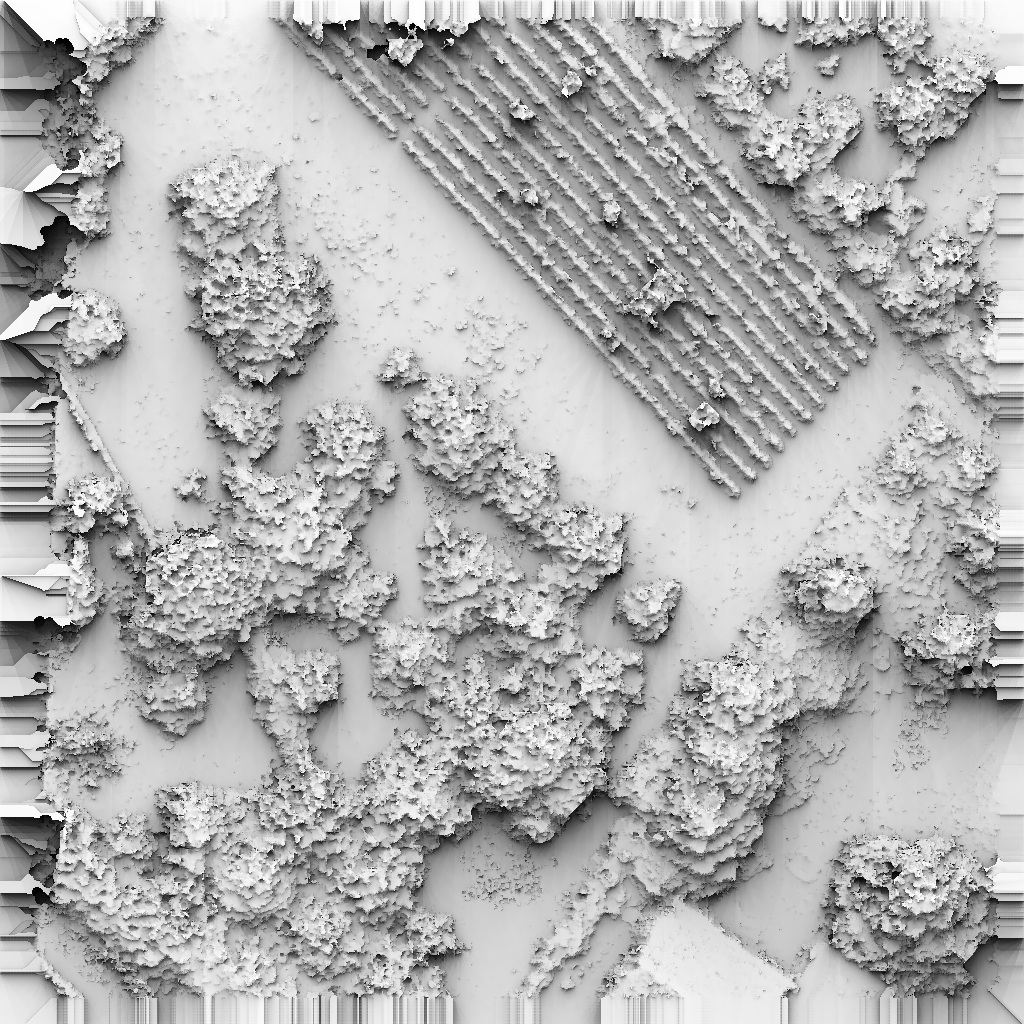}
		\centering{\tiny MICMAC}
	\end{minipage}
	\begin{minipage}[t]{0.19\textwidth}	
		\includegraphics[width=0.098\linewidth]{figures/color_map.png}
		\includegraphics[width=0.85\linewidth]{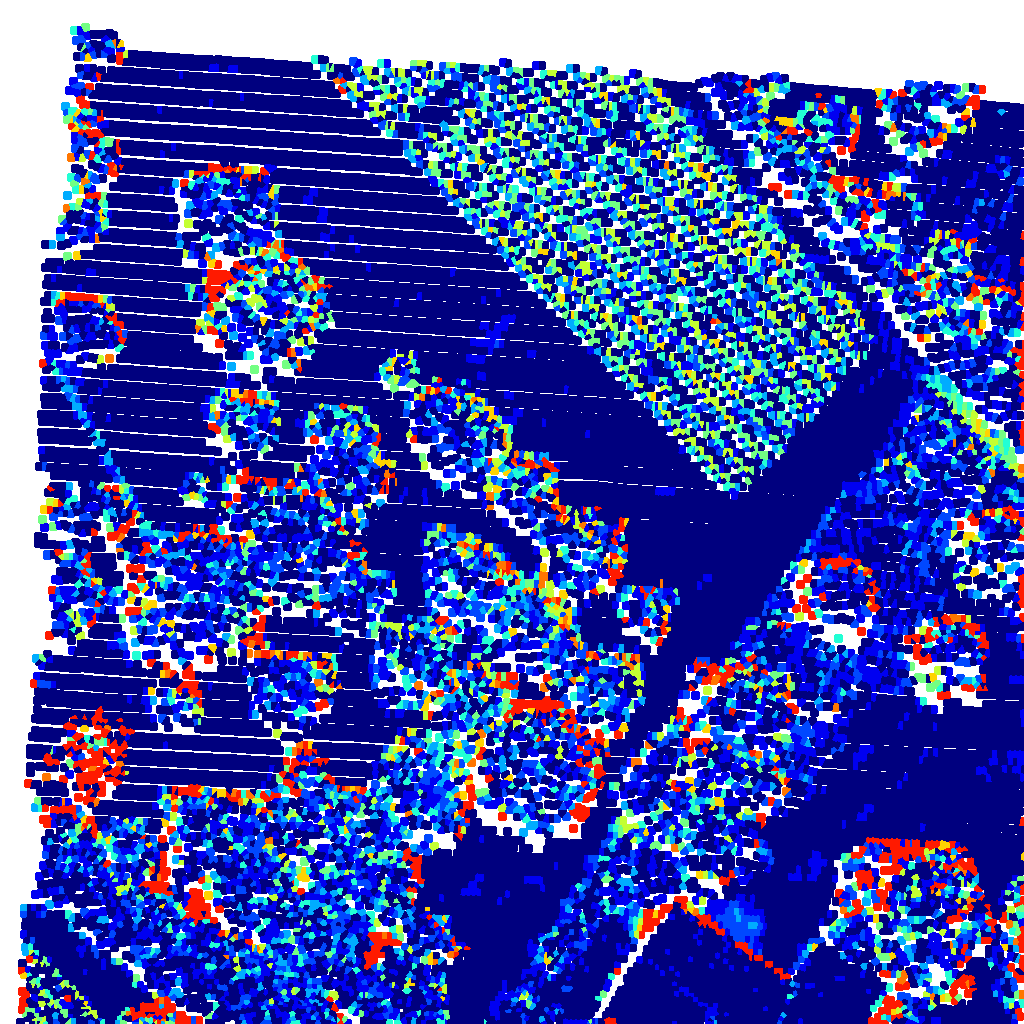}
		\includegraphics[width=\linewidth]{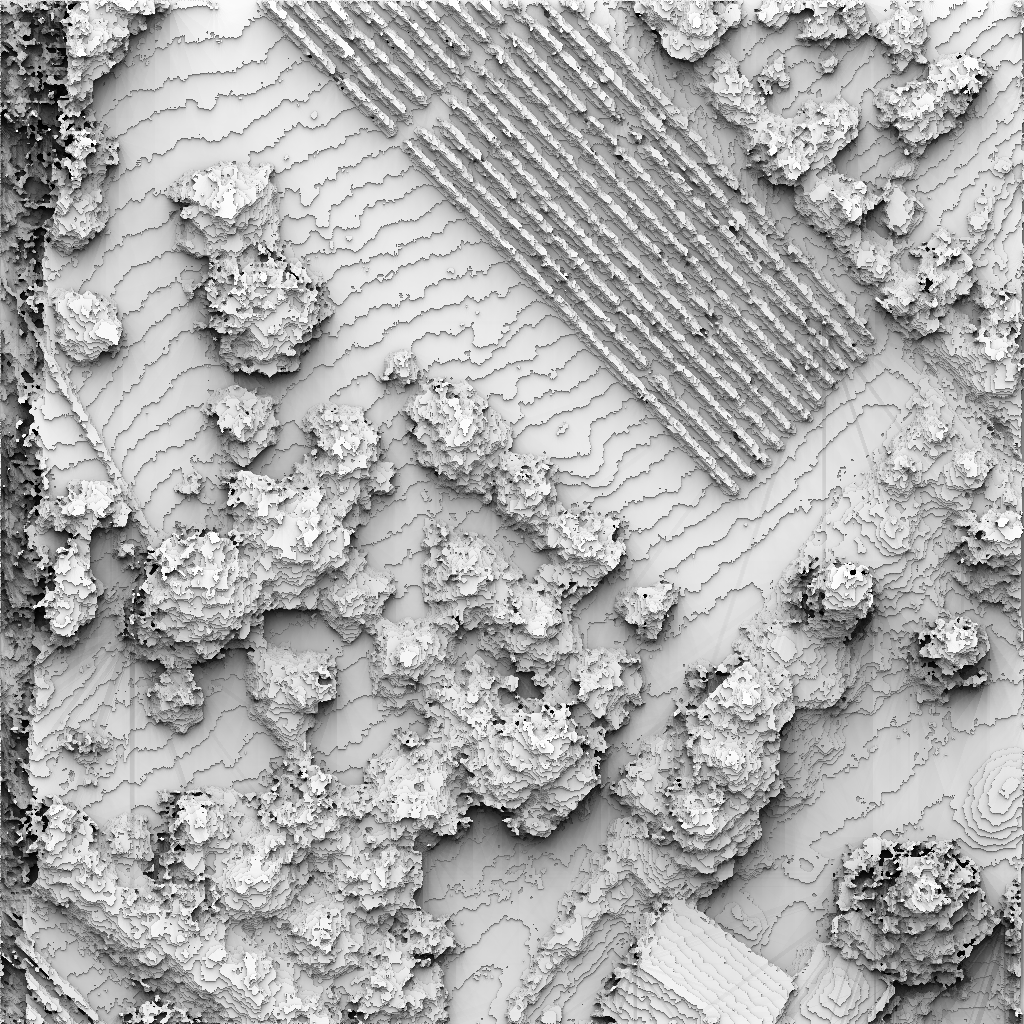}
		\centering{\tiny SGM(CUDA)}
	\end{minipage}
	\begin{minipage}[t]{0.19\textwidth}	
		\includegraphics[width=0.098\linewidth]{figures/color_map.png}
		\includegraphics[width=0.85\linewidth]{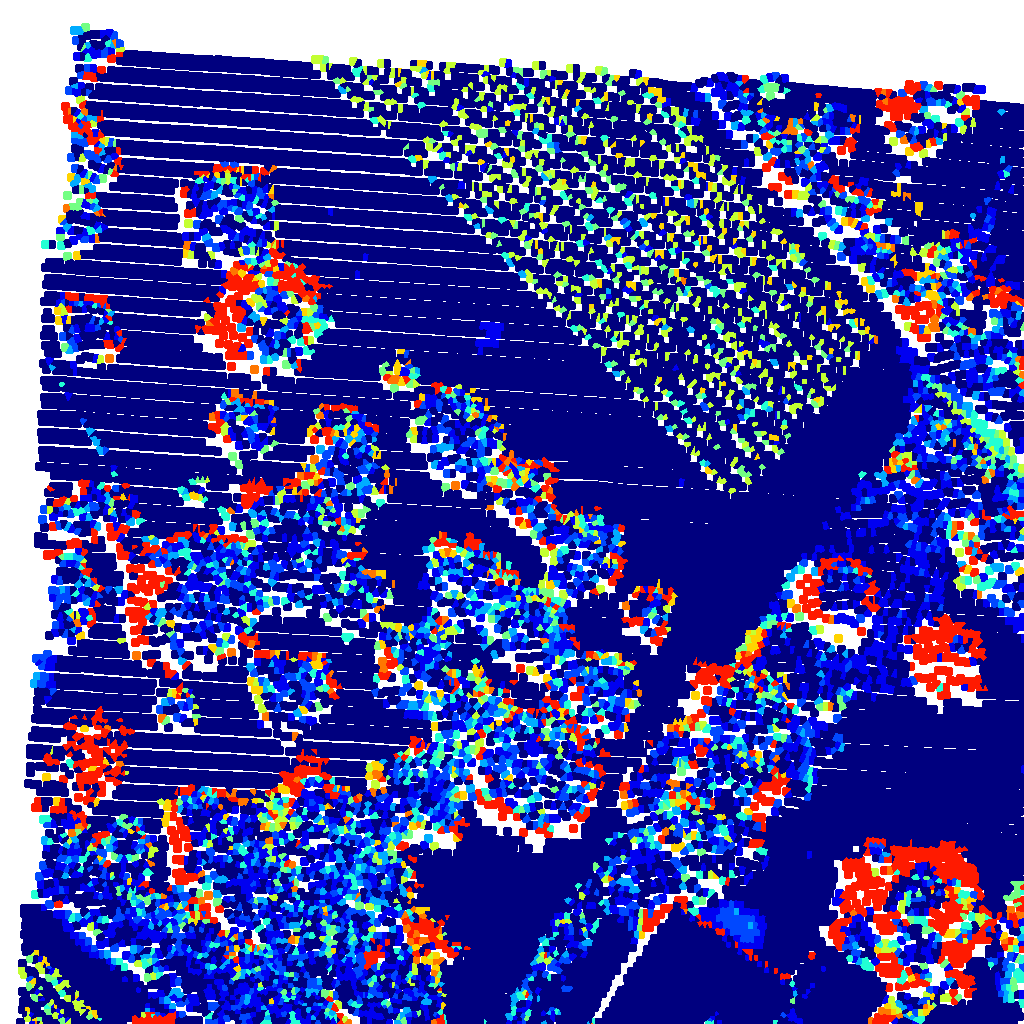}
		\includegraphics[width=\linewidth]{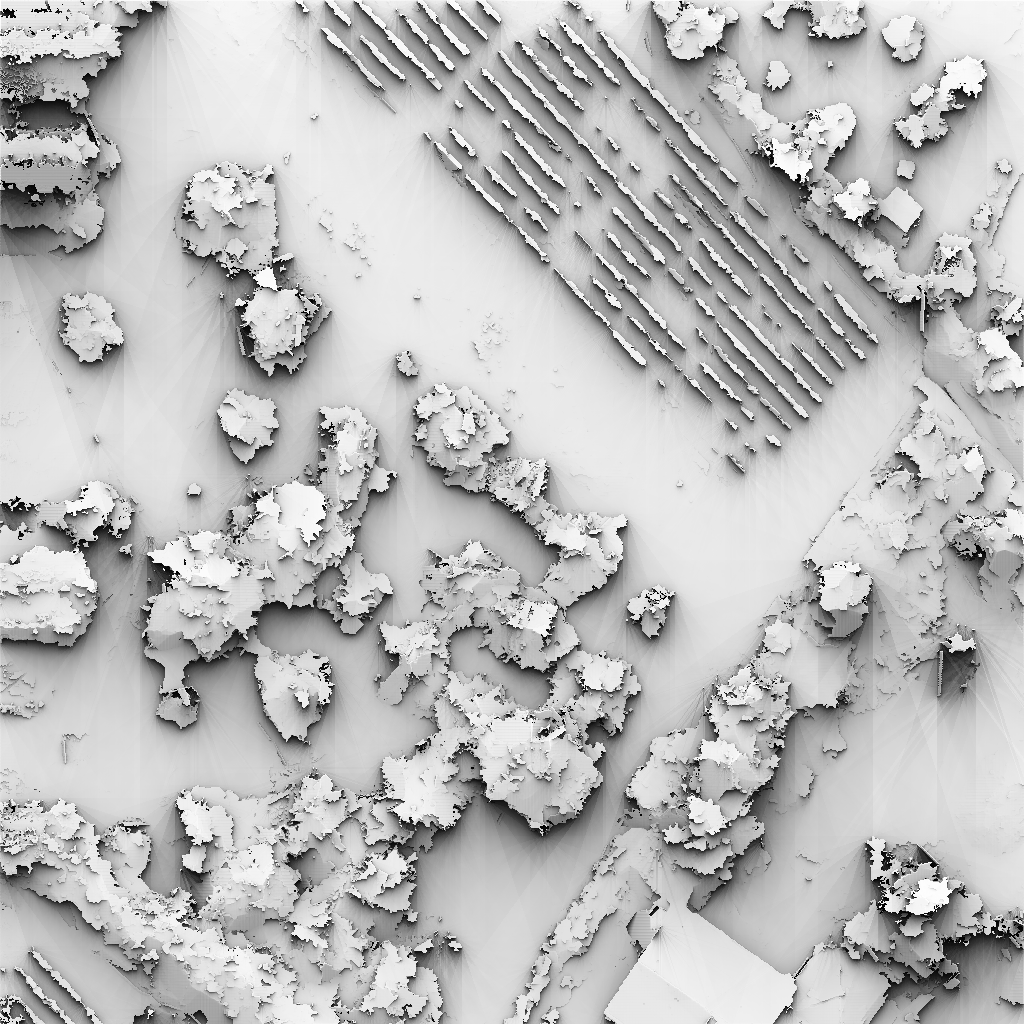}
		\centering{\tiny GraphCuts}
	\end{minipage}
	\begin{minipage}[t]{0.19\textwidth}	
		\includegraphics[width=0.098\linewidth]{figures/color_map.png}
		\includegraphics[width=0.85\linewidth]{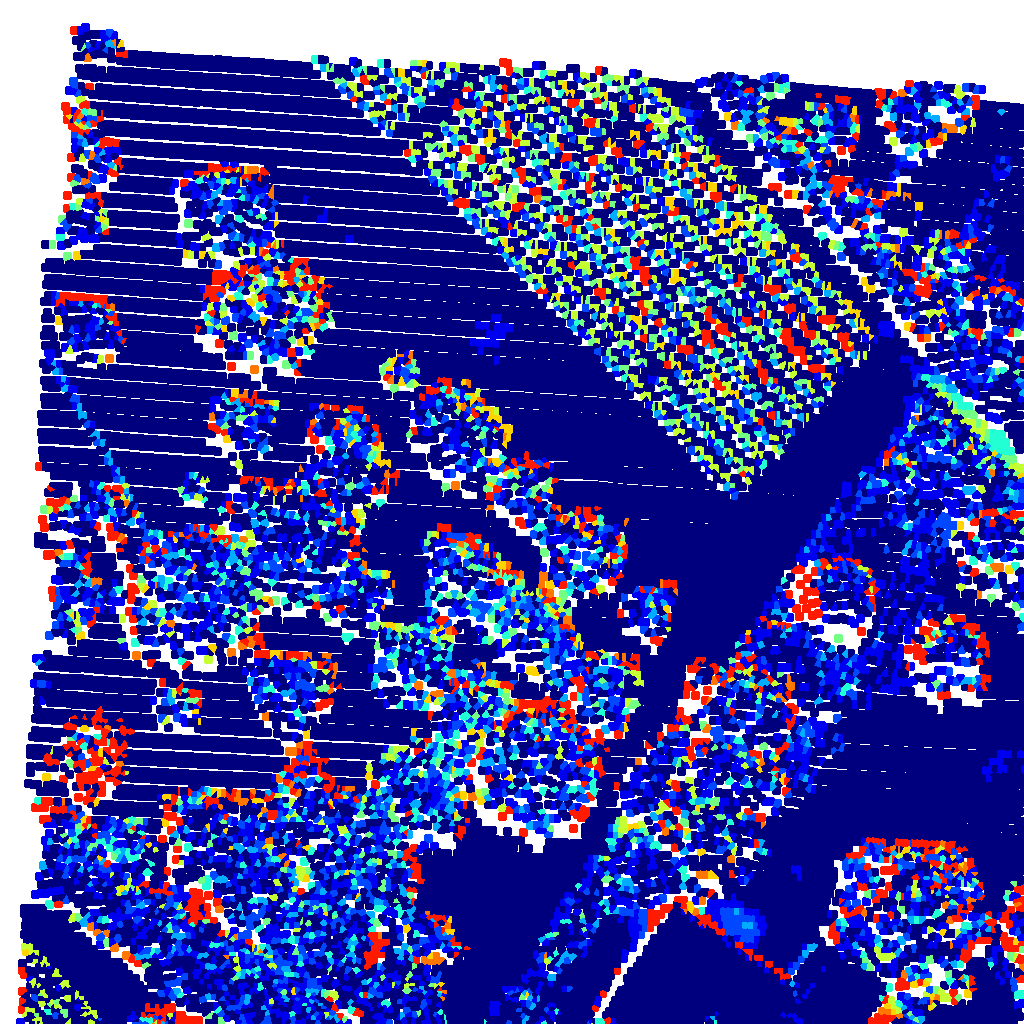}
		\includegraphics[width=\linewidth]{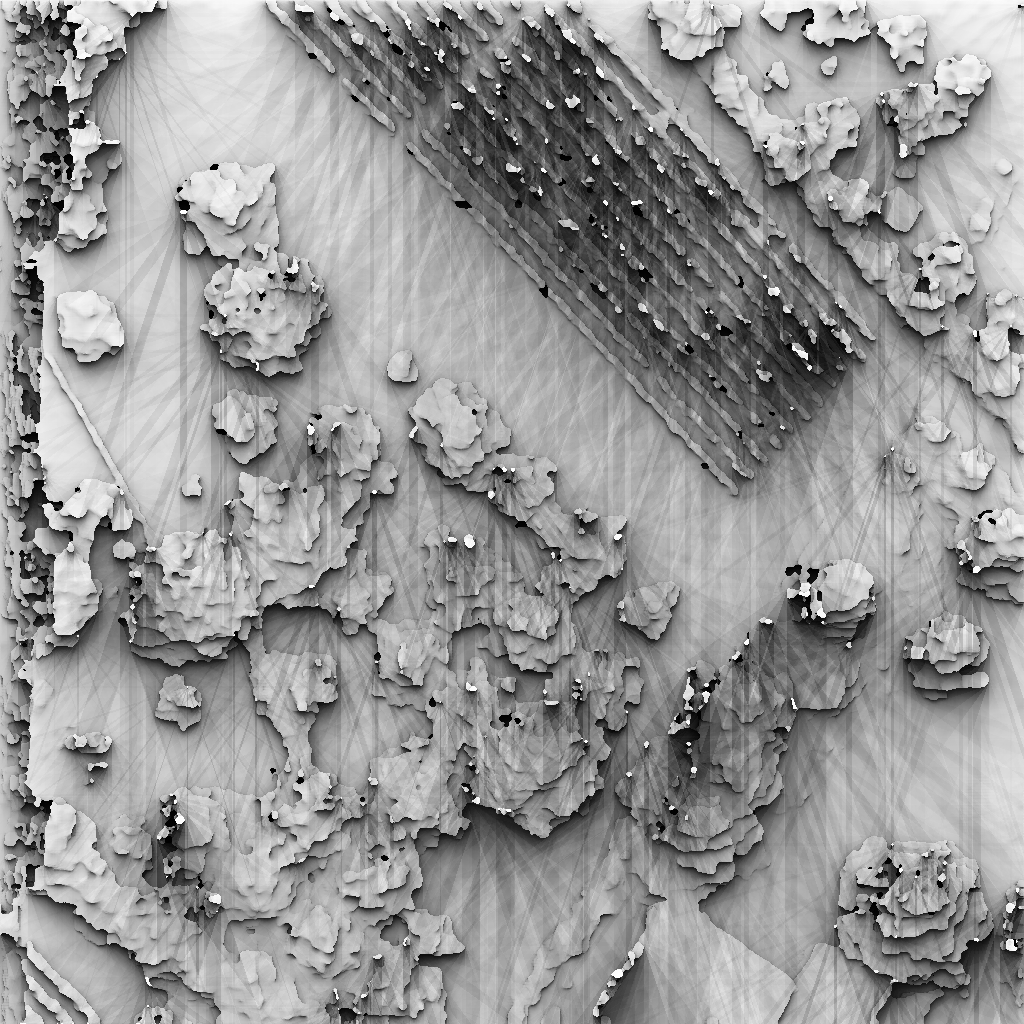}
		\centering{\tiny CBMV(SGM)}
	\end{minipage}
	\begin{minipage}[t]{0.19\textwidth}	
		\includegraphics[width=0.098\linewidth]{figures/color_map.png}
		\includegraphics[width=0.85\linewidth]{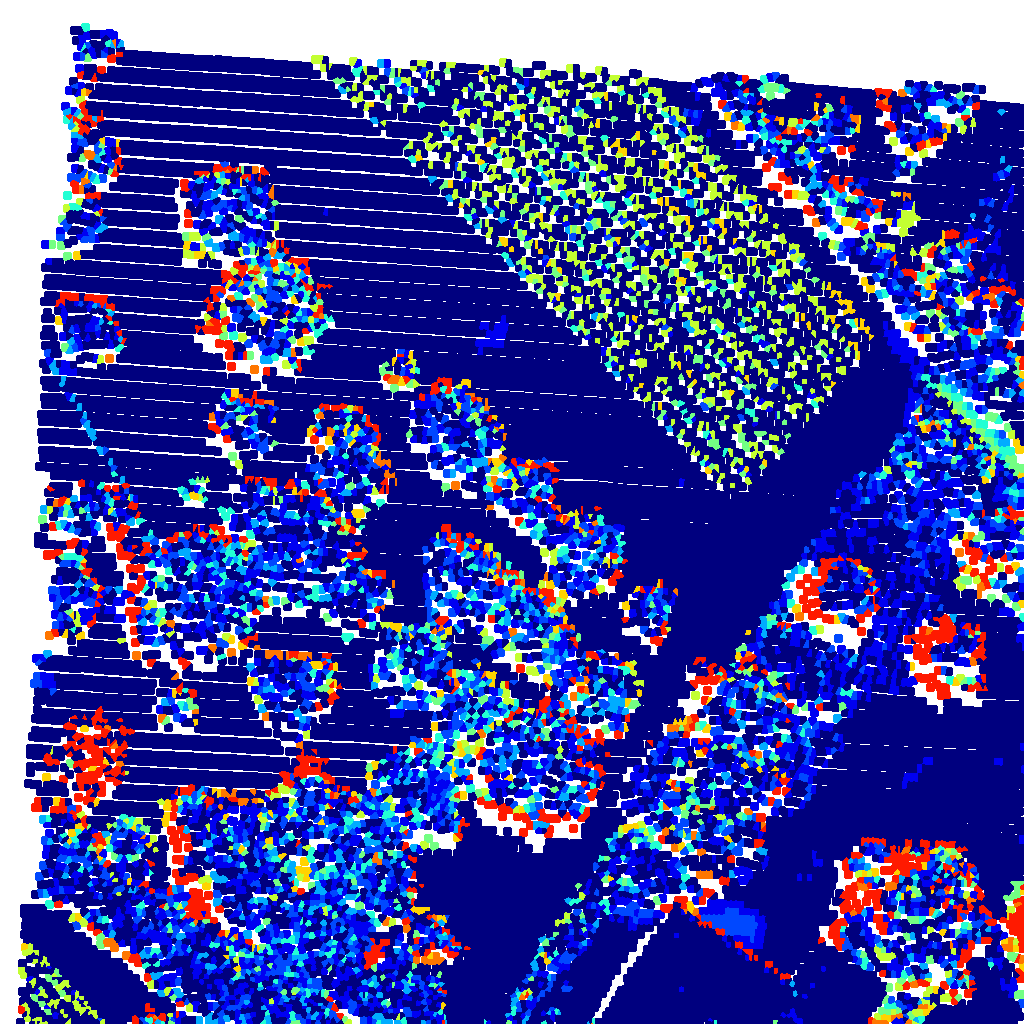}
		\includegraphics[width=\linewidth]{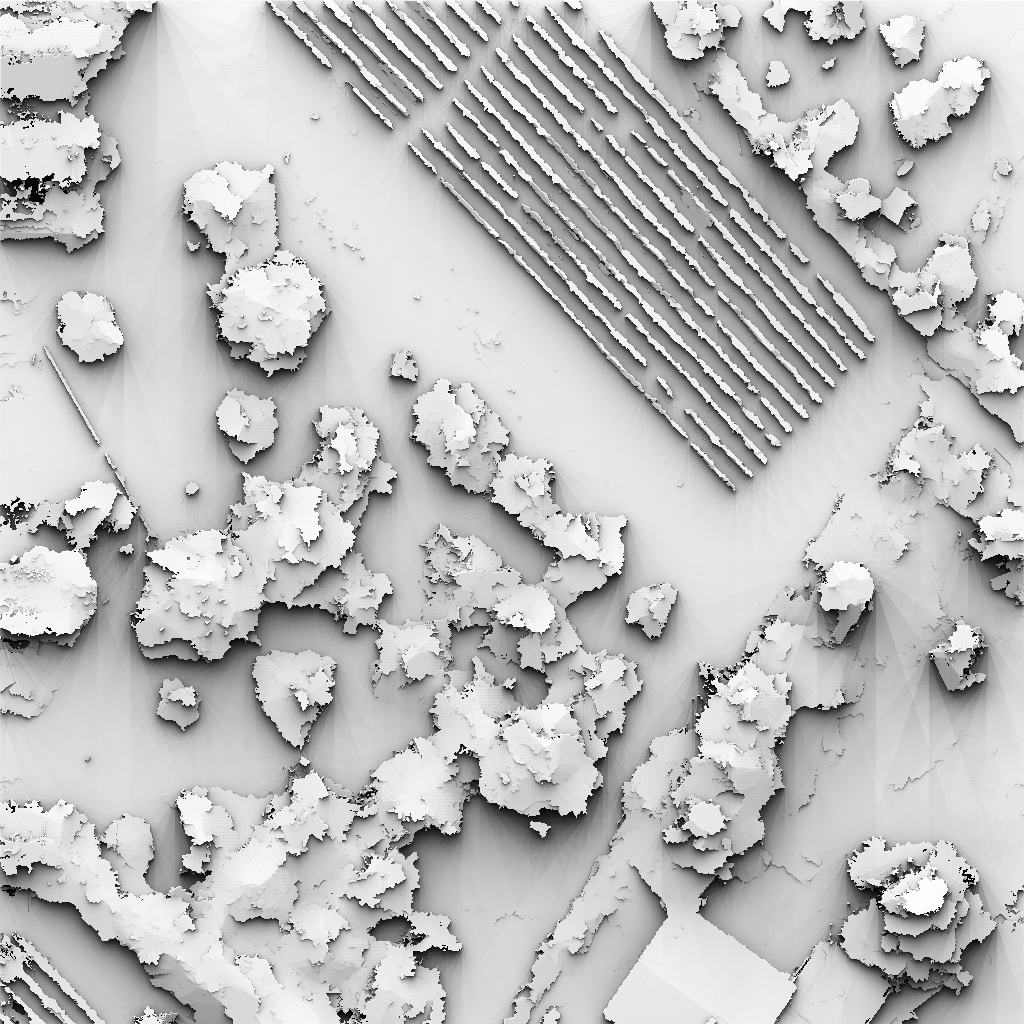}
		\centering{\tiny CBMV(GraphCuts)}
	\end{minipage}
	\begin{minipage}[t]{0.19\textwidth}	
		\includegraphics[width=0.098\linewidth]{figures/color_map.png}
		\includegraphics[width=0.85\linewidth]{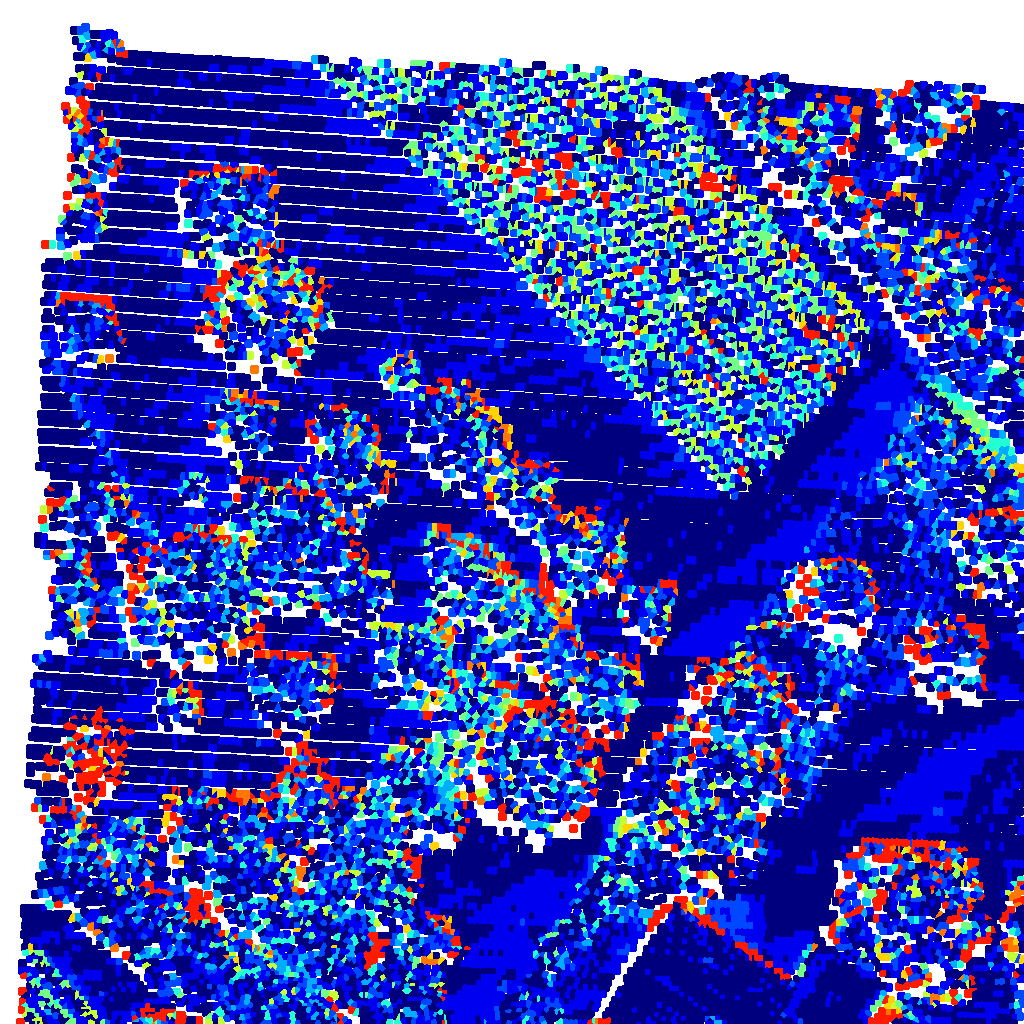}
		\includegraphics[width=\linewidth]{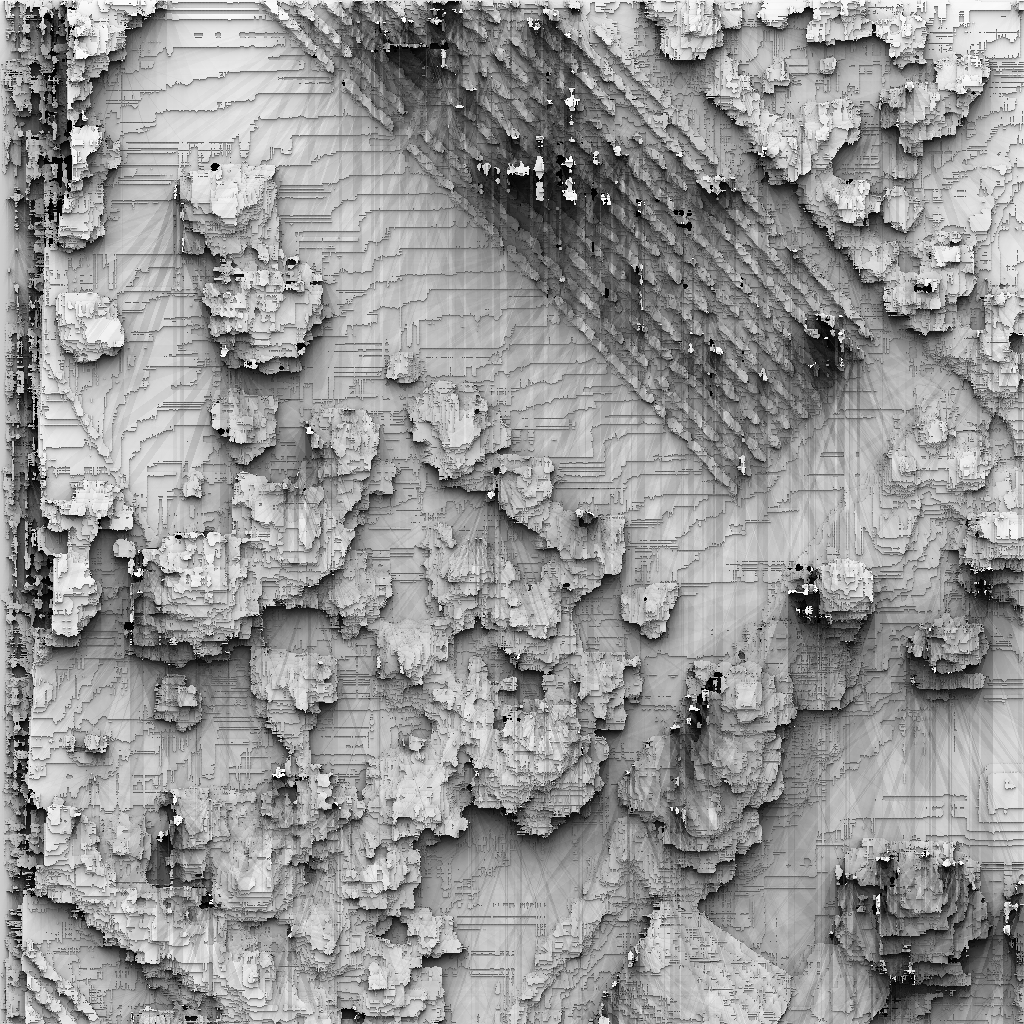}
		\centering{\tiny MC-CNN(KITTI)}
	\end{minipage}
	\begin{minipage}[t]{0.19\textwidth}	
		\includegraphics[width=0.098\linewidth]{figures/color_map.png}
		\includegraphics[width=0.85\linewidth]{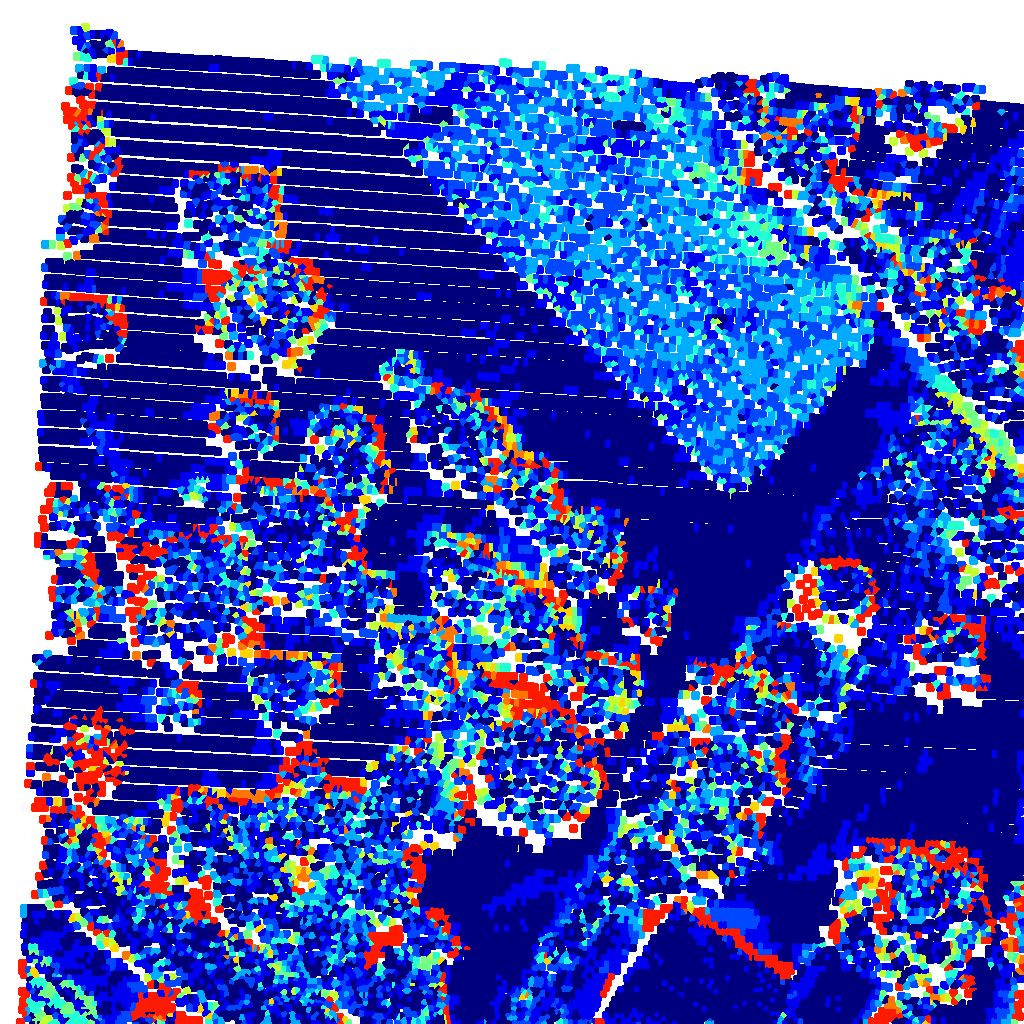}
		\includegraphics[width=\linewidth]{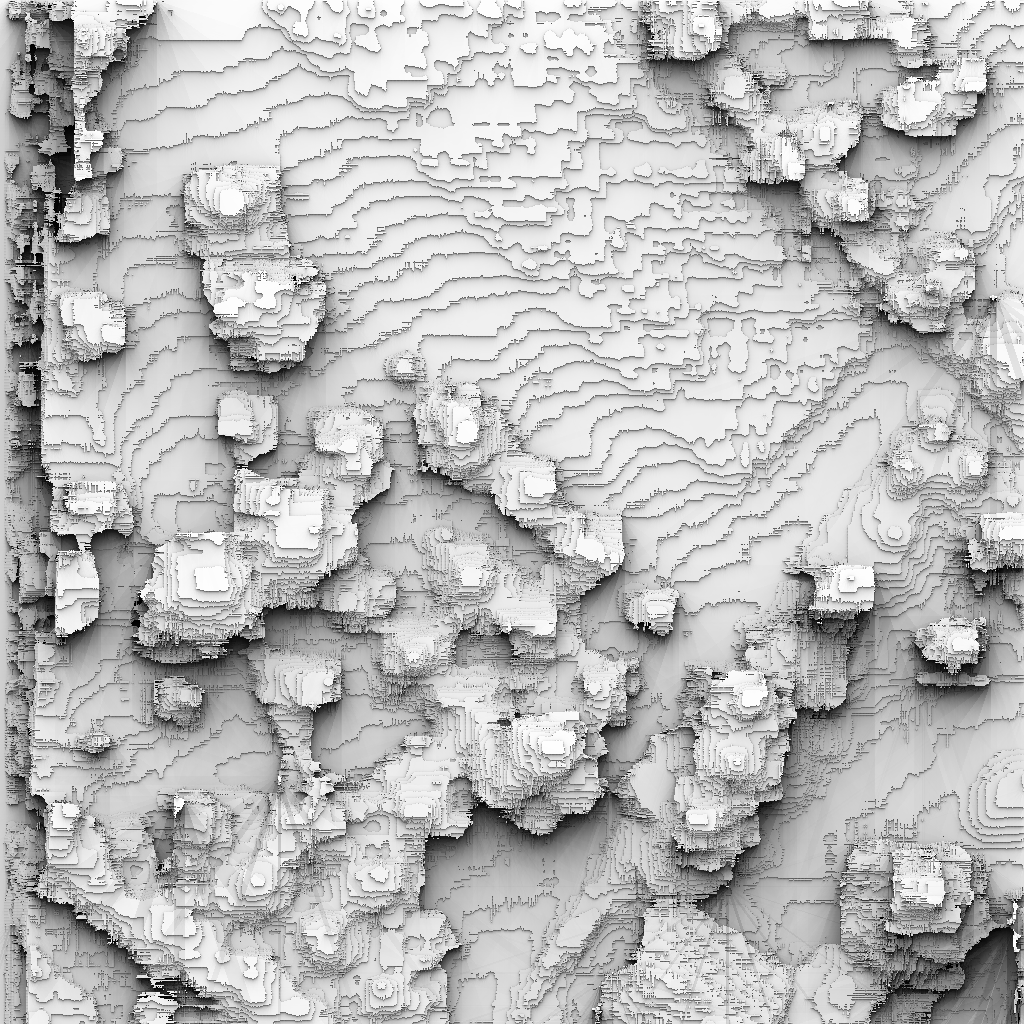}
		\centering{\tiny DeepFeature(KITTI)}
	\end{minipage}
	\begin{minipage}[t]{0.19\textwidth}	
		\includegraphics[width=0.098\linewidth]{figures/color_map.png}
		\includegraphics[width=0.85\linewidth]{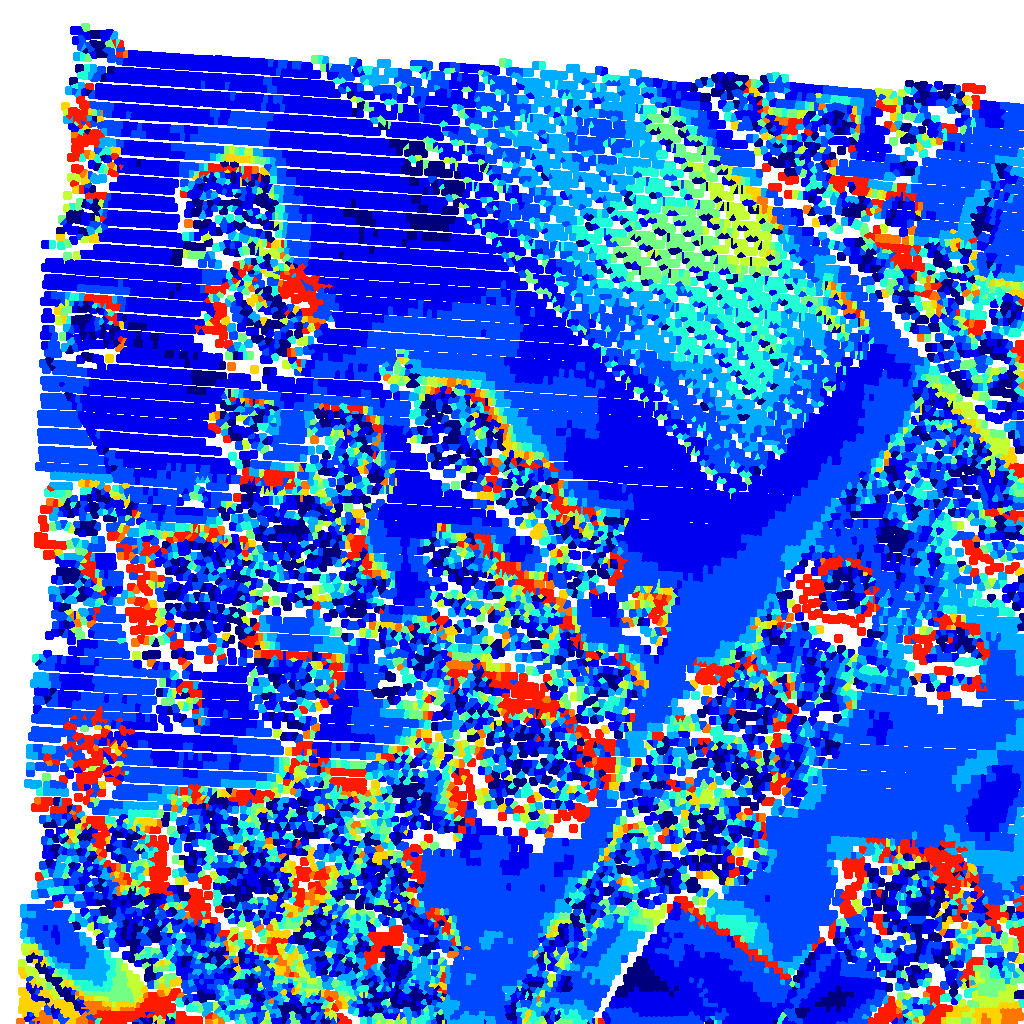}
		\includegraphics[width=\linewidth]{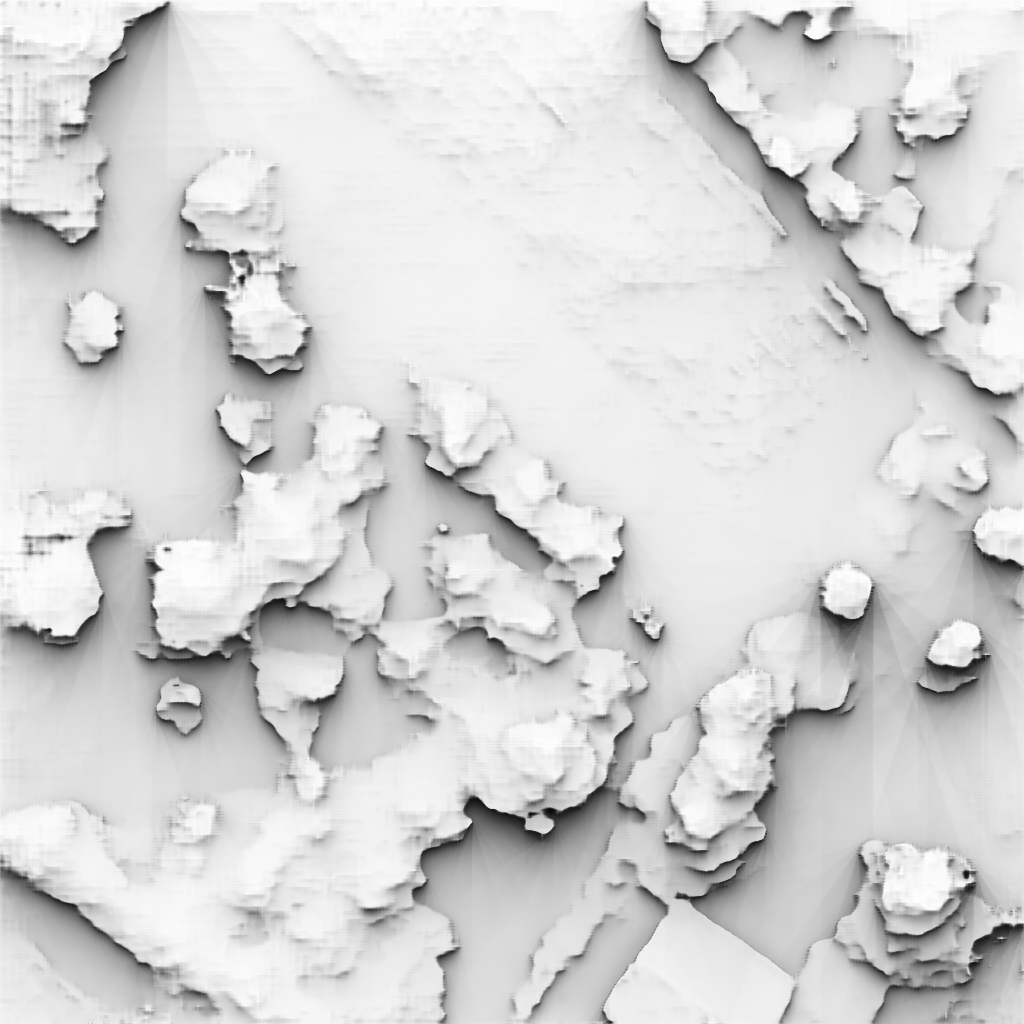}
		\centering{\tiny PSM net(KITTI)}
	\end{minipage}
	\begin{minipage}[t]{0.19\textwidth}	
		\includegraphics[width=0.098\linewidth]{figures/color_map.png}
		\includegraphics[width=0.85\linewidth]{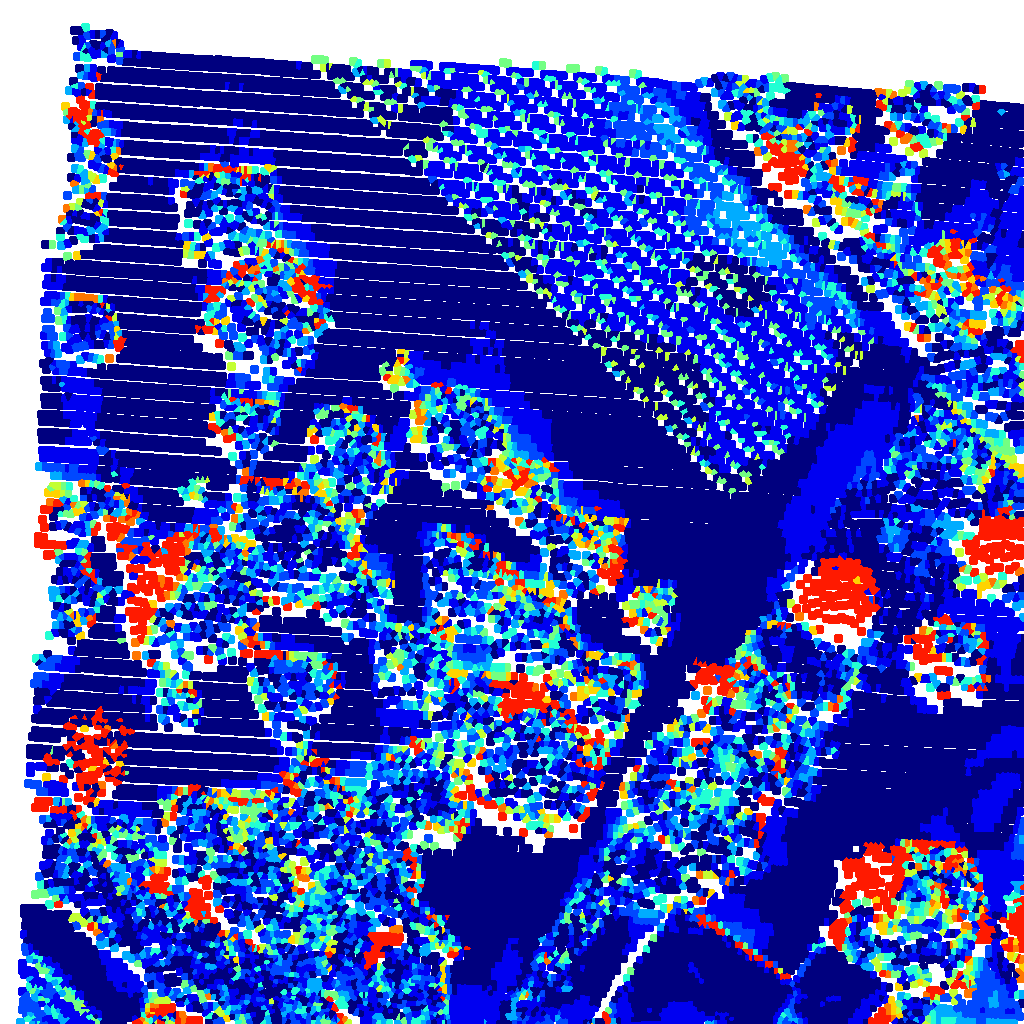}
		\includegraphics[width=\linewidth]{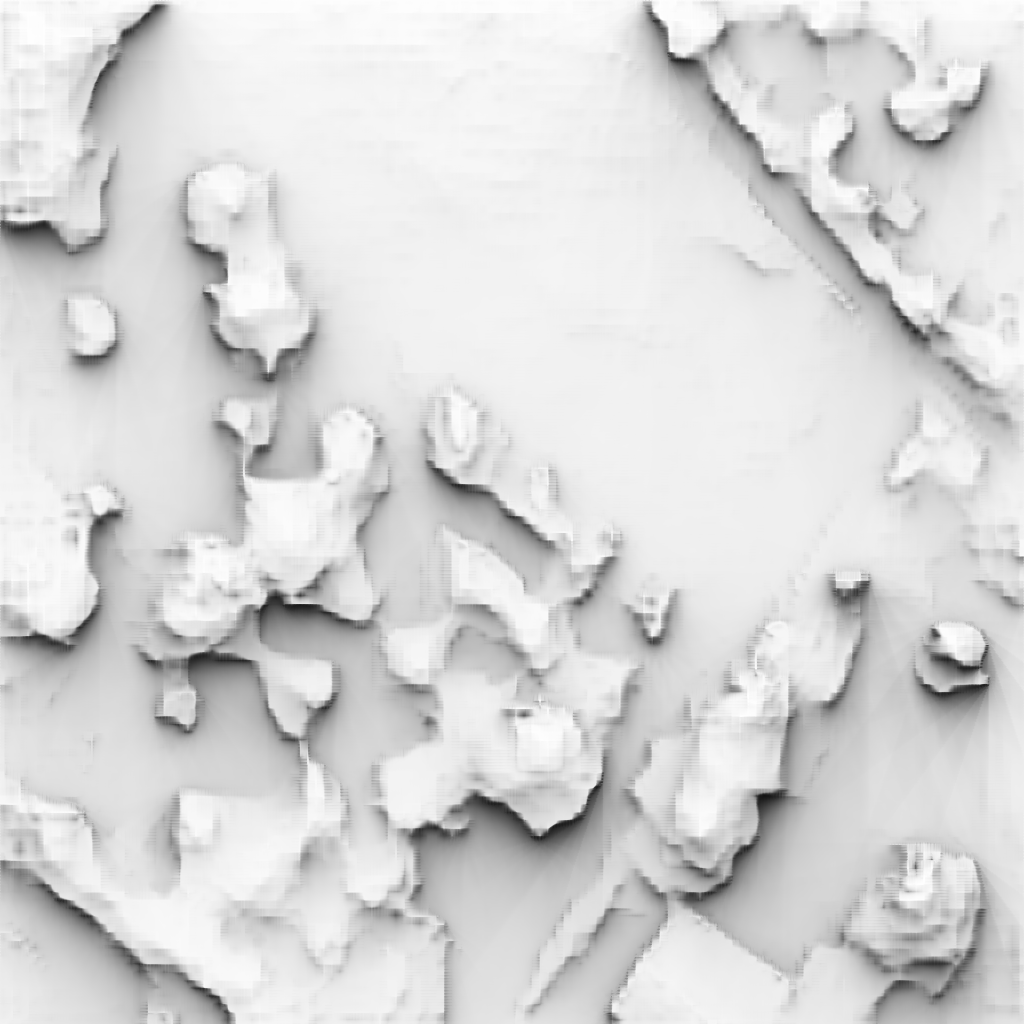}
		\centering{\tiny HRS net(KITTI)}
	\end{minipage}
	\begin{minipage}[t]{0.19\textwidth}	
		\includegraphics[width=0.098\linewidth]{figures/color_map.png}
		\includegraphics[width=0.85\linewidth]{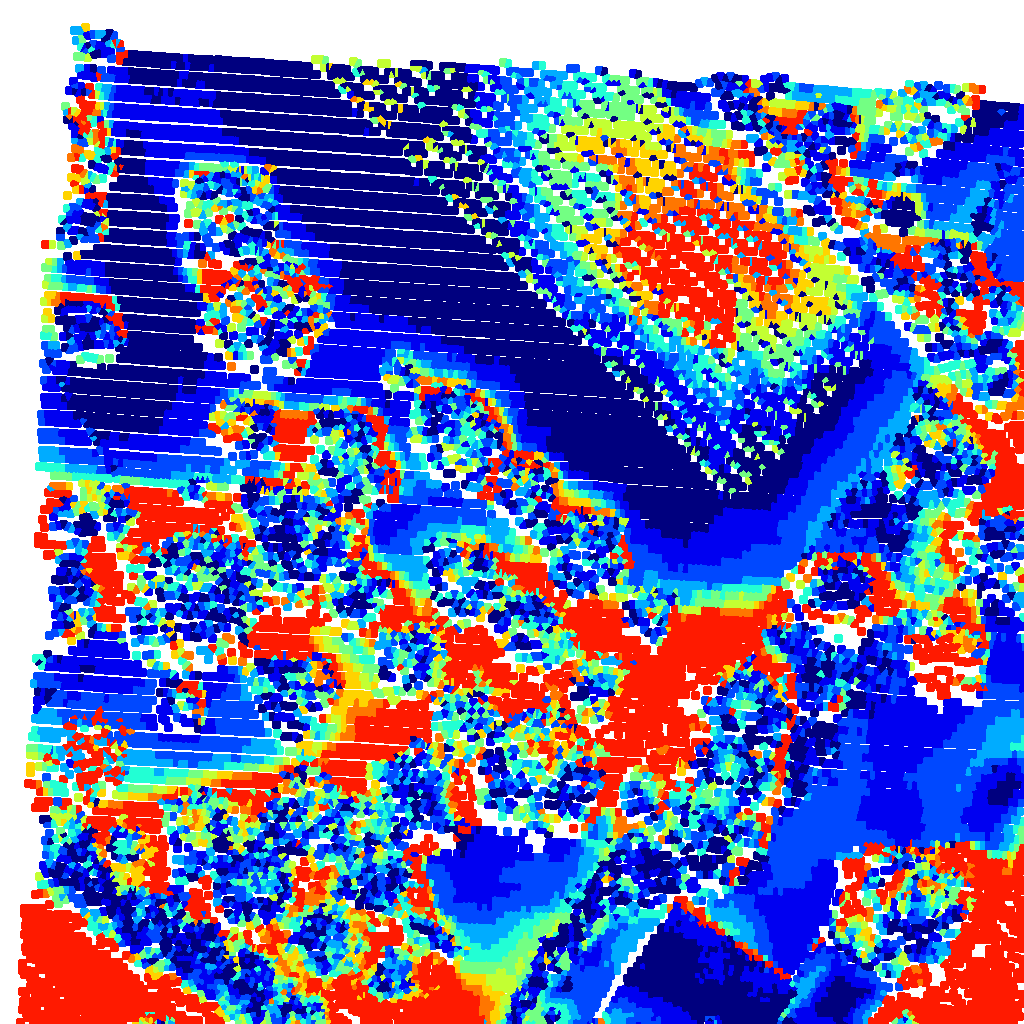}
		\includegraphics[width=\linewidth]{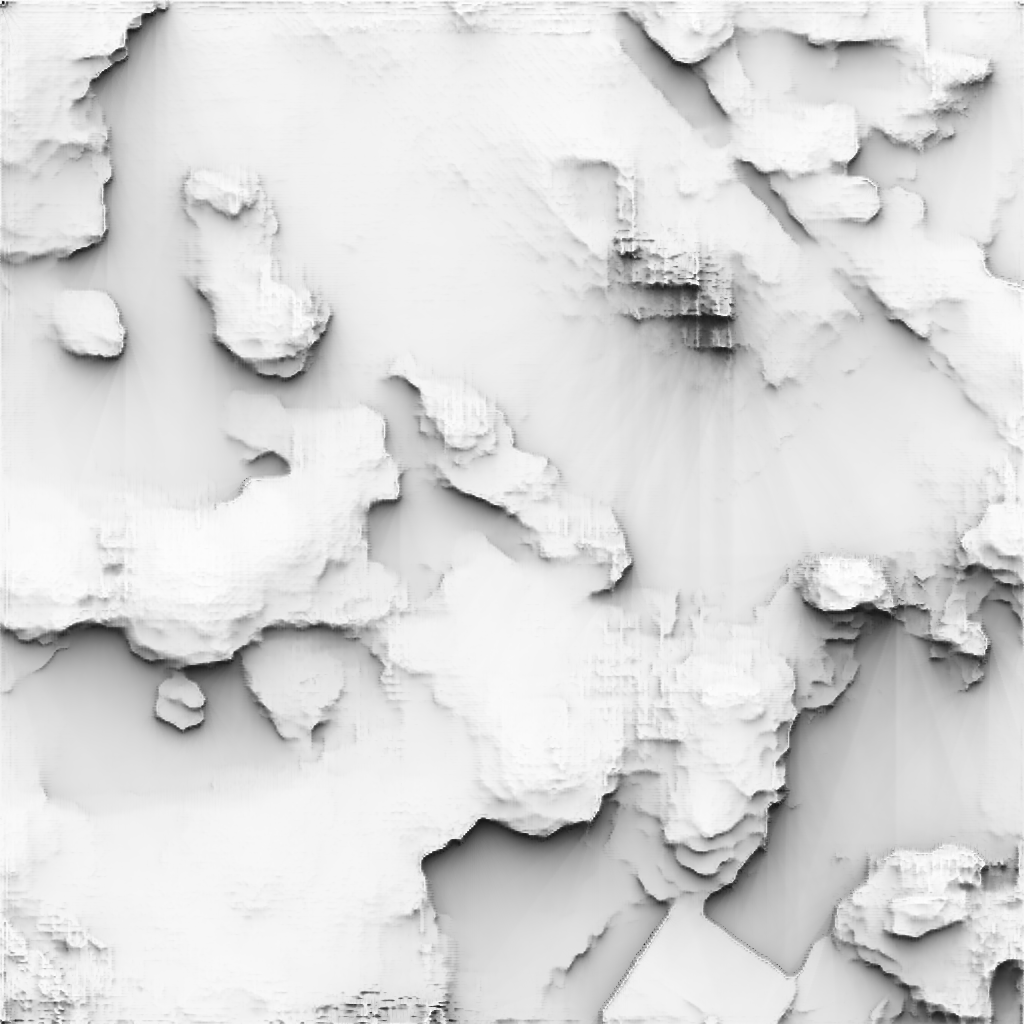}
		\centering{\tiny DeepPruner(KITTI)}
	\end{minipage}
	\begin{minipage}[t]{0.19\textwidth}
		\includegraphics[width=0.098\linewidth]{figures/color_map.png}
		\includegraphics[width=0.85\linewidth]{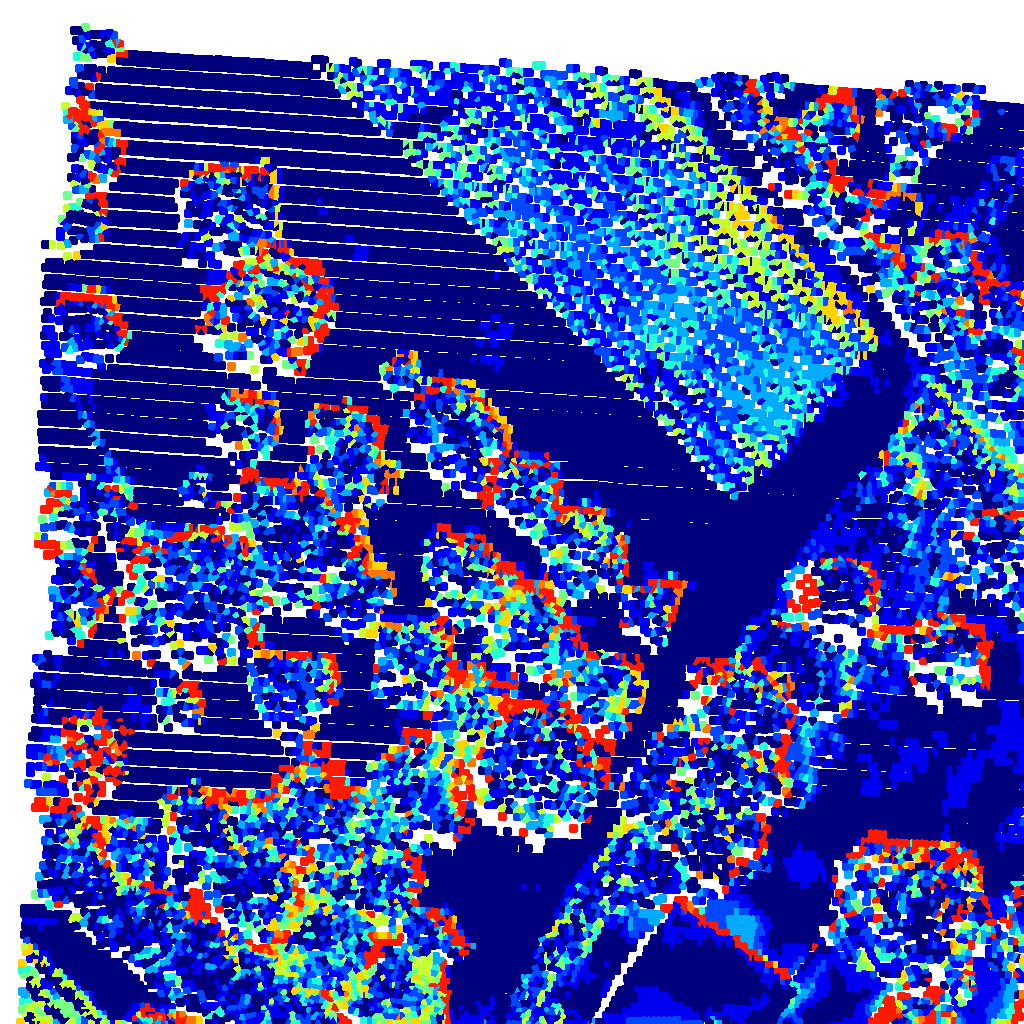}
		\includegraphics[width=\linewidth]{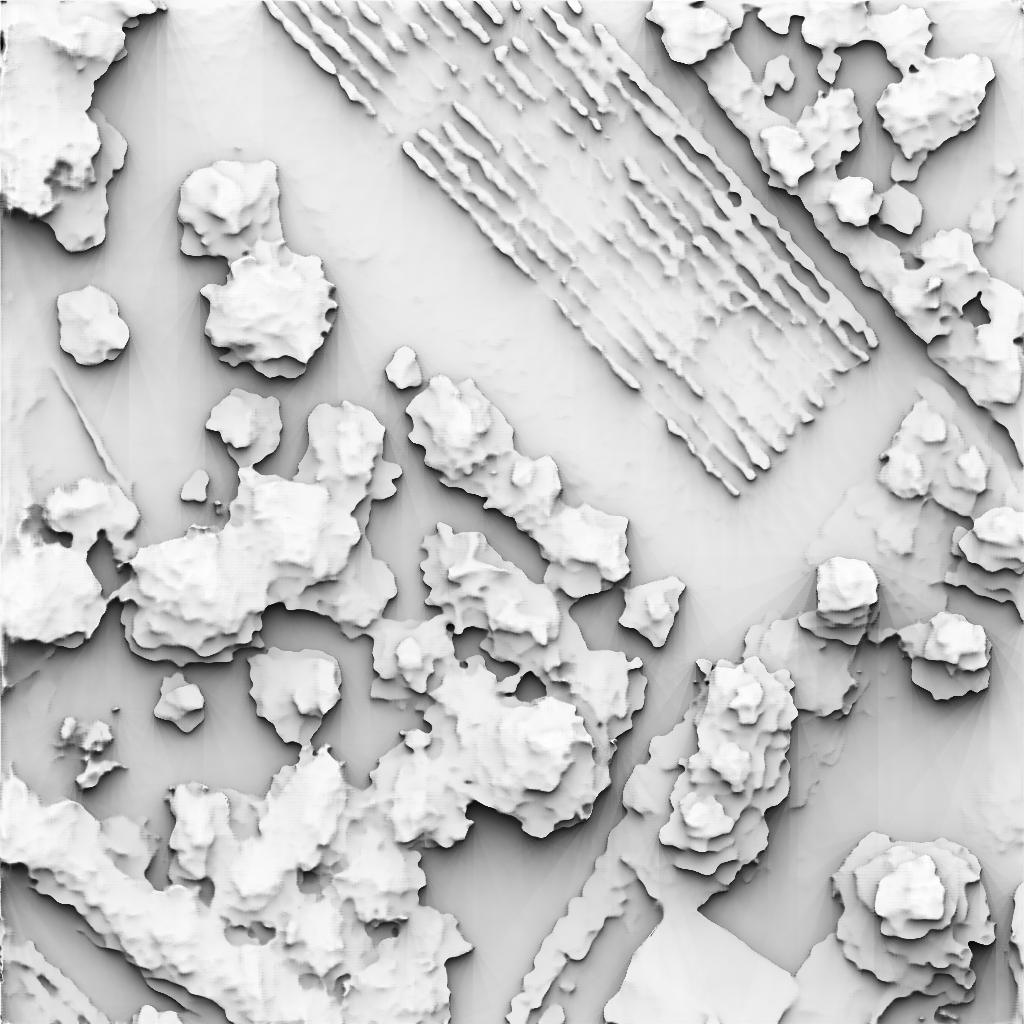}
		\centering{\tiny GANet(KITTI)}
	\end{minipage}
	\begin{minipage}[t]{0.19\textwidth}	
		\includegraphics[width=0.098\linewidth]{figures/color_map.png}
		\includegraphics[width=0.85\linewidth]{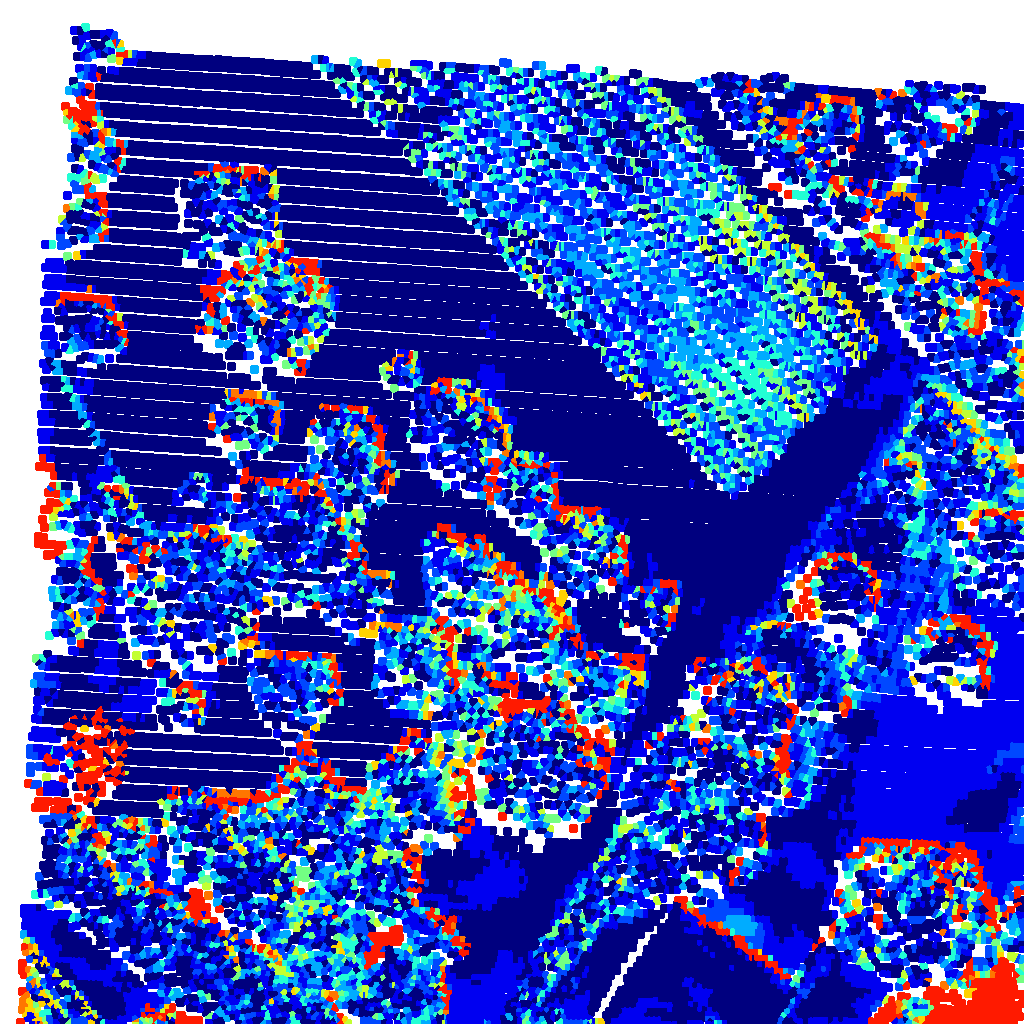}
		\includegraphics[width=\linewidth]{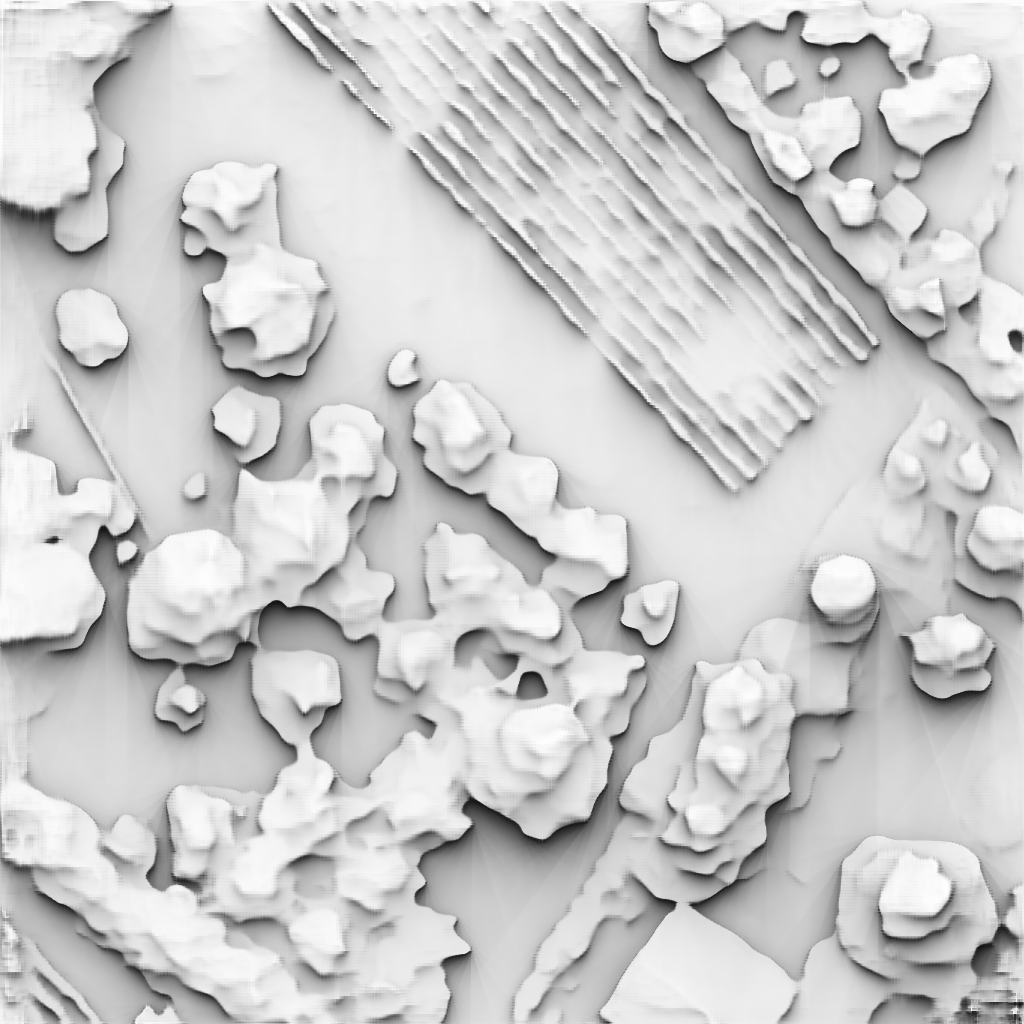}
		\centering{\tiny LEAStereo(KITTI)}
	\end{minipage}
	\begin{minipage}[t]{0.19\textwidth}	
		\includegraphics[width=0.098\linewidth]{figures/color_map.png}
		\includegraphics[width=0.85\linewidth]{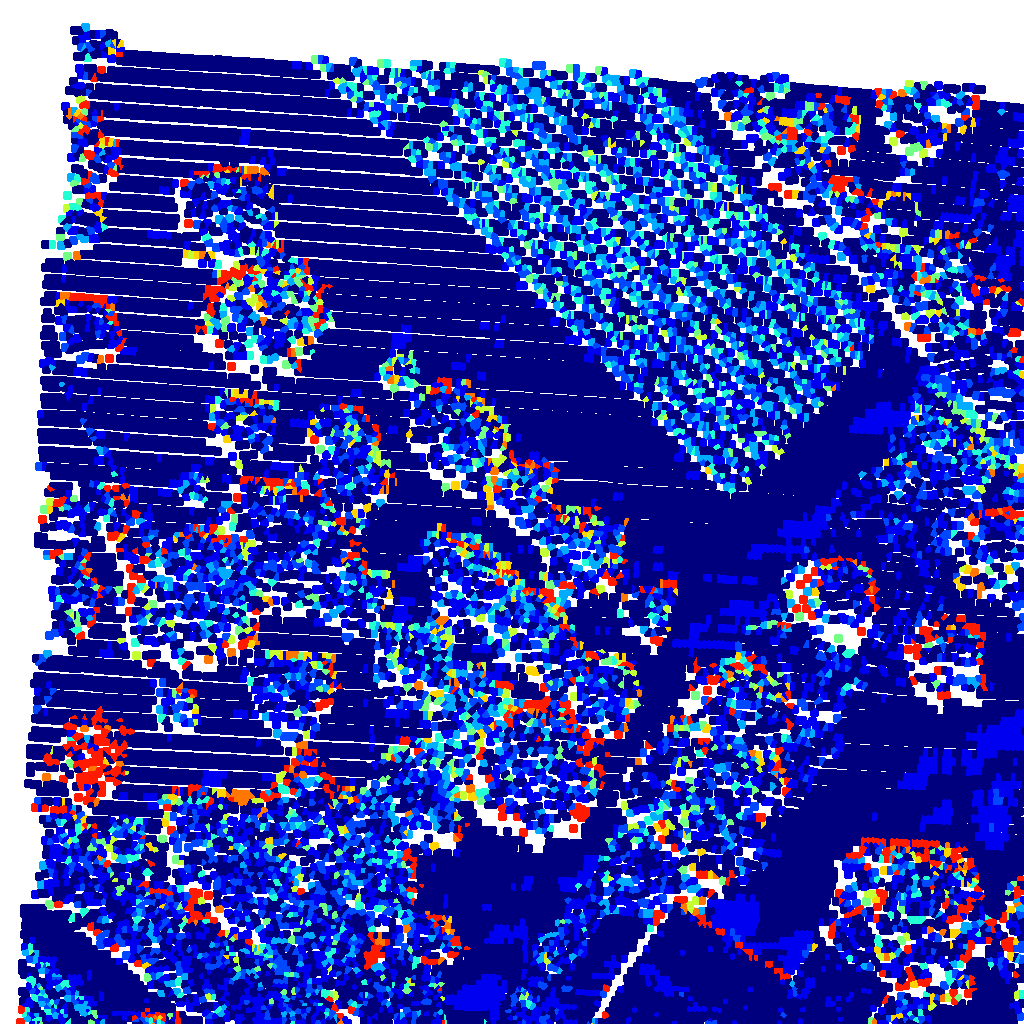}
		\includegraphics[width=\linewidth]{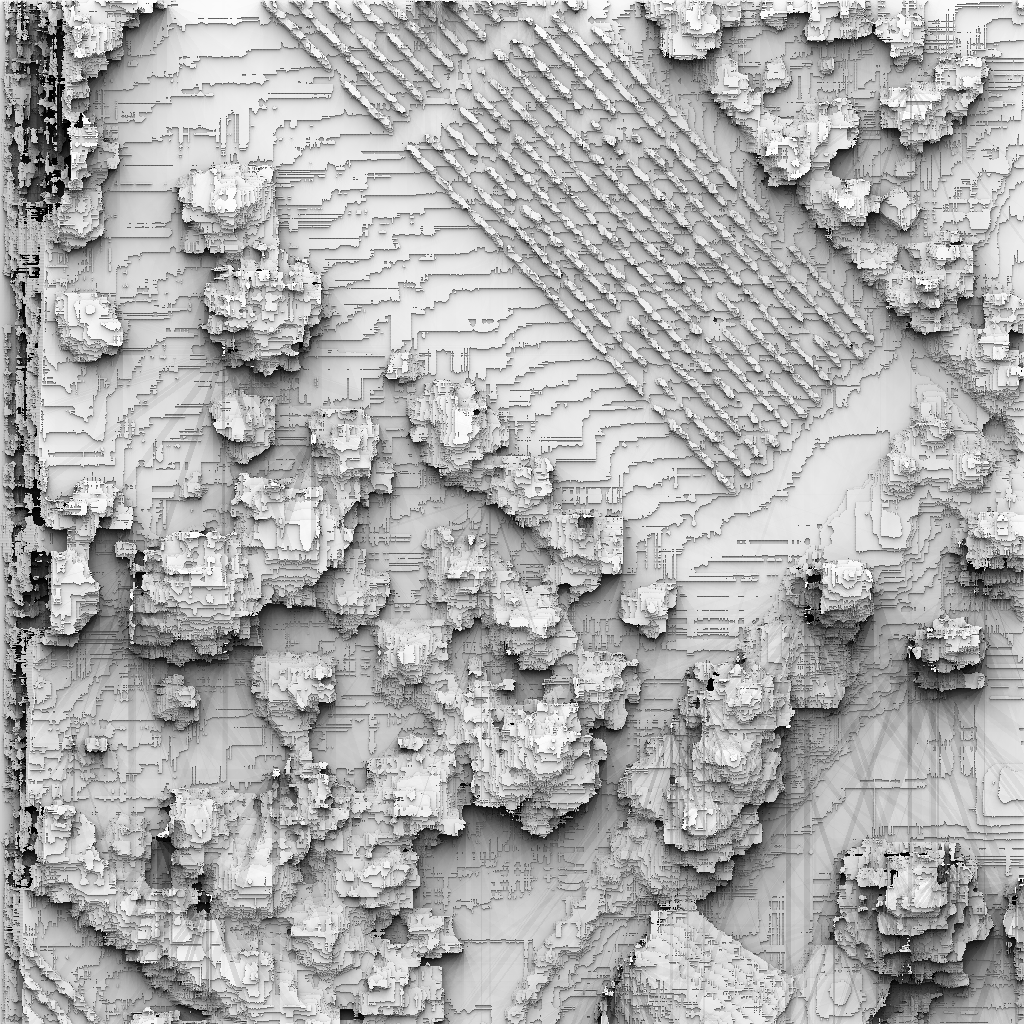}
		\centering{\tiny MC-CNN}
	\end{minipage}
	\begin{minipage}[t]{0.19\textwidth}	
		\includegraphics[width=0.098\linewidth]{figures/color_map.png}
		\includegraphics[width=0.85\linewidth]{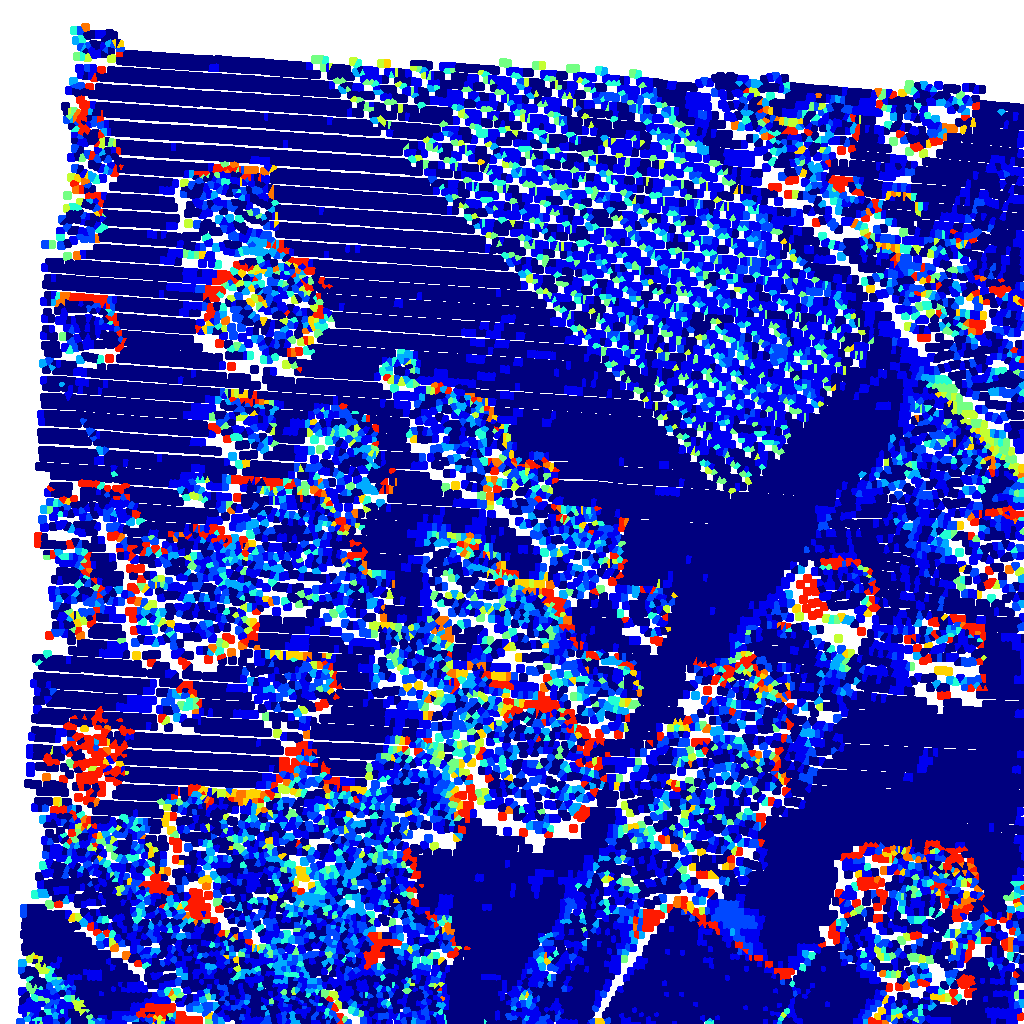}
		\includegraphics[width=\linewidth]{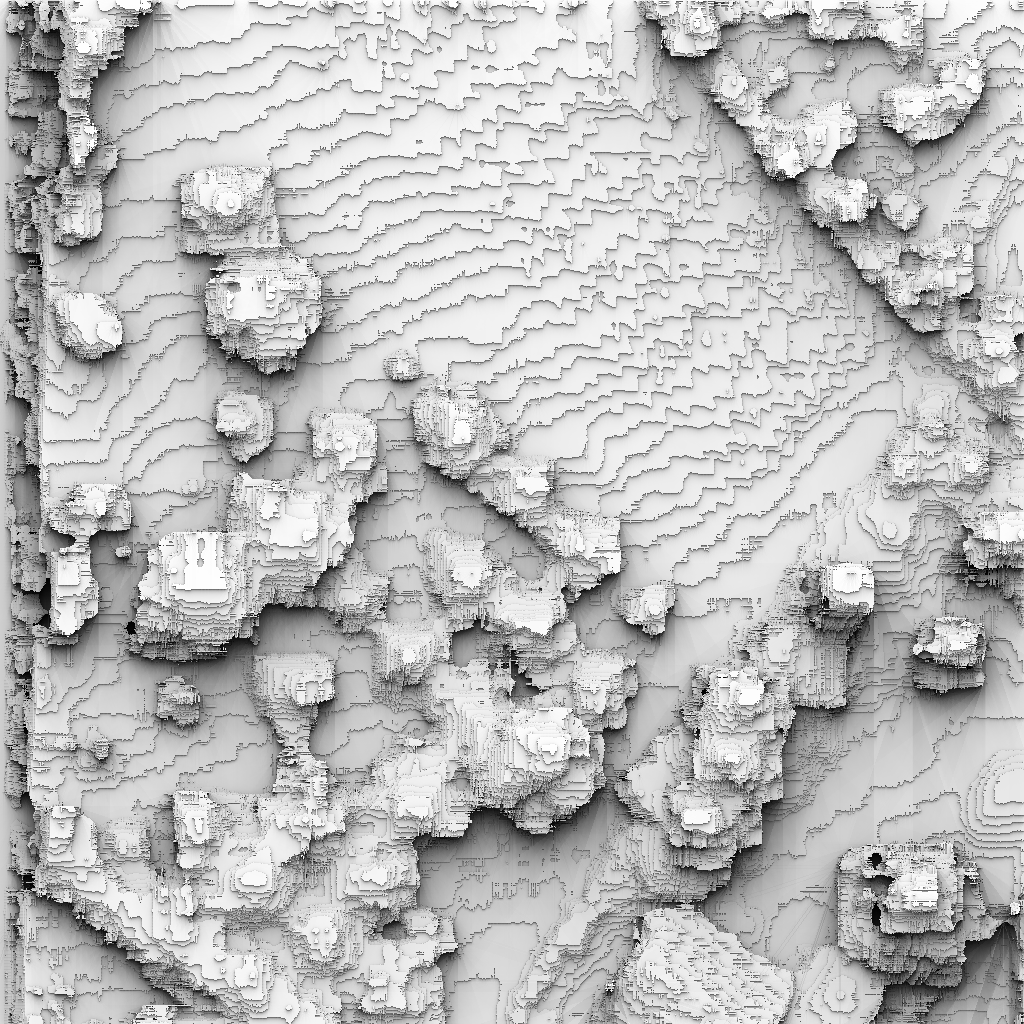}
		\centering{\tiny DeepFeature}
	\end{minipage}
	\begin{minipage}[t]{0.19\textwidth}
		\includegraphics[width=0.098\linewidth]{figures/color_map.png}
		\includegraphics[width=0.85\linewidth]{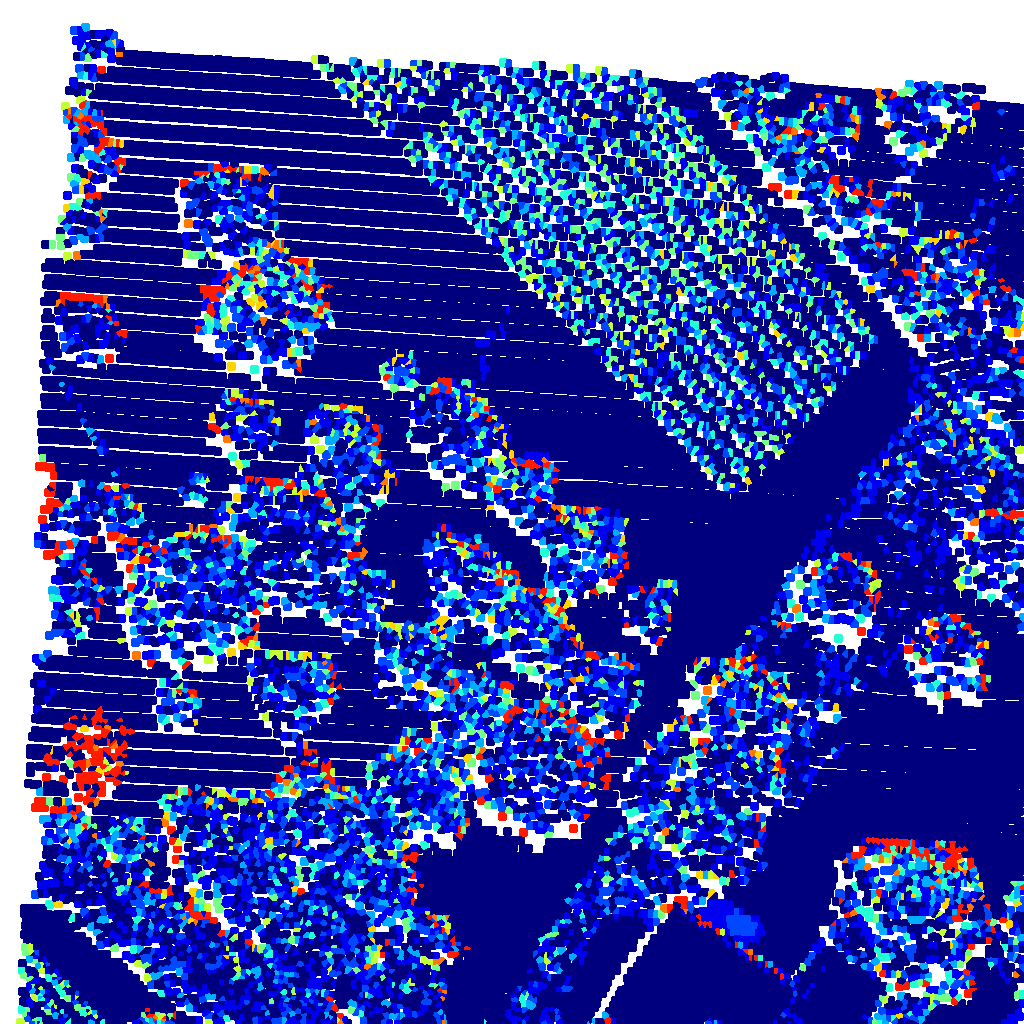}
		\includegraphics[width=\linewidth]{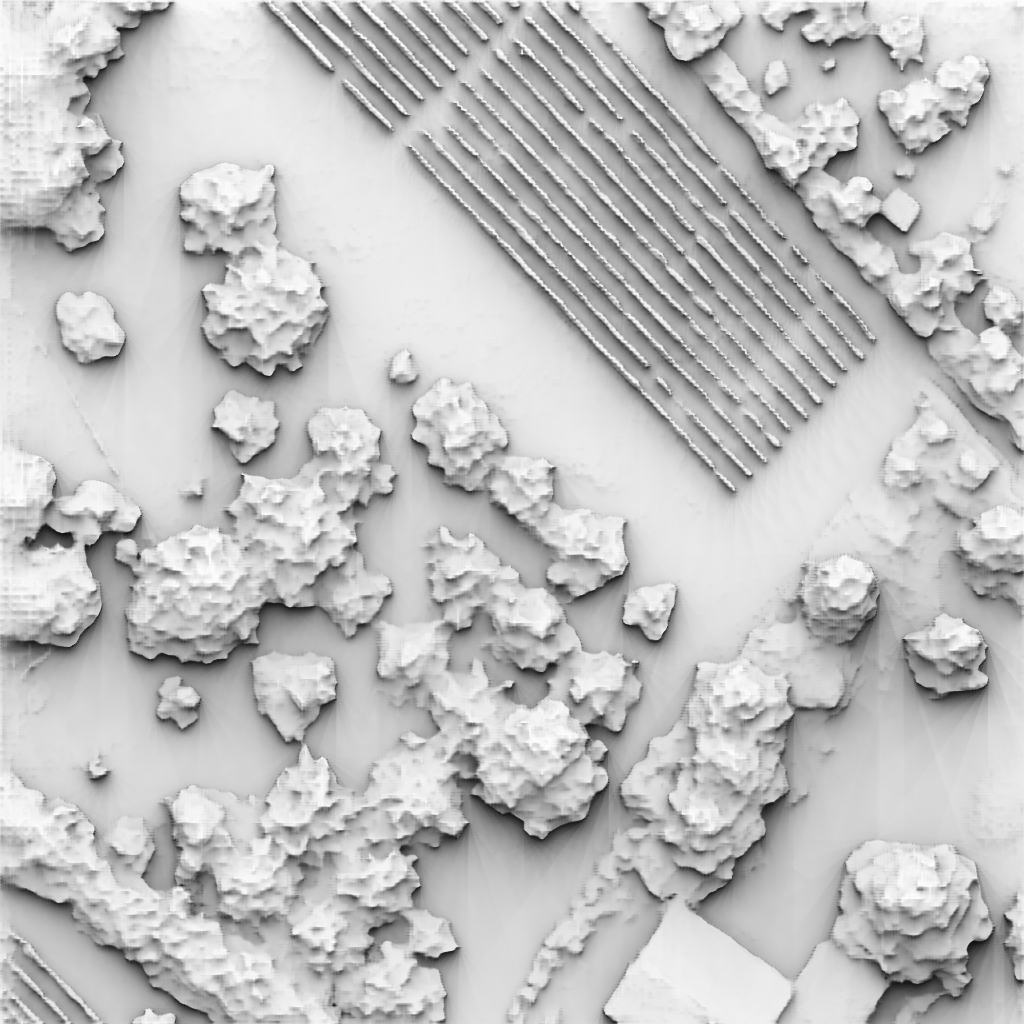}
		\centering{\tiny PSM net}
	\end{minipage}
	\begin{minipage}[t]{0.19\textwidth}	
		\includegraphics[width=0.098\linewidth]{figures/color_map.png}
		\includegraphics[width=0.85\linewidth]{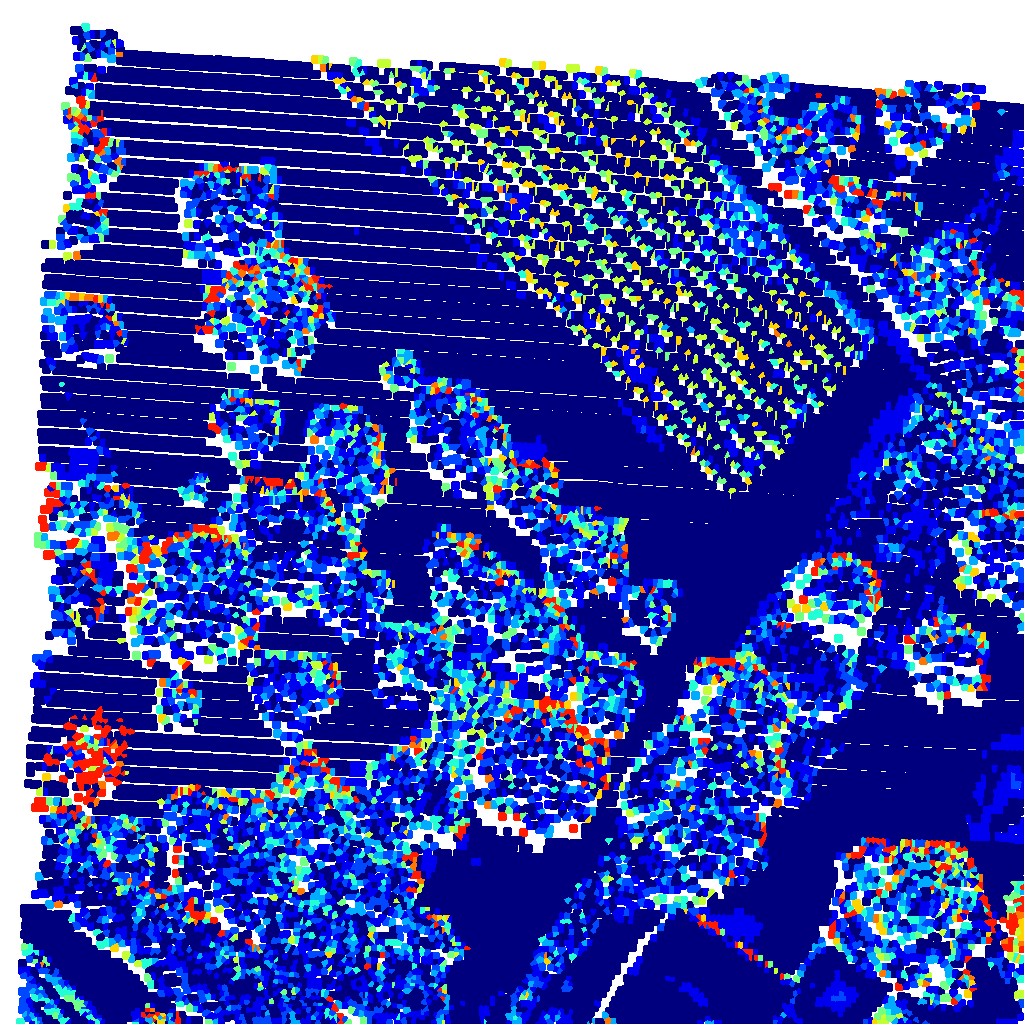}
		\includegraphics[width=\linewidth]{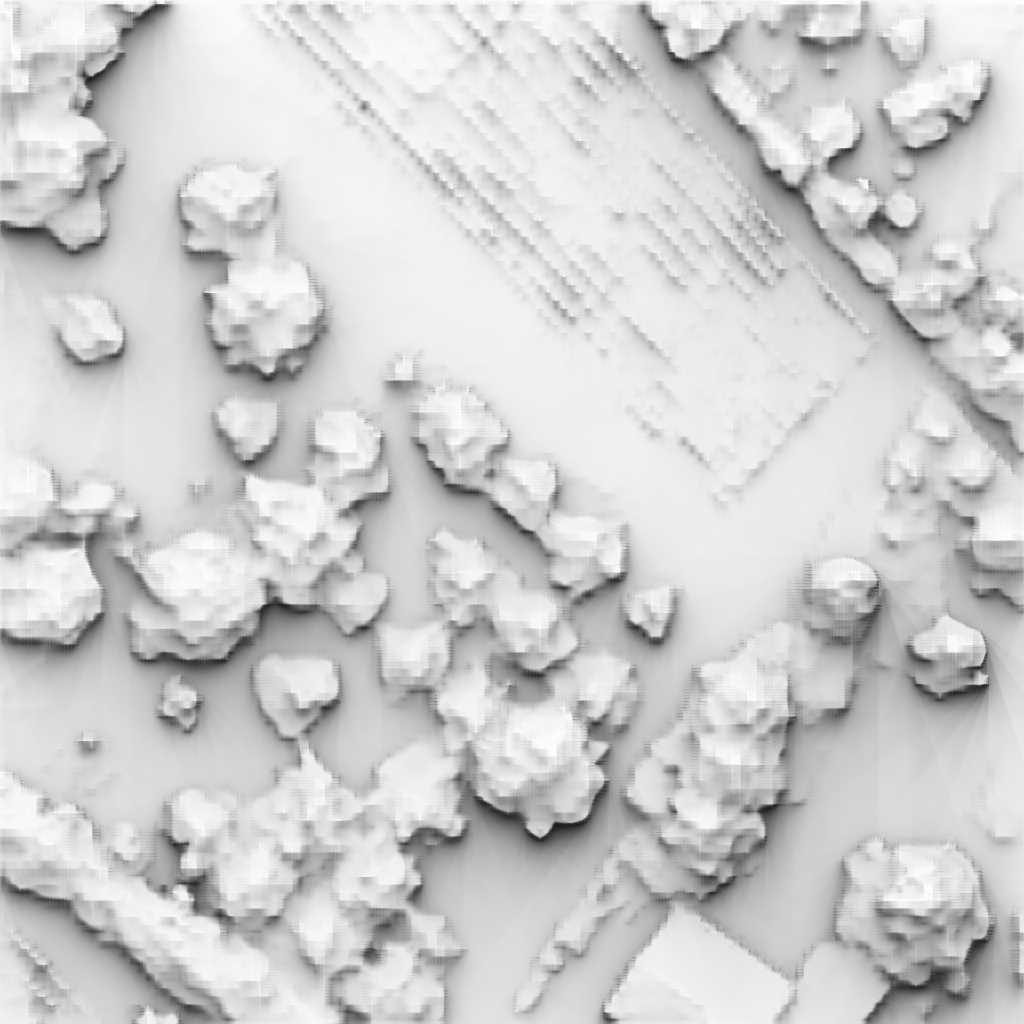}
		\centering{\tiny HRS net}
	\end{minipage}
	\begin{minipage}[t]{0.19\textwidth}	
		\includegraphics[width=0.098\linewidth]{figures/color_map.png}
		\includegraphics[width=0.85\linewidth]{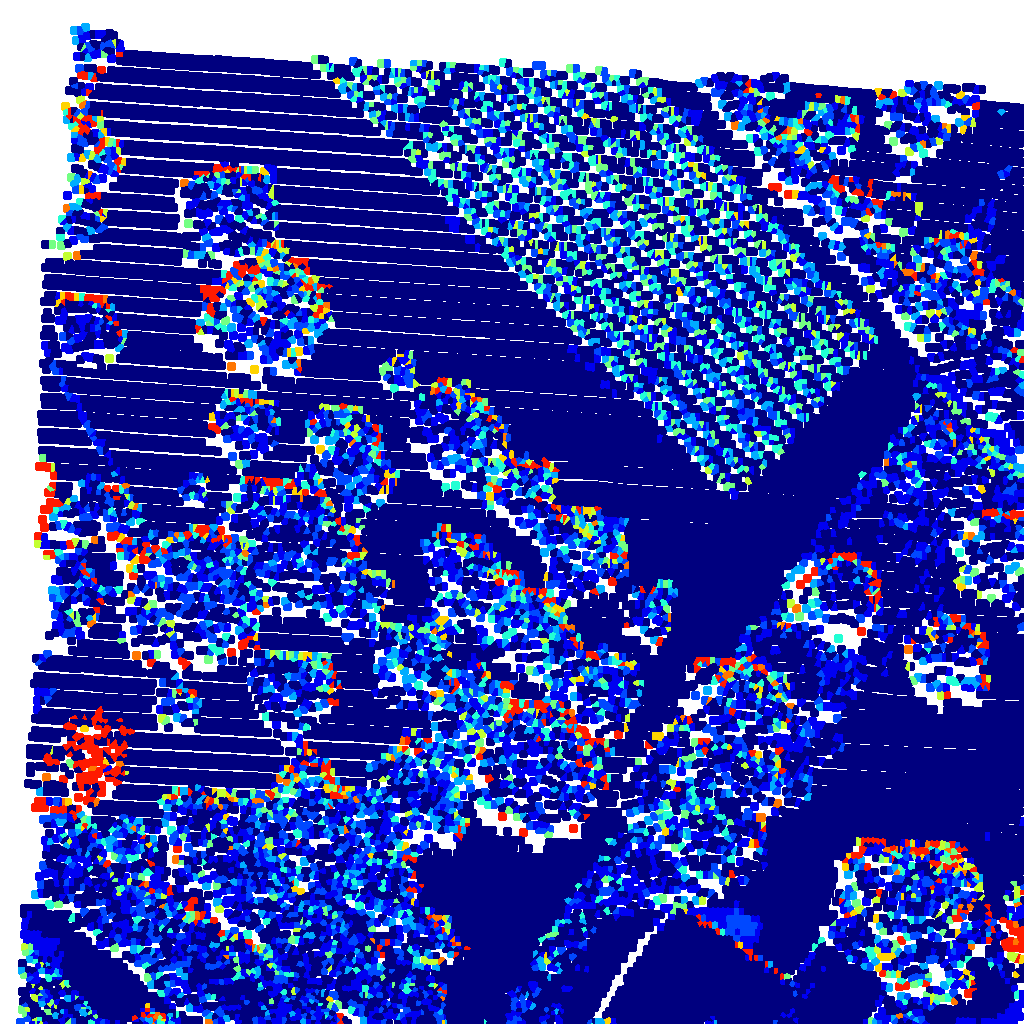}
		\includegraphics[width=\linewidth]{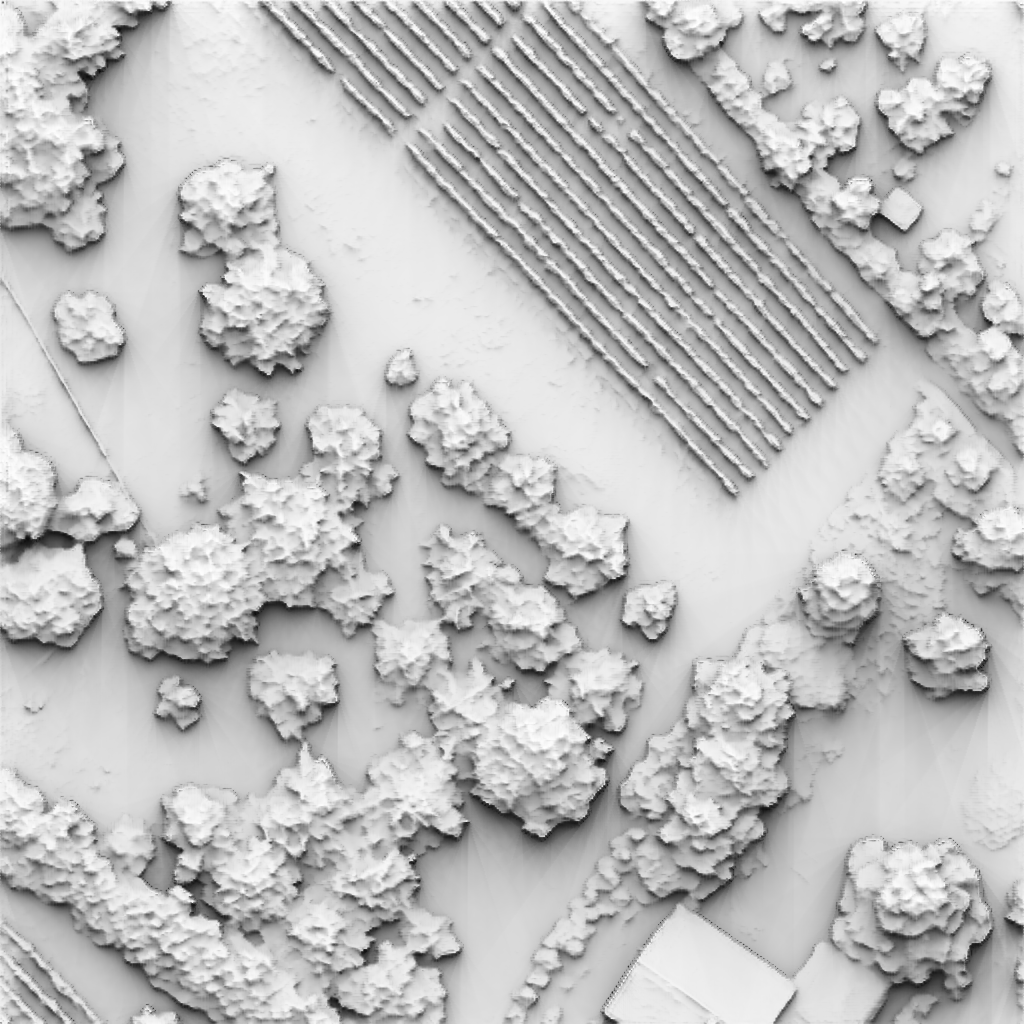}
		\centering{\tiny DeepPruner}
	\end{minipage}
	\begin{minipage}[t]{0.19\textwidth}
		\includegraphics[width=0.098\linewidth]{figures/color_map.png}
		\includegraphics[width=0.85\linewidth]{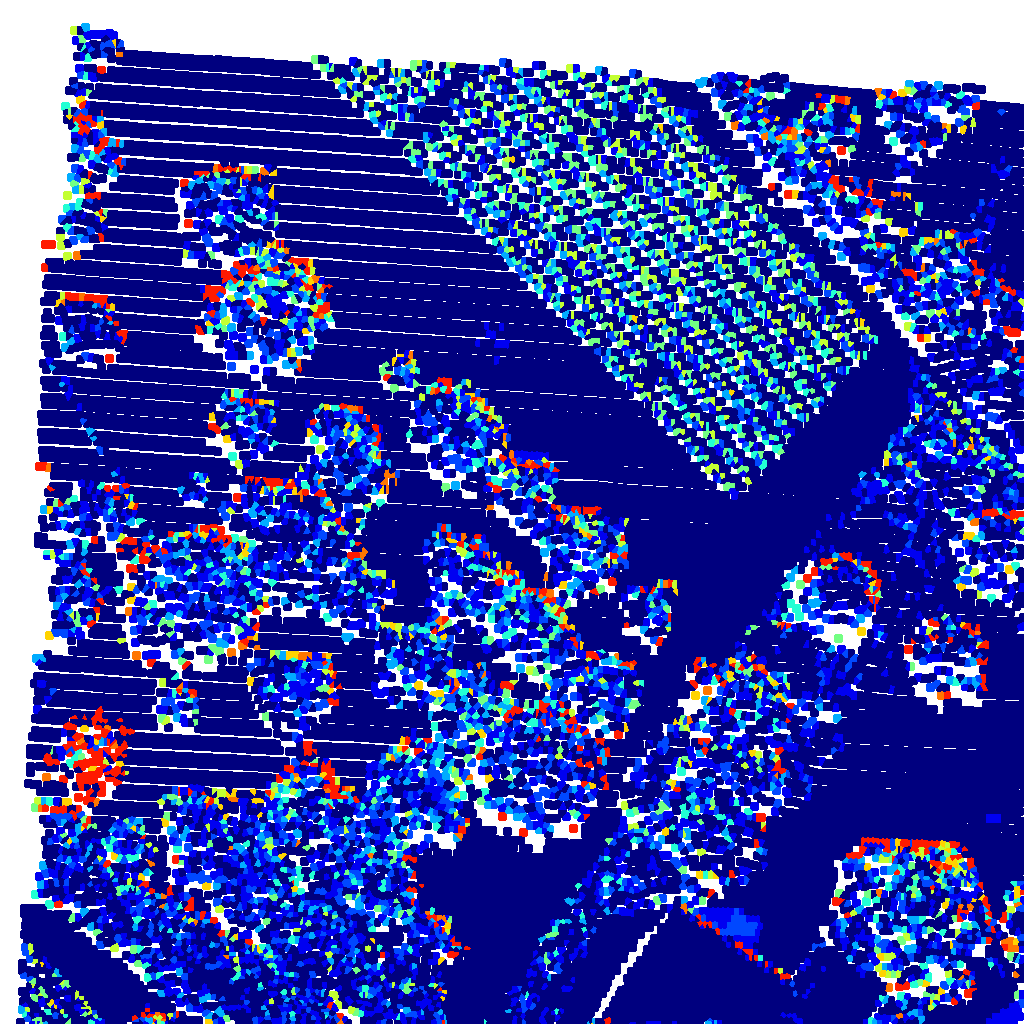}
		\includegraphics[width=\linewidth]{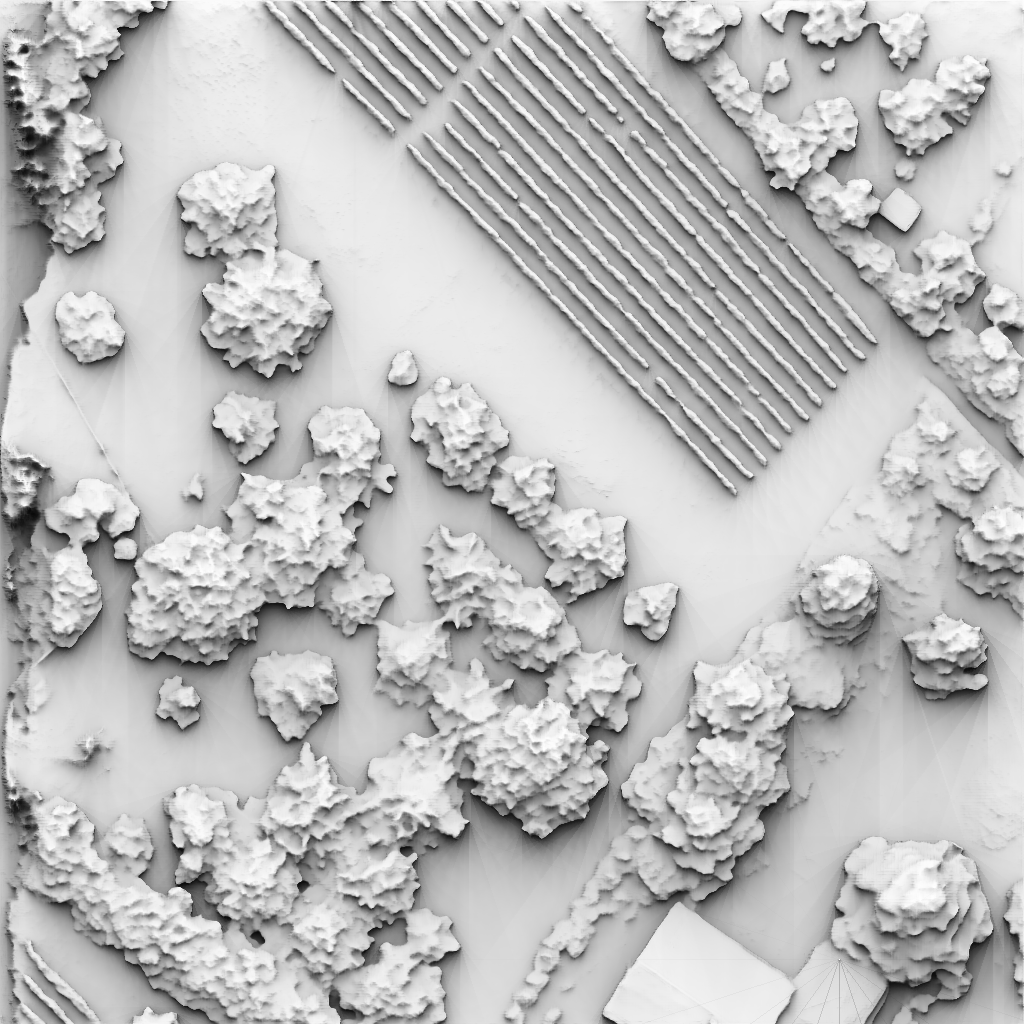}
		\centering{\tiny GANet}
	\end{minipage}
	\begin{minipage}[t]{0.19\textwidth}	
		\includegraphics[width=0.098\linewidth]{figures/color_map.png}
		\includegraphics[width=0.85\linewidth]{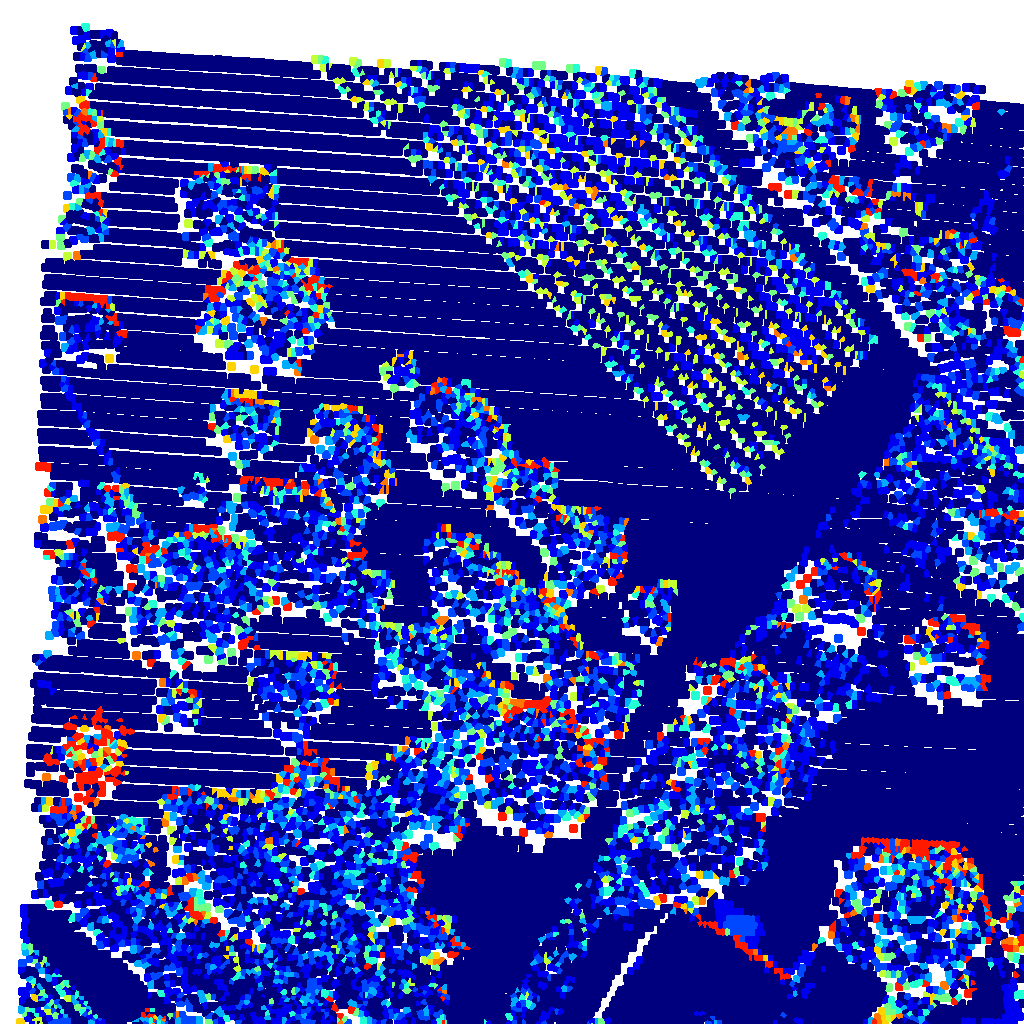}
		\includegraphics[width=\linewidth]{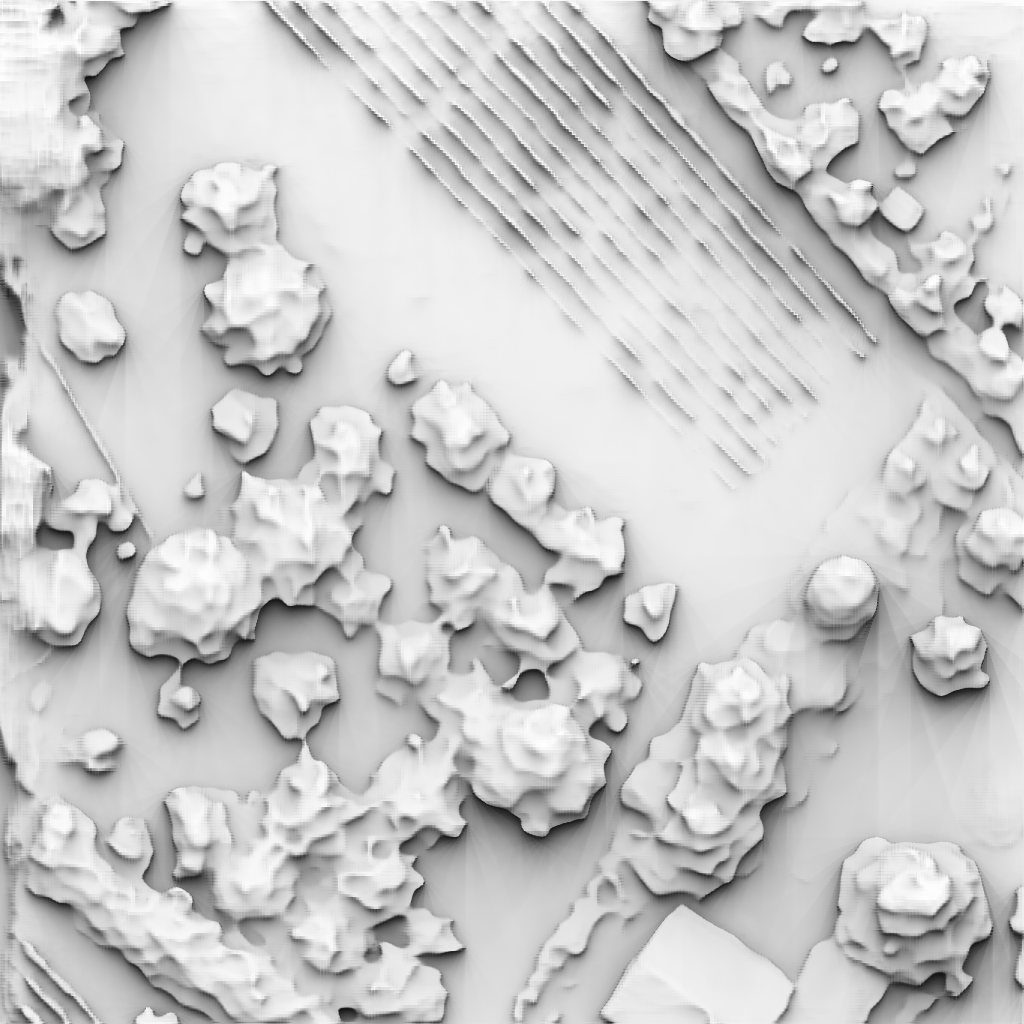}
		\centering{\tiny LEAStereo}
	\end{minipage}
	\caption{Error map and disparity visualization on the vegetable area for ISPRS Vaihingen dataset.}
	\label{Figure.vaihingentree}
\end{figure}

While $N$-pixel error is a statistical metric giving a global evaluation, dense matching results are locally influenced by the scene.
When the scene is complex, errors can show different patterns on various scene types (buildings, ground, vegetation,...).
In order to give a visual assessment of the methods, we show some results in a building scene and a vegetable scene of the ISPRS Vaihingen dataset \Cref{Figure.vaihingen_example}.
The results on the other 5 datasets are provided in the appendix.
Because the ground truth is sparse, the valid evaluated pixel is sparse too but we have interpolated it with the nearest point in order to facilitate the interpretation.
The disparity maps are visualized using an ambient occlusion shading implemented in MicMac.
The result is shown in \Cref{Figure.vaihingenbulding} and \Cref{Figure.vaihingentree}.

\begin{figure}[t]
	\centering
	\subfigure[Left image]{
		\label{Figure.vaihingen_example:a}
		\centering
		\includegraphics[width=0.2\linewidth]{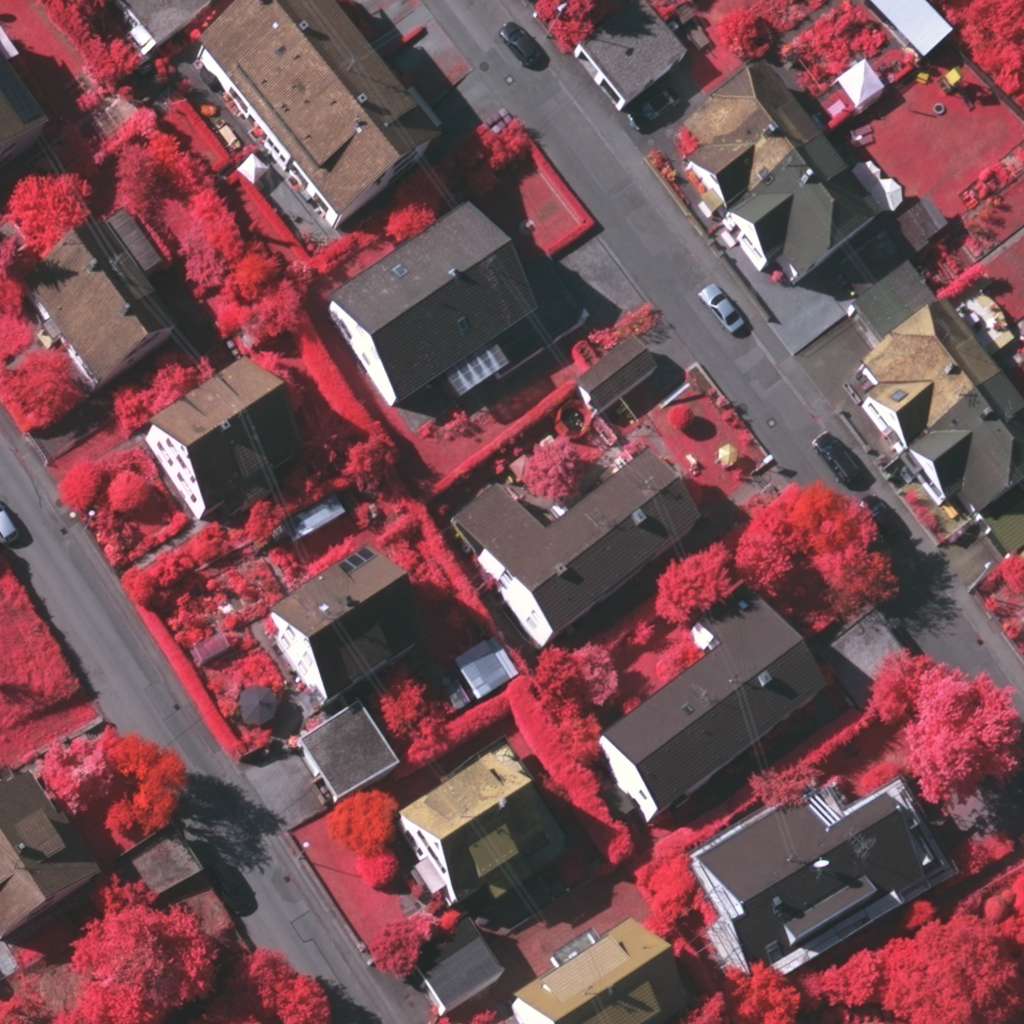}
	}
	\subfigure[GT dispairty]{
		\label{Figure.vaihingen_example:b}
		\centering
		\includegraphics[width=0.2\linewidth]{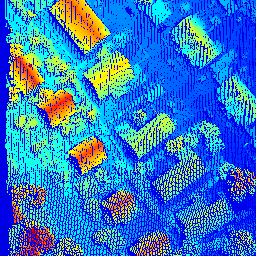}
	}
	\subfigure[Left image]{
		\label{Figure.vaihingen_example:c}
		\centering
		\includegraphics[width=0.2\linewidth]{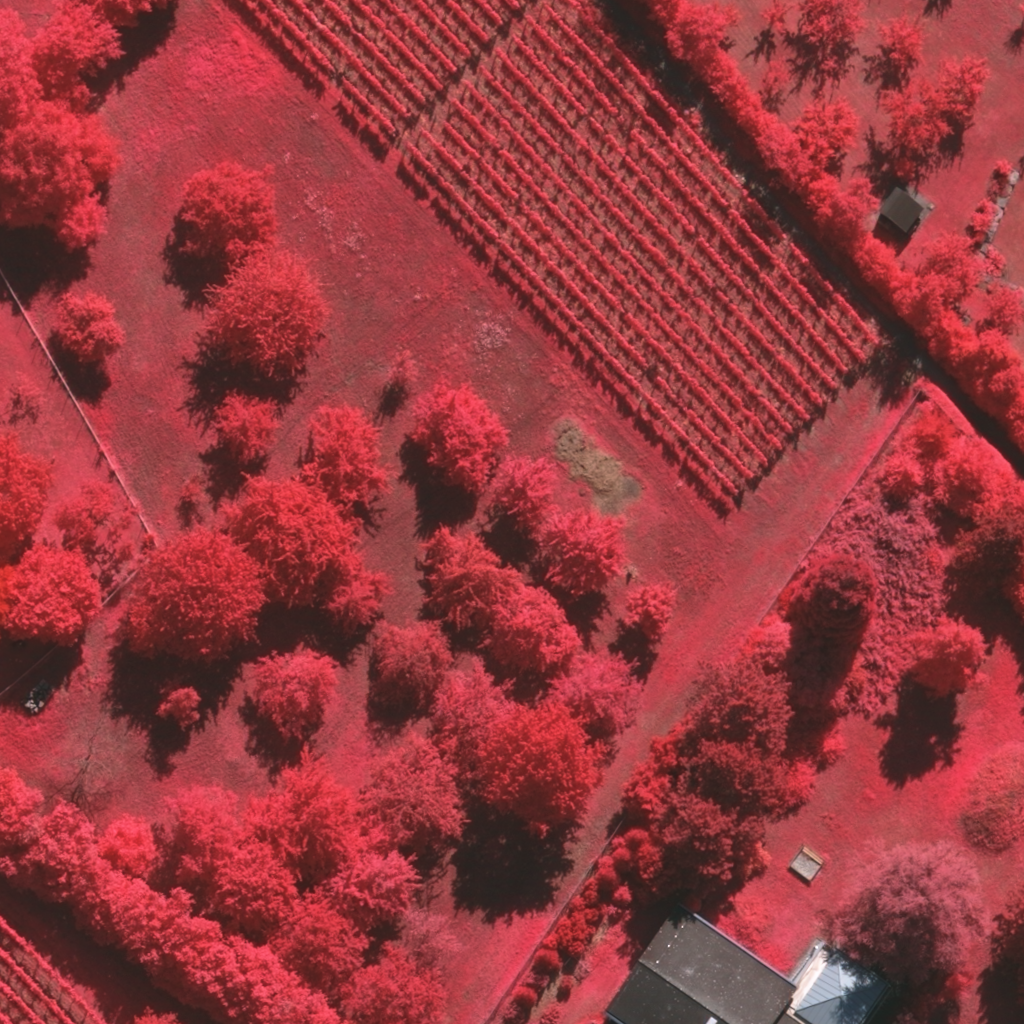}
	}
	\subfigure[GT dispairty]{
		\label{Figure.vaihingen_example:d}
		\centering
		\includegraphics[width=0.2\linewidth]{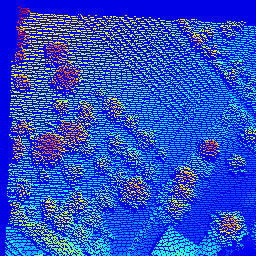}
	}
	\caption{Tree and building extracts from the ISPRS Vaihingen dataset.}
    \label{Figure.vaihingen_example}
\end{figure}

On buildings, the disparity is usually smooth on the roof, with a large discontinuity on the boundary.
The error map and disparity map are shown in \Cref{Figure.vaihingenbulding} and the left image in \Cref{Figure.vaihingen_example:a}.
We see that the GraphCuts-based methods produce a smooth disparity which is good on the roofs but bad on the discontinuity.
Conversely, after fine-tuning, DL-based methods perform well on the building's boundary (disparity discontinuity). 
The pre-trained model is not good, because the scene is quite different from KITTI, among them, the \textit{DeepPruner} gives the worst result, and the fine-tune methods are better than the traditional methods.

The vegetable area is different from the man-made structure, the disparity of vegetable changes frequently, but not large, the error map and disparity map visualization is shown in \Cref{Figure.vaihingentree}, and the left image is shown in \Cref{Figure.vaihingen_example:c}. The pre-trained result of \textit{DeepPruner} is poor, which means that \textit{DeepPruner} is highly dependent on the training dataset. Because there is a plane constraint in the \textit{GraphCuts} method, the result are not satisfactory on vegetable boundaries. 
The end-to-end methods give a smooth result, especially for the bush area.
The \textit{PSM net}, \textit{DeepPruner}, and \textit{GANet} give a good result.

\section{Conclusion}

In this paper, we propose a method to generate training data for DL-based stereo-dense matching methods, and produce a new big dataset from the aerial dataset. The state-of-the-art methods are evaluated on this dataset, containing both traditional and learning base methods. 
Dataset shift is an important research point for the application because it is impossible to obtain training data for every application.
Several factors in dataset shift are investigated, for example different resolutions, different scenes, image color and brightness, and base to height ratio. 
We summary a conclusion from the experiment :
\begin{itemize}
	\item ML depends on the training dataset, especially the end-to-end methods. Using DL-based method can improve performance significantly.
	\item For dataset shift, models from high-resolution datasets have a better performance than low resolution.
	\item Using the same area to train the model can improve performance, even with different resolutions.
	\item Hybrid methods generalize better than end-to-end methods, on the other handle, end-to-end methods improve more when the training and testing data is same type. 
	\item The transferability is not symmetric, and depends on the network structure.
	\item Pre-trained methods work better in building areas than in tree areas.  Using aerial data to do the fine-tuning can improve performance in the building areas with discontinuity. 
\end{itemize}

In the training dataset generation, because the image and LiDAR are not acquired at the same time, building changes can be removed easily, but vegetable change is difficult to remove. 
Use the filtering to remove the low-accuracy stereo pairs, unfortunately, some positive samples are removed. On one handle, the difficult matching areas are removed, for example reflective areas, on the other hand, it is difficult to know whether the vegetable is changed or not, there is another possibility just due to matching difficulty.
Even training can improve the performance in the vegetable areas, but the vegetable areas are removed because of the low evaluation. Semantic information can be utilized as supplement information in future work.
Dataset shift is an important and difficult topic for dense matching in photogrammetry applications, improving the performance of dataset shift is future work.
It is difficult to obtain the image and LiDAR at the same time, self-supervised or unsupervised learning is also a promising solution. 

\section{Acknowledgments}
\label{sec:ack}

This work was fully funded by the AI4GEO project.

\appendix



%
\section{Introduction}
Below we provide additional experiments, including complementary metrics such as 1-pixel and the average errors, as well as the complete data-shift comparison of all of the investigated DL-based dense image matching methods. Visual assessments are also given for all of the studied dataset and  methods.


\section{ Quantitative evaluation and discussion}

\subsection{\textit{1-pixel} and \textit{average} error metrics}


We report the 1-pixel error (cf. \Cref{Table:quantity_1pixel}) and the average error metrics (cf. \cref{Table:quantity_ave_pixel}) because they reflect the image matching accuracy and the robustness of a method, respectively. 
The color-coding in the tables ranges from red to green, where red corresponds to worse and green to better performance in relation to SGM (CUDA) which we consider as the baseline. The training datasets and methods are in rows, and the testing datasets are in columns. In other words, an entire column indicates the transferability of a dataset, while an entire row indicates the transferability of a method.
In particular, for the 1-pixel error, we note that:

\begin{itemize}
    \item Base-to-height ratio ($B/H$)  influences the performance. Compare Toulouse Metropole ($B/H$) and Toulouse Metropole, the transferability of the model trained on Toulouse Metropole ($B/H$) is sometimes worse than that of the model trained on Toulouse Metropole.
    \item For Toulouse Metropole, all rows are in red and orange color, which means training methods can not improve compare to SMG(CUDA) on the small ($B/H$) dataset.
    \item HRS Net method underperforms in dataset shift, but GANet and LEAStereo  generalize better than the other end-to-end methods.
    \item Except for the model KITTI, nearly all the rows of Enschede and Toulouse UMBRA are in positive color, which means training on aerial data can improve performance.
    \item For the columns of Vaihingen EuroSDR and DublinCity, most values are positive compared to the other columns, this means these two datasets are good for training.
\end{itemize}


Contrary to the 1-pixel error, the average error (cf. \Cref{Table:quantity_ave_pixel}) shows less variability in the \textit{relative shift gain} across the tested methods. In particular, we note that:

\begin{itemize}
	\item Toulouse Metropole ($B/H$) gives a better result than Toulouse Metropole.
    \item For training on the aerial dataset, most values are positive, which means training on aerial data can improve the dense matching
    \item End-to-end methods are better than hybrid methods
    \item The large negative values are in DublinCity, which indicates training on low-resolution data can not improve the performance when deployed on high-resolution data.
    \item Globally, PSMNet sometimes performs better than the other DL-based methods, but GANet and LEAStereo are better at transferability to new scenes.
\end{itemize}

\begin{table}[!ht]
	\centering
	\caption{Relative shift gain on 1-pixel error based on SGM(CUDA)}
	\label{Table:quantity_1pixel}
	\resizebox{\textwidth}{!}{
		\begin{tabular}{c|c|c|cc|ccc|c|c|cc}
			\noalign{\hrule height 2pt}
			\multirow{2}{*}{\backslashbox{Test}{Train}} & \multirow{2}{*}{Method} & \multirow{2}{*}{KITTI} &  ISPRS  & EuroSDR  & Touluose & Touluose  & Toulouse  & \multirow{2}{*}{Enschede} & Dublin & Aerial & Aerial \\
			~ & ~ & ~  &  Vaihingen &  Vaihingen &  UMBRA &  Metropole &  Metropole(B/H) & ~ & City & (real) & (all) \\
			\noalign{\hrule height 2pt}
			\multirow{7}{*}{ISPRS} & MC-CNN & \gradient{-9.49} & \gradient{15.60} & \gradient{-9.41} & \gradient{-3.88} & \gradient{-16.54} & \gradient{-4.82} & \gradient{-18.08} & \gradient{-2.41} & \gradient{-10.77} & \gradient{-2.23} \\
			\cline{2-10}
			~ & EfficientDeep & \gradient{-8.00} & \gradient{9.64} & \gradient{-5.86} & \gradient{-7.92} & \gradient{-10.25} & \gradient{-10.02} & \gradient{-12.30} & \gradient{-4.37} & \gradient{-7.04} & \gradient{-6.15} \\
			\cline{2-10}
			~ & PSMNet & \gradient{-76.25} & \gradient{32.23} & \gradient{7.01} & \gradient{-8.78} & \gradient{-13.14} & \gradient{-2.03} & \gradient{-0.07} & \gradient{1.27} & \gradient{3.49} & \gradient{14.44} \\
			\cline{2-10}
			~ & Deeppruner & \gradient{-45.46} & \gradient{29.05} & \gradient{1.72} & \gradient{-9.29} & \gradient{-4.37} & \gradient{-13.81} & \gradient{-5.22} & \gradient{1.76} & \gradient{3.79} & \gradient{15.88} \\
			\cline{2-10}
			~ & HRS Net & \gradient{-15.37} & \gradient{17.12} & \gradient{17.12} & \gradient{-17.23} & \gradient{-10.88} & \gradient{-6.96} & \gradient{-9.20} & \gradient{-12.25} & \gradient{-6.00} & \gradient{3.59} \\
			\cline{2-10}
			~ & GANet & \gradient{-23.00} & \gradient{32.82} & \gradient{8.72} & \gradient{-13.93} & \gradient{-3.27} & \gradient{-5.37} & \gradient{-5.18} & \gradient{-8.20} & \gradient{-5.56} & \gradient{14.68} \\
			\cline{2-10}
			~ & LEAStereo & \gradient{-35.56} & \gradient{15.69} & \gradient{5.51} & \gradient{-2.86} & \gradient{1.53} & \gradient{2.40} & \gradient{6.10} & \gradient{3.30} & \gradient{2.33} & \gradient{5.36} \\
			\cline{2-10}
			\noalign{\hrule height 2pt}
			\multirow{7}{*}{EuroSDR} & MC-CNN & \gradient{-3.43} & \gradient{9.85} & \gradient{4.62} & \gradient{2.37} & \gradient{6.56} & \gradient{-3.14} & \gradient{-0.74} & \gradient{0.58} & \gradient{5.26} & \gradient{4.22} \\
			\cline{2-10}
			~ & EfficientDeep & \gradient{-10.20} & \gradient{-5.36} & \gradient{2.42} & \gradient{-10.51} & \gradient{-7.06} & \gradient{-8.25} & \gradient{2.56} & \gradient{-3.66} & \gradient{-6.71} & \gradient{-6.71} \\
			\cline{2-10}
			~ & PSMNet & \gradient{-76.11} & \gradient{-1.48} & \gradient{18.30} & \gradient{-8.65} & \gradient{-12.87} & \gradient{0.98} & \gradient{3.89} & \gradient{7.50} & \gradient{8.14} & \gradient{14.11} \\
			\cline{2-10}
			~ & Deeppruner & \gradient{-26.33} & \gradient{1.22} & \gradient{18.19} & \gradient{-19.02} & \gradient{2.50} & \gradient{-17.87} & \gradient{5.94} & \gradient{9.77} & \gradient{6.22} & \gradient{14.57} \\
			\cline{2-10}
			~ & HRS Net & \gradient{-21.36} & \gradient{-22.30} & \gradient{5.01} & \gradient{-20.25} & \gradient{-10.71} & \gradient{-14.88} & \gradient{-9.69} & \gradient{-9.17} & \gradient{-0.89} & \gradient{-1.04} \\
			\cline{2-10}
			~ & GANet & \gradient{-14.55} & \gradient{4.15} & \gradient{24.23} & \gradient{-9.23} & \gradient{6.65} & \gradient{0.97} & \gradient{7.80} & \gradient{7.14} & \gradient{8.01} & \gradient{13.80} \\
			\cline{2-10}
			~ & LEAStereo & \gradient{-40.21} & \gradient{1.04} & \gradient{8.27} & \gradient{-7.07} & \gradient{0.39} & \gradient{1.71} & \gradient{6.22} & \gradient{4.39} & \gradient{3.64} & \gradient{5.67} \\
			\cline{2-10}
			\noalign{\hrule height 2pt}
			\multirow{7}{*}{UMBRA} & MC-CNN & \gradient{4.60} & \gradient{3.61} & \gradient{1.83} & \gradient{13.31} & \gradient{6.92} & \gradient{2.94} & \gradient{-0.01} & \gradient{5.91} & \gradient{3.84} & \gradient{10.30} \\
			\cline{2-10}
			~ & EfficientDeep & \gradient{3.50} & \gradient{7.54} & \gradient{0.22} & \gradient{13.28} & \gradient{5.44} & \gradient{3.83} & \gradient{-2.71} & \gradient{3.41} & \gradient{4.24} & \gradient{10.62} \\
			\cline{2-10}
			~ & PSMNet & \gradient{-61.01} & \gradient{1.79} & \gradient{7.58} & \gradient{31.33} & \gradient{3.34} & \gradient{5.93} & \gradient{3.57} & \gradient{15.48} & \gradient{15.53} & \gradient{19.99} \\
			\cline{2-10}
			~ & Deeppruner & \gradient{-25.98} & \gradient{-4.50} & \gradient{9.11} & \gradient{24.67} & \gradient{14.86} & \gradient{2.81} & \gradient{5.38} & \gradient{14.17} & \gradient{15.87} & \gradient{21.06} \\
			\cline{2-10}
			~ & HRS Net & \gradient{-11.59} & \gradient{-8.22} & \gradient{-7.27} & \gradient{12.88} & \gradient{-8.22} & \gradient{-8.13} & \gradient{-6.39} & \gradient{-1.97} & \gradient{3.64} & \gradient{10.20} \\
			\cline{2-10}
			~ & GANet & \gradient{3.50} & \gradient{14.32} & \gradient{12.88} & \gradient{27.96} & \gradient{11.43} & \gradient{2.66} & \gradient{8.97} & \gradient{15.95} & \gradient{17.52} & \gradient{19.73} \\
			\cline{2-10}
			~ & LEAStereo & \gradient{-26.42} & \gradient{6.76} & \gradient{3.57} & \gradient{21.96} & \gradient{4.64} & \gradient{4.14} & \gradient{13.69} & \gradient{13.56} & \gradient{11.22} & \gradient{12.84} \\
			\cline{2-10}
			\noalign{\hrule height 2pt}
			\multirow{7}{*}{Metropole} & MC-CNN & \gradient{-16.80} & \gradient{-7.96} & \gradient{-11.62} & \gradient{-10.88} & \gradient{-1.85} & \gradient{-5.81} & \gradient{-11.77} & \gradient{-11.30} & \gradient{-12.35} & \gradient{-5.60} \\
			\cline{2-10}
			~ & EfficientDeep & \gradient{-11.50} & \gradient{-17.51} & \gradient{-7.13} & \gradient{-10.37} & \gradient{-2.77} & \gradient{-6.38} & \gradient{-11.18} & \gradient{-8.27} & \gradient{-8.64} & \gradient{-8.44} \\
			\cline{2-10}
			~ & PSMNet & \gradient{-85.74} & \gradient{-6.81} & \gradient{-2.33} & \gradient{-10.50} & \gradient{3.26} & \gradient{3.24} & \gradient{-6.91} & \gradient{-6.23} & \gradient{-2.89} & \gradient{-2.89} \\
			\cline{2-10}
			~ & Deeppruner & \gradient{-31.25} & \gradient{-12.77} & \gradient{-3.85} & \gradient{-11.82} & \gradient{5.87} & \gradient{4.01} & \gradient{-7.20} & \gradient{-4.25} & \gradient{-3.51} & \gradient{-0.95} \\
			\cline{2-10}
			~ & HRS Net & \gradient{-20.95} & \gradient{-12.63} & \gradient{-9.73} & \gradient{-19.41} & \gradient{-7.87} & \gradient{-7.63} & \gradient{-13.44} & \gradient{-14.31} & \gradient{-10.29} & \gradient{-10.10} \\
			\cline{2-10}
			~ & GANet & \gradient{-11.50} & \gradient{-4.44} & \gradient{0.98} & \gradient{-11.38} & \gradient{4.71} & \gradient{1.44} & \gradient{-6.48} & \gradient{-4.99} & \gradient{-2.85} & \gradient{1.56} \\
			\cline{2-10}
			~ & LEAStereo & \gradient{-39.57} & \gradient{-2.37} & \gradient{-3.53} & \gradient{-7.38} & \gradient{-1.09} & \gradient{-3.40} & \gradient{-2.94} & \gradient{-3.43} & \gradient{-2.76} & \gradient{-2.91} \\
			\cline{2-10}
			\noalign{\hrule height 2pt}
			\multirow{7}{*}{Metropole(B/H)} & MC-CNN & \gradient{-4.35} & \gradient{0.09} & \gradient{-0.31} & \gradient{0.92} & \gradient{10.29} & \gradient{10.50} & \gradient{0.19} & \gradient{0.59} & \gradient{2.77} & \gradient{23.03} \\
			\cline{2-10}
			~ & EfficientDeep & \gradient{-2.34} & \gradient{-10.70} & \gradient{3.85} & \gradient{0.16} & \gradient{5.98} & \gradient{15.21} & \gradient{25.06} & \gradient{3.82} & \gradient{3.18} & \gradient{4.34} \\
			\cline{2-10}
			~ & PSMNet & \gradient{-72.41} & \gradient{-3.14} & \gradient{7.88} & \gradient{-0.73} & \gradient{18.65} & \gradient{21.74} & \gradient{5.15} & \gradient{7.39} & \gradient{9.01} & \gradient{14.19} \\
			\cline{2-10}
			~ & Deeppruner & \gradient{-24.31} & \gradient{-4.46} & \gradient{7.94} & \gradient{-4.30} & \gradient{13.79} & \gradient{21.28} & \gradient{5.05} & \gradient{7.52} & \gradient{8.36} & \gradient{12.55} \\
			\cline{2-10}
			~ & HRS Net & \gradient{-10.04} & \gradient{-11.72} & \gradient{0.20} & \gradient{-10.43} & \gradient{1.54} & \gradient{4.90} & \gradient{-1.57} & \gradient{-3.76} & \gradient{0.27} & \gradient{2.21} \\
			\cline{2-10}
			~ & GANet & \gradient{-5.13} & \gradient{-3.29} & \gradient{9.77} & \gradient{1.87} & \gradient{14.51} & \gradient{18.02} & \gradient{5.24} & \gradient{7.73} & \gradient{9.47} & \gradient{11.79} \\
			\cline{2-10}
			~ & LEAStereo & \gradient{-34.58} & \gradient{2.56} & \gradient{5.37} & \gradient{0.45} & \gradient{5.94} & \gradient{5.27} & \gradient{5.98} & \gradient{5.80} & \gradient{6.54} & \gradient{5.37} \\
			\cline{2-10}
			\noalign{\hrule height 2pt}
			\multirow{7}{*}{Enschede} & MC-CNN & \gradient{2.56} & \gradient{2.18} & \gradient{8.36} & \gradient{4.41} & \gradient{5.09} & \gradient{4.70} & \gradient{12.17} & \gradient{9.62} & \gradient{9.16} & \gradient{12.03} \\
			\cline{2-10}
			~ & EfficientDeep & \gradient{6.74} & \gradient{3.11} & \gradient{9.05} & \gradient{2.67} & \gradient{8.53} & \gradient{5.31} & \gradient{9.99} & \gradient{8.75} & \gradient{6.95} & \gradient{7.66} \\
			\cline{2-10}
			~ & PSMNet & \gradient{-80.23} & \gradient{11.96} & \gradient{17.79} & \gradient{2.59} & \gradient{2.90} & \gradient{8.00} & \gradient{23.89} & \gradient{16.75} & \gradient{9.68} & \gradient{22.32} \\
			\cline{2-10}
			~ & Deeppruner & \gradient{-33.95} & \gradient{4.80} & \gradient{17.62} & \gradient{1.06} & \gradient{14.39} & \gradient{7.02} & \gradient{24.10} & \gradient{17.67} & \gradient{23.53} & \gradient{21.91} \\
			\cline{2-10}
			~ & HRS Net & \gradient{-3.13} & \gradient{-10.32} & \gradient{6.76} & \gradient{-5.86} & \gradient{6.17} & \gradient{5.43} & \gradient{13.36} & \gradient{4.36} & \gradient{7.83} & \gradient{11.29} \\
			\cline{2-10}
			~ & GANet & \gradient{-24.45} & \gradient{9.36} & \gradient{22.03} & \gradient{3.27} & \gradient{14.18} & \gradient{1.65} & \gradient{23.50} & \gradient{16.47} & \gradient{17.33} & \gradient{20.80} \\
			\cline{2-10}
			~ & LEAStereo & \gradient{-49.49} & \gradient{13.64} & \gradient{14.56} & \gradient{8.35} & \gradient{11.96} & \gradient{11.62} & \gradient{16.96} & \gradient{16.63} & \gradient{16.96} & \gradient{17.71} \\
			\cline{2-10}
			\noalign{\hrule height 2pt}
			\multirow{7}{*}{DublinCity} & MC-CNN & \gradient{2.42} & \gradient{-1.21} & \gradient{3.95} & \gradient{3.49} & \gradient{3.82} & \gradient{1.30} & \gradient{4.36} & \gradient{7.79} & \gradient{4.62} & \gradient{7.48} \\
			\cline{2-10}
			~ & EfficientDeep & \gradient{-1.86} & \gradient{-10.44} & \gradient{7.35} & \gradient{-3.68} & \gradient{3.66} & \gradient{0.22} & \gradient{2.71} & \gradient{10.51} & \gradient{4.84} & \gradient{8.77} \\
			\cline{2-10}
			~ & PSMNet & \gradient{-63.20} & \gradient{-17.19} & \gradient{12.50} & \gradient{-15.15} & \gradient{-45.03} & \gradient{-10.60} & \gradient{11.09} & \gradient{26.29} & \gradient{16.46} & \gradient{22.89} \\
			\cline{2-10}
			~ & Deeppruner & \gradient{-35.15} & \gradient{-0.97} & \gradient{10.52} & \gradient{-15.63} & \gradient{-14.56} & \gradient{-39.53} & \gradient{13.27} & \gradient{28.00} & \gradient{8.31} & \gradient{25.14} \\
			\cline{2-10}
			~ & HRS Net & \gradient{-2.36} & \gradient{-9.22} & \gradient{2.99} & \gradient{-12.40} & \gradient{-20.31} & \gradient{-26.36} & \gradient{4.53} & \gradient{16.17} & \gradient{0.81} & \gradient{14.14} \\
			\cline{2-10}
			~ & GANet & \gradient{-28.69} & \gradient{-9.59} & \gradient{14.07} & \gradient{-2.01} & \gradient{-11.03} & \gradient{-45.30} & \gradient{17.89} & \gradient{24.52} & \gradient{14.59} & \gradient{21.88} \\
			\cline{2-10}
			~ & LEAStereo & \gradient{-30.90} & \gradient{-2.91} & \gradient{8.02} & \gradient{3.13} & \gradient{4.30} & \gradient{5.55} & \gradient{14.68} & \gradient{17.06} & \gradient{11.23} & \gradient{14.77} \\
			\cline{2-10}
			\noalign{\hrule height 2pt}
		\end{tabular}
	}
\end{table}


\begin{table}[!ht]
	\centering
	\caption{Relative shift gain on average pixel error based on SGM(CUDA)}
	\label{Table:quantity_ave_pixel}
	\resizebox{\textwidth}{!}{
            \begin{tabular}{c|c|c|cc|ccc|c|c|cc}
			\noalign{\hrule height 2pt}
			\multirow{2}{*}{\backslashbox{Test}{Train}} & \multirow{2}{*}{Method} & \multirow{2}{*}{KITTI} &  ISPRS  & EuroSDR  & Touluose & Touluose  & Toulouse  & \multirow{2}{*}{Enschede} & Dublin & Aerial & Aerial \\
			~ & ~ & ~  &  Vaihingen &  Vaihingen &  UMBRA &  Metropole &  Metropole(B/H) & ~ & City & (real) & (all) \\
			\noalign{\hrule height 2pt}
			\multirow{7}{*}{ISPRS} & MC-CNN & \gradient{0.86} & \gradient{4.06} & \gradient{0.53} & \gradient{1.06} & \gradient{-0.86} & \gradient{2.05} & \gradient{-1.22} & \gradient{1.32} & \gradient{0.29} & \gradient{1.83} \\
			\cline{2-10}
			~ & EfficientDeep & \gradient{-0.76} & \gradient{2.70} & \gradient{0.69} & \gradient{0.29} & \gradient{-0.58} & \gradient{-0.25} & \gradient{-2.39} & \gradient{-0.02} & \gradient{0.20} & \gradient{0.71} \\
			\cline{2-10}
			~ & PSMNet & \gradient{-6.67} & \gradient{6.92} & \gradient{4.53} & \gradient{0.22} & \gradient{2.63} & \gradient{3.40} & \gradient{2.33} & \gradient{2.96} & \gradient{3.74} & \gradient{6.00} \\
			\cline{2-10}
			~ & Deeppruner & \gradient{-8.80} & \gradient{6.65} & \gradient{4.25} & \gradient{-1.06} & \gradient{2.36} & \gradient{2.82} & \gradient{-1.05} & \gradient{3.28} & \gradient{4.06} & \gradient{5.91} \\
			\cline{2-10}
			~ & HRS Net & \gradient{0.07} & \gradient{5.59} & \gradient{5.59} & \gradient{-1.93} & \gradient{1.84} & \gradient{2.65} & \gradient{2.06} & \gradient{0.77} & \gradient{2.79} & \gradient{4.59} \\
			\cline{2-10}
			~ & GANet & \gradient{-3.58} & \gradient{6.88} & \gradient{4.77} & \gradient{-0.25} & \gradient{2.04} & \gradient{2.88} & \gradient{2.19} & \gradient{1.52} & \gradient{2.75} & \gradient{5.54} \\
			\cline{2-10}
			~ & LEAStereo & \gradient{-3.73} & \gradient{5.35} & \gradient{4.13} & \gradient{0.99} & \gradient{3.54} & \gradient{3.79} & \gradient{4.37} & \gradient{3.30} & \gradient{3.76} & \gradient{4.35} \\
			\cline{2-10}
			\noalign{\hrule height 2pt}
			\multirow{7}{*}{EuroSDR} & MC-CNN & \gradient{1.38} & \gradient{2.76} & \gradient{4.15} & \gradient{1.71} & \gradient{2.17} & \gradient{2.57} & \gradient{2.35} & \gradient{1.66} & \gradient{2.79} & \gradient{3.39} \\
			\cline{2-10}
			~ & EfficientDeep & \gradient{-1.14} & \gradient{0.33} & \gradient{2.67} & \gradient{-1.07} & \gradient{0.20} & \gradient{0.30} & \gradient{2.22} & \gradient{0.02} & \gradient{0.17} & \gradient{0.17} \\
			\cline{2-10}
			~ & PSMNet & \gradient{-6.95} & \gradient{2.70} & \gradient{6.53} & \gradient{-1.75} & \gradient{3.63} & \gradient{4.34} & \gradient{2.53} & \gradient{3.38} & \gradient{4.07} & \gradient{5.74} \\
			\cline{2-10}
			~ & Deeppruner & \gradient{-3.97} & \gradient{1.59} & \gradient{6.58} & \gradient{-1.75} & \gradient{3.44} & \gradient{3.59} & \gradient{2.79} & \gradient{4.19} & \gradient{3.96} & \gradient{5.99} \\
			\cline{2-10}
			~ & HRS Net & \gradient{-1.73} & \gradient{-0.71} & \gradient{5.52} & \gradient{-1.70} & \gradient{2.55} & \gradient{2.93} & \gradient{2.32} & \gradient{1.85} & \gradient{4.23} & \gradient{4.16} \\
			\cline{2-10}
			~ & GANet & \gradient{-3.45} & \gradient{3.48} & \gradient{7.08} & \gradient{-0.11} & \gradient{3.85} & \gradient{4.95} & \gradient{3.62} & \gradient{3.59} & \gradient{4.40} & \gradient{5.20} \\
			\cline{2-10}
			~ & LEAStereo & \gradient{-7.32} & \gradient{2.17} & \gradient{4.68} & \gradient{-0.34} & \gradient{2.48} & \gradient{3.47} & \gradient{3.83} & \gradient{2.97} & \gradient{3.34} & \gradient{3.75} \\
			\cline{2-10}
			\noalign{\hrule height 2pt}
			\multirow{7}{*}{UMBRA} & MC-CNN & \gradient{2.34} & \gradient{3.73} & \gradient{1.30} & \gradient{5.08} & \gradient{1.71} & \gradient{3.08} & \gradient{0.91} & \gradient{2.37} & \gradient{2.53} & \gradient{4.07} \\
			\cline{2-10}
			~ & EfficientDeep & \gradient{2.46} & \gradient{3.52} & \gradient{0.08} & \gradient{4.73} & \gradient{2.17} & \gradient{2.79} & \gradient{-0.12} & \gradient{1.70} & \gradient{2.54} & \gradient{3.63} \\
			\cline{2-10}
			~ & PSMNet & \gradient{-6.88} & \gradient{4.36} & \gradient{4.34} & \gradient{10.19} & \gradient{5.03} & \gradient{5.05} & \gradient{3.25} & \gradient{6.52} & \gradient{6.02} & \gradient{7.56} \\
			\cline{2-10}
			~ & Deeppruner & \gradient{-7.14} & \gradient{3.54} & \gradient{4.68} & \gradient{9.55} & \gradient{5.97} & \gradient{4.03} & \gradient{2.58} & \gradient{5.97} & \gradient{5.90} & \gradient{7.72} \\
			\cline{2-10}
			~ & HRS Net & \gradient{1.55} & \gradient{2.33} & \gradient{1.97} & \gradient{8.44} & \gradient{2.33} & \gradient{3.56} & \gradient{3.86} & \gradient{4.33} & \gradient{5.61} & \gradient{7.40} \\
			\cline{2-10}
			~ & GANet & \gradient{2.46} & \gradient{5.35} & \gradient{3.99} & \gradient{9.05} & \gradient{5.20} & \gradient{4.48} & \gradient{3.81} & \gradient{5.99} & \gradient{5.78} & \gradient{7.18} \\
			\cline{2-10}
			~ & LEAStereo & \gradient{-6.44} & \gradient{5.19} & \gradient{4.22} & \gradient{8.69} & \gradient{4.07} & \gradient{4.14} & \gradient{6.14} & \gradient{5.88} & \gradient{5.44} & \gradient{6.06} \\
			\cline{2-10}
			\noalign{\hrule height 2pt}
			\multirow{7}{*}{Metropole} & MC-CNN & \gradient{-2.93} & \gradient{-1.78} & \gradient{-1.01} & \gradient{-2.10} & \gradient{-0.49} & \gradient{-0.26} & \gradient{-1.44} & \gradient{-2.67} & \gradient{-0.47} & \gradient{0.07} \\
			\cline{2-10}
			~ & EfficientDeep & \gradient{-2.00} & \gradient{-2.75} & \gradient{-1.05} & \gradient{-1.87} & \gradient{-1.00} & \gradient{-1.02} & \gradient{-2.03} & \gradient{-1.63} & \gradient{-1.48} & \gradient{-1.20} \\
			\cline{2-10}
			~ & PSMNet & \gradient{-8.19} & \gradient{-0.24} & \gradient{0.83} & \gradient{-1.42} & \gradient{4.39} & \gradient{4.18} & \gradient{1.30} & \gradient{0.93} & \gradient{2.57} & \gradient{2.57} \\
			\cline{2-10}
			~ & Deeppruner & \gradient{-5.19} & \gradient{-2.24} & \gradient{1.13} & \gradient{-2.47} & \gradient{4.38} & \gradient{4.25} & \gradient{-0.24} & \gradient{1.93} & \gradient{1.62} & \gradient{3.05} \\
			\cline{2-10}
			~ & HRS Net & \gradient{-3.54} & \gradient{-1.01} & \gradient{-0.29} & \gradient{-2.91} & \gradient{2.61} & \gradient{3.16} & \gradient{0.70} & \gradient{0.65} & \gradient{1.30} & \gradient{2.02} \\
			\cline{2-10}
			~ & GANet & \gradient{-2.00} & \gradient{0.89} & \gradient{1.94} & \gradient{-0.24} & \gradient{3.74} & \gradient{4.09} & \gradient{1.61} & \gradient{1.65} & \gradient{1.12} & \gradient{3.19} \\
			\cline{2-10}
			~ & LEAStereo & \gradient{-6.01} & \gradient{0.23} & \gradient{0.92} & \gradient{-0.13} & \gradient{2.32} & \gradient{2.00} & \gradient{1.52} & \gradient{1.05} & \gradient{1.00} & \gradient{1.54} \\
			\cline{2-10}
			\noalign{\hrule height 2pt}
			\multirow{7}{*}{Metropole(B/H)} & MC-CNN & \gradient{0.06} & \gradient{0.95} & \gradient{1.63} & \gradient{0.39} & \gradient{3.89} & \gradient{4.95} & \gradient{1.59} & \gradient{-0.20} & \gradient{2.98} & \gradient{7.55} \\
			\cline{2-10}
			~ & EfficientDeep & \gradient{0.89} & \gradient{0.44} & \gradient{2.24} & \gradient{0.81} & \gradient{2.99} & \gradient{5.69} & \gradient{6.09} & \gradient{2.16} & \gradient{2.90} & \gradient{2.86} \\
			\cline{2-10}
			~ & PSMNet & \gradient{-5.98} & \gradient{4.41} & \gradient{5.54} & \gradient{1.24} & \gradient{10.97} & \gradient{11.31} & \gradient{6.71} & \gradient{6.70} & \gradient{8.12} & \gradient{9.65} \\
			\cline{2-10}
			~ & Deeppruner & \gradient{-5.21} & \gradient{2.15} & \gradient{6.00} & \gradient{0.77} & \gradient{10.24} & \gradient{10.85} & \gradient{5.42} & \gradient{6.23} & \gradient{7.11} & \gradient{8.80} \\
			\cline{2-10}
			~ & HRS Net & \gradient{-0.52} & \gradient{1.54} & \gradient{5.02} & \gradient{1.20} & \gradient{7.20} & \gradient{9.15} & \gradient{5.20} & \gradient{4.07} & \gradient{6.37} & \gradient{7.65} \\
			\cline{2-10}
			~ & GANet & \gradient{-0.62} & \gradient{4.30} & \gradient{6.67} & \gradient{3.64} & \gradient{9.95} & \gradient{10.58} & \gradient{6.25} & \gradient{5.47} & \gradient{6.58} & \gradient{8.62} \\
			\cline{2-10}
			~ & LEAStereo & \gradient{-7.11} & \gradient{3.08} & \gradient{5.23} & \gradient{2.37} & \gradient{5.41} & \gradient{6.16} & \gradient{5.86} & \gradient{4.92} & \gradient{5.46} & \gradient{5.23} \\
			\cline{2-10}
			\noalign{\hrule height 2pt}
			\multirow{7}{*}{Enschede} & MC-CNN & \gradient{0.32} & \gradient{-0.42} & \gradient{2.50} & \gradient{-1.86} & \gradient{-2.35} & \gradient{-0.69} & \gradient{3.16} & \gradient{1.59} & \gradient{2.37} & \gradient{2.91} \\
			\cline{2-10}
			~ & EfficientDeep & \gradient{1.11} & \gradient{1.36} & \gradient{2.34} & \gradient{0.27} & \gradient{1.25} & \gradient{1.59} & \gradient{2.41} & \gradient{1.47} & \gradient{1.68} & \gradient{2.03} \\
			\cline{2-10}
			~ & PSMNet & \gradient{-11.05} & \gradient{5.68} & \gradient{5.77} & \gradient{-0.62} & \gradient{6.28} & \gradient{6.21} & \gradient{8.44} & \gradient{4.89} & \gradient{4.70} & \gradient{7.61} \\
			\cline{2-10}
			~ & Deeppruner & \gradient{-10.67} & \gradient{4.66} & \gradient{5.76} & \gradient{-0.56} & \gradient{5.88} & \gradient{5.30} & \gradient{8.30} & \gradient{5.37} & \gradient{8.07} & \gradient{7.39} \\
			\cline{2-10}
			~ & HRS Net & \gradient{1.62} & \gradient{3.60} & \gradient{4.87} & \gradient{-1.08} & \gradient{5.61} & \gradient{5.59} & \gradient{7.06} & \gradient{3.55} & \gradient{5.04} & \gradient{6.06} \\
			\cline{2-10}
			~ & GANet & \gradient{-6.65} & \gradient{5.59} & \gradient{6.54} & \gradient{0.04} & \gradient{6.11} & \gradient{5.96} & \gradient{8.10} & \gradient{4.40} & \gradient{5.33} & \gradient{6.47} \\
			\cline{2-10}
			~ & LEAStereo & \gradient{-8.56} & \gradient{5.97} & \gradient{5.72} & \gradient{1.63} & \gradient{5.18} & \gradient{5.30} & \gradient{5.55} & \gradient{4.76} & \gradient{5.55} & \gradient{5.76} \\
			\cline{2-10}
			\noalign{\hrule height 2pt}
			\multirow{7}{*}{DublinCity} & MC-CNN & \gradient{-0.15} & \gradient{-3.31} & \gradient{0.90} & \gradient{-3.25} & \gradient{-1.51} & \gradient{-0.09} & \gradient{1.07} & \gradient{2.76} & \gradient{-0.25} & \gradient{2.46} \\
			\cline{2-10}
			~ & EfficientDeep & \gradient{-0.59} & \gradient{-4.67} & \gradient{0.89} & \gradient{-4.57} & \gradient{-0.13} & \gradient{-0.02} & \gradient{0.32} & \gradient{2.02} & \gradient{0.73} & \gradient{1.78} \\
			\cline{2-10}
			~ & PSMNet & \gradient{-8.89} & \gradient{0.90} & \gradient{2.61} & \gradient{-10.92} & \gradient{-15.30} & \gradient{-1.23} & \gradient{2.42} & \gradient{6.66} & \gradient{5.24} & \gradient{6.12} \\
			\cline{2-10}
			~ & Deeppruner & \gradient{-16.47} & \gradient{1.78} & \gradient{3.79} & \gradient{-9.14} & \gradient{-9.41} & \gradient{-8.99} & \gradient{2.26} & \gradient{6.70} & \gradient{4.15} & \gradient{6.22} \\
			\cline{2-10}
			~ & HRS Net & \gradient{1.66} & \gradient{0.77} & \gradient{3.00} & \gradient{-4.45} & \gradient{-3.35} & \gradient{0.80} & \gradient{3.44} & \gradient{5.66} & \gradient{3.96} & \gradient{5.26} \\
			\cline{2-10}
			~ & GANet & \gradient{-8.63} & \gradient{-2.28} & \gradient{2.96} & \gradient{-0.49} & \gradient{0.84} & \gradient{-1.01} & \gradient{4.38} & \gradient{6.15} & \gradient{4.80} & \gradient{5.77} \\
			\cline{2-10}
			~ & LEAStereo & \gradient{-8.07} & \gradient{2.84} & \gradient{4.12} & \gradient{1.81} & \gradient{2.25} & \gradient{3.13} & \gradient{4.97} & \gradient{5.17} & \gradient{3.51} & \gradient{5.00} \\
			\cline{2-10}
			\noalign{\hrule height 2pt}
		\end{tabular}
    }
\end{table}

 
\subsection{Dataset shift comparisons across tested methods}

\paragraph{MC-CNN}

\begin{figure}[tp]
	\centering
	\subfigure[Result on ISPRS Vaihingen]{
		\label{Figure.fusion_mccnn:a}
		\centering
		\includegraphics[width=0.45\linewidth]{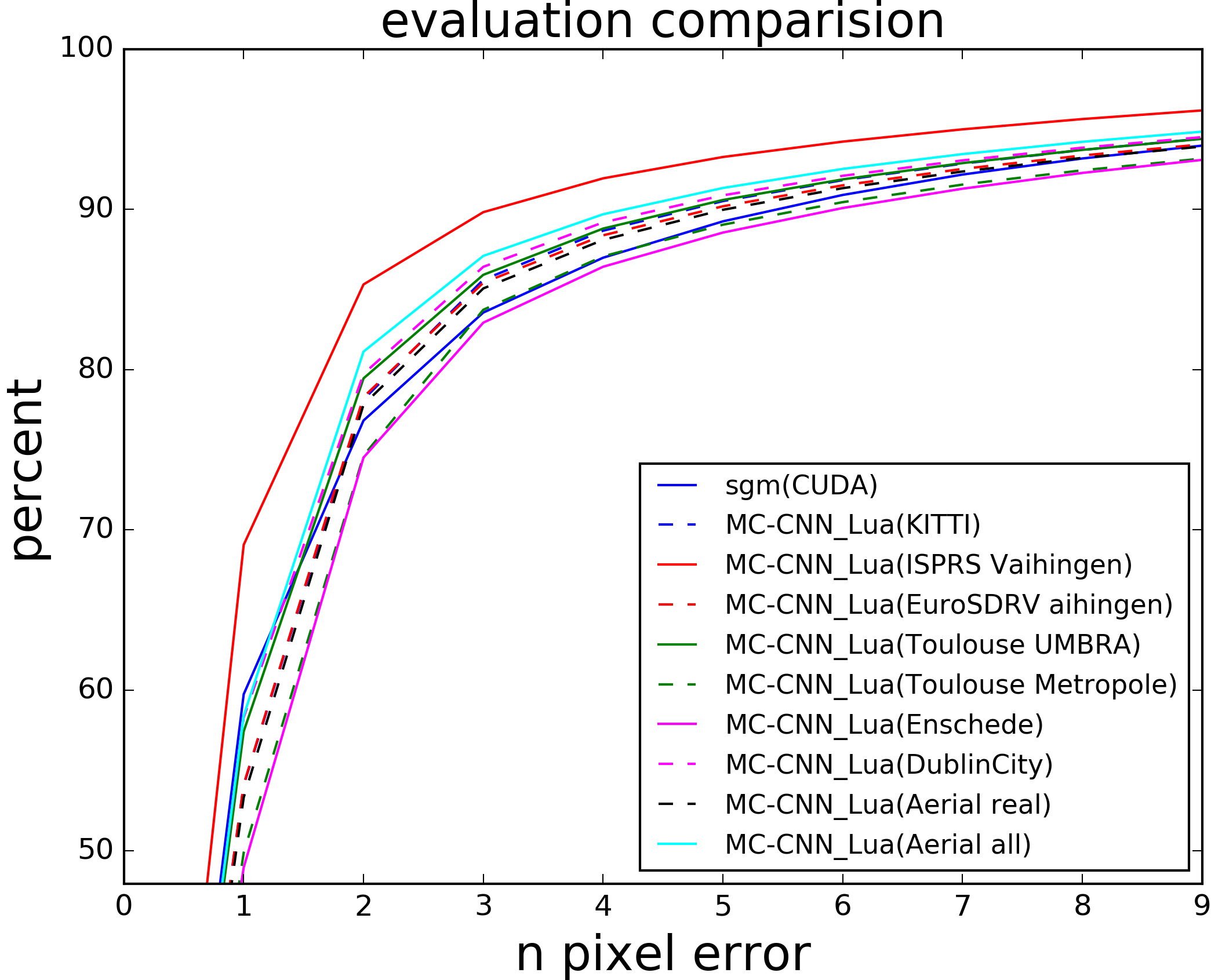}
	}
	\subfigure[Result on EuroSDR Vaihingen]{
		\label{Figure.fusion_mccnn:b}
		\centering
		\includegraphics[width=0.45\linewidth]{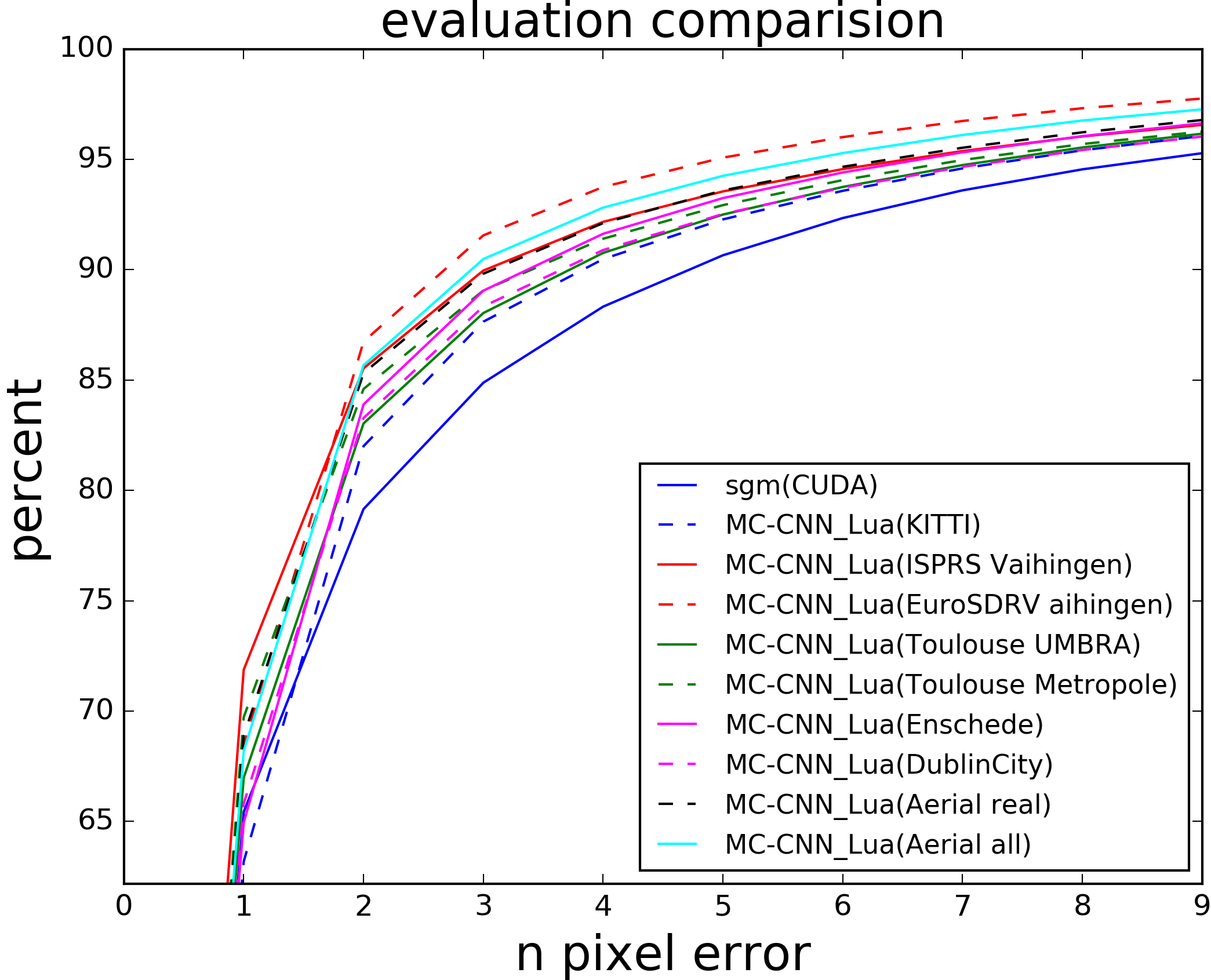}
	}
	
	\subfigure[Result on Toulouse Metropole]{
		\label{Figure.fusion_mccnn:c}
		\centering
		\includegraphics[width=0.45\linewidth]{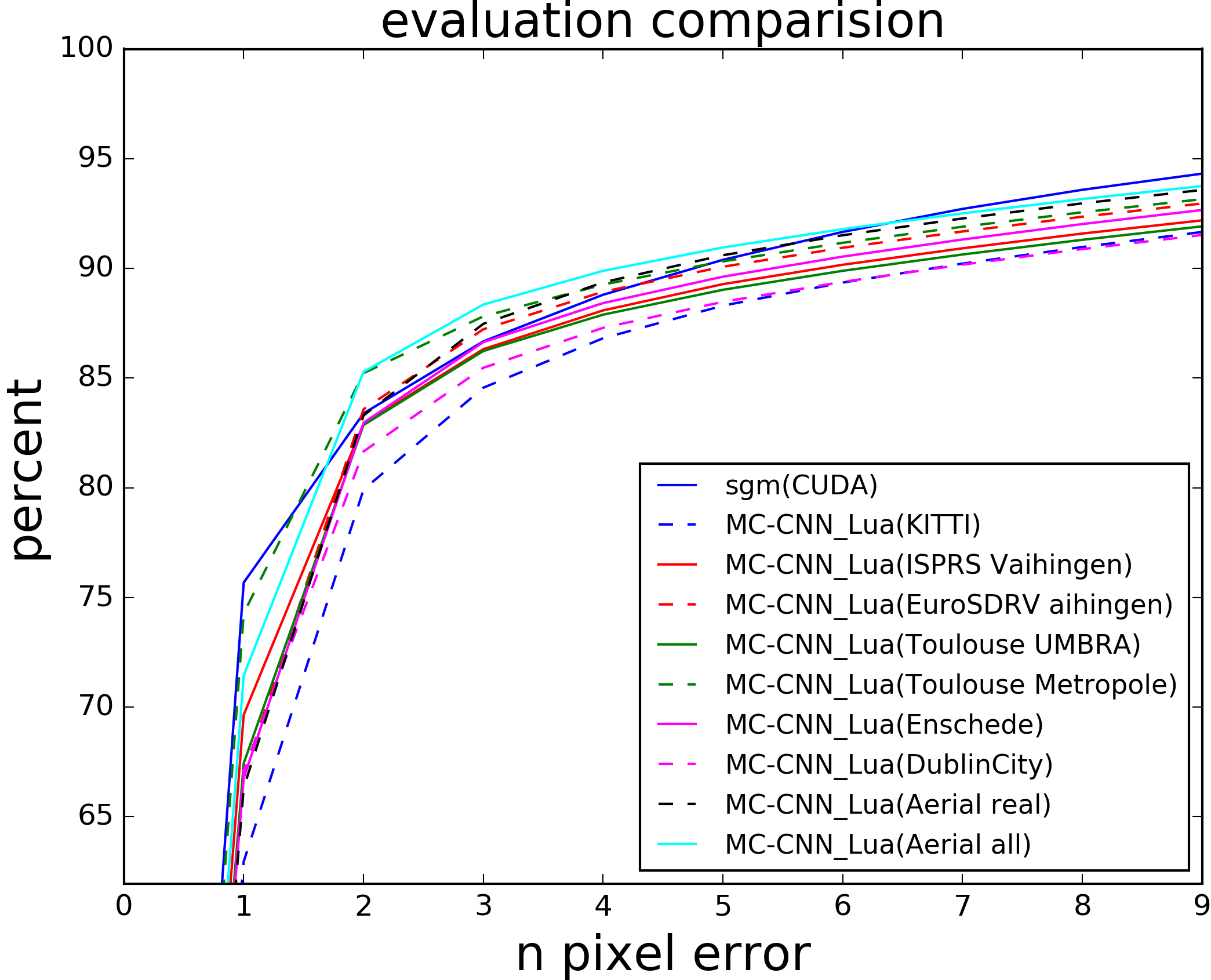}
	}
	\subfigure[Result on Toulouse UMBRA]{
		\label{Figure.fusion_mccnn:d}
		\centering
		\includegraphics[width=0.45\linewidth]{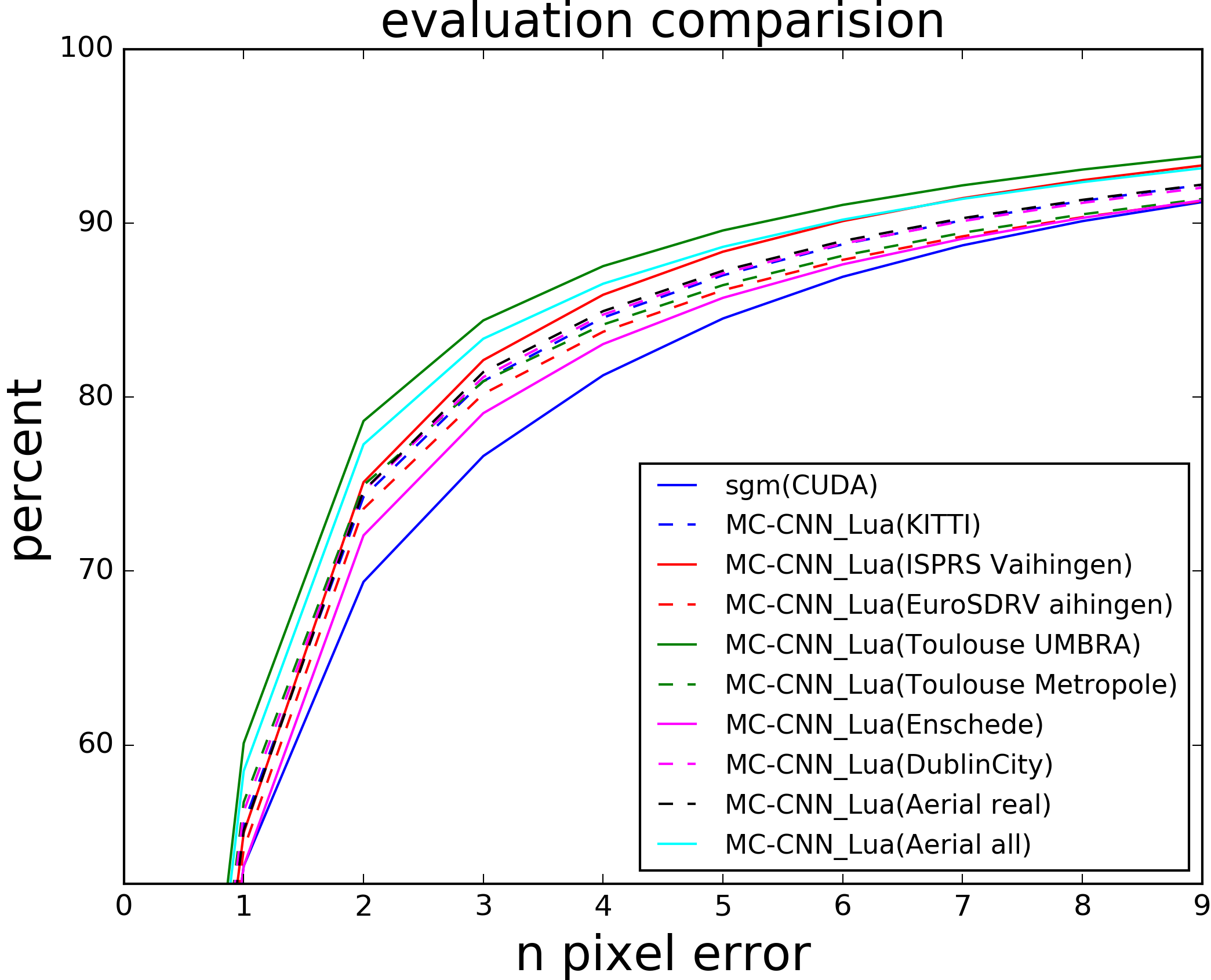}
	}
	
	\subfigure[Result on Enschede]{
		\label{Figure.fusion_mccnn:e}
		\centering
		\includegraphics[width=0.45\linewidth]{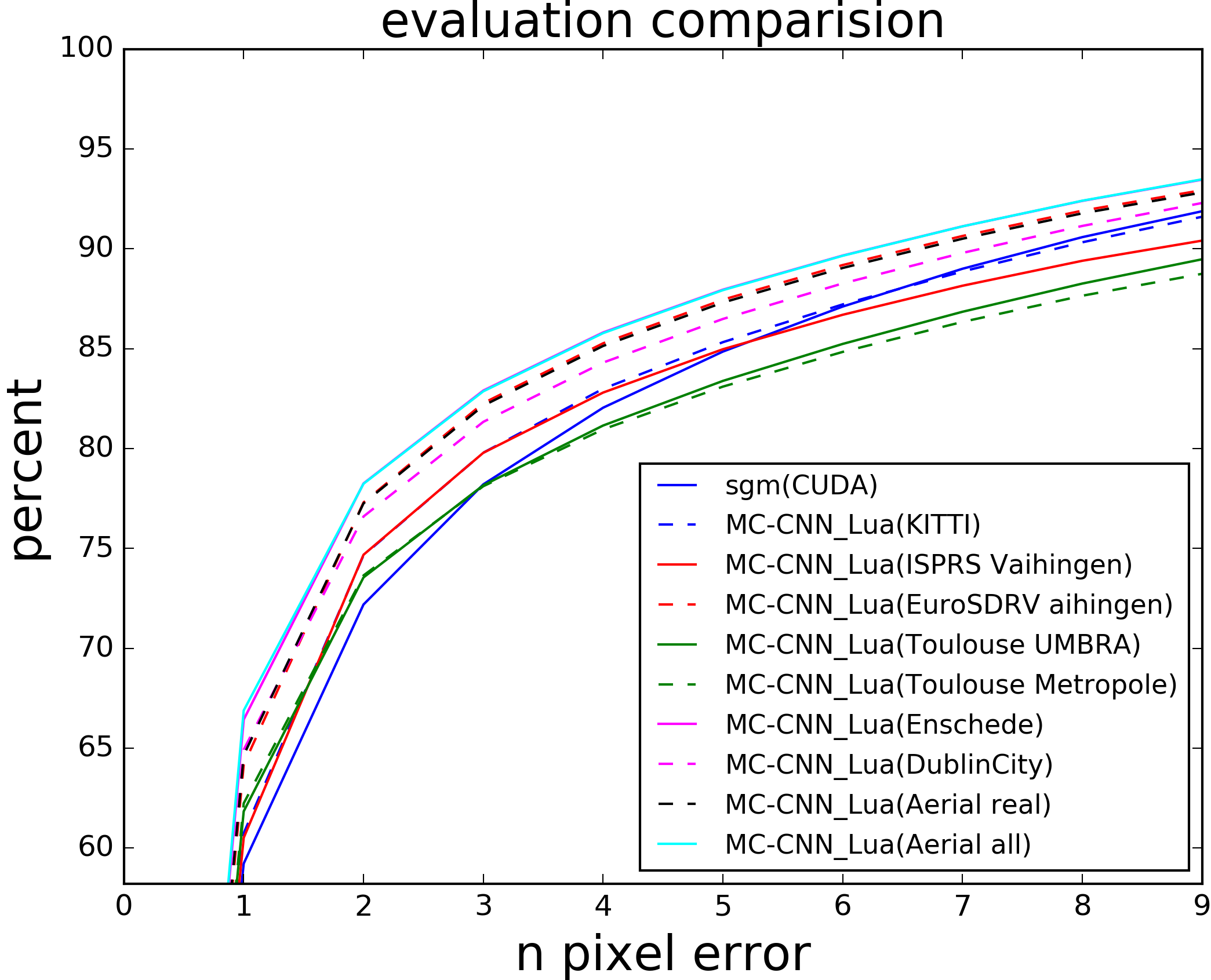}
	}
	\subfigure[Result on DublinCity]{
		\label{Figure.fusion_mccnn:f}
		\centering
		\includegraphics[width=0.45\linewidth]{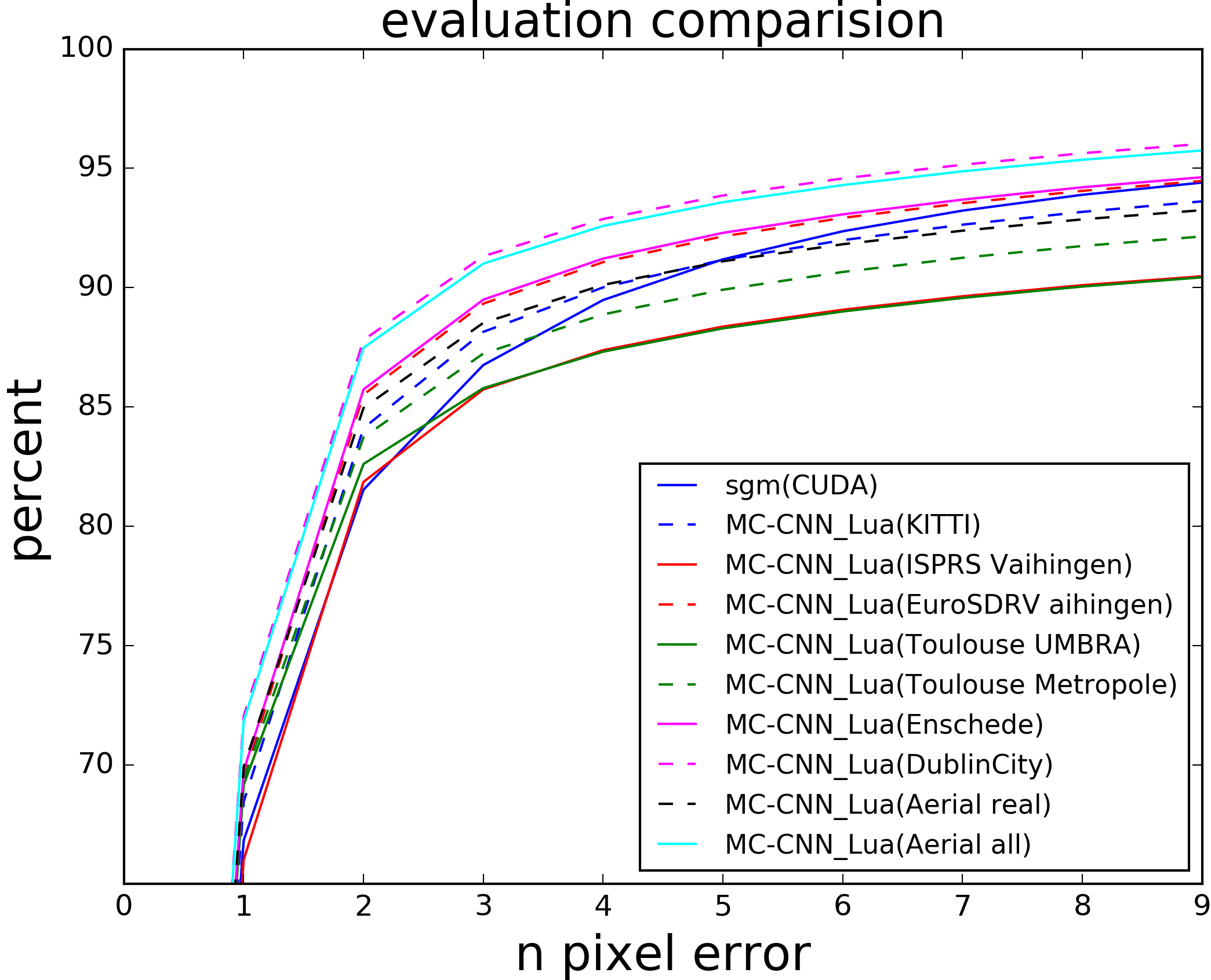}
	}
	\caption{Training from multi-dataset using MC-CNN.}
	\label{Figure.fusion_mccnn}
\end{figure}

The performance of the MC-CNN method varies slightly across datasets, even when training on the mobile mapping data KITTI and testing on aerial datasets. We attribute this to the fact that it is a hybrid method, where the DL-extracted features are followed by the traditional SGM step. All in all, we can draw the following general conclusions:

\begin{itemize}
    \item The higher resolution of the images, the better the models trained on them (for \textit{EuroSDR Vaihingen} in \Cref{Figure.fusion_mccnn:b} the best is the model trained on \textit{ISPRS Vaihingen})
    \item Augmenting the training set with other different datasets increases the error variance (see \textbf{\textit{Aerial real}} for \textit{ISPRS Vaihingen} in \Cref{Figure.fusion_mccnn:a});
    \item MC-CNN is on par with SGM(CUDA) on the small $B/H$ images of the \textit{Toulouse Metropole} in \Cref{Figure.fusion_mccnn:c}. 
    \item MC-CNN performance on large $B/H$ outperforms the classical SGM(CUDA) (see the \textit{Enschede} dataset in \Cref{Figure.fusion_mccnn:e}). 
    \item The performance of pretrained model(KITTI) varies, for example, in dataset \textit{Toulouse Metropole}, pretained model performs worst, but in dataset \textit{DublinCity} shown in \Cref{Figure.fusion_mccnn:f}, pretrained model is better than SGM(CUDA).
\end{itemize}





\paragraph{EfficientDeep}

\begin{figure}[tp]
	\centering
	\subfigure[Result on ISPRS Vaihingen]{
		\label{Figure.fusion_effcnn:a}
		\centering
		\includegraphics[width=0.45\linewidth]{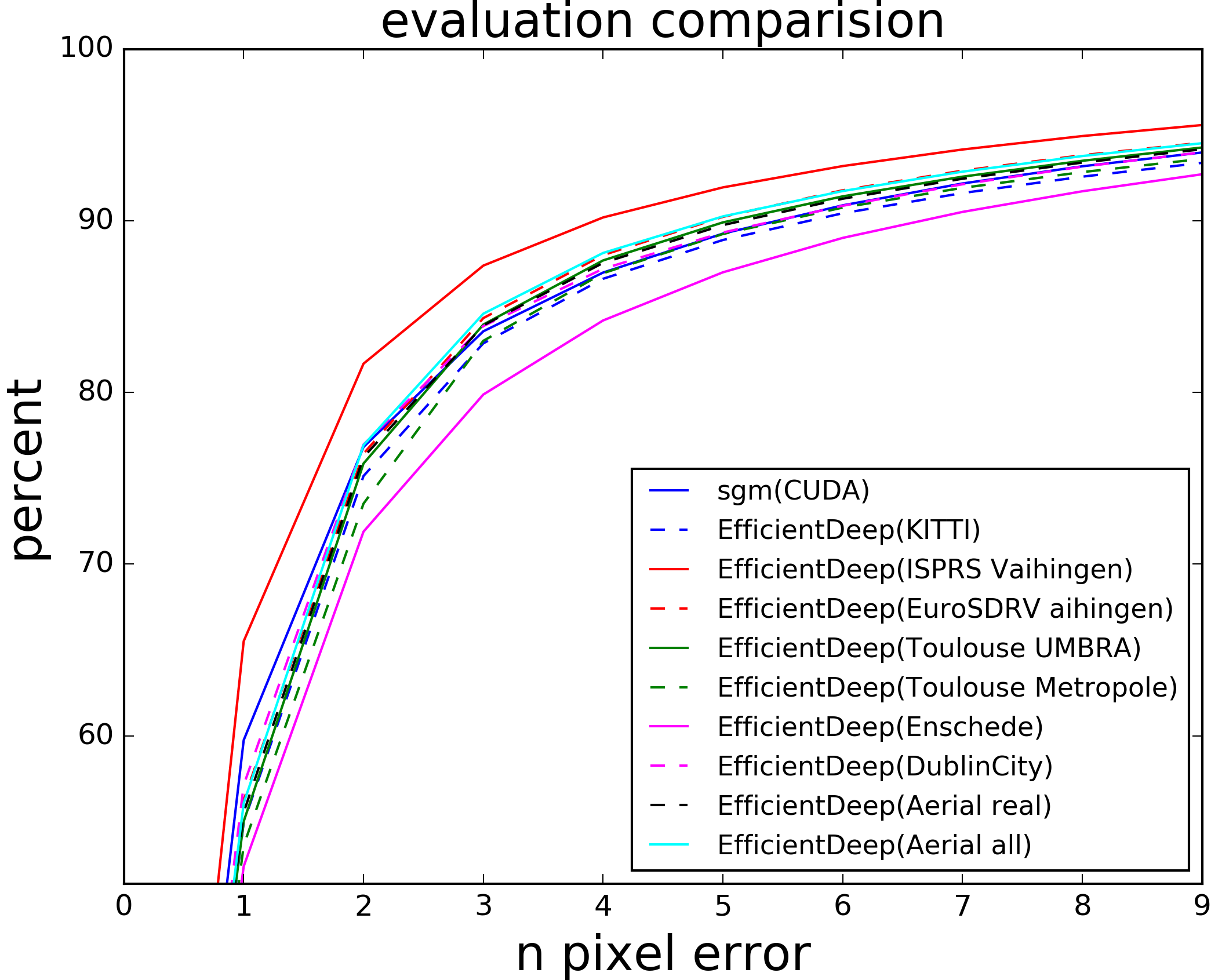}
	}
	\subfigure[Result on EuroSDR Vaihingen]{
		\label{Figure.fusion_effcnn:b}
		\centering
		\includegraphics[width=0.45\linewidth]{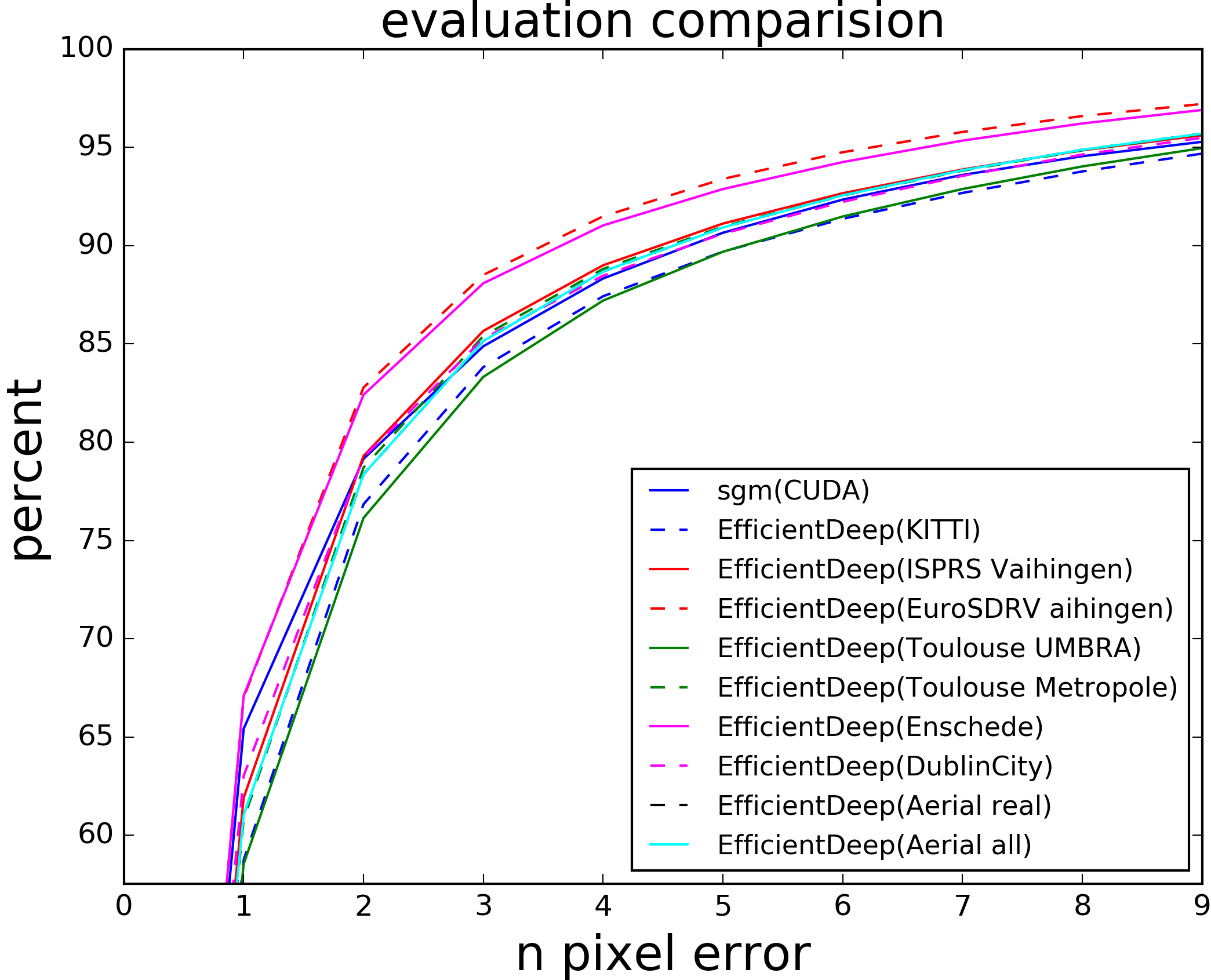}
	}
	
	\subfigure[Result on Toulouse Metropole]{
		\label{Figure.fusion_effcnn:c}
		\centering
		\includegraphics[width=0.45\linewidth]{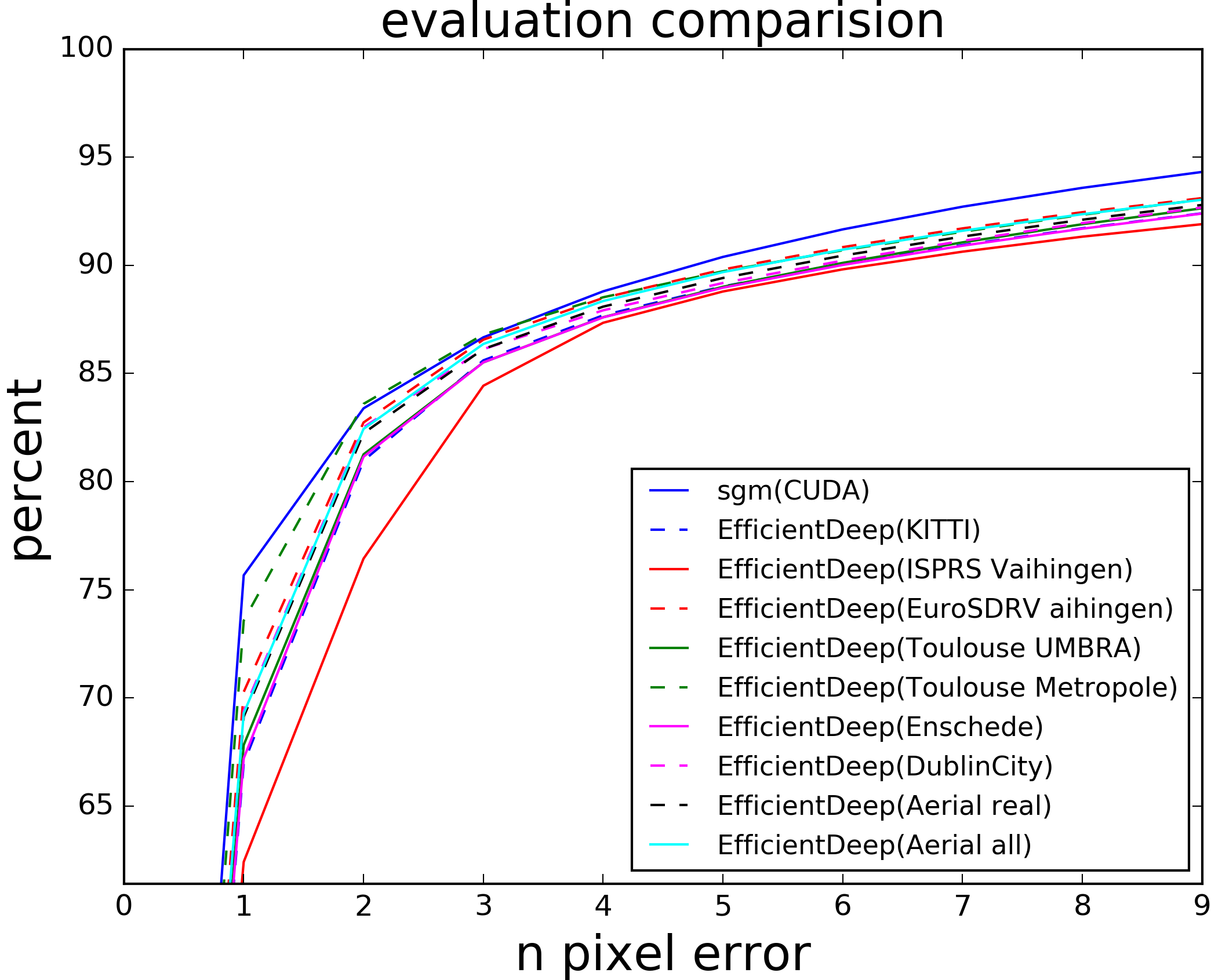}
	}
	\subfigure[Result on Toulouse UMBRA]{
		\label{Figure.fusion_effcnn:d}
		\centering
		\includegraphics[width=0.45\linewidth]{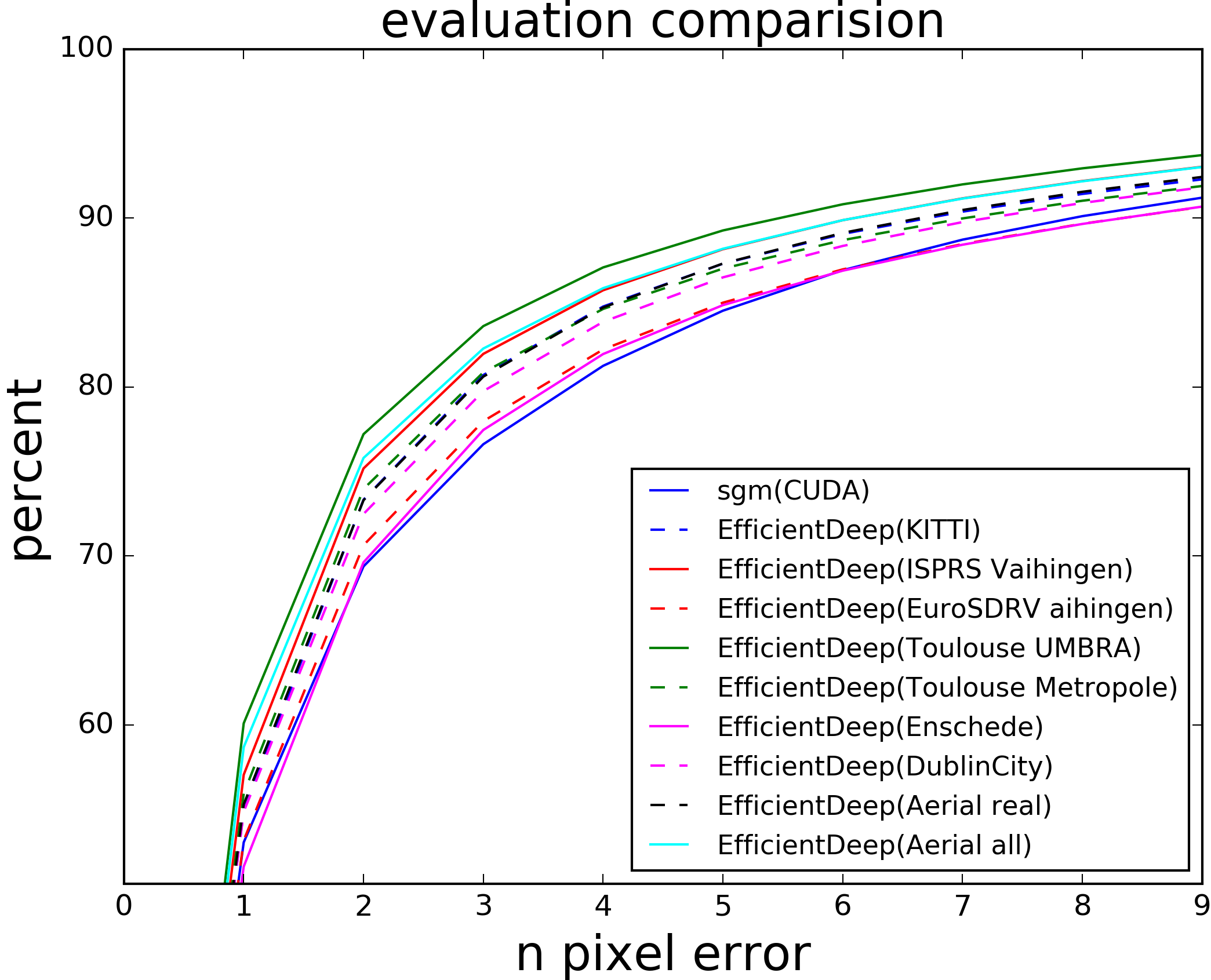}
	}
	
	\subfigure[Result on Enschede]{
		\label{Figure.fusion_effcnn:e}
		\centering
		\includegraphics[width=0.45\linewidth]{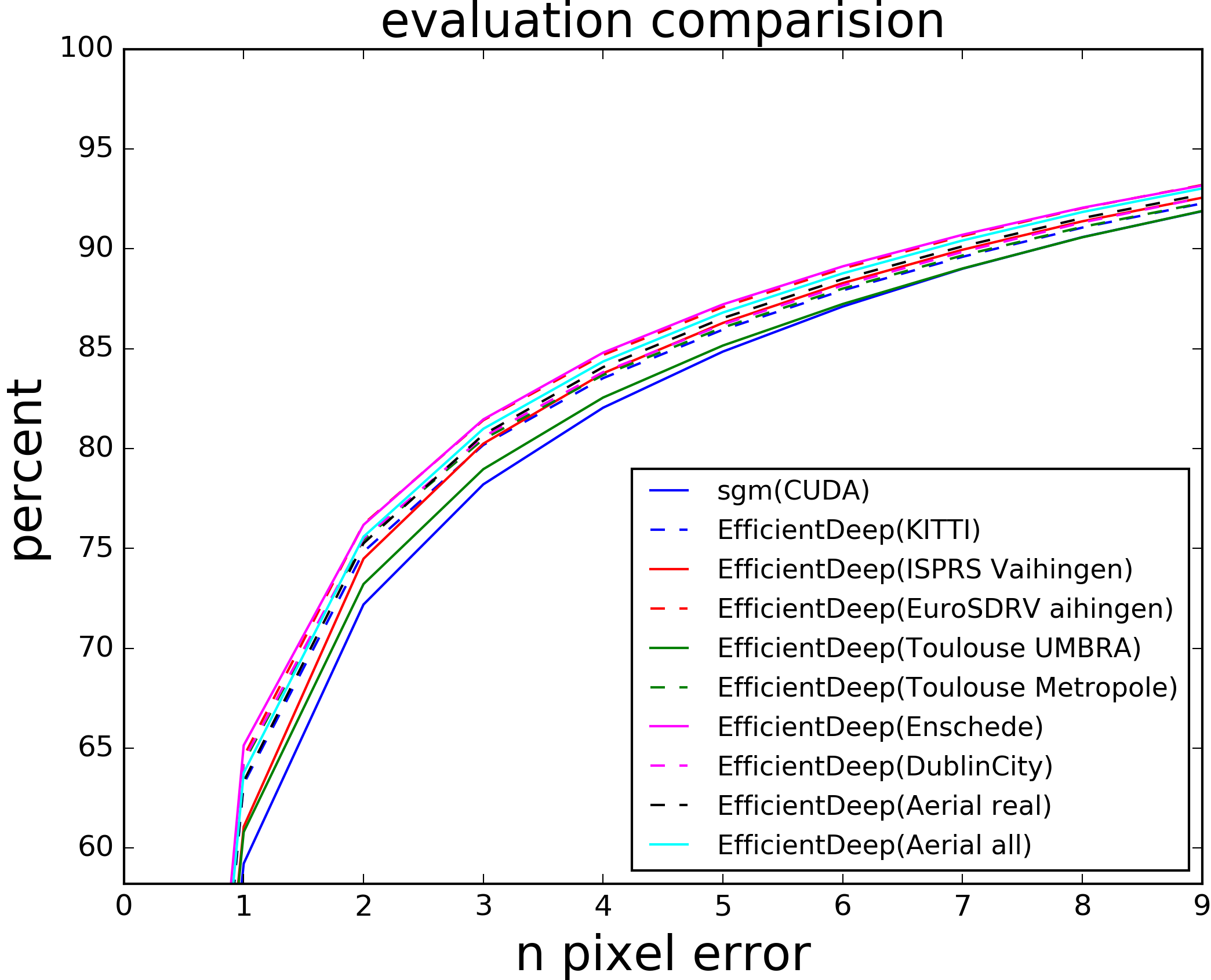}
	}
	\subfigure[Result on DublinCity]{
		\label{Figure.fusion_effcnn:f}
		\centering
		\includegraphics[width=0.45\linewidth]{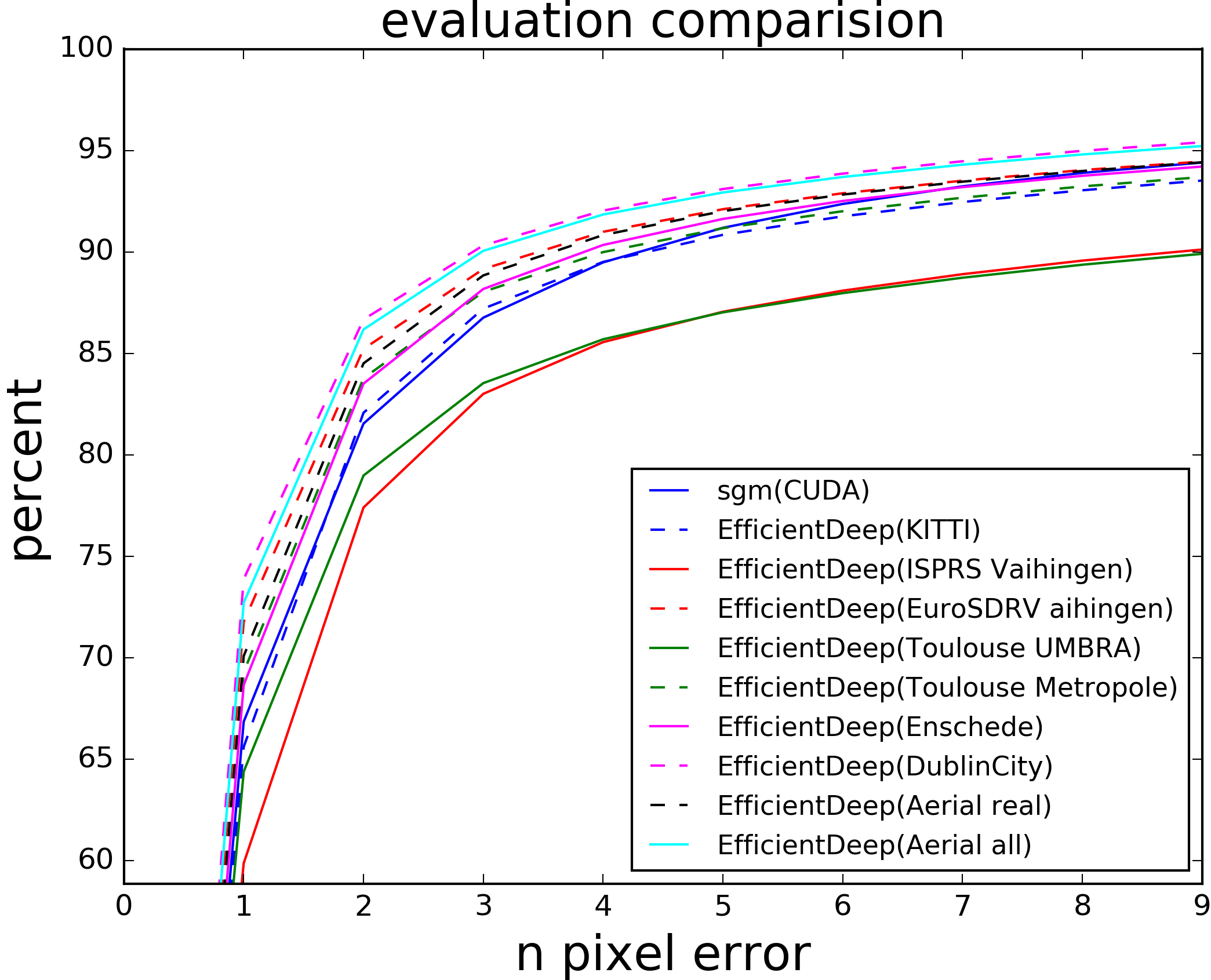}
	}
	
	\caption{Training from multi-dataset using Efficient DL feature.}
	\label{Figure.fusion_effcnn}
\end{figure}

Because \textit{EfficientDeep} is also a hybrid method, the results are similar to that of \textit{MC-CNN} (see \Cref{Figure.fusion_effcnn}). We draw the following generic conclusion from our experiments:
\begin{itemize}
    \item The variance of training data influences the performance of the model, as shown in \Cref{Figure.fusion_effcnn:a}, the performance of the model of Aerial(all) is far lower than that of the model of \textit{ISPRS Vaihingen}, the model of \textit{EuroSDR Vaihingen} is a little better than SGM(CUDA).
    \item Even though \textit{EfficientDeep} is similar to \textit{MC-CNN}, different network performs differently, for example, for dataset \textit{EuroSDR Vaihingen}, compared to SGM(CUDA), fine-trained models of \textit{MC-CNN} improve the performance, but fine-trained models of \textit{EfficientDeep}  don't improve (see \Cref{Figure.fusion_mccnn:b} and \Cref{Figure.fusion_effcnn:b}).
    \item The higher image resolution has the more precise result. In \textit{Toulouse UMBRA} (cf. \Cref{Figure.fusion_effcnn:c}) both \textit{ISPRS Vaihingen} and \textit{DublinCity}  perform best. 
    \item For dataset \textit{Toulouse UMBRA} and \textit{Enschede} shown in \Cref{Figure.fusion_effcnn:d,Figure.fusion_effcnn:e}, using \textit{EfficientDeep} to learning the feature can improve the result compared to SGM(CUDA).
\end{itemize}








\paragraph{DeepPruner}

\begin{figure}[tp]
	\centering
	\subfigure[Result on ISPRS Vaihingen]{
		\label{Figure.fusion_deepprune:a}
		\centering
		\includegraphics[width=0.45\linewidth]{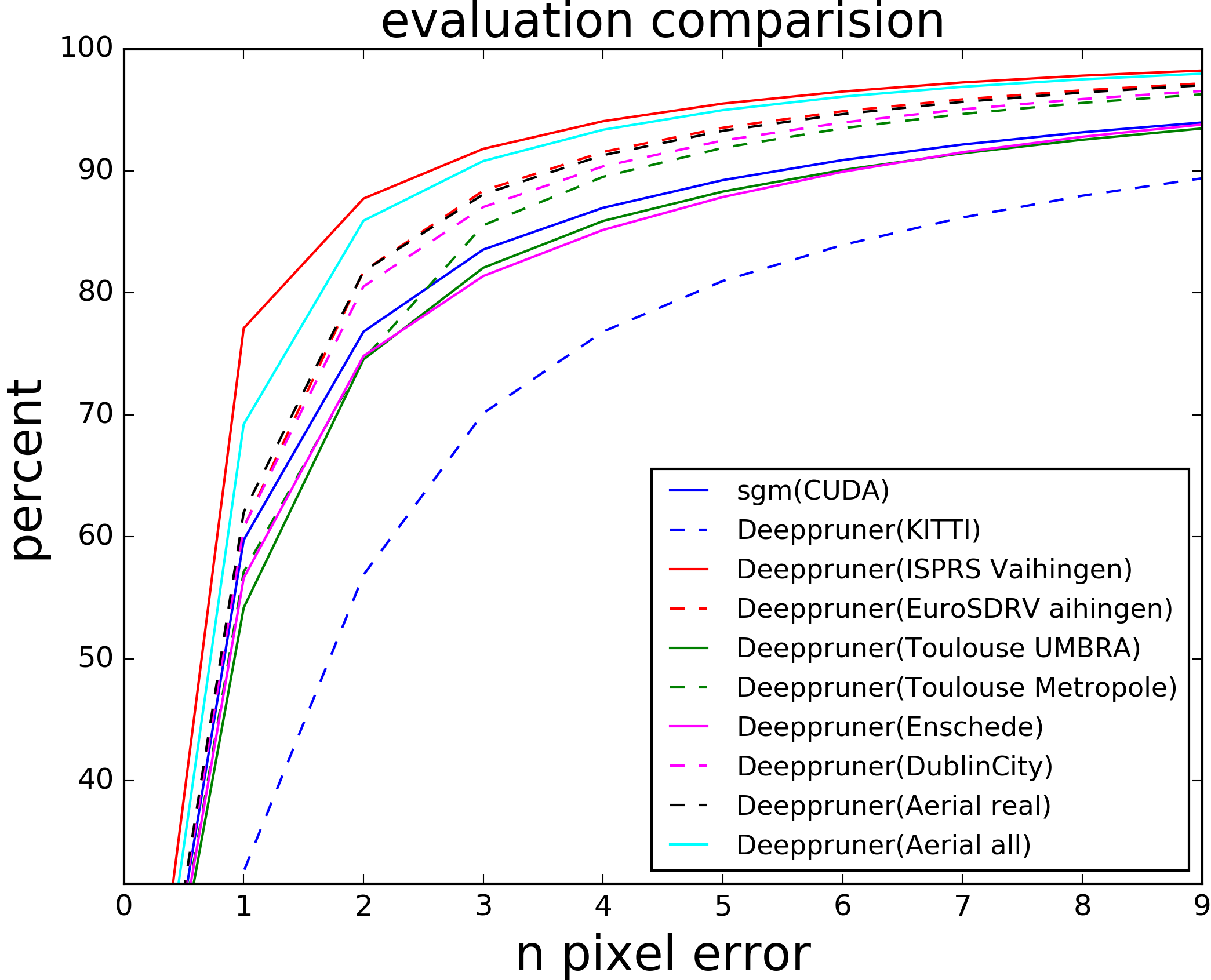}
	}
	\subfigure[Result on EuroSDR Vaihingen]{
		\label{Figure.fusion_deepprune:b}
		\centering
		\includegraphics[width=0.45\linewidth]{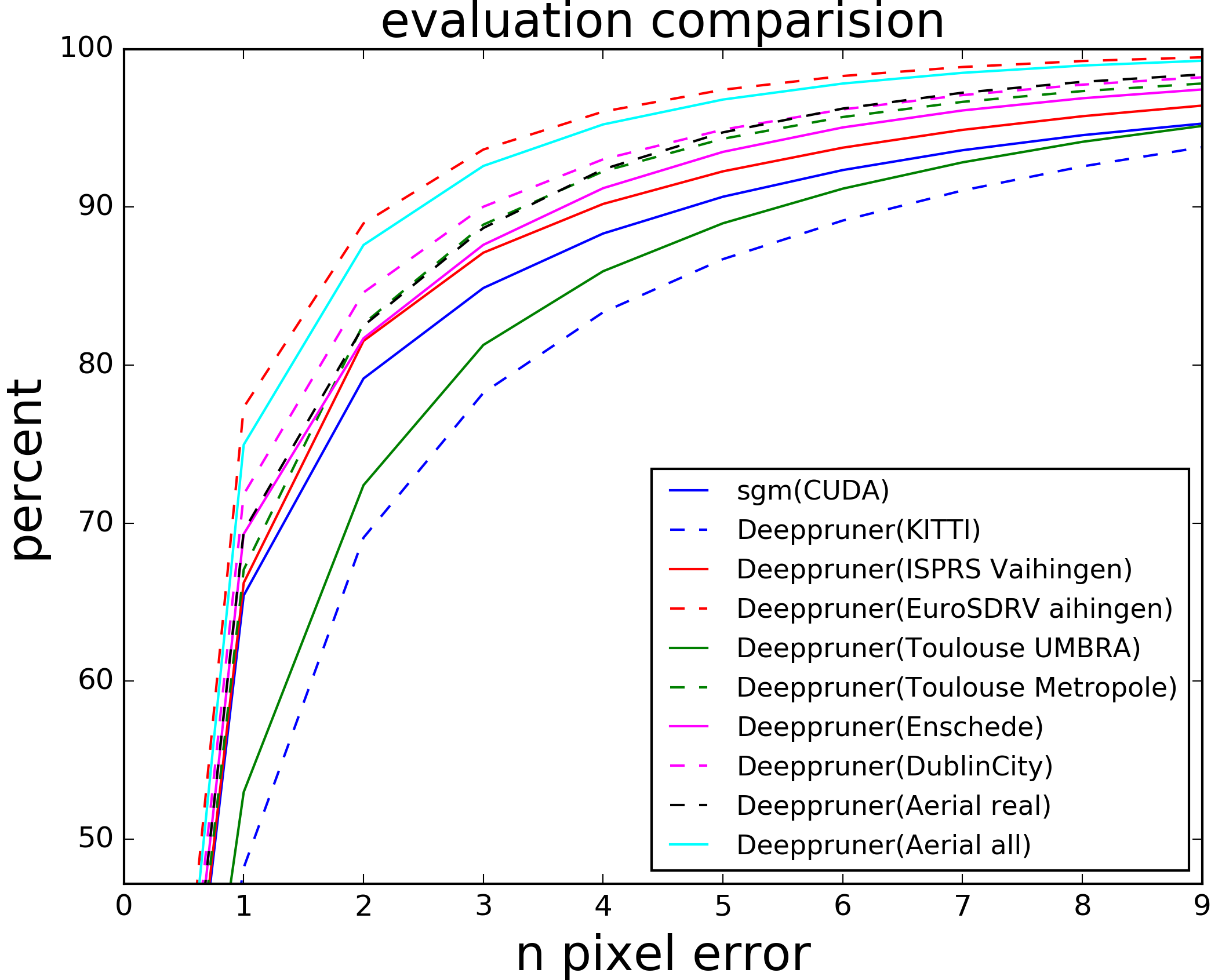}
	}
	
	\subfigure[Result on Toulouse Metropole]{
		\label{Figure.fusion_deepprune:c}
		\centering
		\includegraphics[width=0.45\linewidth]{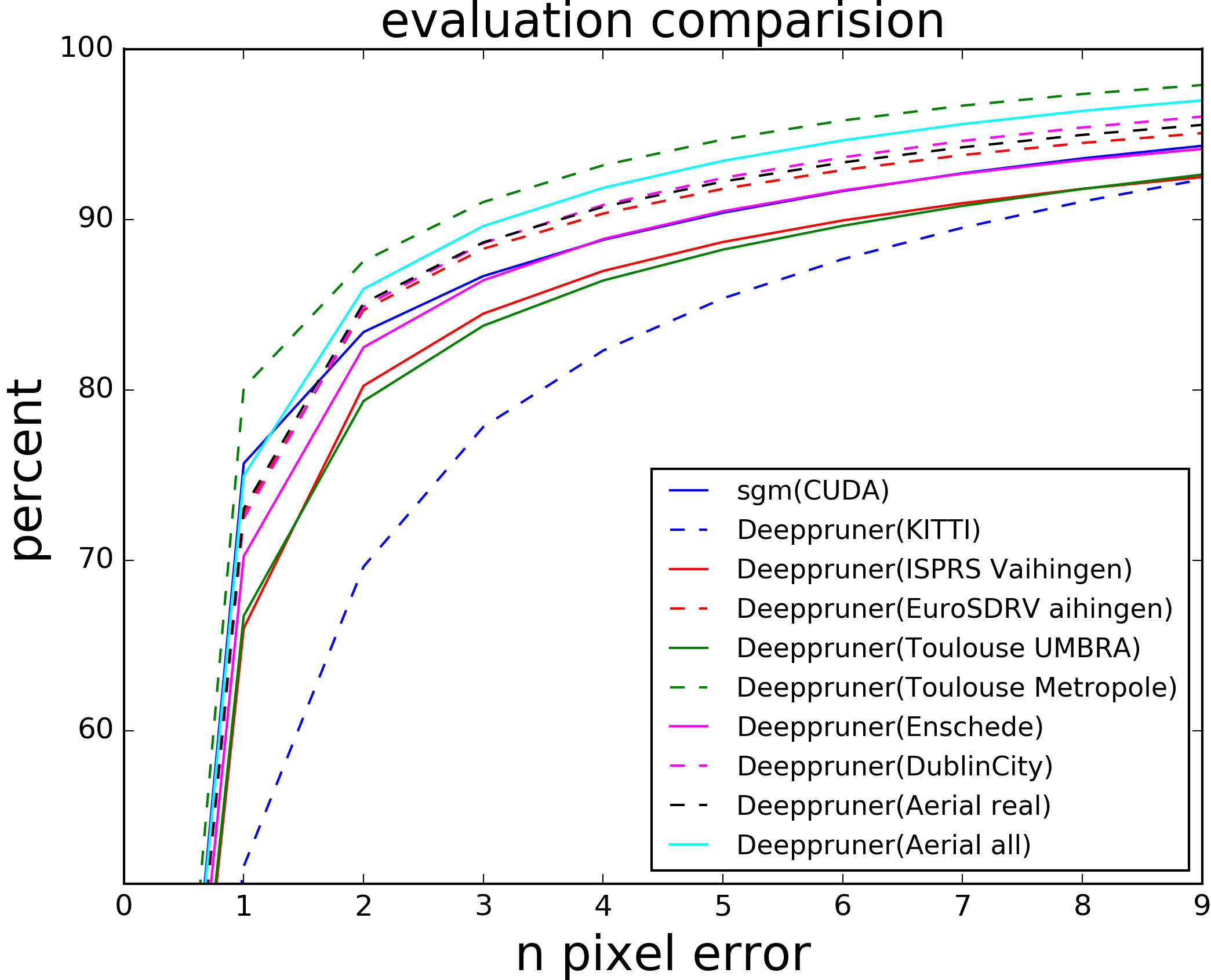}
	}
	\subfigure[Result on Toulouse UMBRA]{
		\label{Figure.fusion_deepprune:d}
		\centering
		\includegraphics[width=0.45\linewidth]{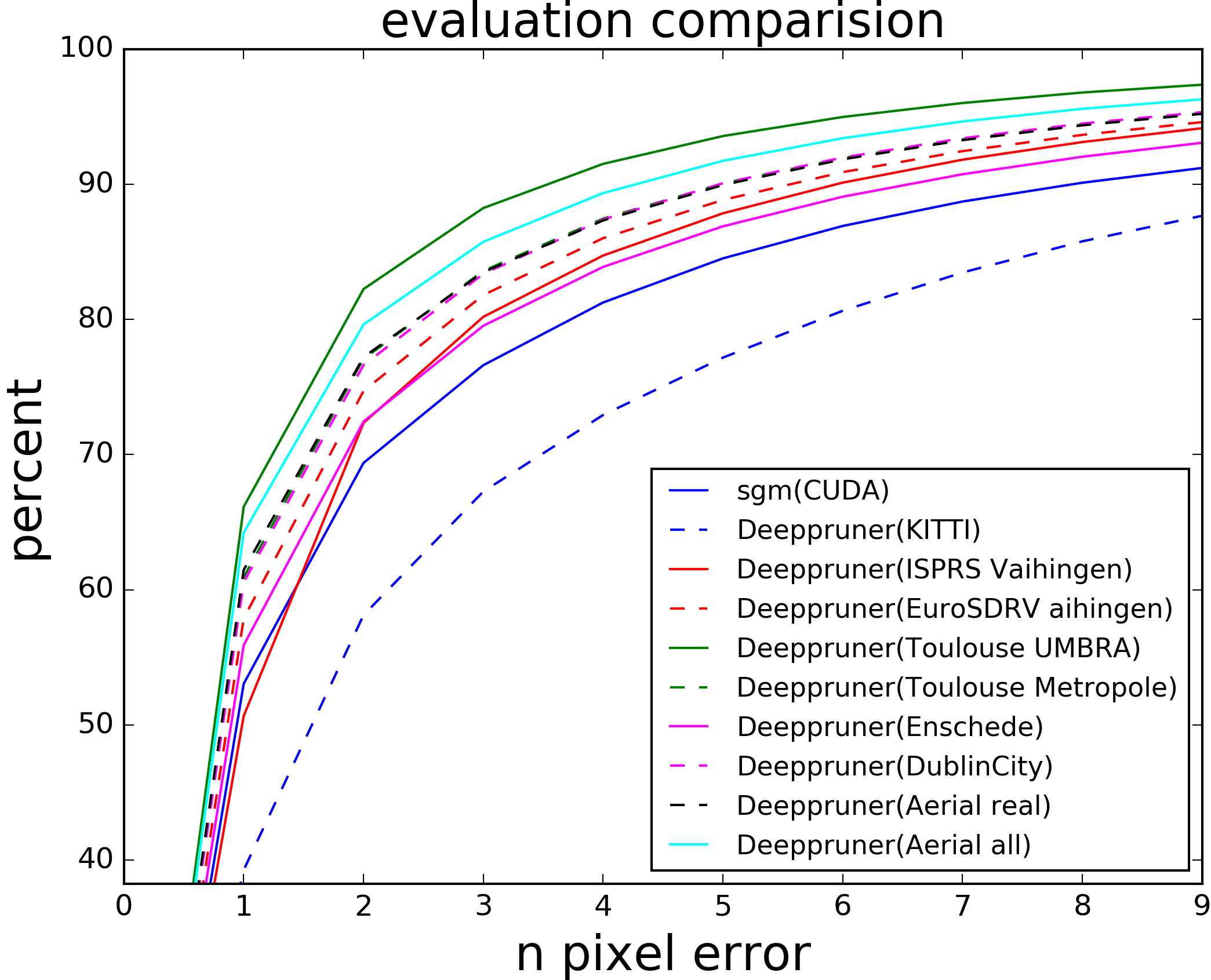}
	}
	
	\subfigure[Result on Enschede]{
		\label{Figure.fusion_deepprune:e}
		\centering
		\includegraphics[width=0.45\linewidth]{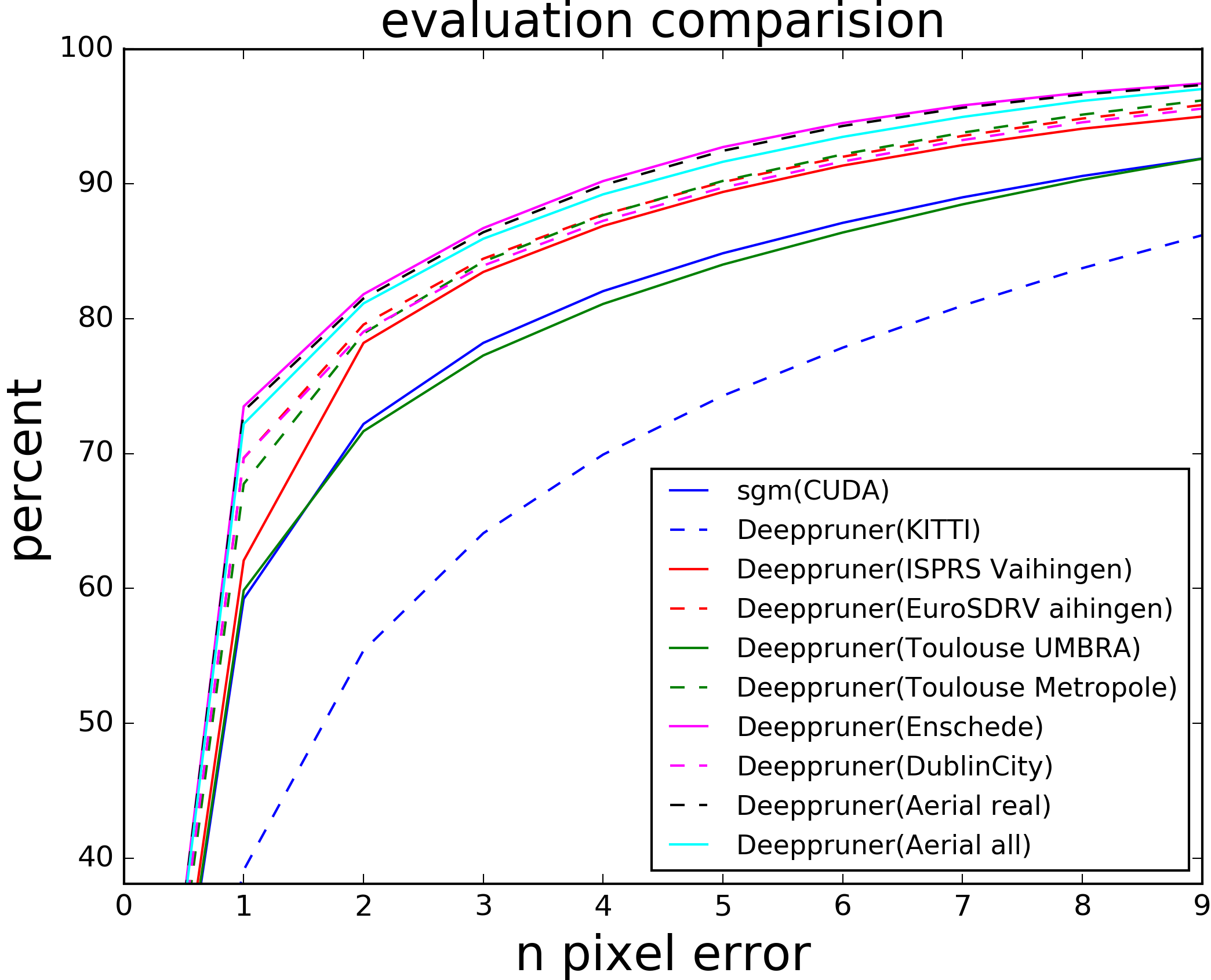}
	}
	\subfigure[Result on DublinCity]{
		\label{Figure.fusion_deepprune:f}
		\centering
		\includegraphics[width=0.45\linewidth]{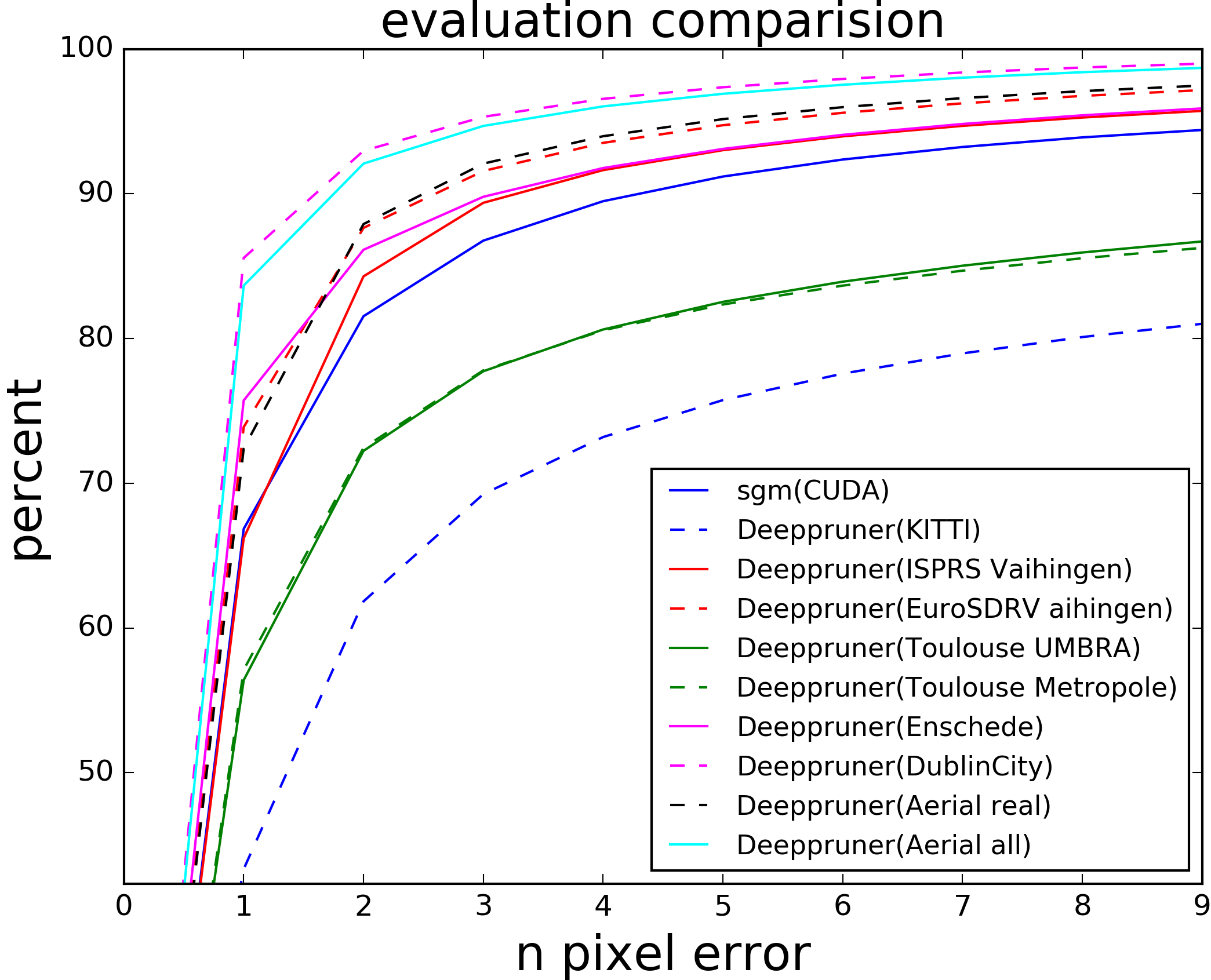}
	}
	
	\caption{Training from multi-dataset using DeepPruner.}
	\label{Figure.fusion_deepprune}
\end{figure}

Unlike \textit{MC-CNN} or \textit{EfficientDeep}, \textit{DeepPruner} is an end-to-end method and the results show different tendencies (cf. \Cref{Figure.fusion_deepprune}). We conclude that:
\begin{itemize}
    \item The KITTI-trained models perform worst of all, while the models trained on \textit{Aerial(real)} produce good results overall across all datasets, slightly worse than training on the same dataset. 
    \item Augmenting the training with different datasets slightly deteriorates the results compared to a model trained and tested on images coming from the same acquisition. However,  a universal model trained on heterogeneous scenes still remains a good compromise and an interesting solution toward a one-for-all model. Compare  the \textit{Aerial(all)} (which includes ISPRS Vaihingen) with the training using exclusively the \textit{ISPRS Vaihingen} in \Cref{Figure.fusion_deepprune:a}.
    \item \textit{DeepPruner} produces smoother results than the traditional SGM (CUDA). Note that the better performance compare to SGM (CUDA) on large pixel error in \Cref{Figure.fusion_deepprune:c}. 
    \item Transferring the \textit{DeepPruner} models to datasets different from their training samples deteriorate the matching performance. For example, the model from \textit{Toulouse Metropole} works better on \textit{Toulouse UMBRA} than on the others.
    \item For difficult dataset where SGM(CUDA) performs badly, fine-trained models of textit{DeepPruner} give great improvement, for example  \textit{Toulouse UMBRA} and \textit{Enschede}.
\end{itemize}








\paragraph{HRS net}

\begin{figure}[tp]
	\centering
	\subfigure[Result on ISPRS Vaihingen]{
		\label{Figure.fusion_hrs:a}
		\centering
		\includegraphics[width=0.45\linewidth]{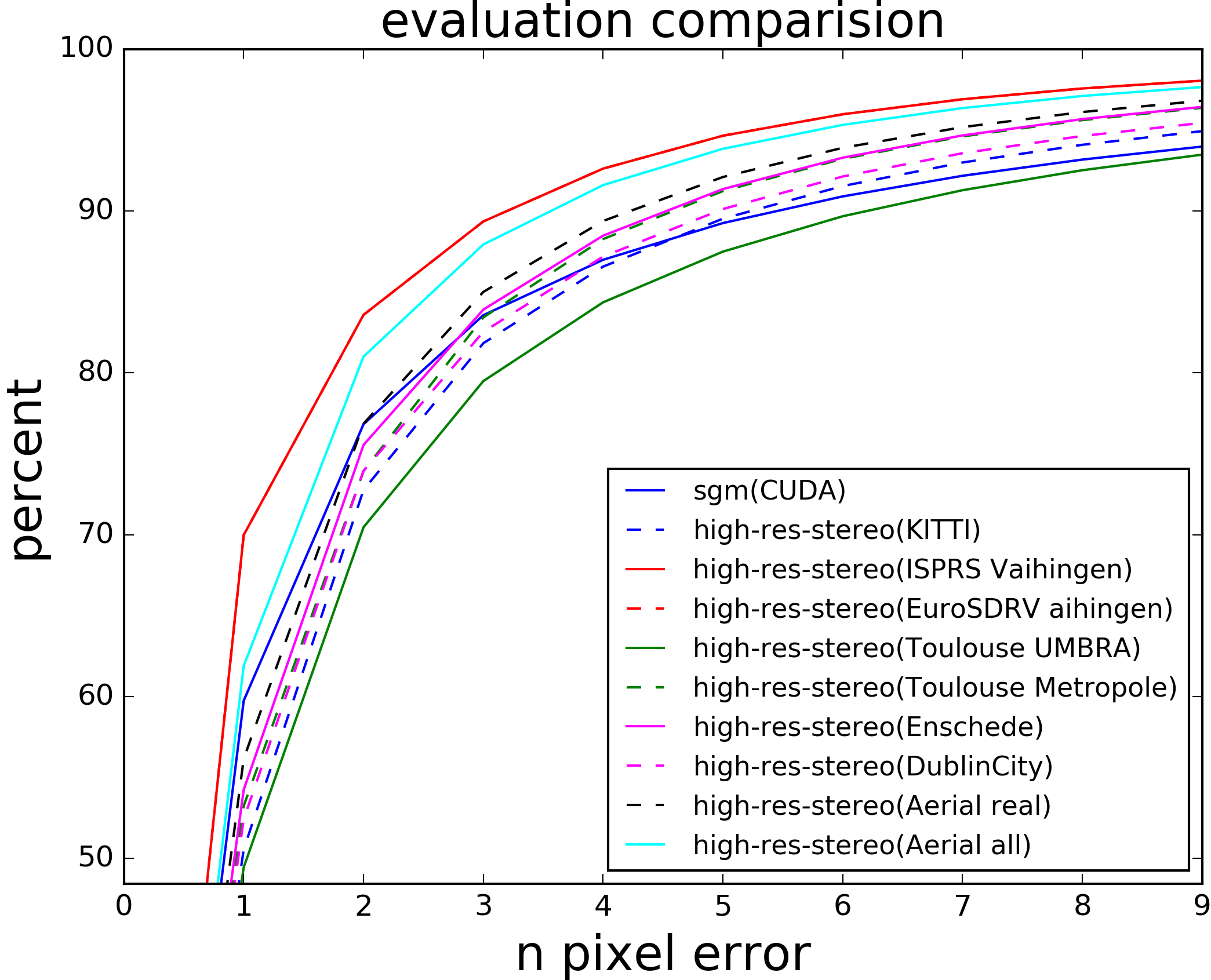}
	}
	\subfigure[Result on EuroSDR Vaihingen]{
		\label{Figure.fusion_hrs:b}
		\centering
		\includegraphics[width=0.45\linewidth]{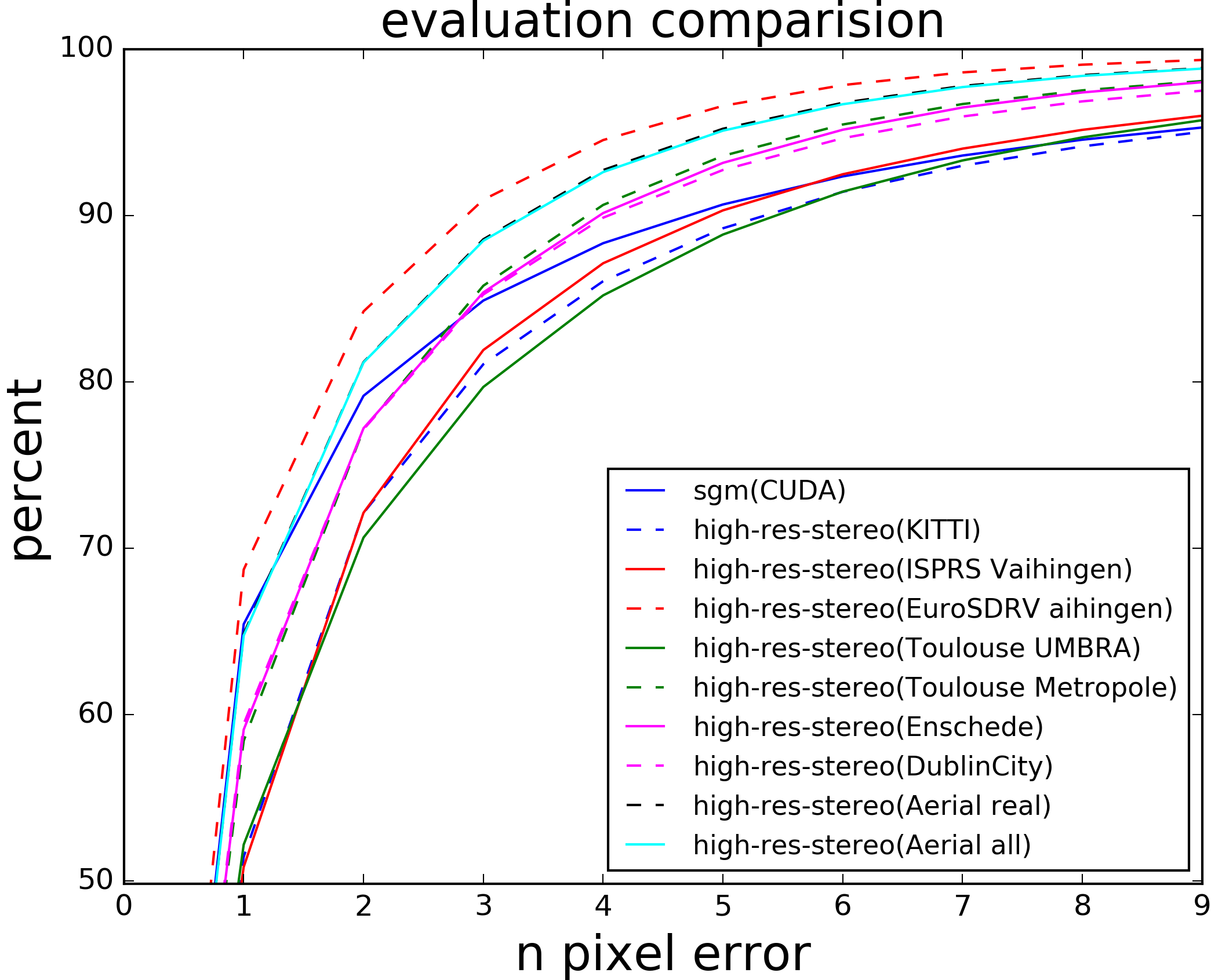}
	}
	
	\subfigure[Result on Toulouse Metropole]{
		\label{Figure.fusion_hrs:c}
		\centering
		\includegraphics[width=0.45\linewidth]{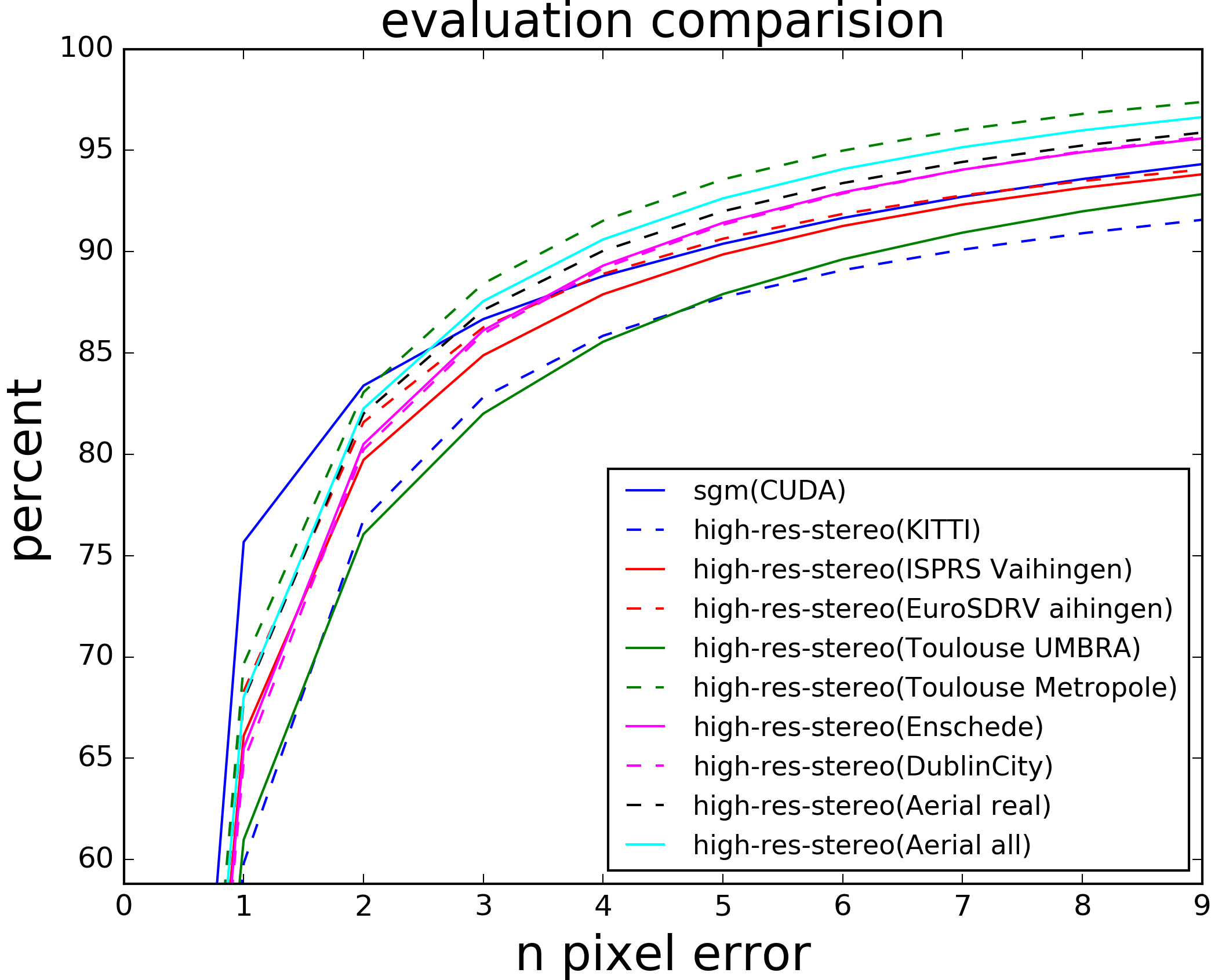}
	}
	\subfigure[Result on Toulouse UMBRA]{
		\label{Figure.fusion_hrs:d}
		\centering
		\includegraphics[width=0.45\linewidth]{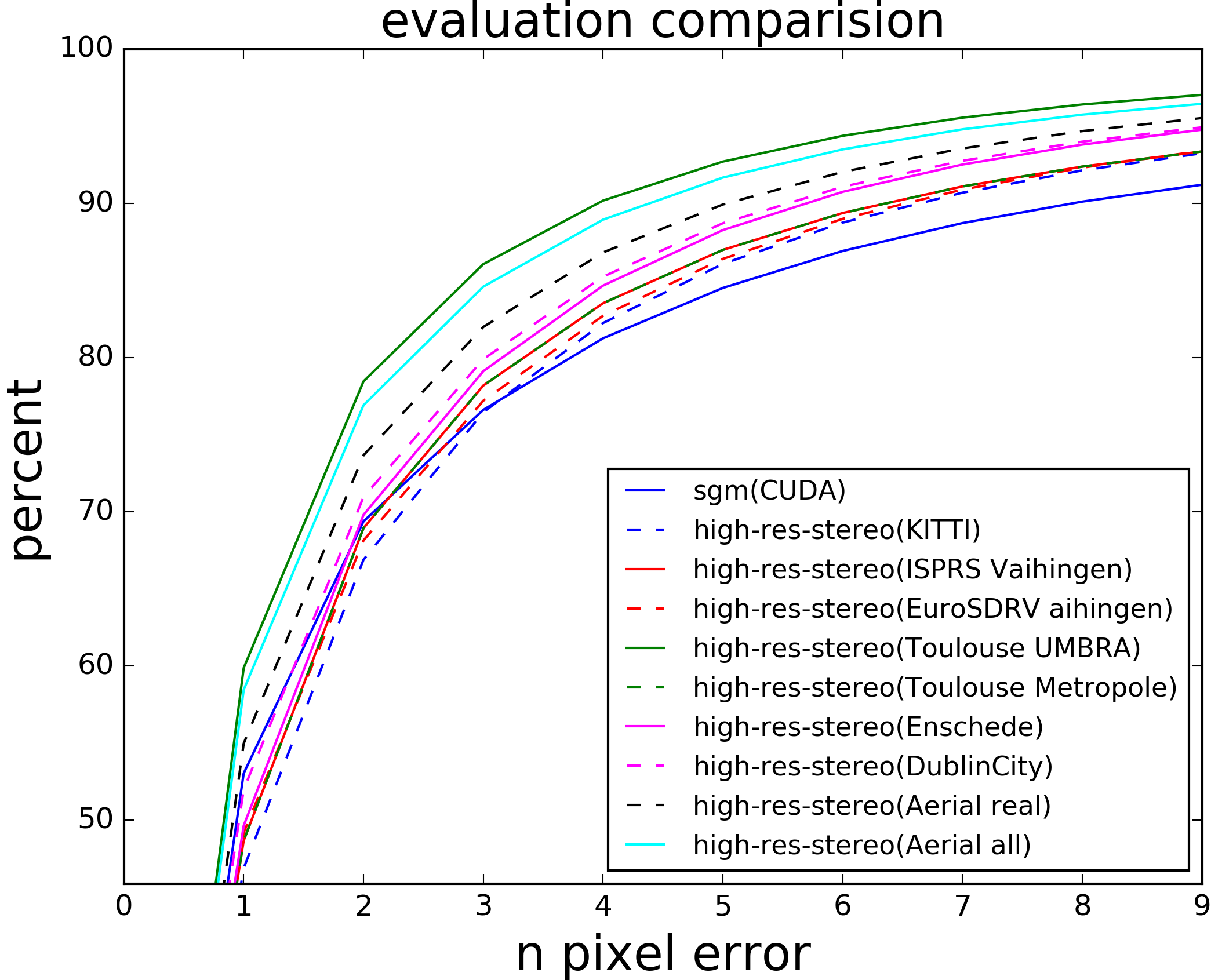}
	}
	
	\subfigure[Result on Enschede]{
		\label{Figure.fusion_hrs:e}
		\centering
		\includegraphics[width=0.45\linewidth]{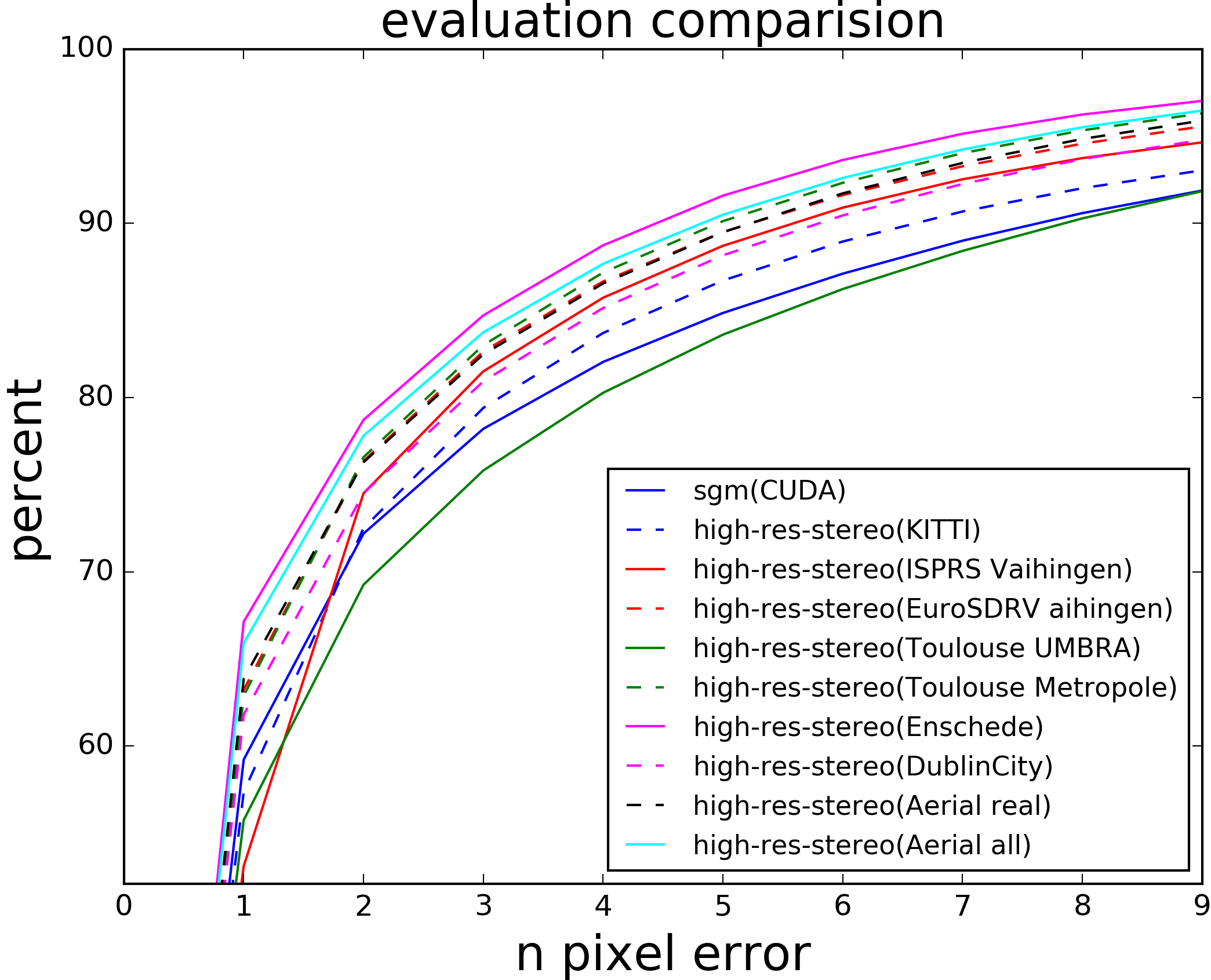}
	}
	\subfigure[Result on DublinCity]{
		\label{Figure.fusion_hrs:f}
		\centering
		\includegraphics[width=0.45\linewidth]{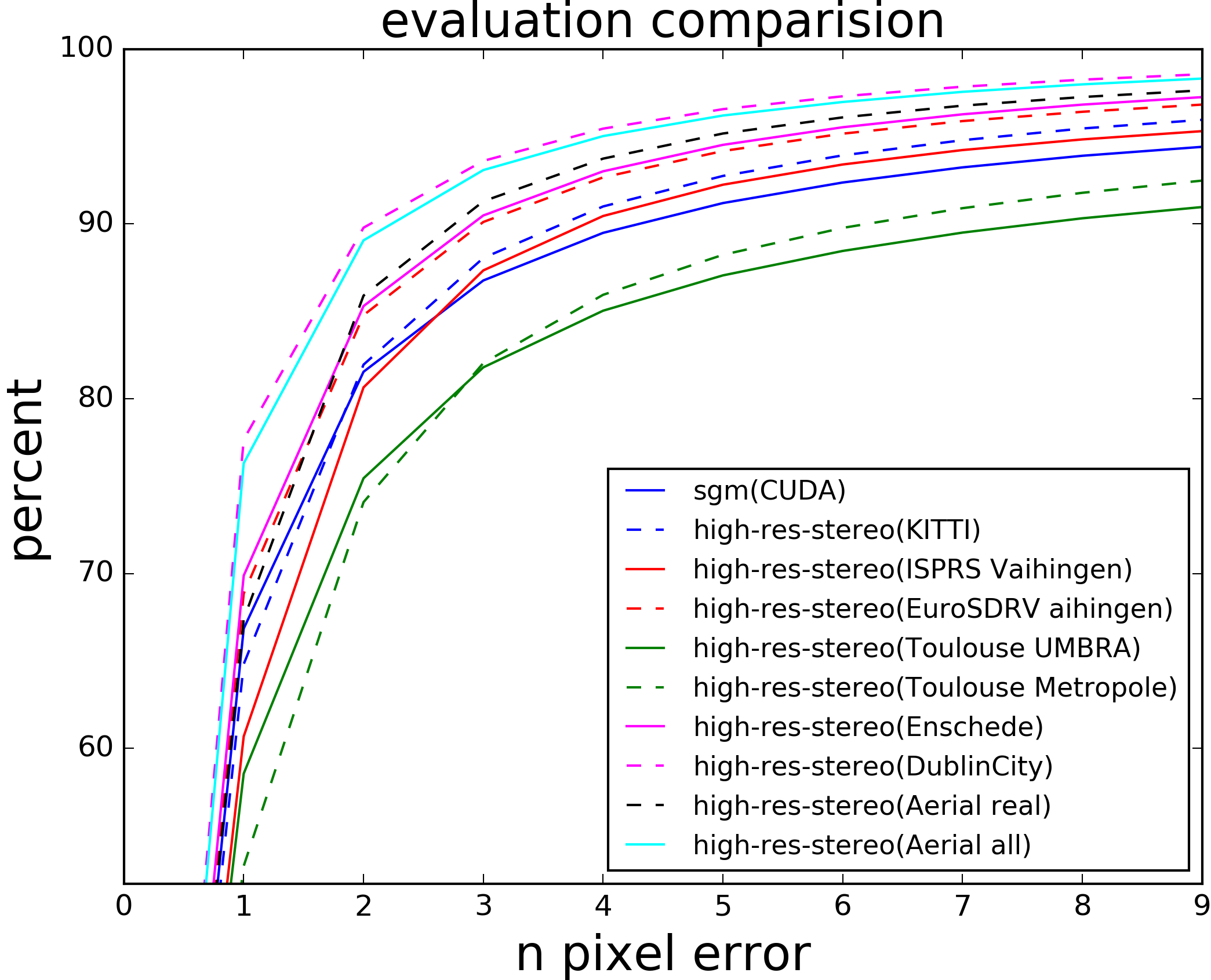}
	}
	
	\caption{Training from multi-dataset using HRS net.}
	\label{Figure.fusion_hrs}
\end{figure}

This architecture and performance is similar to that of \textit{DeepPruner}, the main difference is that \textit{HRS net} uses hierarchical feature volume decoder to simulate a coarse to fine strategy. Based on \Cref{Figure.fusion_hrs} we conclude the following:
\begin{itemize}
    \item For \textit{HRS net}, the model of \textit{KITTI} is as good as the other aerial models, indicating that \textit{HRS net} is less dependent on the type of the training data. 
    \item \textit{HRS net} produces a smooth resul (see  \Cref{Figure.fusion_hrs:c}) with good performance on large pixel errors. 
\end{itemize}







\paragraph{GANet}

\begin{figure}[tp]
	\centering
	\subfigure[Result on ISPRS Vaihingen]{
		\label{Figure.fusion_ganet:a}
		\centering
		\includegraphics[width=0.45\linewidth]{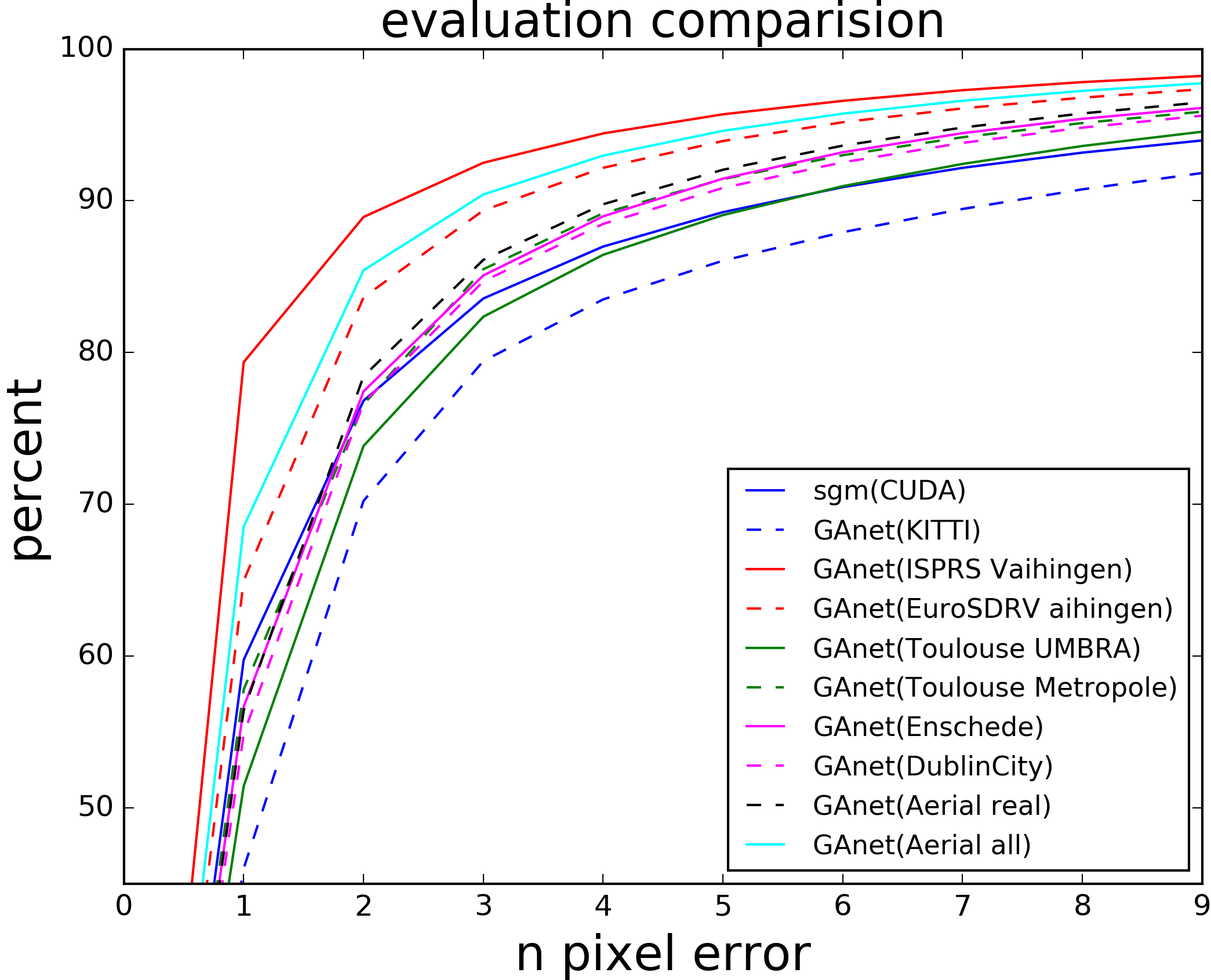}
	}
	\subfigure[Result on EuroSDR Vaihingen]{
		\label{Figure.fusion_ganet:b}
		\centering
		\includegraphics[width=0.45\linewidth]{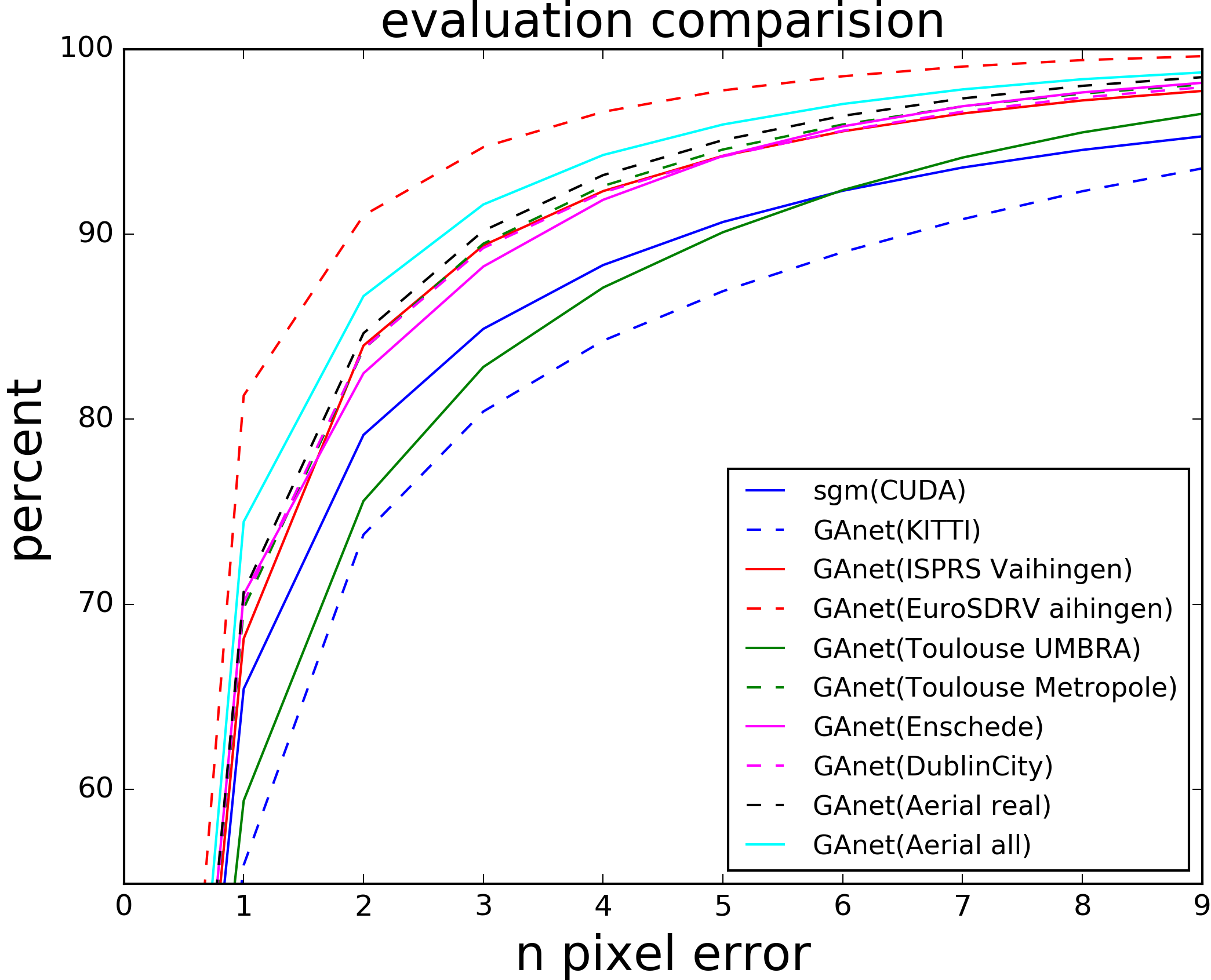}
	}
	
	\subfigure[Result on Toulouse Metropole]{
		\label{Figure.fusion_ganet:c}
		\centering
		\includegraphics[width=0.45\linewidth]{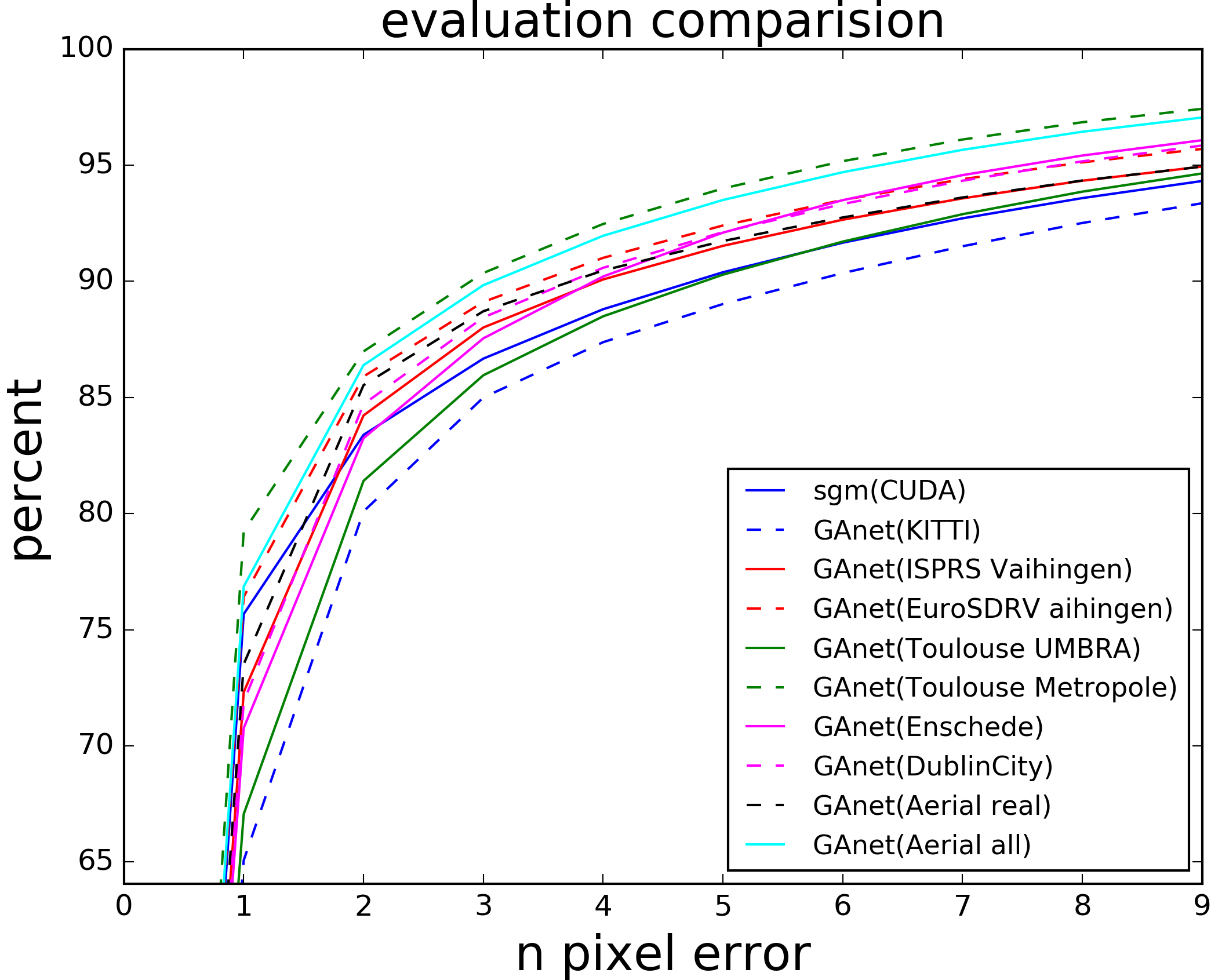}
	}
	\subfigure[Result on Toulouse UMBRA]{
		\label{Figure.fusion_ganet:d}
		\centering
		\includegraphics[width=0.45\linewidth]{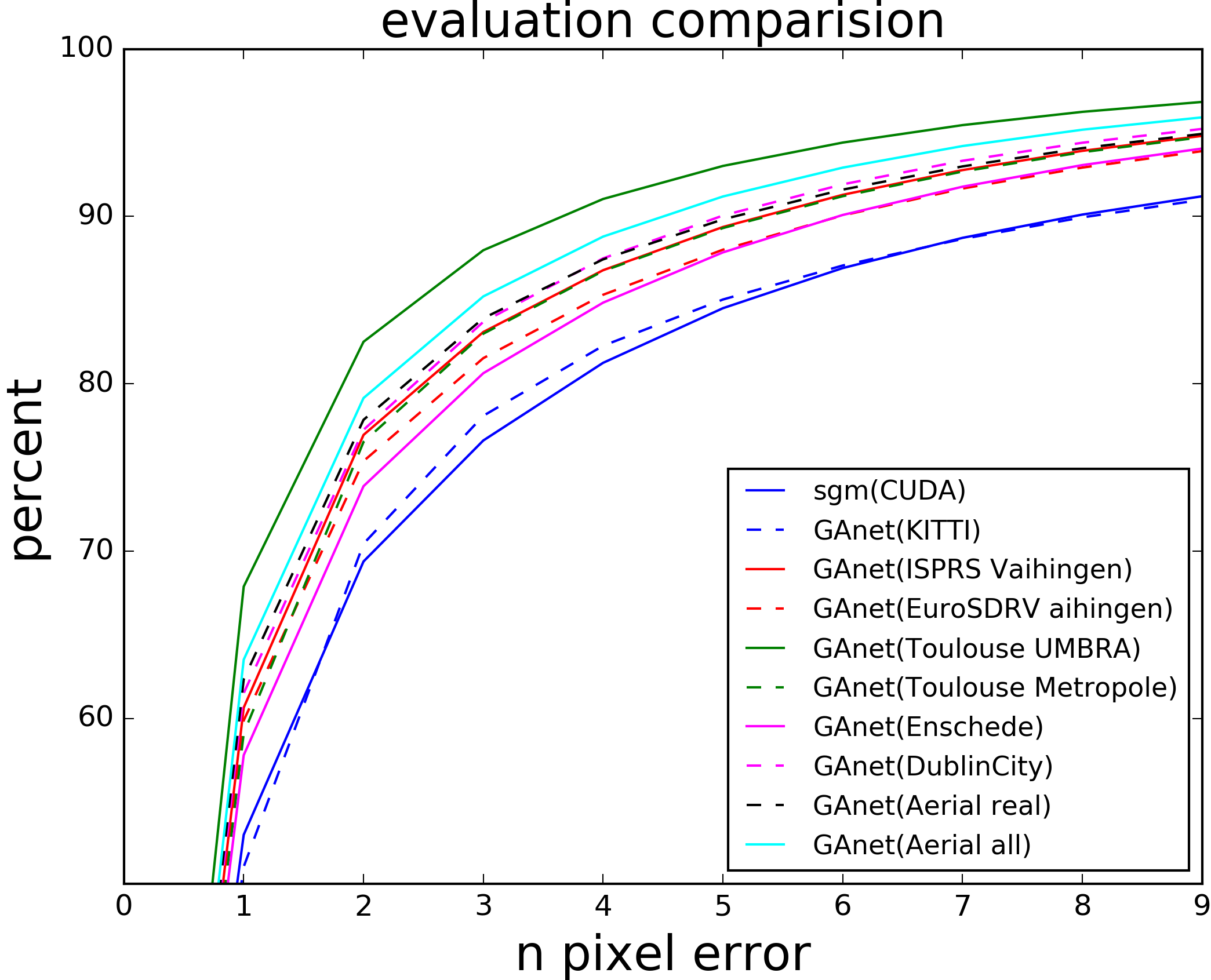}
	}
	
	\subfigure[Result on Enschede]{
		\label{Figure.fusion_ganet:e}
		\centering
		\includegraphics[width=0.45\linewidth]{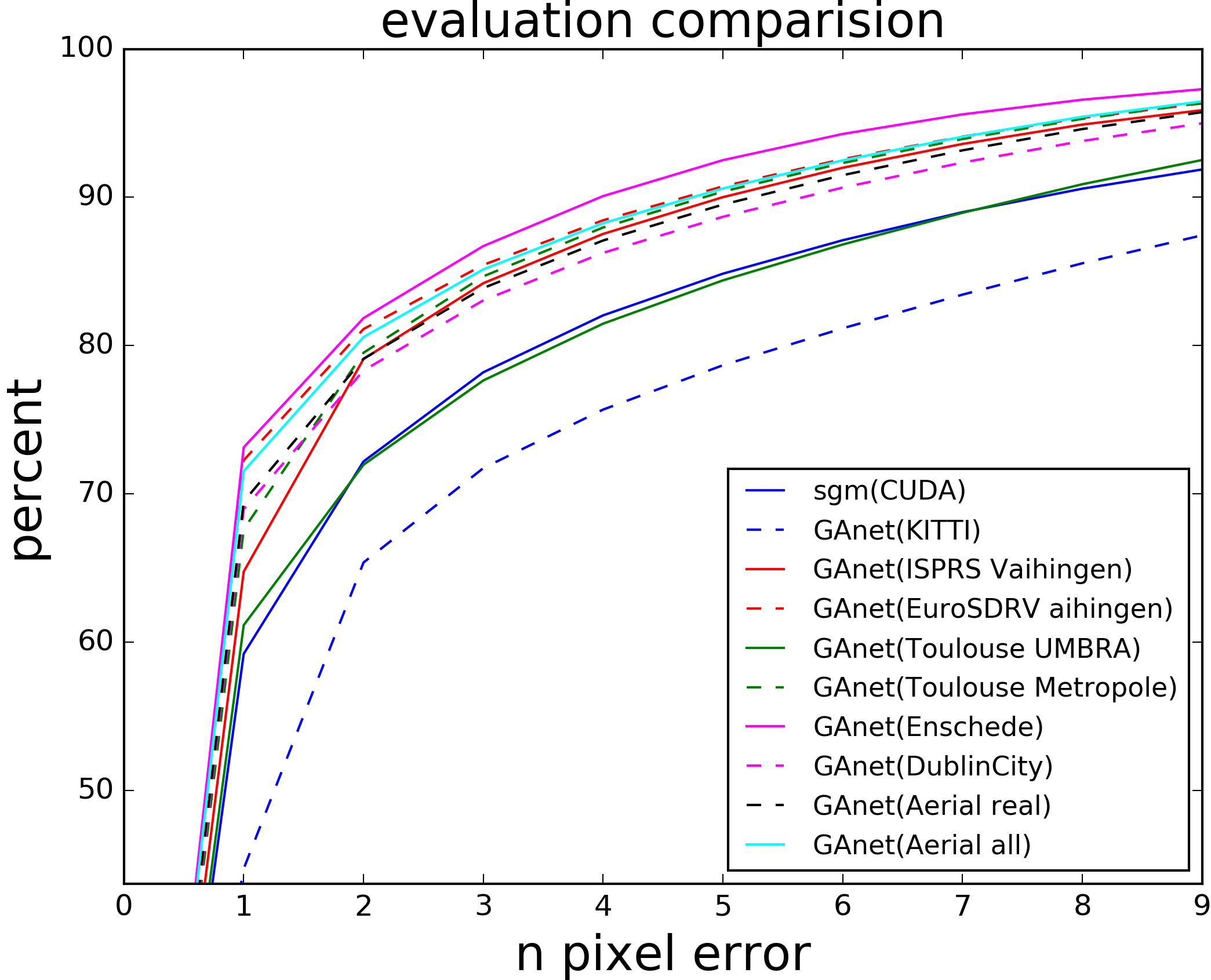}
	}
	\subfigure[Result on DublinCity]{
		\label{Figure.fusion_ganet:f}
		\centering
		\includegraphics[width=0.45\linewidth]{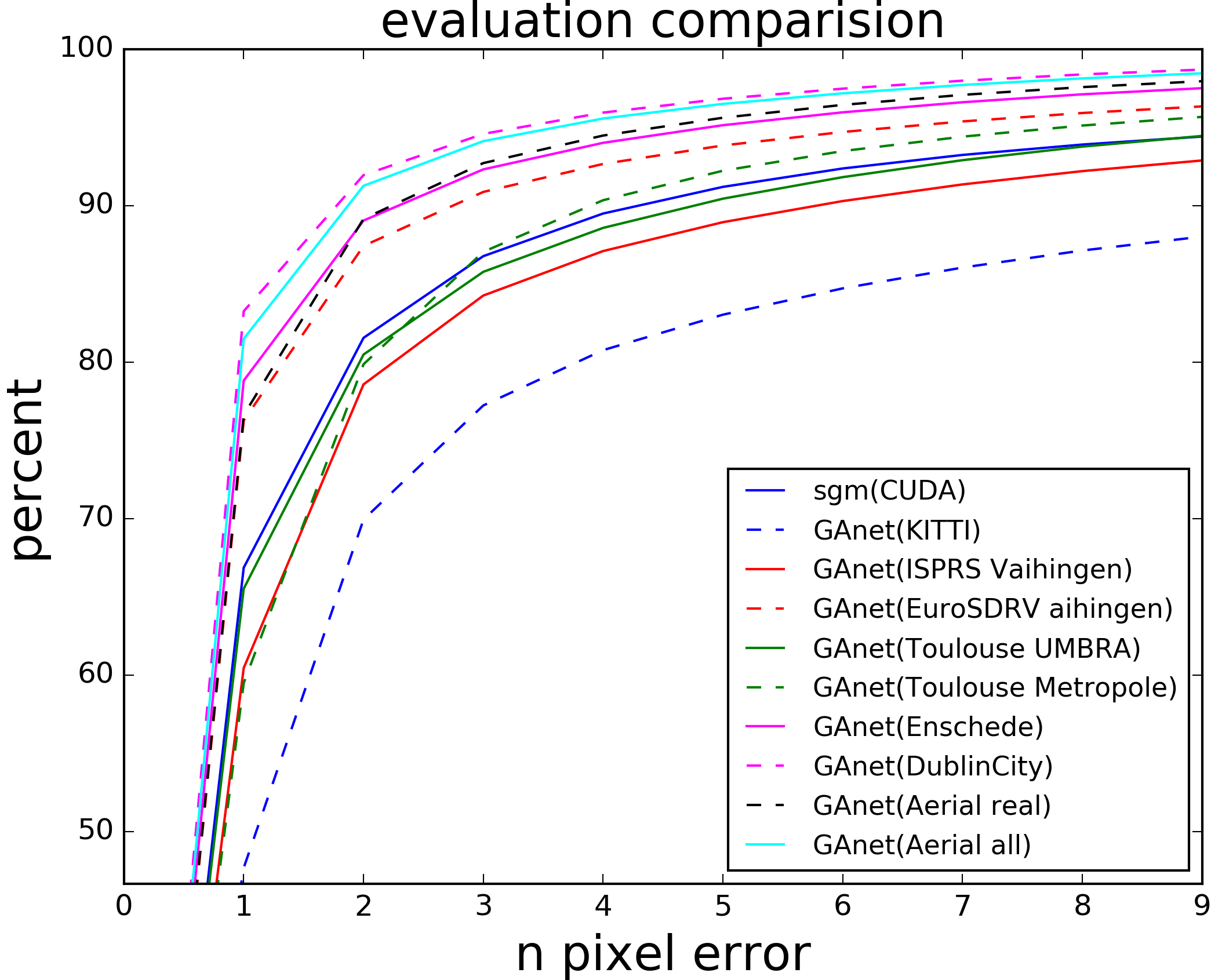}
	}
	
	\caption{Training from multi-dataset using GANet.}
	\label{Figure.fusion_ganet}
\end{figure}

\textit{GANet} focuses on 3D cost volume processing by adding guided aggregation layer. Based on \Cref{Figure.fusion_ganet} we conclude the following:

\begin{itemize}
    \item The \textit{KITTI} model   performs worst than the other models, meaning \textit{GANet} is sensitive to the choice of training data.
    \item It  produces  smooth results, with low scores on 1-pixel error and high scores on large pixel errors (see \Cref{Figure.fusion_ganet:a})
    \item Overall \textit{GANet} performs well and has good transferability, most models are better than SGM(CUDA). 
\end{itemize}






\paragraph{LEAStereo}

\begin{figure}[tp]
	\centering
	\subfigure[Result on ISPRS Vaihingen]{
		\label{Figure.fusion_leastereo:a}
		\centering
		\includegraphics[width=0.45\linewidth]{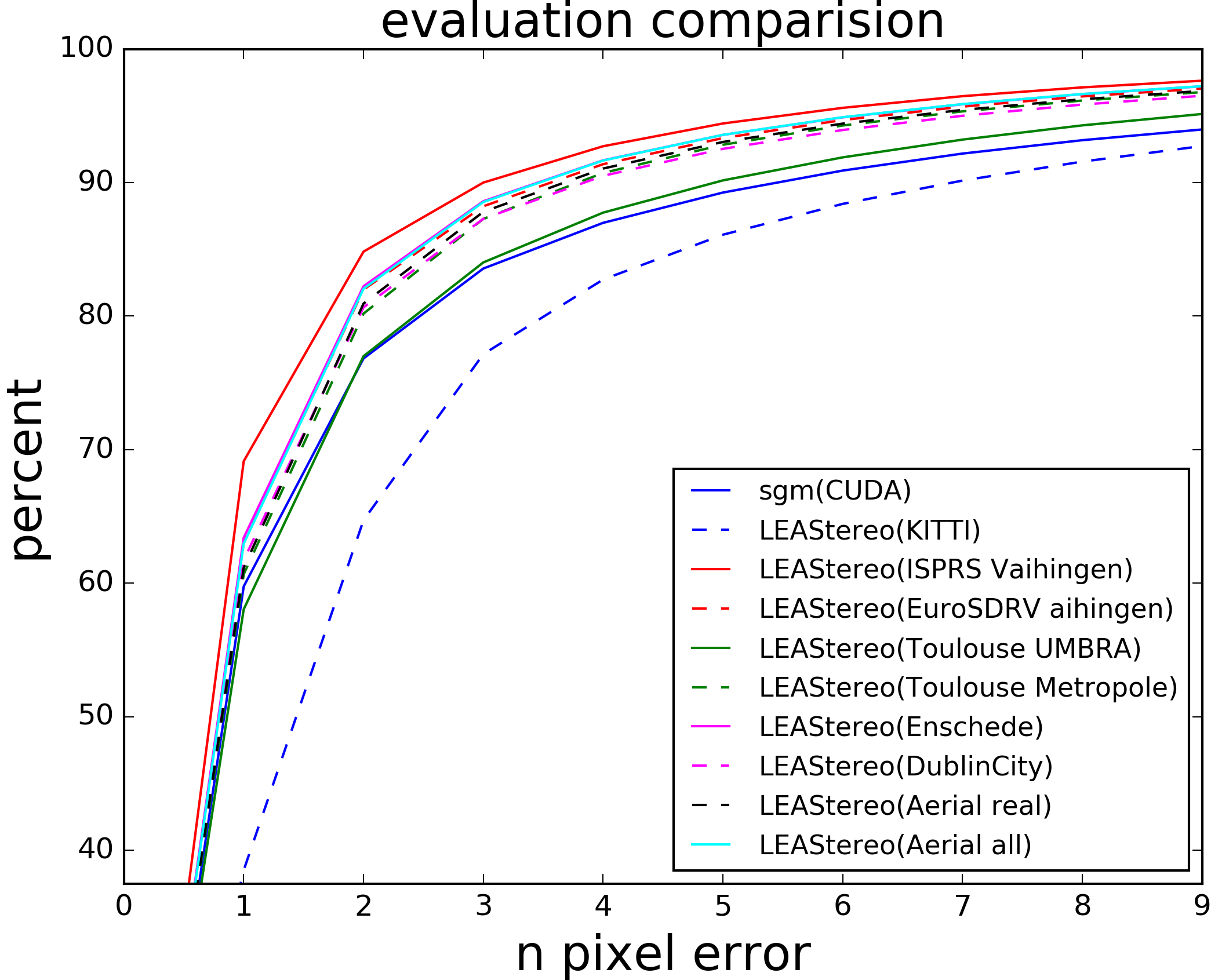}
	}
	\subfigure[Result on EuroSDR Vaihingen]{
		\label{Figure.fusion_leastereo:b}
		\centering
		\includegraphics[width=0.45\linewidth]{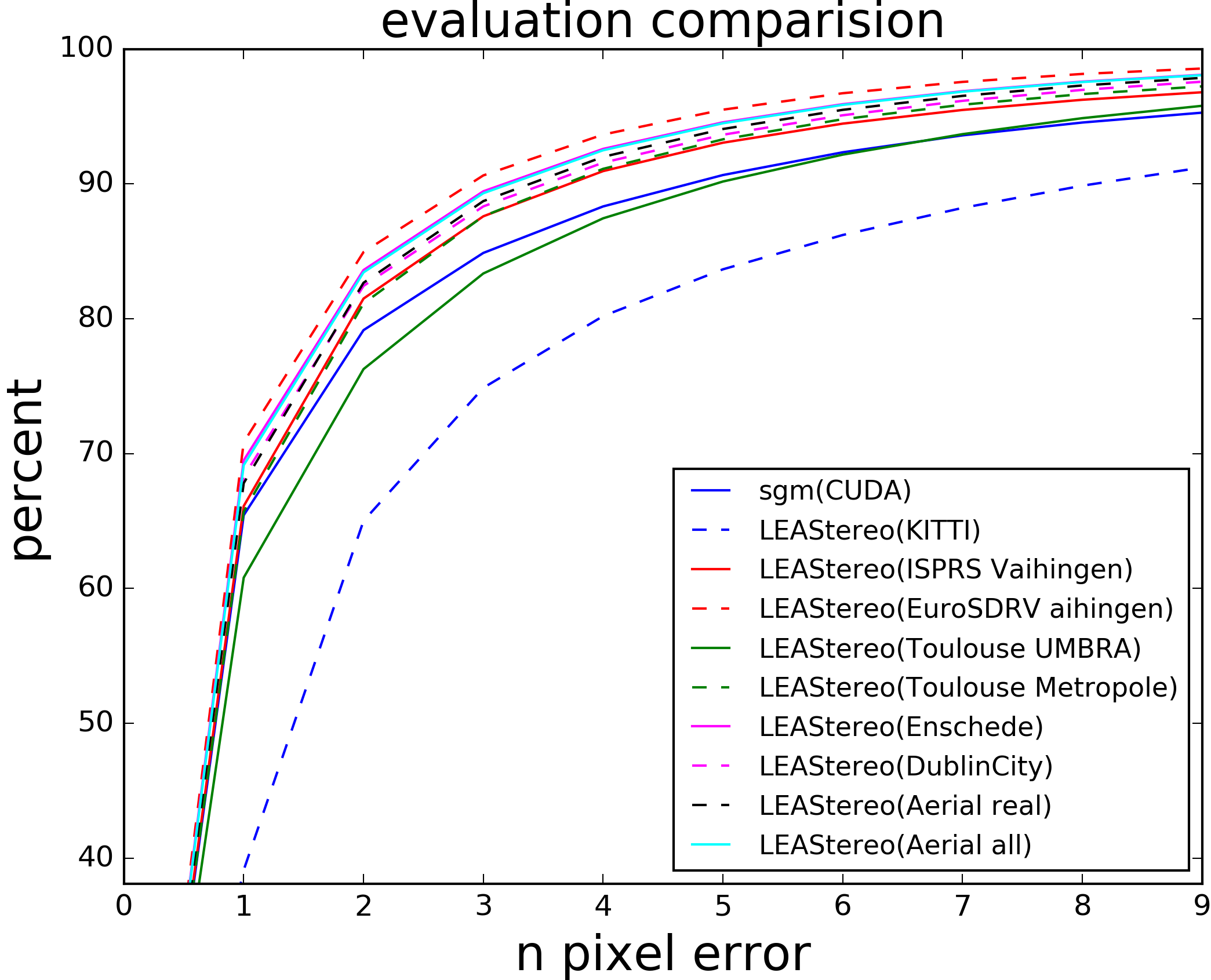}
	}
	
	\subfigure[Result on Toulouse Metropole]{
		\label{Figure.fusion_leastereo:c}
		\centering
		\includegraphics[width=0.45\linewidth]{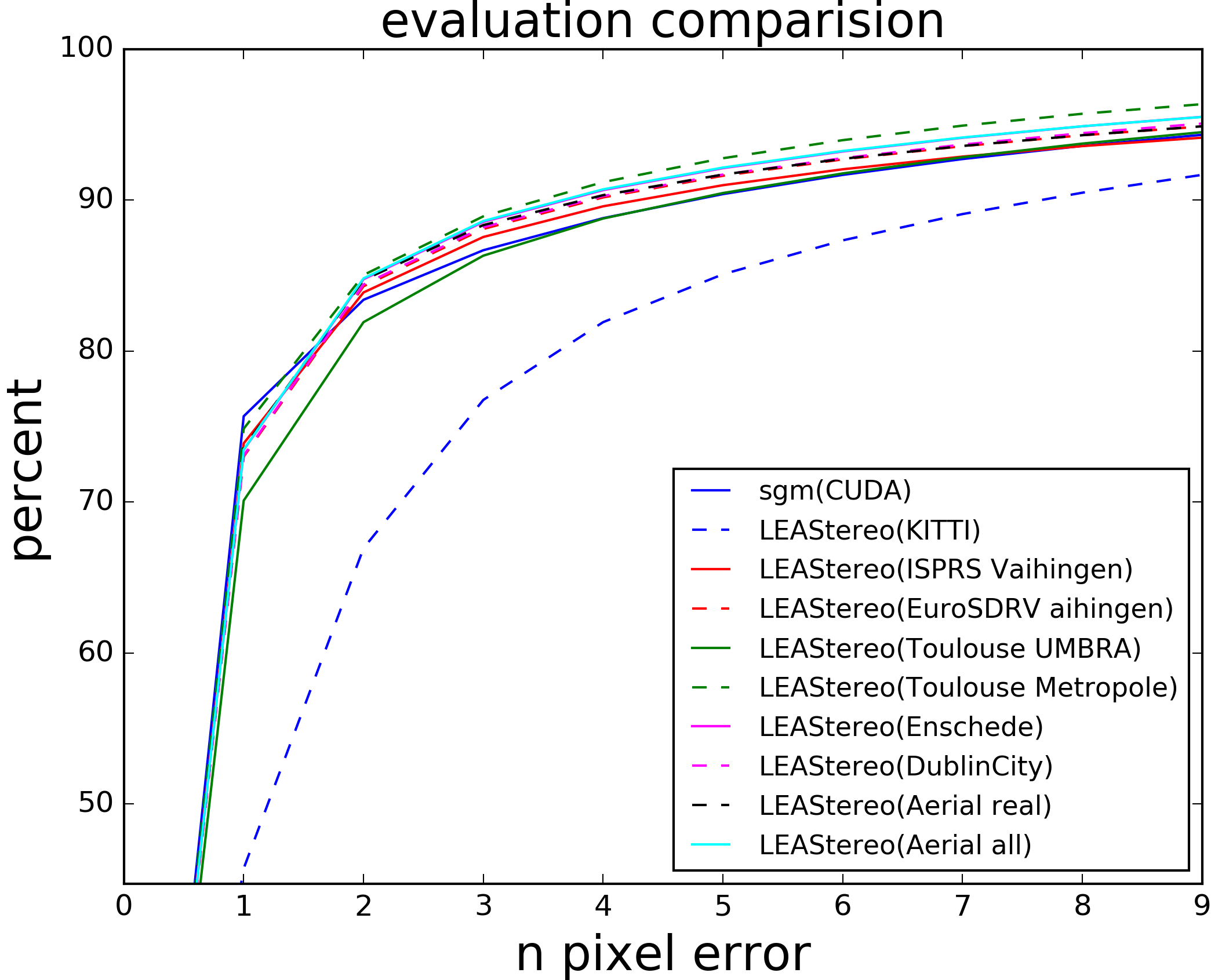}
	}
	\subfigure[Result on Toulouse UMBRA]{
		\label{Figure.fusion_leastereo:d}
		\centering
		\includegraphics[width=0.45\linewidth]{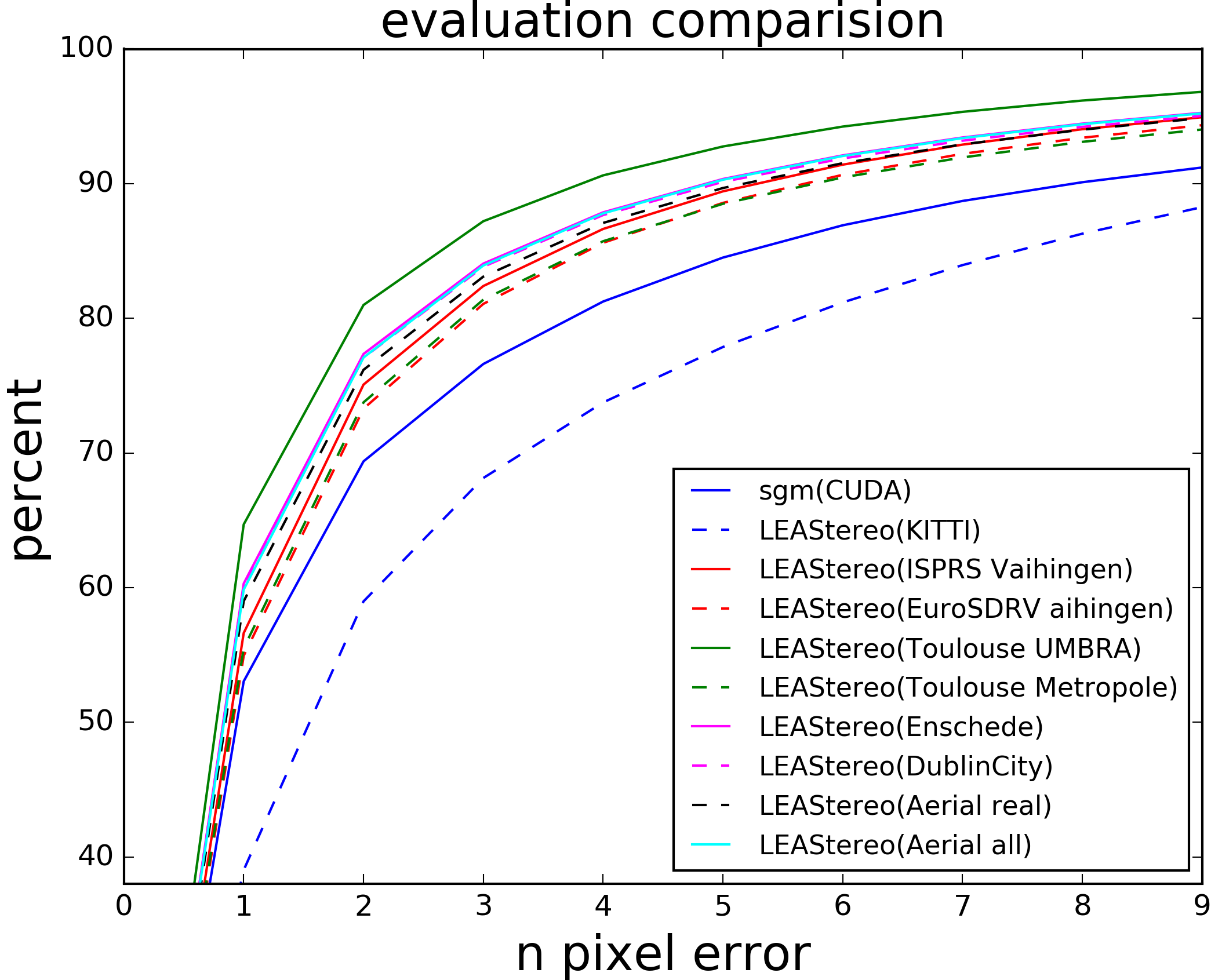}
	}
	
	\subfigure[Result on Enschede]{
		\label{Figure.fusion_leastereo:e}
		\centering
		\includegraphics[width=0.45\linewidth]{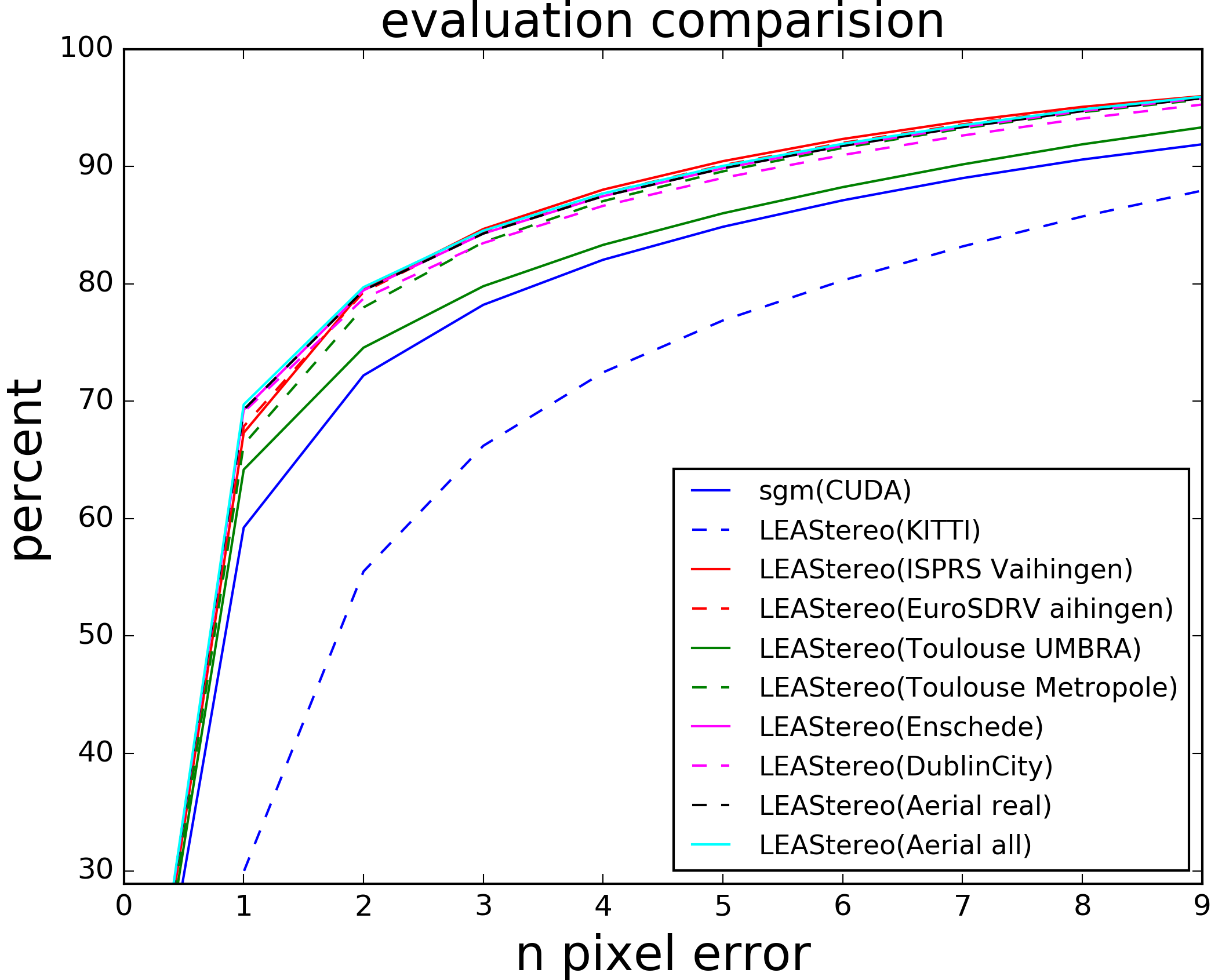}
	}
	\subfigure[Result on DublinCity]{
		\label{Figure.fusion_leastereo:f}
		\centering
		\includegraphics[width=0.45\linewidth]{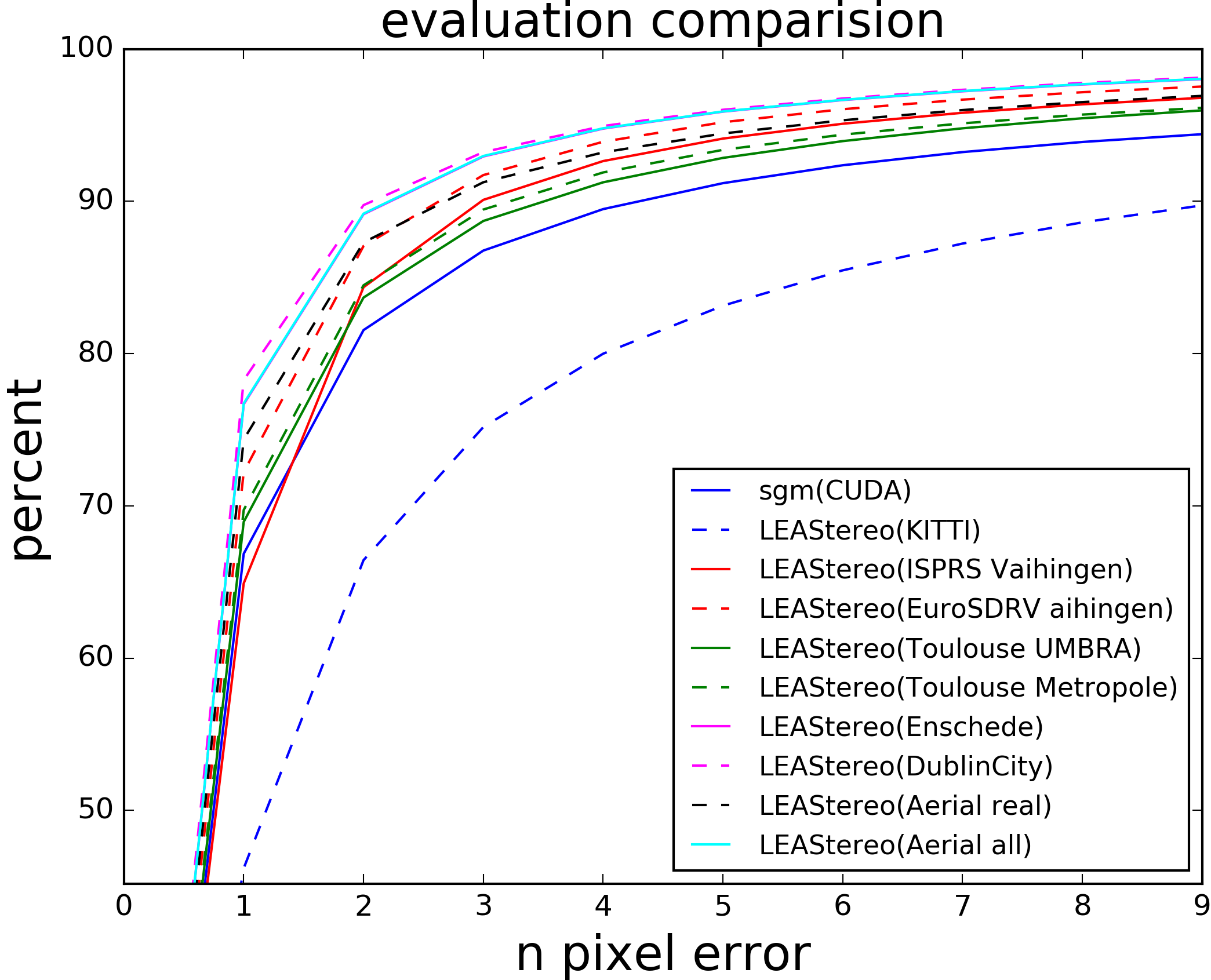}
	}
	
	\caption{Training with multi-dataset using LEAStereo net.}
	\label{Figure.fusion_leastereo}
\end{figure}

\textit{LEAStereo} is an end-to-end method based on Neural Architecture Search, which tries to learn neural network architectures automatically.  
We present the results in \Cref{Figure.fusion_leastereo} and conclude the following:

\begin{itemize}
    \item It  highly depends on the training data, models from KITTI perform worst, and training on aerial data leads to a significant improvement.
    \item It has good transferability among the aerial dataset, these lines are quite close.
    \item It produces  smooth results. As shown in \Cref{Figure.fusion_leastereo:c}, for the 1-pixel error, the fine-trained models do not have any advantage, but at large pixel error, the training models on aerial data outperform the SGM(CUDA).
\end{itemize}






\subsection{Discussion of Dataset shift}
In general, we can make a conclusion:
\begin{itemize}
    \item The transferability of hybrid methods is better than that of end-to-end methods on different types of datasets. Training on \textit{KITTI} is not far from training from aerial data (see \Cref{Figure.fusion_mccnn} and \Cref{Figure.fusion_effcnn}). But for the same type of dataset, in the paper for the aerial scene, the end-to-end methods have better transferability.
    \item End-to-end methods highly depend on the training dataset, training on \textit{KITTI} performs poorly when tested on aerial data.
    \item DL-base methods usually give better results on high pixel error.
    \item The transferability is not symmetric, \textit{DublinCity} works well on Toulouse data, but Toulouse data performs less well on \textit{DublinCity}(cf \Cref{Table:quantity_ave_pixel}).
    \item The transferability depends on the network structure, the scene, the resolution, and the base to height ratio, generally, \textit{DeepPruner} performs better than \textit{ HRS net}, but sometimes it gives bad results, for example, Toulouse data on \textit{DublinCity} in \Cref{Figure.fusion_deepprune:f}.
\end{itemize}

\section{Visual assessment}

\begin{figure}[tp]
	\begin{minipage}[t]{0.19\textwidth}
 		\includegraphics[width=\linewidth]{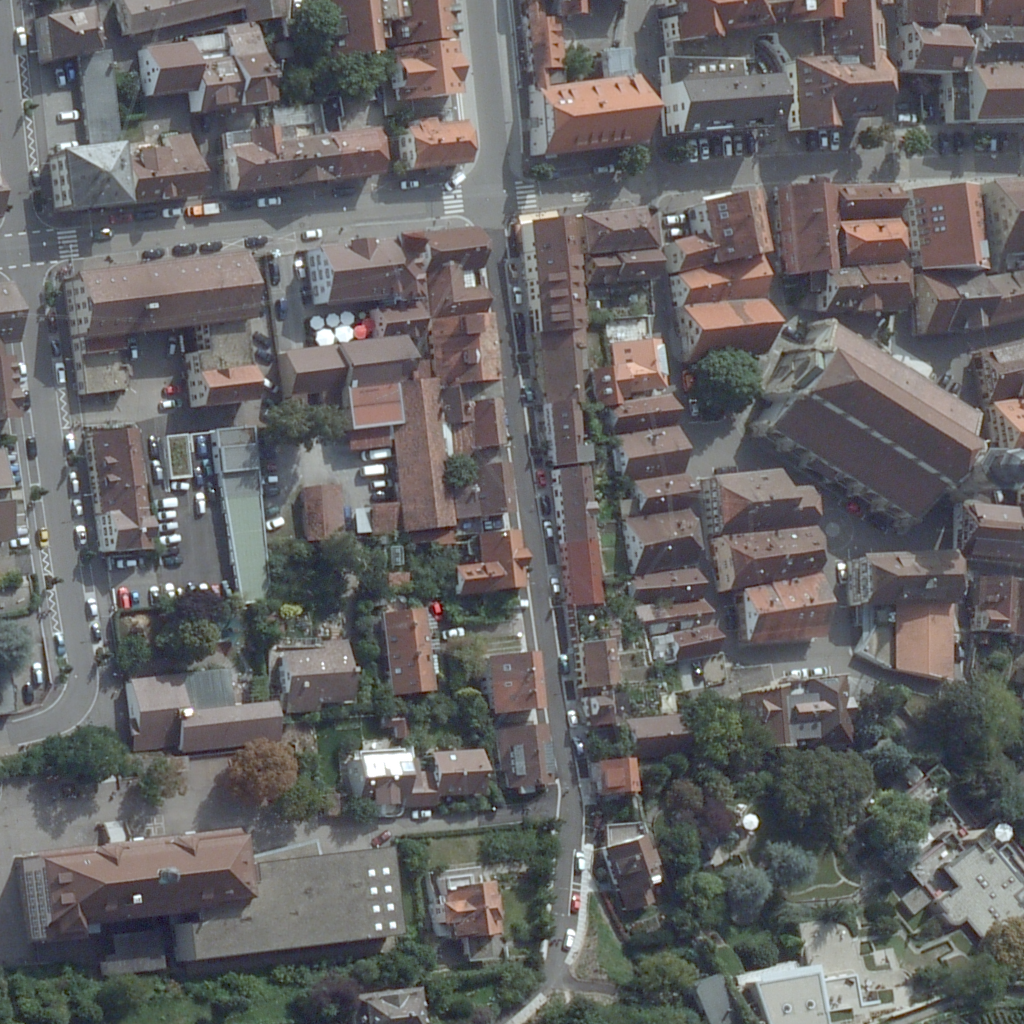} \\
 		\includegraphics[width=\linewidth]{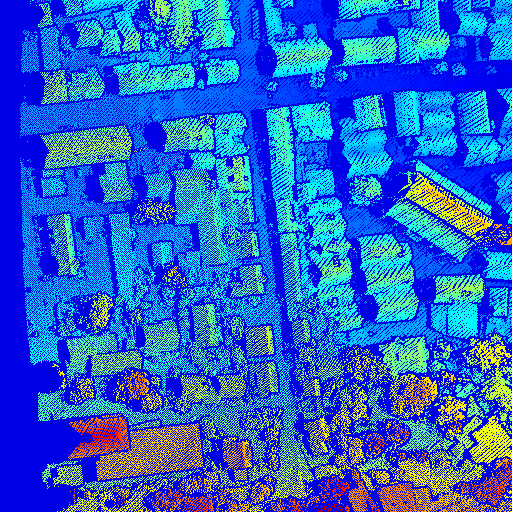} \\
 		\centering{\tiny (a)EuroSDR Vaihingen}
	\end{minipage}
 	\begin{minipage}[t]{0.19\textwidth}
 		\includegraphics[width=\linewidth]{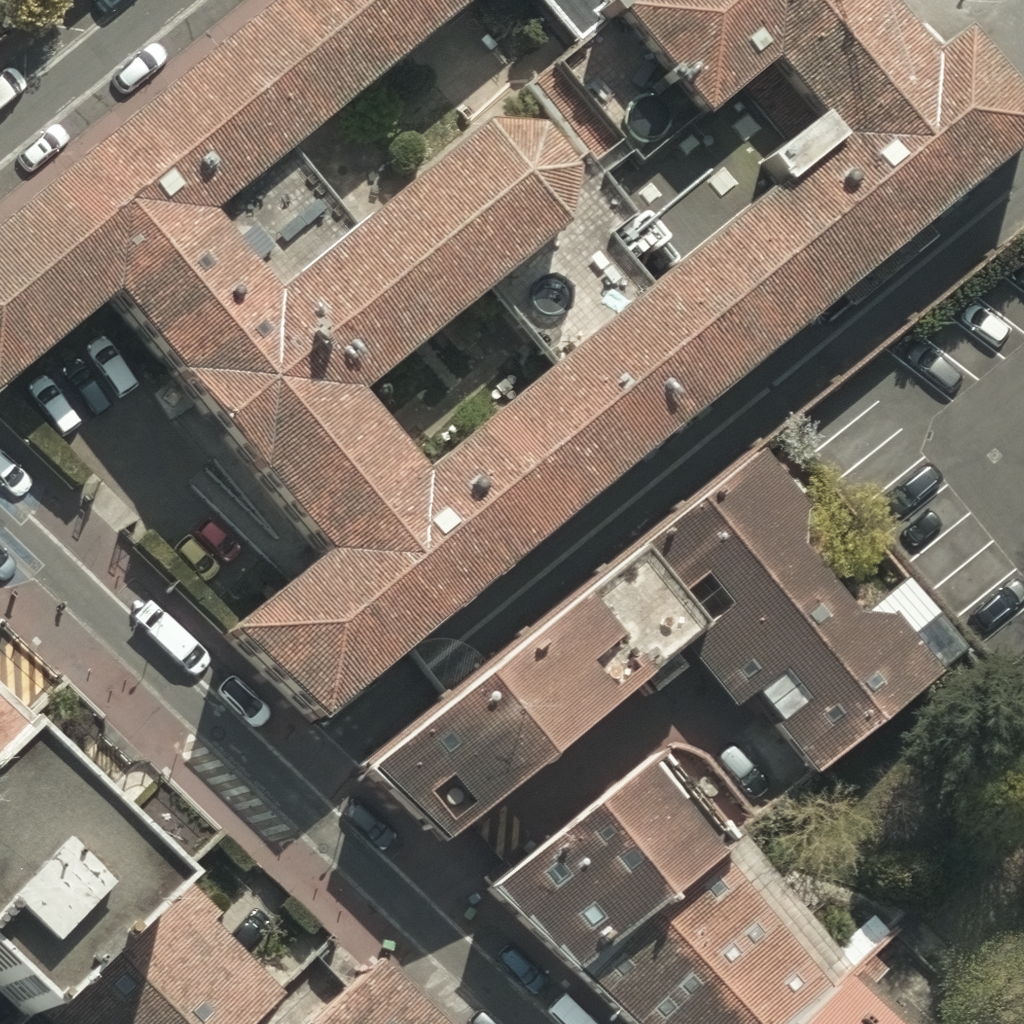} \\
 		\includegraphics[width=\linewidth]{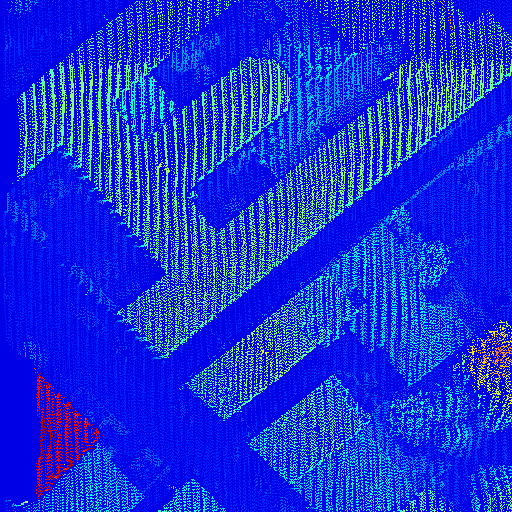} \\
 		\centering{\tiny (b)Toulouse Metropole}
 	\end{minipage}
 	\begin{minipage}[t]{0.19\textwidth}
 		\includegraphics[width=\linewidth]{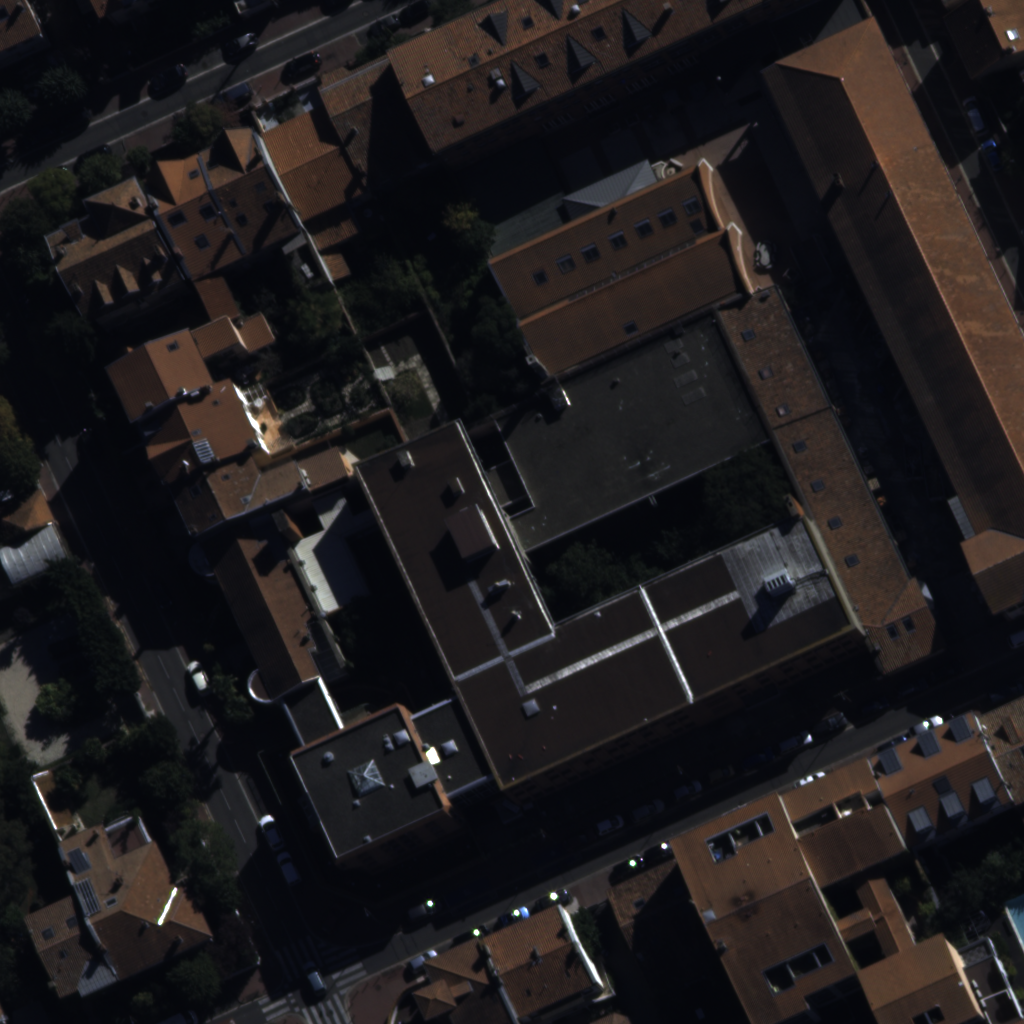}\\
 		\includegraphics[width=\linewidth]{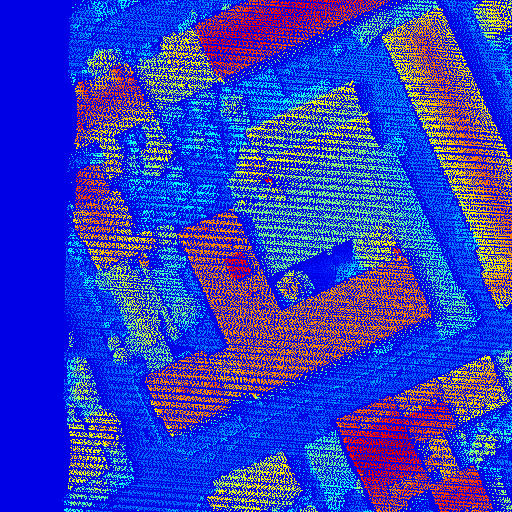}\\
 		\centering{\tiny (c)Toulouse UMBRA}
 	\end{minipage}
	\begin{minipage}[t]{0.19\textwidth}
		\includegraphics[width=\linewidth]{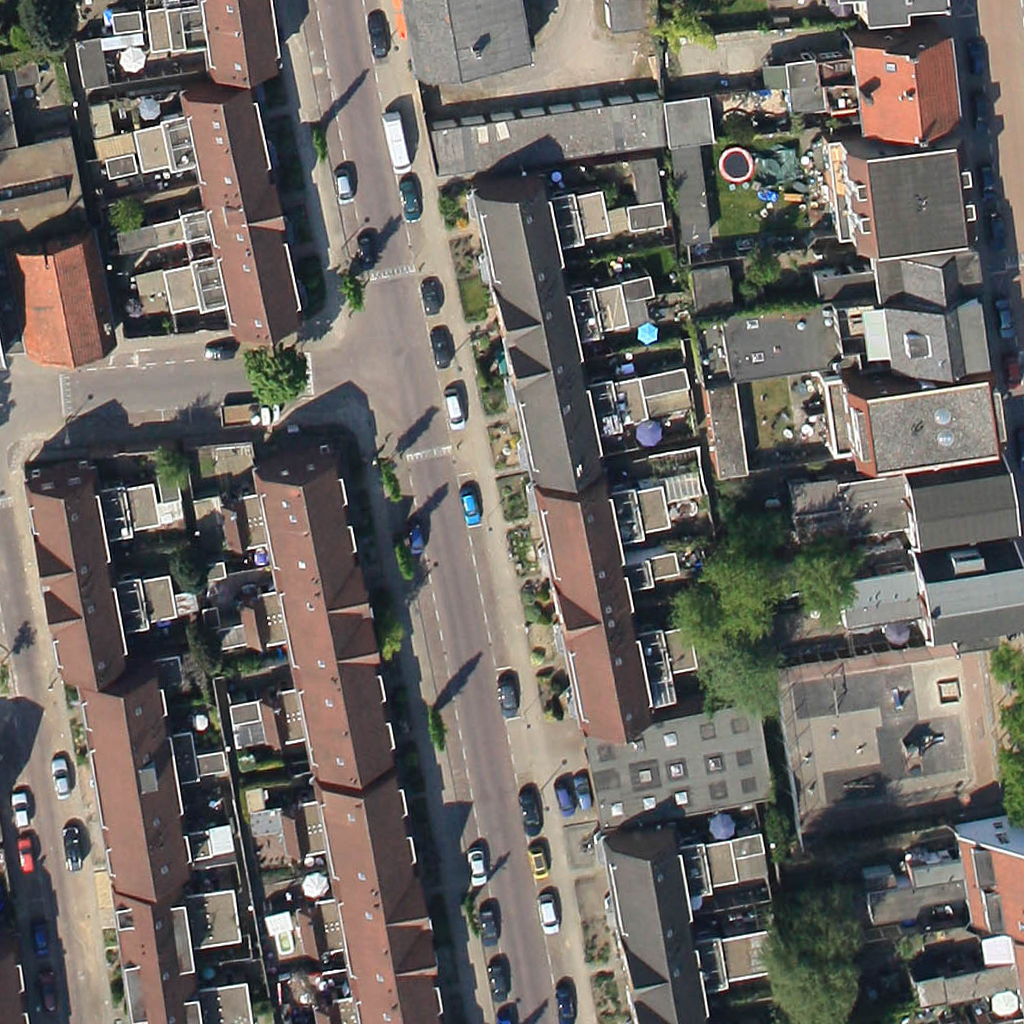}\\
		\includegraphics[width=\linewidth]{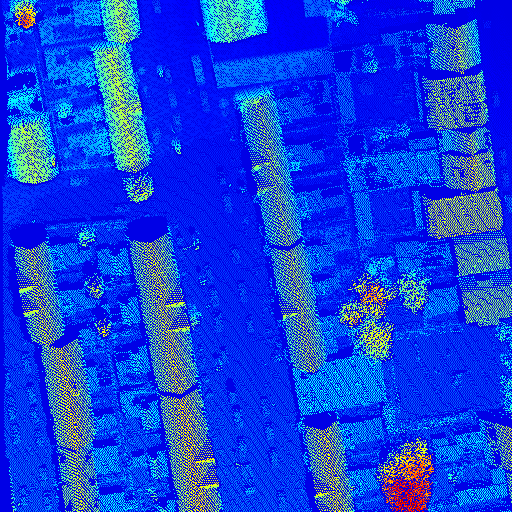}\\
		\centering{\tiny (d)Enschede}
	\end{minipage}
	\begin{minipage}[t]{0.19\textwidth}
		\includegraphics[width=\linewidth]{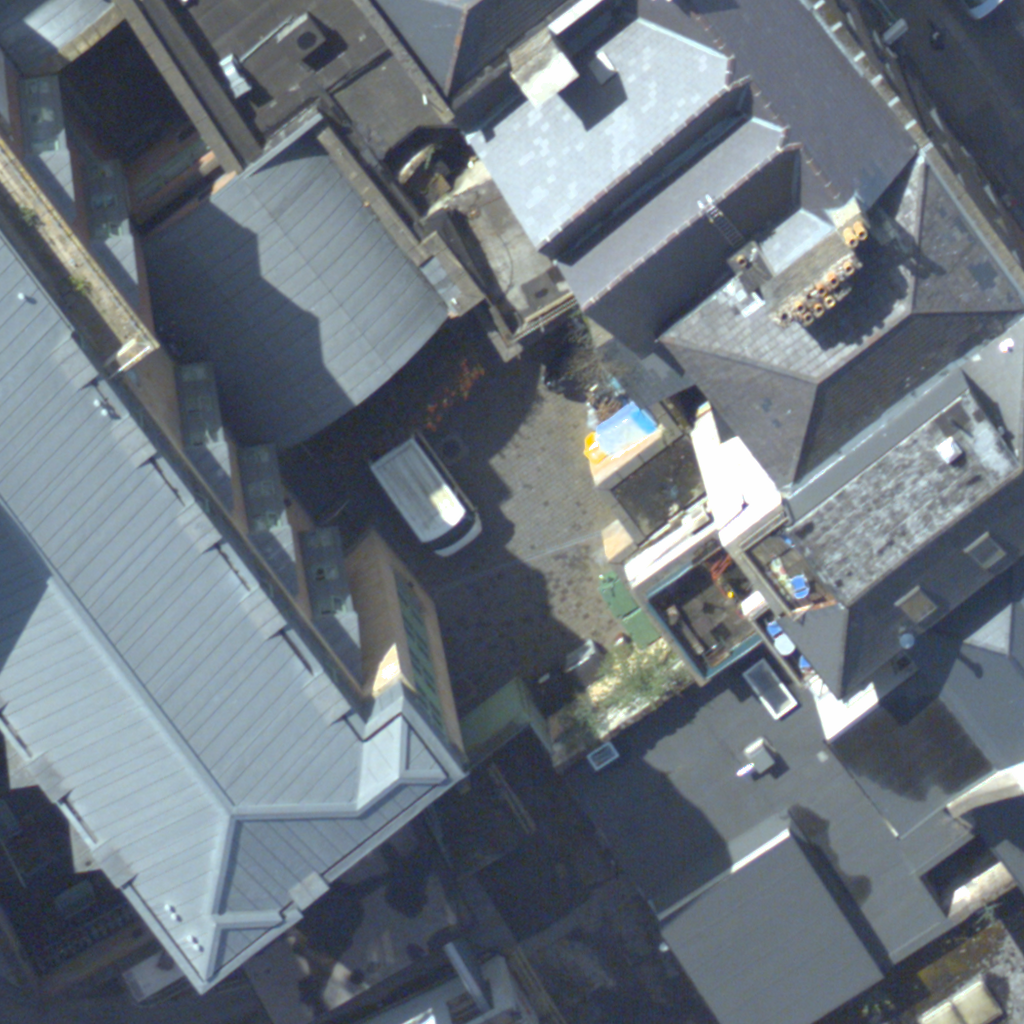}\\
		\includegraphics[width=\linewidth]{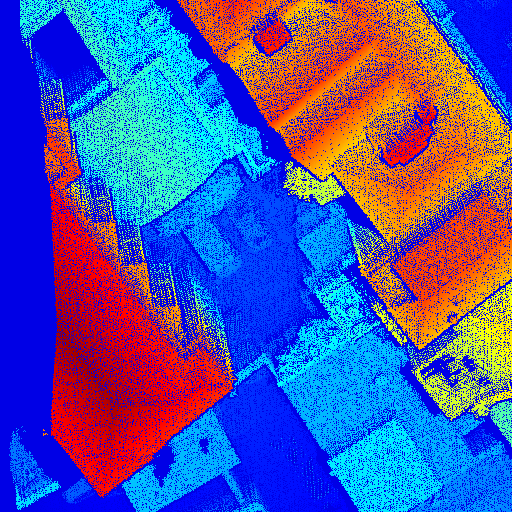}\\
	\centering{\tiny (e)DublinCity}
	\end{minipage}
        \centering
 	\caption{Examples for the building area for each dataset. The left image is shown in the first row, and the disparity is shown in the second row in jet color, from blue to red indicating the disparity from small to large.}
 	\label{Figure.example_bulding}
\end{figure}

\begin{figure}[tp]
	\begin{minipage}[t]{0.19\textwidth}
		\includegraphics[width=\linewidth]{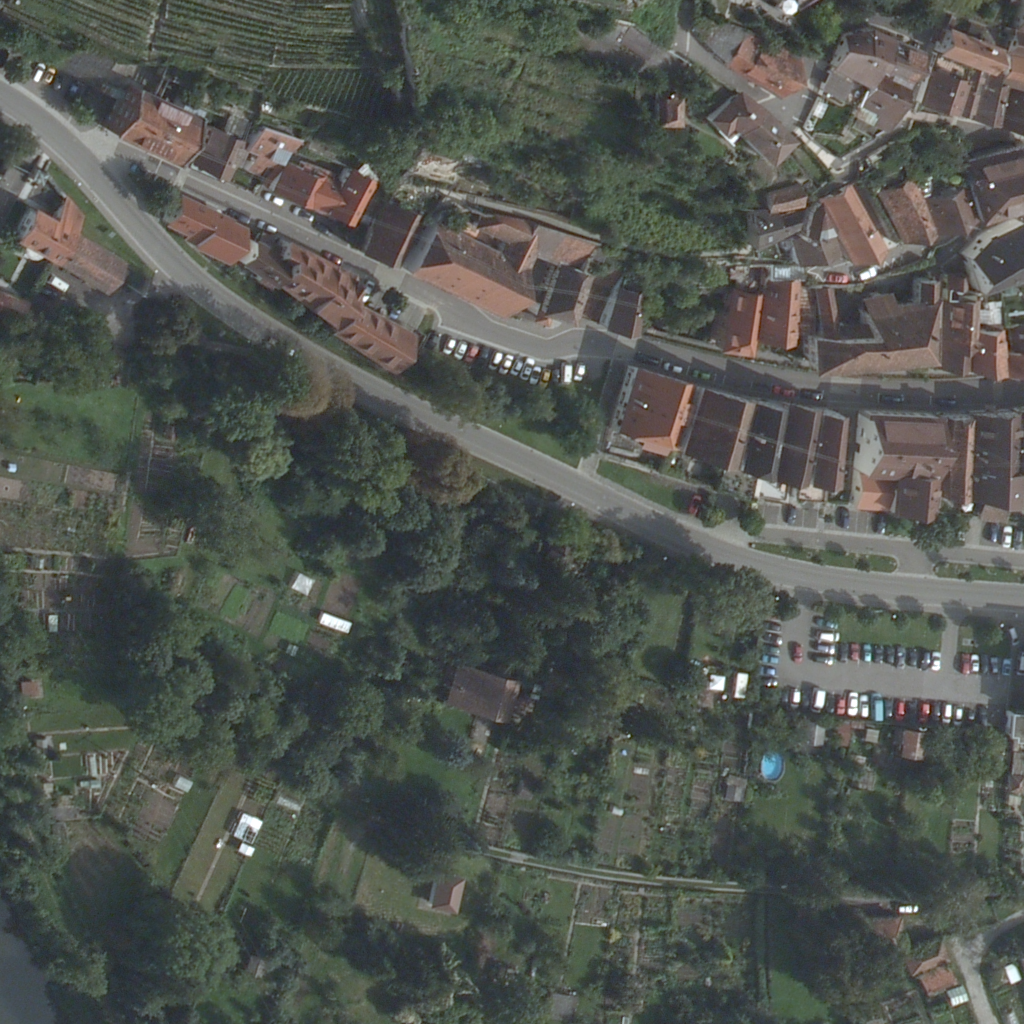} \\
		\includegraphics[width=\linewidth]{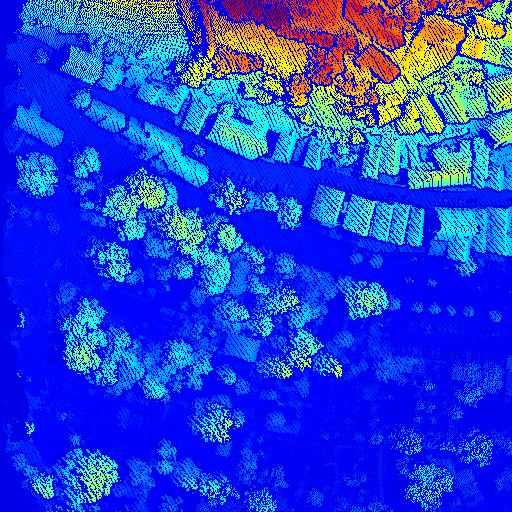} \\
		\centering{\tiny (a)EuroSDR Vaihingen}
	\end{minipage}
	\begin{minipage}[t]{0.19\textwidth}
		\includegraphics[width=\linewidth]{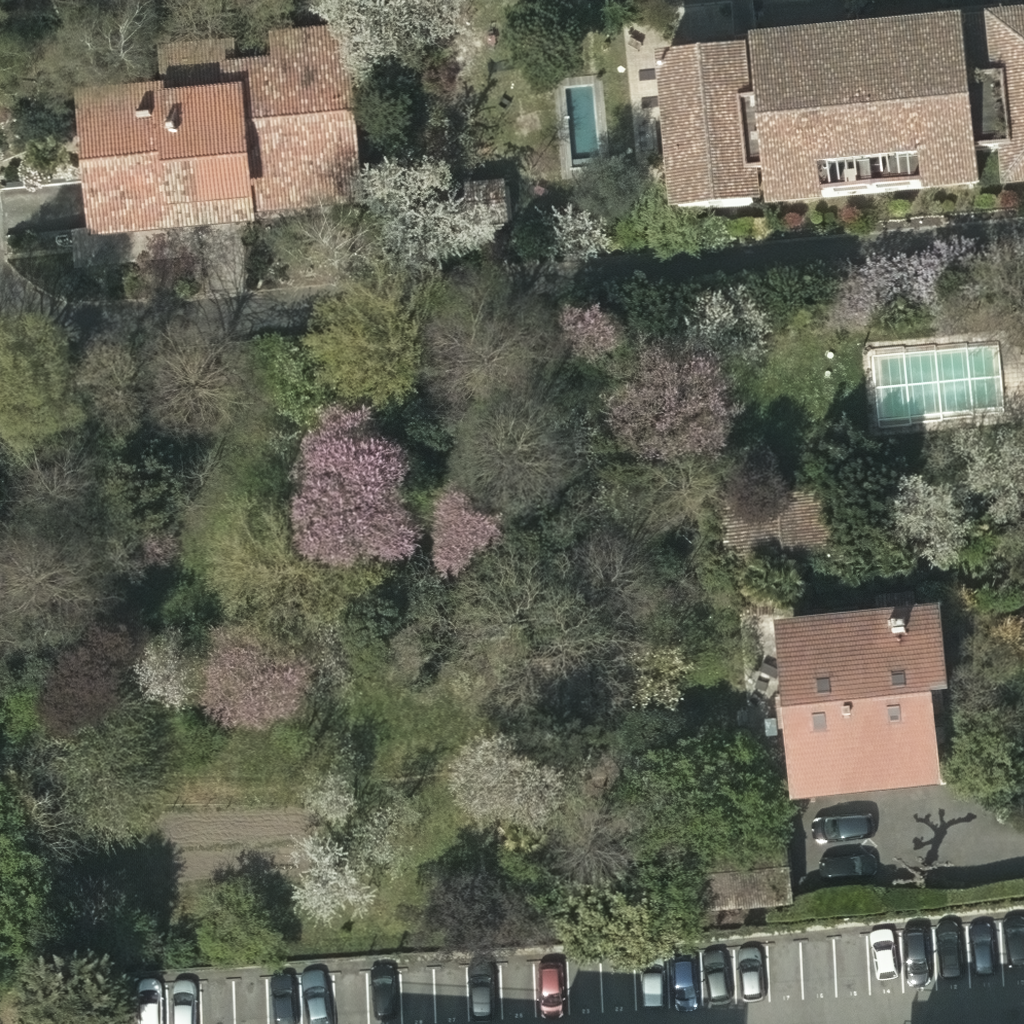} \\
		\includegraphics[width=\linewidth]{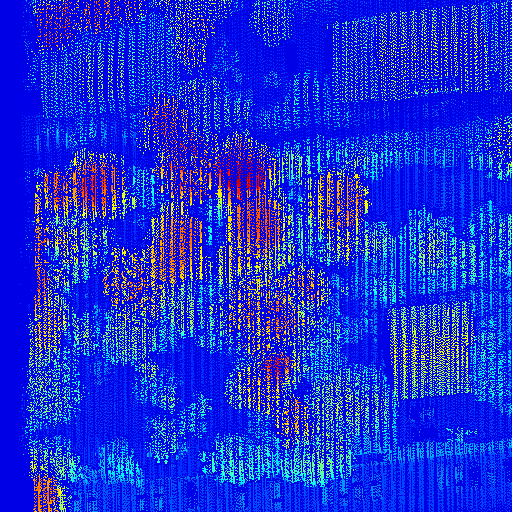} \\
		\centering{\tiny (b)Toulouse Metropole}
	\end{minipage}
	\begin{minipage}[t]{0.19\textwidth}
		\includegraphics[width=\linewidth]{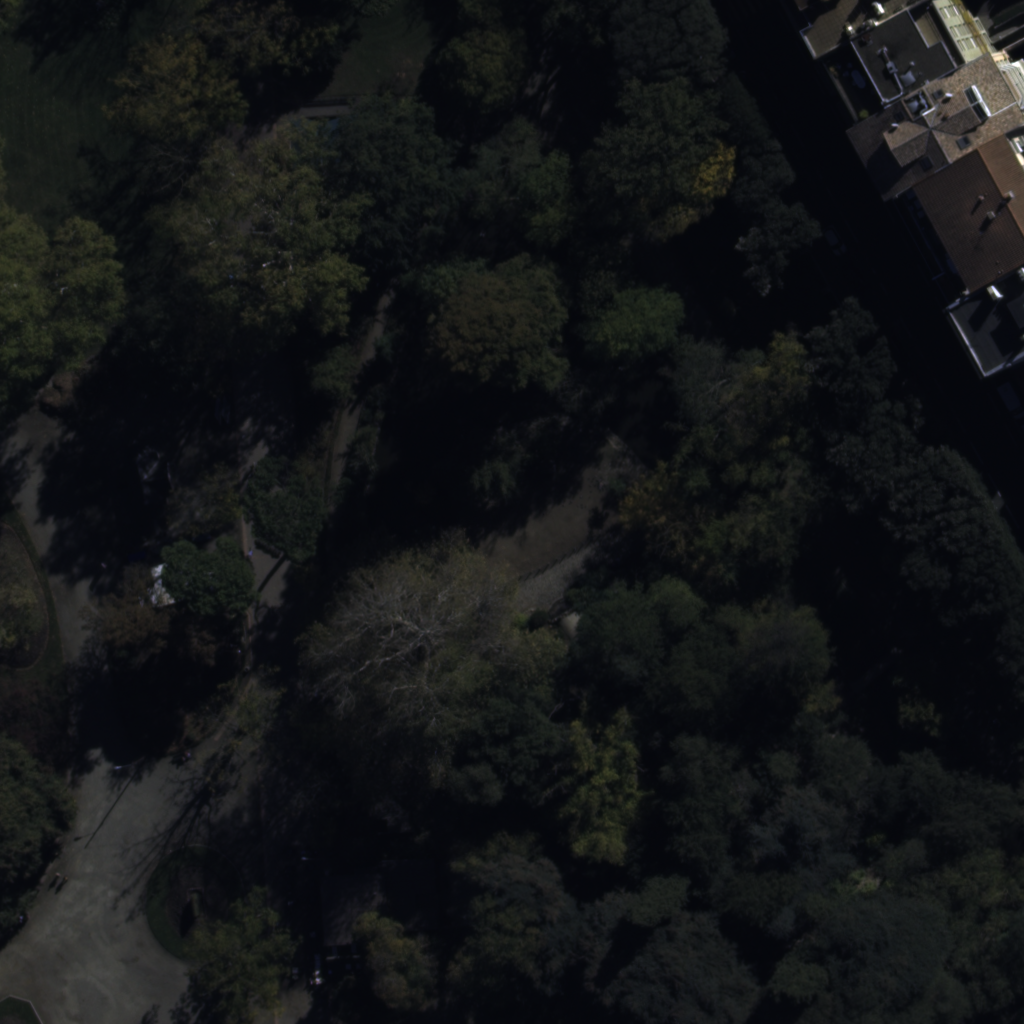}\\
		\includegraphics[width=\linewidth]{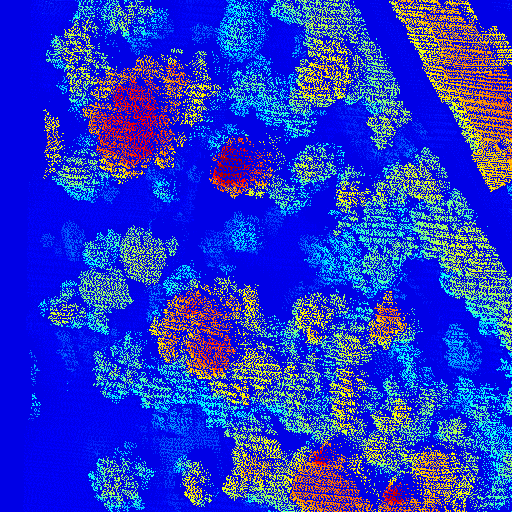}\\
		\centering{\tiny (c)Toulouse UMBRA}
	\end{minipage}
	\begin{minipage}[t]{0.19\textwidth}
		\includegraphics[width=\linewidth]{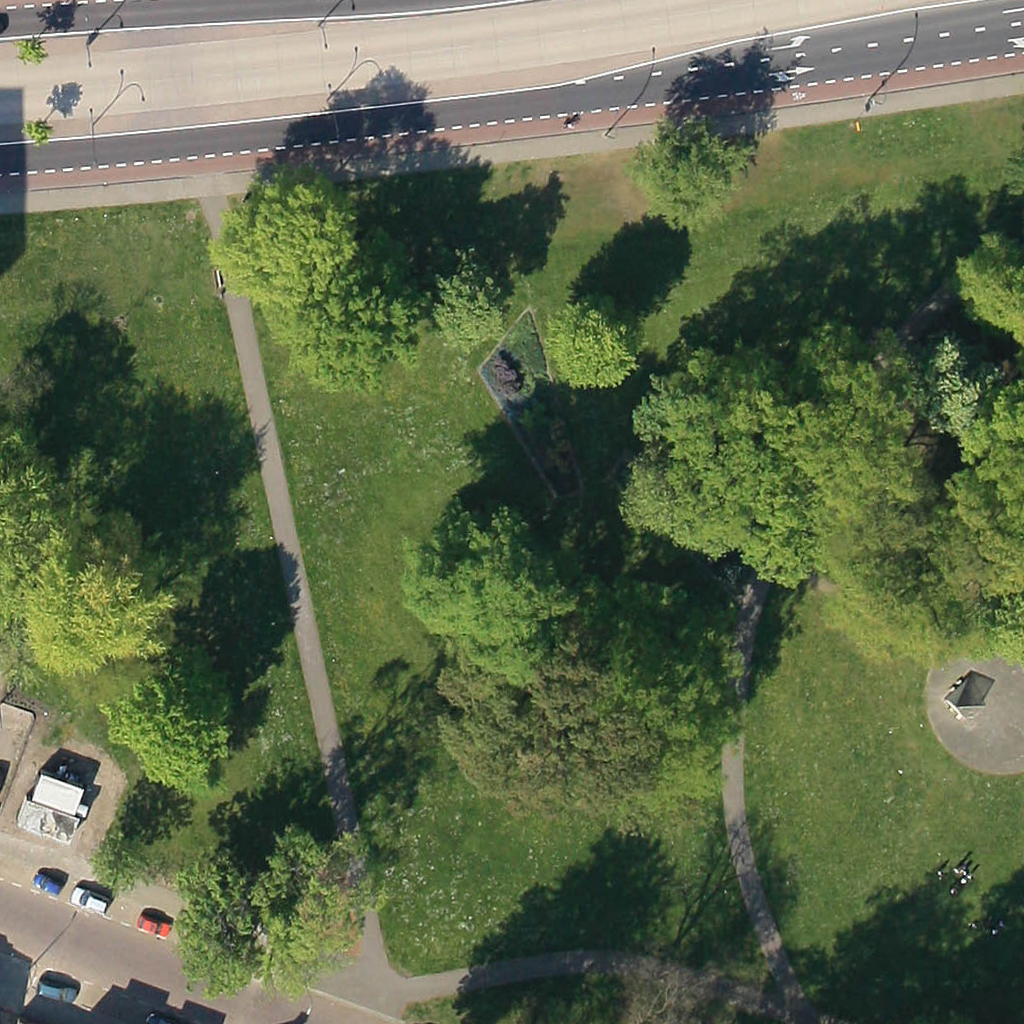}\\
		\includegraphics[width=\linewidth]{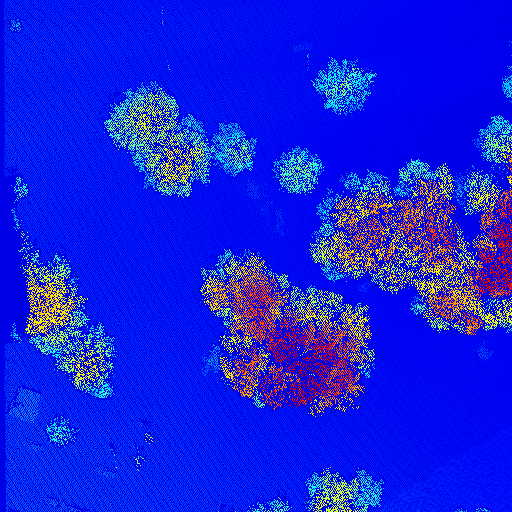}\\
		\centering{\tiny (d)Enschede}
	\end{minipage}
	\begin{minipage}[t]{0.19\textwidth}
		\includegraphics[width=\linewidth]{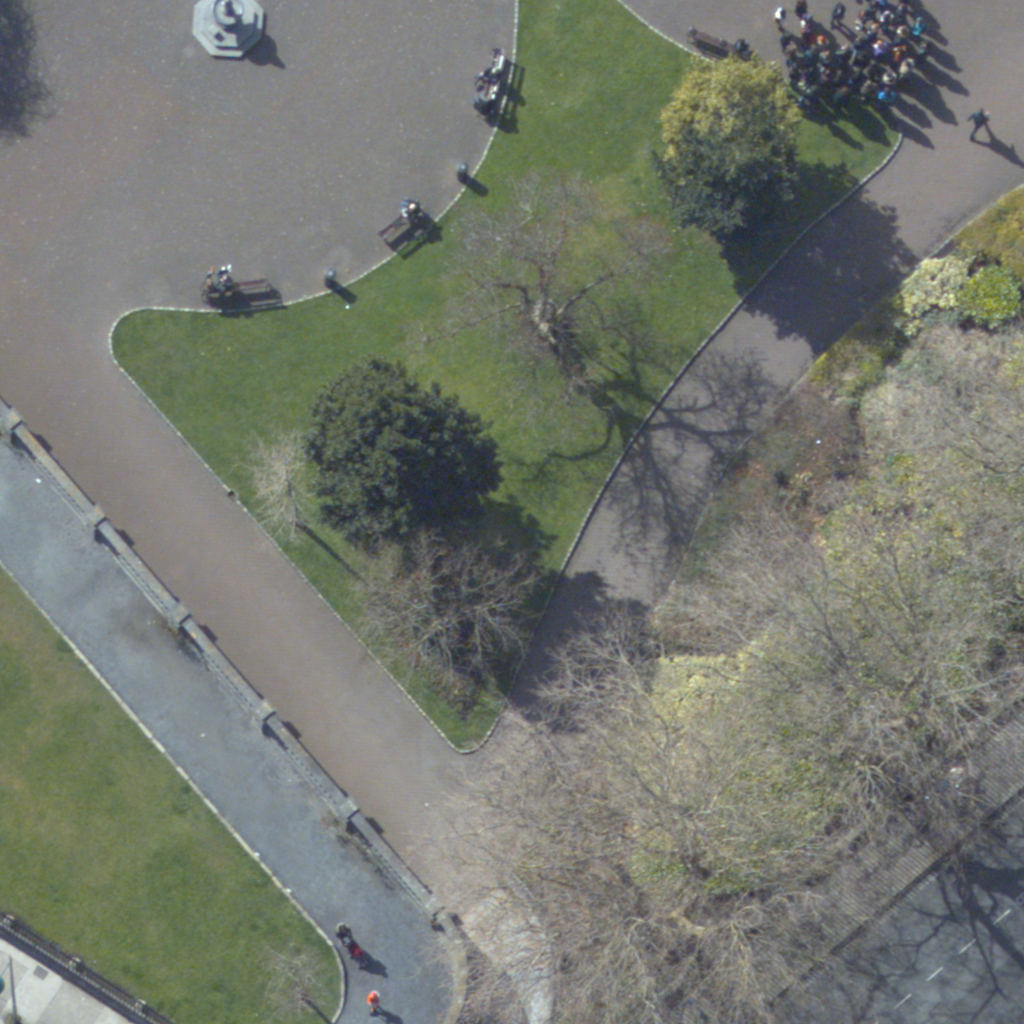}\\
		\includegraphics[width=\linewidth]{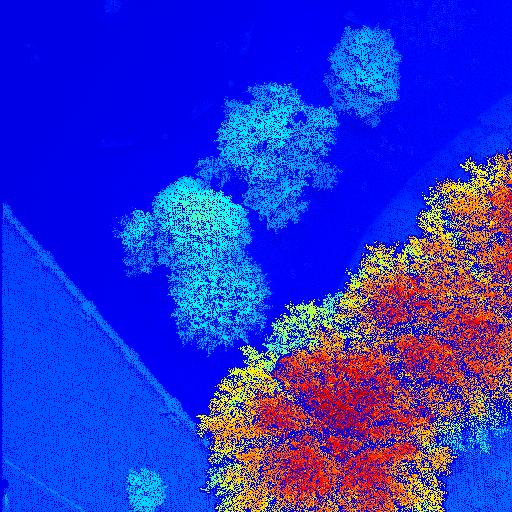}\\
	\centering{\tiny (e)DublinCity}
	\end{minipage}
        \centering
	\caption{Examples for the tree area for each dataset. The left image is shown in the first row, and the disparity is shown in the second row in jet color, from blue to red indicating the disparity from small to large.}
	\label{Figure.example_tree}
\end{figure}

The pixel error metric describes the precision and robustness of a method through a global score. Visual assessment, on the other hand, gives insights into the quality of depth prediction performance across different surface types, e.g. in built-up areas or in natural environments. In the following, we reveal the performance of the other 5 datasets (the 1$^{st}$ being already covered in the main paper) over buildings (cf. \Cref{Figure.example_bulding}) and vegetation (cf. \Cref{Figure.example_tree}).  \Cref{Table:visual_table} sums up the methods' performance at selected scene's objects such as small buildings, shadows, discontinuities, leaves and branches.

\begin{table}[tp]
   \centering
   \caption{Summary of performance of reconstruction in visual assessment}
   \label{Table:visual_table}
    \begin{center}
       \resizebox{\textwidth}{!}{
        \begin{tabular}{c|ccccc|cc}
           	\noalign{\hrule height 1pt}
           	Method & small building & shadow & discontinuity & leaf & branch & pre-trained & trained \\
           	\noalign{\hrule height 1pt}
           	SGM(CUDA) & \Checkmark & \XSolid & \XSolid &  \XSolid & \XSolid & -- & --\\
           	MICMAC &  \Checkmark & \XSolid & \XSolid &  \Checkmark & \XSolid & -- & --\\
           	GraphCuts & \XSolid & \XSolid & \XSolid &  \XSolid  & \XSolid & -- & --\\
           	\noalign{\hrule height 1pt}
           	CBMV(SGM) & \Checkmark & \Checkmark & \XSolid &  \Checkmark & \XSolid & -- & \Checkmark \\
           	CBMV(GraphCuts) & \Checkmark & \Checkmark & \XSolid &  \XSolid & \XSolid & -- & \Checkmark \\
           	\noalign{\hrule height 1pt}
           	MC-CNN(KITTI) &  \XSolid & \Checkmark & \XSolid &  \XSolid & \XSolid & \Checkmark & -- \\
           	DeepFeature(KITTI) &  \XSolid & \Checkmark & \XSolid &  \XSolid & \XSolid & \Checkmark & -- \\
           	\hline
           	MC-CNN & \Checkmark & \Checkmark & \XSolid &  \XSolid & \XSolid & -- & \Checkmark \\
           	DeepFeature & \Checkmark & \Checkmark & \XSolid &  \XSolid & \XSolid & -- & \Checkmark \\
           	\noalign{\hrule height 1pt}
           	PSM Net(KITTI) &  \XSolid & \XSolid & \XSolid &  \XSolid & \XSolid & \Checkmark & -- \\
           	HRS Net(KITTI) &  \XSolid & \XSolid & \XSolid &  \XSolid & \XSolid & \Checkmark & -- \\
           	DeepPruner(KITTI) &  \XSolid & \XSolid & \XSolid &  \XSolid & \XSolid & \Checkmark & -- \\
           	GANet(KITTI) &  \XSolid & \XSolid & \XSolid &  \XSolid & \XSolid & \Checkmark & -- \\
           	LEAStereo(KITTI) & \XSolid & \XSolid & \XSolid &  \XSolid & \XSolid & \Checkmark & -- \\
           	\hline
           	PSM Net & \Checkmark &  \XSolid & \Checkmark &   \Checkmark  & \XSolid & -- & \Checkmark \\
           	HRS Net & \Checkmark & \Checkmark & \XSolid &   \Checkmark & \XSolid & -- & \Checkmark \\
           	DeepPruner & \Checkmark & \Checkmark & \XSolid &  \Checkmark  & \XSolid & -- & \Checkmark \\
           	GANet & \Checkmark & \Checkmark & \Checkmark &  \Checkmark   & \XSolid & -- & \Checkmark \\
           	LEAStereo & \Checkmark & \Checkmark & \XSolid &   \Checkmark & \XSolid & -- & \Checkmark \\
           	\noalign{\hrule height 1pt}
        \end{tabular}
       	}
    \end{center}
\end{table}

\paragraph{EuroSDR Vaihingen}

\begin{figure}[tp]
	\begin{minipage}[t]{0.19\textwidth}
		\includegraphics[width=0.098\linewidth]{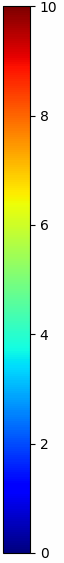}
		\includegraphics[width=0.85\linewidth]{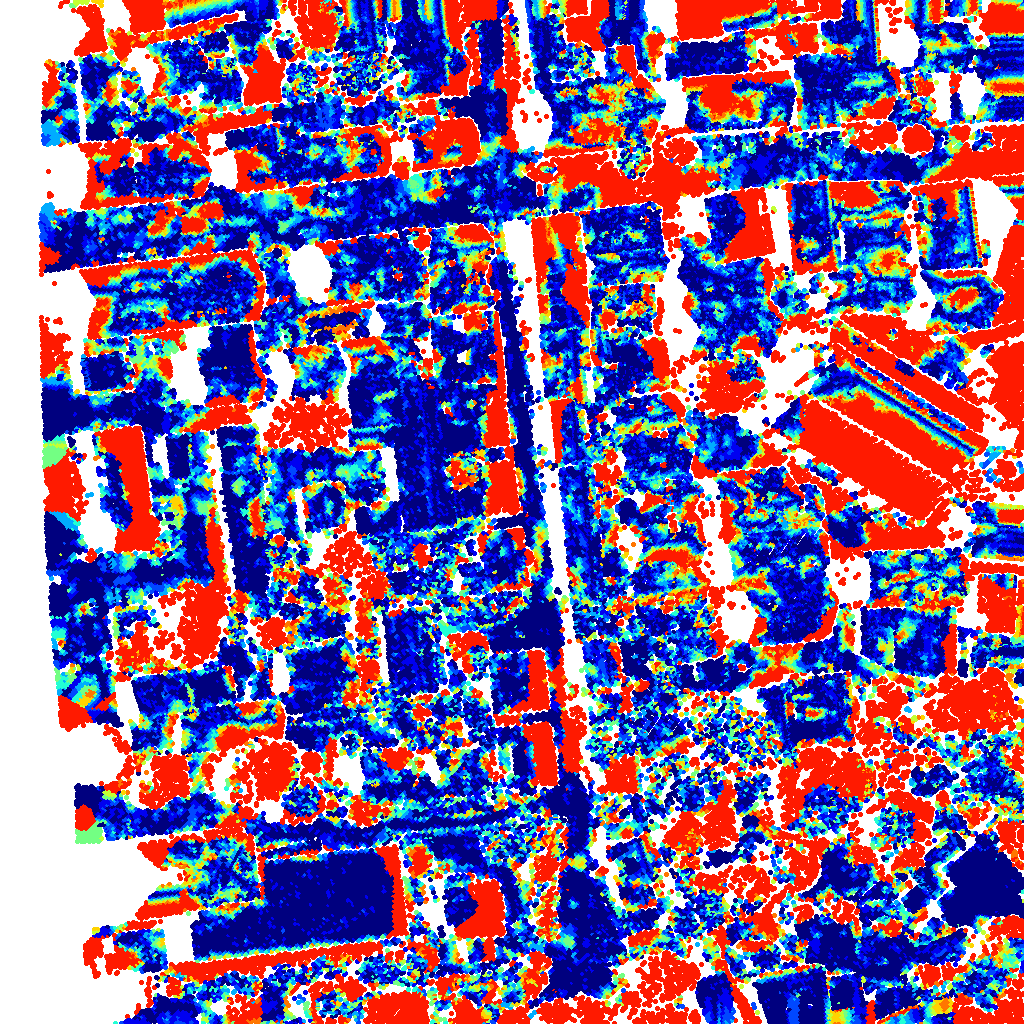} \\
		\includegraphics[width=\linewidth]{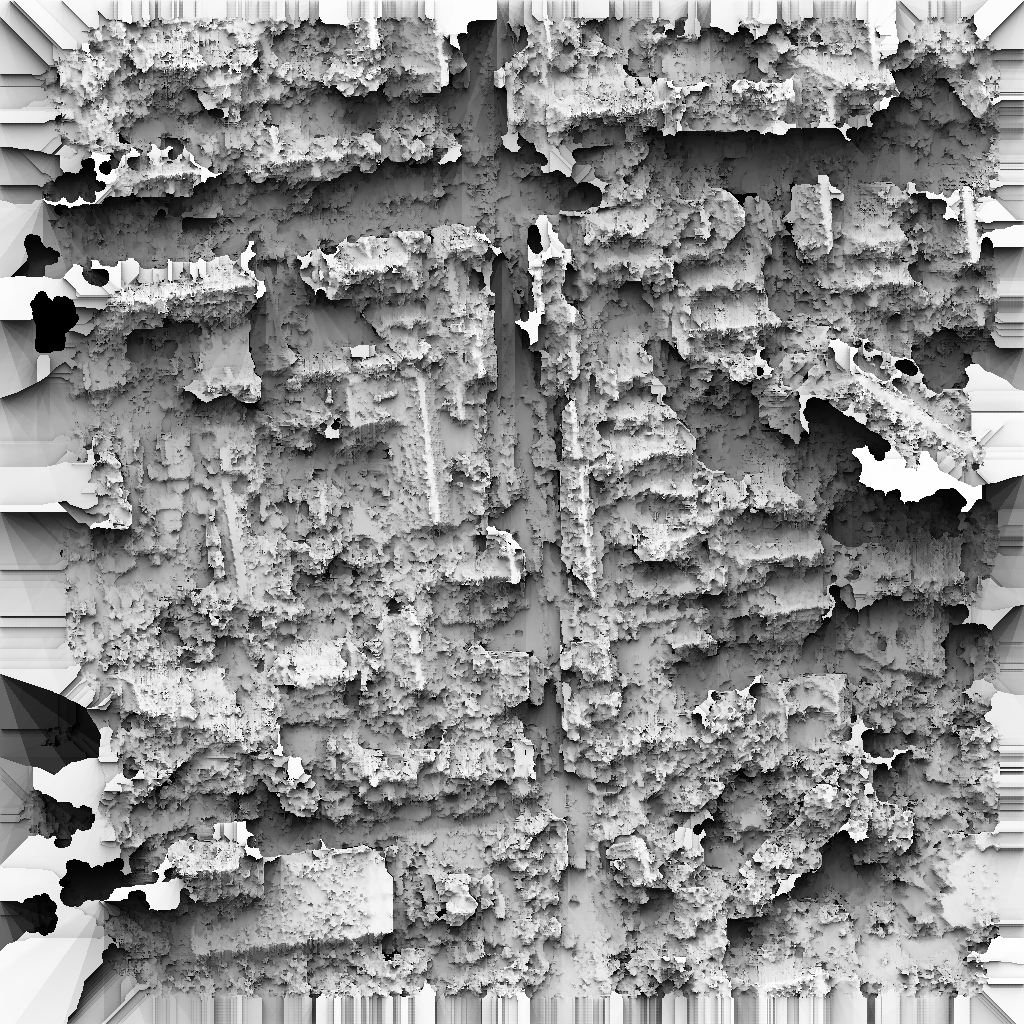}\\
		\centering{\tiny MICMAC}
	\end{minipage}
	\begin{minipage}[t]{0.19\textwidth}
		\includegraphics[width=0.098\linewidth]{figures_supp/color_map.png}
		\includegraphics[width=0.85\linewidth]{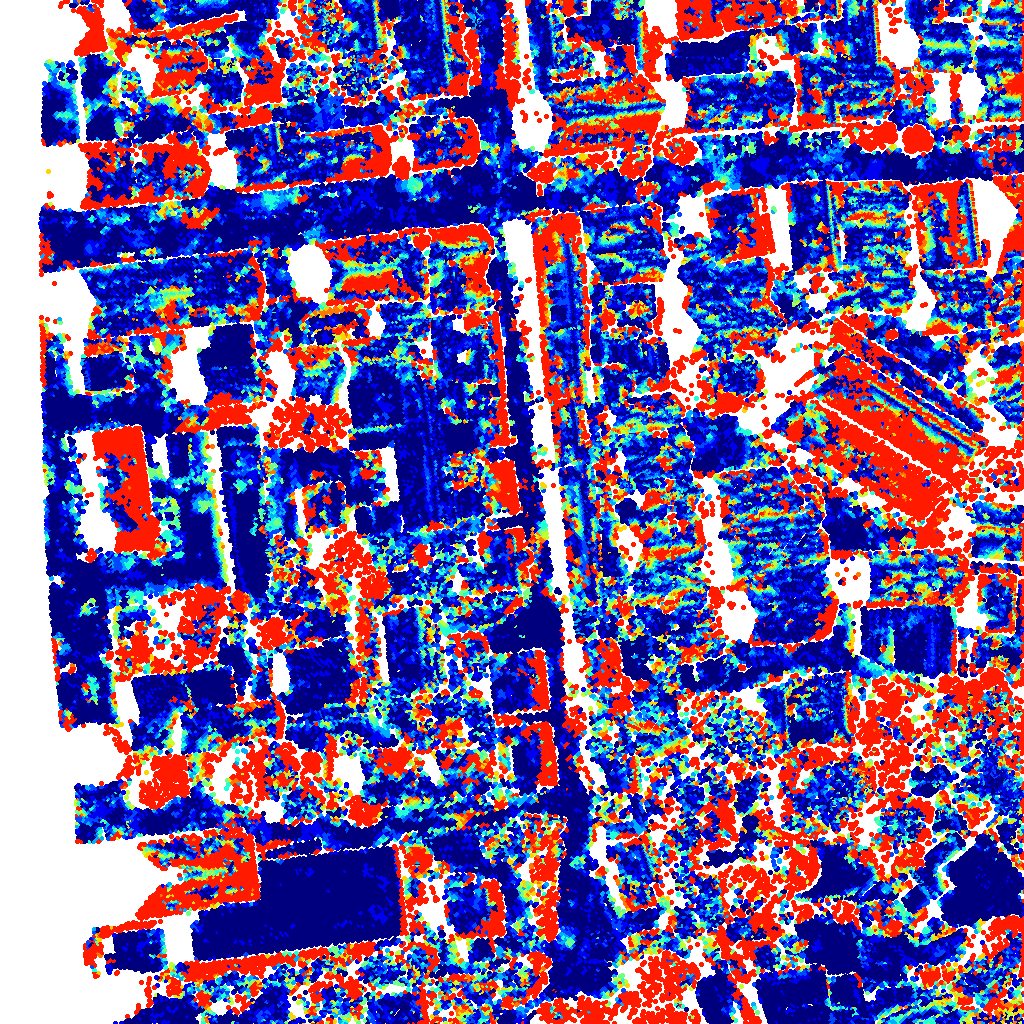} \\
		\includegraphics[width=\linewidth]{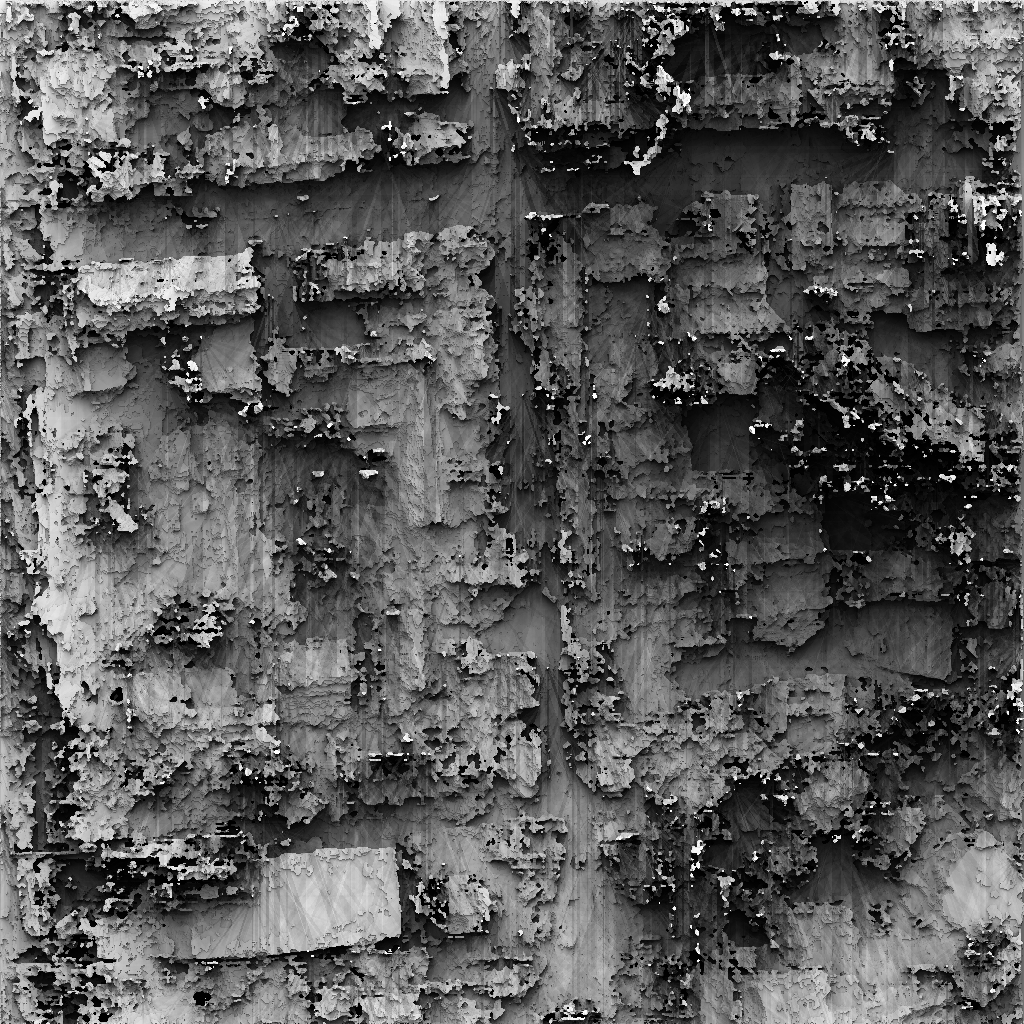} \\
		\centering{\tiny SGM(CUDA)}
	\end{minipage}
	\begin{minipage}[t]{0.19\textwidth}
		\includegraphics[width=0.098\linewidth]{figures_supp/color_map.png}
		\includegraphics[width=0.85\linewidth]{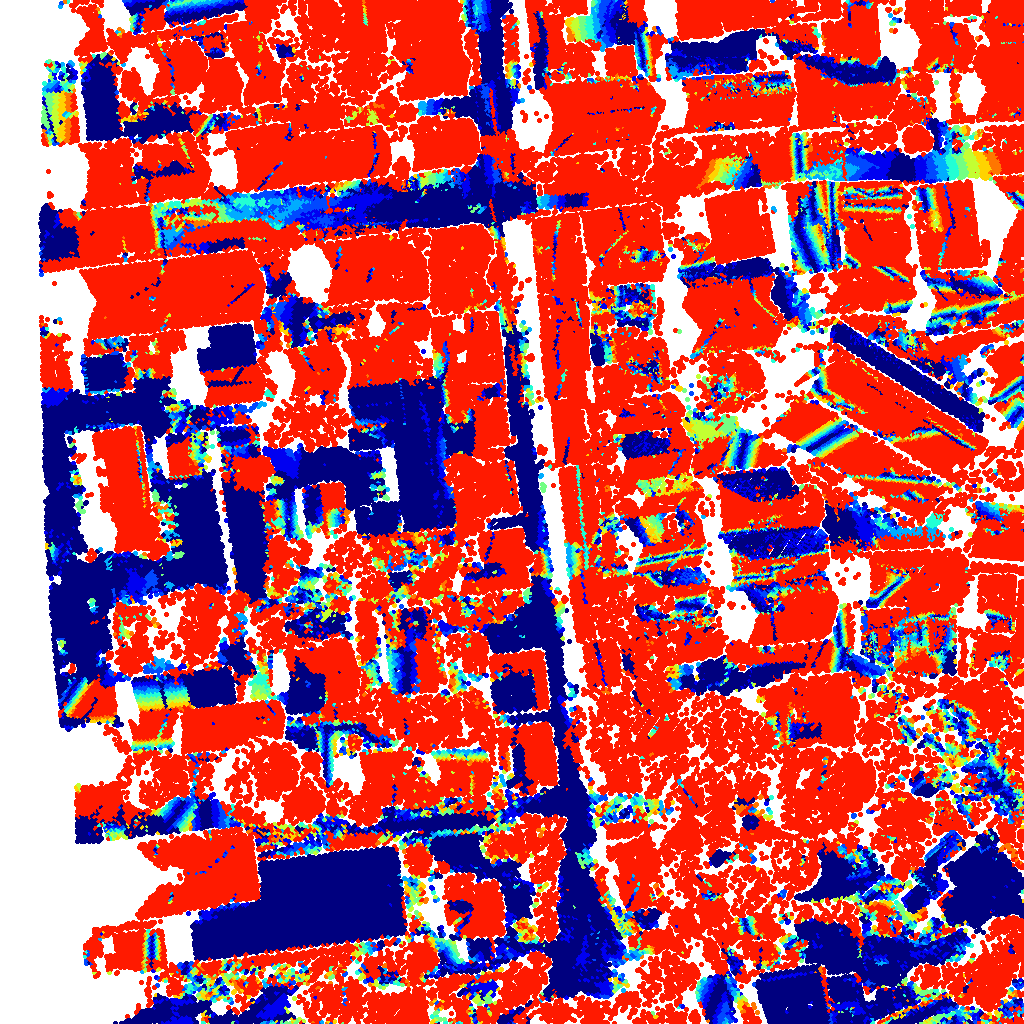} \\
		\includegraphics[width=\linewidth]{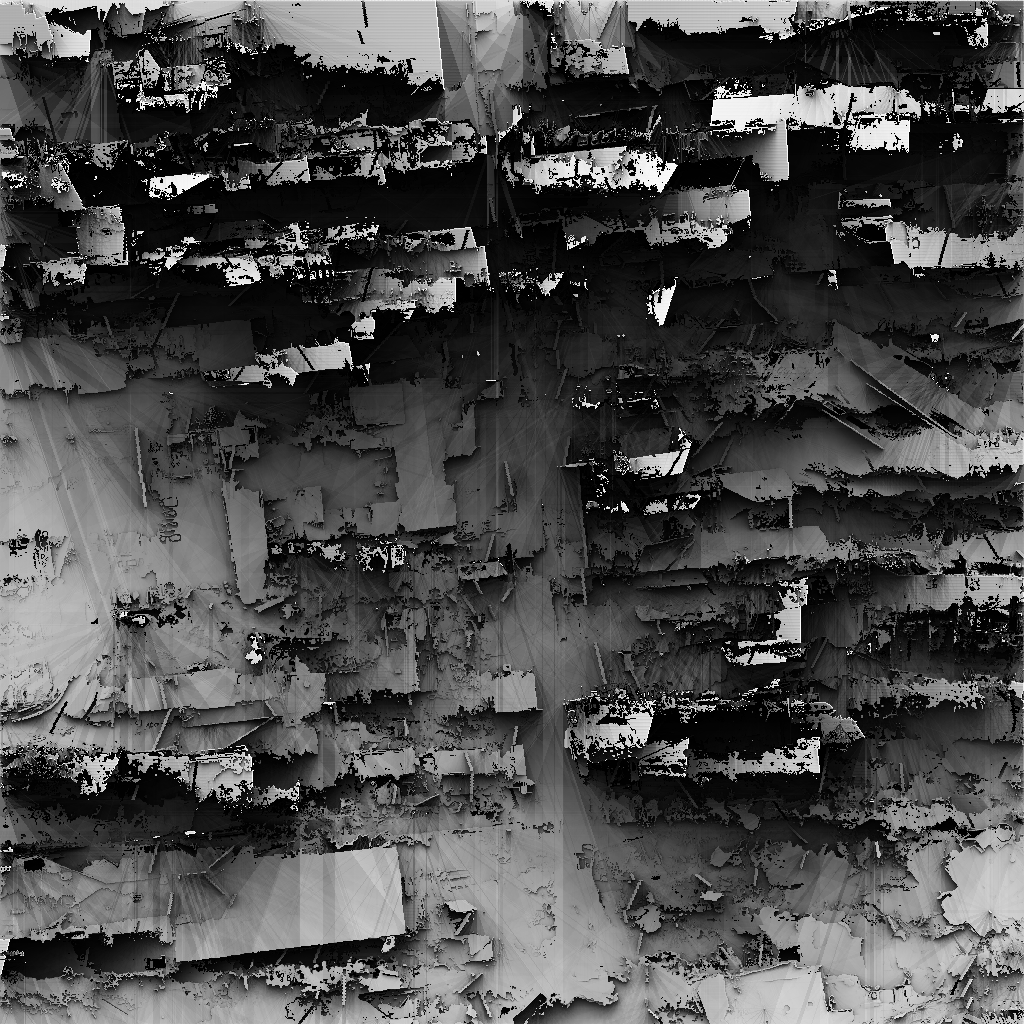}
		\centering{\tiny GraphCuts}
	\end{minipage}
	\begin{minipage}[t]{0.19\textwidth}
		\includegraphics[width=0.098\linewidth]{figures_supp/color_map.png}
		\includegraphics[width=0.85\linewidth]{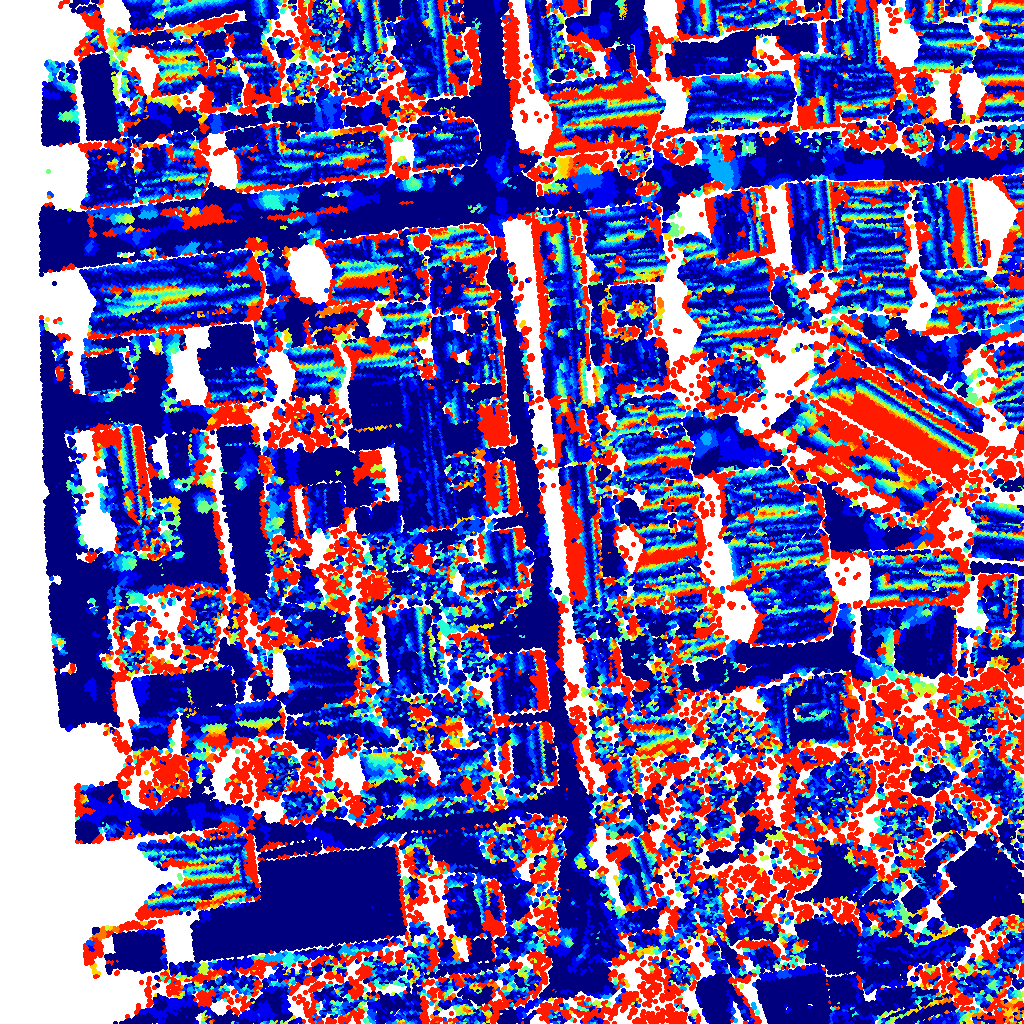} \\
		\includegraphics[width=\linewidth]{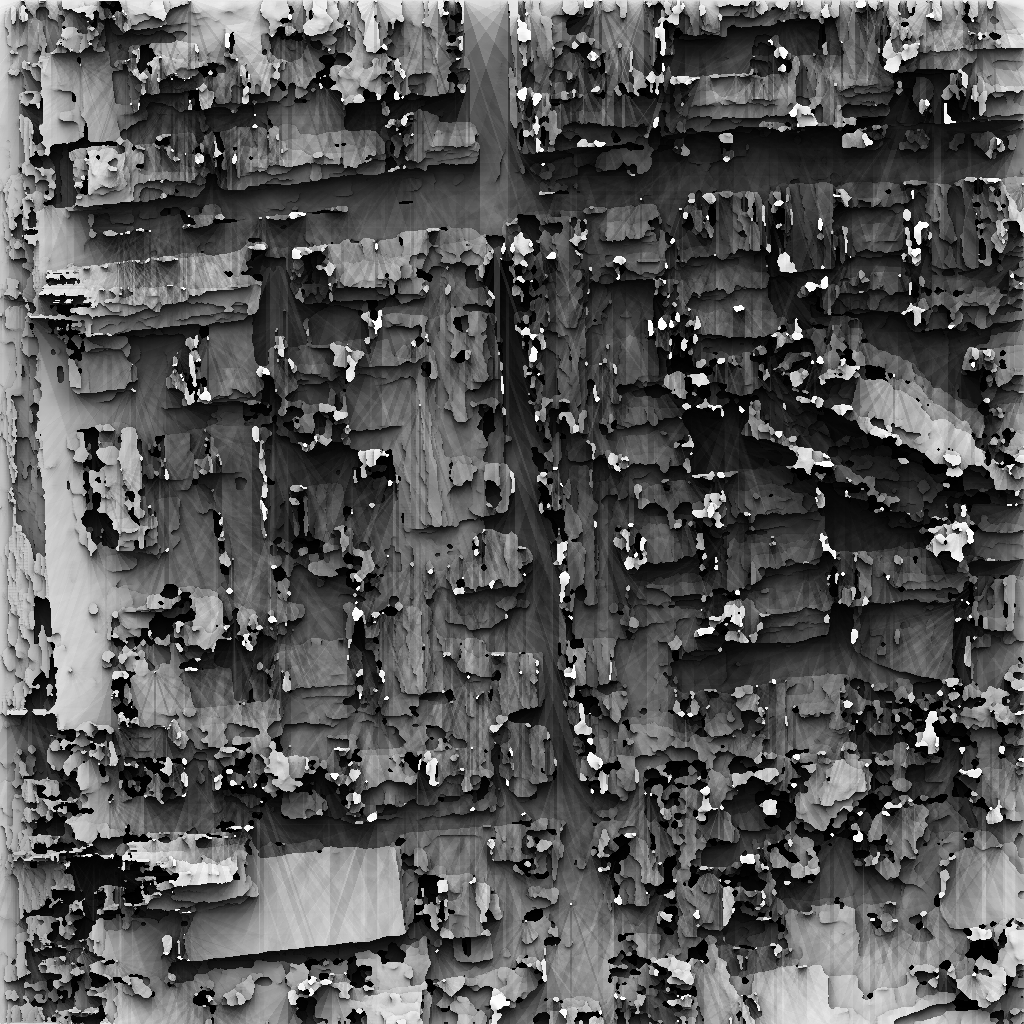}
		\centering{\tiny CBMV(SGM)}
	\end{minipage}
	\begin{minipage}[t]{0.19\textwidth}
		\includegraphics[width=0.098\linewidth]{figures_supp/color_map.png}
		\includegraphics[width=0.85\linewidth]{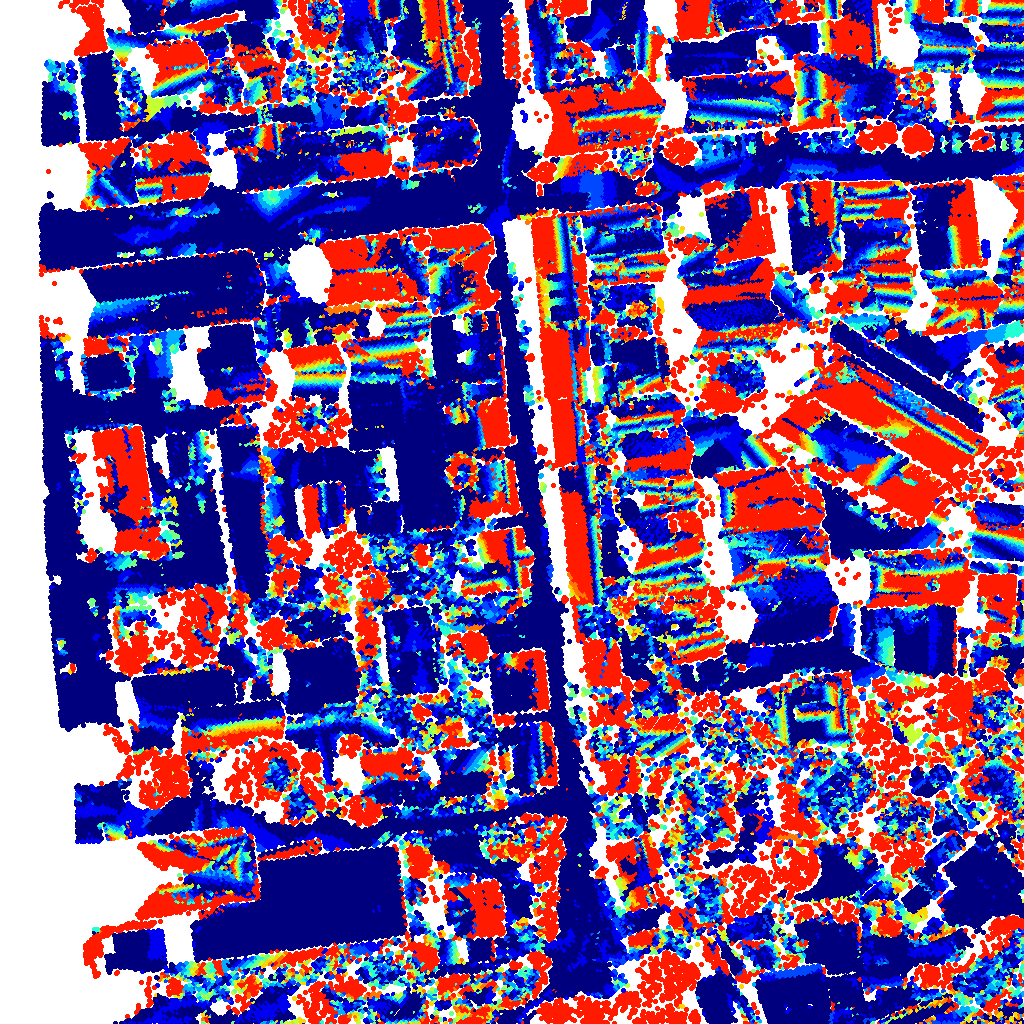} \\
		\includegraphics[width=\linewidth]{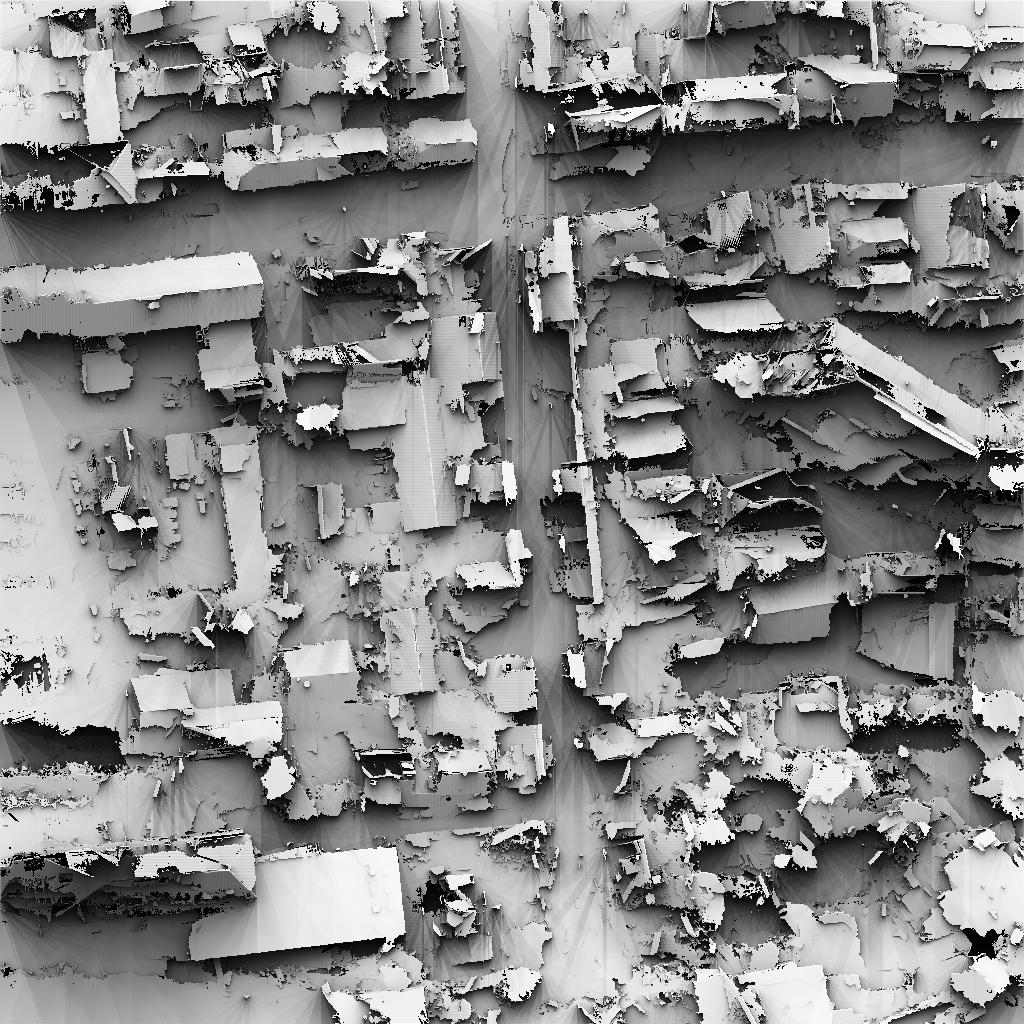}
		\centering{\tiny CBMV(GraphCuts)}
	\end{minipage}

	\begin{minipage}[t]{0.19\textwidth}
		\includegraphics[width=0.098\linewidth]{figures_supp/color_map.png}
		\includegraphics[width=0.85\linewidth]{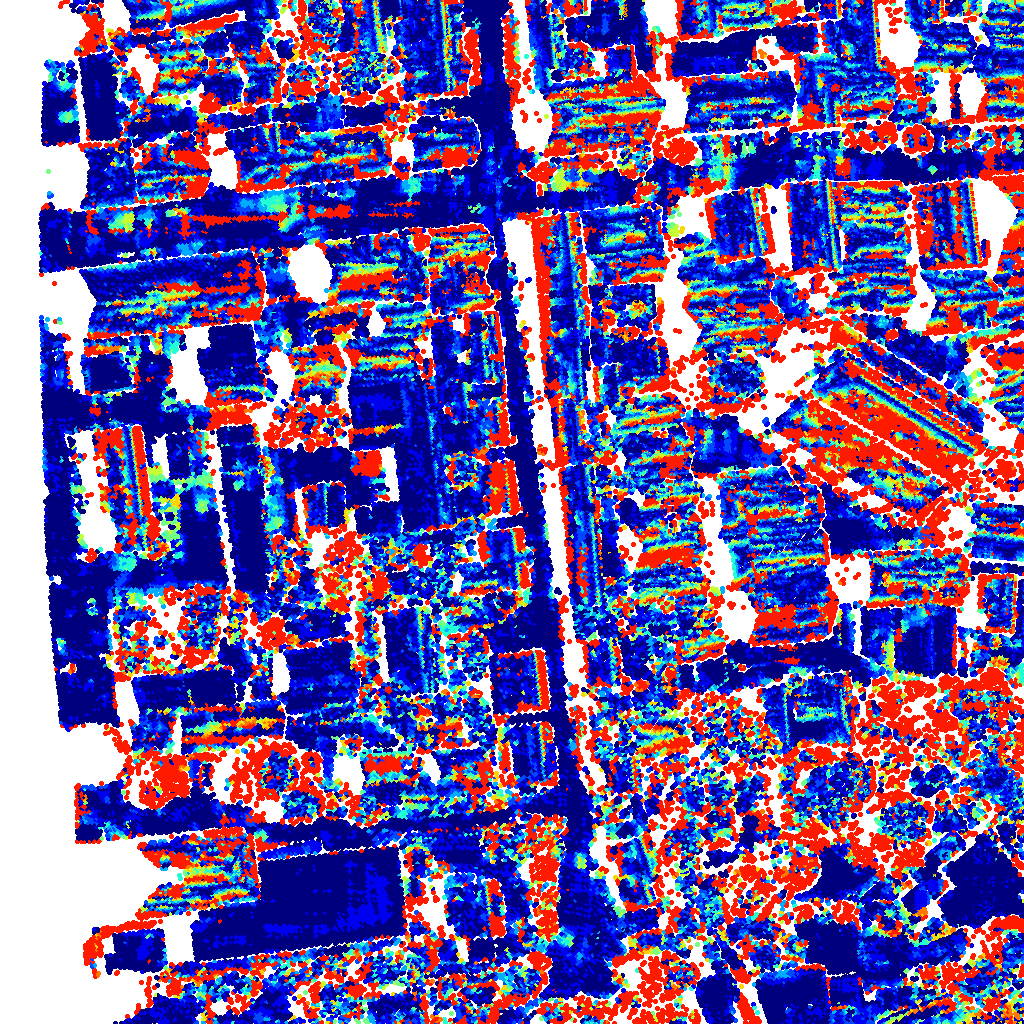} \\
		\includegraphics[width=\linewidth]{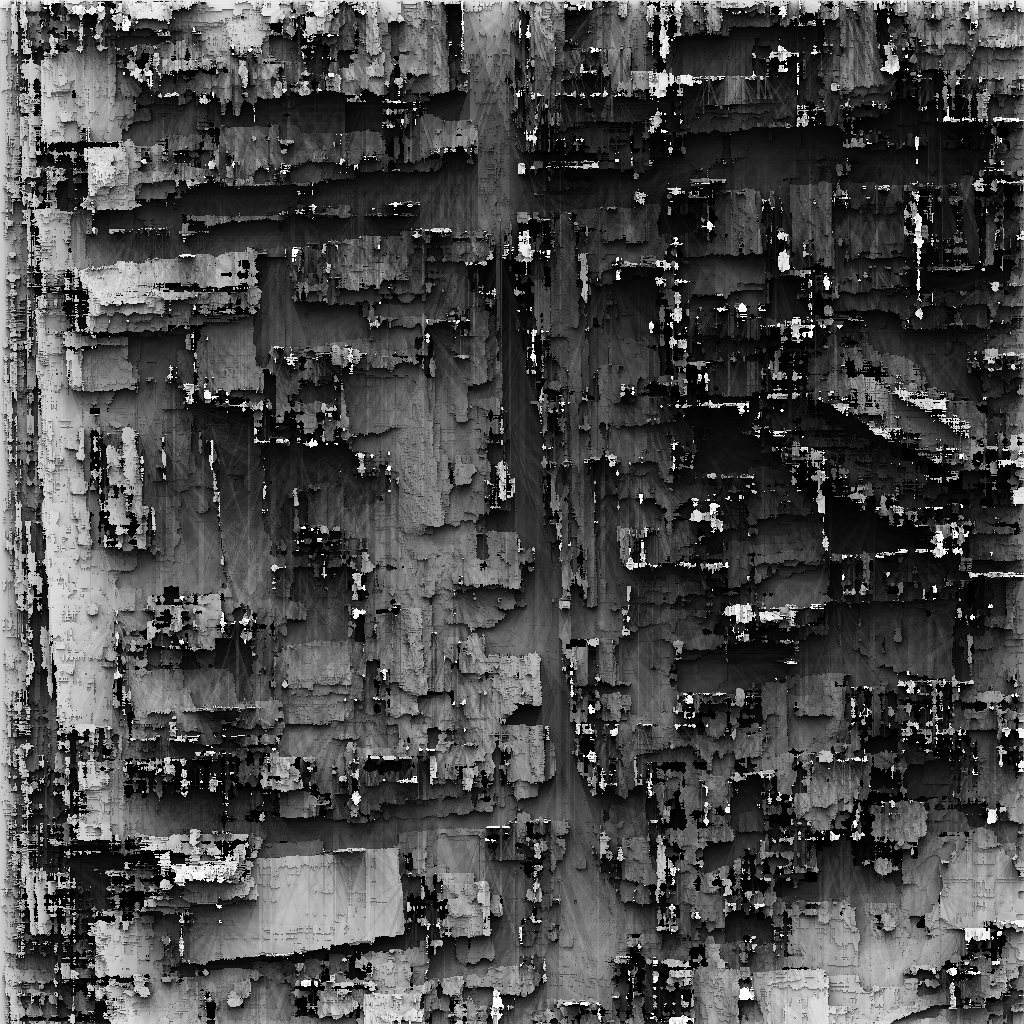}
		\centering{\tiny MC-CNN(KITTI)}
	\end{minipage}
	\begin{minipage}[t]{0.19\textwidth}
		\includegraphics[width=0.098\linewidth]{figures_supp/color_map.png}
		\includegraphics[width=0.85\linewidth]{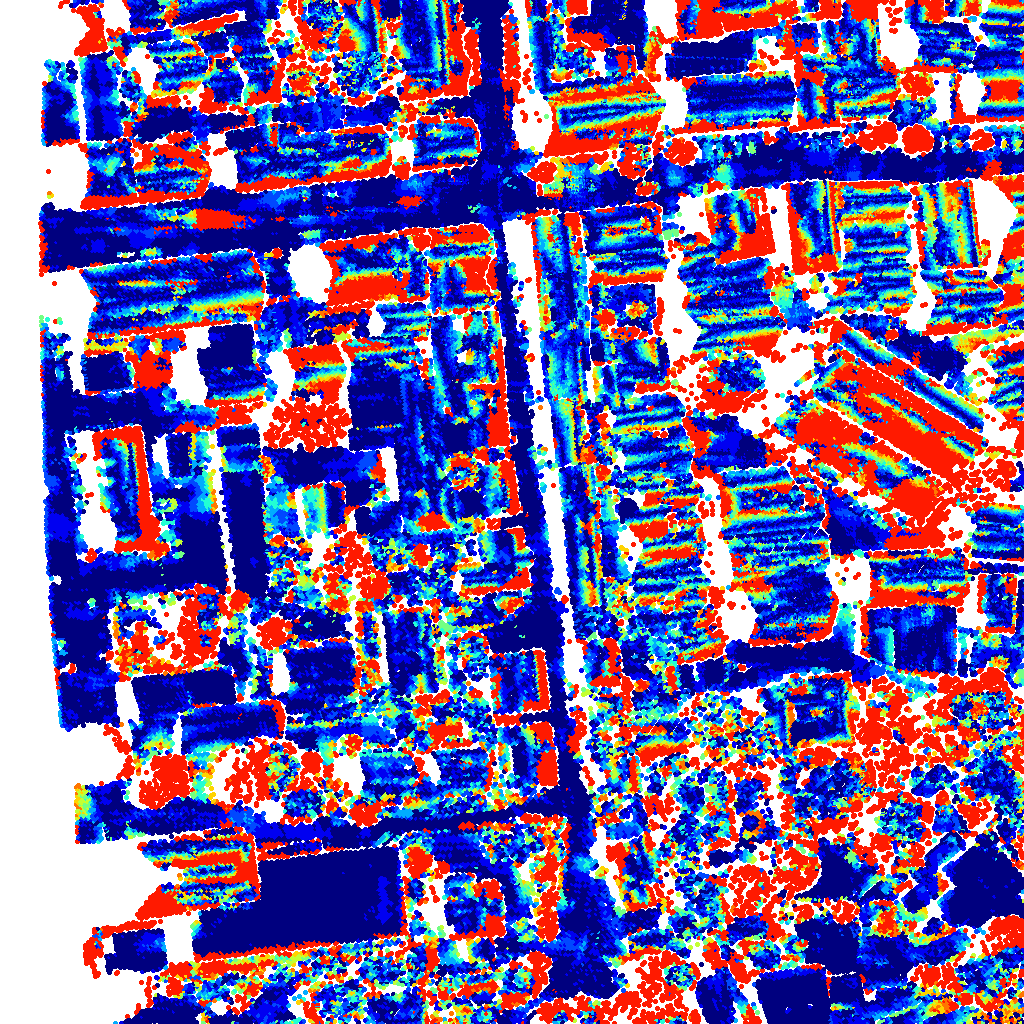}
		\includegraphics[width=\linewidth]{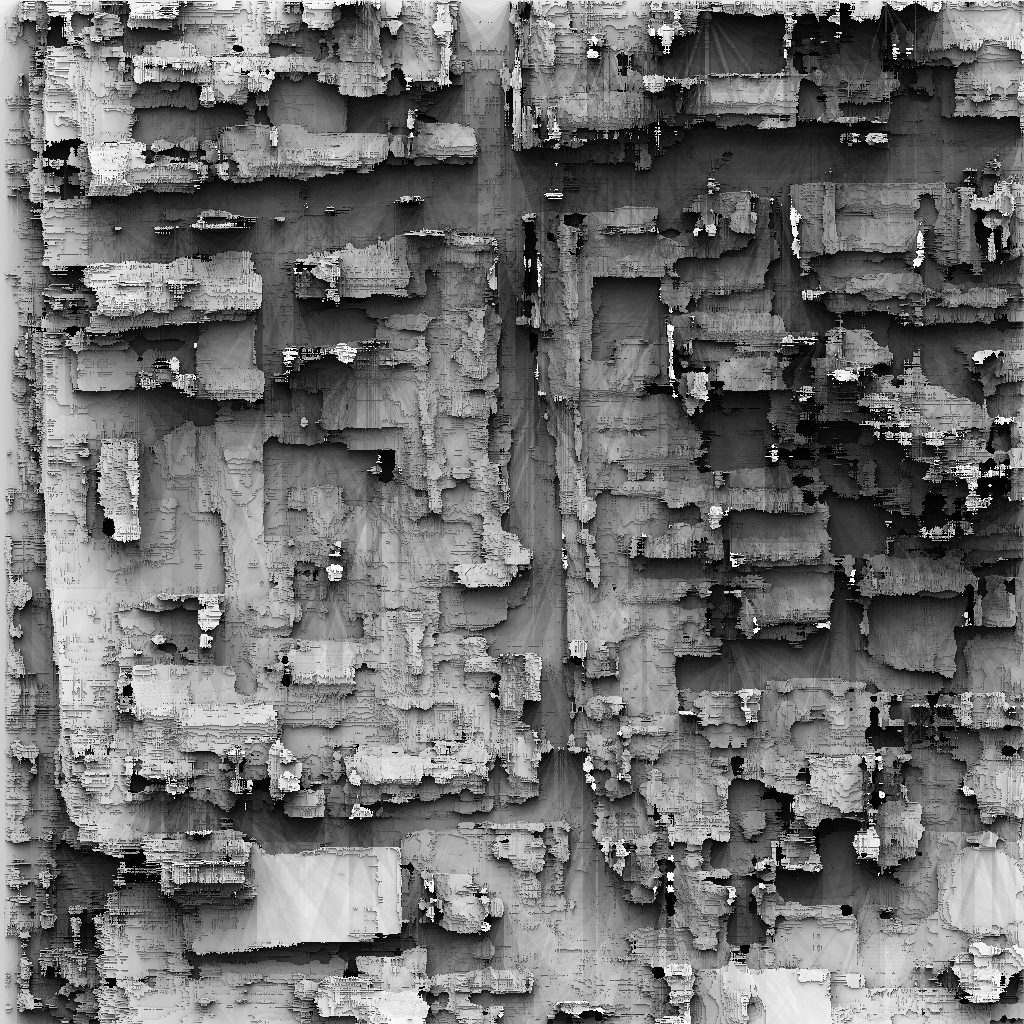}
		\centering{\tiny DeepFeature(KITTI)}
	\end{minipage}
	\begin{minipage}[t]{0.19\textwidth}
		\includegraphics[width=0.098\linewidth]{figures_supp/color_map.png}
		\includegraphics[width=0.85\linewidth]{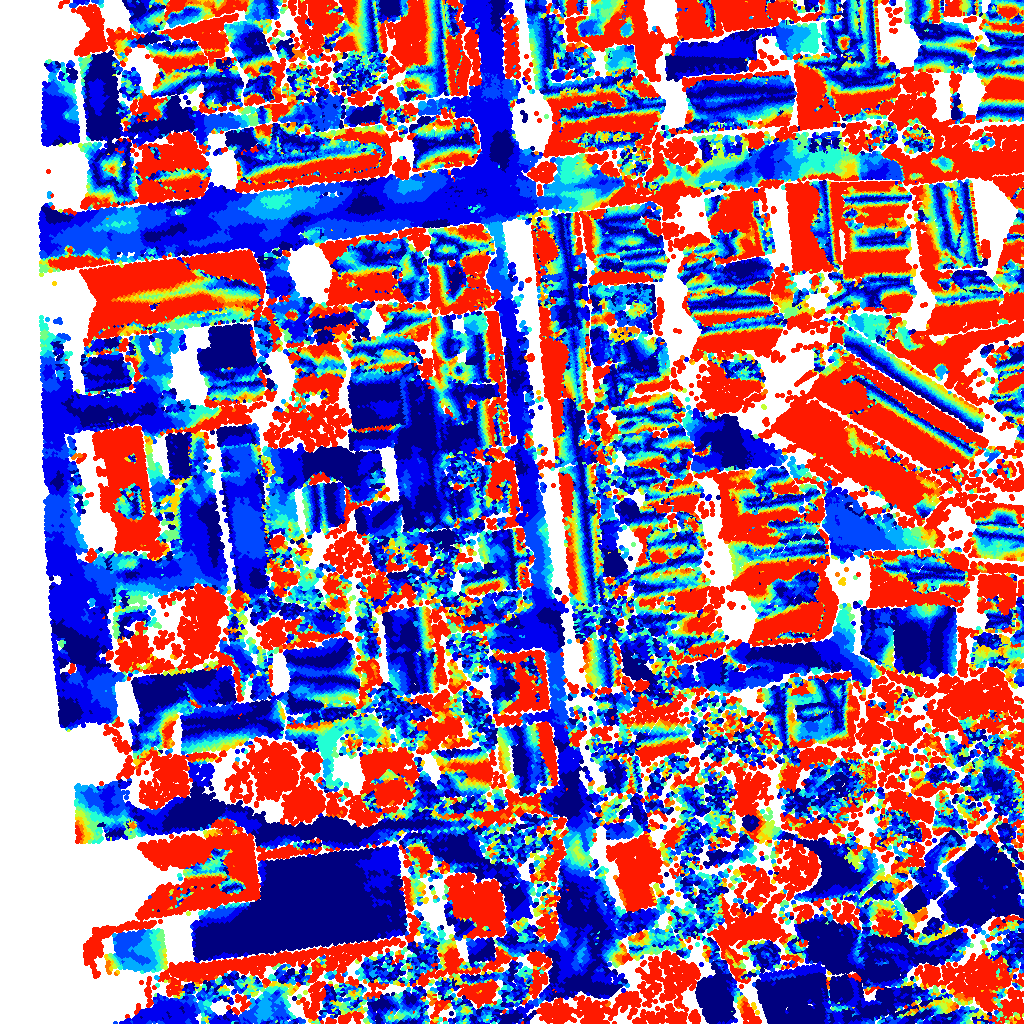}
		\includegraphics[width=\linewidth]{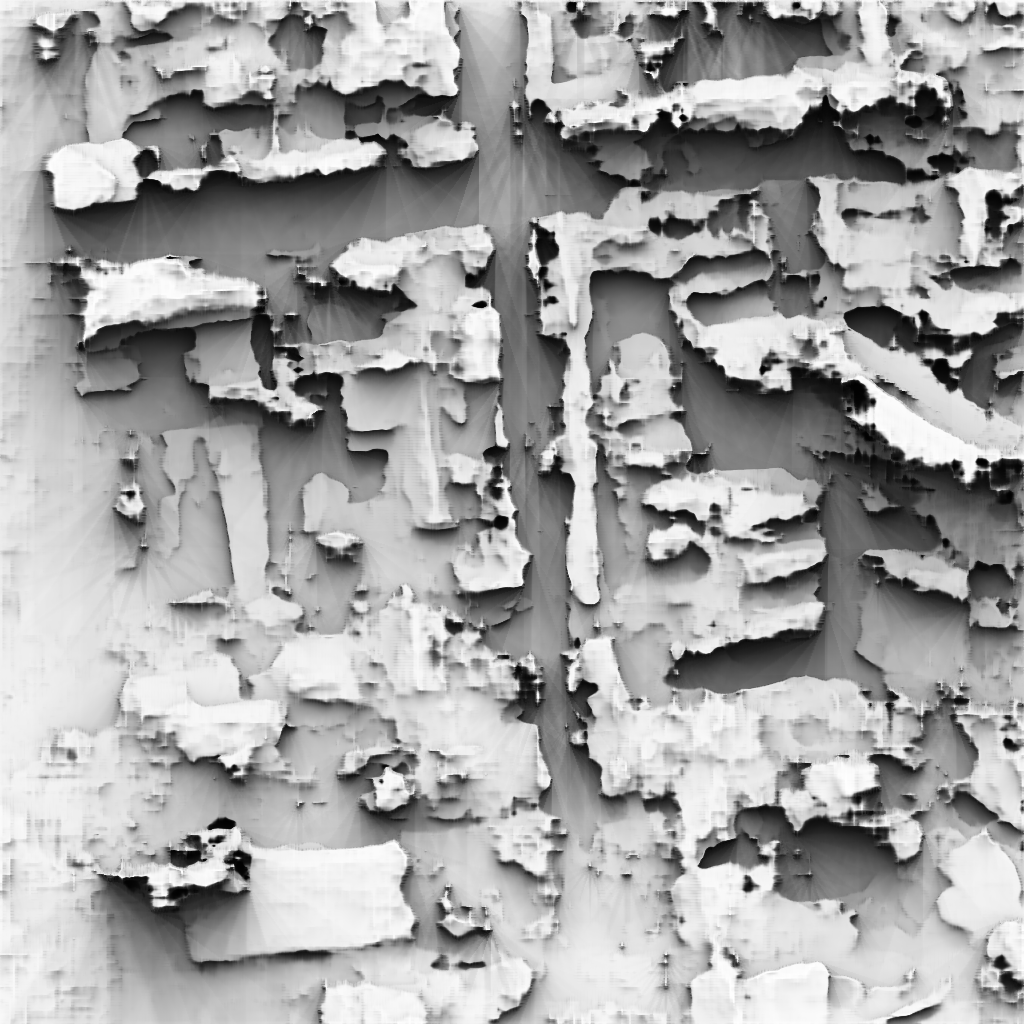}
		\centering{\tiny PSM Net(KITTI)}
	\end{minipage}
	\begin{minipage}[t]{0.19\textwidth}	
		\includegraphics[width=0.098\linewidth]{figures_supp/color_map.png}
		\includegraphics[width=0.85\linewidth]{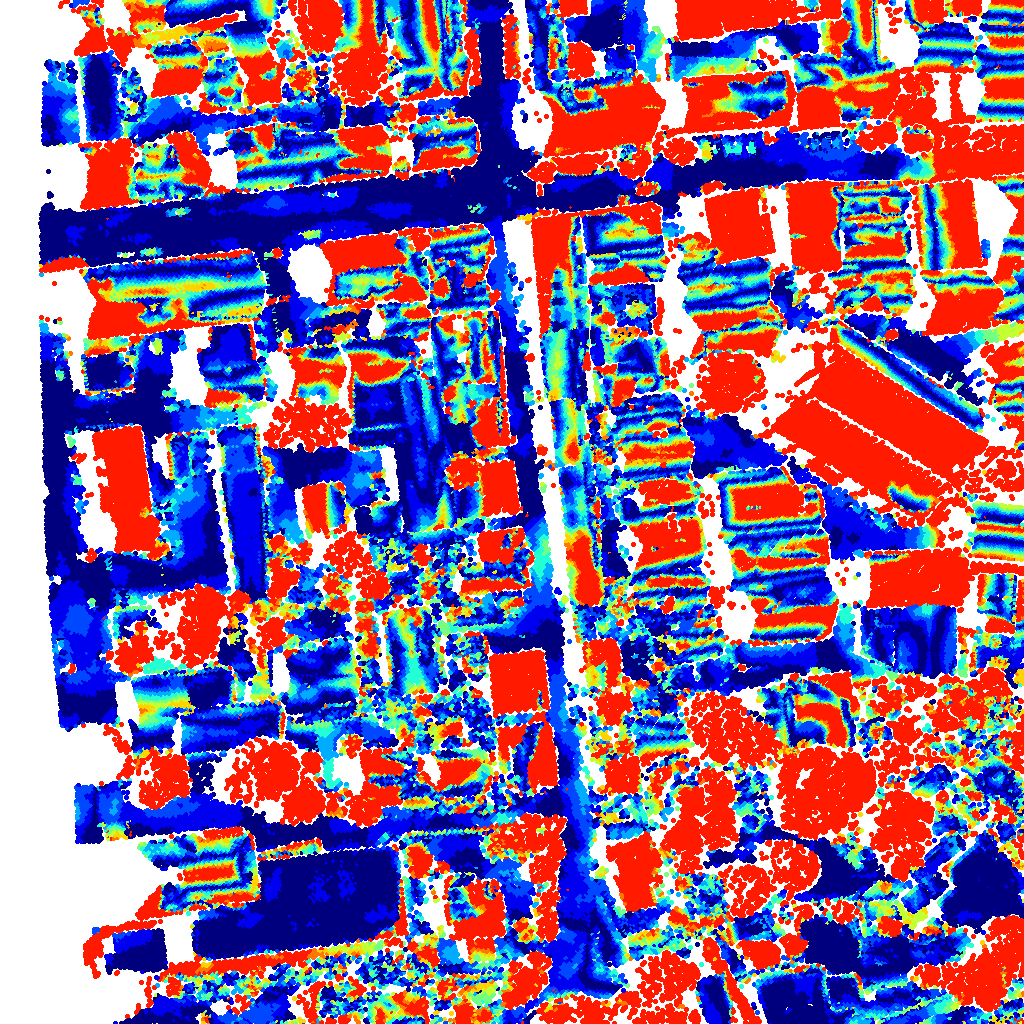} \\
		\includegraphics[width=\linewidth]{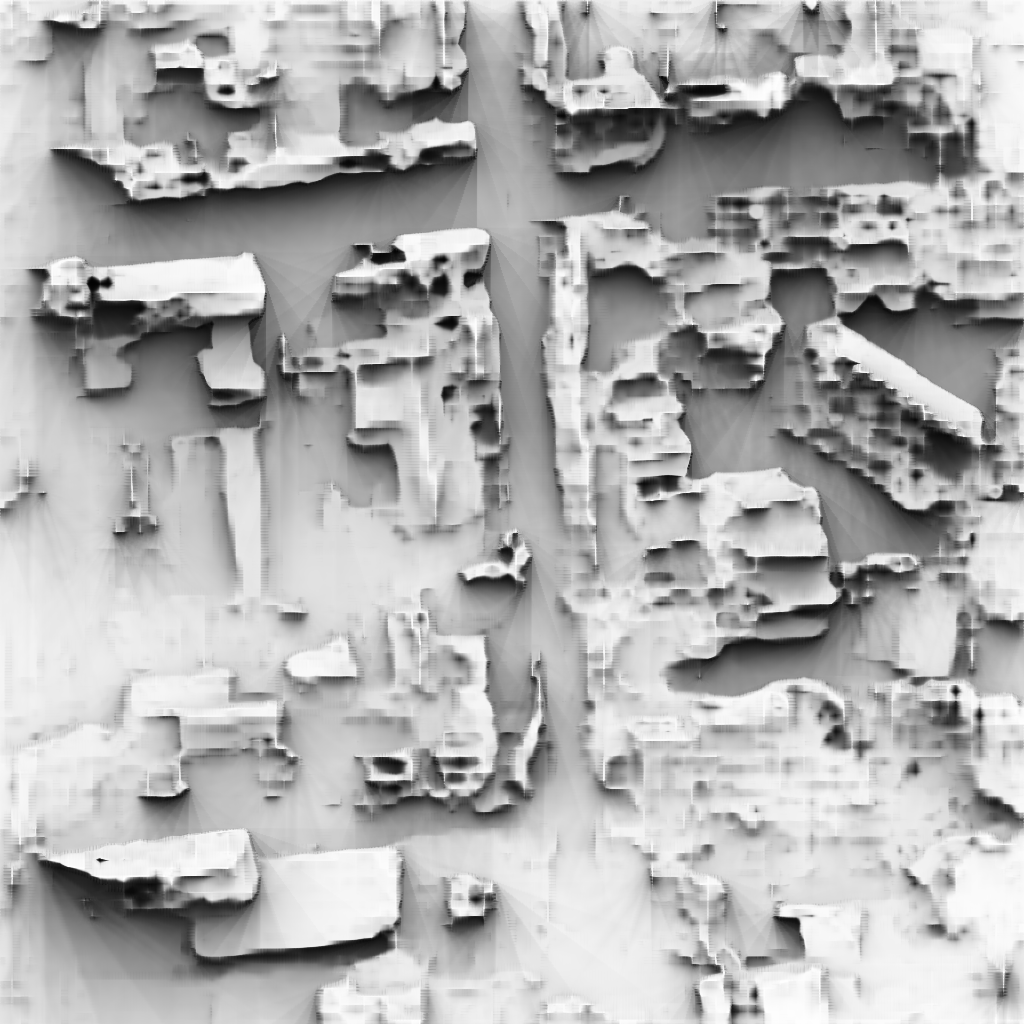}
		\centering{\tiny HRS Net(KITTI)}
	\end{minipage}	
	\begin{minipage}[t]{0.19\textwidth}	
		\includegraphics[width=0.098\linewidth]{figures_supp/color_map.png}
		\includegraphics[width=0.85\linewidth]{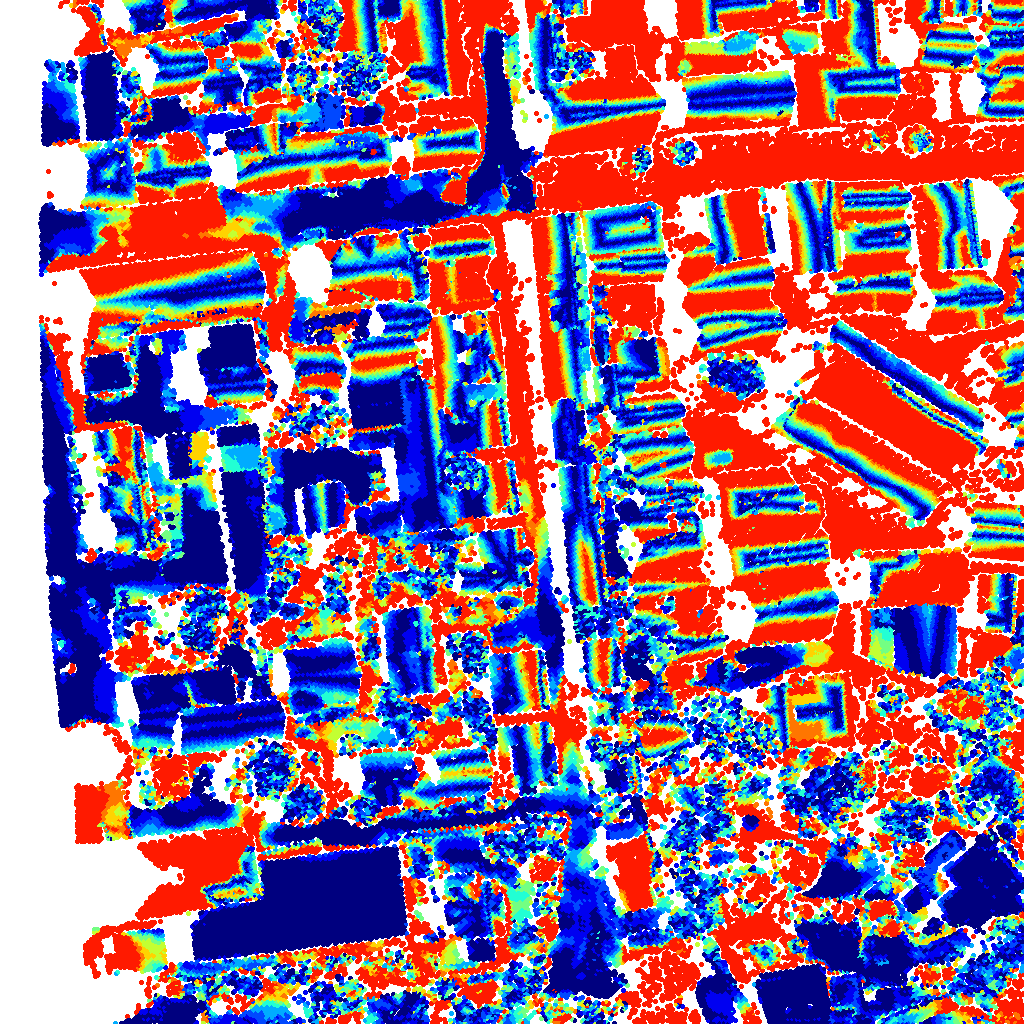}
		\includegraphics[width=\linewidth]{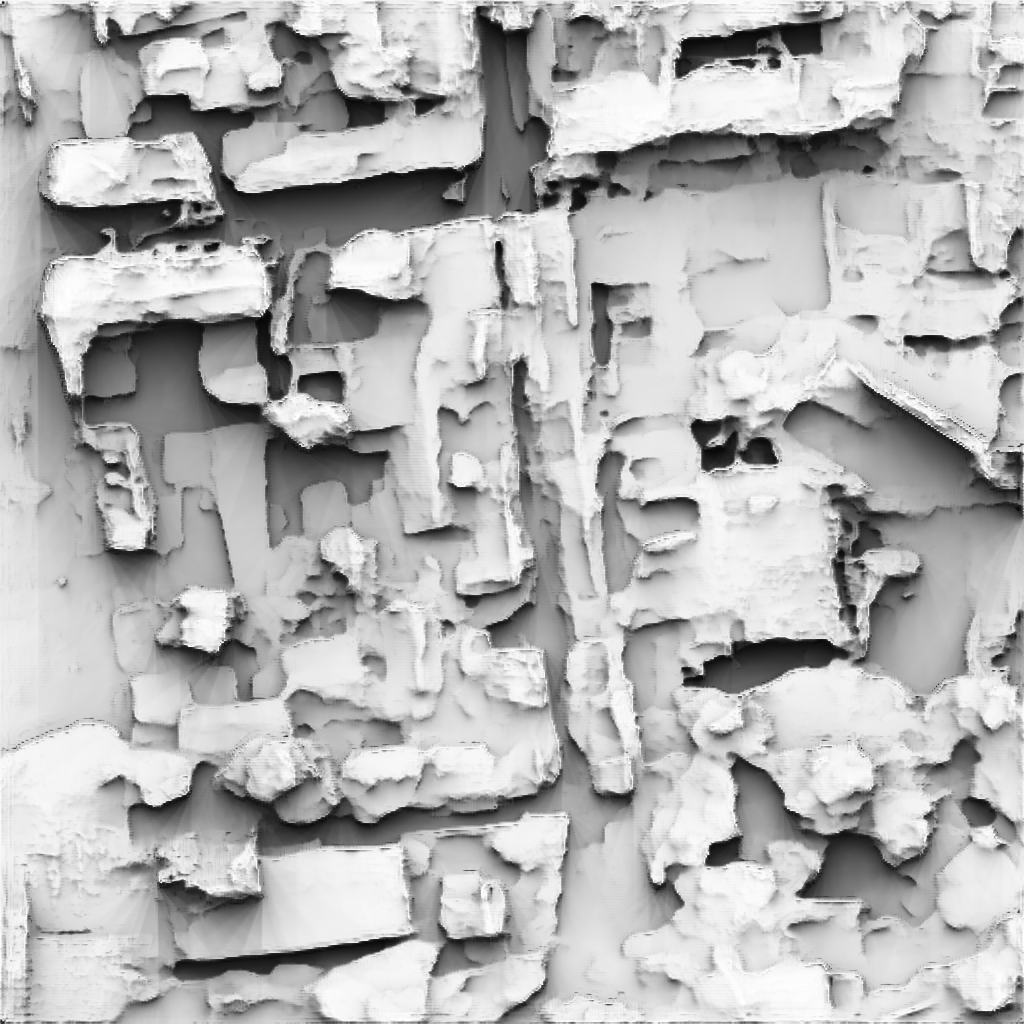}
		\centering{\tiny DeepPruner(KITTI)}
	\end{minipage}	
	\begin{minipage}[t]{0.19\textwidth}	
		\includegraphics[width=0.098\linewidth]{figures_supp/color_map.png}
		\includegraphics[width=0.85\linewidth]{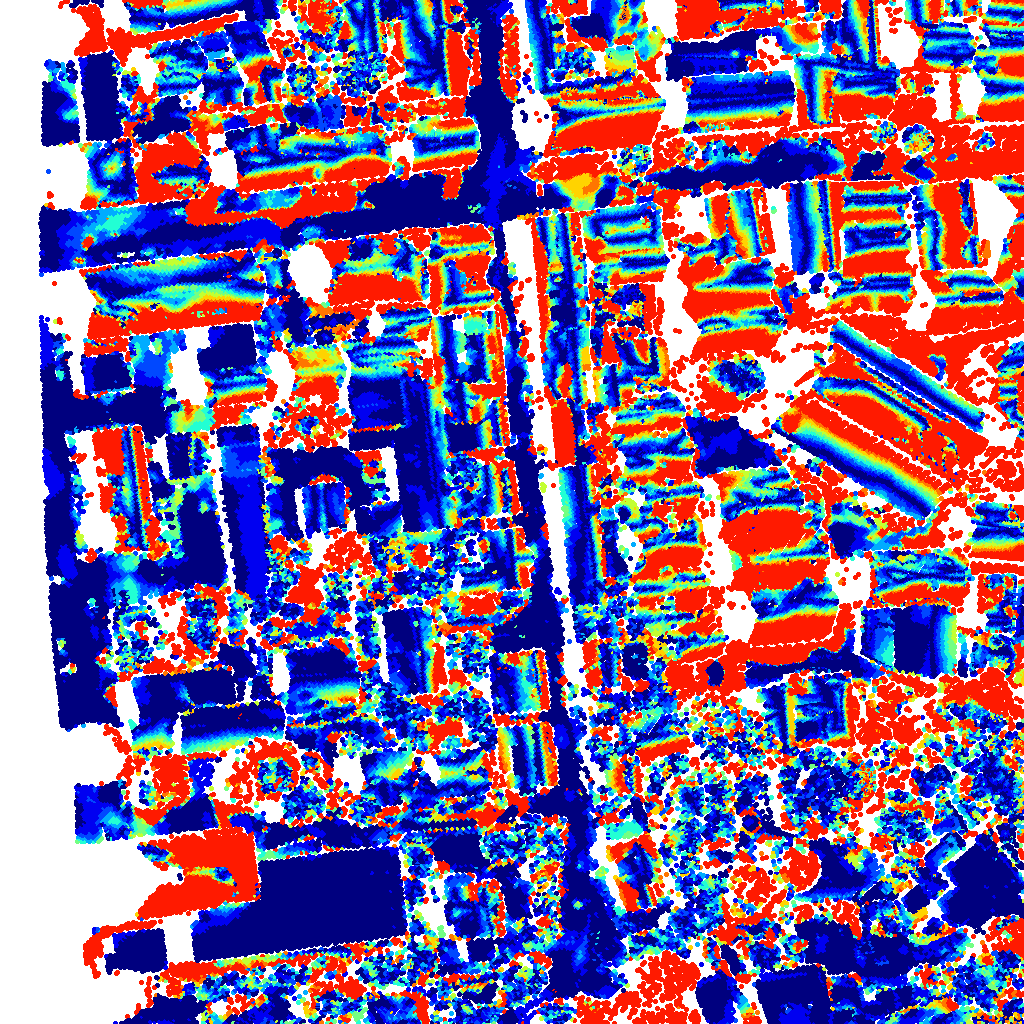}
		\includegraphics[width=\linewidth]{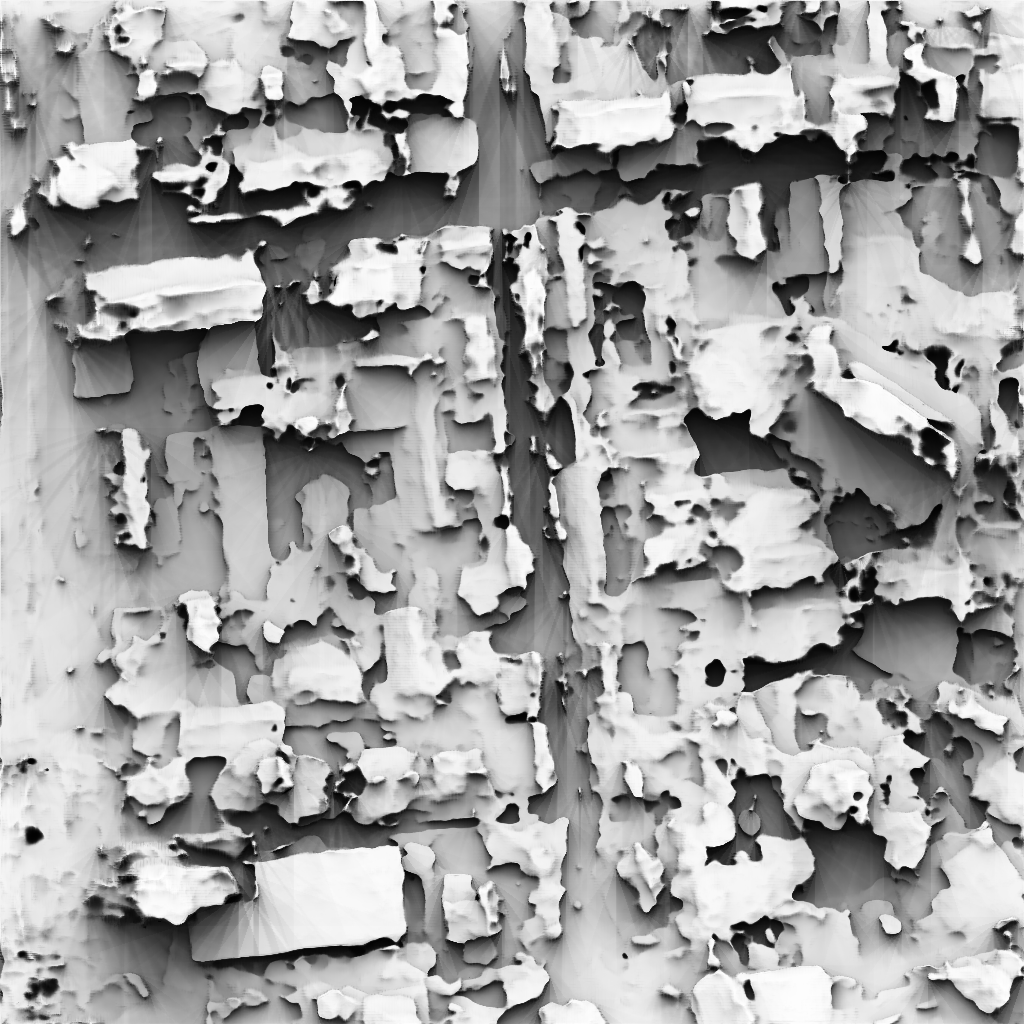}
		\centering{\tiny GANet(KITTI)}
	\end{minipage}	
	\begin{minipage}[t]{0.19\textwidth}	
		\includegraphics[width=0.098\linewidth]{figures_supp/color_map.png}
		\includegraphics[width=0.85\linewidth]{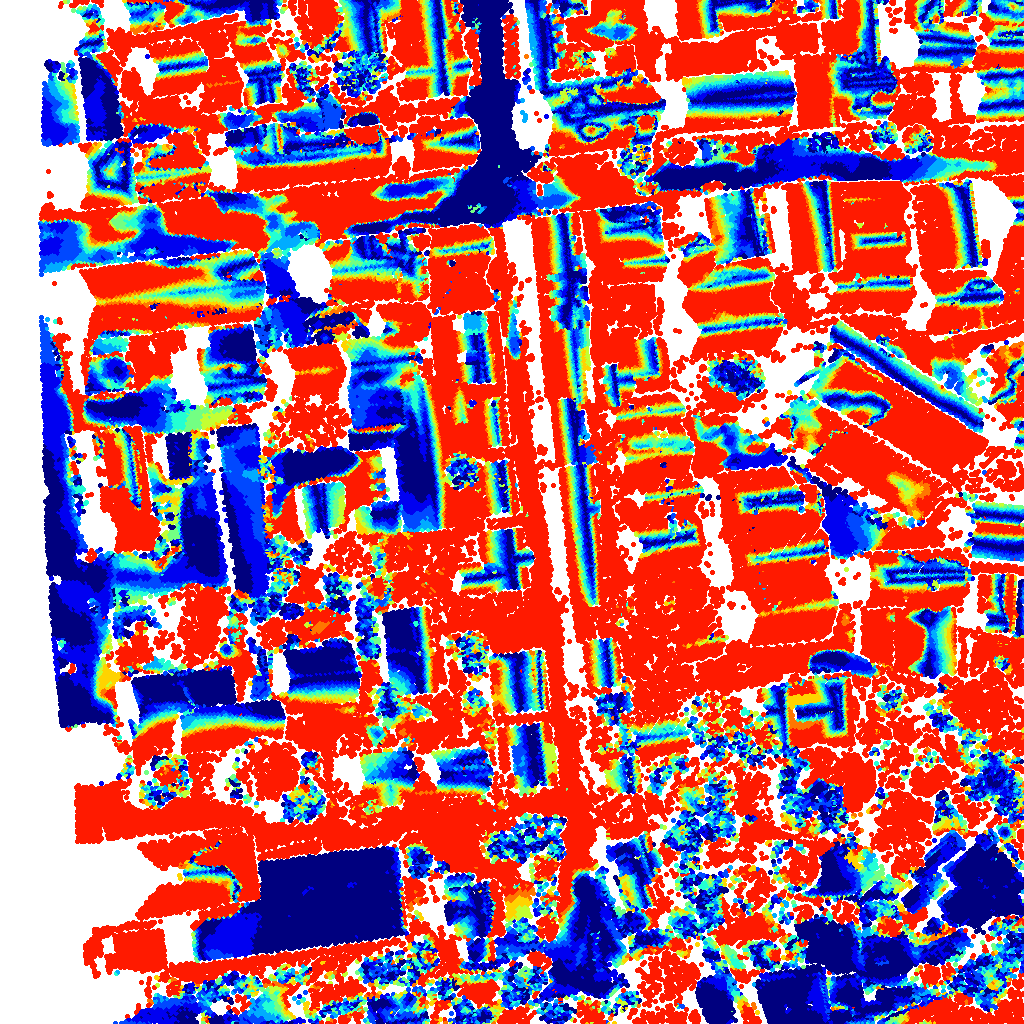}
		\includegraphics[width=\linewidth]{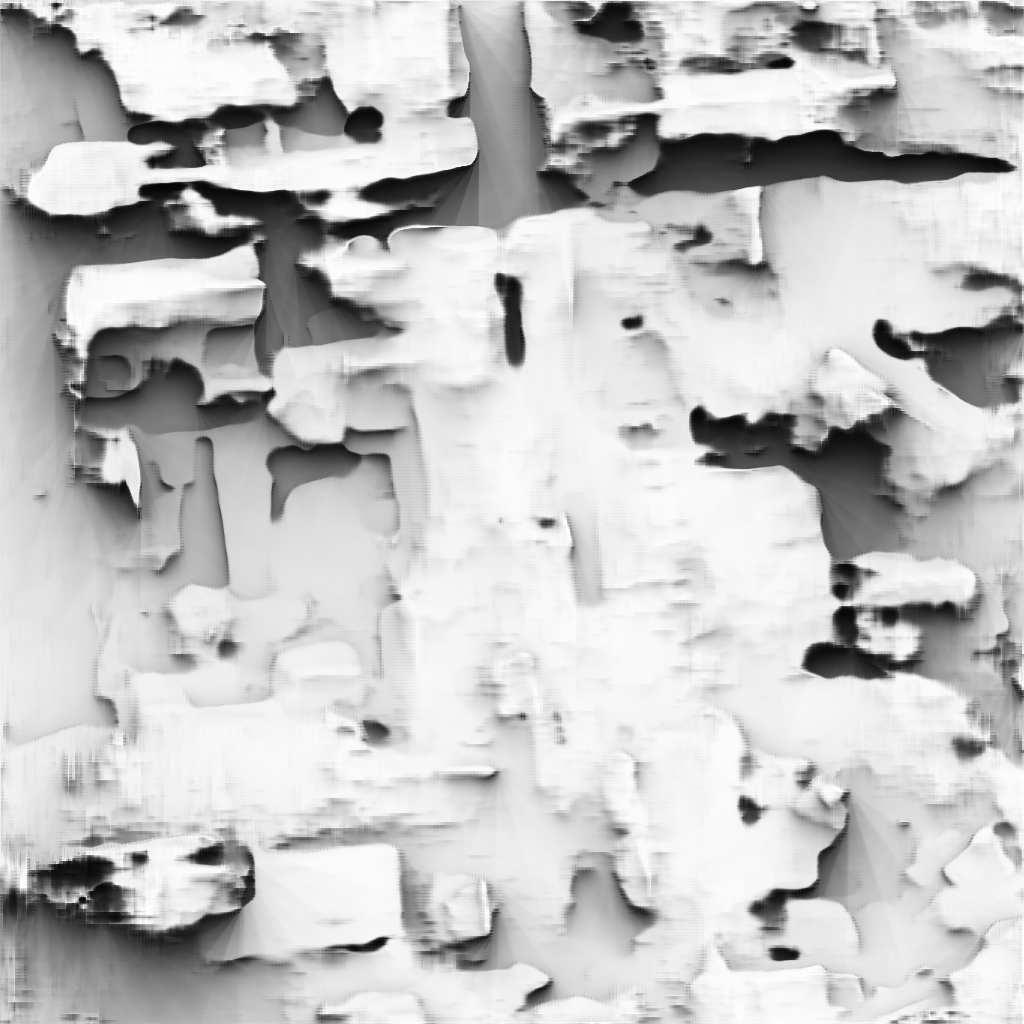}
		\centering{\tiny LEAStereo(KITTI)}
	\end{minipage}
	\begin{minipage}[t]{0.19\textwidth}
		\includegraphics[width=0.098\linewidth]{figures_supp/color_map.png}
		\includegraphics[width=0.85\linewidth]{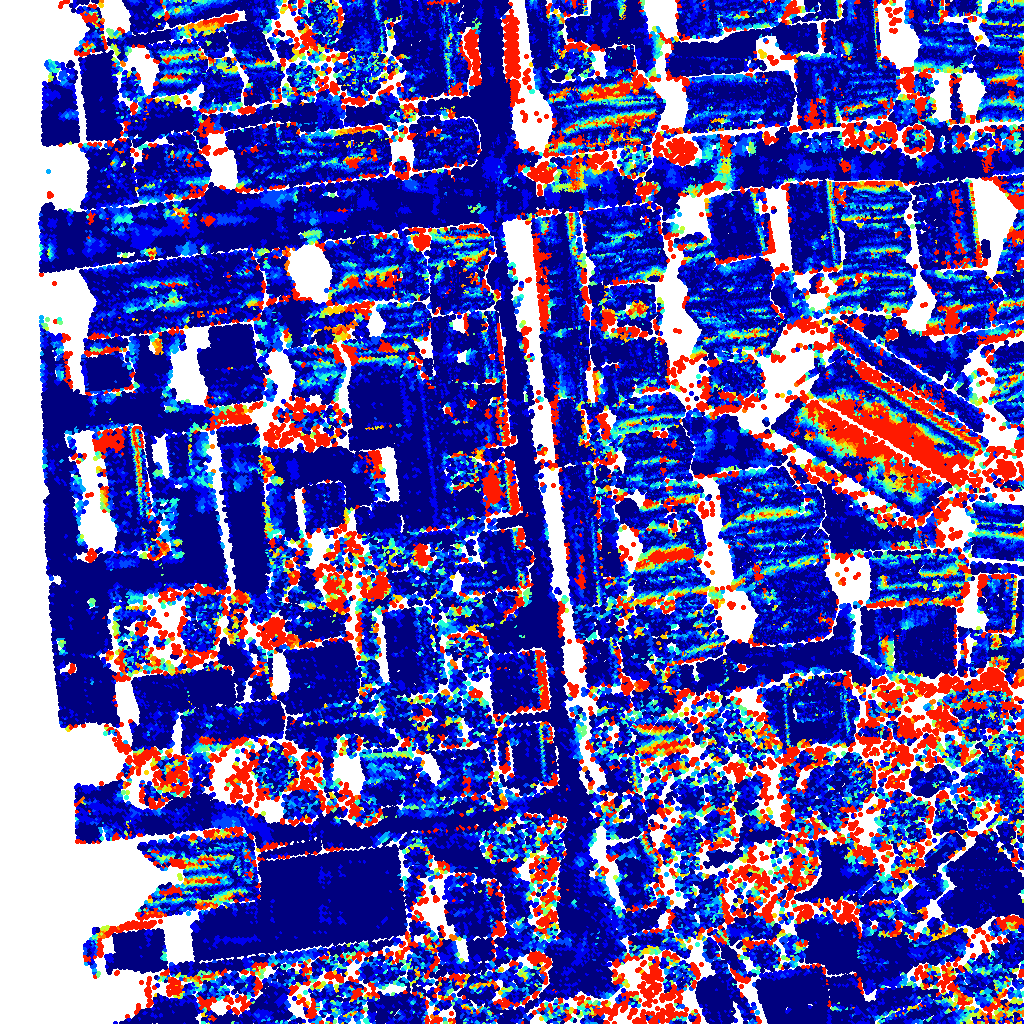} \\
		\includegraphics[width=\linewidth]{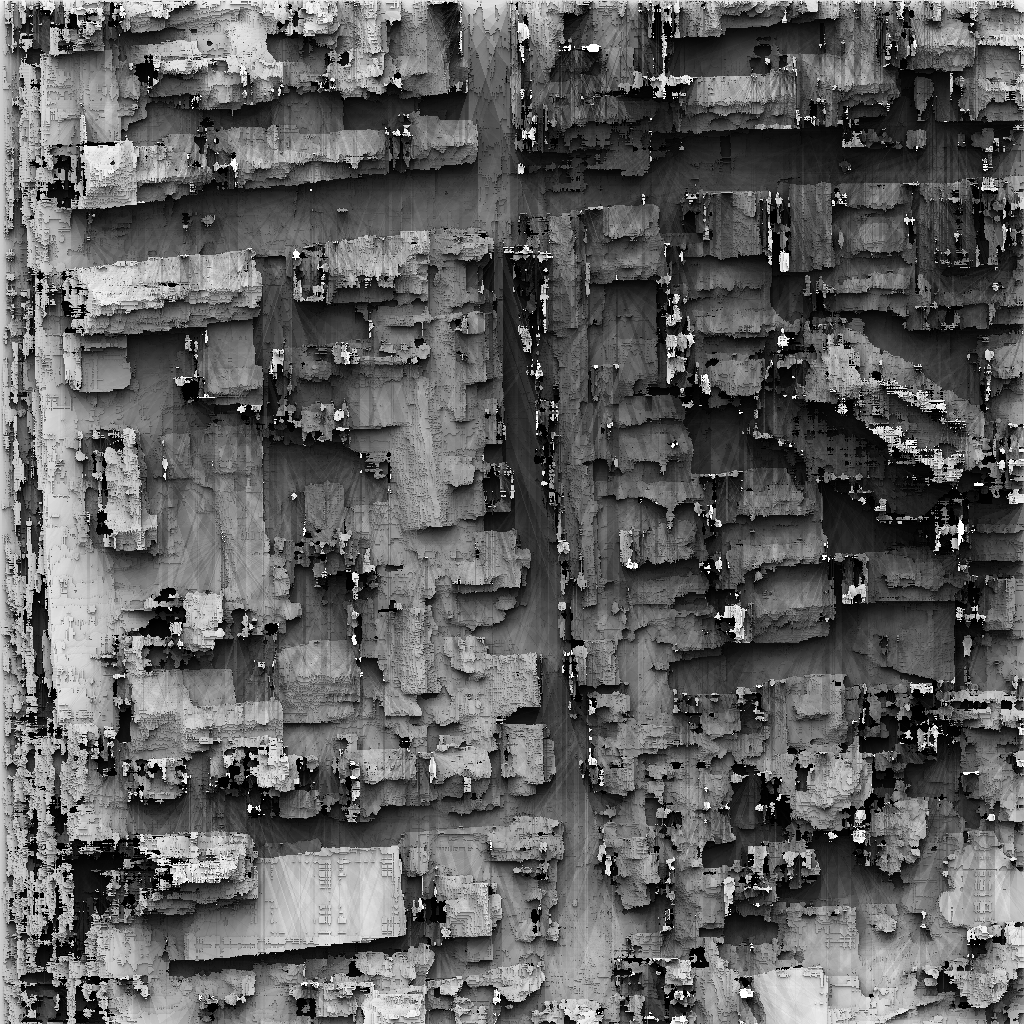}
		\centering{\tiny MC-CNN}
	\end{minipage}
		\begin{minipage}[t]{0.19\textwidth}
		\includegraphics[width=0.098\linewidth]{figures_supp/color_map.png}
		\includegraphics[width=0.85\linewidth]{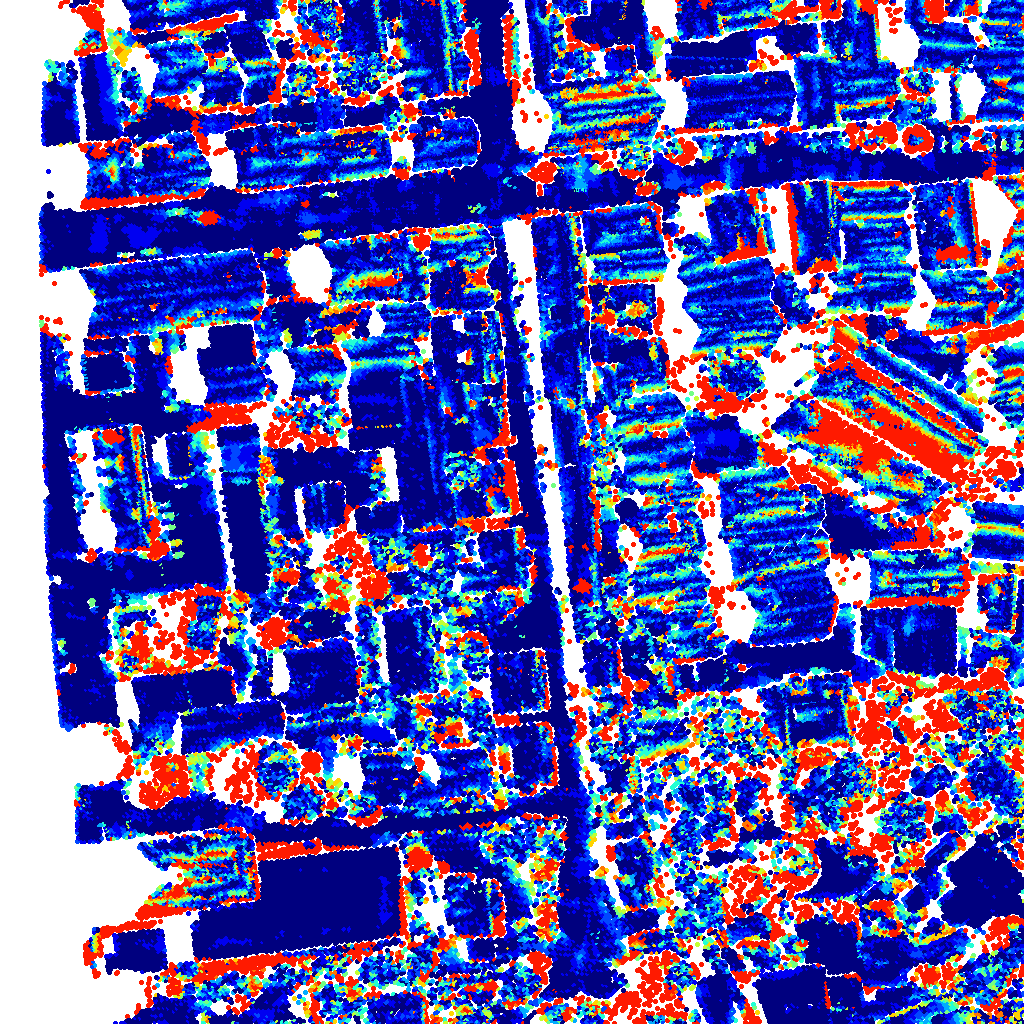}
		\includegraphics[width=\linewidth]{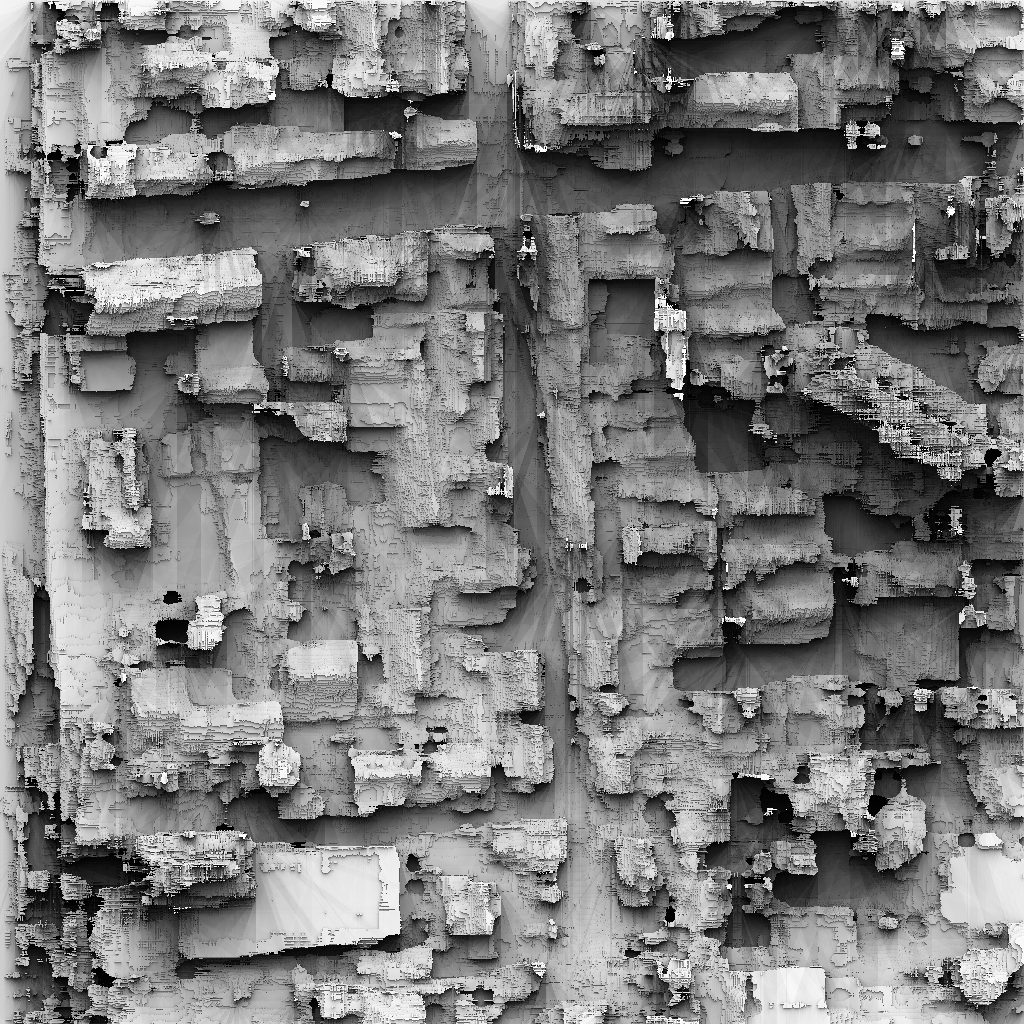}
		\centering{\tiny DeepFeature}
	\end{minipage}
	\begin{minipage}[t]{0.19\textwidth}
		\includegraphics[width=0.098\linewidth]{figures_supp/color_map.png}
		\includegraphics[width=0.85\linewidth]{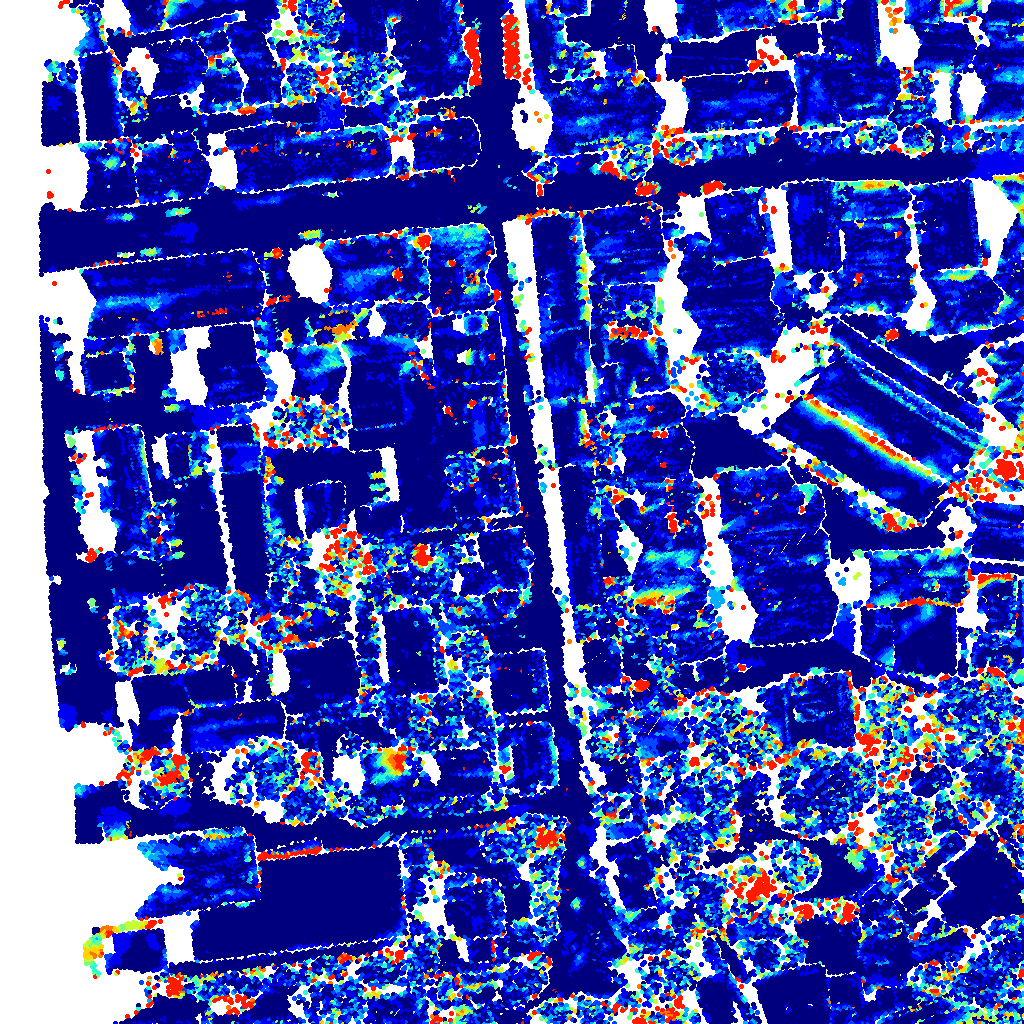}
		\includegraphics[width=\linewidth]{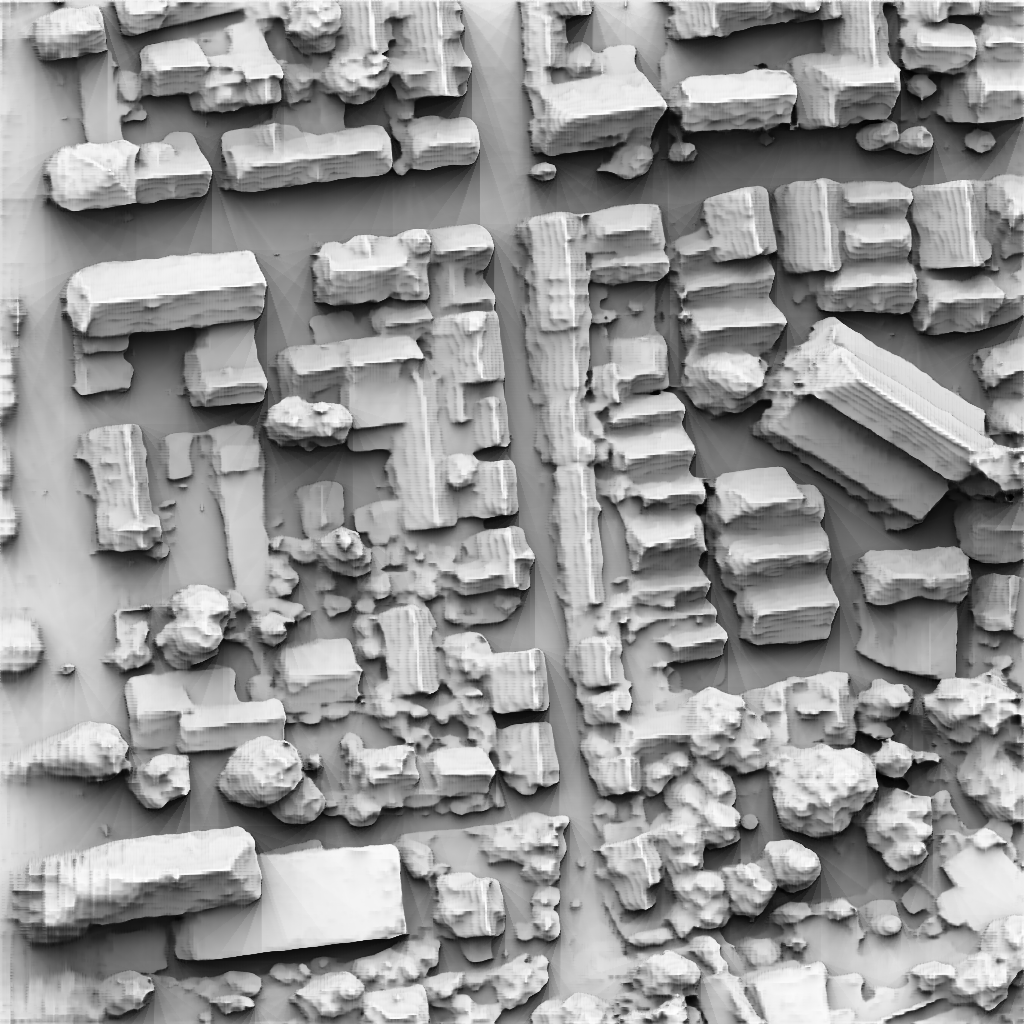}
		\centering{\tiny PSM Net}
	\end{minipage}
	\begin{minipage}[t]{0.19\textwidth}	
		\includegraphics[width=0.098\linewidth]{figures_supp/color_map.png}
		\includegraphics[width=0.85\linewidth]{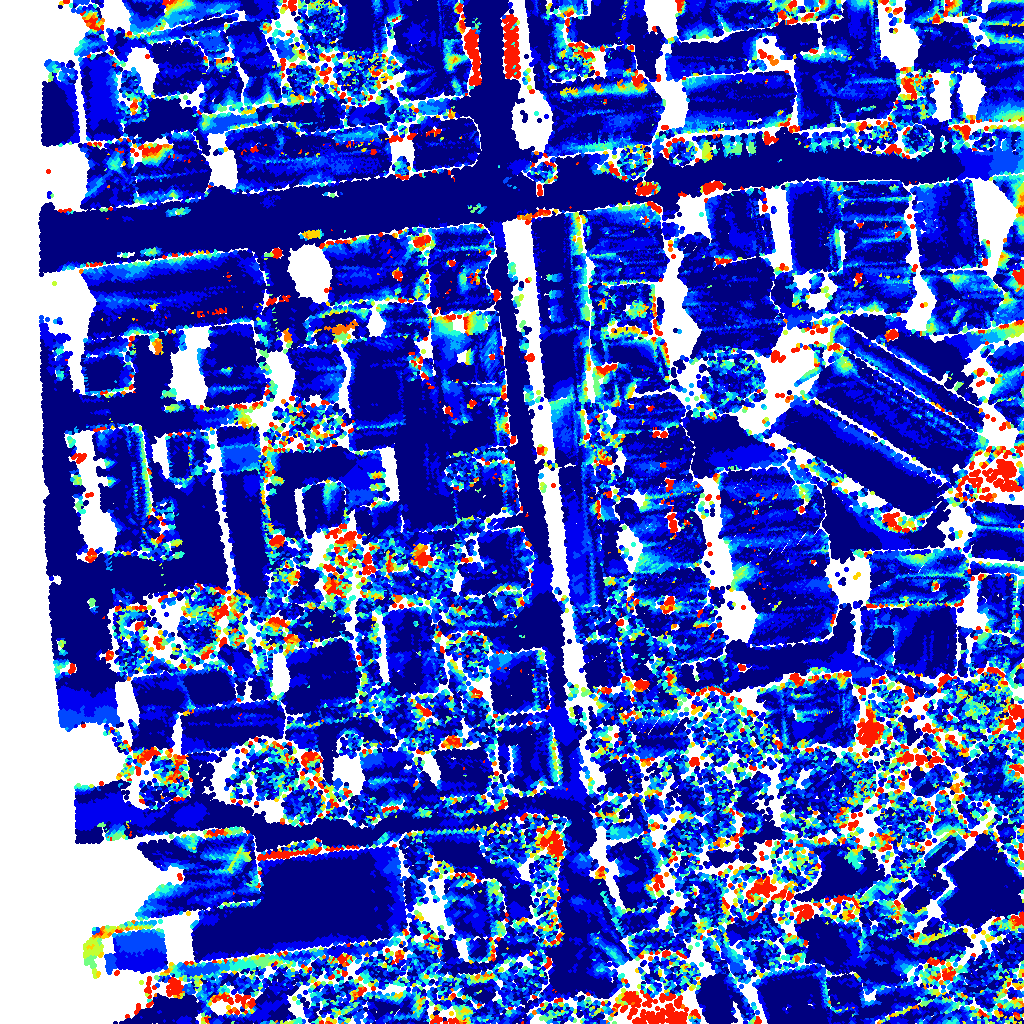} \\
		\includegraphics[width=\linewidth]{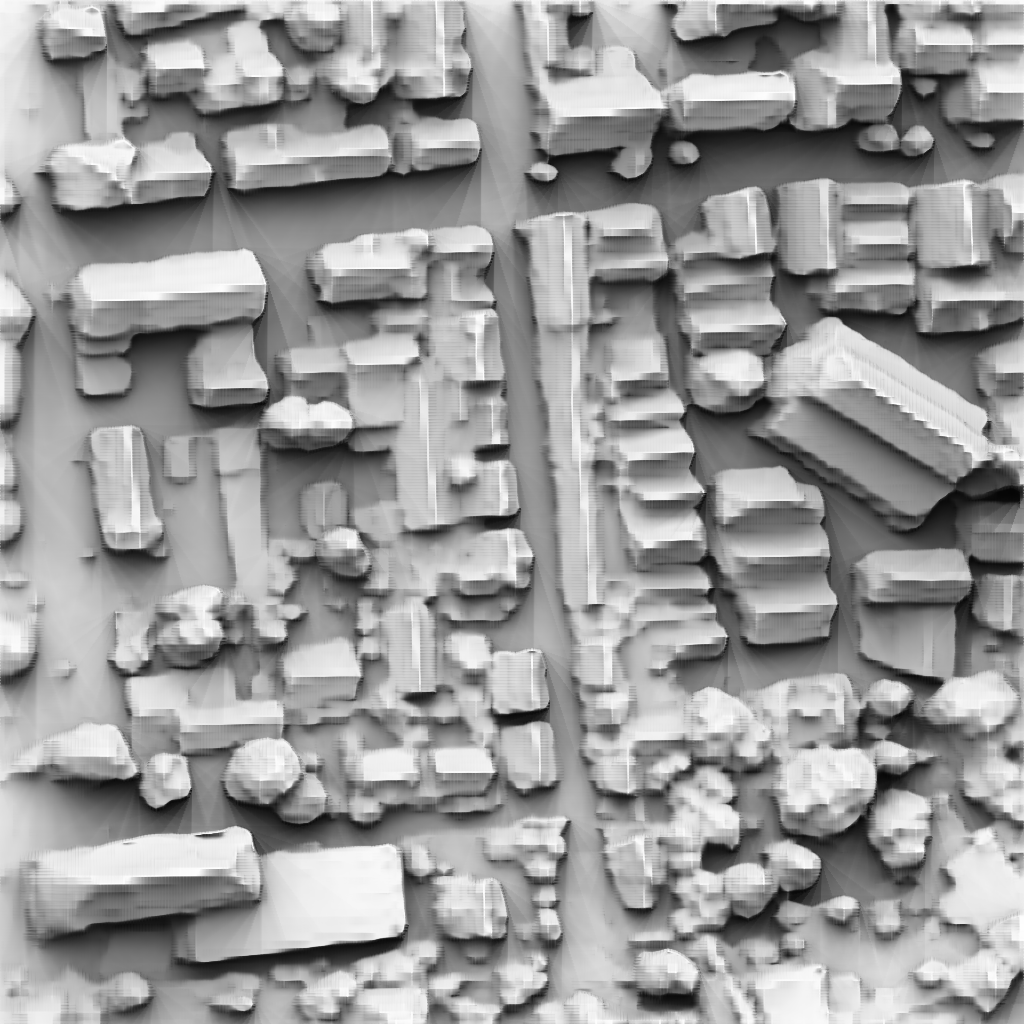}
		\centering{\tiny HRS Net}
	\end{minipage}	
	\begin{minipage}[t]{0.19\textwidth}	
		\includegraphics[width=0.098\linewidth]{figures_supp/color_map.png}
		\includegraphics[width=0.85\linewidth]{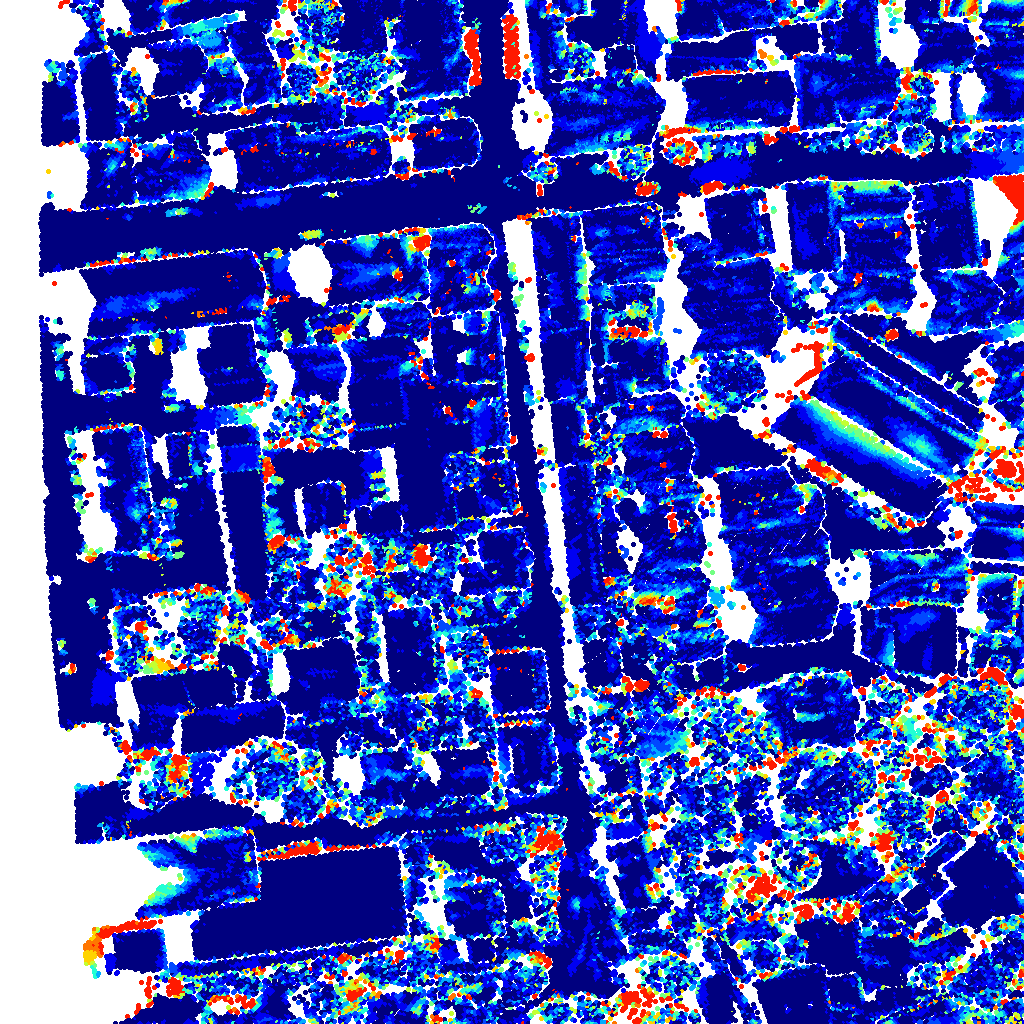}
		\includegraphics[width=\linewidth]{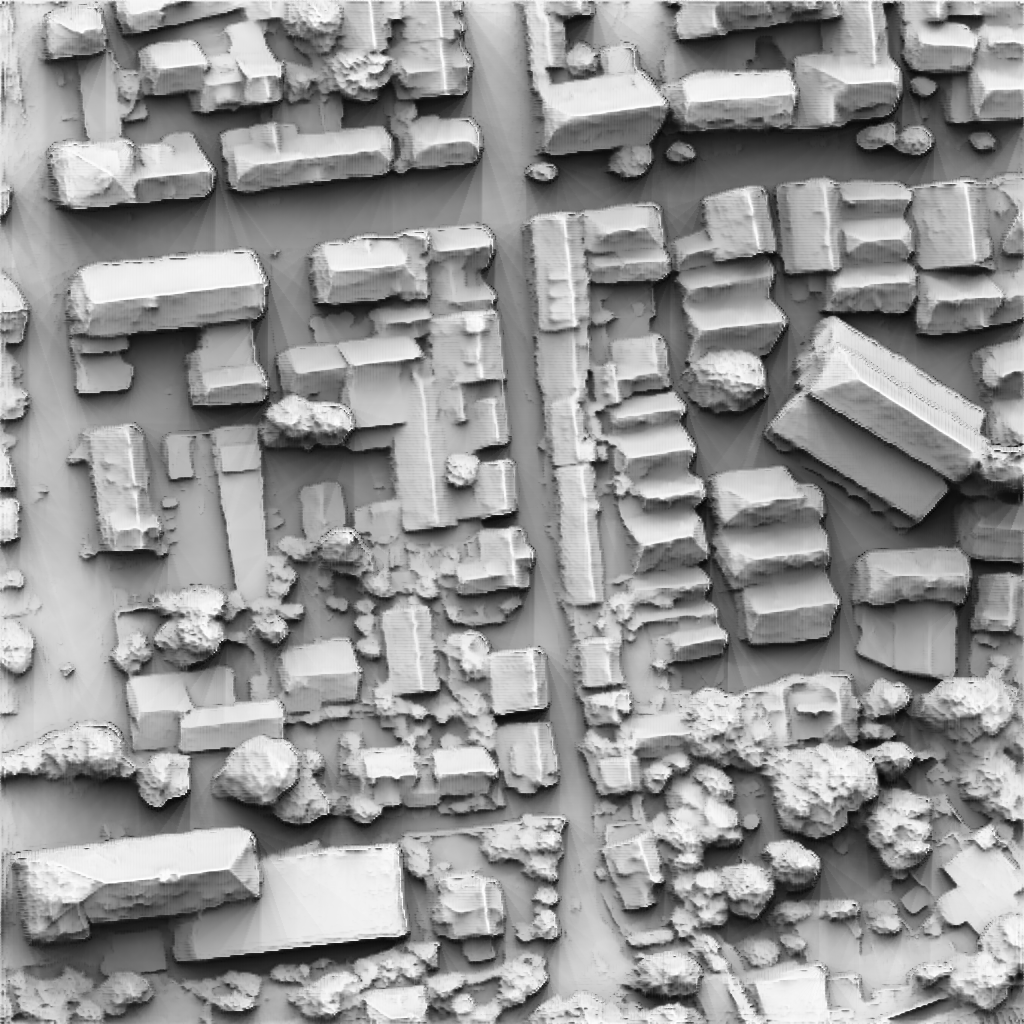}
		\centering{\tiny DeepPruner}
	\end{minipage}	
	\begin{minipage}[t]{0.19\textwidth}	
		\includegraphics[width=0.098\linewidth]{figures_supp/color_map.png}
		\includegraphics[width=0.85\linewidth]{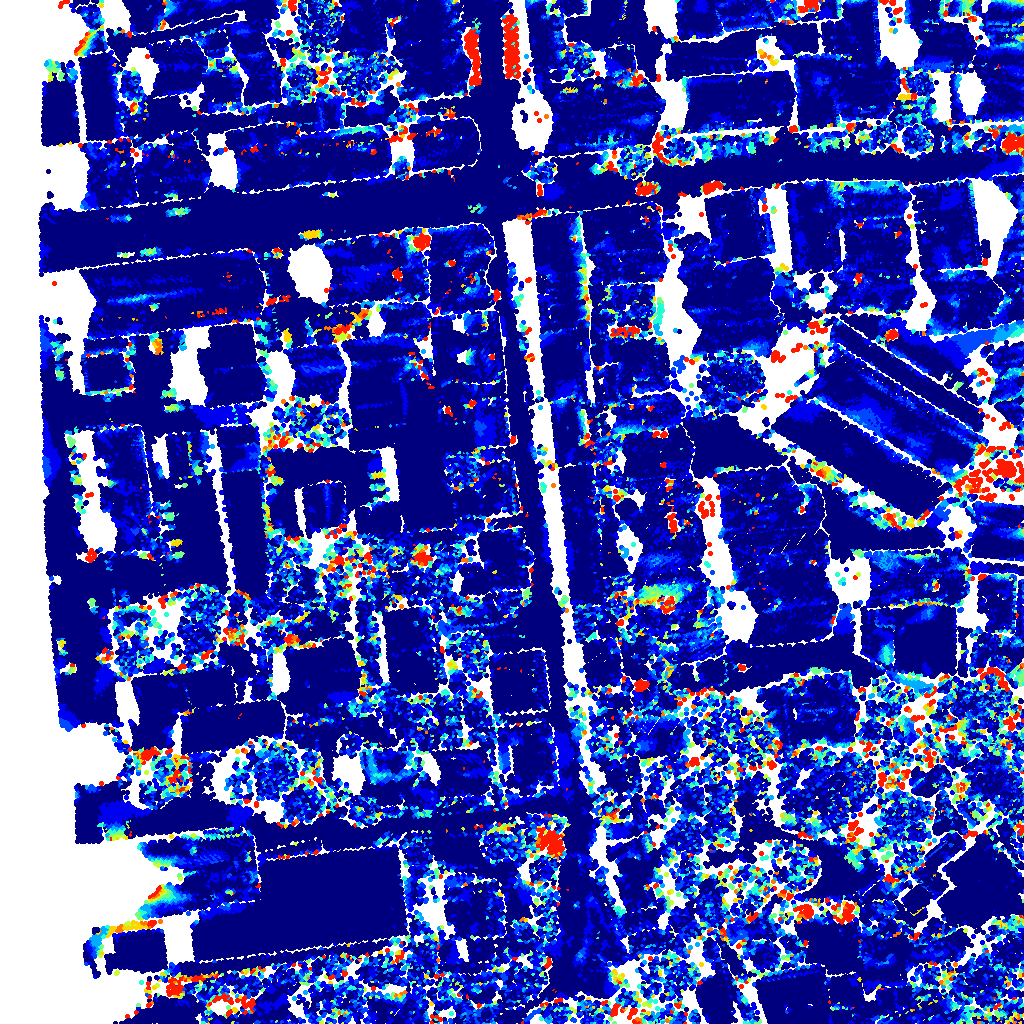}
		\includegraphics[width=\linewidth]{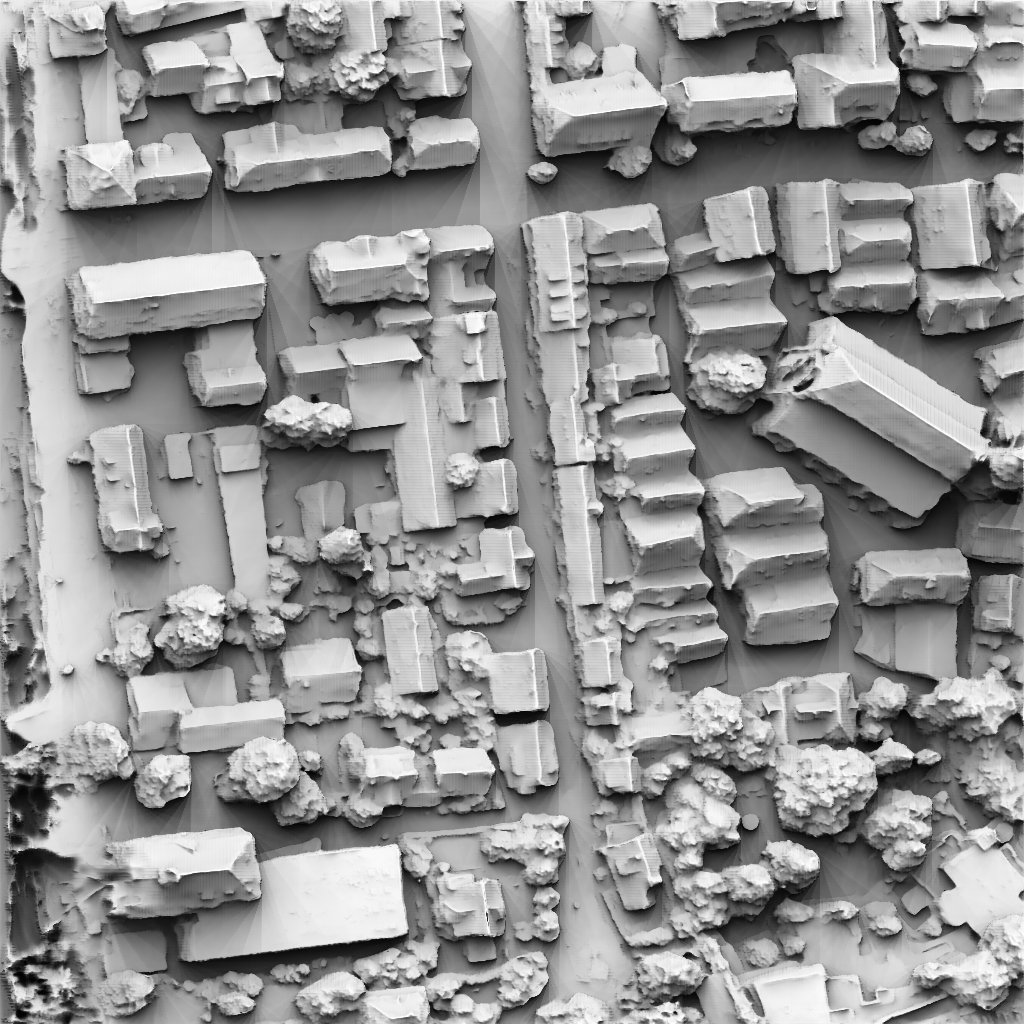}
		\centering{\tiny GANet}
	\end{minipage}	
	\begin{minipage}[t]{0.19\textwidth}	
		\includegraphics[width=0.098\linewidth]{figures_supp/color_map.png}
		\includegraphics[width=0.85\linewidth]{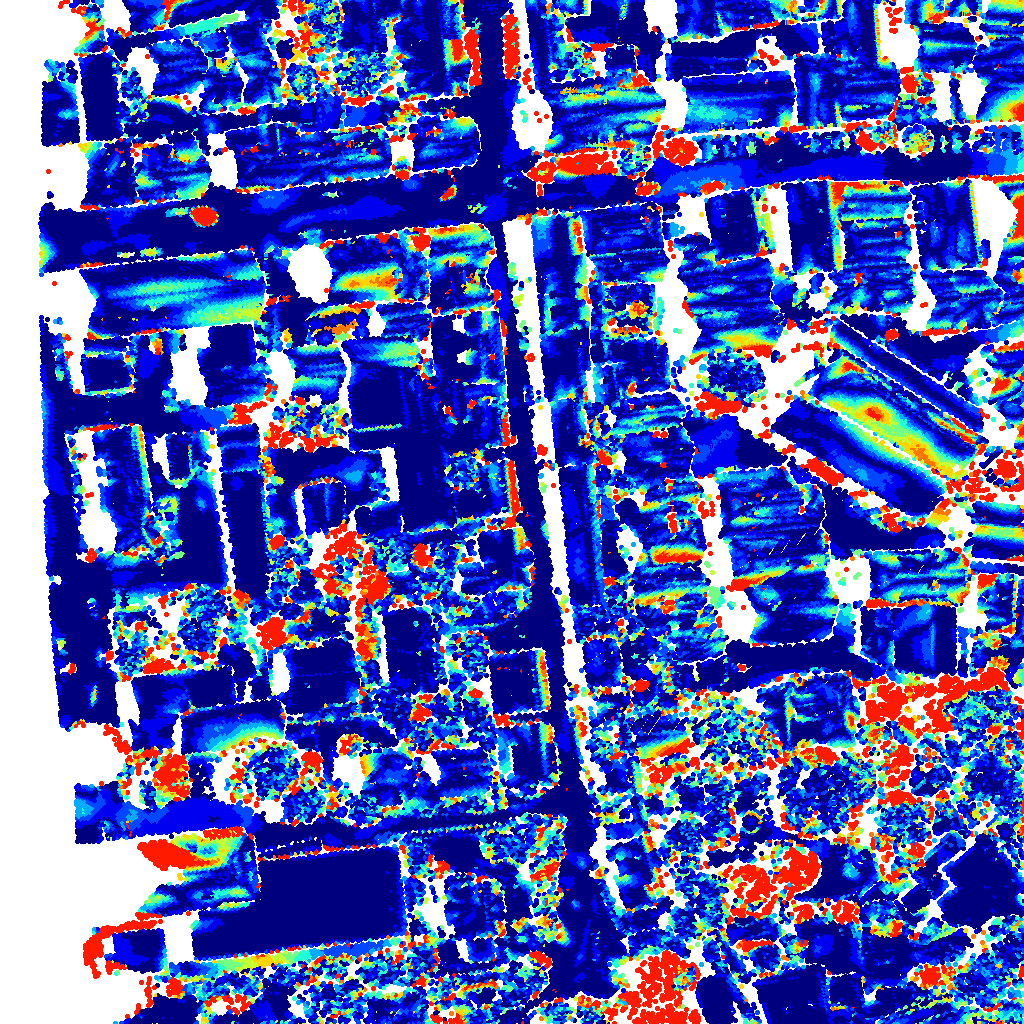}
		\includegraphics[width=\linewidth]{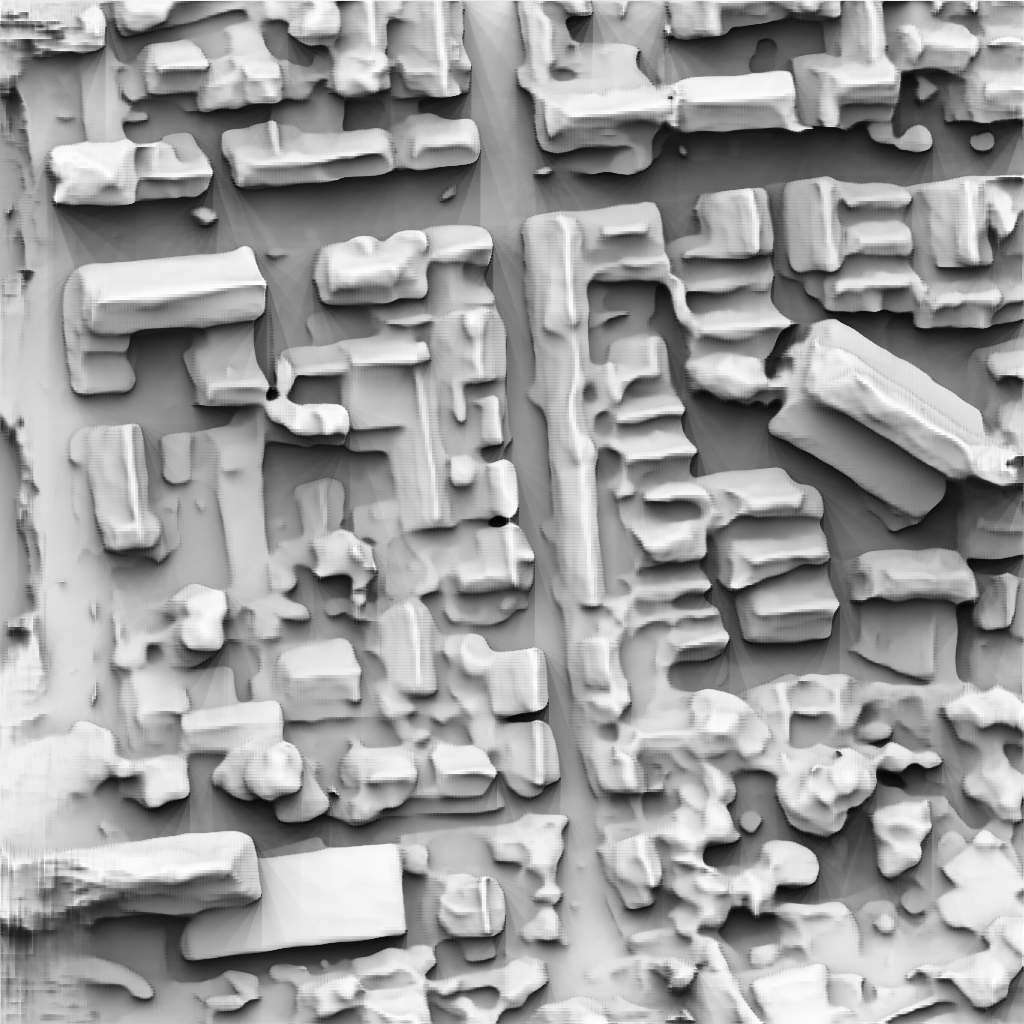}
		\centering{\tiny LEAStereo}
	\end{minipage}	
	\caption{Error map and disparity visualization on building area for EuroSDR Vaihingen.}
	\label{Figure.eurosdr_bulding}
\end{figure}

\begin{figure}[tp]
	\begin{minipage}[t]{0.19\textwidth}
		\includegraphics[width=0.098\linewidth]{figures_supp/color_map.png}
		\includegraphics[width=0.85\linewidth]{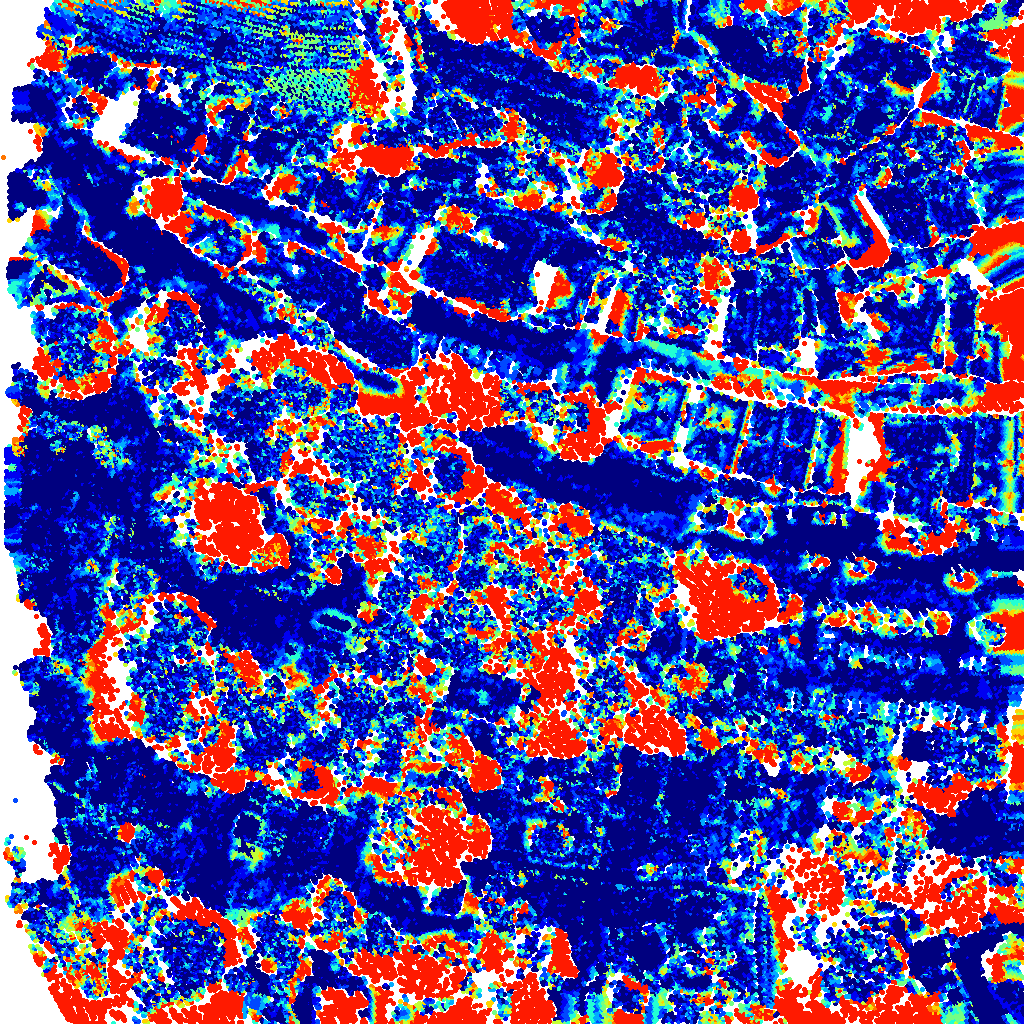} \\
		\includegraphics[width=\linewidth]{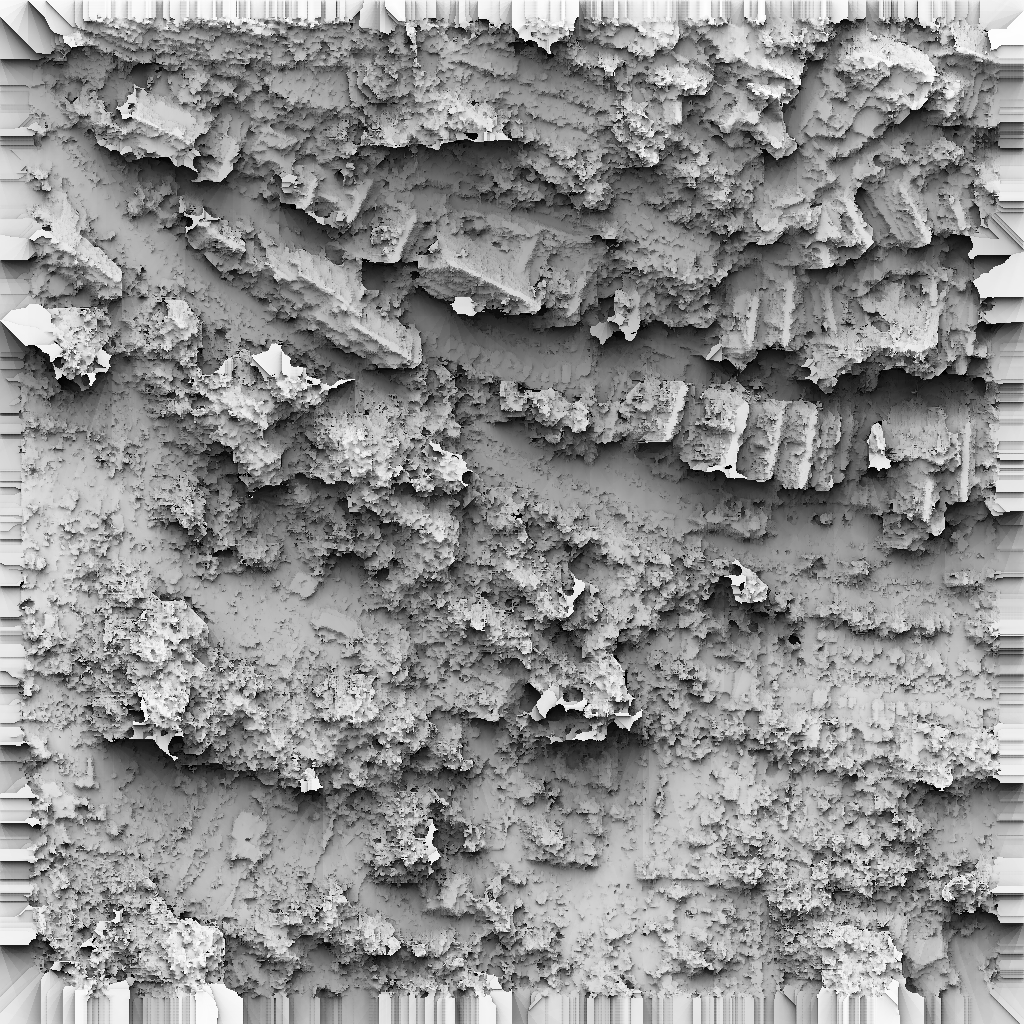}\\
		\centering{\tiny MICMAC}
	\end{minipage}
	\begin{minipage}[t]{0.19\textwidth}
		\includegraphics[width=0.098\linewidth]{figures_supp/color_map.png}
		\includegraphics[width=0.85\linewidth]{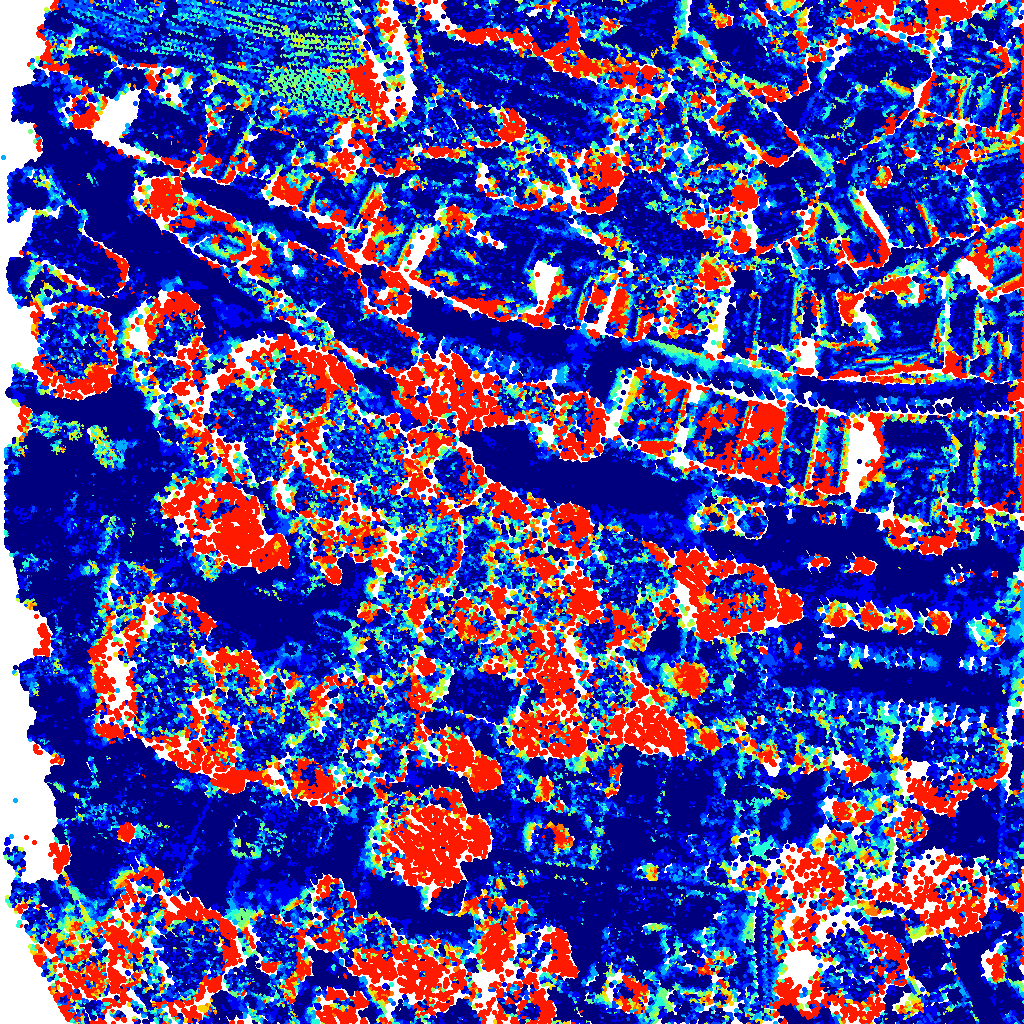} \\
		\includegraphics[width=\linewidth]{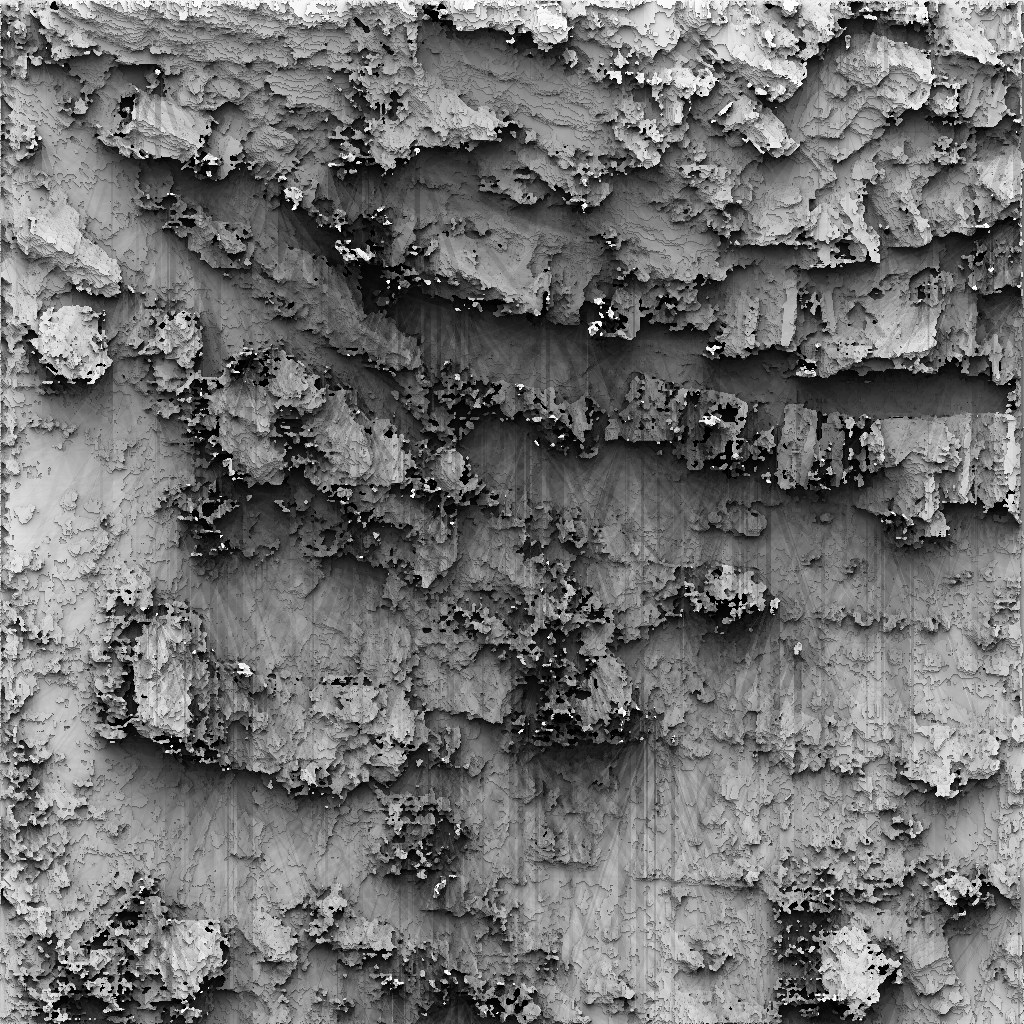} \\
		\centering{\tiny SGM(CUDA)}
	\end{minipage}
	\begin{minipage}[t]{0.19\textwidth}
		\includegraphics[width=0.098\linewidth]{figures_supp/color_map.png}
		\includegraphics[width=0.85\linewidth]{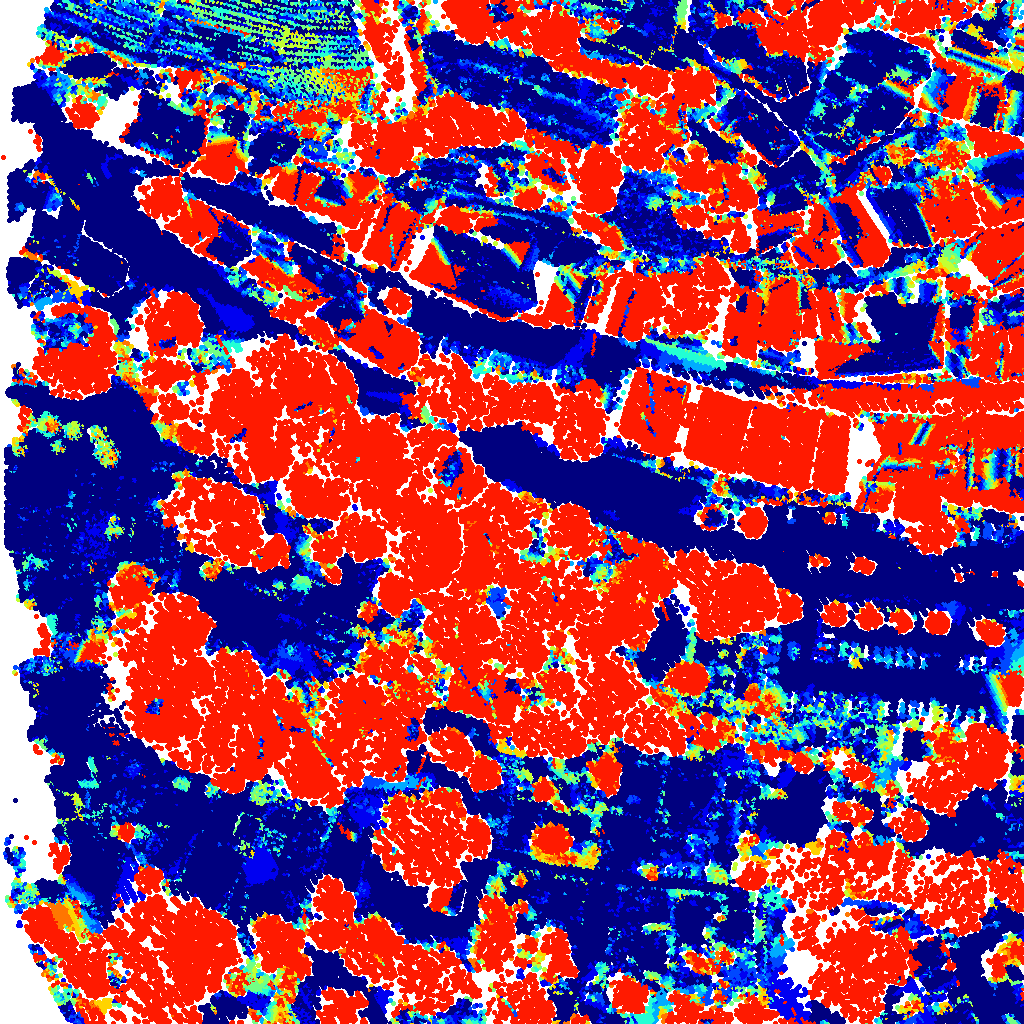} \\
		\includegraphics[width=\linewidth]{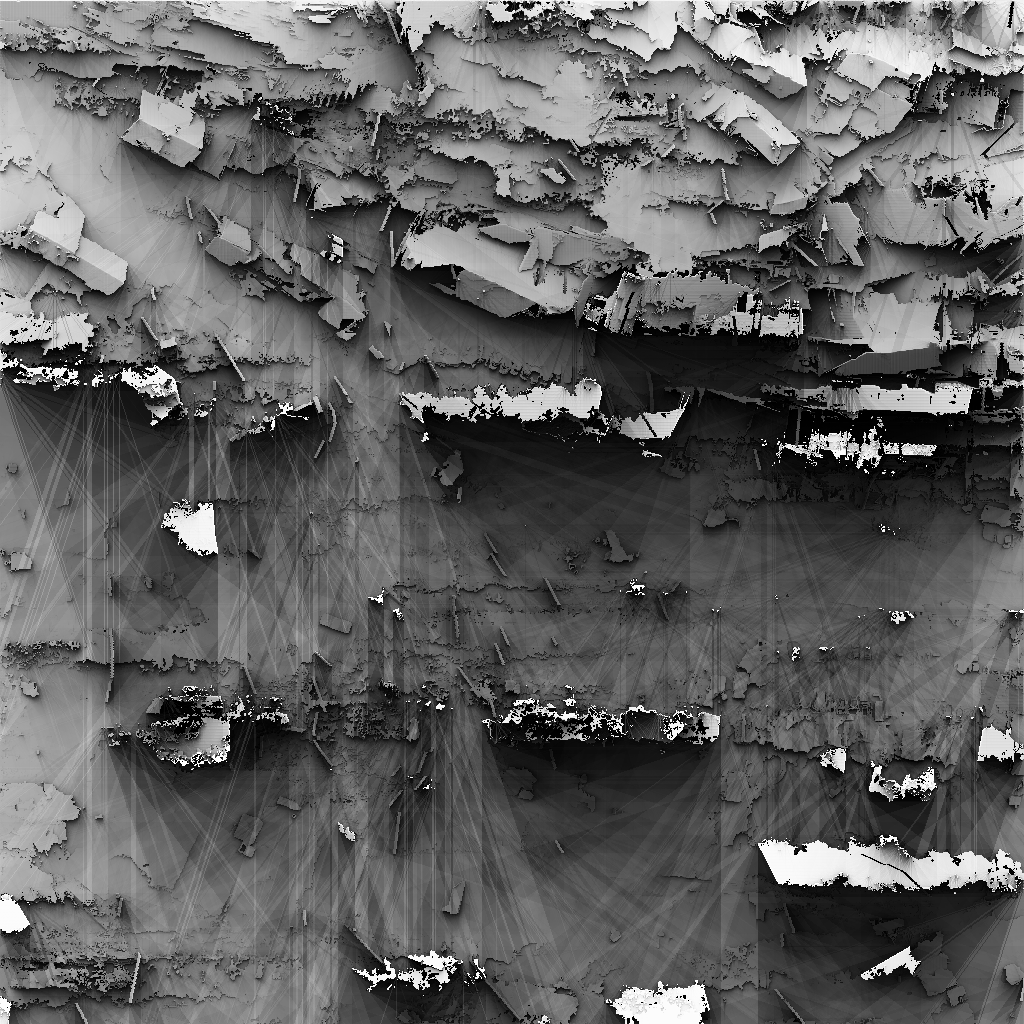}
		\centering{\tiny GraphCuts}
	\end{minipage}
	\begin{minipage}[t]{0.19\textwidth}
		\includegraphics[width=0.098\linewidth]{figures_supp/color_map.png}
		\includegraphics[width=0.85\linewidth]{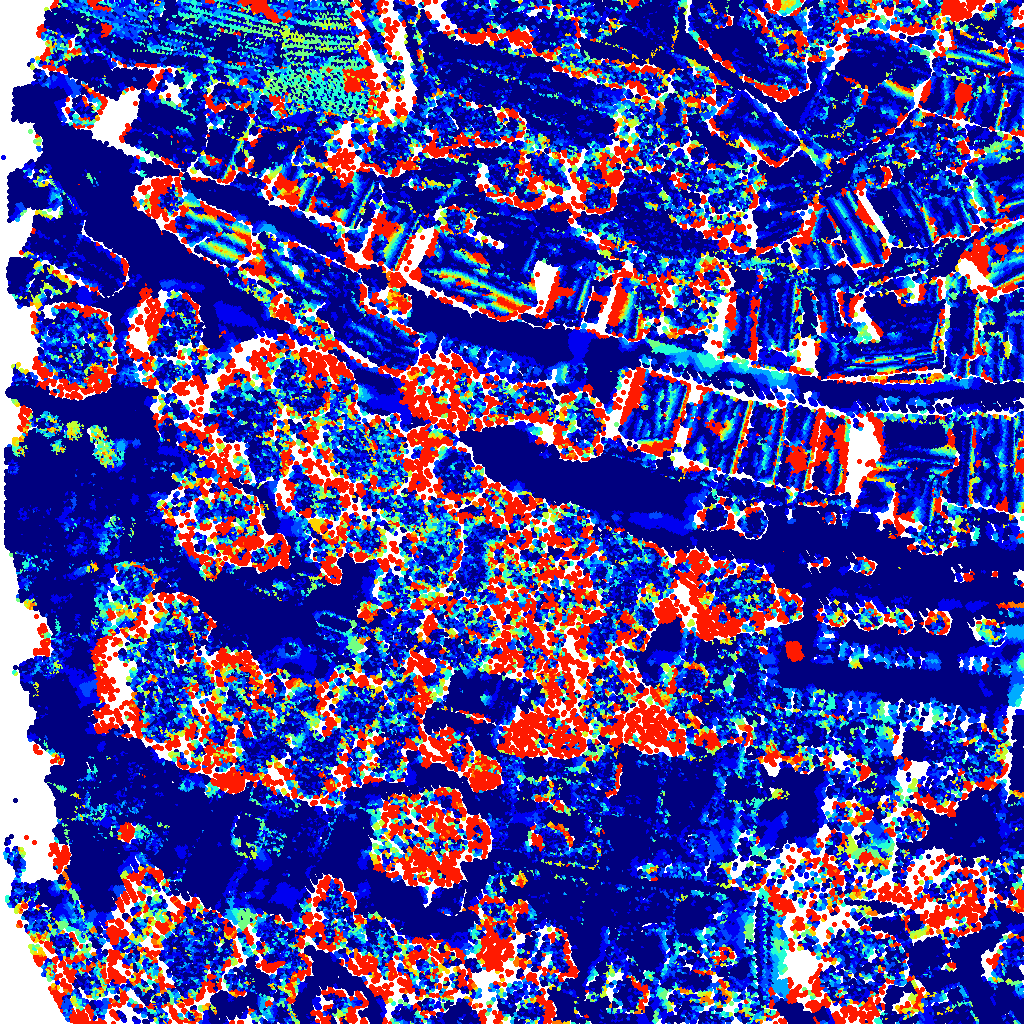} \\
		\includegraphics[width=\linewidth]{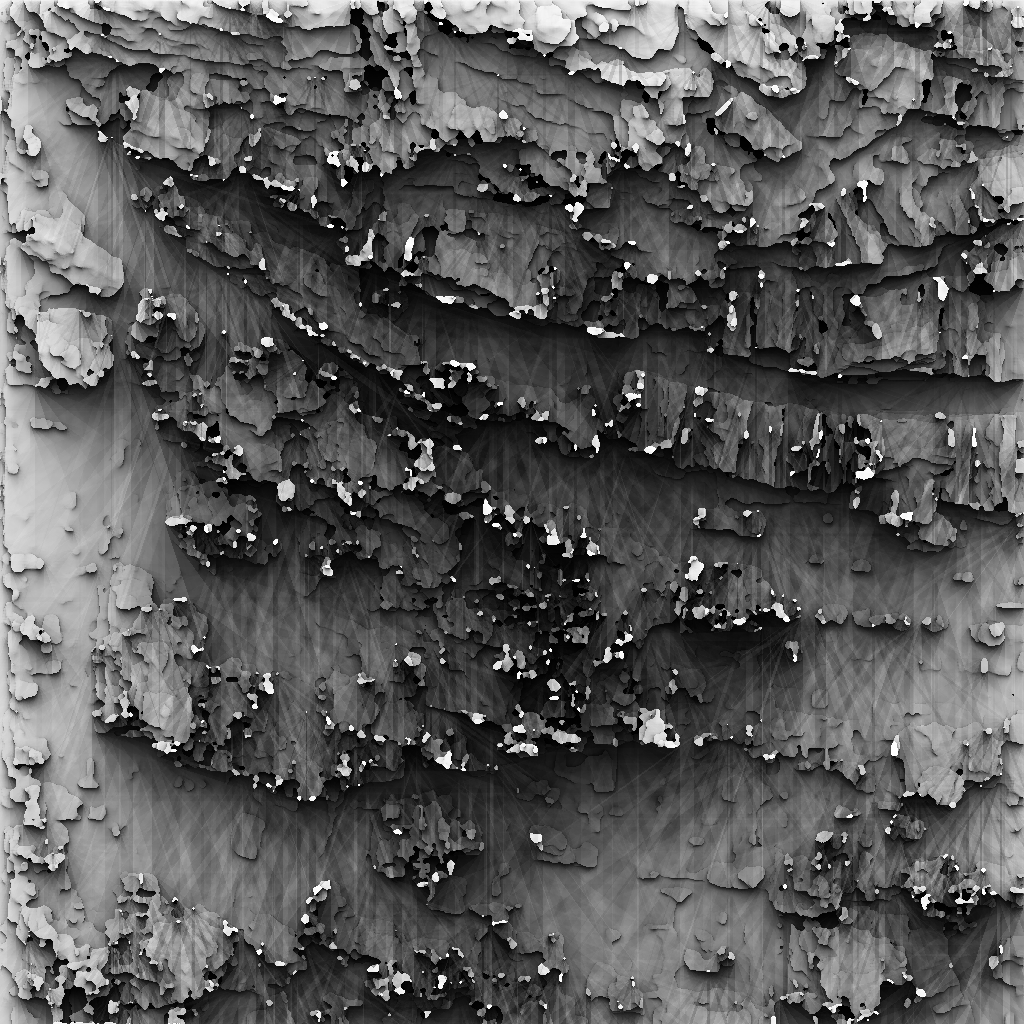}
		\centering{\tiny CBMV(SGM)}
	\end{minipage}
	\begin{minipage}[t]{0.19\textwidth}
		\includegraphics[width=0.098\linewidth]{figures_supp/color_map.png}
		\includegraphics[width=0.85\linewidth]{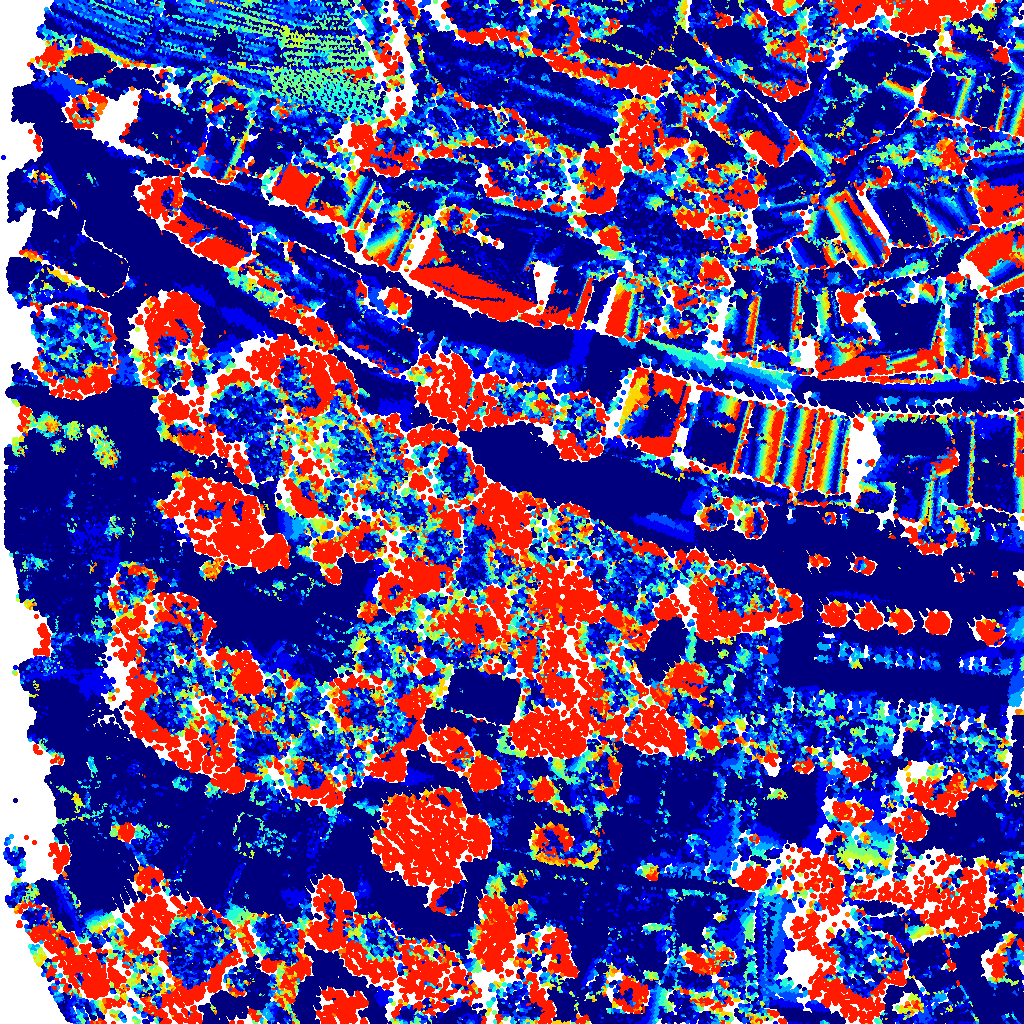} \\
		\includegraphics[width=\linewidth]{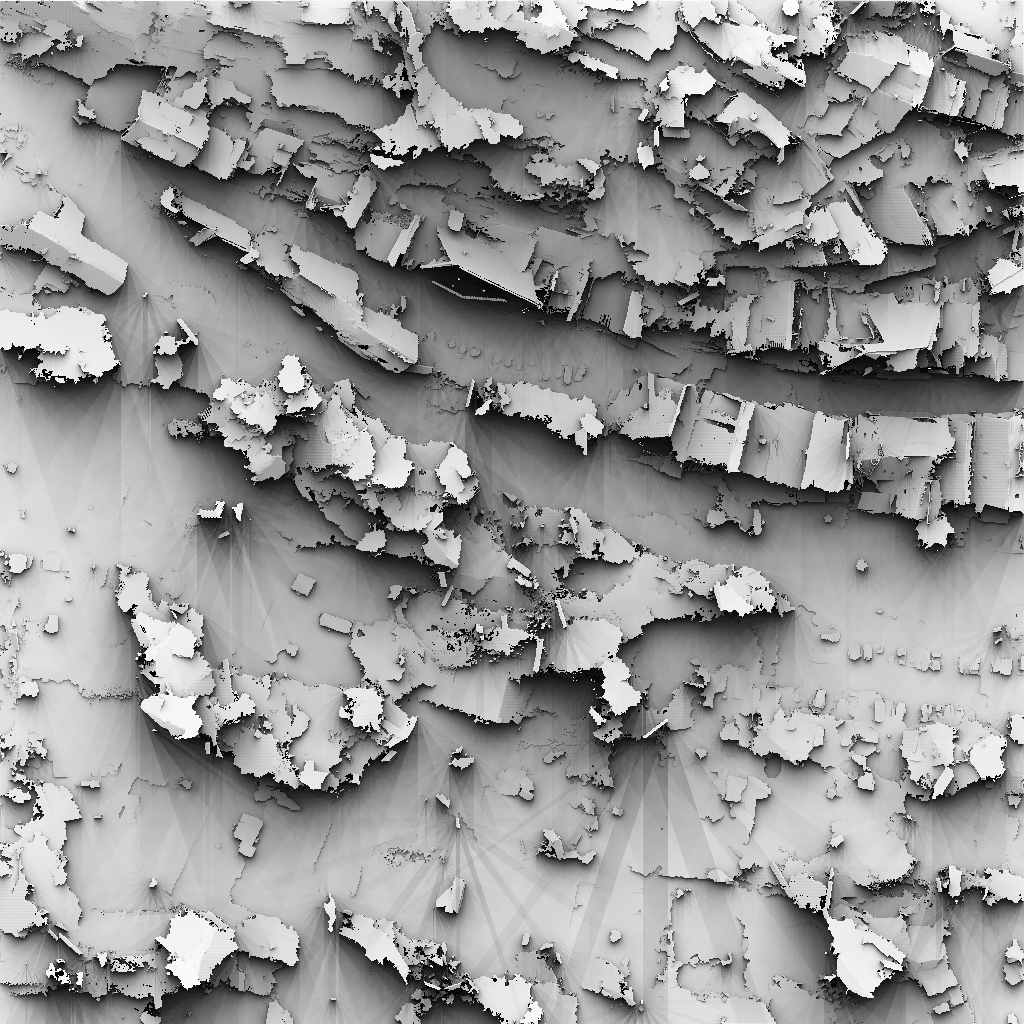}
		\centering{\tiny CBMV(GraphCuts)}
	\end{minipage}
	
	\begin{minipage}[t]{0.19\textwidth}
		\includegraphics[width=0.098\linewidth]{figures_supp/color_map.png}
		\includegraphics[width=0.85\linewidth]{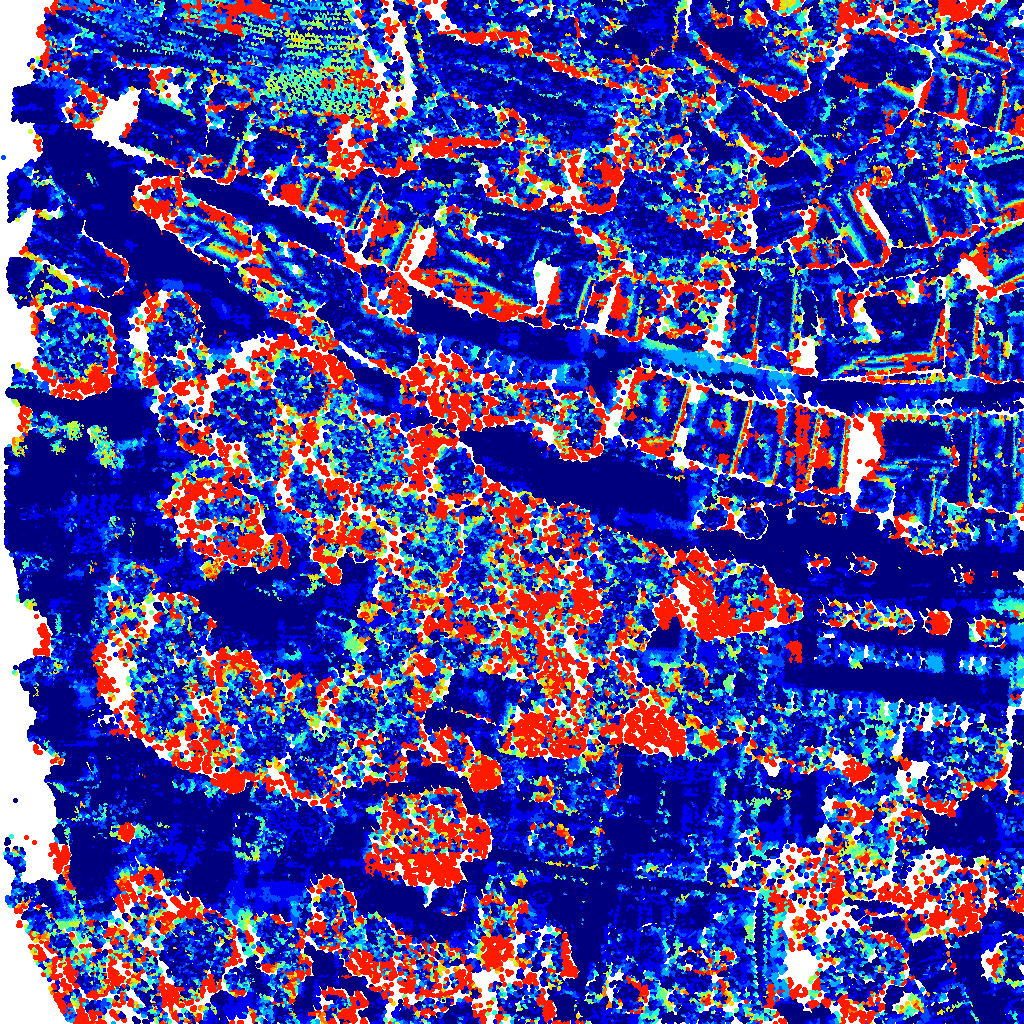} \\
		\includegraphics[width=\linewidth]{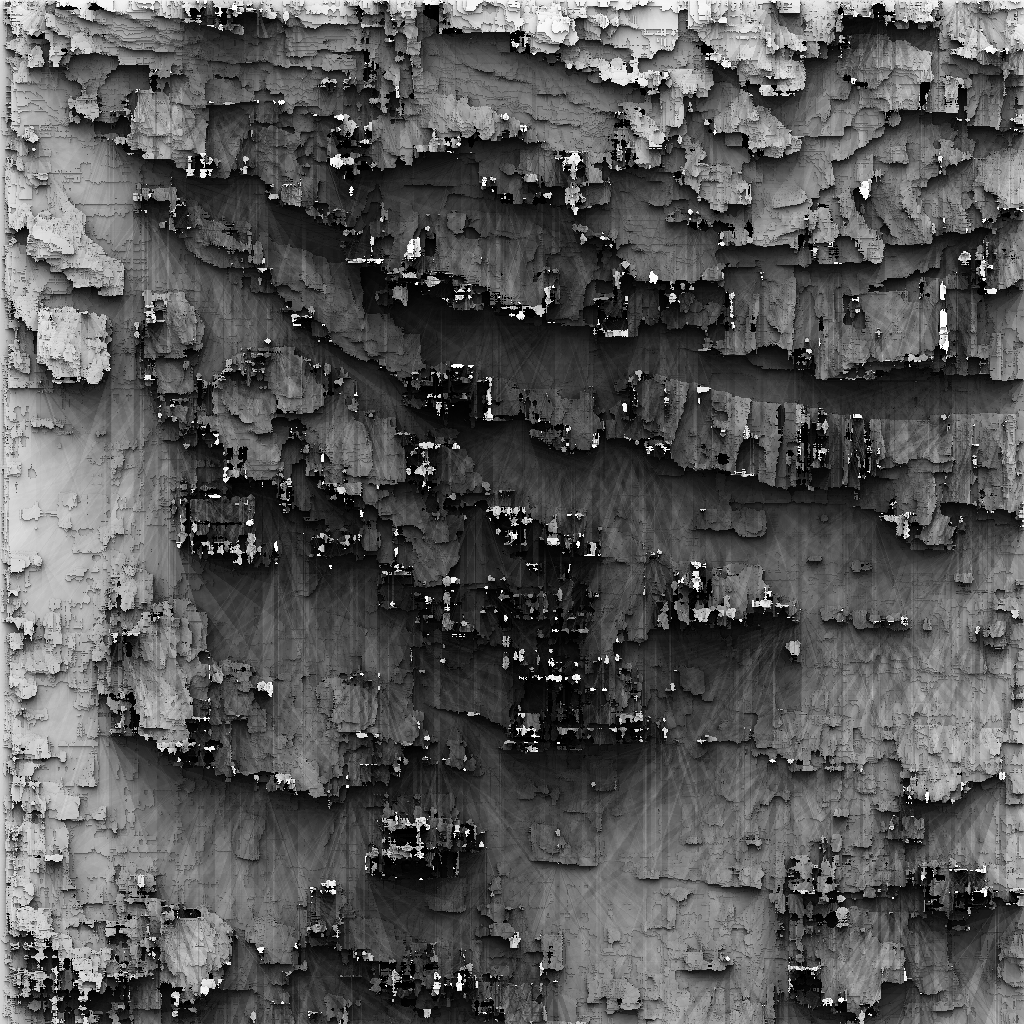}
		\centering{\tiny MC-CNN(KITTI)}
	\end{minipage}
	\begin{minipage}[t]{0.19\textwidth}
		\includegraphics[width=0.098\linewidth]{figures_supp/color_map.png}
		\includegraphics[width=0.85\linewidth]{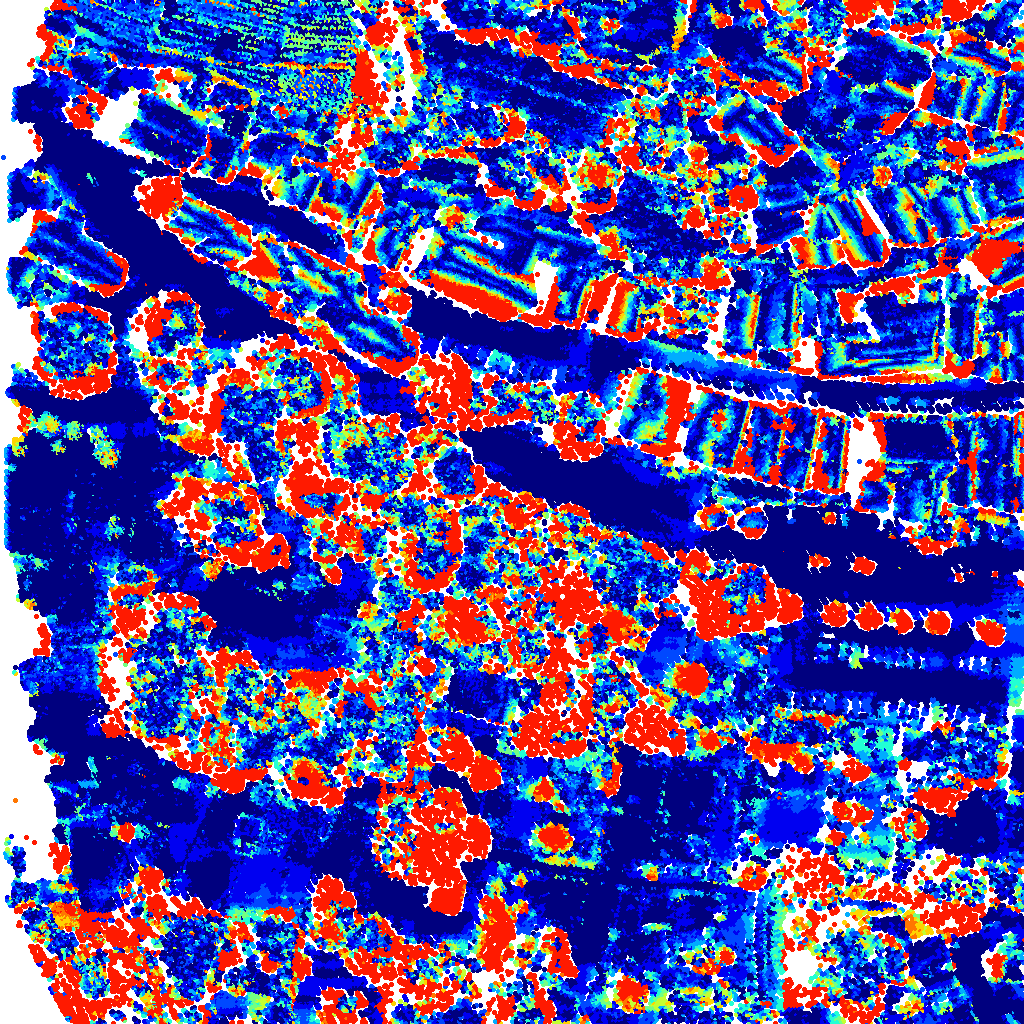}
		\includegraphics[width=\linewidth]{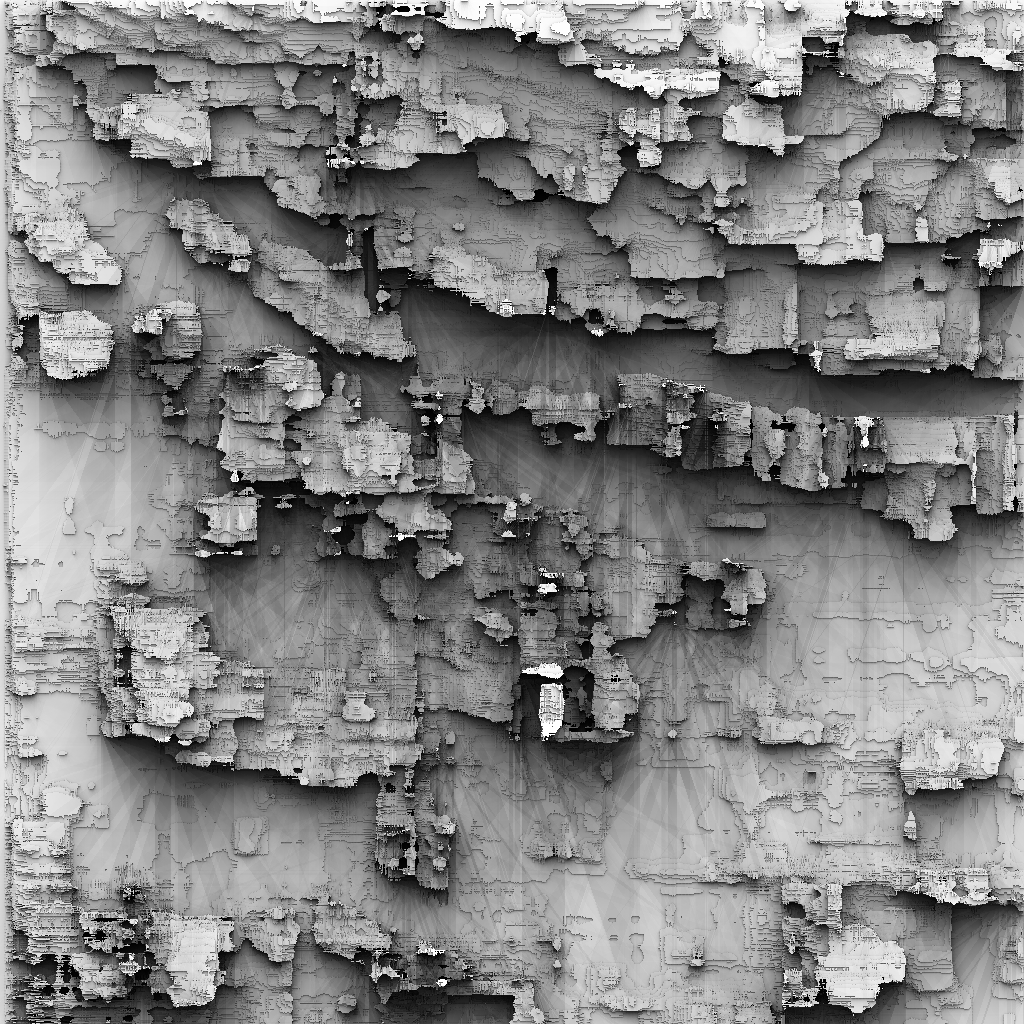}
		\centering{\tiny DeepFeature(KITTI)}
	\end{minipage}
	\begin{minipage}[t]{0.19\textwidth}
		\includegraphics[width=0.098\linewidth]{figures_supp/color_map.png}
		\includegraphics[width=0.85\linewidth]{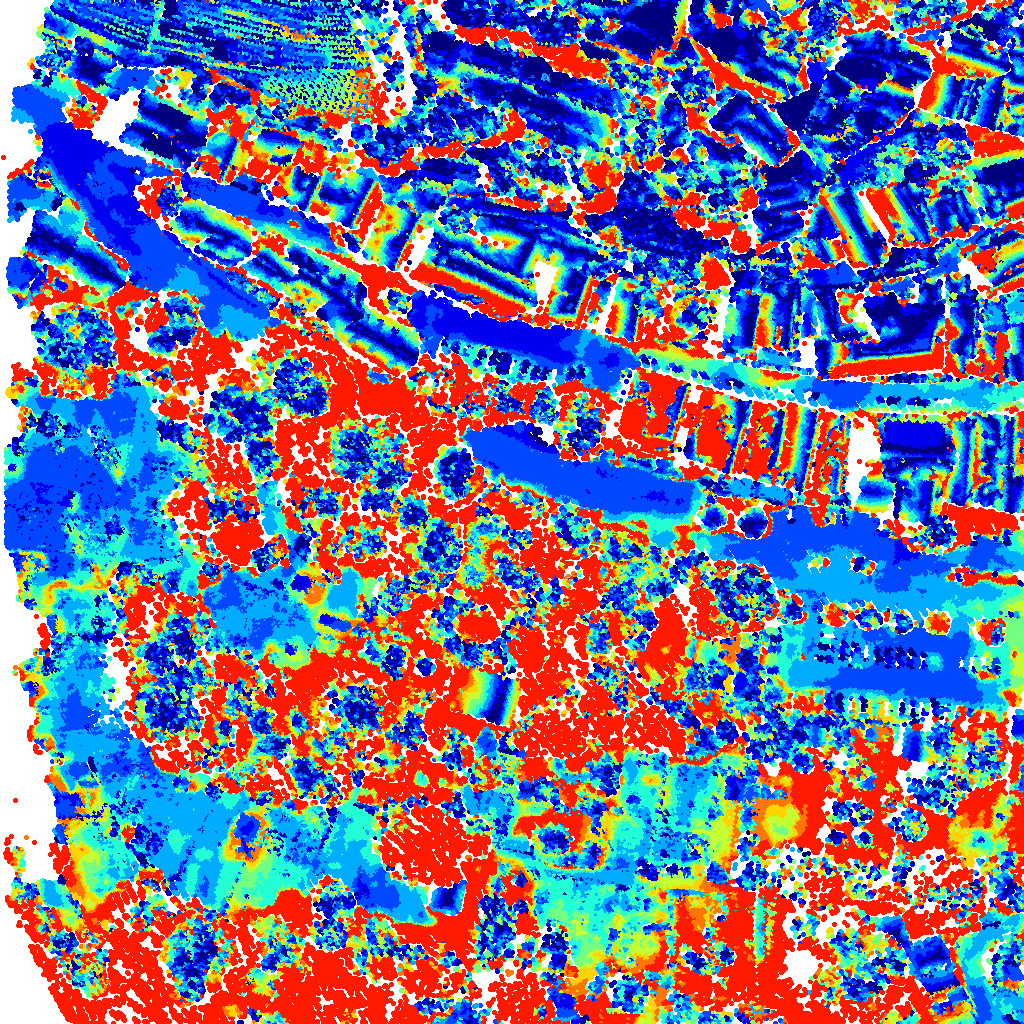}
		\includegraphics[width=\linewidth]{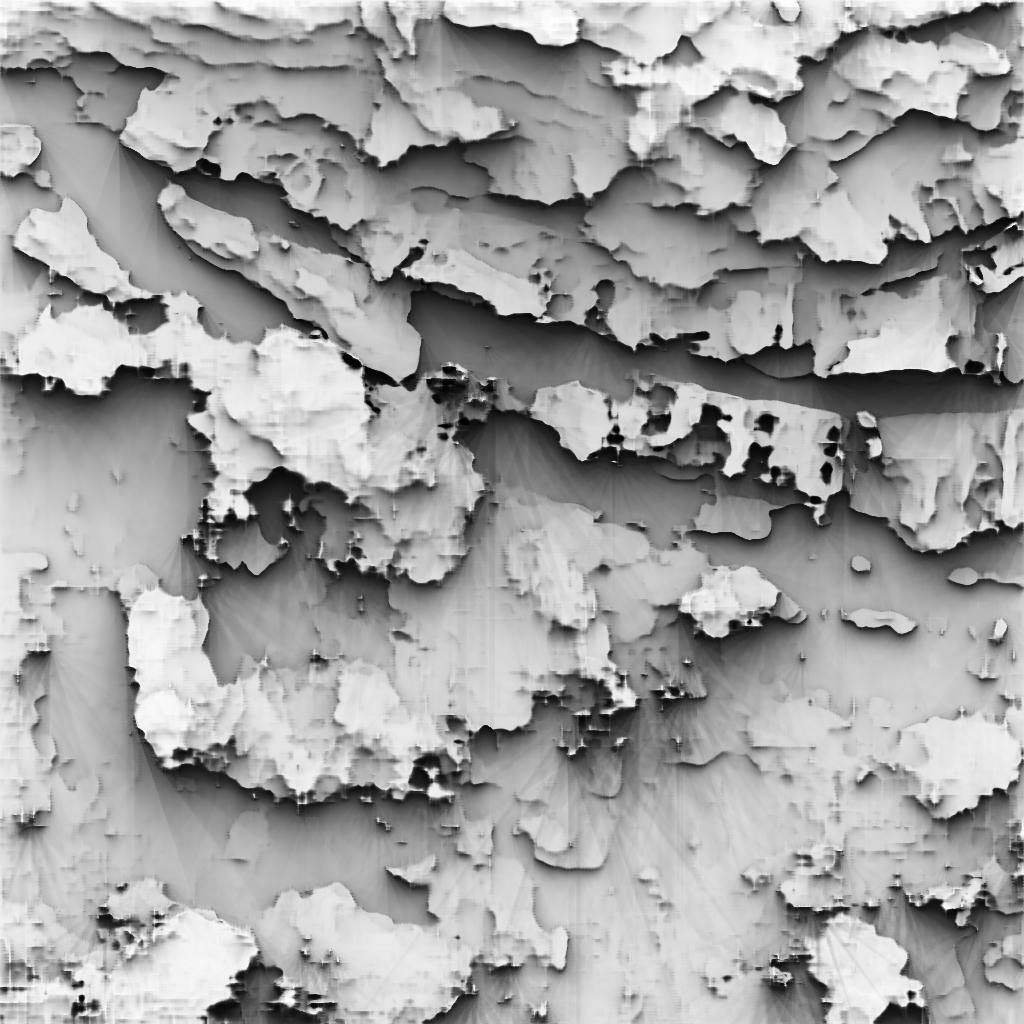}
		\centering{\tiny PSM Net(KITTI)}
	\end{minipage}
	\begin{minipage}[t]{0.19\textwidth}	
		\includegraphics[width=0.098\linewidth]{figures_supp/color_map.png}
		\includegraphics[width=0.85\linewidth]{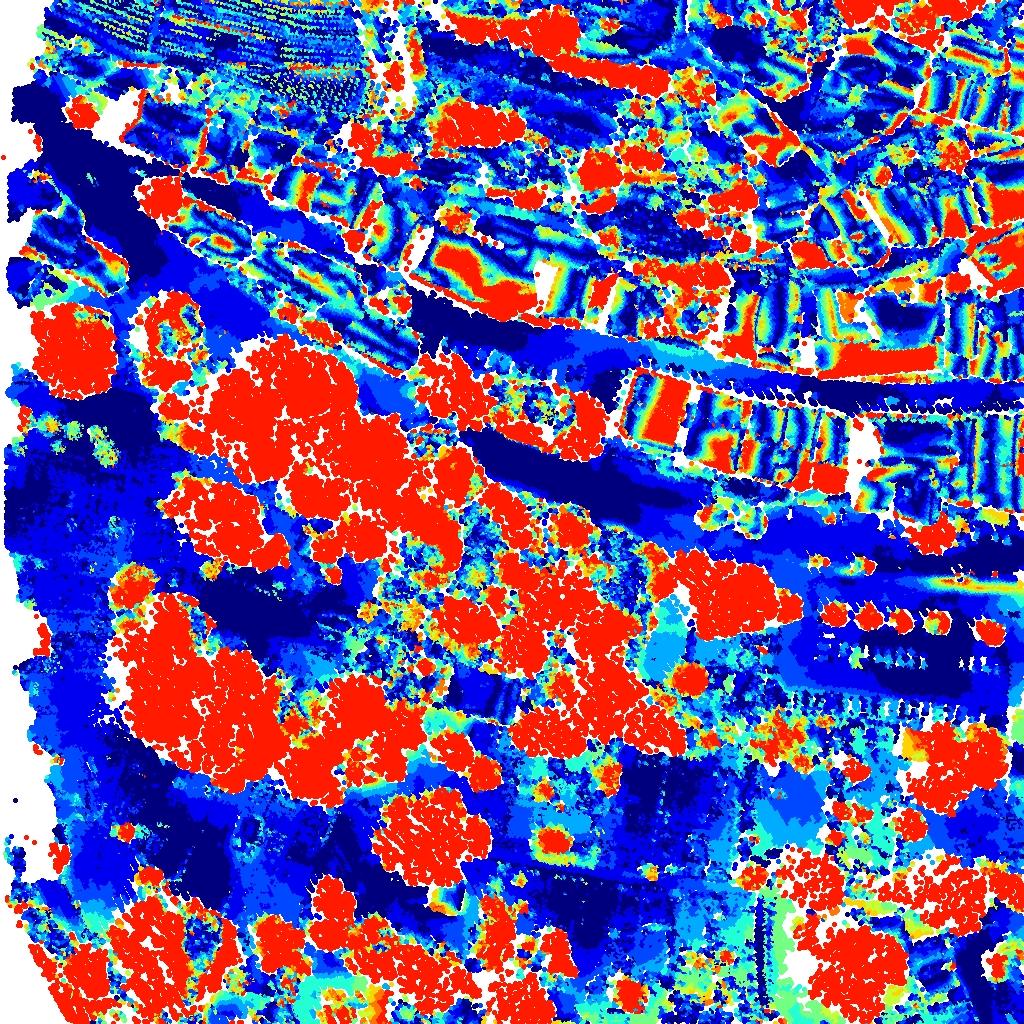} \\
		\includegraphics[width=\linewidth]{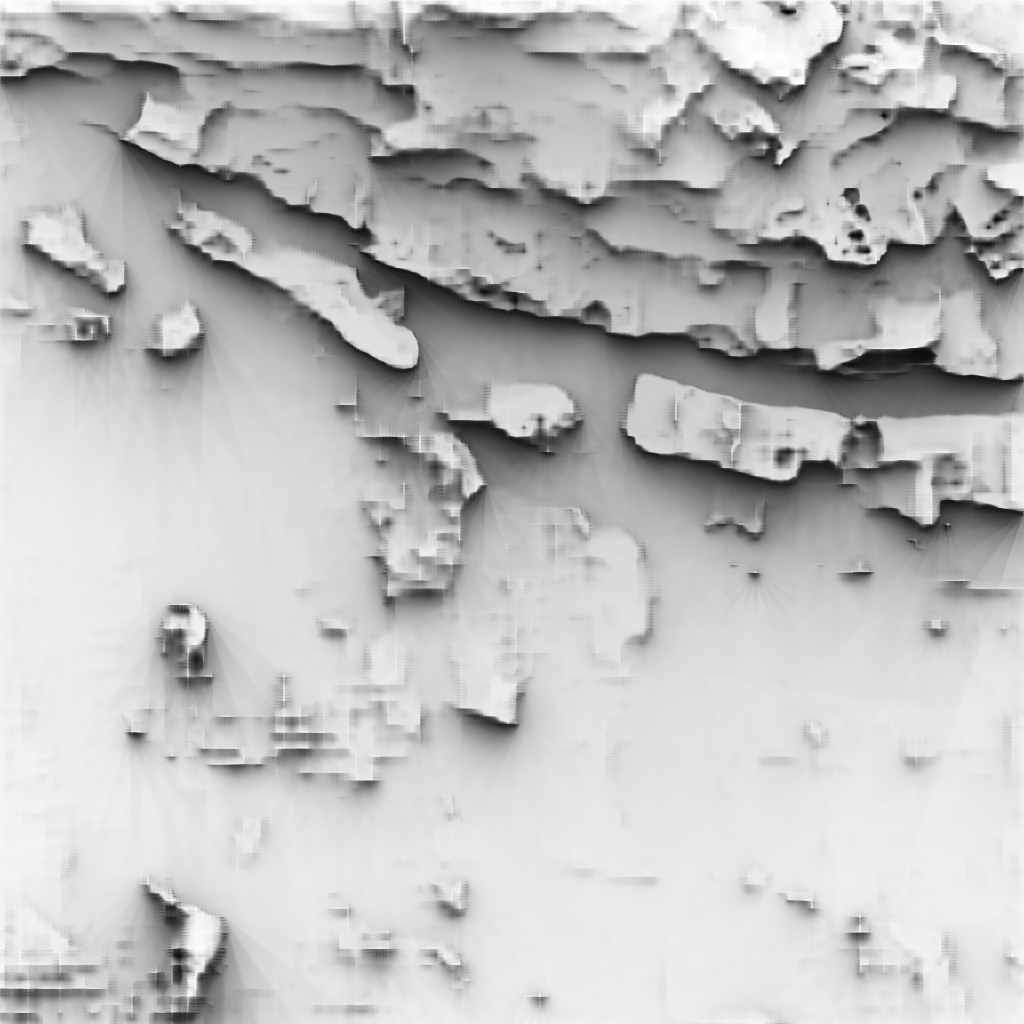}
		\centering{\tiny HRS Net(KITTI)}
	\end{minipage}	
	\begin{minipage}[t]{0.19\textwidth}	
		\includegraphics[width=0.098\linewidth]{figures_supp/color_map.png}
		\includegraphics[width=0.85\linewidth]{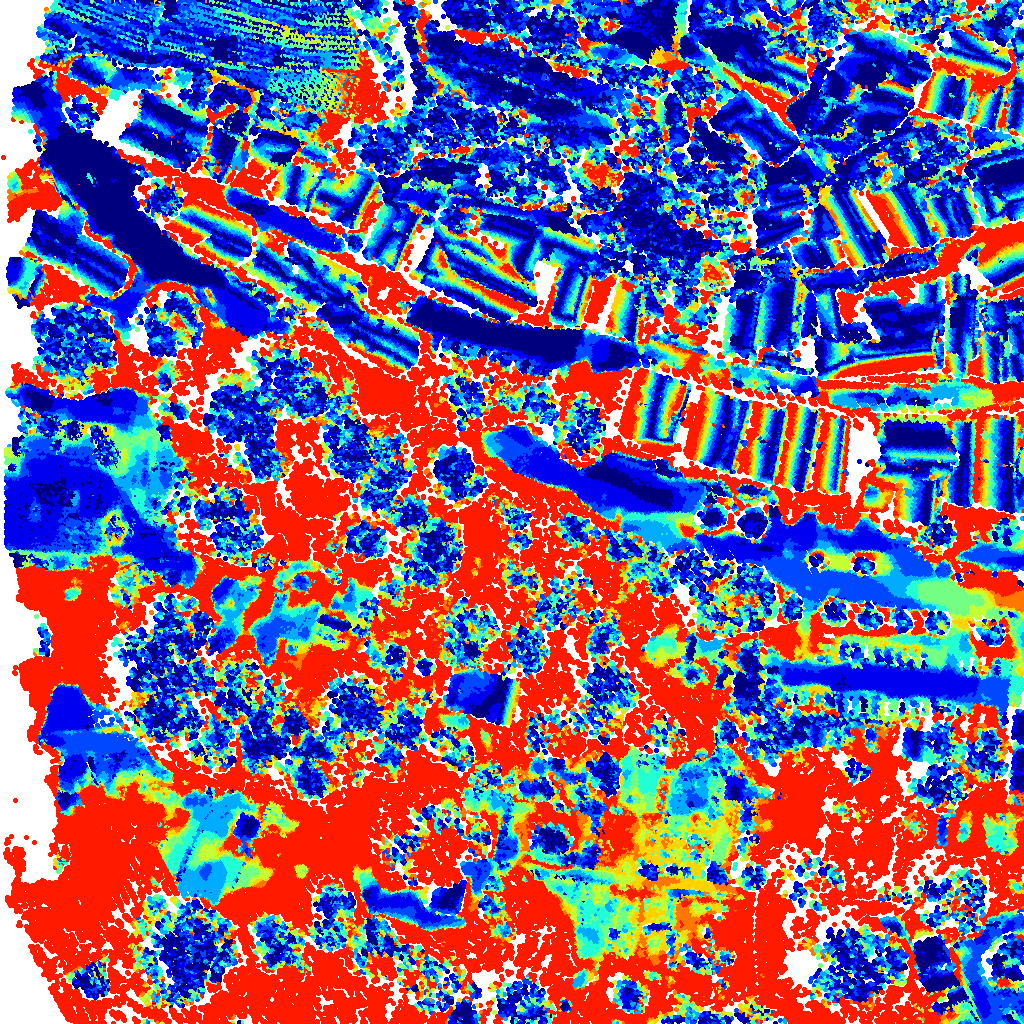}
		\includegraphics[width=\linewidth]{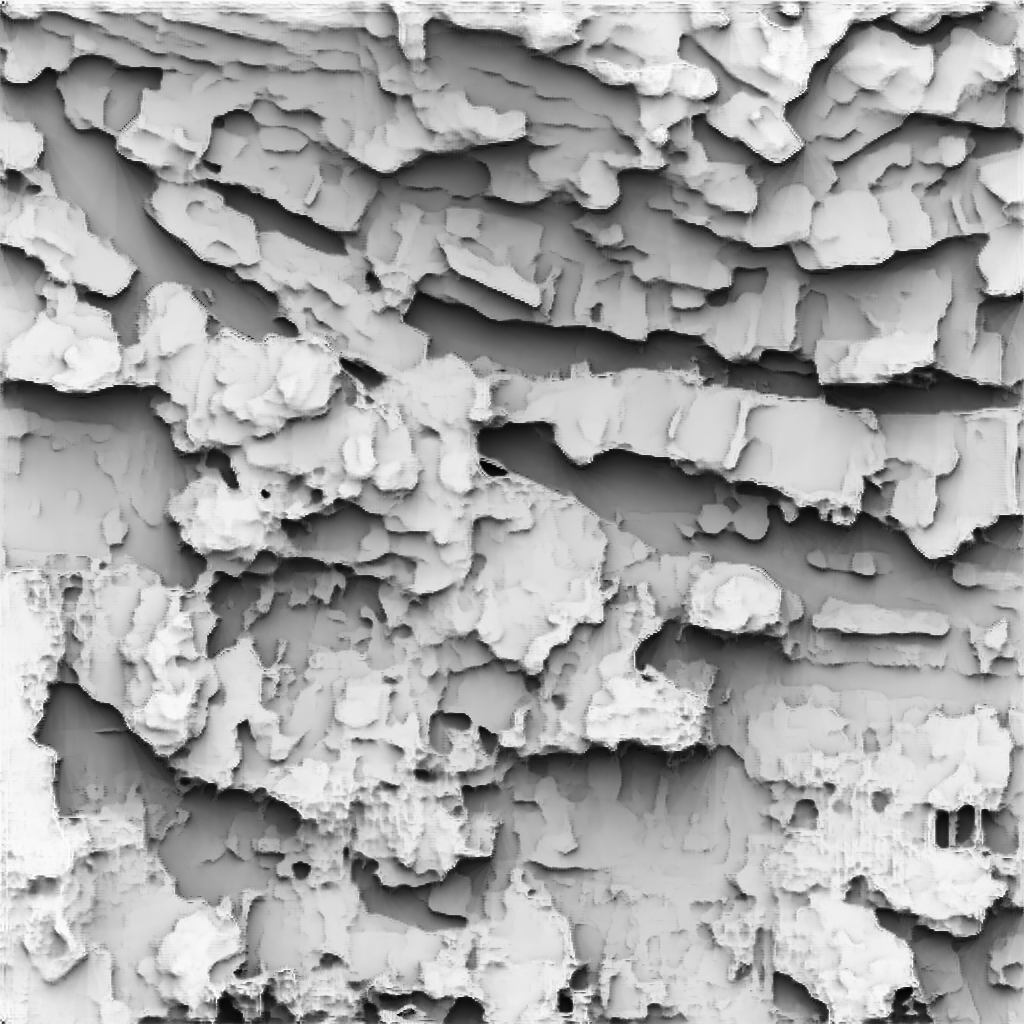}
		\centering{\tiny DeepPruner(KITTI)}
	\end{minipage}	
	\begin{minipage}[t]{0.19\textwidth}	
		\includegraphics[width=0.098\linewidth]{figures_supp/color_map.png}
		\includegraphics[width=0.85\linewidth]{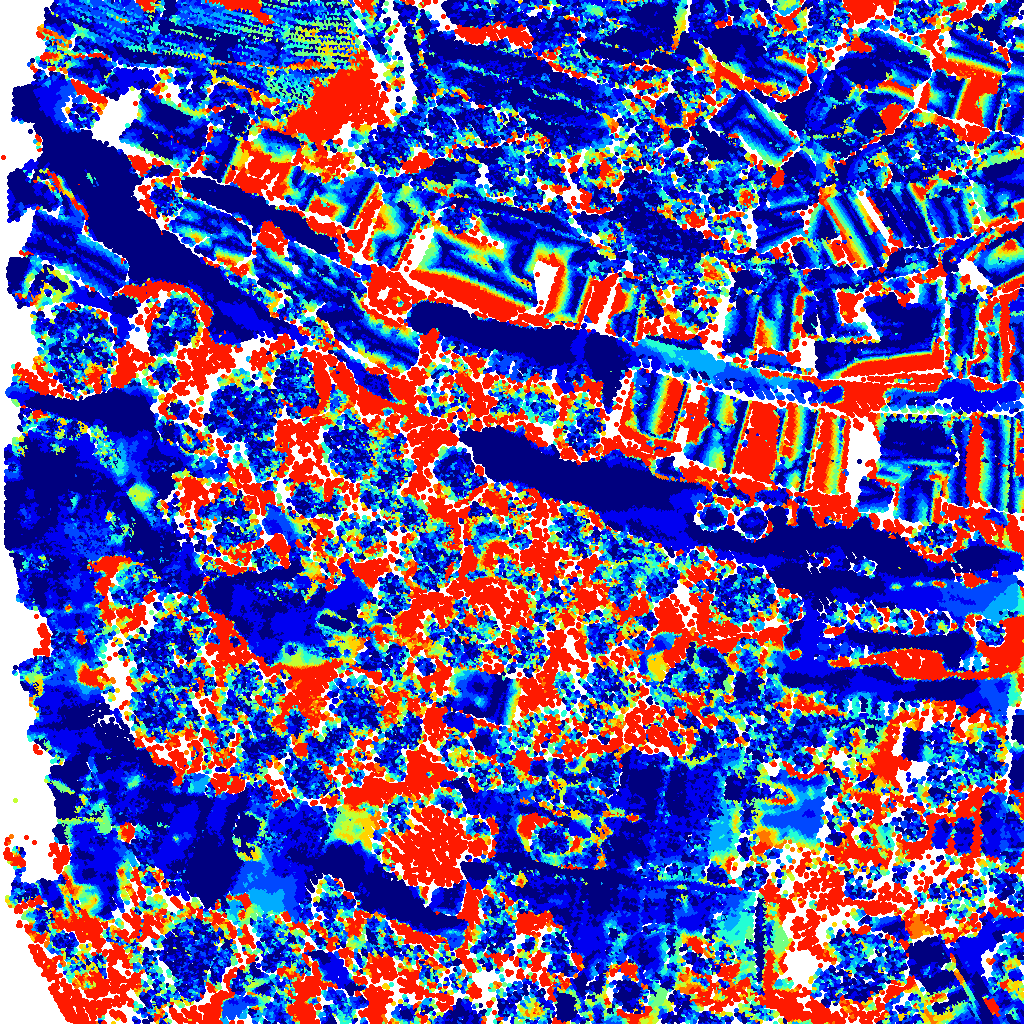}
		\includegraphics[width=\linewidth]{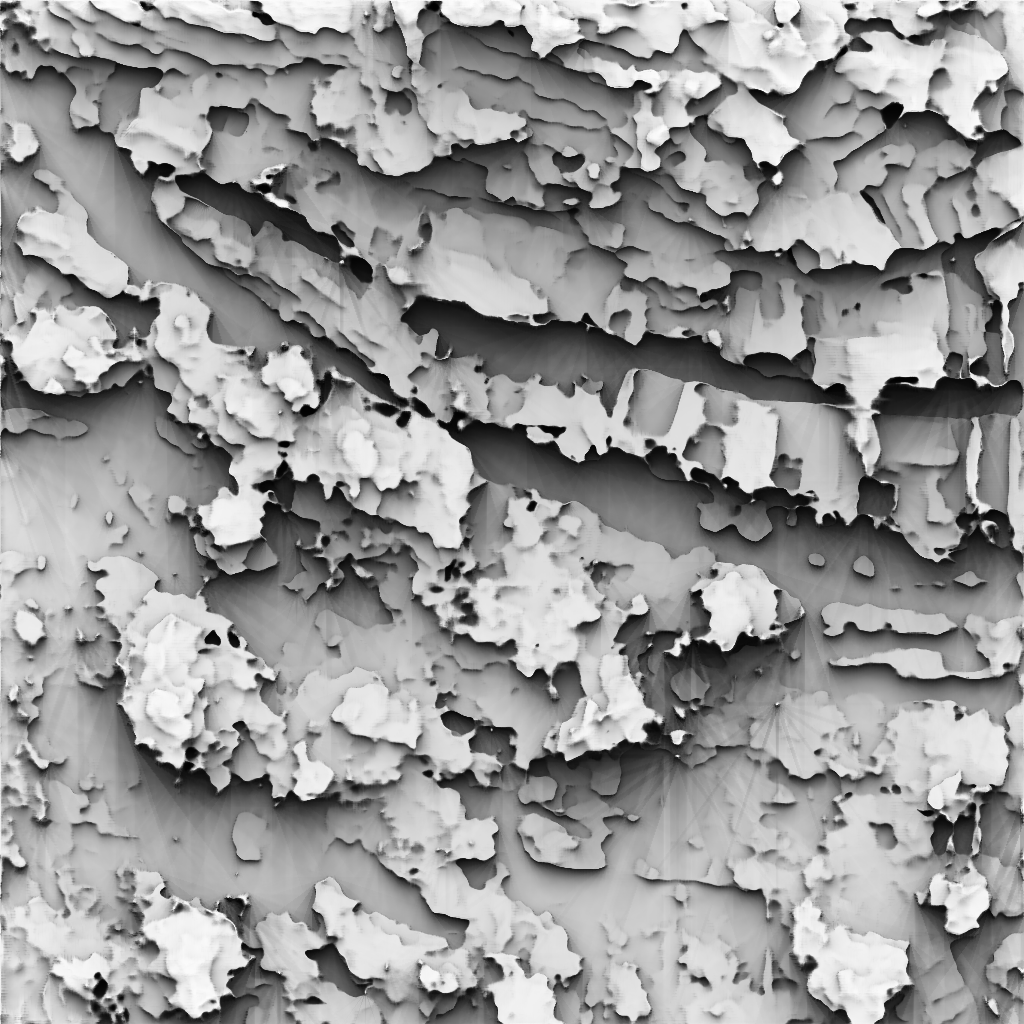}
		\centering{\tiny GANet(KITTI)}
	\end{minipage}	
	\begin{minipage}[t]{0.19\textwidth}	
		\includegraphics[width=0.098\linewidth]{figures_supp/color_map.png}
		\includegraphics[width=0.85\linewidth]{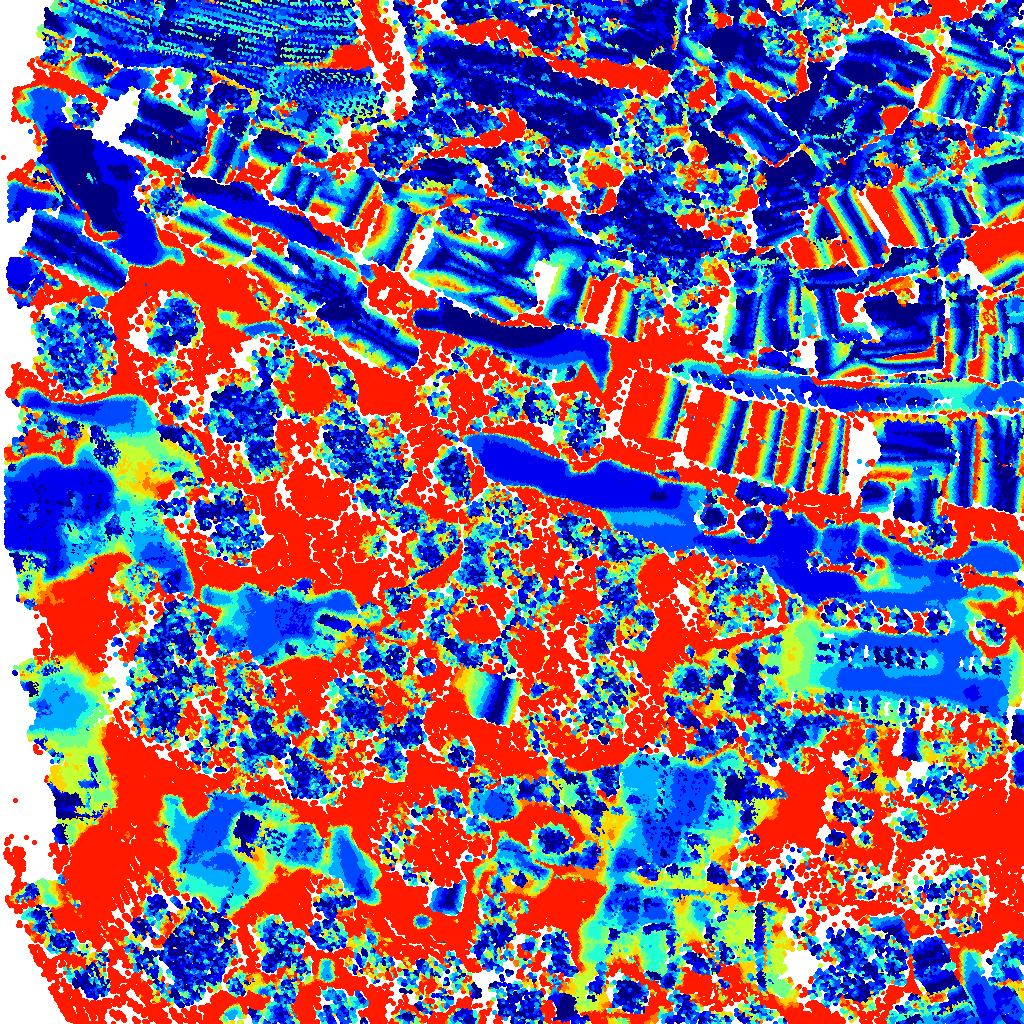}
		\includegraphics[width=\linewidth]{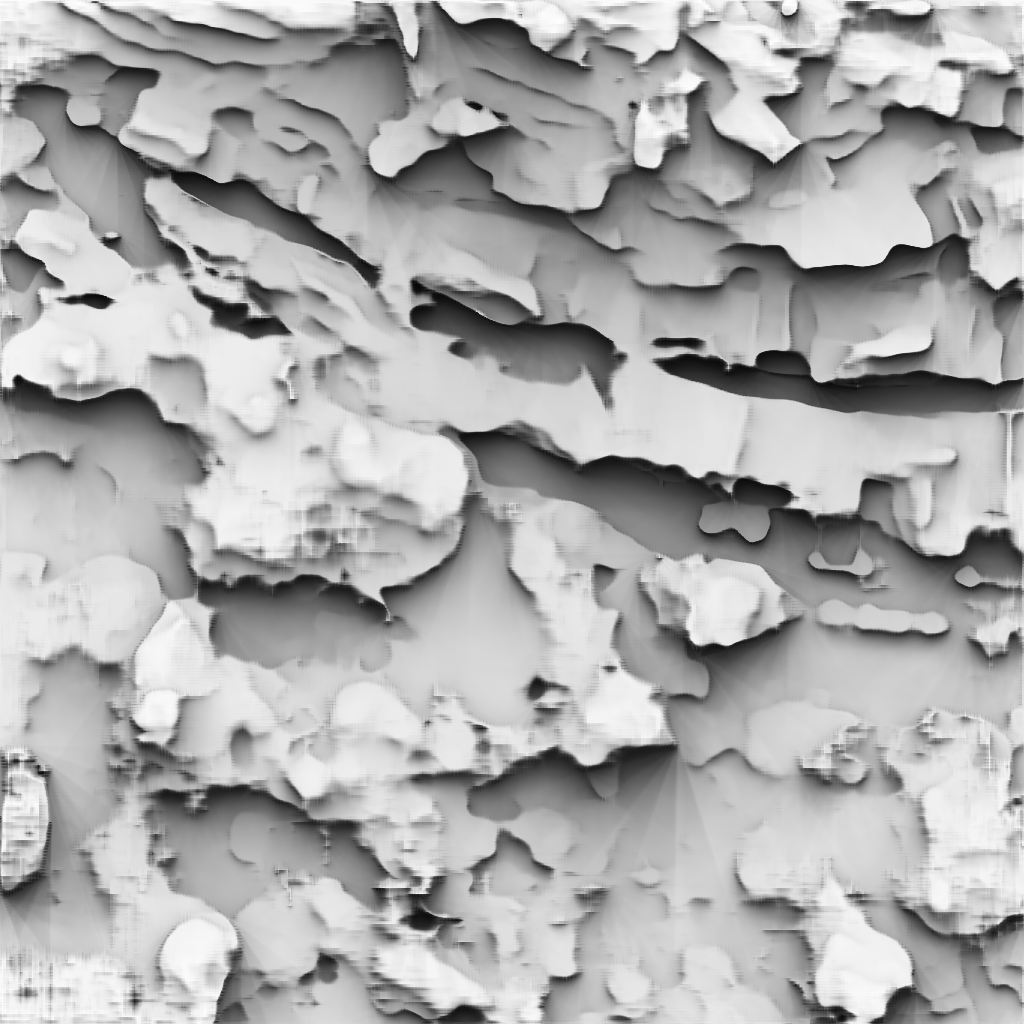}
		\centering{\tiny LEAStereo(KITTI)}
	\end{minipage}
	\begin{minipage}[t]{0.19\textwidth}
		\includegraphics[width=0.098\linewidth]{figures_supp/color_map.png}
		\includegraphics[width=0.85\linewidth]{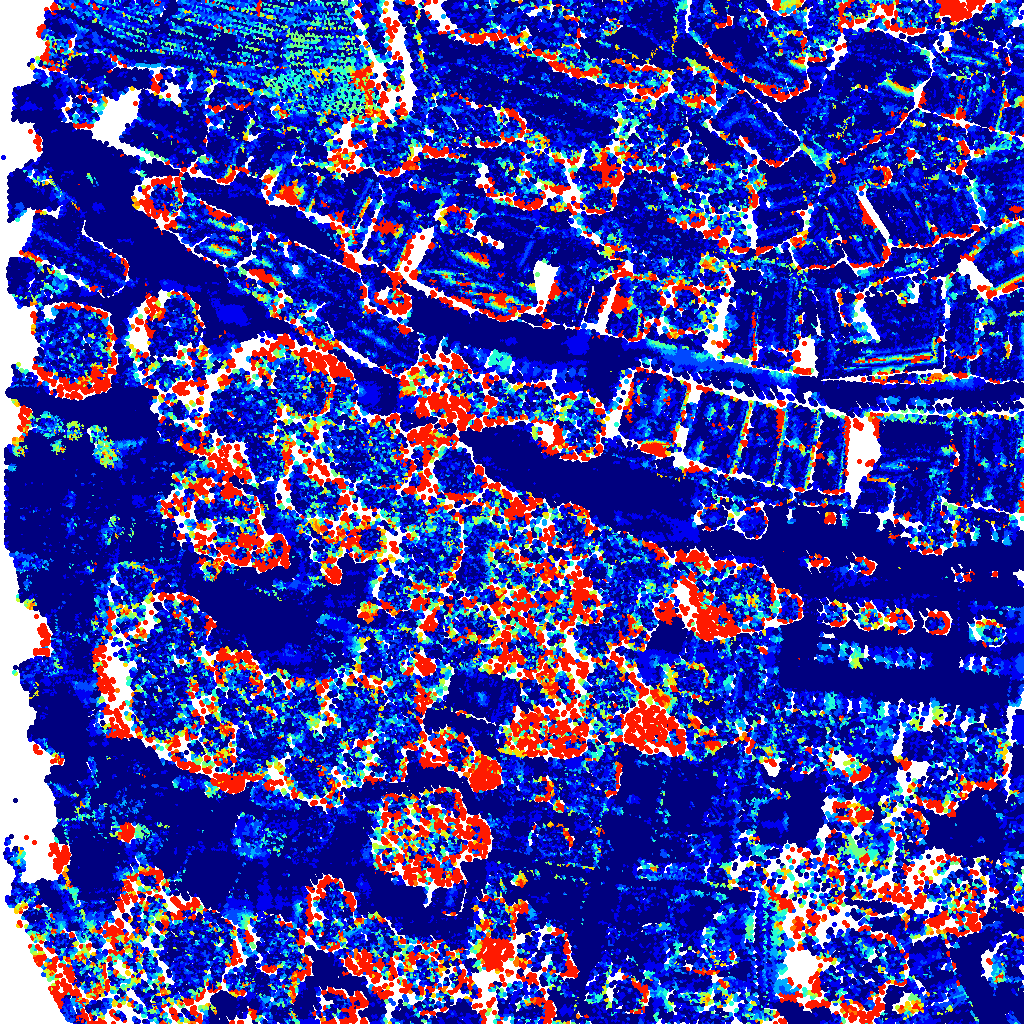} \\
		\includegraphics[width=\linewidth]{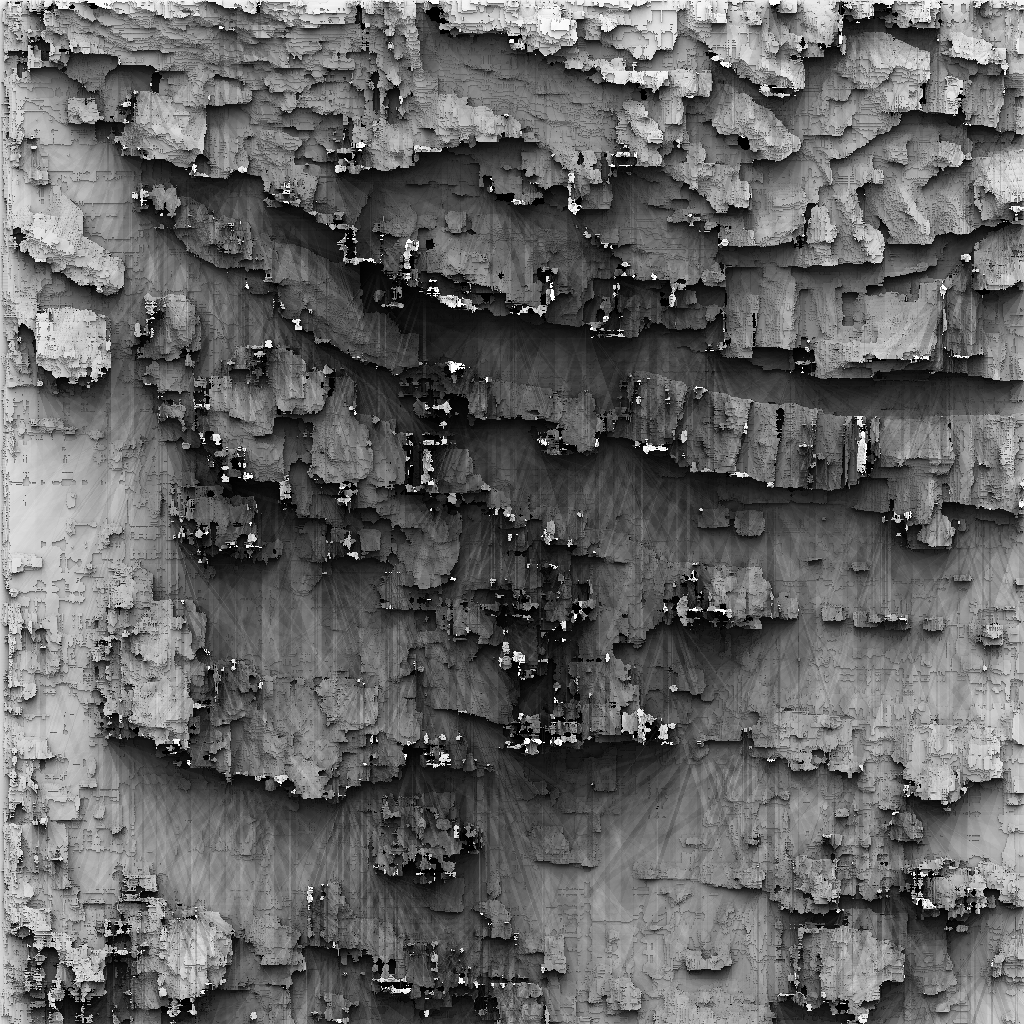}
		\centering{\tiny MC-CNN}
	\end{minipage}
	\begin{minipage}[t]{0.19\textwidth}
		\includegraphics[width=0.098\linewidth]{figures_supp/color_map.png}
		\includegraphics[width=0.85\linewidth]{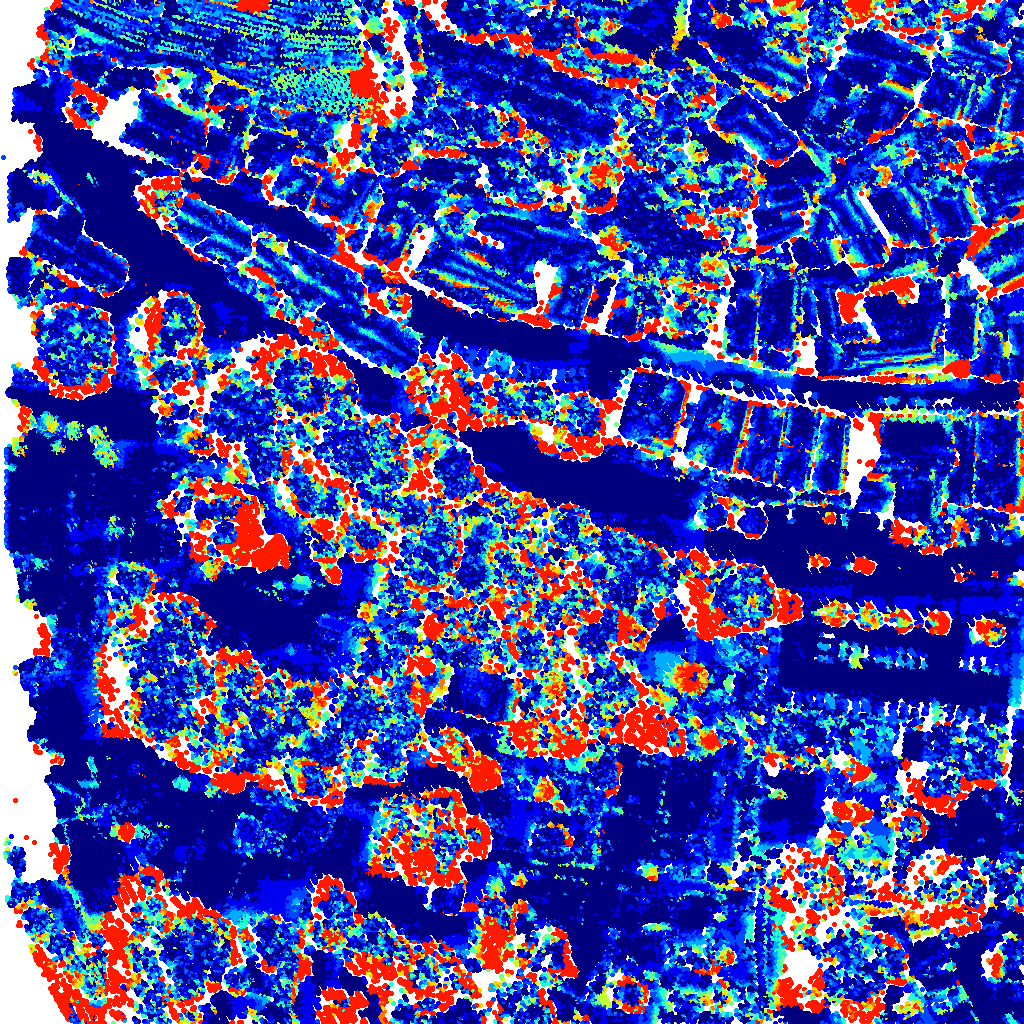}
		\includegraphics[width=\linewidth]{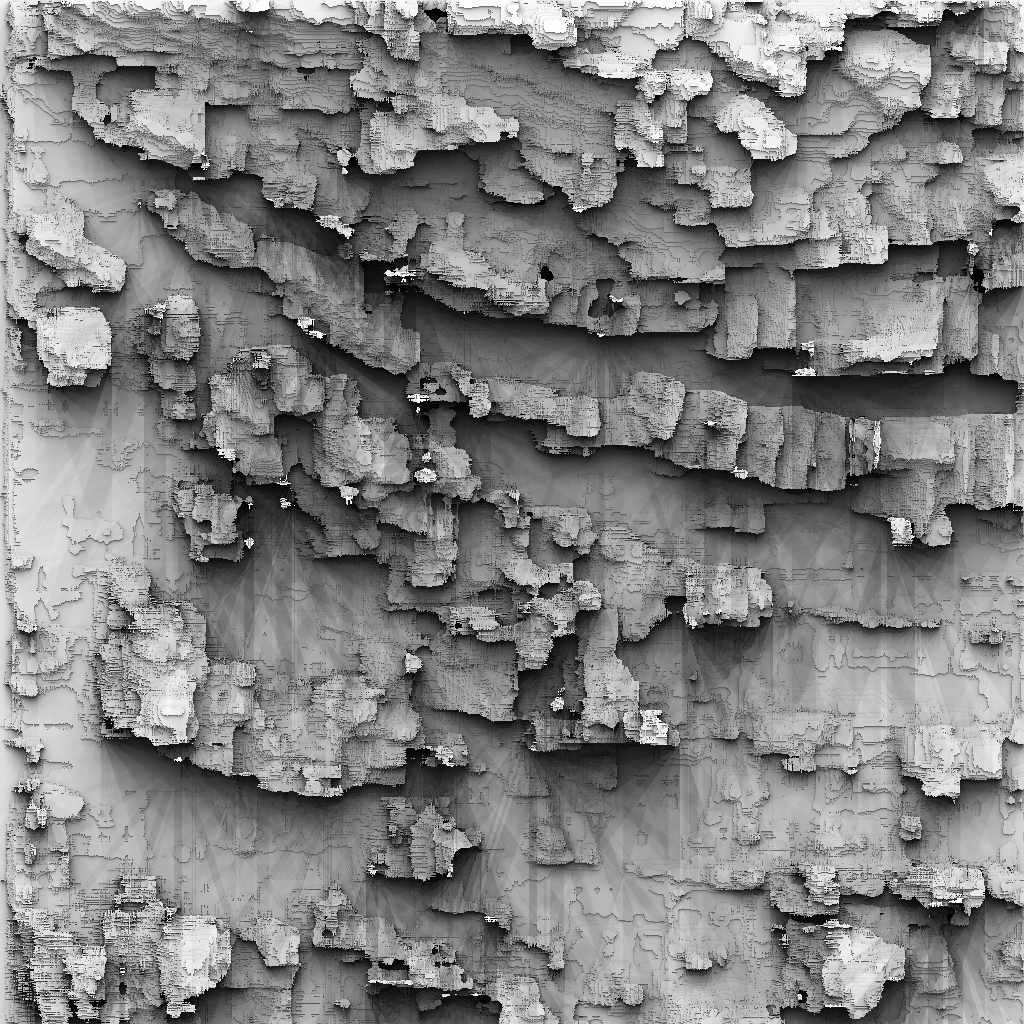}
		\centering{\tiny DeepFeature}
	\end{minipage}
	\begin{minipage}[t]{0.19\textwidth}
		\includegraphics[width=0.098\linewidth]{figures_supp/color_map.png}
		\includegraphics[width=0.85\linewidth]{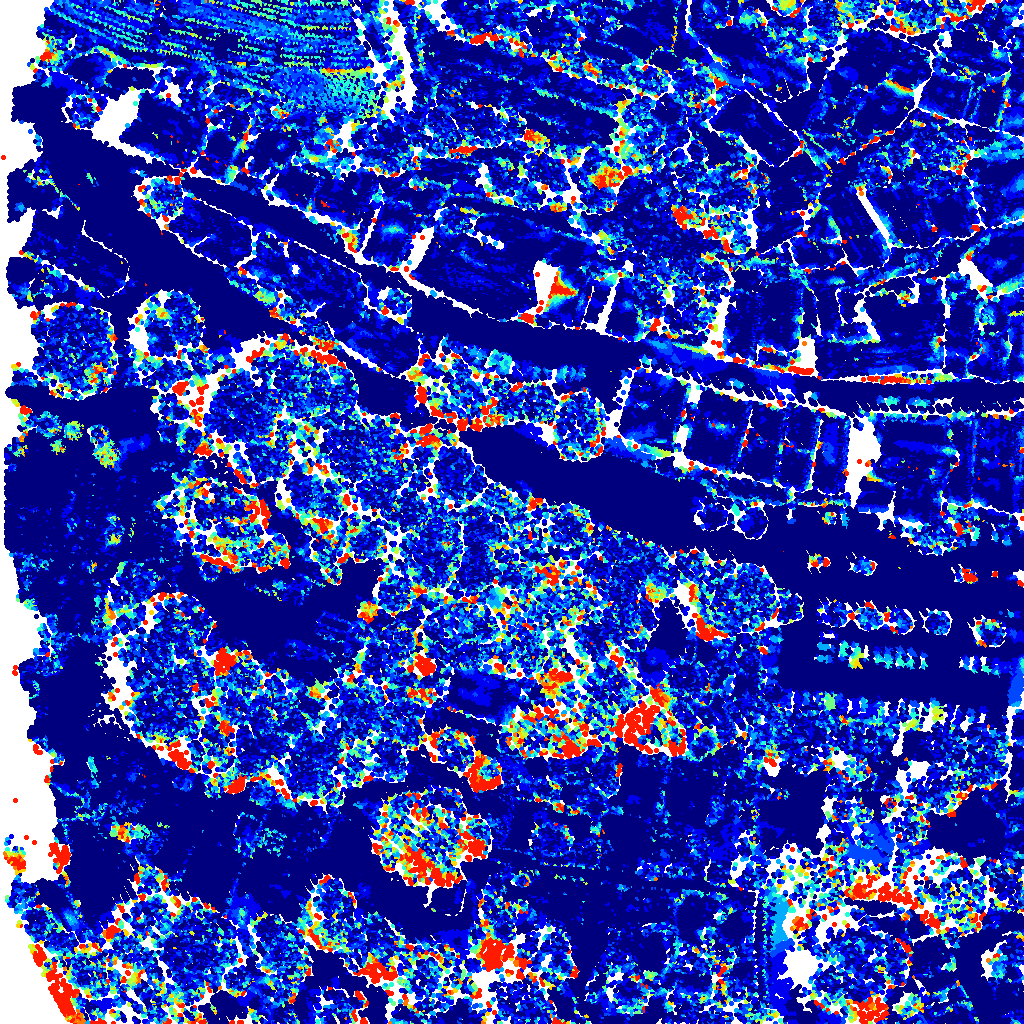}
		\includegraphics[width=\linewidth]{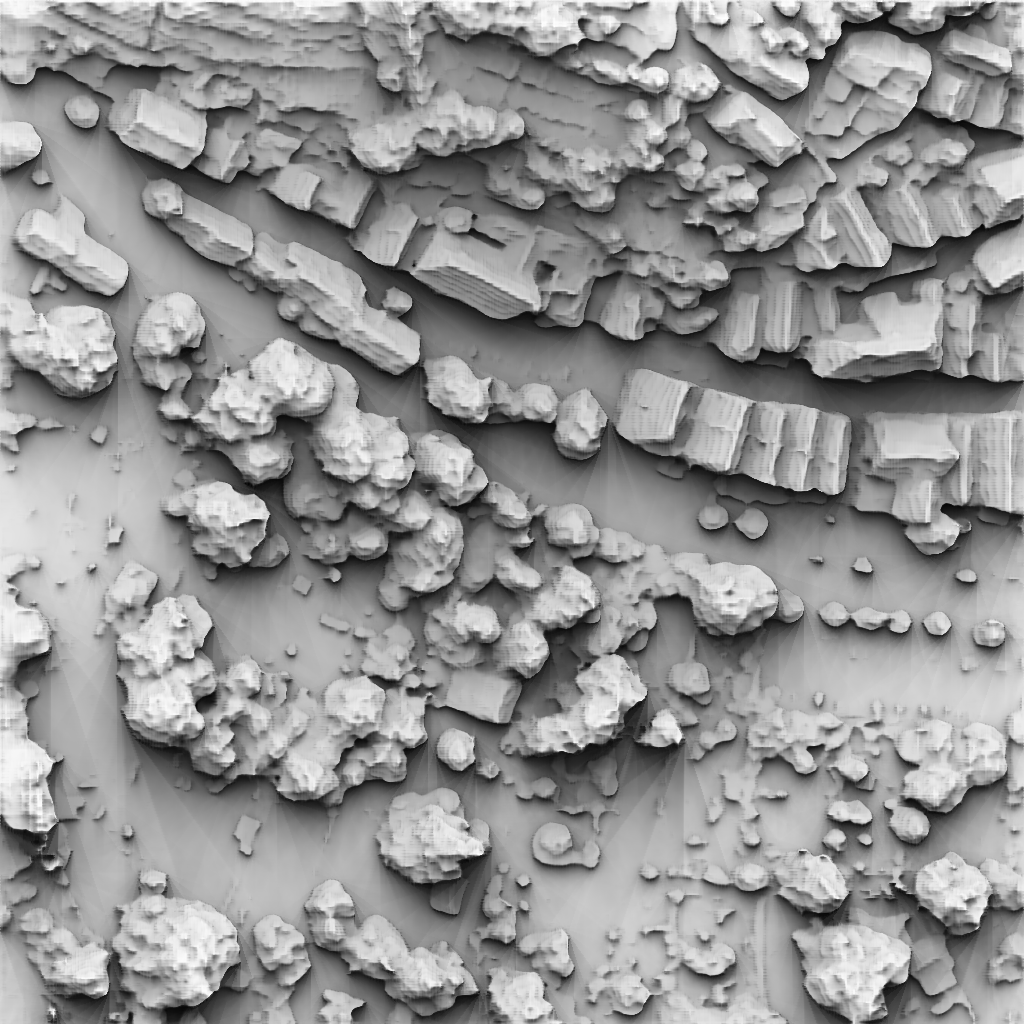}
		\centering{\tiny PSM Net}
	\end{minipage}
	\begin{minipage}[t]{0.19\textwidth}	
		\includegraphics[width=0.098\linewidth]{figures_supp/color_map.png}
		\includegraphics[width=0.85\linewidth]{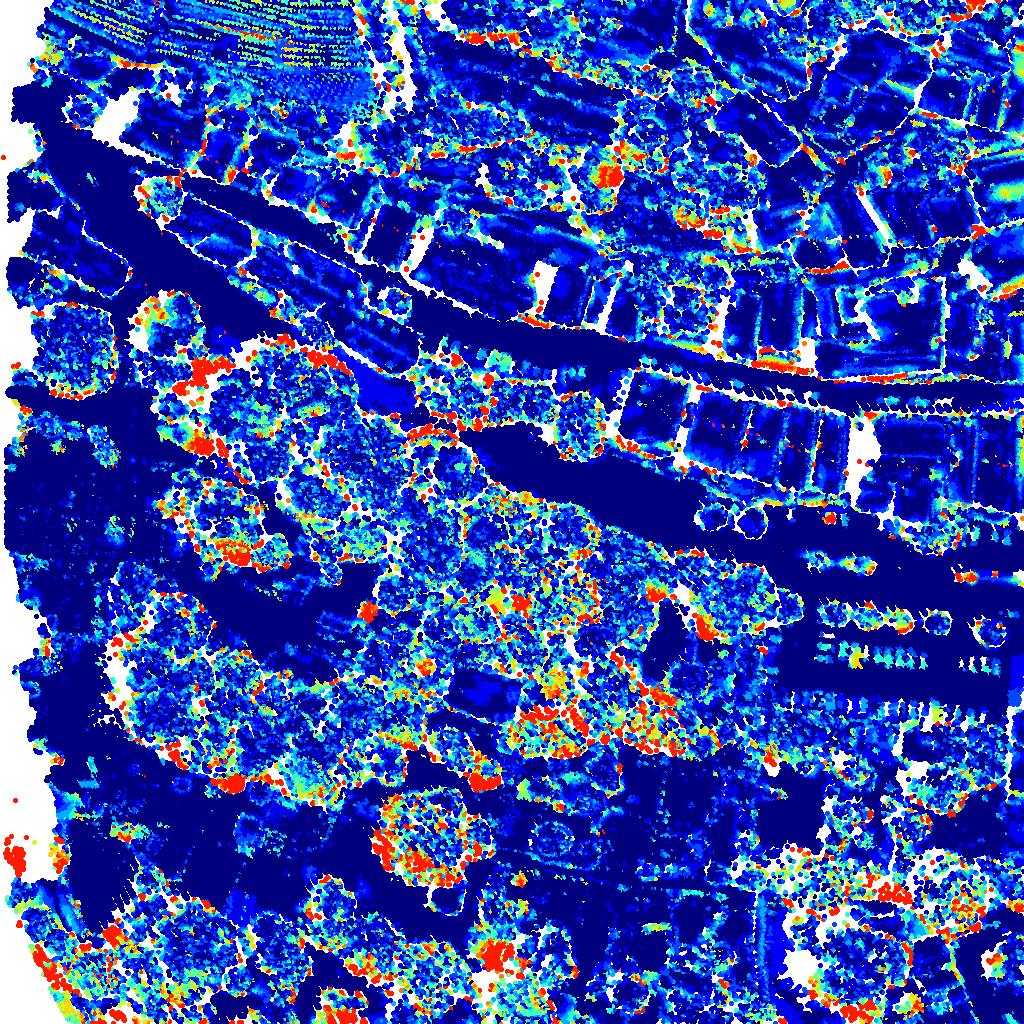} \\
		\includegraphics[width=\linewidth]{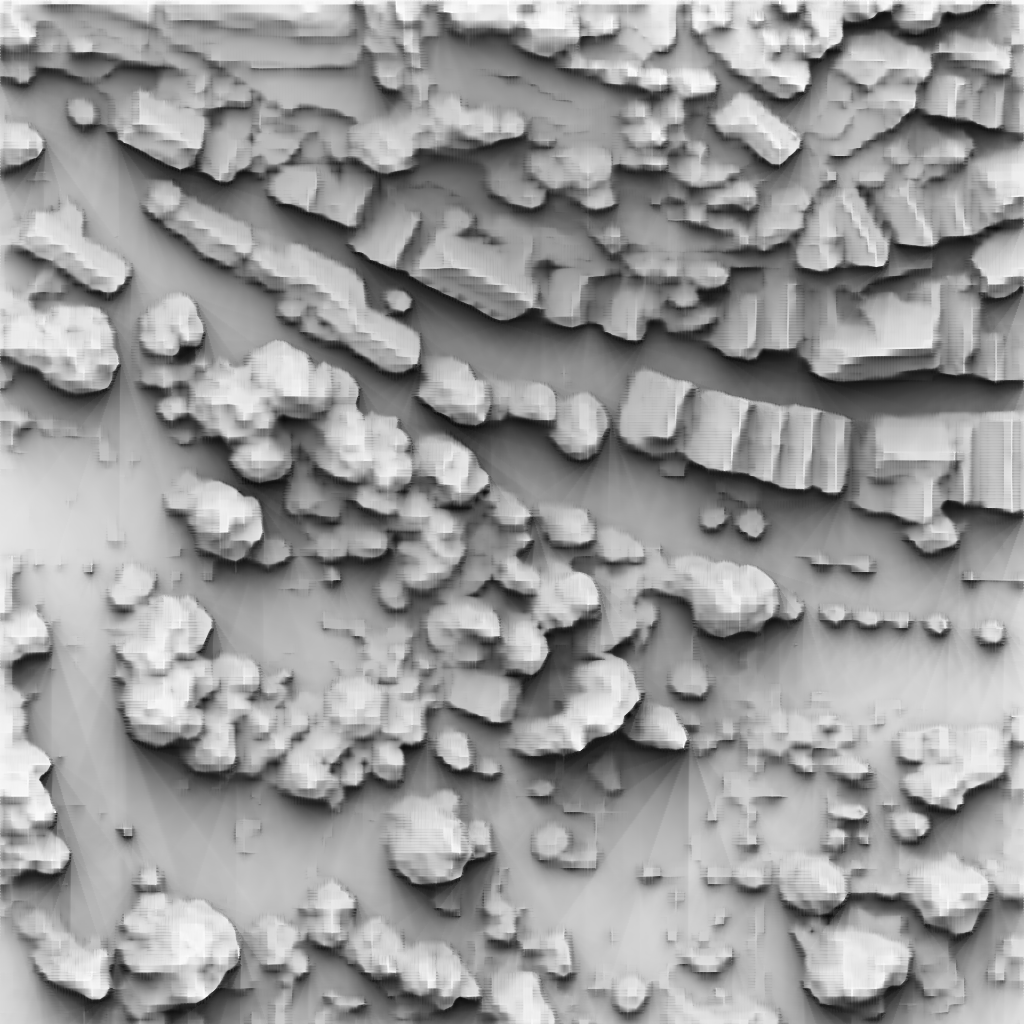}
		\centering{\tiny HRS Net}
	\end{minipage}	
	\begin{minipage}[t]{0.19\textwidth}	
		\includegraphics[width=0.098\linewidth]{figures_supp/color_map.png}
		\includegraphics[width=0.85\linewidth]{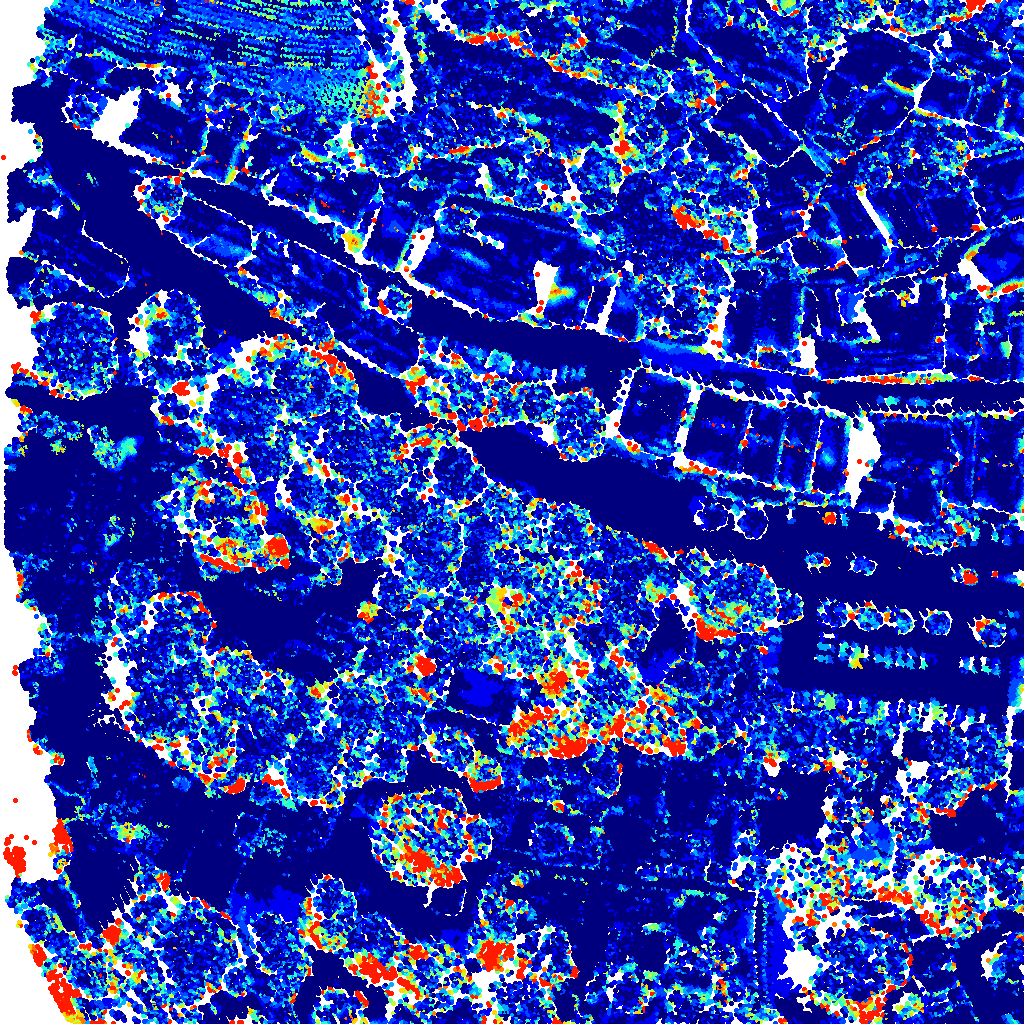}
		\includegraphics[width=\linewidth]{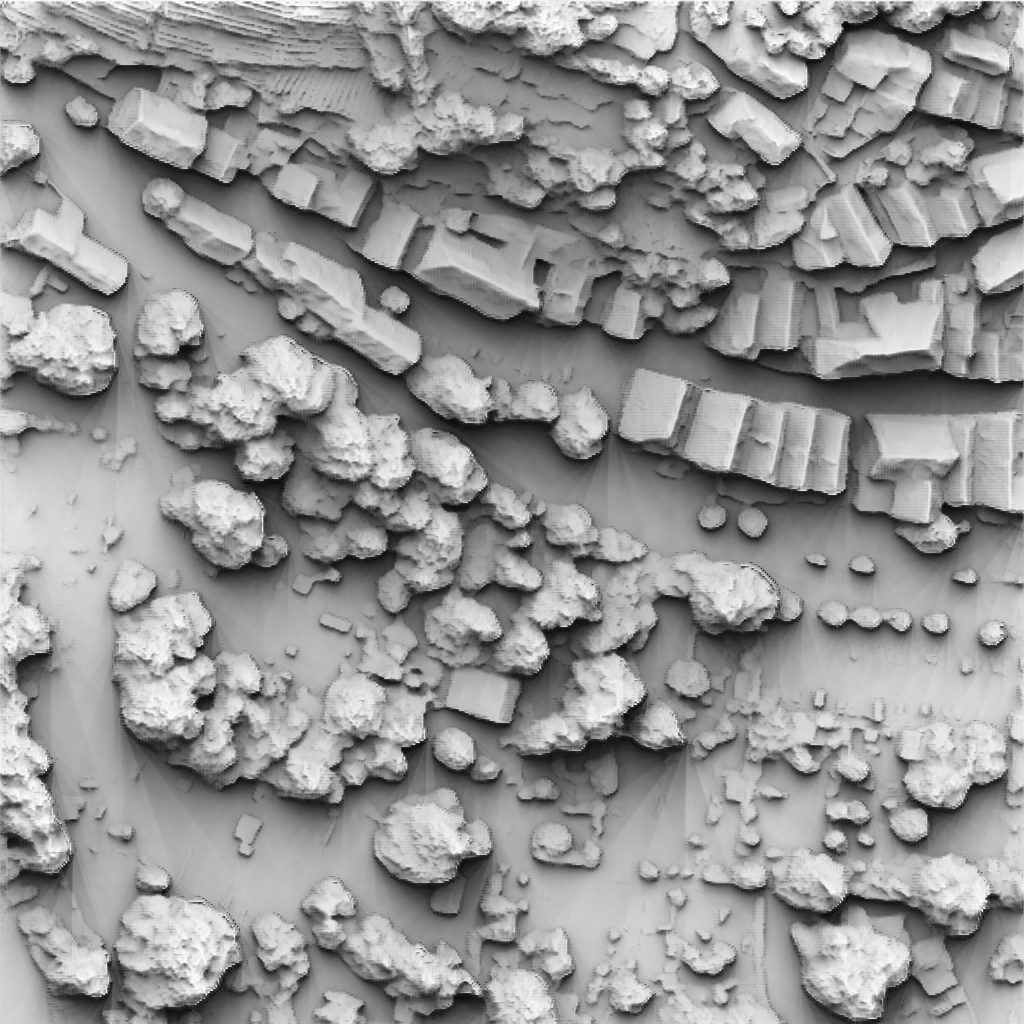}
		\centering{\tiny DeepPruner}
	\end{minipage}	
	\begin{minipage}[t]{0.19\textwidth}	
		\includegraphics[width=0.098\linewidth]{figures_supp/color_map.png}
		\includegraphics[width=0.85\linewidth]{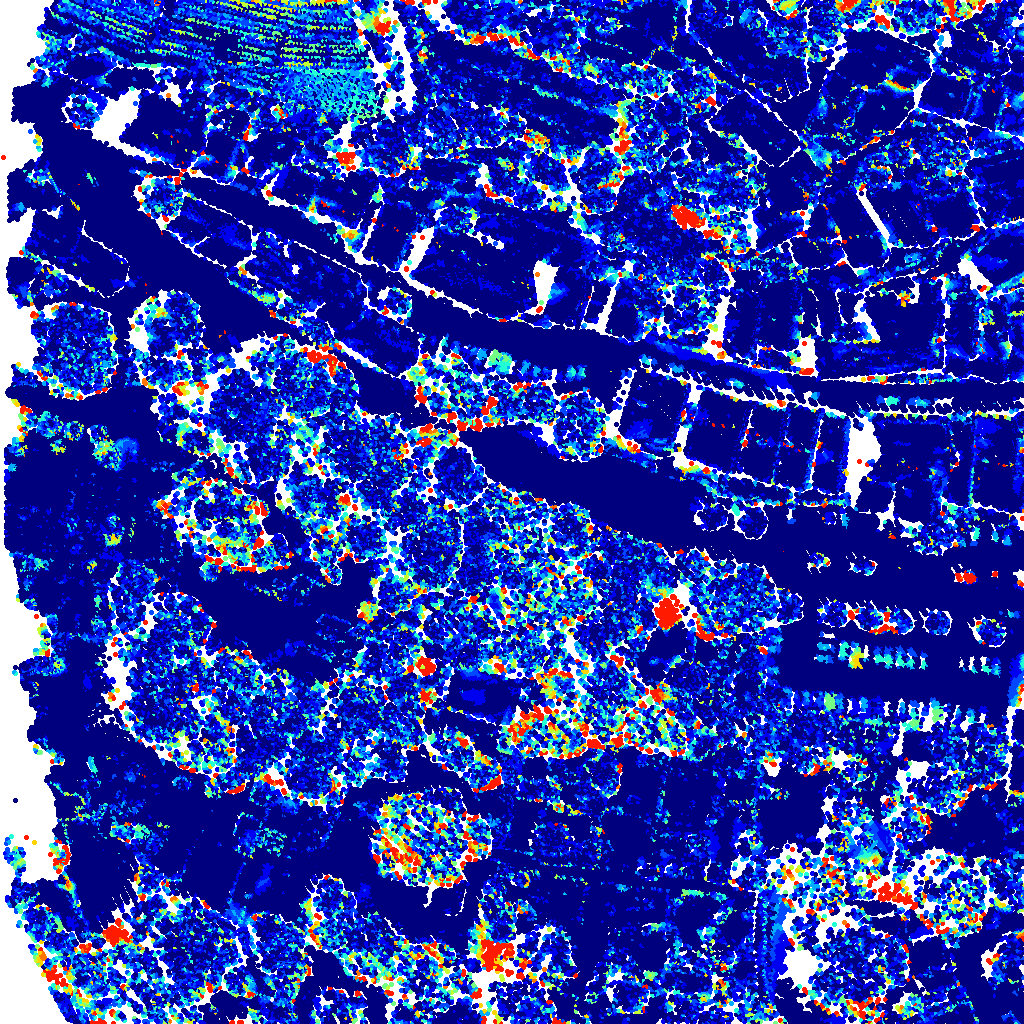}
		\includegraphics[width=\linewidth]{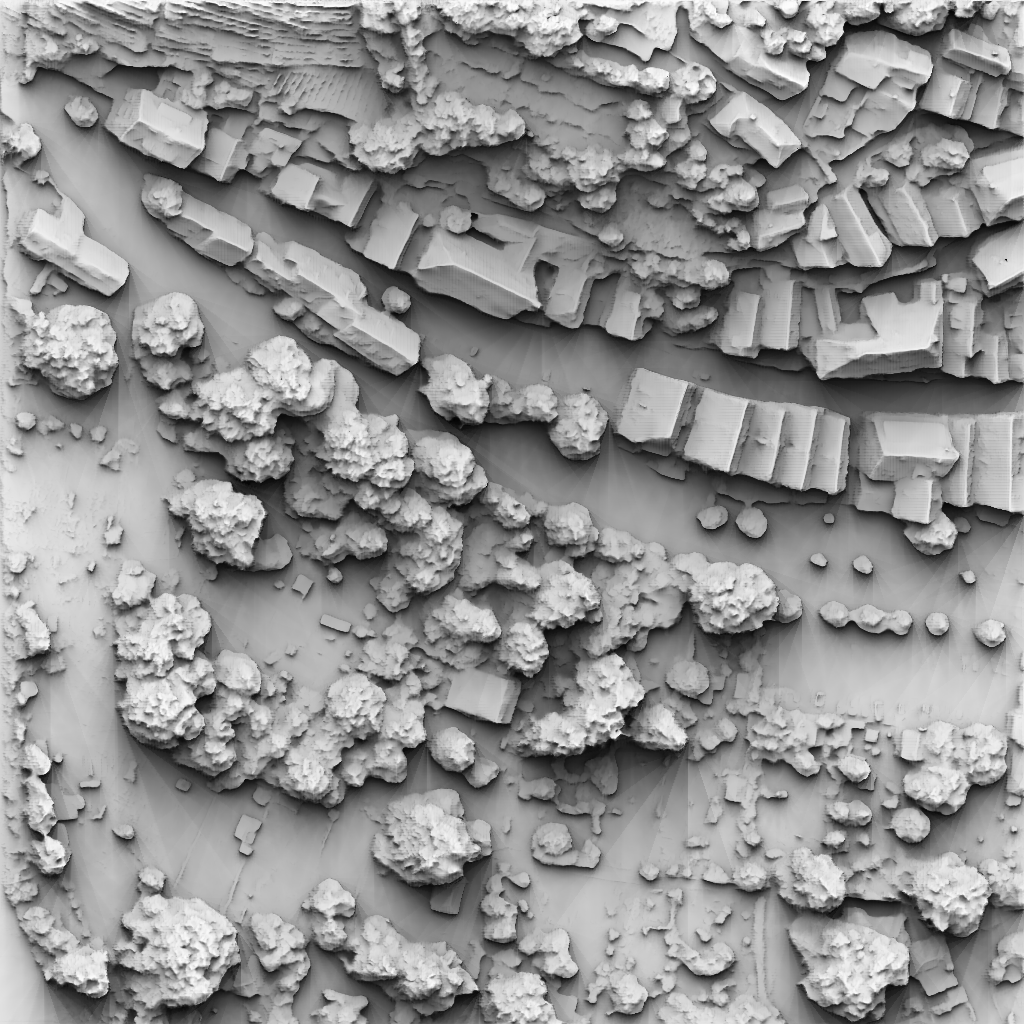}
		\centering{\tiny GANet}
	\end{minipage}	
	\begin{minipage}[t]{0.19\textwidth}	
		\includegraphics[width=0.098\linewidth]{figures_supp/color_map.png}
		\includegraphics[width=0.85\linewidth]{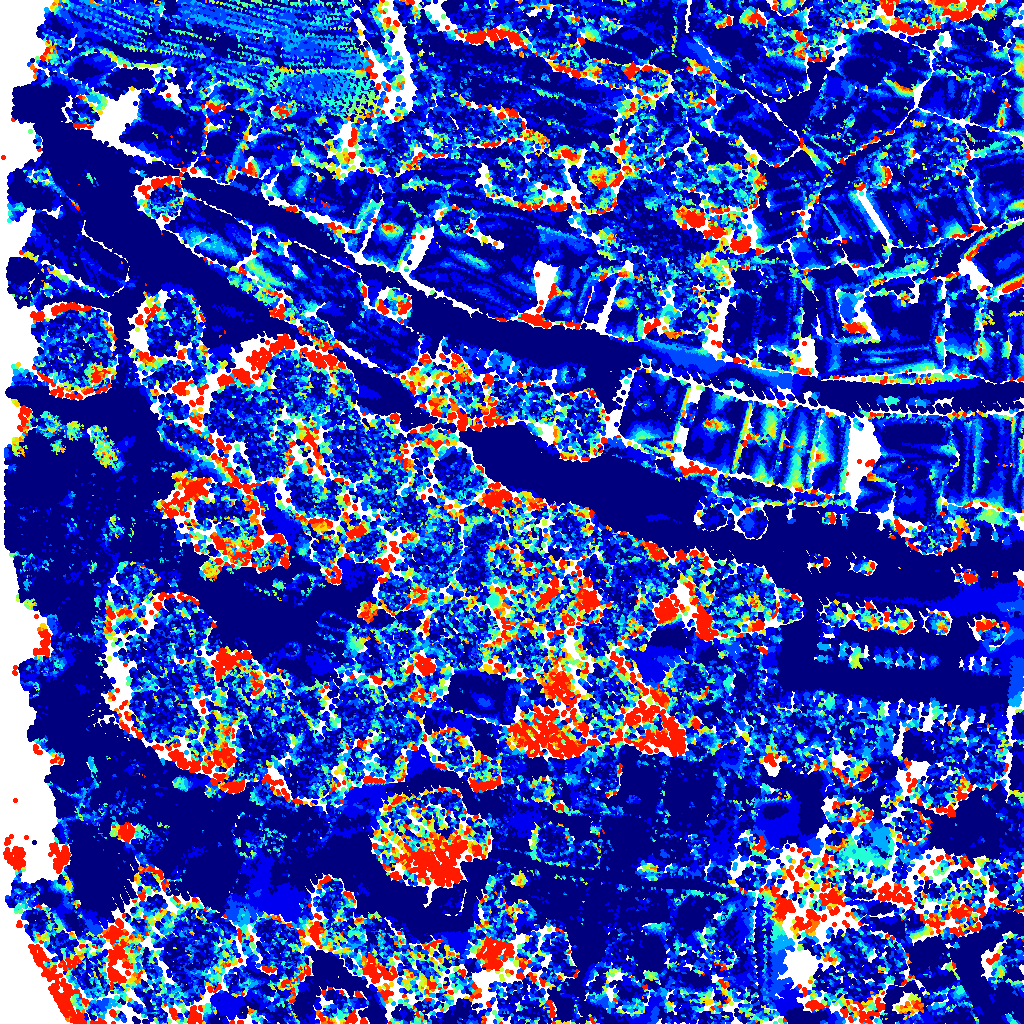}
		\includegraphics[width=\linewidth]{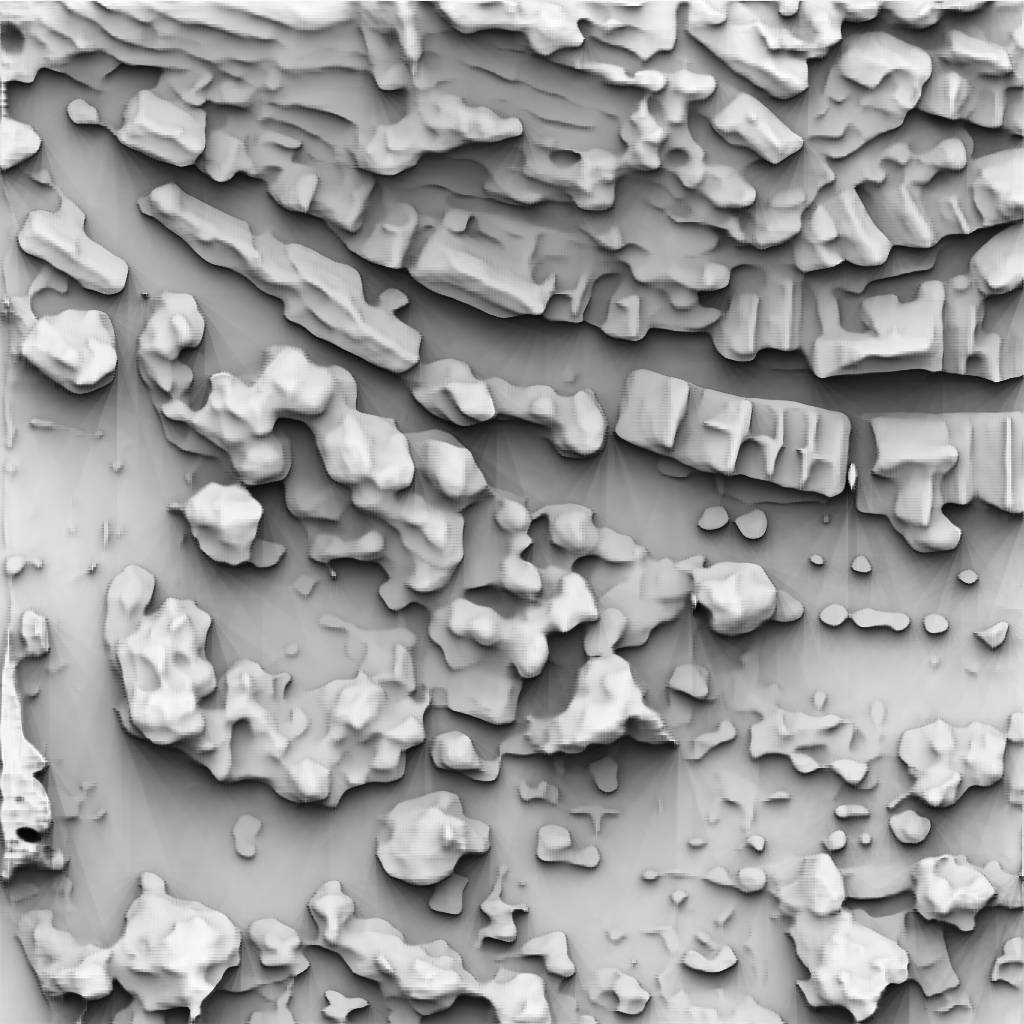}
		\centering{\tiny LEAStereo}
	\end{minipage}	
	\caption{Error map and disparity visualization on tree area for EuroSDR Vaihingen.}
	\label{Figure.eurosdr_tree}
\end{figure}

Reconstruction of building contours (cf. \Cref{Figure.eurosdr_bulding}) or trees (cf. \Cref{Figure.eurosdr_tree}) from models pre-trained on KITTI performs poorly. We believe this is due to the large discontinuities present in this dataset. Fine-tuning significantly improves the results.

%



\begin{figure}[tp]
	\begin{minipage}[t]{0.19\textwidth}
		\includegraphics[width=0.098\linewidth]{figures_supp/color_map.png}
		\includegraphics[width=0.85\linewidth]{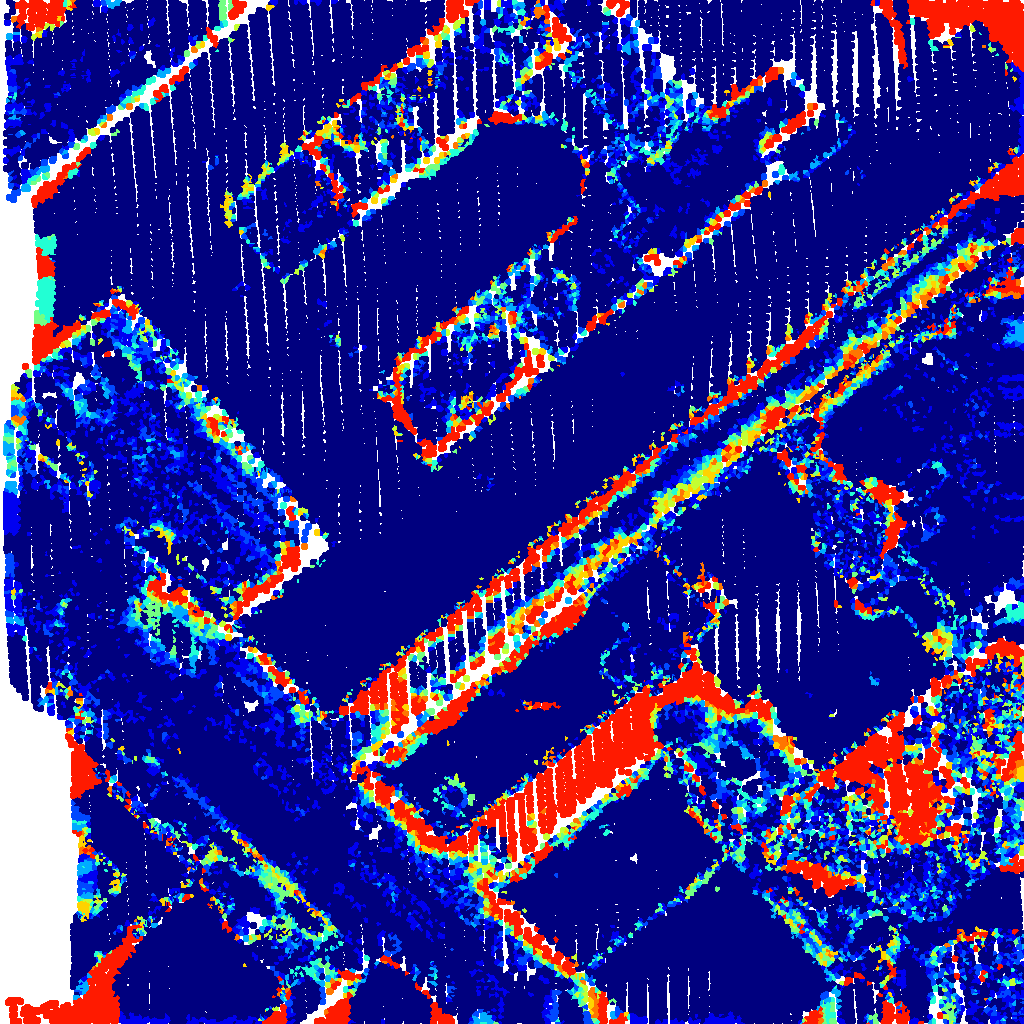}
		\includegraphics[width=\linewidth]{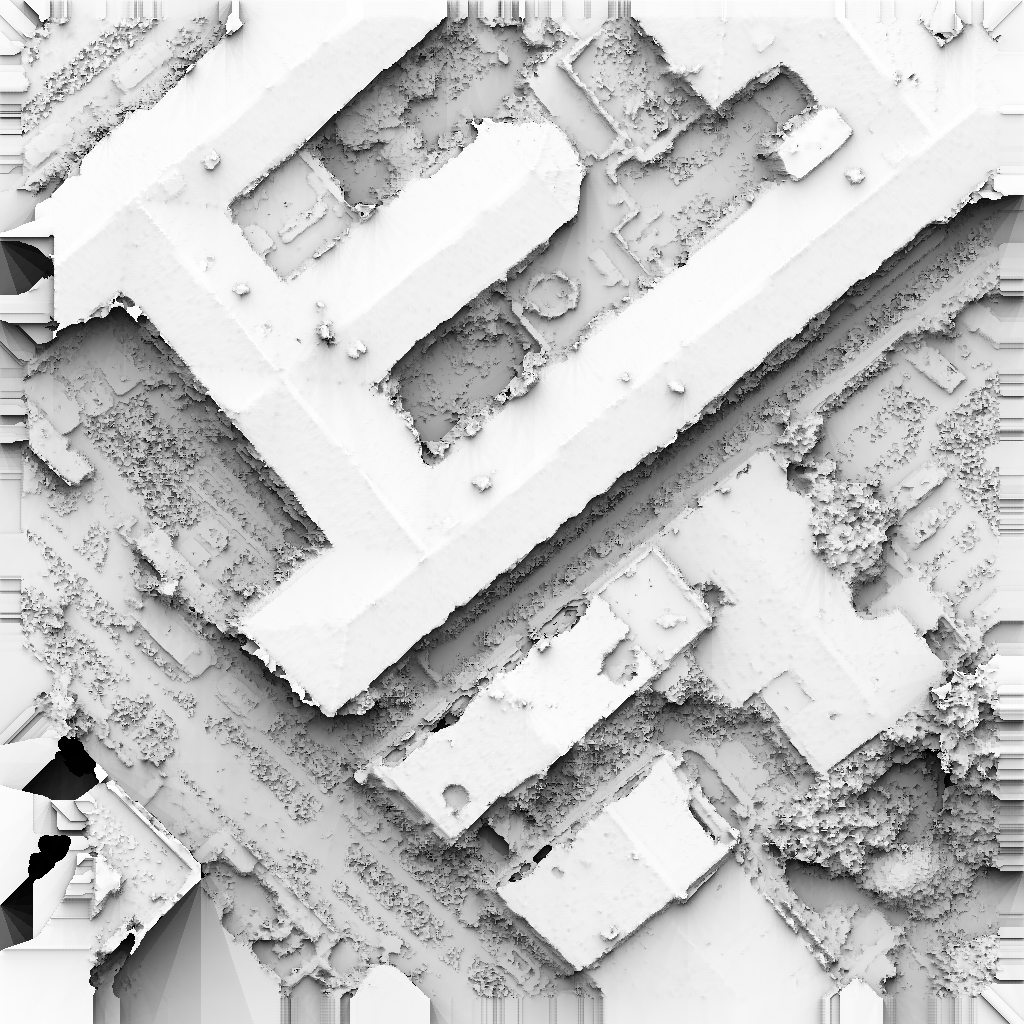}
		\centering{\tiny MICMAC}
	\end{minipage}
	\begin{minipage}[t]{0.19\textwidth}
		\includegraphics[width=0.098\linewidth]{figures_supp/color_map.png}
		\includegraphics[width=0.85\linewidth]{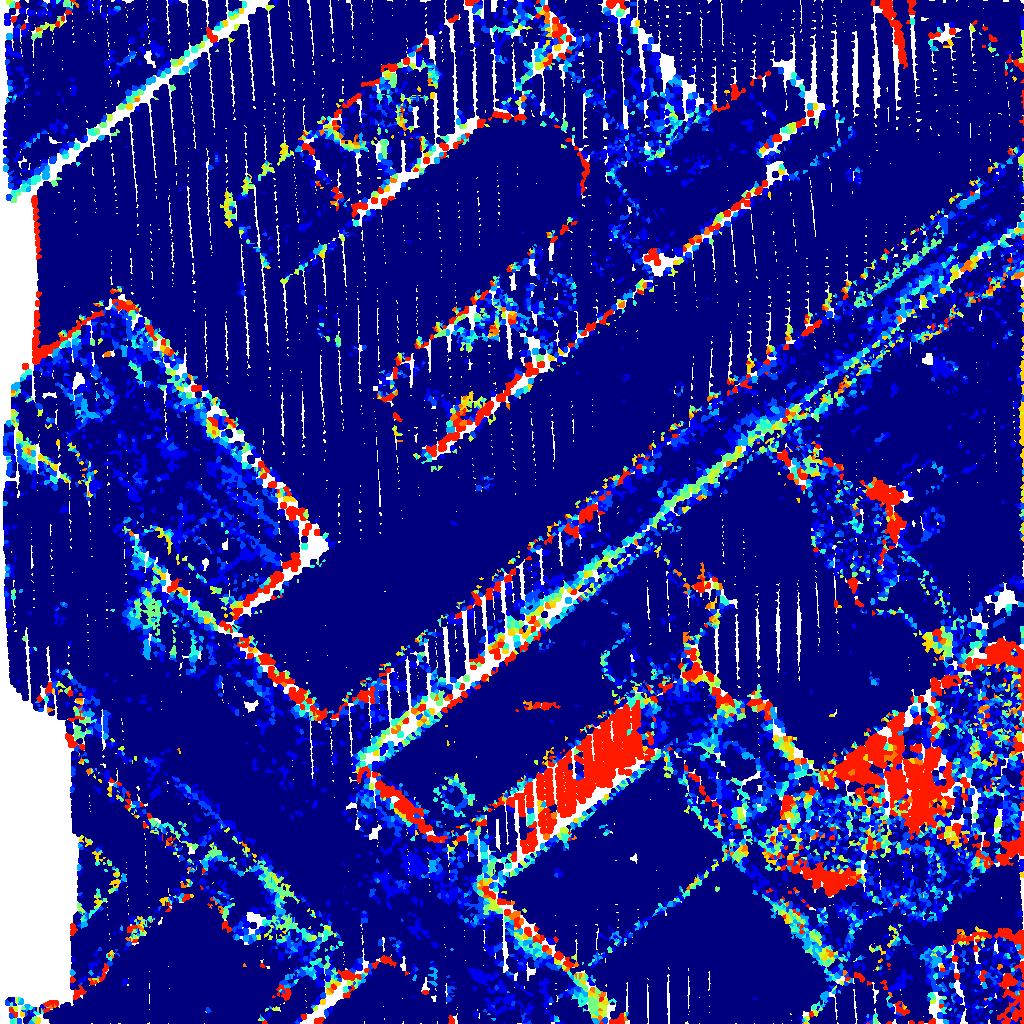}
		\includegraphics[width=\linewidth]{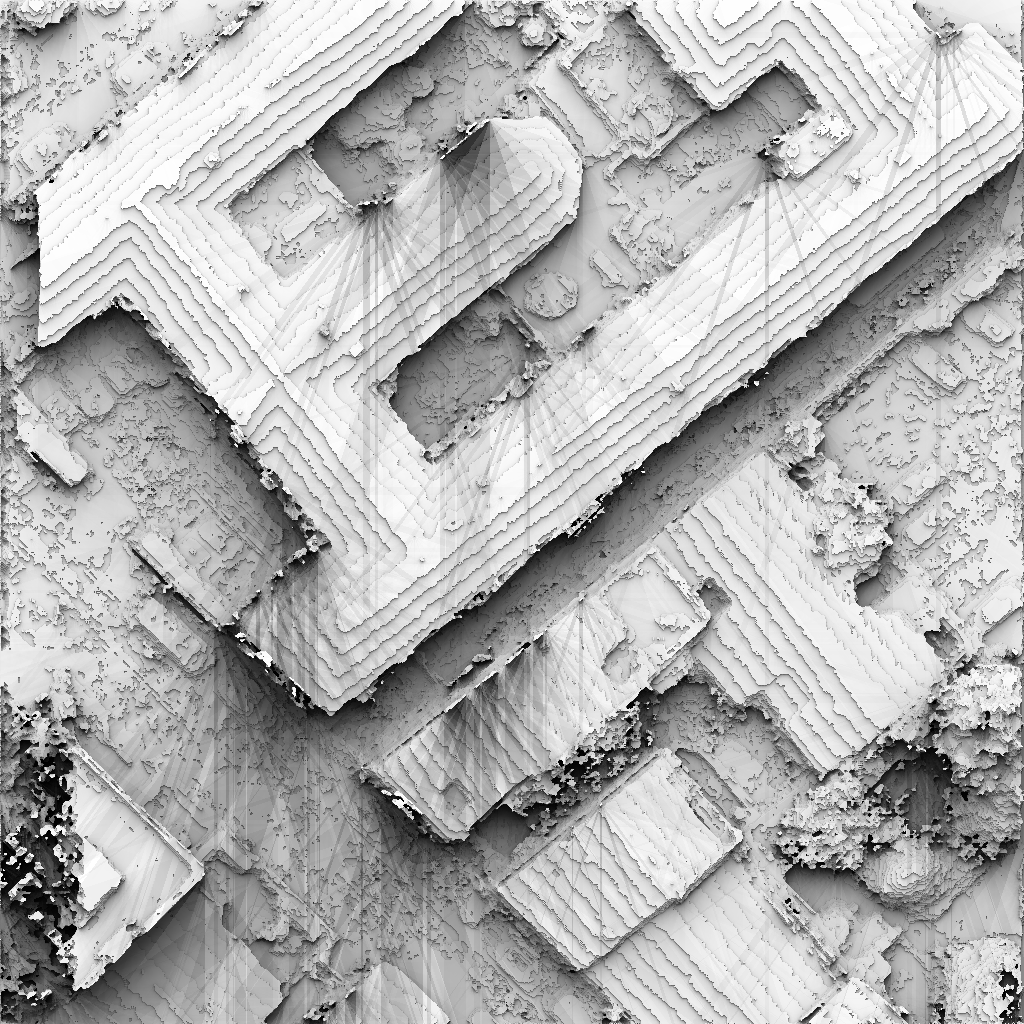}
		\centering{\tiny SGM(CUDA)}
	\end{minipage}
	\begin{minipage}[t]{0.19\textwidth}
		\includegraphics[width=0.098\linewidth]{figures_supp/color_map.png}
		\includegraphics[width=0.85\linewidth]{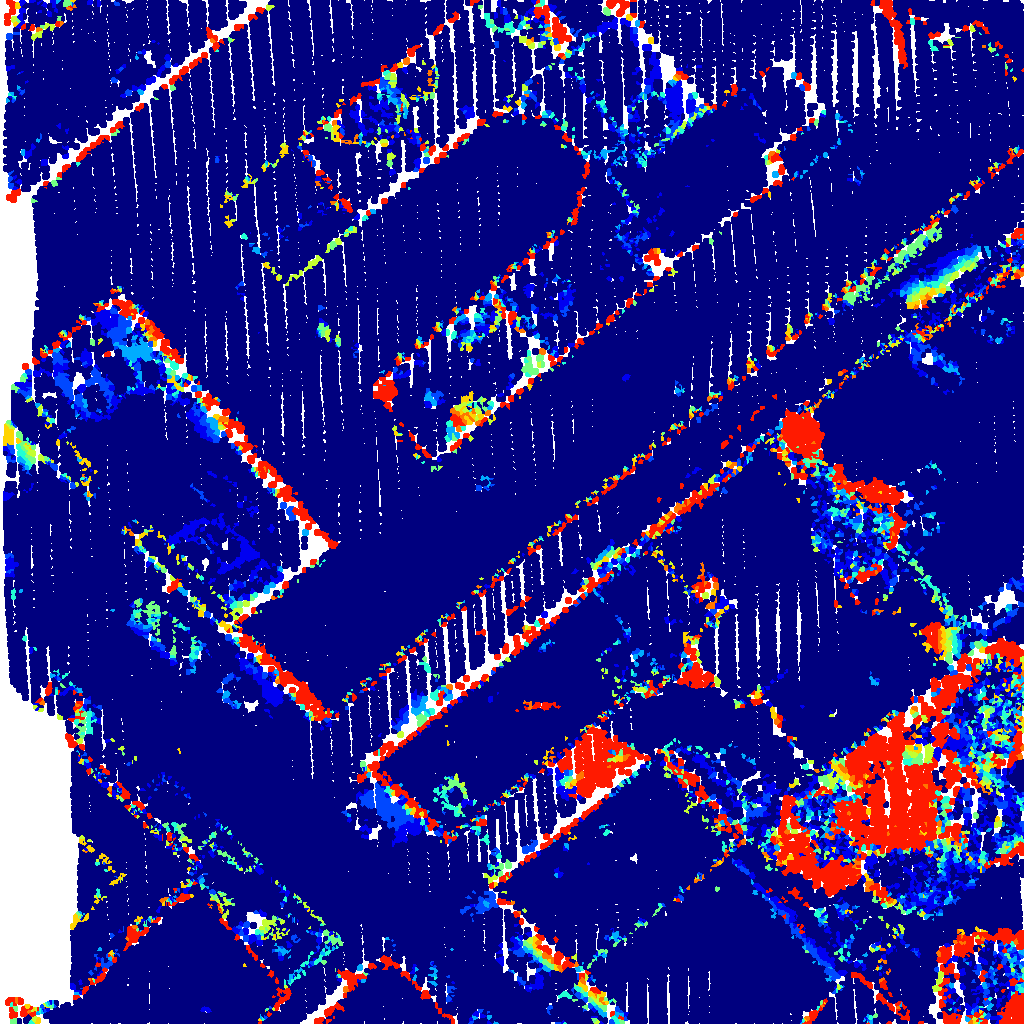}
		\includegraphics[width=\linewidth]{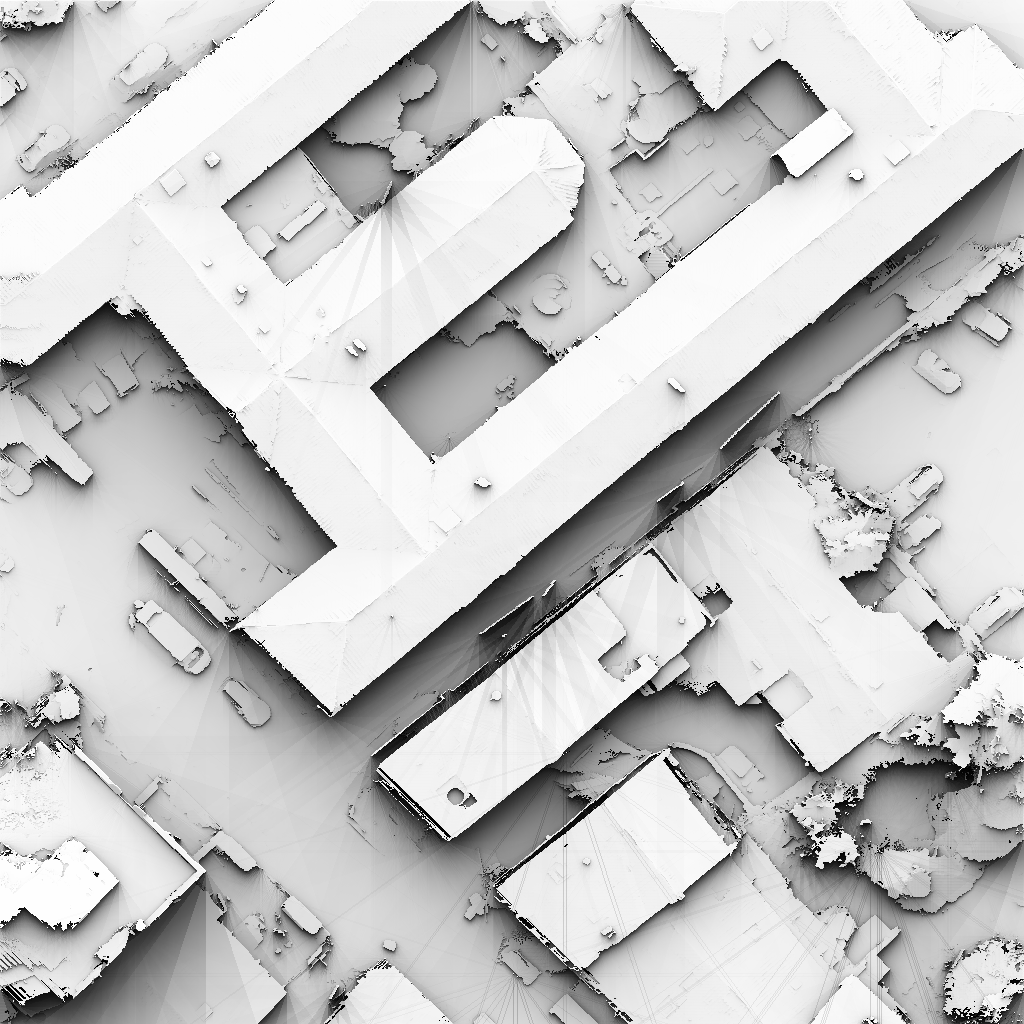}
		\centering{\tiny GraphCuts}
	\end{minipage}
	\begin{minipage}[t]{0.19\textwidth}
		\includegraphics[width=0.098\linewidth]{figures_supp/color_map.png}
		\includegraphics[width=0.85\linewidth]{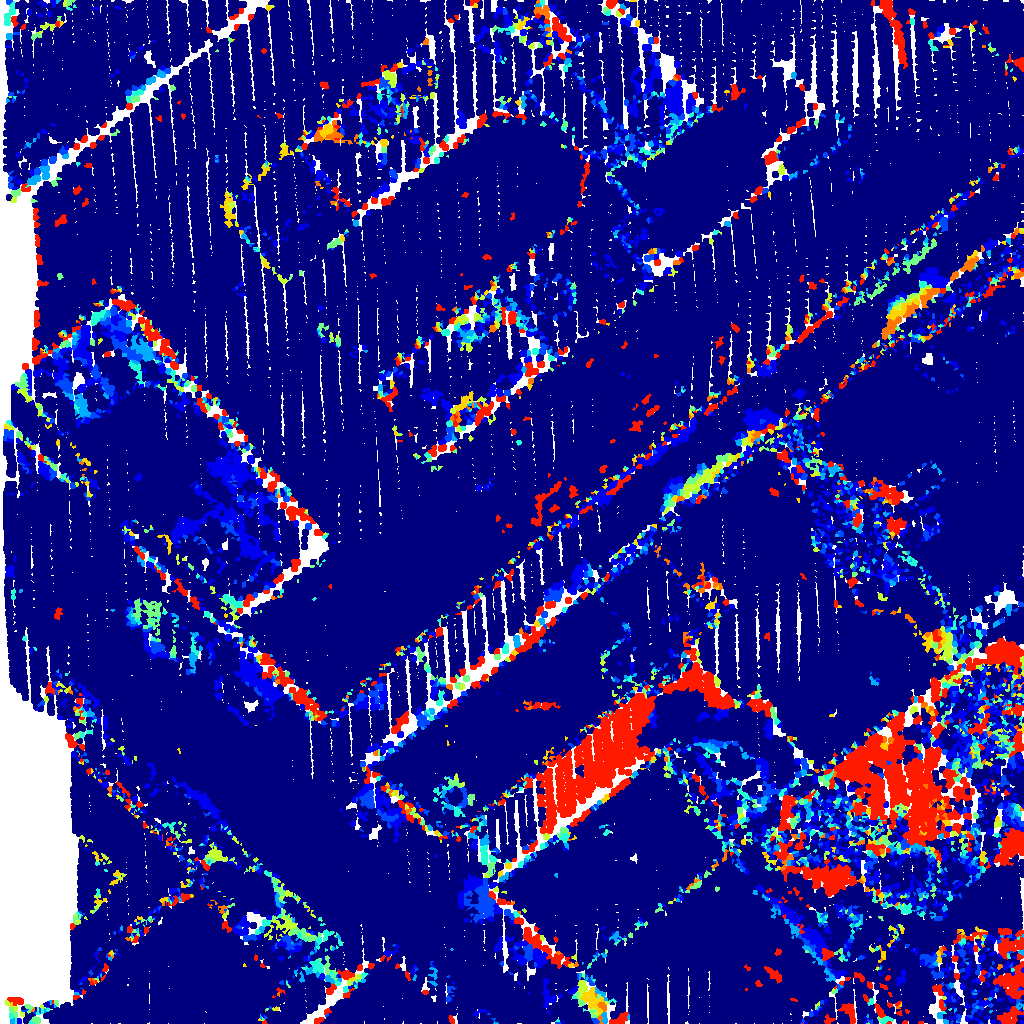}
		\includegraphics[width=\linewidth]{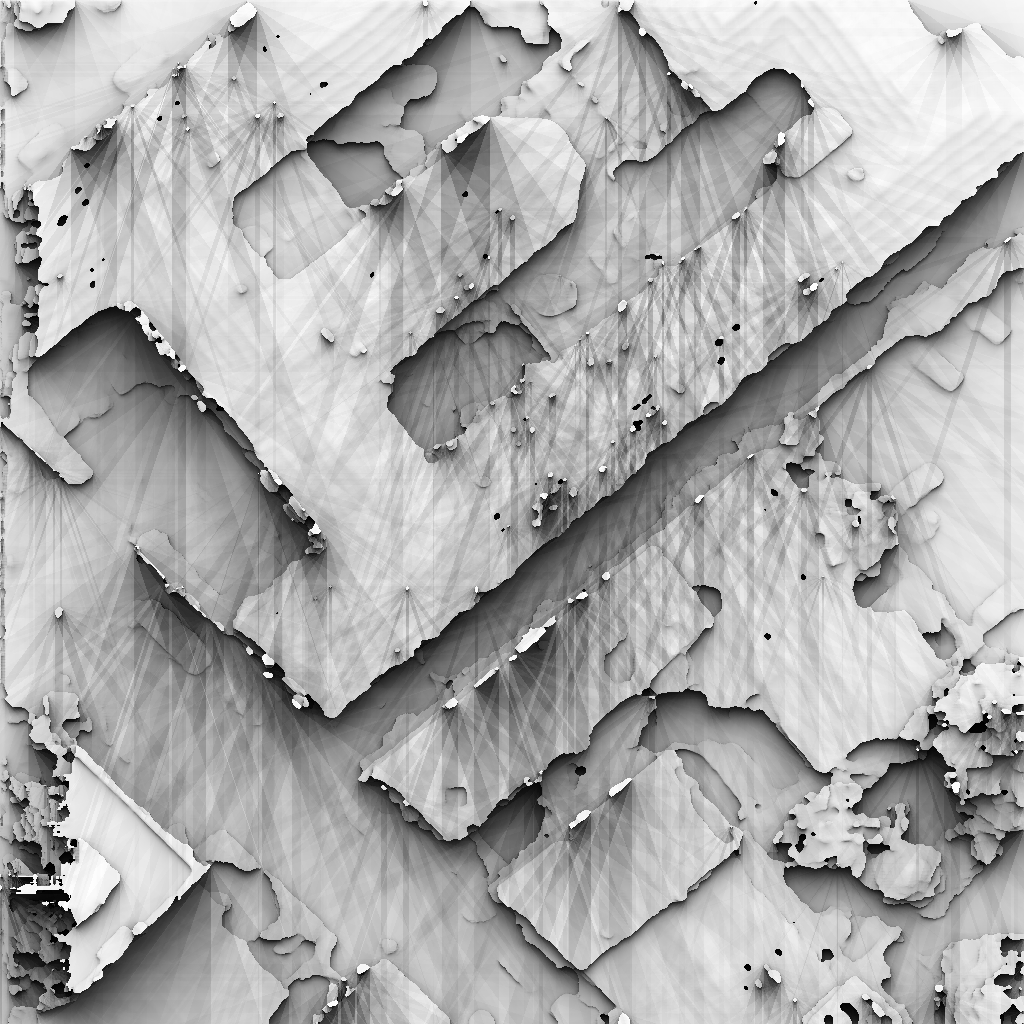}
		\centering{\tiny CBMV(SGM)}
	\end{minipage}
	\begin{minipage}[t]{0.19\textwidth}
		\includegraphics[width=0.098\linewidth]{figures_supp/color_map.png}
		\includegraphics[width=0.85\linewidth]{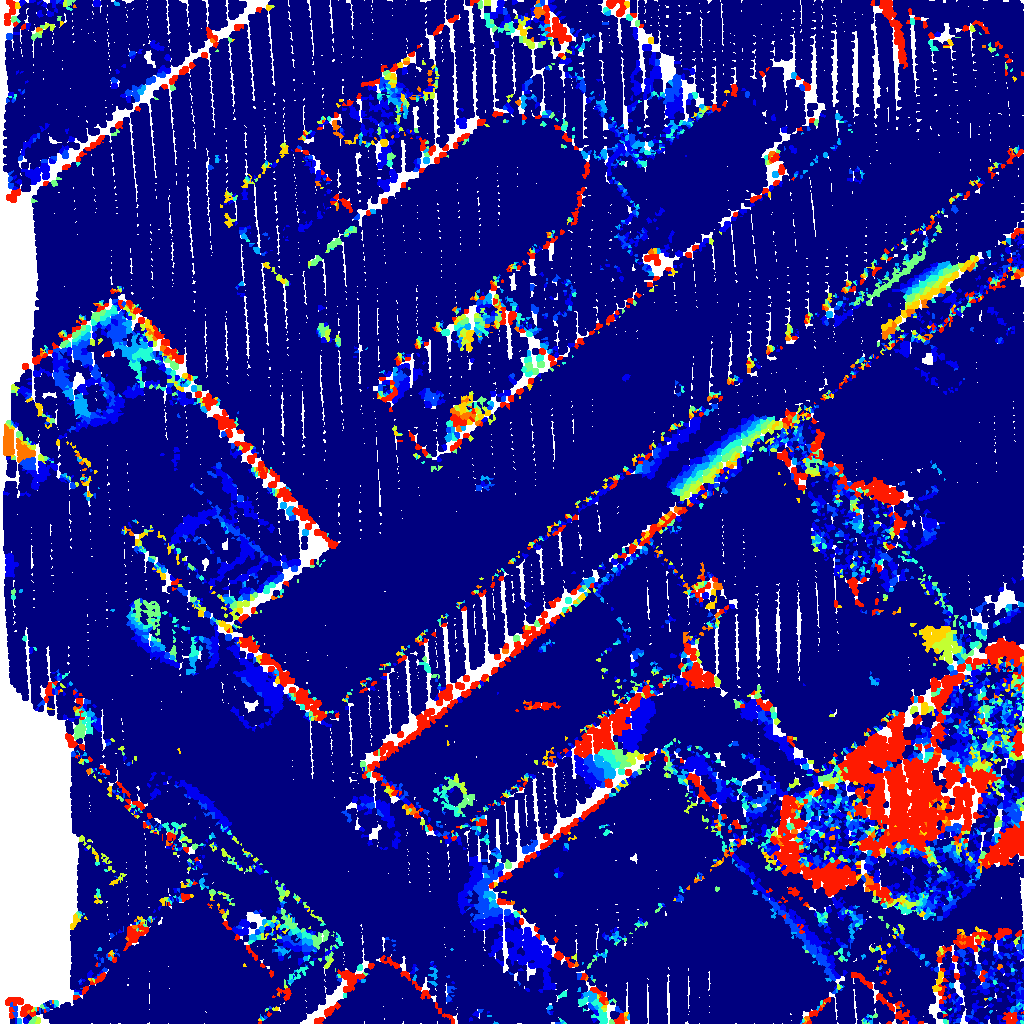}
		\includegraphics[width=\linewidth]{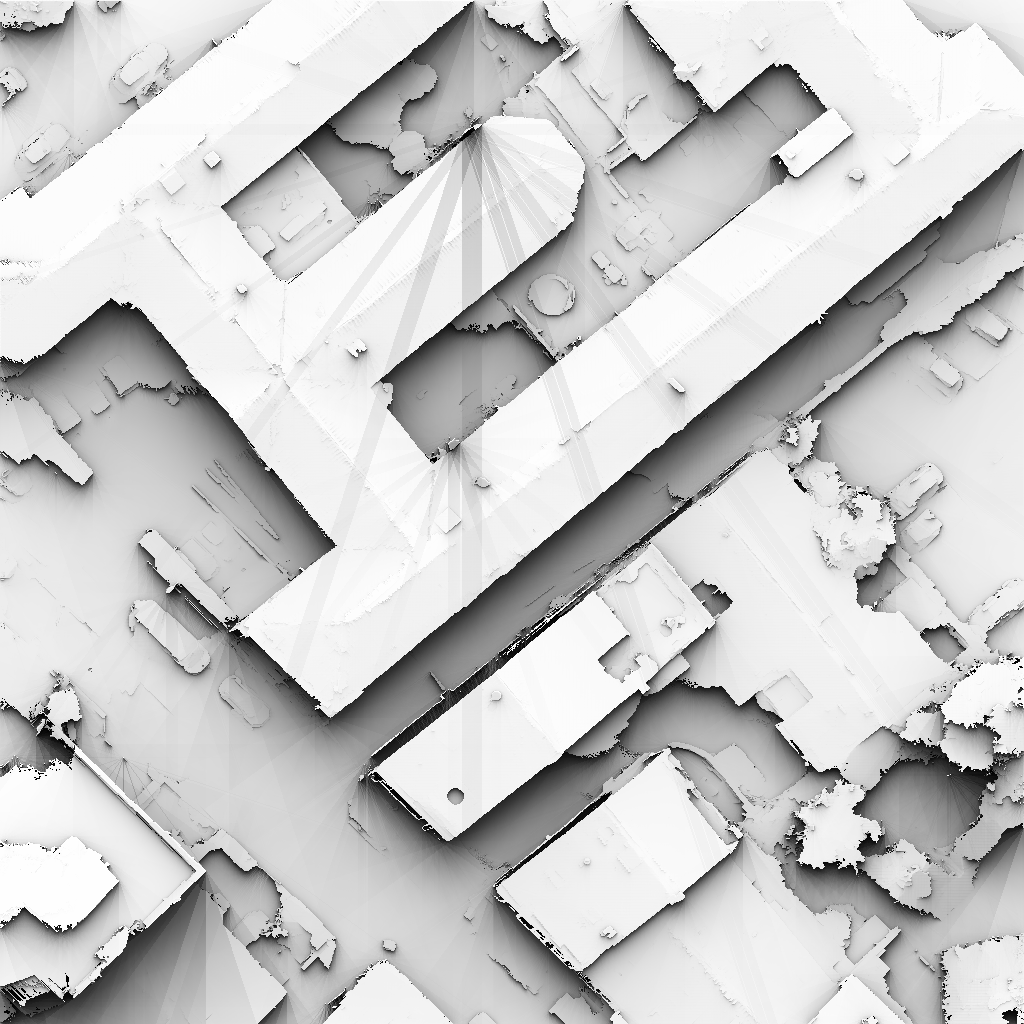}
		\centering{\tiny CBMV(GraphCuts)}
	\end{minipage}
	\begin{minipage}[t]{0.19\textwidth}
		\includegraphics[width=0.098\linewidth]{figures_supp/color_map.png}
		\includegraphics[width=0.85\linewidth]{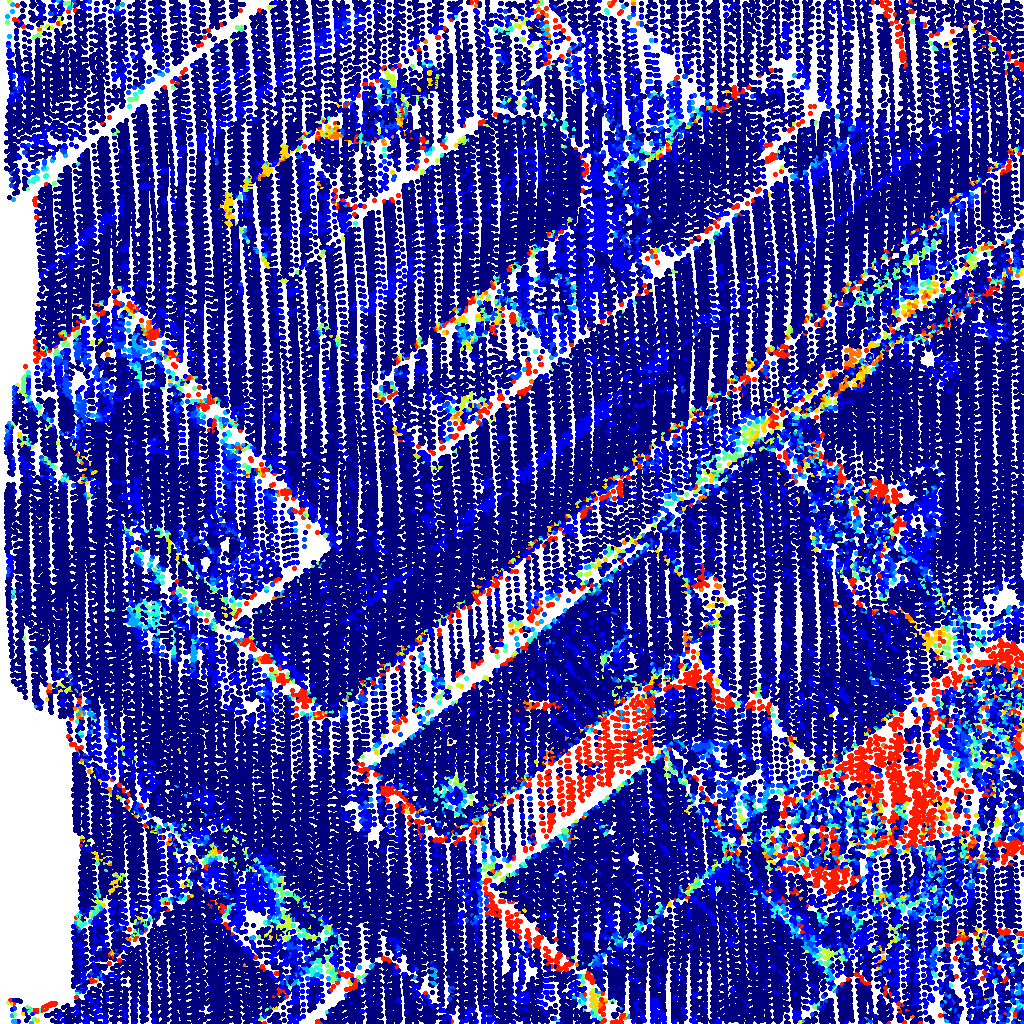}
		\includegraphics[width=\linewidth]{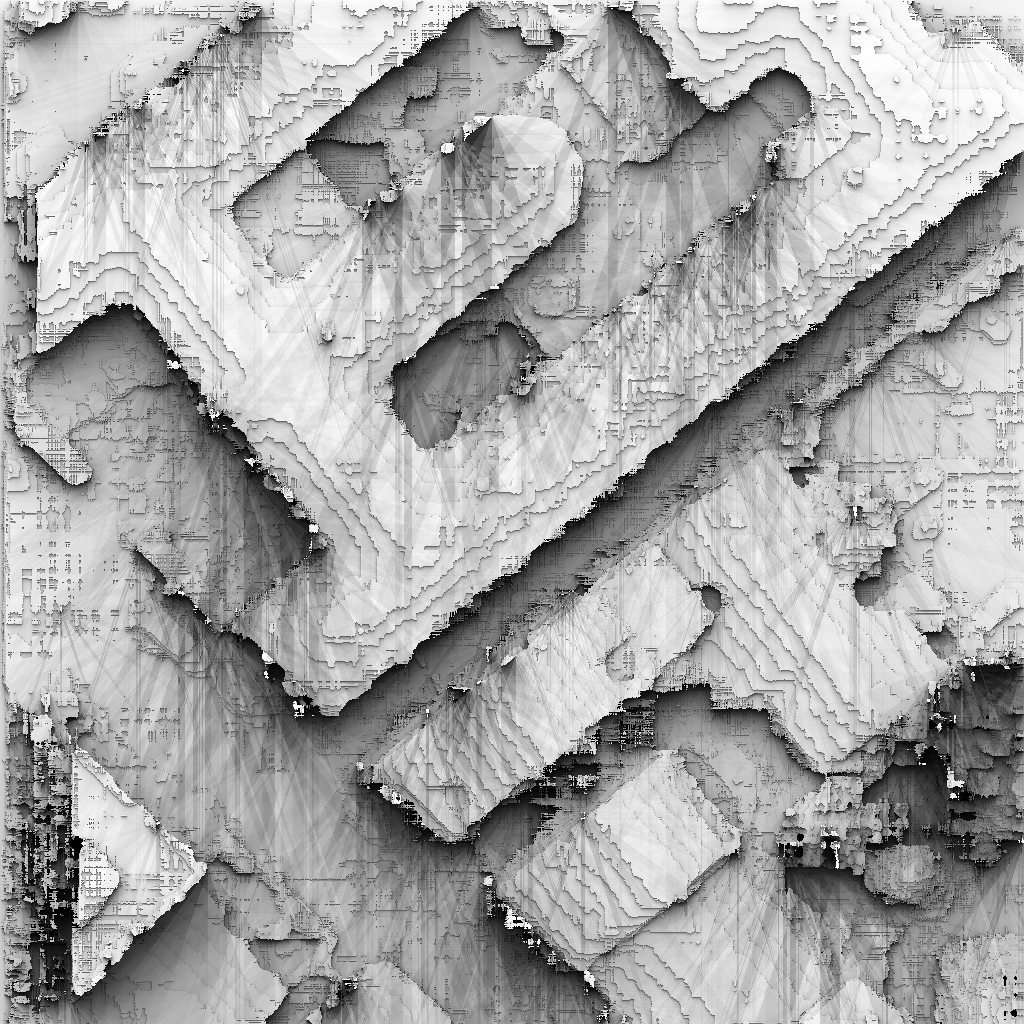}
		\centering{\tiny MC-CNN(KITTI)}
	\end{minipage}
	\begin{minipage}[t]{0.19\textwidth}
		\includegraphics[width=0.098\linewidth]{figures_supp/color_map.png}
		\includegraphics[width=0.85\linewidth]{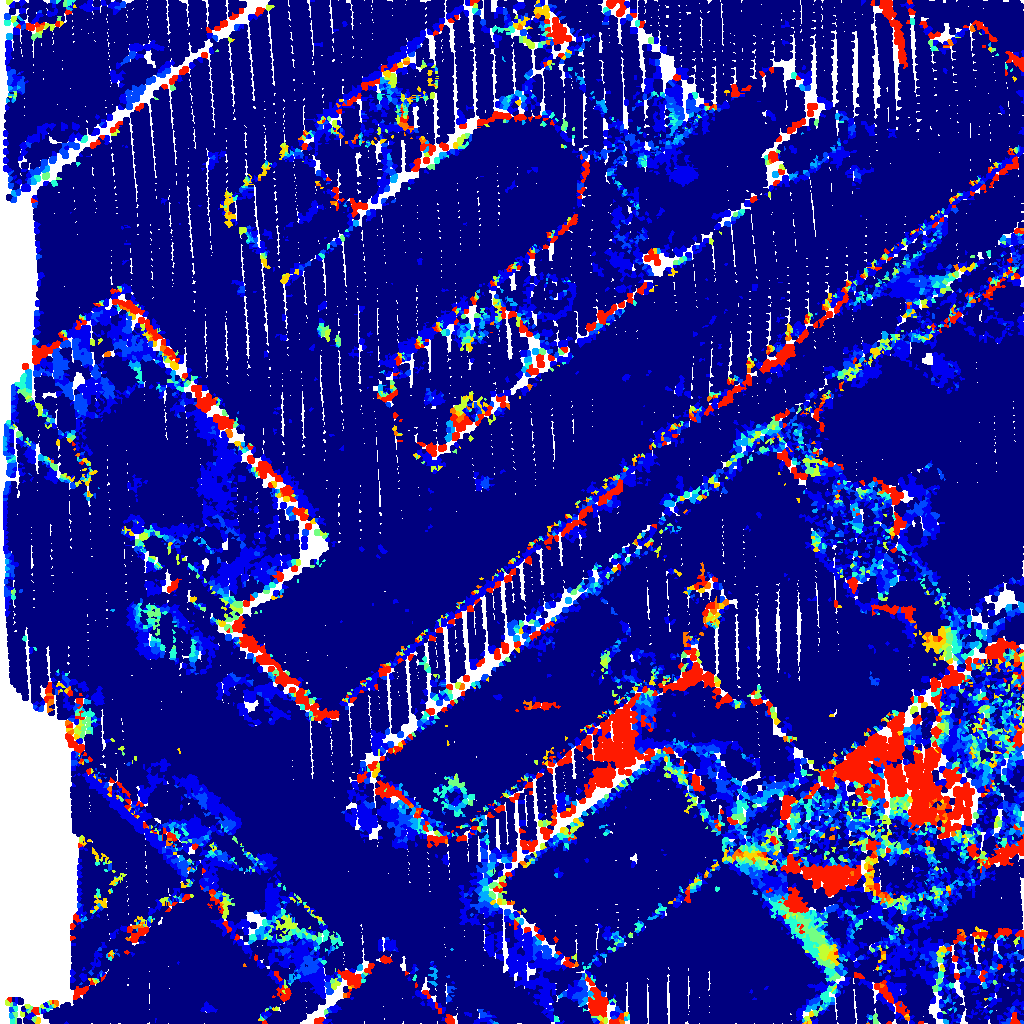}
		\includegraphics[width=\linewidth]{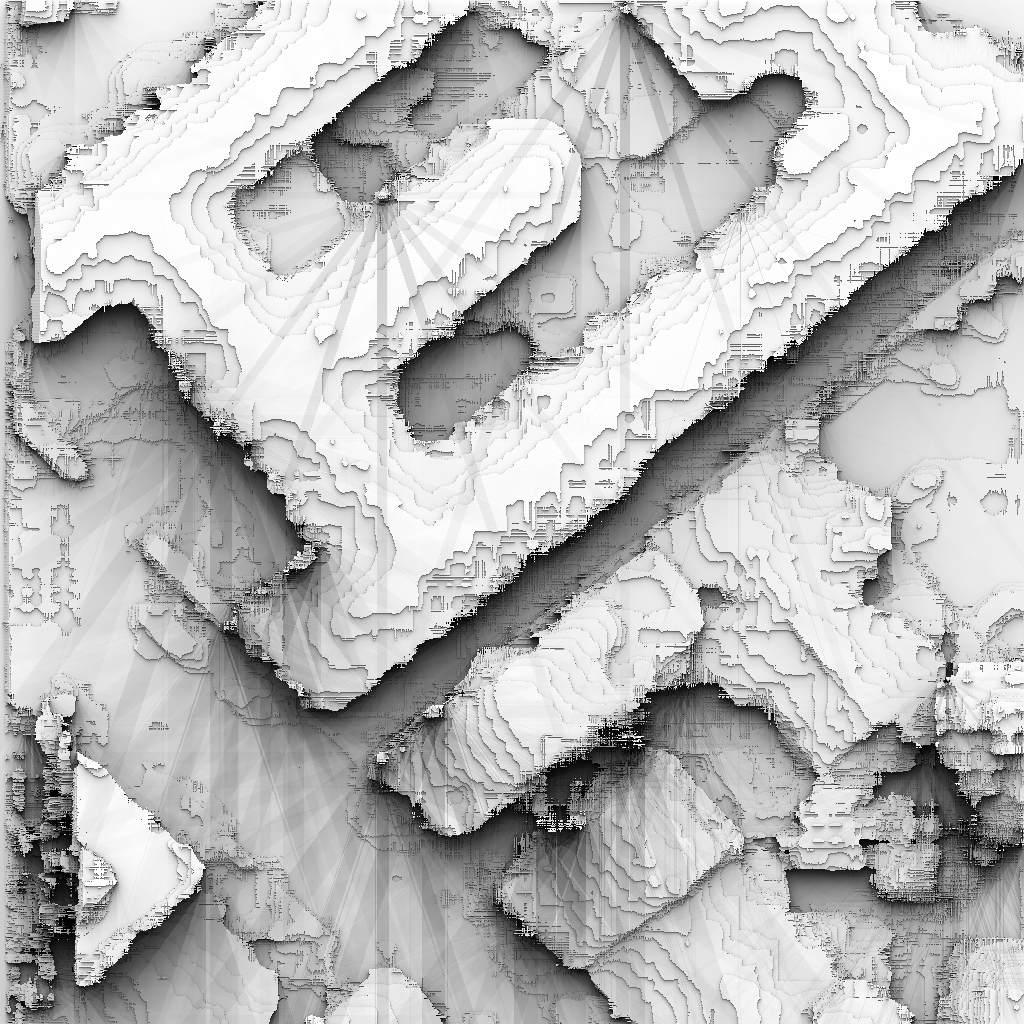}
		\centering{\tiny DeepFeature(KITTI)}
	\end{minipage}
	\begin{minipage}[t]{0.19\textwidth}
		\includegraphics[width=0.098\linewidth]{figures_supp/color_map.png}
		\includegraphics[width=0.85\linewidth]{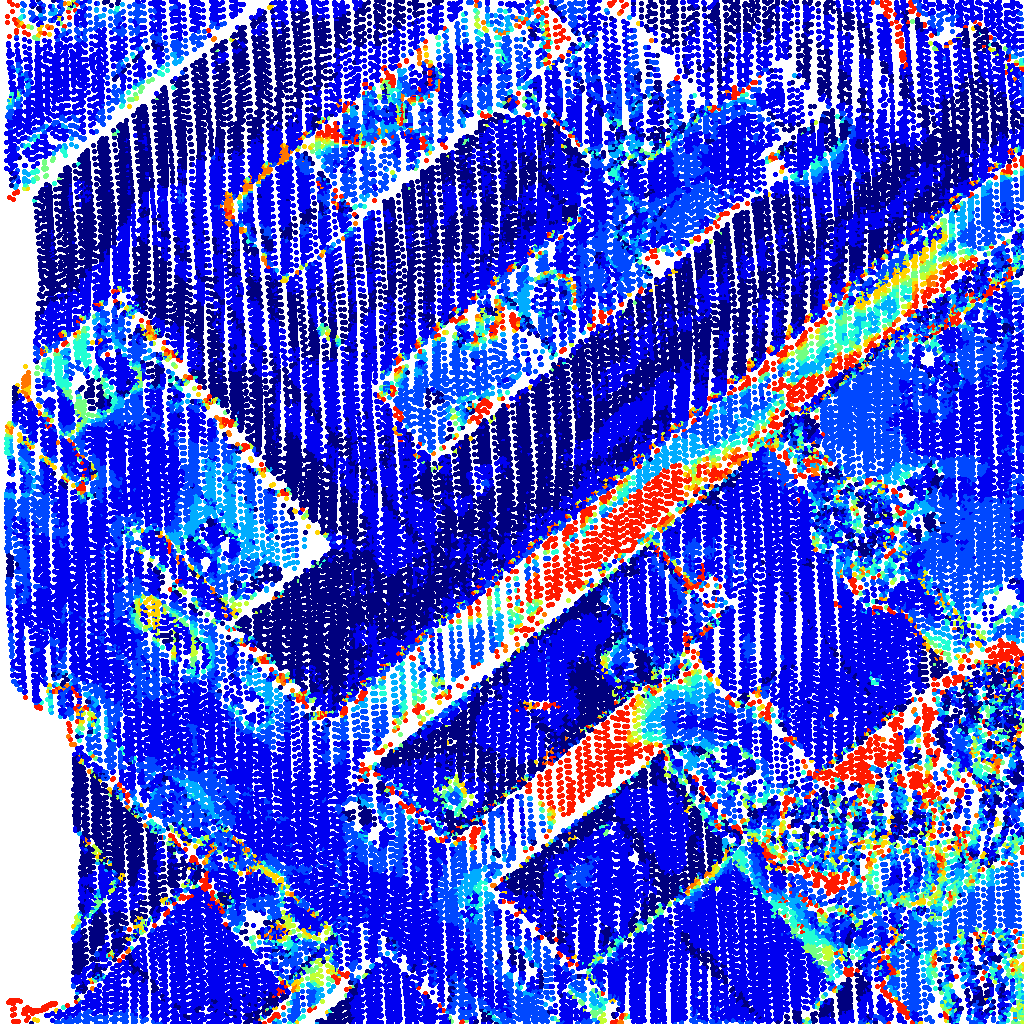}
		\includegraphics[width=\linewidth]{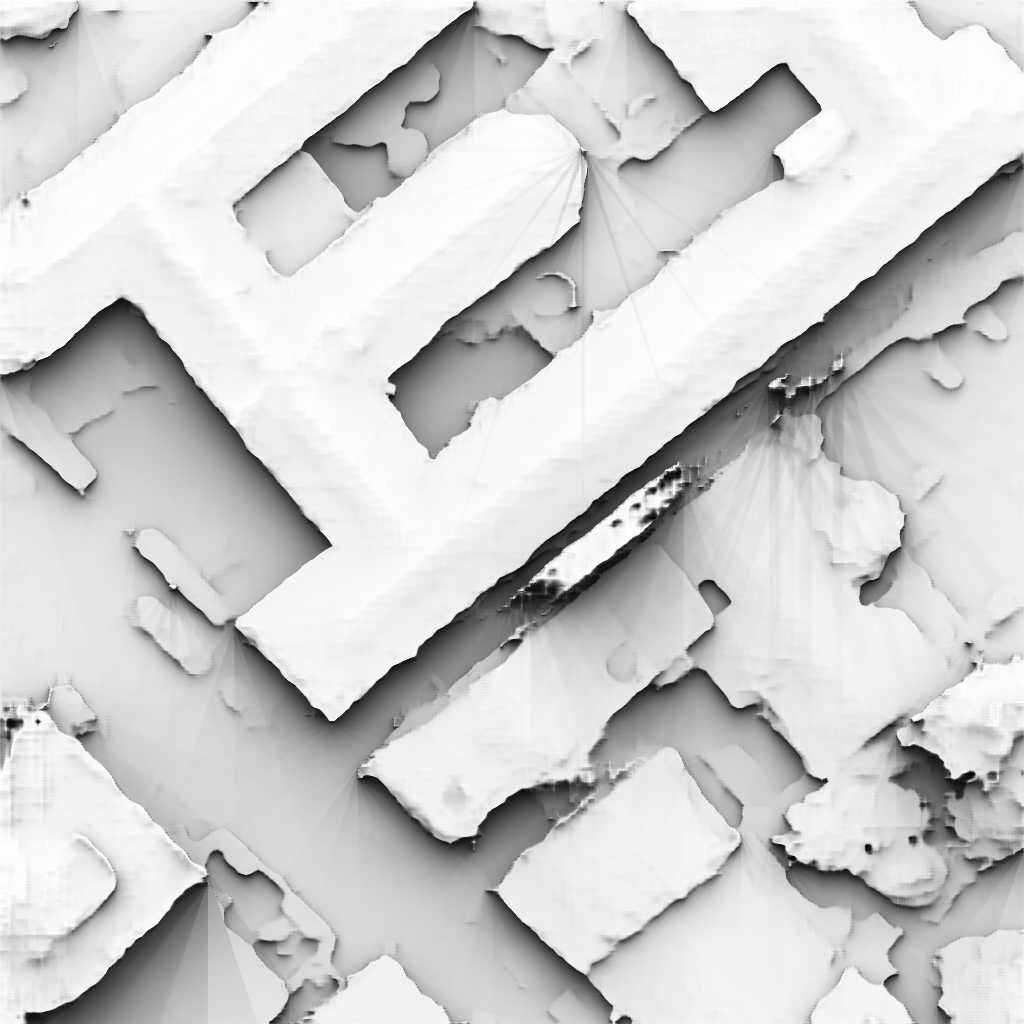}
		\centering{\tiny PSM net(KITTI)}
	\end{minipage}
	\begin{minipage}[t]{0.19\textwidth}
		\includegraphics[width=0.098\linewidth]{figures_supp/color_map.png}
		\includegraphics[width=0.85\linewidth]{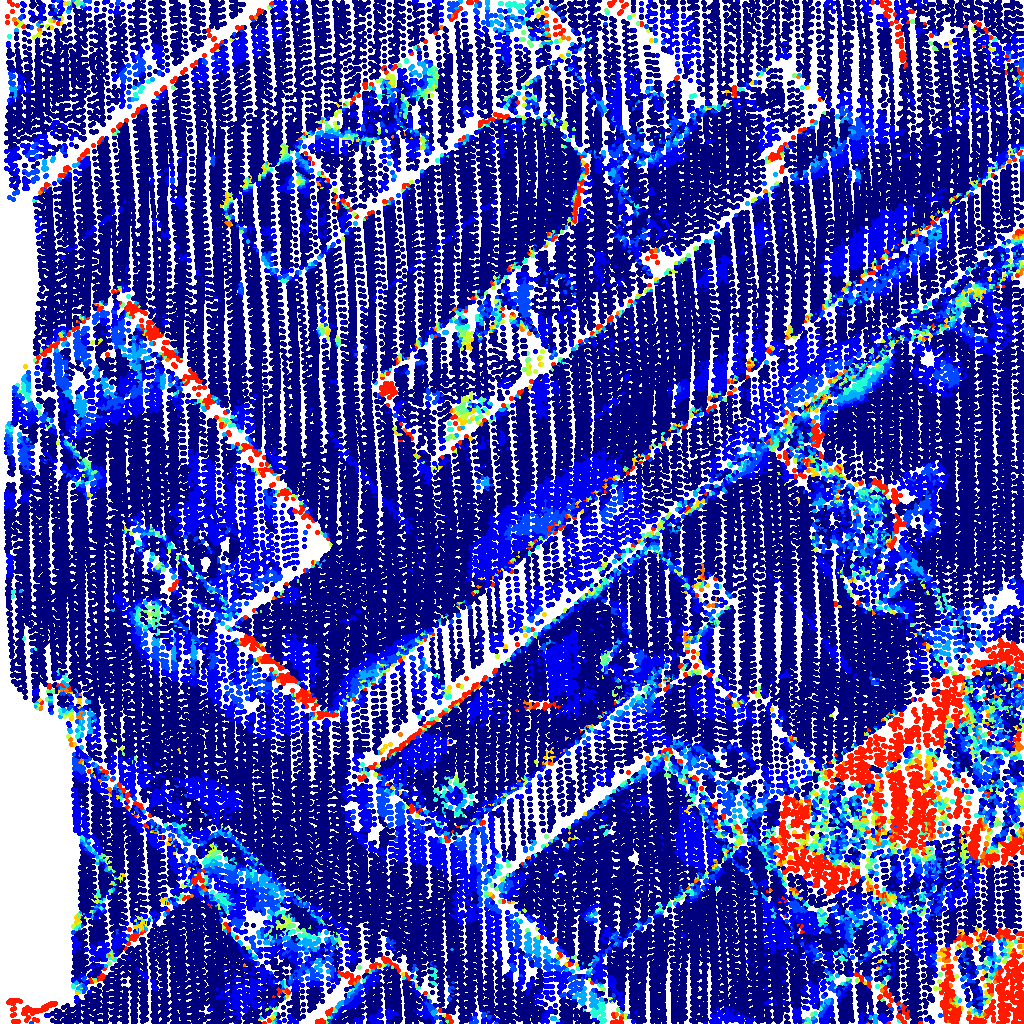}
		\includegraphics[width=\linewidth]{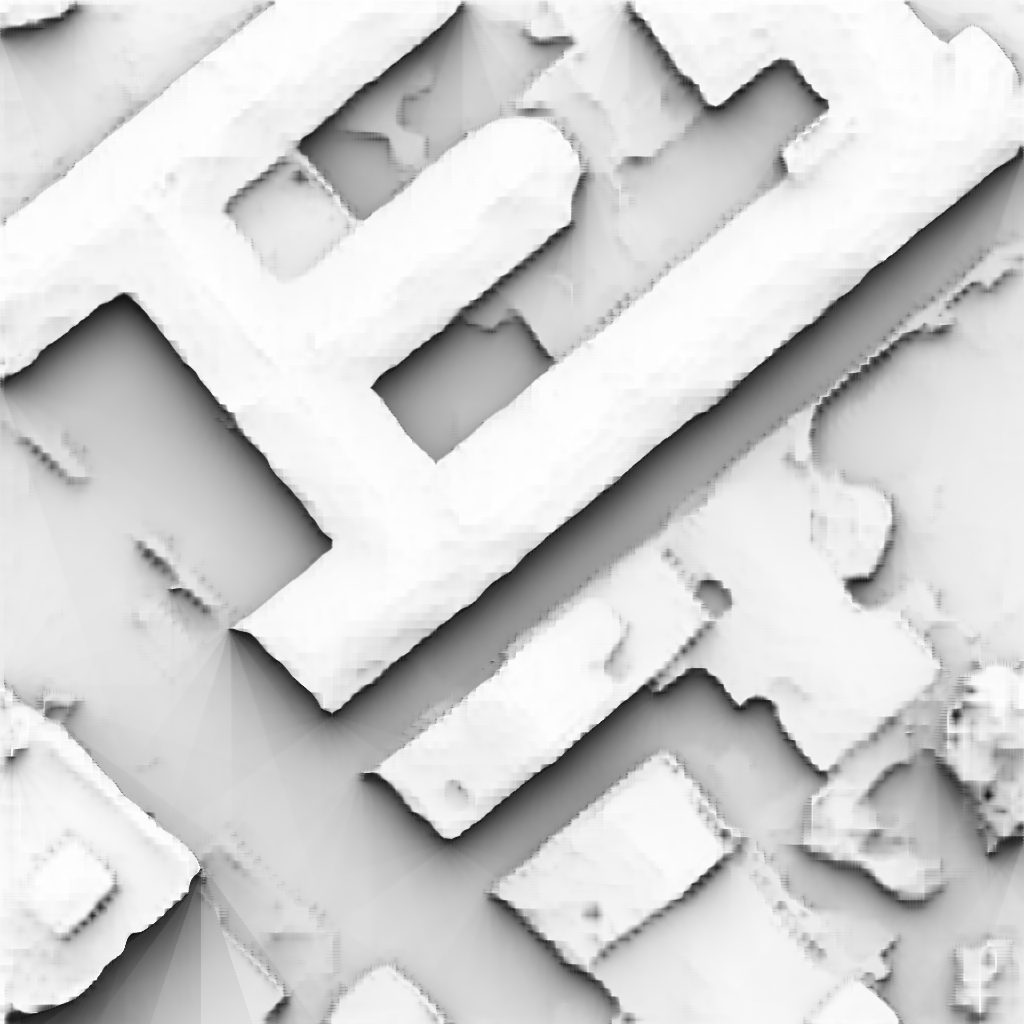}
		\centering{\tiny HRS net(KITTI)}
	\end{minipage}
	\begin{minipage}[t]{0.19\textwidth}	
		\includegraphics[width=0.098\linewidth]{figures_supp/color_map.png}
		\includegraphics[width=0.85\linewidth]{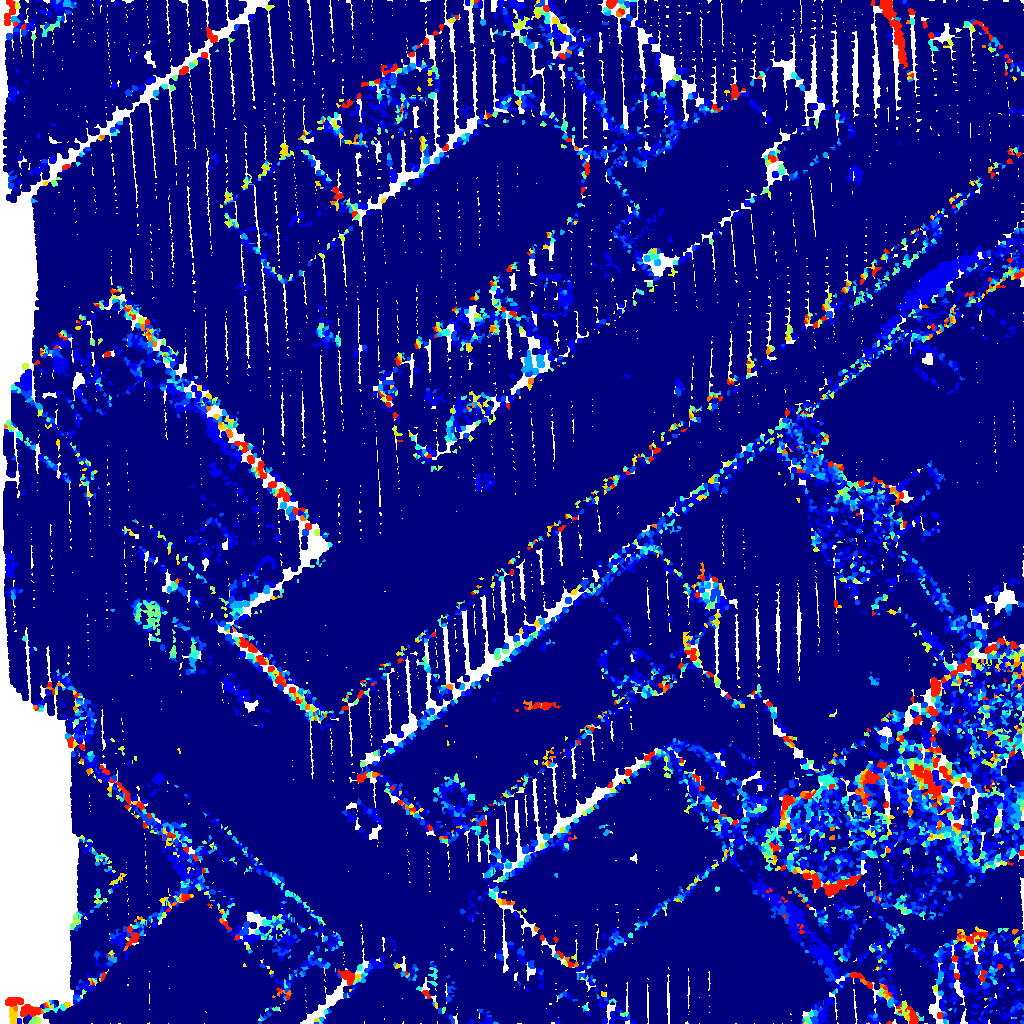}
		\includegraphics[width=\linewidth]{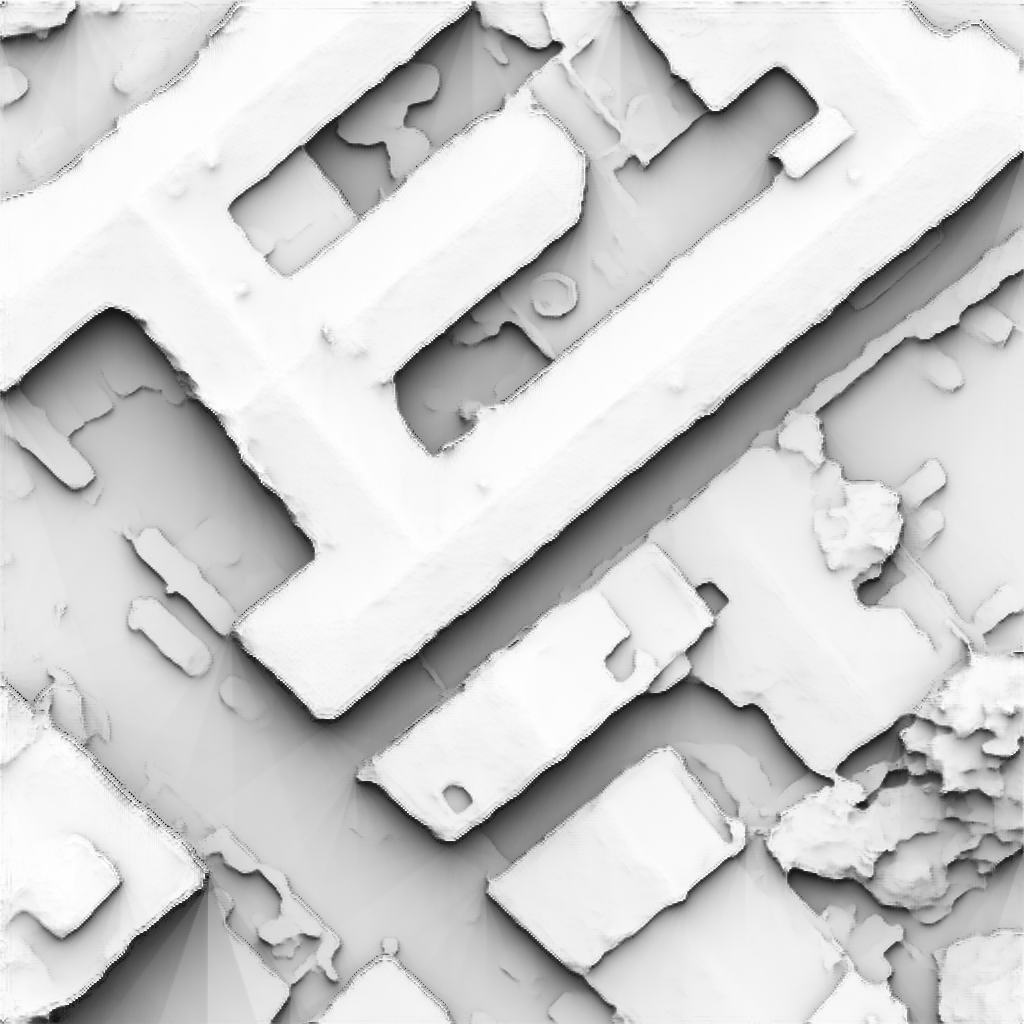}
		\centering{\tiny DeepPruner(KITTI)}
	\end{minipage}
	\begin{minipage}[t]{0.19\textwidth}
		\includegraphics[width=0.098\linewidth]{figures_supp/color_map.png}
		\includegraphics[width=0.85\linewidth]{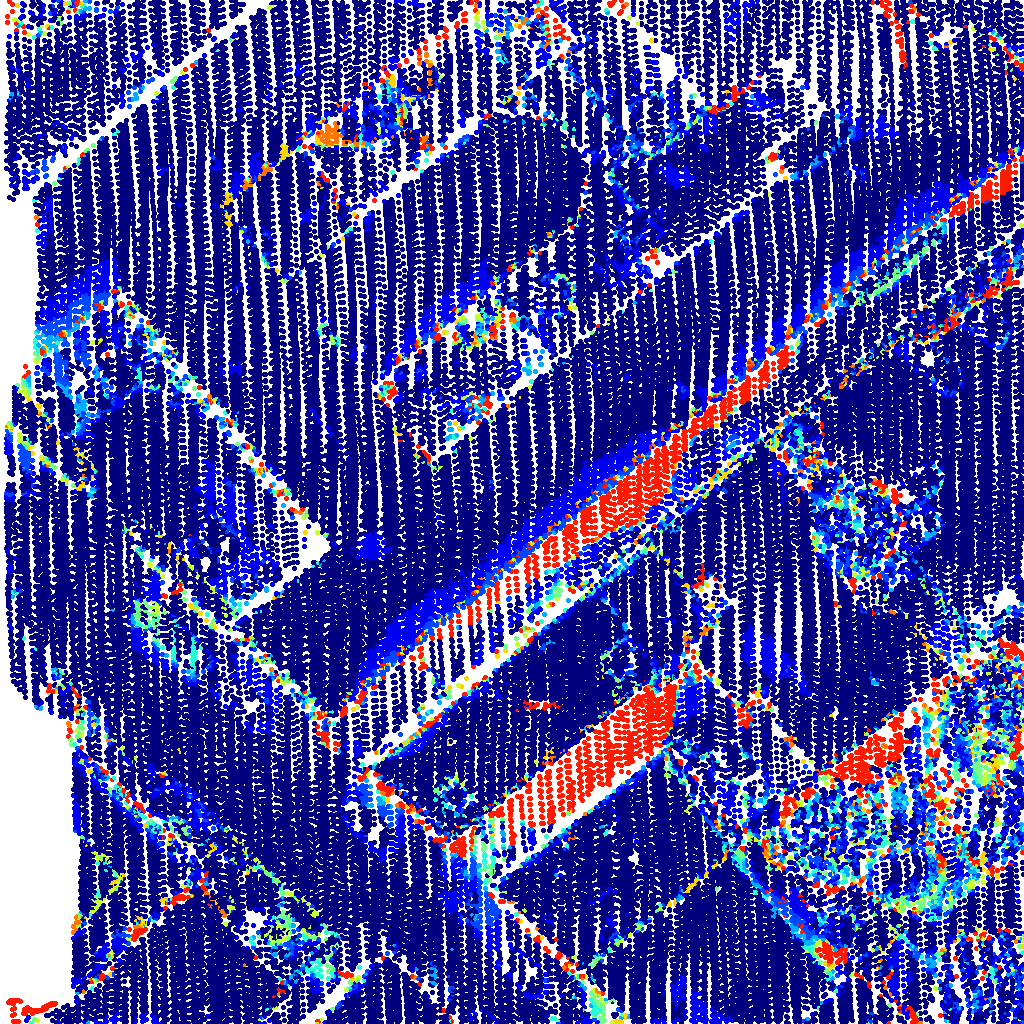}
		\includegraphics[width=\linewidth]{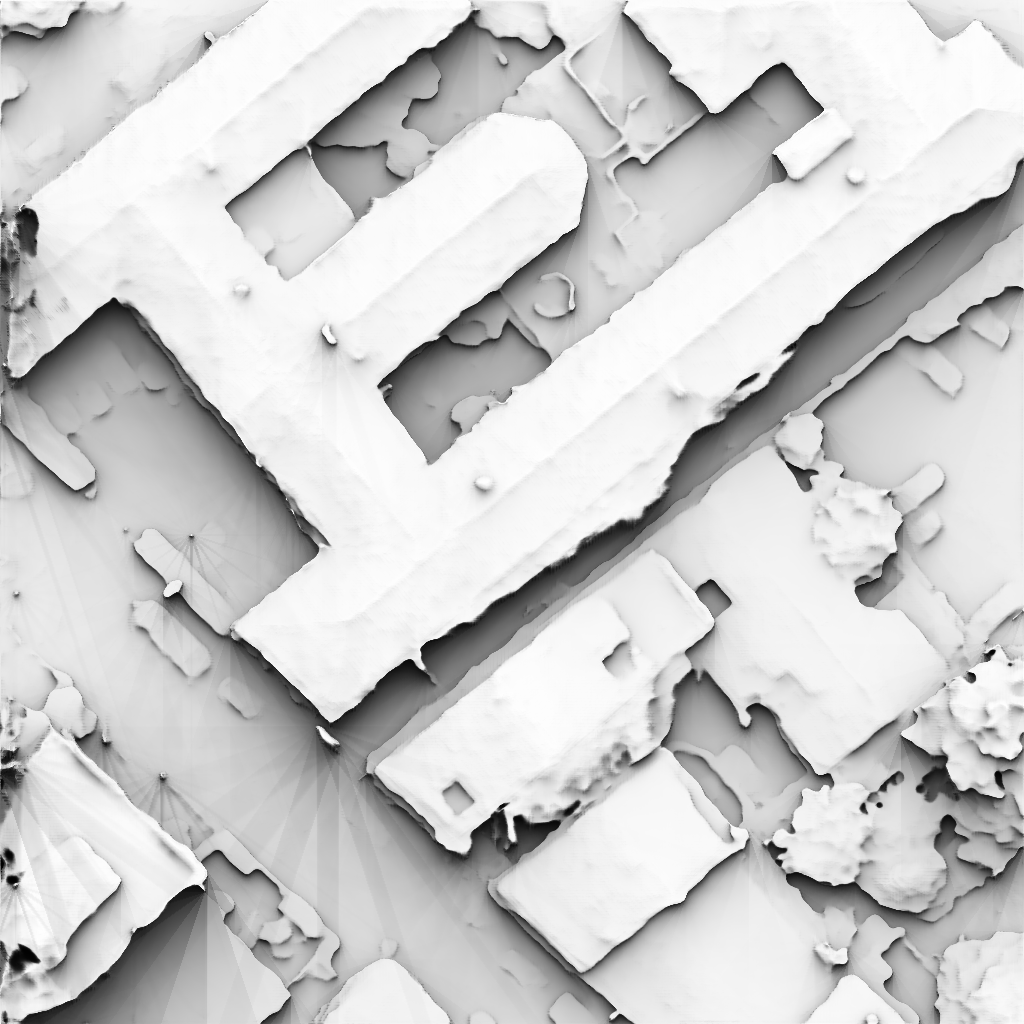}
		\centering{\tiny GANet(KITTI)}
	\end{minipage}
	\begin{minipage}[t]{0.19\textwidth}	
		\includegraphics[width=0.098\linewidth]{figures_supp/color_map.png}
		\includegraphics[width=0.85\linewidth]{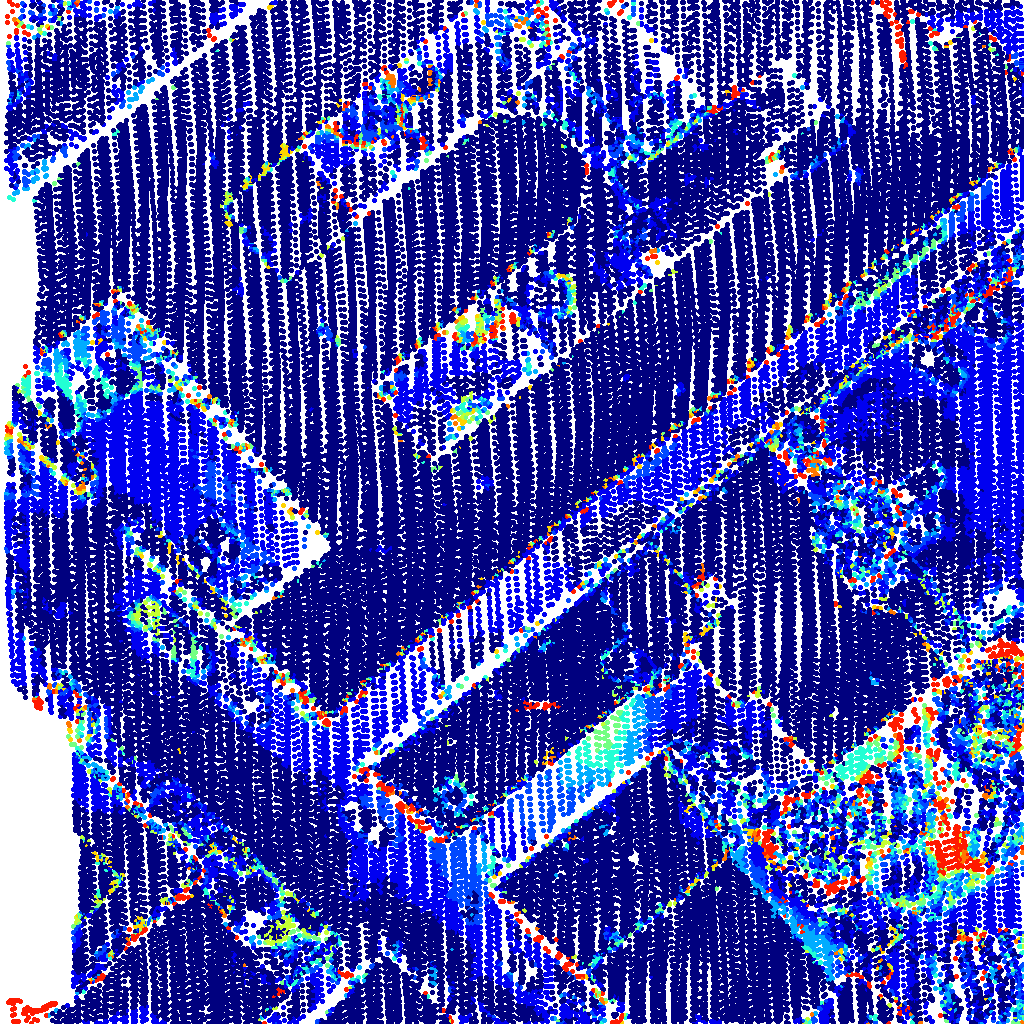}
		\includegraphics[width=\linewidth]{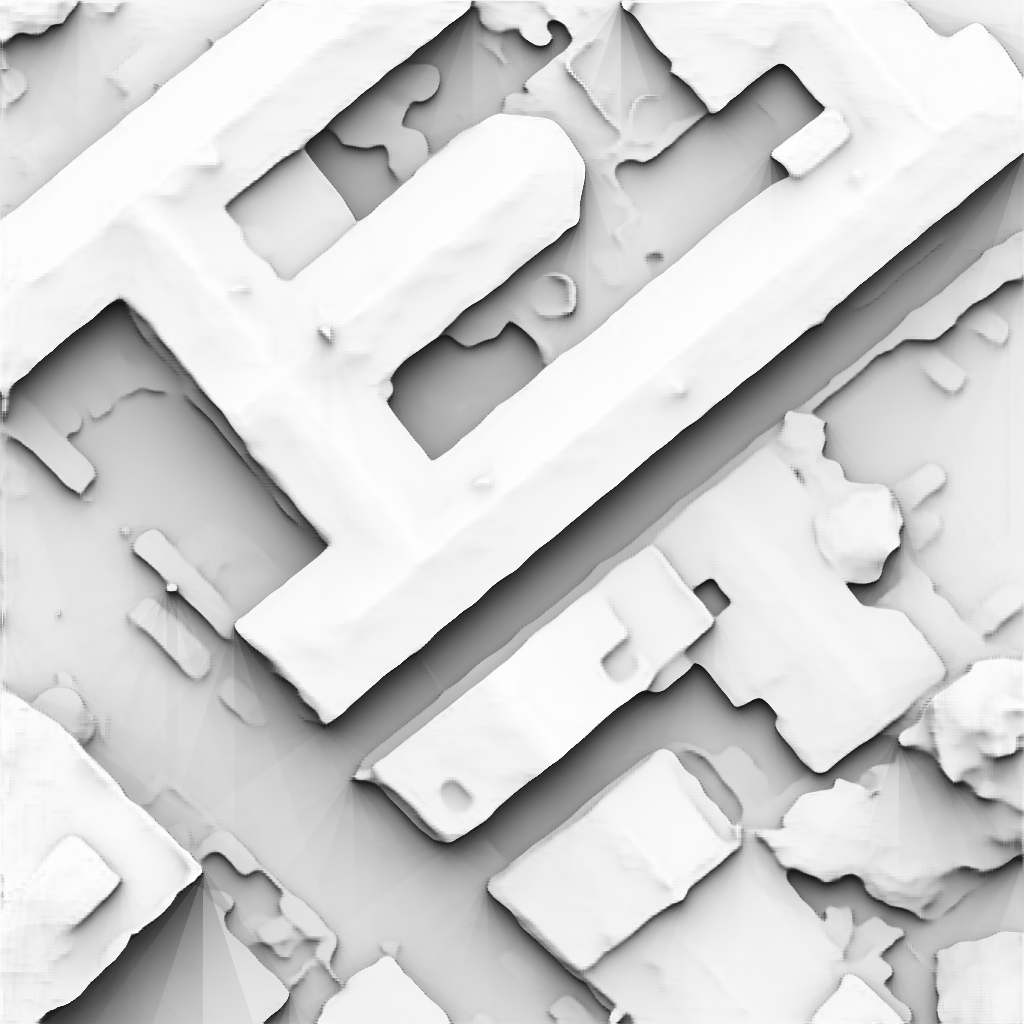}
		\centering{\tiny LEAStereo(KITTI)}
	\end{minipage}
	\begin{minipage}[t]{0.19\textwidth}
		\includegraphics[width=0.098\linewidth]{figures_supp/color_map.png}
		\includegraphics[width=0.85\linewidth]{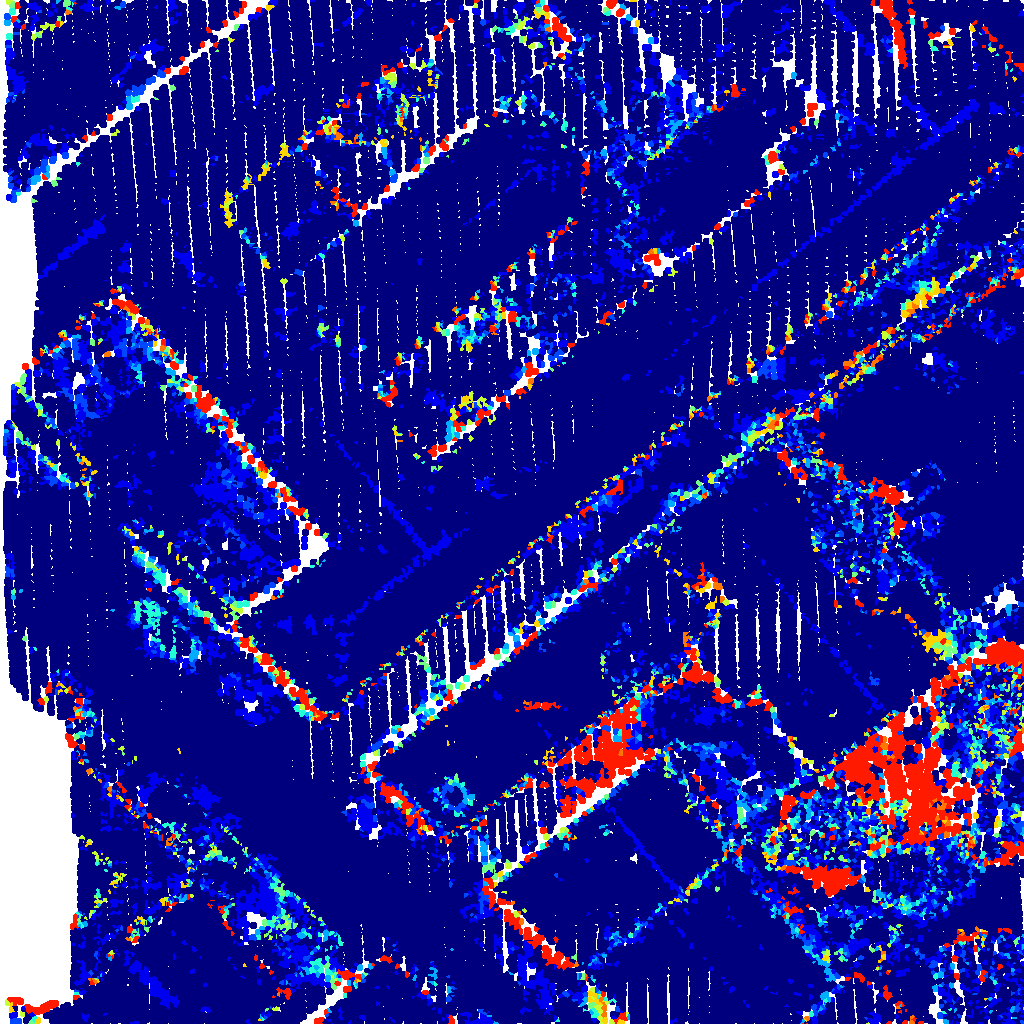}
		\includegraphics[width=\linewidth]{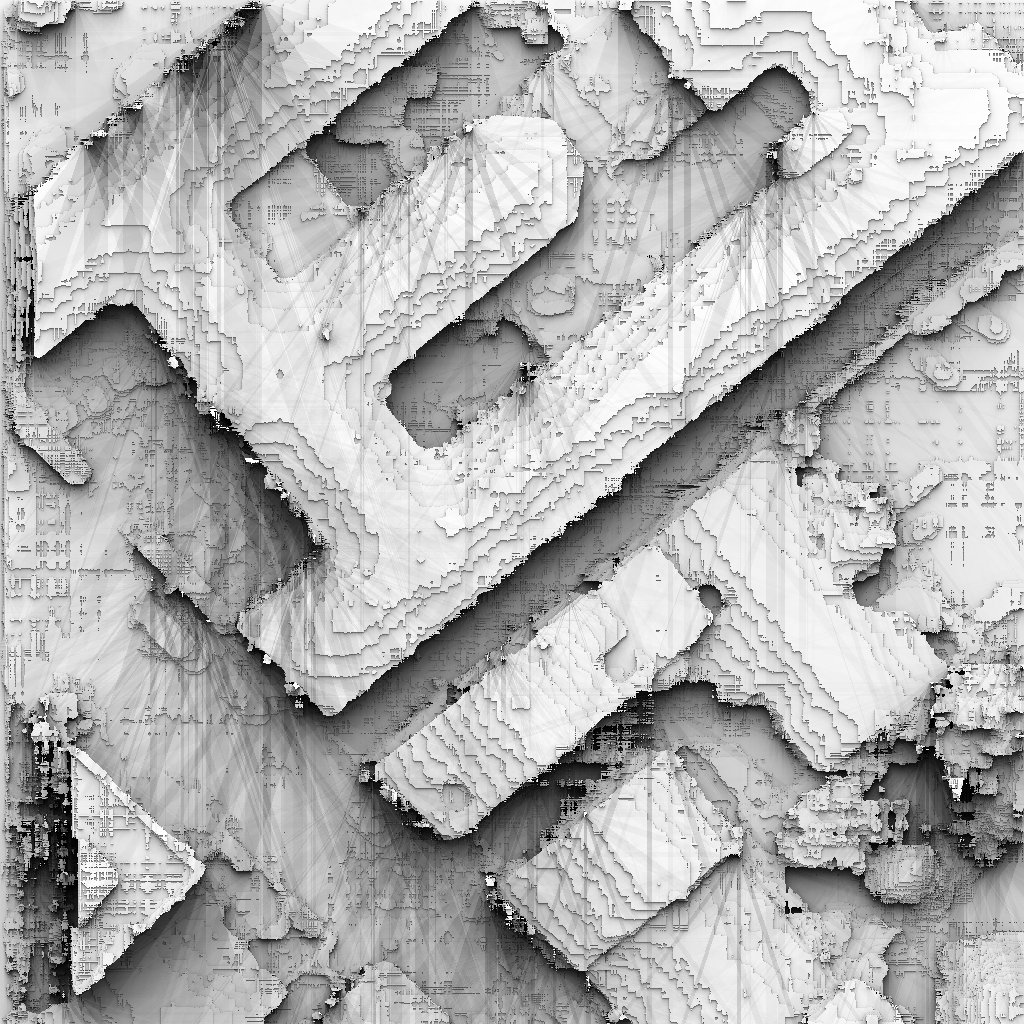}
		\centering{\tiny MC-CNN}
	\end{minipage}
	\begin{minipage}[t]{0.19\textwidth}
		\includegraphics[width=0.098\linewidth]{figures_supp/color_map.png}
		\includegraphics[width=0.85\linewidth]{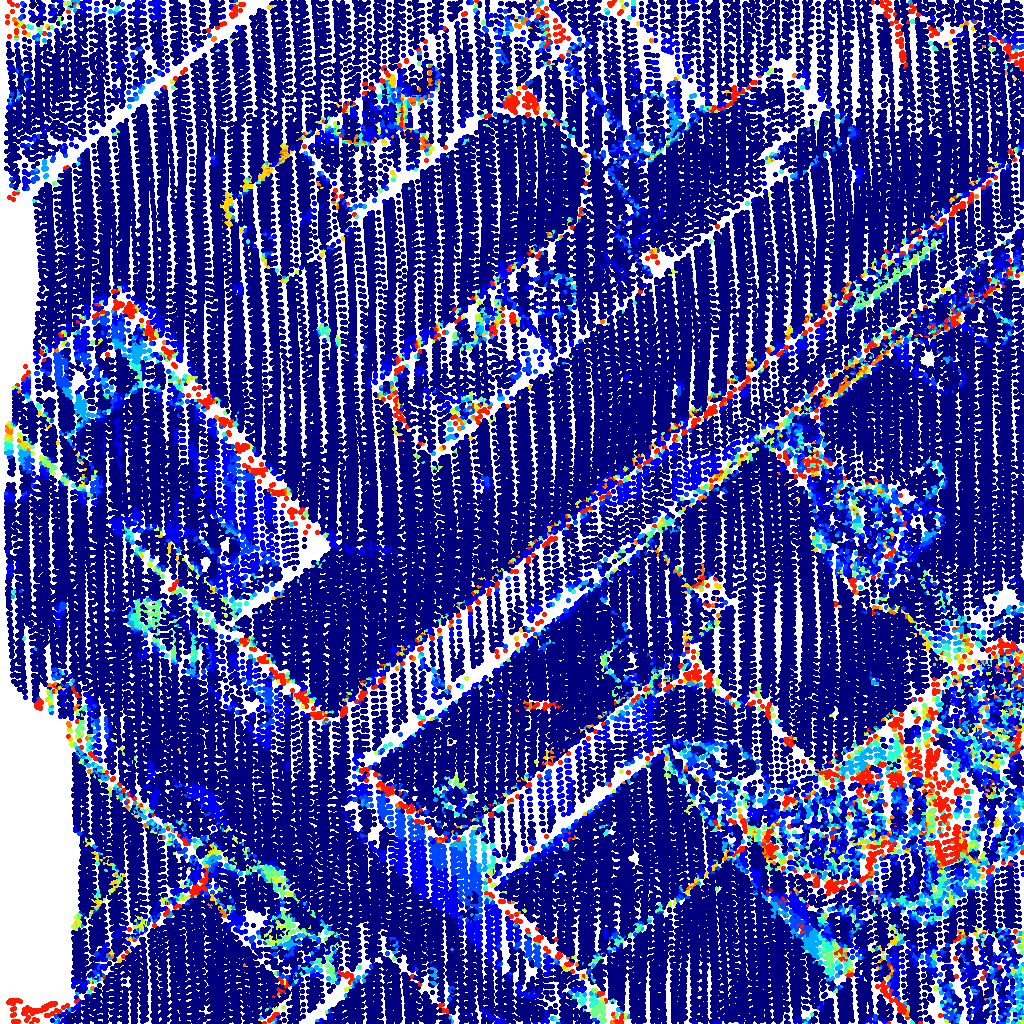}
		\includegraphics[width=\linewidth]{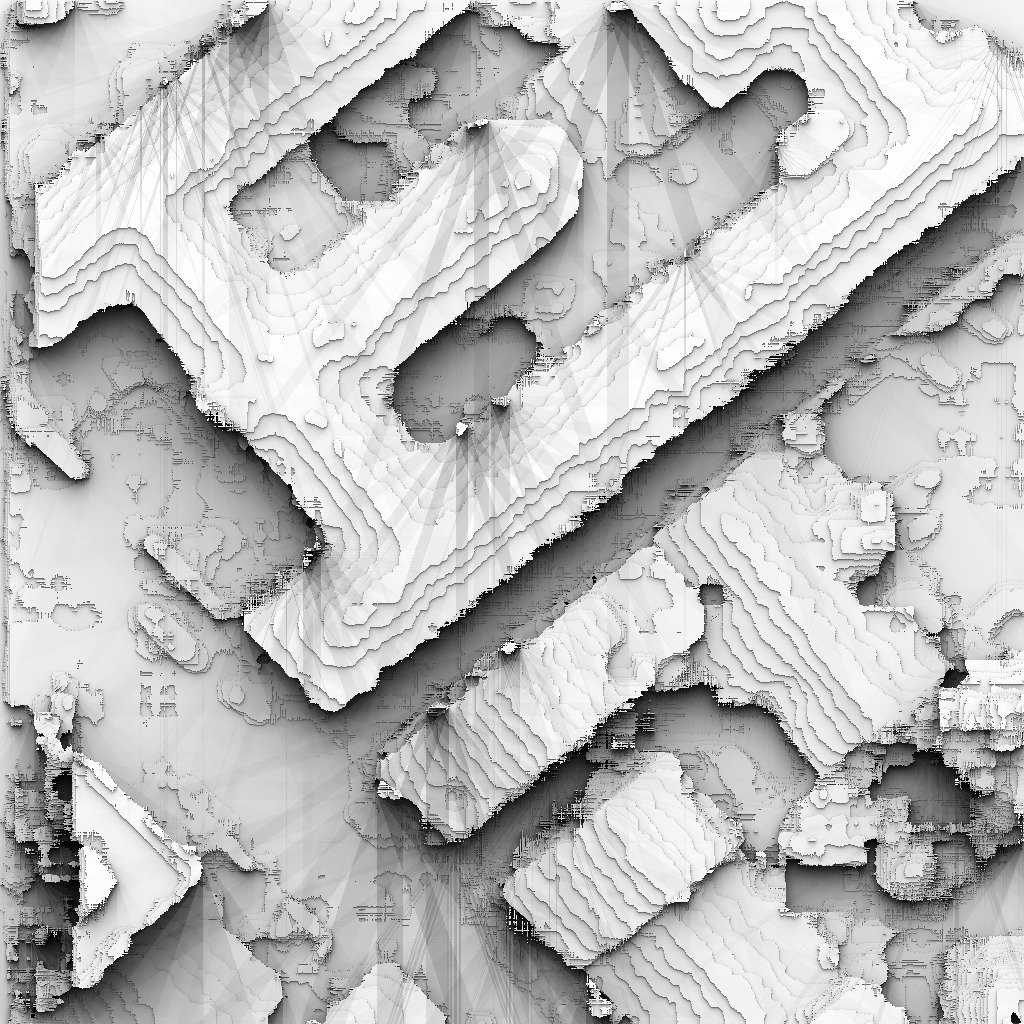}
		\centering{\tiny DeepFeature}
	\end{minipage}
	\begin{minipage}[t]{0.19\textwidth}
		\includegraphics[width=0.098\linewidth]{figures_supp/color_map.png}
		\includegraphics[width=0.85\linewidth]{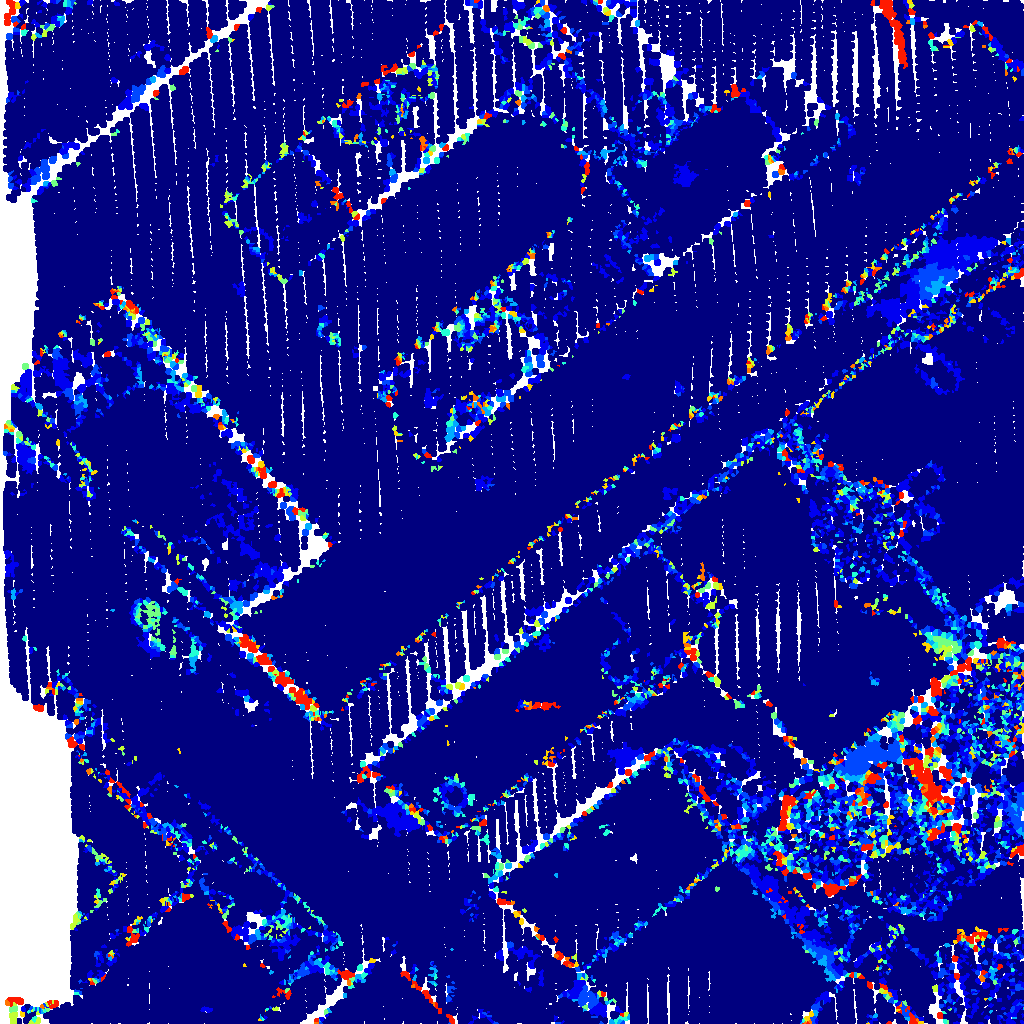}
		\includegraphics[width=\linewidth]{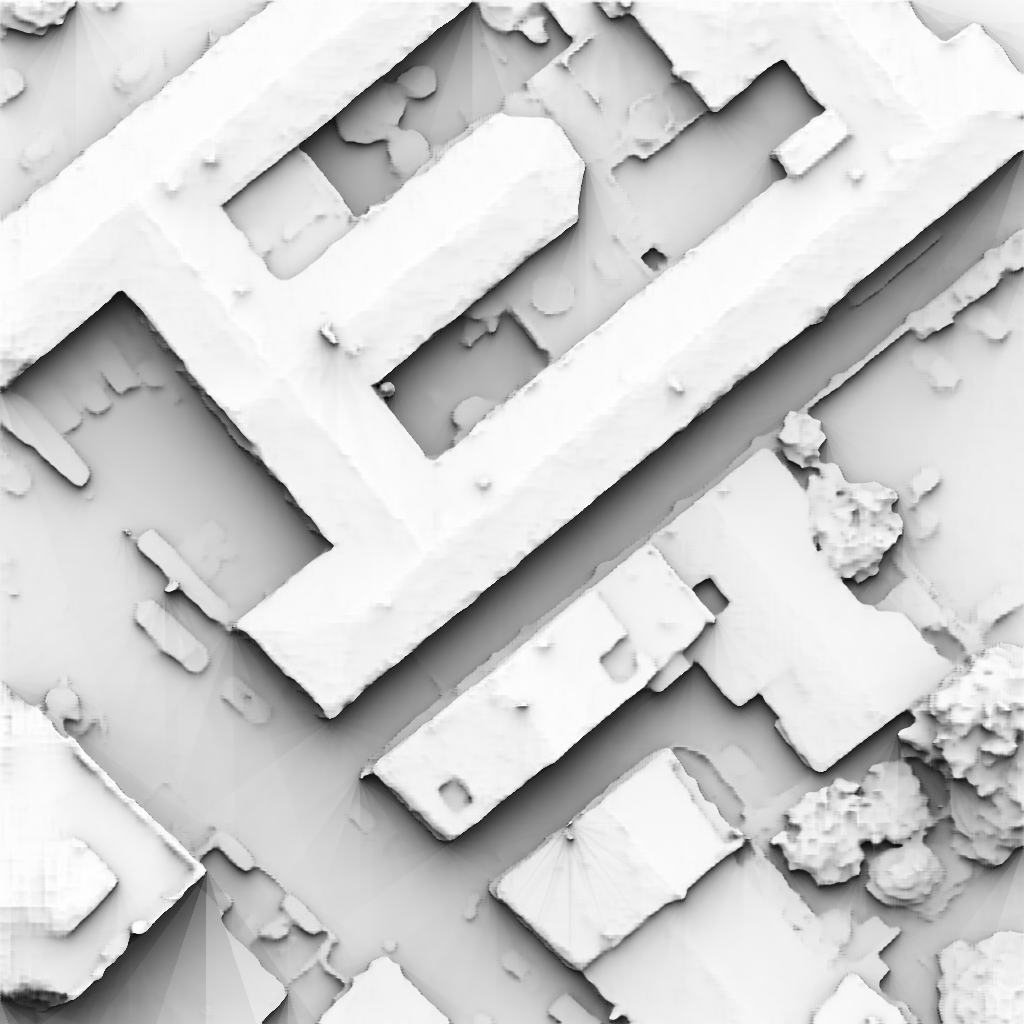}
		\centering{\tiny PSM net}
	\end{minipage}
	\begin{minipage}[t]{0.19\textwidth}
		\includegraphics[width=0.098\linewidth]{figures_supp/color_map.png}
		\includegraphics[width=0.85\linewidth]{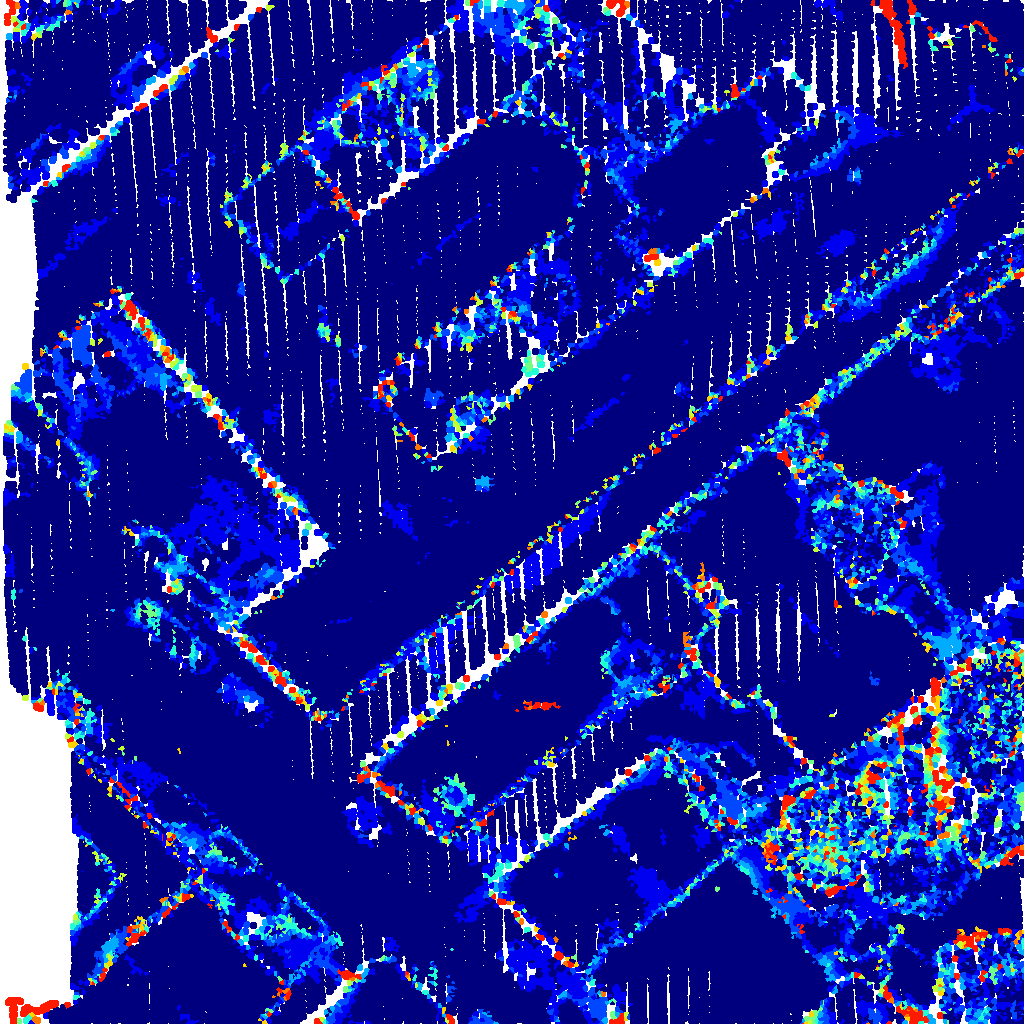}
		\includegraphics[width=\linewidth]{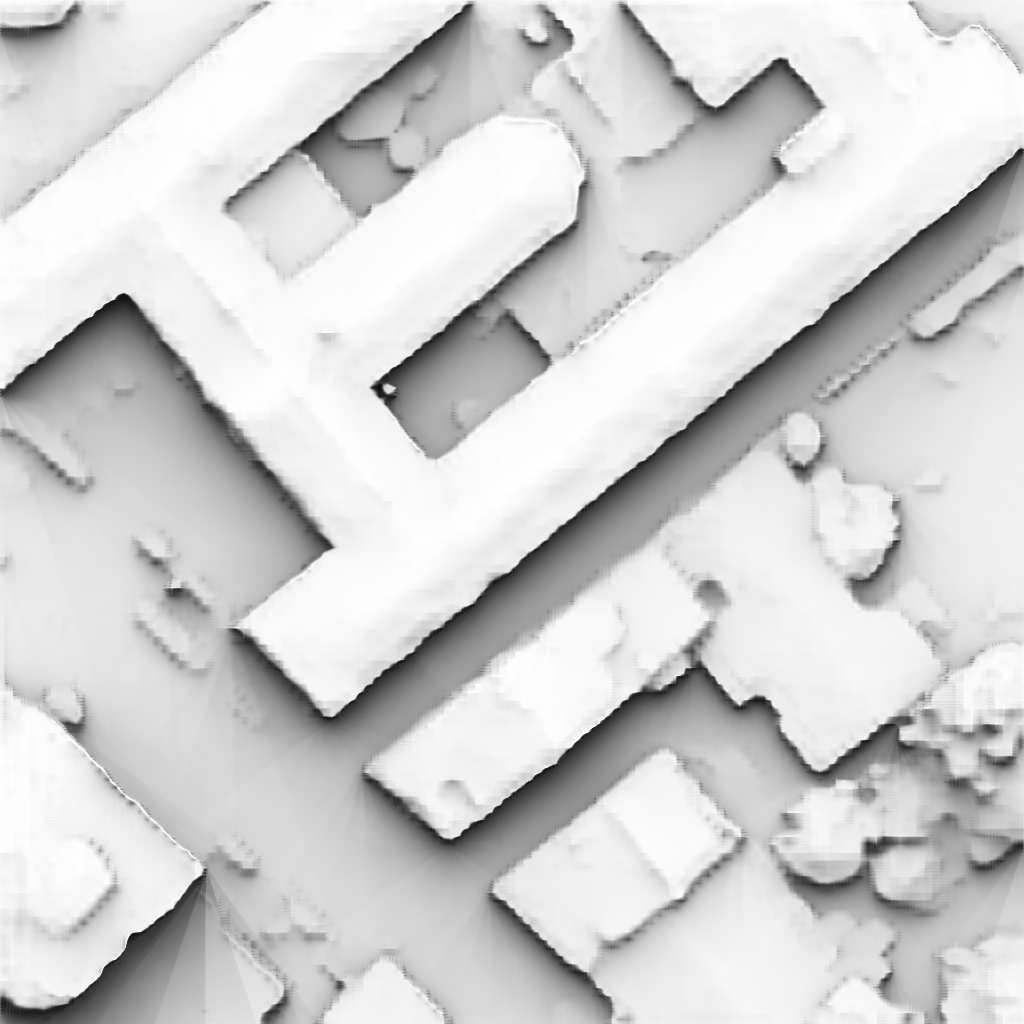}
		\centering{\tiny HRS net}
	\end{minipage}
	\begin{minipage}[t]{0.19\textwidth}	
		\includegraphics[width=0.098\linewidth]{figures_supp/color_map.png}
		\includegraphics[width=0.85\linewidth]{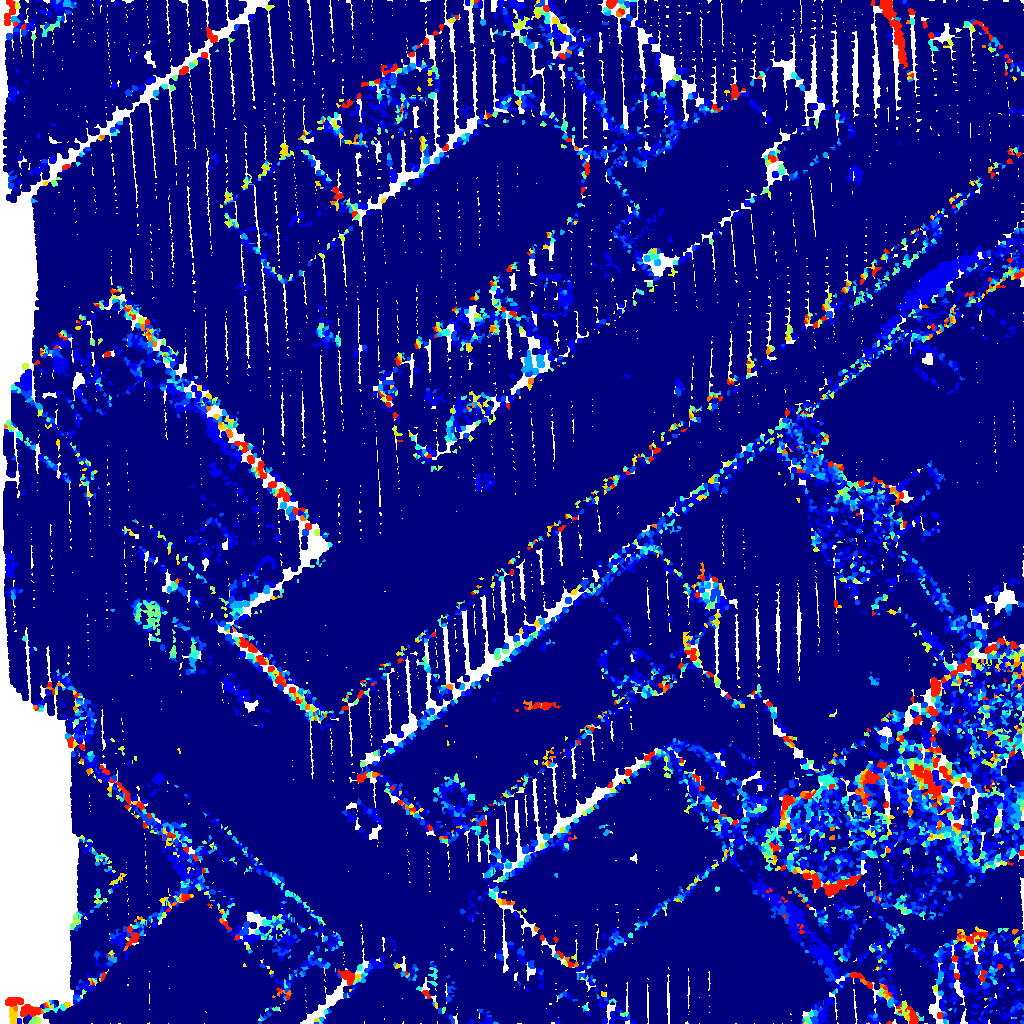}
		\includegraphics[width=\linewidth]{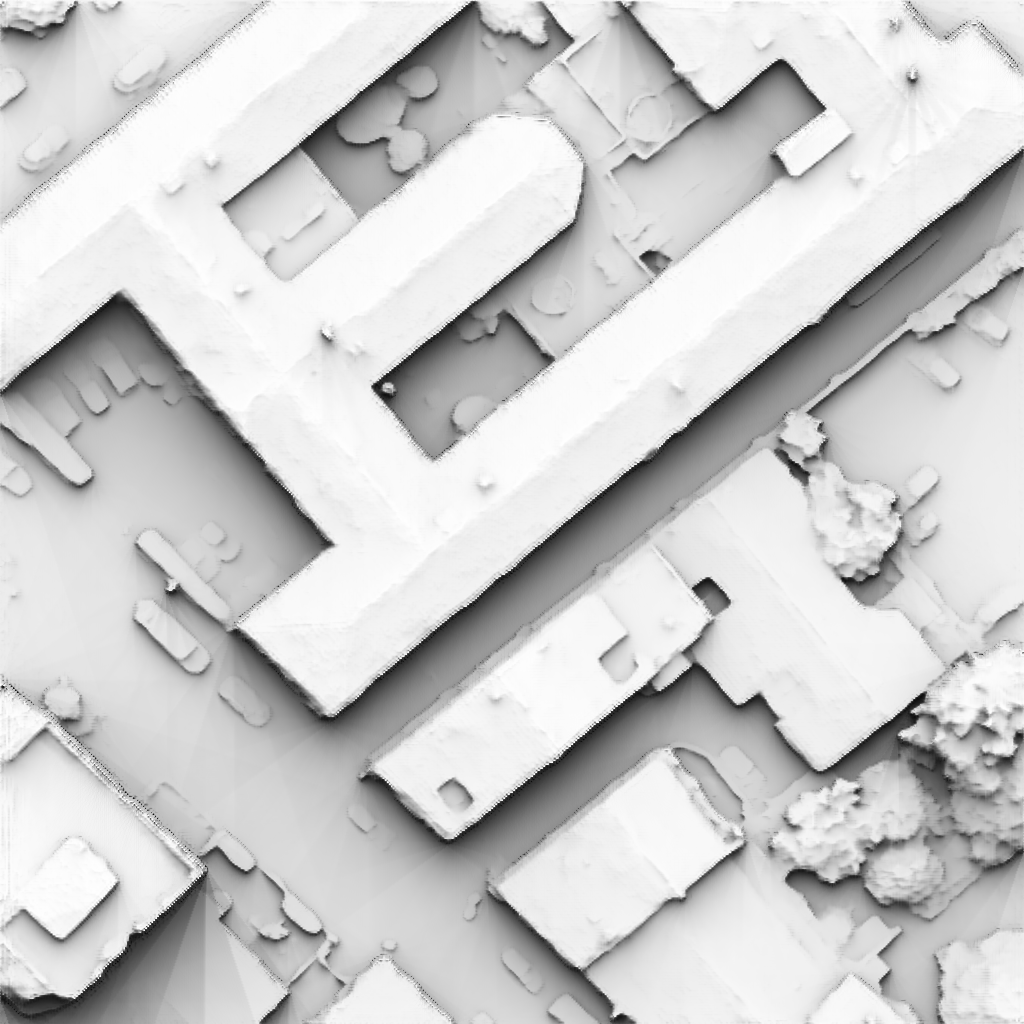}
		\centering{\tiny DeepPruner}
	\end{minipage}
	\begin{minipage}[t]{0.19\textwidth}
		\includegraphics[width=0.098\linewidth]{figures_supp/color_map.png}
		\includegraphics[width=0.85\linewidth]{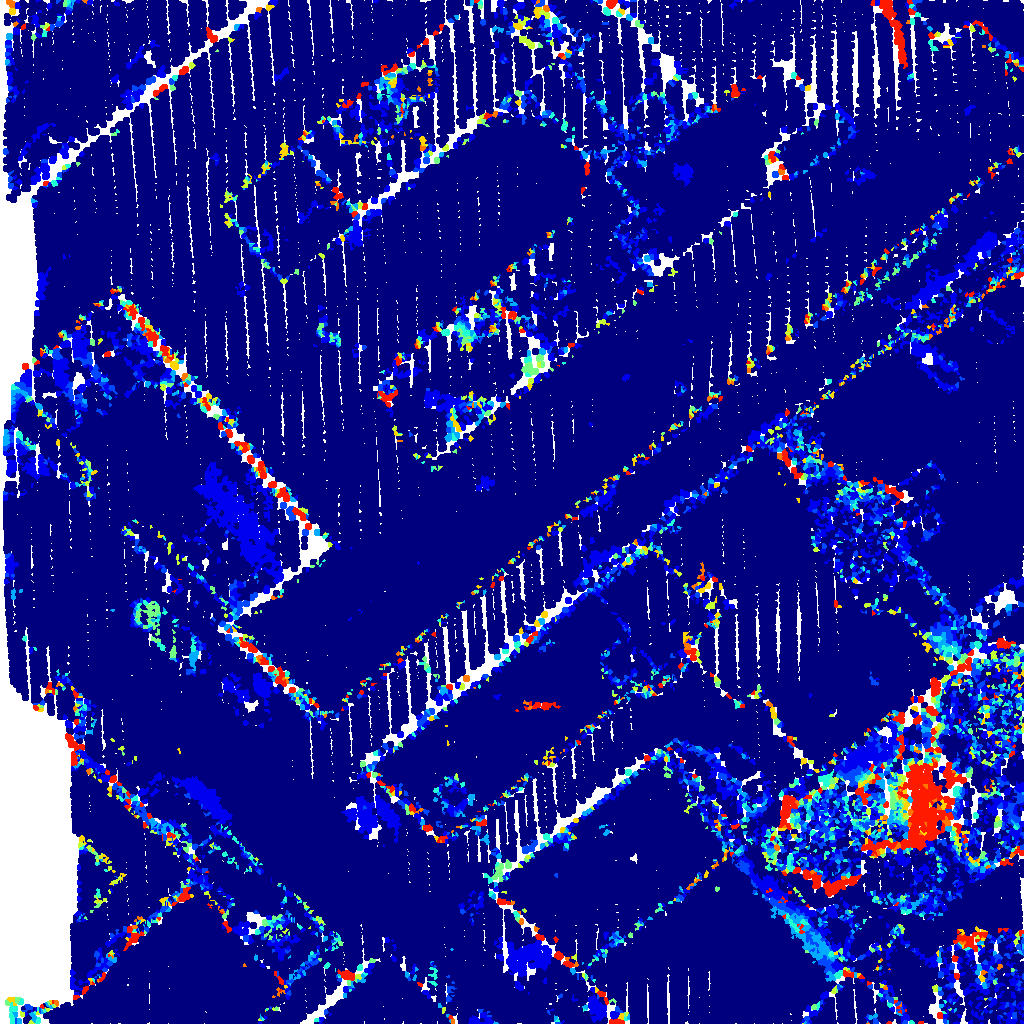}
		\includegraphics[width=\linewidth]{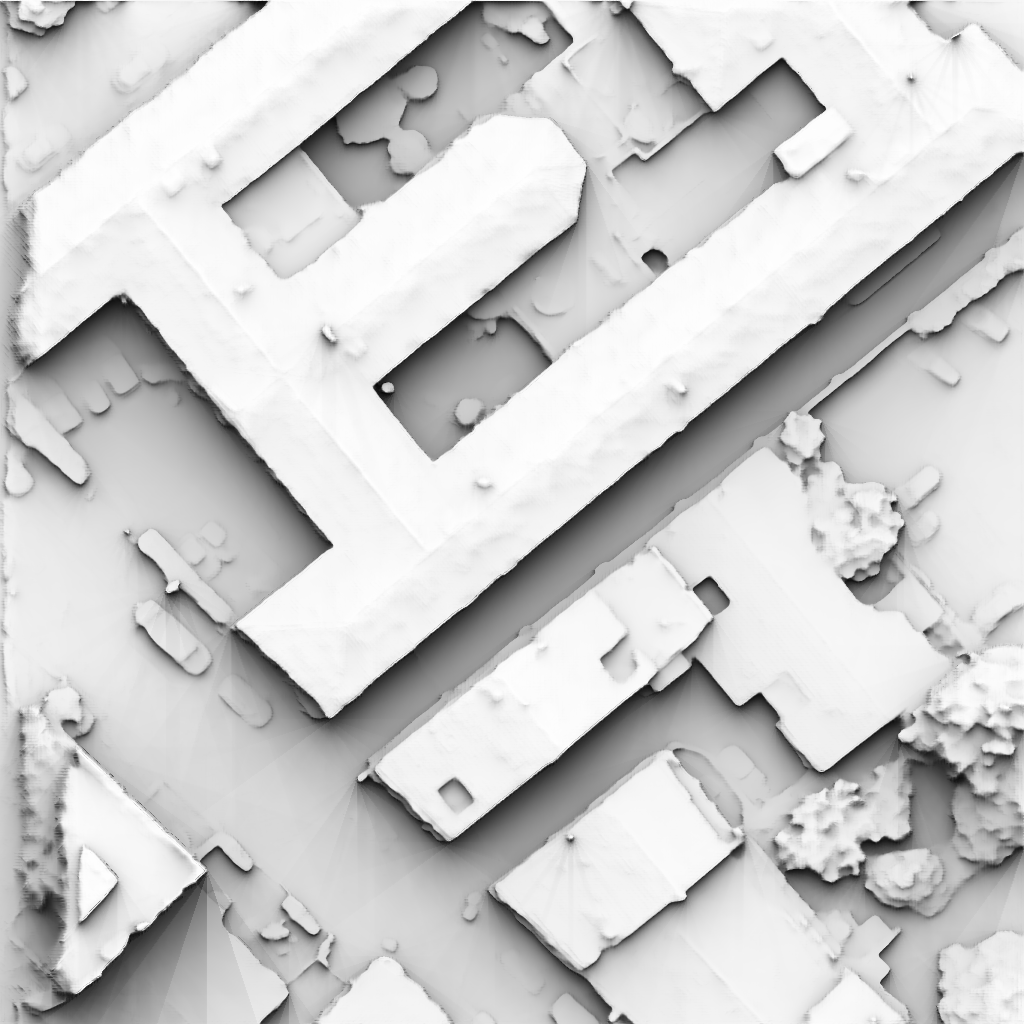}
		\centering{\tiny GANet}
	\end{minipage}
	\begin{minipage}[t]{0.19\textwidth}	
		\includegraphics[width=0.098\linewidth]{figures_supp/color_map.png}
		\includegraphics[width=0.85\linewidth]{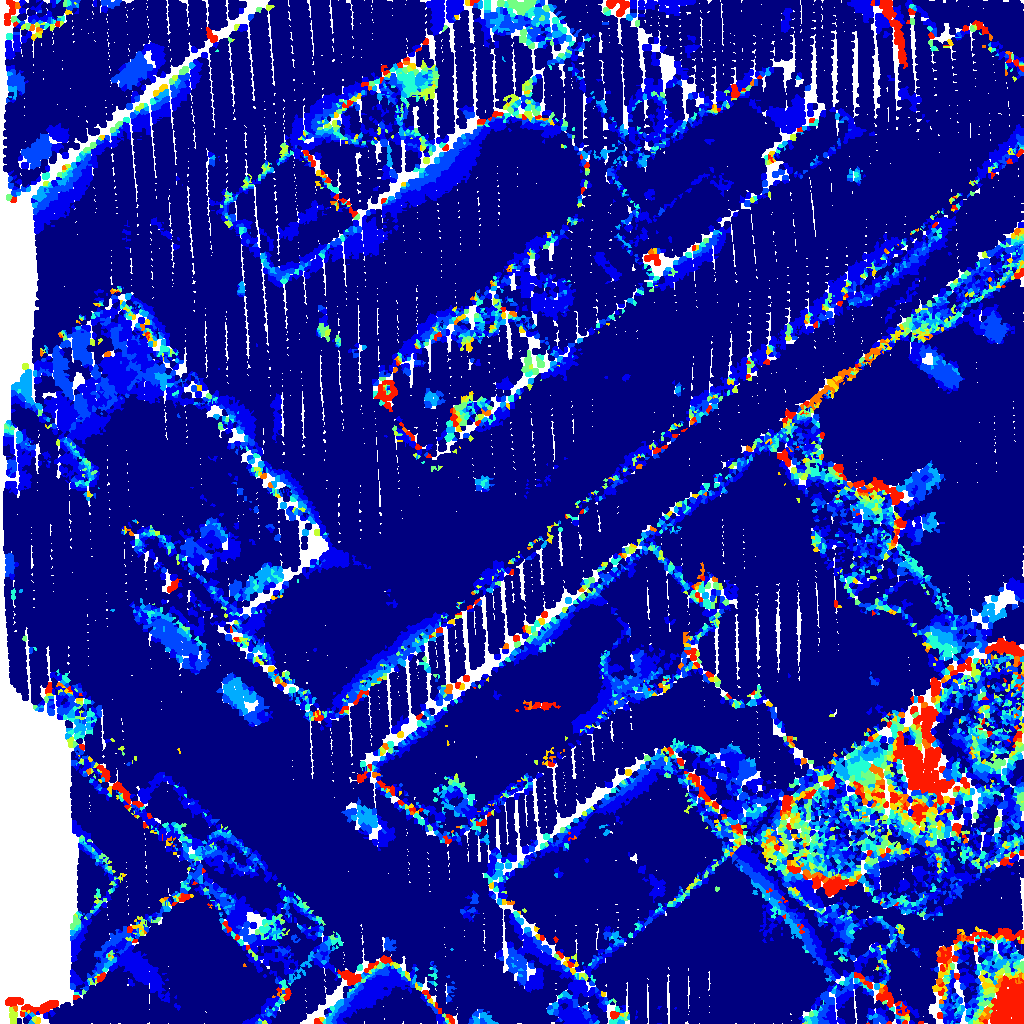}
		\includegraphics[width=\linewidth]{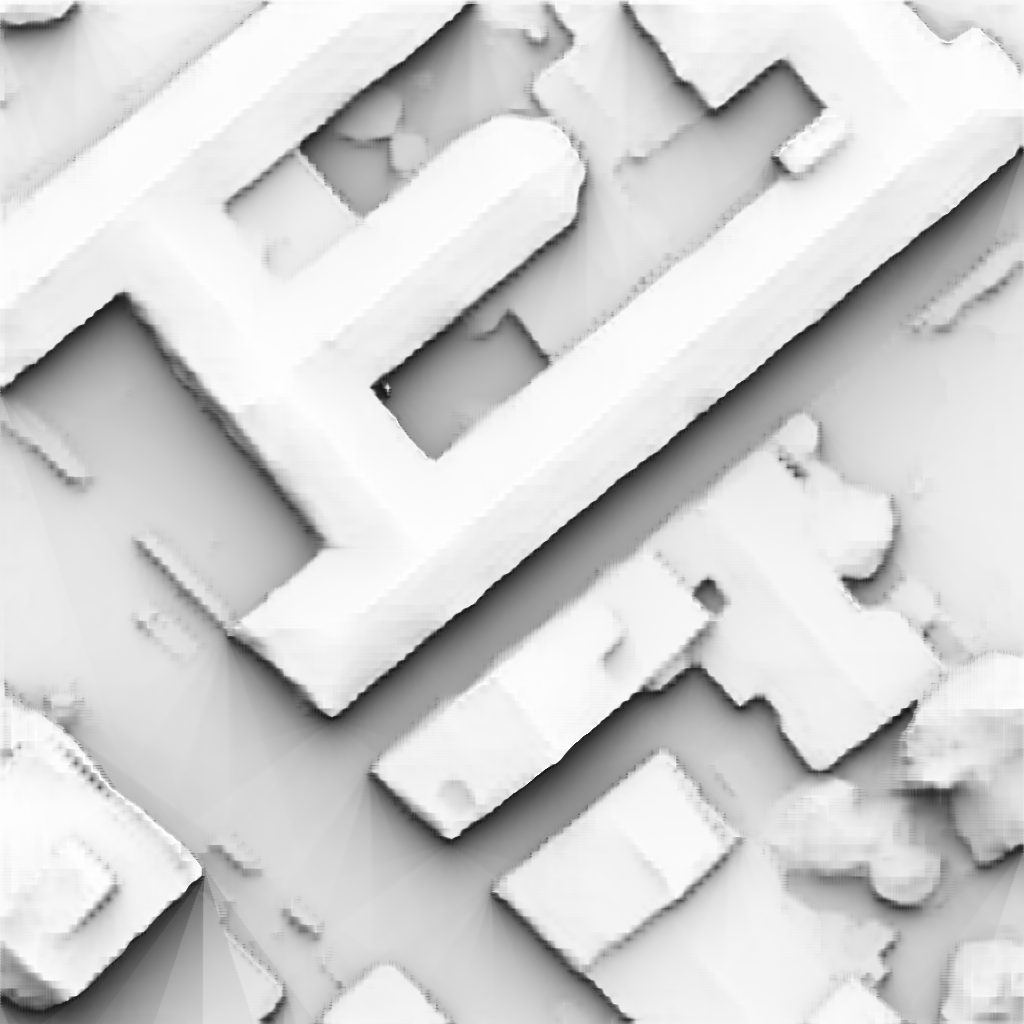}
		\centering{\tiny LEAStereo}
	\end{minipage}
	\caption{Error map and disparity visualization on building area for Toulouse Metropole.}
	\label{Figure.mlsebulding}
\end{figure}

\begin{figure}[tp]
	\begin{minipage}[t]{0.19\textwidth}
		\includegraphics[width=0.098\linewidth]{figures_supp/color_map.png}
		\includegraphics[width=0.85\linewidth]{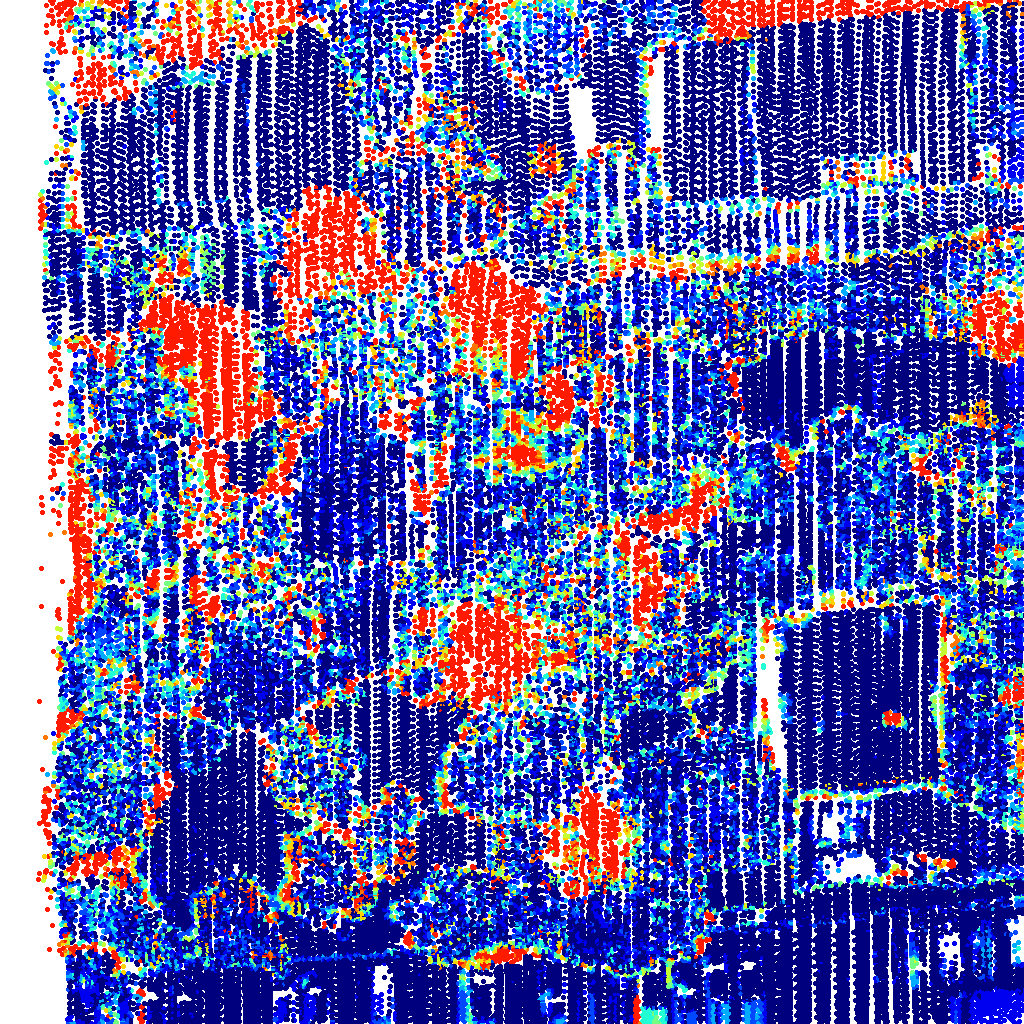}
		\includegraphics[width=\linewidth]{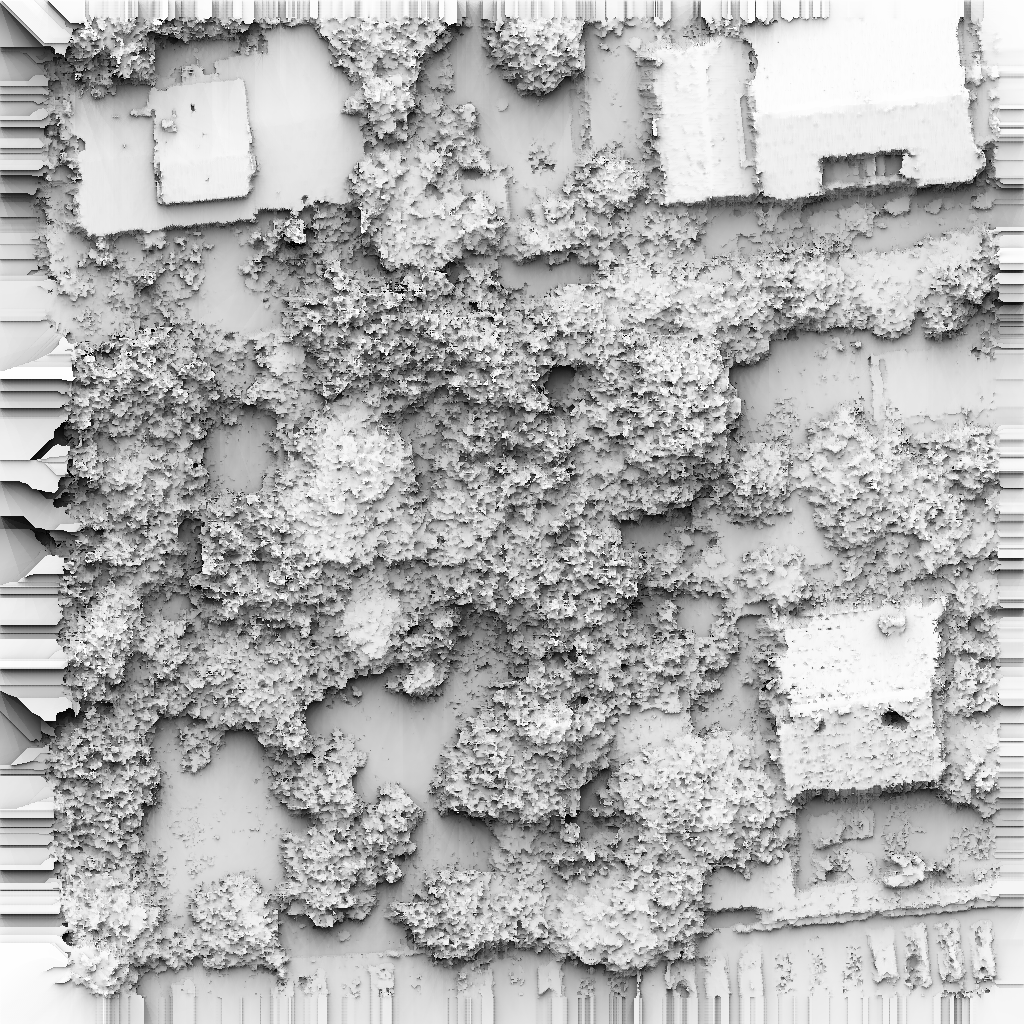}
		\centering{\tiny MICMAC}
	\end{minipage}
	\begin{minipage}[t]{0.19\textwidth}
		\includegraphics[width=0.098\linewidth]{figures_supp/color_map.png}
		\includegraphics[width=0.85\linewidth]{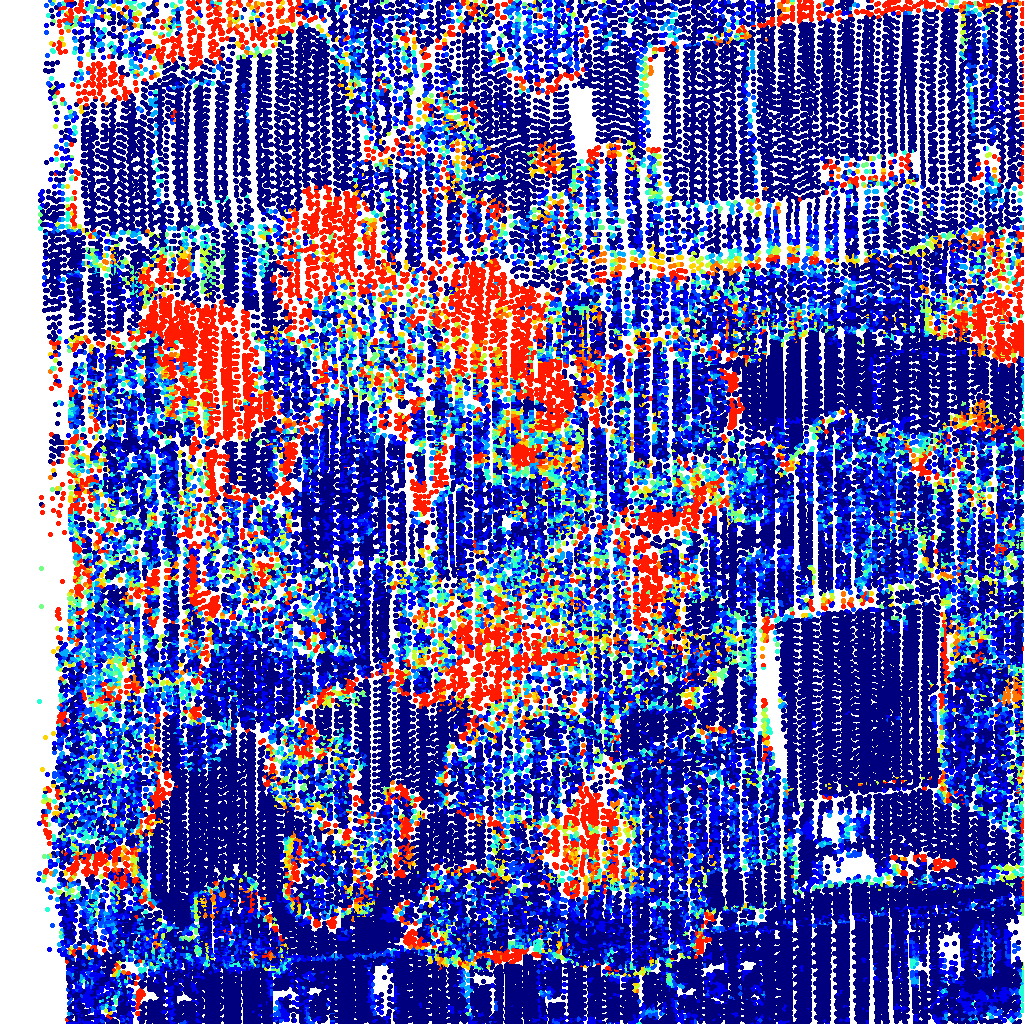}
		\includegraphics[width=\linewidth]{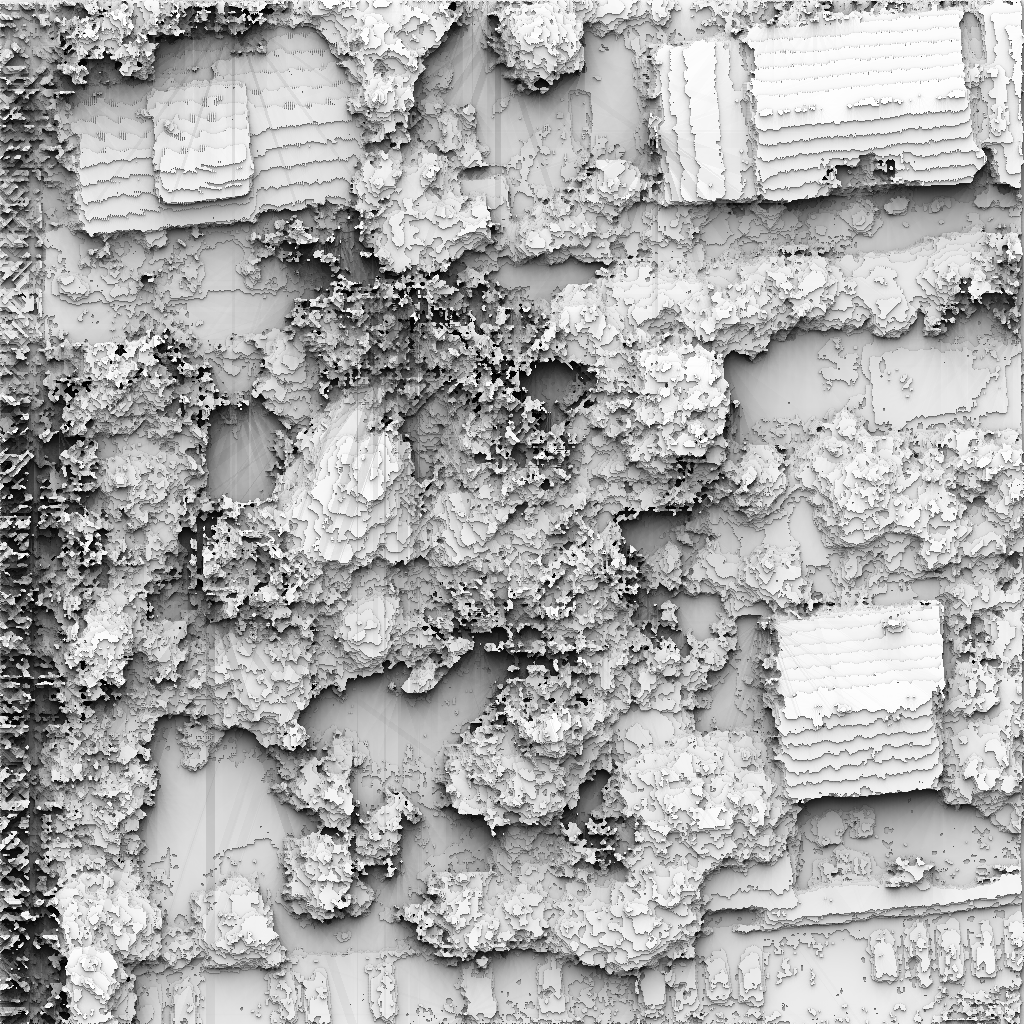}
		\centering{\tiny SGM(CUDA)}
	\end{minipage}
	\begin{minipage}[t]{0.19\textwidth}
		\includegraphics[width=0.098\linewidth]{figures_supp/color_map.png}
		\includegraphics[width=0.85\linewidth]{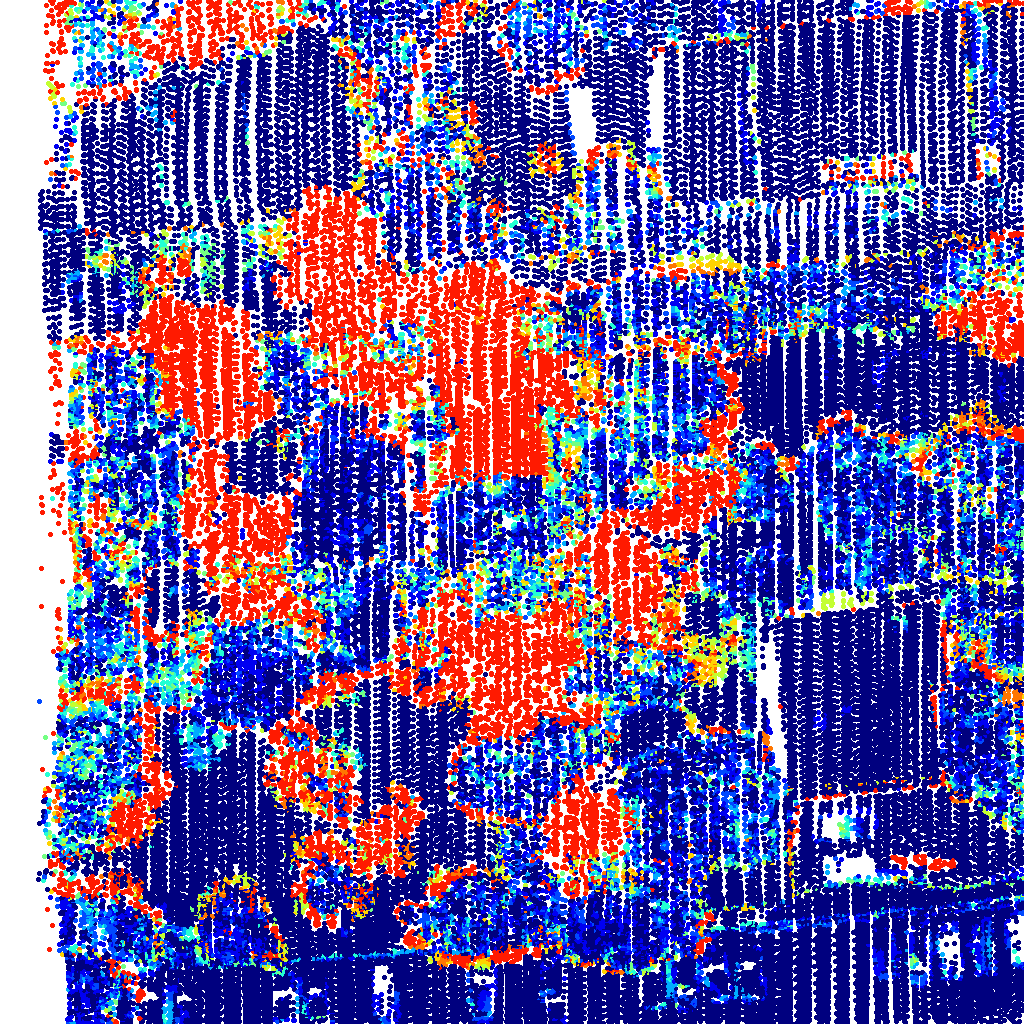}
		\includegraphics[width=\linewidth]{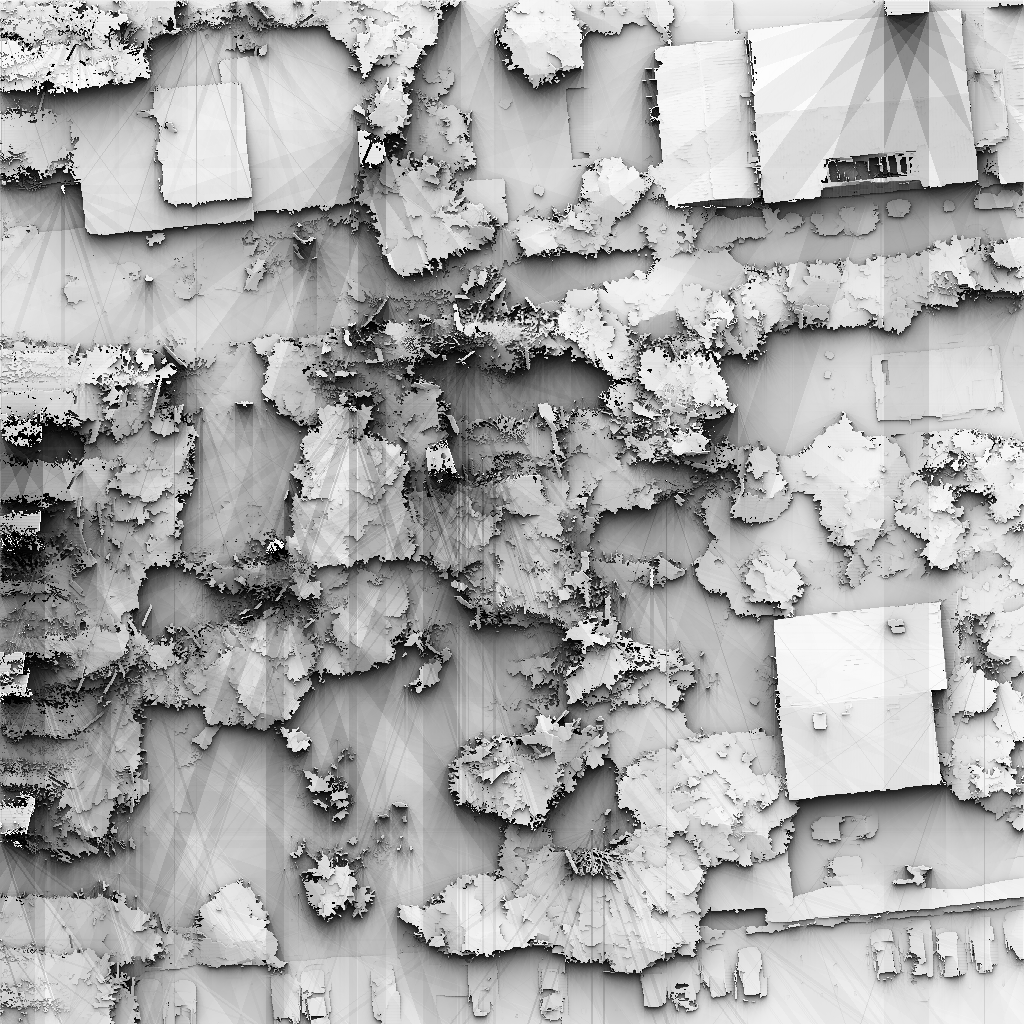}
		\centering{\tiny GraphCuts}
	\end{minipage}
	\begin{minipage}[t]{0.19\textwidth}
		\includegraphics[width=0.098\linewidth]{figures_supp/color_map.png}
		\includegraphics[width=0.85\linewidth]{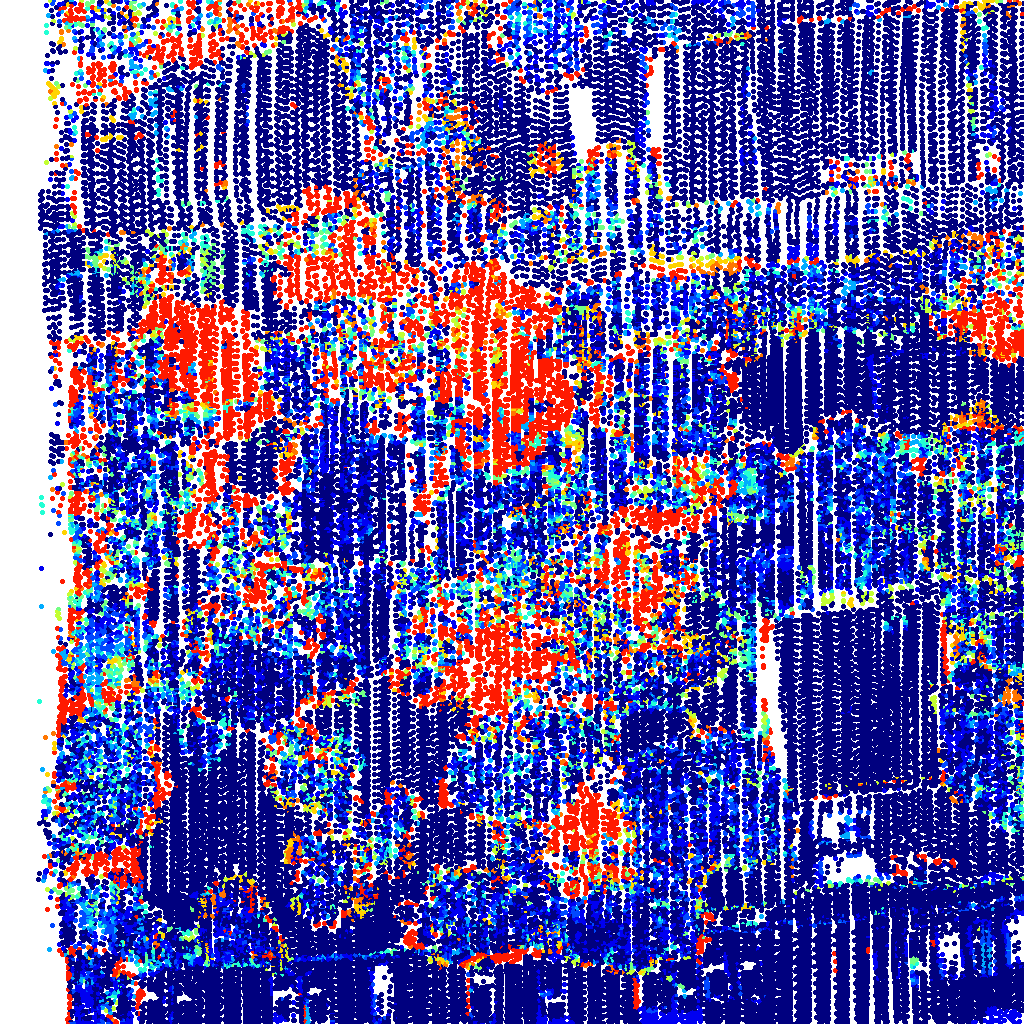}
		\includegraphics[width=\linewidth]{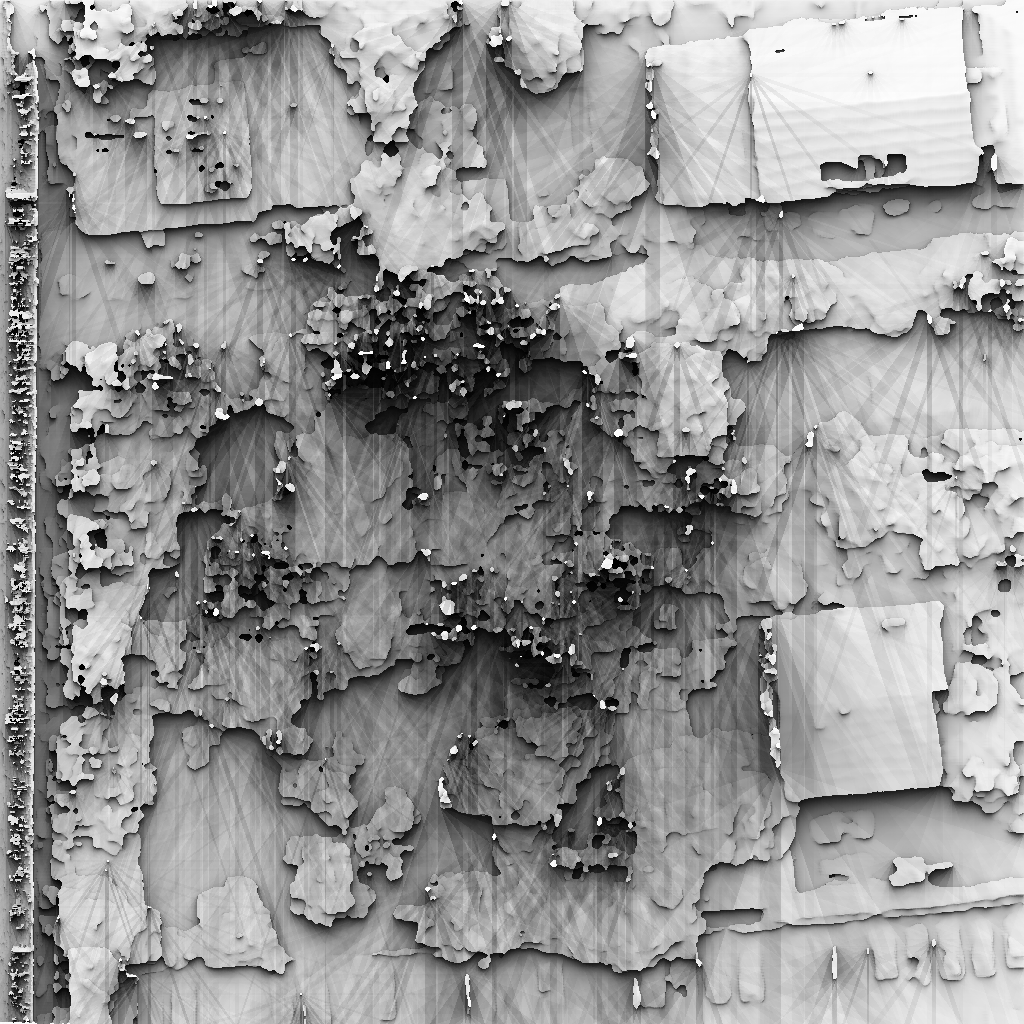}
		\centering{\tiny CBMV(SGM)}
	\end{minipage}
	\begin{minipage}[t]{0.19\textwidth}
		\includegraphics[width=0.098\linewidth]{figures_supp/color_map.png}
		\includegraphics[width=0.85\linewidth]{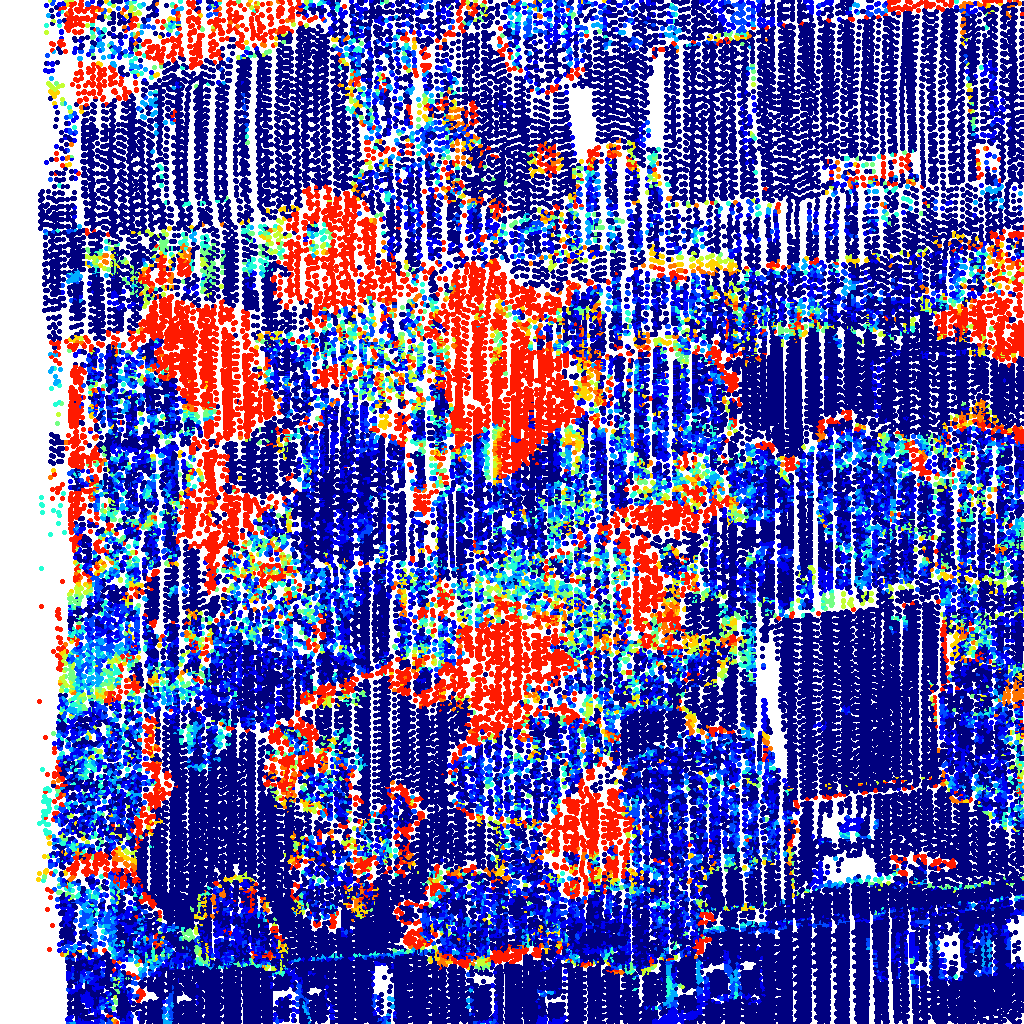}
		\includegraphics[width=\linewidth]{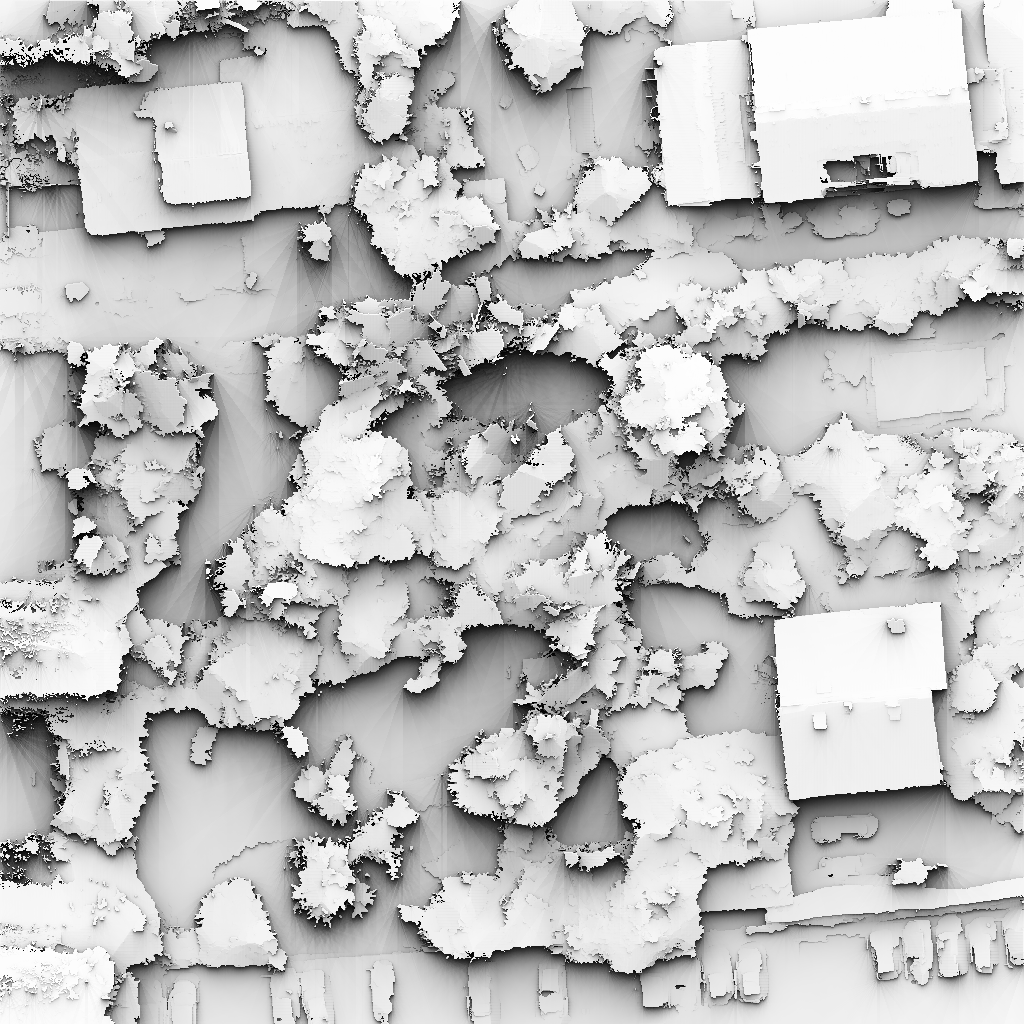}
		\centering{\tiny CBMV(GraphCuts)}
	\end{minipage}
	\begin{minipage}[t]{0.19\textwidth}
		\includegraphics[width=0.098\linewidth]{figures_supp/color_map.png}
		\includegraphics[width=0.85\linewidth]{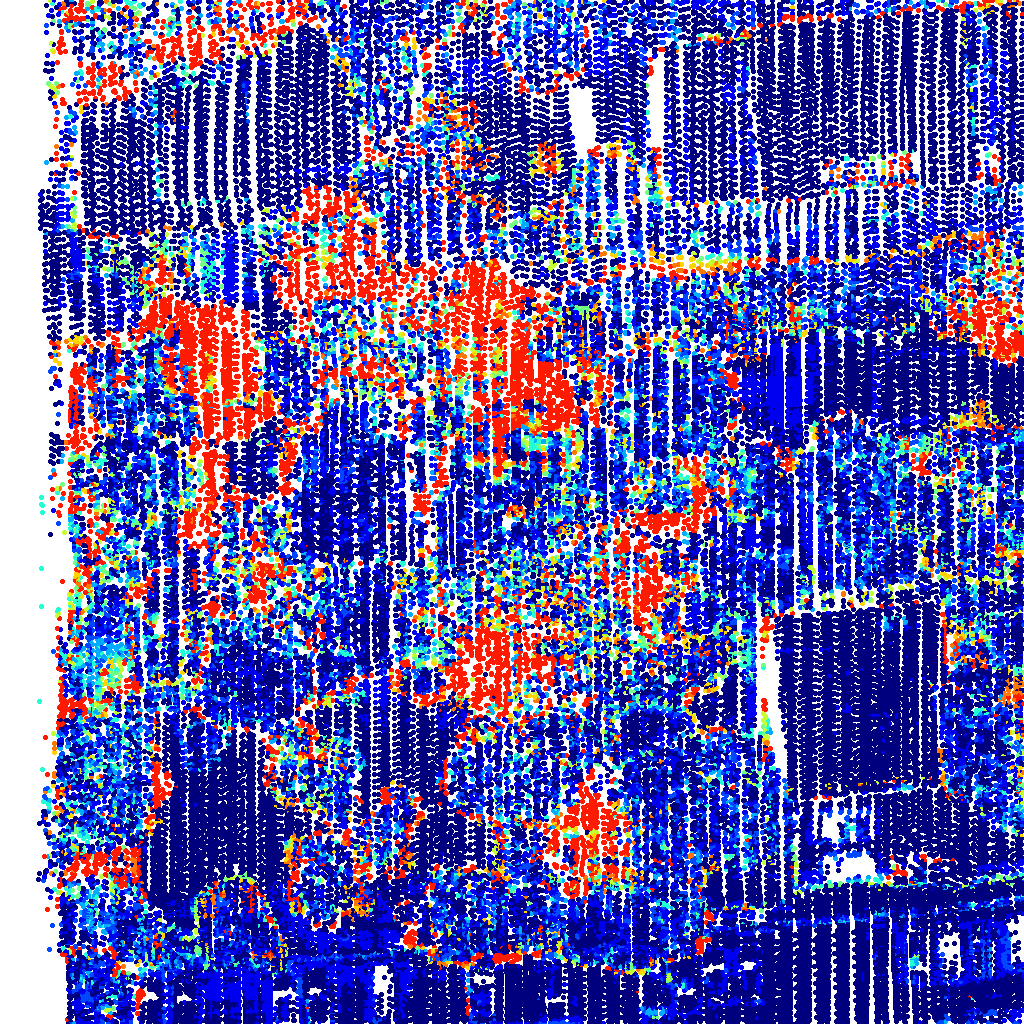}
		\includegraphics[width=\linewidth]{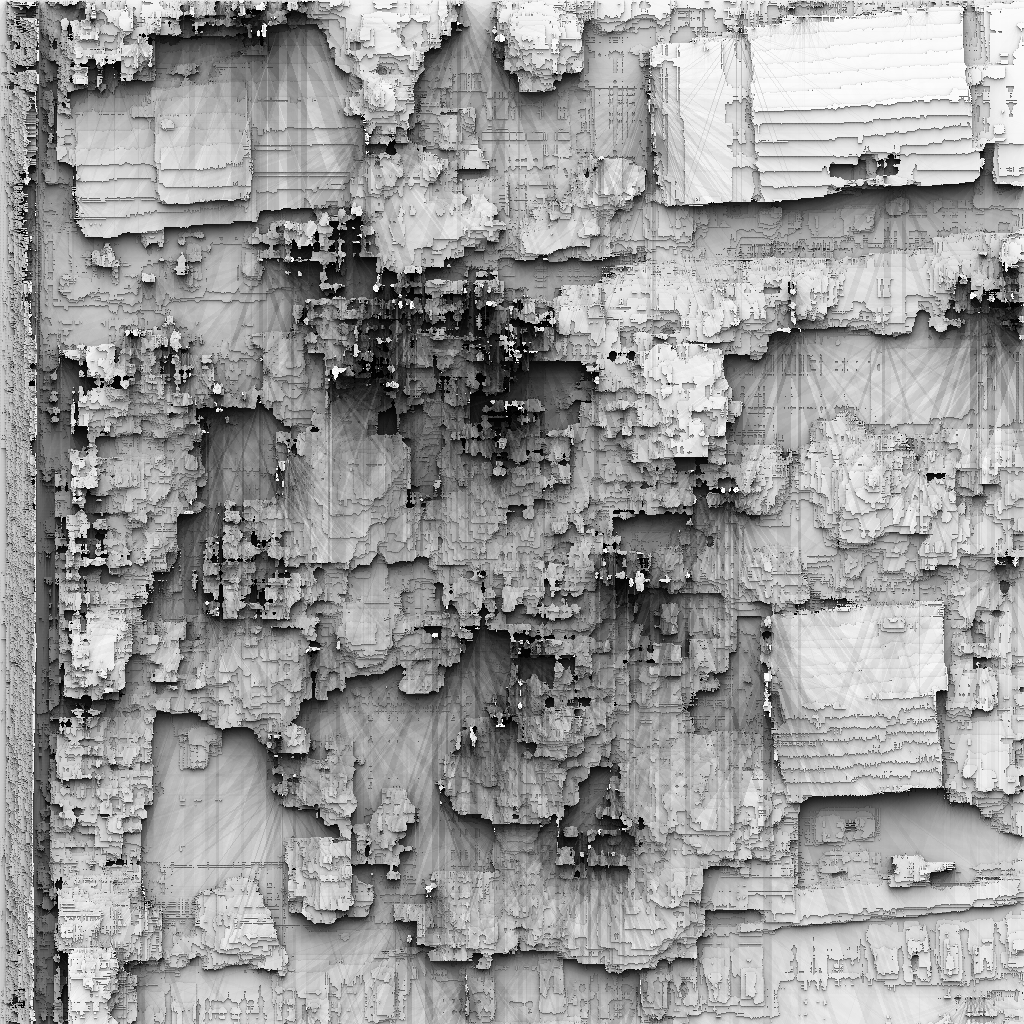}
		\centering{\tiny MC-CNN(KITTI)}
	\end{minipage}
	\begin{minipage}[t]{0.19\textwidth}
		\includegraphics[width=0.098\linewidth]{figures_supp/color_map.png}
		\includegraphics[width=0.85\linewidth]{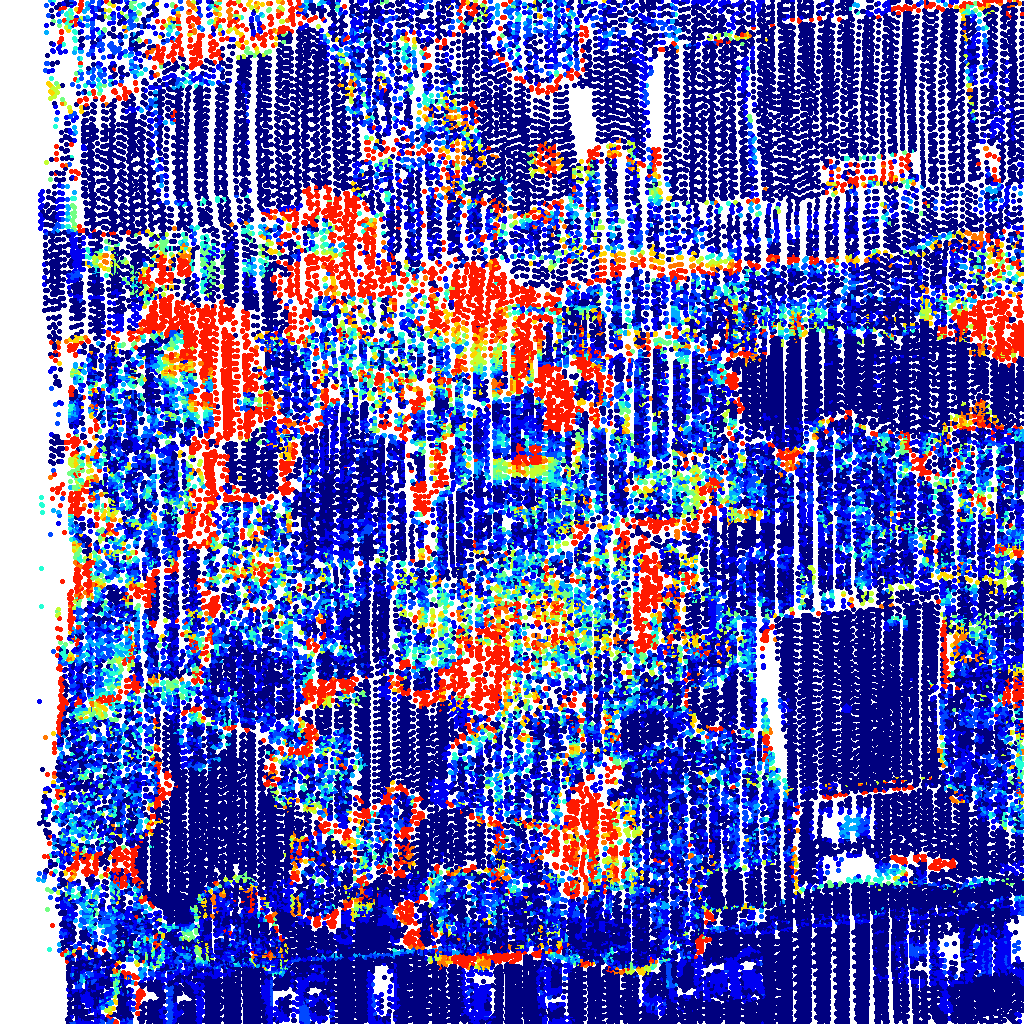}
		\includegraphics[width=\linewidth]{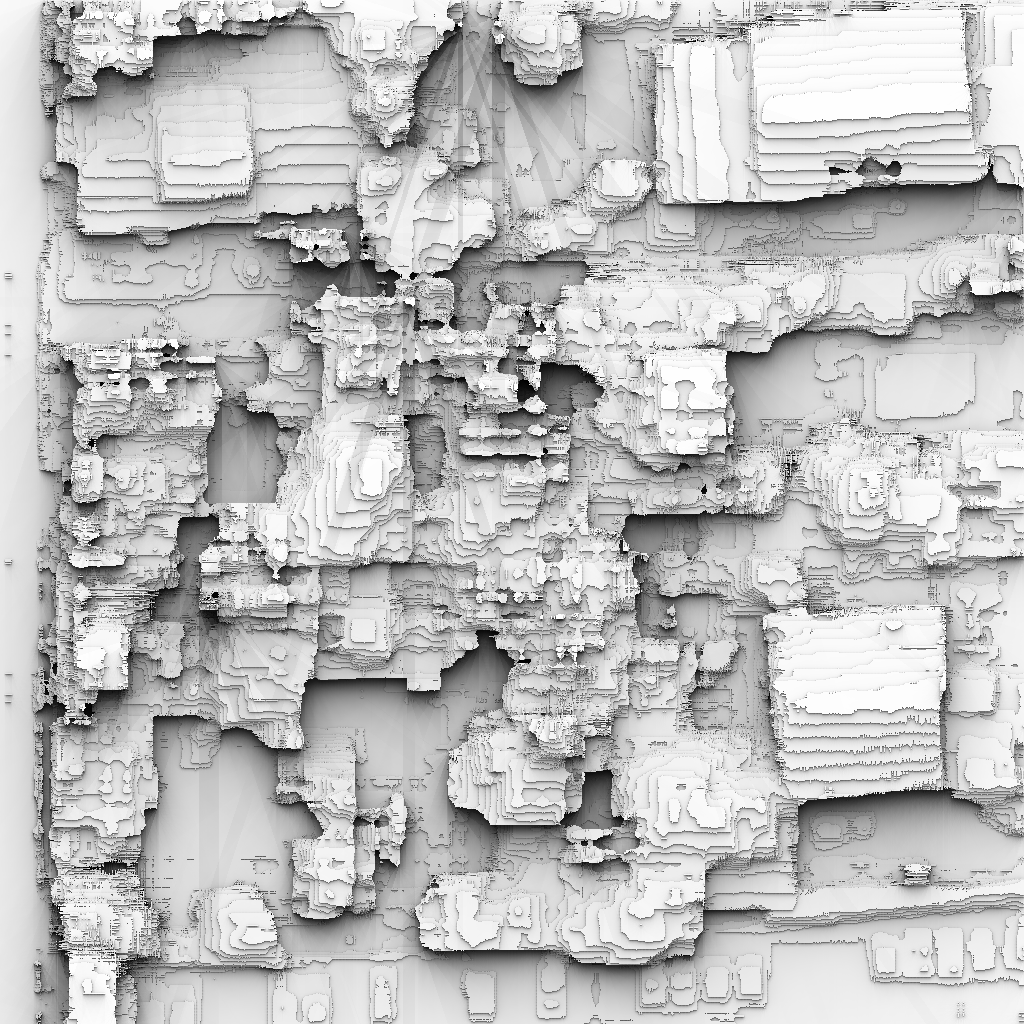}
		\centering{\tiny DeepFeature(KITTI)}
	\end{minipage}
	\begin{minipage}[t]{0.19\textwidth}
		\includegraphics[width=0.098\linewidth]{figures_supp/color_map.png}
		\includegraphics[width=0.85\linewidth]{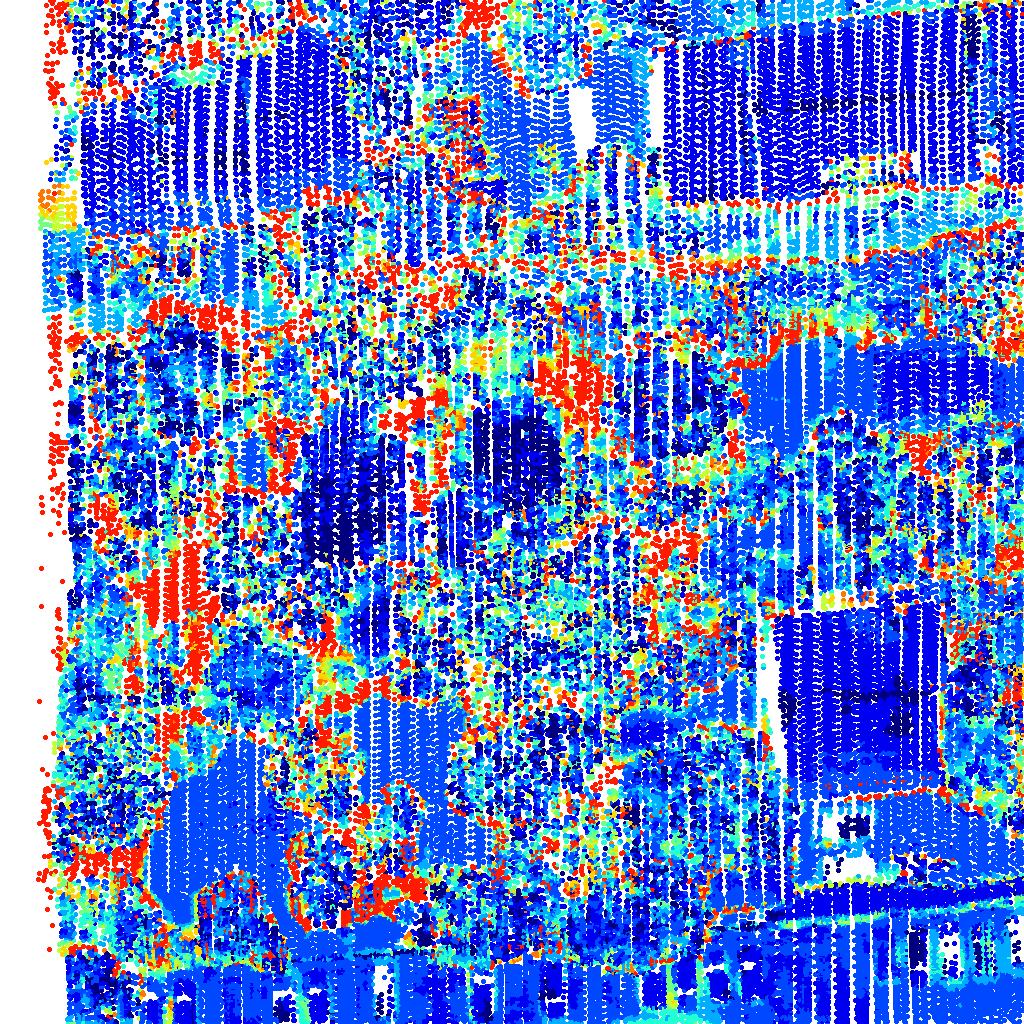}
		\includegraphics[width=\linewidth]{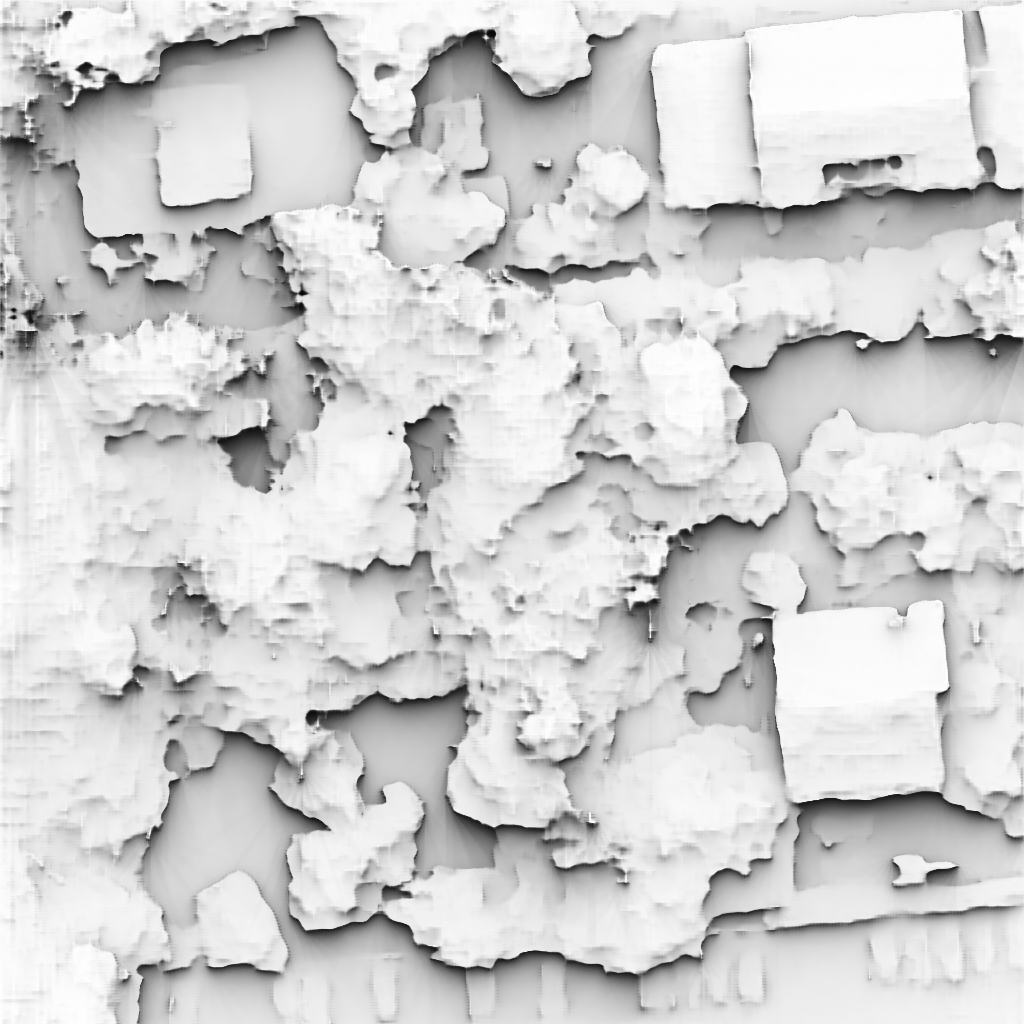}
		\centering{\tiny PSM net(KITTI)}
	\end{minipage}
	\begin{minipage}[t]{0.19\textwidth}
		\includegraphics[width=0.098\linewidth]{figures_supp/color_map.png}
		\includegraphics[width=0.85\linewidth]{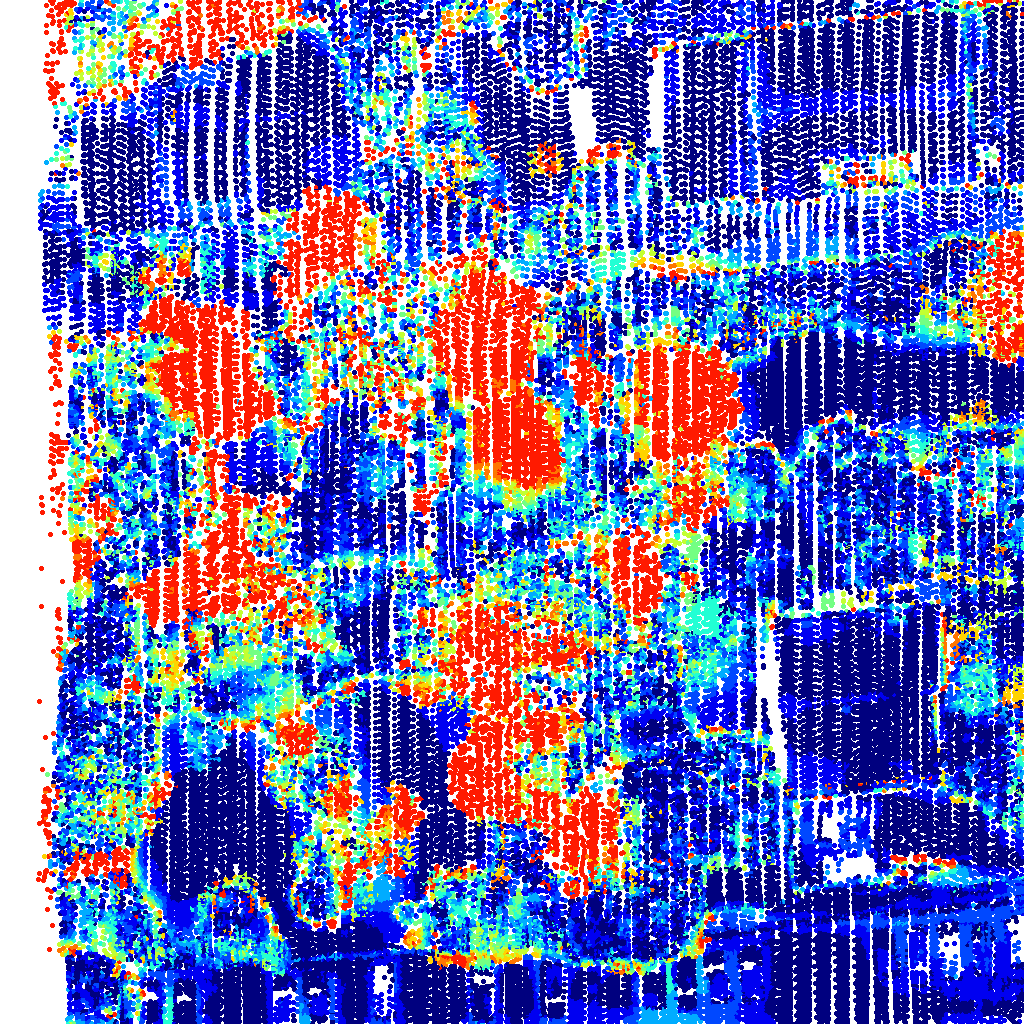}
		\includegraphics[width=\linewidth]{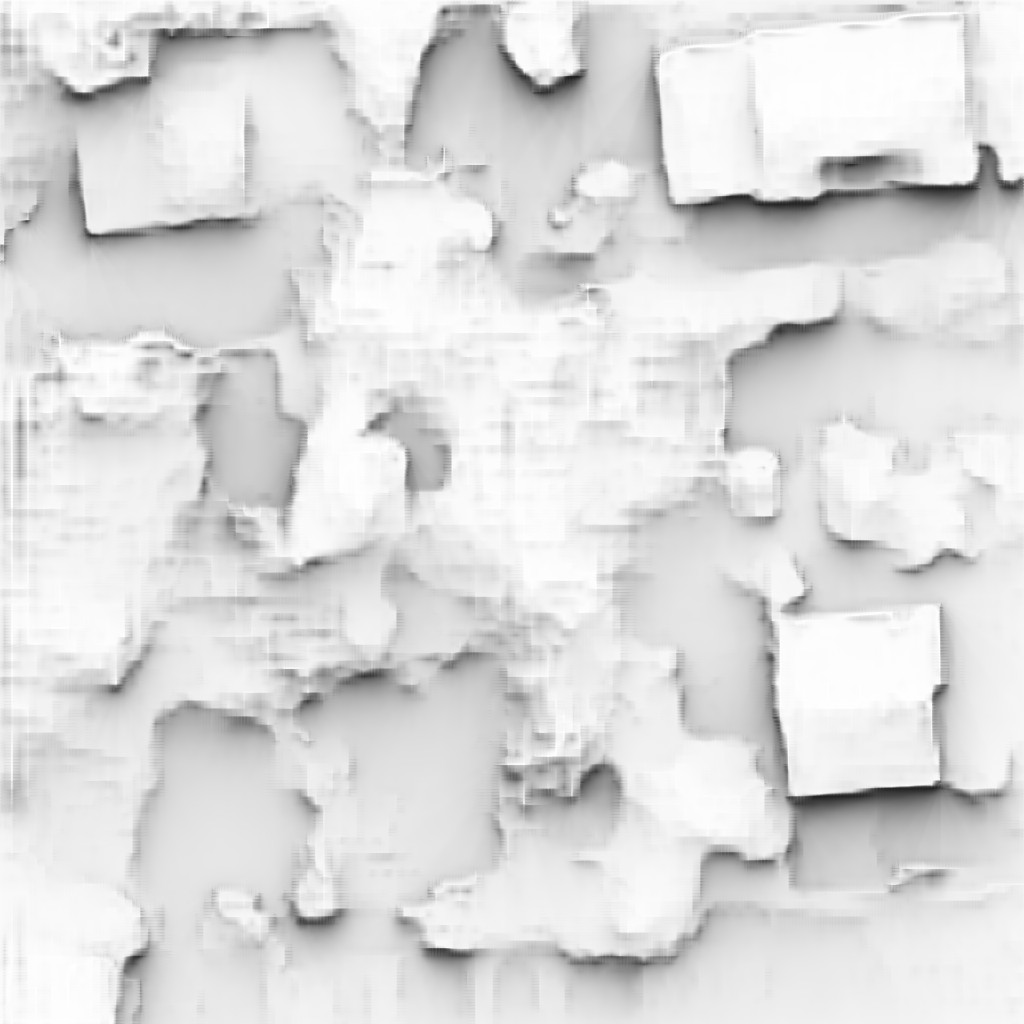}
		\centering{\tiny HRS net(KITTI)}
	\end{minipage}
	\begin{minipage}[t]{0.19\textwidth}	
		\includegraphics[width=0.098\linewidth]{figures_supp/color_map.png}
		\includegraphics[width=0.85\linewidth]{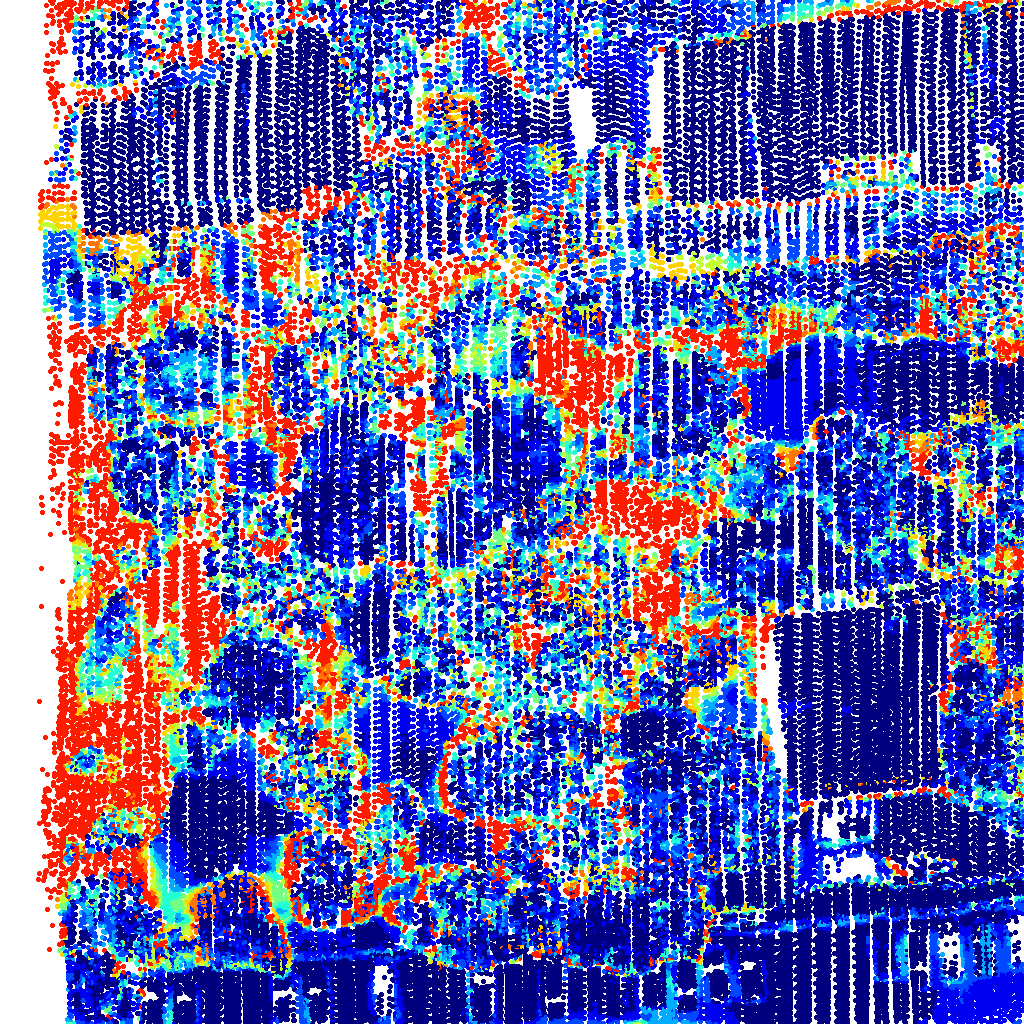}
		\includegraphics[width=\linewidth]{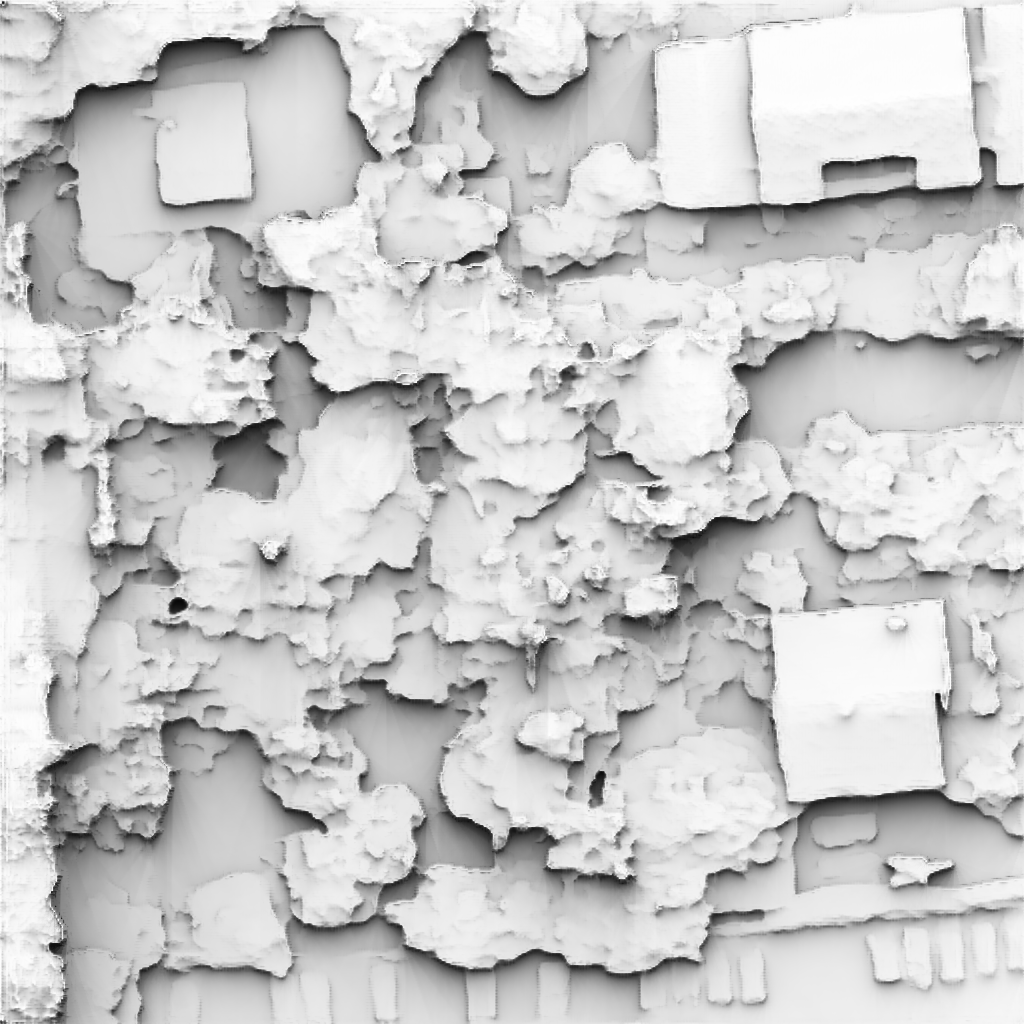}
		\centering{\tiny DeepPruner(KITTI)}
	\end{minipage}
	\begin{minipage}[t]{0.19\textwidth}
		\includegraphics[width=0.098\linewidth]{figures_supp/color_map.png}
		\includegraphics[width=0.85\linewidth]{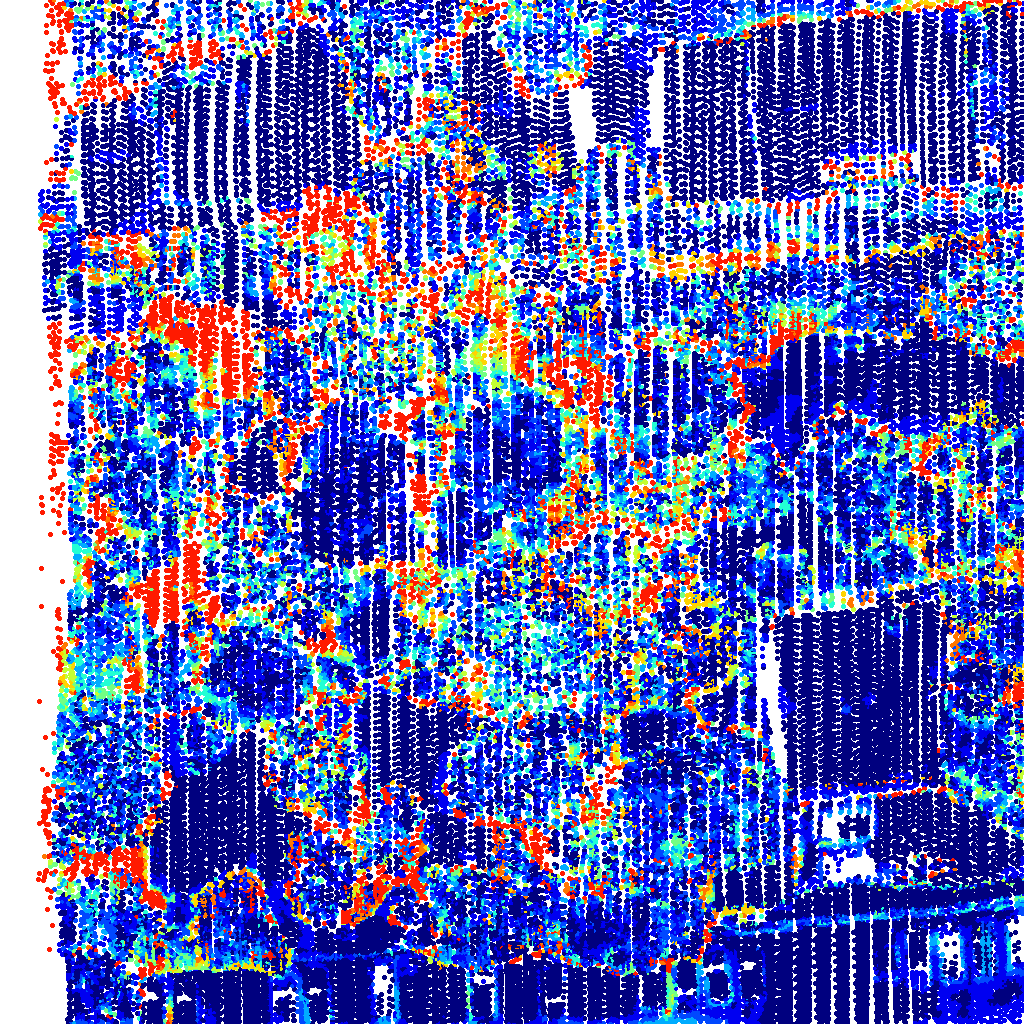}
		\includegraphics[width=\linewidth]{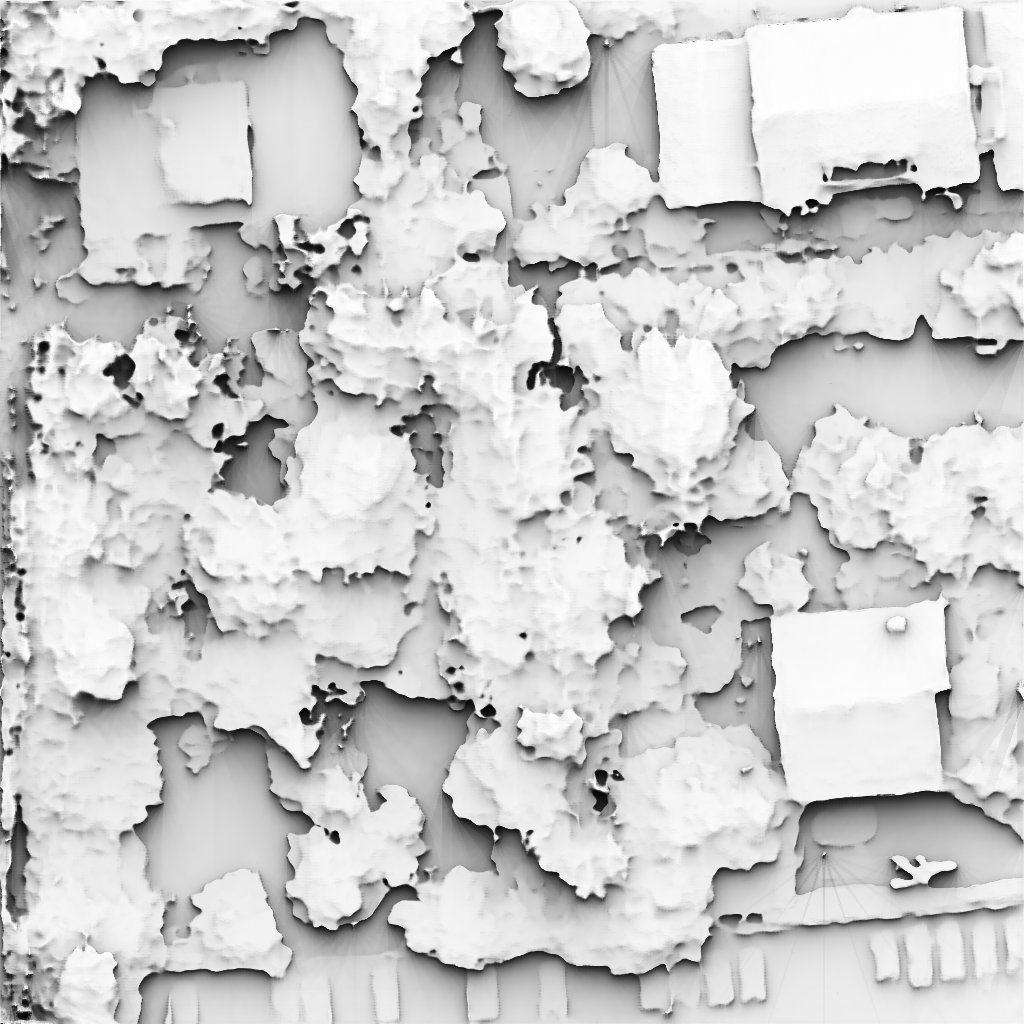}
		\centering{\tiny GANet(KITTI)}
	\end{minipage}
	\begin{minipage}[t]{0.19\textwidth}	
		\includegraphics[width=0.098\linewidth]{figures_supp/color_map.png}
		\includegraphics[width=0.85\linewidth]{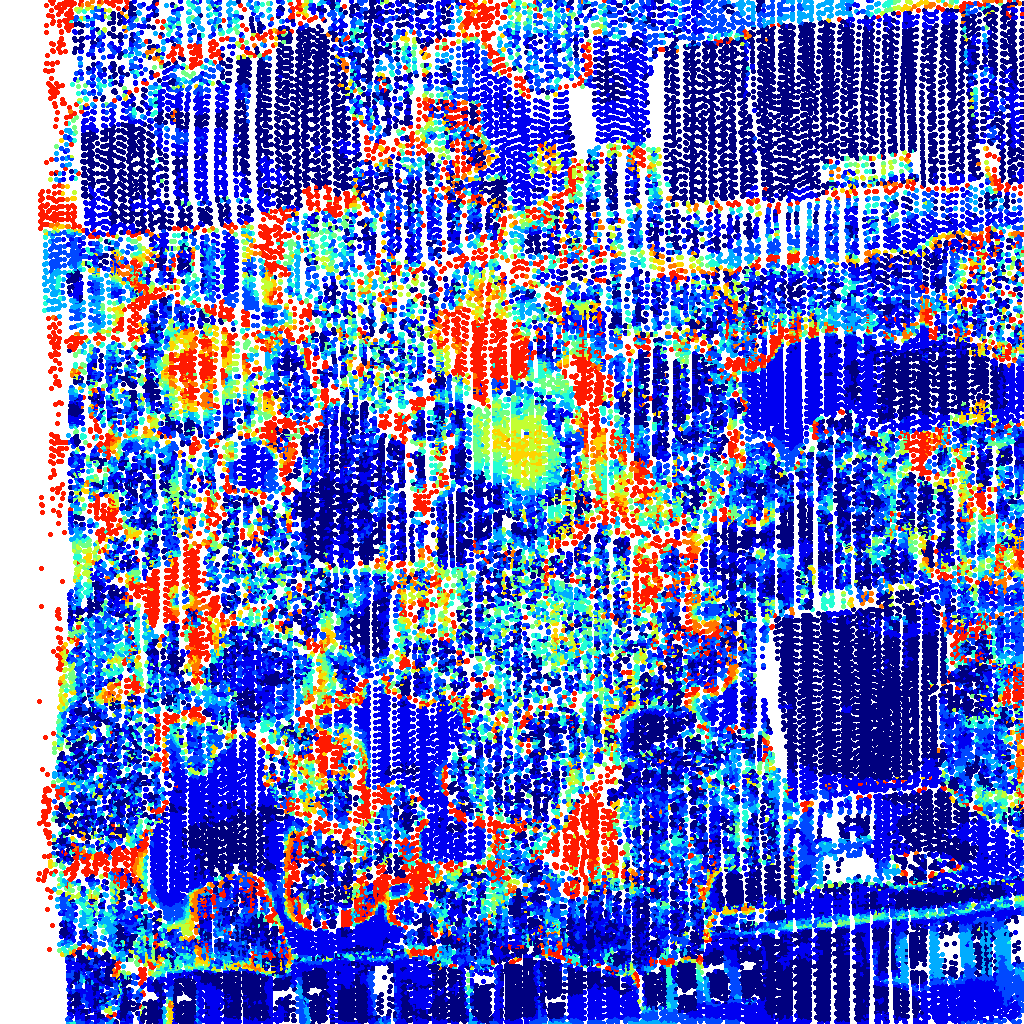}
		\includegraphics[width=\linewidth]{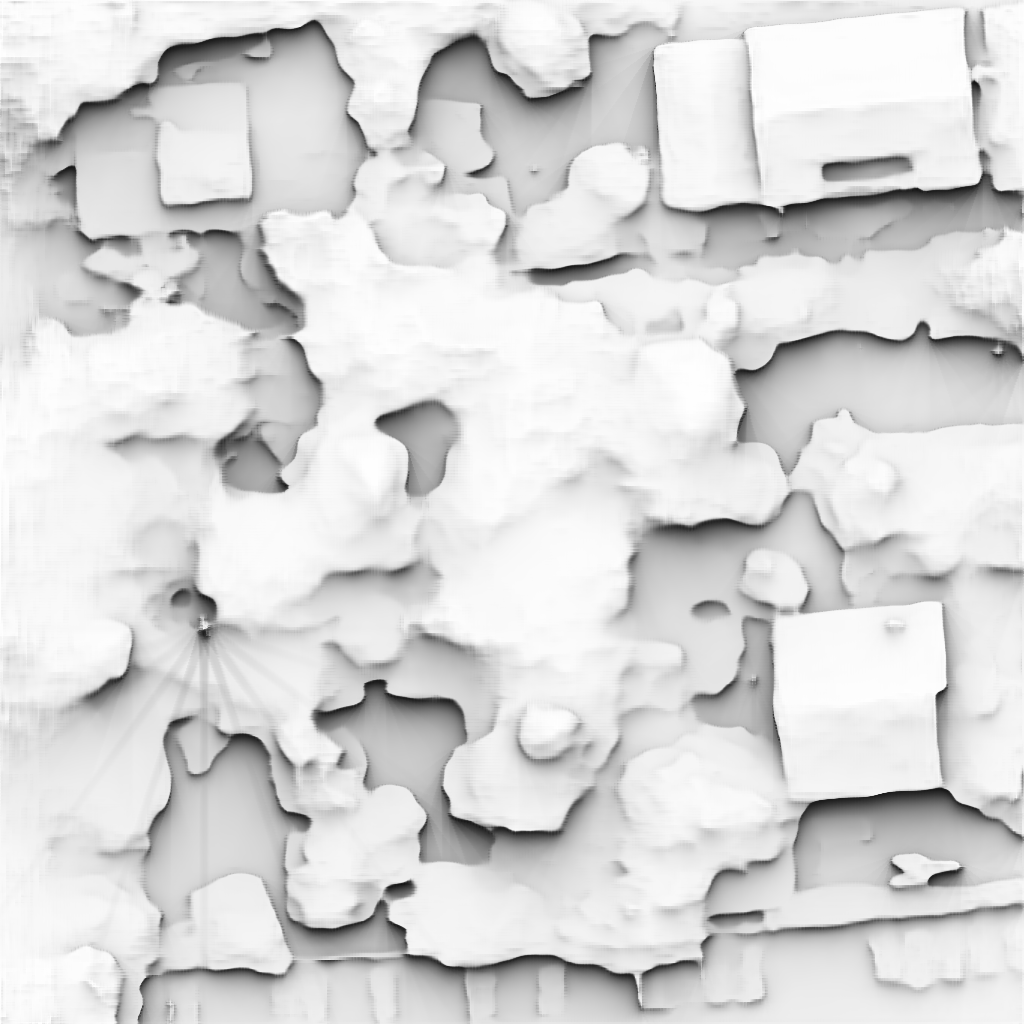}
		\centering{\tiny LEAStereo(KITTI)}
	\end{minipage}
	\begin{minipage}[t]{0.19\textwidth}
		\includegraphics[width=0.098\linewidth]{figures_supp/color_map.png}
		\includegraphics[width=0.85\linewidth]{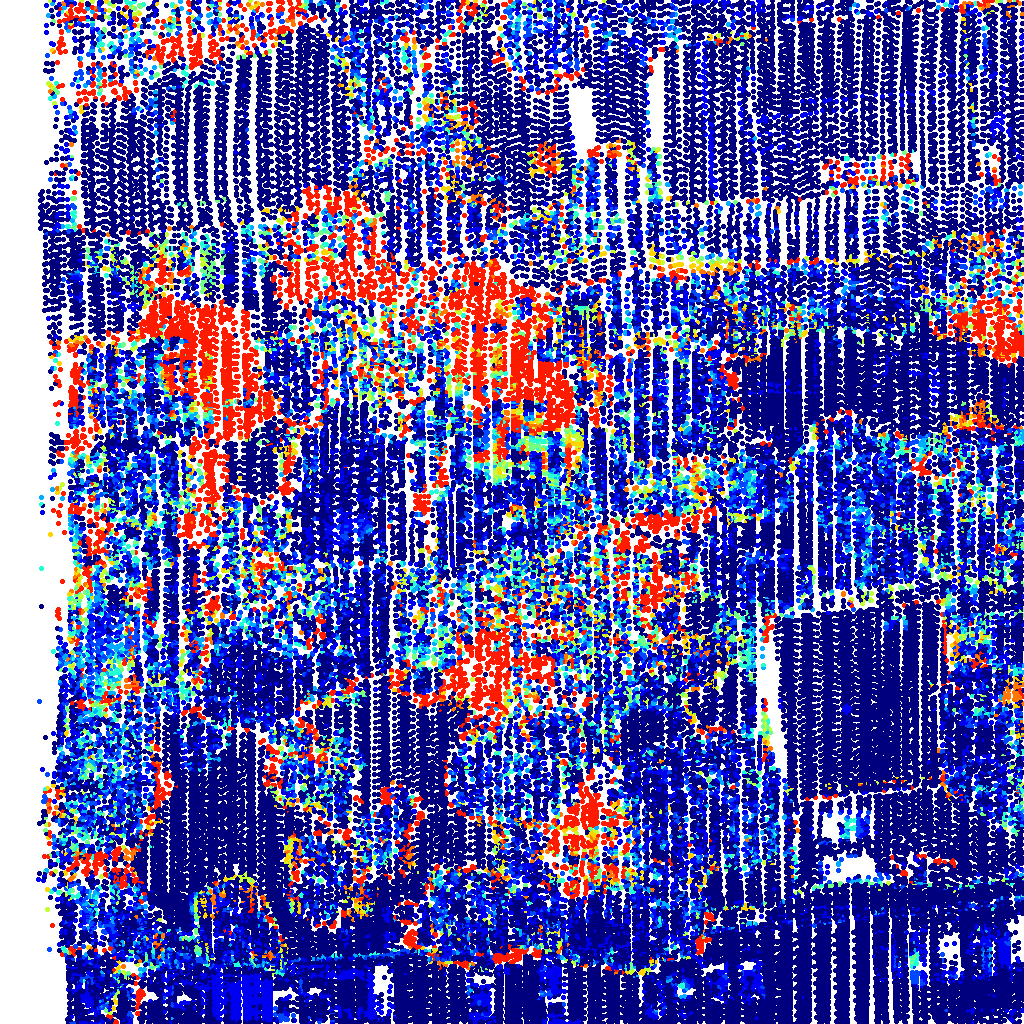}
		\includegraphics[width=\linewidth]{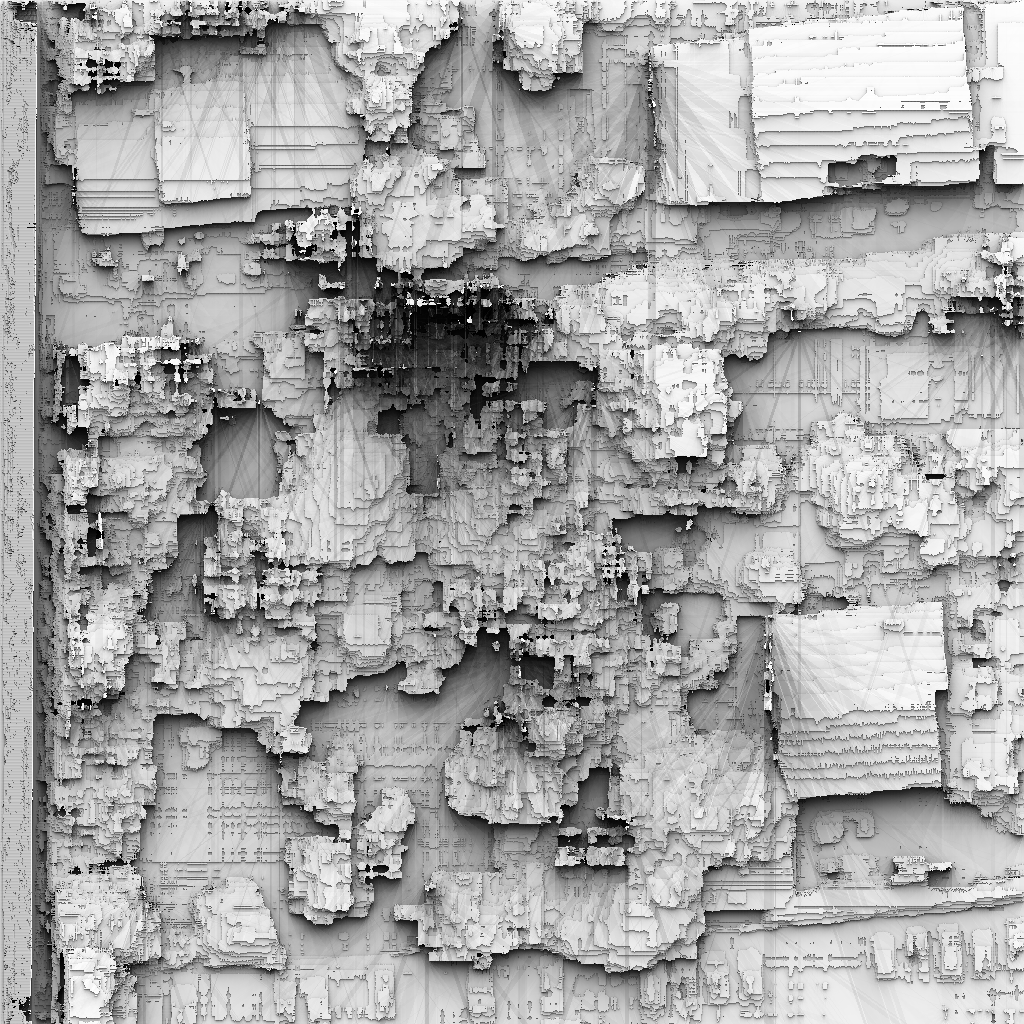}
		\centering{\tiny MC-CNN}
	\end{minipage}
	\begin{minipage}[t]{0.19\textwidth}
		\includegraphics[width=0.098\linewidth]{figures_supp/color_map.png}
		\includegraphics[width=0.85\linewidth]{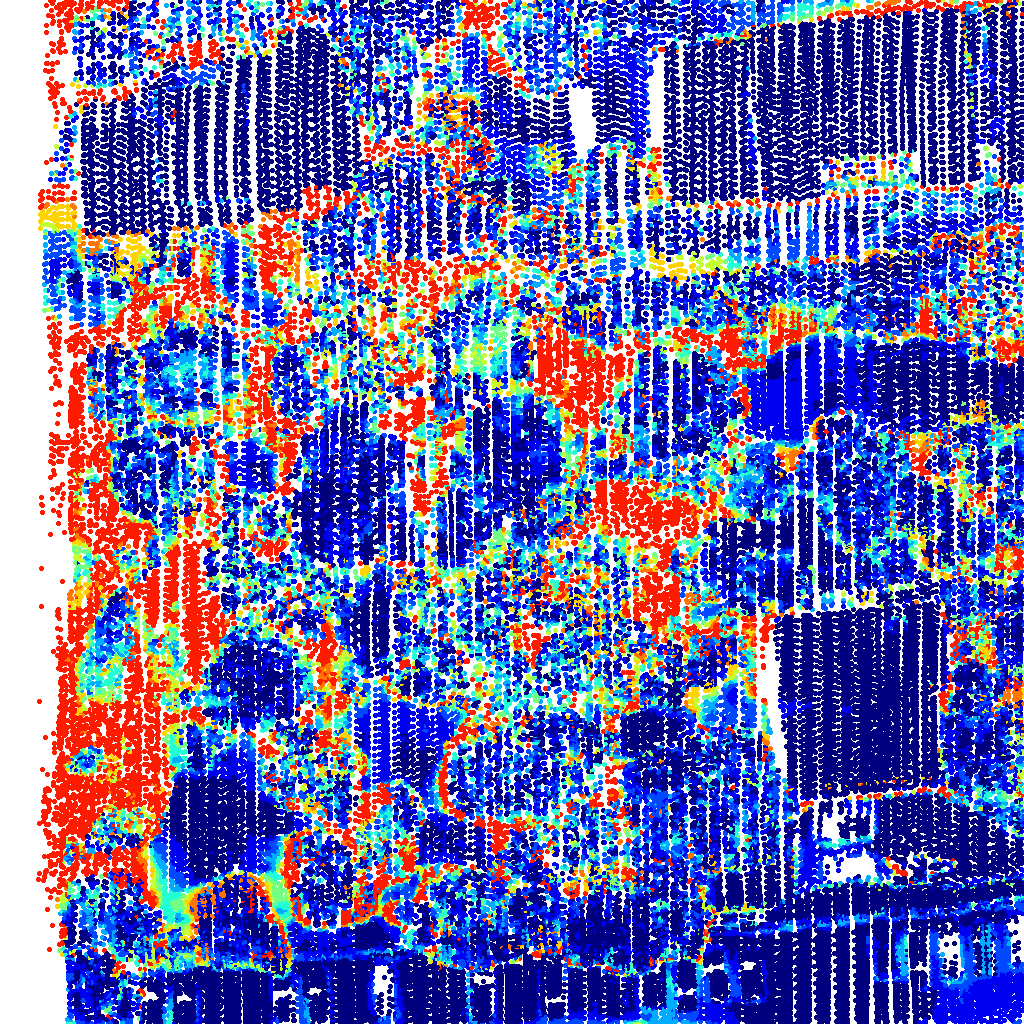}
		\includegraphics[width=\linewidth]{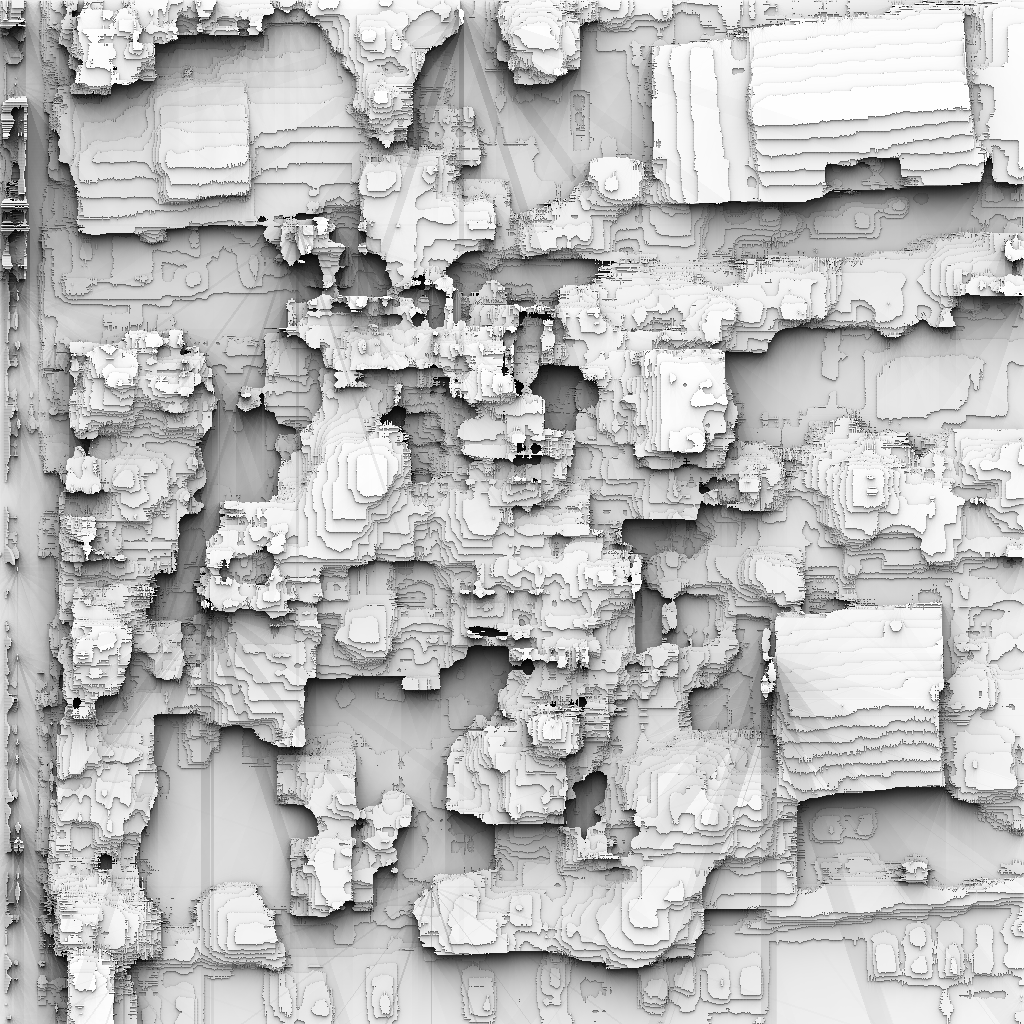}
		\centering{\tiny DeepFeature}
	\end{minipage}
	\begin{minipage}[t]{0.19\textwidth}
		\includegraphics[width=0.098\linewidth]{figures_supp/color_map.png}
		\includegraphics[width=0.85\linewidth]{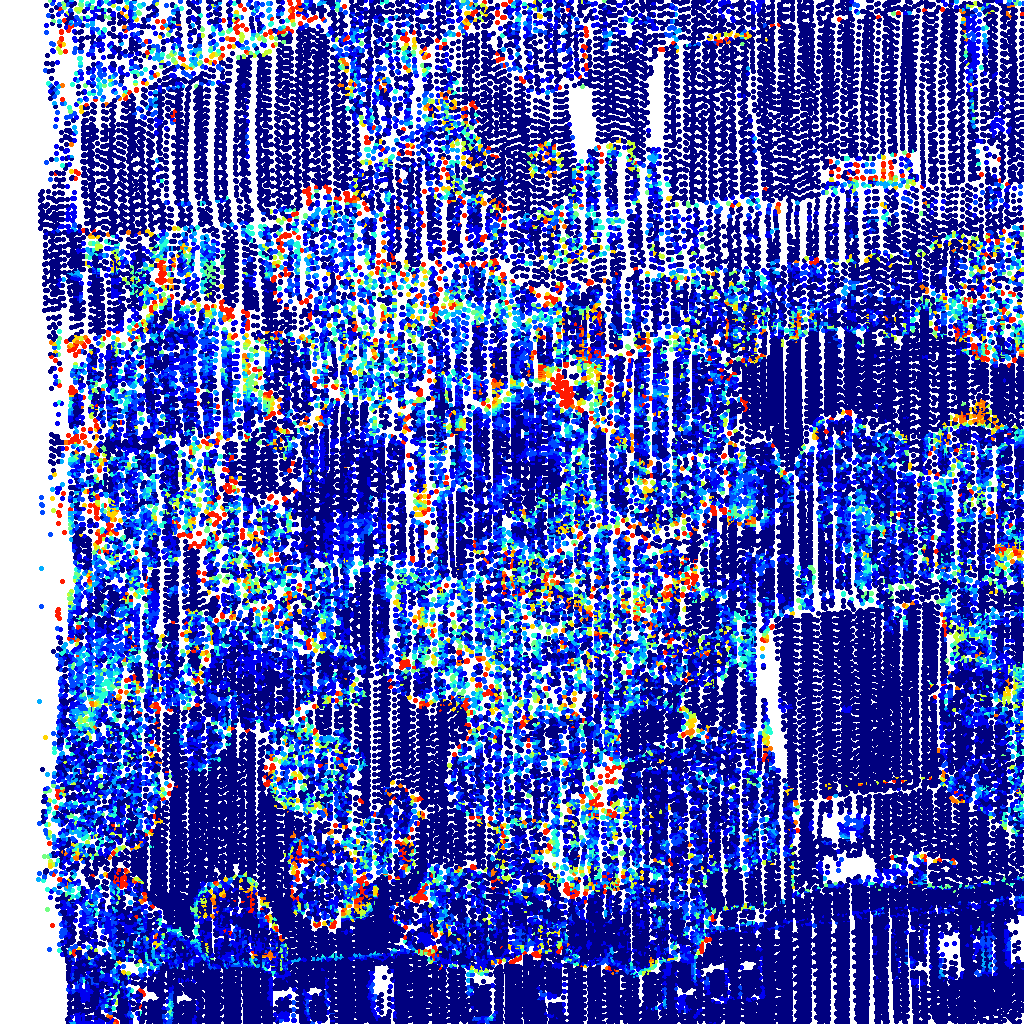}
		\includegraphics[width=\linewidth]{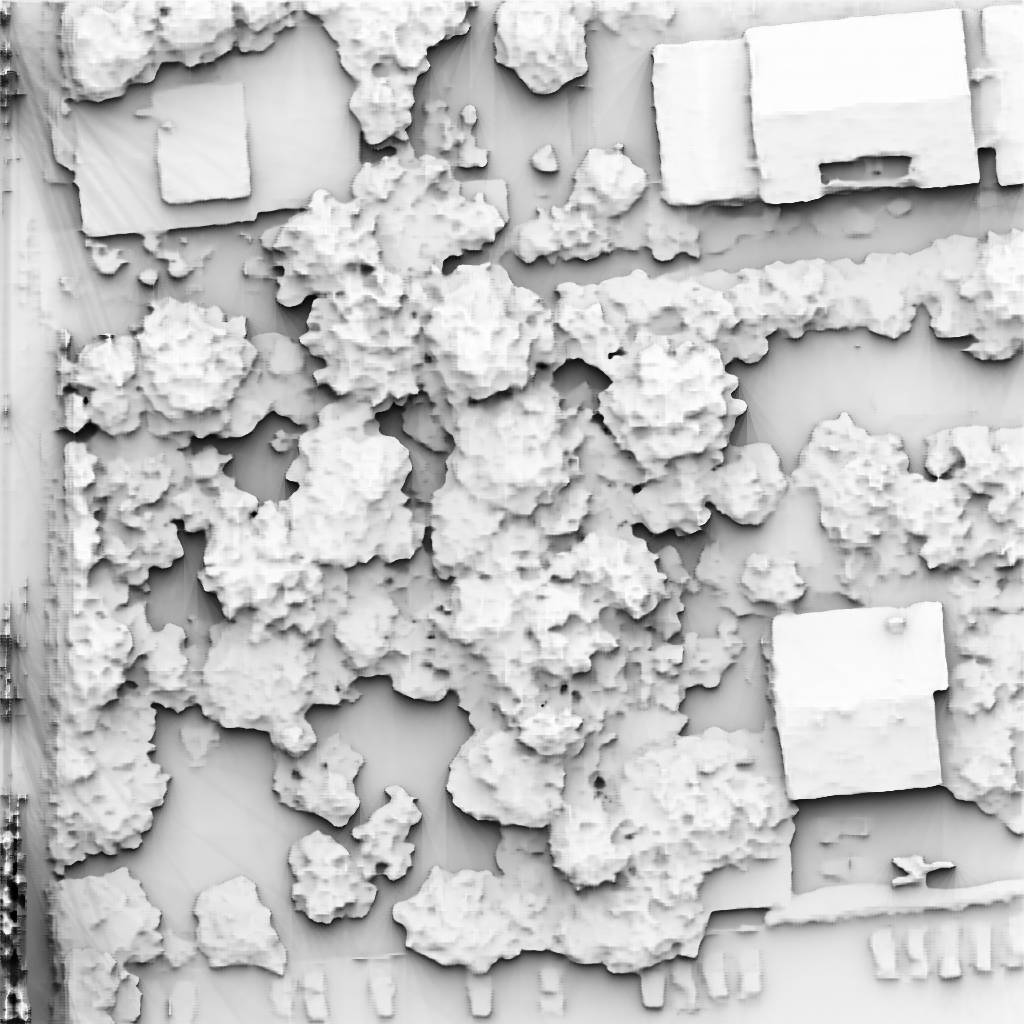}
		\centering{\tiny PSM net}
	\end{minipage}
	\begin{minipage}[t]{0.19\textwidth}
		\includegraphics[width=0.098\linewidth]{figures_supp/color_map.png}
		\includegraphics[width=0.85\linewidth]{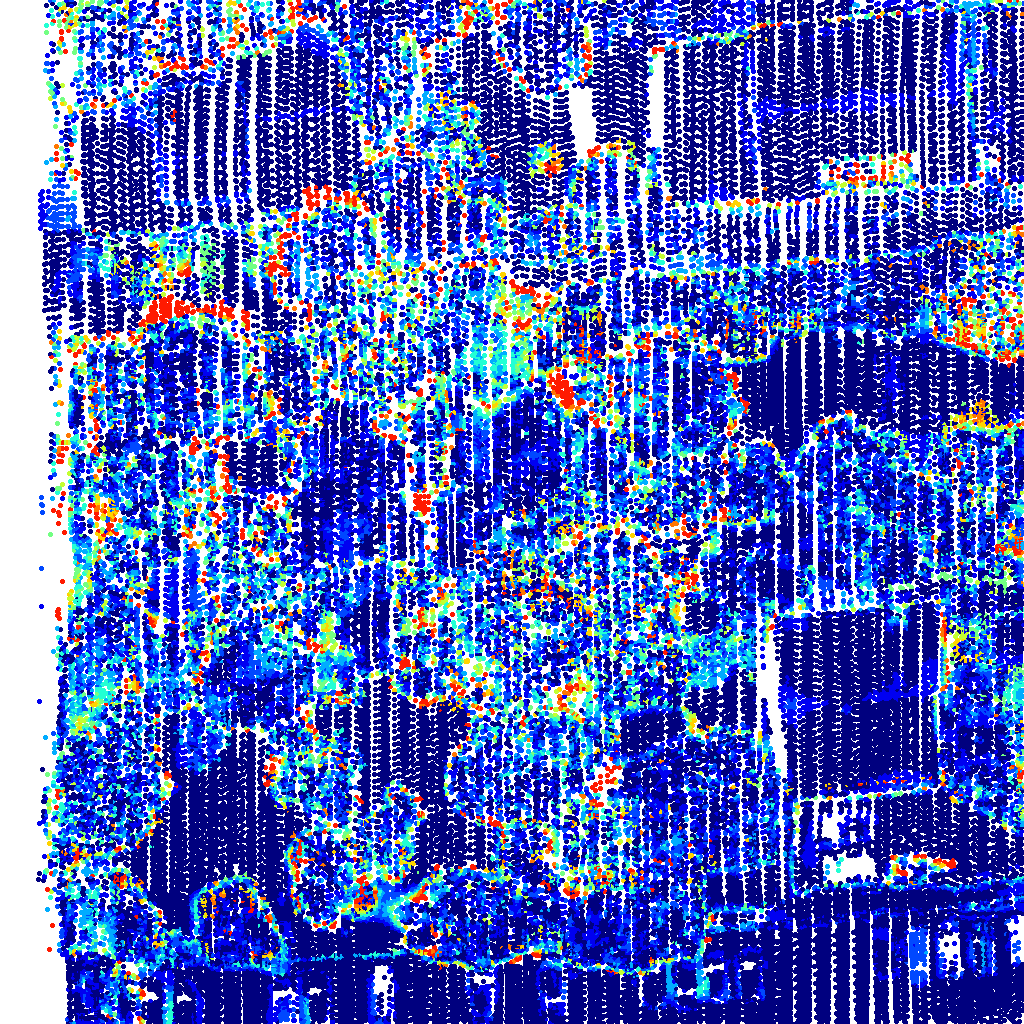}
		\includegraphics[width=\linewidth]{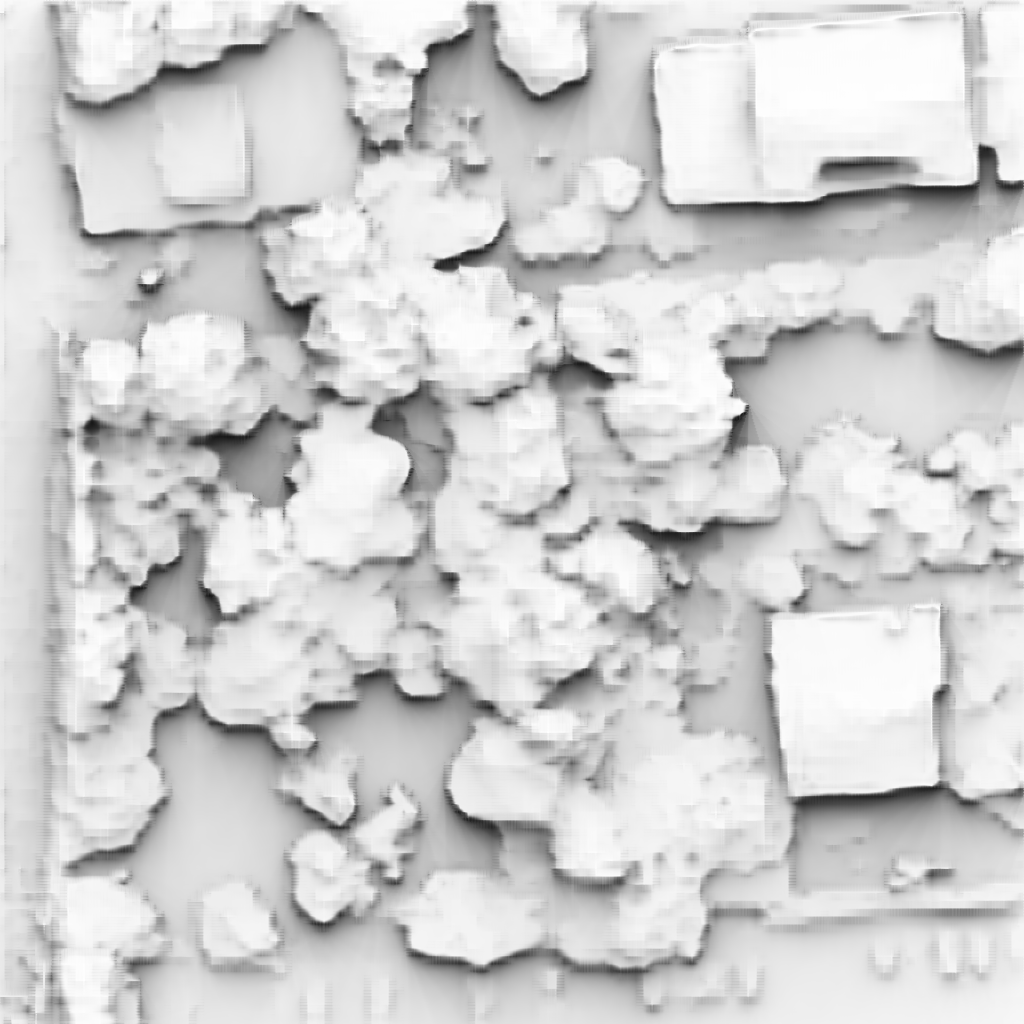}
		\centering{\tiny HRS net}
	\end{minipage}
	\begin{minipage}[t]{0.19\textwidth}	
		\includegraphics[width=0.098\linewidth]{figures_supp/color_map.png}
		\includegraphics[width=0.85\linewidth]{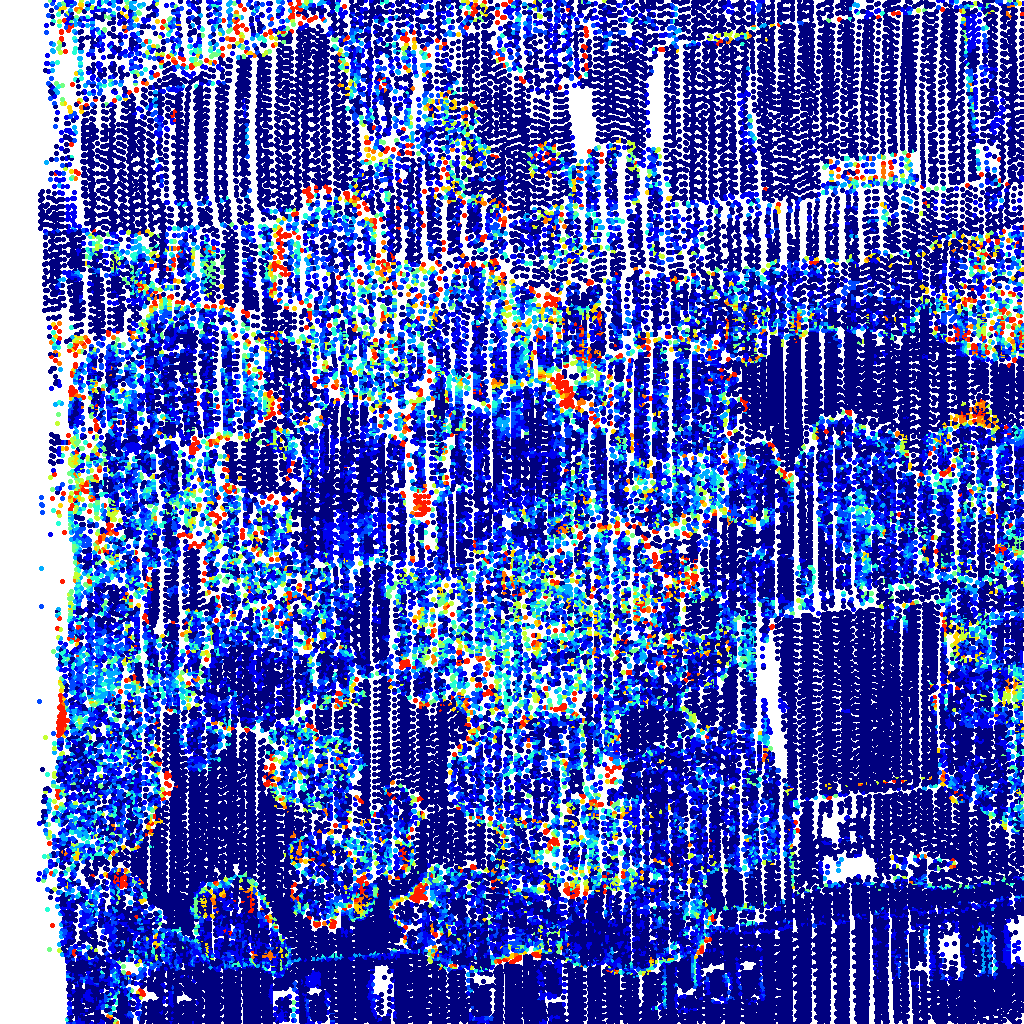}
		\includegraphics[width=\linewidth]{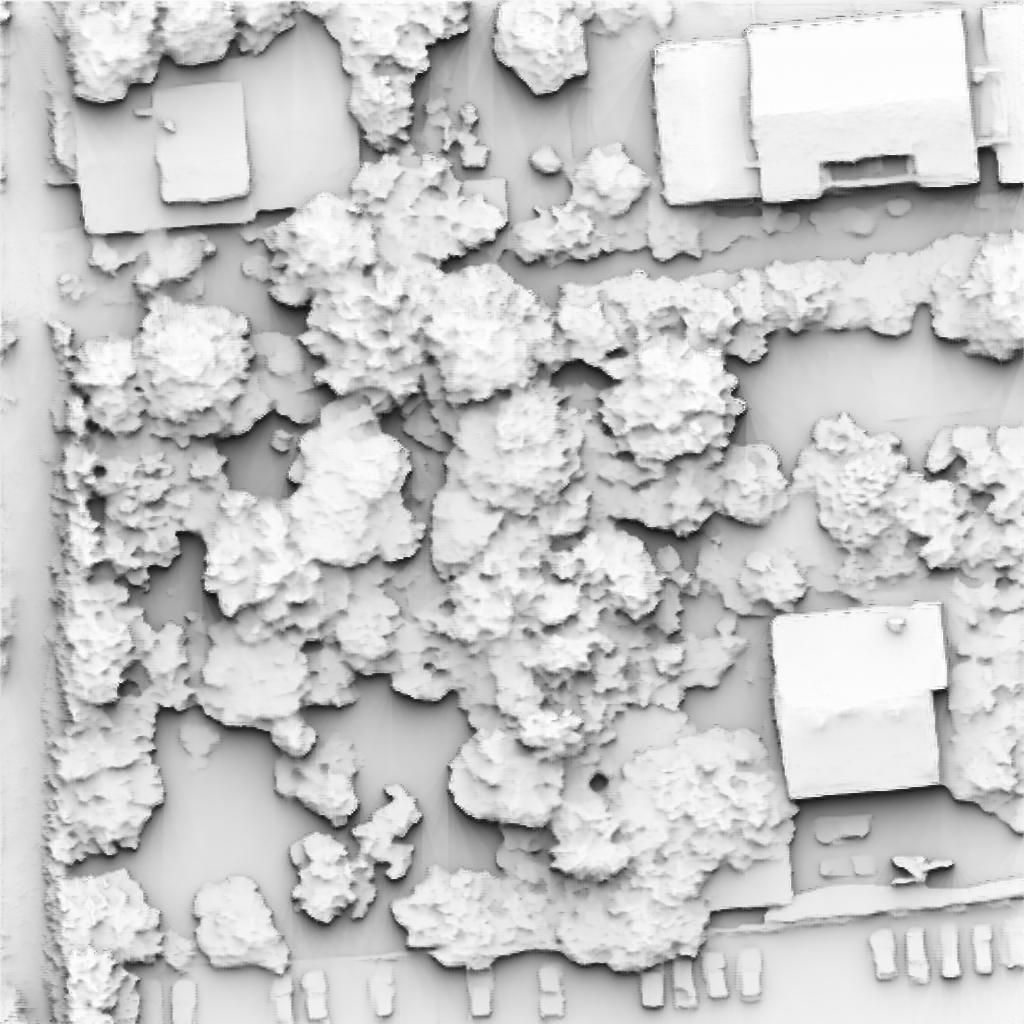}
		\centering{\tiny DeepPruner}
	\end{minipage}
	\begin{minipage}[t]{0.19\textwidth}
		\includegraphics[width=0.098\linewidth]{figures_supp/color_map.png}
		\includegraphics[width=0.85\linewidth]{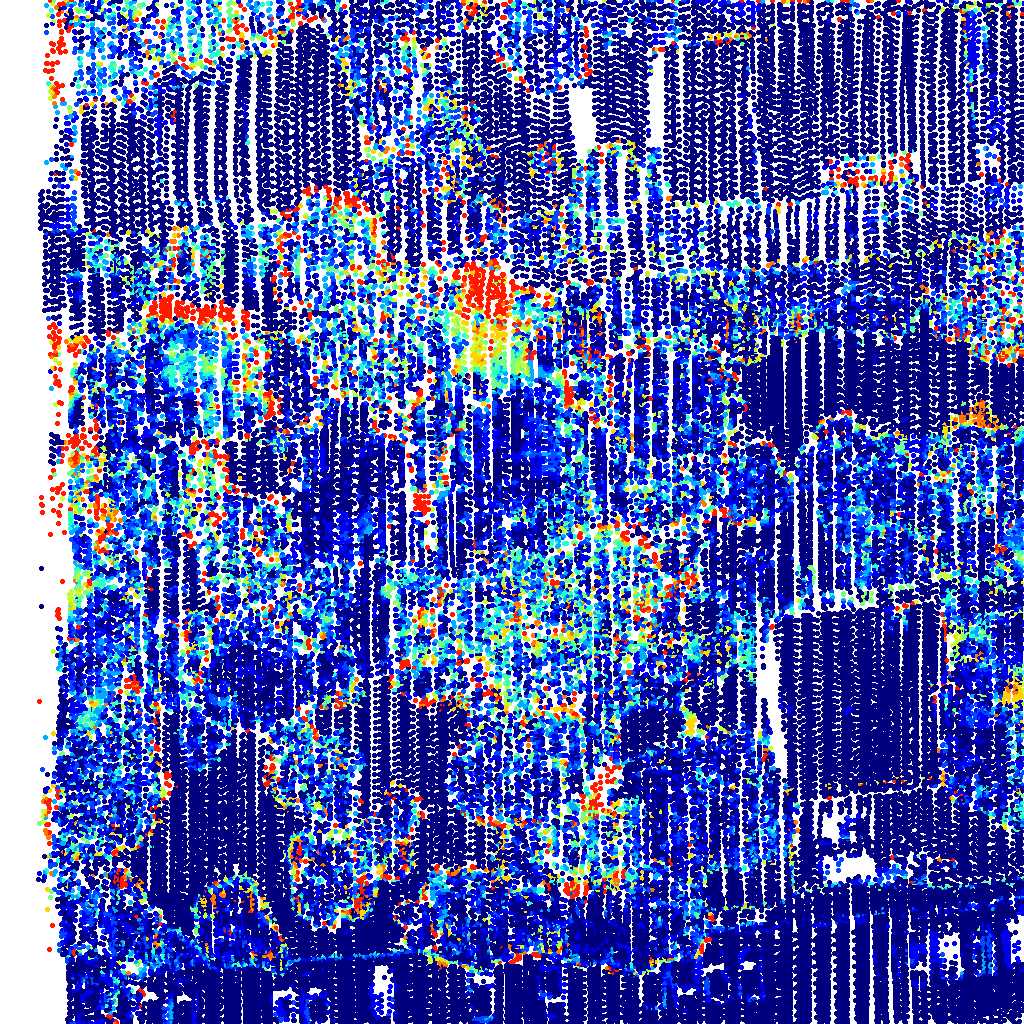}
		\includegraphics[width=\linewidth]{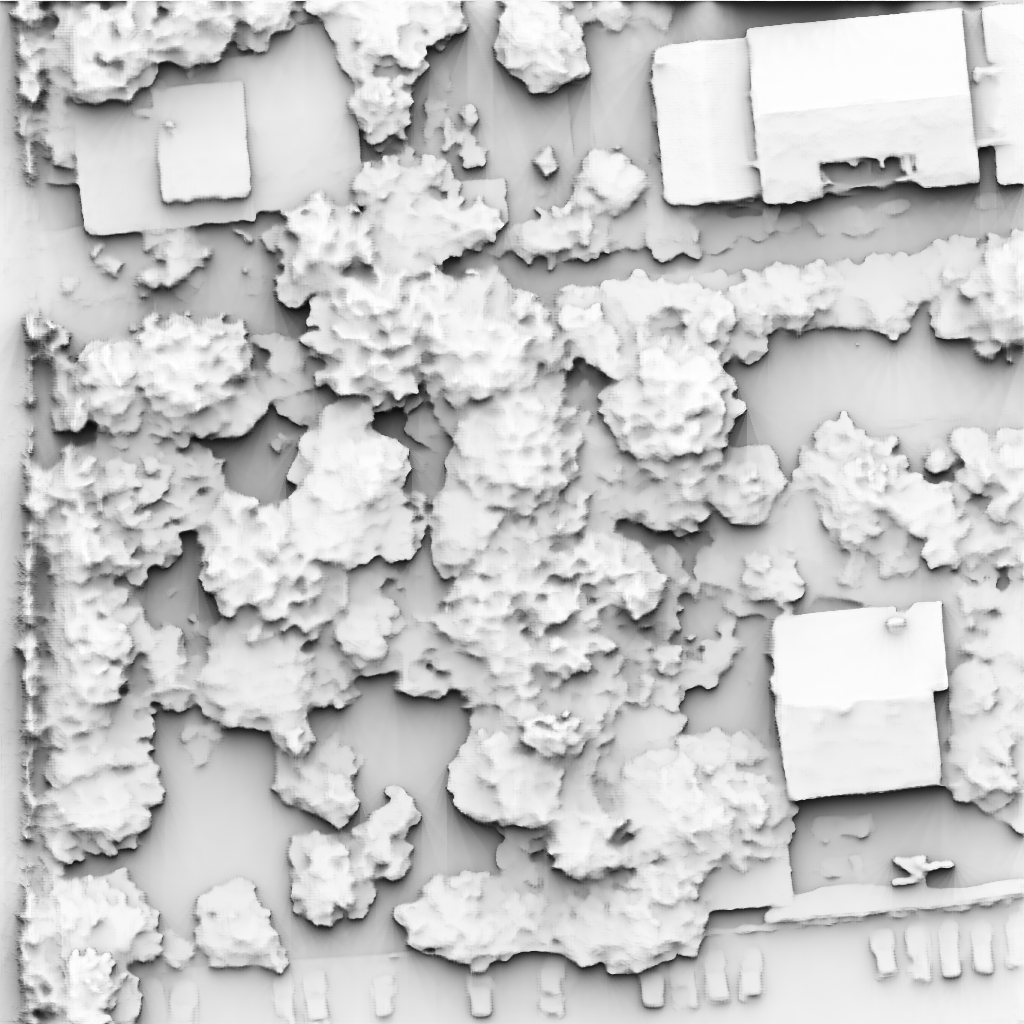}
		\centering{\tiny GANet}
	\end{minipage}
	\begin{minipage}[t]{0.19\textwidth}	
		\includegraphics[width=0.098\linewidth]{figures_supp/color_map.png}
		\includegraphics[width=0.85\linewidth]{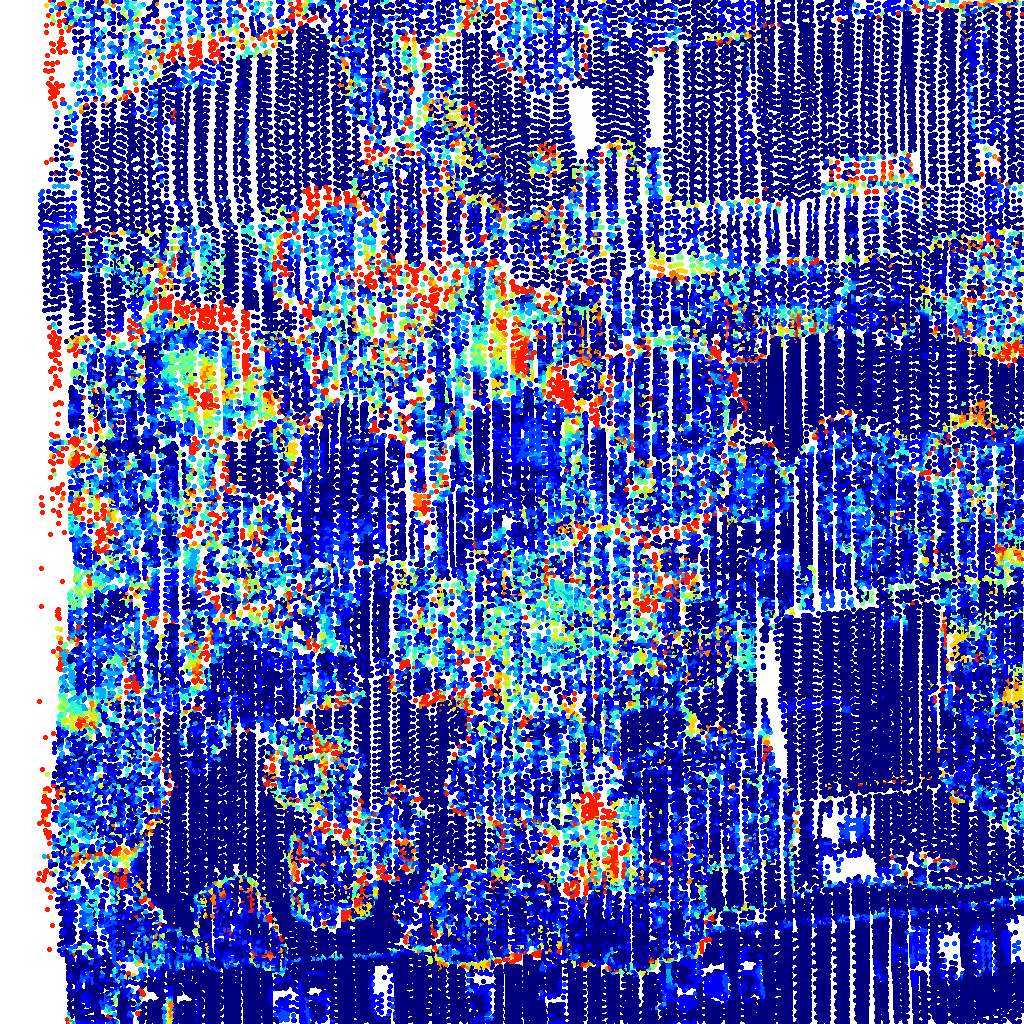}
		\includegraphics[width=\linewidth]{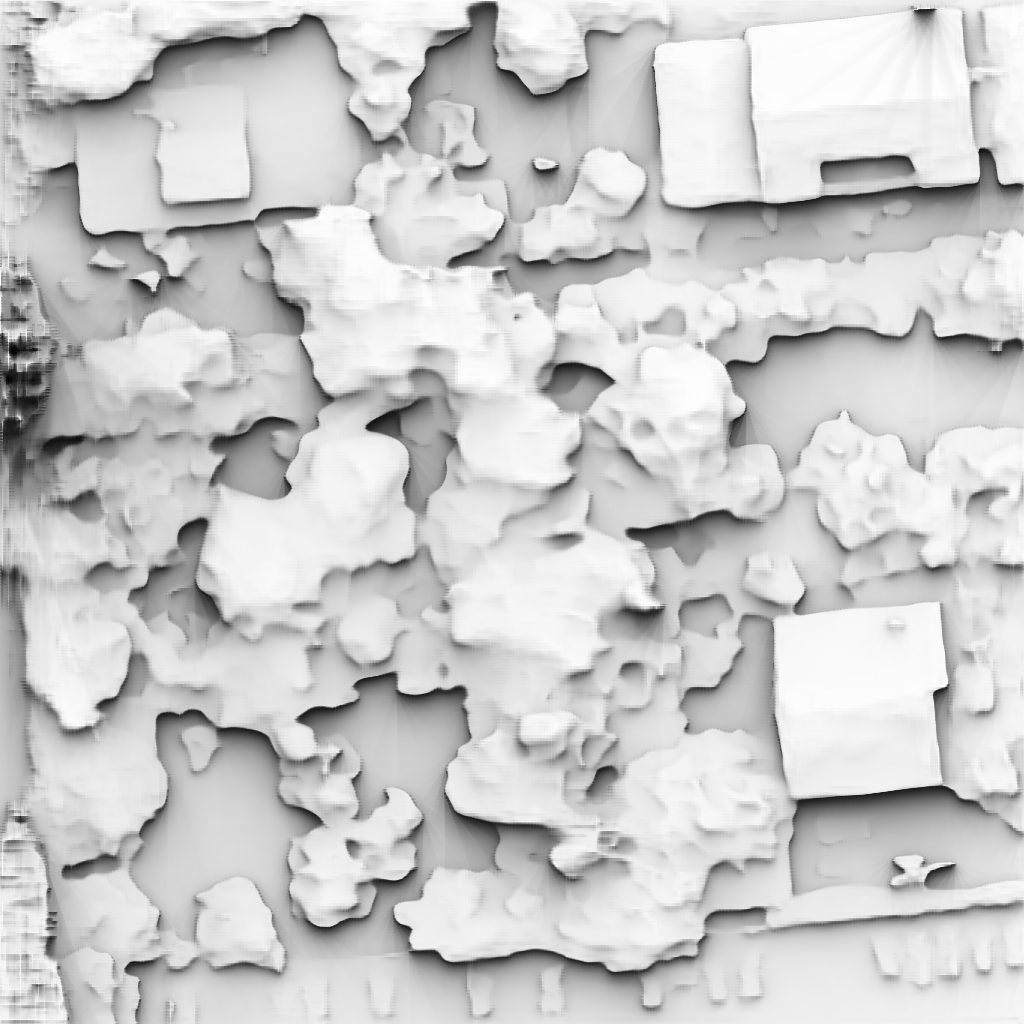}
		\centering{\tiny LEAStereo}
	\end{minipage}
	\caption{Error map and disparity visualization on tree area for Toulouse Metropole.}
	\label{Figure.mlsetree}
\end{figure}

\paragraph{Toulouse Metropole}
Over man-made objects all the methods perform well, even on pre-trained models. Fine-tuning further enhances the results, especially at discontinuities (cd. \Cref{Figure.mlsebulding}). As far as the trees go, both the traditional SGM and DL-based methods on pre-trained models produce mediocre outcomes. Fine-tuning helps to recover the tree canopies (cf. \Cref{Figure.mlsetree}). PSMNet performs best of all.
%

\begin{figure}[tp]
	\begin{minipage}[t]{0.19\textwidth}	
		\includegraphics[width=0.098\linewidth]{figures_supp/color_map.png}
		\includegraphics[width=0.85\linewidth]{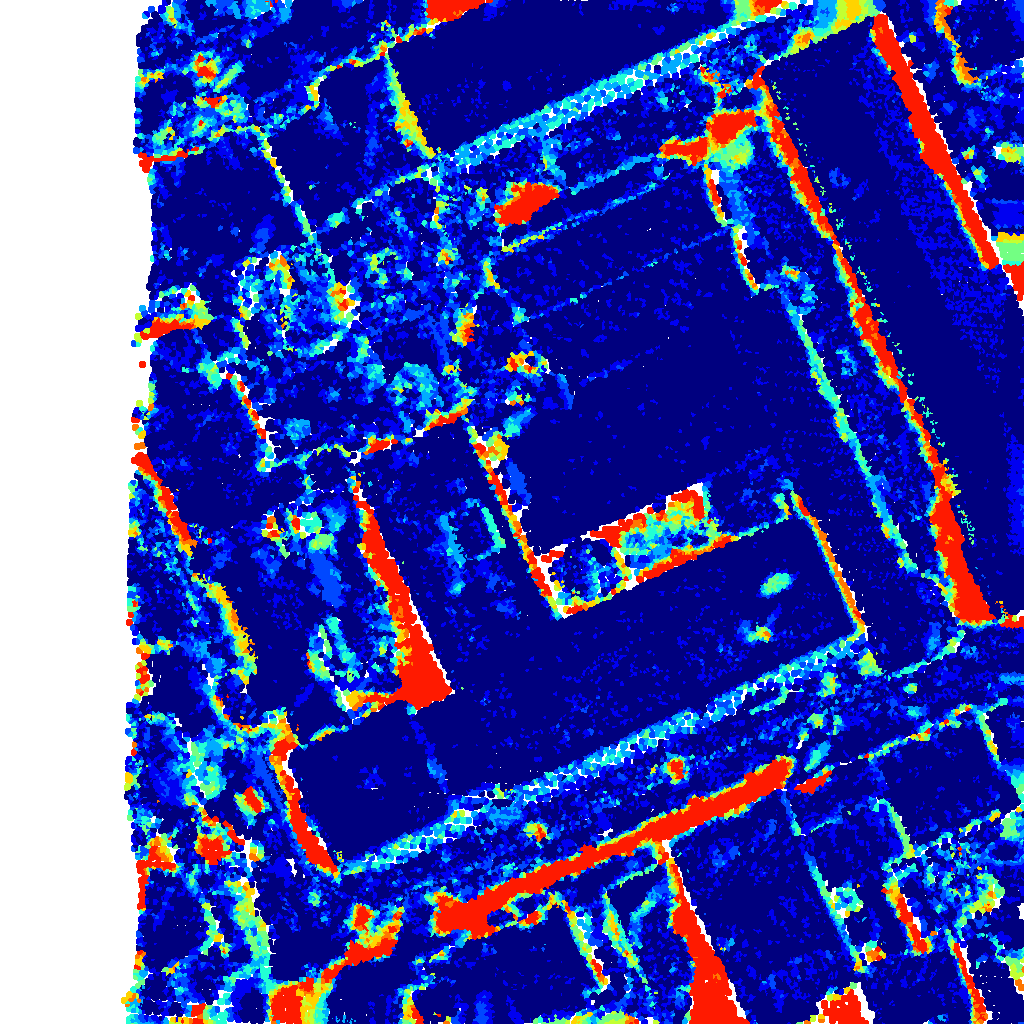}
		\includegraphics[width=\linewidth]{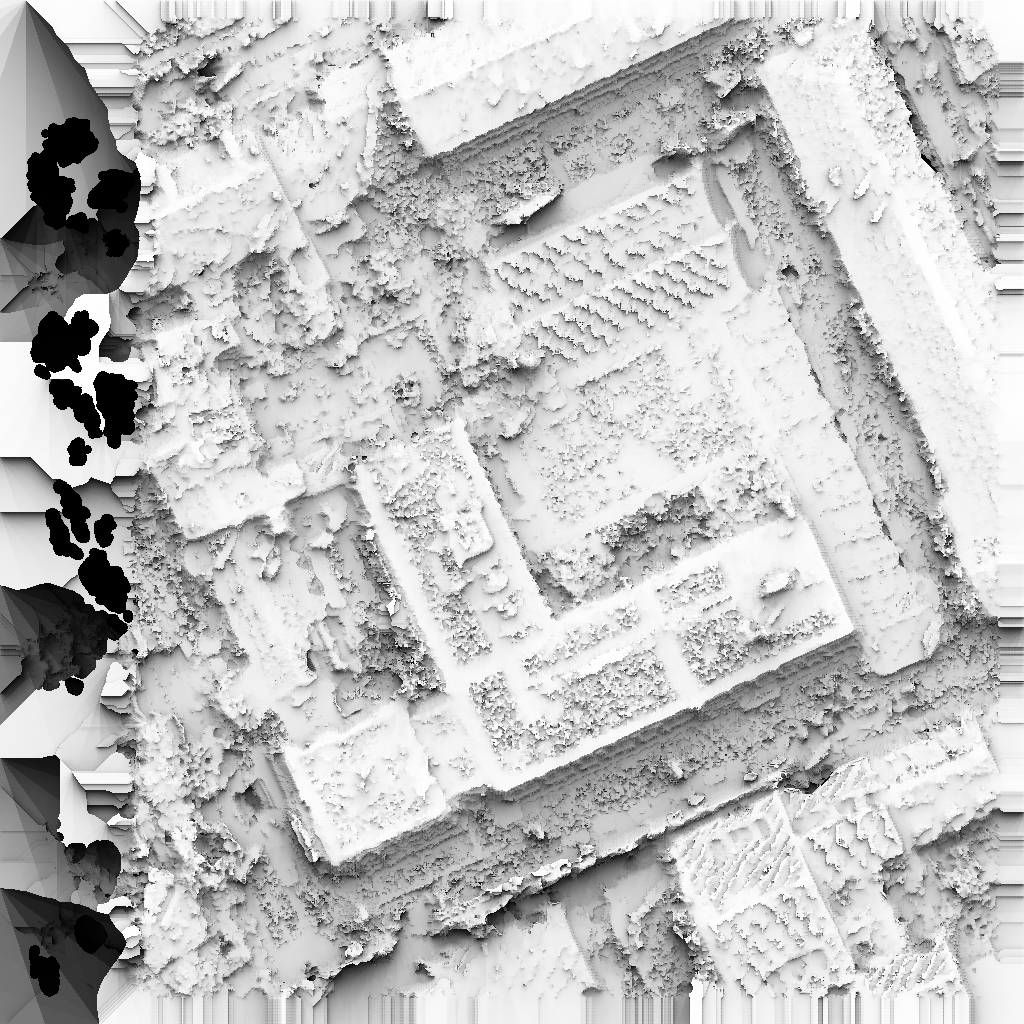}
		\centering{\tiny MICMAC}
	\end{minipage}
	\begin{minipage}[t]{0.19\textwidth}	
		\includegraphics[width=0.098\linewidth]{figures_supp/color_map.png}
		\includegraphics[width=0.85\linewidth]{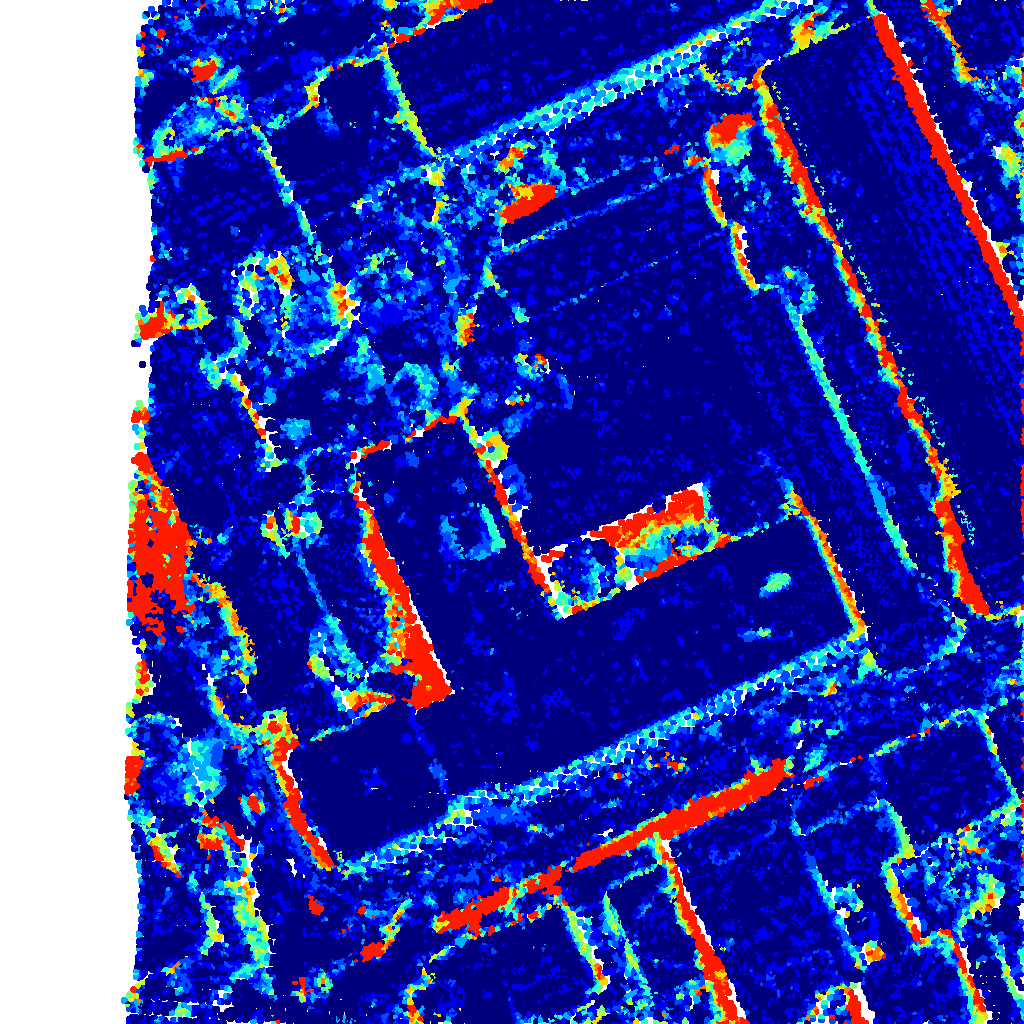}
		\includegraphics[width=\linewidth]{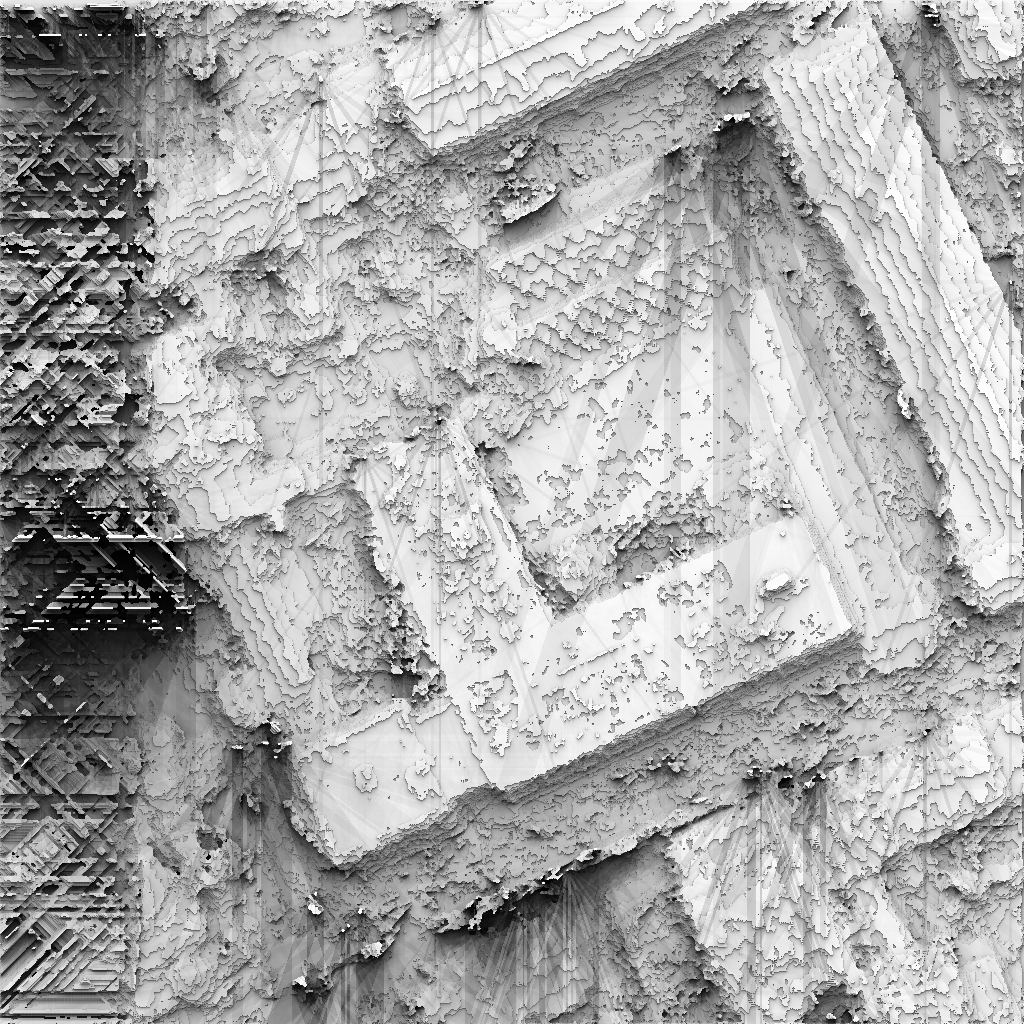}
		\centering{\tiny SGM(CUDA)}
	\end{minipage}
	\begin{minipage}[t]{0.19\textwidth}	
		\includegraphics[width=0.098\linewidth]{figures_supp/color_map.png}
		\includegraphics[width=0.85\linewidth]{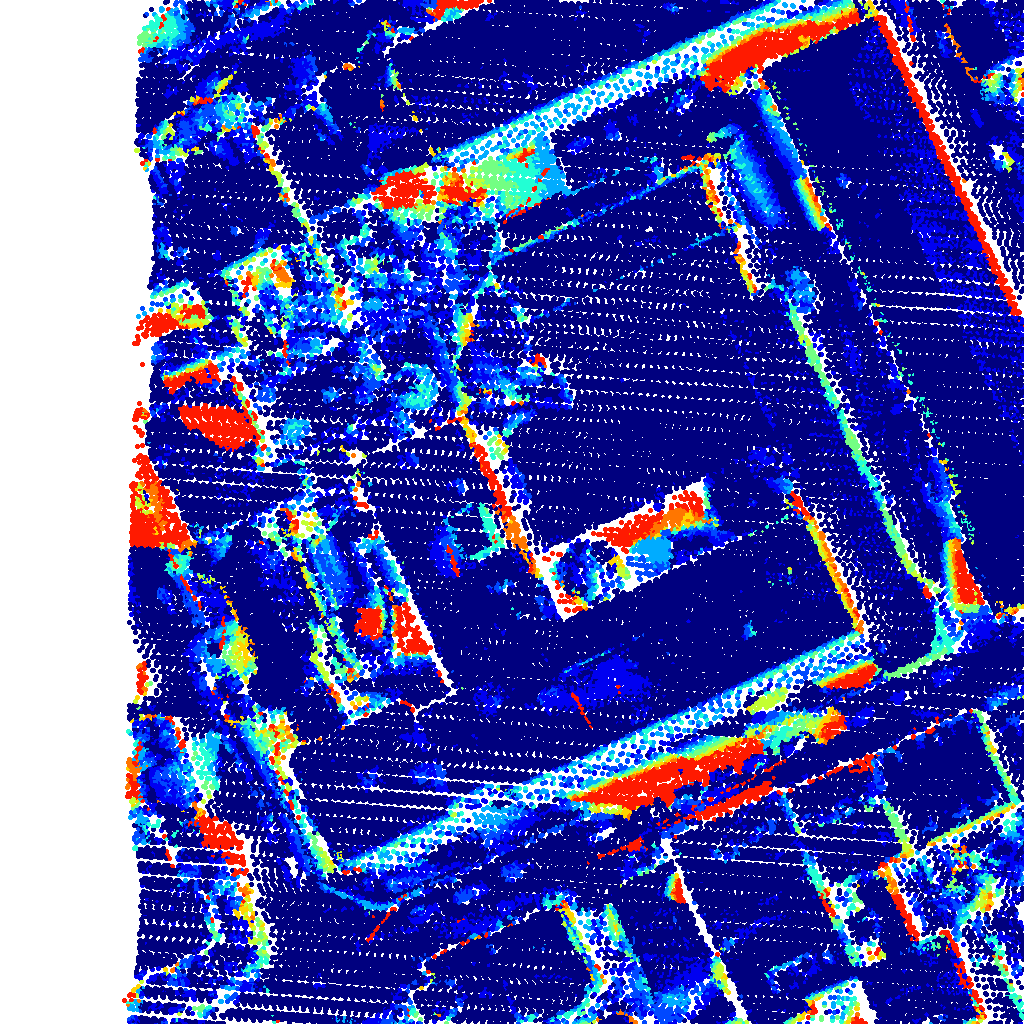}
		\includegraphics[width=\linewidth]{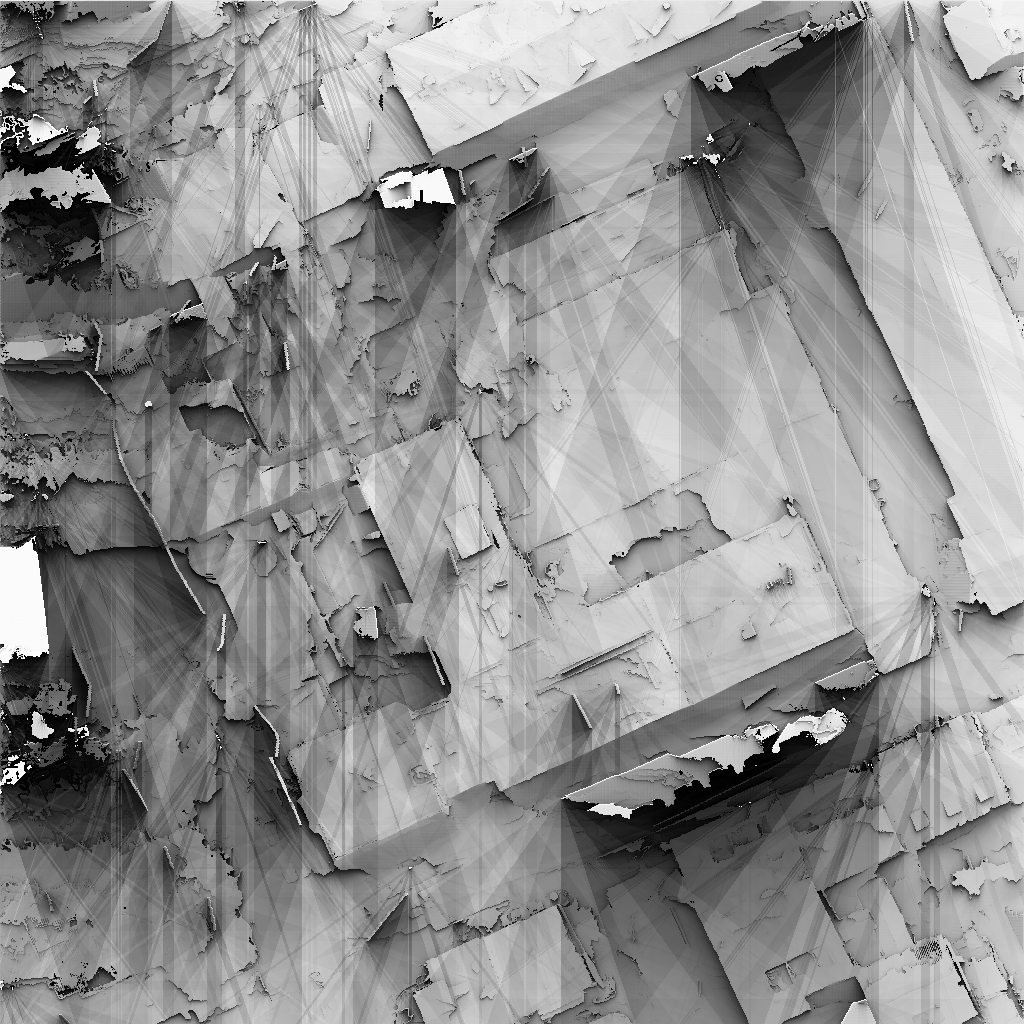}
		\centering{\tiny GraphCuts}
	\end{minipage}
	\begin{minipage}[t]{0.19\textwidth}	
		\includegraphics[width=0.098\linewidth]{figures_supp/color_map.png}
		\includegraphics[width=0.85\linewidth]{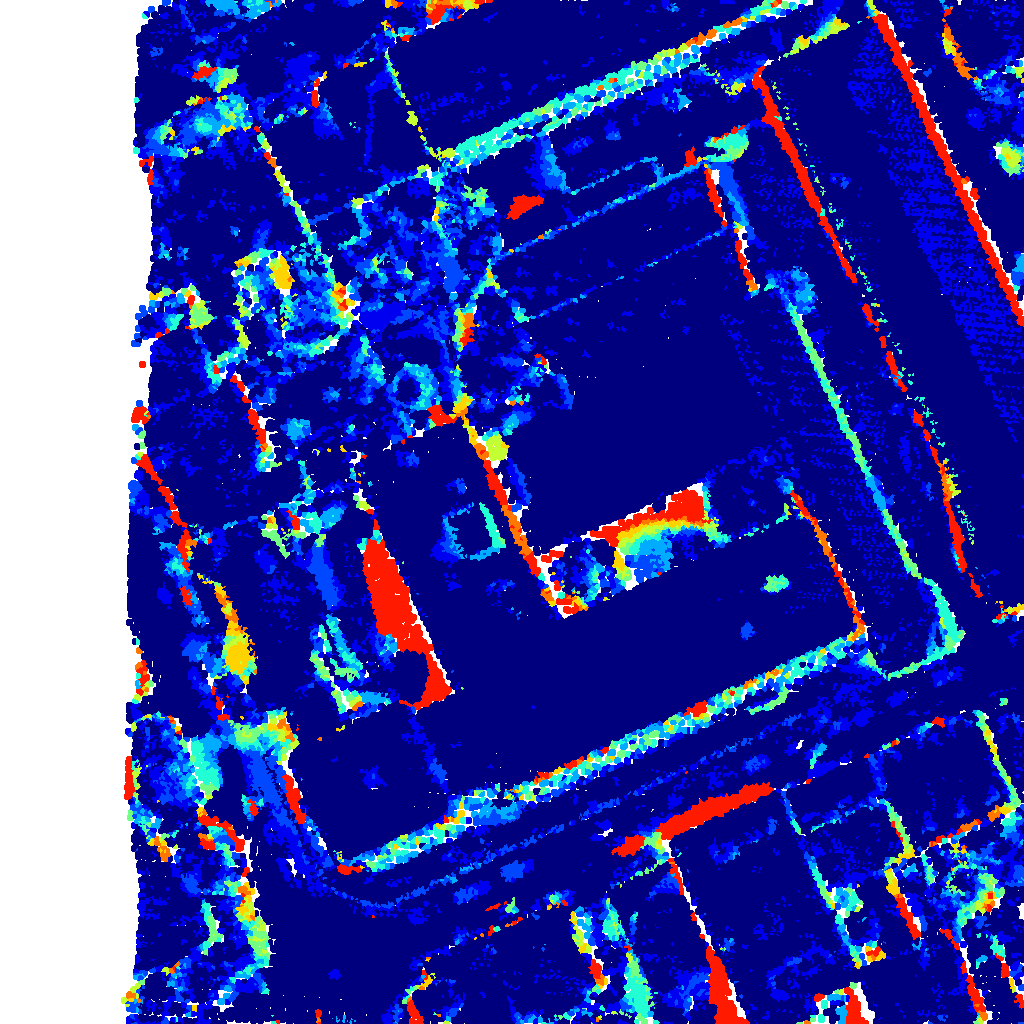}
		\includegraphics[width=\linewidth]{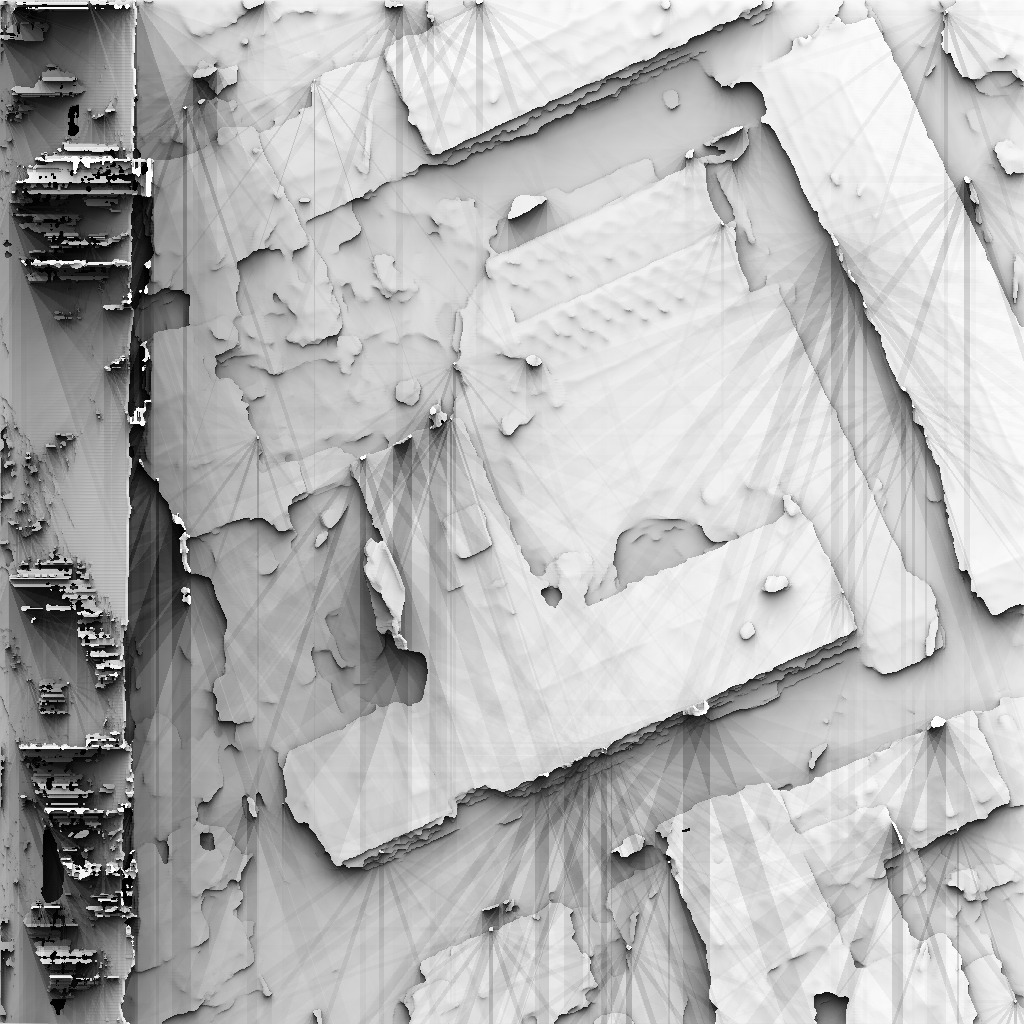}
		\centering{\tiny CBMV(SGM)}
	\end{minipage}
	\begin{minipage}[t]{0.19\textwidth}	
		\includegraphics[width=0.098\linewidth]{figures_supp/color_map.png}
		\includegraphics[width=0.85\linewidth]{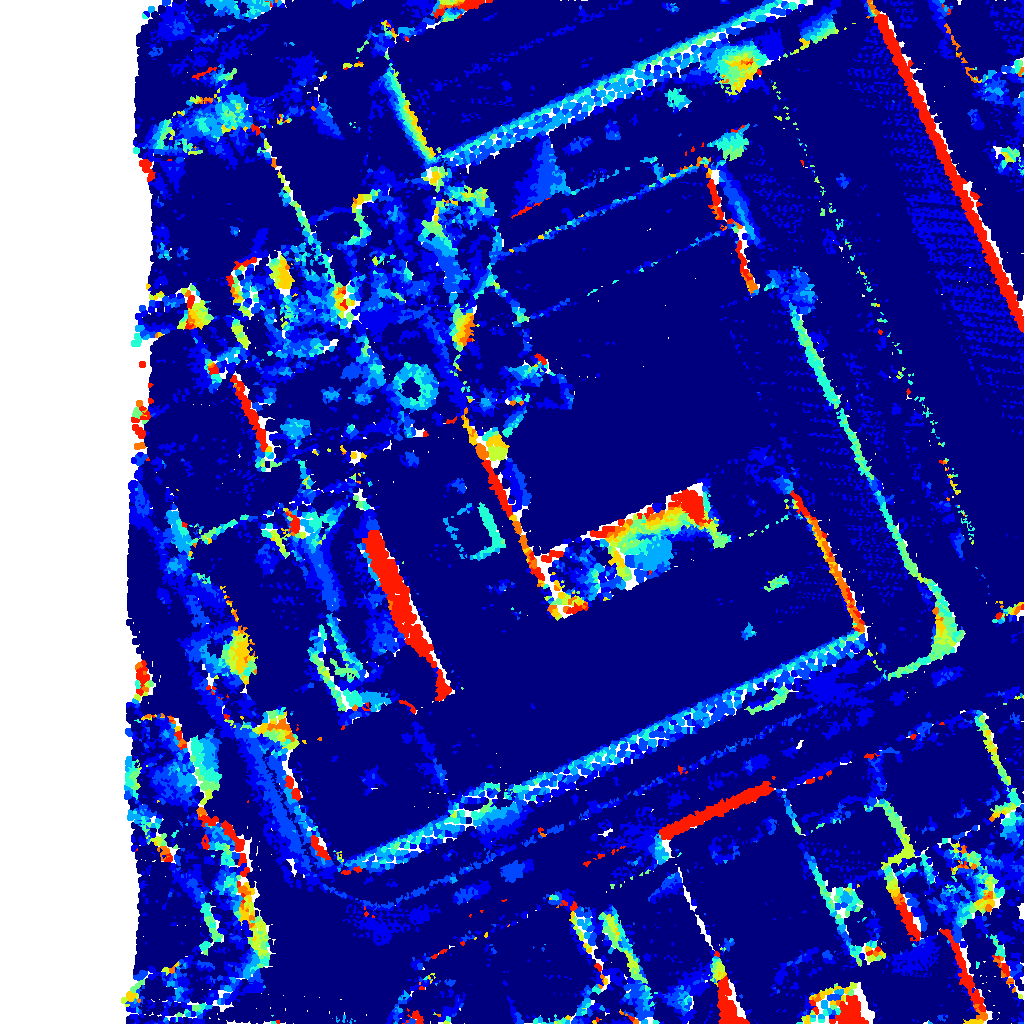}
		\includegraphics[width=\linewidth]{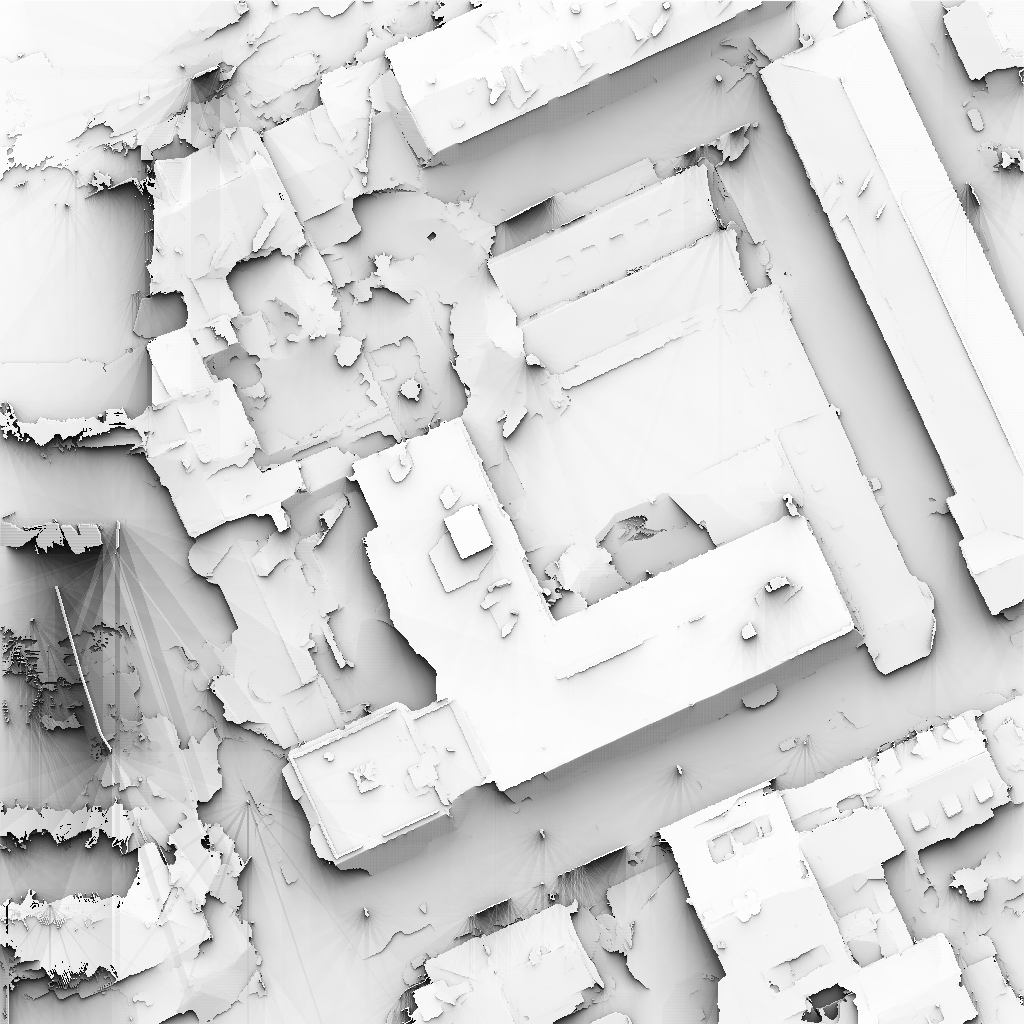}
		\centering{\tiny CBMV(GraphCuts)}
	\end{minipage}
	\begin{minipage}[t]{0.19\textwidth}	
		\includegraphics[width=0.098\linewidth]{figures_supp/color_map.png}
		\includegraphics[width=0.85\linewidth]{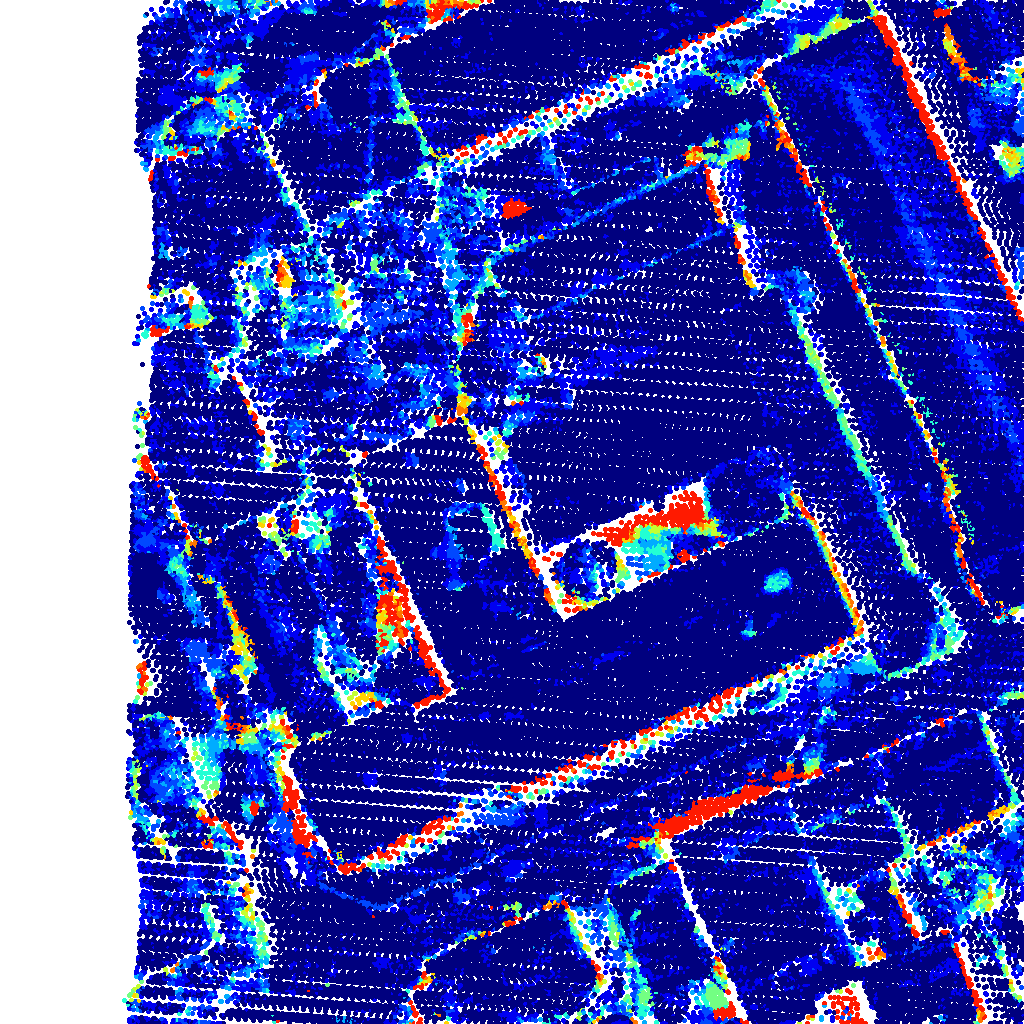}
		\includegraphics[width=\linewidth]{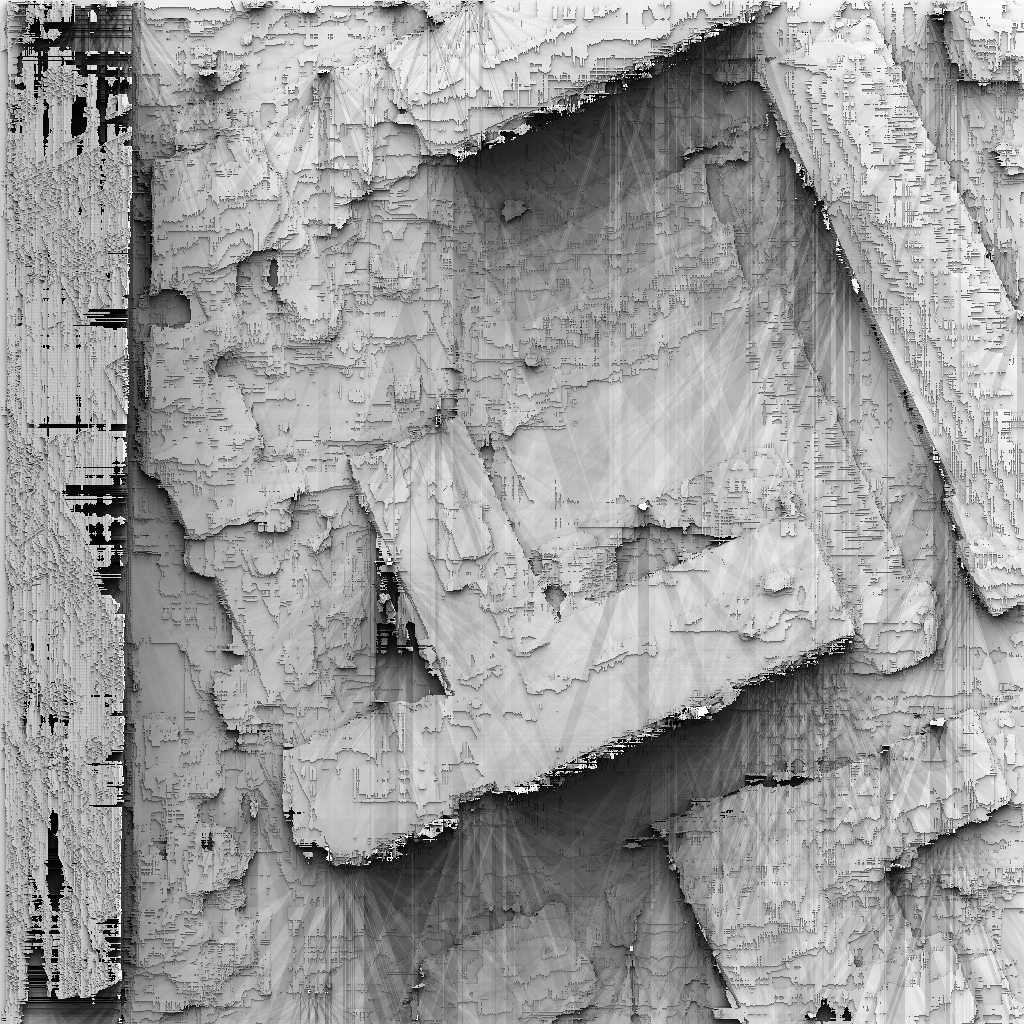}
		\centering{\tiny MC-CNN(KITTI)}
	\end{minipage}
	\begin{minipage}[t]{0.19\textwidth}	
		\includegraphics[width=0.098\linewidth]{figures_supp/color_map.png}
		\includegraphics[width=0.85\linewidth]{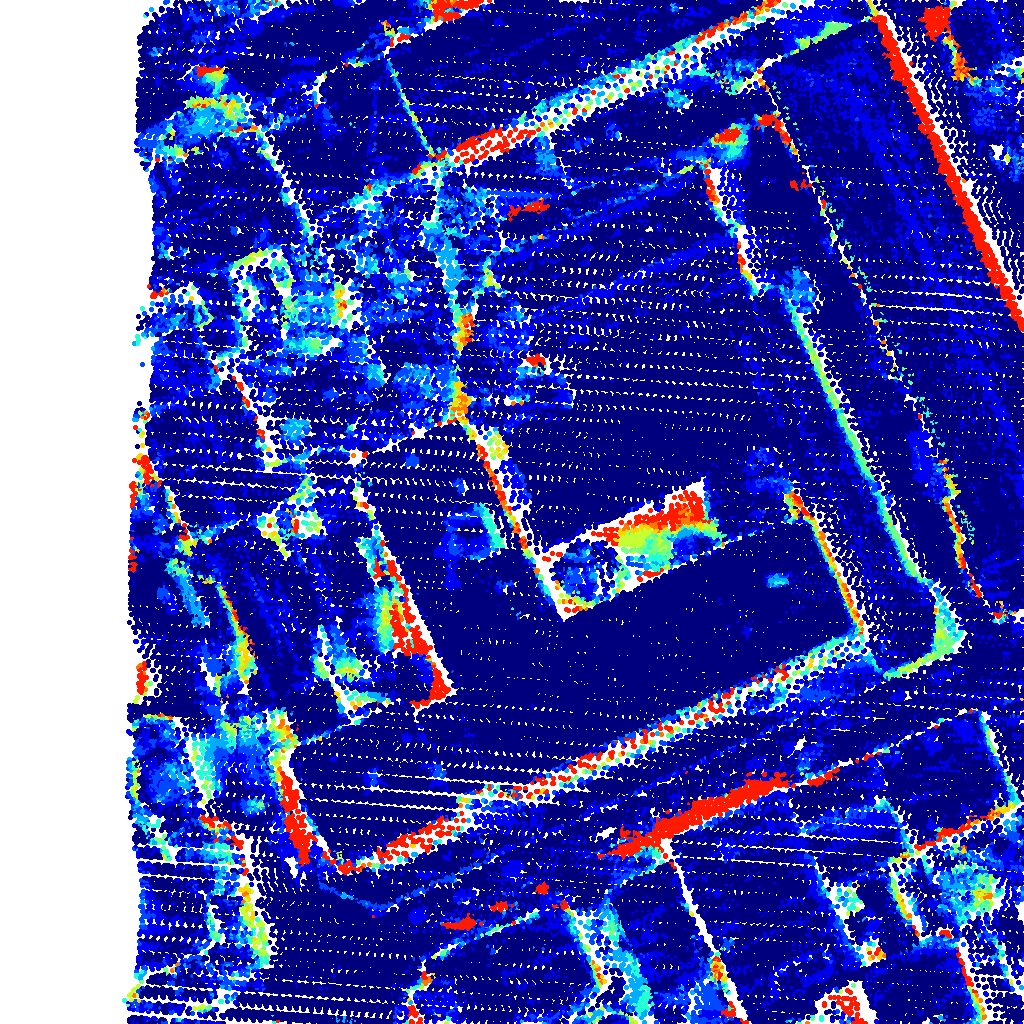}
		\includegraphics[width=\linewidth]{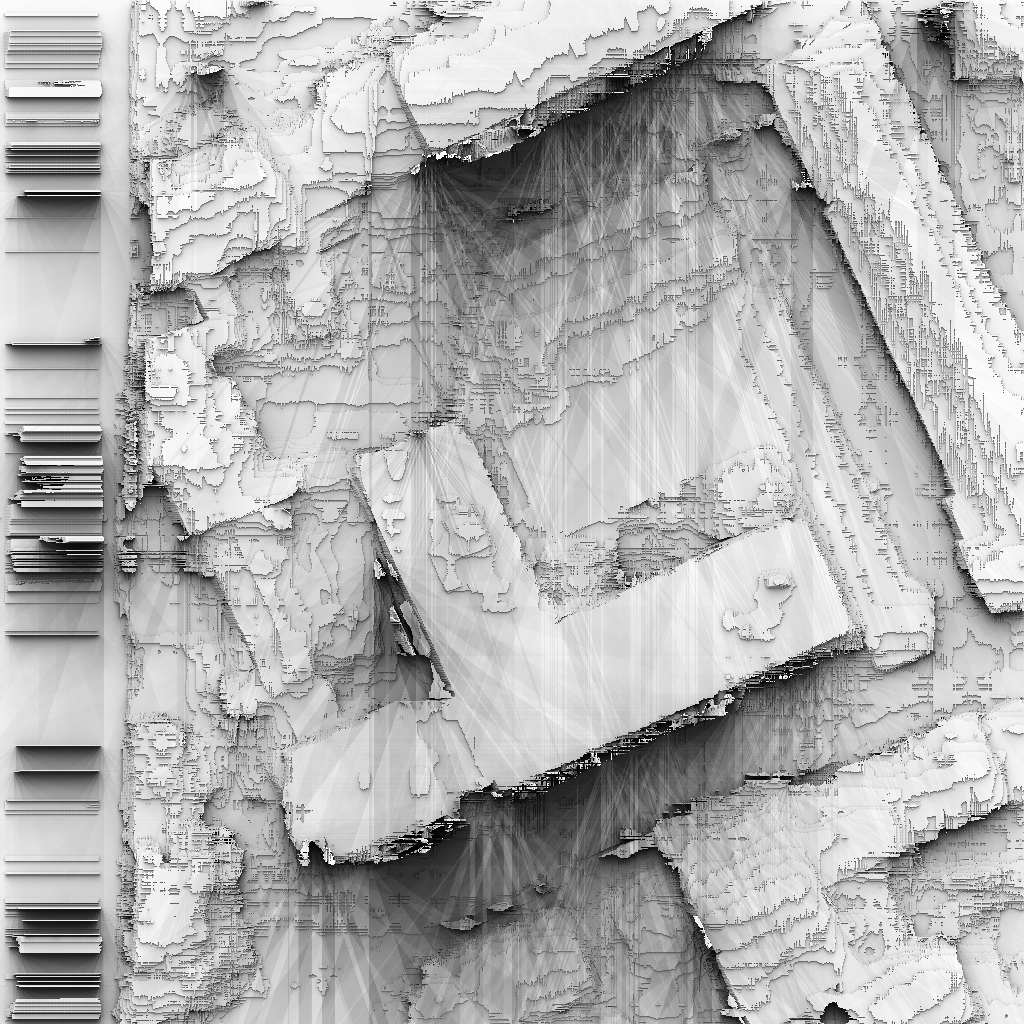}
		\centering{\tiny DeepFeature(KITTI)}
	\end{minipage}
	\begin{minipage}[t]{0.19\textwidth}	
		\includegraphics[width=0.098\linewidth]{figures_supp/color_map.png}
		\includegraphics[width=0.85\linewidth]{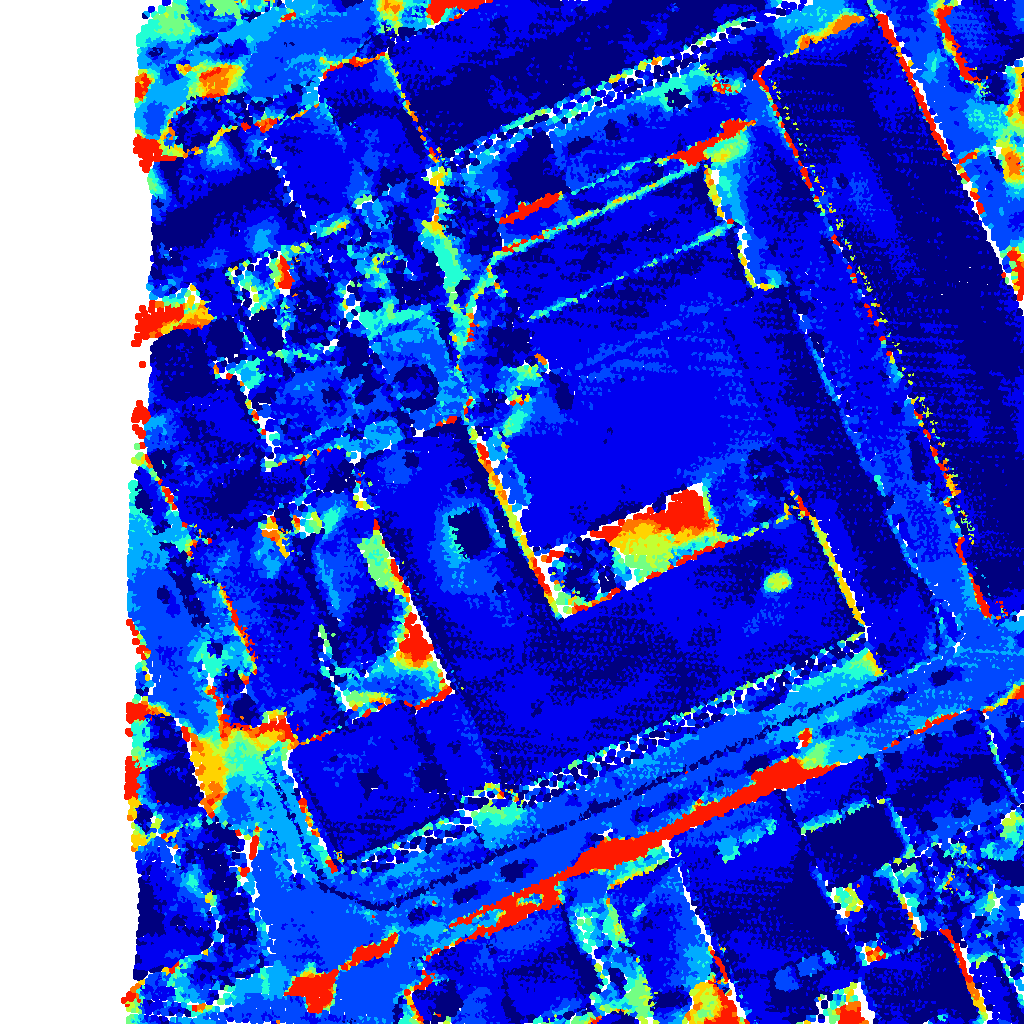}
		\includegraphics[width=\linewidth]{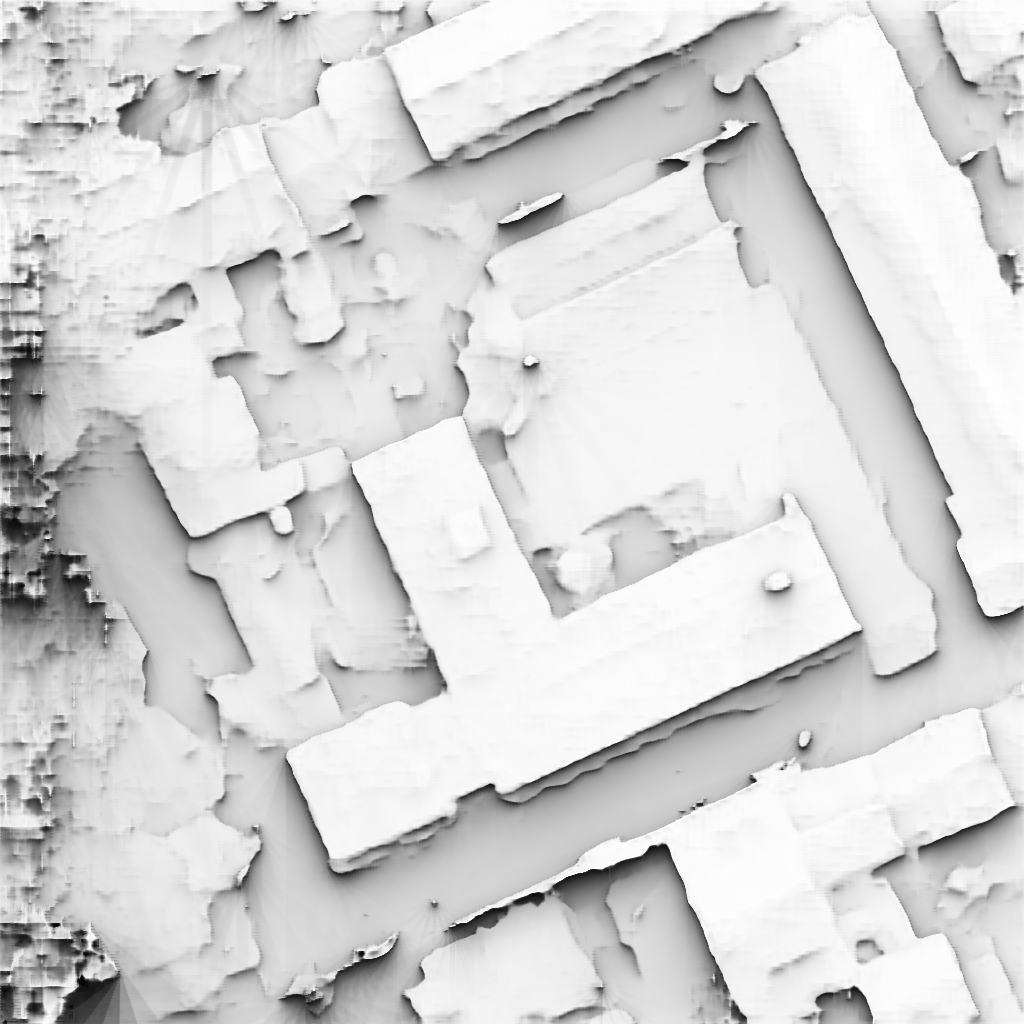}
		\centering{\tiny PSM net(KITTI)}
	\end{minipage}
	\begin{minipage}[t]{0.19\textwidth}		
		\includegraphics[width=0.098\linewidth]{figures_supp/color_map.png}
		\includegraphics[width=0.85\linewidth]{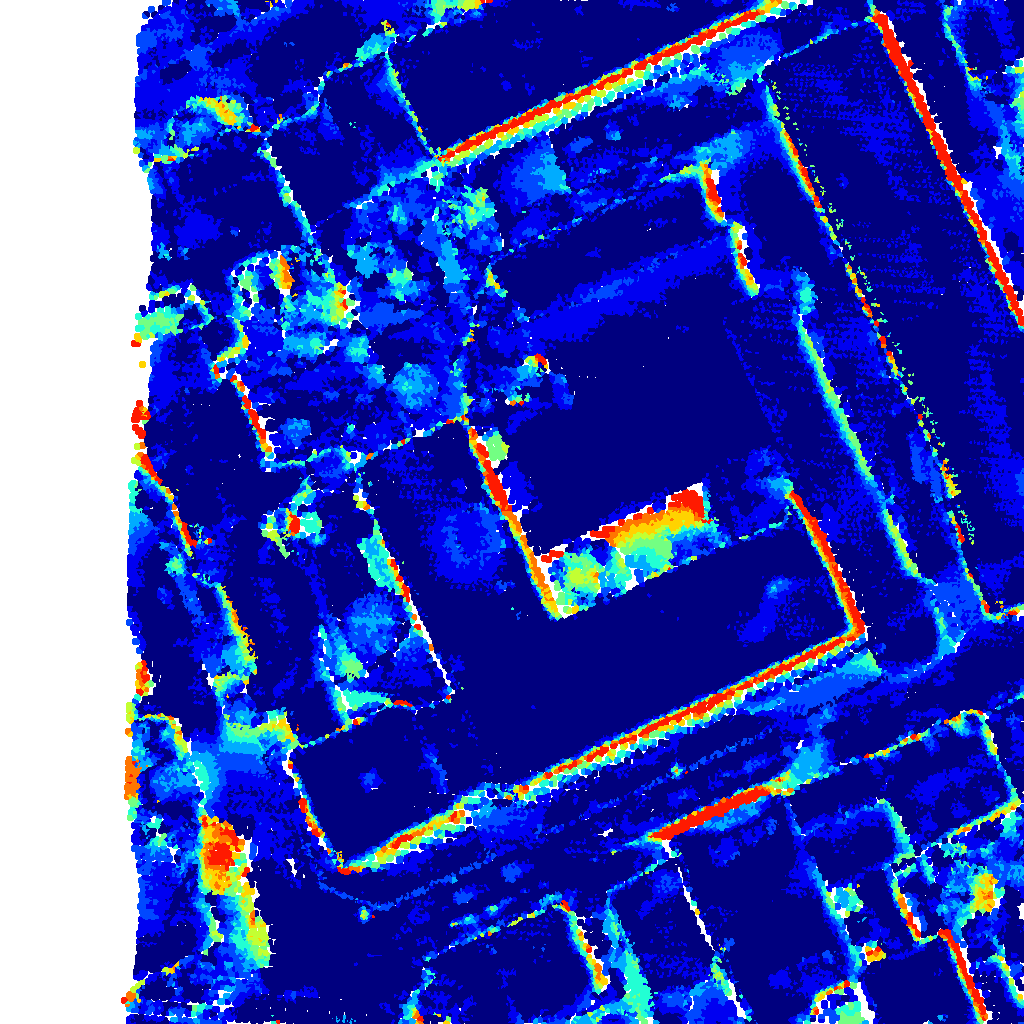}
		\includegraphics[width=\linewidth]{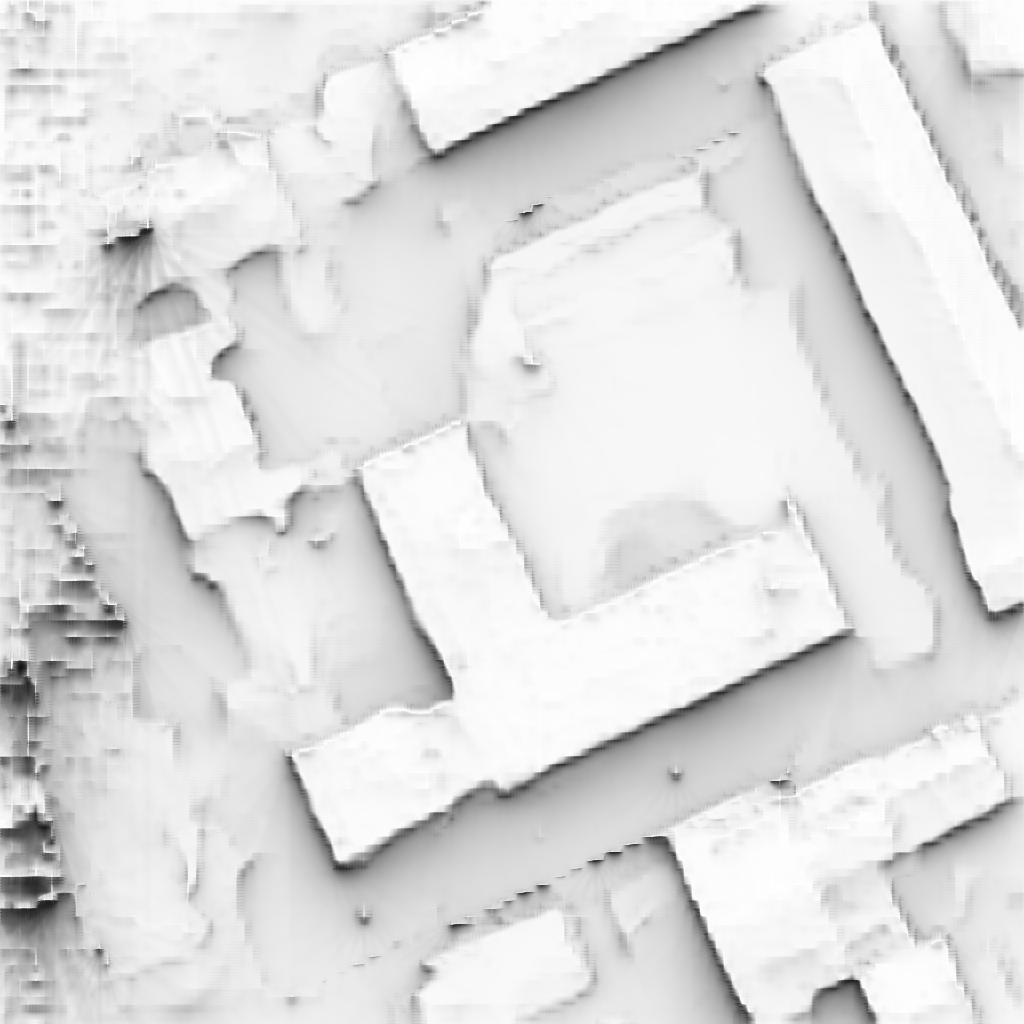}
		\centering{\tiny HRS net(KITTI)}
	\end{minipage}
	\begin{minipage}[t]{0.19\textwidth}	
		\includegraphics[width=0.098\linewidth]{figures_supp/color_map.png}
		\includegraphics[width=0.85\linewidth]{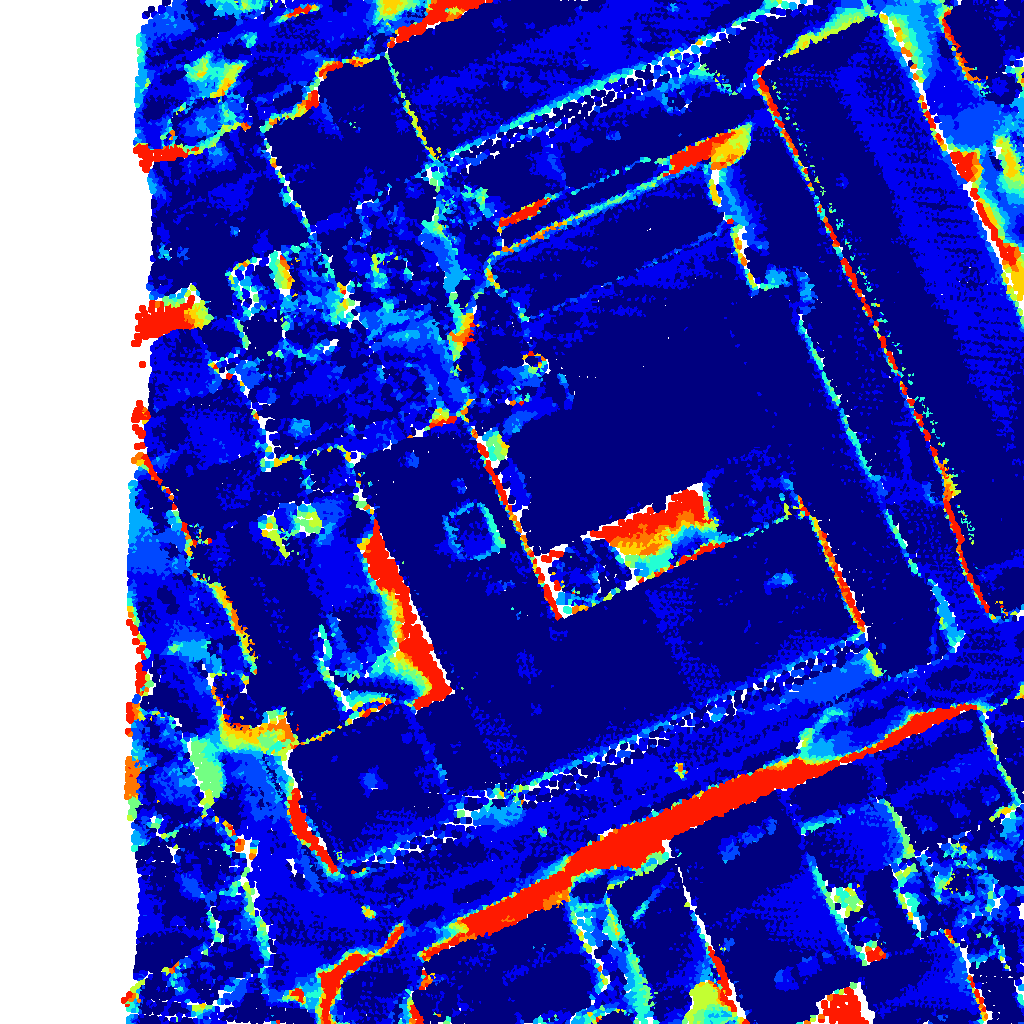}
		\includegraphics[width=\linewidth]{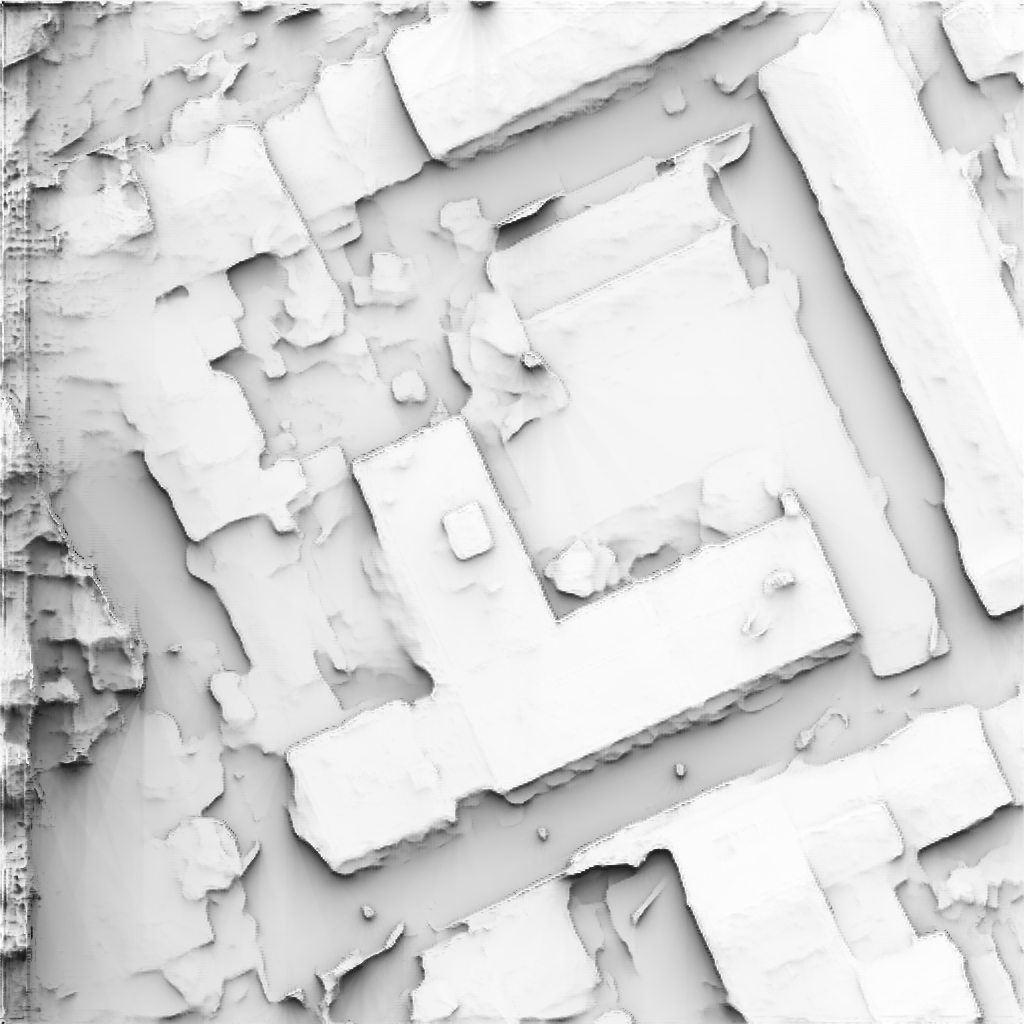}
		\centering{\tiny DeepPruner(KITTI)}
	\end{minipage}
	\begin{minipage}[t]{0.19\textwidth}		
		\includegraphics[width=0.098\linewidth]{figures_supp/color_map.png}
		\includegraphics[width=0.85\linewidth]{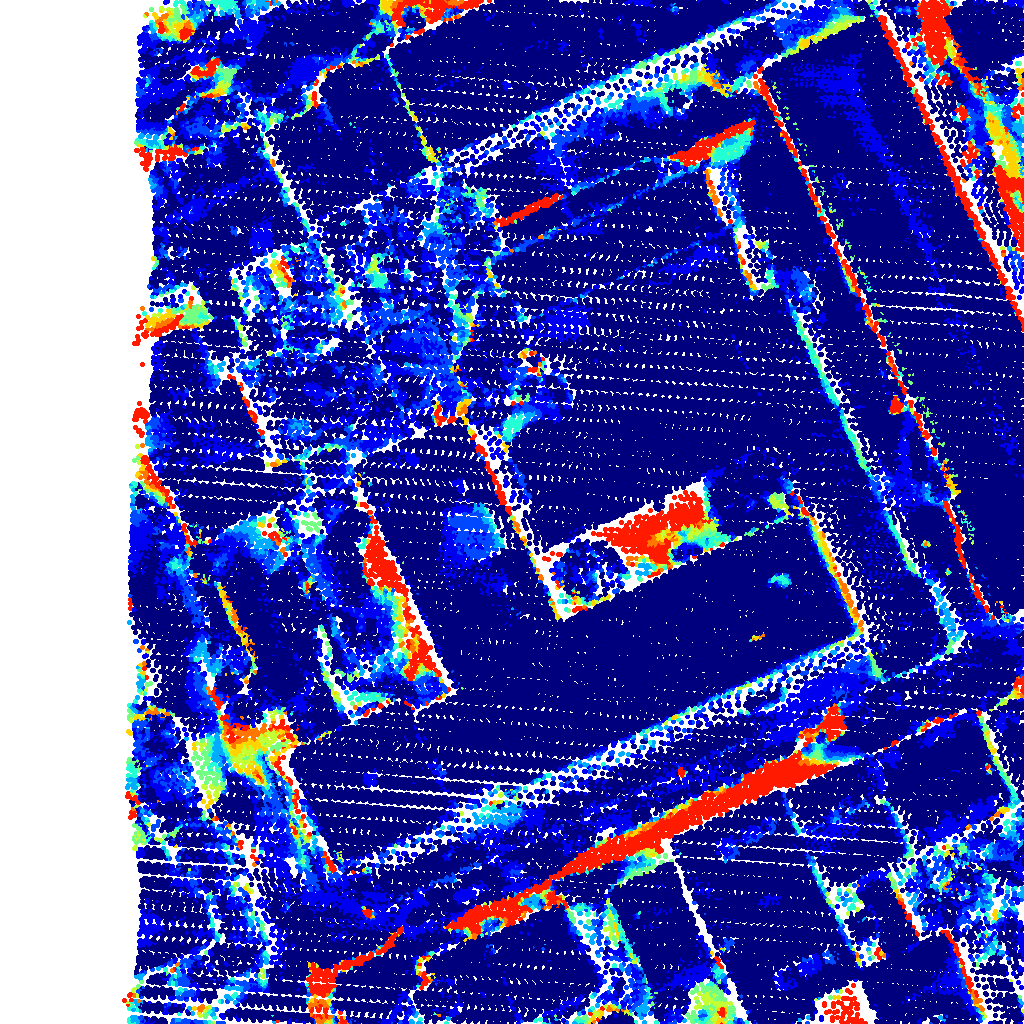}
		\includegraphics[width=\linewidth]{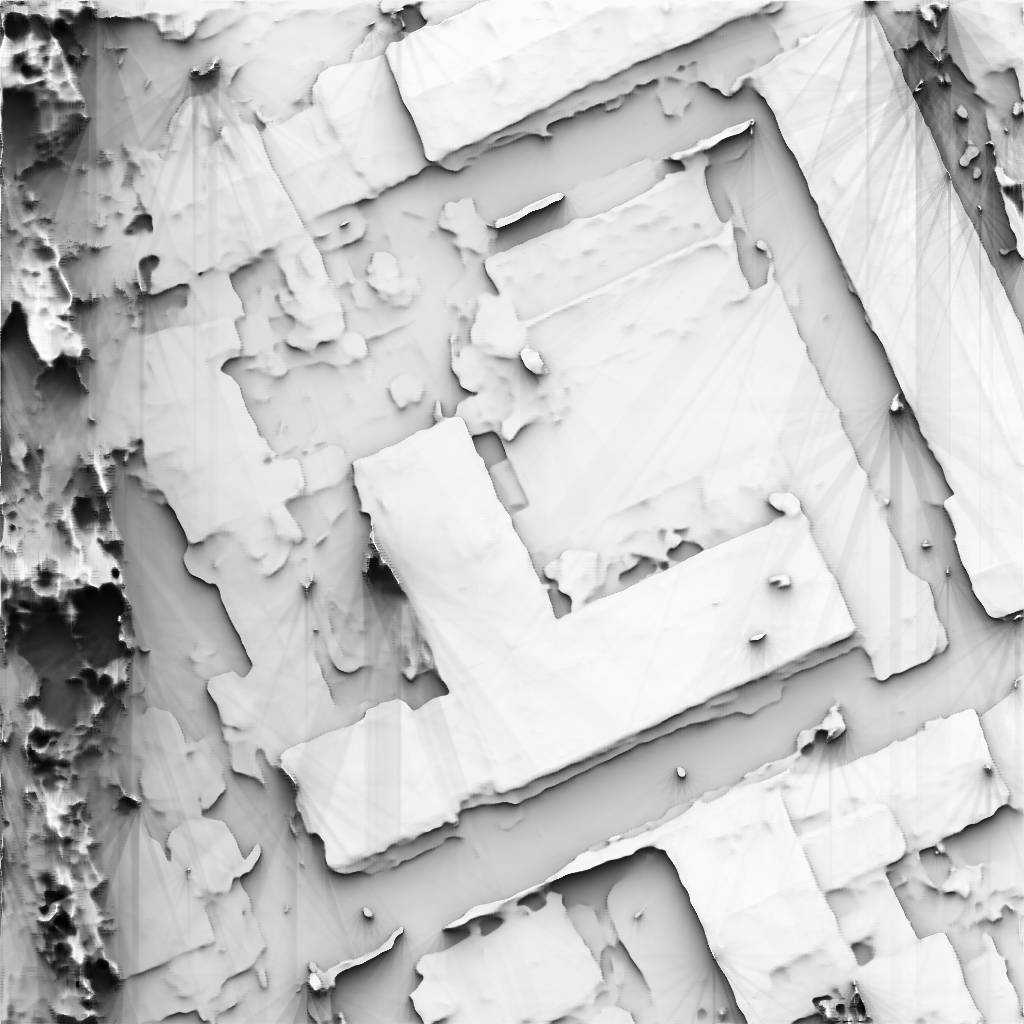}
		\centering{\tiny GAnet(KITTI)}
	\end{minipage}
	\begin{minipage}[t]{0.19\textwidth}	
		\includegraphics[width=0.098\linewidth]{figures_supp/color_map.png}
		\includegraphics[width=0.85\linewidth]{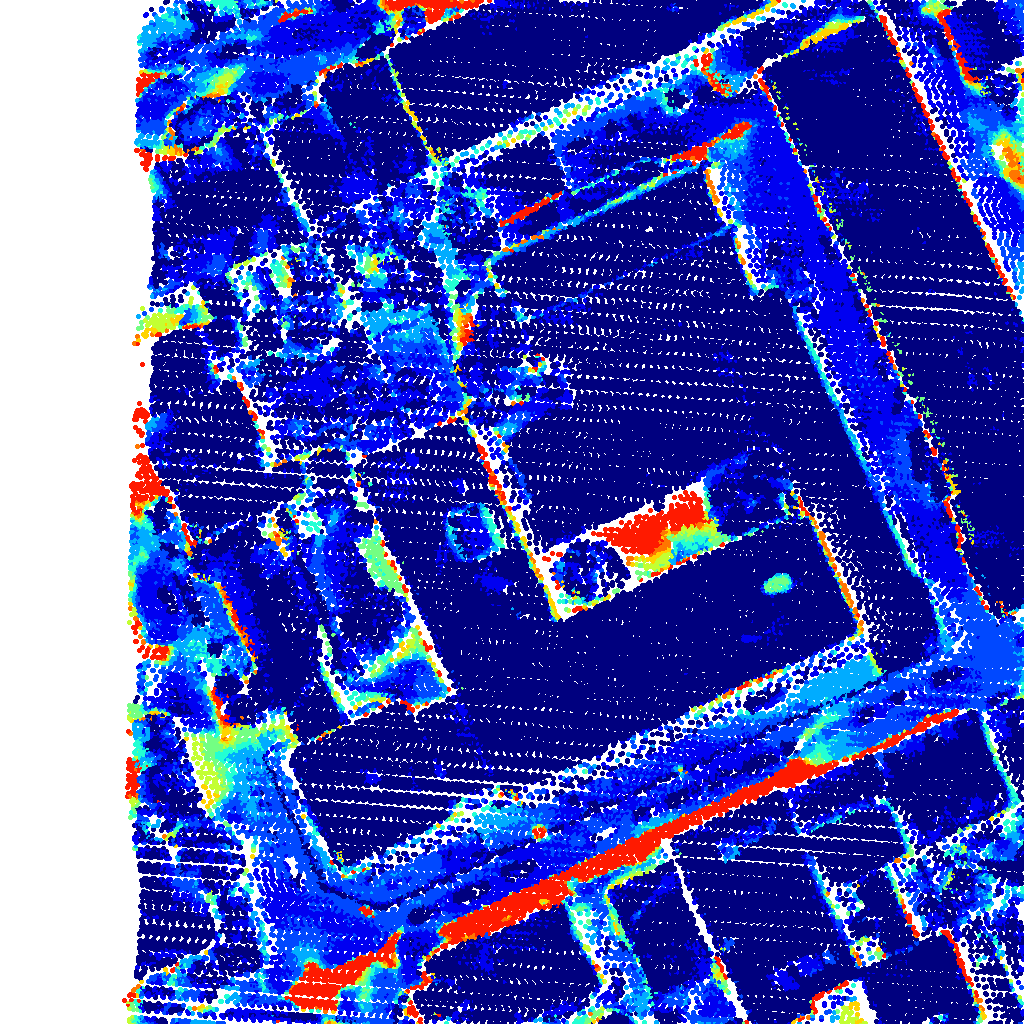}
		\includegraphics[width=\linewidth]{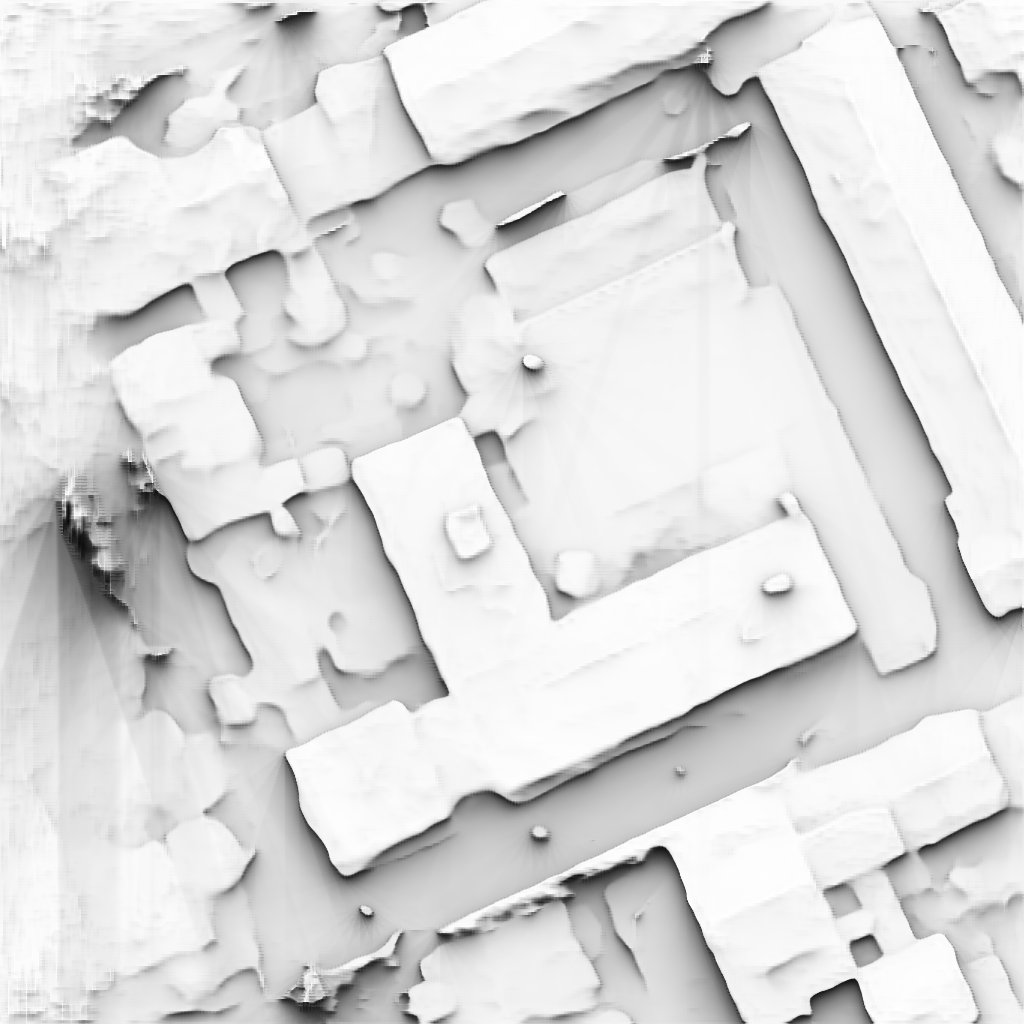}
		\centering{\tiny LEAStereo(KITTI)}
	\end{minipage}
		\begin{minipage}[t]{0.19\textwidth}	
		\includegraphics[width=0.098\linewidth]{figures_supp/color_map.png}
		\includegraphics[width=0.85\linewidth]{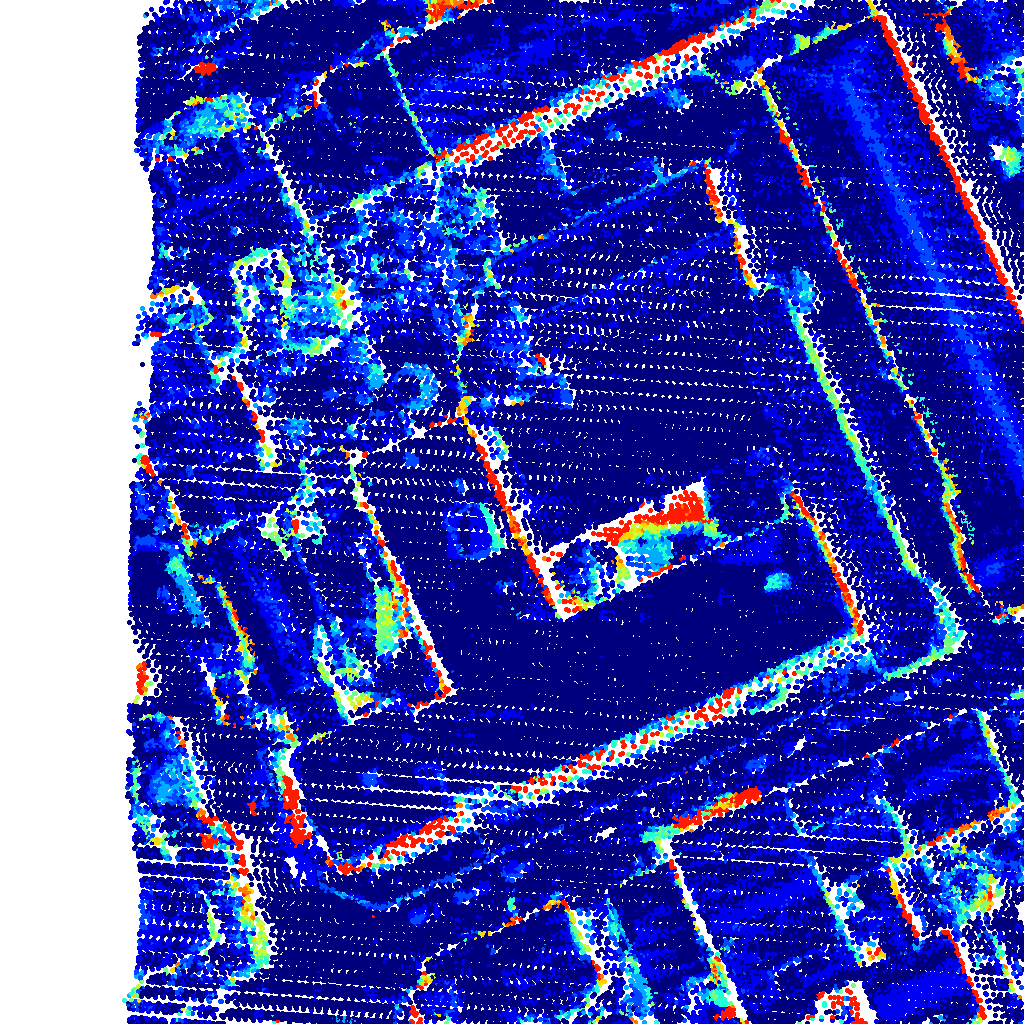}
		\includegraphics[width=\linewidth]{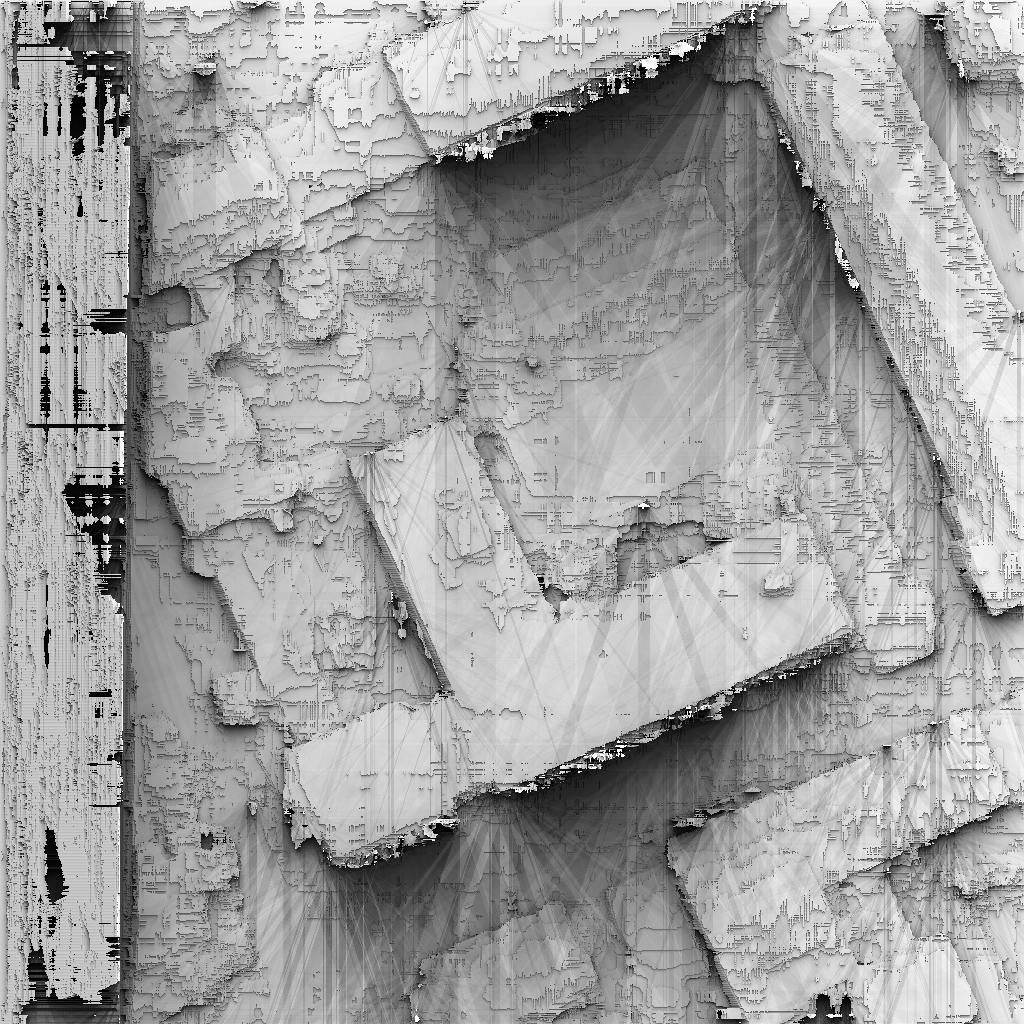}
		\centering{\tiny MC-CNN}
	\end{minipage}
	\begin{minipage}[t]{0.19\textwidth}	
		\includegraphics[width=0.098\linewidth]{figures_supp/color_map.png}
		\includegraphics[width=0.85\linewidth]{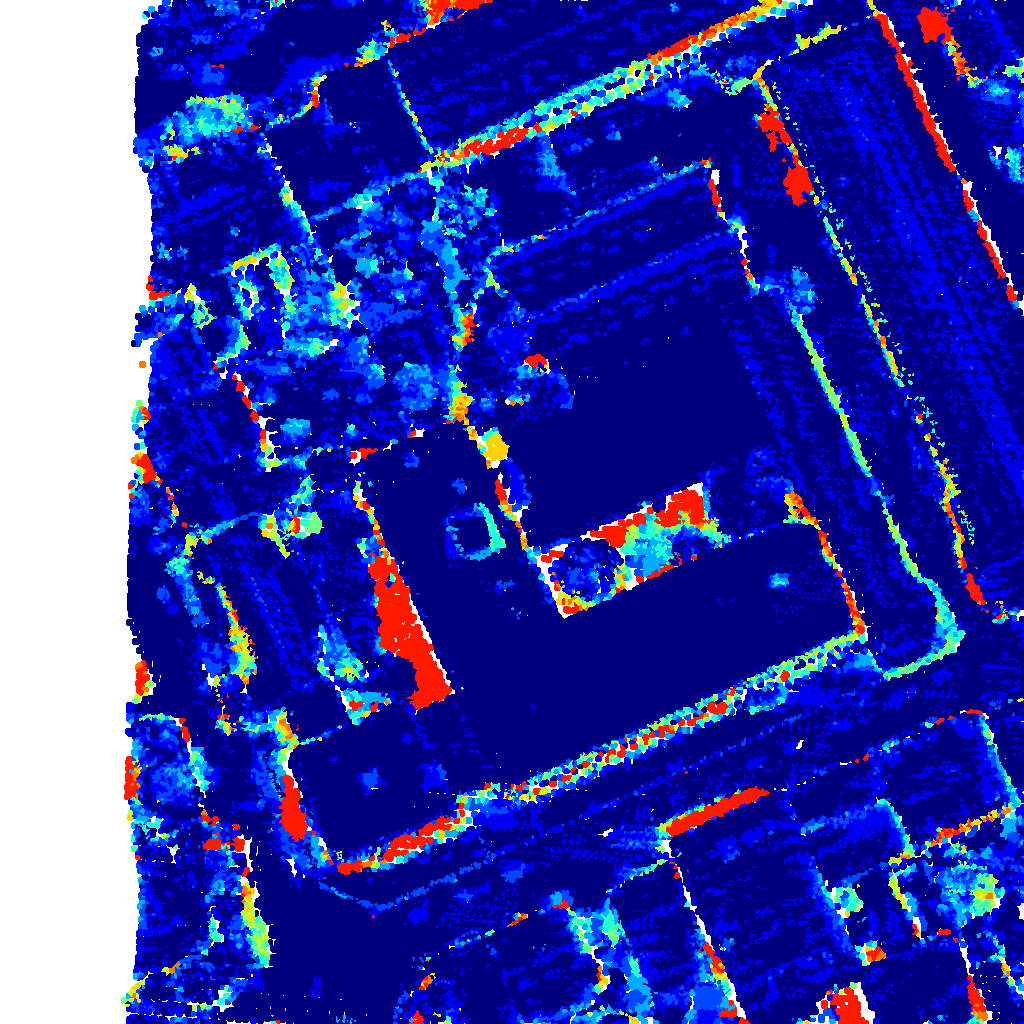}
		\includegraphics[width=\linewidth]{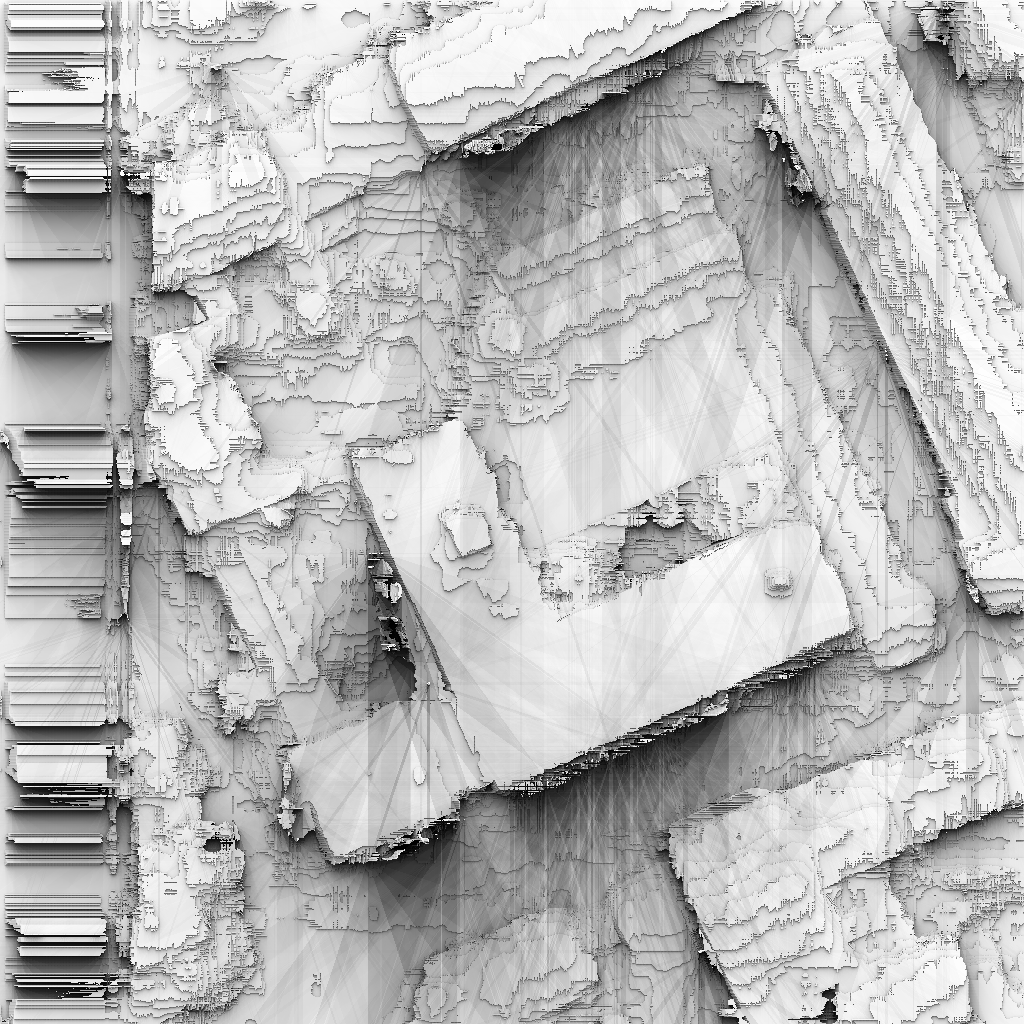}
		\centering{\tiny DeepFeature}
	\end{minipage}
	\begin{minipage}[t]{0.19\textwidth}	
		\includegraphics[width=0.098\linewidth]{figures_supp/color_map.png}
		\includegraphics[width=0.85\linewidth]{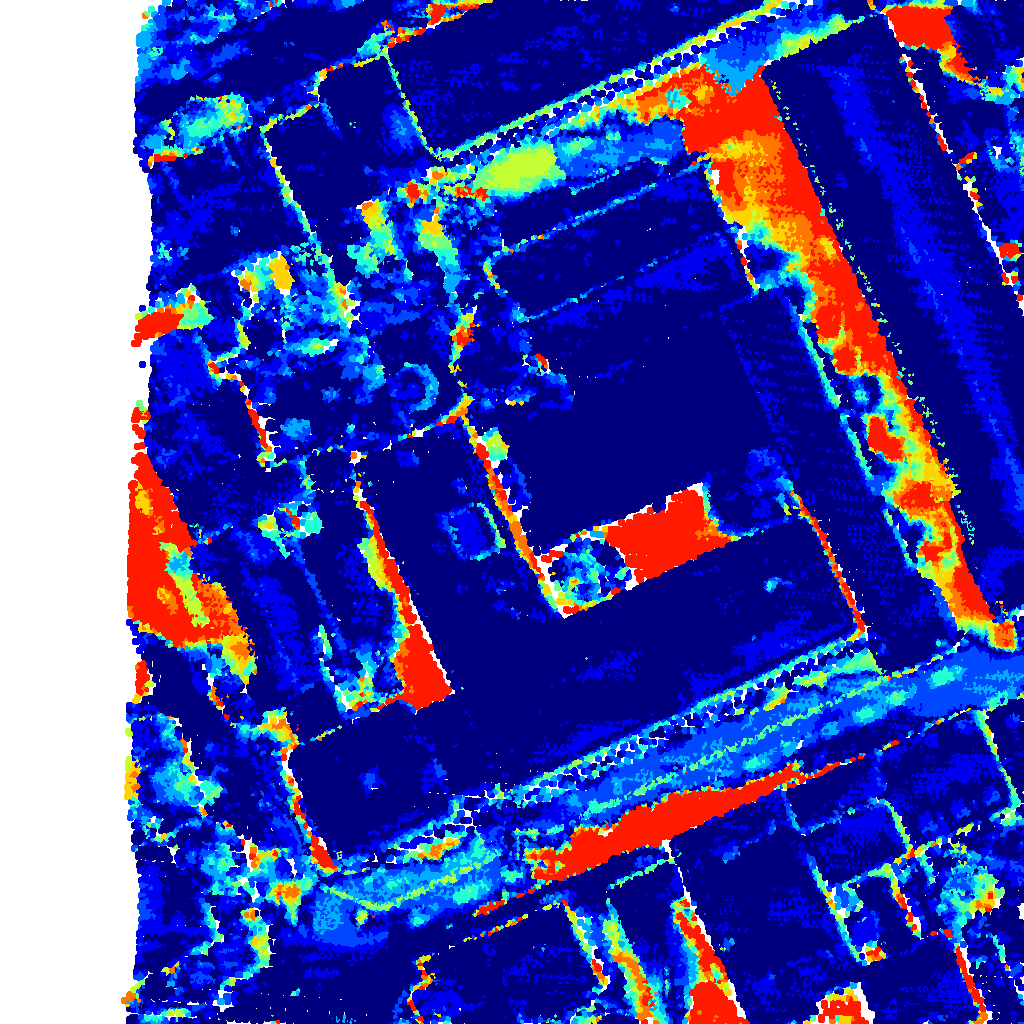}
		\includegraphics[width=\linewidth]{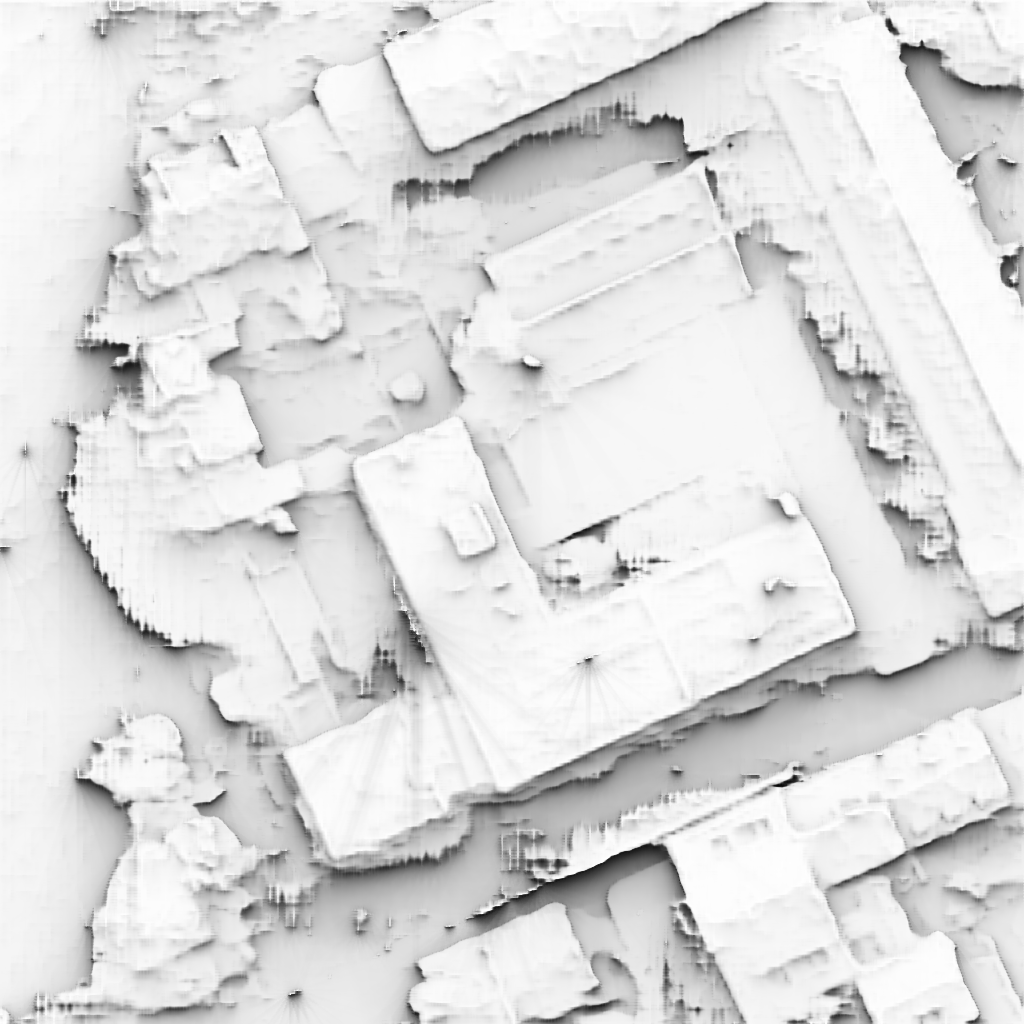}
		\centering{\tiny PSM net}
	\end{minipage}
	\begin{minipage}[t]{0.19\textwidth}		
		\includegraphics[width=0.098\linewidth]{figures_supp/color_map.png}
		\includegraphics[width=0.85\linewidth]{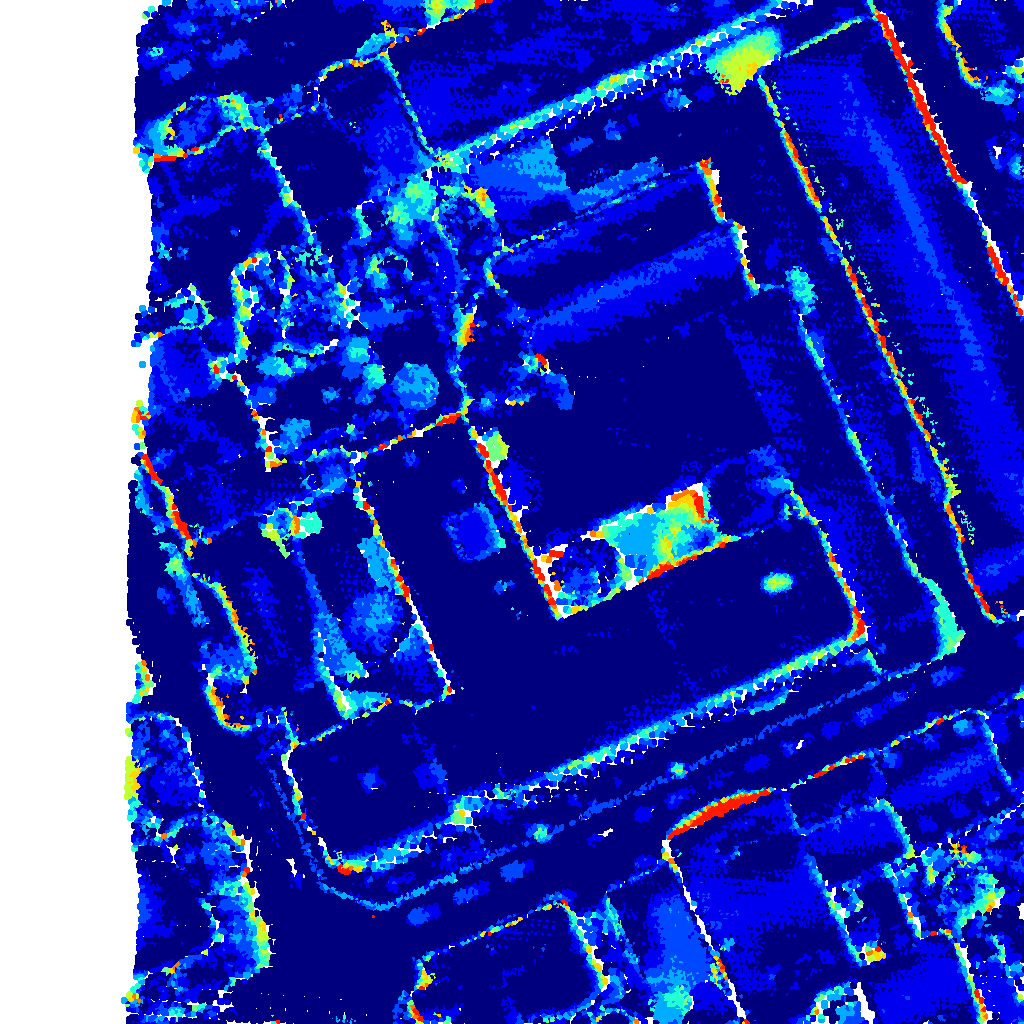}
		\includegraphics[width=\linewidth]{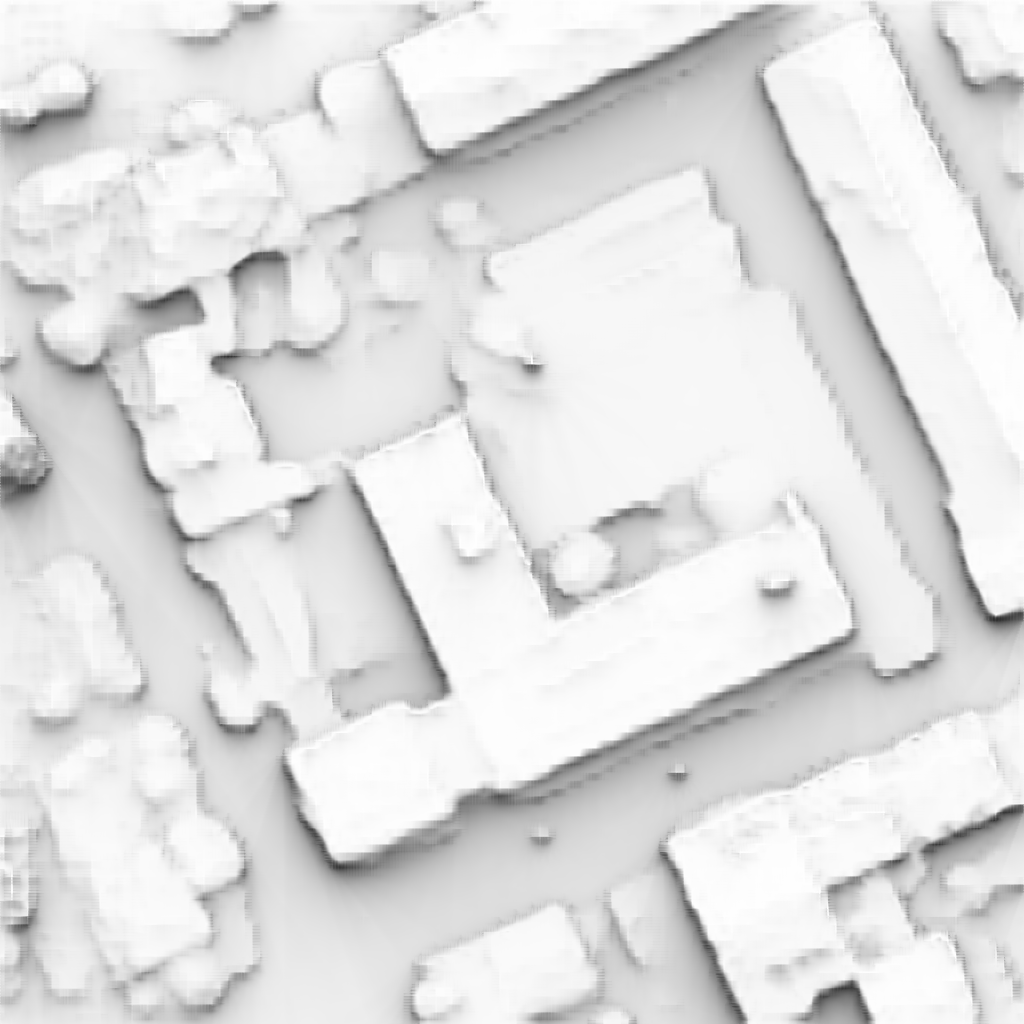}
		\centering{\tiny HRS net}
	\end{minipage}
	\begin{minipage}[t]{0.19\textwidth}	
		\includegraphics[width=0.098\linewidth]{figures_supp/color_map.png}
		\includegraphics[width=0.85\linewidth]{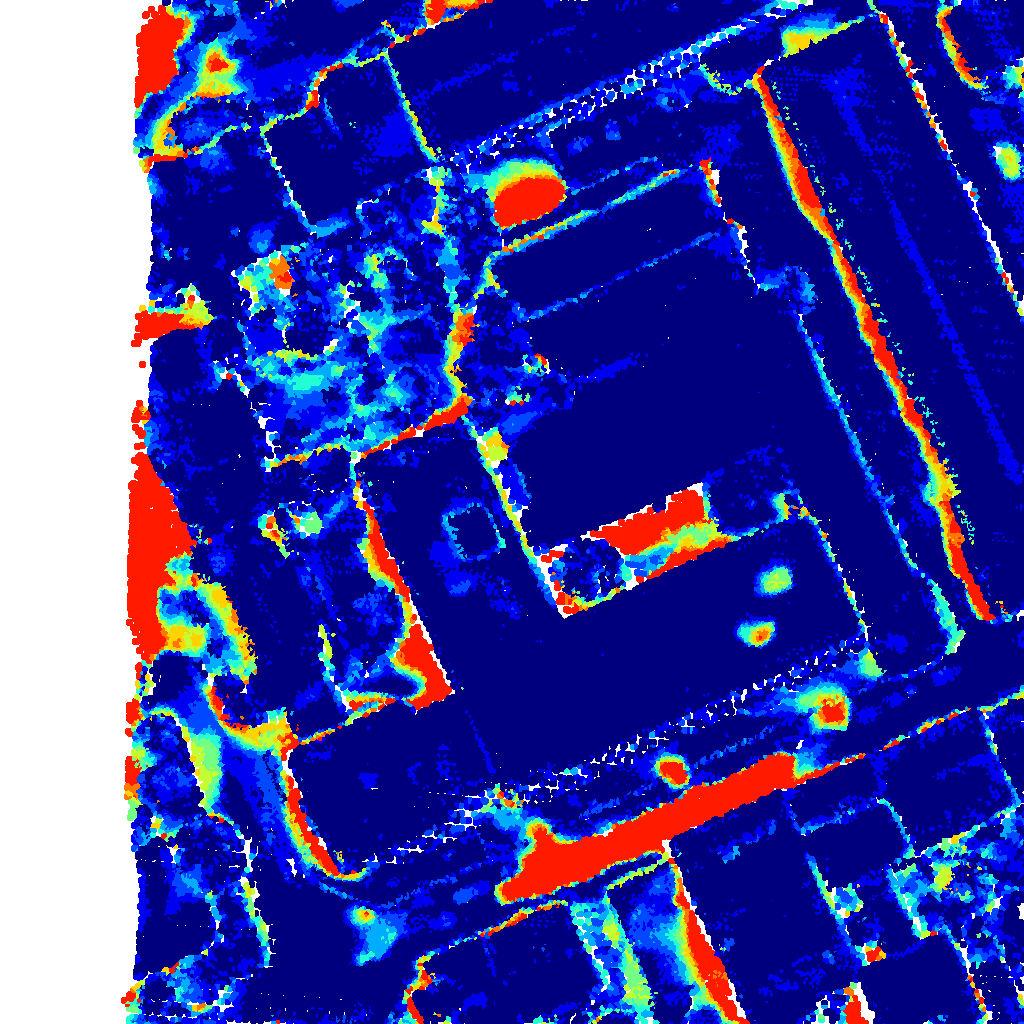}
		\includegraphics[width=\linewidth]{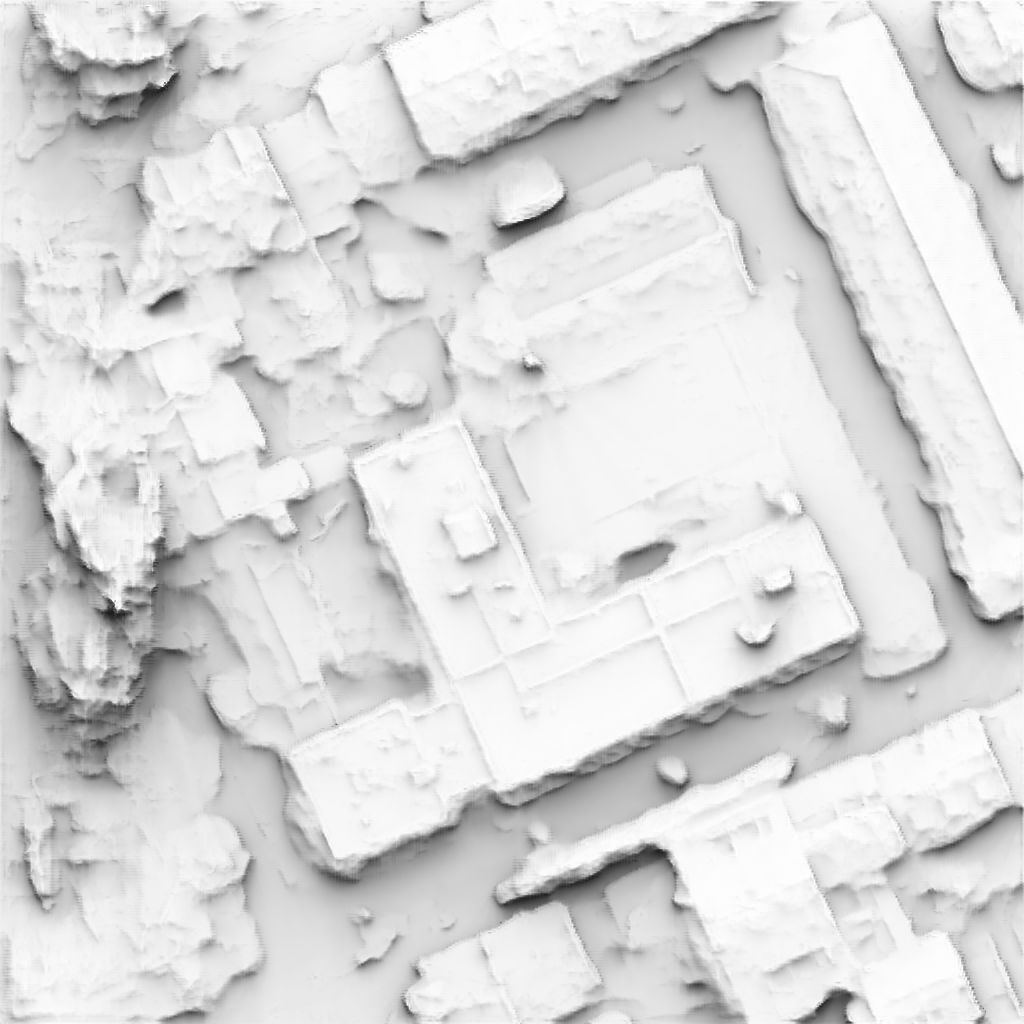}
		\centering{\tiny DeepPruner}
	\end{minipage}
	\begin{minipage}[t]{0.19\textwidth}		
		\includegraphics[width=0.098\linewidth]{figures_supp/color_map.png}
		\includegraphics[width=0.85\linewidth]{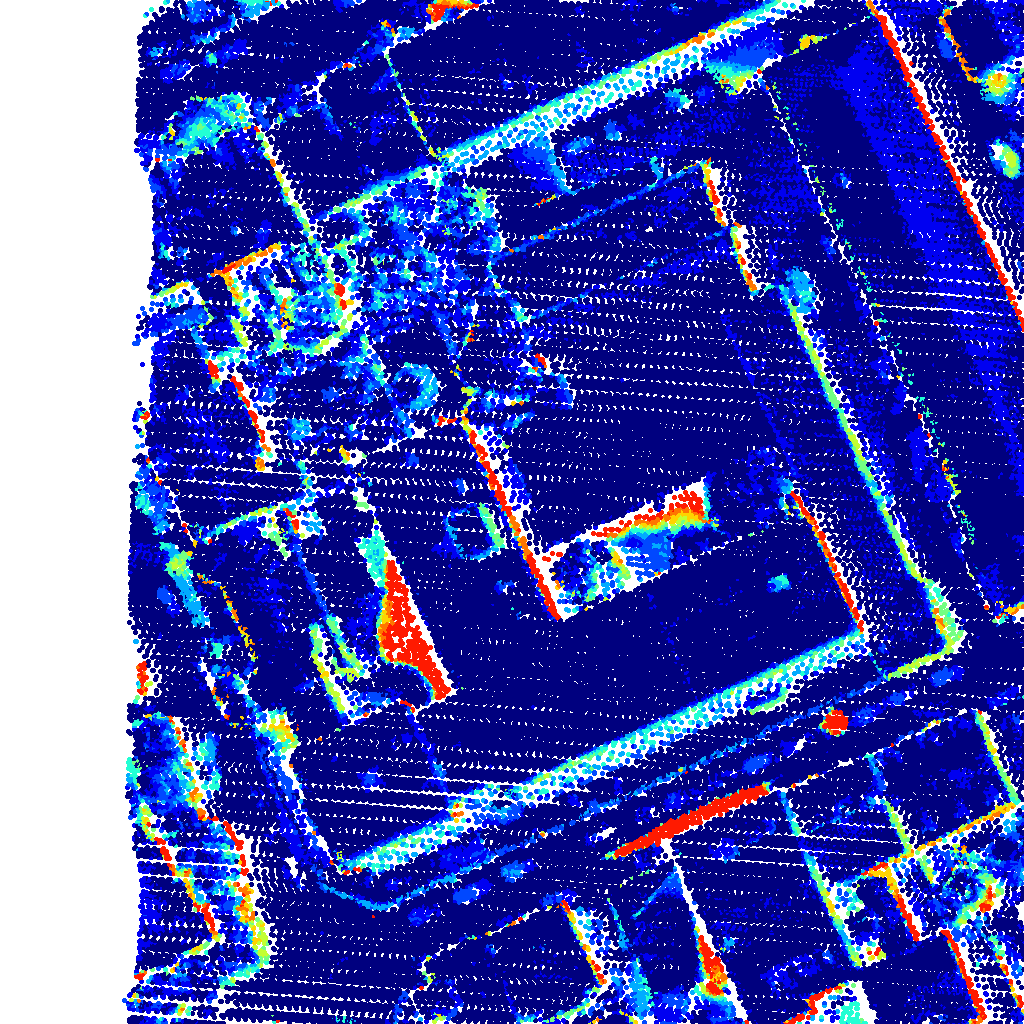}
		\includegraphics[width=\linewidth]{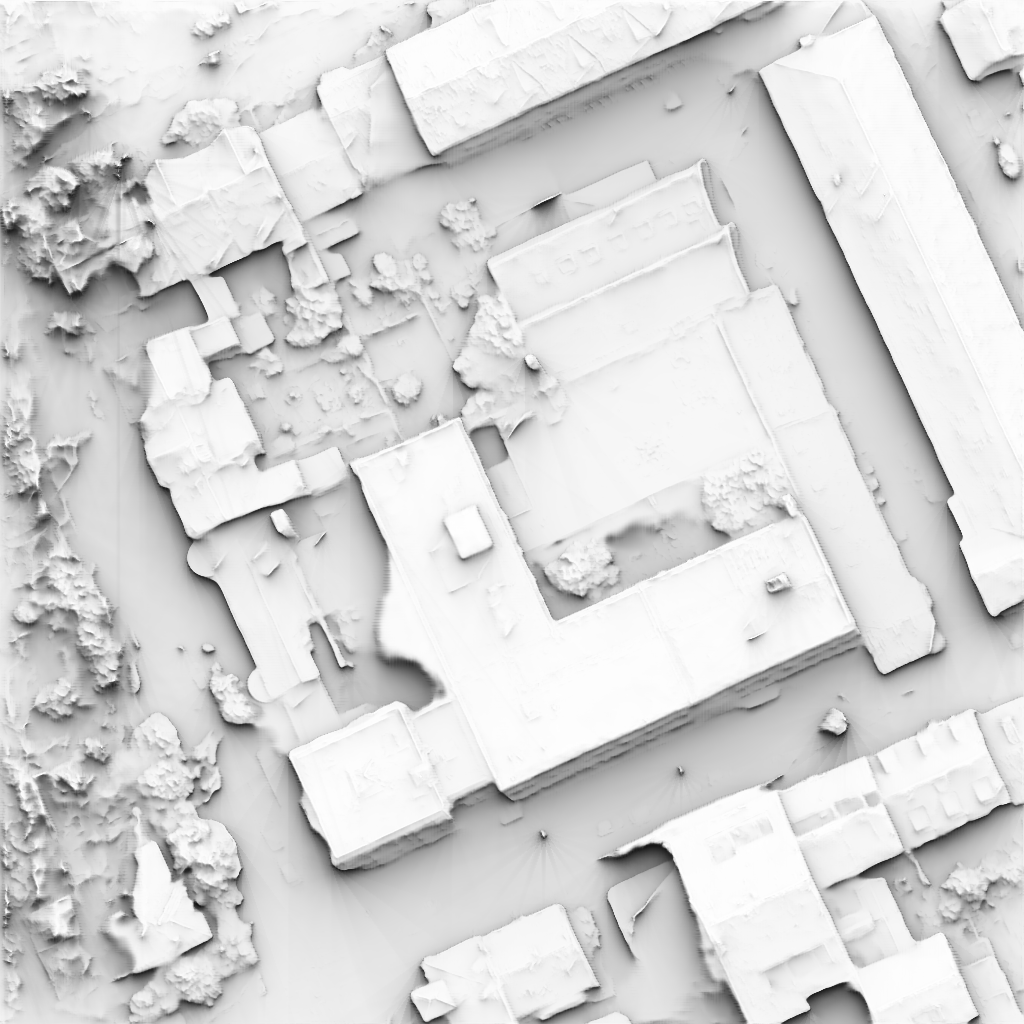}
		\centering{\tiny GAnet}
	\end{minipage}
	\begin{minipage}[t]{0.19\textwidth}	
		\includegraphics[width=0.098\linewidth]{figures_supp/color_map.png}
		\includegraphics[width=0.85\linewidth]{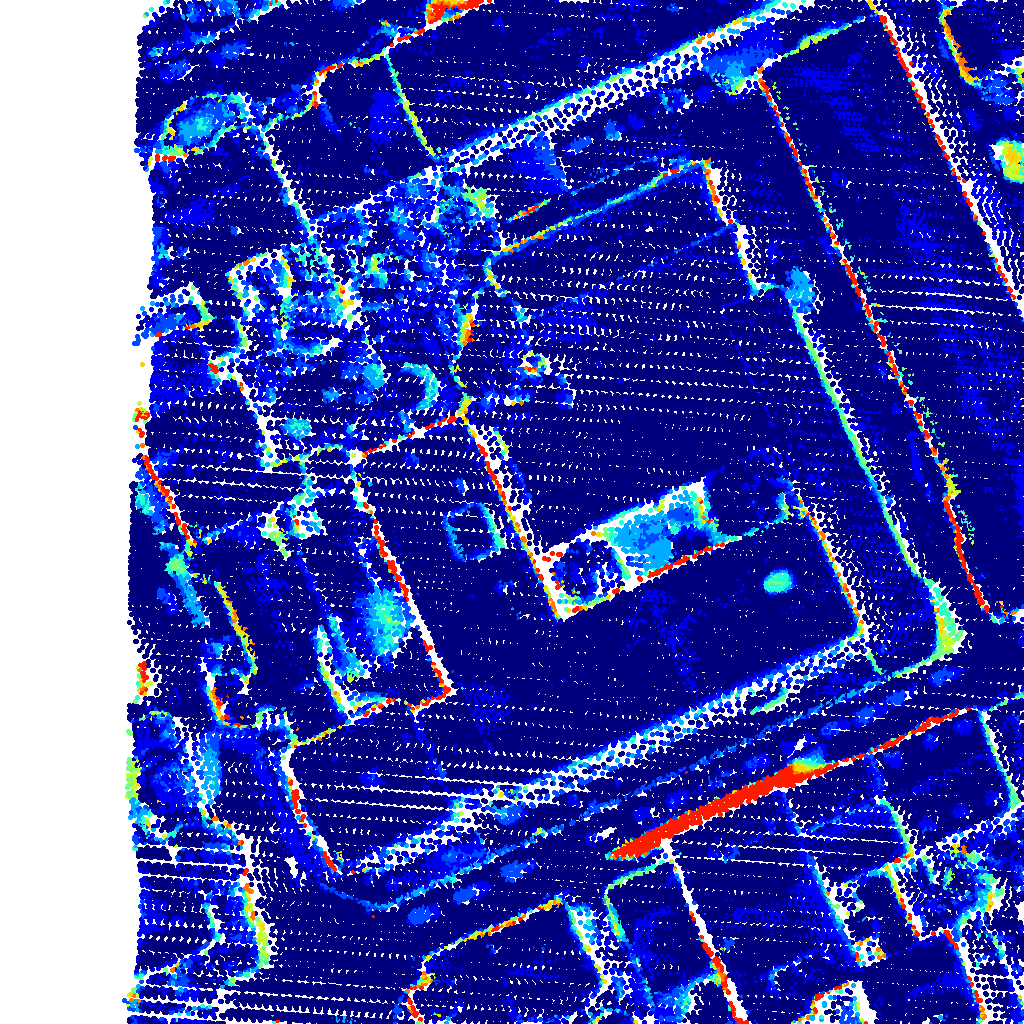}
		\includegraphics[width=\linewidth]{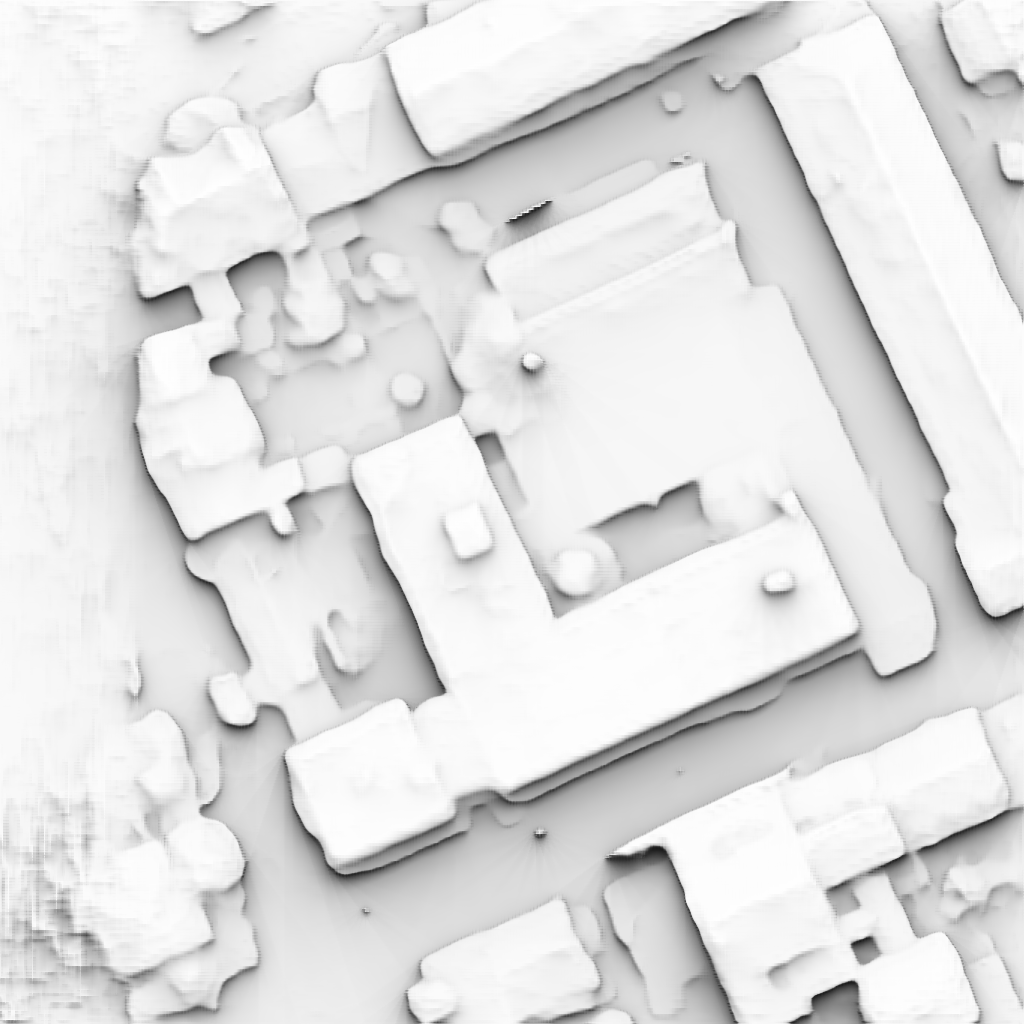}
		\centering{\tiny LEAStereo}
	\end{minipage}
	\caption{Error map and disparity visualization on building area for Toulouse UMBRA.}
	\label{Figure.umbrabulding}
\end{figure}

\paragraph{Toulouse UMBRA}
On large man-made objects (with the exception of discontinuities) all the methods perform relatively well (cf. \Cref{Figure.umbrabulding}). We also observe the following tendencies:
\begin{itemize}
    \item GraphCuts-based methods outrun the SGM-based methods.
    \item \textit{PSM net} performs poorly in dark shadow areas.
    \item \textit{LEAStereo} performs best among the DL-based methods.
\end{itemize}


\begin{figure}[tp]
	\begin{minipage}[t]{0.19\textwidth}	
		\includegraphics[width=0.098\linewidth]{figures_supp/color_map.png}
		\includegraphics[width=0.85\linewidth]{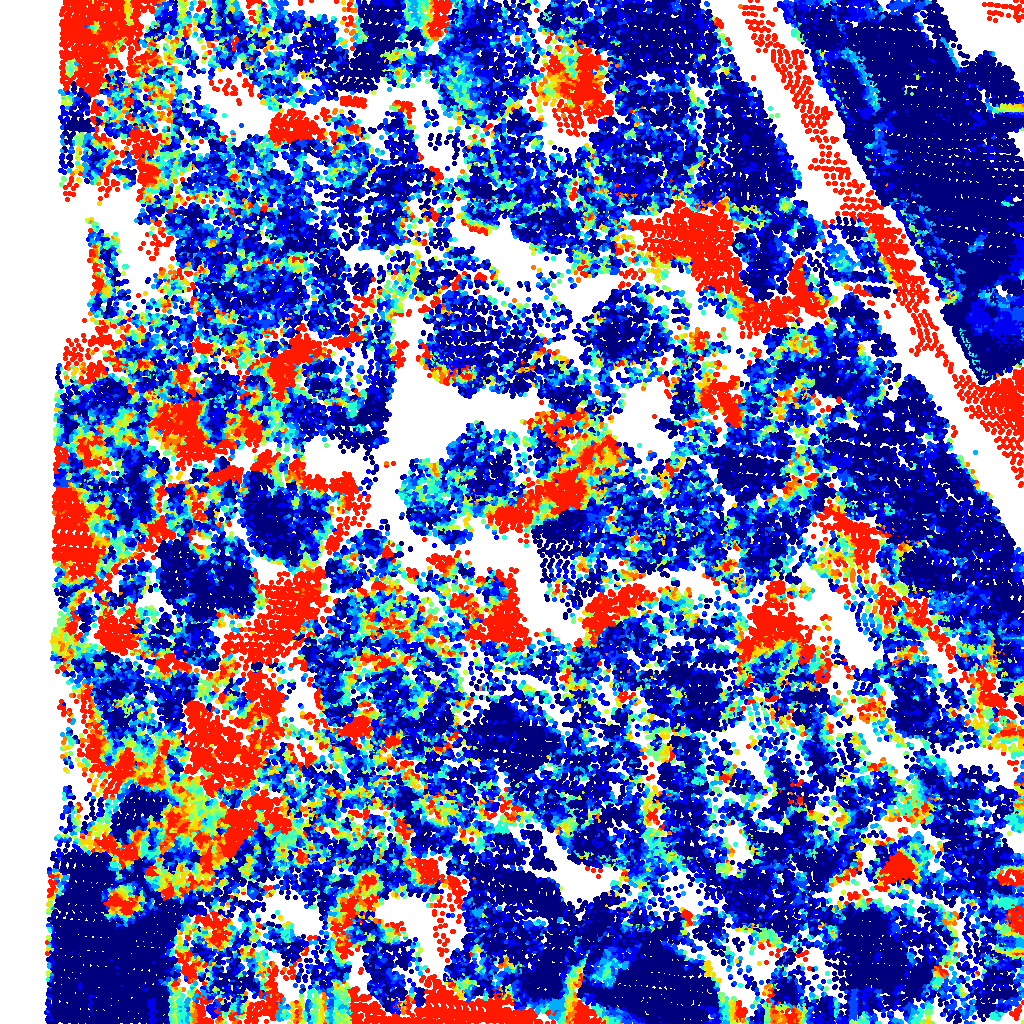}
		\includegraphics[width=\linewidth]{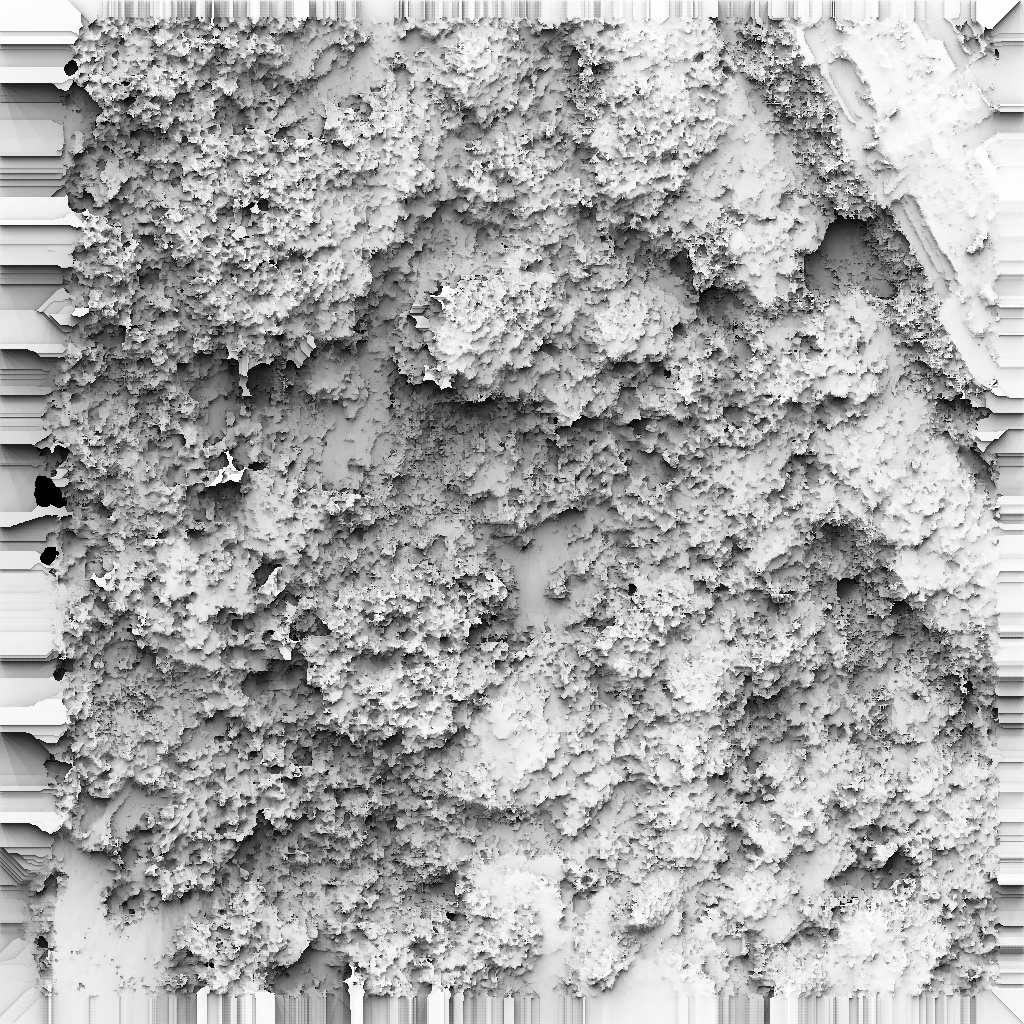}
		\centering{\tiny MICMAC}
	\end{minipage}
	\begin{minipage}[t]{0.19\textwidth}	
		\includegraphics[width=0.098\linewidth]{figures_supp/color_map.png}
		\includegraphics[width=0.85\linewidth]{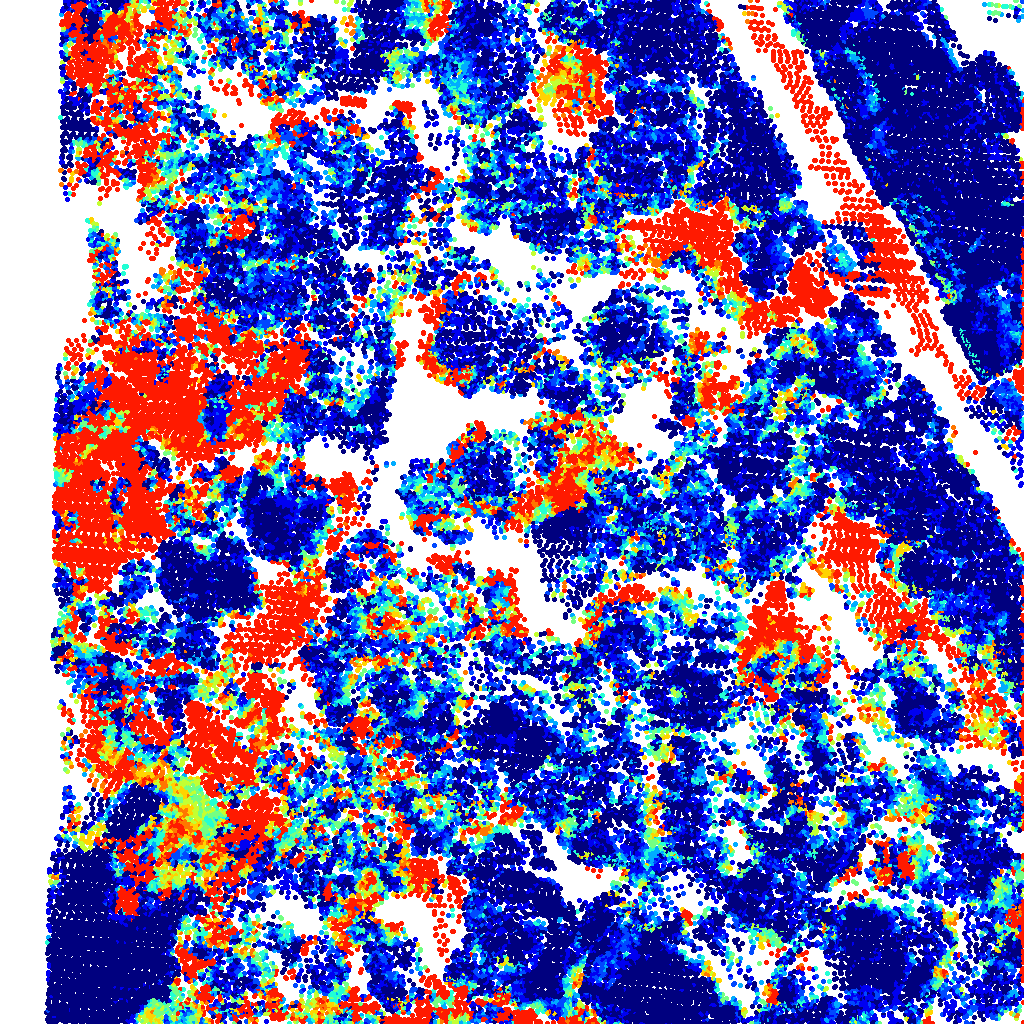}
		\includegraphics[width=\linewidth]{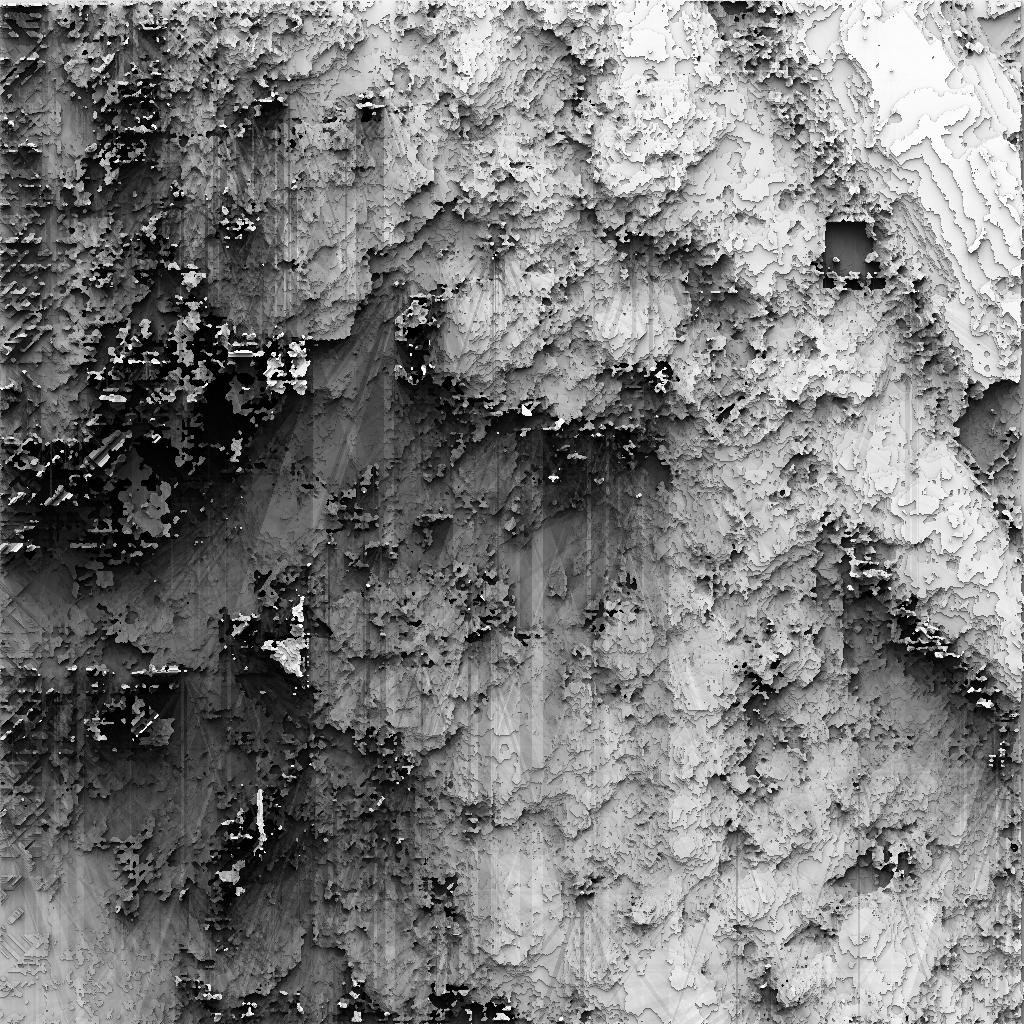}
		\centering{\tiny SGM(CUDA)}
	\end{minipage}
	\begin{minipage}[t]{0.19\textwidth}	
		\includegraphics[width=0.098\linewidth]{figures_supp/color_map.png}
		\includegraphics[width=0.85\linewidth]{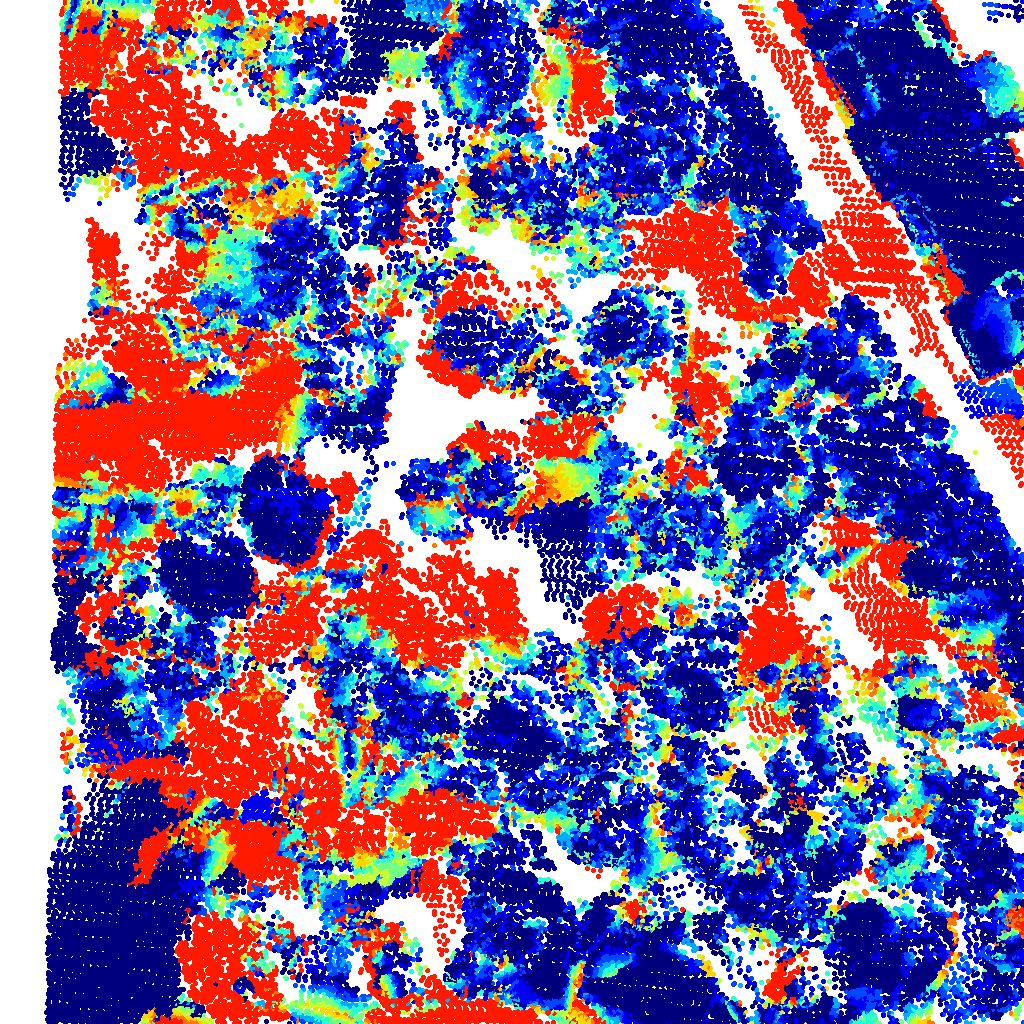}
		\includegraphics[width=\linewidth]{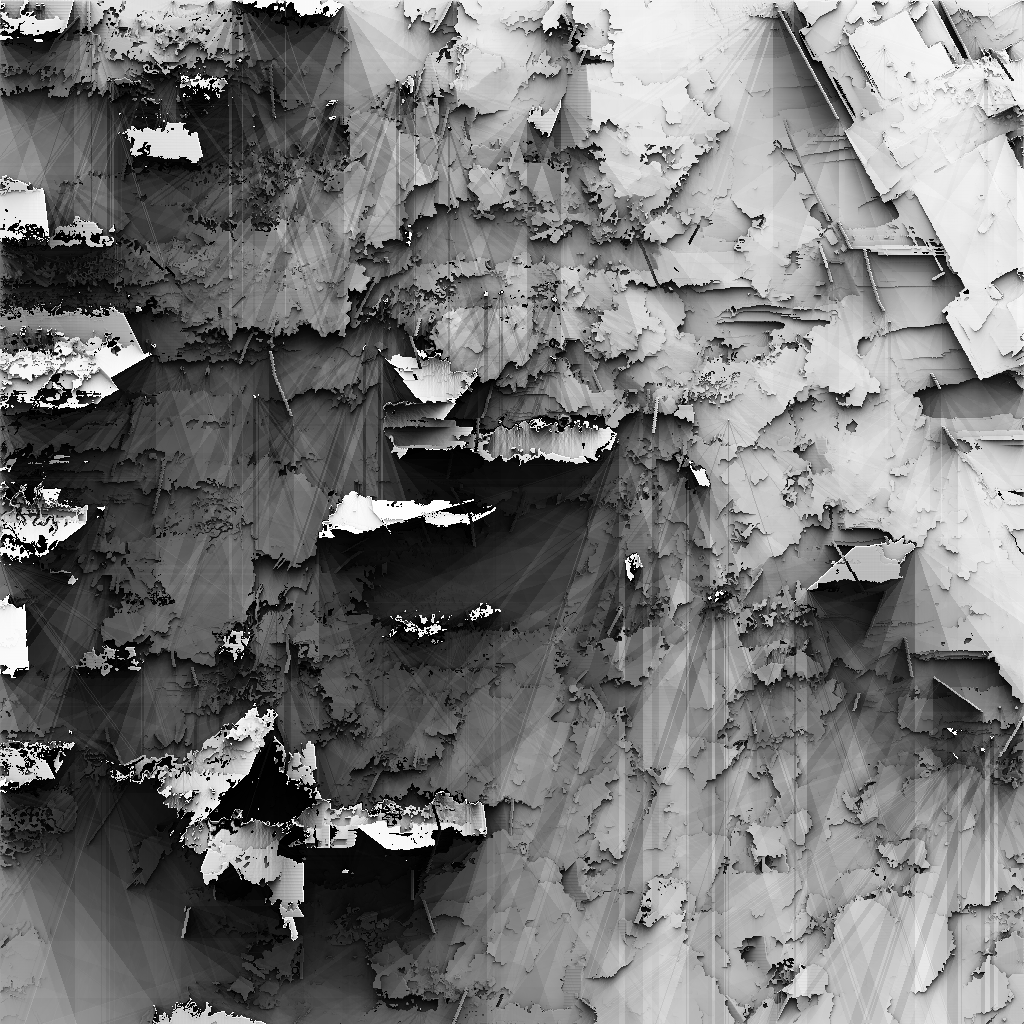}
		\centering{\tiny GraphCuts}
	\end{minipage}
	\begin{minipage}[t]{0.19\textwidth}	
		\includegraphics[width=0.098\linewidth]{figures_supp/color_map.png}
		\includegraphics[width=0.85\linewidth]{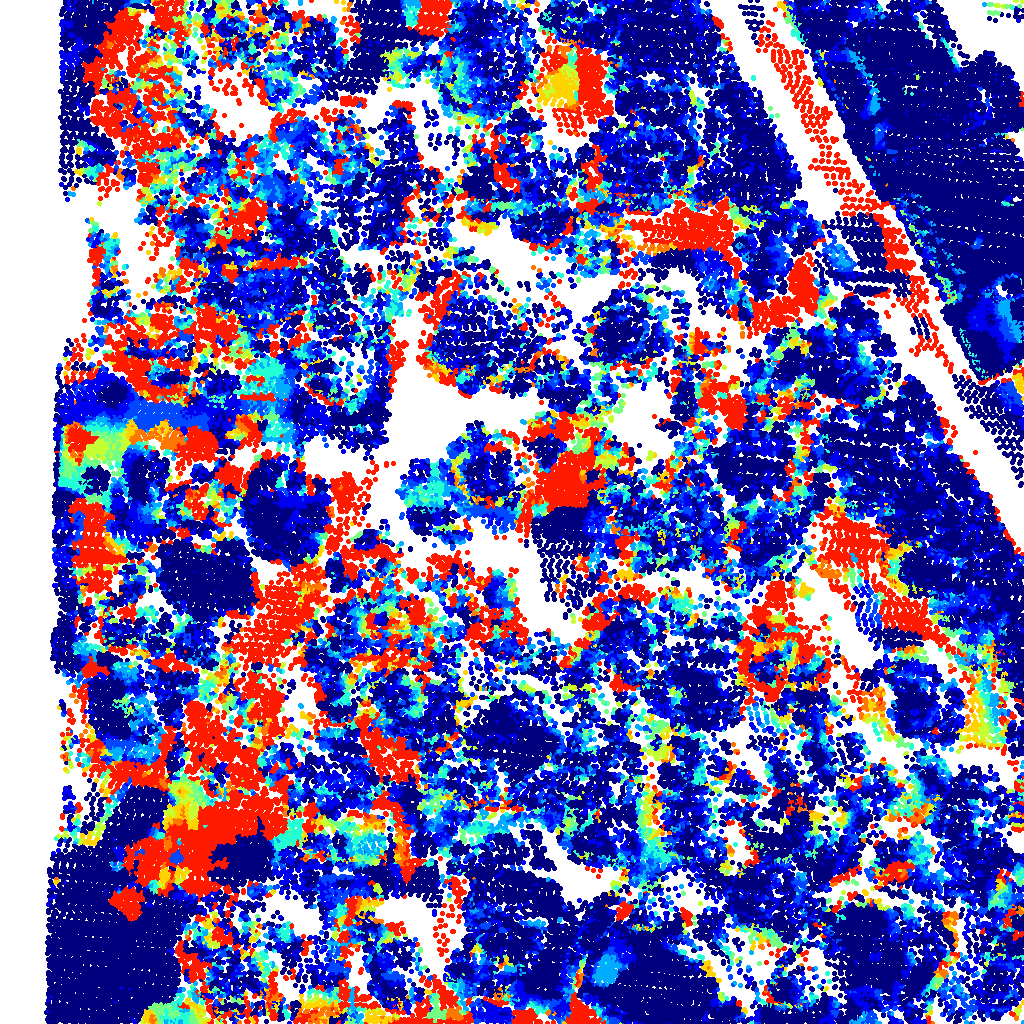}
		\includegraphics[width=\linewidth]{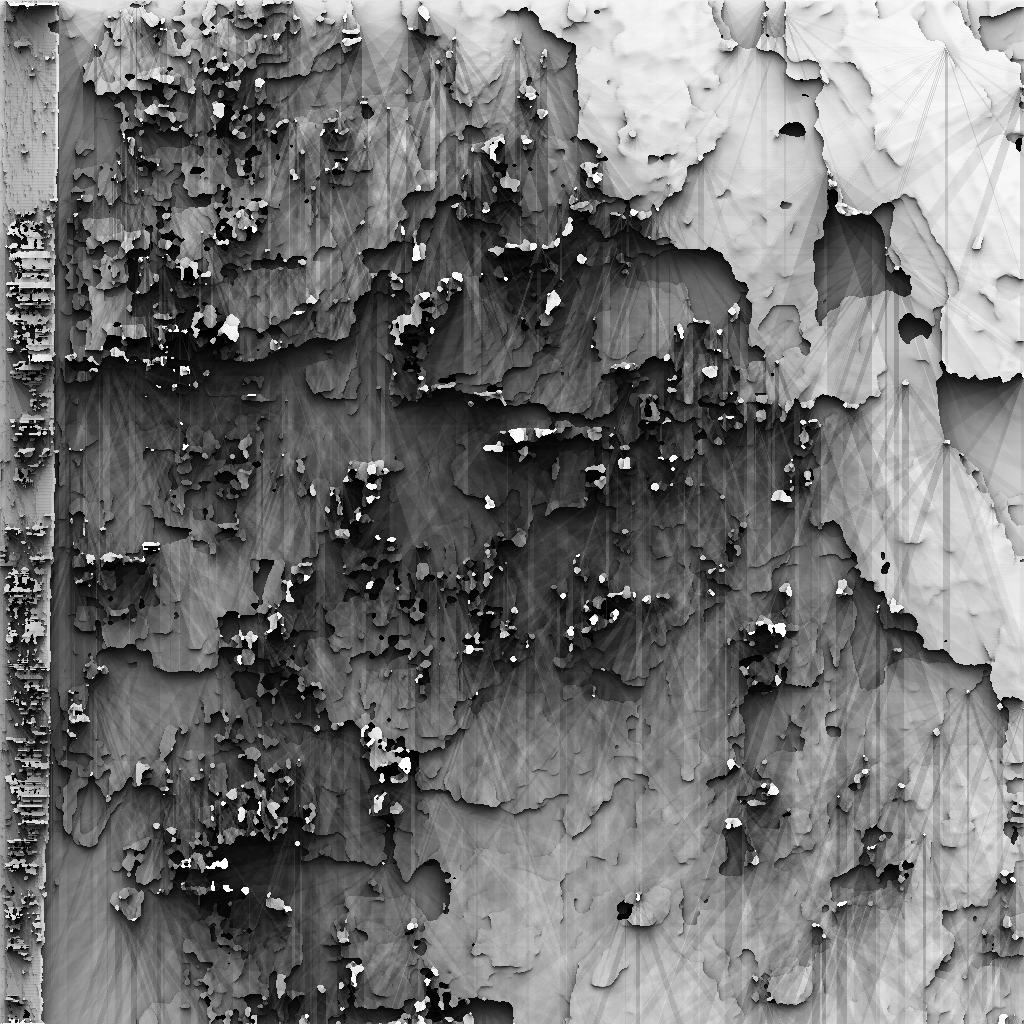}
		\centering{\tiny CBMV(SGM)}
	\end{minipage}
	\begin{minipage}[t]{0.19\textwidth}	
		\includegraphics[width=0.098\linewidth]{figures_supp/color_map.png}
		\includegraphics[width=0.85\linewidth]{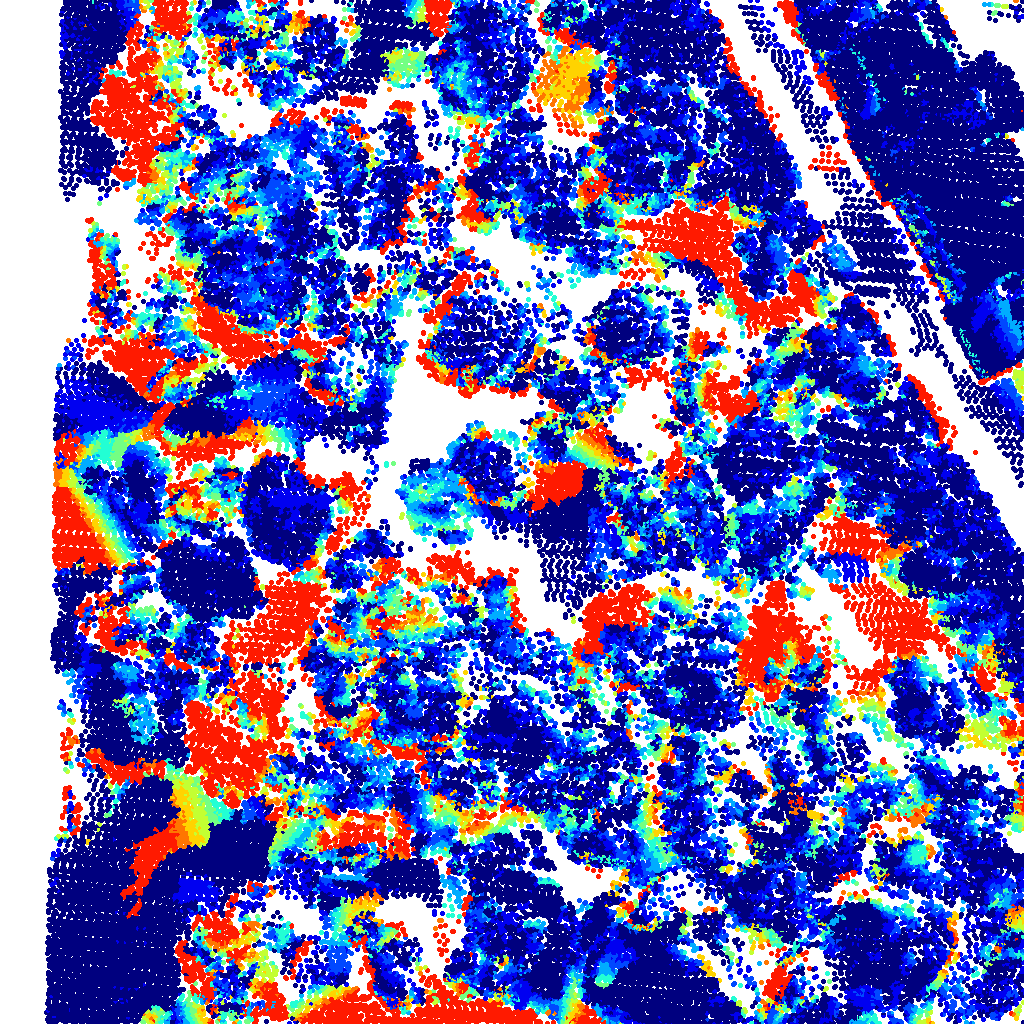}
		\includegraphics[width=\linewidth]{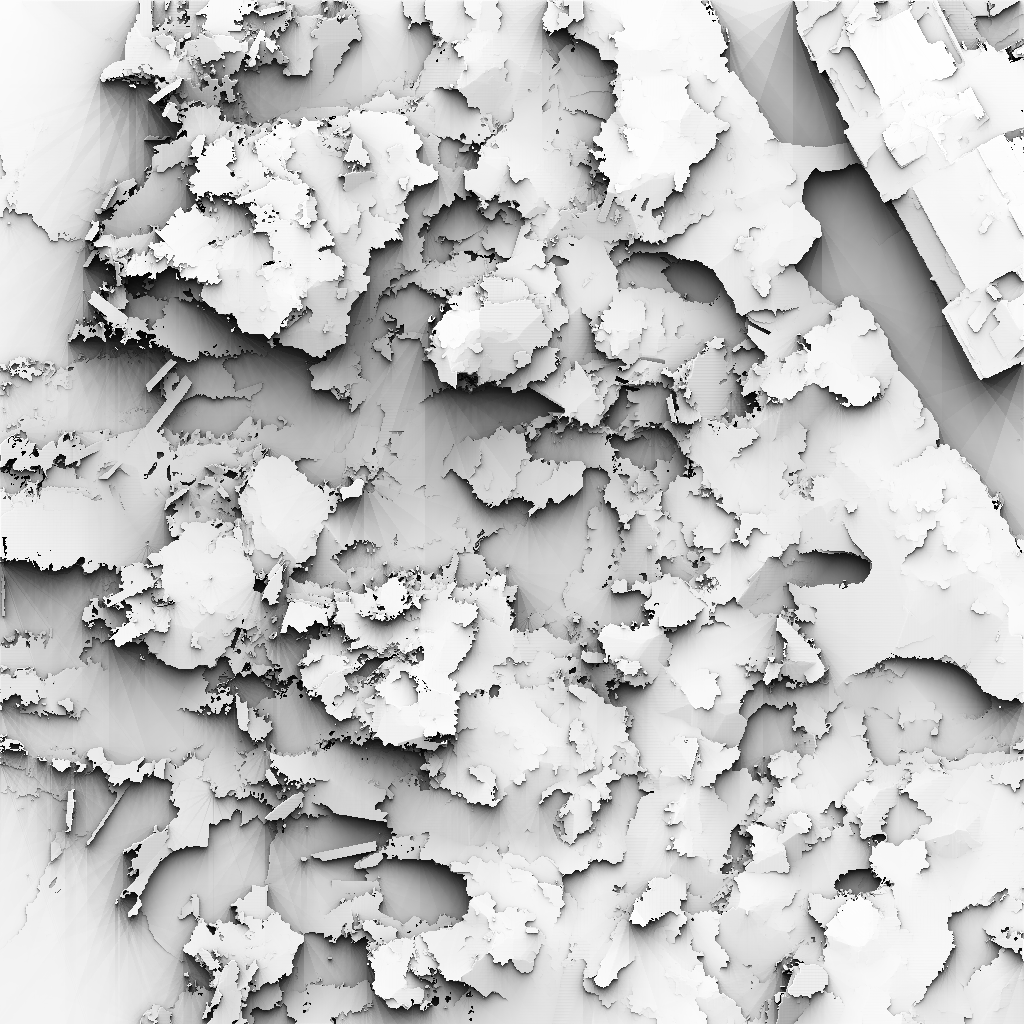}
		\centering{\tiny CBMV(GraphCuts)}
	\end{minipage}
	\begin{minipage}[t]{0.19\textwidth}	
		\includegraphics[width=0.098\linewidth]{figures_supp/color_map.png}
		\includegraphics[width=0.85\linewidth]{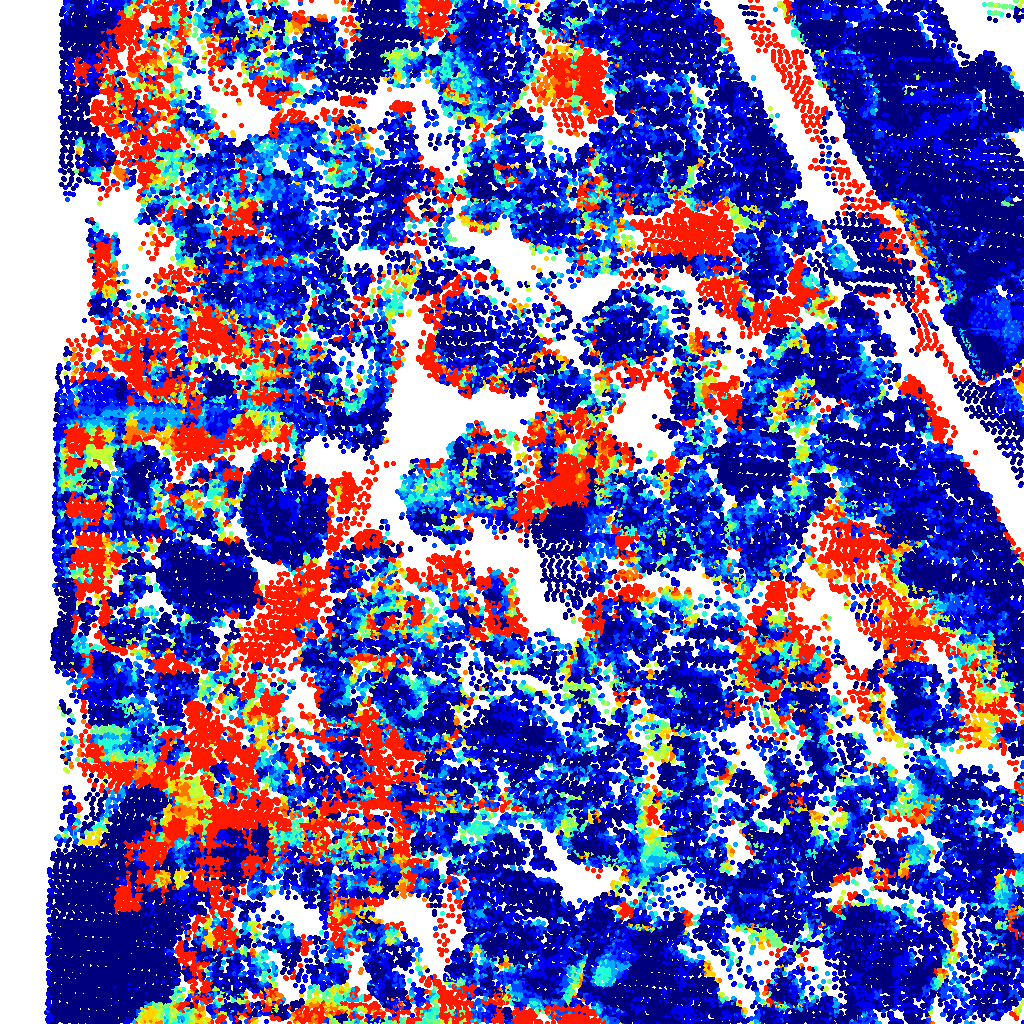}
		\includegraphics[width=\linewidth]{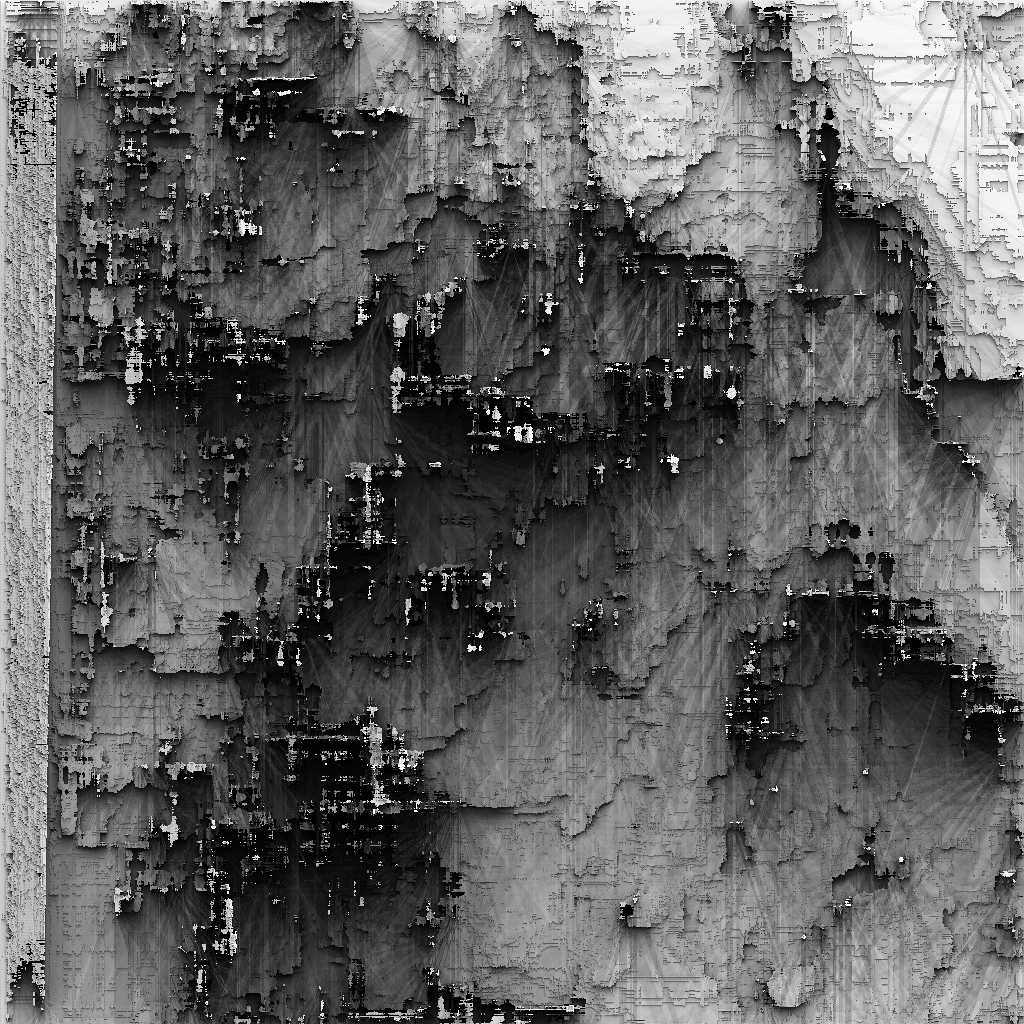}
		\centering{\tiny MC-CNN(KITTI)}
	\end{minipage}
	\begin{minipage}[t]{0.19\textwidth}	
		\includegraphics[width=0.098\linewidth]{figures_supp/color_map.png}
		\includegraphics[width=0.85\linewidth]{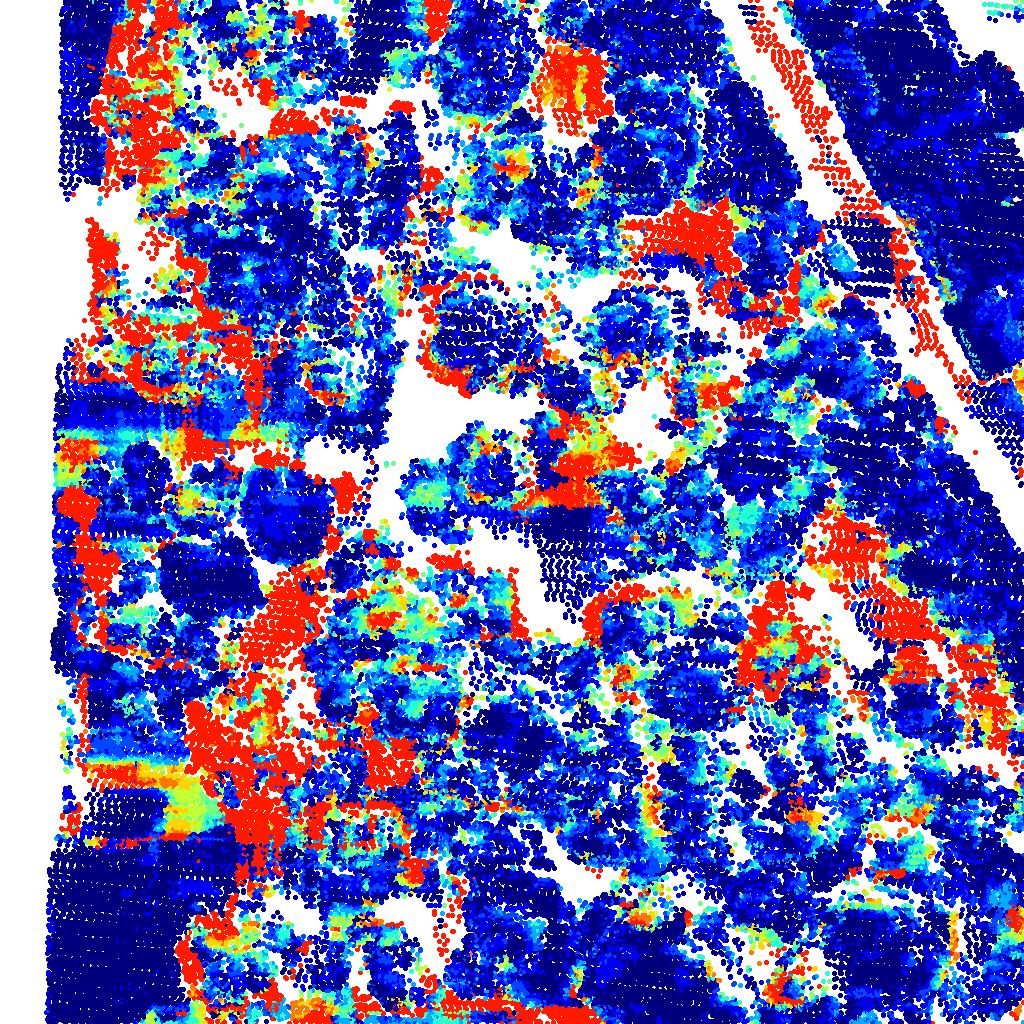}
		\includegraphics[width=\linewidth]{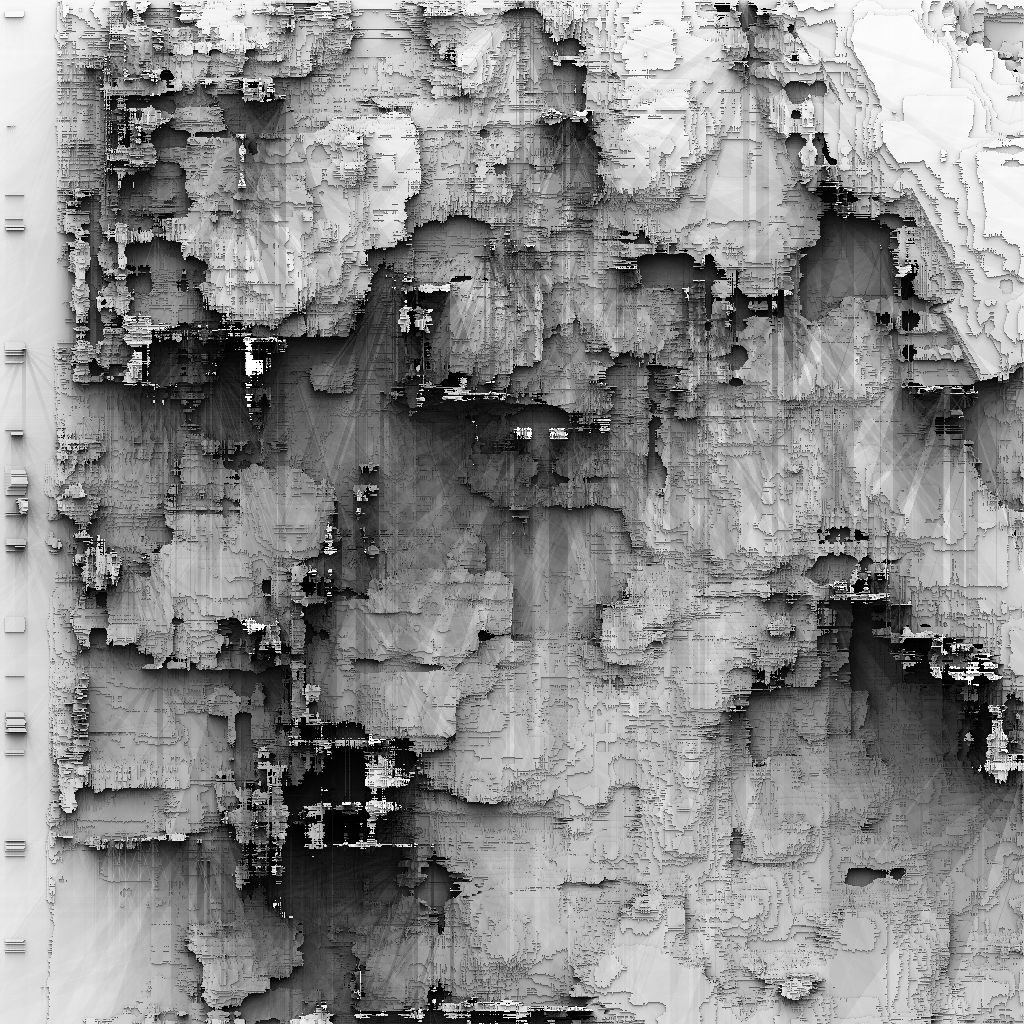}
		\centering{\tiny DeepFeature(KITTI)}
	\end{minipage}
	\begin{minipage}[t]{0.19\textwidth}	
		\includegraphics[width=0.098\linewidth]{figures_supp/color_map.png}
		\includegraphics[width=0.85\linewidth]{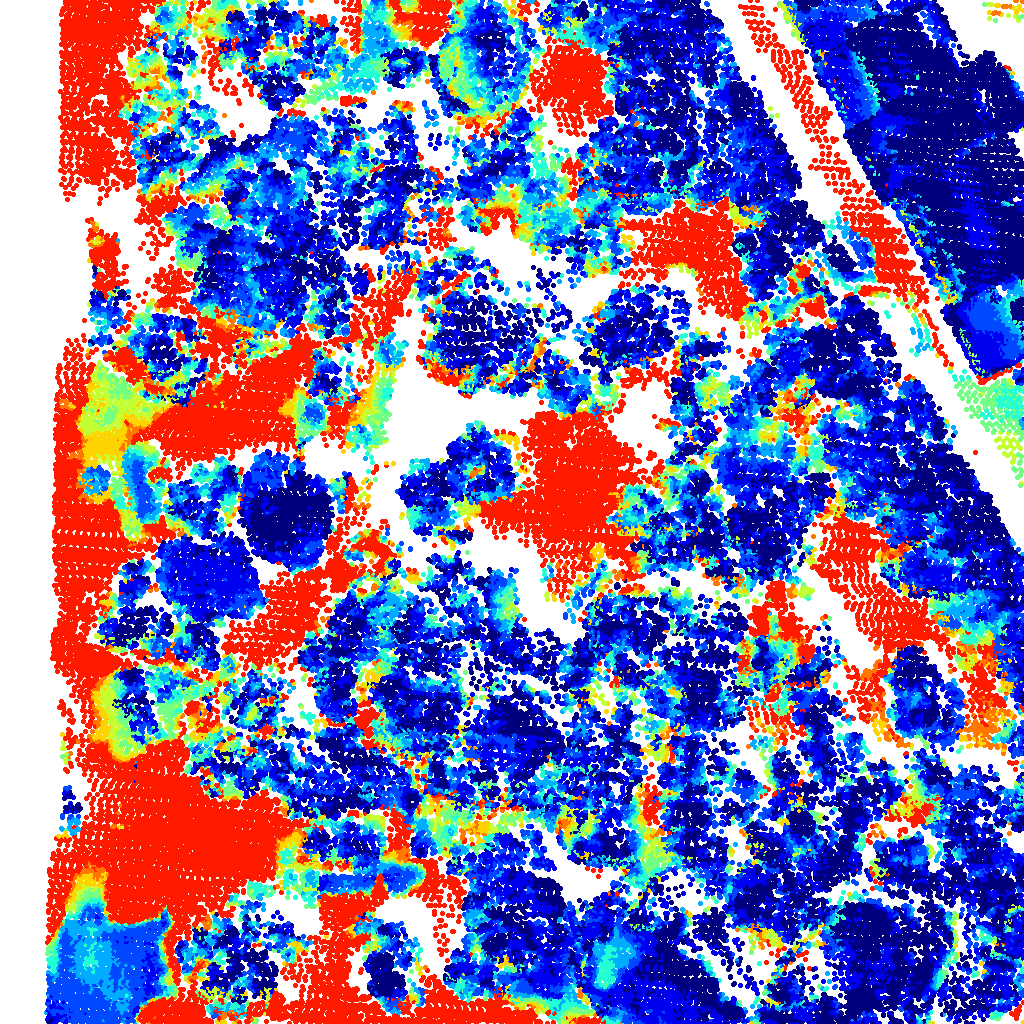}
		\includegraphics[width=\linewidth]{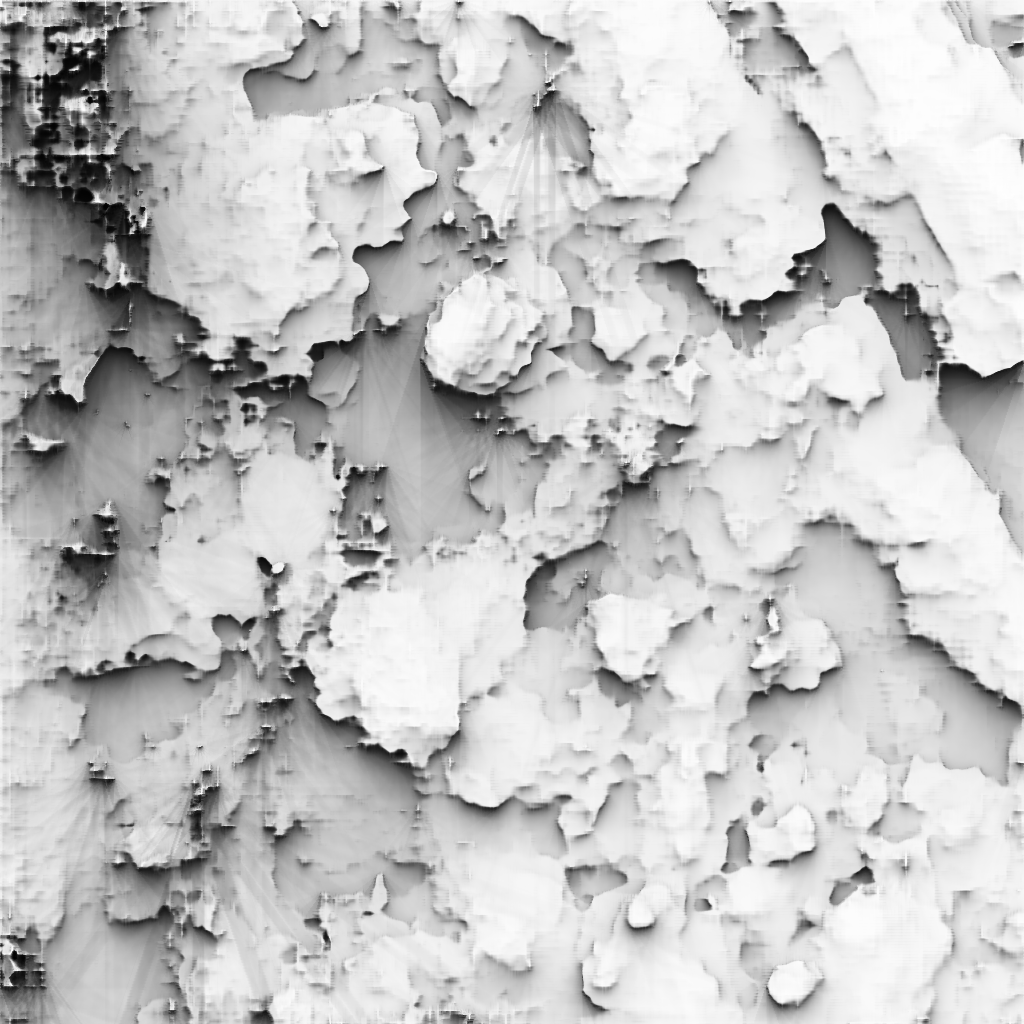}
		\centering{\tiny PSM net(KITTI)}
	\end{minipage}
	\begin{minipage}[t]{0.19\textwidth}		
		\includegraphics[width=0.098\linewidth]{figures_supp/color_map.png}
		\includegraphics[width=0.85\linewidth]{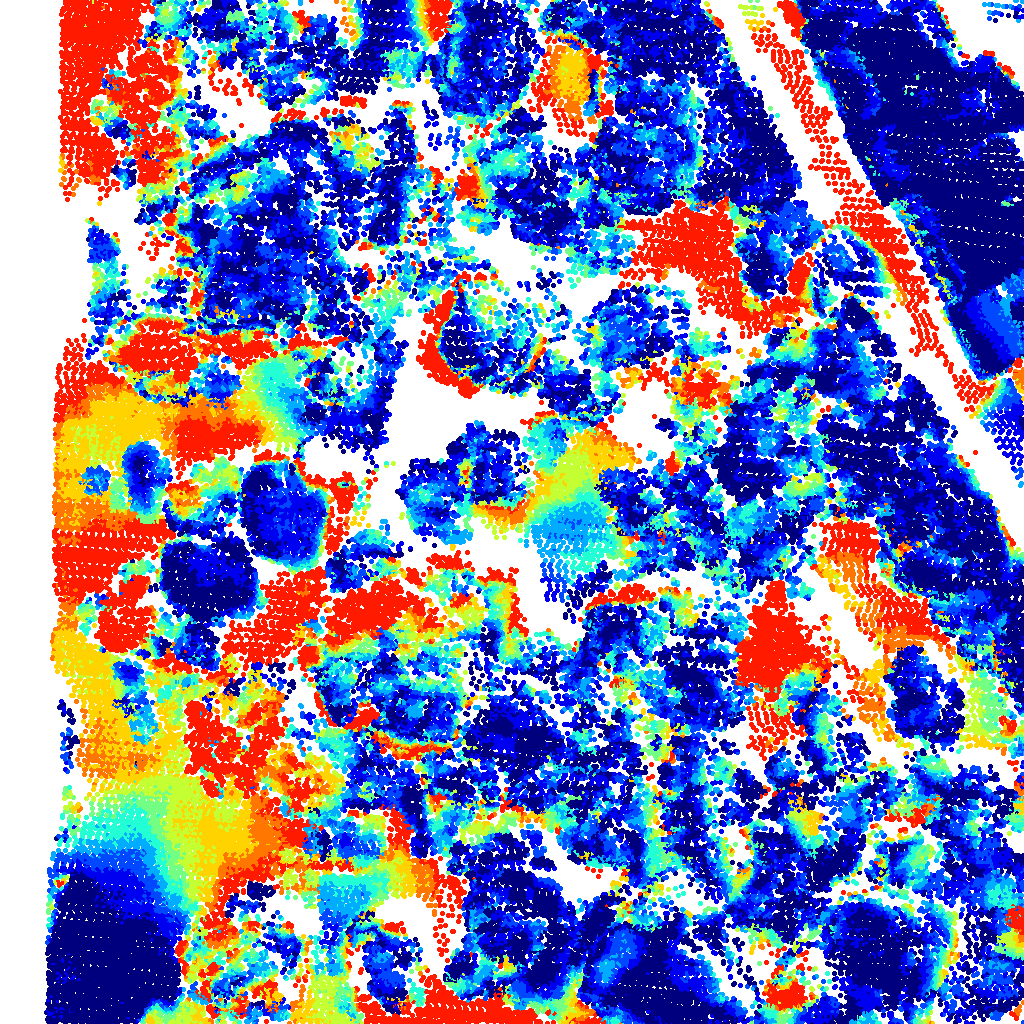}
		\includegraphics[width=\linewidth]{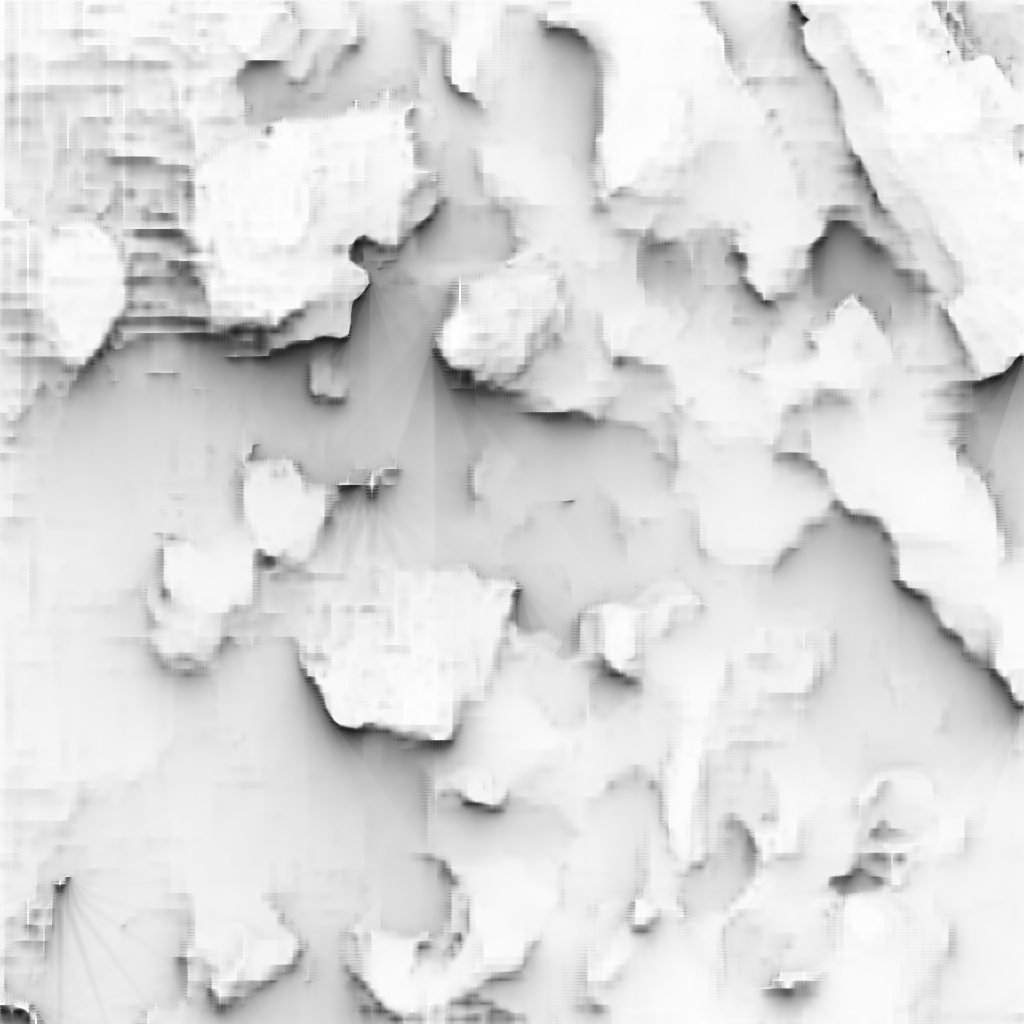}
		\centering{\tiny HRS net(KITTI)}
	\end{minipage}
	\begin{minipage}[t]{0.19\textwidth}	
		\includegraphics[width=0.098\linewidth]{figures_supp/color_map.png}
		\includegraphics[width=0.85\linewidth]{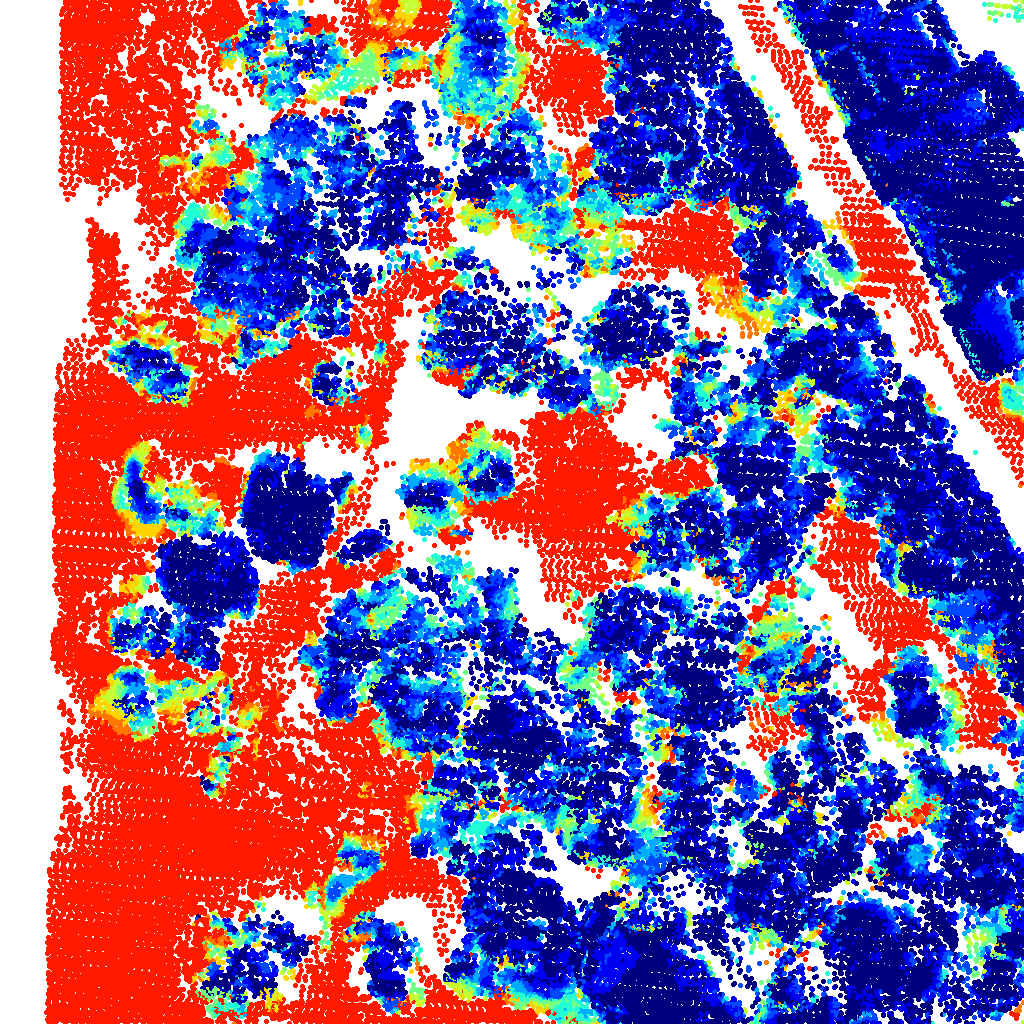}
		\includegraphics[width=\linewidth]{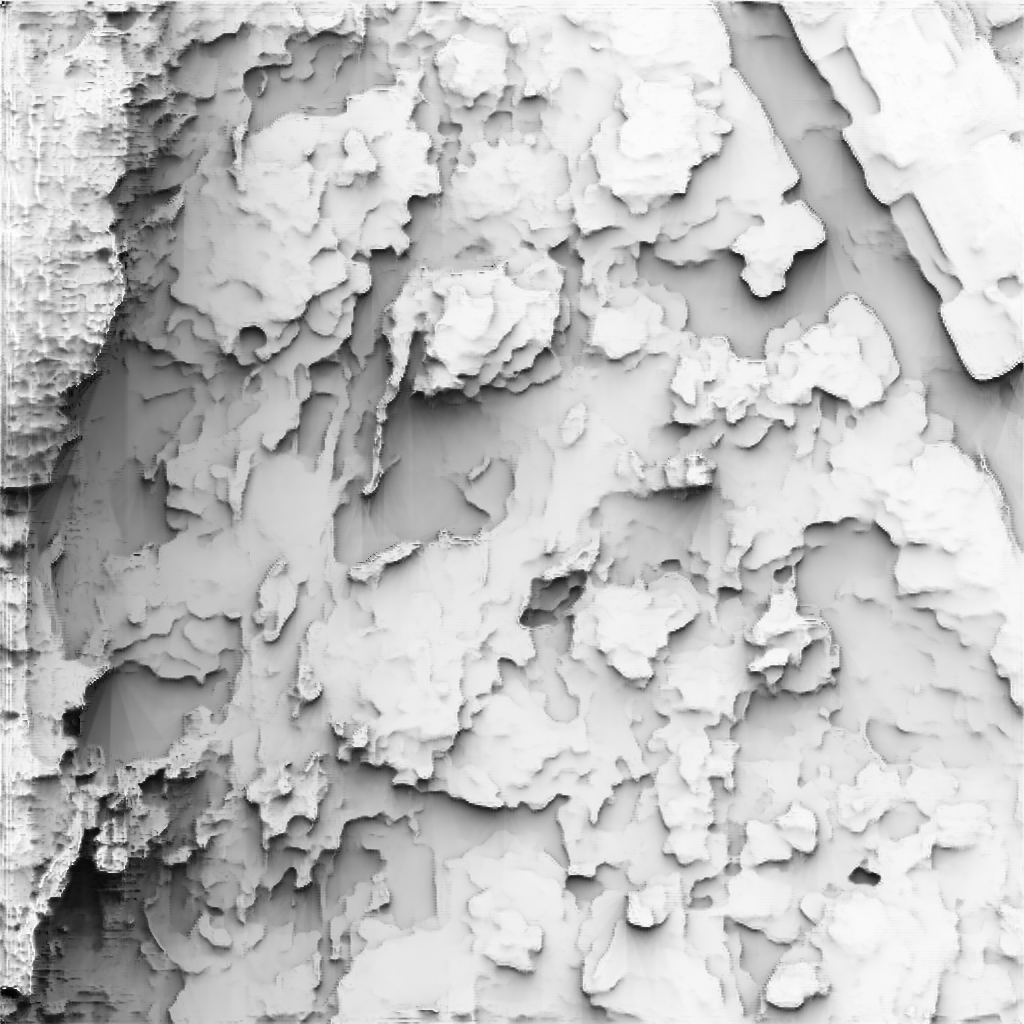}
		\centering{\tiny DeepPruner(KITTI)}
	\end{minipage}
	\begin{minipage}[t]{0.19\textwidth}		
		\includegraphics[width=0.098\linewidth]{figures_supp/color_map.png}
		\includegraphics[width=0.85\linewidth]{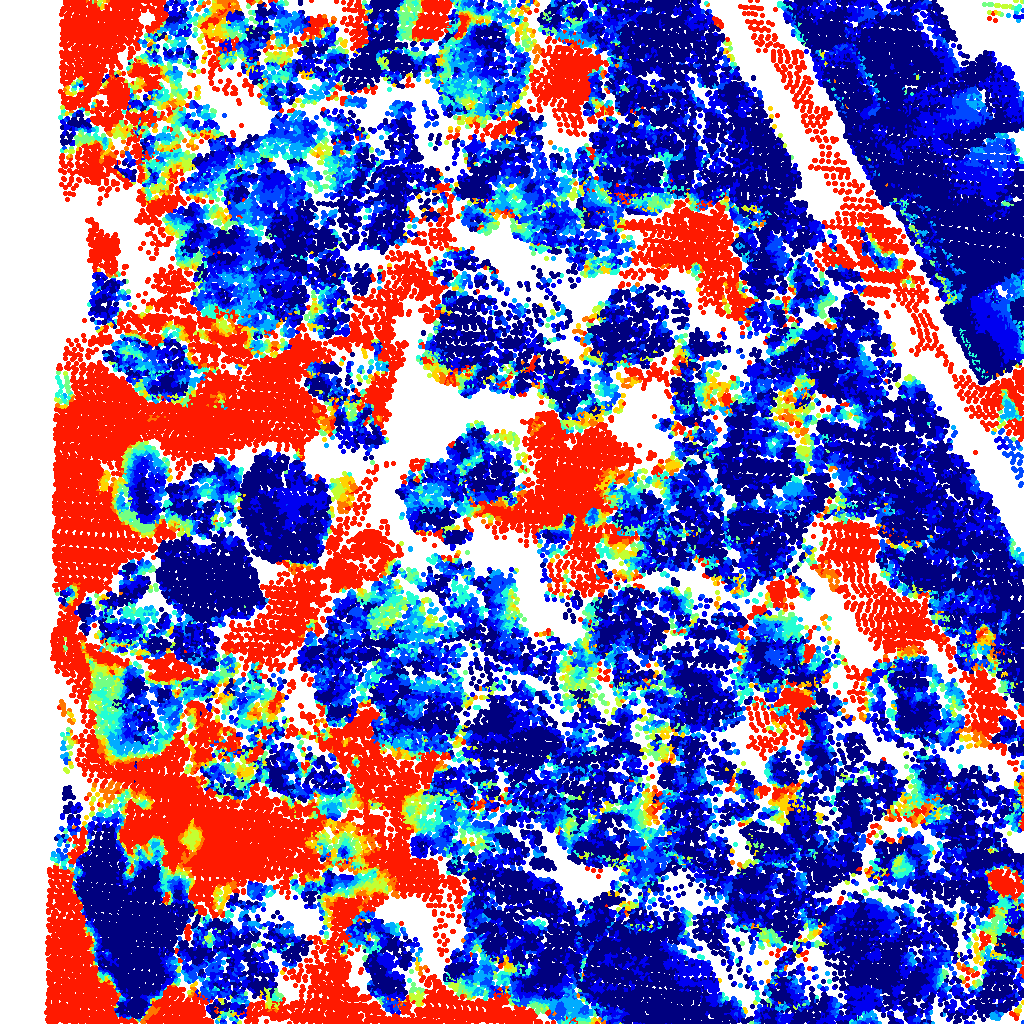}
		\includegraphics[width=\linewidth]{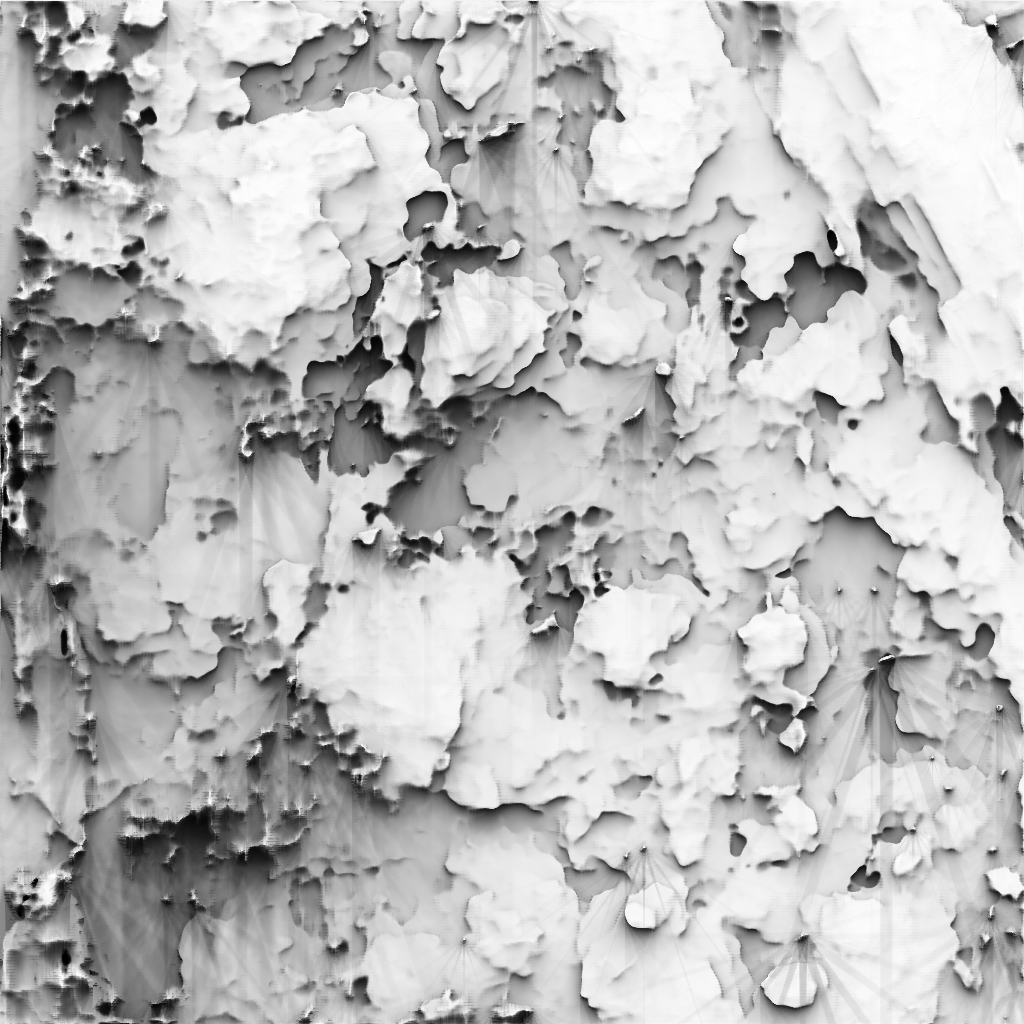}
		\centering{\tiny GAnet(KITTI)}
	\end{minipage}
	\begin{minipage}[t]{0.19\textwidth}	
		\includegraphics[width=0.098\linewidth]{figures_supp/color_map.png}
		\includegraphics[width=0.85\linewidth]{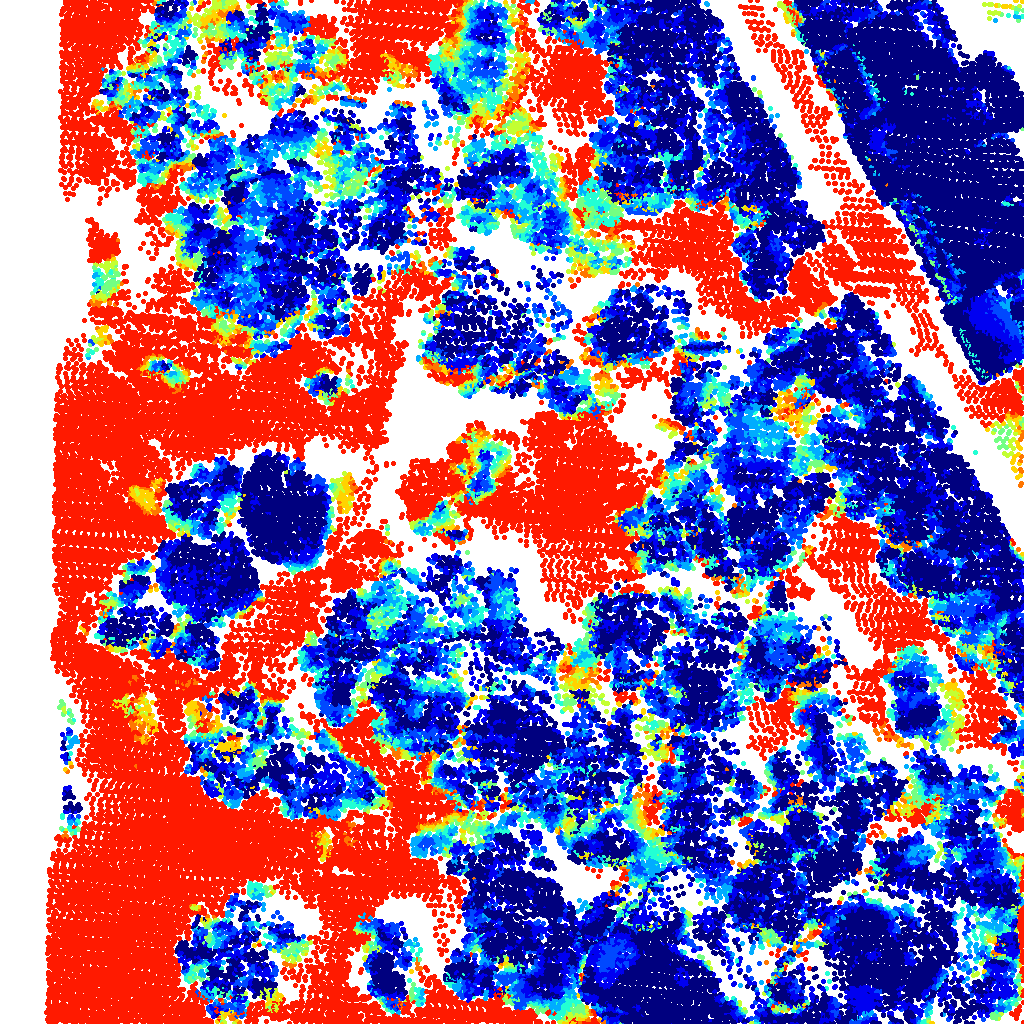}
		\includegraphics[width=\linewidth]{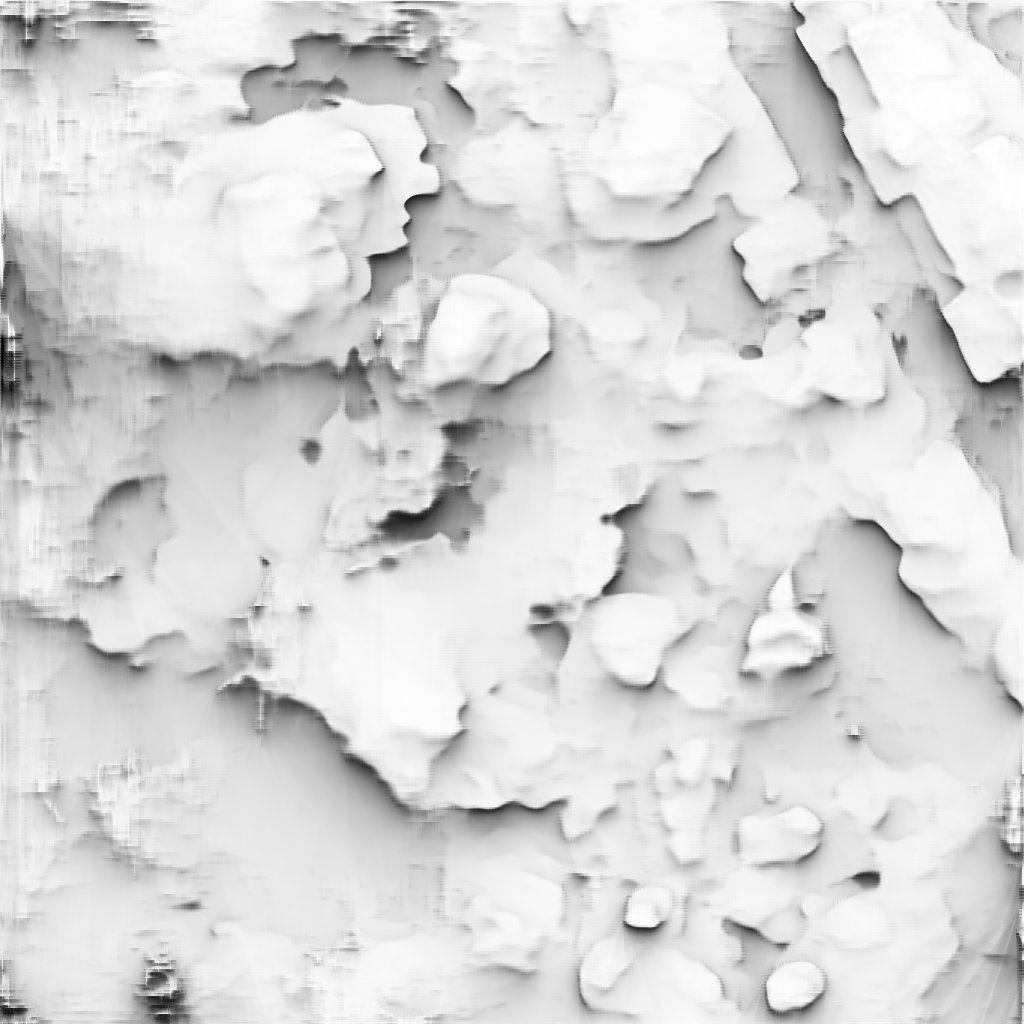}
		\centering{\tiny LEAStereo(KITTI)}
	\end{minipage}
	\begin{minipage}[t]{0.19\textwidth}	
		\includegraphics[width=0.098\linewidth]{figures_supp/color_map.png}
		\includegraphics[width=0.85\linewidth]{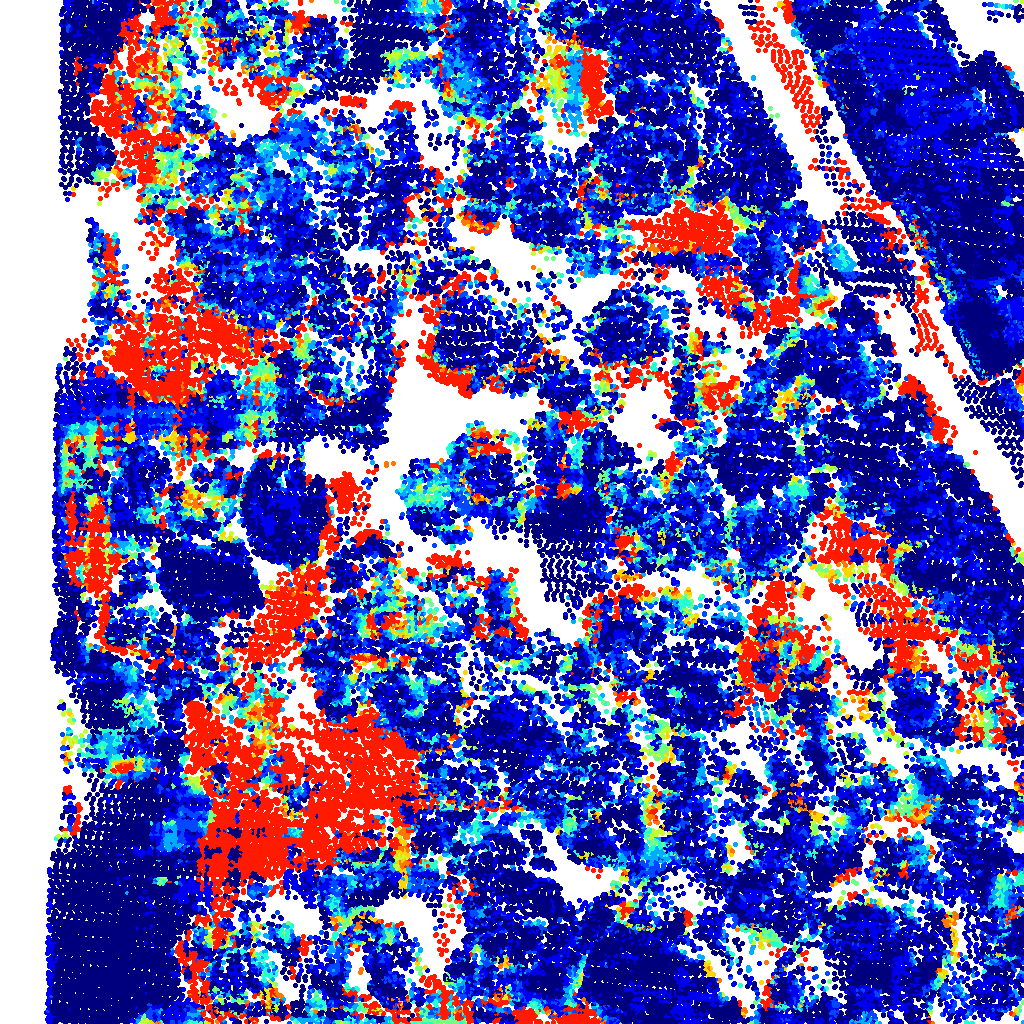}
		\includegraphics[width=\linewidth]{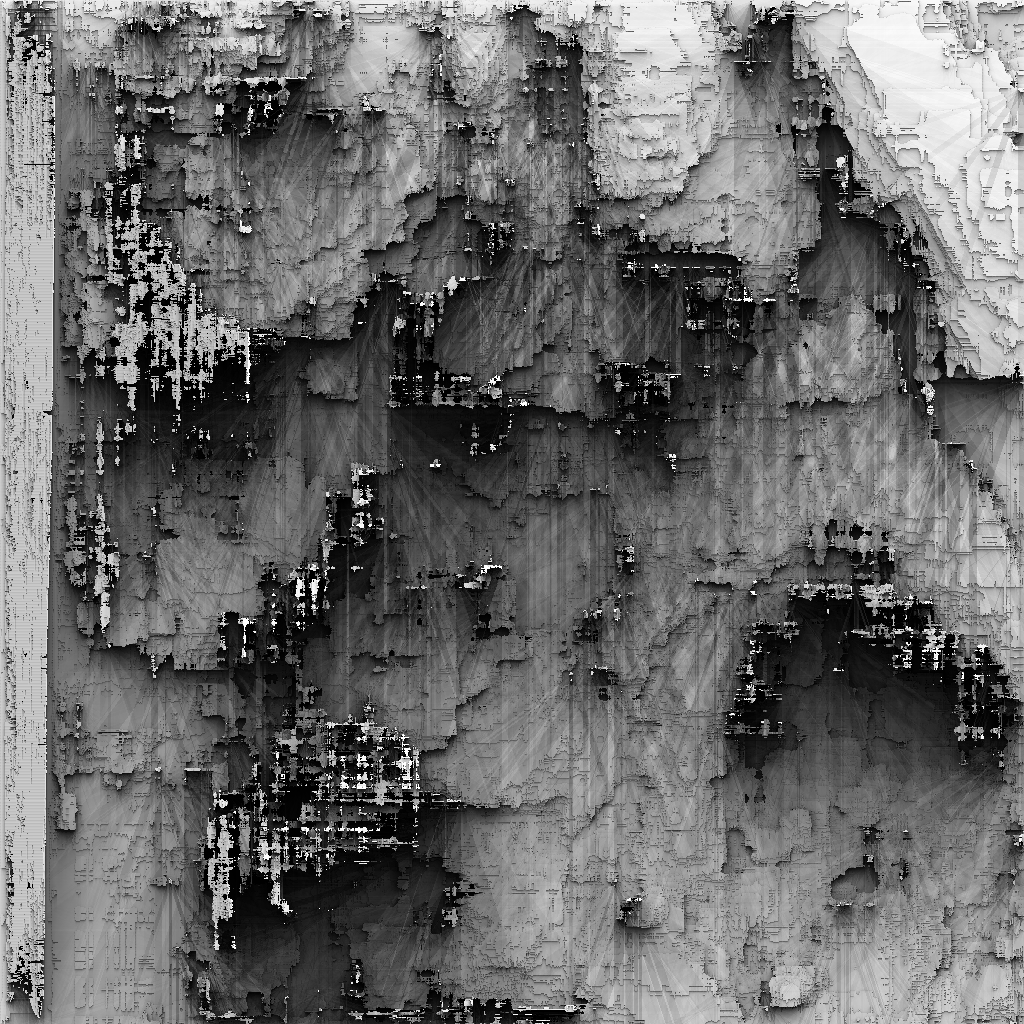}
		\centering{\tiny MC-CNN}
	\end{minipage}
	\begin{minipage}[t]{0.19\textwidth}	
		\includegraphics[width=0.098\linewidth]{figures_supp/color_map.png}
		\includegraphics[width=0.85\linewidth]{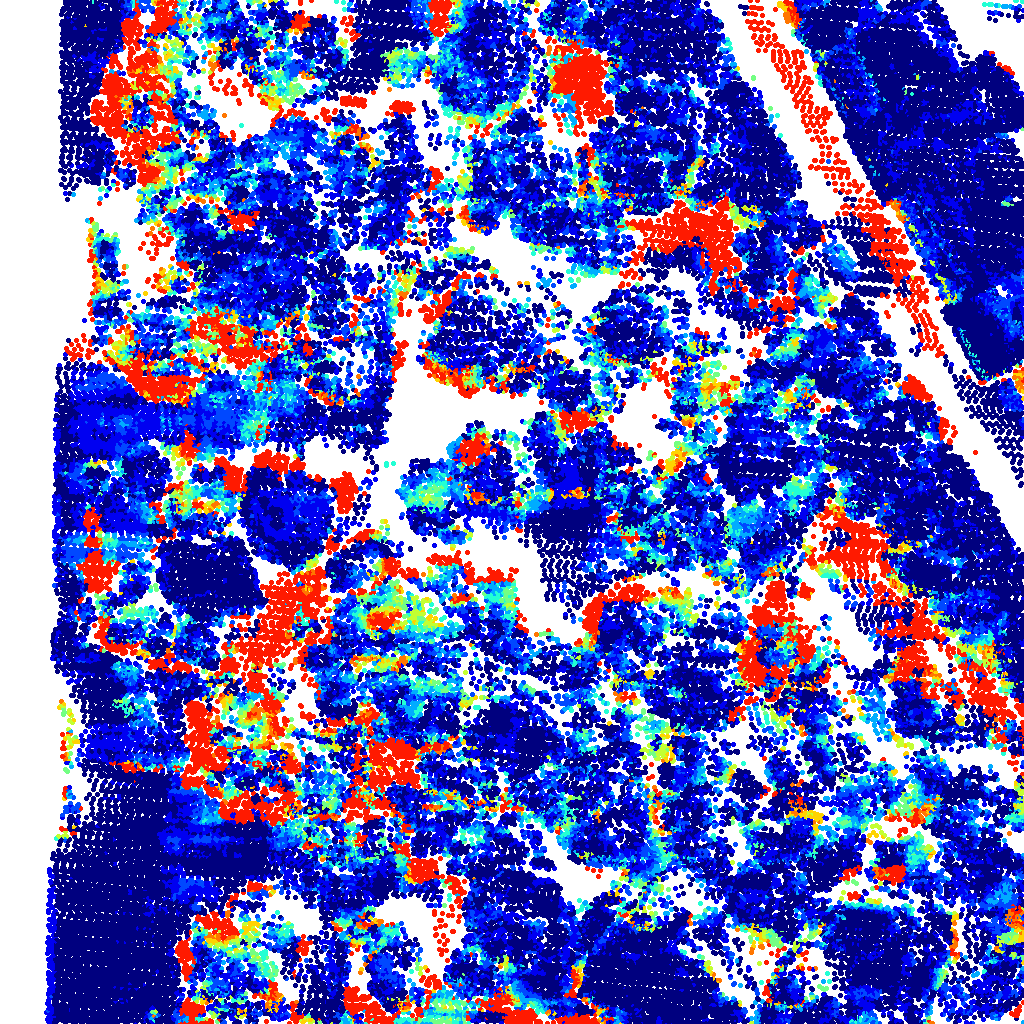}
		\includegraphics[width=\linewidth]{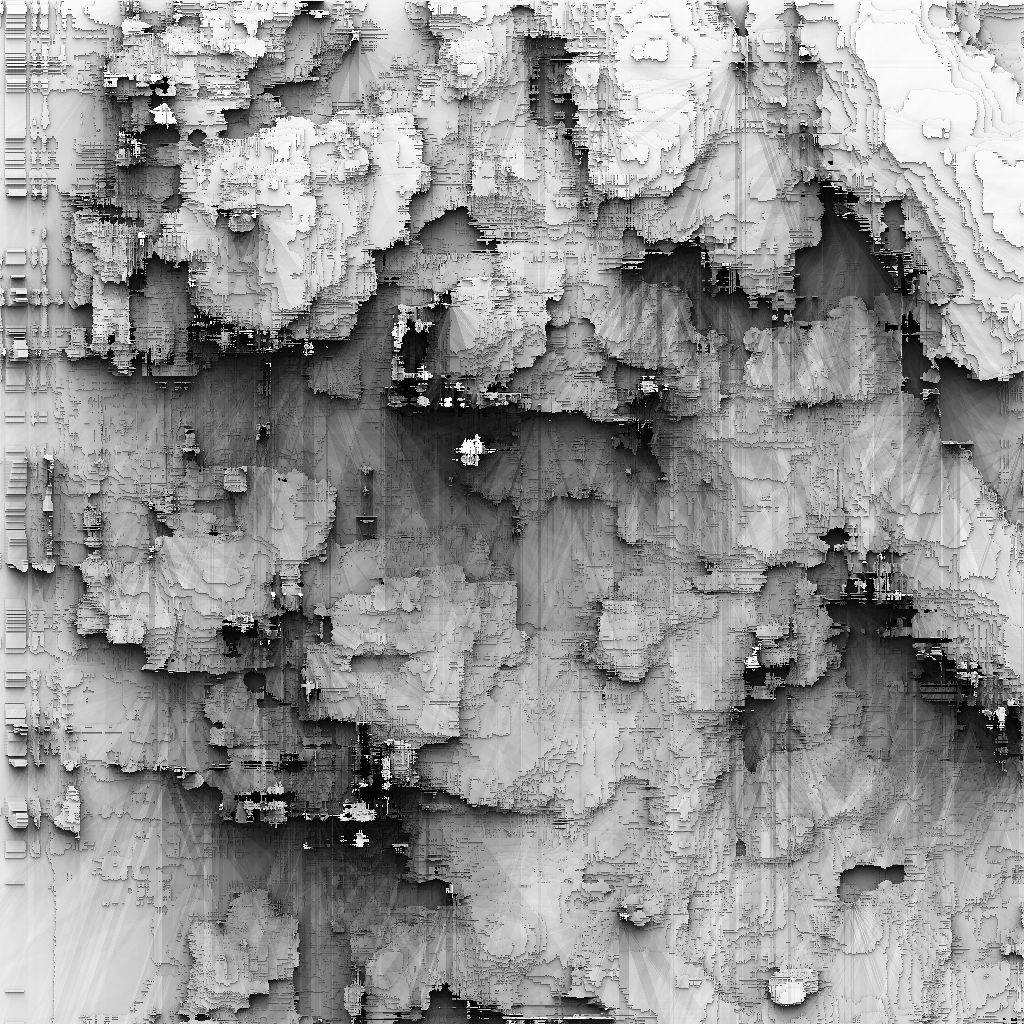}
		\centering{\tiny DeepFeature}
	\end{minipage}
	\begin{minipage}[t]{0.19\textwidth}	
		\includegraphics[width=0.098\linewidth]{figures_supp/color_map.png}
		\includegraphics[width=0.85\linewidth]{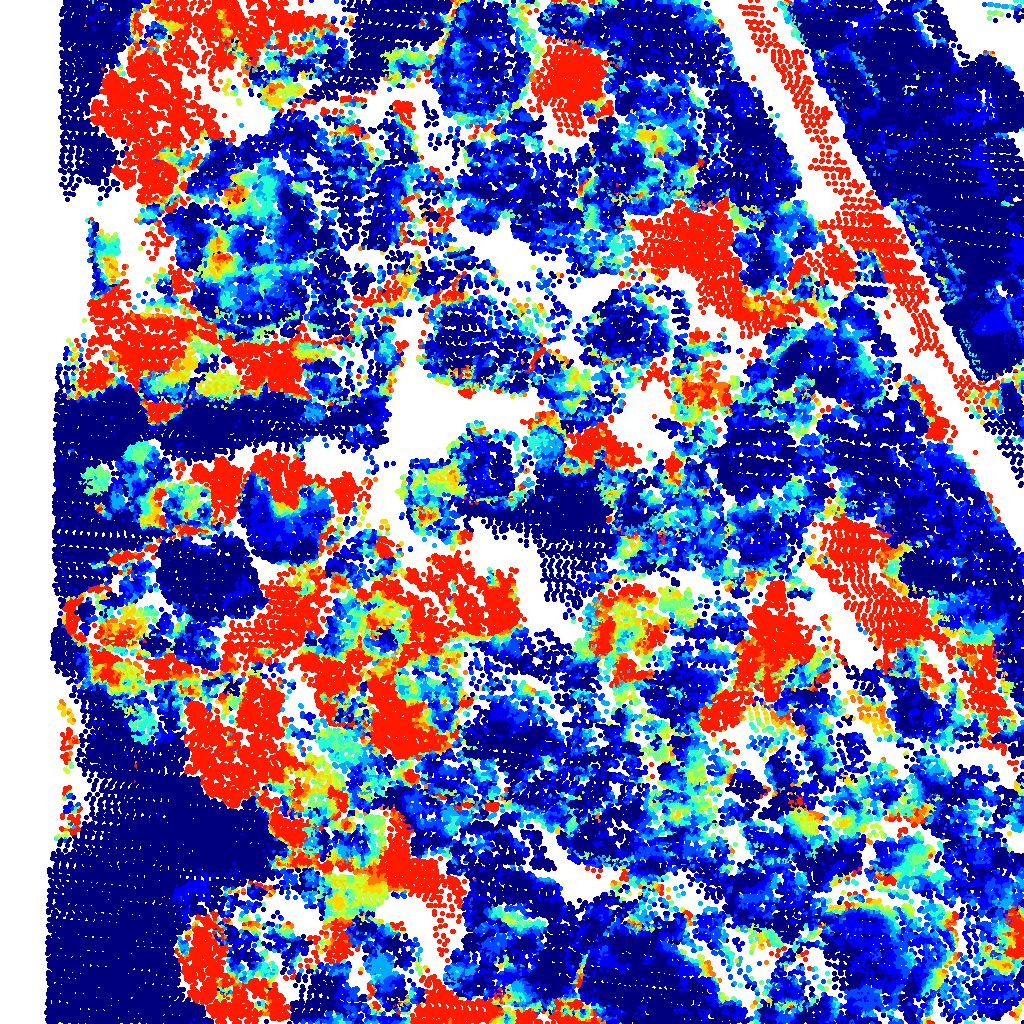}
		\includegraphics[width=\linewidth]{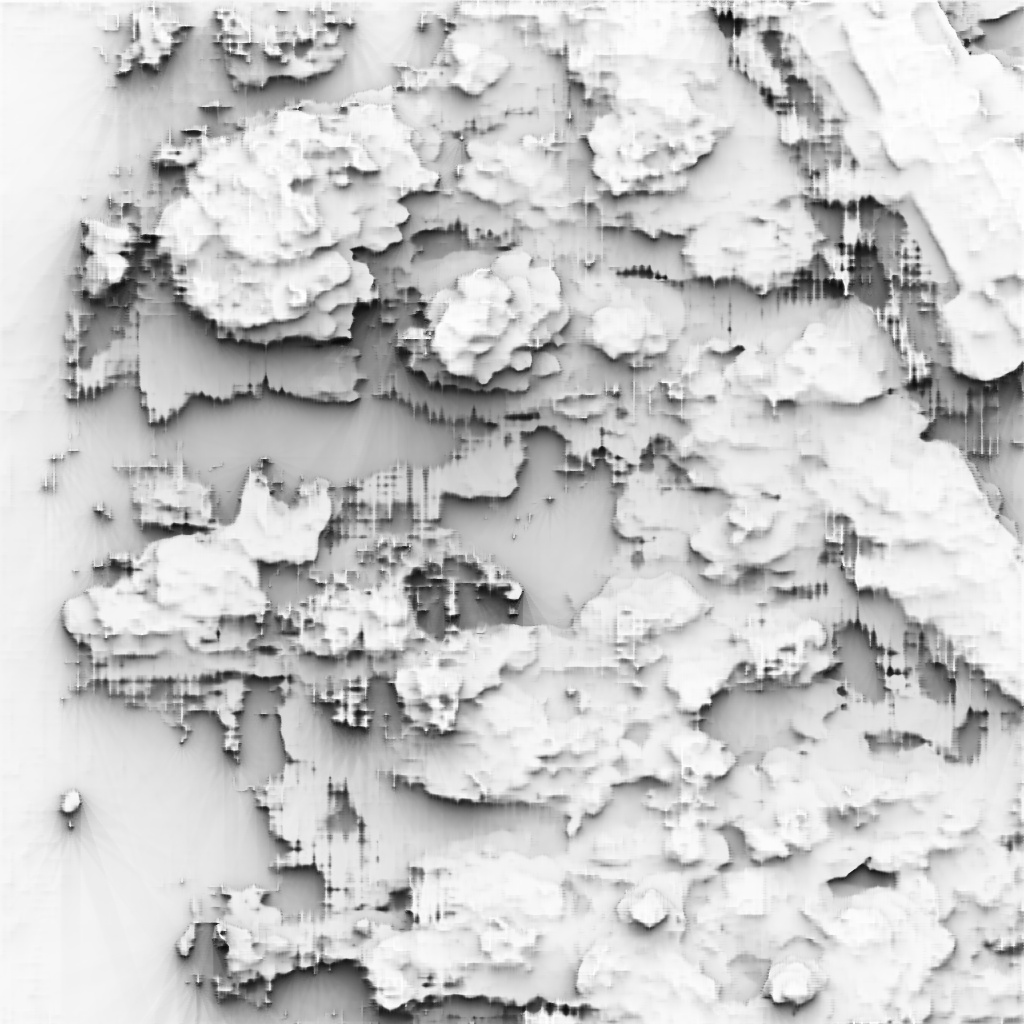}
		\centering{\tiny PSM net}
	\end{minipage}
	\begin{minipage}[t]{0.19\textwidth}		
		\includegraphics[width=0.098\linewidth]{figures_supp/color_map.png}
		\includegraphics[width=0.85\linewidth]{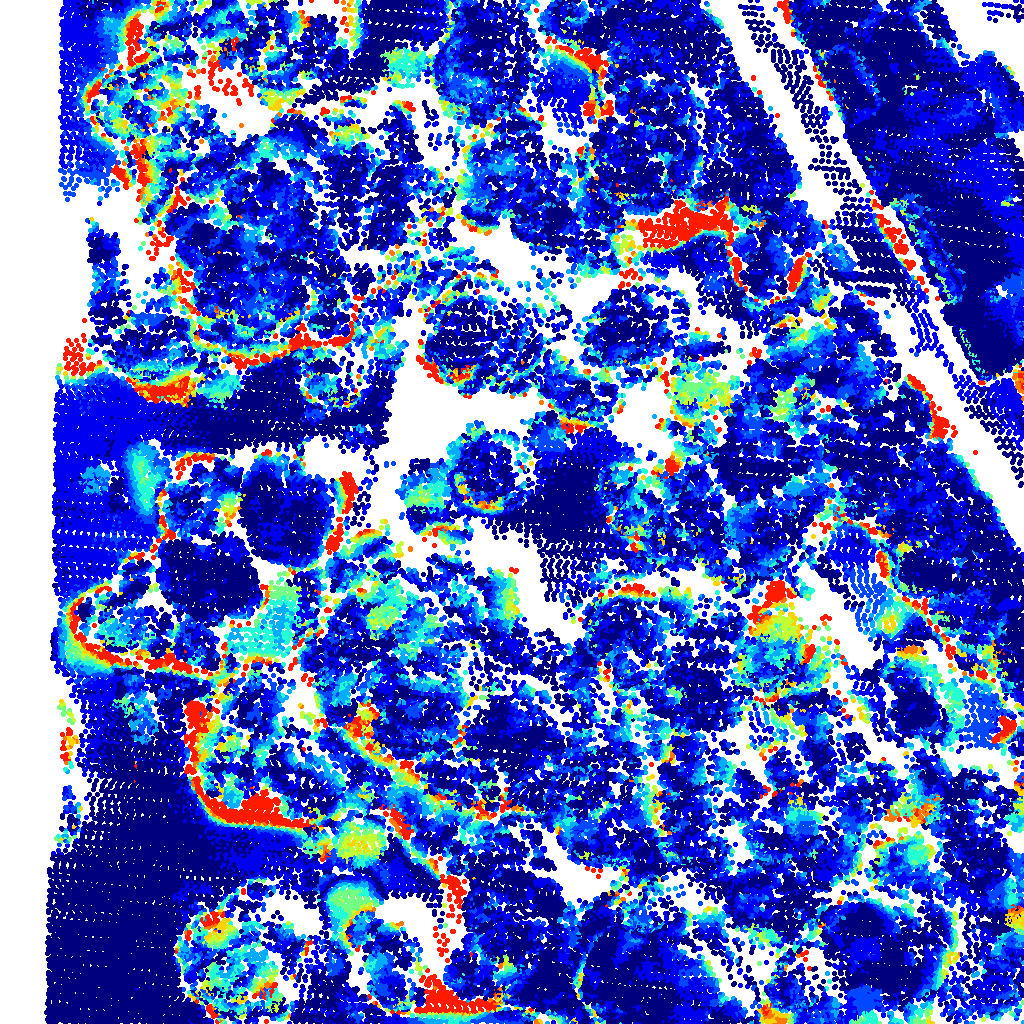}
		\includegraphics[width=\linewidth]{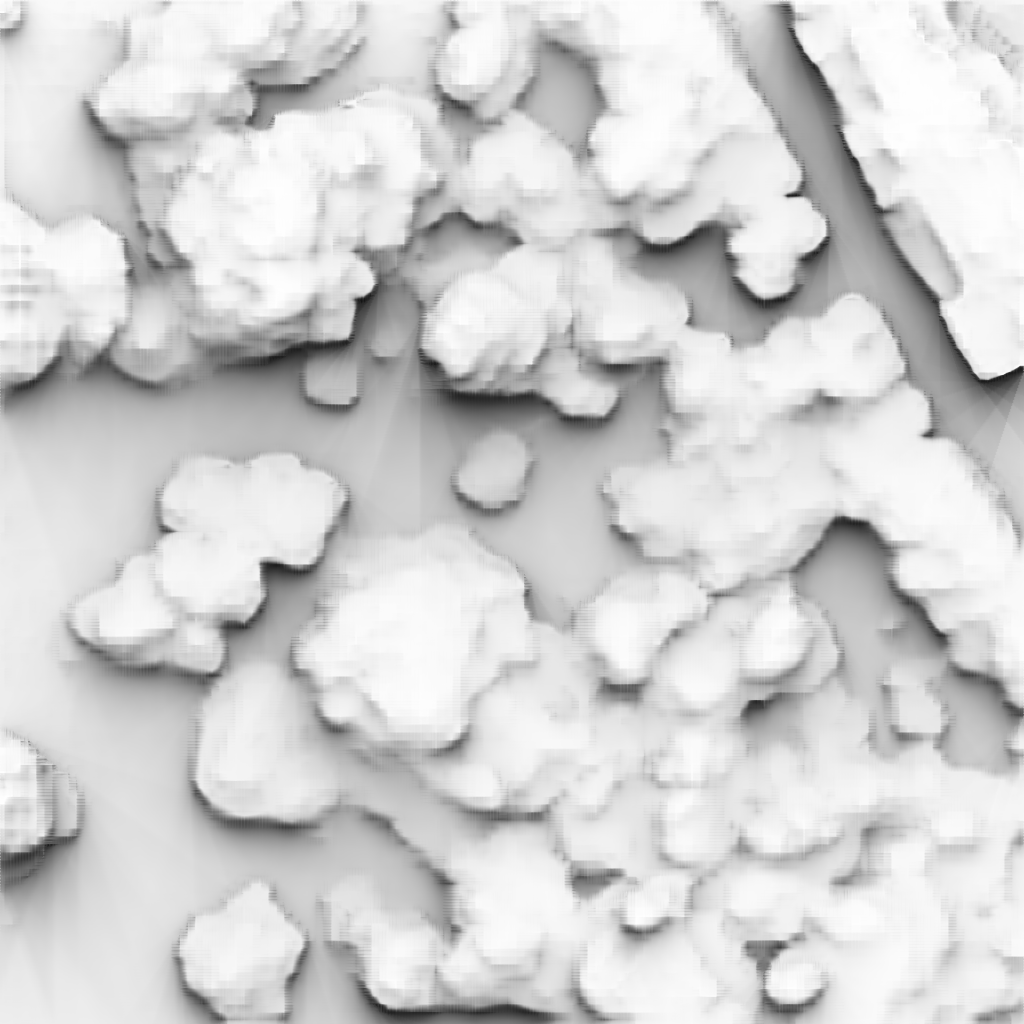}
		\centering{\tiny HRS net}
	\end{minipage}
	\begin{minipage}[t]{0.19\textwidth}	
		\includegraphics[width=0.098\linewidth]{figures_supp/color_map.png}
		\includegraphics[width=0.85\linewidth]{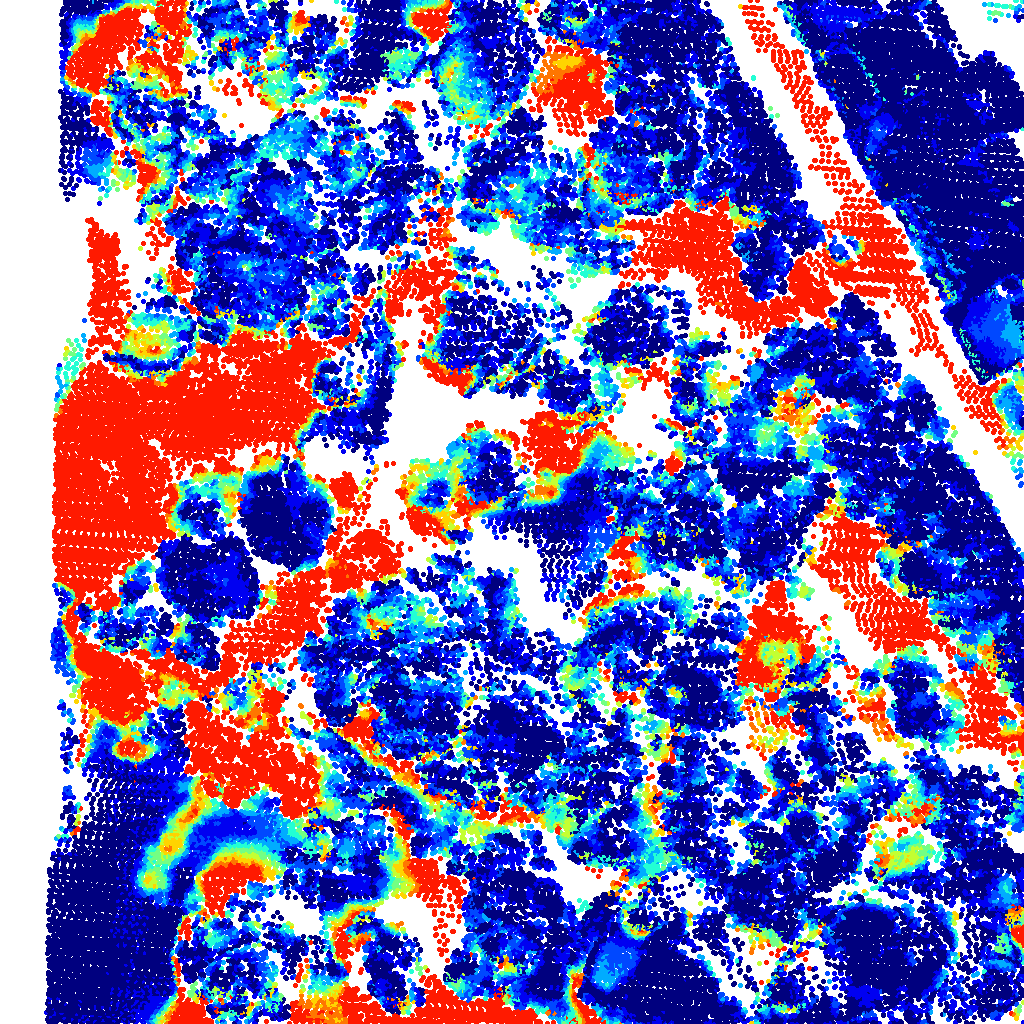}
		\includegraphics[width=\linewidth]{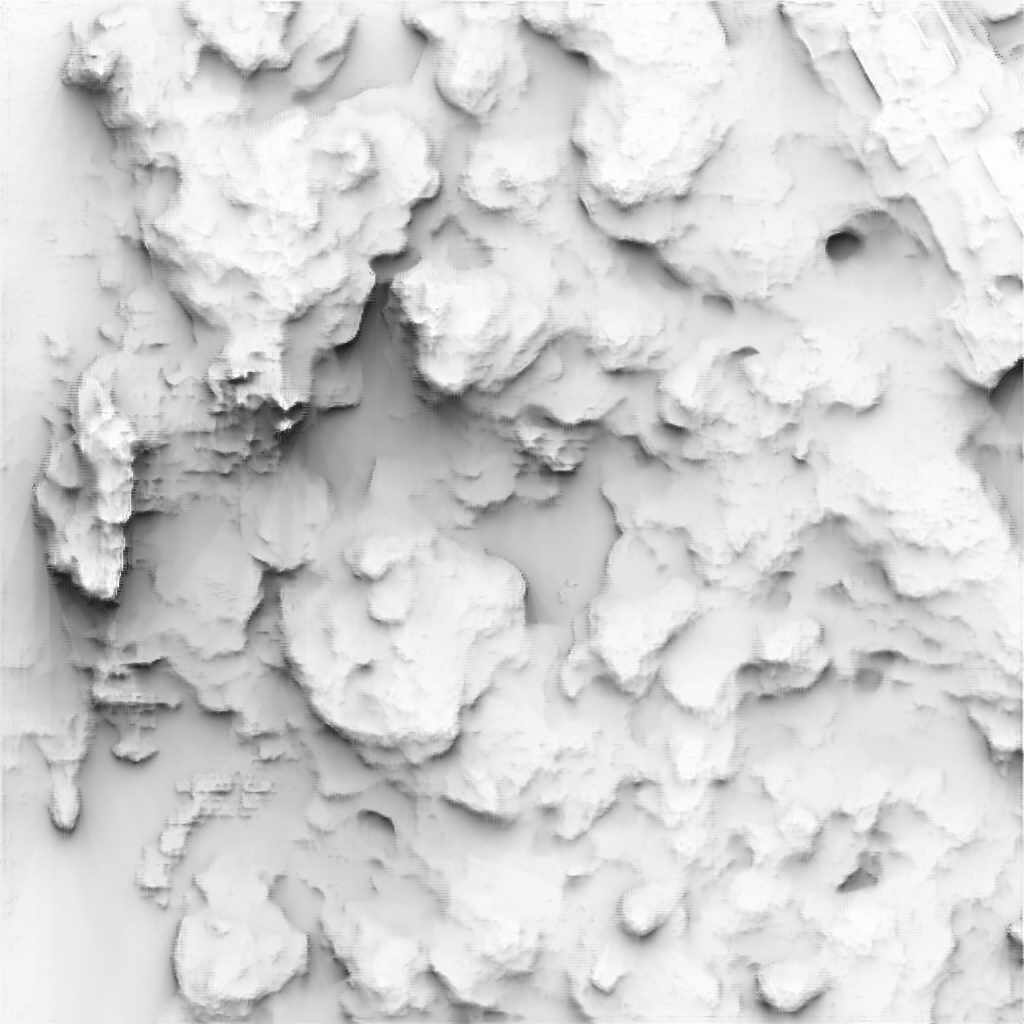}
		\centering{\tiny DeepPruner}
	\end{minipage}
	\begin{minipage}[t]{0.19\textwidth}		
		\includegraphics[width=0.098\linewidth]{figures_supp/color_map.png}
		\includegraphics[width=0.85\linewidth]{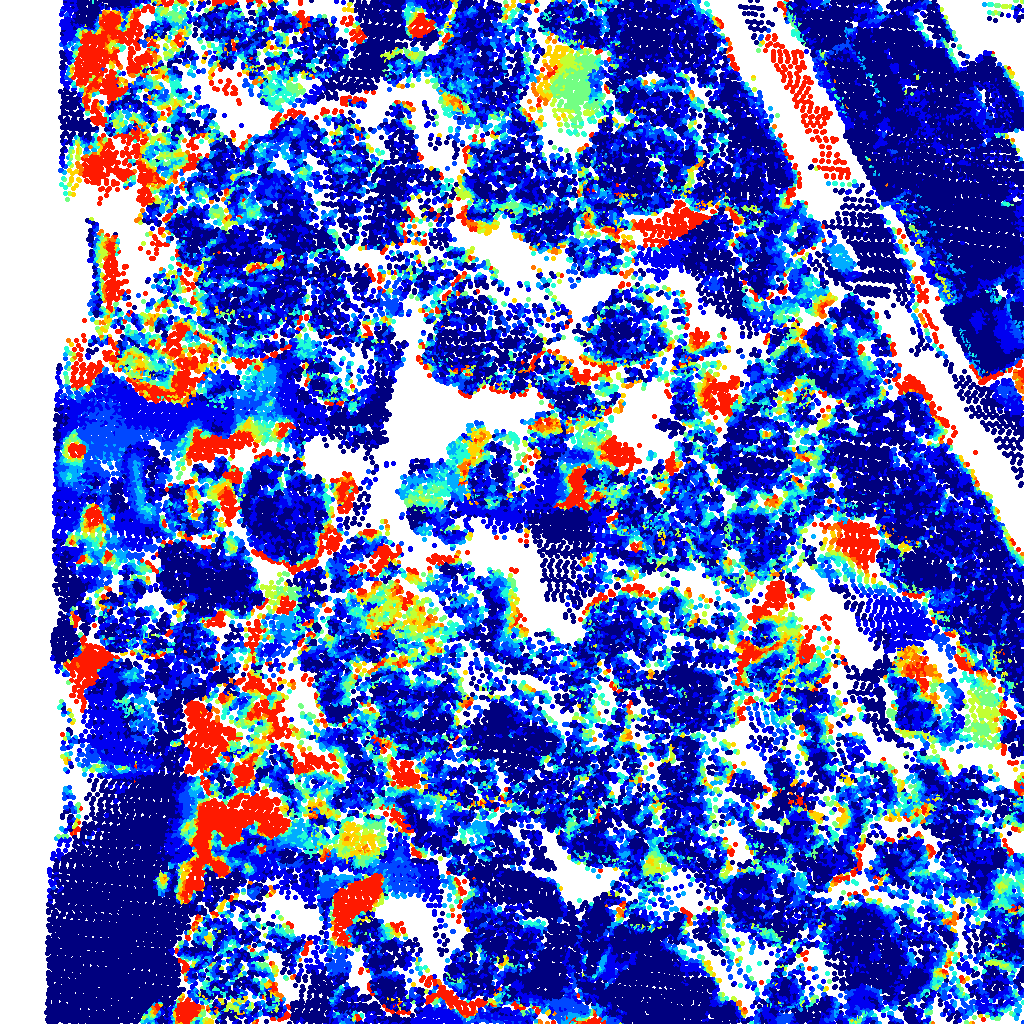}
		\includegraphics[width=\linewidth]{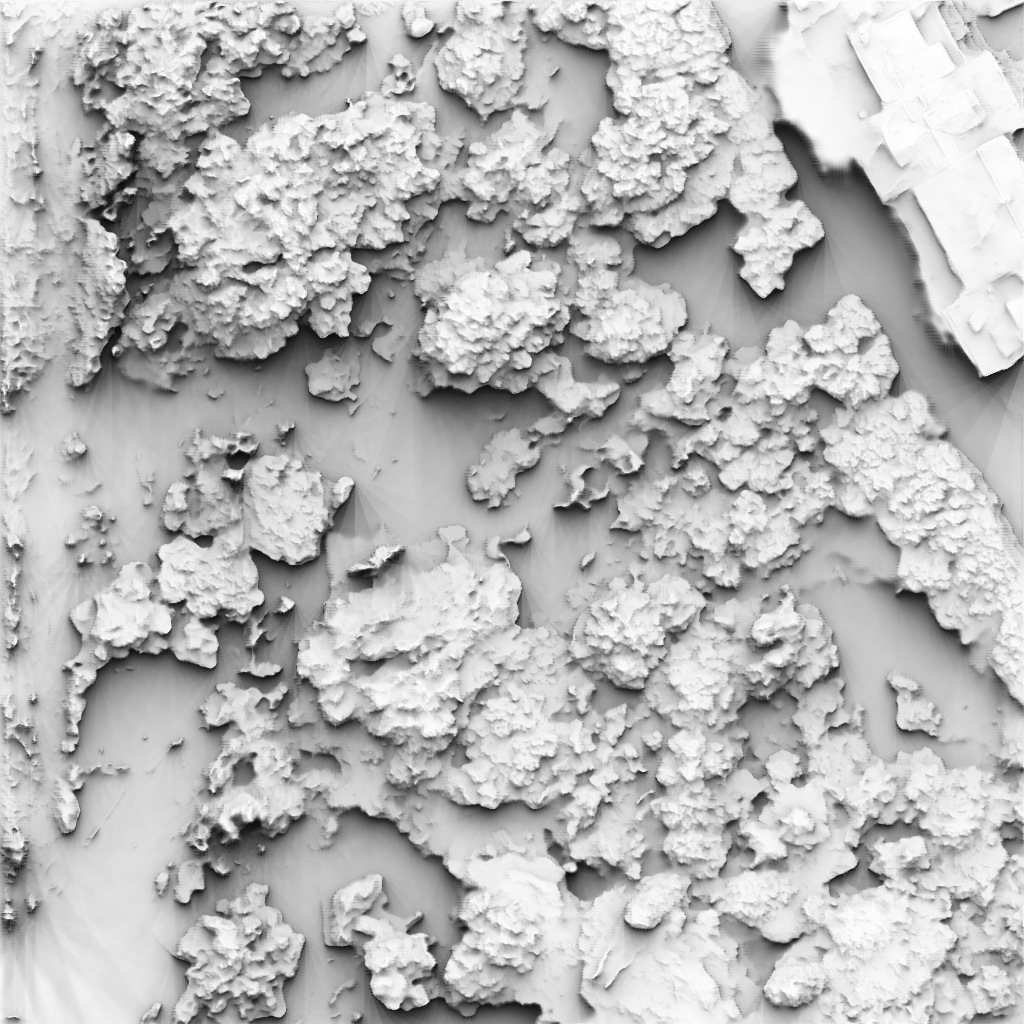}
		\centering{\tiny GAnet}
	\end{minipage}
	\begin{minipage}[t]{0.19\textwidth}	
		\includegraphics[width=0.098\linewidth]{figures_supp/color_map.png}
		\includegraphics[width=0.85\linewidth]{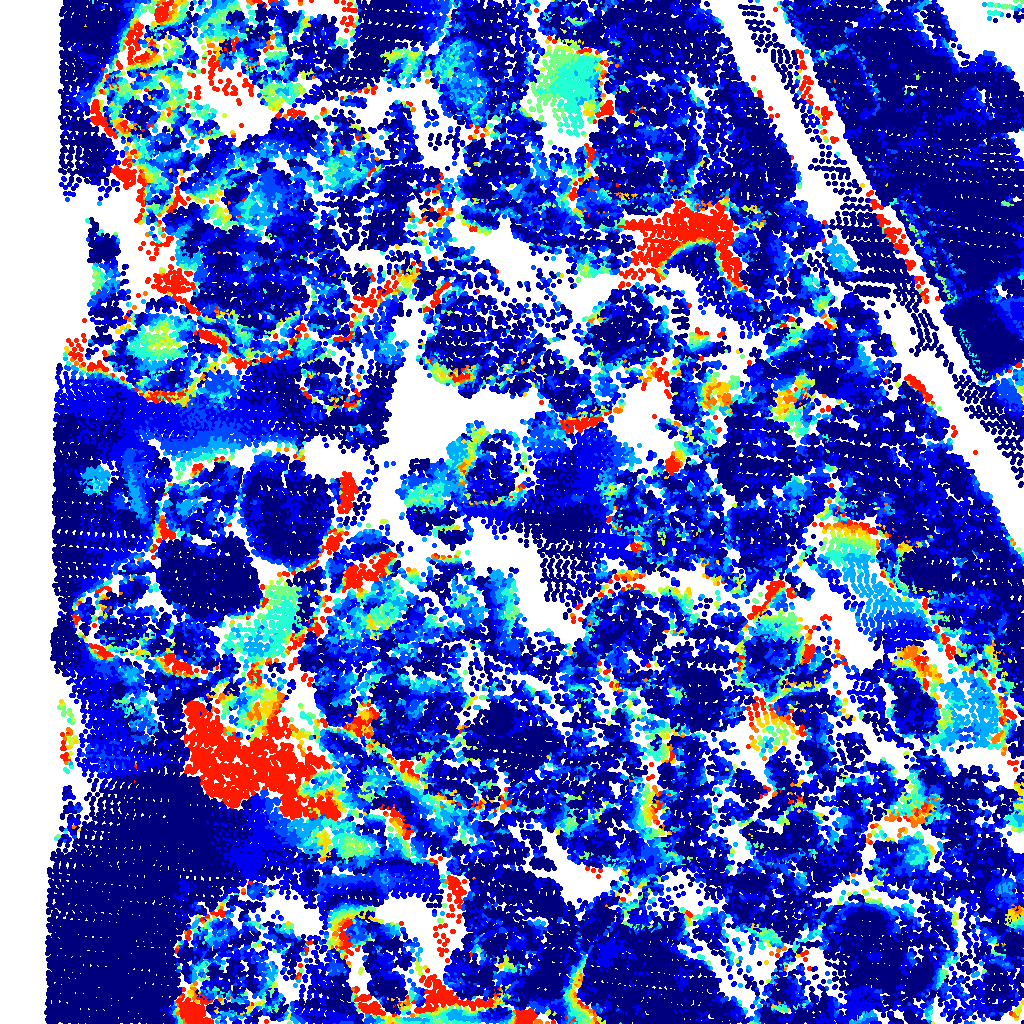}
		\includegraphics[width=\linewidth]{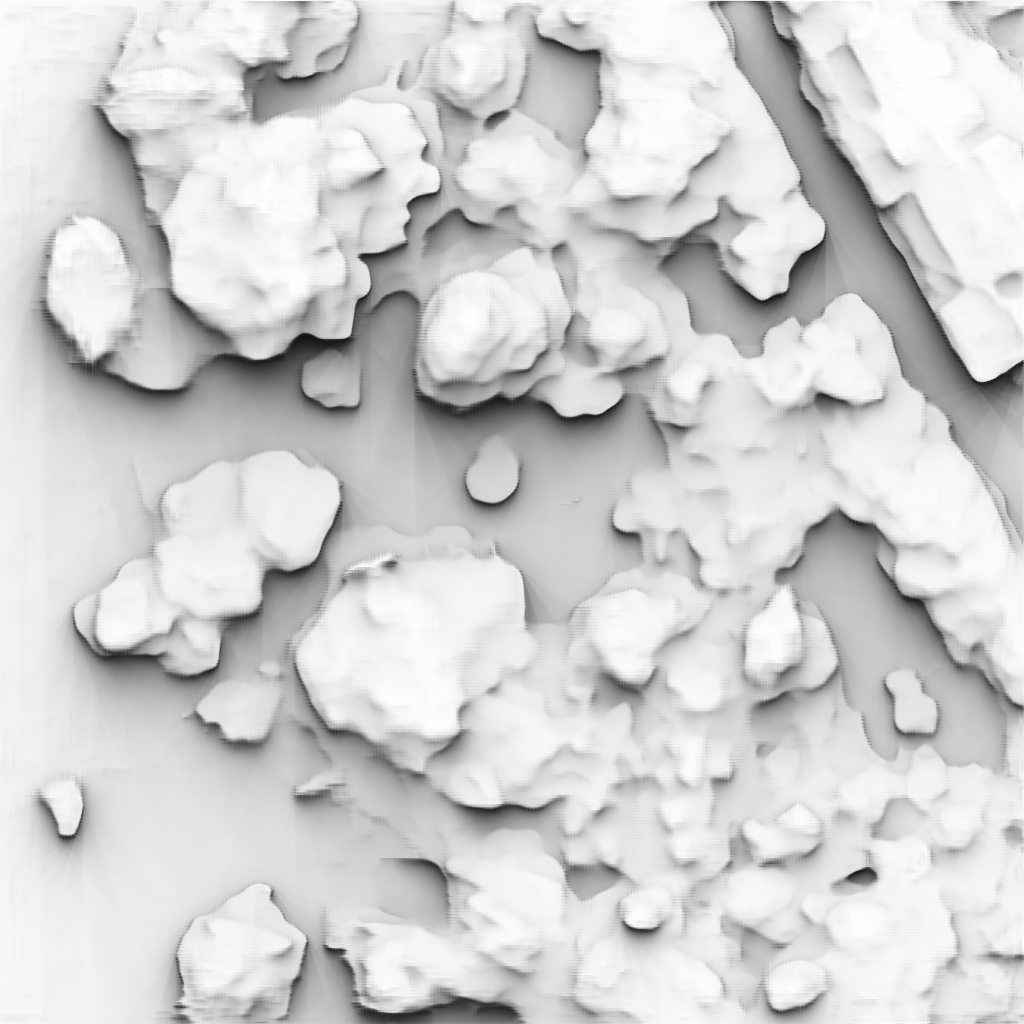}
		\centering{\tiny LEAStereo}
	\end{minipage}
	\caption{Error map and disparity visualization on tree area for Toulouse UMBRA.}
	\label{Figure.umbratree}
\end{figure}

Another example from the tree area is shown in \Cref{Figure.umbratree}, 
all the trees are flourishing, so the large disparity discontinuity is not frequent in tree areas, but there are big error pixels for all the methods, SGM based methods have big errors on the disparity discontinuity areas, even with DL-based features. For the end-to-end methods, pre-trained models work badly, even after fine-tuning, \textit{PSM net} and \textit{DeepPruner} also have a lot of errors.
After fine-tuning, \textit{GAnet} obtains the best result, and the other end-to-end methods produce a smooth result.

\begin{figure}[tp]
	\begin{minipage}[t]{0.19\textwidth}
		\includegraphics[width=0.098\linewidth]{figures_supp/color_map.png}
		\includegraphics[width=0.85\linewidth]{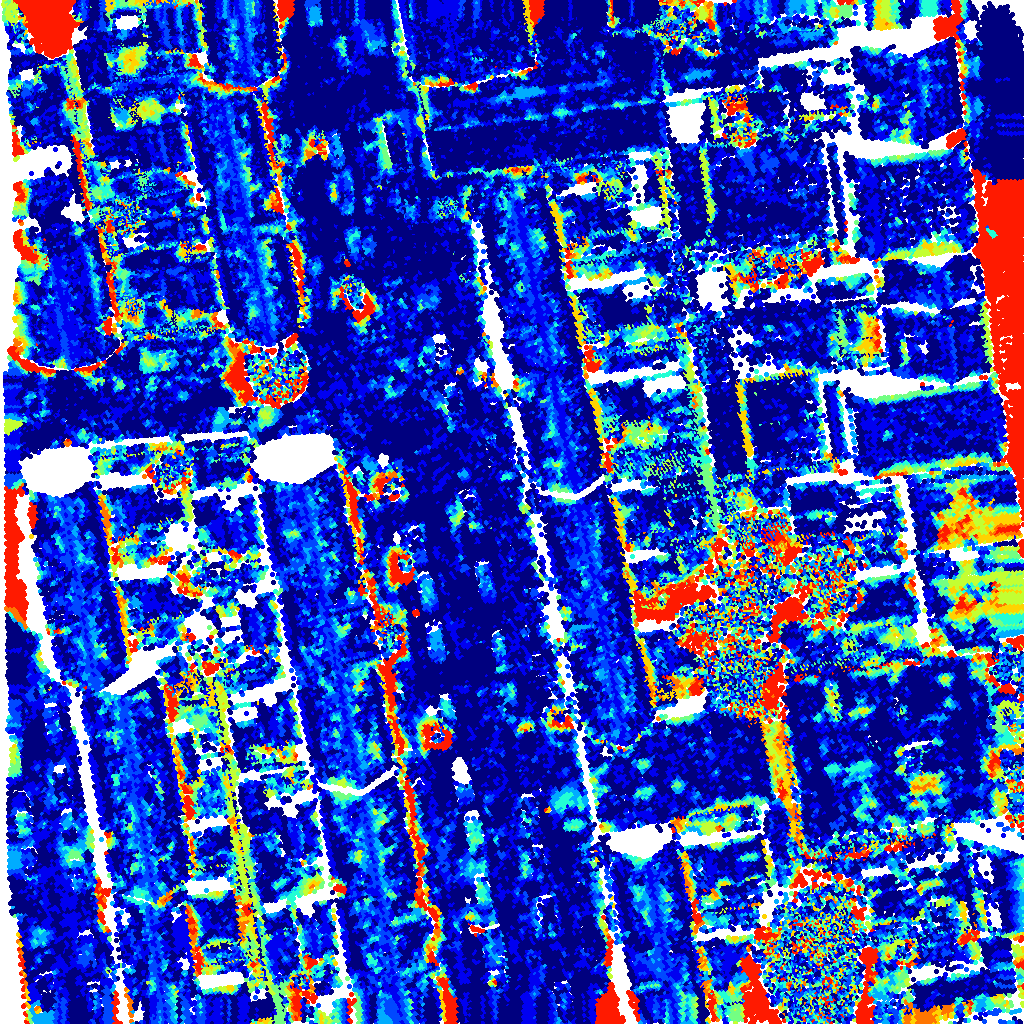}
		\includegraphics[width=\linewidth]{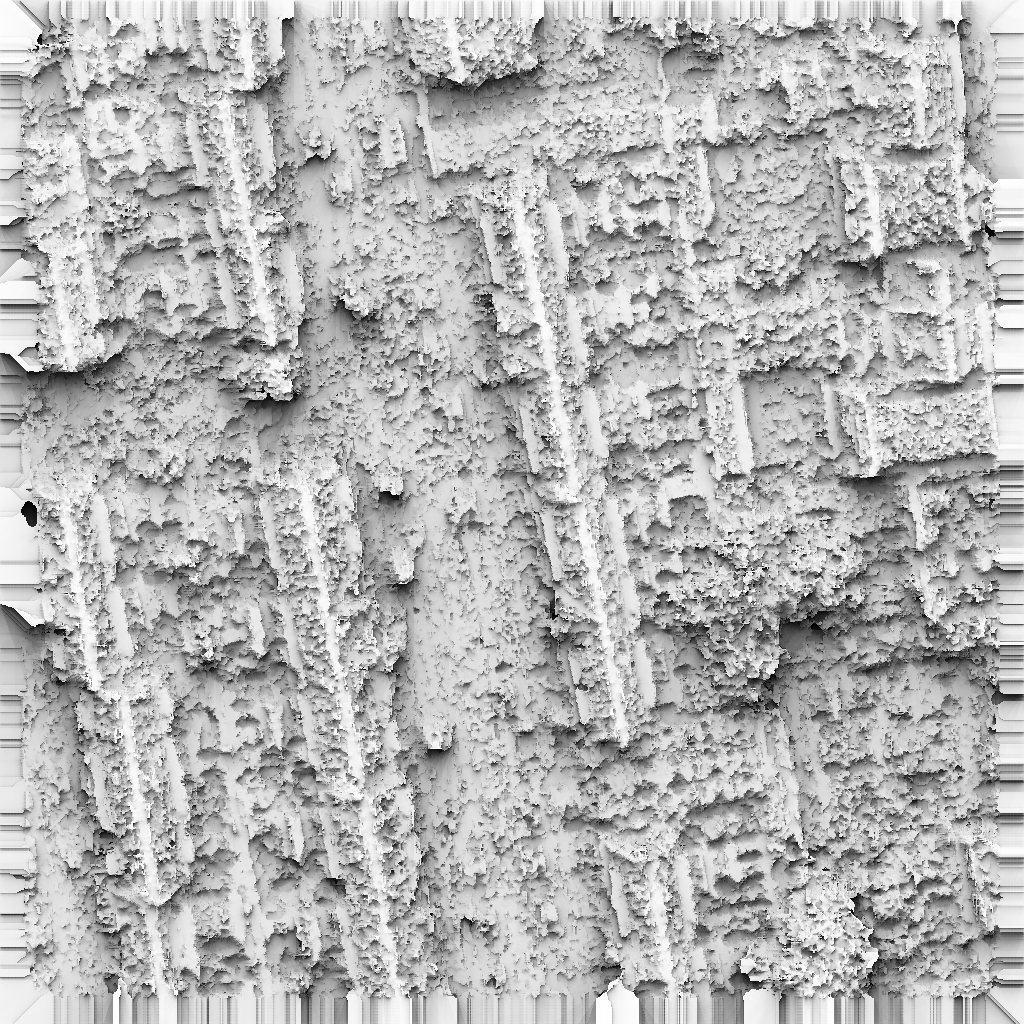}
		\centering{\tiny MICMAC}
	\end{minipage}
	\begin{minipage}[t]{0.19\textwidth}	
		\includegraphics[width=0.098\linewidth]{figures_supp/color_map.png}
		\includegraphics[width=0.85\linewidth]{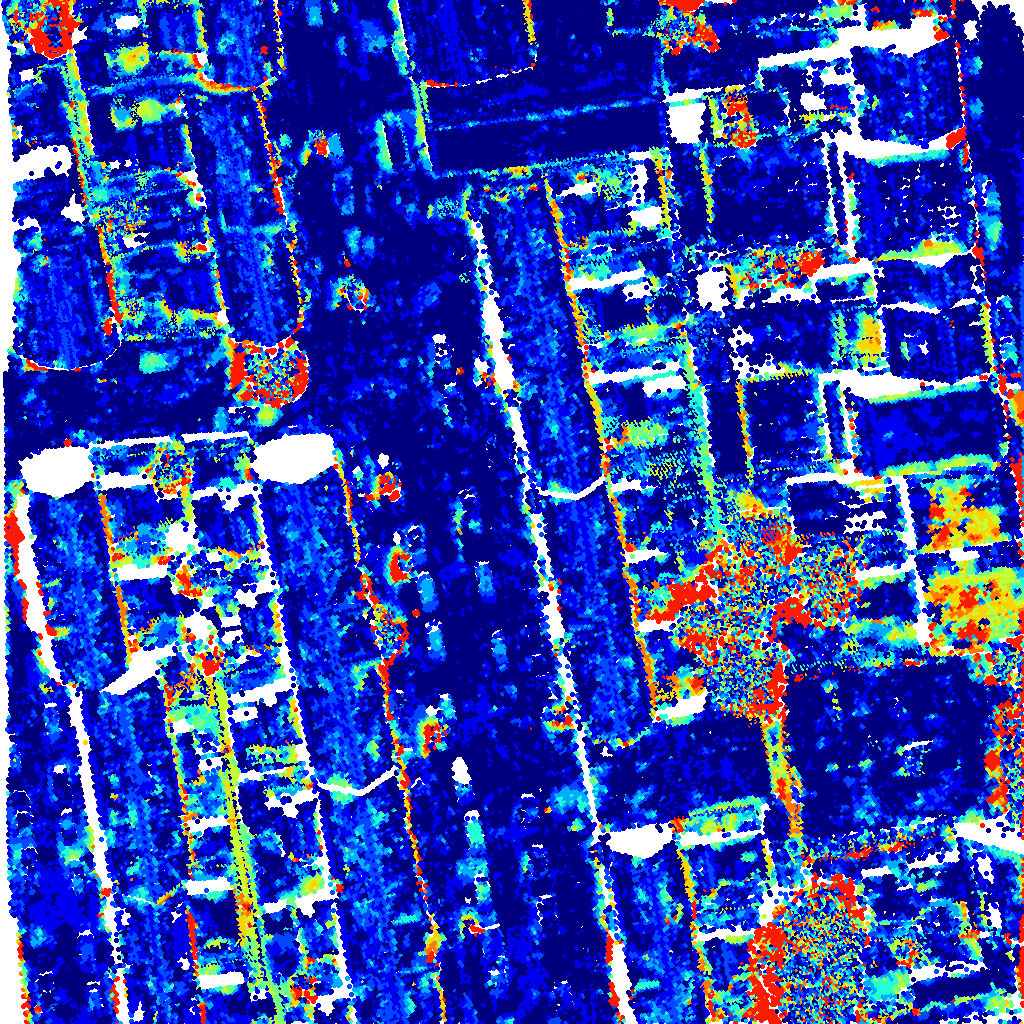}
		\includegraphics[width=\linewidth]{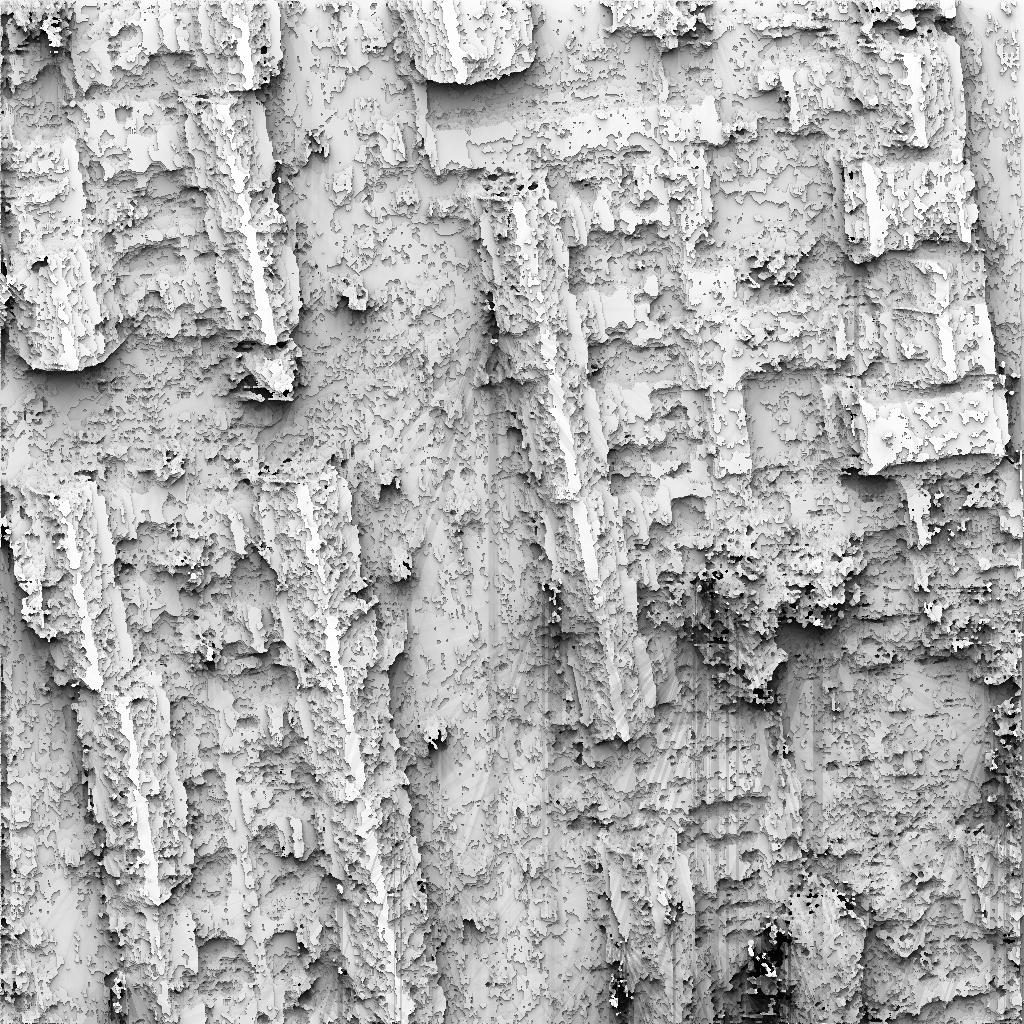}
		\centering{\tiny SGM(CUDA)}
	\end{minipage}
	\begin{minipage}[t]{0.19\textwidth}	
		\includegraphics[width=0.098\linewidth]{figures_supp/color_map.png}
		\includegraphics[width=0.85\linewidth]{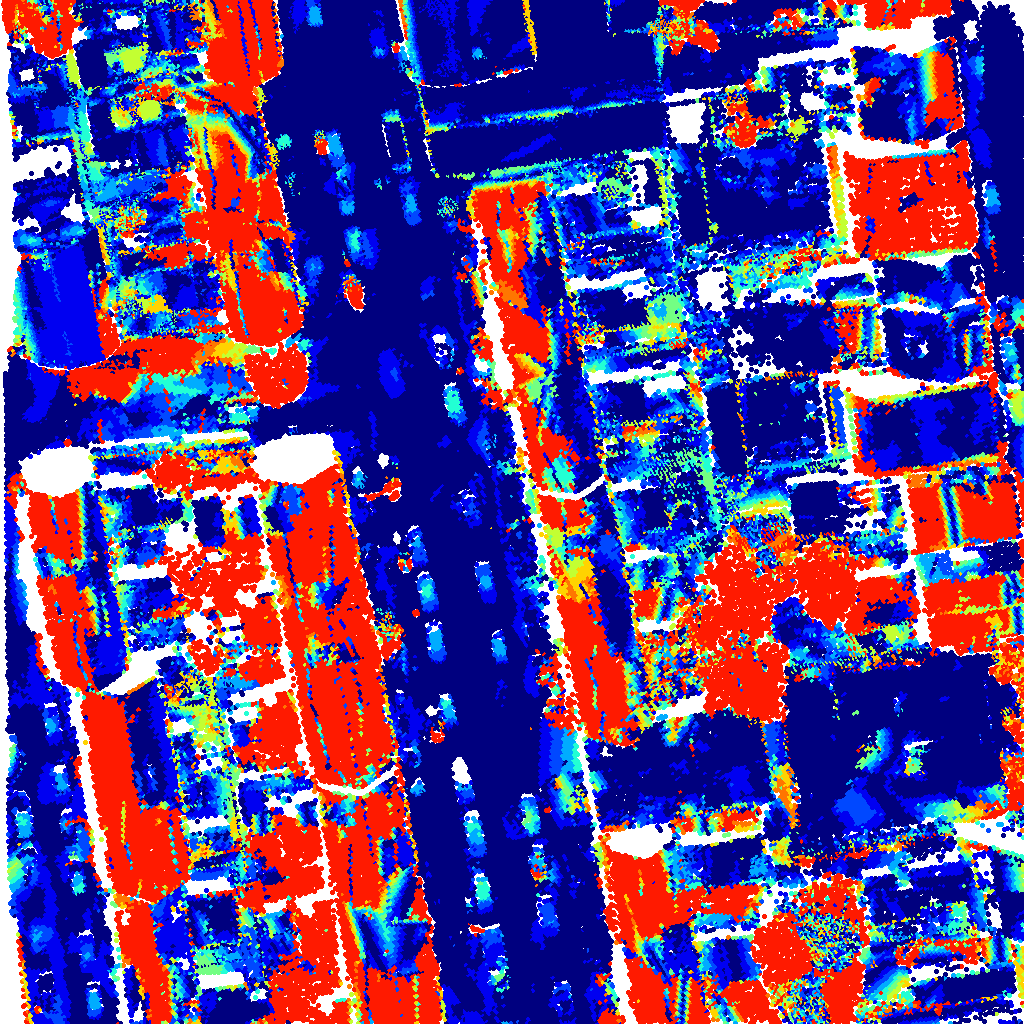}
		\includegraphics[width=\linewidth]{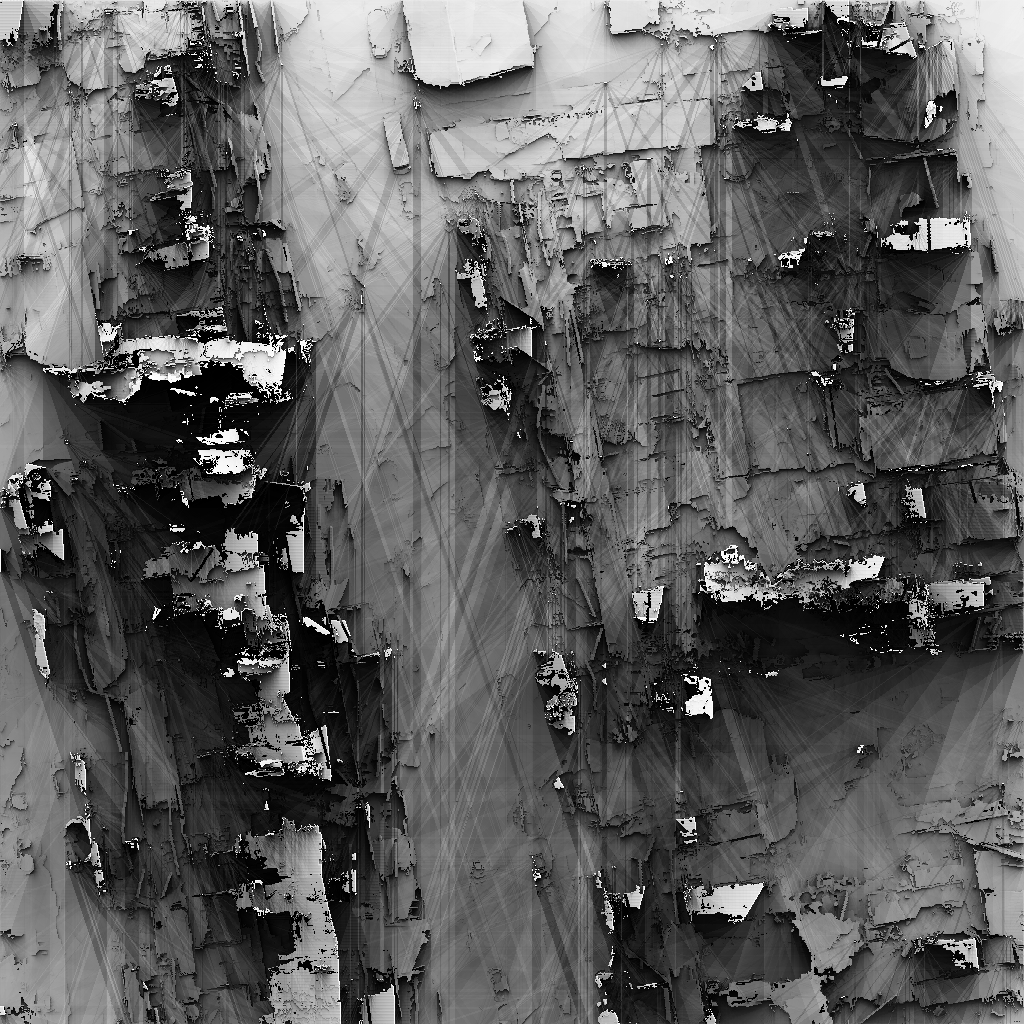}
		\centering{\tiny GraphCuts}
	\end{minipage}
	\begin{minipage}[t]{0.19\textwidth}	
		\includegraphics[width=0.098\linewidth]{figures_supp/color_map.png}
		\includegraphics[width=0.85\linewidth]{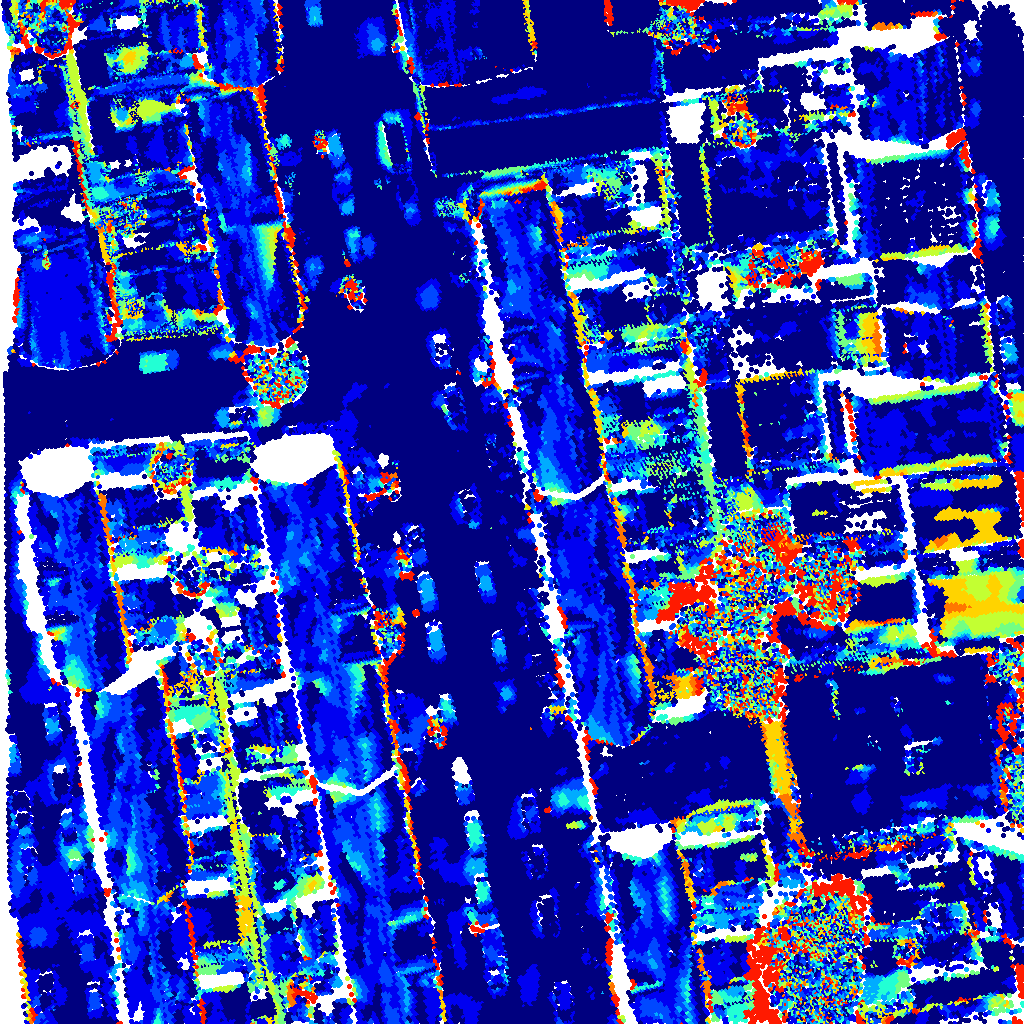}
		\includegraphics[width=\linewidth]{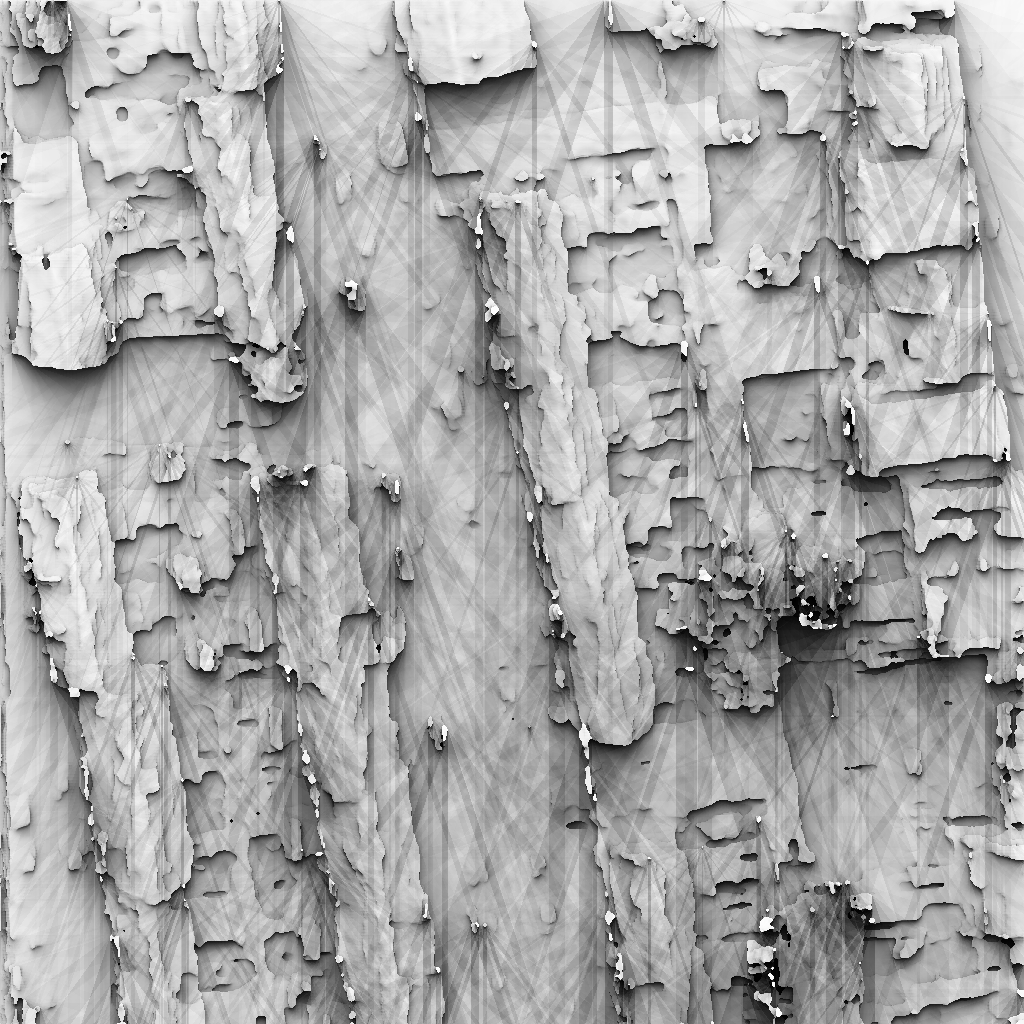}
		\centering{\tiny CBMV(SGM)}
	\end{minipage}
	\begin{minipage}[t]{0.19\textwidth}	
		\includegraphics[width=0.098\linewidth]{figures_supp/color_map.png}
		\includegraphics[width=0.85\linewidth]{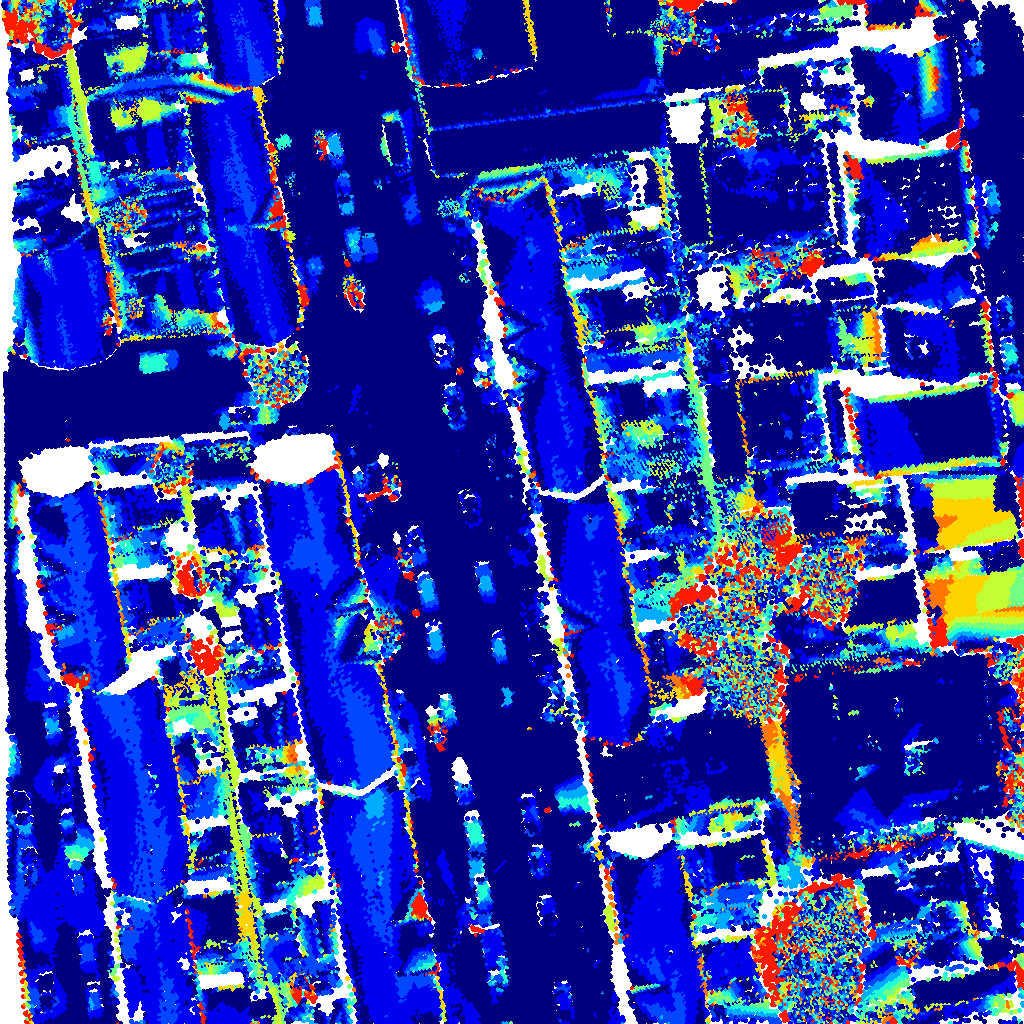}
		\includegraphics[width=\linewidth]{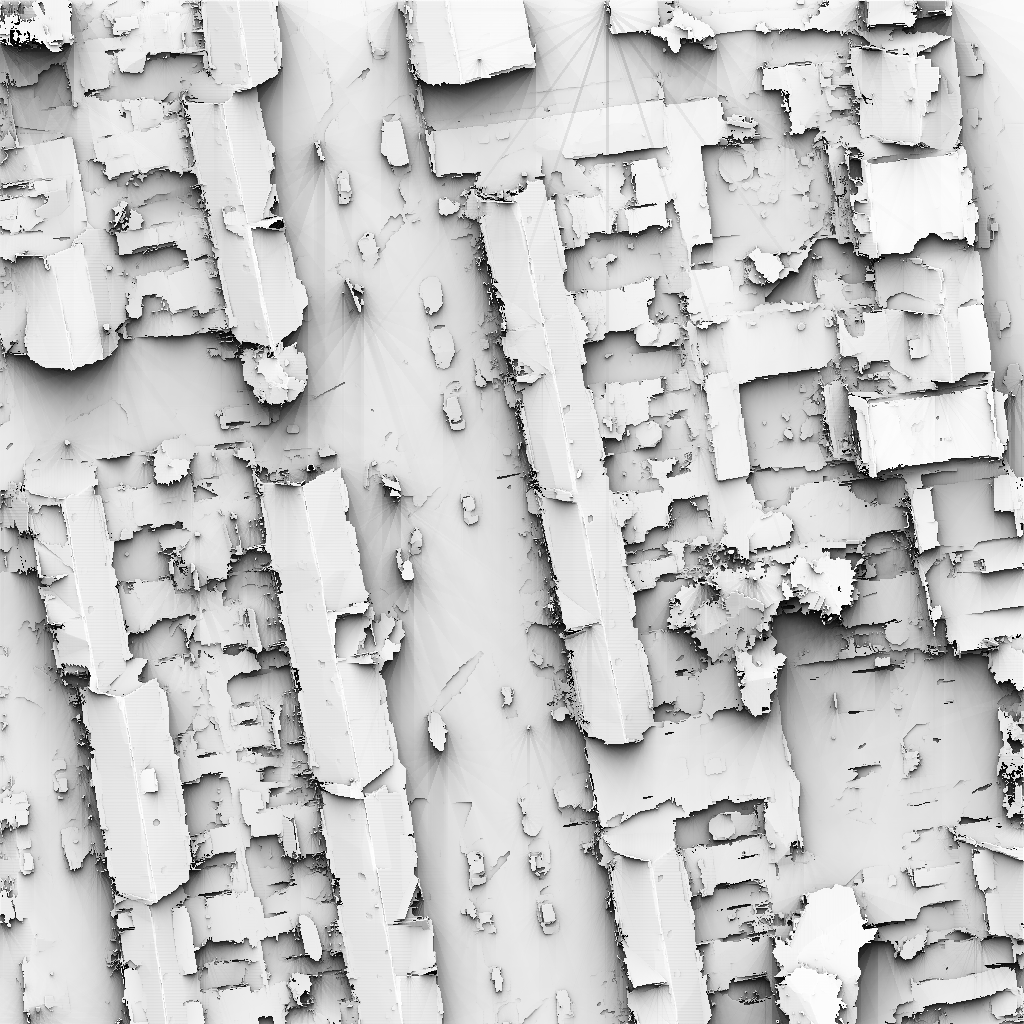}
		\centering{\tiny CBMV(GraphCuts)}
	\end{minipage}
	\begin{minipage}[t]{0.19\textwidth}	
		\includegraphics[width=0.098\linewidth]{figures_supp/color_map.png}
		\includegraphics[width=0.85\linewidth]{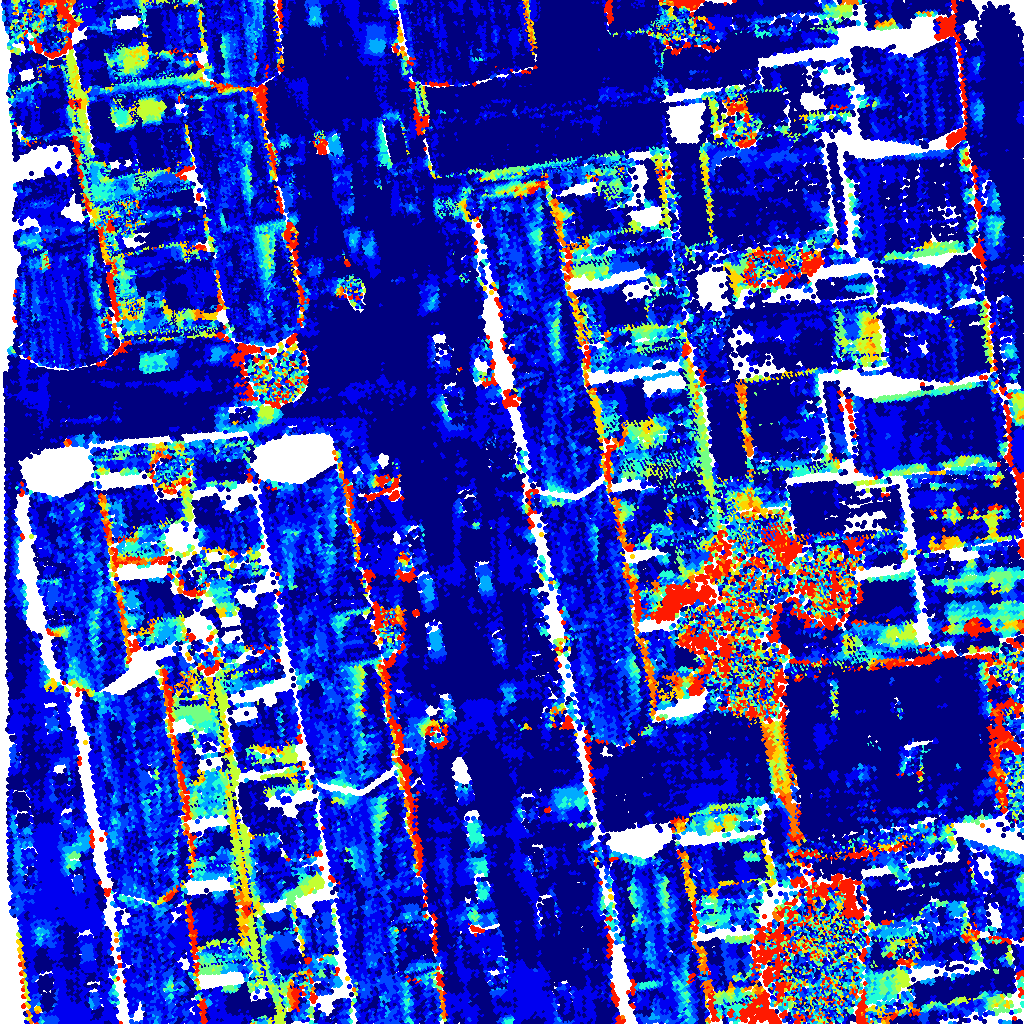}
		\includegraphics[width=\linewidth]{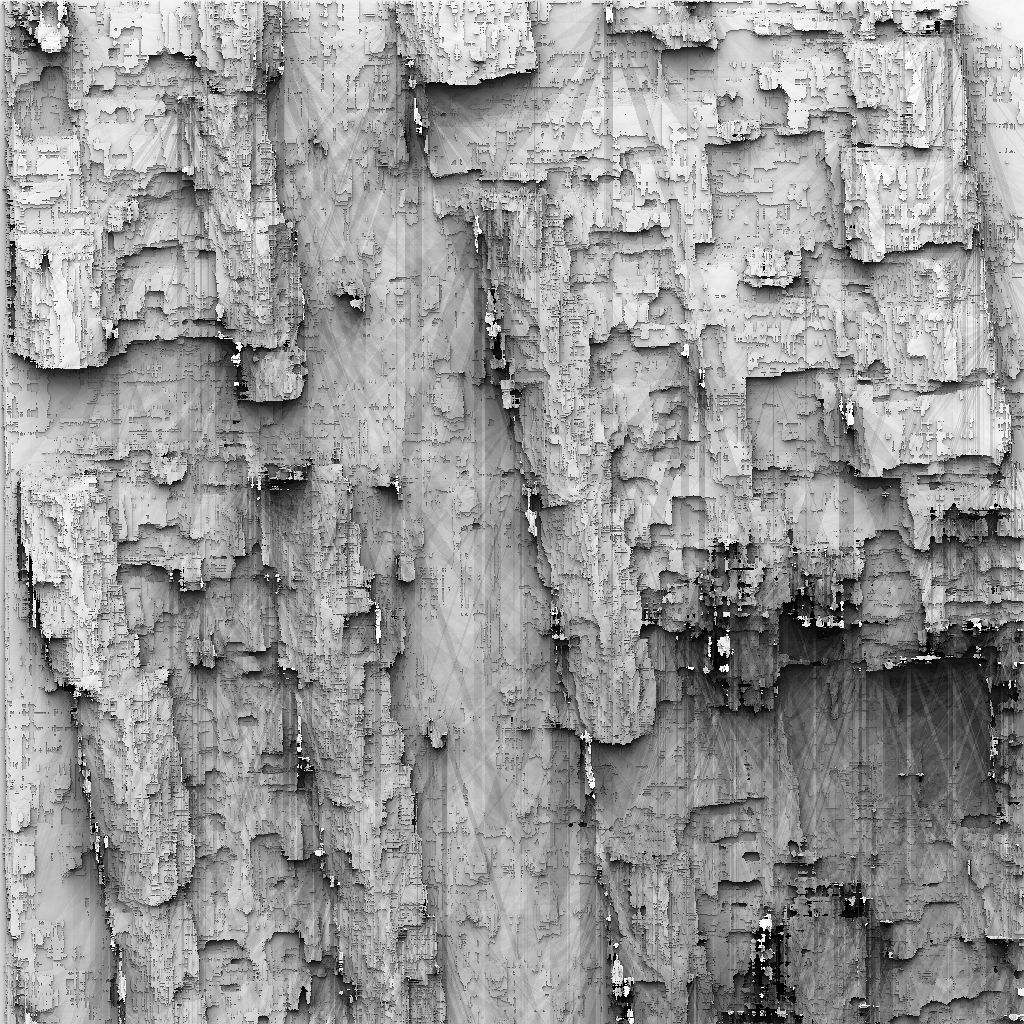}
		\centering{\tiny MC-CNN(KITTI)}
	\end{minipage}
	\begin{minipage}[t]{0.19\textwidth}	
		\includegraphics[width=0.098\linewidth]{figures_supp/color_map.png}
		\includegraphics[width=0.85\linewidth]{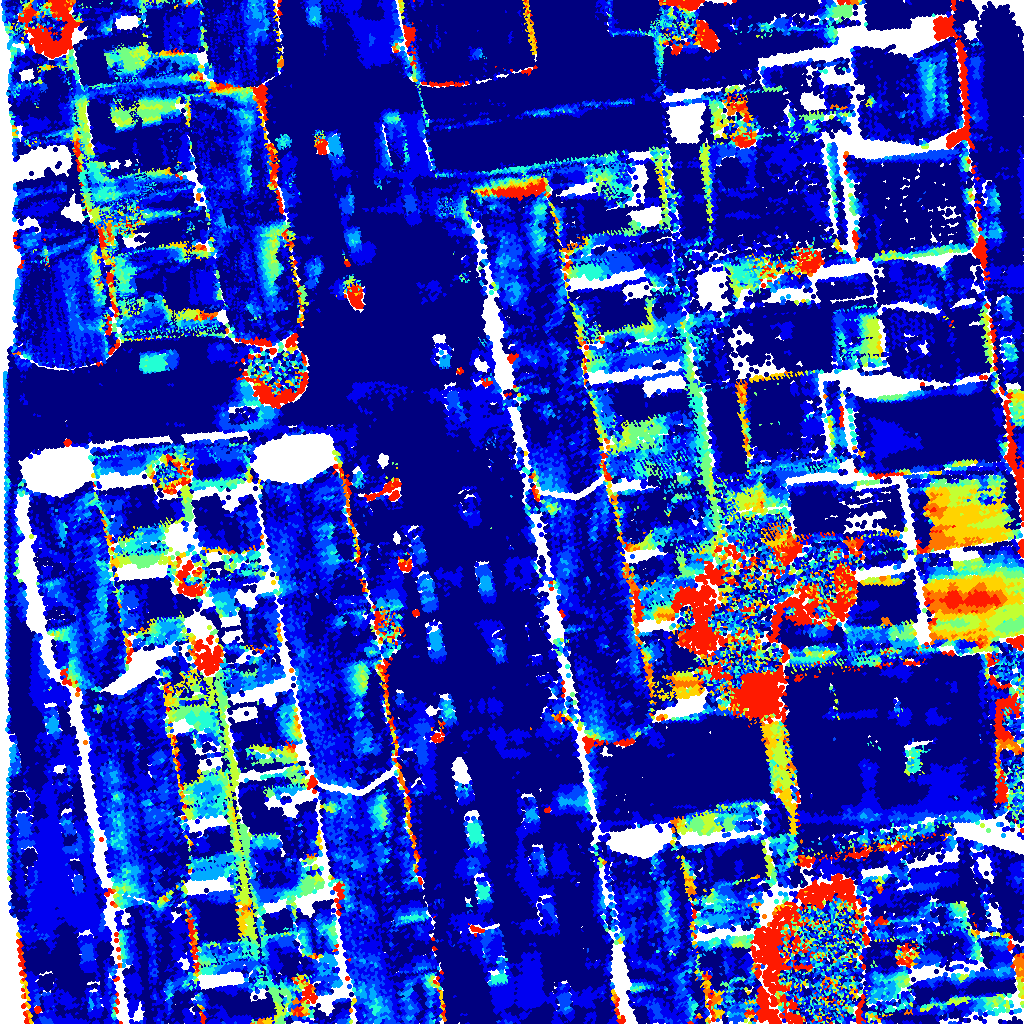}
		\includegraphics[width=\linewidth]{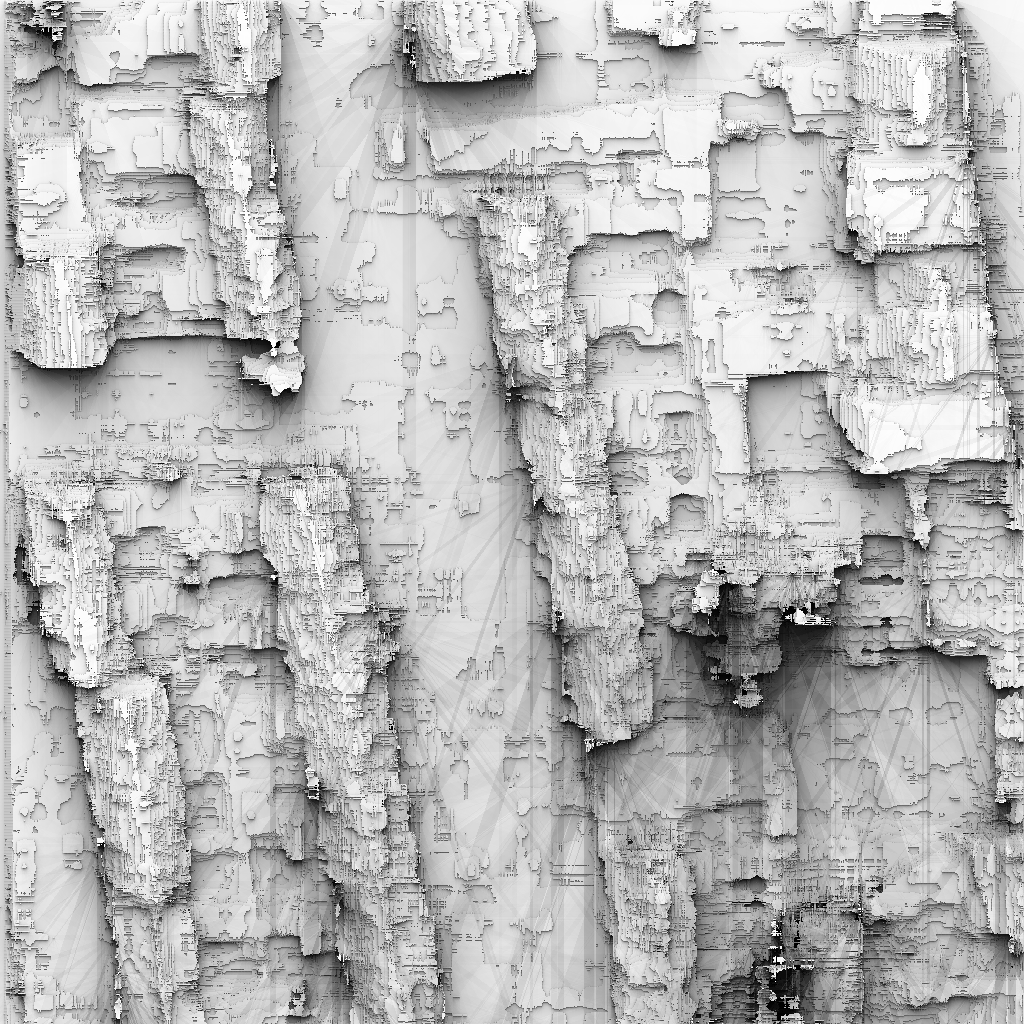}
		\centering{\tiny DeepFeature(KITTI)}
	\end{minipage}
	\begin{minipage}[t]{0.19\textwidth}	
		\includegraphics[width=0.098\linewidth]{figures_supp/color_map.png}
		\includegraphics[width=0.85\linewidth]{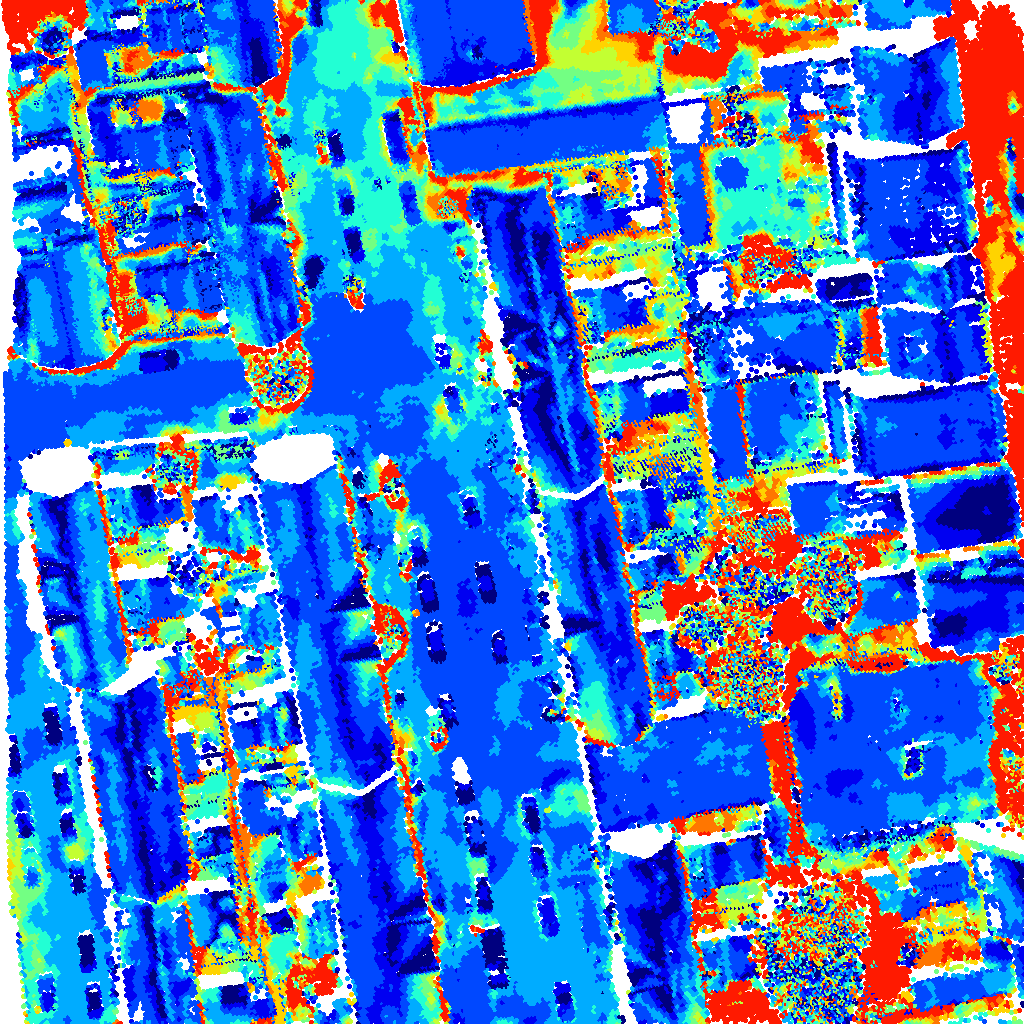}
		\includegraphics[width=\linewidth]{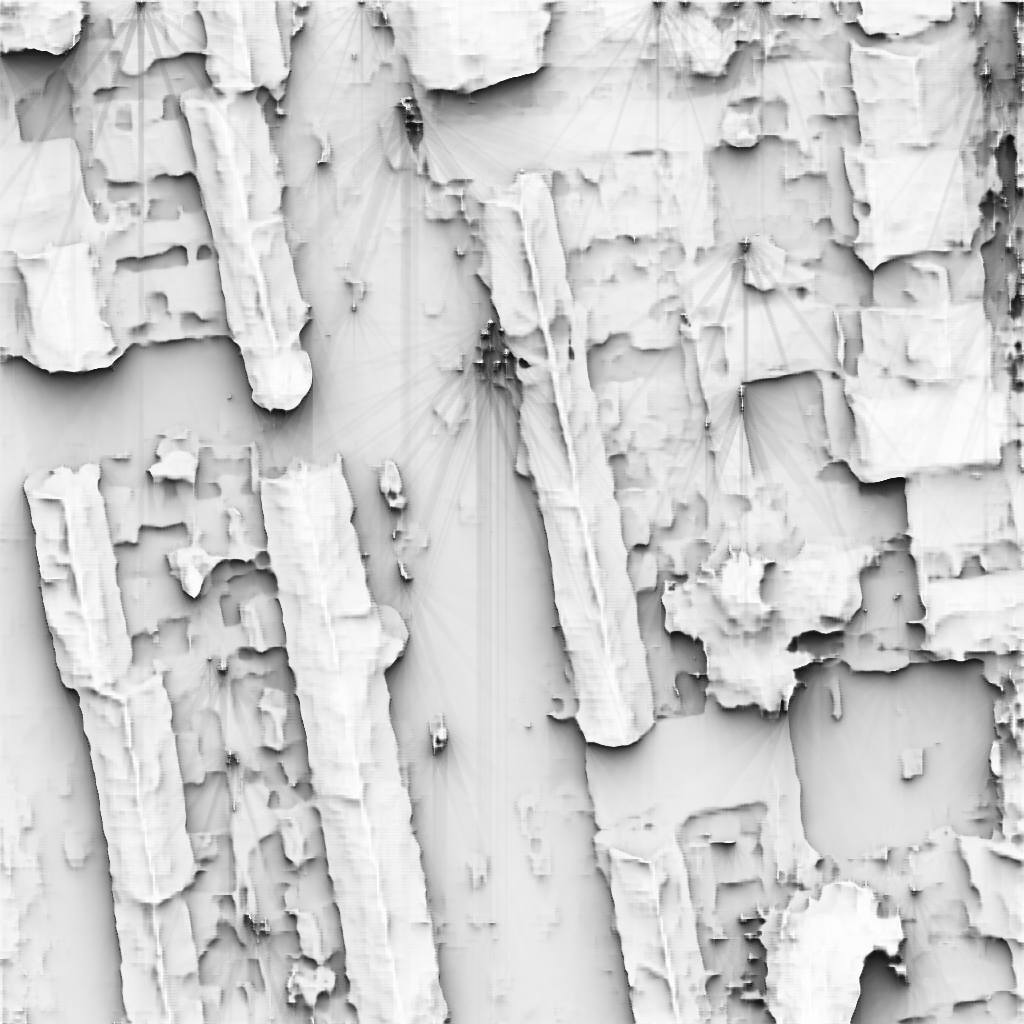}
		\centering{\tiny PSM net(KITTI)}
	\end{minipage}
	\begin{minipage}[t]{0.19\textwidth}	
		\includegraphics[width=0.098\linewidth]{figures_supp/color_map.png}
		\includegraphics[width=0.85\linewidth]{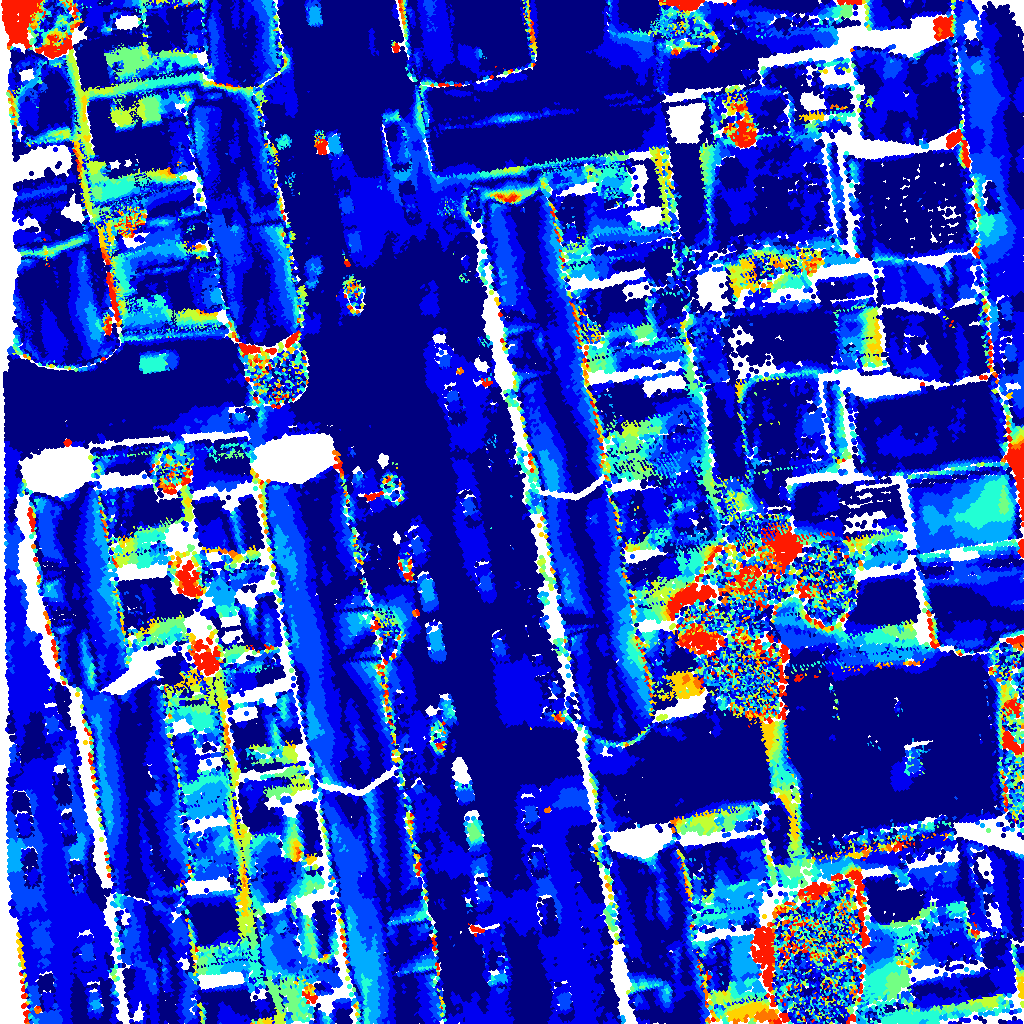}
		\includegraphics[width=\linewidth]{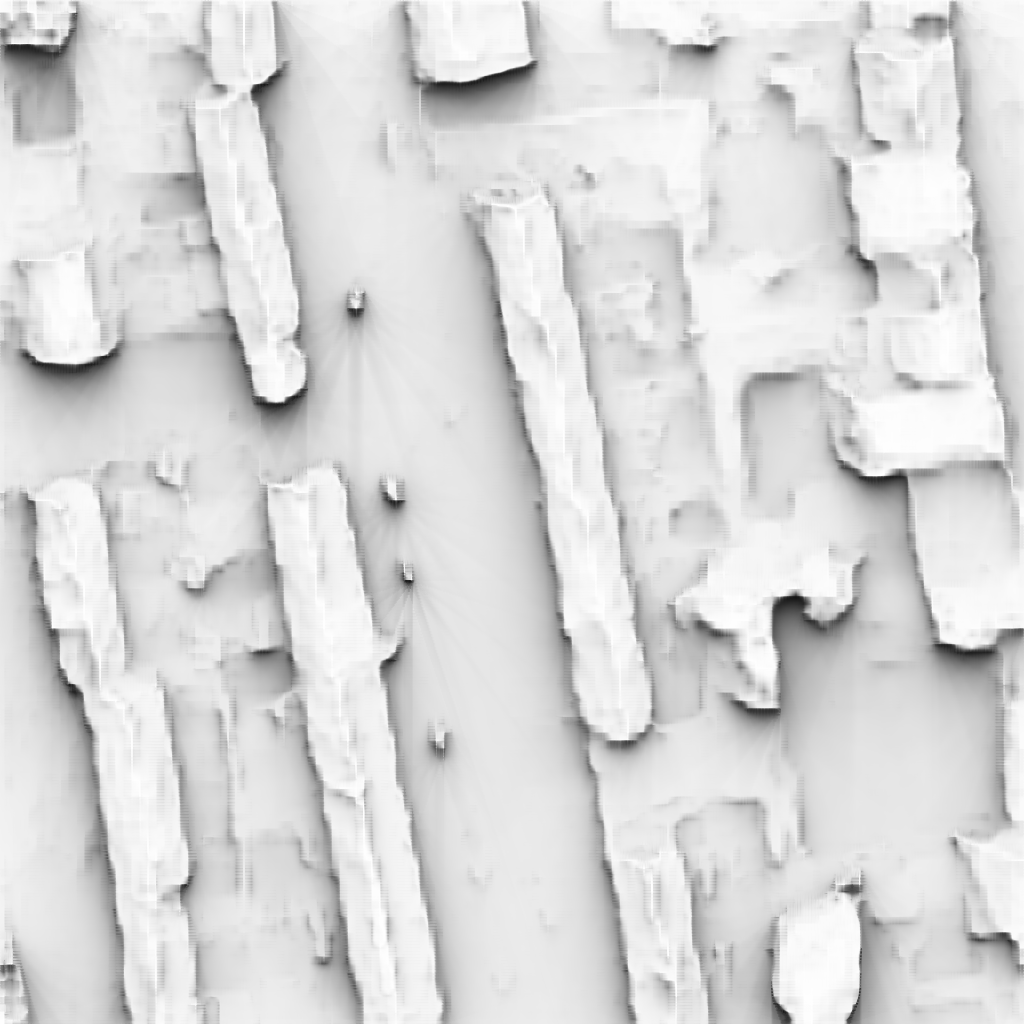}
		\centering{\tiny HRS net(KITTI)}
	\end{minipage}
	\begin{minipage}[t]{0.19\textwidth}	
		\includegraphics[width=0.098\linewidth]{figures_supp/color_map.png}
		\includegraphics[width=0.85\linewidth]{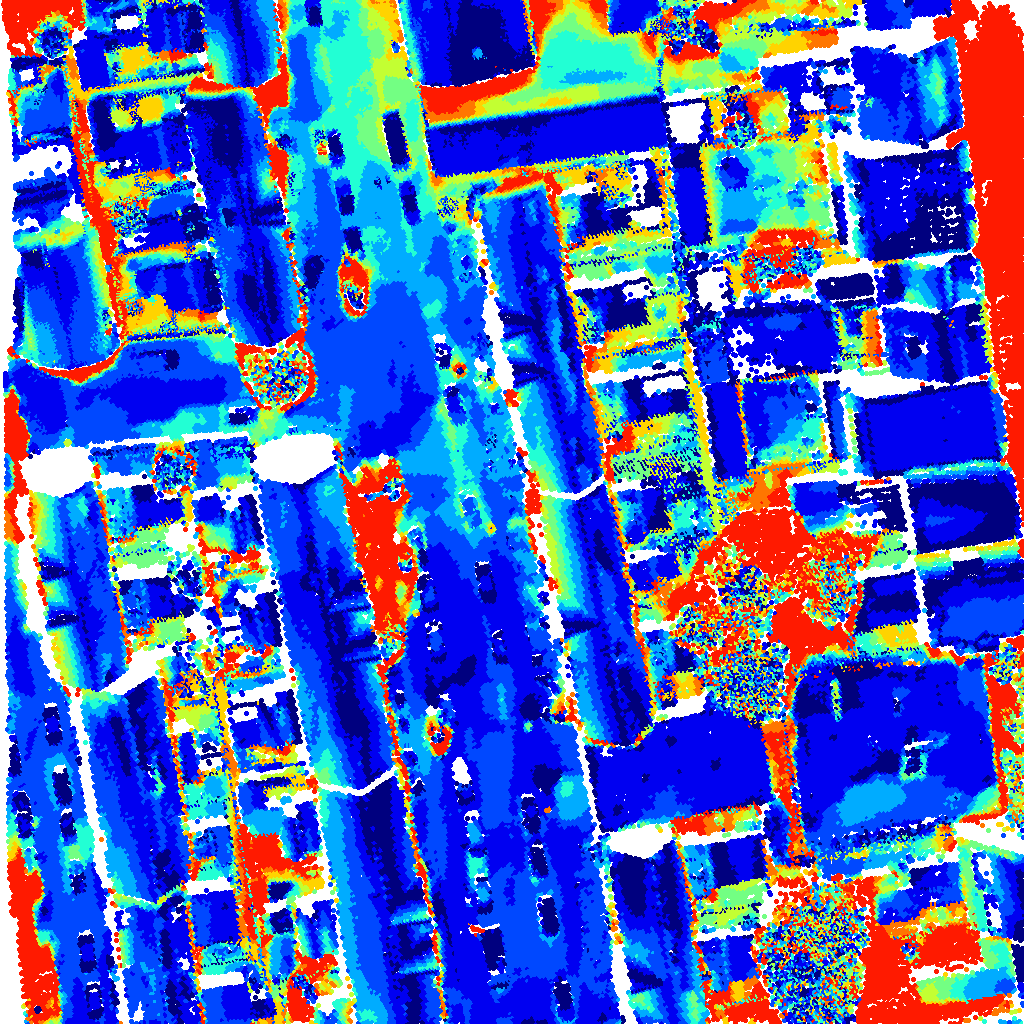}
		\includegraphics[width=\linewidth]{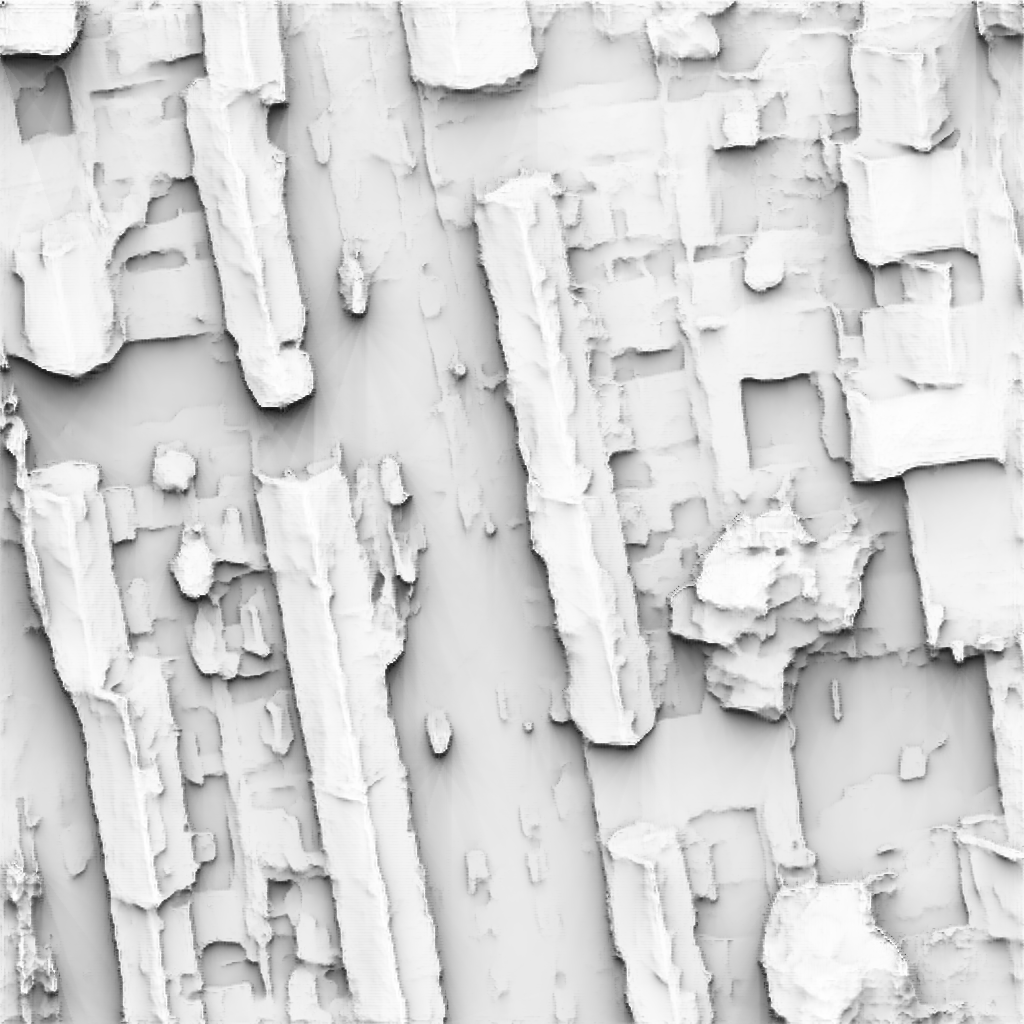}
		\centering{\tiny DeepPruner(KITTI)}
	\end{minipage}
	\begin{minipage}[t]{0.19\textwidth}
		\includegraphics[width=0.098\linewidth]{figures_supp/color_map.png}
		\includegraphics[width=0.85\linewidth]{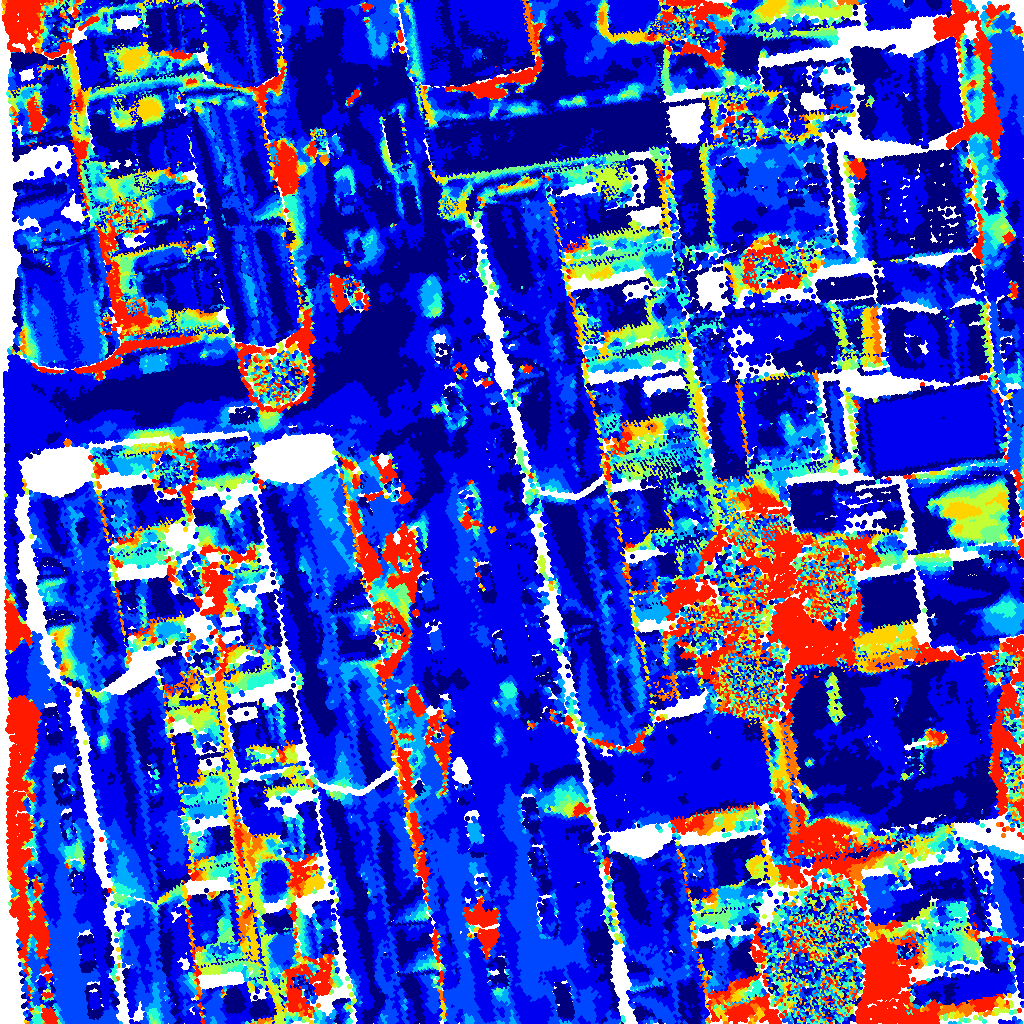}
		\includegraphics[width=\linewidth]{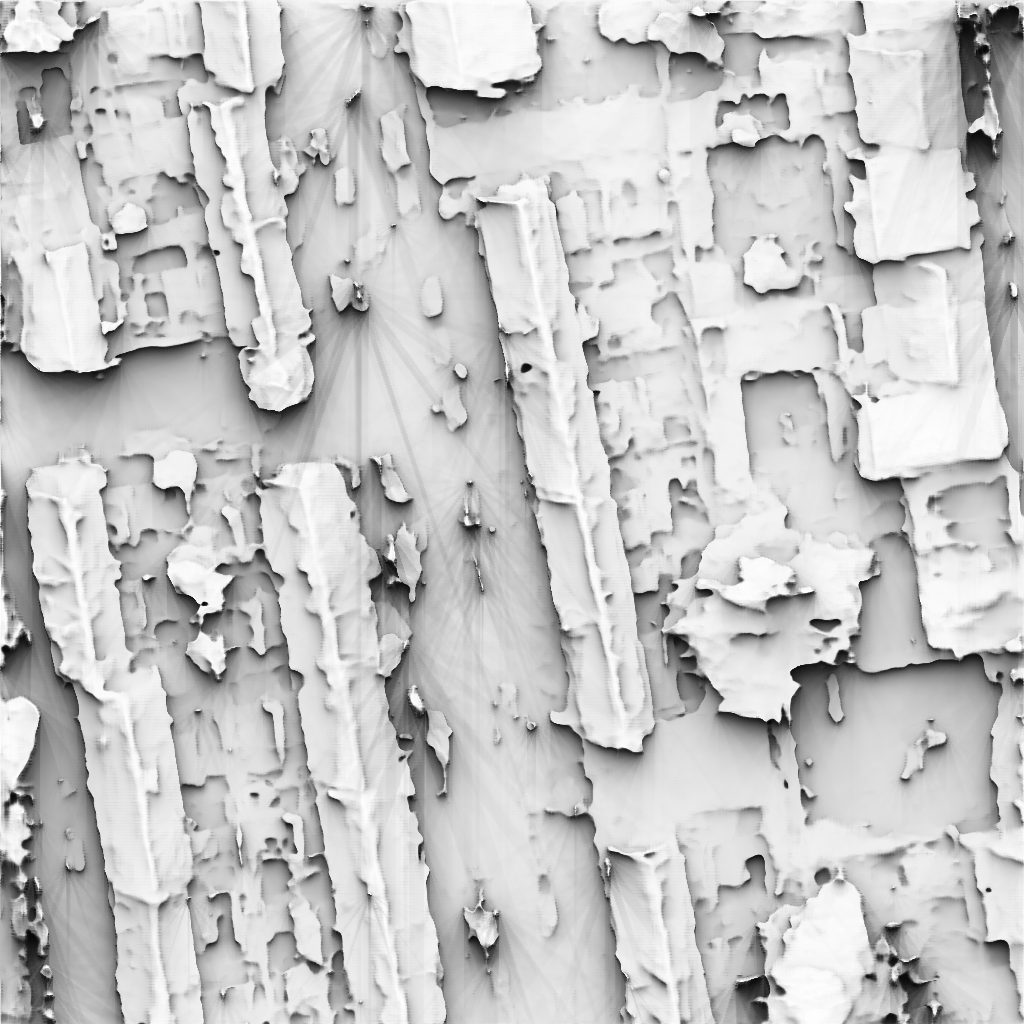}
		\centering{\tiny GANet(KITTI)}
	\end{minipage}
		\begin{minipage}[t]{0.19\textwidth}	
		\includegraphics[width=0.098\linewidth]{figures_supp/color_map.png}
		\includegraphics[width=0.85\linewidth]{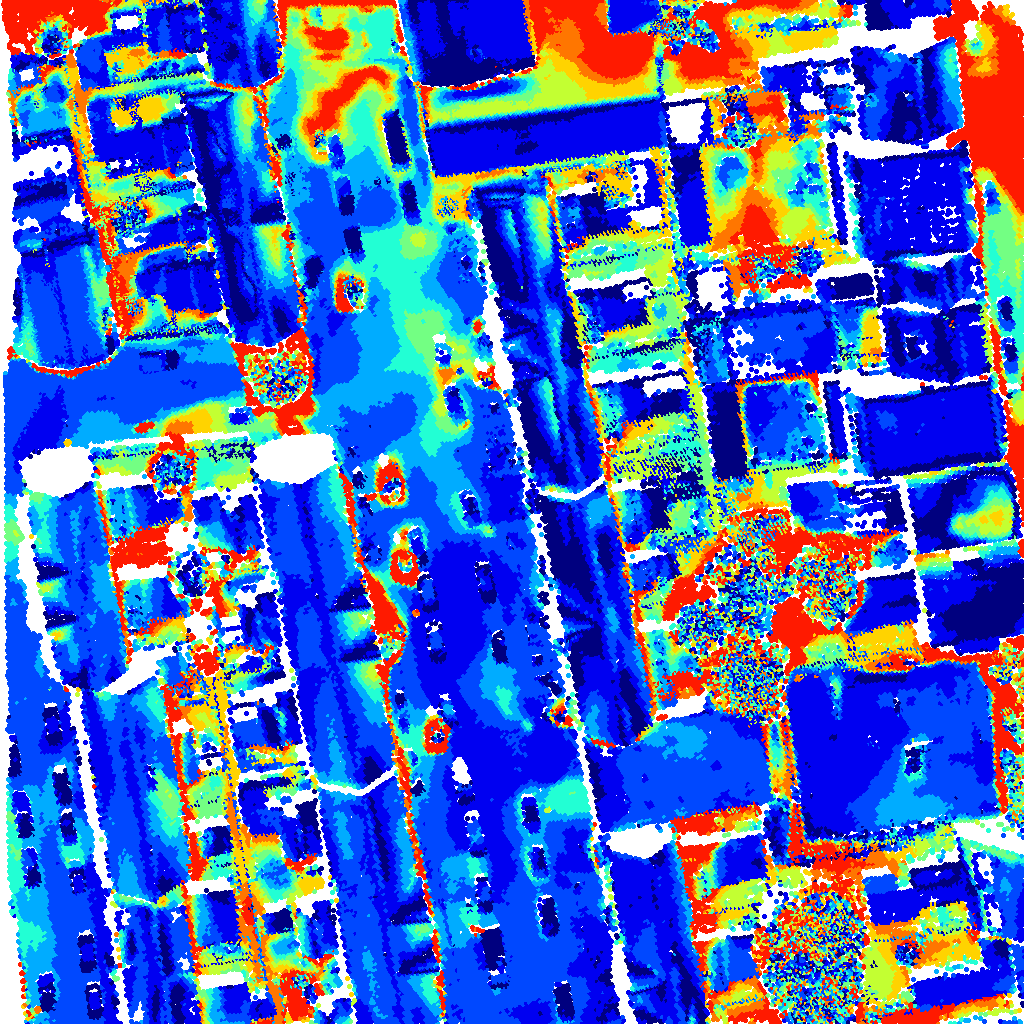}
		\includegraphics[width=\linewidth]{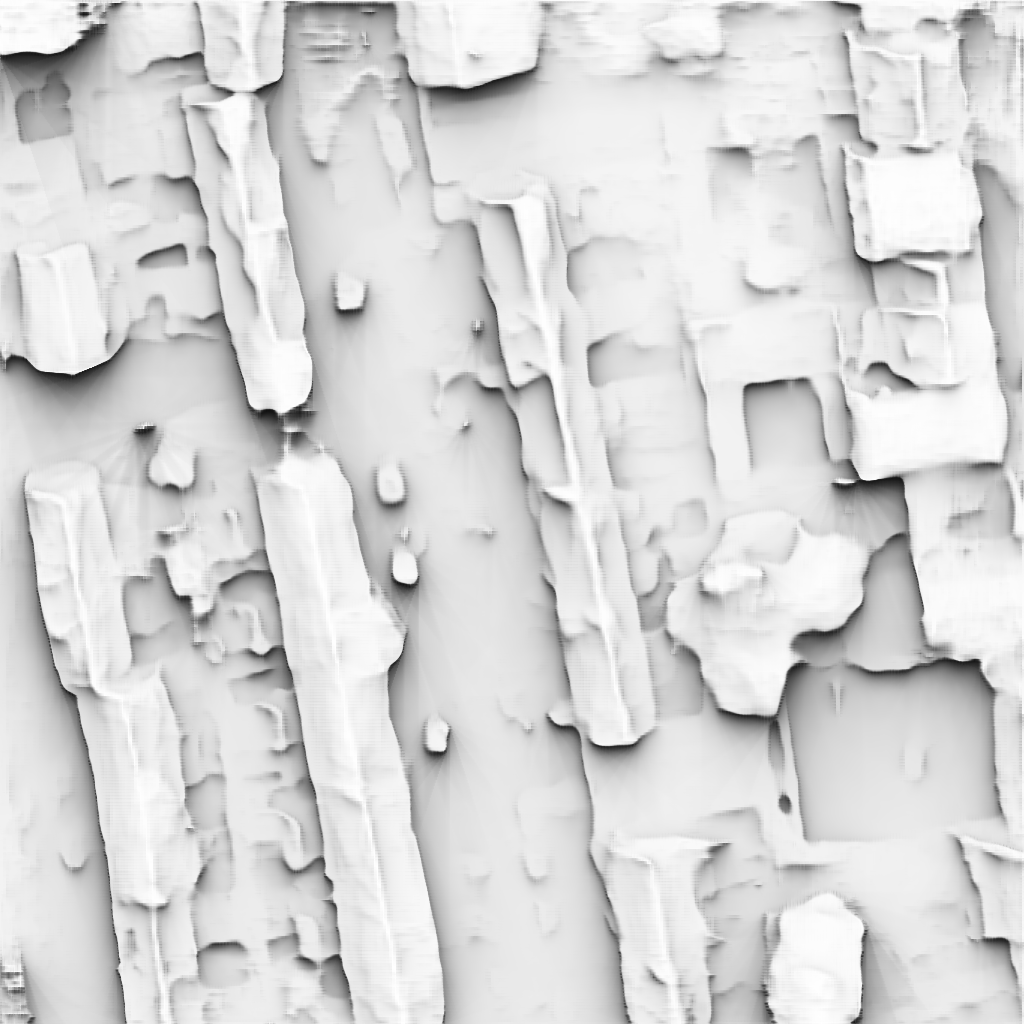}
		\centering{\tiny LEAStereo(KITTI)}
	\end{minipage}
		\begin{minipage}[t]{0.19\textwidth}	
		\includegraphics[width=0.098\linewidth]{figures_supp/color_map.png}
		\includegraphics[width=0.85\linewidth]{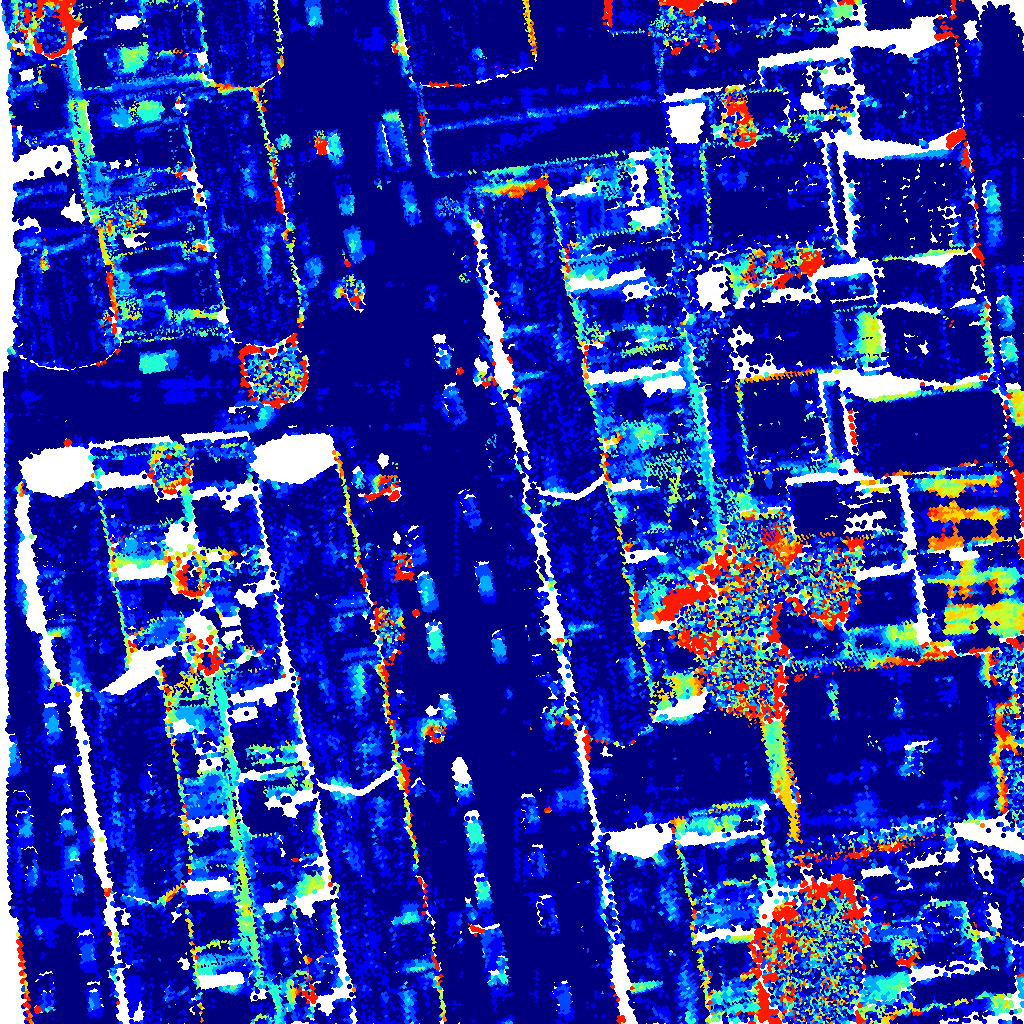}
		\includegraphics[width=\linewidth]{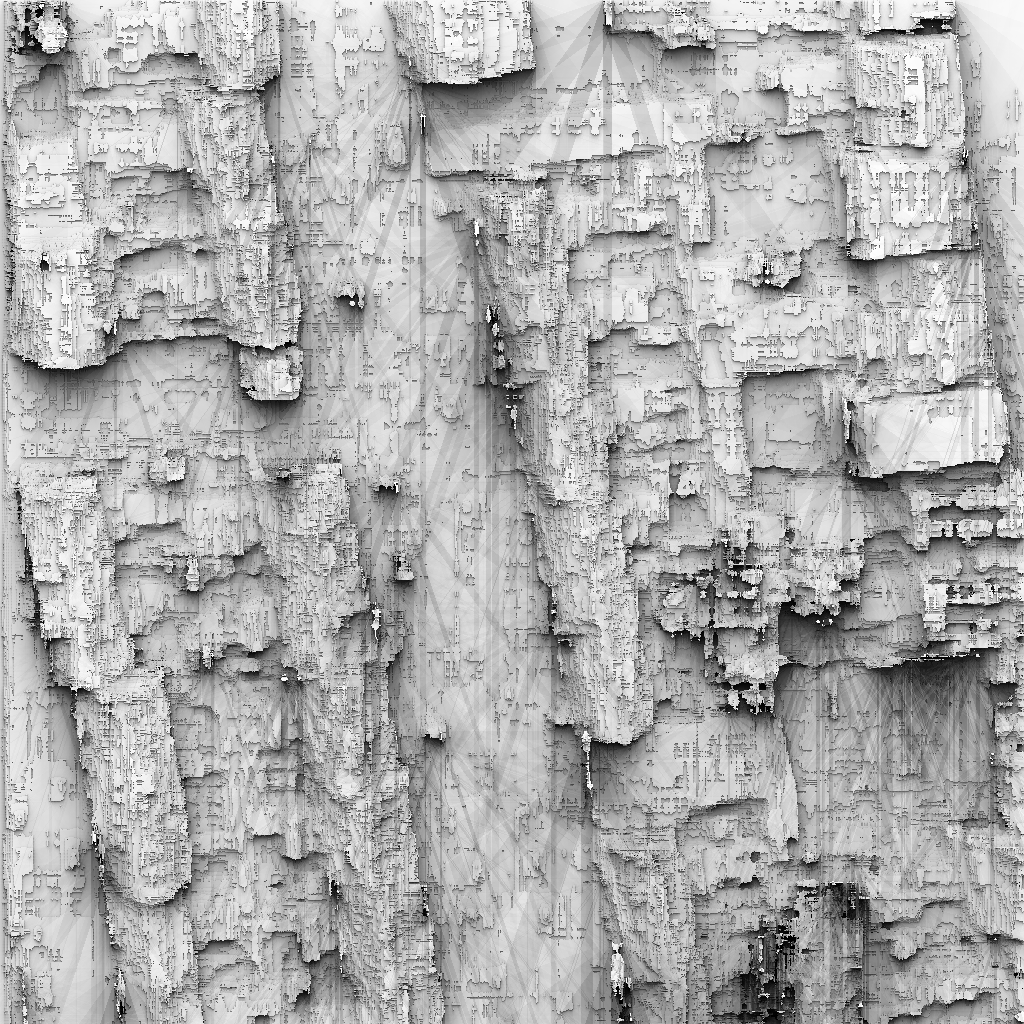}
		\centering{\tiny MC-CNN}
	\end{minipage}
	\begin{minipage}[t]{0.19\textwidth}	
		\includegraphics[width=0.098\linewidth]{figures_supp/color_map.png}
		\includegraphics[width=0.85\linewidth]{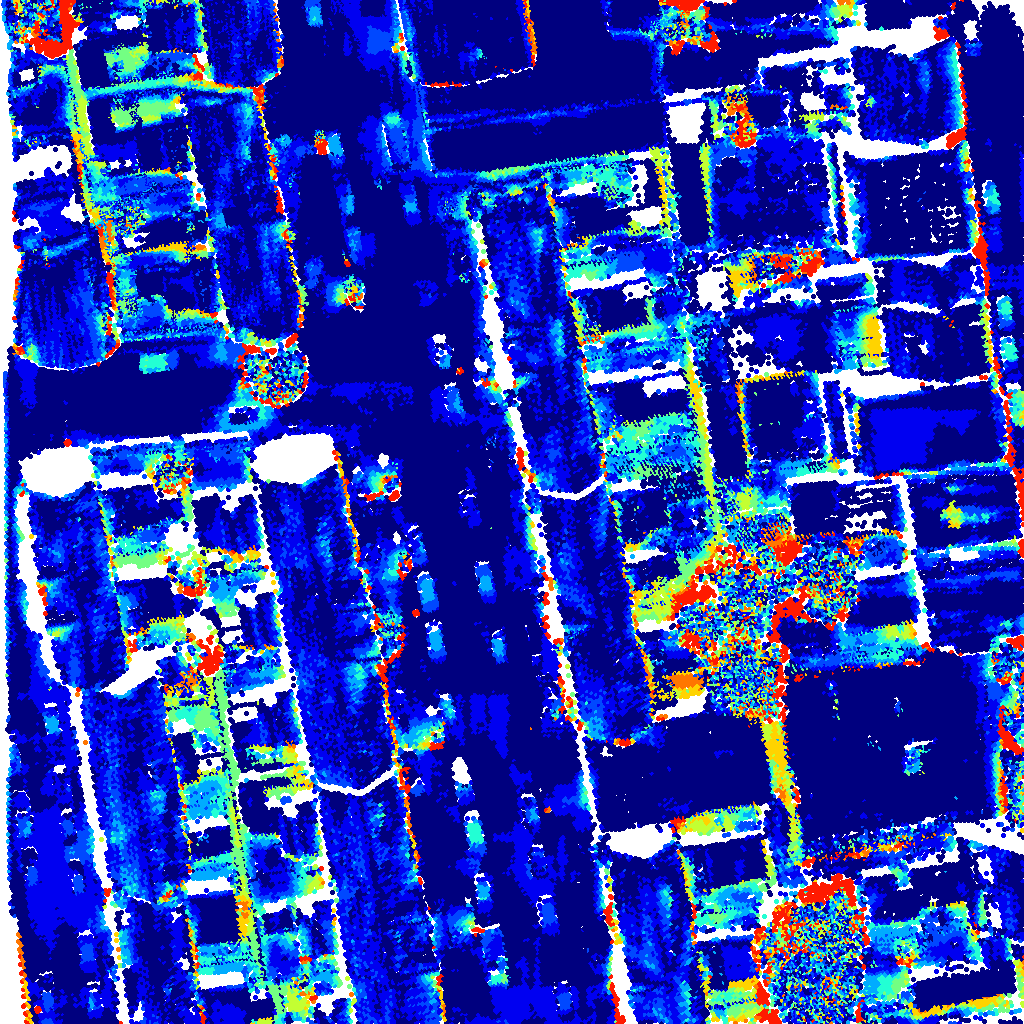}
		\includegraphics[width=\linewidth]{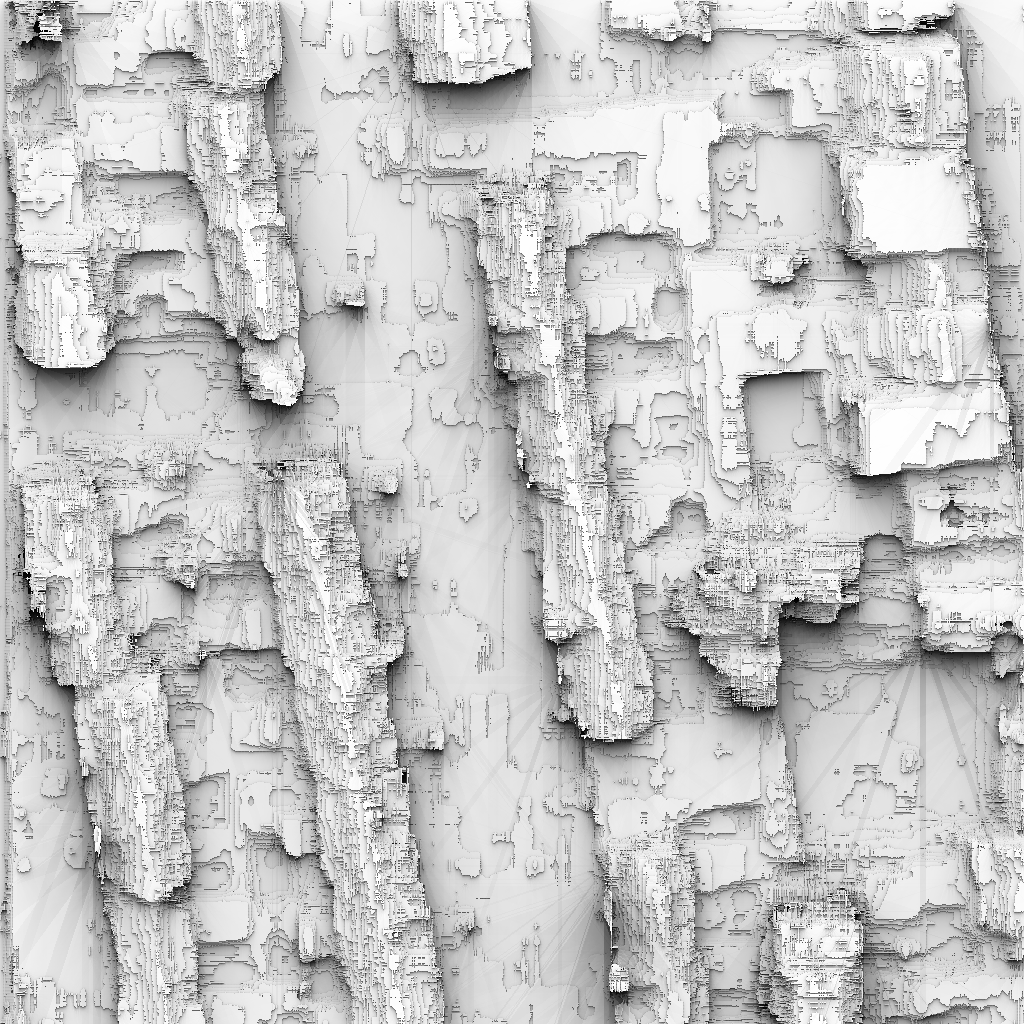}
		\centering{\tiny DeepFeature}
	\end{minipage}
	\begin{minipage}[t]{0.19\textwidth}	
		\includegraphics[width=0.098\linewidth]{figures_supp/color_map.png}
		\includegraphics[width=0.85\linewidth]{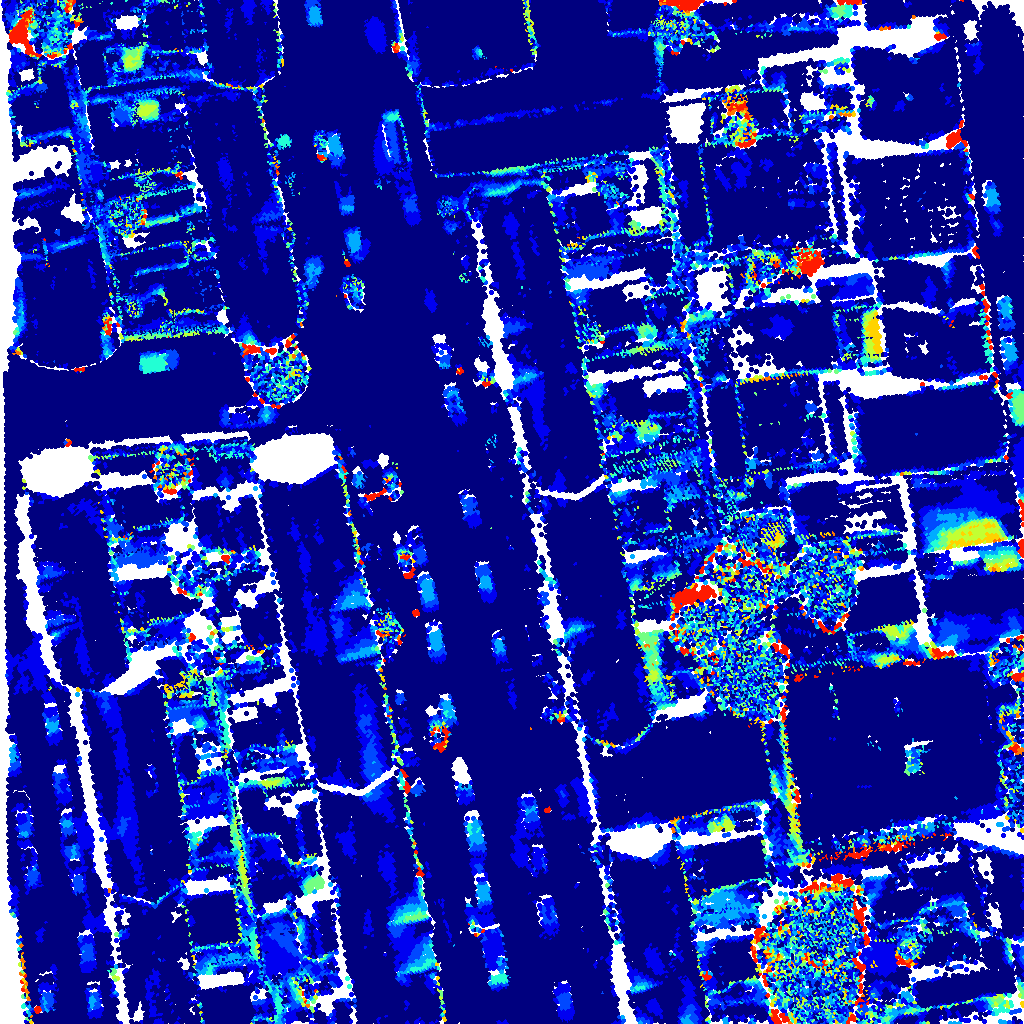}
		\includegraphics[width=\linewidth]{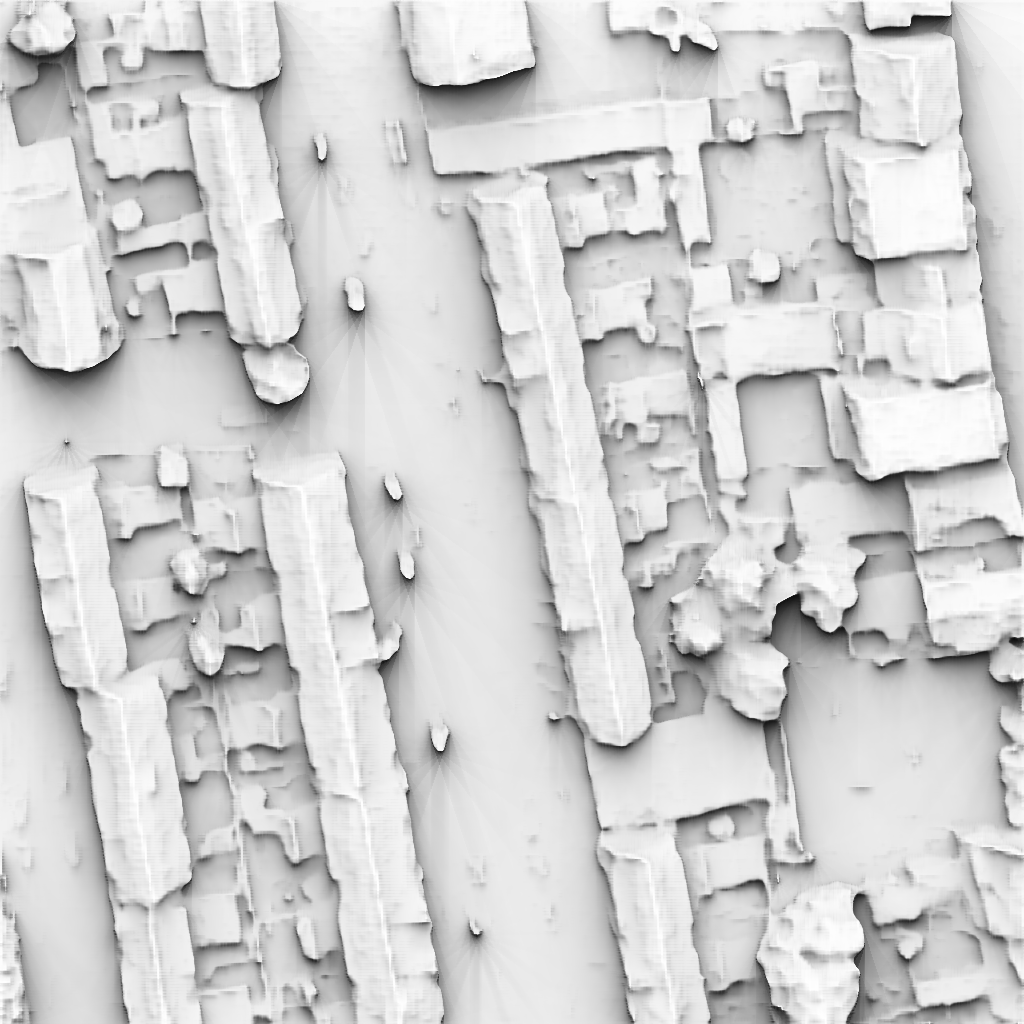}
		\centering{\tiny PSM net}
	\end{minipage}
	\begin{minipage}[t]{0.19\textwidth}	
		\includegraphics[width=0.098\linewidth]{figures_supp/color_map.png}
		\includegraphics[width=0.85\linewidth]{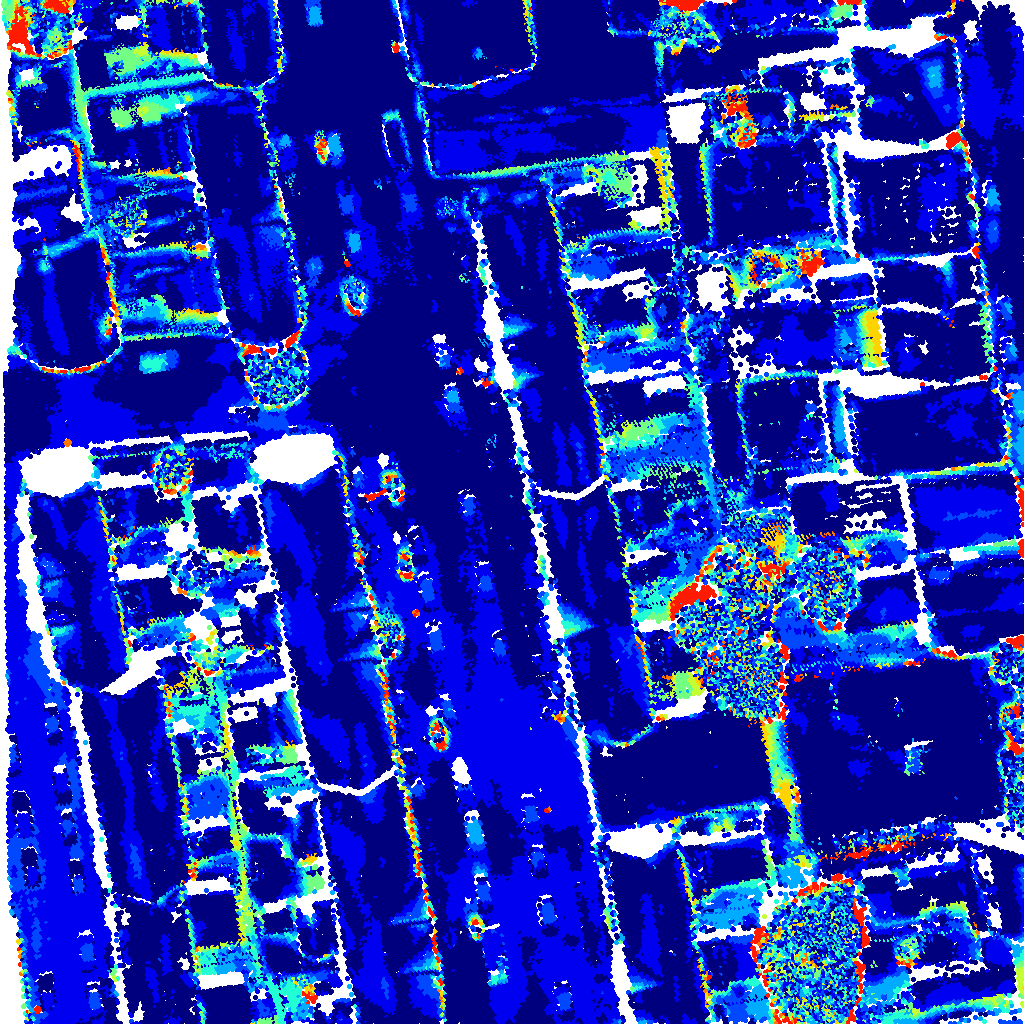}
		\includegraphics[width=\linewidth]{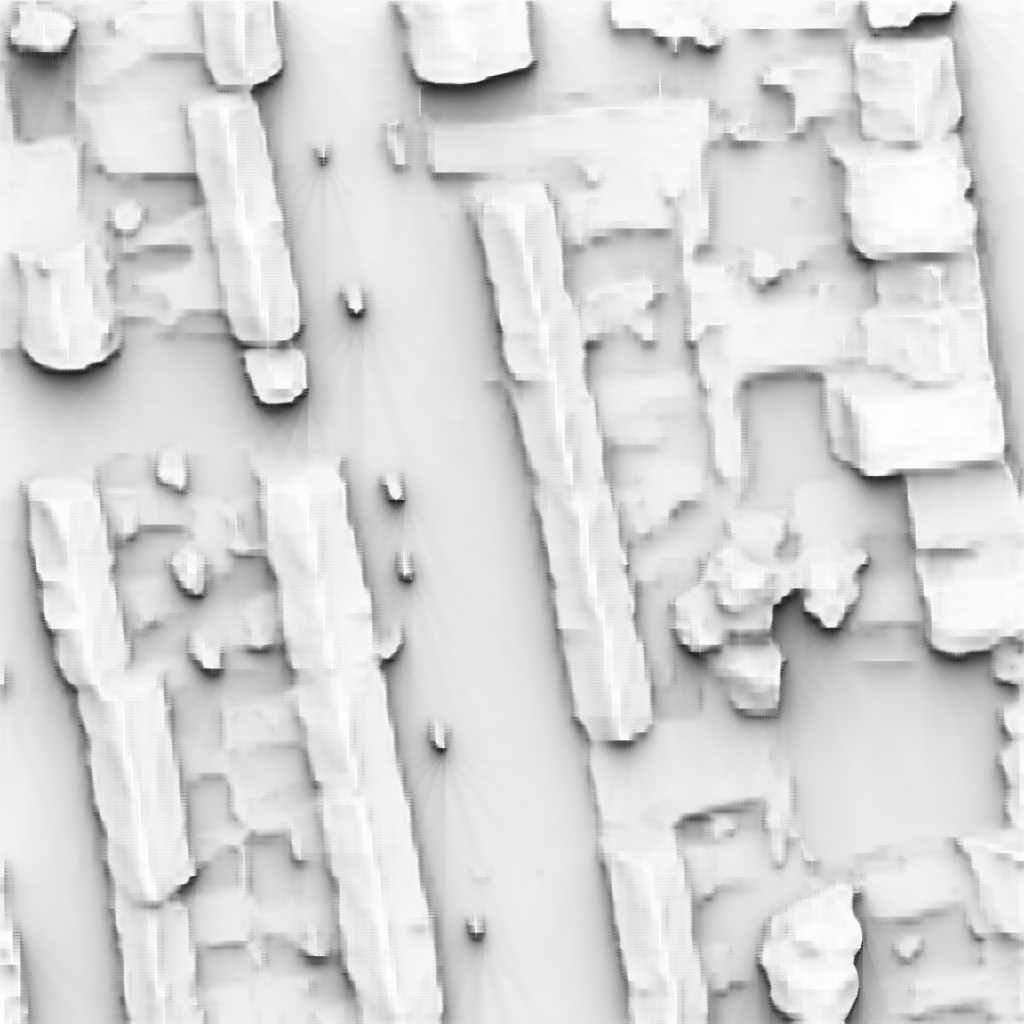}
		\centering{\tiny HRS net}
	\end{minipage}
	\begin{minipage}[t]{0.19\textwidth}	
		\includegraphics[width=0.098\linewidth]{figures_supp/color_map.png}
		\includegraphics[width=0.85\linewidth]{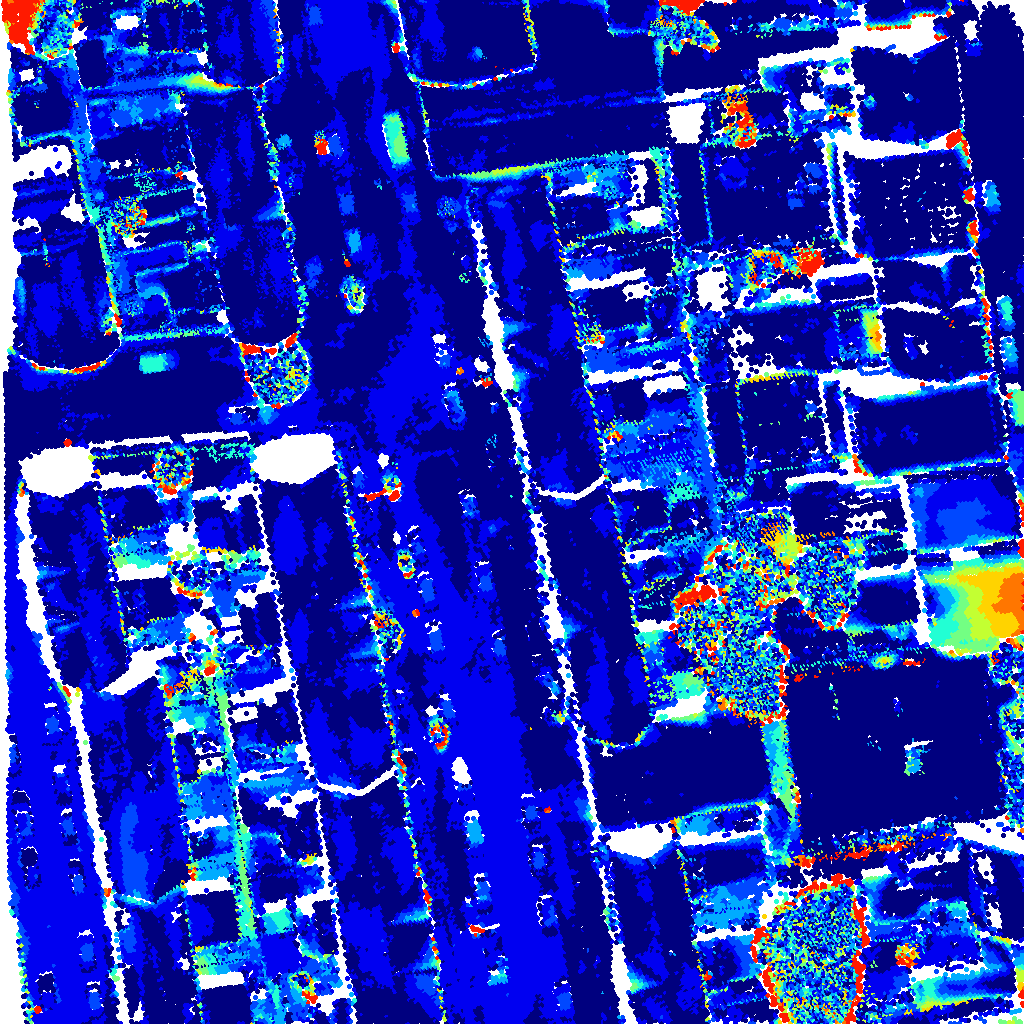}
		\includegraphics[width=\linewidth]{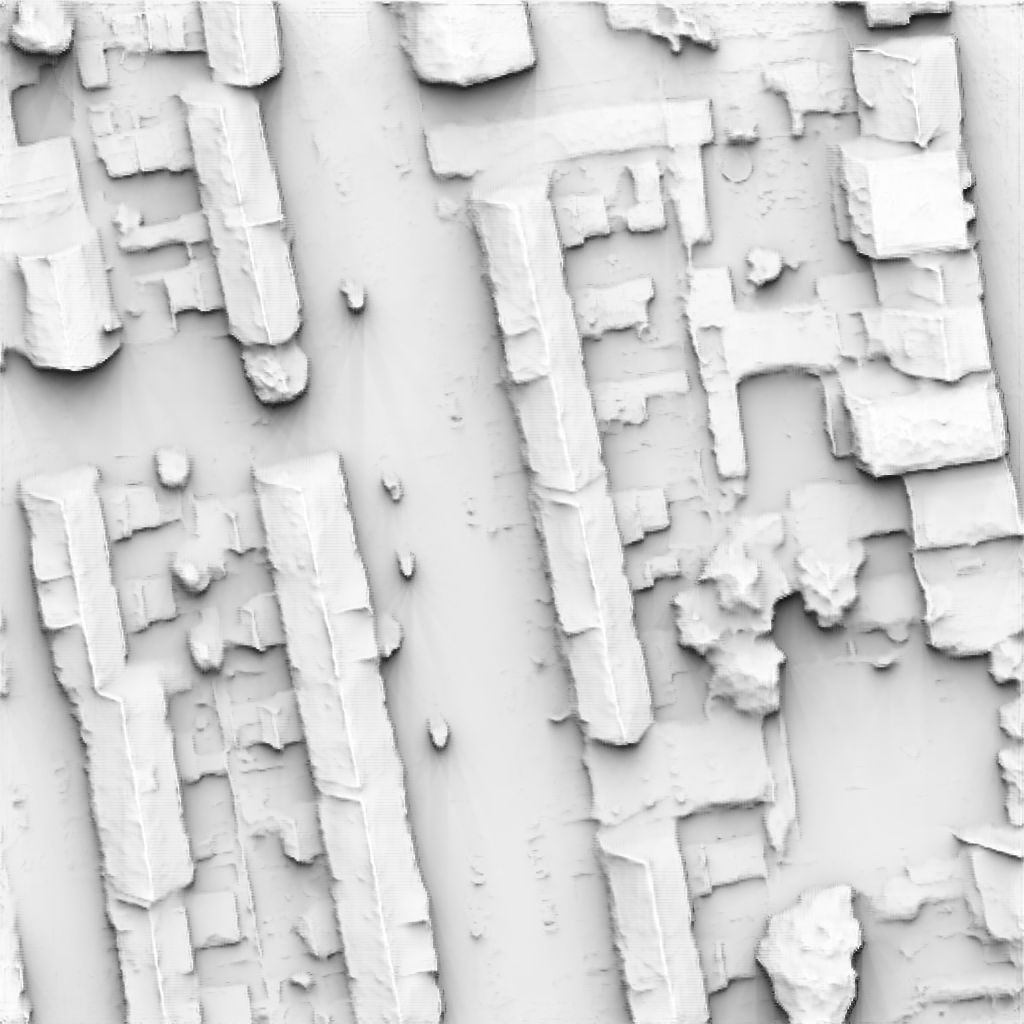}
		\centering{\tiny DeepPruner}
	\end{minipage}
	\begin{minipage}[t]{0.19\textwidth}
		\includegraphics[width=0.098\linewidth]{figures_supp/color_map.png}
		\includegraphics[width=0.85\linewidth]{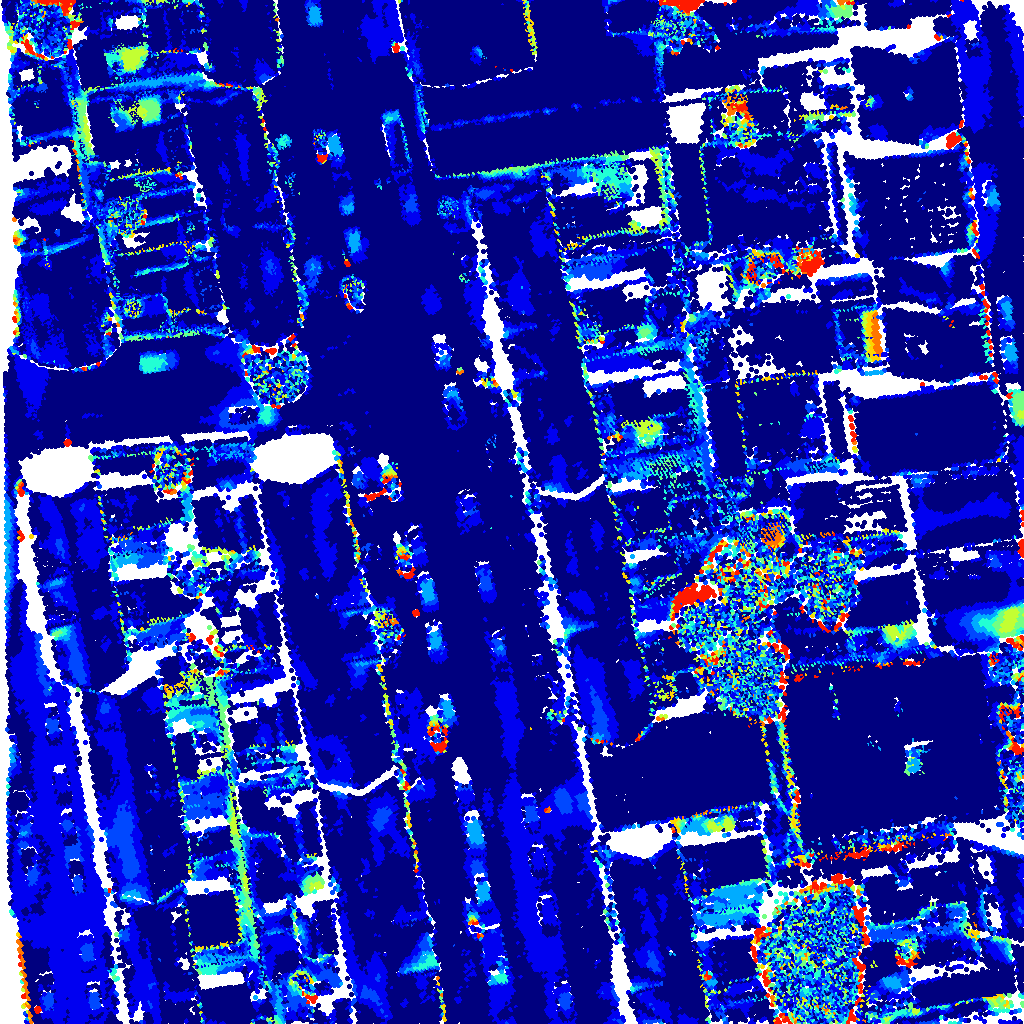}
		\includegraphics[width=\linewidth]{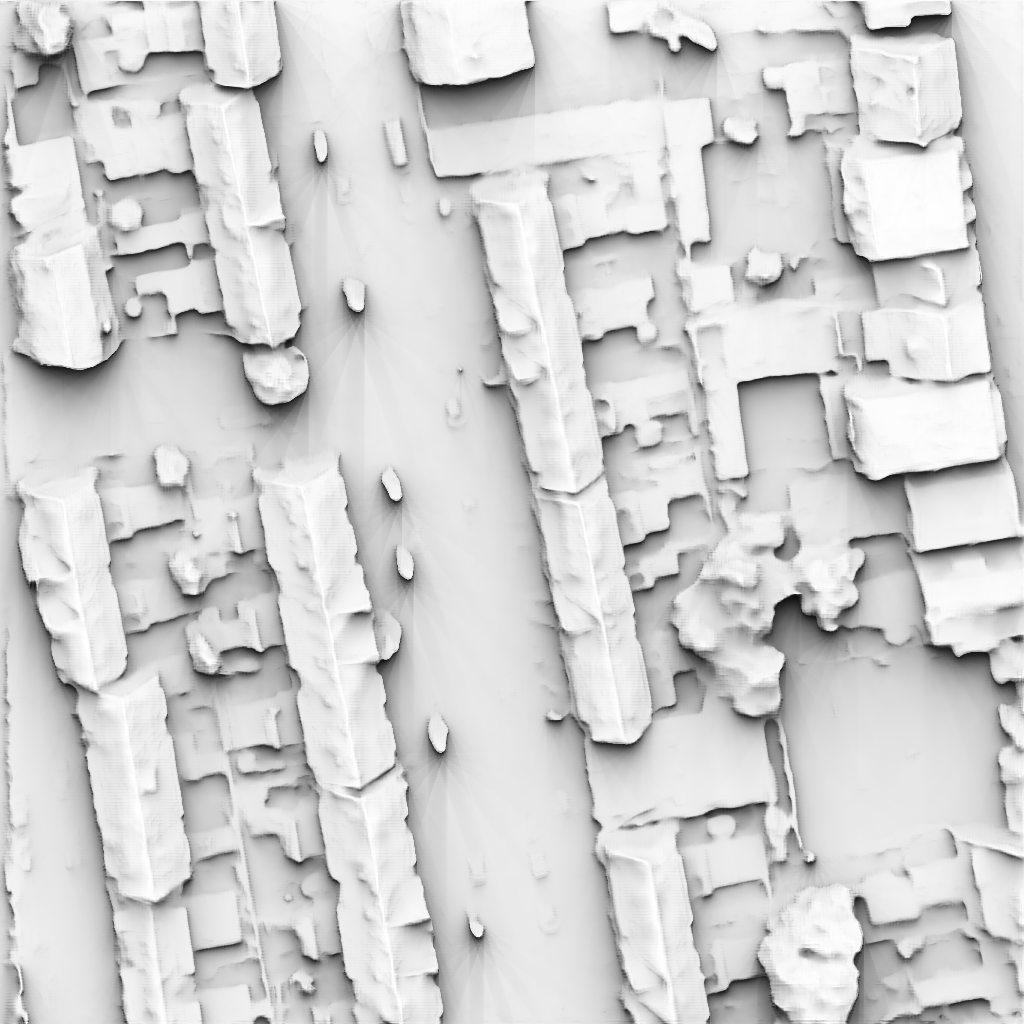}
		\centering{\tiny GANet}
	\end{minipage}
	\begin{minipage}[t]{0.19\textwidth}	
		\includegraphics[width=0.098\linewidth]{figures_supp/color_map.png}
		\includegraphics[width=0.85\linewidth]{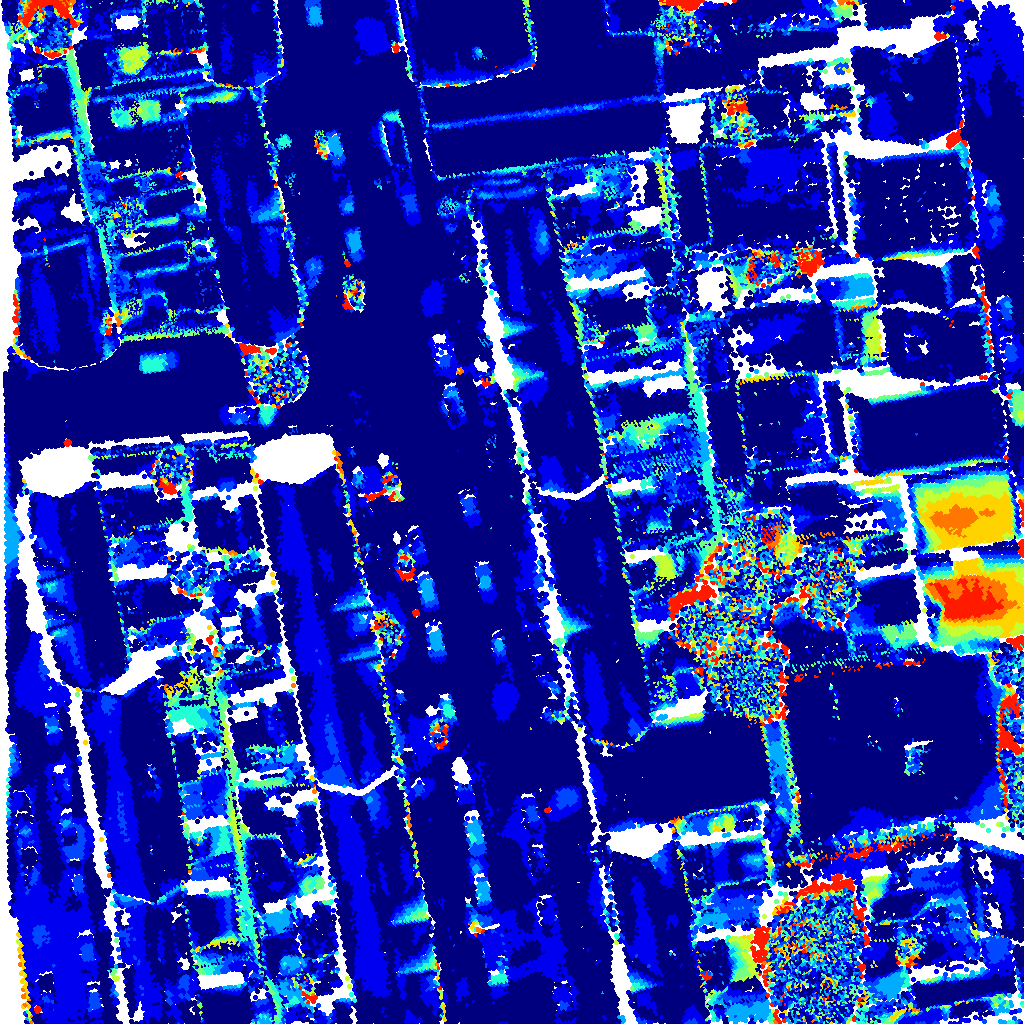}
		\includegraphics[width=\linewidth]{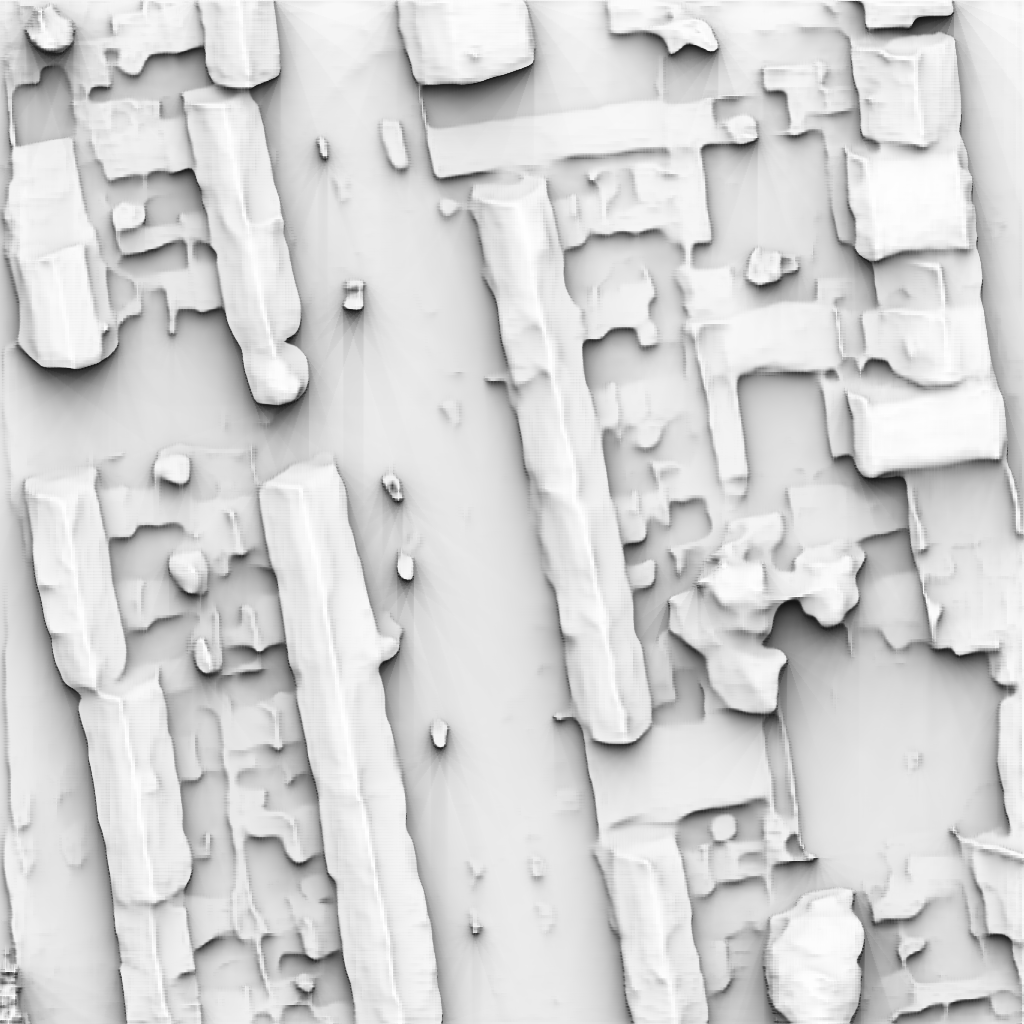}
		\centering{\tiny LEAStereo}
	\end{minipage}
	\caption{Error map and disparity visualization on building area for Enschede dataset.}
	\label{Figure.enschedebulding}
\end{figure}

\begin{figure}[tp]
	\begin{minipage}[t]{0.19\textwidth}
		\includegraphics[width=0.098\linewidth]{figures_supp/color_map.png}
		\includegraphics[width=0.85\linewidth]{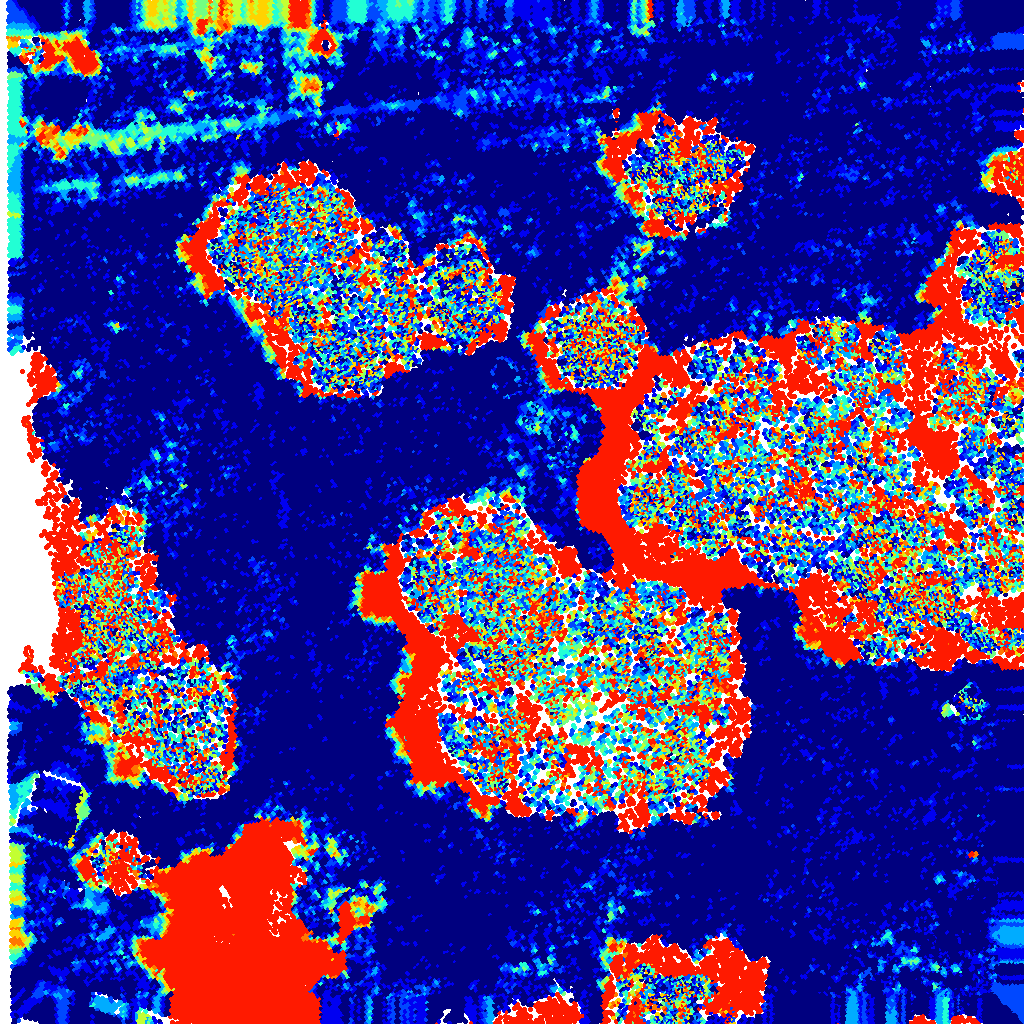}
		\includegraphics[width=\linewidth]{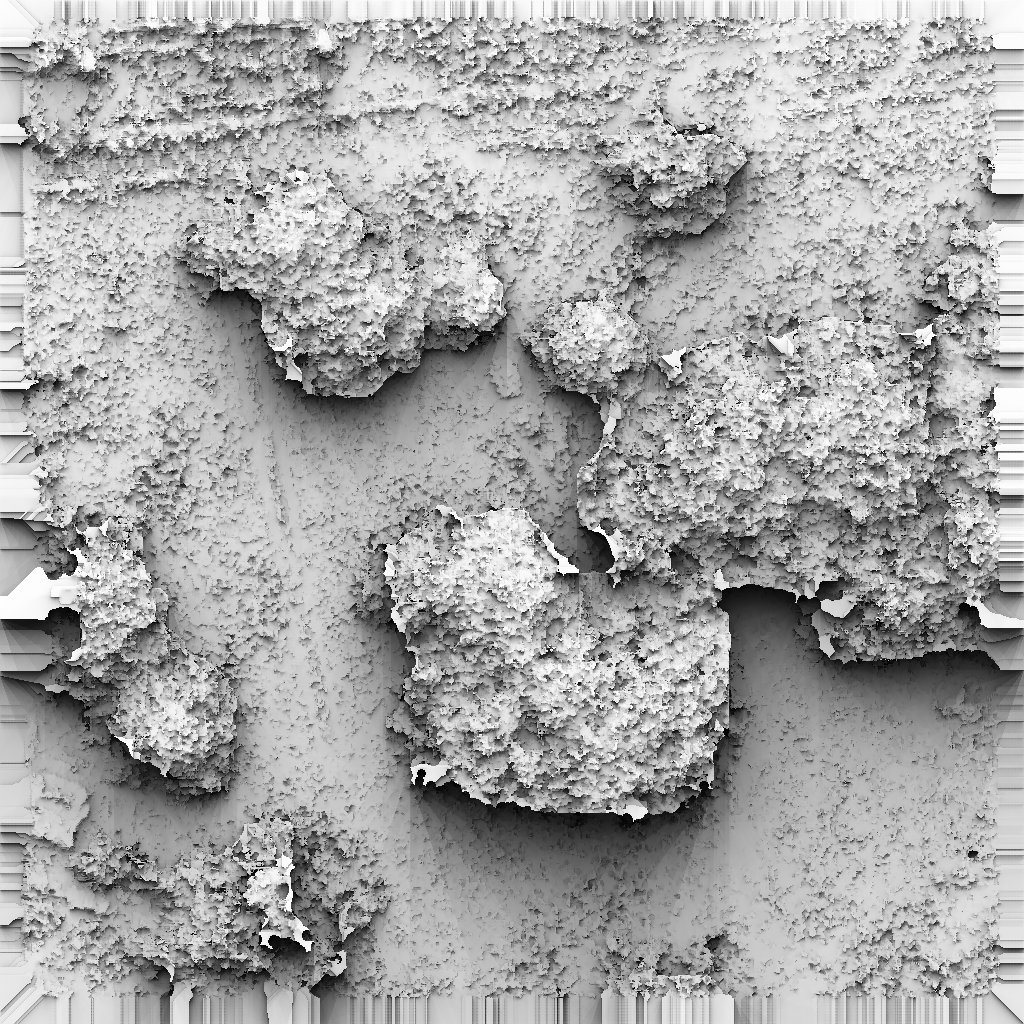}
		\centering{\tiny MICMAC}
	\end{minipage}
	\begin{minipage}[t]{0.19\textwidth}	
		\includegraphics[width=0.098\linewidth]{figures_supp/color_map.png}
		\includegraphics[width=0.85\linewidth]{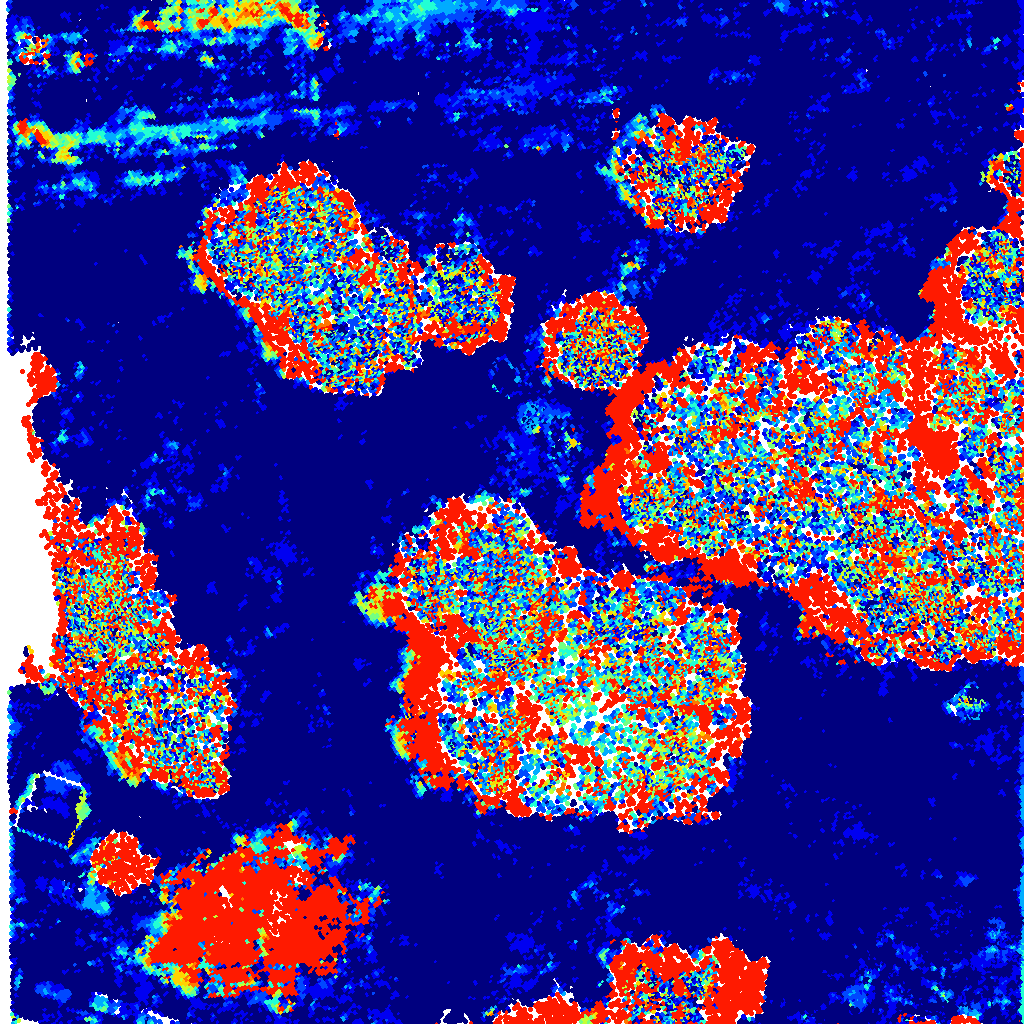}
		\includegraphics[width=\linewidth]{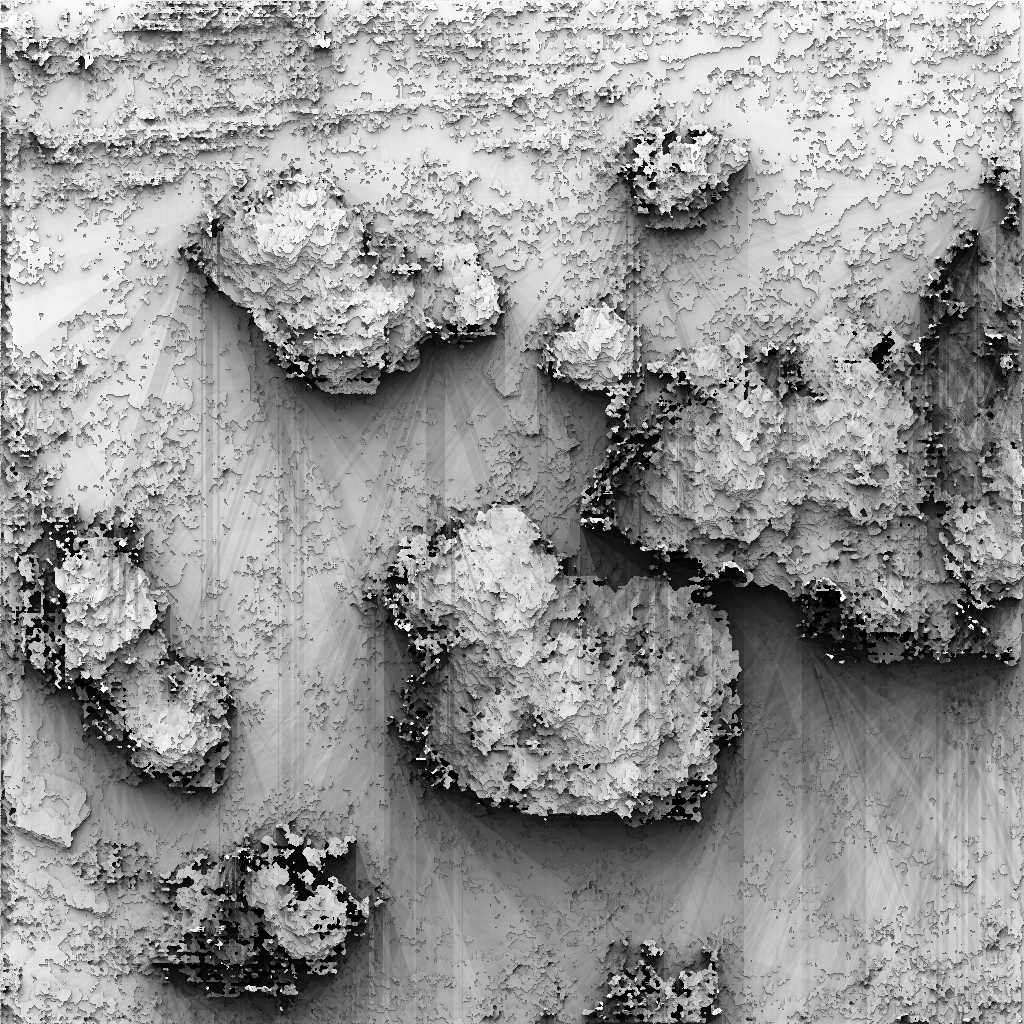}
		\centering{\tiny SGM(CUDA)}
	\end{minipage}
	\begin{minipage}[t]{0.19\textwidth}	
		\includegraphics[width=0.098\linewidth]{figures_supp/color_map.png}
		\includegraphics[width=0.85\linewidth]{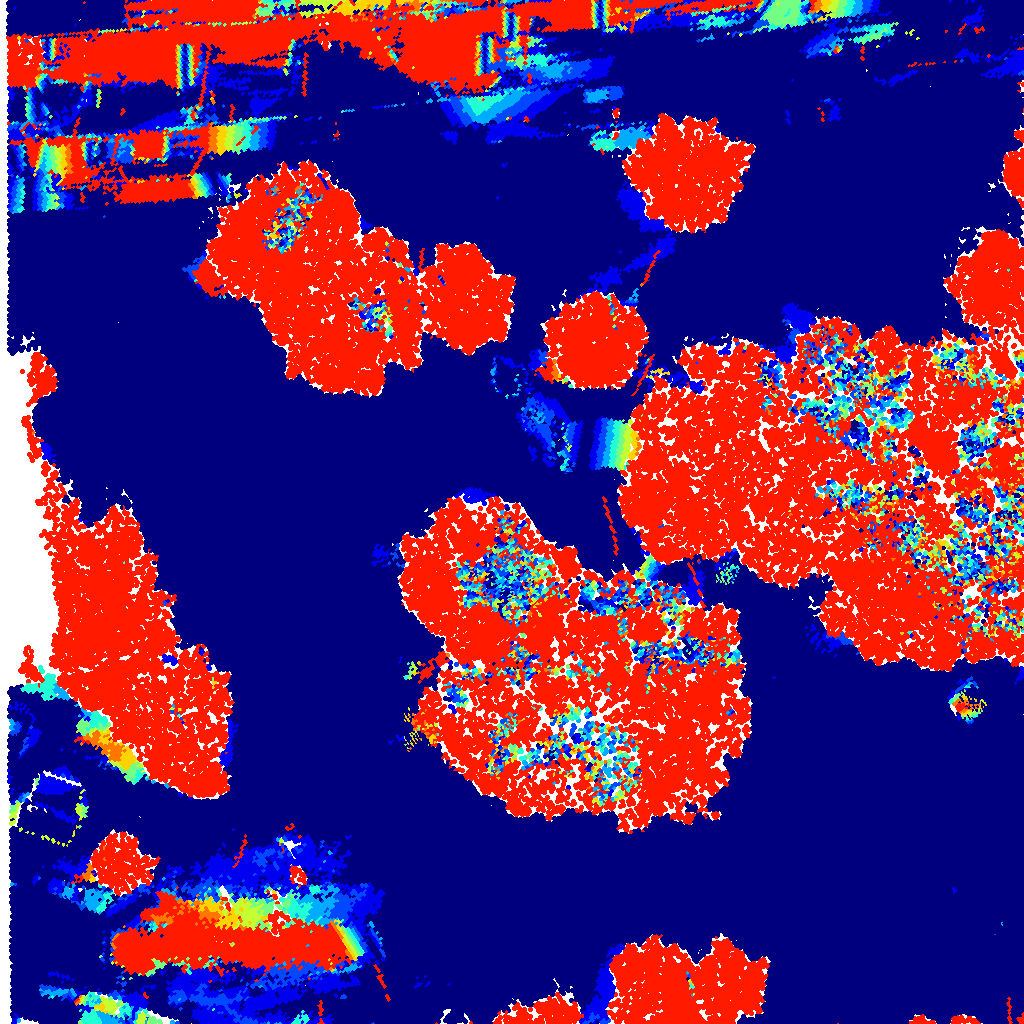}
		\includegraphics[width=\linewidth]{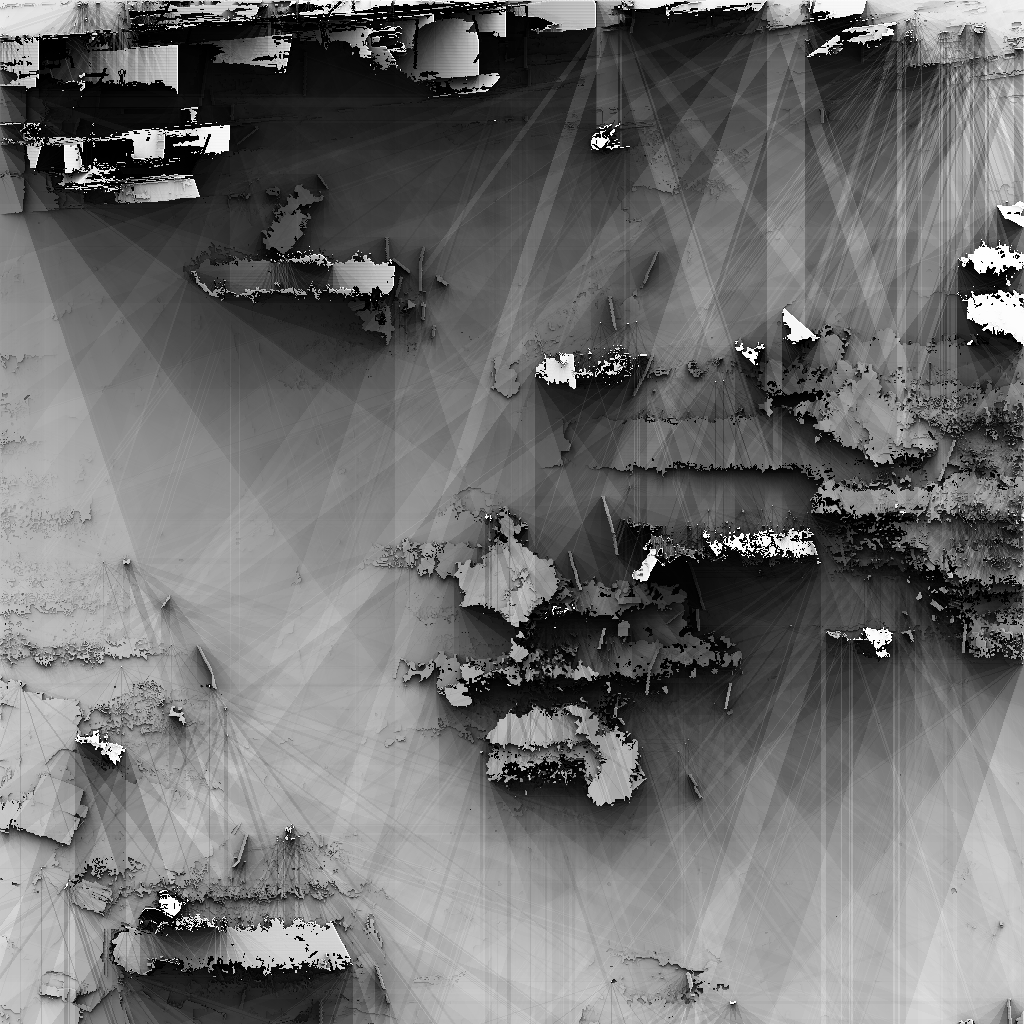}
		\centering{\tiny GraphCuts}
	\end{minipage}
	\begin{minipage}[t]{0.19\textwidth}	
		\includegraphics[width=0.098\linewidth]{figures_supp/color_map.png}
		\includegraphics[width=0.85\linewidth]{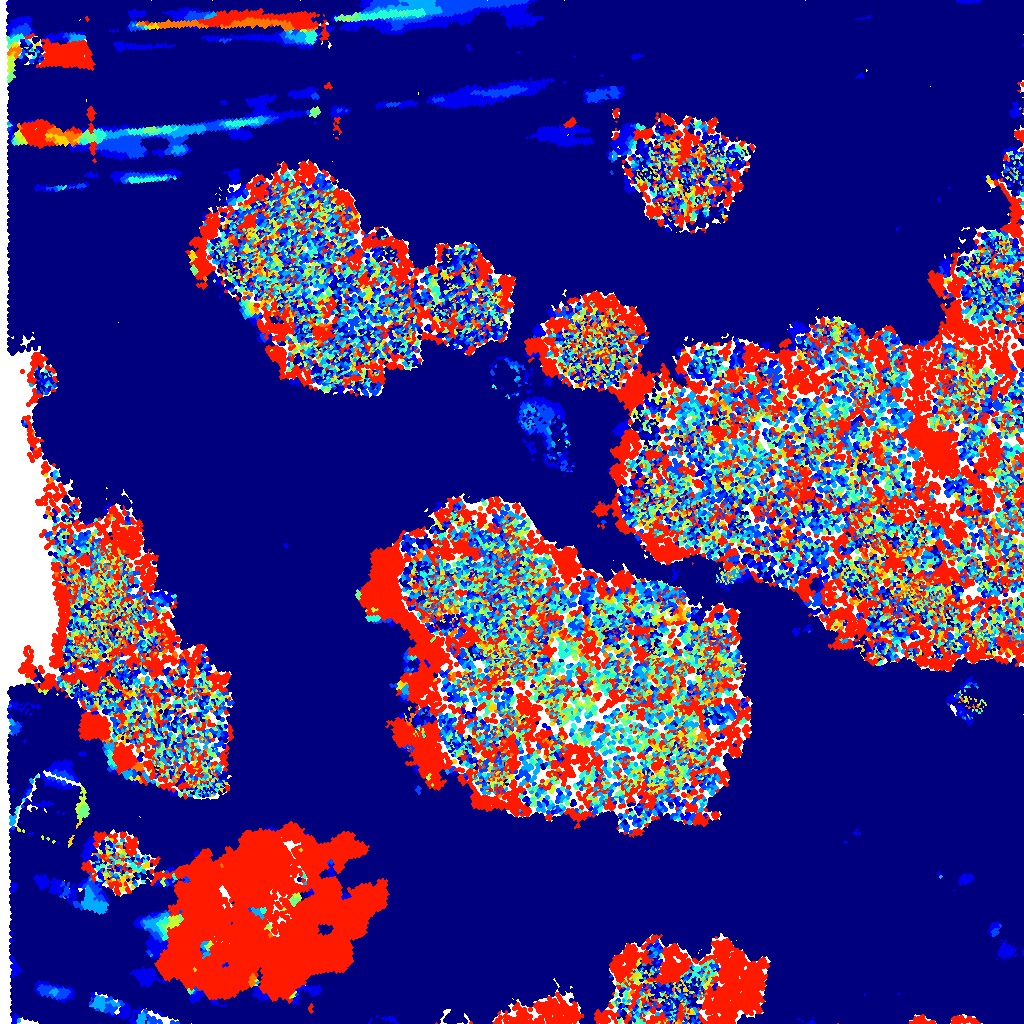}
		\includegraphics[width=\linewidth]{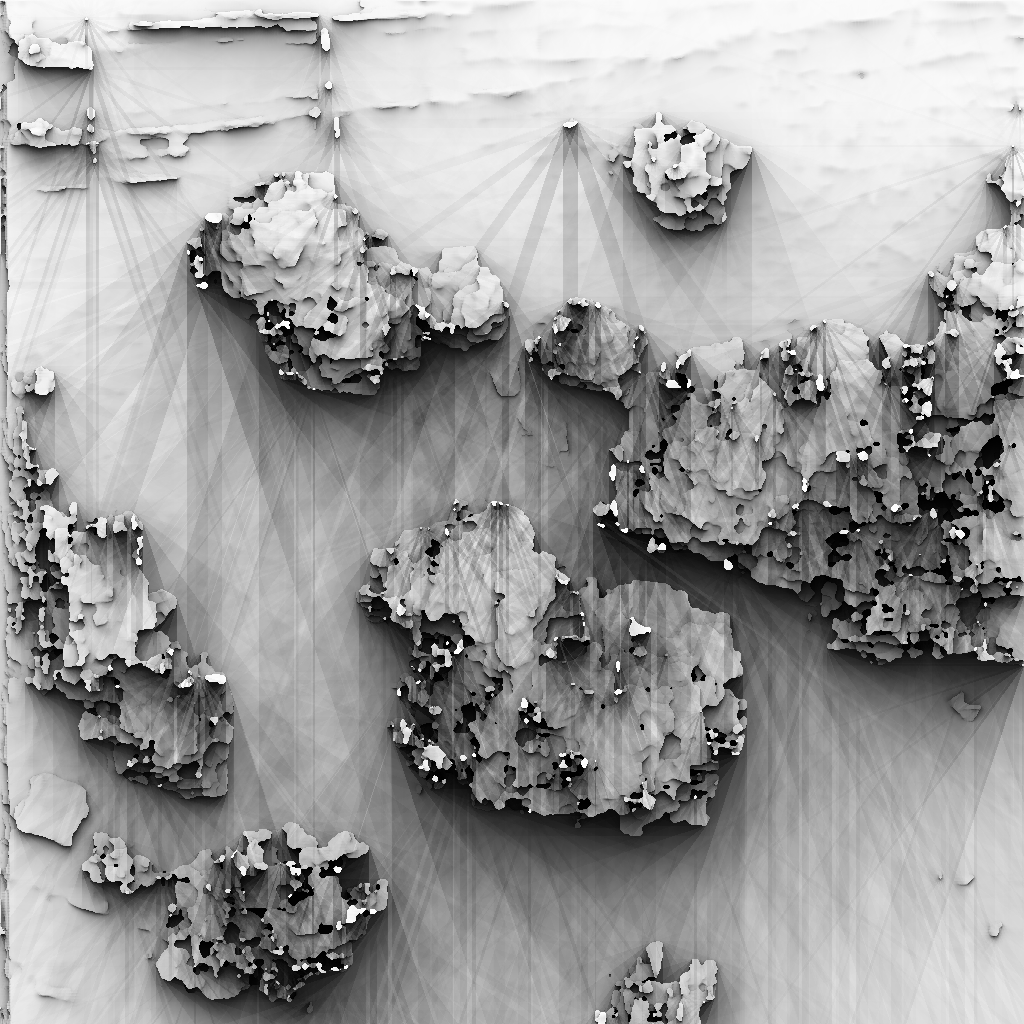}
		\centering{\tiny CBMV(SGM)}
	\end{minipage}
	\begin{minipage}[t]{0.19\textwidth}	
		\includegraphics[width=0.098\linewidth]{figures_supp/color_map.png}
		\includegraphics[width=0.85\linewidth]{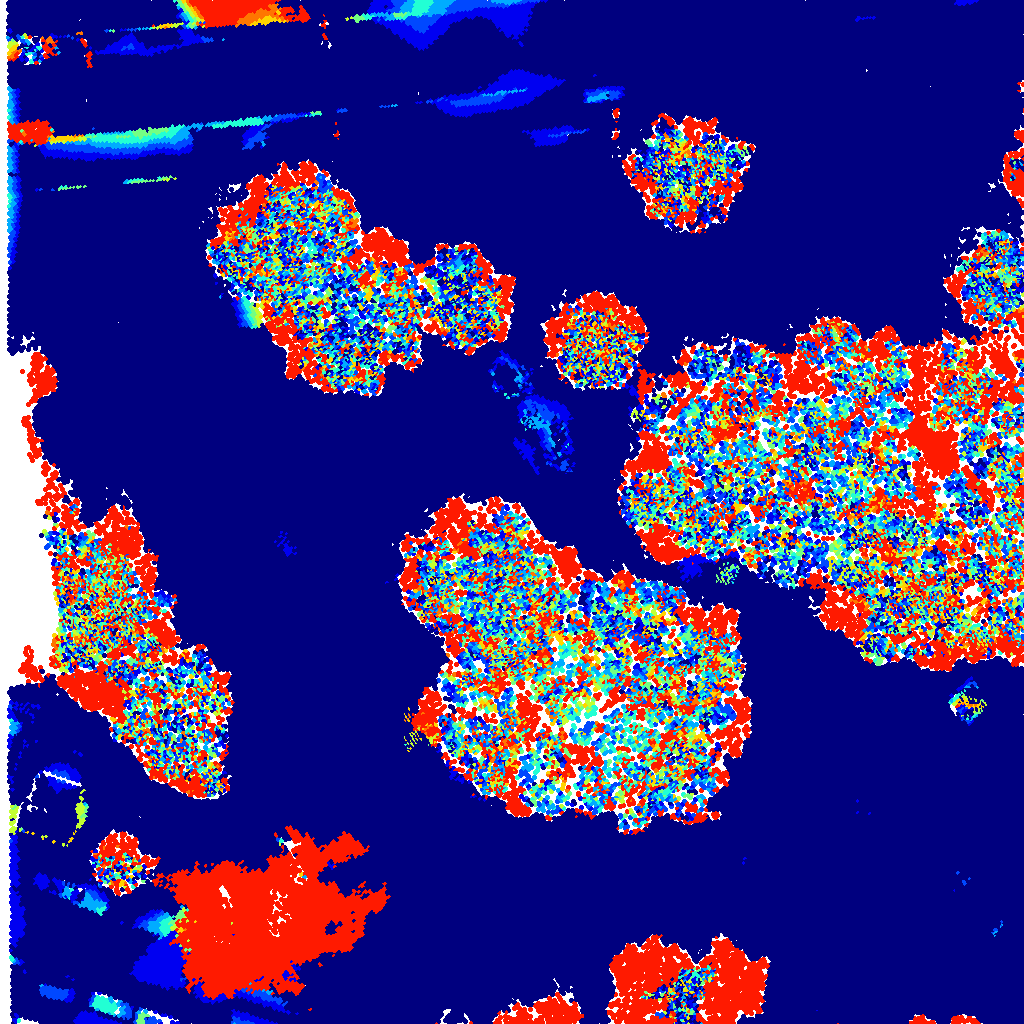}
		\includegraphics[width=\linewidth]{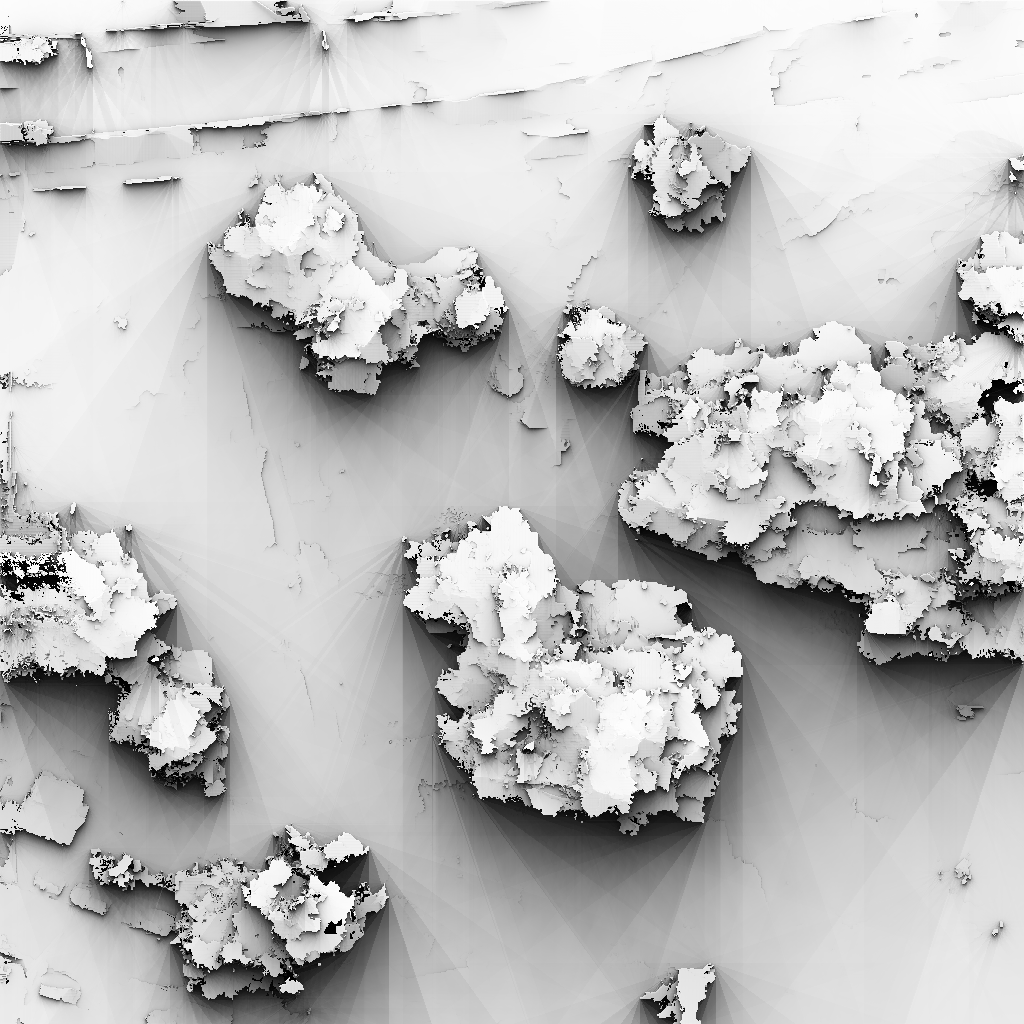}
		\centering{\tiny CBMV(GraphCuts)}
	\end{minipage}
	\begin{minipage}[t]{0.19\textwidth}	
		\includegraphics[width=0.098\linewidth]{figures_supp/color_map.png}
		\includegraphics[width=0.85\linewidth]{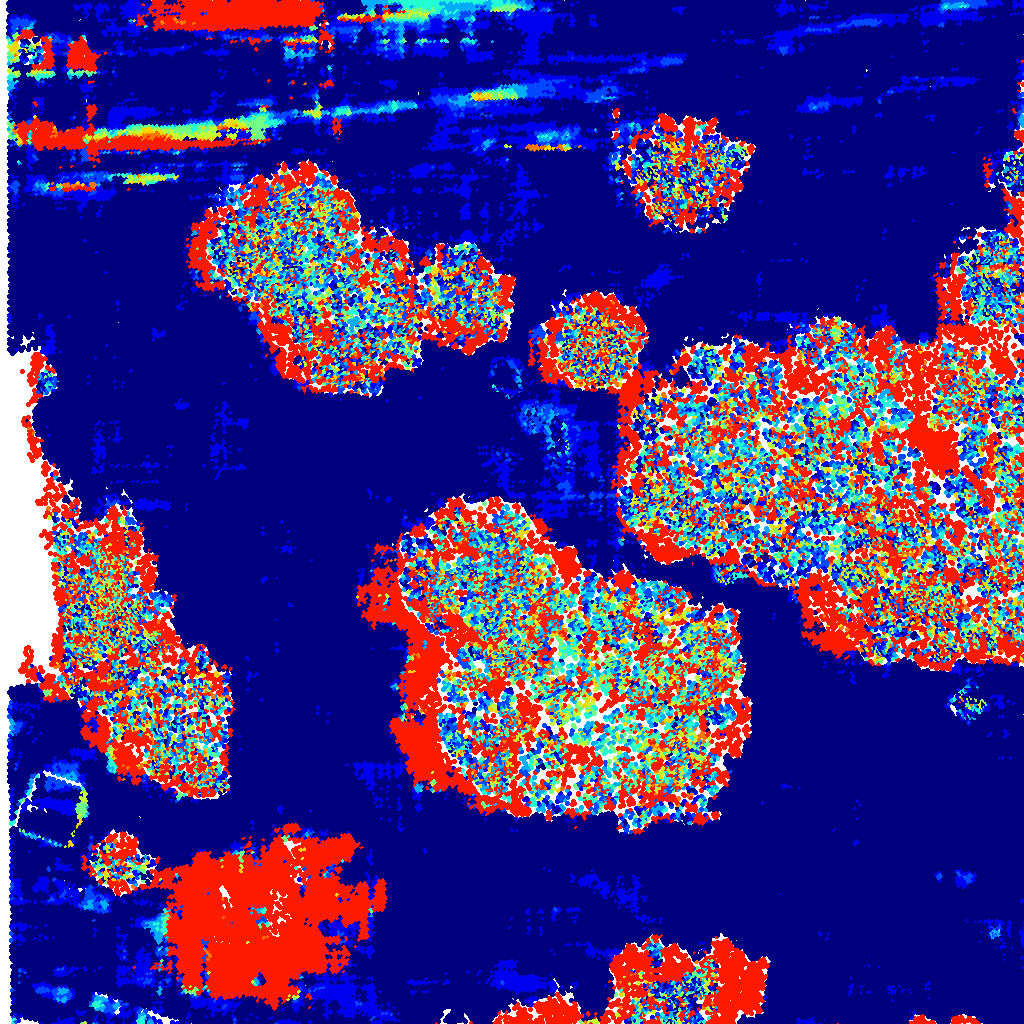}
		\includegraphics[width=\linewidth]{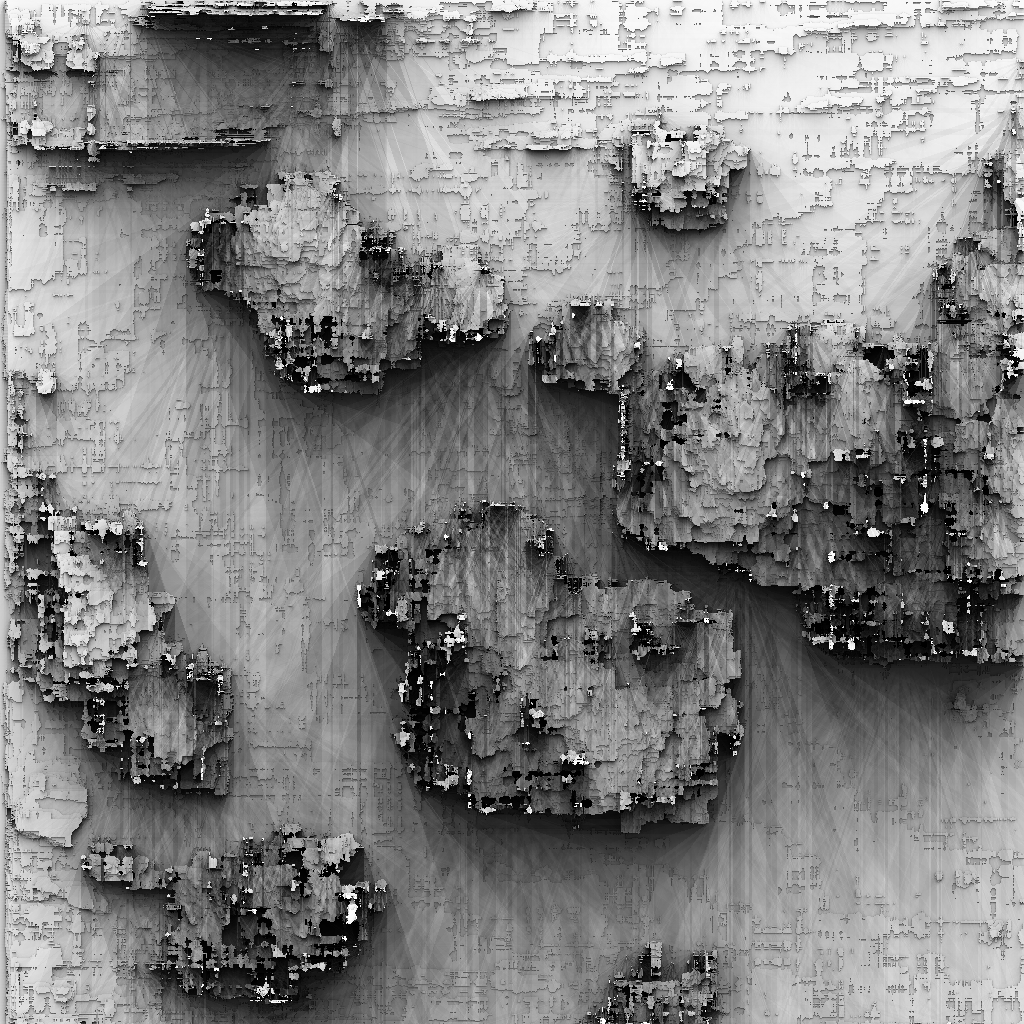}
		\centering{\tiny MC-CNN(KITTI)}
	\end{minipage}
	\begin{minipage}[t]{0.19\textwidth}	
		\includegraphics[width=0.098\linewidth]{figures_supp/color_map.png}
		\includegraphics[width=0.85\linewidth]{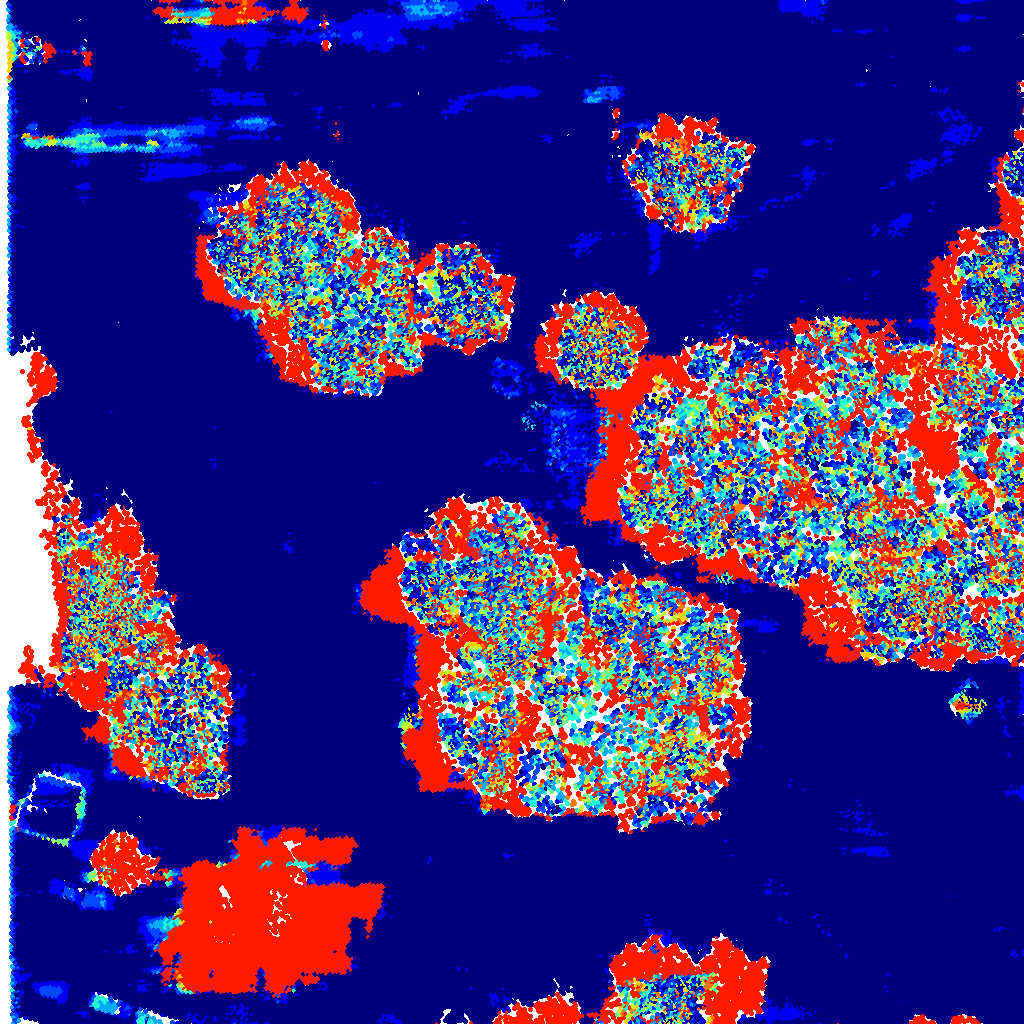}
		\includegraphics[width=\linewidth]{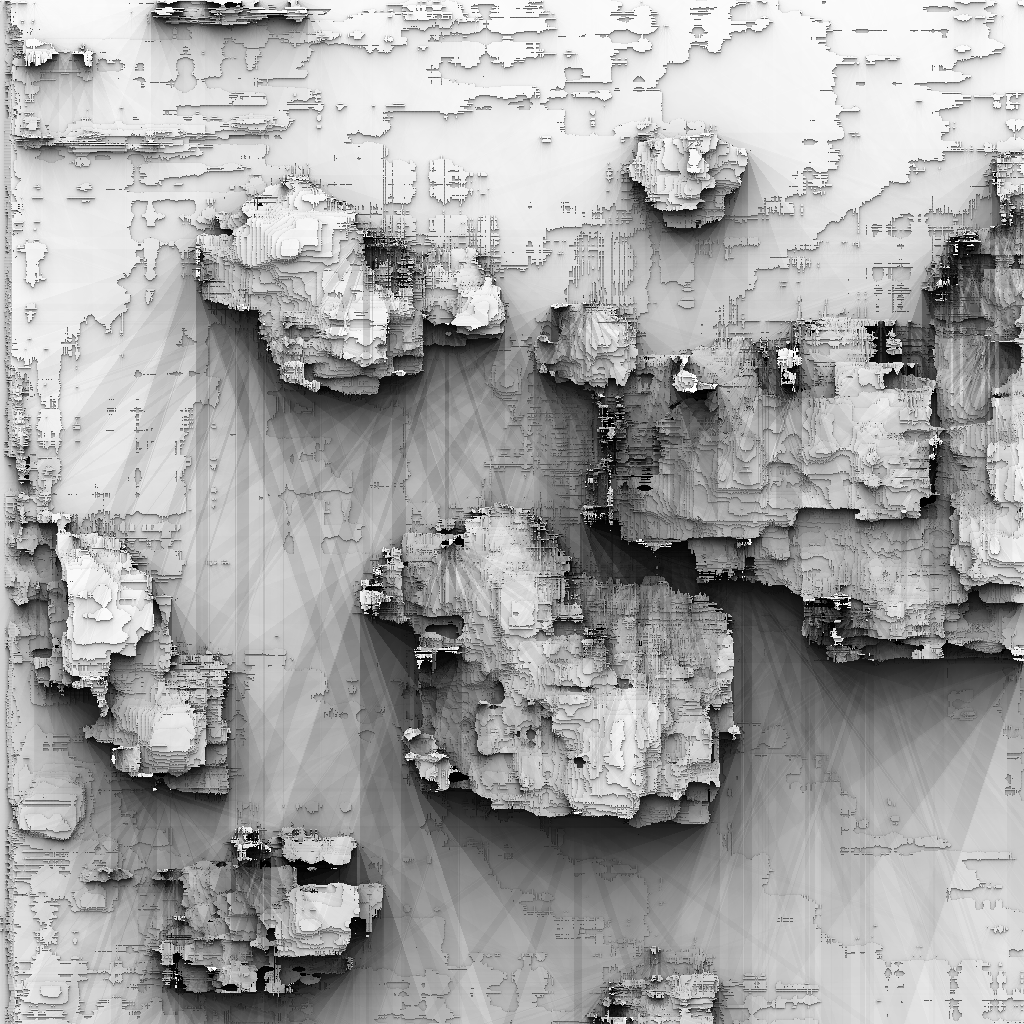}
		\centering{\tiny DeepFeature(KITTI)}
	\end{minipage}
	\begin{minipage}[t]{0.19\textwidth}	
		\includegraphics[width=0.098\linewidth]{figures_supp/color_map.png}
		\includegraphics[width=0.85\linewidth]{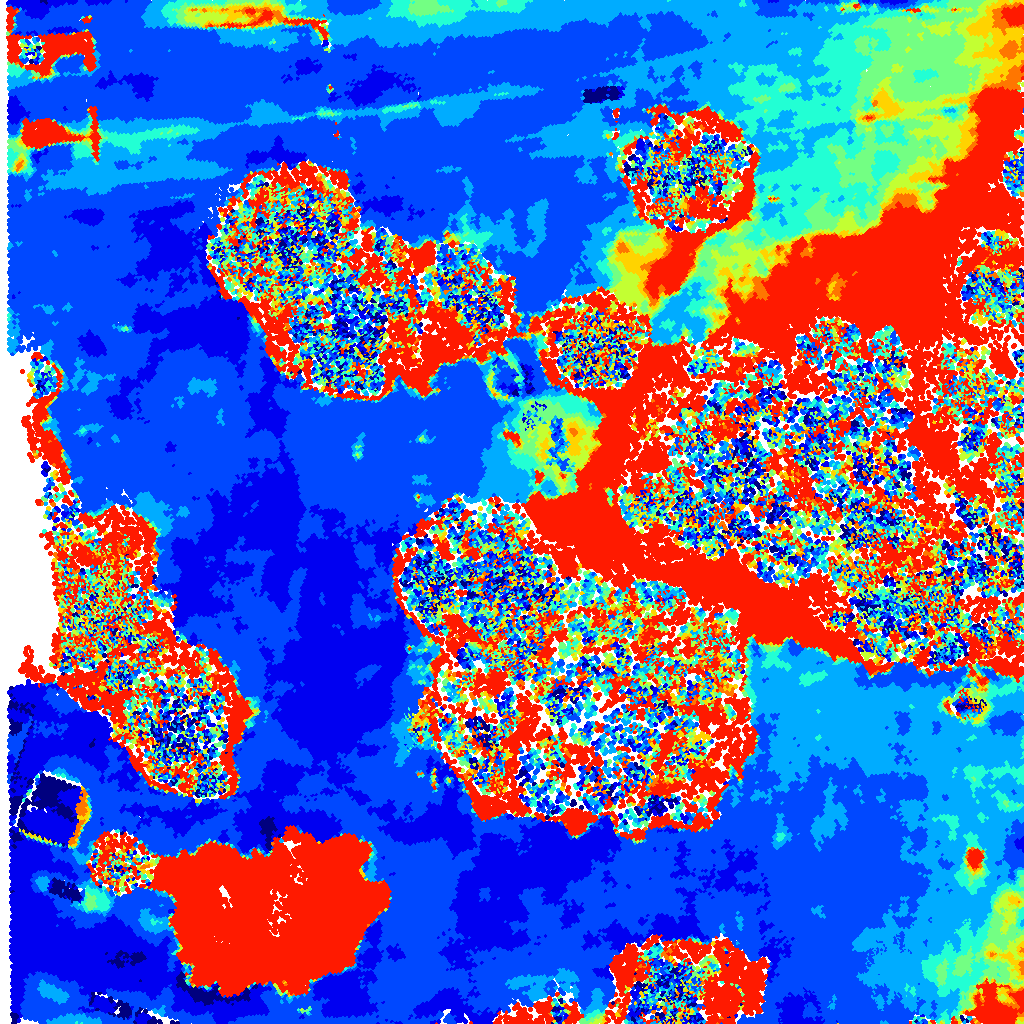}
		\includegraphics[width=\linewidth]{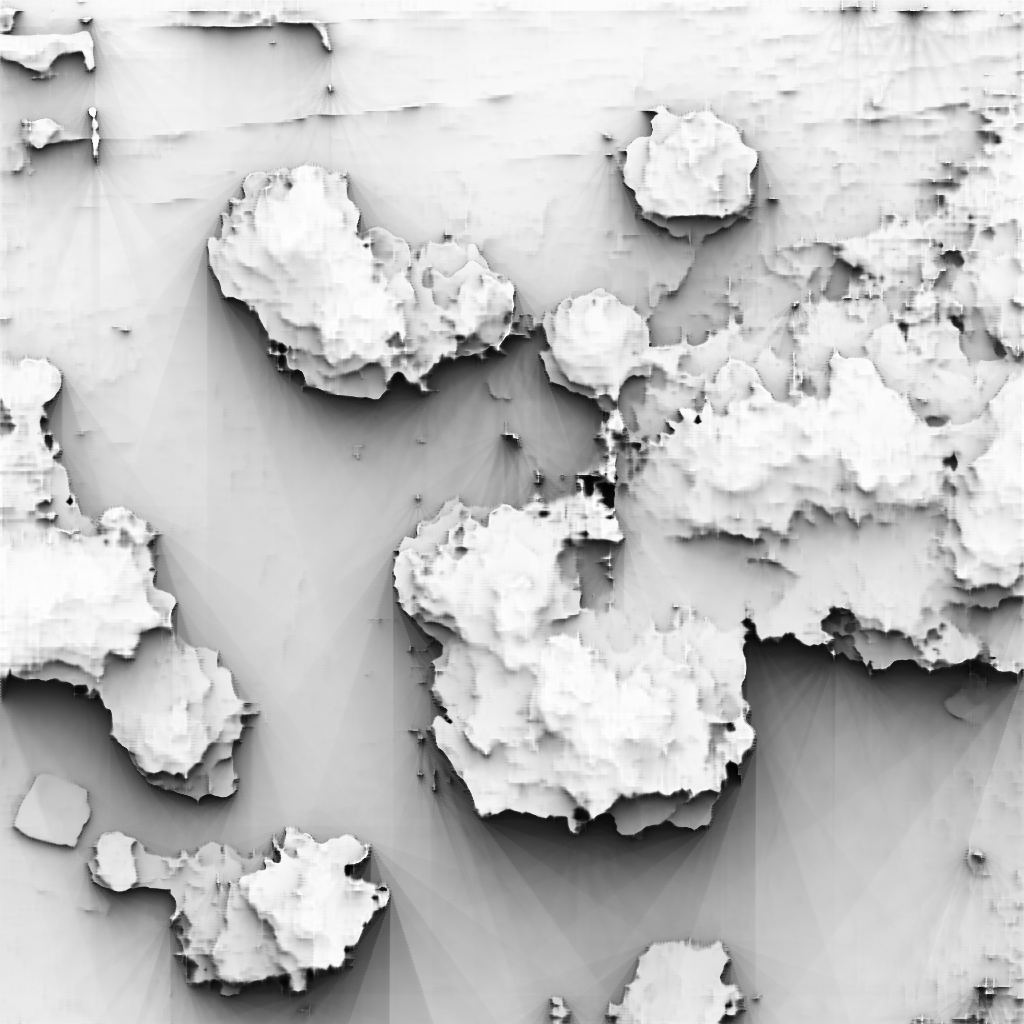}
		\centering{\tiny PSM net(KITTI)}
	\end{minipage}
	\begin{minipage}[t]{0.19\textwidth}	
		\includegraphics[width=0.098\linewidth]{figures_supp/color_map.png}
		\includegraphics[width=0.85\linewidth]{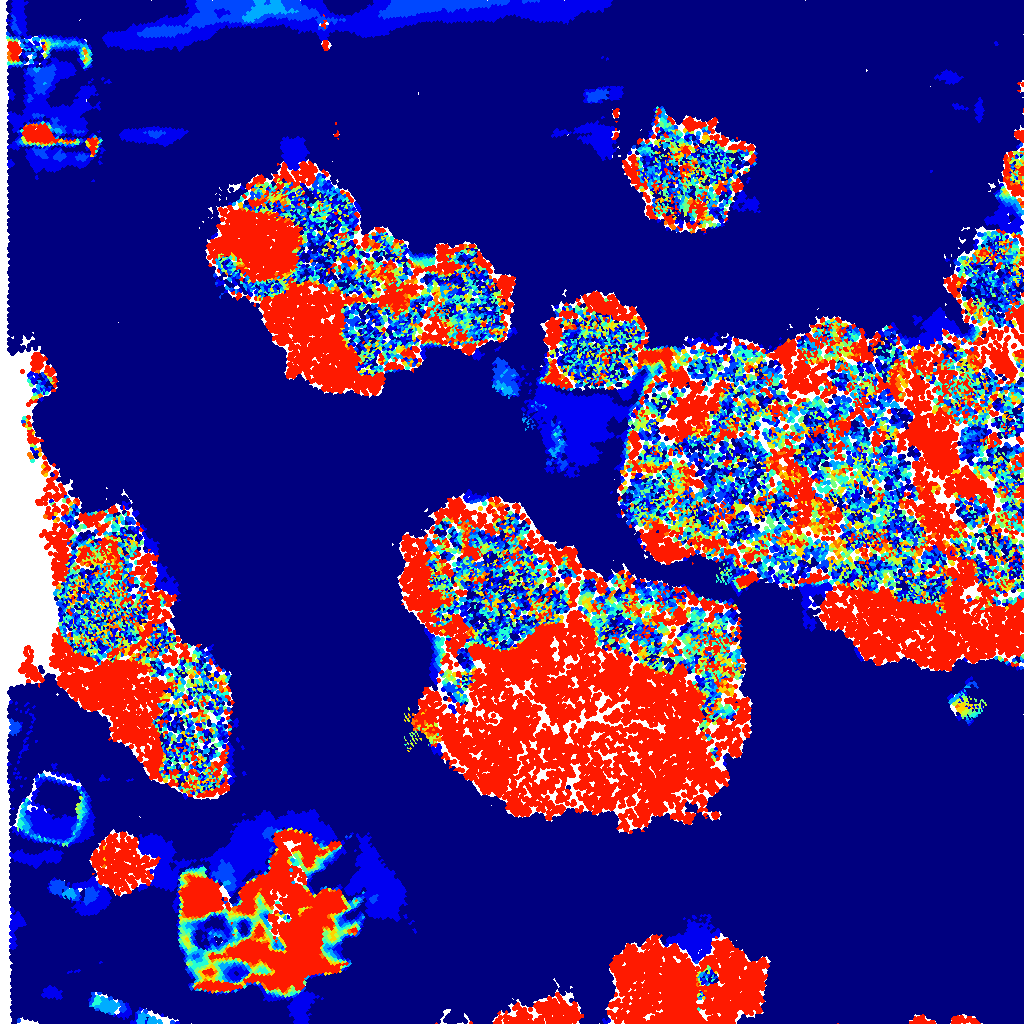}
		\includegraphics[width=\linewidth]{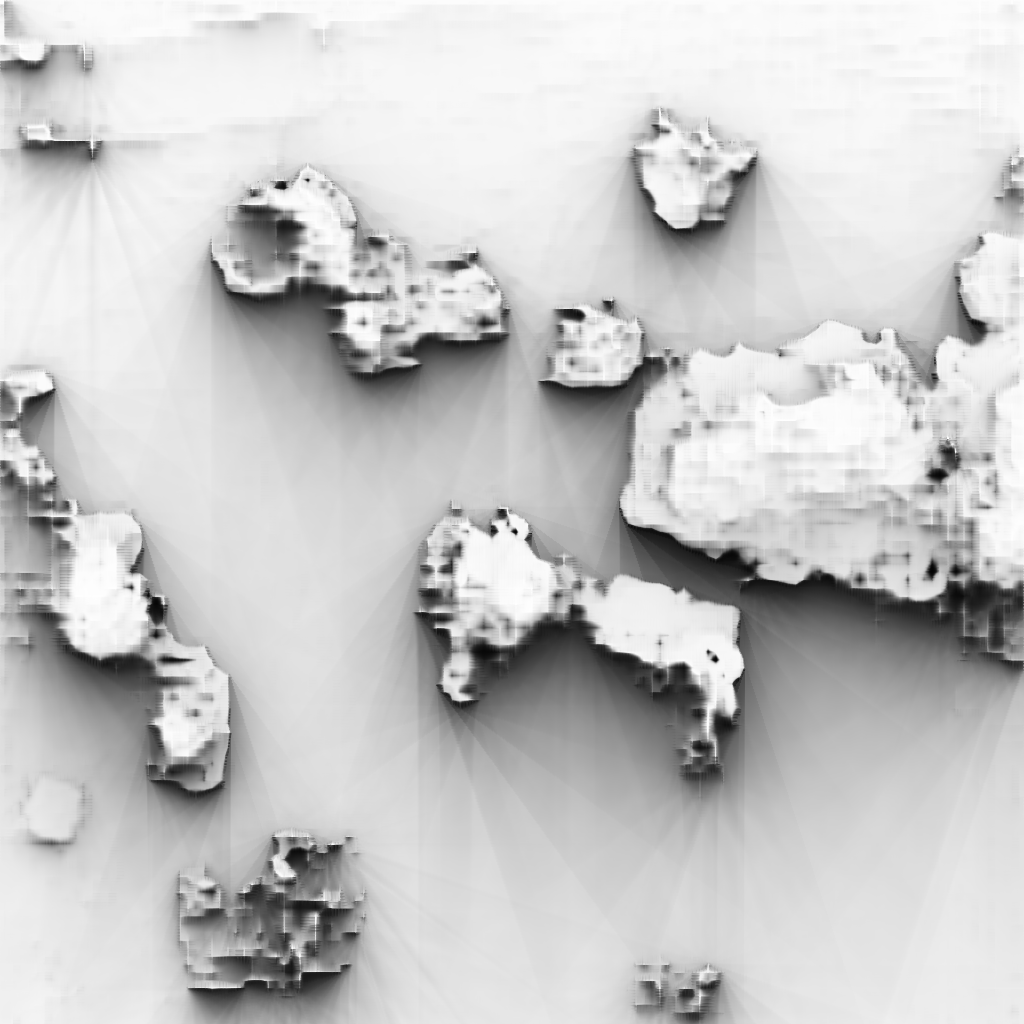}
		\centering{\tiny HRS net(KITTI)}
	\end{minipage}
	\begin{minipage}[t]{0.19\textwidth}	
		\includegraphics[width=0.098\linewidth]{figures_supp/color_map.png}
		\includegraphics[width=0.85\linewidth]{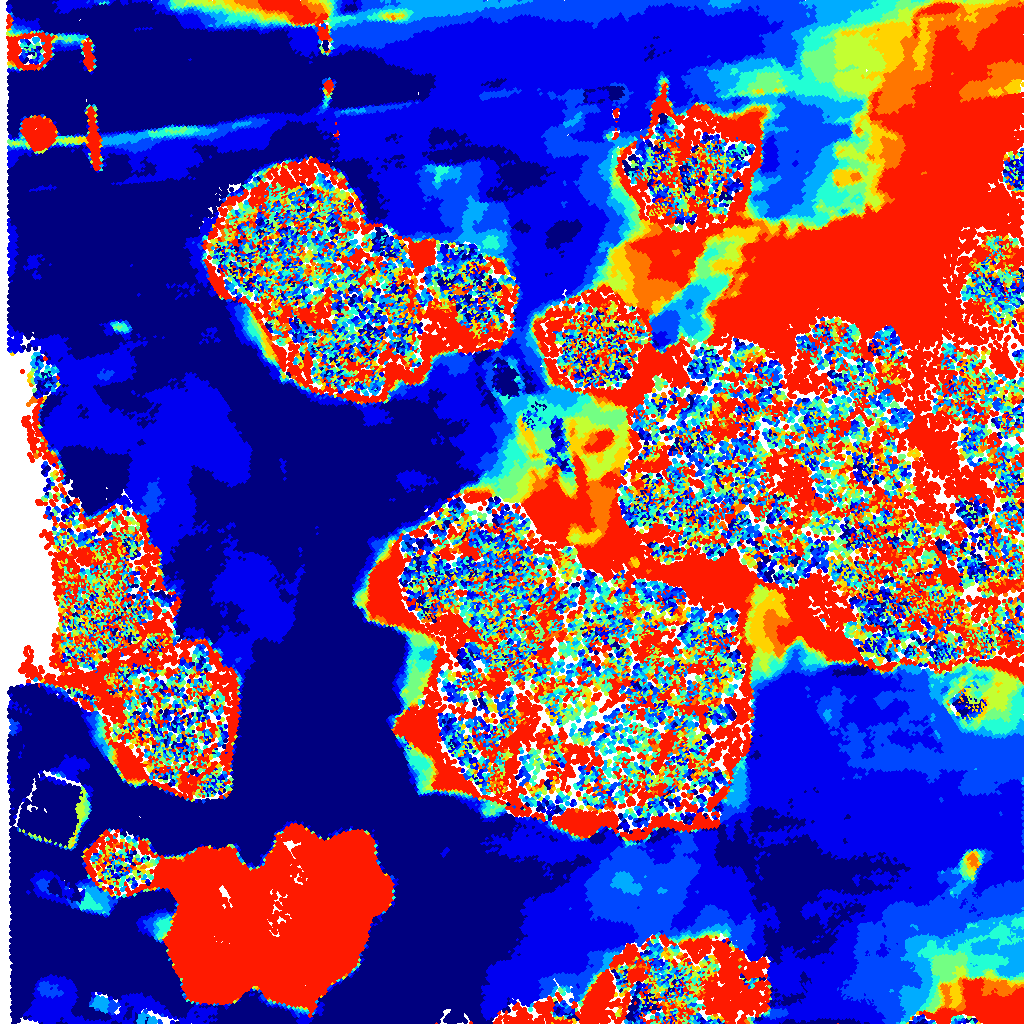}
		\includegraphics[width=\linewidth]{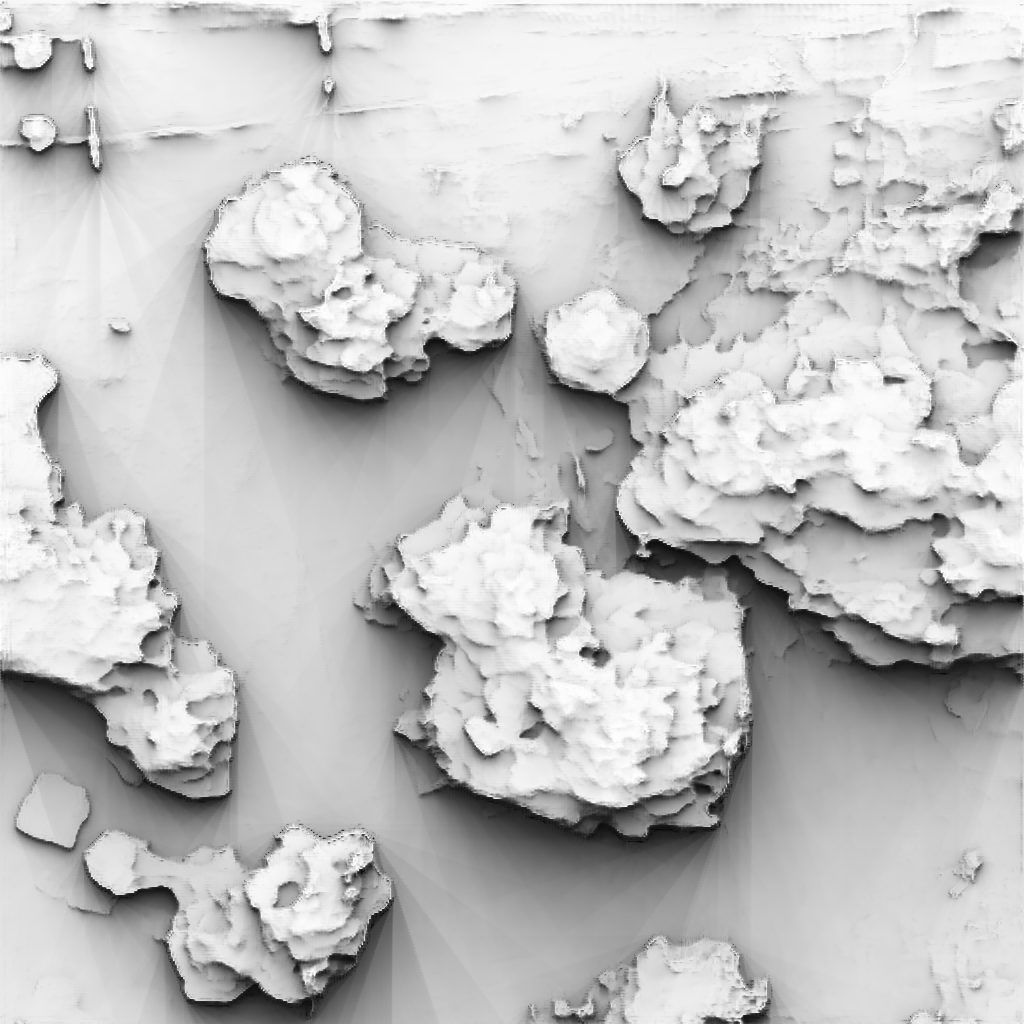}
		\centering{\tiny DeepPruner(KITTI)}
	\end{minipage}
	\begin{minipage}[t]{0.19\textwidth}
		\includegraphics[width=0.098\linewidth]{figures_supp/color_map.png}
		\includegraphics[width=0.85\linewidth]{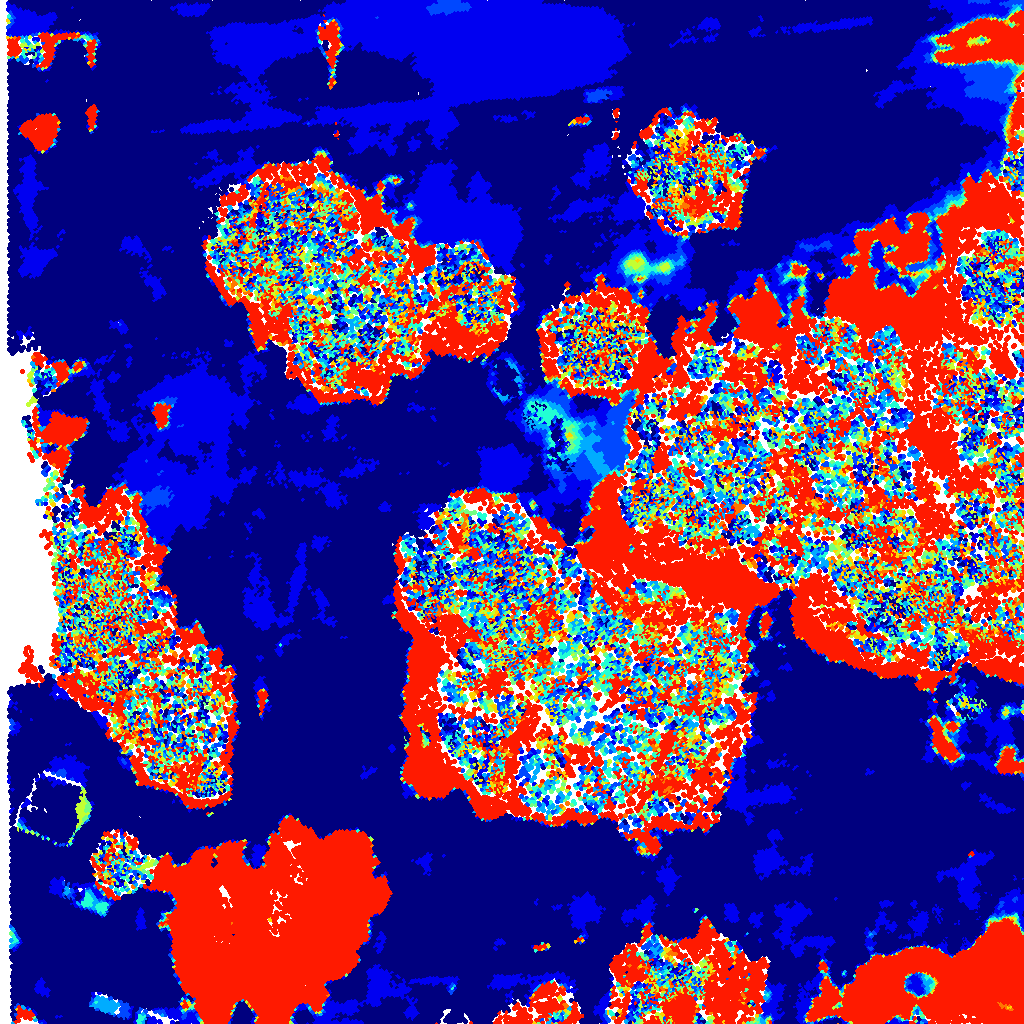}
		\includegraphics[width=\linewidth]{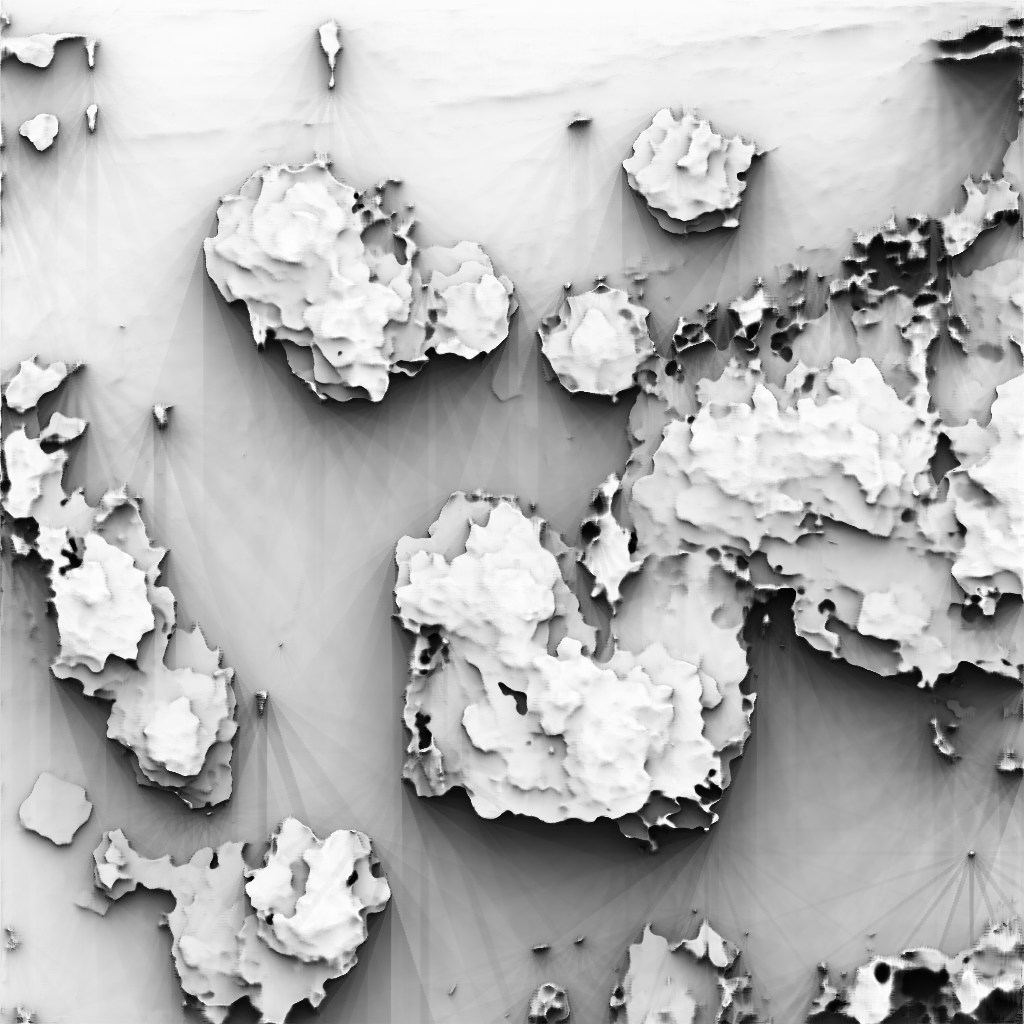}
		\centering{\tiny GANet(KITTI)}
	\end{minipage}
	\begin{minipage}[t]{0.19\textwidth}	
		\includegraphics[width=0.098\linewidth]{figures_supp/color_map.png}
		\includegraphics[width=0.85\linewidth]{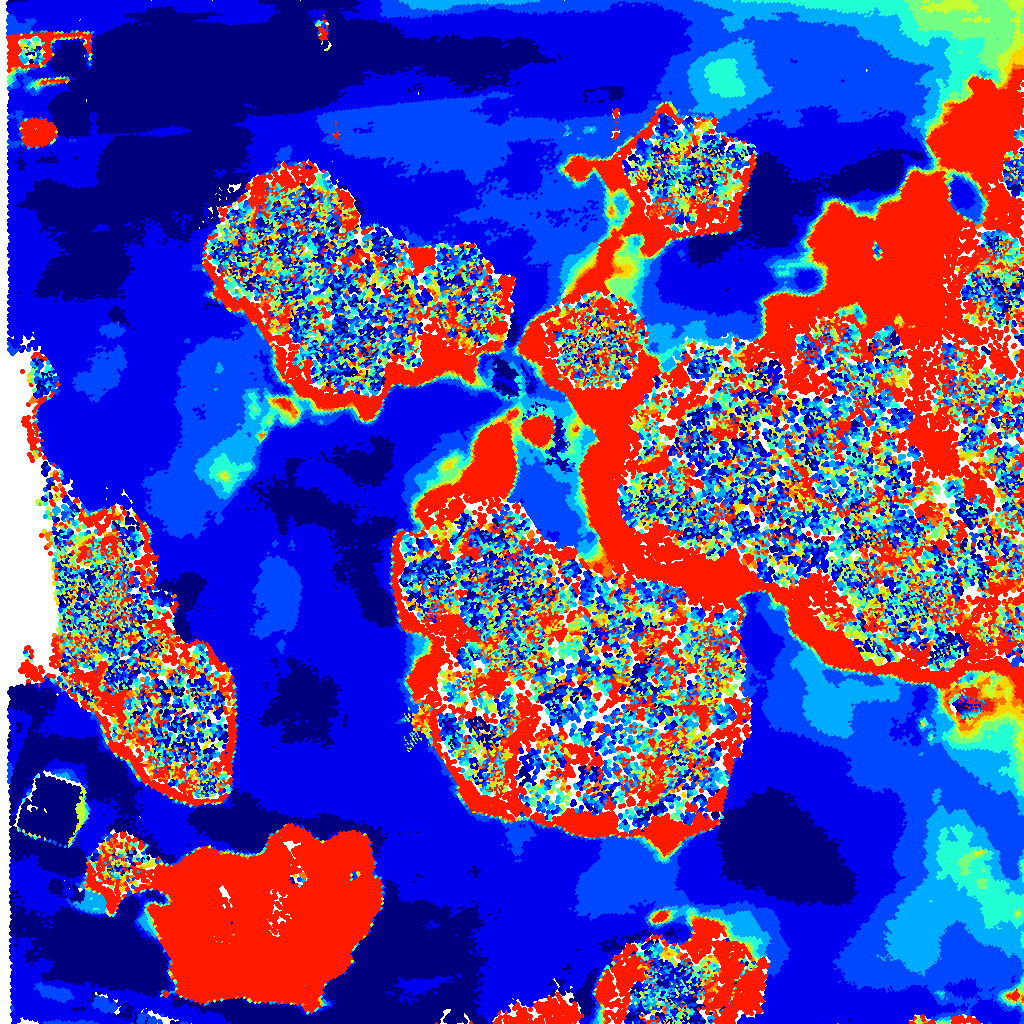}
		\includegraphics[width=\linewidth]{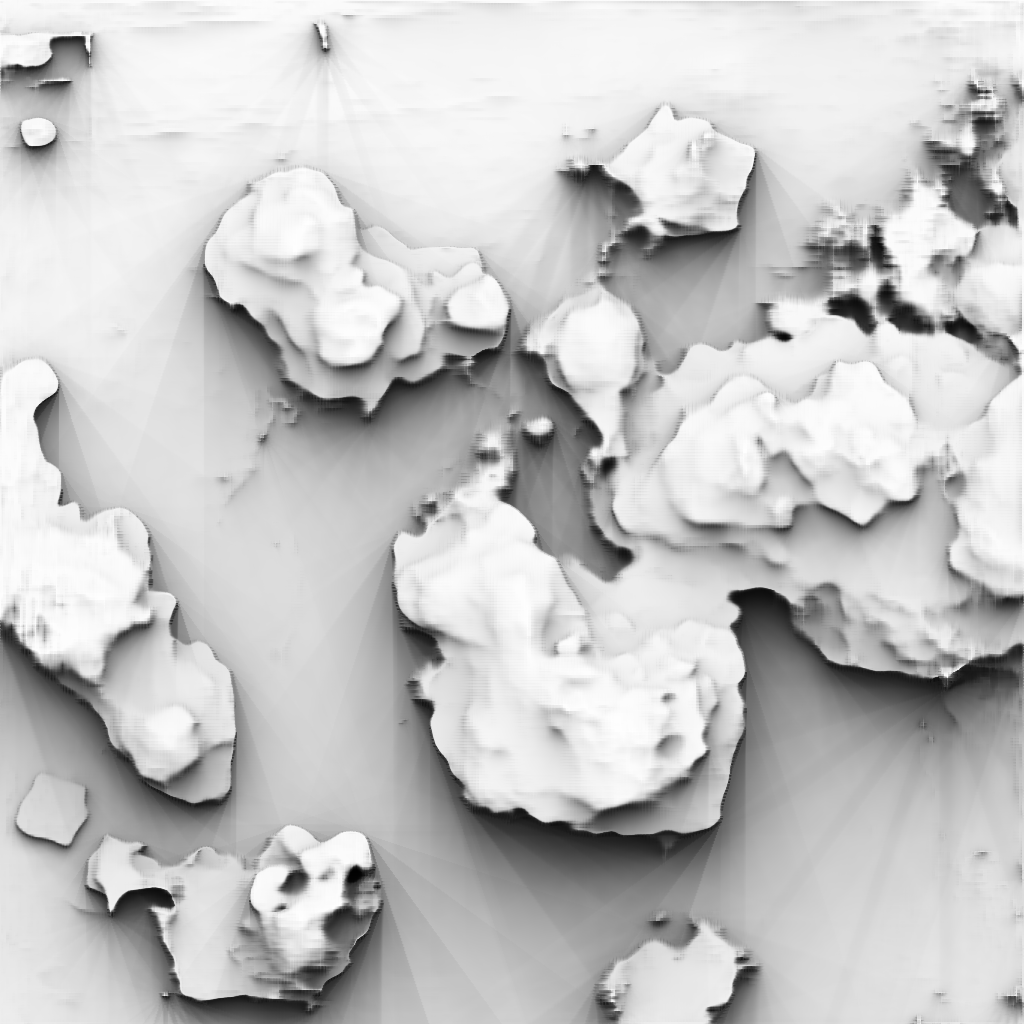}
		\centering{\tiny LEAStereo(KITTI)}
	\end{minipage}
	\begin{minipage}[t]{0.19\textwidth}	
		\includegraphics[width=0.098\linewidth]{figures_supp/color_map.png}
		\includegraphics[width=0.85\linewidth]{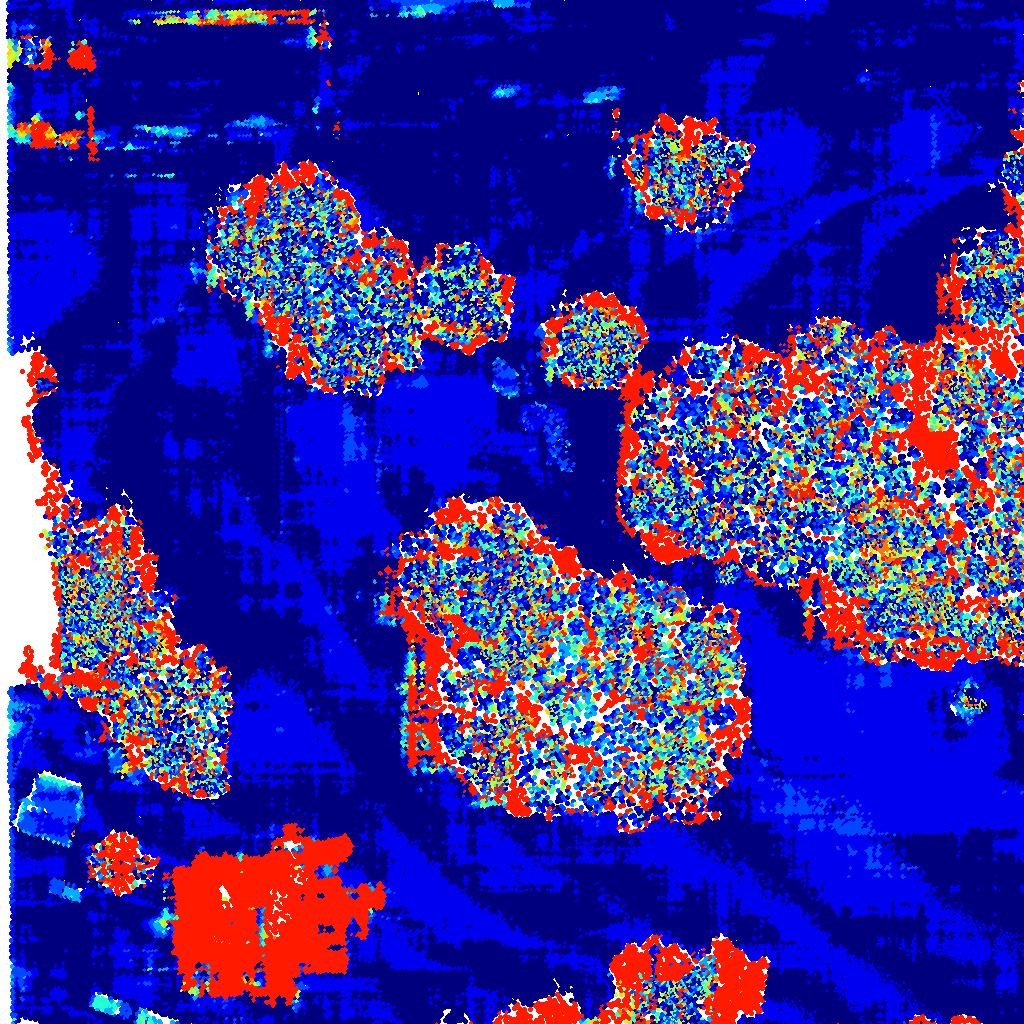}
		\includegraphics[width=\linewidth]{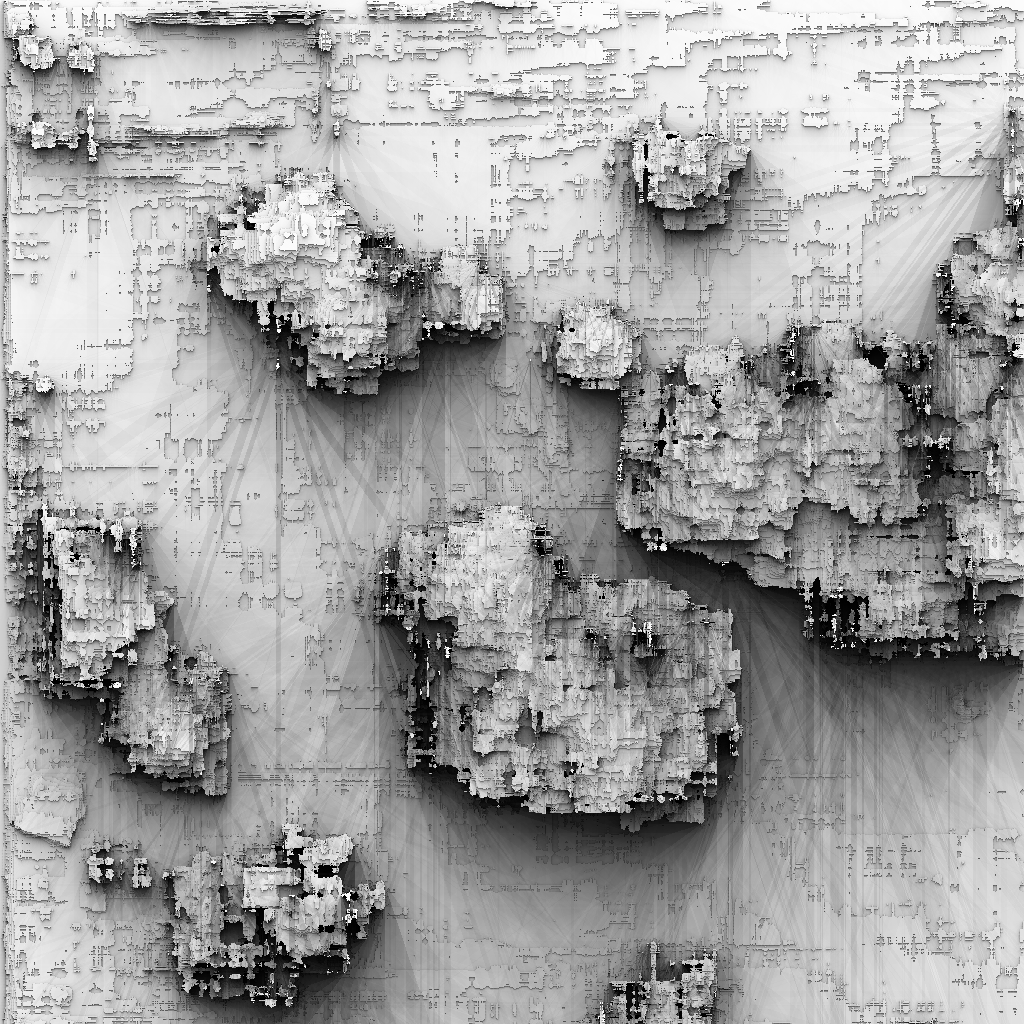}
		\centering{\tiny MC-CNN}
	\end{minipage}
	\begin{minipage}[t]{0.19\textwidth}	
		\includegraphics[width=0.098\linewidth]{figures_supp/color_map.png}
		\includegraphics[width=0.85\linewidth]{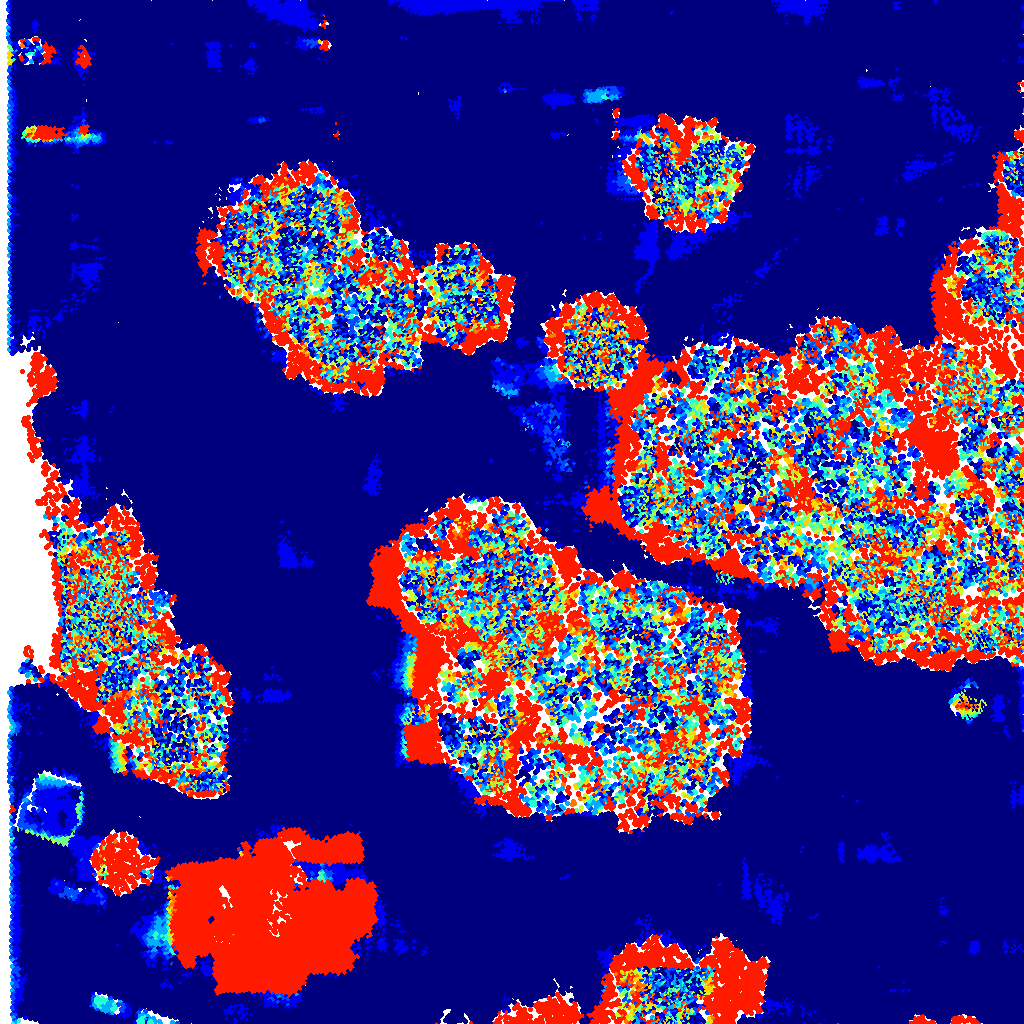}
		\includegraphics[width=\linewidth]{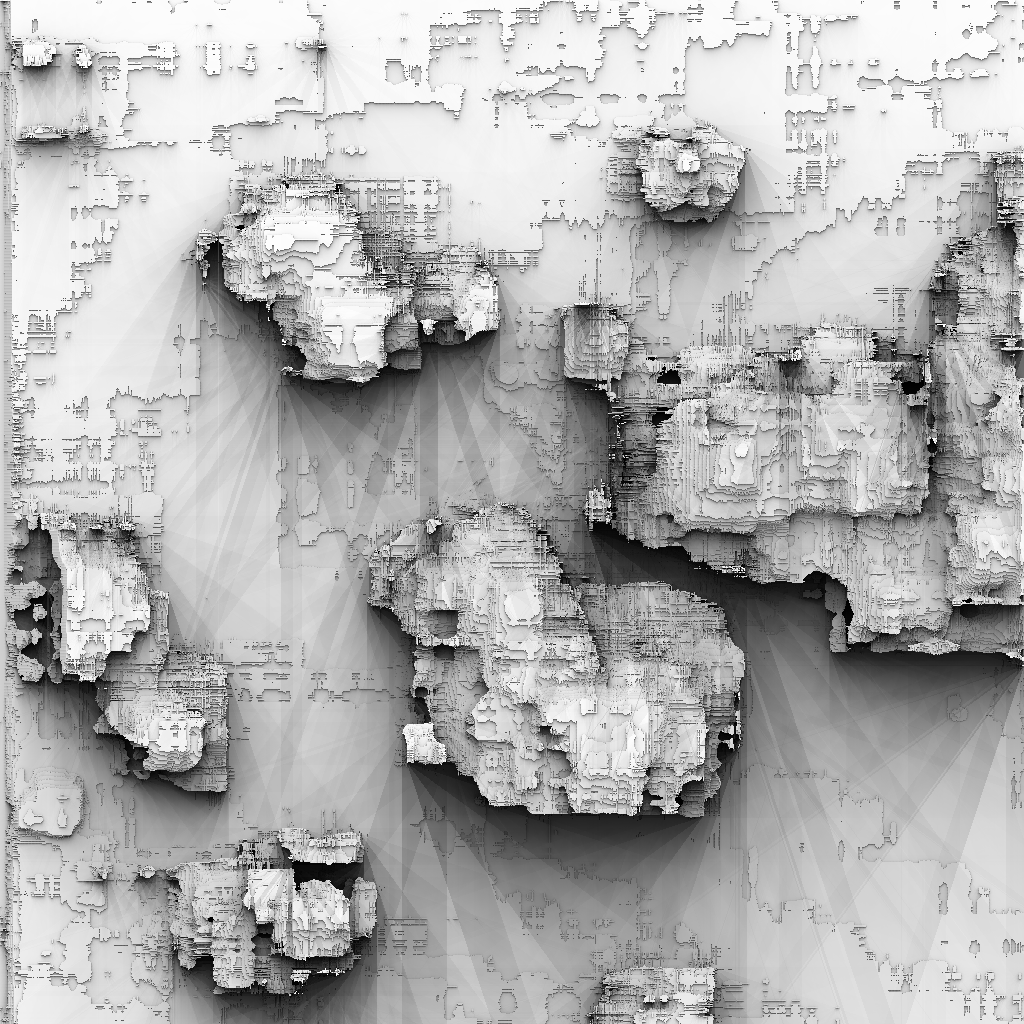}
		\centering{\tiny DeepFeature}
	\end{minipage}
	\begin{minipage}[t]{0.19\textwidth}	
		\includegraphics[width=0.098\linewidth]{figures_supp/color_map.png}
		\includegraphics[width=0.85\linewidth]{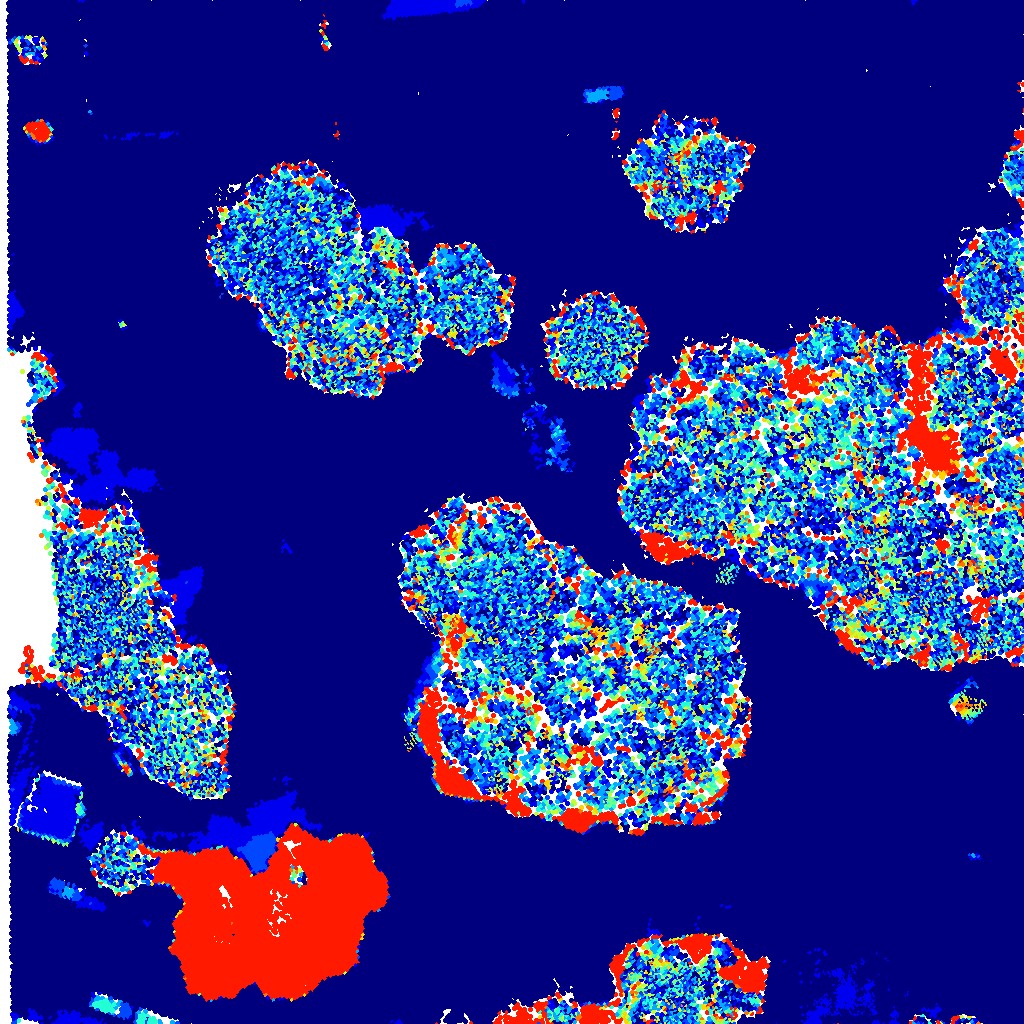}
		\includegraphics[width=\linewidth]{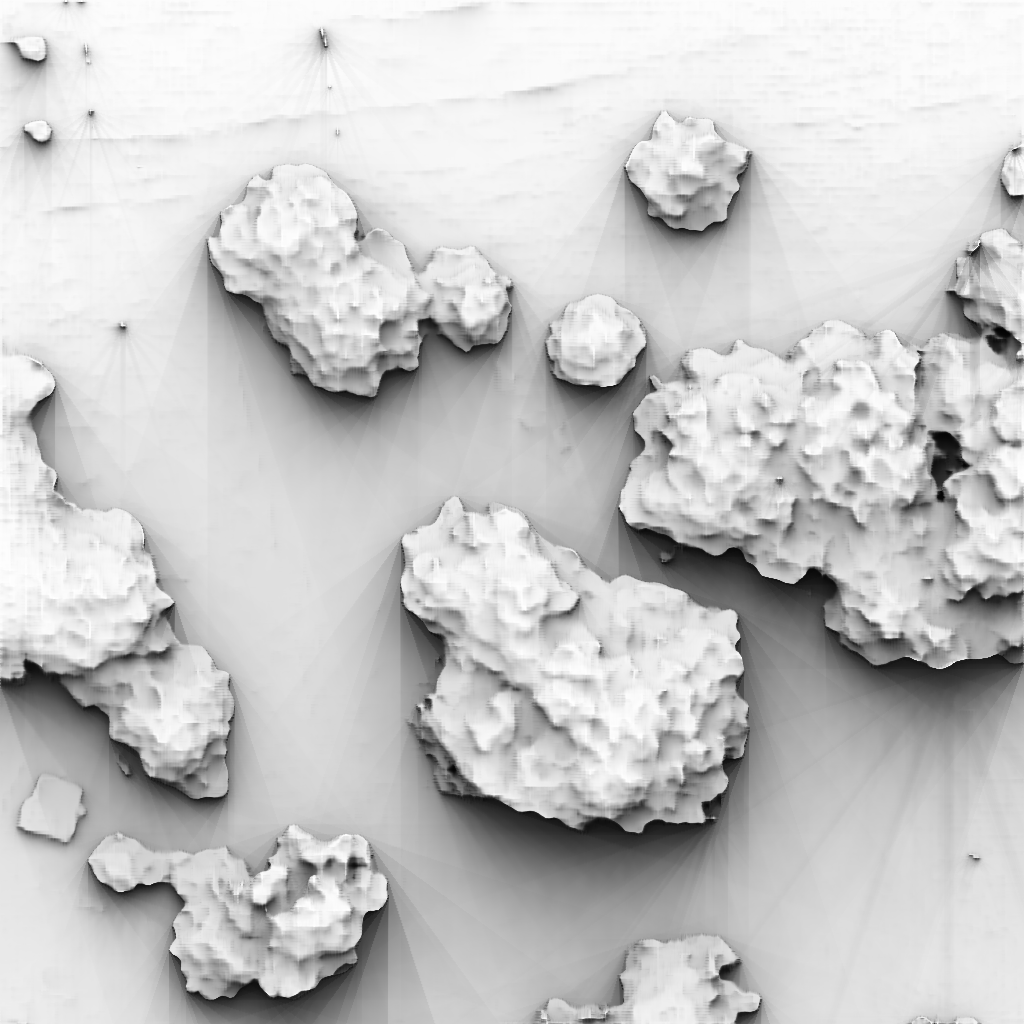}
		\centering{\tiny PSM net}
	\end{minipage}
	\begin{minipage}[t]{0.19\textwidth}	
		\includegraphics[width=0.098\linewidth]{figures_supp/color_map.png}
		\includegraphics[width=0.85\linewidth]{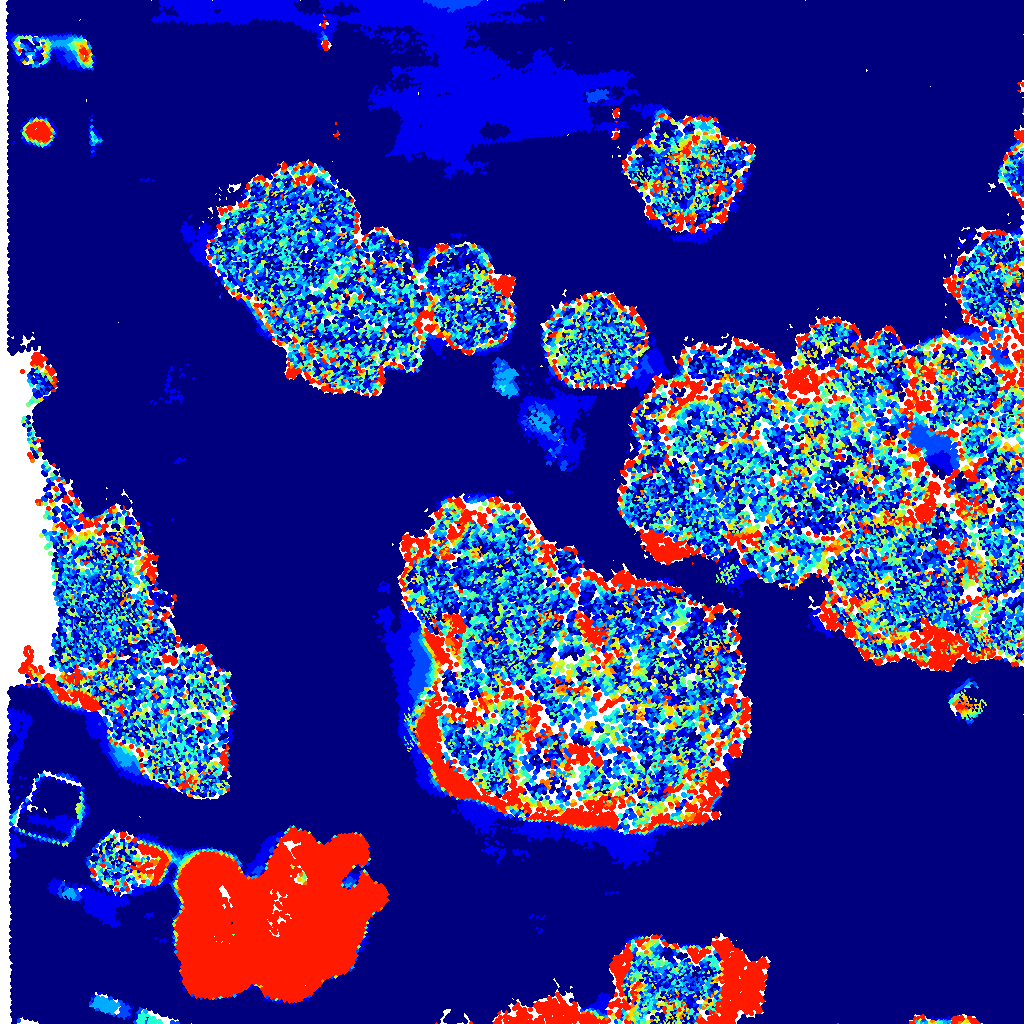}
		\includegraphics[width=\linewidth]{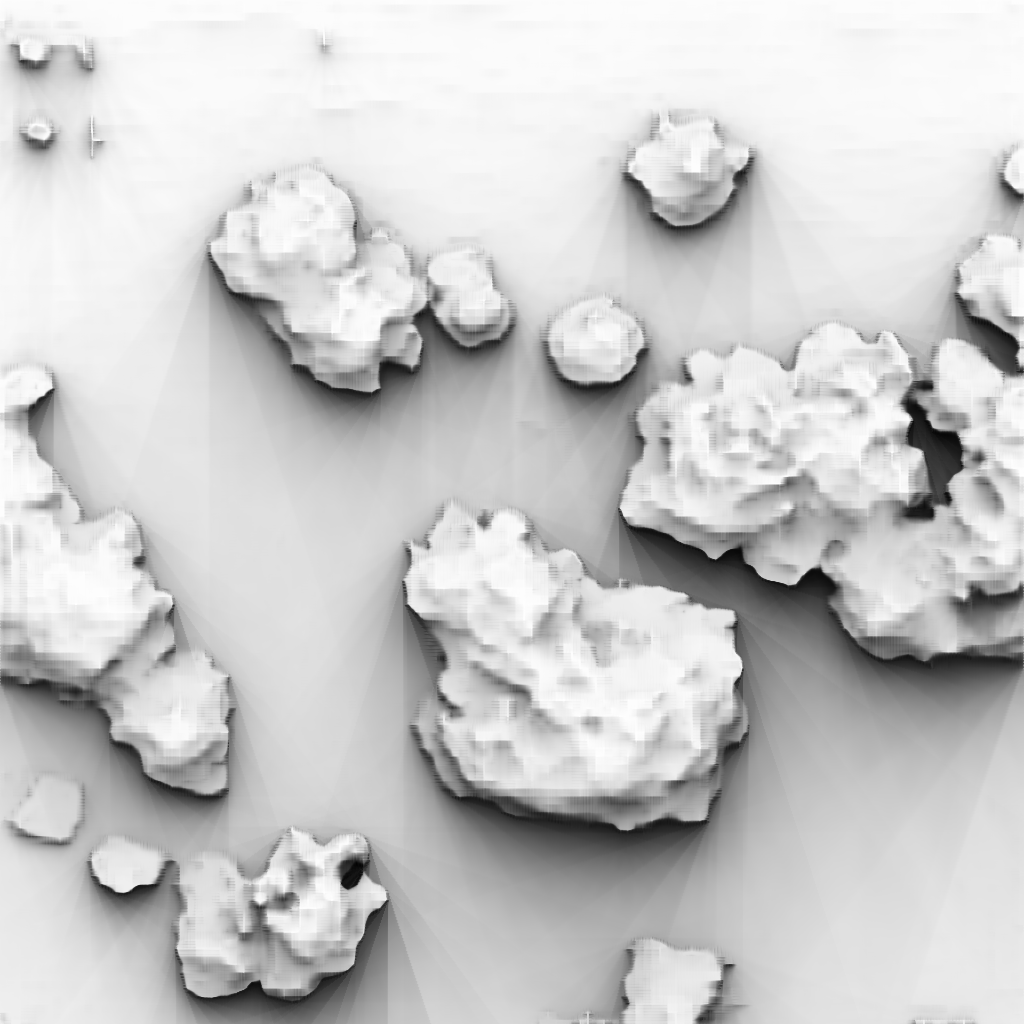}
		\centering{\tiny HRS net}
	\end{minipage}
	\begin{minipage}[t]{0.19\textwidth}	
		\includegraphics[width=0.098\linewidth]{figures_supp/color_map.png}
		\includegraphics[width=0.85\linewidth]{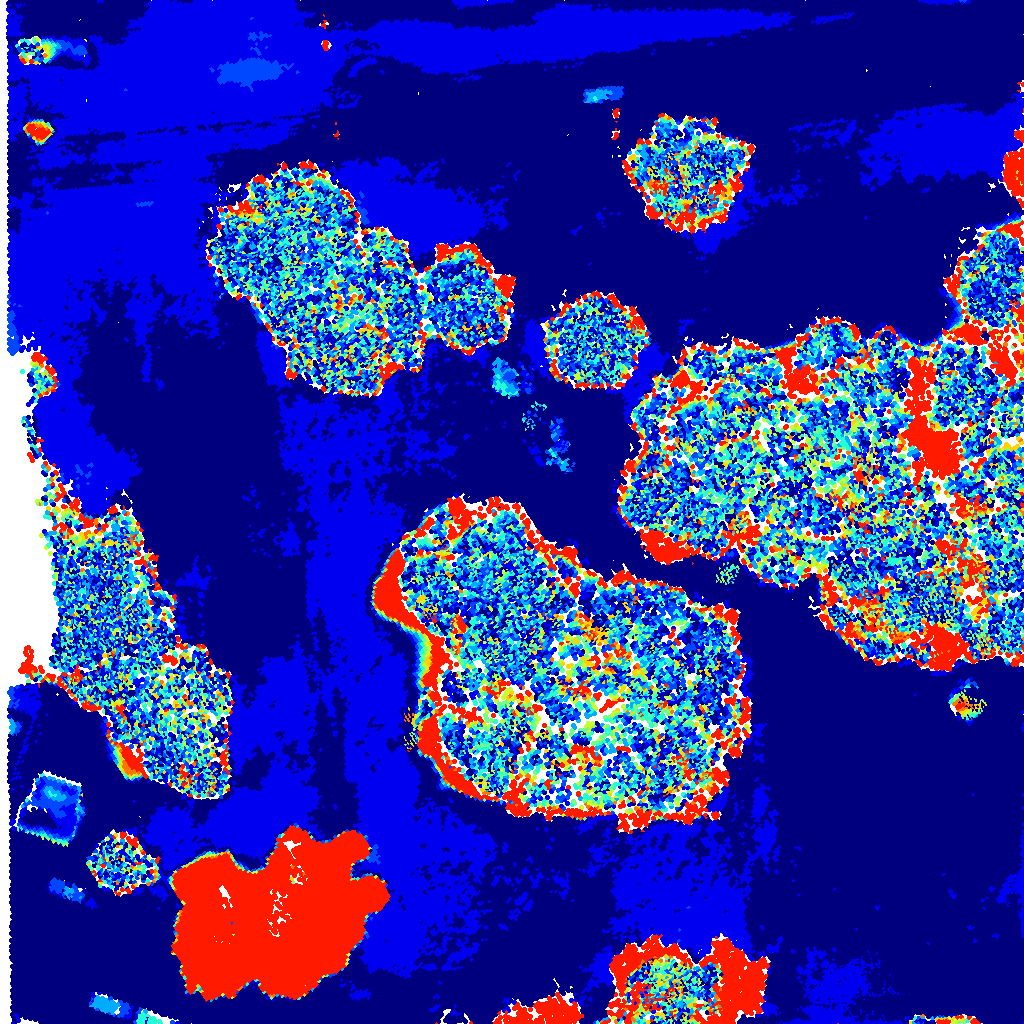}
		\includegraphics[width=\linewidth]{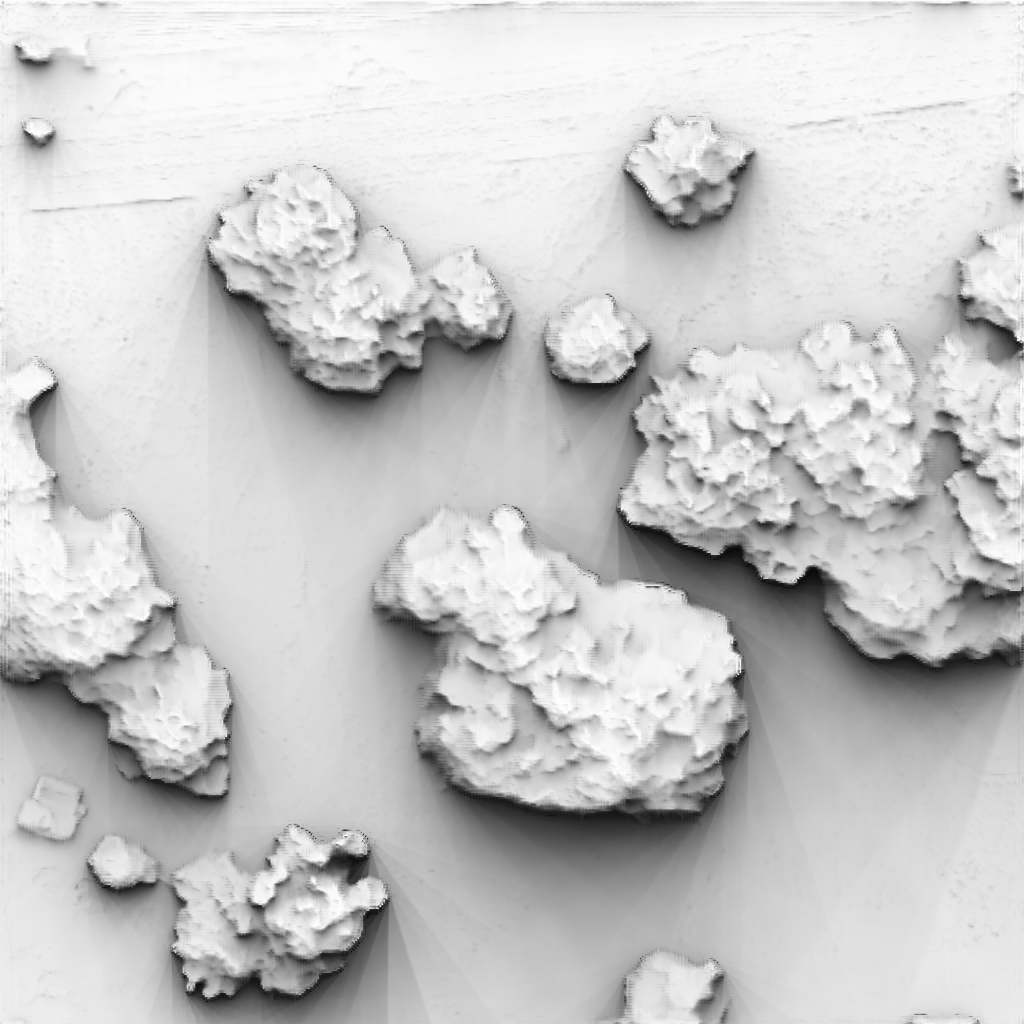}
		\centering{\tiny DeepPruner}
	\end{minipage}
	\begin{minipage}[t]{0.19\textwidth}
		\includegraphics[width=0.098\linewidth]{figures_supp/color_map.png}
		\includegraphics[width=0.85\linewidth]{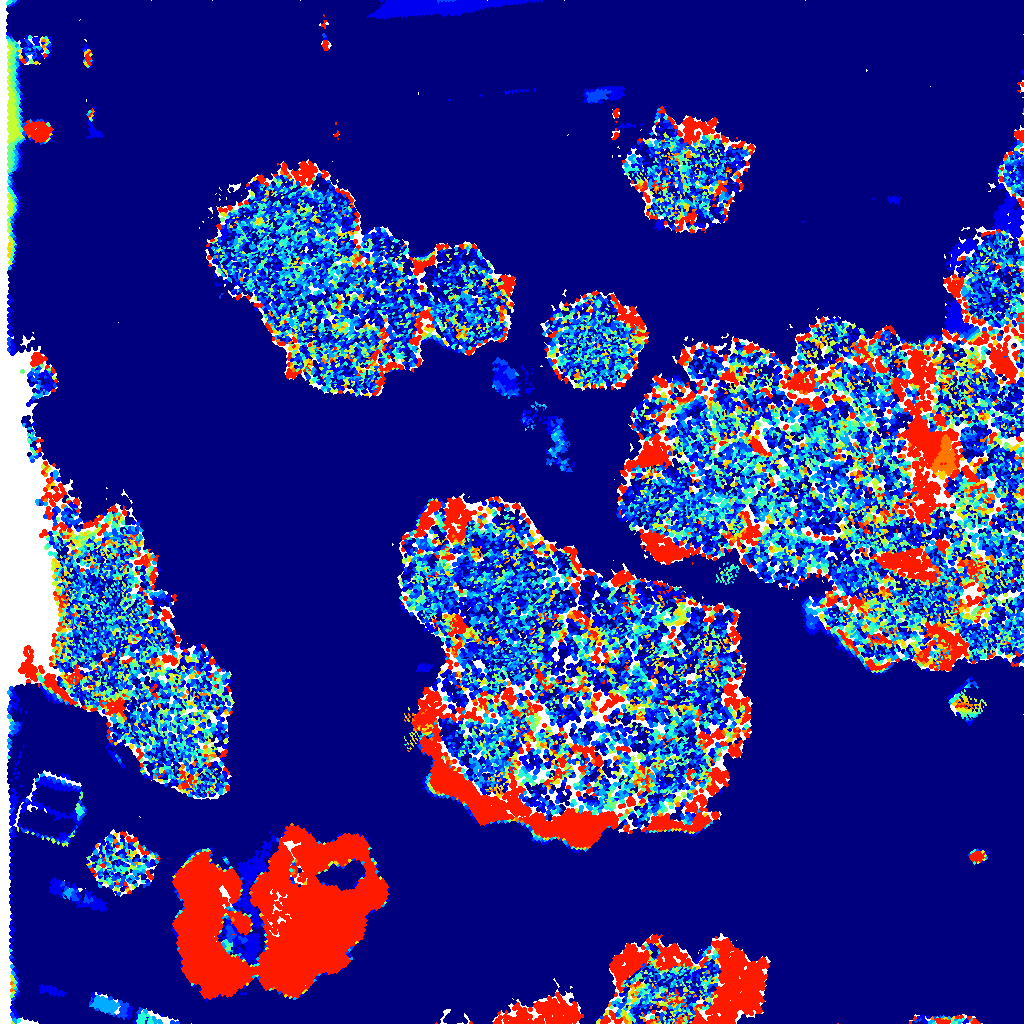}
		\includegraphics[width=\linewidth]{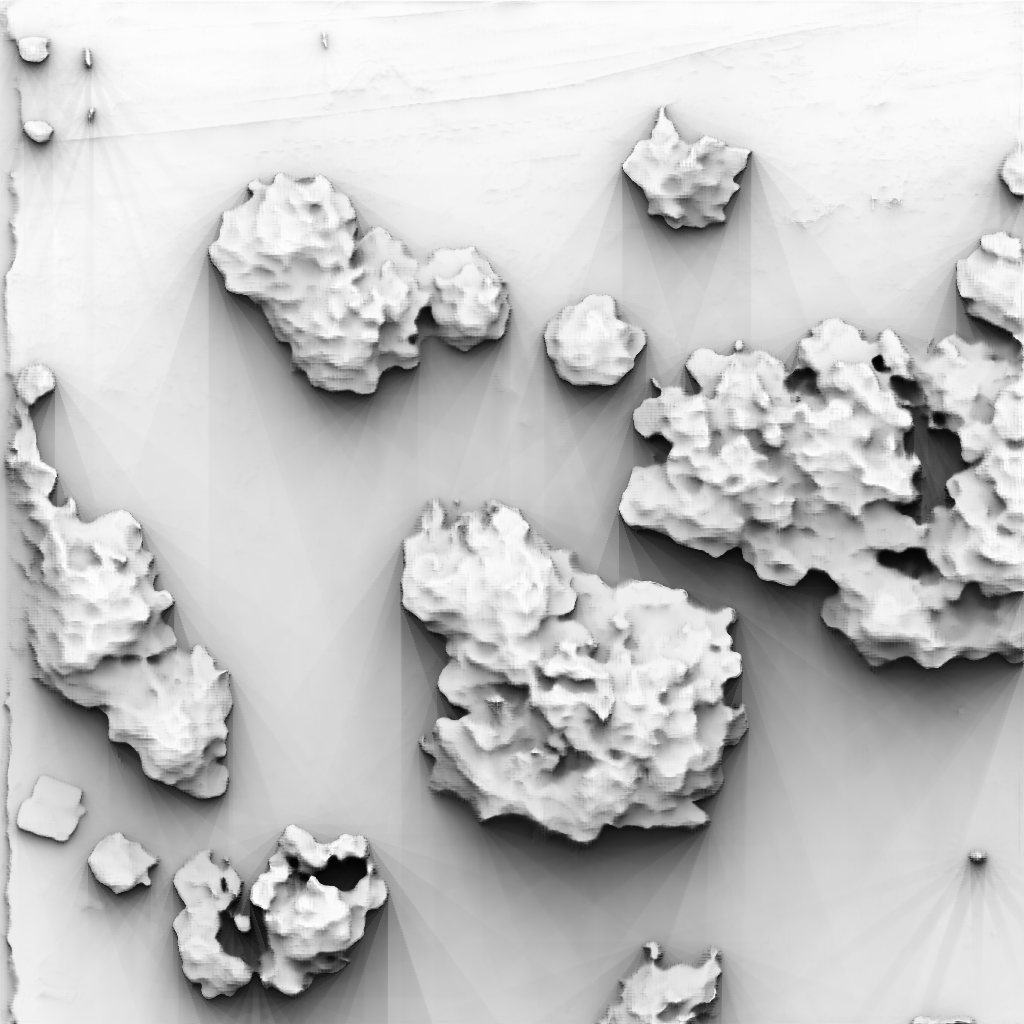}
		\centering{\tiny GANet}
	\end{minipage}
	\begin{minipage}[t]{0.19\textwidth}	
		\includegraphics[width=0.098\linewidth]{figures_supp/color_map.png}
		\includegraphics[width=0.85\linewidth]{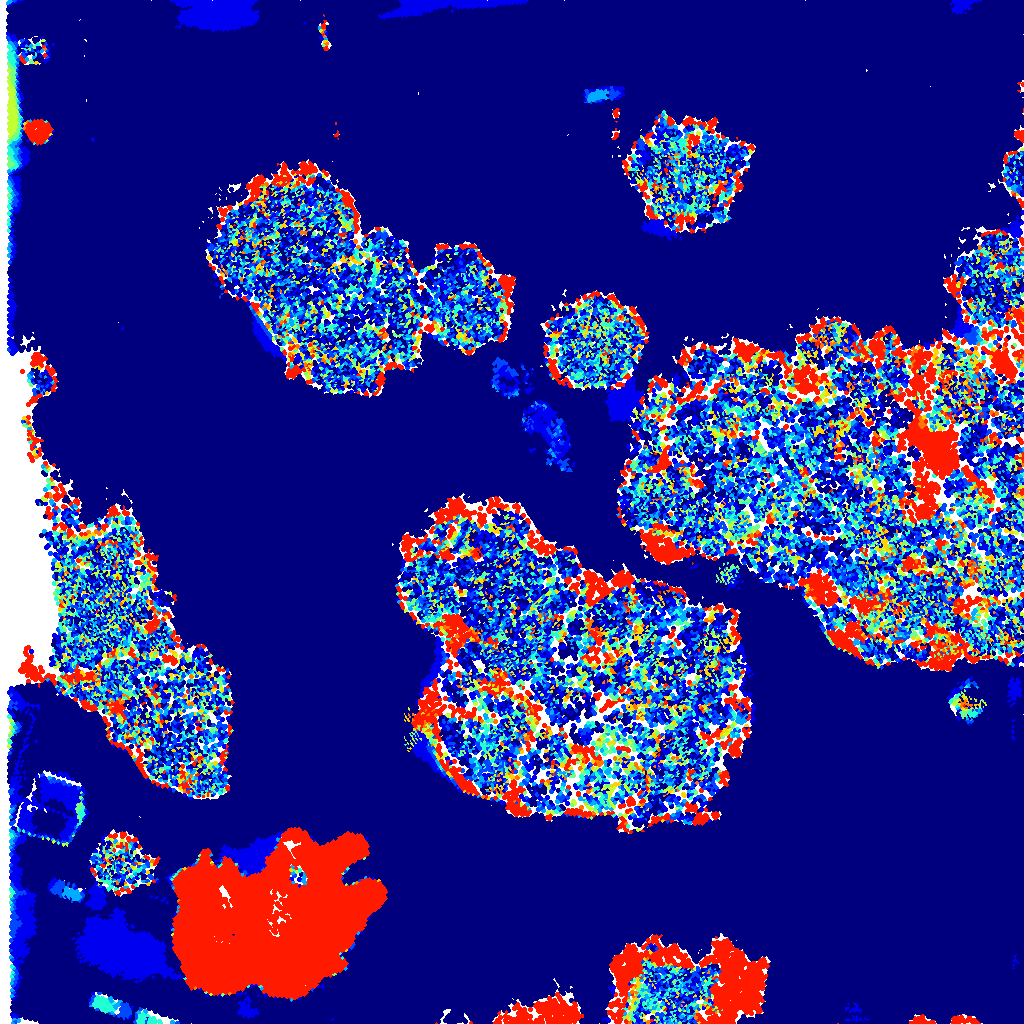}
		\includegraphics[width=\linewidth]{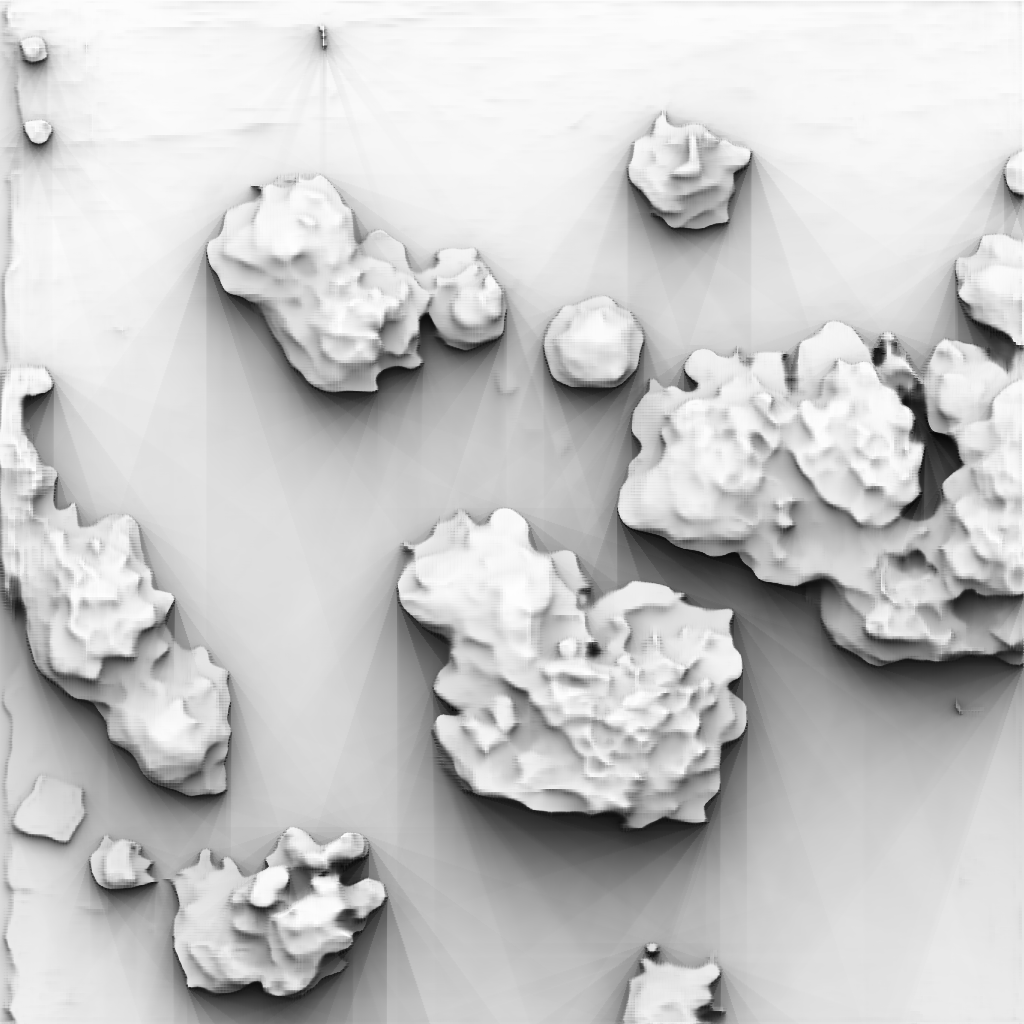}
		\centering{\tiny LEAStereo}
	\end{minipage}
	\caption{Error map and disparity visualization on tree area for Enschede dataset.}
	\label{Figure.enschedetree}
\end{figure}

\paragraph{Enschede}
The traditional SGM-based methods perform well on man-made objects, and so does the \textit{CBMV(GraphCuts)} after re-learning the features. The fine-tuned end-to-end methods are equally good, in particular the \textit{PSM net} and \textit{GAnet} (cf. \Cref{Figure.enschedebulding}). On vegetated areas, contrary to the end-to-end methods, the hybrid methods perform well even on pre-trained models (cf.  \Cref{Figure.enschedetree}). We observe that disparities predicted from pre-trained models of \textit{PSM net} and \textit{HRS net} differ in quality which points to their sensitivity to the training data.


For the result, on the bottom left, we observe an inconsistency between the scene represented in ground truth data and the images. This is more likely due to the temporal difference between the two acquisitions, i.e. LiDAR and images.

\begin{figure}[tp]
	\begin{minipage}[t]{0.19\textwidth}
		\includegraphics[width=0.098\linewidth]{figures_supp/color_map.png}
		\includegraphics[width=0.85\linewidth]{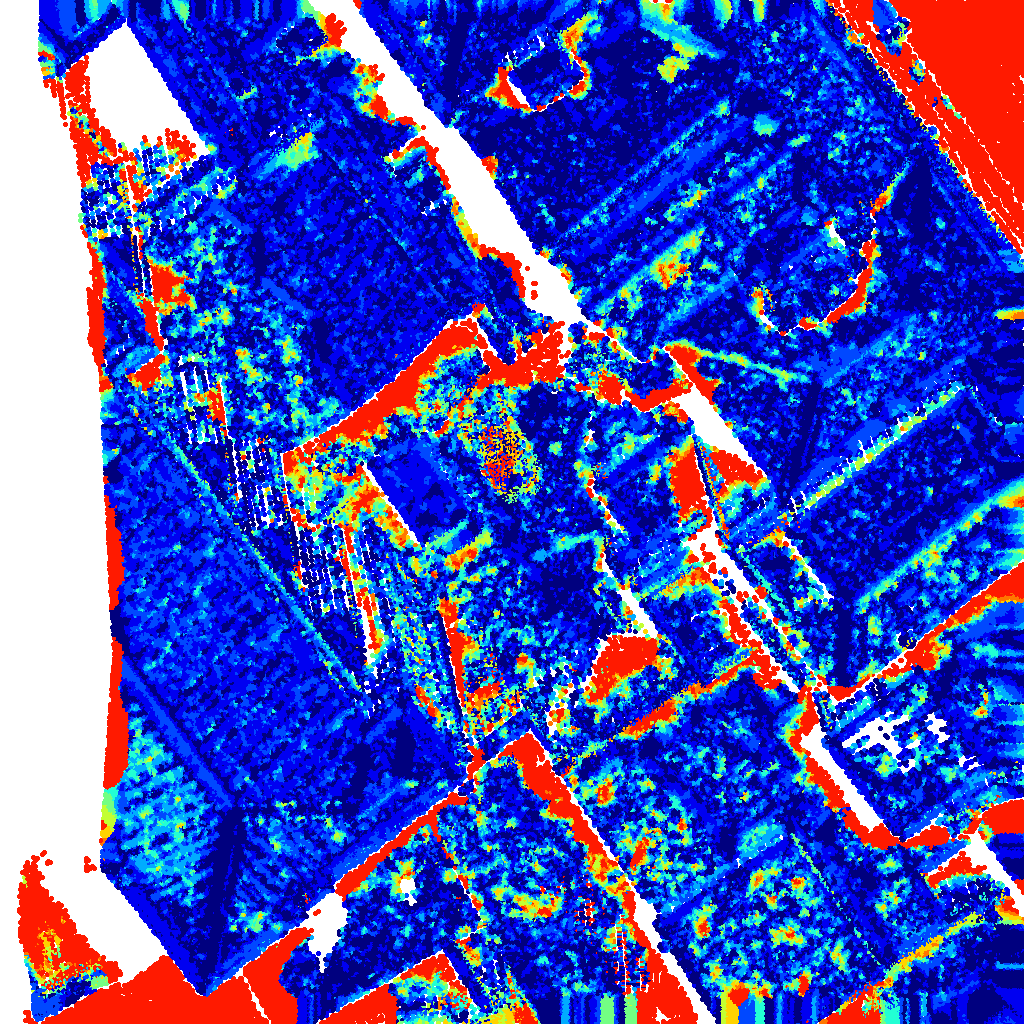}
		\includegraphics[width=\linewidth]{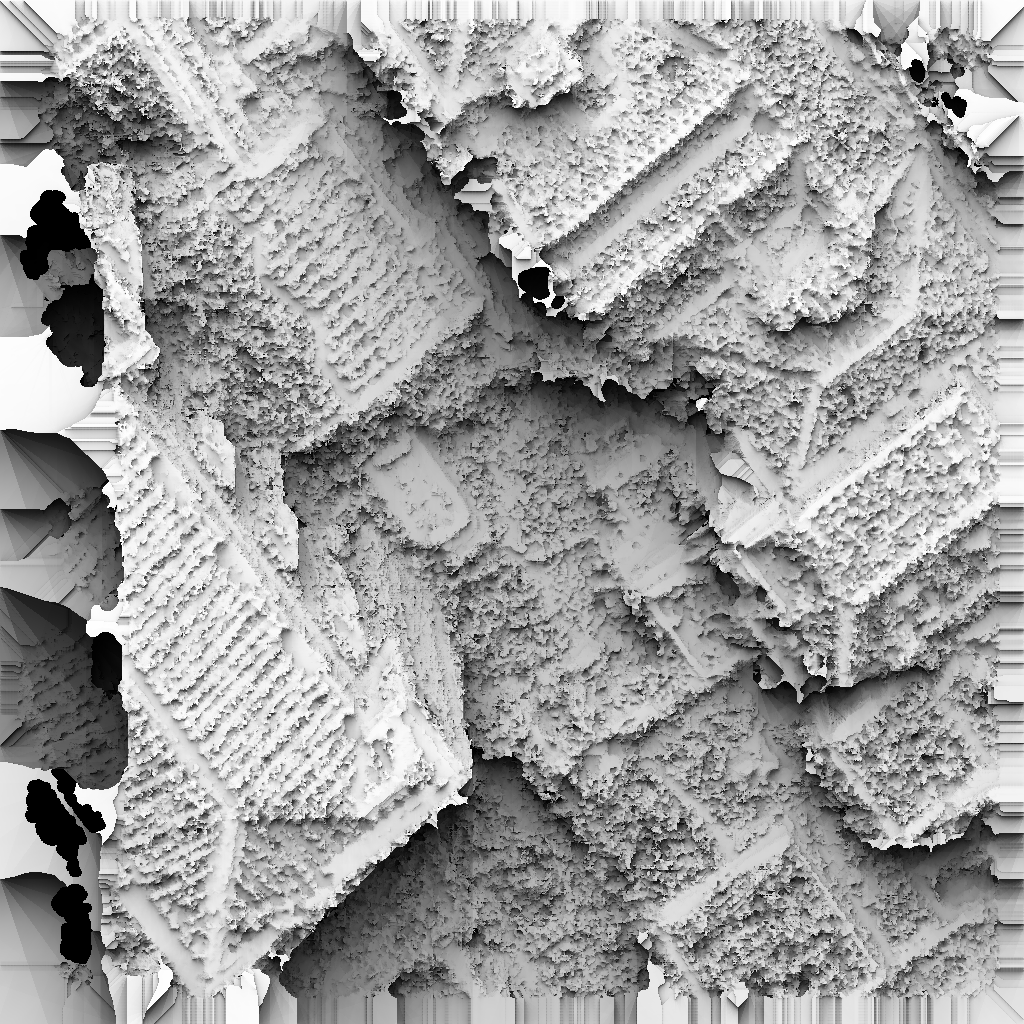}
		\centering{\tiny MICMAC}
	\end{minipage}
	\begin{minipage}[t]{0.19\textwidth}	
		\includegraphics[width=0.098\linewidth]{figures_supp/color_map.png}
		\includegraphics[width=0.85\linewidth]{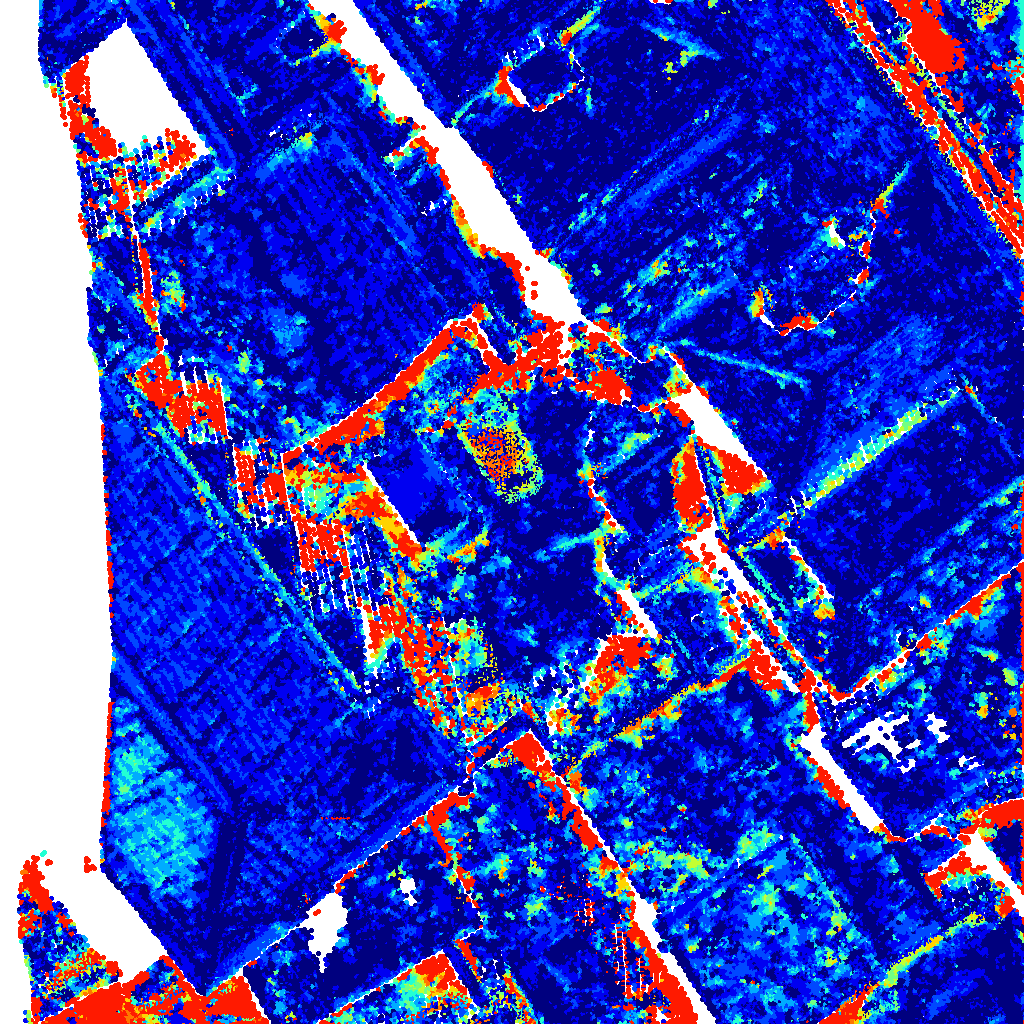}
		\includegraphics[width=\linewidth]{figures_supp/dublin/pyr_micmac/3489_DUBLIN_AREA_2KM2_rgb_124885_id278c1_20150326120951_3489_DUBLIN_AREA_2KM2_rgb_124888_id281c1_20150326120954_0005_shade.png}
		\centering{\tiny SGM(CUDA)}
	\end{minipage}
	\begin{minipage}[t]{0.19\textwidth}	
		\includegraphics[width=0.098\linewidth]{figures_supp/color_map.png}
		\includegraphics[width=0.85\linewidth]{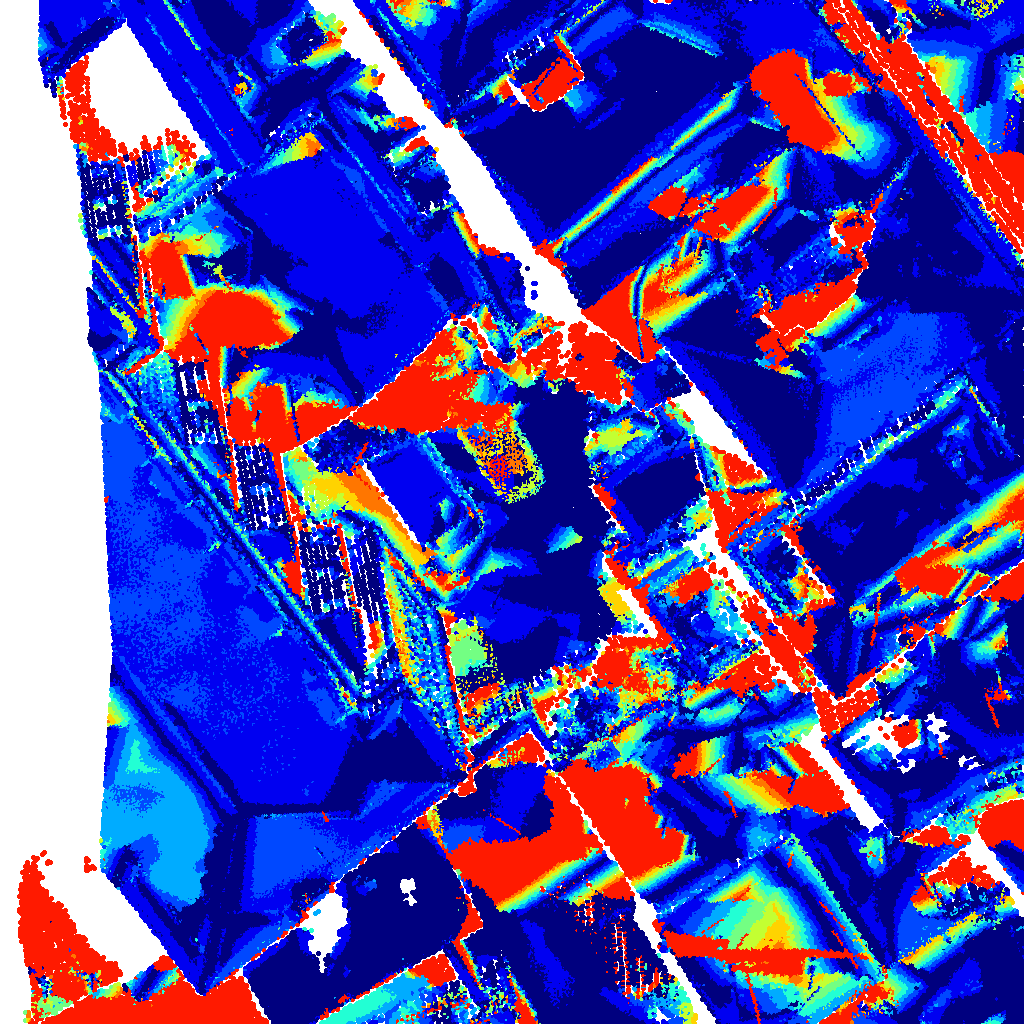}
		\includegraphics[width=\linewidth]{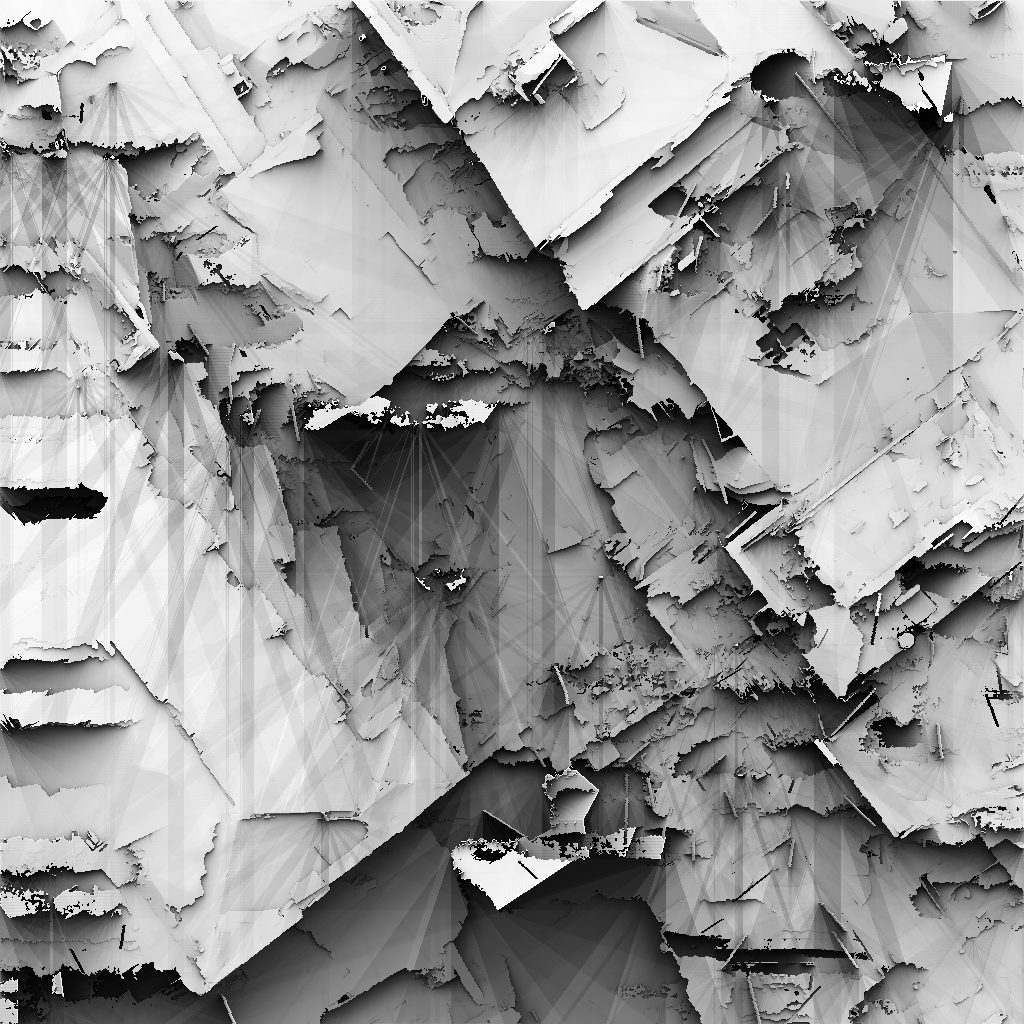}
		\centering{\tiny GraphCuts}
	\end{minipage}
	\begin{minipage}[t]{0.19\textwidth}	
		\includegraphics[width=0.098\linewidth]{figures_supp/color_map.png}
		\includegraphics[width=0.85\linewidth]{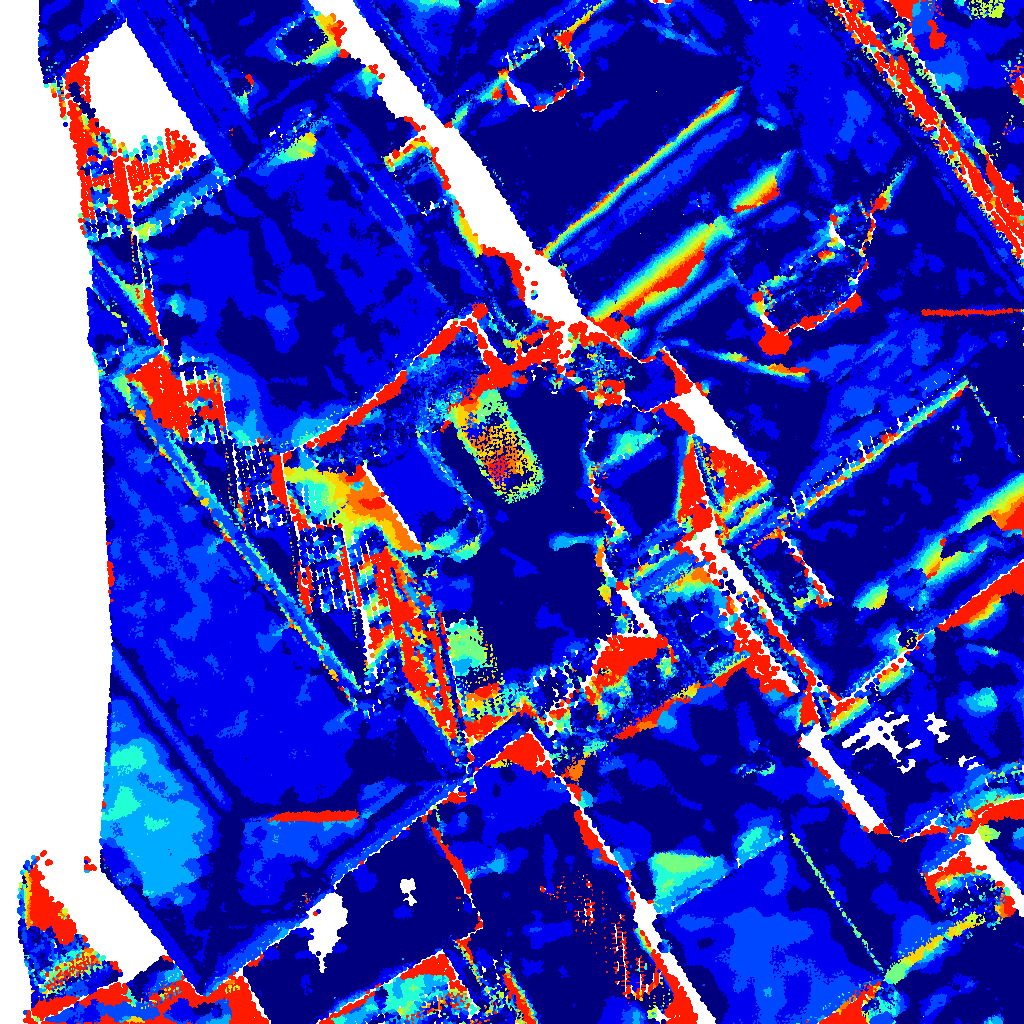}
		\includegraphics[width=\linewidth]{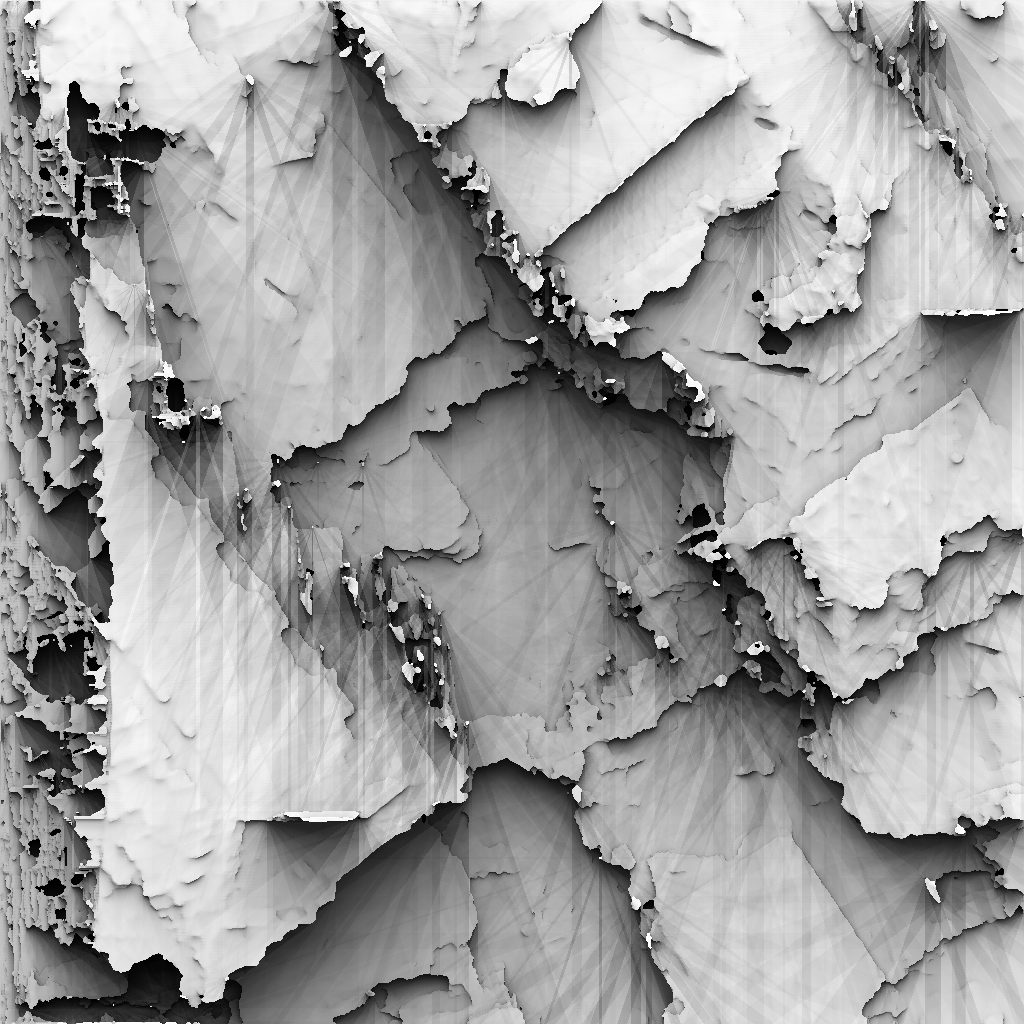}
		\centering{\tiny CBMV(SGM)}
	\end{minipage}
	\begin{minipage}[t]{0.19\textwidth}
		\includegraphics[width=0.098\linewidth]{figures_supp/color_map.png}
		\includegraphics[width=0.85\linewidth]{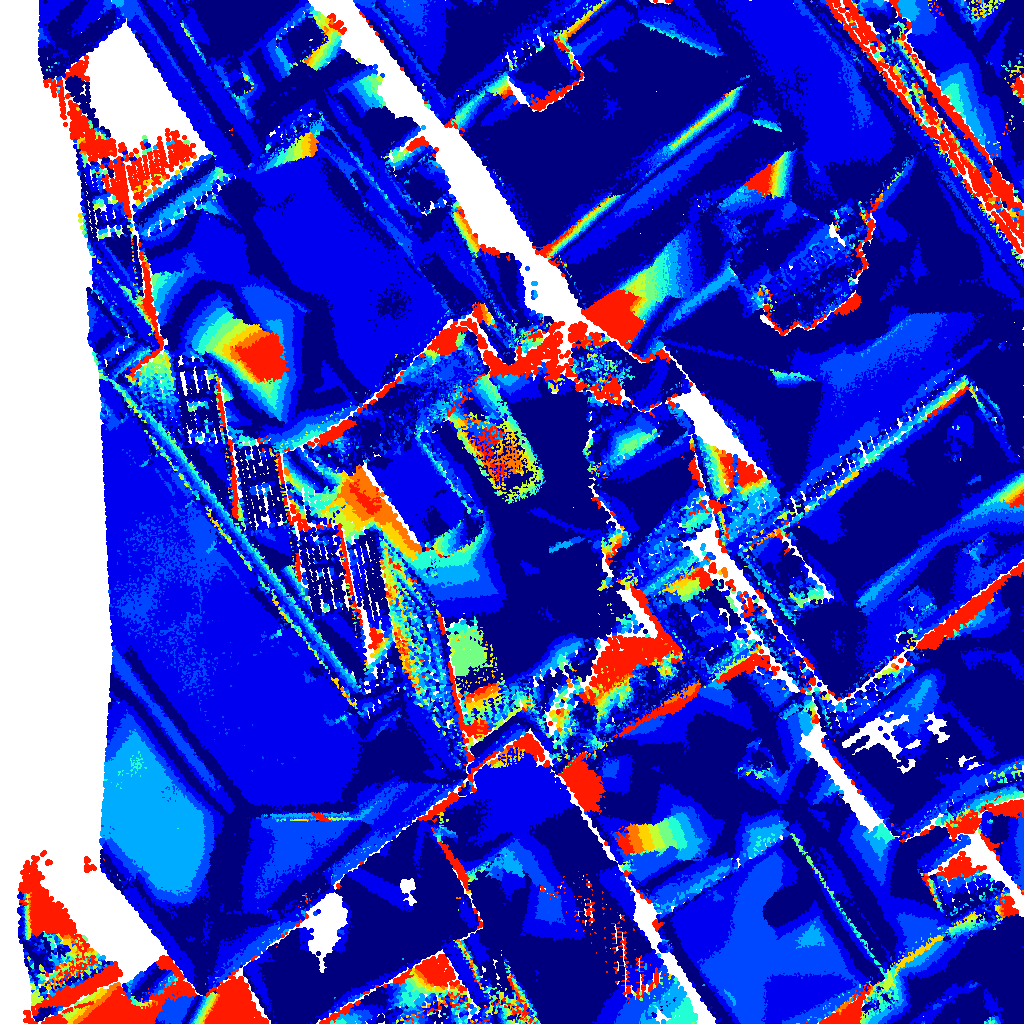}
		\includegraphics[width=\linewidth]{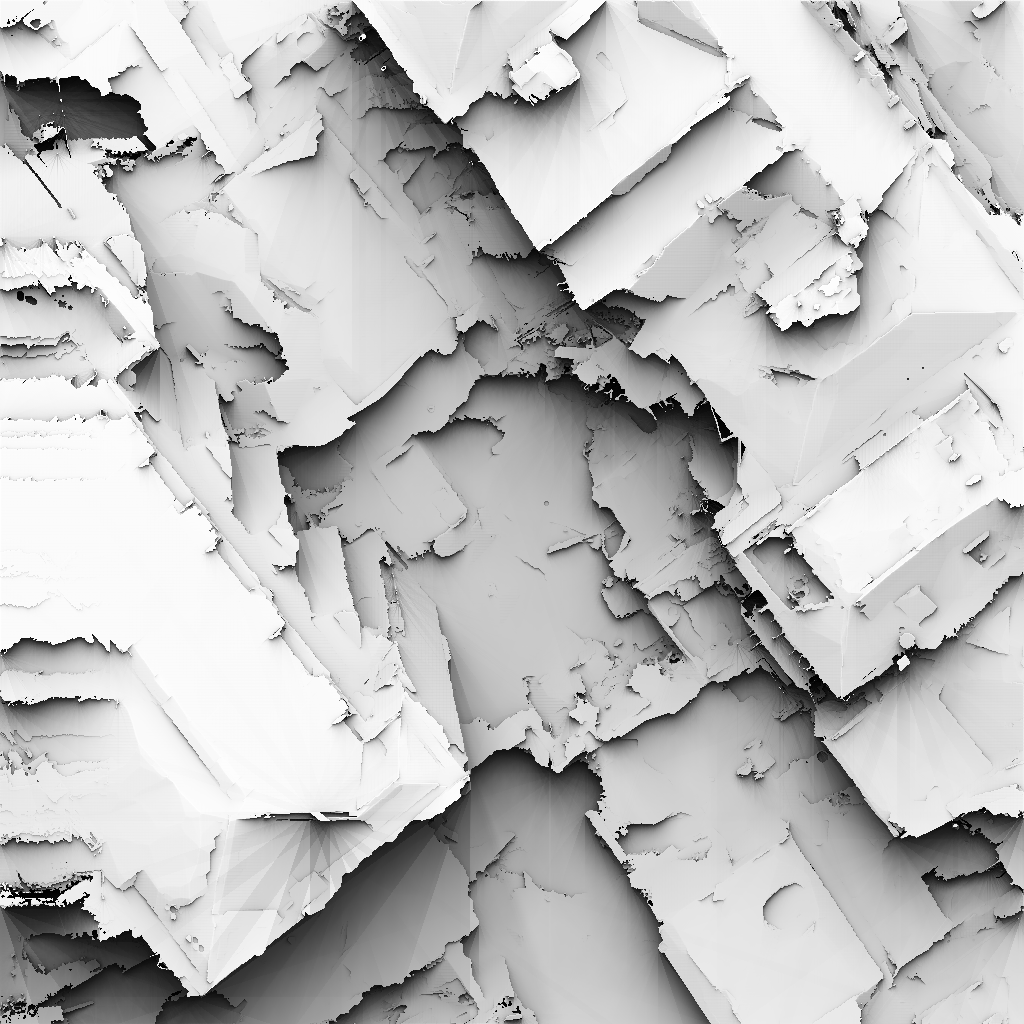}
		\centering{\tiny CBMV(GraphCuts)}
	\end{minipage}
	\begin{minipage}[t]{0.19\textwidth}
		\includegraphics[width=0.098\linewidth]{figures_supp/color_map.png}
		\includegraphics[width=0.85\linewidth]{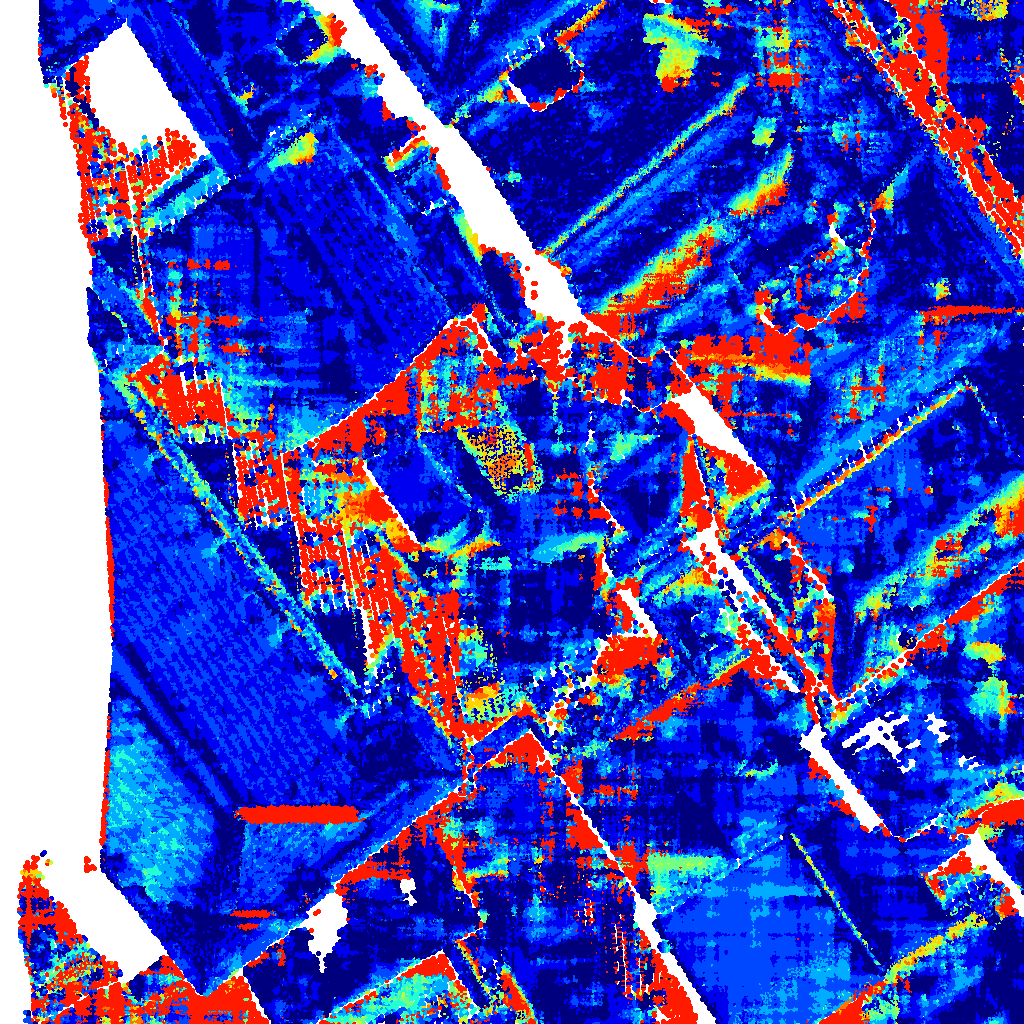}
		\includegraphics[width=\linewidth]{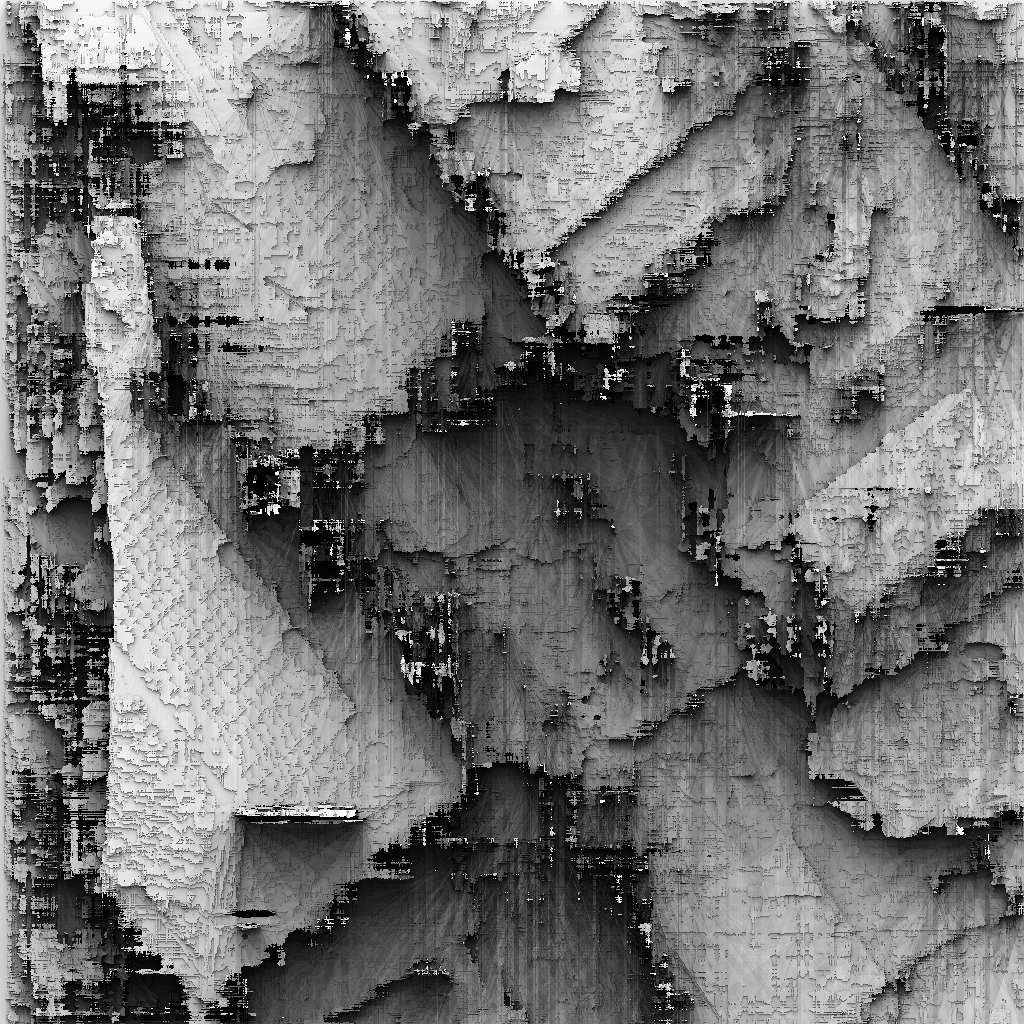}
		\centering{\tiny MC-CNN(KITTI)}
	\end{minipage}
	\begin{minipage}[t]{0.19\textwidth}
		\includegraphics[width=0.098\linewidth]{figures_supp/color_map.png}
		\includegraphics[width=0.85\linewidth]{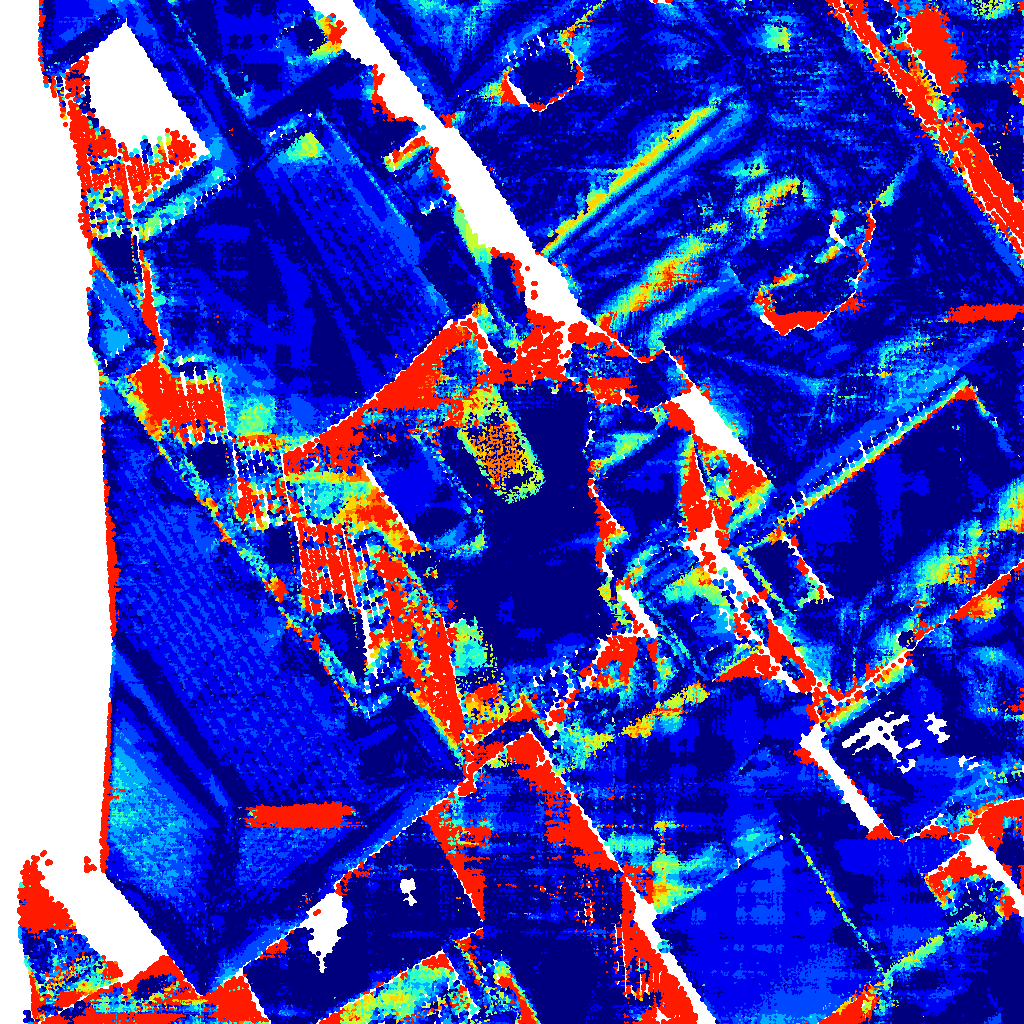}
		\includegraphics[width=\linewidth]{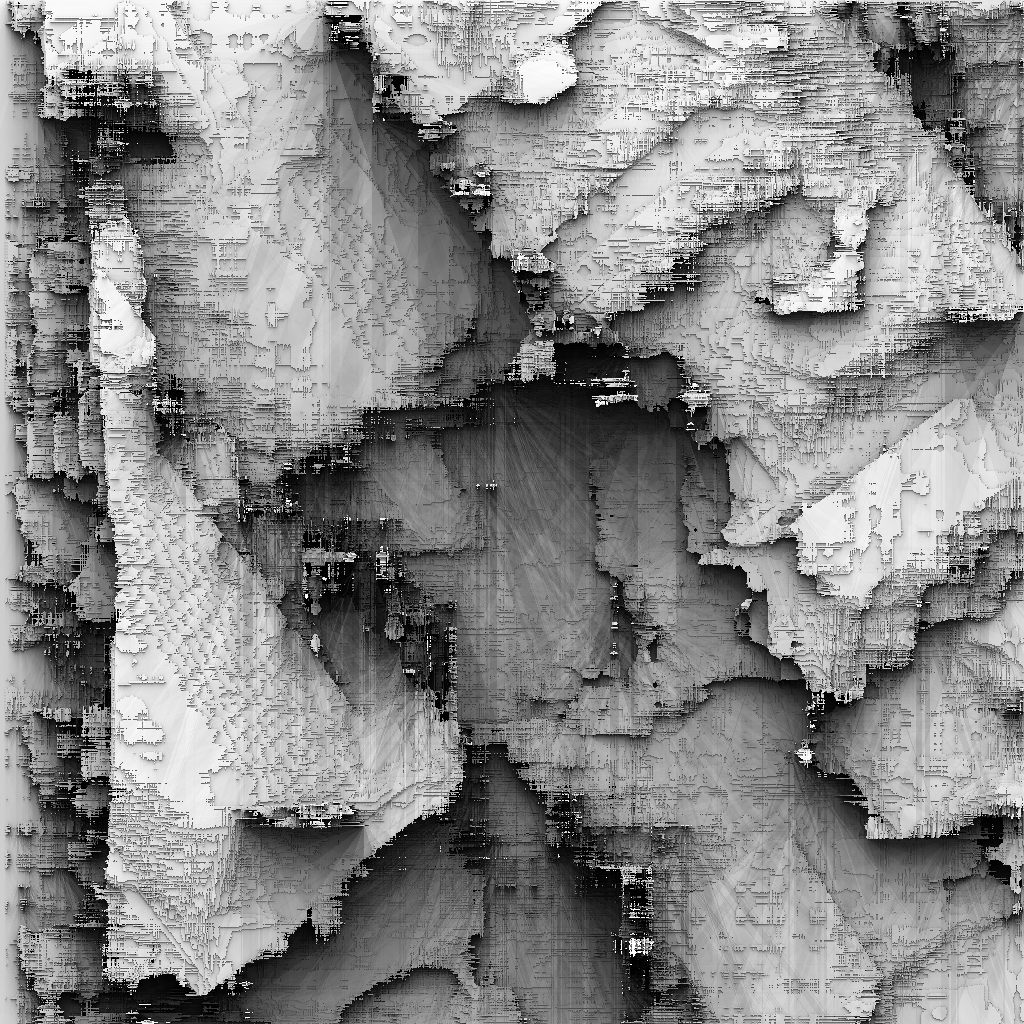}
		\centering{\tiny EfficientDeep(KITTI)}
	\end{minipage}
	\begin{minipage}[t]{0.19\textwidth}
		\includegraphics[width=0.098\linewidth]{figures_supp/color_map.png}
		\includegraphics[width=0.85\linewidth]{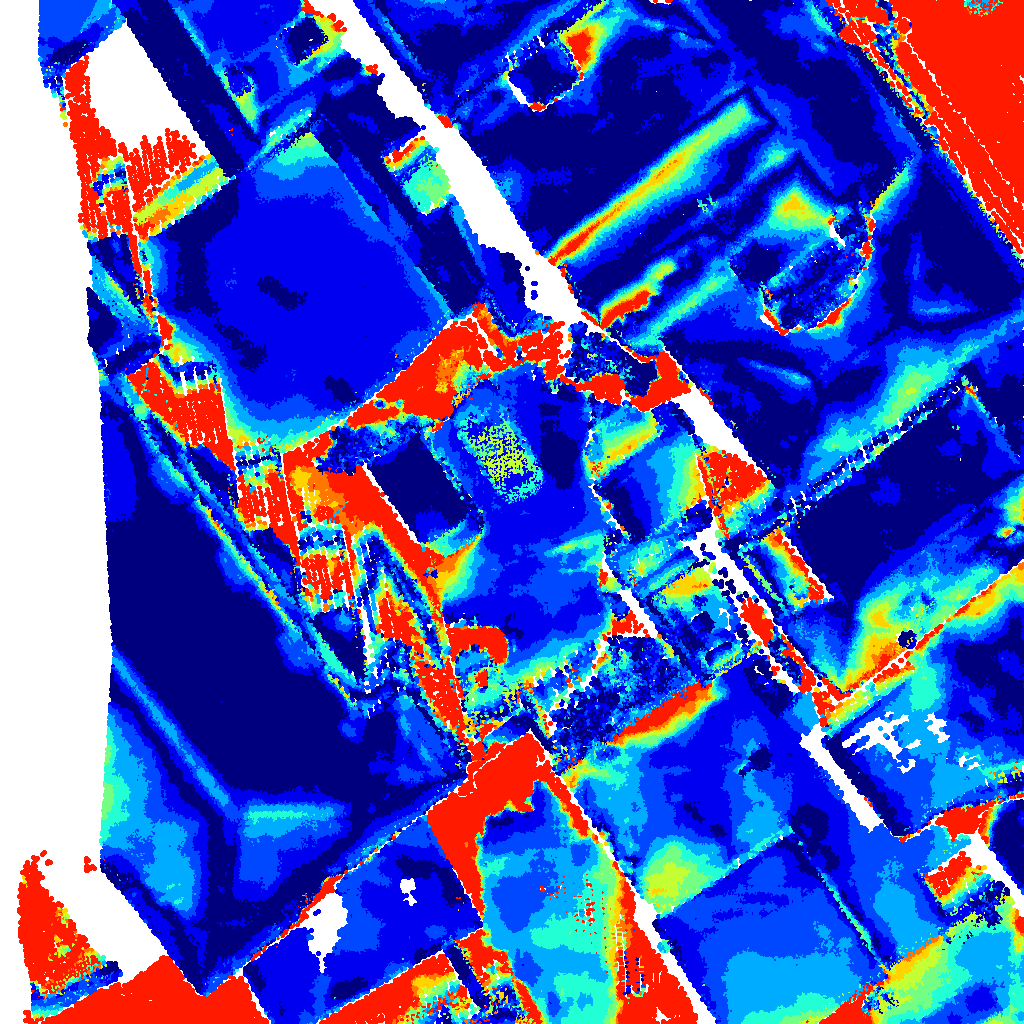}
		\includegraphics[width=\linewidth]{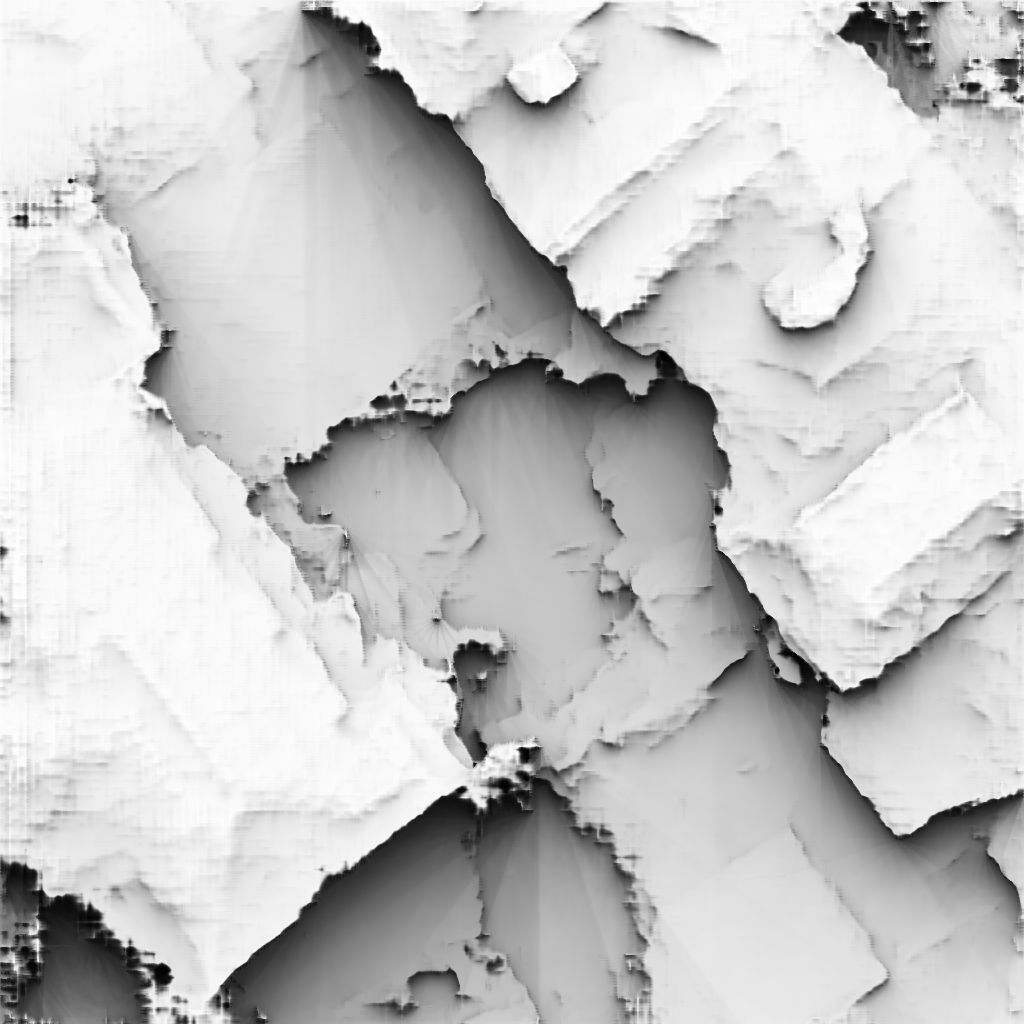}
		\centering{\tiny PSM net(KITTI)}
	\end{minipage}
	\begin{minipage}[t]{0.19\textwidth}	
		\includegraphics[width=0.098\linewidth]{figures_supp/color_map.png}
		\includegraphics[width=0.85\linewidth]{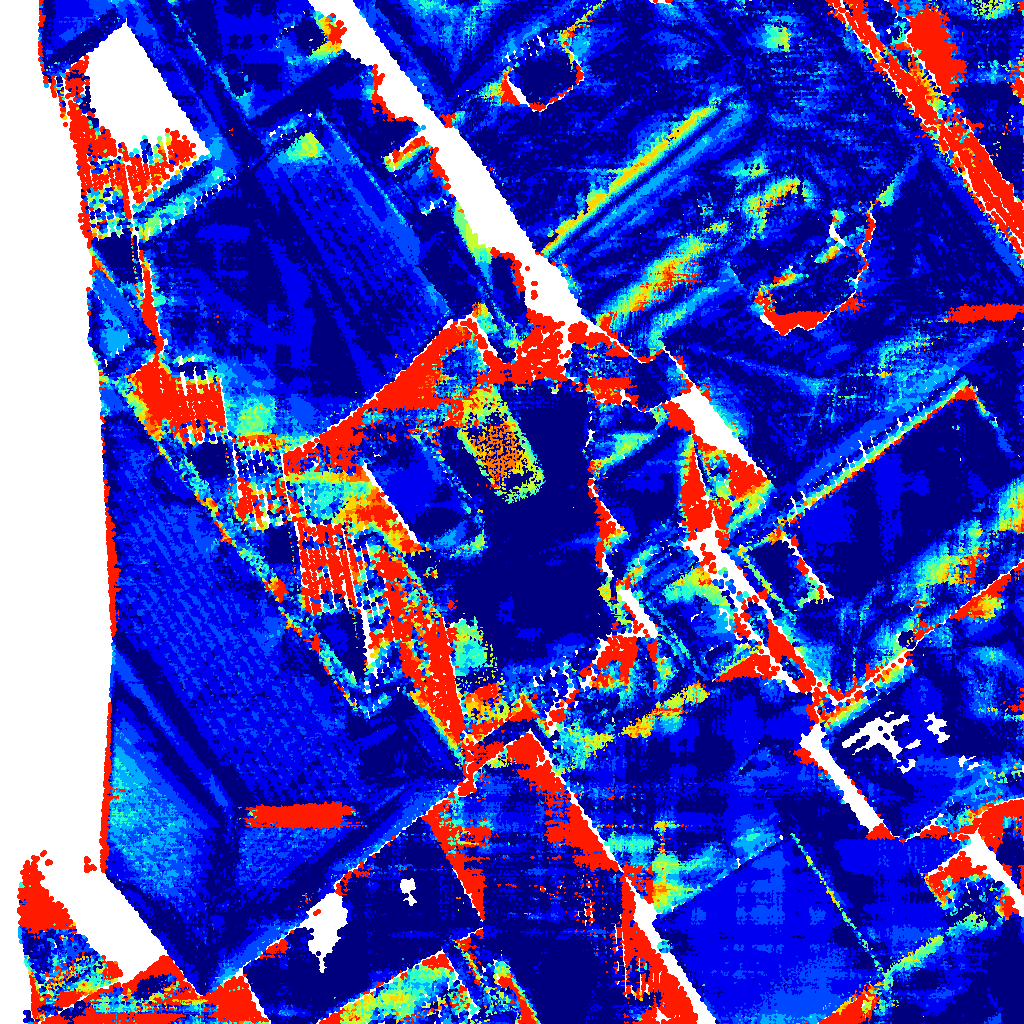}
		\includegraphics[width=\linewidth]{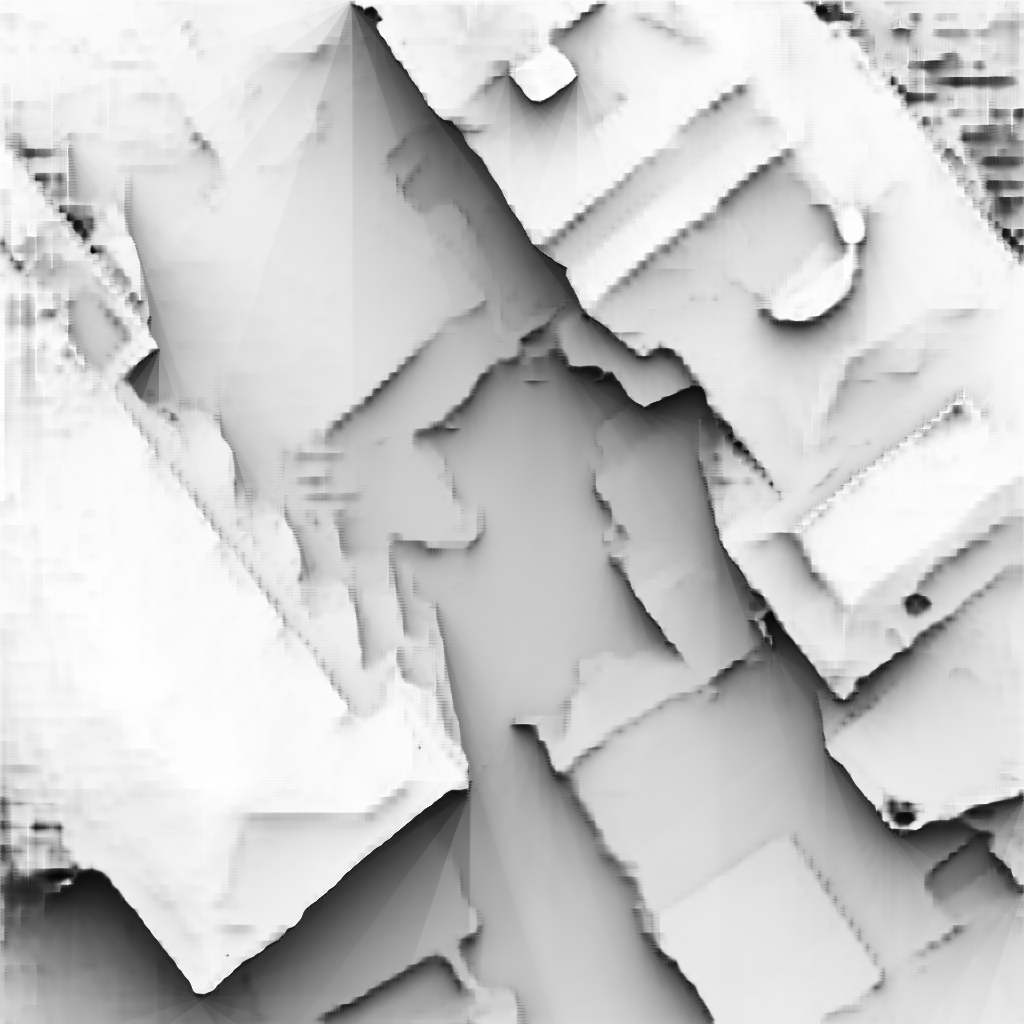}
			\centering{\tiny HRS net(KITTI)}
	\end{minipage}
	\begin{minipage}[t]{0.19\textwidth}	
		\includegraphics[width=0.098\linewidth]{figures_supp/color_map.png}
		\includegraphics[width=0.85\linewidth]{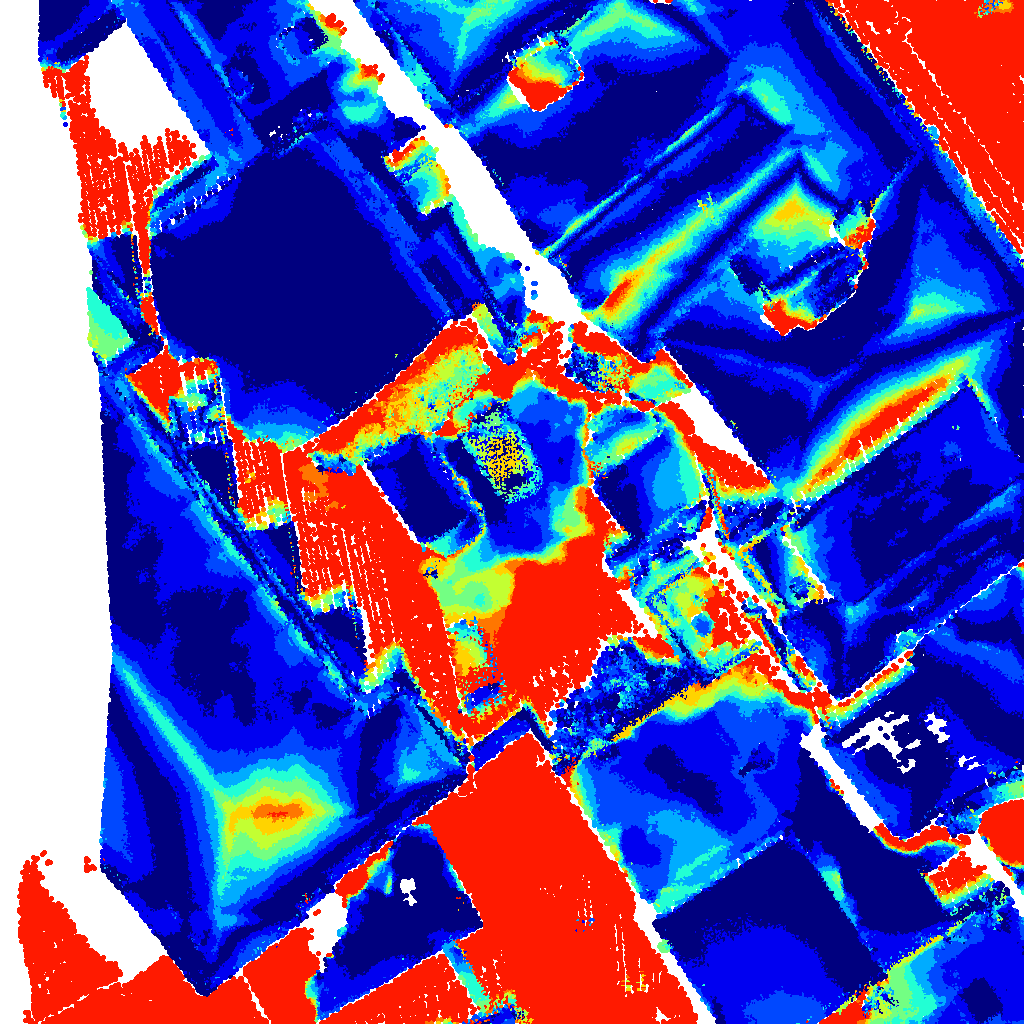}
		\includegraphics[width=\linewidth]{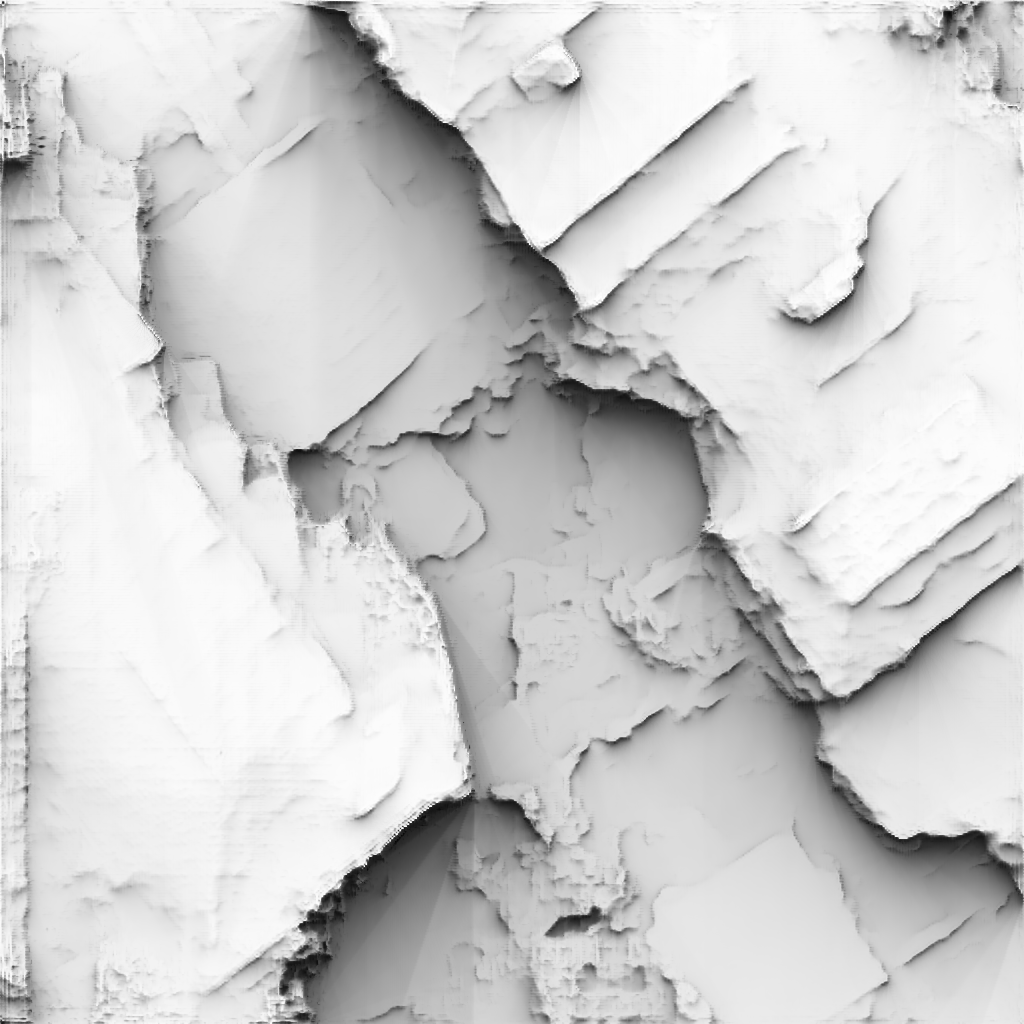}
		\centering{\tiny DeepPruner(KITTI)}
	\end{minipage}
	\begin{minipage}[t]{0.19\textwidth}	
		\includegraphics[width=0.098\linewidth]{figures_supp/color_map.png}
		\includegraphics[width=0.85\linewidth]{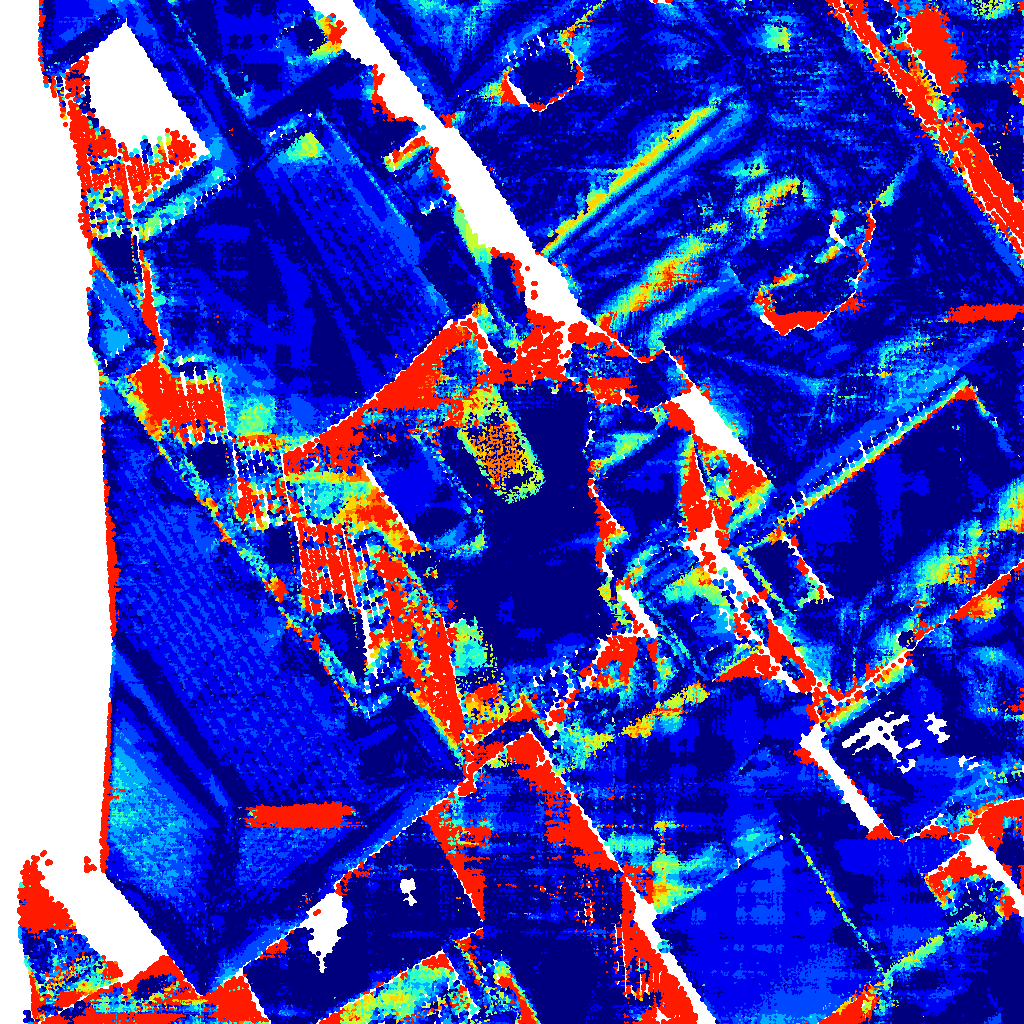}
		\includegraphics[width=\linewidth]{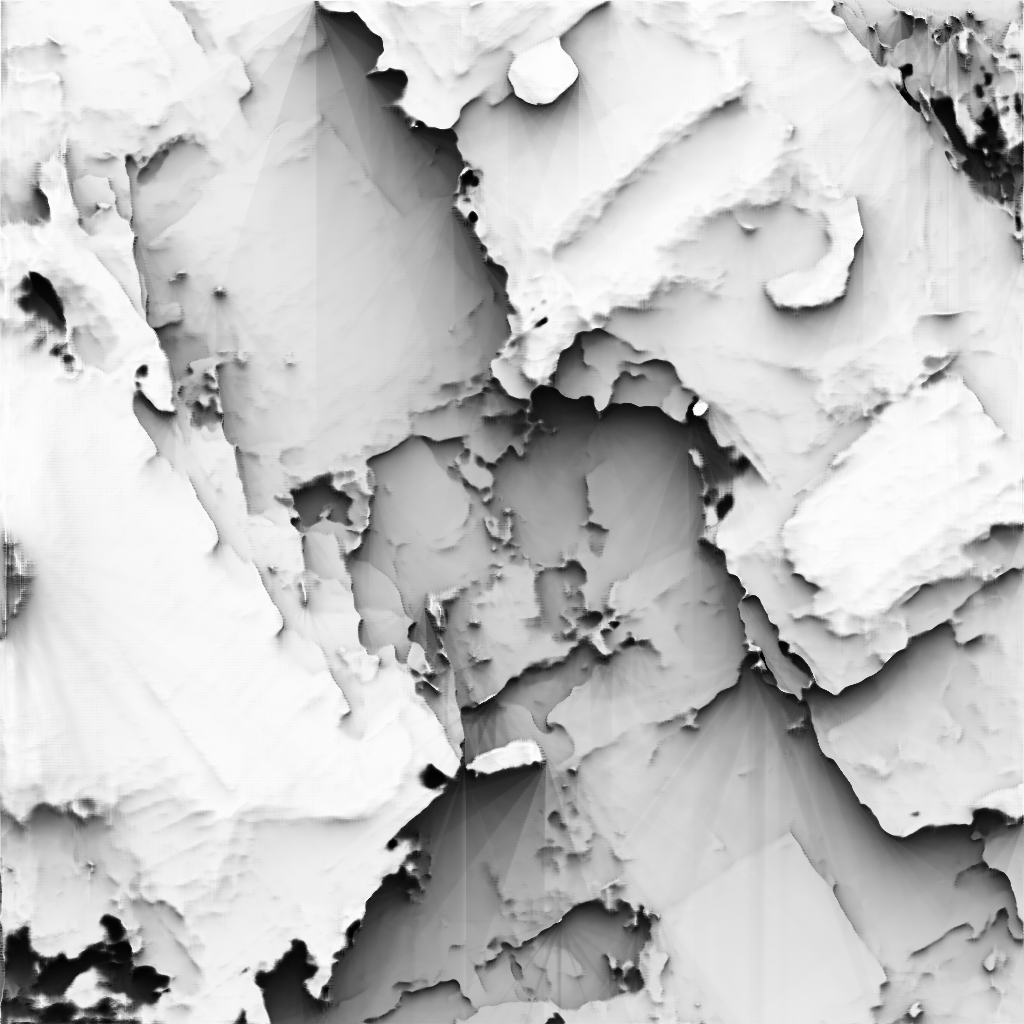}
		\centering{\tiny GANet(KITTI)}
	\end{minipage}
	\begin{minipage}[t]{0.19\textwidth}	
		\includegraphics[width=0.098\linewidth]{figures_supp/color_map.png}
		\includegraphics[width=0.85\linewidth]{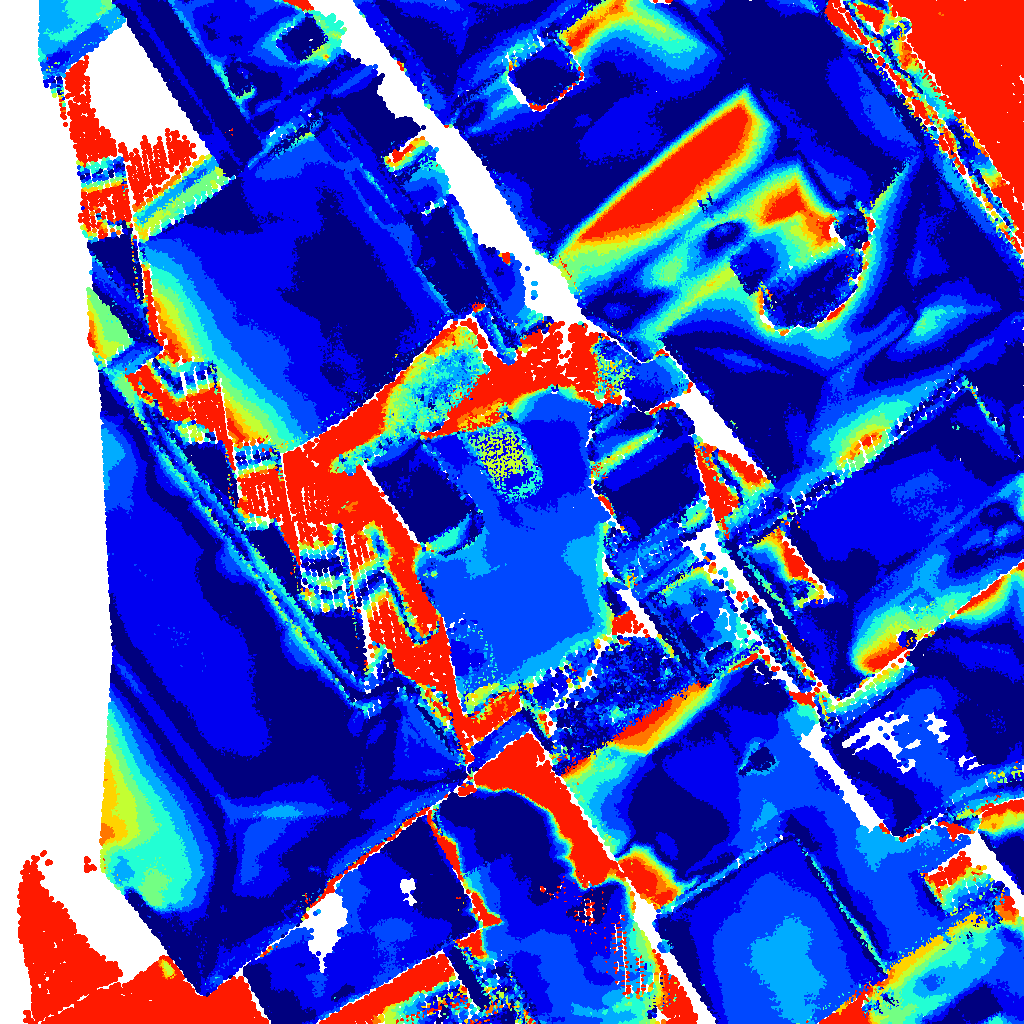}
		\includegraphics[width=\linewidth]{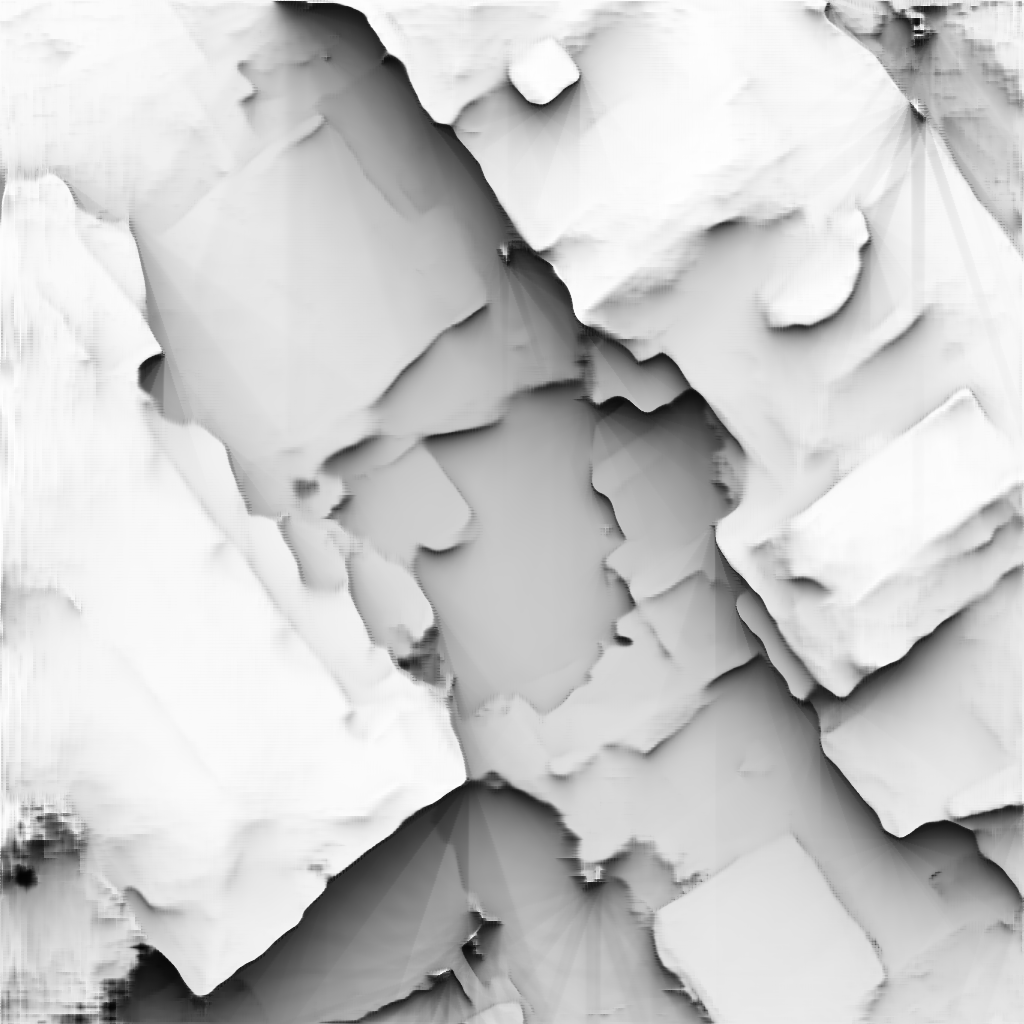}
		\centering{\tiny LEAStereo(KITTI)}
	\end{minipage}
	\begin{minipage}[t]{0.19\textwidth}
		\includegraphics[width=0.098\linewidth]{figures_supp/color_map.png}
		\includegraphics[width=0.85\linewidth]{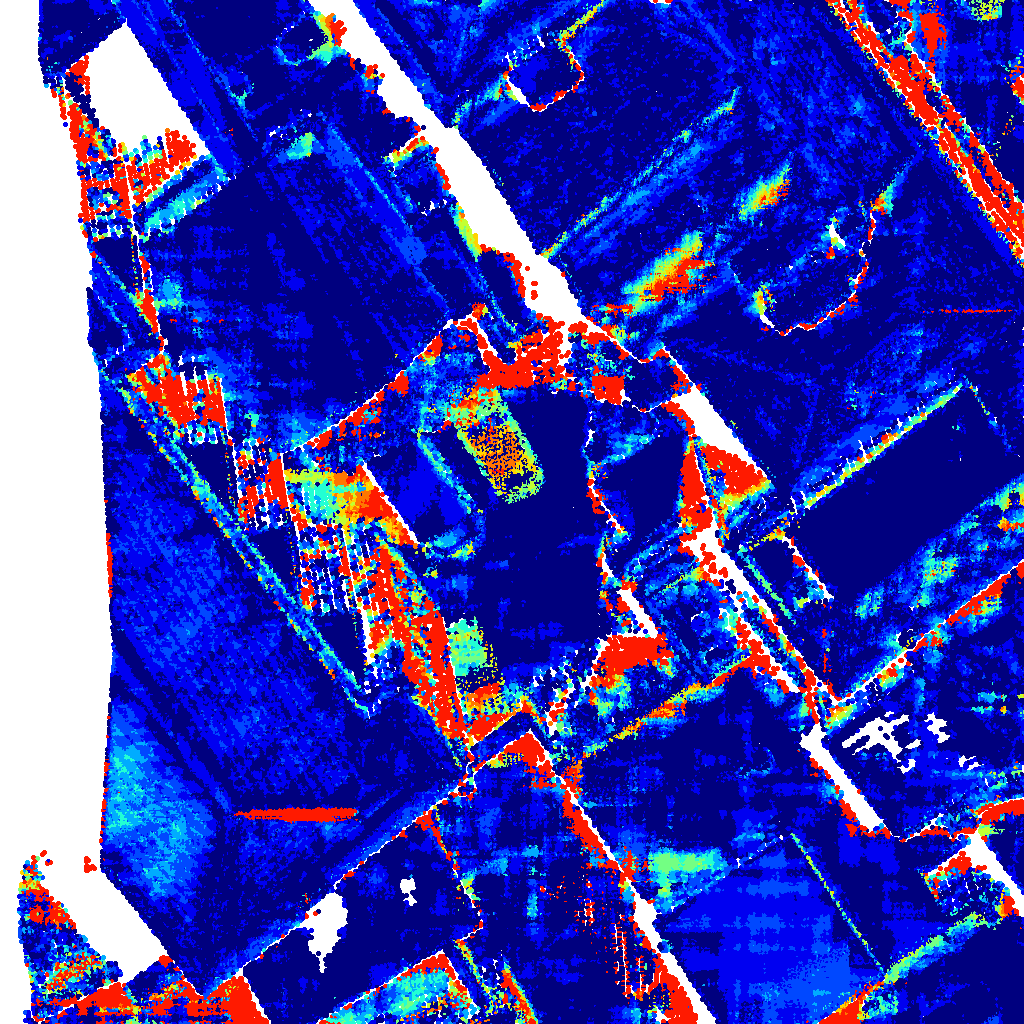}
		\includegraphics[width=\linewidth]{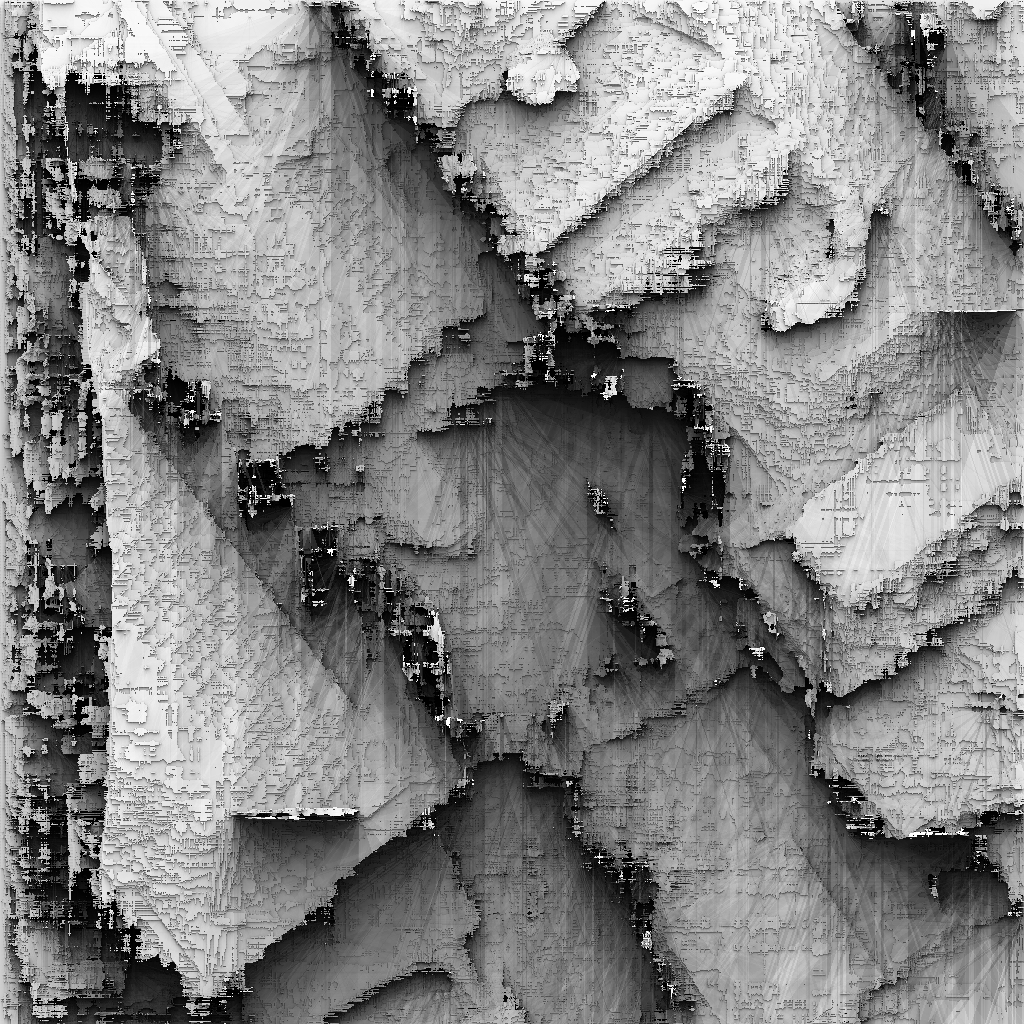}
		\centering{\tiny MC-CNN}
	\end{minipage}
	\begin{minipage}[t]{0.19\textwidth}
		\includegraphics[width=0.098\linewidth]{figures_supp/color_map.png}
		\includegraphics[width=0.85\linewidth]{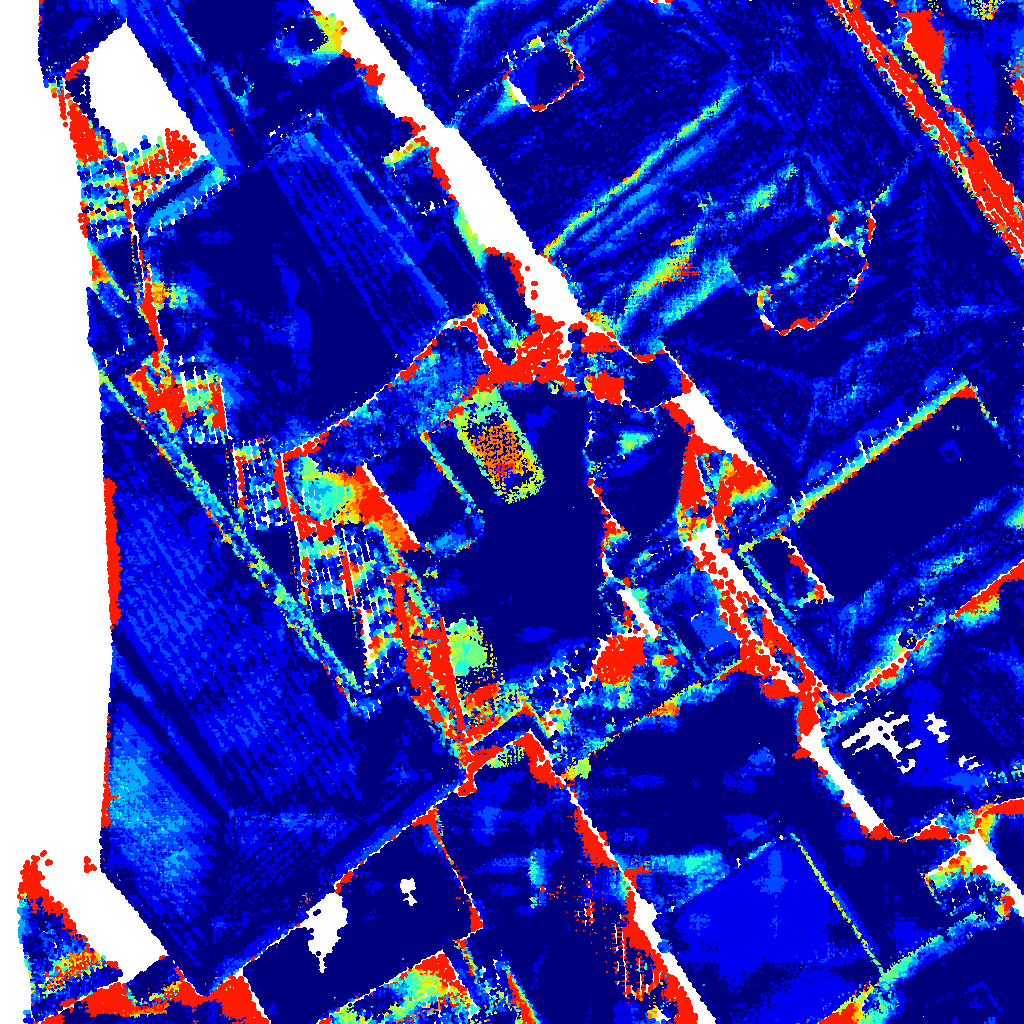}
		\includegraphics[width=\linewidth]{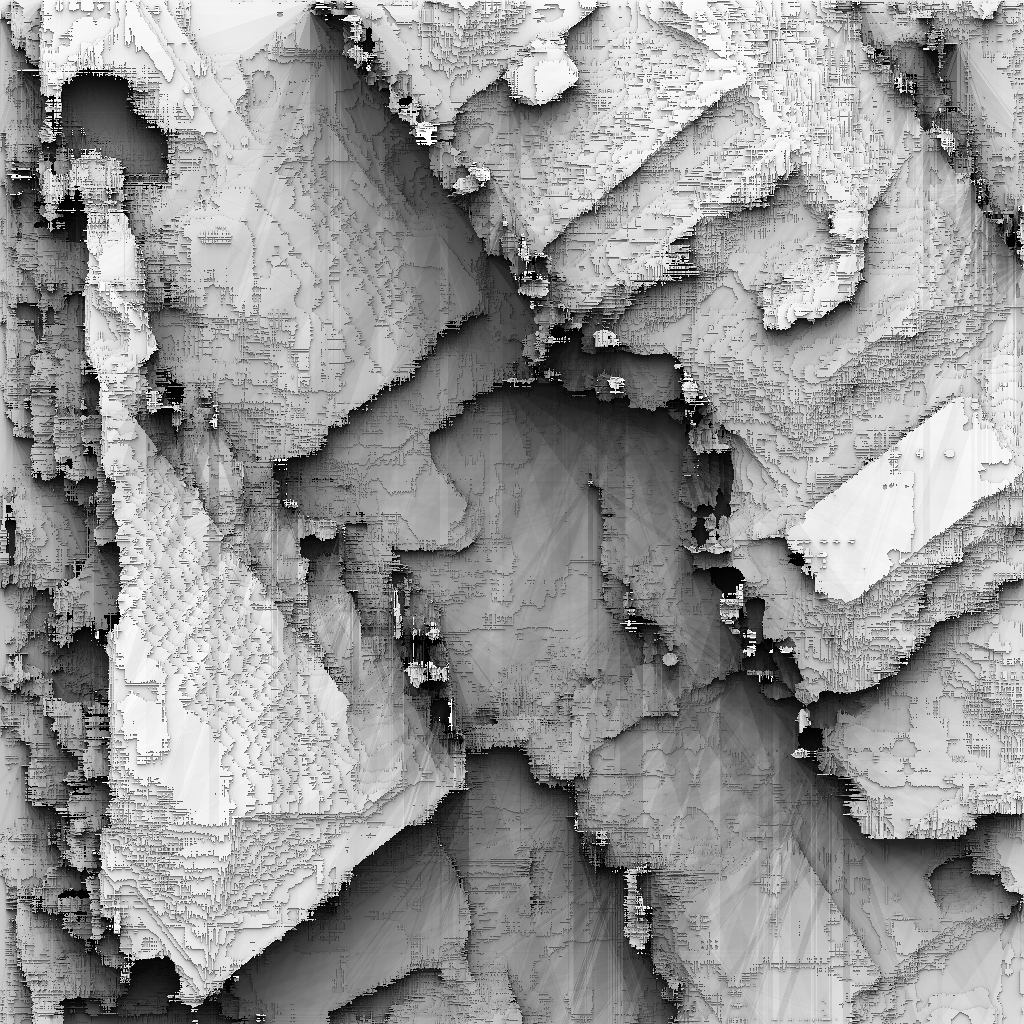}
		\centering{\tiny EfficientDeep}
	\end{minipage}
	\begin{minipage}[t]{0.19\textwidth}
		\includegraphics[width=0.098\linewidth]{figures_supp/color_map.png}
		\includegraphics[width=0.85\linewidth]{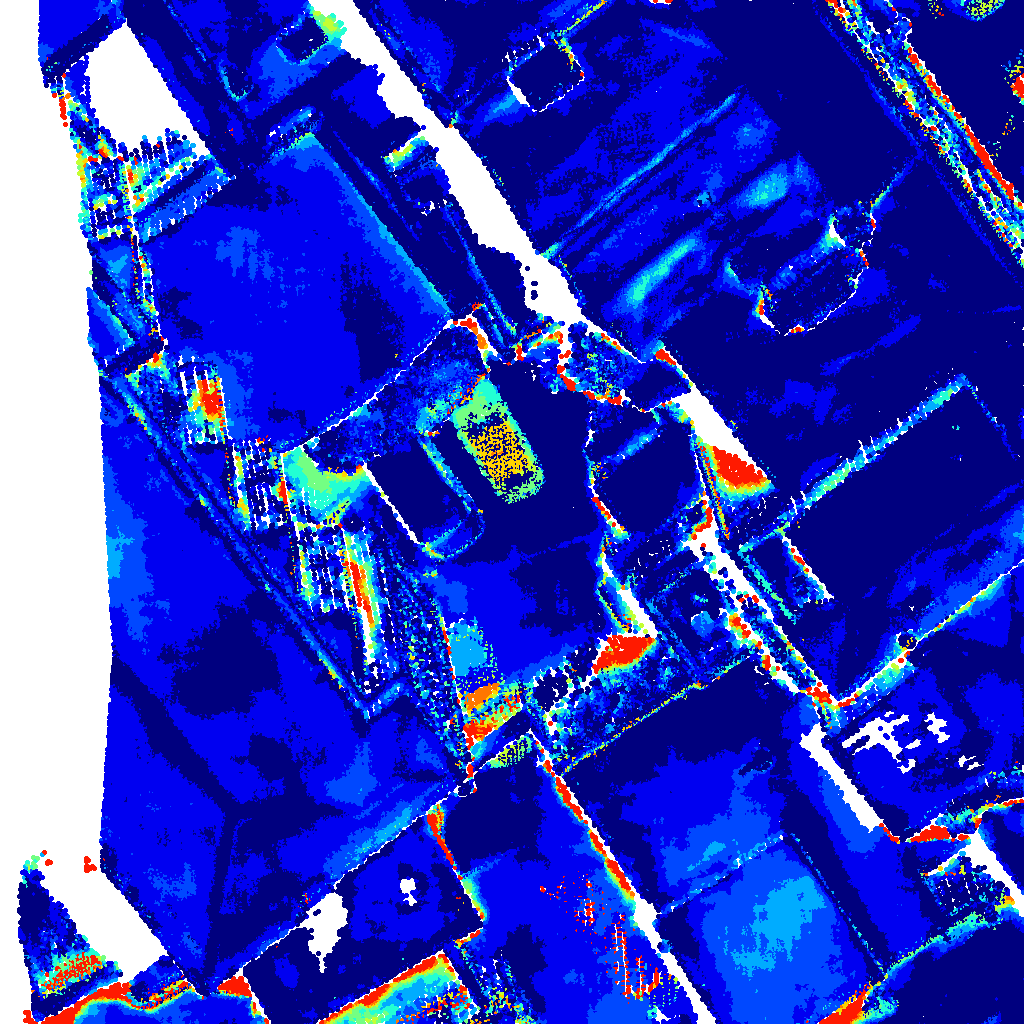}
		\includegraphics[width=\linewidth]{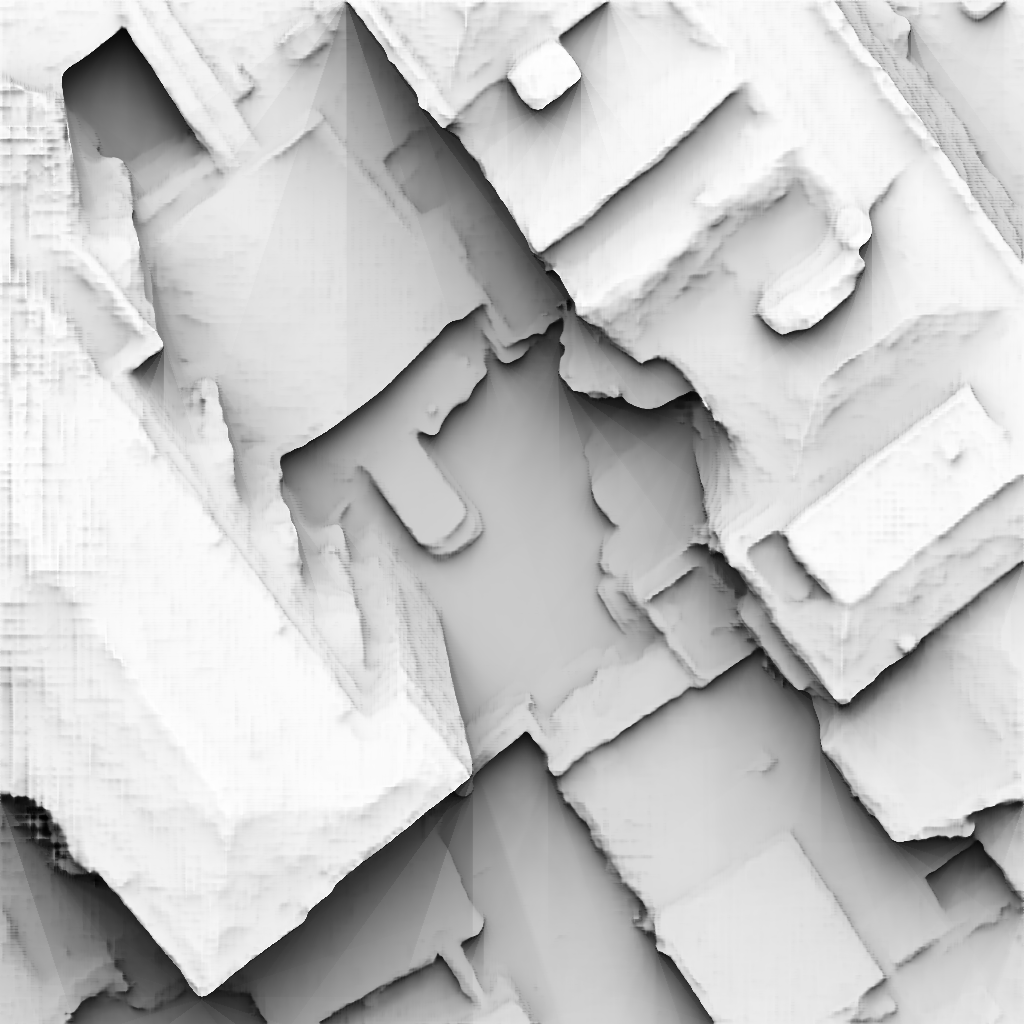}
		\centering{\tiny PSM net}
	\end{minipage}
	\begin{minipage}[t]{0.19\textwidth}	
		\includegraphics[width=0.098\linewidth]{figures_supp/color_map.png}
		\includegraphics[width=0.85\linewidth]{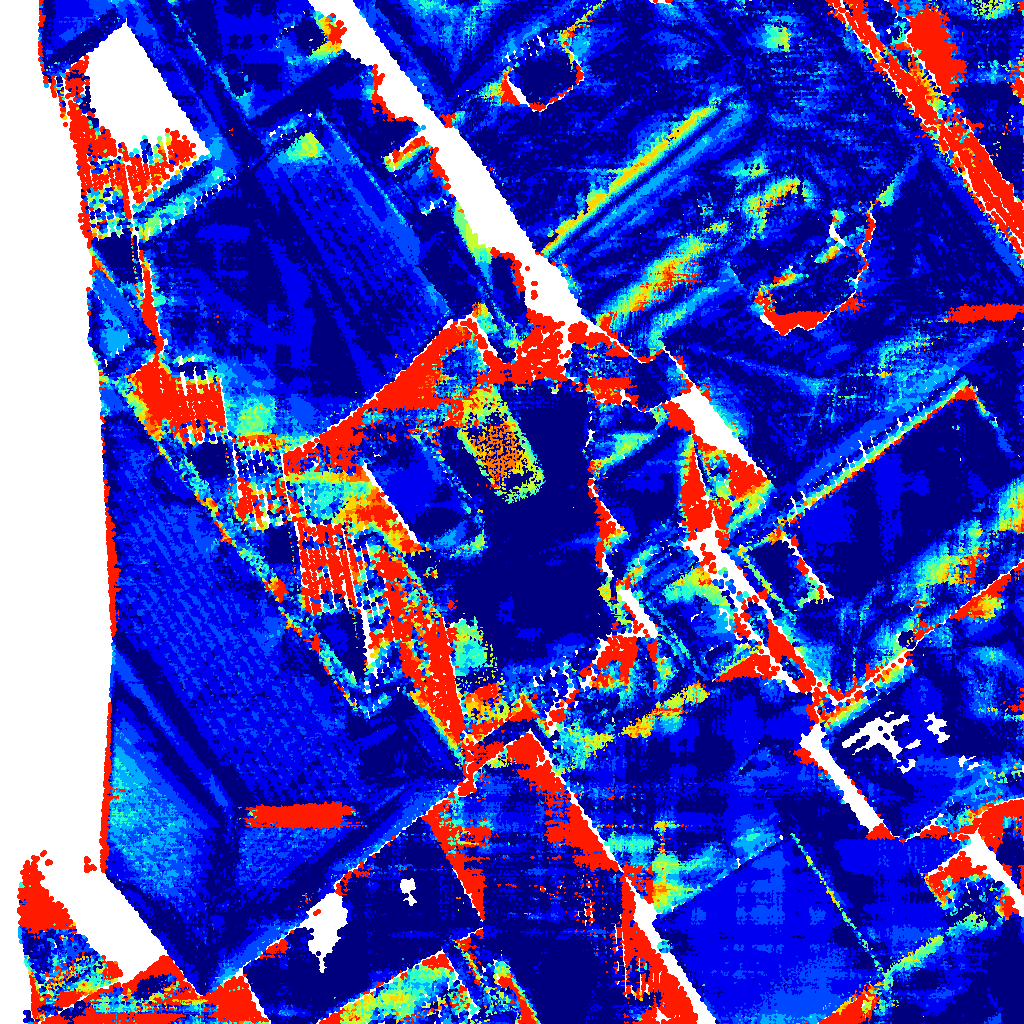}
		\includegraphics[width=\linewidth]{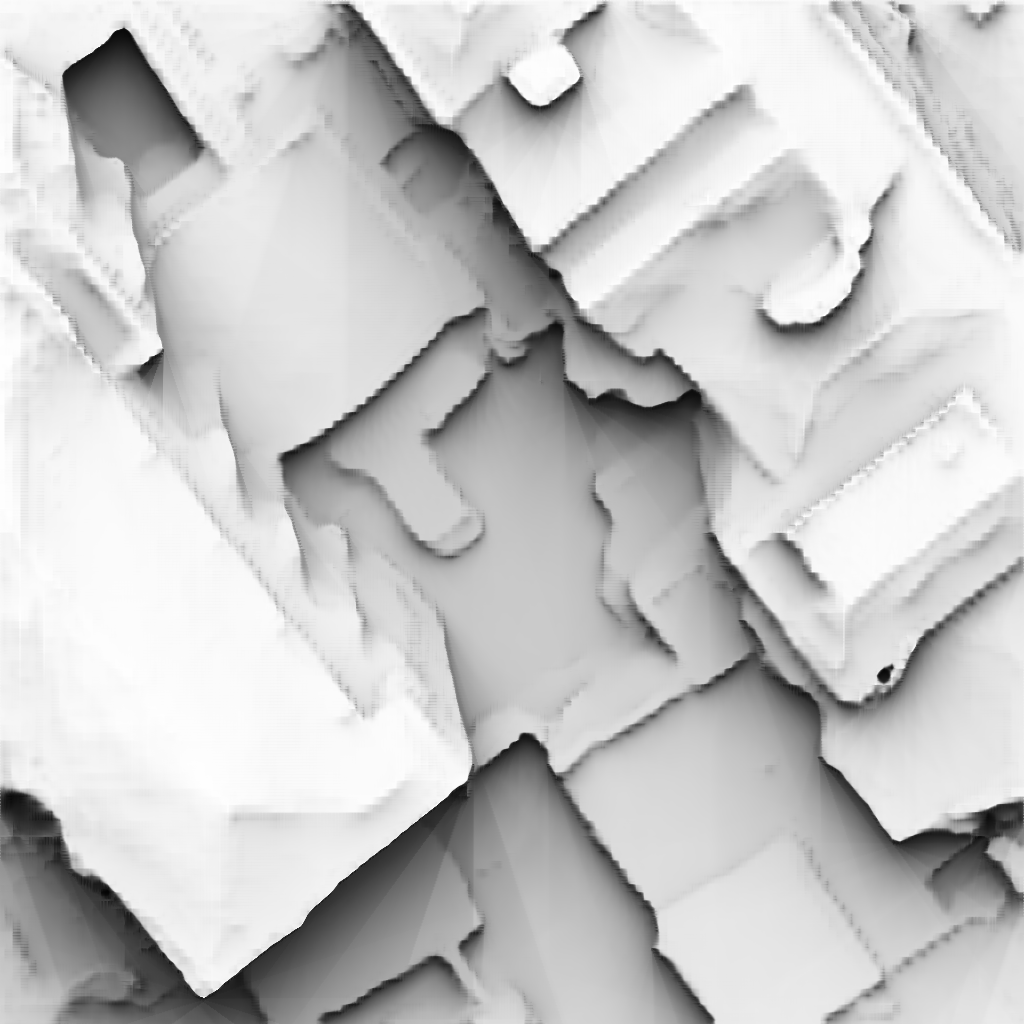}
		\centering{\tiny HRS net}
	\end{minipage}
	\begin{minipage}[t]{0.19\textwidth}	
		\includegraphics[width=0.098\linewidth]{figures_supp/color_map.png}
		\includegraphics[width=0.85\linewidth]{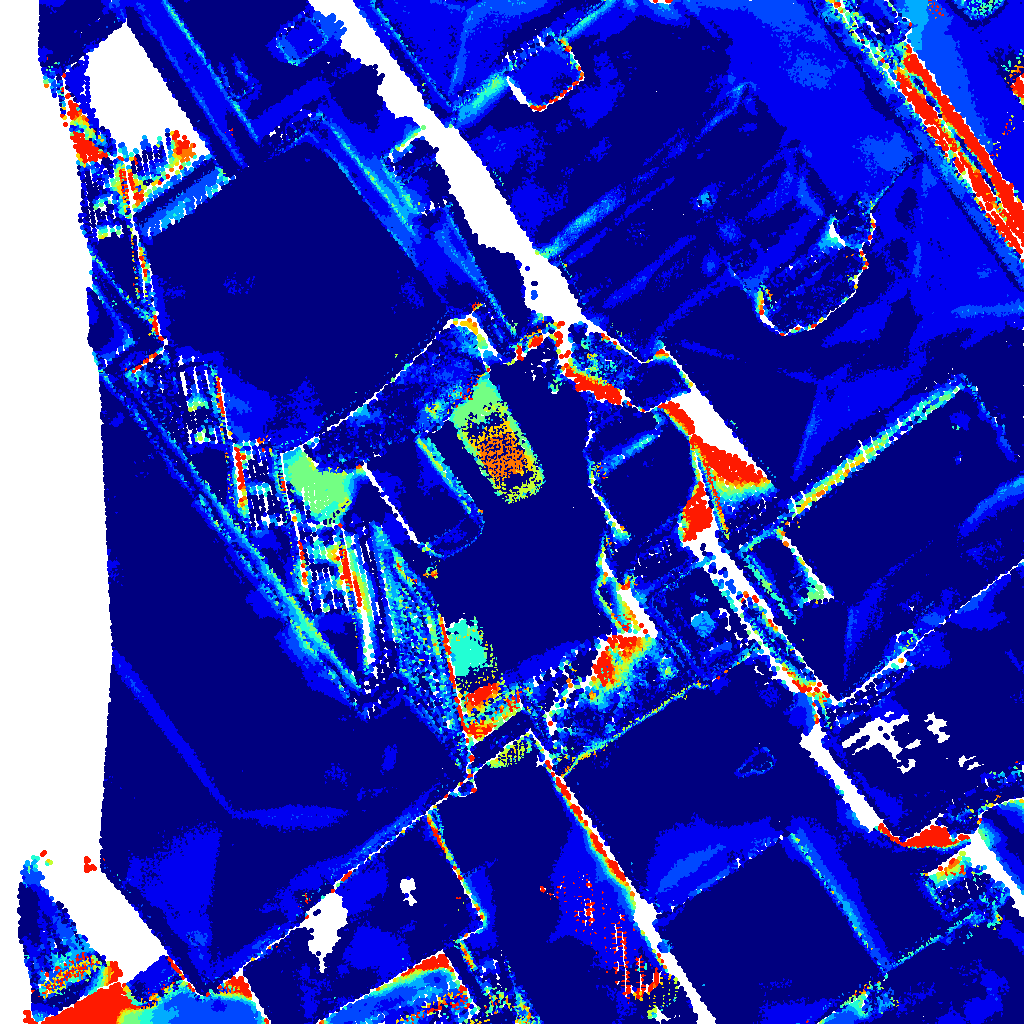}
		\includegraphics[width=\linewidth]{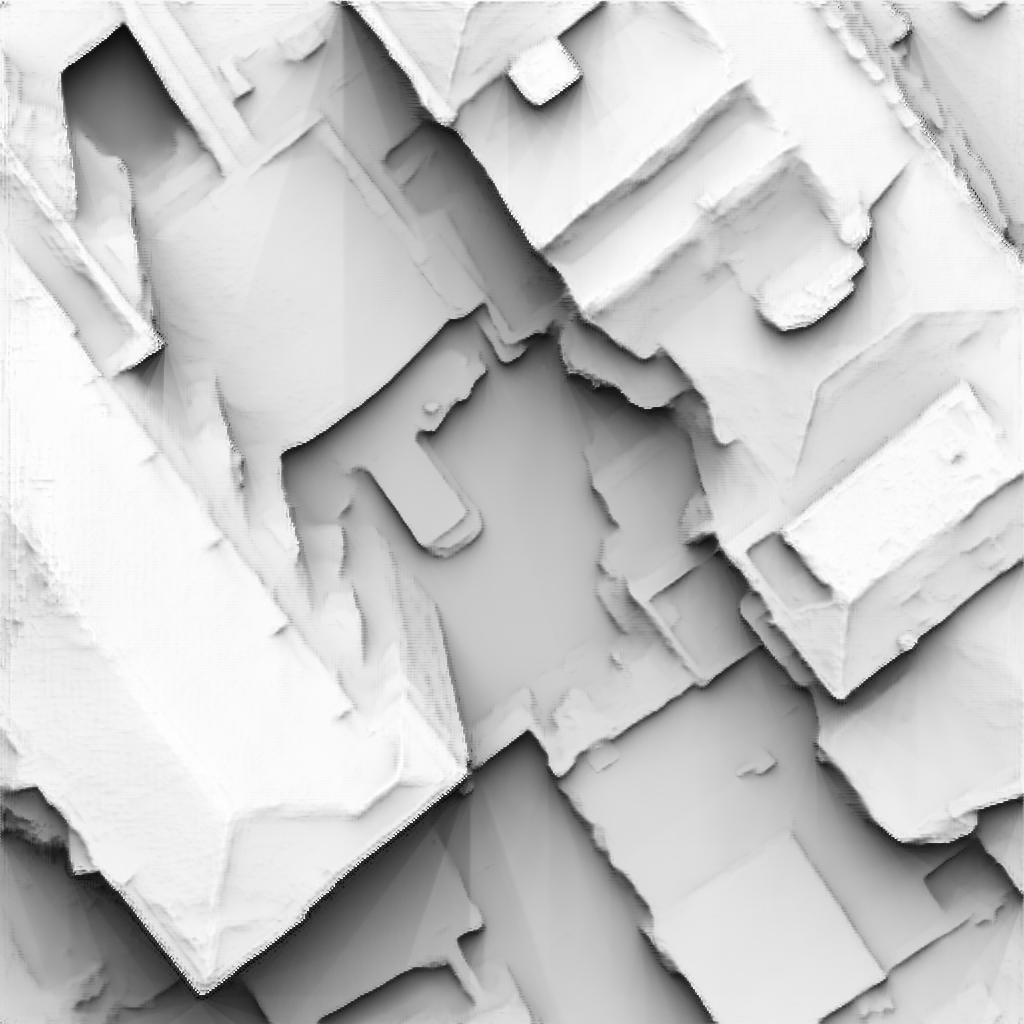}
		\centering{\tiny DeepPruner}
	\end{minipage}
	\begin{minipage}[t]{0.19\textwidth}	
		\includegraphics[width=0.098\linewidth]{figures_supp/color_map.png}
		\includegraphics[width=0.85\linewidth]{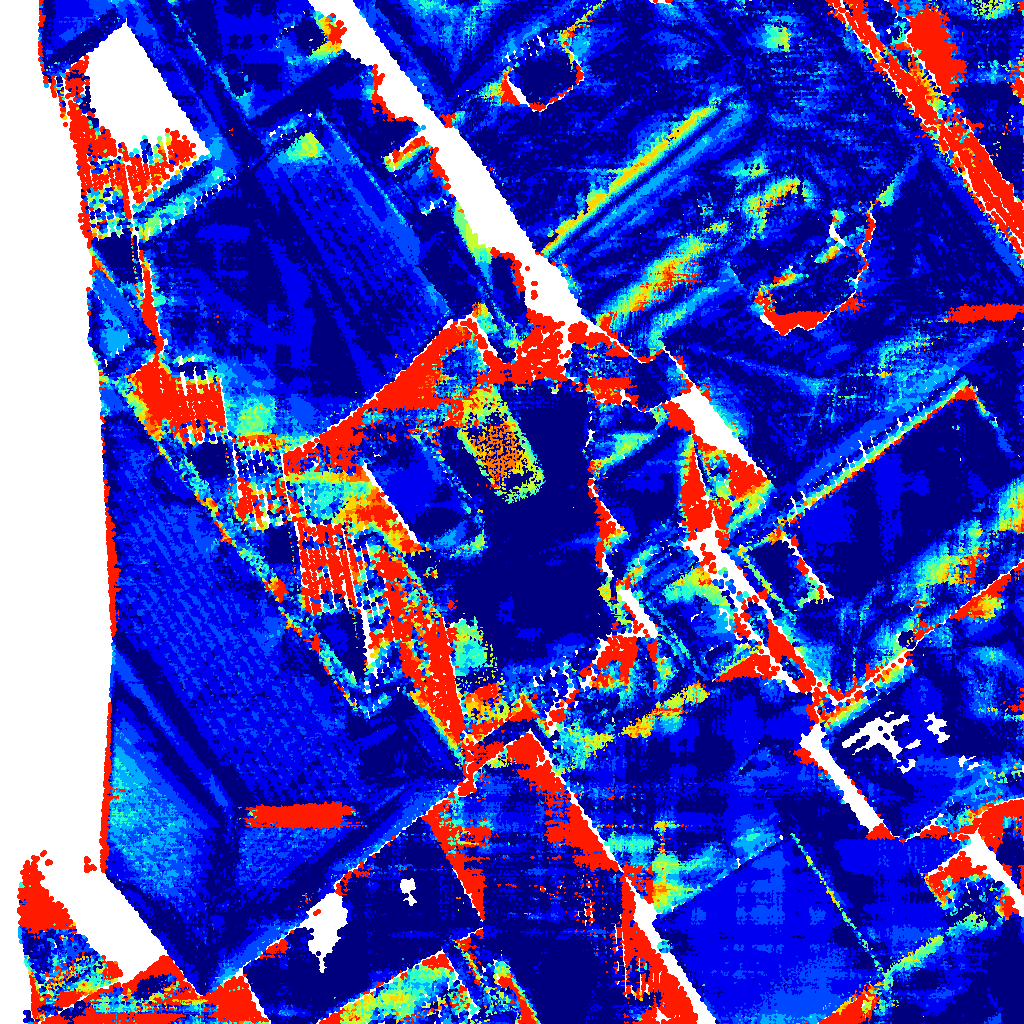}
		\includegraphics[width=\linewidth]{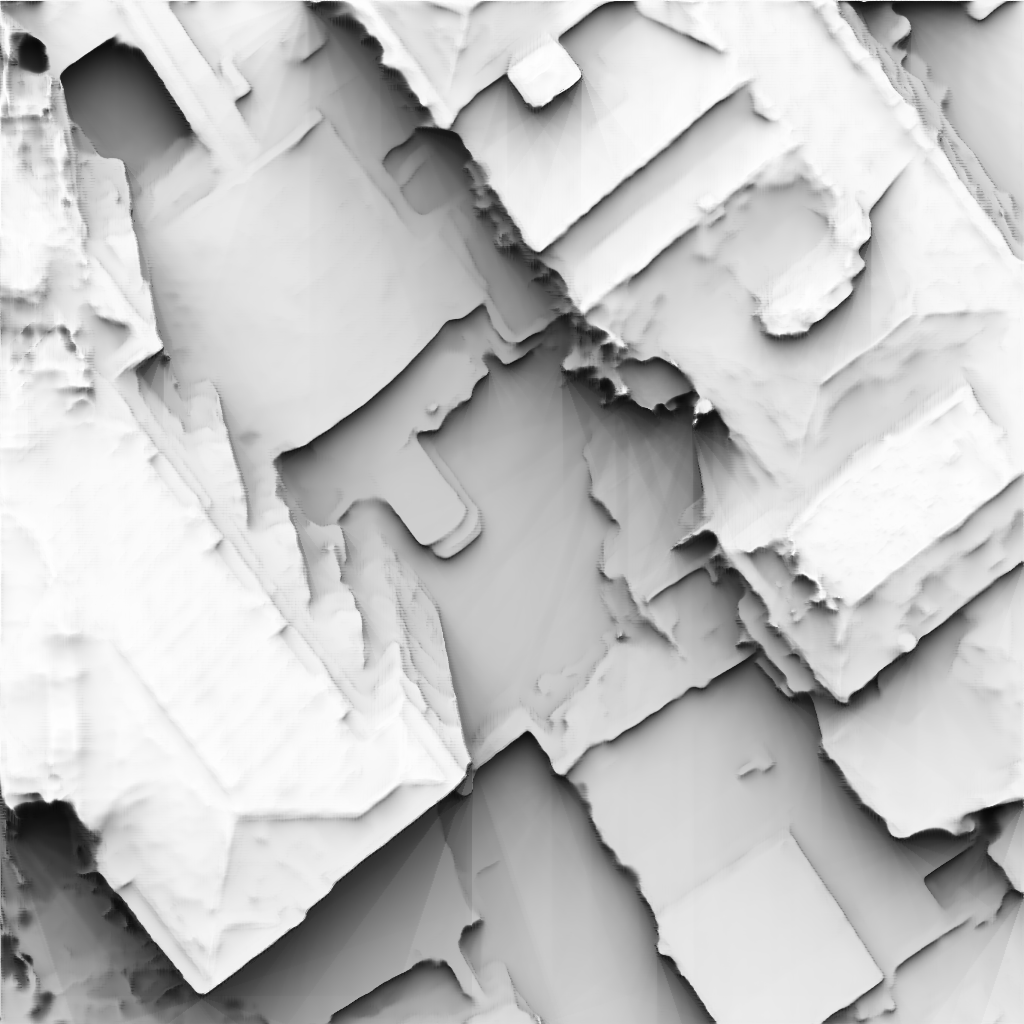}
		\centering{\tiny GANet}
	\end{minipage}
	\begin{minipage}[t]{0.19\textwidth}	
		\includegraphics[width=0.098\linewidth]{figures_supp/color_map.png}
		\includegraphics[width=0.85\linewidth]{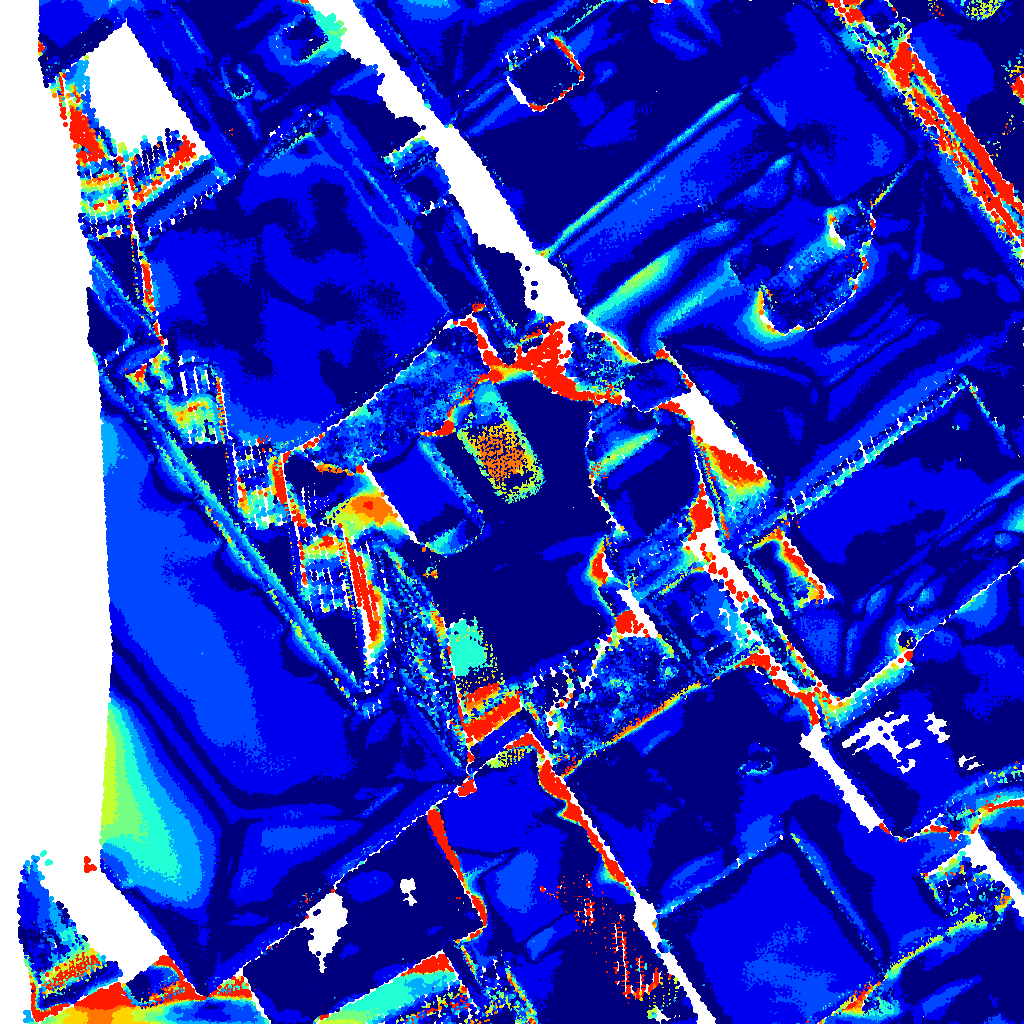}
		\includegraphics[width=\linewidth]{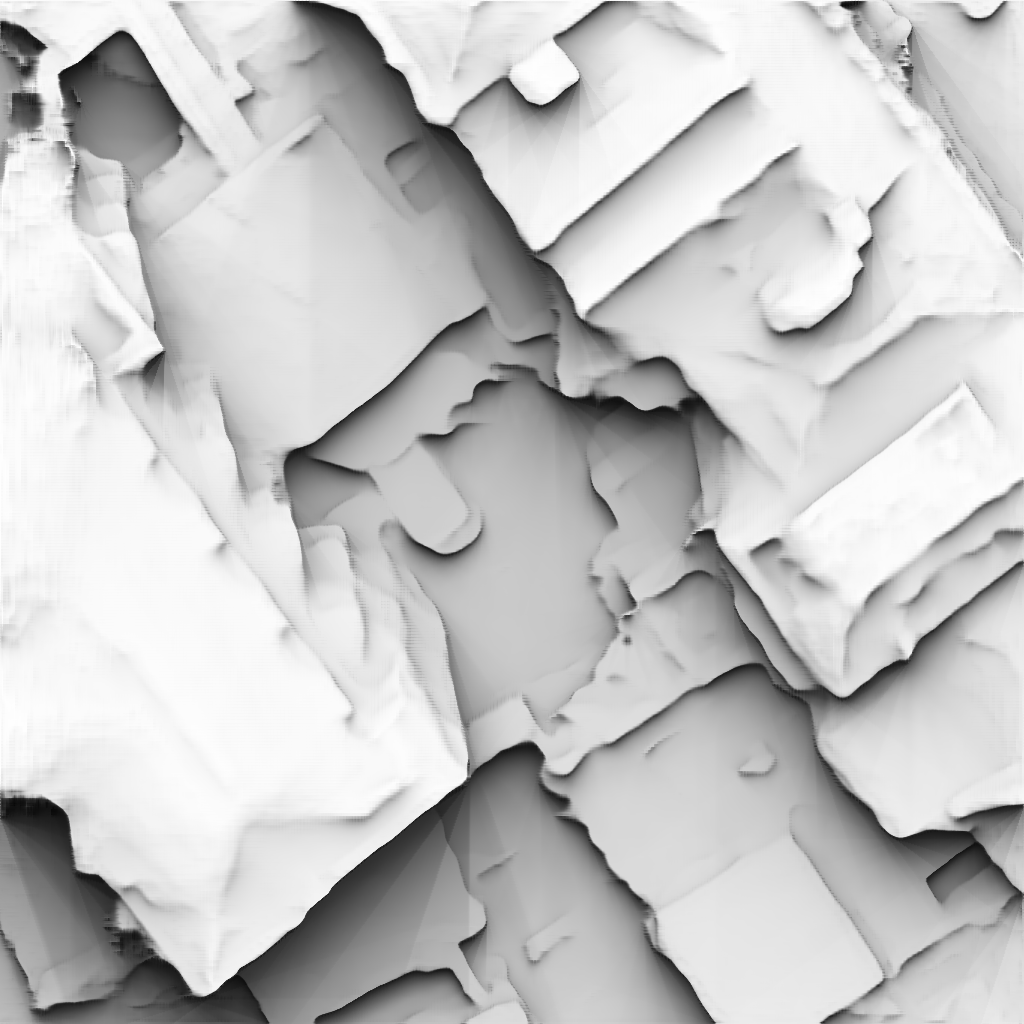}
		\centering{\tiny LEAStereo}
	\end{minipage}
	\caption{Error map and disparity visualization on building area for DublinCity dataset.}
	\label{Figure.dublinbulding}
\end{figure}

\begin{figure}[tp]
	\begin{minipage}[t]{0.19\textwidth}
		\includegraphics[width=0.098\linewidth]{figures_supp/color_map.png}
		\includegraphics[width=0.85\linewidth]{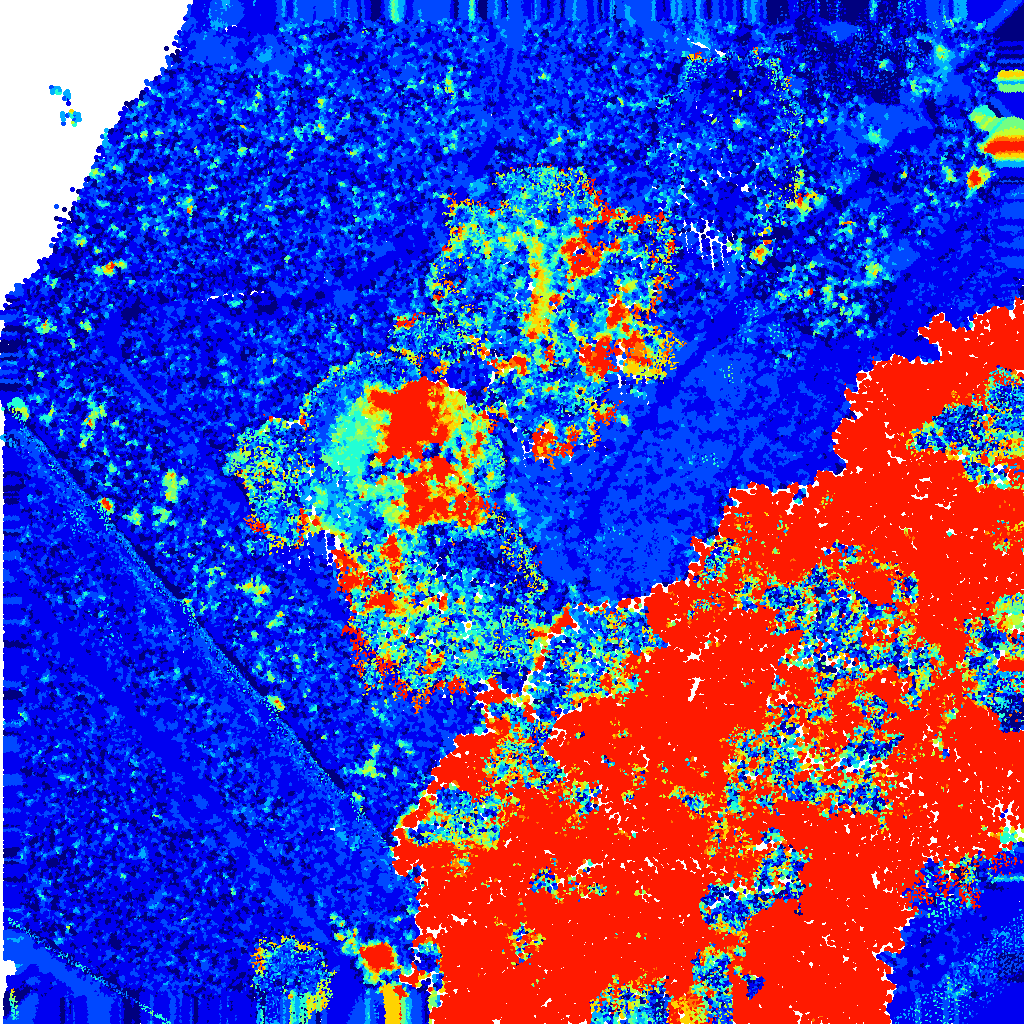}
		\includegraphics[width=\linewidth]{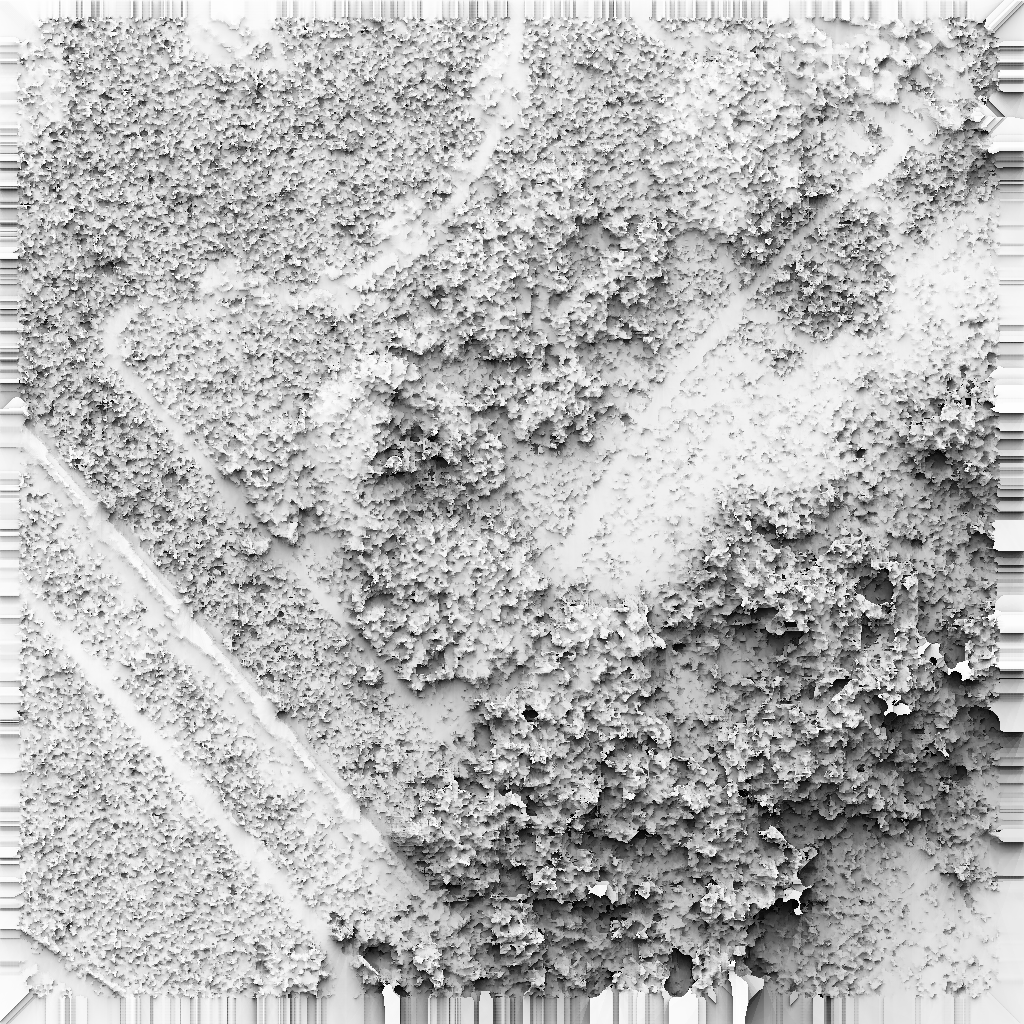}
		\centering{\tiny MICMAC}
	\end{minipage}
	\begin{minipage}[t]{0.19\textwidth}	
		\includegraphics[width=0.098\linewidth]{figures_supp/color_map.png}
		\includegraphics[width=0.85\linewidth]{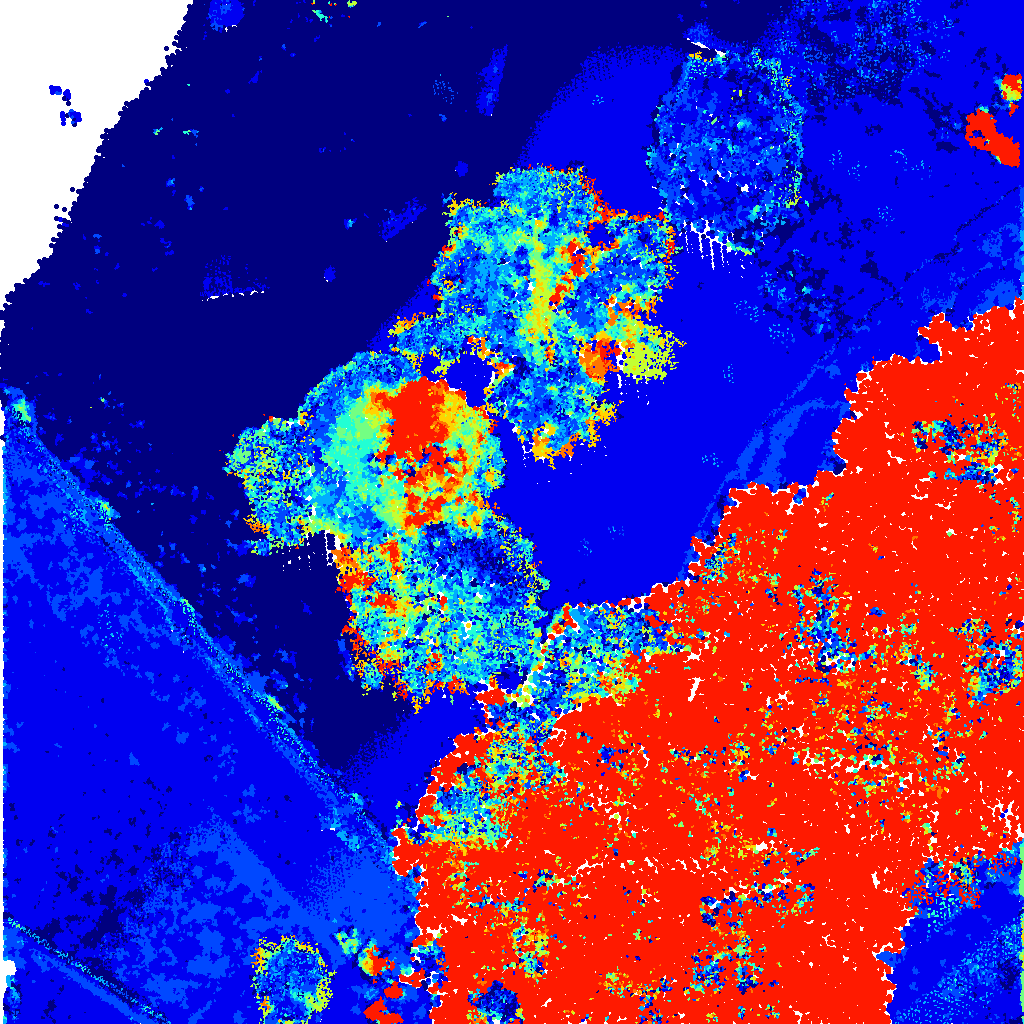}
		\includegraphics[width=\linewidth]{figures_supp/dublin/pyr_micmac/3489_DUBLIN_AREA_2KM2_rgb_124791_id184c1_20150326120501_3489_DUBLIN_AREA_2KM2_rgb_124792_id185c1_20150326120502_0009_shade.png}
		\centering{\tiny SGM(CUDA)}
	\end{minipage}
	\begin{minipage}[t]{0.19\textwidth}	
		\includegraphics[width=0.098\linewidth]{figures_supp/color_map.png}
		\includegraphics[width=0.85\linewidth]{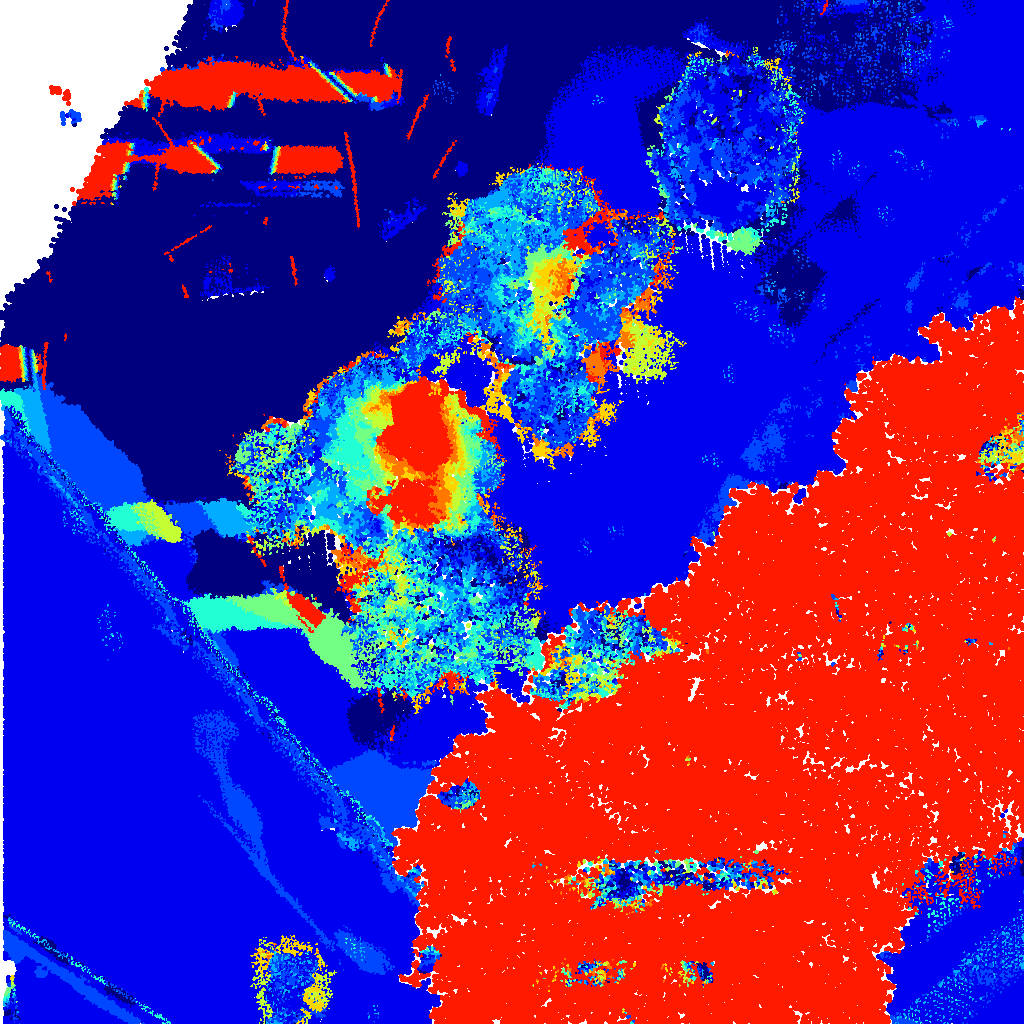}
		\includegraphics[width=\linewidth]{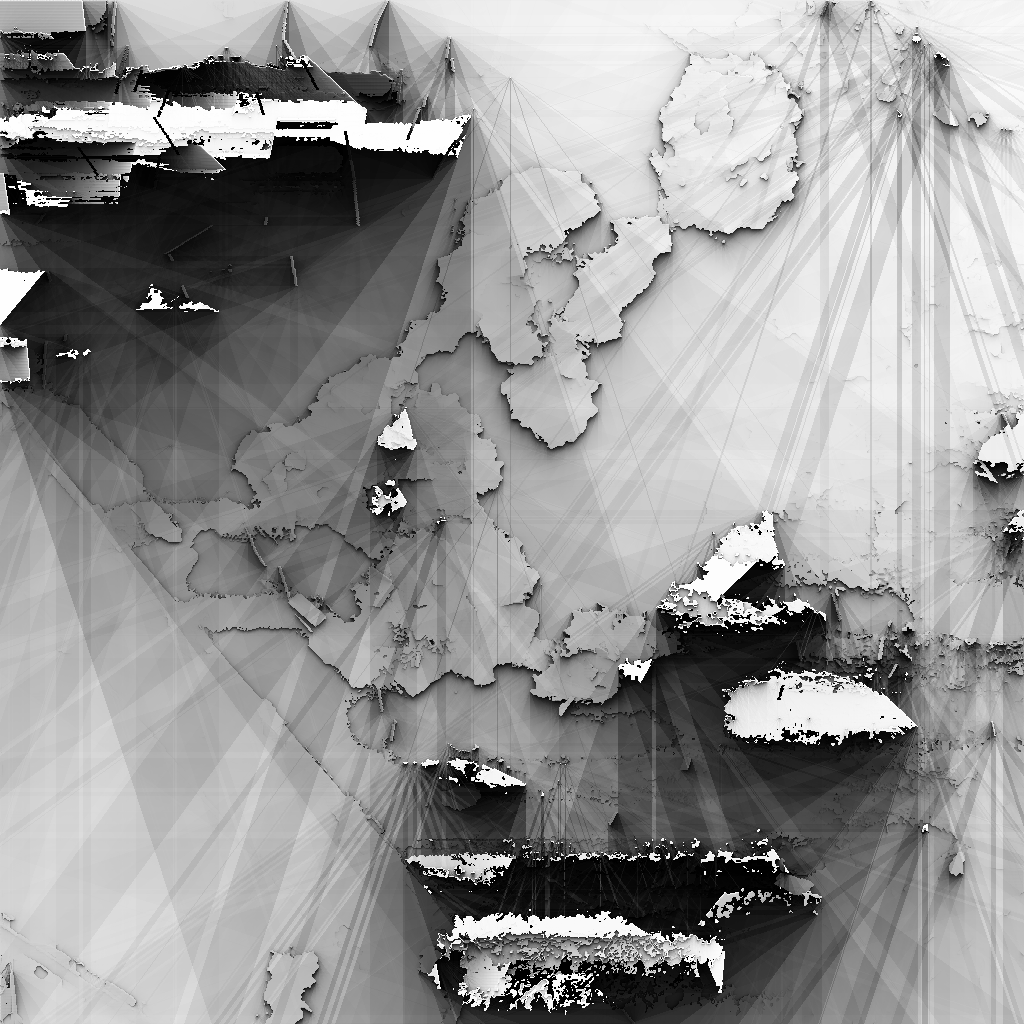}
		\centering{\tiny GraphCuts}
	\end{minipage}
	\begin{minipage}[t]{0.19\textwidth}	
		\includegraphics[width=0.098\linewidth]{figures_supp/color_map.png}
		\includegraphics[width=0.85\linewidth]{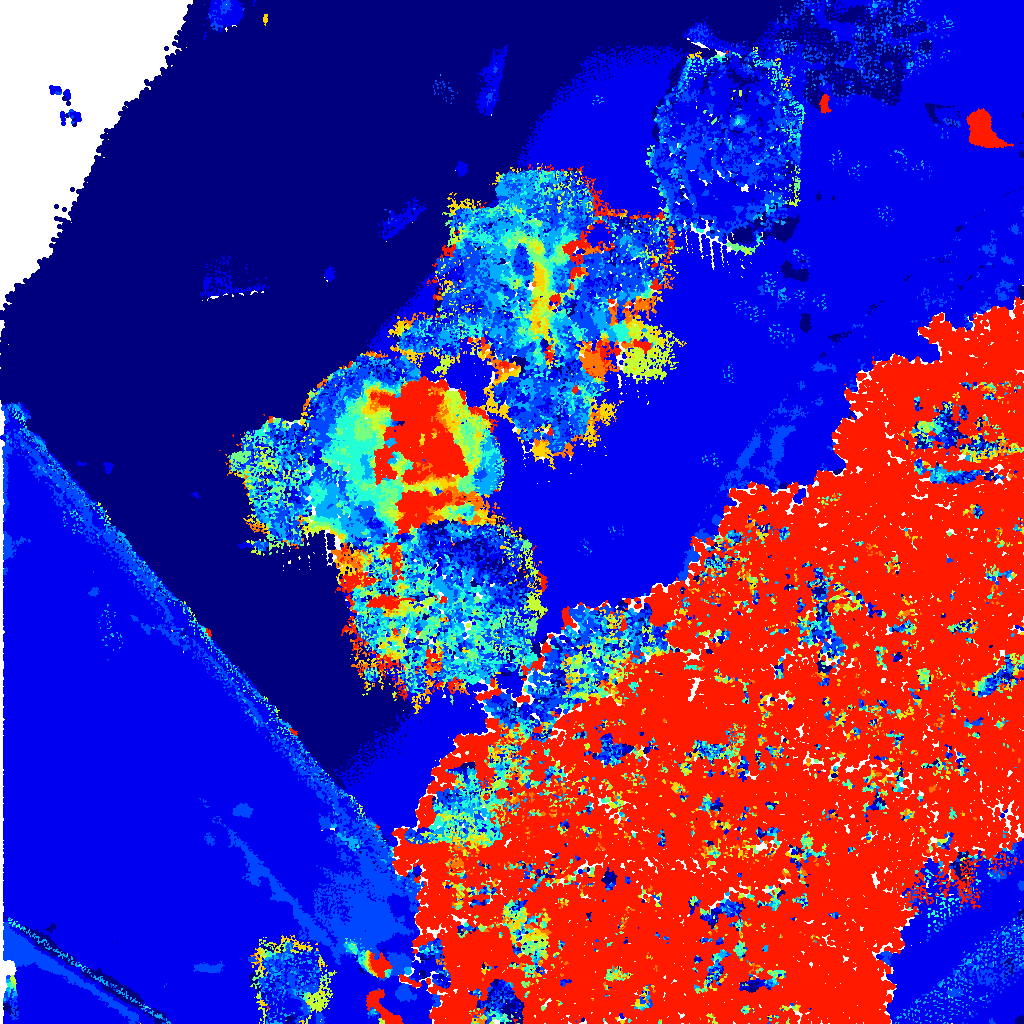}
		\includegraphics[width=\linewidth]{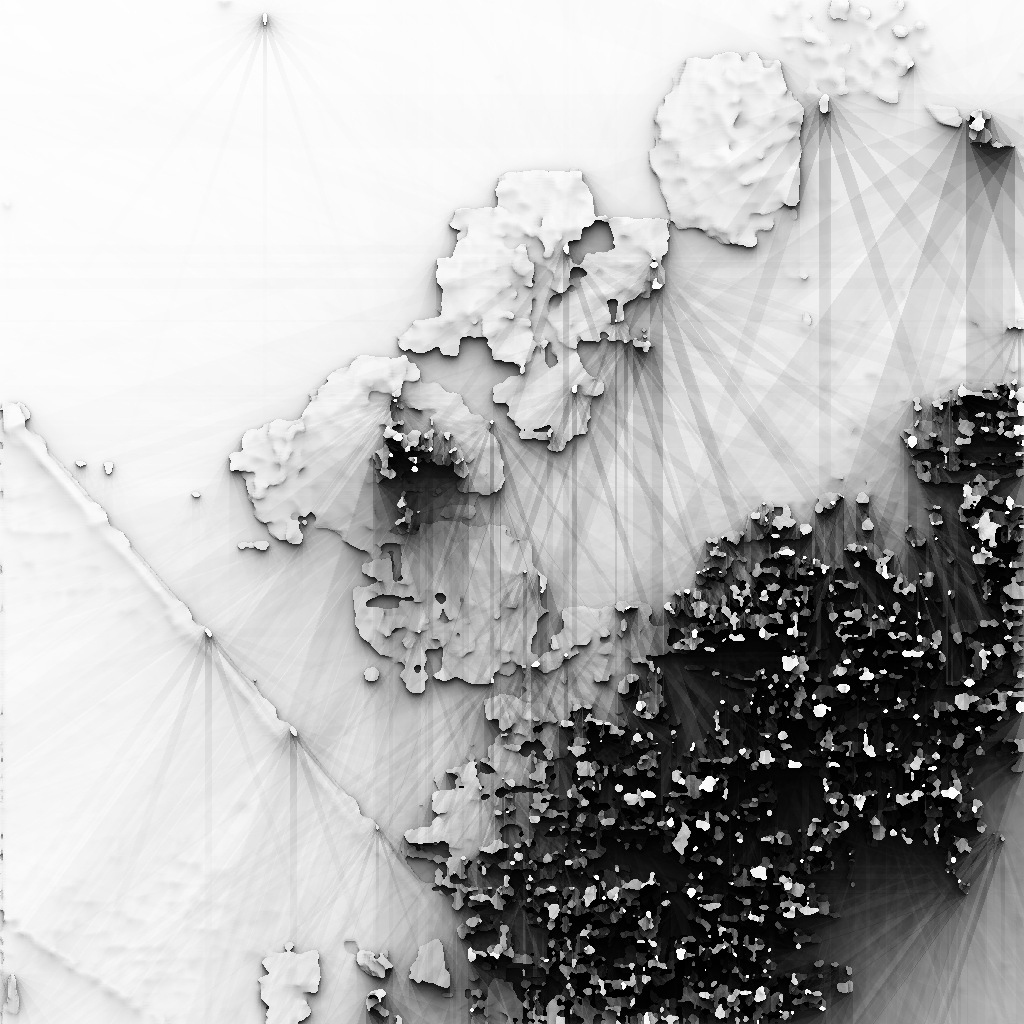}
		\centering{\tiny CBMV(SGM)}
	\end{minipage}
	\begin{minipage}[t]{0.19\textwidth} 
		\includegraphics[width=0.098\linewidth]{figures_supp/color_map.png}\includegraphics[width=0.85\linewidth]{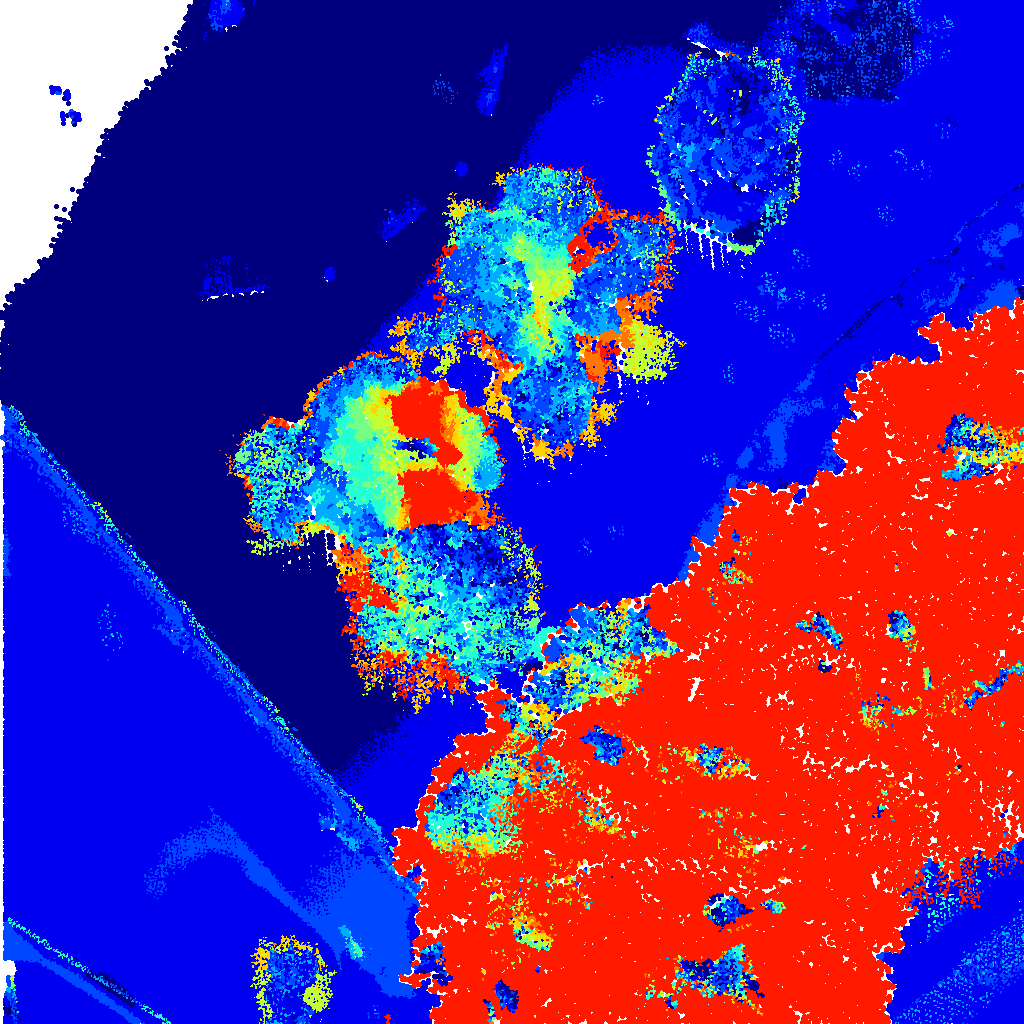}
		\includegraphics[width=\linewidth]{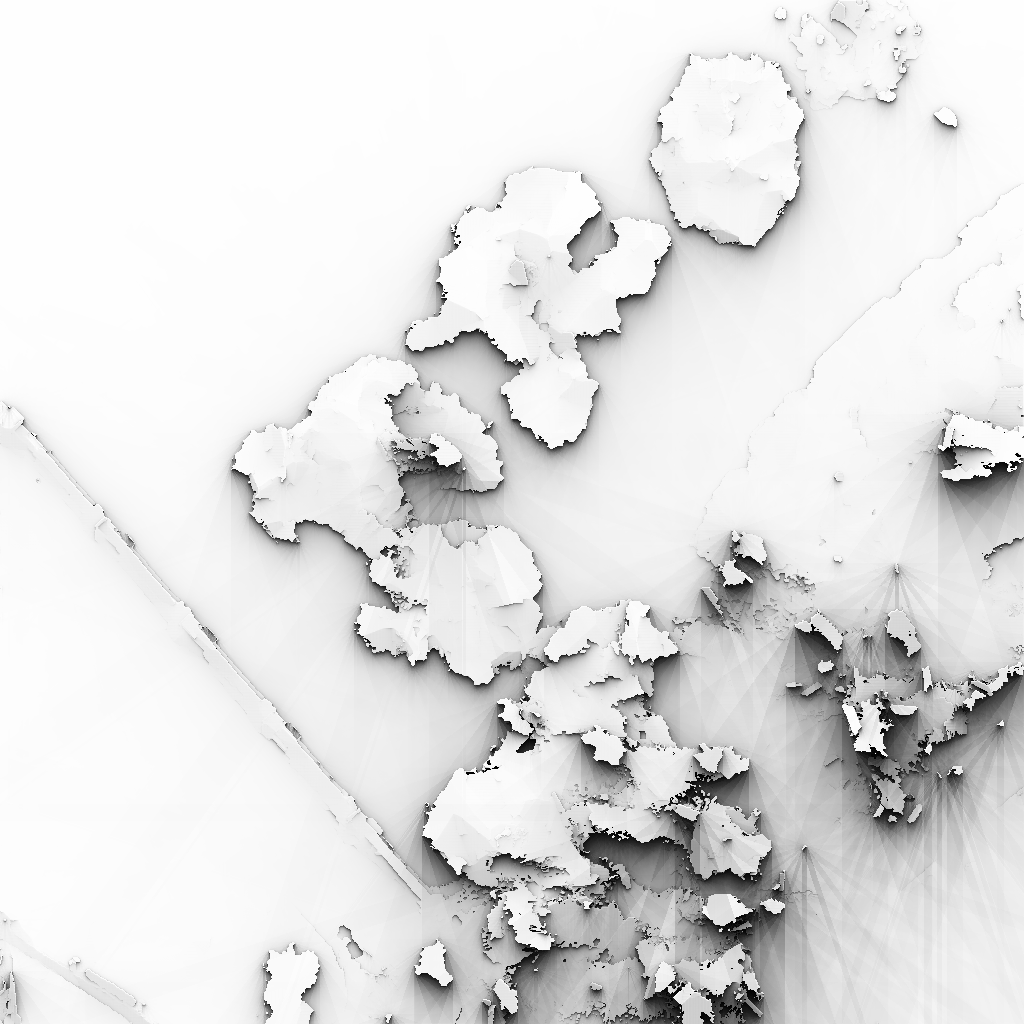}
		\centering{\tiny CBMV(GraphCuts)}
	\end{minipage}
	\begin{minipage}[t]{0.19\textwidth}
		\includegraphics[width=0.098\linewidth]{figures_supp/color_map.png}
		\includegraphics[width=0.85\linewidth]{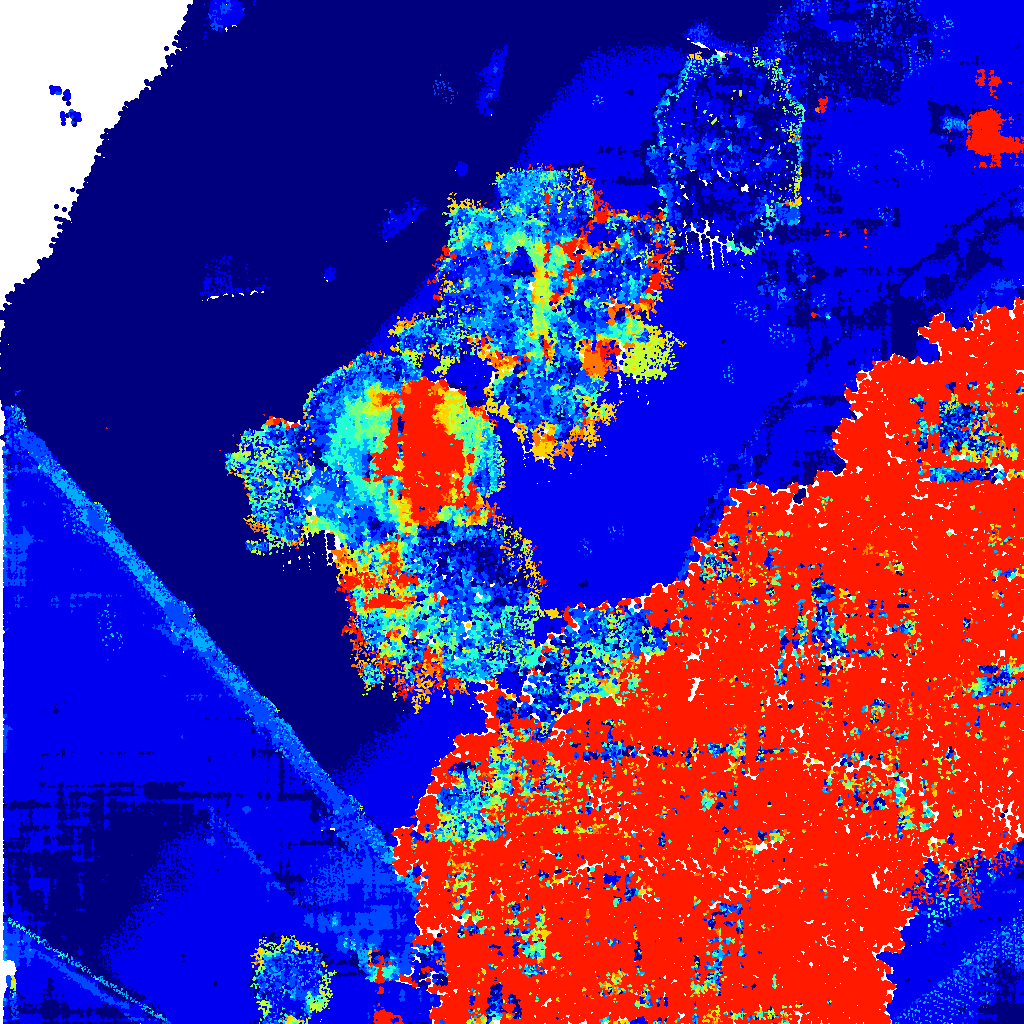}
		\includegraphics[width=\linewidth]{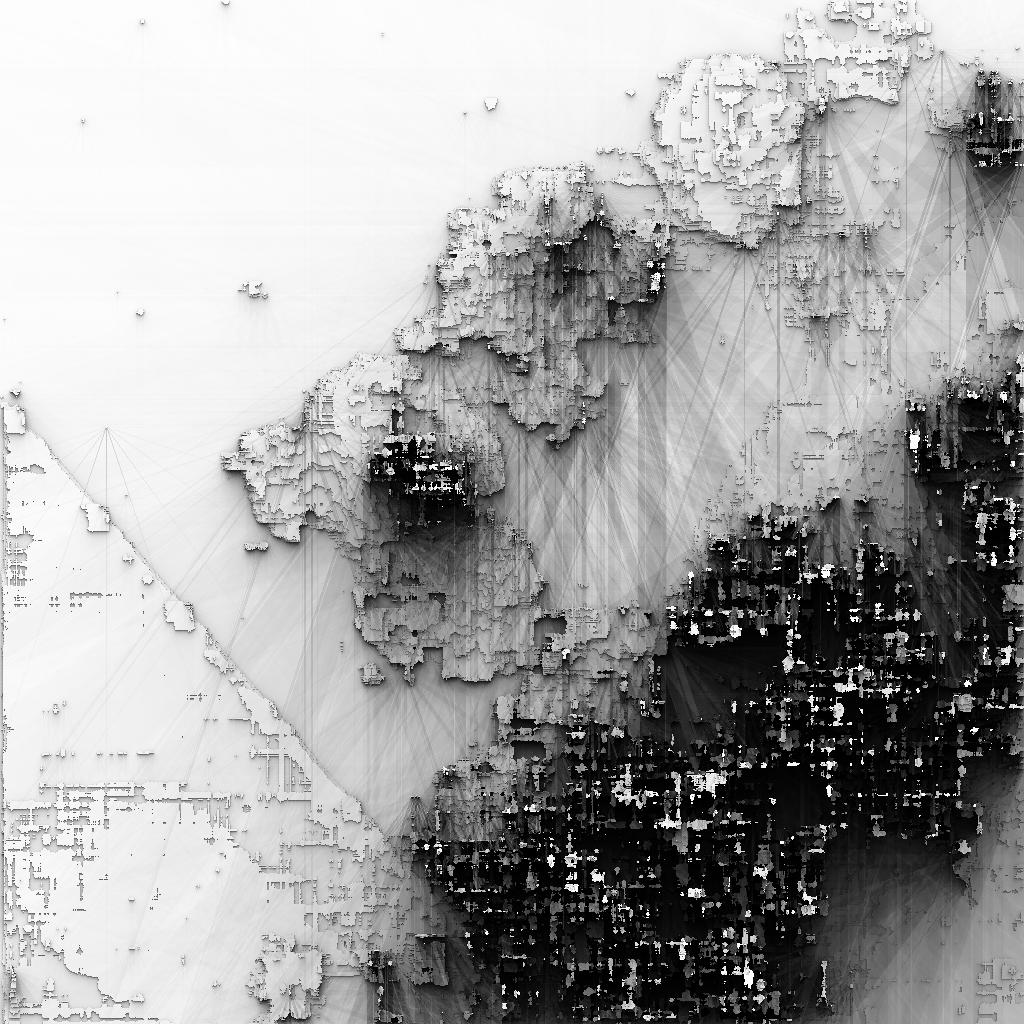}
		\centering{\tiny MC-CNN(KITTI)}
	\end{minipage}
	\begin{minipage}[t]{0.19\textwidth}
		\includegraphics[width=0.098\linewidth]{figures_supp/color_map.png}
		\includegraphics[width=0.85\linewidth]{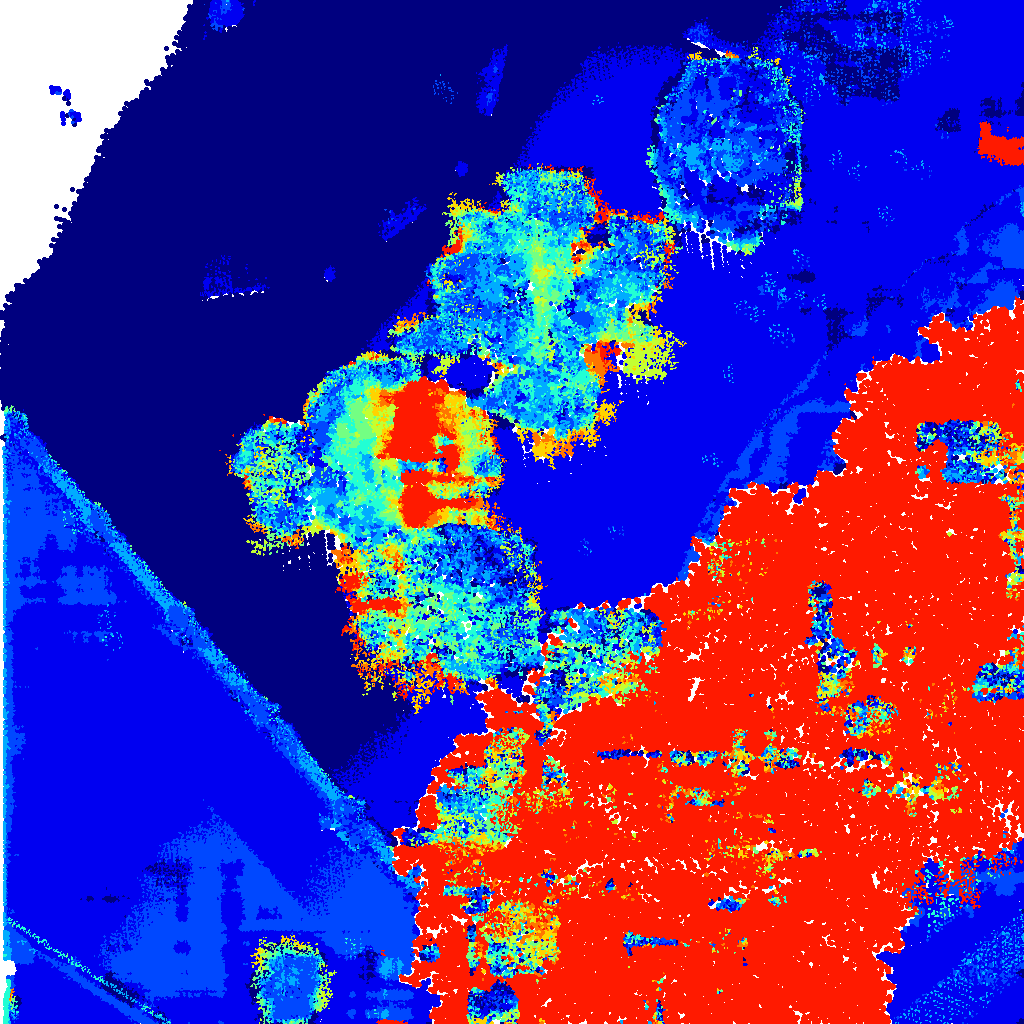}
		\includegraphics[width=\linewidth]{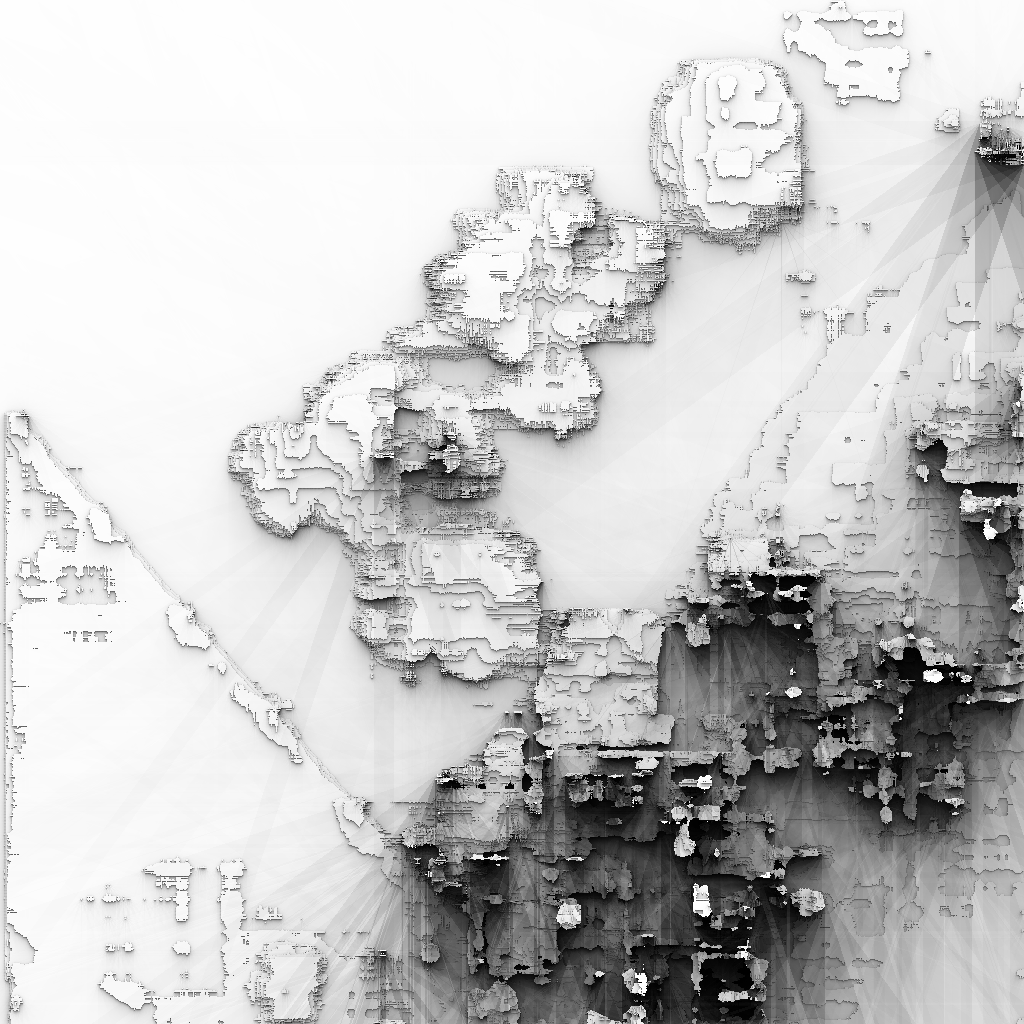}
		\centering{\tiny EfficientDeep(KITTI)}
	\end{minipage}
	\begin{minipage}[t]{0.19\textwidth}
		\includegraphics[width=0.098\linewidth]{figures_supp/color_map.png}
		\includegraphics[width=0.85\linewidth]{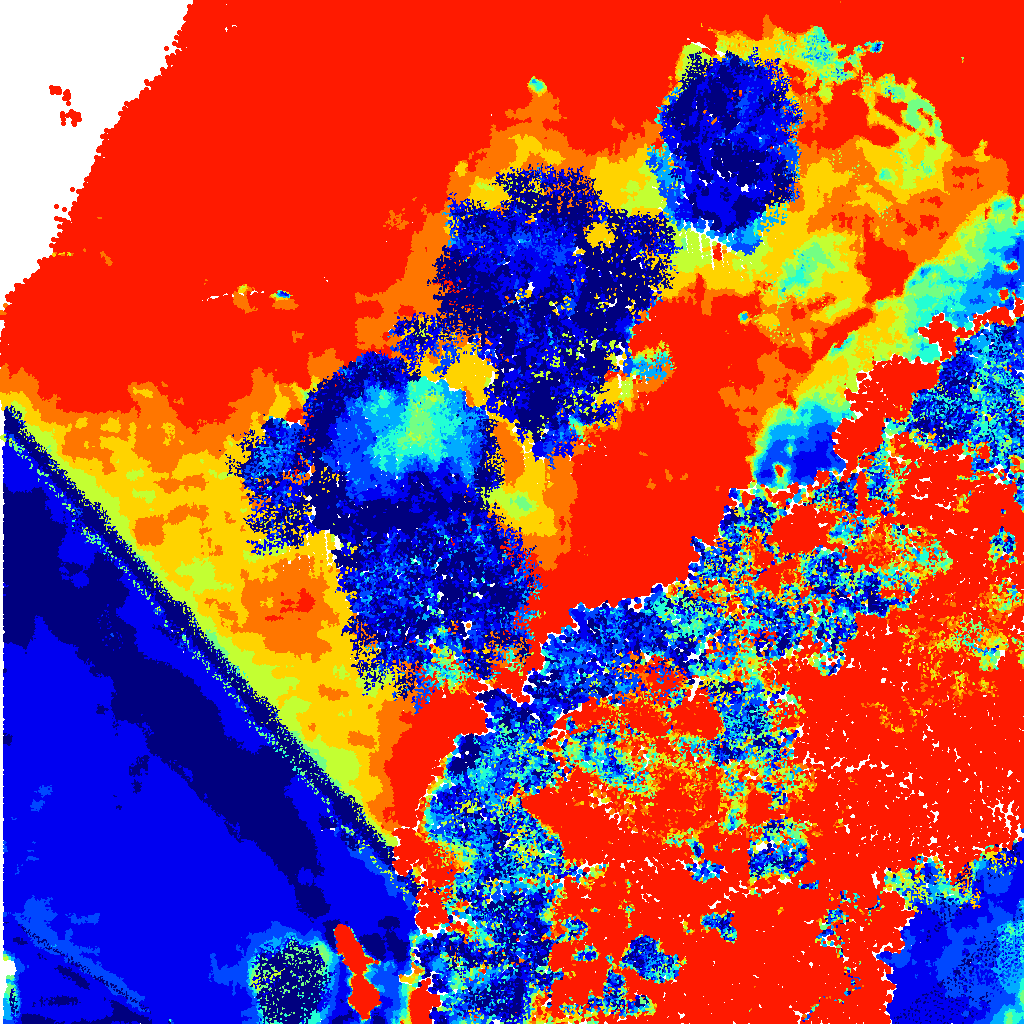}
		\includegraphics[width=\linewidth]{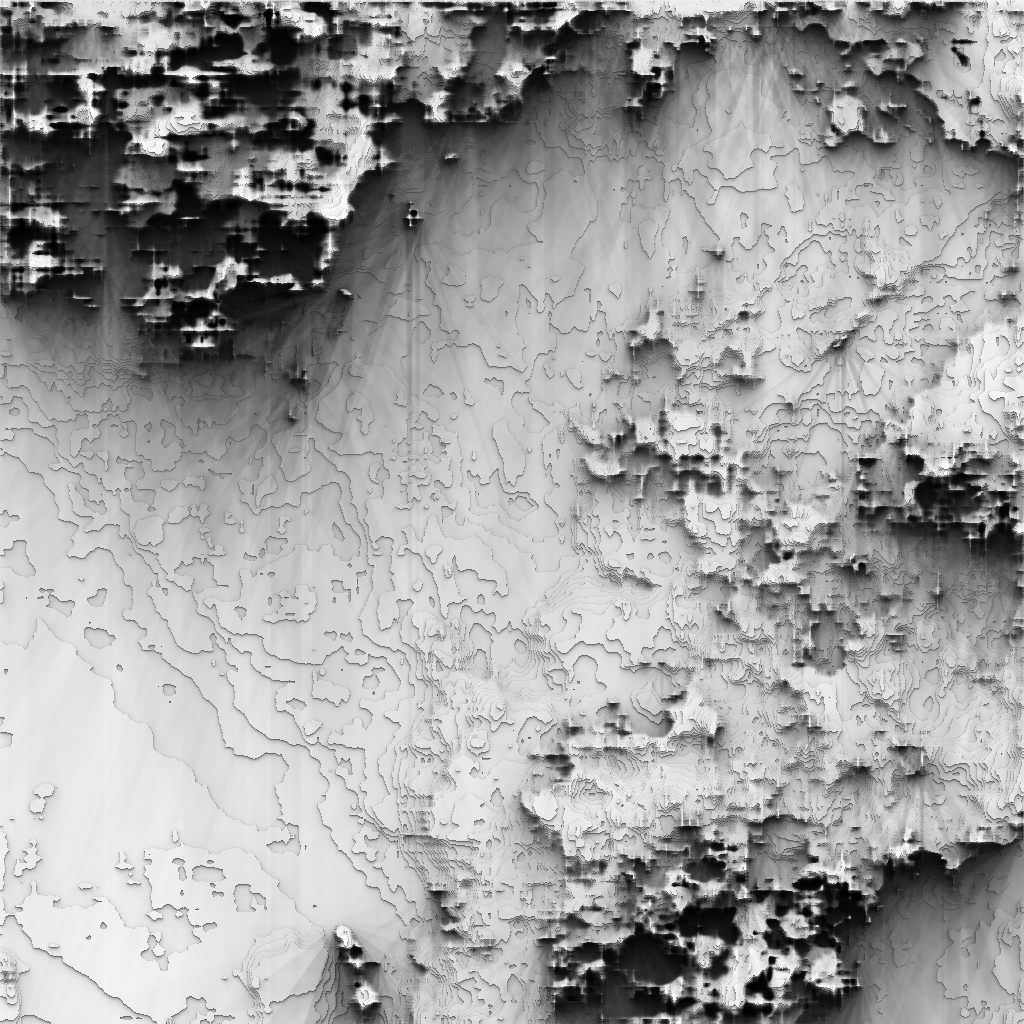}
		\centering{\tiny PSM net(KITTI)}
	\end{minipage}
	\begin{minipage}[t]{0.19\textwidth}	
		\includegraphics[width=0.098\linewidth]{figures_supp/color_map.png}
		\includegraphics[width=0.85\linewidth]{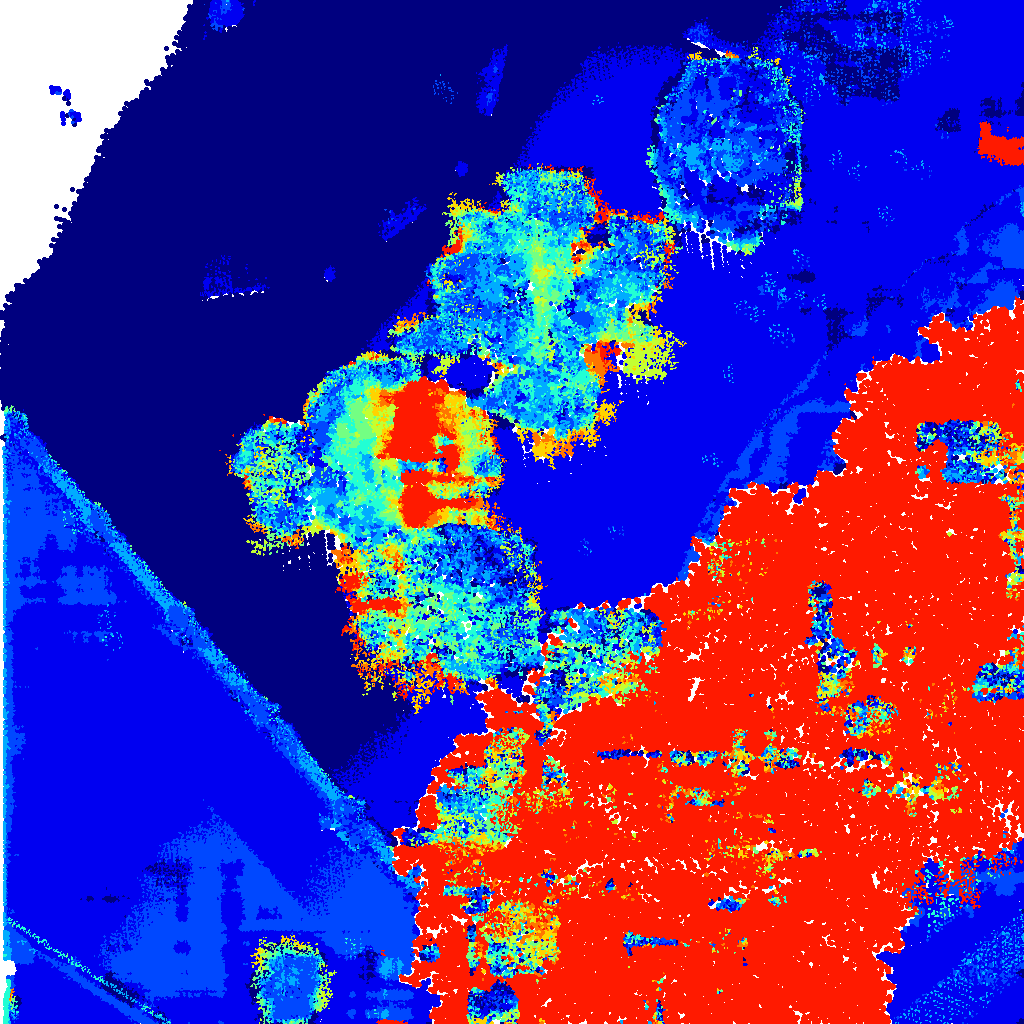}
		\includegraphics[width=\linewidth]{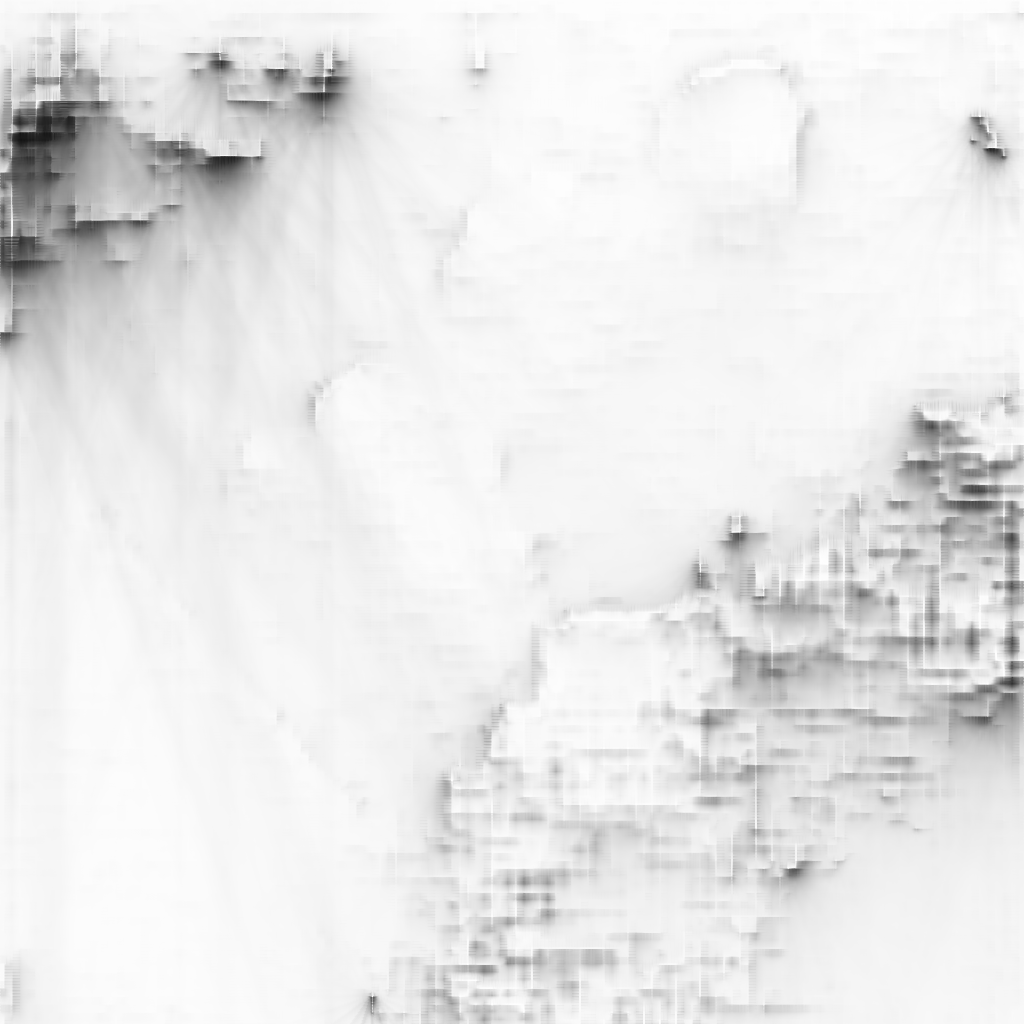}
		\centering{\tiny HRS net(KITTI)}
	\end{minipage}
	\begin{minipage}[t]{0.19\textwidth}	
		\includegraphics[width=0.098\linewidth]{figures_supp/color_map.png}
		\includegraphics[width=0.85\linewidth]{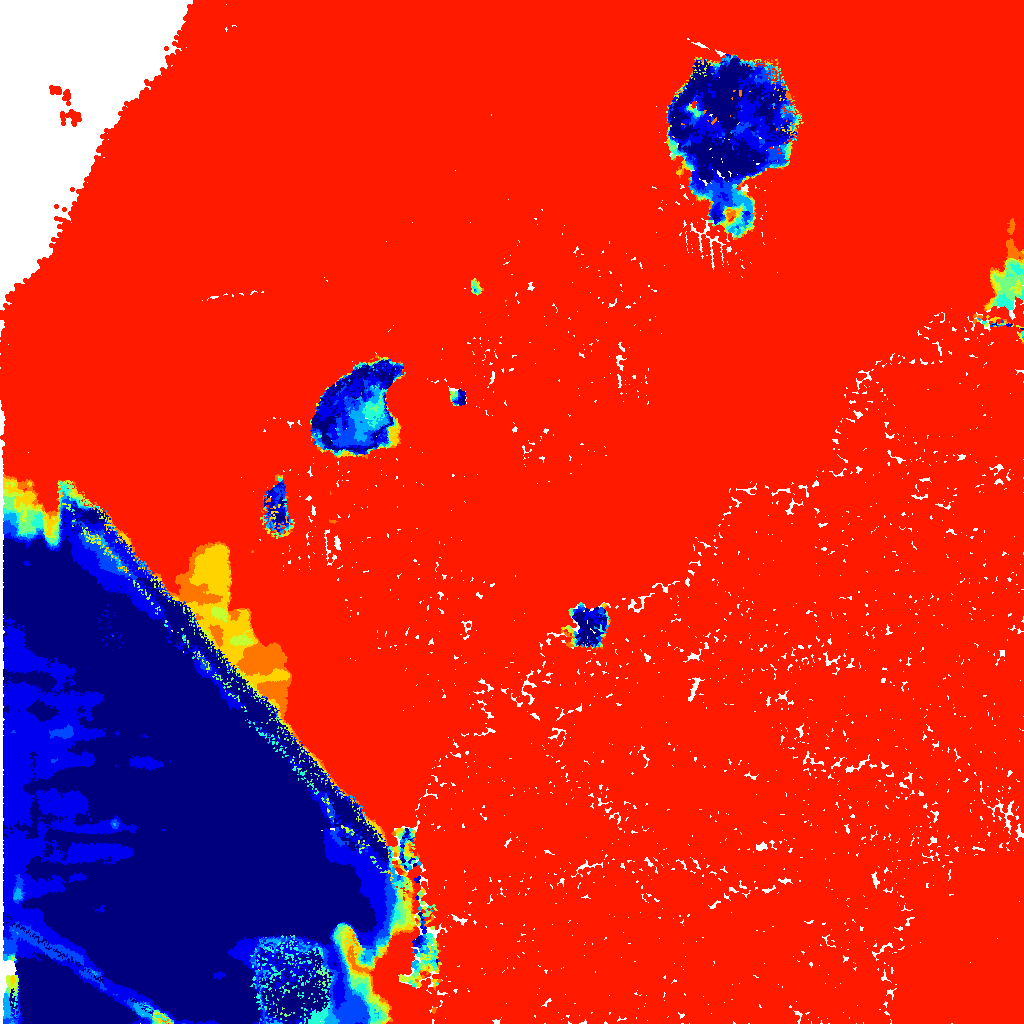}
		\includegraphics[width=\linewidth]{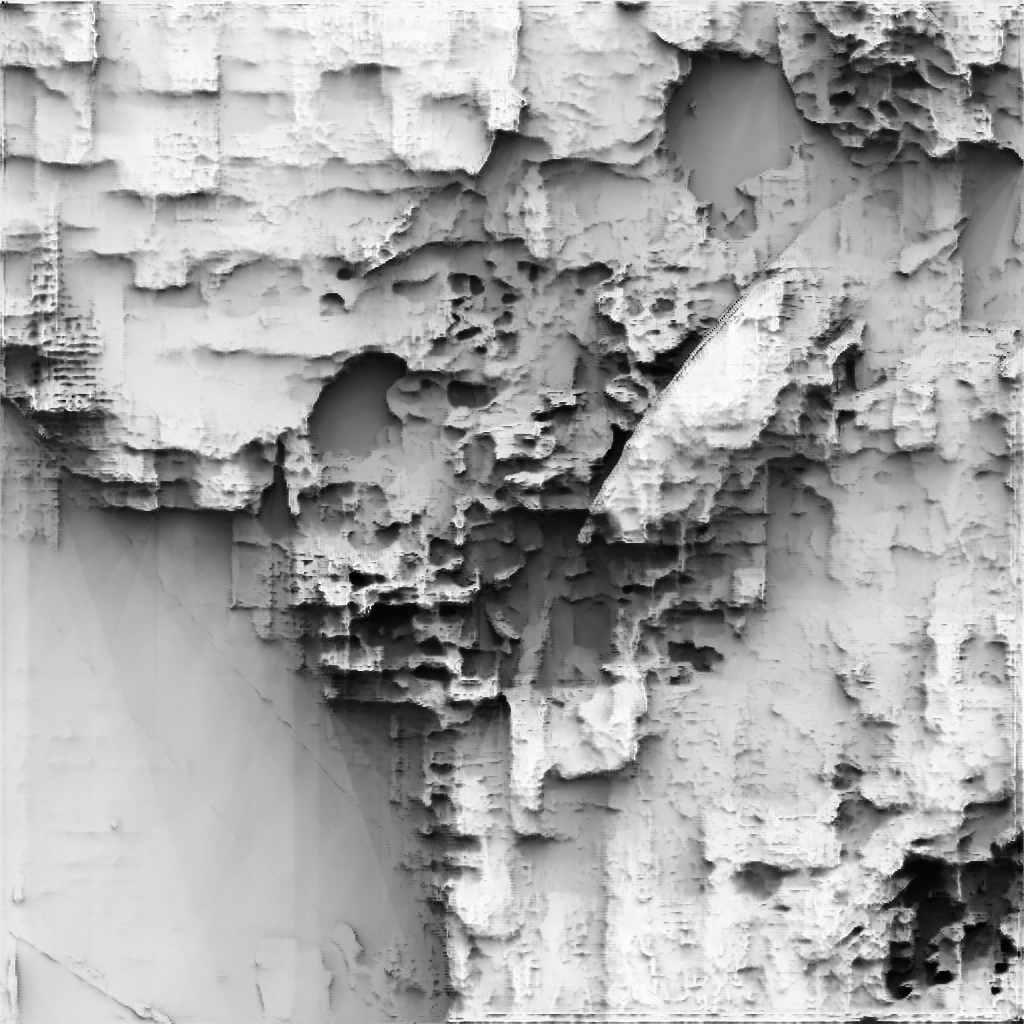}
		\centering{\tiny DeepPruner(KITTI)}
	\end{minipage}
	\begin{minipage}[t]{0.19\textwidth}	
		\includegraphics[width=0.098\linewidth]{figures_supp/color_map.png}
		\includegraphics[width=0.85\linewidth]{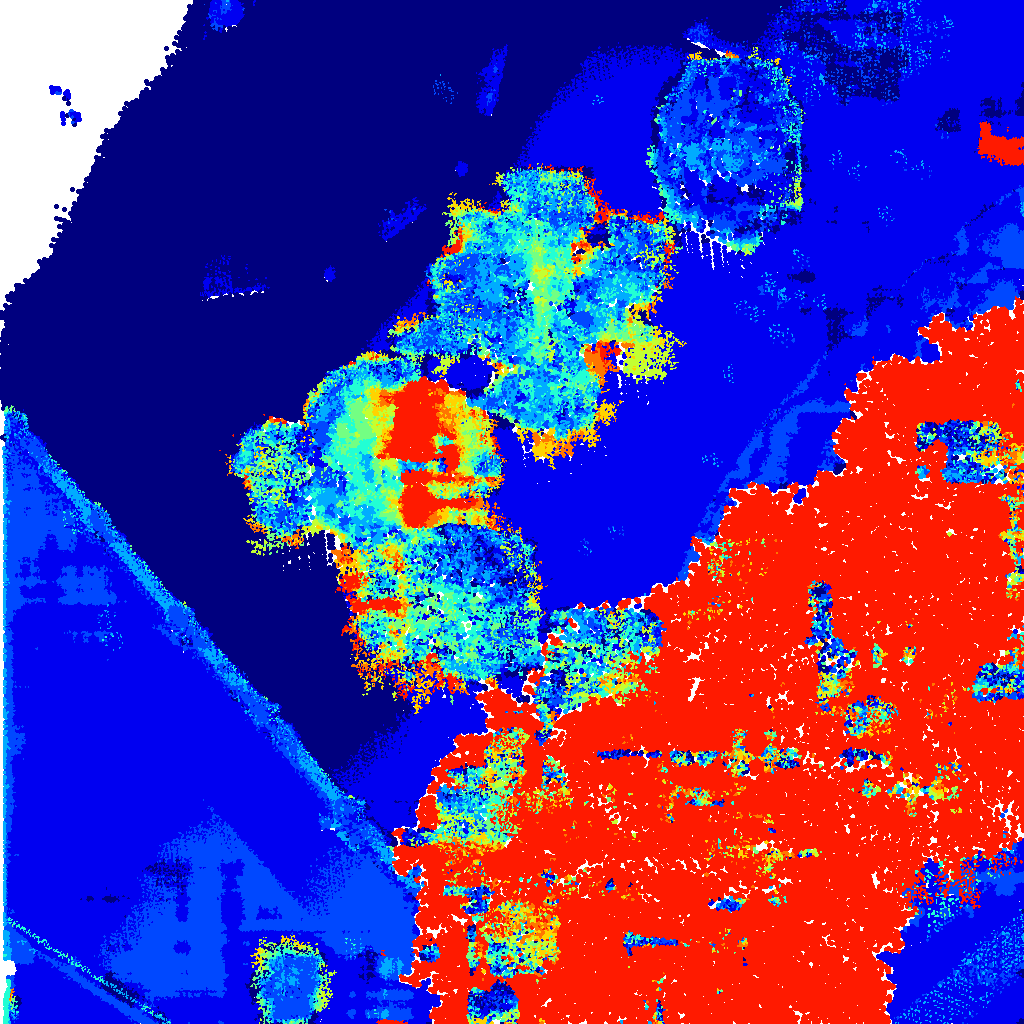}
		\includegraphics[width=\linewidth]{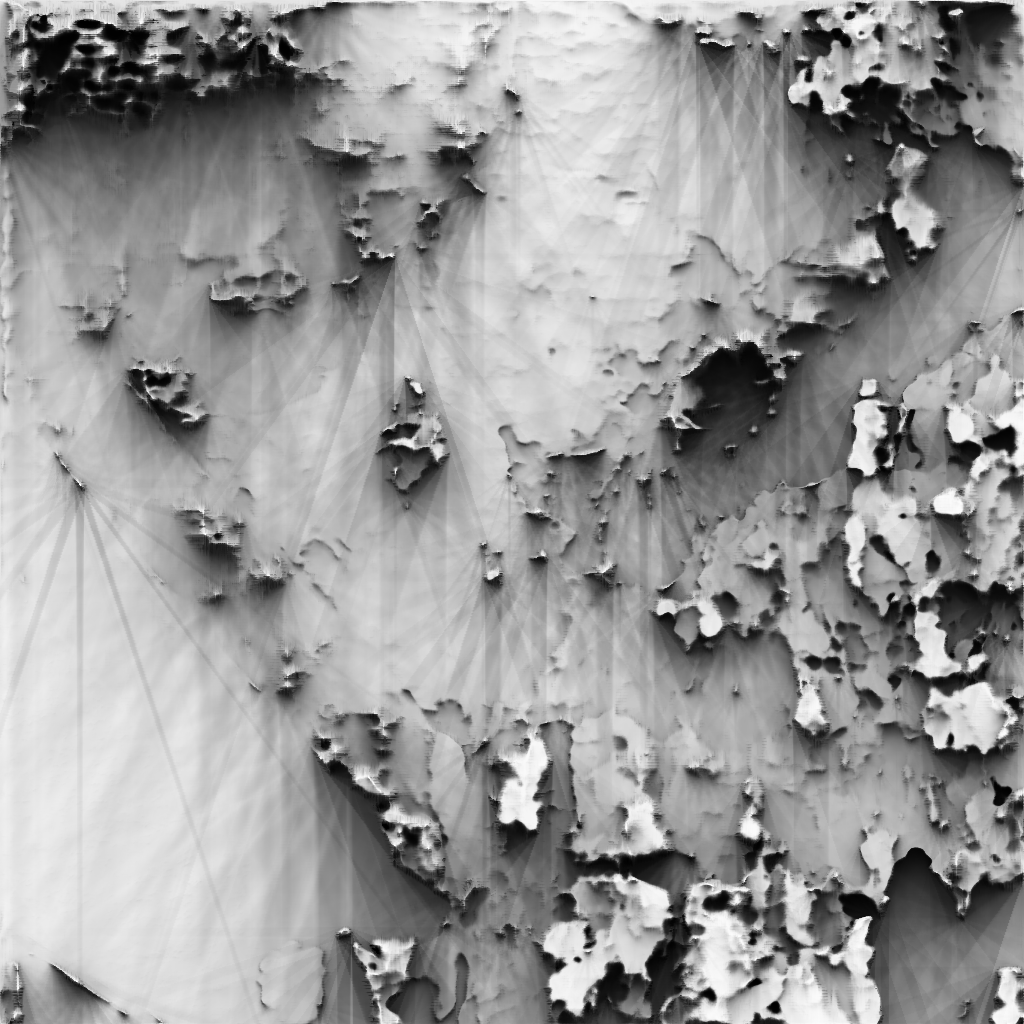}
		\centering{\tiny GANet(KITTI)}
	\end{minipage}
	\begin{minipage}[t]{0.19\textwidth}	
		\includegraphics[width=0.098\linewidth]{figures_supp/color_map.png}
		\includegraphics[width=0.85\linewidth]{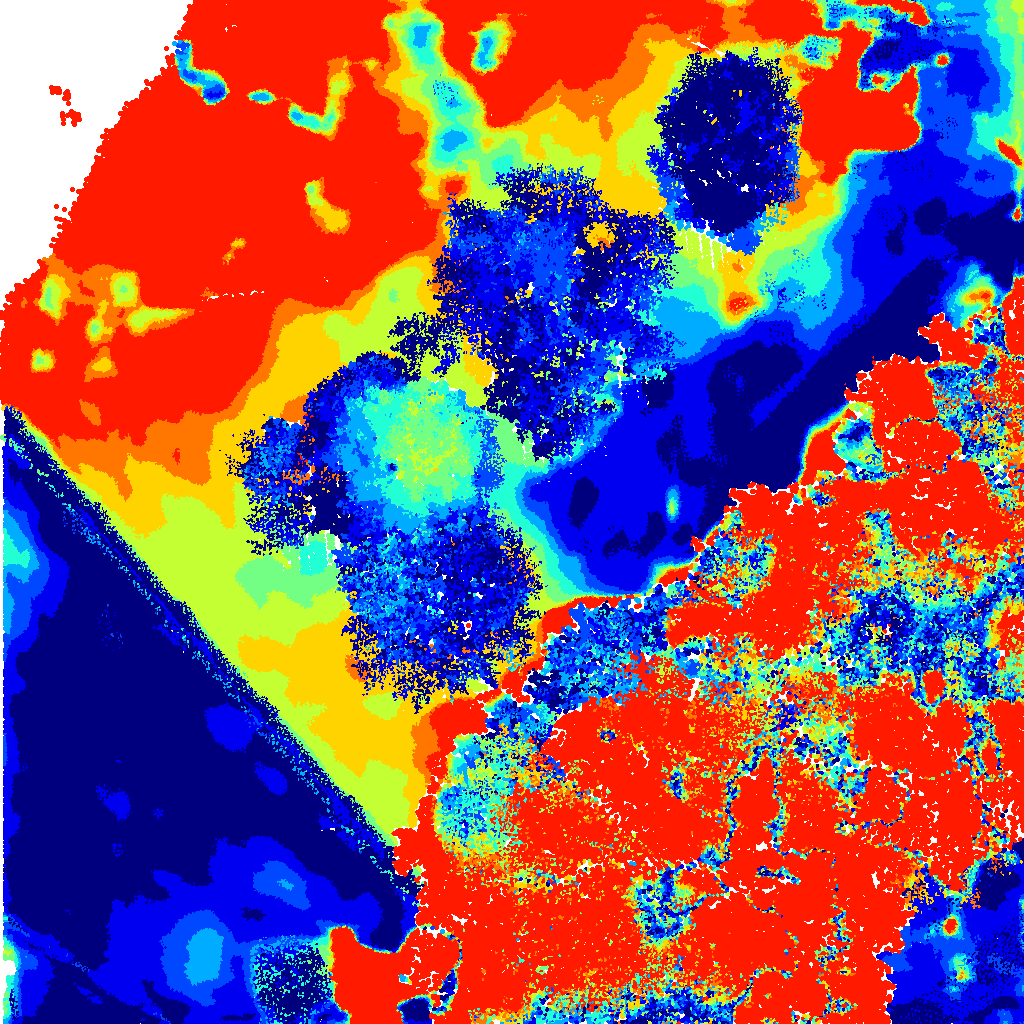}
		\includegraphics[width=\linewidth]{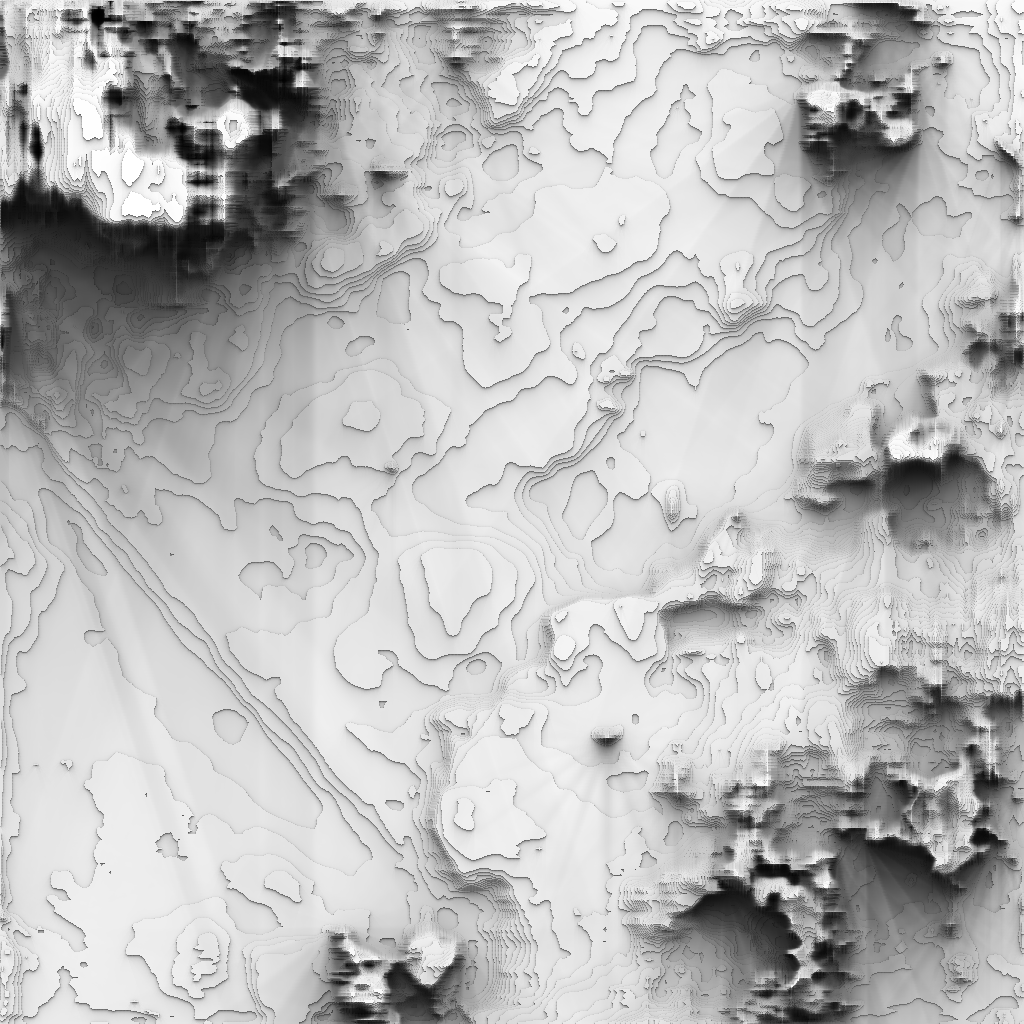}
		\centering{\tiny LEAStereo(KITTI)}
	\end{minipage}
	\begin{minipage}[t]{0.19\textwidth}
		\includegraphics[width=0.098\linewidth]{figures_supp/color_map.png}
		\includegraphics[width=0.85\linewidth]{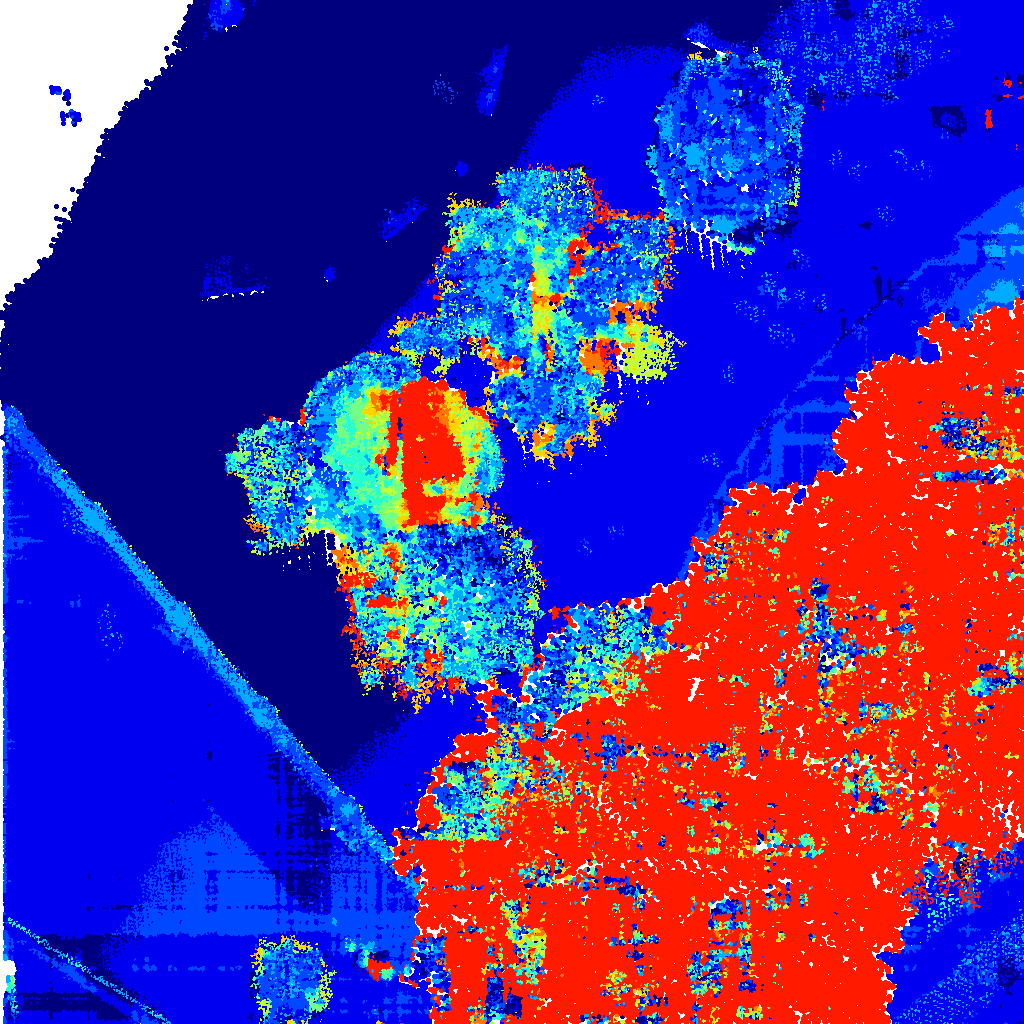}
		\includegraphics[width=\linewidth]{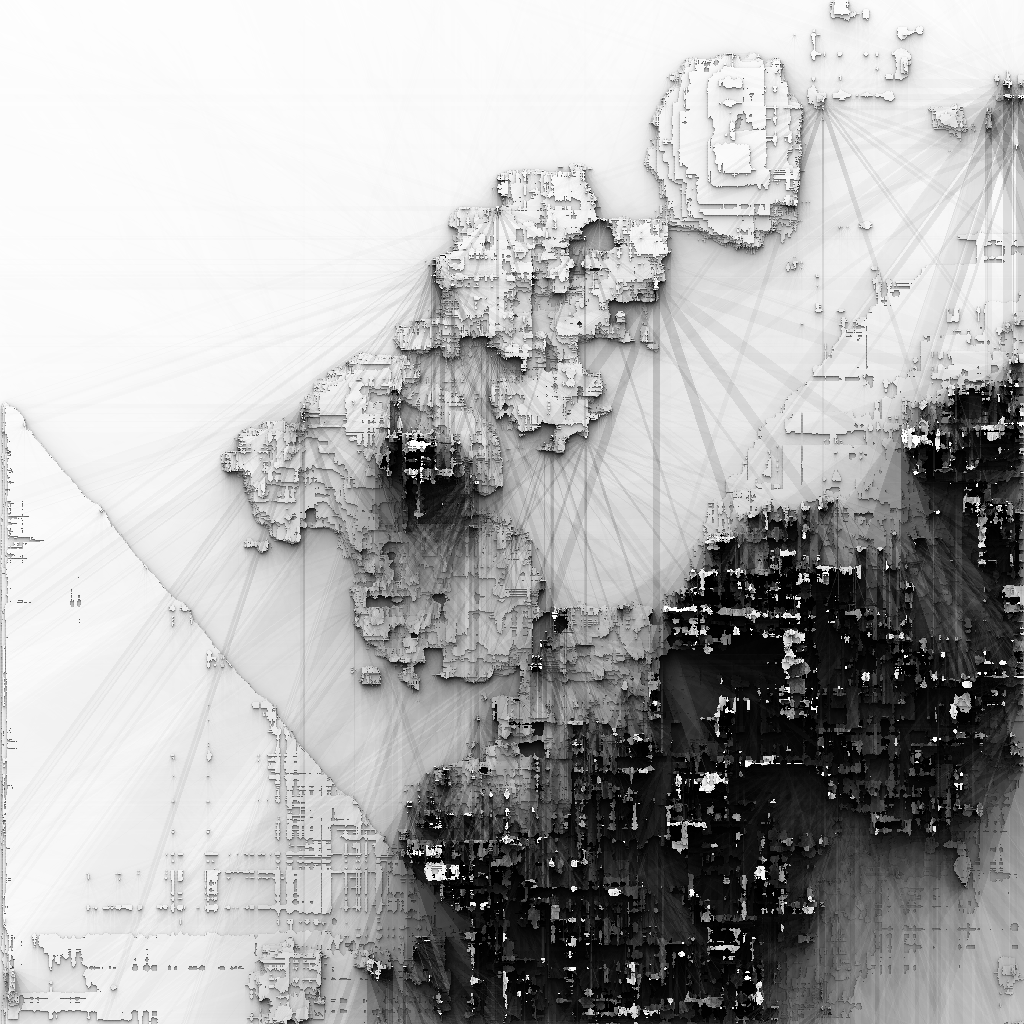}
		\centering{\tiny MC-CNN}
	\end{minipage}
	\begin{minipage}[t]{0.19\textwidth}
		\includegraphics[width=0.098\linewidth]{figures_supp/color_map.png}
		\includegraphics[width=0.85\linewidth]{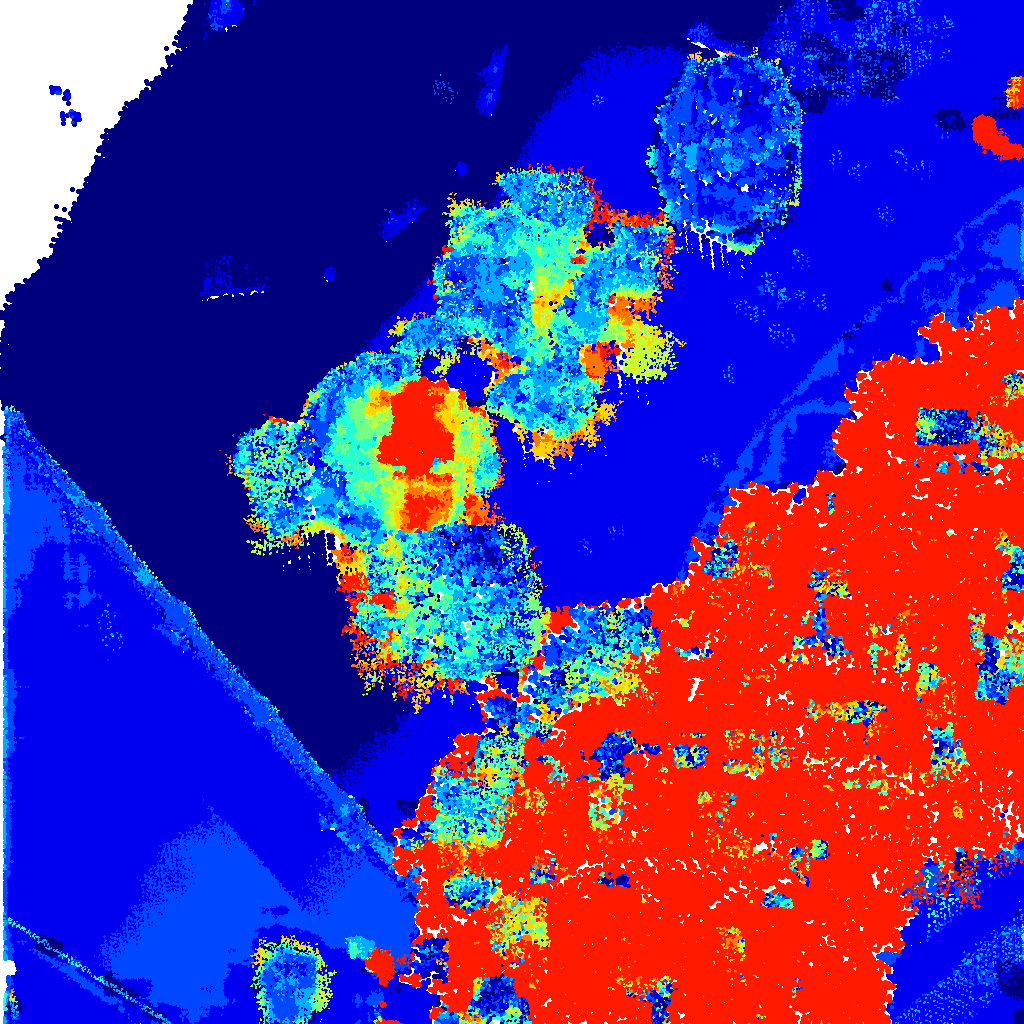}
		\includegraphics[width=\linewidth]{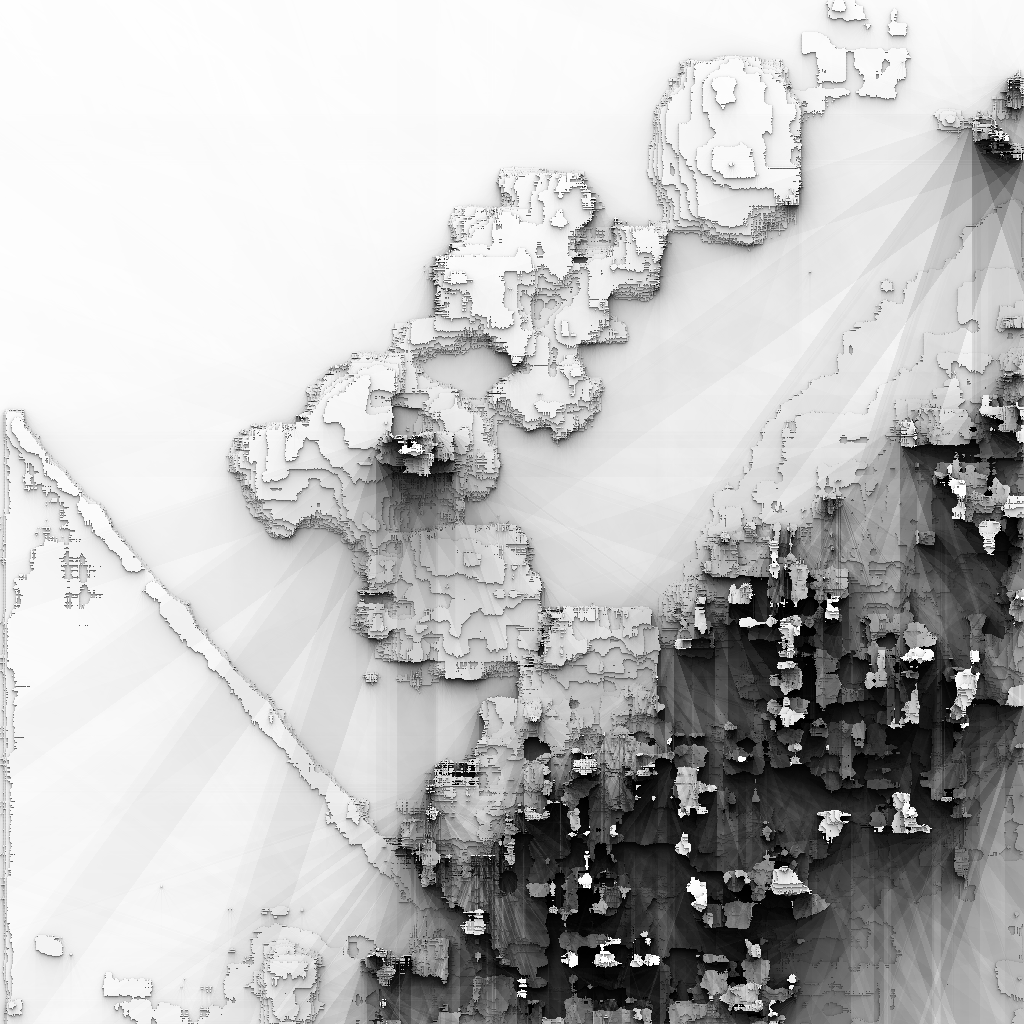}
		\centering{\tiny EfficientDeep}
	\end{minipage}
	\begin{minipage}[t]{0.19\textwidth}
		\includegraphics[width=0.098\linewidth]{figures_supp/color_map.png}
		\includegraphics[width=0.85\linewidth]{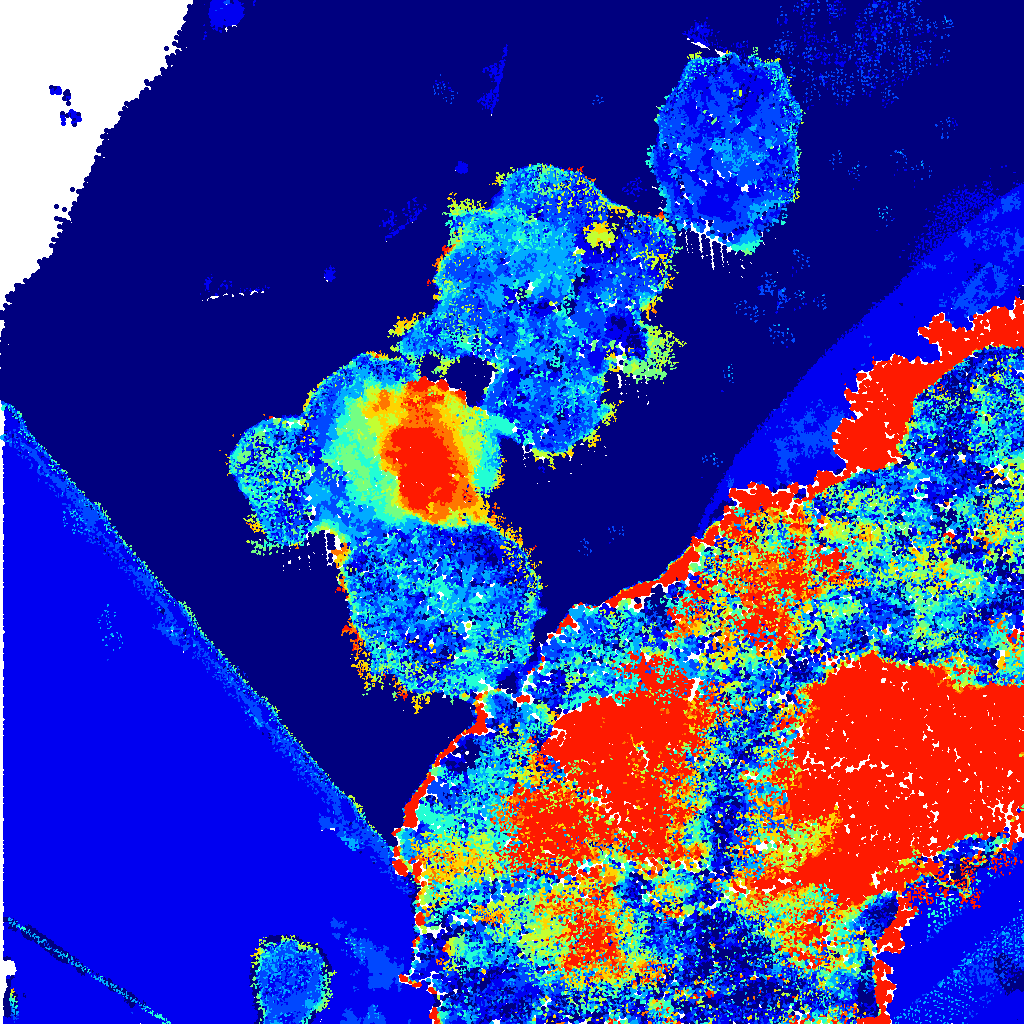}
		\includegraphics[width=\linewidth]{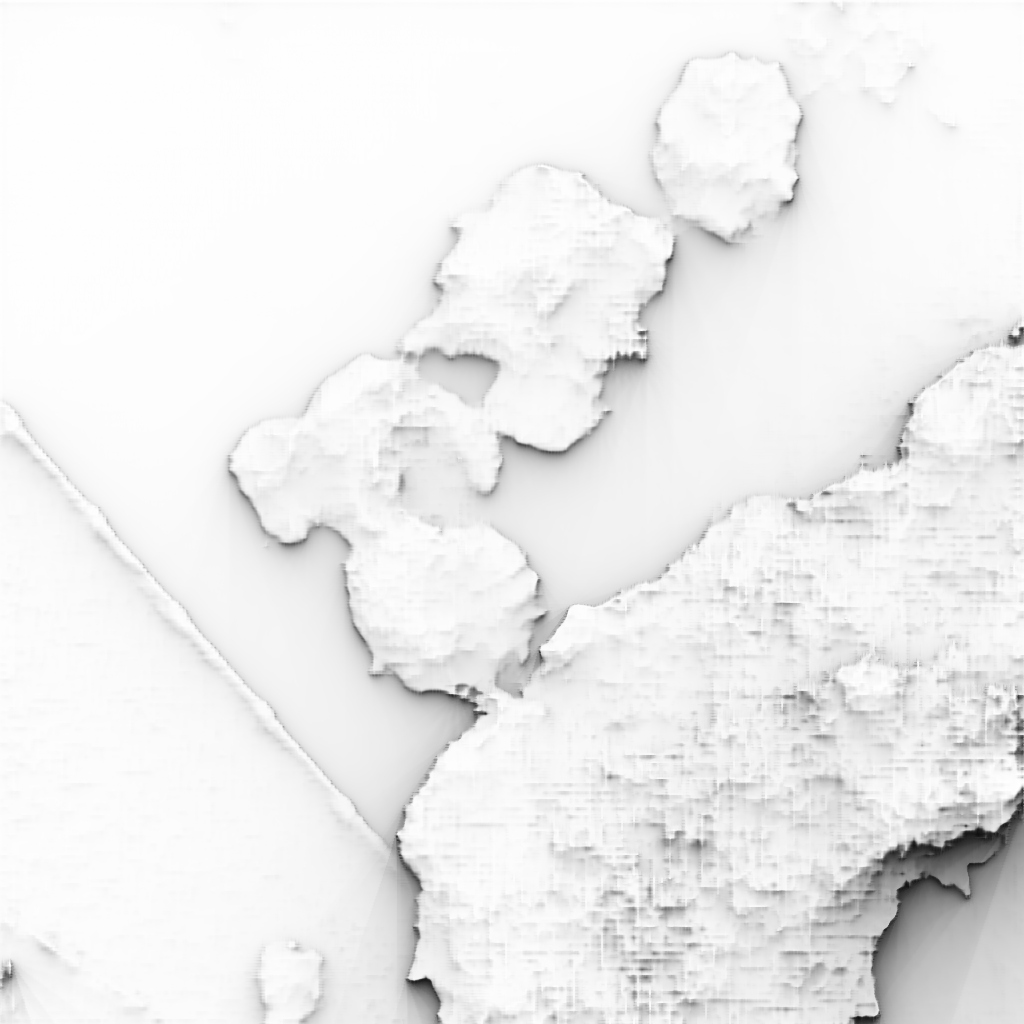}
		\centering{\tiny PSM net}
	\end{minipage}
	\begin{minipage}[t]{0.19\textwidth}	
		\includegraphics[width=0.098\linewidth]{figures_supp/color_map.png}
		\includegraphics[width=0.85\linewidth]{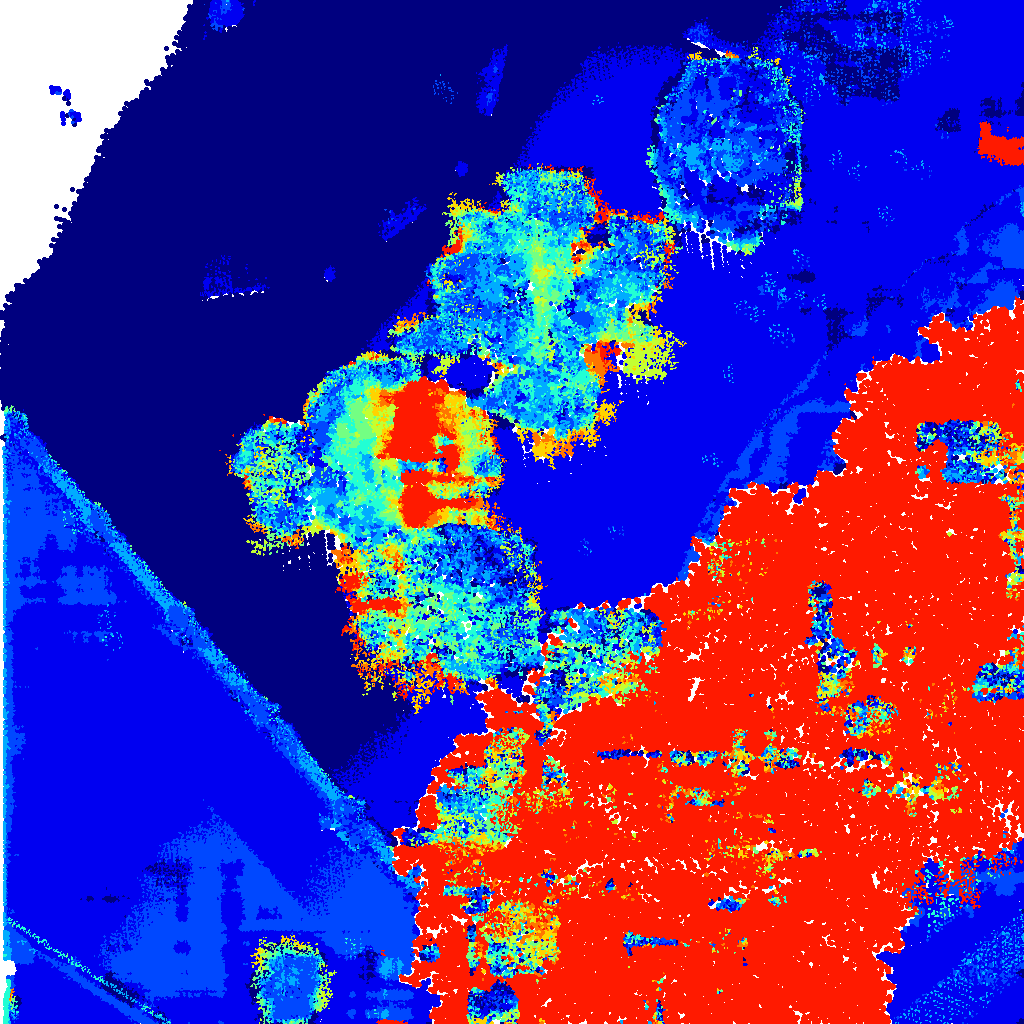}
		\includegraphics[width=\linewidth]{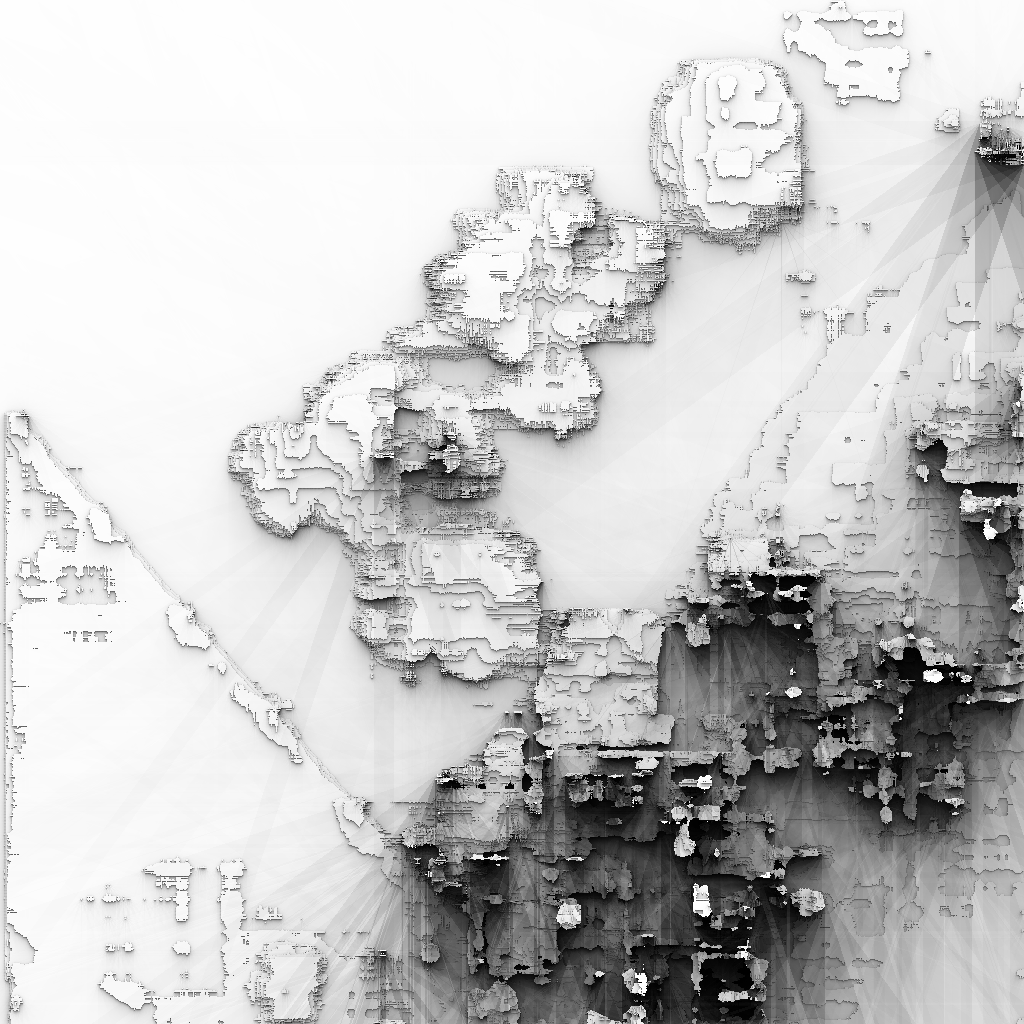}
		\centering{\tiny HRS net}
	\end{minipage}
	\begin{minipage}[t]{0.19\textwidth}	
		\includegraphics[width=0.098\linewidth]{figures_supp/color_map.png}
		\includegraphics[width=0.85\linewidth]{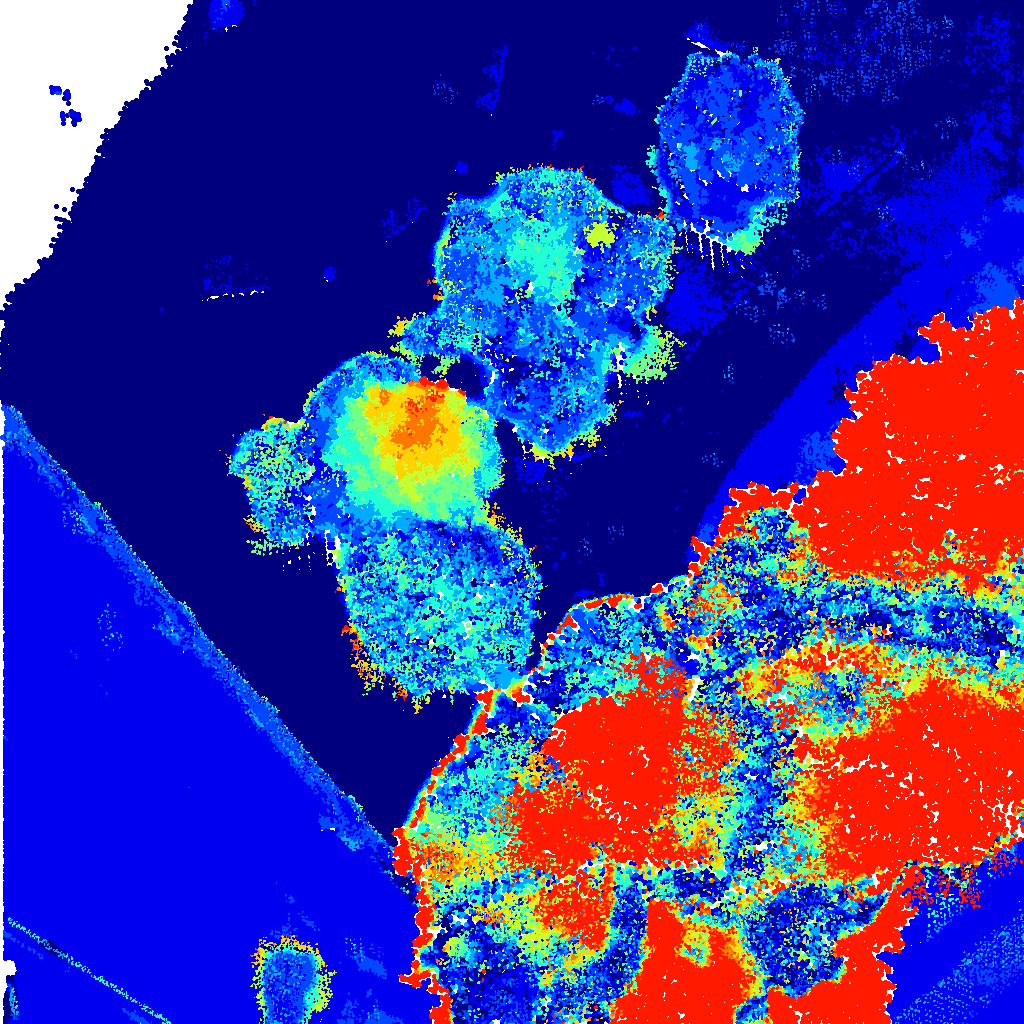}
		\includegraphics[width=\linewidth]{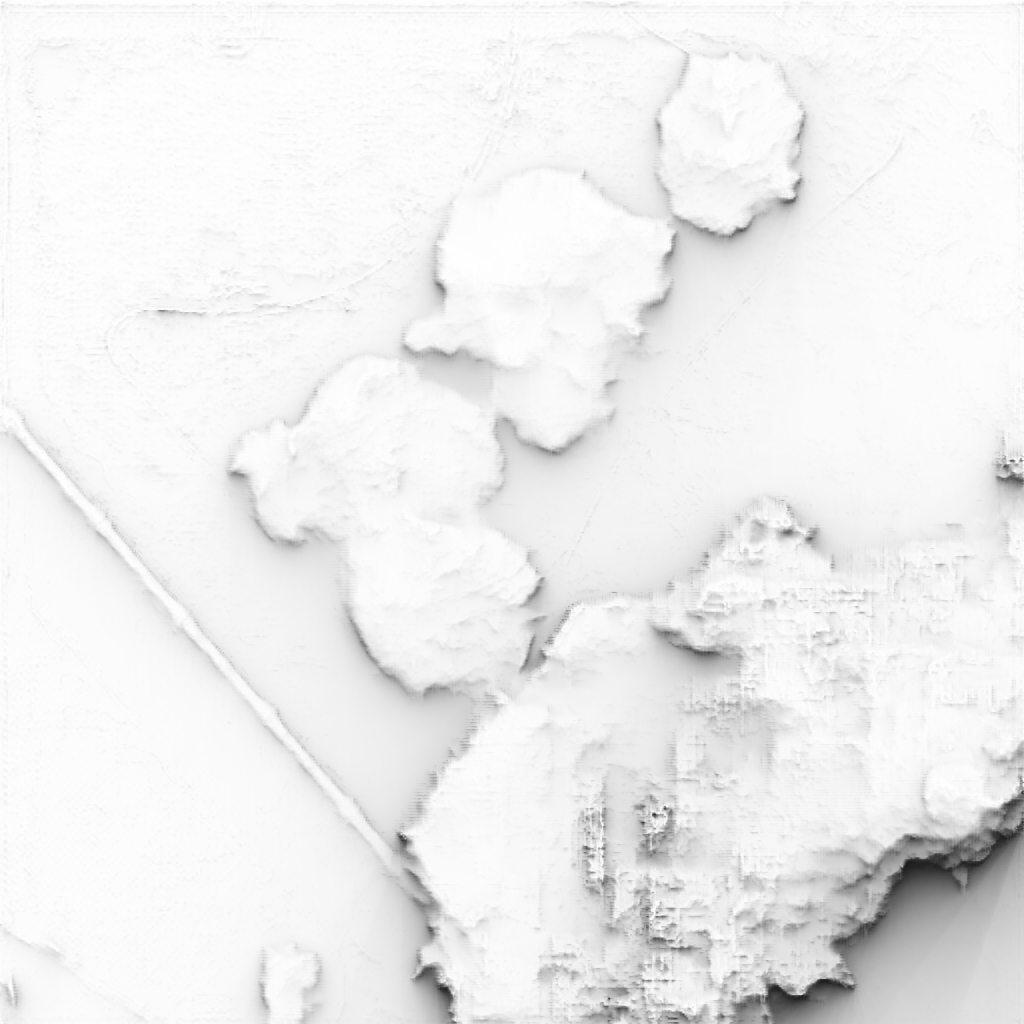}
		\centering{\tiny DeepPruner}
	\end{minipage}
	\begin{minipage}[t]{0.19\textwidth}	
		\includegraphics[width=0.098\linewidth]{figures_supp/color_map.png}
		\includegraphics[width=0.85\linewidth]{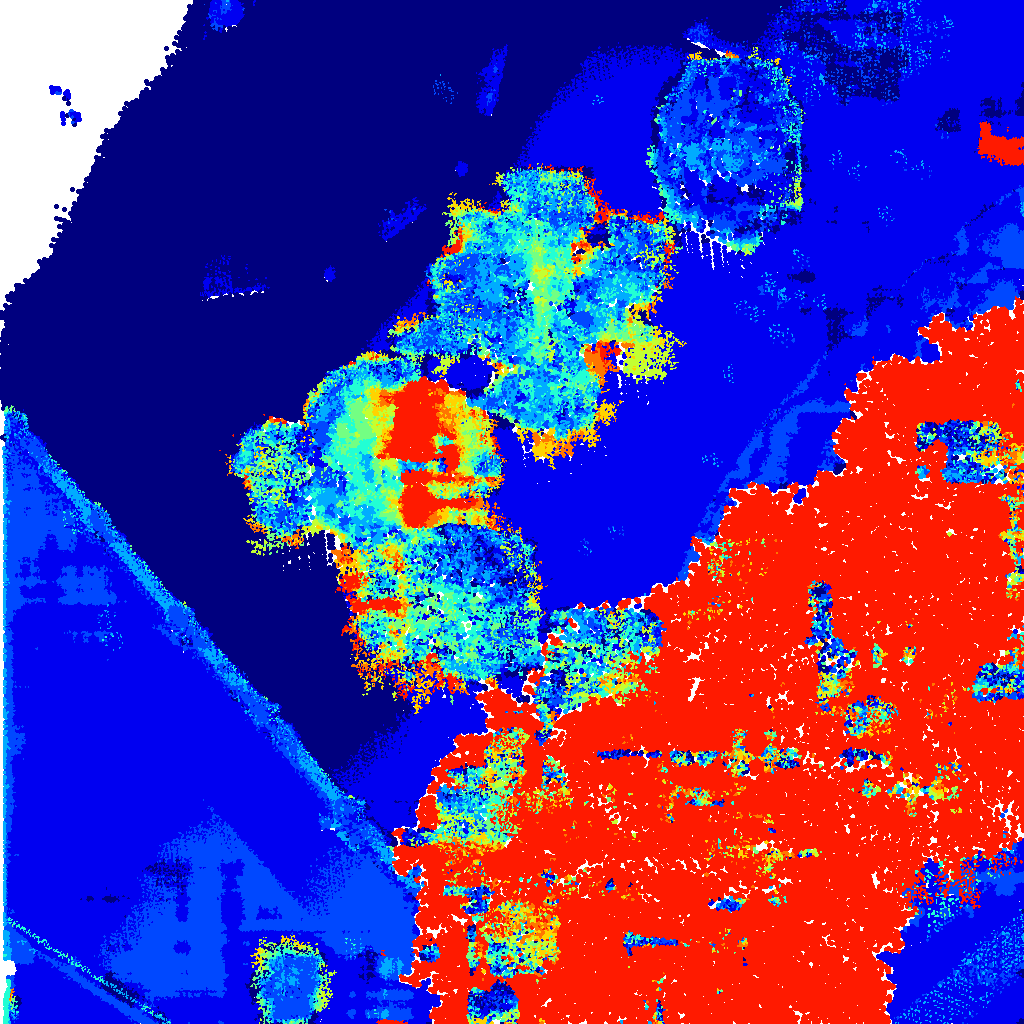}
		\includegraphics[width=\linewidth]{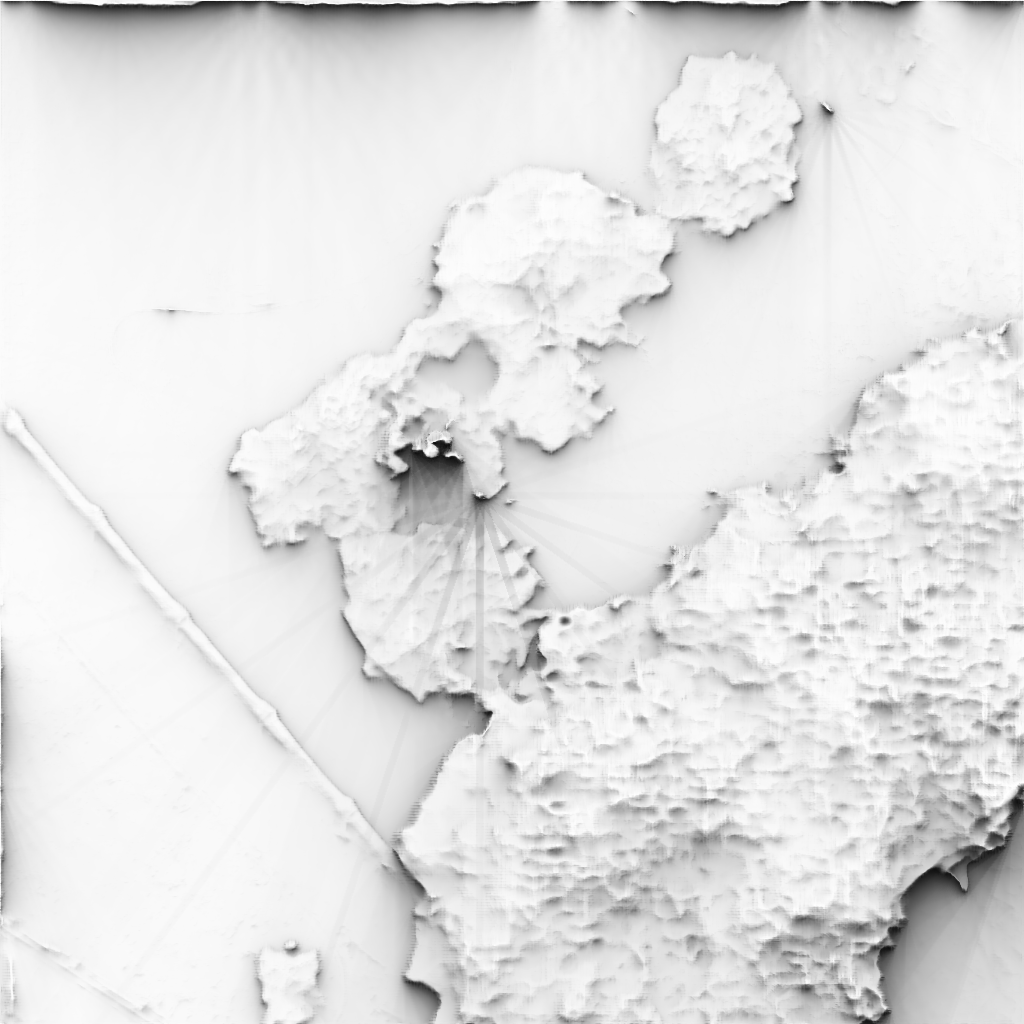}
		\centering{\tiny GANet}
	\end{minipage}
	\begin{minipage}[t]{0.19\textwidth}	
		\includegraphics[width=0.098\linewidth]{figures_supp/color_map.png}
		\includegraphics[width=0.85\linewidth]{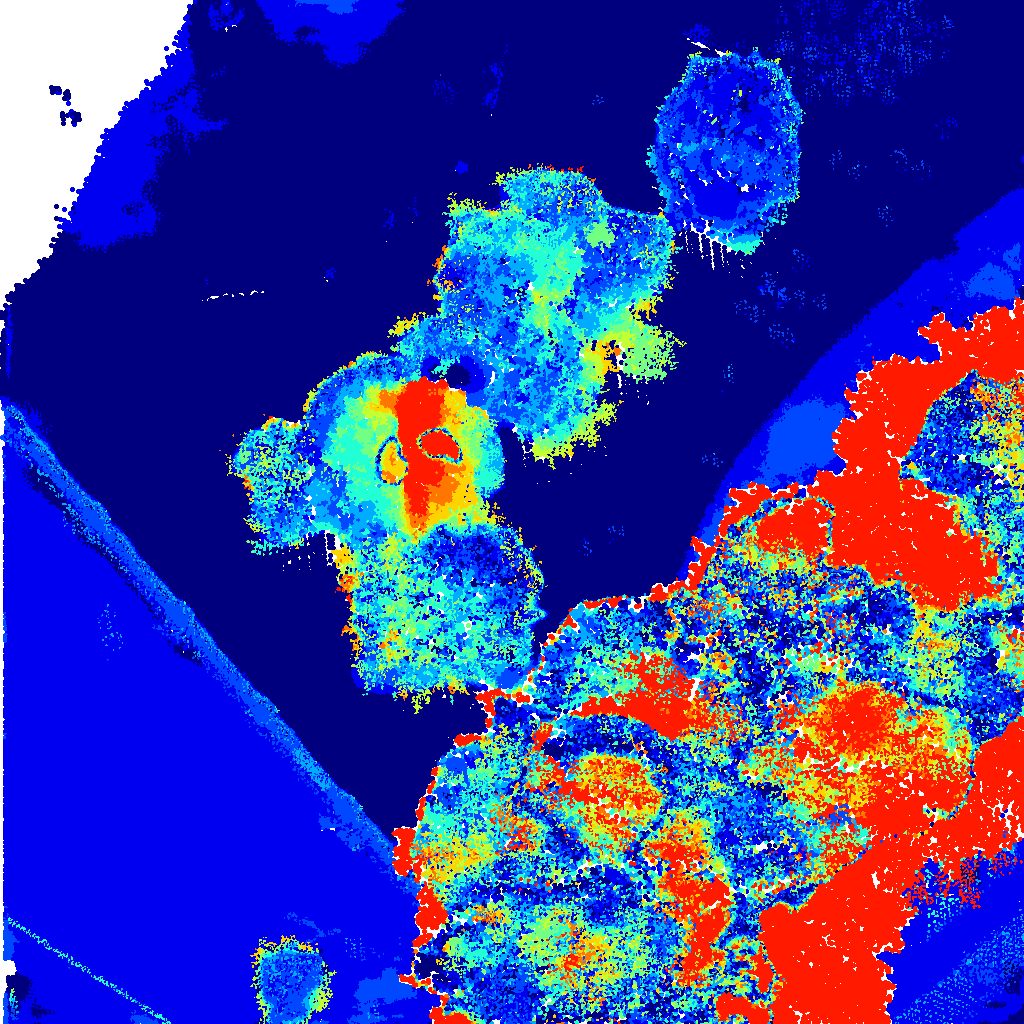}
		\includegraphics[width=\linewidth]{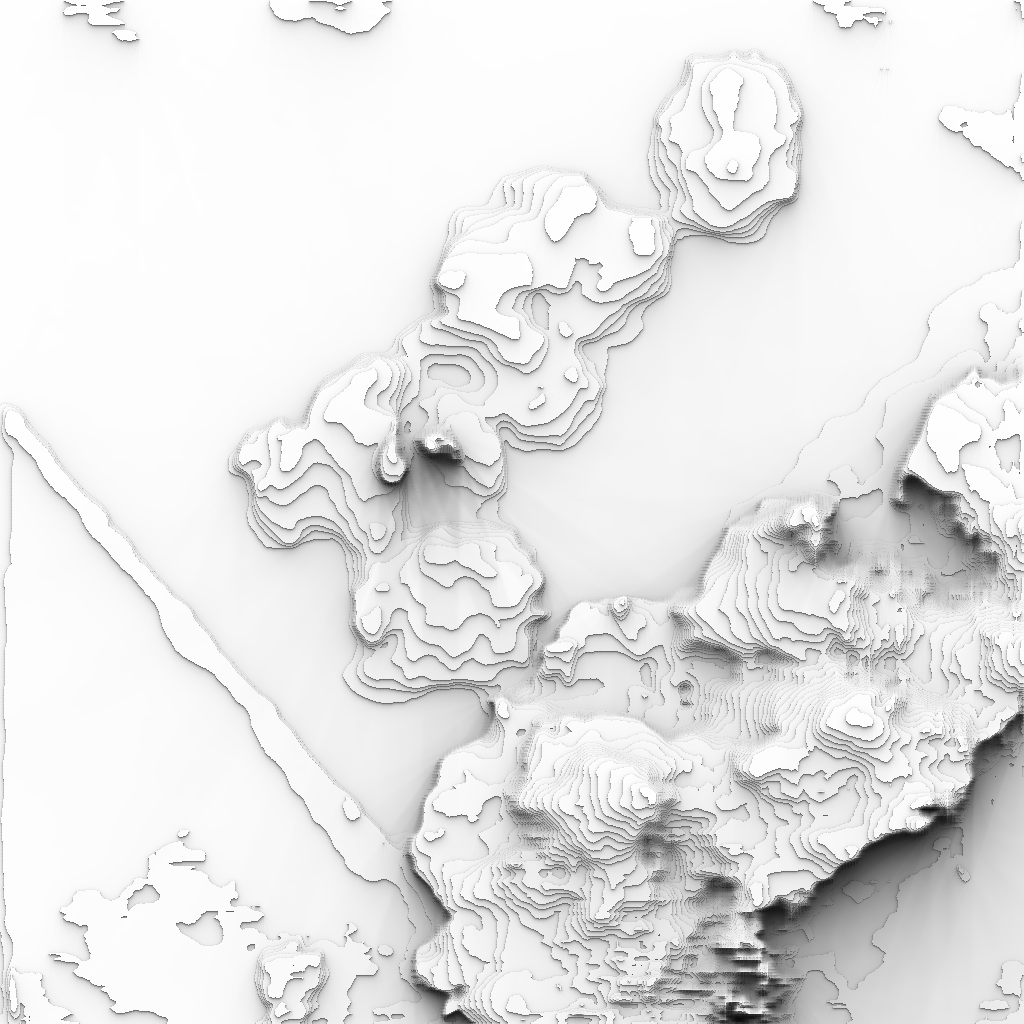}
		\centering{\tiny LEAStereo}
	\end{minipage}
	\caption{Error map and disparity visualization on tree area for DublinCity dataset.}
	\label{Figure.dublintree}
\end{figure}

\paragraph{DublinCity}
All in all in build-up areas the traditional methods show good results, while among the DL-based methods, the pertained model of \textit{DeepPruner} is the worst, after fine-tuning, \textit{GANet} is the best (cf. \Cref{Figure.dublinbulding}). 
DublinCity images are high resolution and thus the buildings' discontinuities span large disparities. This in turn affects the performance of \textit{GANet}. The performance in the vegetated area differs depending on whether the trees are leafy and leafless (cf. \Cref{Figure.example_bulding}). Neither of the methods is capable of reconstructing the leafless trees, even after fine-tuning, for the tree leafy, the training model improves much(cf. \Cref{Figure.dublintree}). 


\subsection{Discussion of Visual assessment}

From the visual assessment, generally, we can make a conclusion:
\begin{itemize}
    \item Using aerial data to do the fine-tuning can improve performance in the building areas with discontinuity. 
    \item Pre-trained methods work better in building areas than in tree areas.
    \item For the tree area, using the pre-trained model on KITTI of DL methods usually produces a bad result. Using fine-tuning can improve the result, but the performance of the building area is better than the tree area.
    \item For the building area, the traditional methods give good results, but after training, DL-based methods also improve the results much. 
    \item For the tree area without leaves, because the discontinuity is too large, DL-based methods do not improve much.
\end{itemize}




\section{Reference}

\bibliography{review}
\bibliographystyle{elsarticle-harv}


\end{document}